\pdfoutput=1
\documentclass[11pt]{article}
\usepackage{fibres}

\title{A Categorical Atlas of\texorpdfstring{\\}{ } Parameter-Efficient Fine-Tuning}
\subtitle{Cones, Orbits, and Gauges\textsuperscript{\,\textdagger}}
\author{\fbauthors{%
  \fbauthor{Xiaoyu Li}{University of New South Wales}{xiaoyu.li2@unsw.edu.au} &
  \fbauthor{Zhizhou Sha}{University of Texas at Austin}{zhizhousha@utexas.edu}\\[0.35em]
  \fbauthor{Chiwun Yang}{City University of Hong Kong}{christiannyang37@gmail.com} &
  \fbauthor{Dai Shi}{University of Cambridge}{ds2213@cam.ac.uk}}%
  \fbwith{with Claude Opus 5.5 (Anthropic), in the spirit of Grothendieck's rising sea}}
\date{Technical report\textsuperscript{\,\textdaggerdbl} \enspace·\enspace October 2026 \enspace·\enspace Version 1}
\companion{Interactive companion: \url{https://xiaoyulics.com/peft-atlas/}\par Video: \url{https://xiaoyulics.com/peft-atlas/film/}}
\hypersetup{pdftitle={A Categorical Atlas of Parameter-Efficient Fine-Tuning: Cones, Orbits, and Gauges},pdfauthor={Xiaoyu Li, Zhizhou Sha, Chiwun Yang, Dai Shi}}

\begin{document}
\newcommand{\nMethods}{382}
\newcommand{\nPapers}{138}
\newcommand{\nHF}{82}
\newcommand{\nCore}{142}
\newcommand{\nArrows}{403}
\newcommand{\nObstructions}{485}
\newcommand{\nPropositions}{41}
\newcommand{\nBib}{630}

\maketitle
{\renewcommand{\thefootnote}{\textdagger}\footnotetext{The title pays homage to \citet{bronstein2021geometric}, \emph{Geometric Deep Learning: Grids, Groups, Graphs, Geodesics, and Gauges}, whose programme of reading architectures through their symmetries this atlas extends to fine-tuning.}}
{\renewcommand{\thefootnote}{\textdaggerdbl}\footnotetext{This is a technical report rather than a journal or conference submission. We share it, with its proofs, catalogue and interactive companion, as an open resource for everyone working on parameter-efficient fine-tuning, and we welcome corrections and suggestions for future versions.}}
\begin{abstract}
Parameter-efficient fine-tuning has hundreds of methods, usually compared by parameter count and benchmark score. We compare them as maps. A pretrained checkpoint is a base point in weight space, and a weight-space method is a smooth map from a small parameter space that starts at that point. These maps form a slice category in which the frozen model is initial and full fine-tuning terminal. Three invariants of the map do most of the work. Its image near the base point governs expressivity, through the first-order image, the dimension, a min-cut bound on rank and whether that point is smooth or an apex. Its fibres, with the optimizer's metric, govern training dynamics. Base change moves the base point and covers splitting initialisations, quantisation, restarts and merging. We prove, for example, that zero-initialised LoRA, and Householder reflection adaptation at its default start under stated rank and parity conditions, begin at the apex of a cone; that no exact initialisation lets LoRA reach an update of rank above twice its own; that exactly orthogonal methods never change a weight's singular values in exact arithmetic, however often they are merged; that under Adam with no epsilon, weight decay or clipping and one shared schedule, LoRA+ from the standard zero initialisation follows exactly the trajectory of plain LoRA with its scale multiplied by the learning-rate ratio; that Adam with shared hyperparameters respects only the signed permutations inside LoRA's continuous symmetry group of full-rank factors; and that the activation function decides when a scaling adapter folds into preceding weights. We classify \nMethods{} methods from \nPapers{} papers by seven coordinates read off their formulas, with \nArrows{} simulation arrows and \nObstructions{} obstructions. Every numbered result is proved in full, and results due to others are credited where we re-derive them.
\end{abstract}

\clearpage
{\small\tableofcontents}
\clearpage

\startatzero
\section{Thesis}\label{sec:thesis}

\begin{flushright}\small\itshape
The right generality turns a hard theorem into a remark,\\
and the remark into a design rule.\\[3pt]
{\upshape\footnotesize after Grothendieck}
\end{flushright}

Parameter-efficient fine-tuning (PEFT) has hundreds of methods, usually compared by parameter count and benchmark score. Those two numbers leave open what a method can reach, how it trains, and what happens to it when it is initialised differently, quantised, restarted or merged. Each of them becomes a precise mathematical question once a method is treated as a map, and many then have exact answers. This paper compares methods that way.

\begin{figure}[tbp]
\centering
\includegraphics[width=\linewidth]{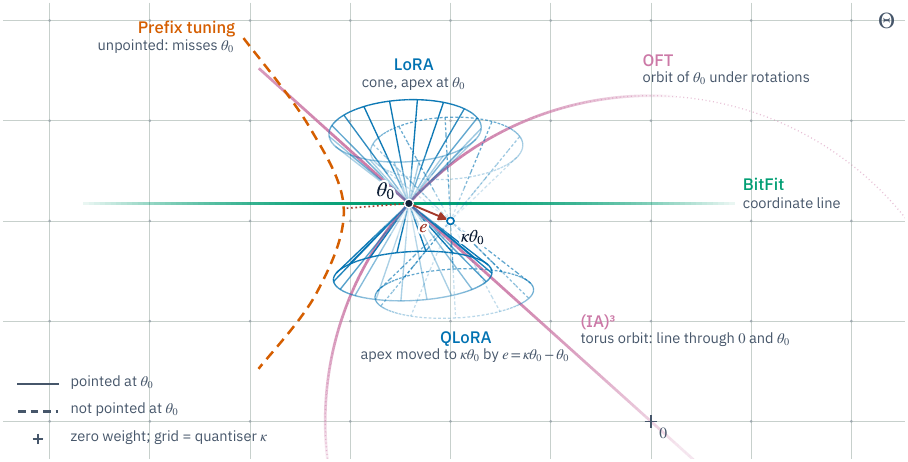}
\caption{\textbf{Fine-tuning moves a point.} Weight space $\Theta$ is drawn as a chart around the pretrained point $\theta_0$; the grid is the lattice of a round-to-nearest quantiser $\kappa$ through the zero weight $0$ (marked $+$). Solid curves are image germs of methods pointed at $\theta_0$ (\thmref{Definition}{I.2}), drawn darkest near $\theta_0$ because only the germ there matters. LoRA \citep{hu2021lora} is the cone $\theta_0+\Mr$ with its apex at $\theta_0$ (\thmref{Theorem}{II.2}), shown as the rank-$\le1$ double cone $x^2=y^2+z^2$ of the symmetric $2\times2$ slice in perspective. OFT \citep{qiu2023controlling} is the orbit of $\theta_0$ under rotations, an arc of the circle about $0$ through $\theta_0$, the rest of which is dotted. (IA)$^3$ \citep{liu2022few} is a torus orbit, whose closure is the line through $0$ and $\theta_0$ (\thmref{Theorem}{II.5}). BitFit \citep{benzaken2021bitfit} moves along a coordinate line, parallel to the grid. The dashed curves are not pointed at $\theta_0$. Prefix tuning \citep{li2021prefix} extends the network and is unpointed (\thmref{Definition}{VI.1}); drawn in weight space it is a curve that misses $\theta_0$, at the distance marked by the dotted segment. QLoRA \citep{dettmers2023qlora} is LoRA pointed at the nearest grid node $\kappa\theta_0$, so its cone is LoRA's cone translated by the defect $e=\kappa\theta_0-\theta_0$ (red arrow), and $\theta_0$ is off its image unless $\rk e\le r$ (\thmref{Proposition}{V.2}).}
\label{fig:teaser}
\end{figure}

Fine-tuning moves a point. A pretrained checkpoint $\theta_0$ is a \emph{base point} in weight space $\Theta$, and a weight-space PEFT method is a \emph{relative object} over it, namely a pointed smooth map
\[
  \rho:(Q,q_0)\longrightarrow(\Theta,\theta_0),\qquad \rho(q_0)=\theta_0,
\]
from a small space $Q$ of trainable parameters into weight space. These maps form the pointed slice category $\Meth(\Theta,\theta_0)$ (\thmref{Definition}{I.2}). A morphism $h:M\to N$ is a pointed smooth map with $\rho_N\circ h=\rho_M$, read ``$N$ simulates $M$''. The frozen model is initial and full fine-tuning is terminal (\thmref{Observation}{I.4}), so every method pointed at $\theta_0$ sits in the interval between them. Following Grothendieck's relative point of view, the object of study is the morphism $\rho$ itself. Almost all the work is done by three invariants of the morphism, plus one further layer for architectures. Figure~\ref{fig:teaser} draws the picture.

The bicategory $\Para$ and the lens reading of backpropagation are due to \citet{fong2017backprop} and \citet{cruttwell2021categorical}. The closest earlier unified frame for PEFT is that of \citet{he2021unified}, who read adapters, prefix tuning and LoRA as modifications of specific hidden states. Ours works on the weight side, where adapters and prefixes return as extensions of the network (\thmref{Definition}{VI.1}).

\subsubsection{Three invariants and one further layer}

The three invariants answer three practical questions: what a method can say, how it learns, and how it travels.

\paragraph{The image germ at $\theta_0$ governs expressivity}
This is what a method can say. Near $\theta_0$, the weights a method reaches form a linear space, a cone, a group orbit, a torus saturation, or a meet of these. Four data are read off this germ: the first-order image $S_1$, the expressive dimension $d(M)$, the min-cut bound on rank (the cut-rank), and the type of the base point, which is either a smooth point or an apex. Zero-initialised LoRA \citep{hu2021lora} starts at the apex of the cone $\theta_0+\Mr$, where first steps miss directions that the method reaches later; a PiSSA-style split \citep{meng2024pissa} starts at a smooth point. Exposé~\ref{sec:image} treats the germ in three parts: strata of initialisations (\ref{sec:strata}), invariants of group-type methods (\ref{sec:erlangen}), and cut and volume (\ref{sec:cut}).

\paragraph{The fibres, together with a metric, govern dynamics}
This is how a method learns. Many parameter values give the same weight, and the fibres of $\rho$ are gauge orbits; on LoRA's full-rank locus they are the orbits of $\GL_r$ acting by $g\cdot(B,A)=(Bg^{-1},gA)$. Combined with the optimizer's learning rates $\Lambda\succ0$, the parametrisation induces a cometric on weight space,
\[
  K=D\rho\,\Lambda\,D\rho\T ,
\]
so isomorphic methods can train differently. For LoRA with per-block rates, the compact half of the gauge acts by isometries and the non-compact half gives charges conserved under gradient flow. The optimizer's equivariance group decides what descends to the model (Exposé~\ref{sec:fibres}). The lens structure of backpropagation sorts tangent-side from cotangent-side methods (Exposé~\ref{sec:lens}). Through this lens, GaLore \citep{zhao2024galore} with no weight decay and no full-gradient clipping is one-sided LoRA with a moving frozen factor and carried-over optimizer state (\thmref{Proposition}{IV.2}); this identity is due to \citet{xtorrobahennigen2025duality}. This layer rests on work on factorised and linear networks: balancedness \citep{du2018algorithmic,arora2018optimization}, symmetries and conservation laws \citep{kunin2020neural,zhao2022symmetries}, and imbalance at initialisation \citep{xtarmoun2021understanding,xmin2021explicit}.

\paragraph{Base change moves $\theta_0$}
This is how a method travels. Initialisation by splitting, quantisation, restarts and merging all move the base point (Exposé~\ref{sec:base}). QLoRA \citep{dettmers2023qlora} is LoRA pointed at the quantised weights $\kappa\theta_0$, with defect $e=\kappa\theta_0-\theta_0$, and its image misses $\theta_0$ whenever $\rk e>r$, which is the generic case (\thmref{Proposition}{V.2}). Merge-and-restart with a fixed adapter shape reaches, in the limit, the orbit of the monoid generated by one cycle. For ReLoRA \citep{lialin2023relora} this means that $K$ cycles reach exactly the updates of rank at most $\min(Kr,m,n)$ (\thmref{Theorem}{V.3}).

\paragraph{Extensions and mergeability}
The fourth layer is the one place where a method changes the network $f$ itself; the network otherwise enters only through its realisation map and the backpropagation lens (\thmref{Remark}{I.3}). Adapters, prefixes and ReFT first extend $f$. Merging is then a lifting problem, certified by a local rewriting calculus of seven rules with side conditions, which gives sufficient conditions and box-local obstructions (Exposé~\ref{sec:merge}).

\subsubsection{Contributions}

Each result below is proved in Appendix~\ref{app:proofs}, and the scope stated here is the scope of the proof. Where a result, or a part of one, is due to others, the item says so. We re-derive such results in our language and add only their reading for fine-tuning.

\begin{enumerate}[label=\textbf{\arabic*.},leftmargin=1.9em]
\item \textbf{A relative frame for PEFT.} Weight-space methods are the objects of the pointed slice $\Meth(\Theta,\theta_0)=\Diff_*/(\Theta,\theta_0)$. Every such method lies in the interval [frozen, full FT], and ``$N$ simulates $M$'' becomes an explicit pointed map $h$ with $\rho_N\circ h=\rho_M$ (\thmref{Definition}{I.2}, \thmref{Observation}{I.4}). Isomorphism invariants finer than the image, such as $S_1$ and the singular locus, separate initialisations that the image cannot (\thmref{Observation}{I.6}, \thmref{Theorem}{II.2}). A categorical product produces a named method. Freezing either frame of LoRA gives two one-sided methods whose product is the sandwich $R\mapsto W_0+B_0RA_0$, which is LoRA-XS \citep{baazy2024lora} in the SVD frames of $W_0$ (\thmref{Proposition}{I.8}). Image-level meets, which are not products, organise others, such as HRA below. The bicategory $\Para$ is that of \citet{fong2017backprop} and \citet{cruttwell2021categorical}; the slice for PEFT and its invariants are new.

\item \textbf{No initialisation buys rank above $2r$} (\thmref{Theorem}{II.1}). For an additive method, moving the starting point while subtracting a frozen residual keeps the method pointed at $\theta_0$, and the union of all such re-pointed images is $\theta_0$ plus the difference set $C-C$ of its update shape. For $\mathrm{LoRA}_r$ that difference set is the matrices of rank at most $\min(2r,m,n)$. So no exact pointing of $\mathrm{LoRA}_r$, zero-init or split (PiSSA, OLoRA \citep{buyukakyuz2024olora}, CorDA \citep{yang2024corda}, LoRA-GA \citep{wang2024lorab}), makes an update of rank above $2r$ reachable in any adapted matrix, and split initialisations attain $2r$ when $2r\le\min(m,n)$. The statement is elementary; the $2r$ bound underlies the PiSSA-to-LoRA conversion of \citet{meng2024pissa}, implemented in Hugging Face (HF) PEFT.

\item \textbf{Three strata of initialisations, and two apexes} (\thmref{Theorem}{II.2}, \thmref{Theorem}{II.7}; Figure~\ref{fig:apex}). For LoRA of rank $r<\min(m,n)$ on one $m\times n$ weight there are three strata of exact full-rank pointings: $B_0=0$ with full-rank $A_0$ (vanilla LoRA on linear layers, EVA \citep{paischer2024parameter}); $A_0=0$ with full-rank $B_0$ (Init[B] \citep{hayou2024impact}, HF's default on embedding layers); and a rank-$r$ product $B_0A_0$ subtracted from the frozen weight (PiSSA, MiLoRA \citep{wang2024milora}, OLoRA, CorDA, LoRA-GA, Astra \citep{liu2026astra}). Two pointed LoRA methods in these strata are isomorphic over weight space exactly when they share $\operatorname{row}A_0$, $\operatorname{col}B_0$ or $B_0A_0$ respectively, so as a set the classes are $\Gr(r,n)\sqcup\Gr(r,m)\sqcup\MM_r$. Zero-factor pointings sit at the apex of the cone $W_0+\Mr$ and miss $r(n-r)$ or $r(m-r)$ first-order directions; split pointings sit at a smooth point, and neither image contains the other. The classification is set-level and in exact arithmetic. Rank-deficient pointings such as HF's \texttt{orthogonal} init, frozen-factor sub-objects (LoRA-FA \citep{zhang2023lora}, MiCA \citep{rudiger2026mica}) and quantised-base defects (QLoRA, LoftQ \citep{li2023loftq}) lie outside the three strata.

Let $W_0$ have full column rank and let $r$ be even with $2\le r\le n-2$. Then the image of $r$-reflection HRA \citep{yuan2024bridging} without Gram--Schmidt, HF PEFT's default (\texttt{apply\_GS=False}), is exactly $\{W_0R:R\in\SO(n),\ \rk(W_0R-W_0)\le r\}$. This is the image-level intersection of $\mathrm{LoRA}_r$ with the rotation orbit $W_0\cdot\SO(n)$, of dimension $rn-r(r+1)/2$, and HRA is not the categorical product of the two. At HF's default paired initialisation ($u_{2i-1}=u_{2i}$), $W_0$ is a singular vertex of the image and the first-order image has dimension $kn-k(k+1)/2$ with $k=r/2$, so that initialisation is an apex like zero-init LoRA. Both theorems are new.

\item \textbf{Exactly orthogonal methods keep the spectrum} (\thmref{Corollary}{II.6}, \thmref{Theorem}{V.3}). If every weight a method produces, including after merge-and-restart cycles, has the form $W_0R$ with $R$ exactly orthogonal on the input side, then in exact arithmetic the neuron Gram matrix $WW\T$ never changes. Every neuron norm, every angle between neurons (and with them the hyperspherical energy) and every singular value of $W_0$ stay fixed, however often the method is merged, and block OFT with a fixed partition also fixes each block Gram. This covers exact-Cayley OFT \citep{qiu2023controlling} and block OFT, strict GOFT \citep{ma2024parameter} without its optional scaler, HRA, and ETHER \citep{bini2024ether} as defined in its paper. It does not cover variants with a per-neuron or per-input scale (HF BOFT \citep{liu2023parameter}, HF OFT 0.14 to 0.15, GOFT's default code), output-side variants, or approximately orthogonal ones such as HF PEFT's default Cayley--Neumann OFT, qGOFT and ETHER+. With small generators each Cayley--Neumann merge is a contraction, so repeated merging can only shrink neuron norms and singular values; low-precision merges add rounding drift. The invariance itself is elementary, and preserving hyperspherical energy is the motivation of OFT \citep{qiu2023controlling}; new is the sorting of methods and implementations by whether they satisfy its hypothesis.

\item \textbf{Rank is a cut, dimension is a volume} (\thmref{Theorem}{II.10}, \thmref{Proposition}{II.12}; Figure~\ref{fig:rankvol}). The rank of a map computed by a tensor network is at most the smallest product of wire dimensions across any cut separating inputs from outputs (standard). The expressive dimension is at most the budget minus the dimension of the generic gauge orbit, with equality when the gauge is locally transitive on generic fibres. At equal budget, with $2r\le\min(m,n)$, $r^2\le\min(m,n)$ and $2r^2<m+n-1$, $\mathrm{LoHa}_{r,r}$ \citep{hyeonwoo2021fedpara} reaches at least $(m+n-1)-2r^2$ fewer dimensions than $\mathrm{LoRA}_{2r}$, about $m+n$ when $r^2\ll(m+n)/2$. It buys rank up to $r^2$ in exchange for $r\ge3$, and it is strictly dominated for $r\le2$. When $2r^2\ge m+n-1$, for example $r=64$ on $4096\times4096$ or LyCORIS-style $r=32$ on 320- to 640-wide layers, the dimension bound no longer separates the two, and wherever it is attained, as the numerical checks find, LoHa matches or exceeds $\mathrm{LoRA}_{2r}$ in dimension as well as in rank. GraLoRA \citep{jung2025gralora} ($k\times k$ blocks of rank $r/k$, HF default \texttt{hybrid\_r=0}) keeps LoRA's budget and dimension $r(m+n-r)$ and lifts the maximum rank to $\min(kr,m,n)$, but caps each block at rank $r/k$; for $k\ge2$ and $kr\le\min(m,n)$ neither image contains the other. The census is new.

\item \textbf{Kronecker sums cut the other way} (\thmref{Theorem}{II.11}; Figure~\ref{fig:cut}). Split the dimensions as $m=m_1m_2$ and $n=n_1n_2$. A $k$-term Kronecker sum $\sum_iC_i\otimes D_i$ is exactly a matrix of rank at most $k$ after the rearrangement of Van Loan and Pitsianis~\citeyearpar{loan1993approximation}, so unconstrained Kronecker methods are LoRA across a different cut of the same four-leg tensor. With equal square splits of a $d\times d$ weight, a generic rank-one update has the largest Kronecker rank, $d$, while the identity has Kronecker rank one. Hence zero-init $\mathrm{LoRA}_r$ and unconstrained $k$-term Kronecker sums, for $r,k<d$, have incomparable images. LoKr \citep{yeh2023navigating} updates $C\otimes BA$ with inner rank $r'\le\min(m_2,n_2)$ have rank at most $\min(m_1,n_1)\,r'$, attained, so LoKr's image lies inside that of zero-init $\mathrm{LoRA}_r$ exactly when $r\ge\min(m_1,n_1)\,r'$, and is incomparable with it otherwise when $\min(m_1,m_2)\min(n_1,n_2)>1$. The linear algebra is standard (the rearrangement and the operator-Schmidt decomposition); the PEFT reading is new.

\item \textbf{A conserved charge} (\thmref{Theorem}{III.5}, \thmref{Corollary}{III.6}, \thmref{Theorem}{III.9}; Figure~\ref{fig:noether}). With per-block rates, LoRA's charge
\[
  \Phi=B\T B/\eta_B-AA\T/\eta_A
\]
is conserved by every continuous-time update $\dot q=-\Lambda D\rho_q\T\gamma(t)$ with constant $\Lambda\succ0$, gradient or not, including minibatches, global-norm clipping and DoRA's \citep{liu2024dora} detached norm. The compact half of the gauge gives $\Lambda$-isometries that leave $K$ unchanged, so gauge-related initialisations give identical weight trajectories whenever the signal depends on $q$ only through $(t,\rho(q))$. With per-rank-channel rates, off-diagonal charges and isometries exist exactly between channels with equal products $\eta_{B,j}\eta_{A,j}$. Discrete steps make the charge drift, with relative drift of second order in the step size, and Adam and momentum are not covered. The law is a mild extension of known results: \citet[Prop.~5.1]{zhao2022symmetries} and \citet{marcotte2023abide}, after the balancedness principle of \citet{du2018algorithmic} and \citet{arora2018optimization} and the symmetry analysis of \citet{kunin2020neural}. New here is only the remark that $\gamma$ need not be a gradient. Under an isotropic charge the weight dynamics have a closed form, with a crossover scale below which zero-$B$ LoRA behaves like LoRA-FA; this closed form is due to \citet{xtarmoun2021understanding} and \citet{xmin2021explicit}, and the PEFT reading is ours.

\item \textbf{LoRA+ is $\alpha$ in disguise} (\thmref{Theorem}{III.10}, \thmref{Corollary}{III.11}; Figure~\ref{fig:orbit}). For zero-init LoRA under Adam with $\varepsilon=0$, no weight decay or clipping, one learning-rate schedule for both groups, and the same data order and dropout masks, LoRA+ \citep{hayou2024lora} with ratio $\lambda$ follows, in exact arithmetic, the same weight trajectory as LoRA with $\alpha$ multiplied by $\lambda$. Per-block $\varepsilon$ and decoupled weight decay keep the statement once they are rescaled too, and under SGD the factor is $\sqrt\lambda$. At a fixed rank, rsLoRA \citep{kalajdzievski2023rank} is LoRA with a rescaled $\alpha$ or $B$ learning rate; how $\alpha$ should scale with rank is a separate question. Both statements are instances of a hyperparameter torus: for a method whose update is $s$ times a multihomogeneous function of its parameter blocks, rescaling each block's initial scale and learning rate together with the overall scale $s$ (and, under Adam, the per-block $\varepsilon$, decoupled decay and clipping thresholds) leaves the weight trajectory exactly unchanged under Adam, given the same data order and dropout masks; under SGD the same holds with learning rates scaled by the square of the block factor. The LoRA/Adam case is due to \citet{schulman2025lora}; the general multihomogeneous torus and the SGD law are new.

\item \textbf{Adam sees a finite shadow of the gauge} (\thmref{Theorem}{III.3}, \thmref{Theorem}{III.12}, \thmref{Proposition}{III.4}; Figure~\ref{fig:gauge}). Gradient descent is natural exactly where cometrics agree, which for LoRA's linear gauge with block-scalar learning rates, at points where $A$ has full row rank and $r<n$, means exactly along $\Ort(r)$. On LoRA's full-rank locus with channel-uniform hyperparameters, the largest subgroup of $\GL_r$ under which the parameter update is equivariant is $\Ort(r)$ for SGD, with or without momentum, and the signed permutations $B_r$, of order $2^rr!$, for Adam and AdamW with any $\beta$'s, any $\varepsilon\ge0$ and decoupled decay. Positive diagonal rescalings become symmetries of Adam only jointly with per-rank-channel rescaling of the learning rates (and of $\varepsilon$ and learning-rate-coupled decay when these are nonzero), and only without global-norm clipping. Undamped scaled GD (Riemannian preconditioning) and LoRA-RITE \citep{yen2024lora} are equivariant under all of $\GL_r$; damping reduces scaled GD to $\Ort(r)$, and wrapping it in Adam reduces it to $B_r$. Combination rules must be gauge-natural as well. A rule for combining LoRA factors is a well-defined function of the adapters exactly when it is constant on the factorisation fibres, which task arithmetic on the products satisfies and post-hoc product-of-sums rules such as LoraHub \citep{huang2023lorahub} do not (\thmref{Proposition}{V.4}). The hierarchy is new; scaled GD is due to \citet{xtong2021accelerating} and, for LoRA, \citet{zhang2024riemannian}.

\item \textbf{GaLore is one-sided LoRA with a moving frame} (\thmref{Proposition}{IV.2}). If an adapter is linear in its trainable tensor ($W=W_k+\mathcal LC$ on the current base $W_k$, with $C_0=0$) and the optimizer depends only on the gradients it receives (SGD or momentum, Adam without weight decay, Adafactor without parameter scaling, no global-norm clipping, the same implementation on both sides), then training $C$ and merging gives exactly the same weights as running that optimizer on the compressed gradient $\mathcal L^*\nabla_WL$ and mapping back with $\mathcal L$. In particular, GaLore with weight decay $0$ is exactly one-sided LoRA whose frozen factor is GaLore's projector matrix, merged and re-initialised every period with the trainable factor's optimizer state carried over. Decoupled weight decay, parameter-scaled Adafactor, trust-ratio optimizers and full-gradient clipping break the identity, and Flora, which compresses only the momentum, is not covered. Known: \citet{xtorrobahennigen2025duality} for GaLore, and \citet{hao2024flora} for random projections.

\item \textbf{Base change has explicit defects and closures} (\thmref{Proposition}{V.2}, \thmref{Theorem}{V.3}; Figure~\ref{fig:rebase}). Quantising the frozen base moves the base point. At initialisation QLoRA misses the pretrained weight $W$ by the generically full-rank error $e=\kappa W-W$, and it can reach $W$ only if $\rk e\le r$. Splitting or error-fitting initialisations change this defect in an order-dependent way: split-then-quantise (QPiSSA) leaves $\kappa(W_{\mathrm{res}})-W_{\mathrm{res}}$, and one-step LoftQ leaves $e-s\,\mathrm{SVD}_r(e)$, which is $e$ minus its best rank-$r$ approximation only when the LoRA scale $s=\alpha/r$ is $1$, because HF's LoftQ initialisation ignores the scale. After $K$ merge-and-restart cycles with a fixed shape, the reachable set is $\theta_0+C^{+K}$ for an additive shape $C$, or $S^K\theta_0$ for a set $S$ of group elements containing the identity. Methods whose shape is a fixed subspace, subgroup or submonoid gain nothing, and ReLoRA reaches exactly $W_0+\{\rk\le\min(Kr,m,n)\}$. Exact-Cayley block OFT stays in $W_0\cdot\prod\SO(b)$, up to closure, unless its block hypergraph is connected, which BOFT's butterflies achieve only at full depth, and re-HRA reaches $(W_0+\MM_{\le Kr})\cap W_0\cdot\SO(n)$. New.

\item \textbf{The activation decides the merge} (\thmref{Theorem}{VI.3}, \thmref{Theorem}{VI.5}; Figure~\ref{fig:prefix}). A few exact rewrite rules give sufficient conditions for folding a scaling or affine PEFT edit into existing weights with no change in output. The feed-forward scale of (IA)$^3$ \citep{liu2022few} always folds into $W_{\mathrm{down}}$. It folds into $W_{\mathrm{up}}$ under SwiGLU or GeGLU for every scale and under ReLU for nonnegative ones, while under GELU only entries $0$ and $1$ fold. Position-selective edits (ReFT \citep{wu2024reft}, Activated LoRA \citep{greenewald2025activated}) and nonlinear adapters are not box-locally mergeable, and exactly merging a generic QLoRA or LoftQ update leaves the quantised format, with QA-LoRA \citep{xu2023qa} as the exception. For attention, a prefix \citep{li2021prefix} acts as an input-gated parallel adapter \citep{he2021unified} that rescales the content attention weights within its layer and never reorders them \citep{petrov2023prompting}, while LLaMA-Adapter's \citep{zhang2023llama} separate softmax with a zero-initialised gate leaves those weights unchanged. The rules are elementary, and the merge into $W_{\mathrm{down}}$ is noted by \citet{liu2022few}; SmoothQuant \citep{xxiao2023smoothquant} and AWQ \citep{xlin2024awq} already migrate scales through activations and meet the GELU obstruction. New is only the packaging as a merge calculus for PEFT, with explicit side conditions. The calculus gives sufficient conditions and box-local obstructions; deciding mergeability of a whole network lies outside its scope.

\item \textbf{An atlas} (Exposés~\ref{sec:examples}--\ref{sec:rosetta}, Appendix~\ref{app:catalogue}; Figures~\ref{fig:timeline} and~\ref{fig:graph}). Fourteen methods are worked through the three invariants. Seven coordinates, each read off a method's formula (kind, shape of the image germ, base point, gauge, covariance group, merge type and rebasing closure), classify \nMethods{} methods from \nPapers{} papers, \nHF{} of them in HF PEFT. The atlas records \nArrows{} simulation arrows between methods, each exhibited by an explicit map $h$ with $\rho_N\circ h=\rho_M$, and \nObstructions{} obstructions, each naming the test that rules an arrow out. The \nCore{} core methods form the periodic table and the category drawn in Figure~\ref{fig:graph}. A Rosetta stone translates between the PEFT and the categorical vocabulary.

\item \textbf{Predictions and limits} (Exposés~\ref{sec:predictions} and~\ref{sec:analogy}). The theory makes falsifiable predictions, each with its optimizer scope, and states plainly which of its words are analogies and which problems are open.
\end{enumerate}

Some further results are classical or standard and carry their credit in their headings: the complete invariants of the basic actions (\thmref{Theorem}{II.5}), the Kempf--Ness slice of the LoRA gauge (\thmref{Proposition}{III.7}; \citealp{xkempf1979length}) and the privileged basis fixed by an activation (\thmref{Lemma}{VI.2}).

\paragraph{Where the category theory earns its keep}
There are seven places.
\begin{enumerate}[label=\arabic*.]
\item The pointed slice gives the interval [frozen, full FT] for weight-space methods and makes ``simulation'' precise (Exposé~\ref{sec:slice}).
\item Isomorphism invariants finer than the image ($S_1$, the singular locus) classify initialisations (\thmref{Theorem}{II.2}).
\item A categorical product produces a named method (LoRA-XS, \thmref{Proposition}{I.8}), and image-level meets organise others (HRA, \thmref{Theorem}{II.7}).
\item Base change separates function-preserving reparametrisations, which act on the slice, from maps that move the base point (quantisers). The resulting pointing defects are elementary, but the slice makes their order dependence explicit (\thmref{Observation}{II.15}, \thmref{Proposition}{V.2}).
\item The lens 2-functor $\Para(\mathsf R)$ separates tangent-side from cotangent-side methods. The degree-one collapse shows that every cotangent method whose optimizer reads only its gradients (no weight decay, no parameter scaling, no full-gradient clipping) is, period by period, a forward linear method on a moving base (\thmref{Proposition}{IV.2}).
\item Naturality: gradient descent is natural exactly where cometrics agree (for LoRA's linear gauge with block-scalar rates, along $\Ort(r)$ where $A$ has full row rank and $r<n$), and combination rules must be gauge-natural (\thmref{Theorem}{III.3}, \thmref{Proposition}{V.4}).
\item Symmetric monoidal structure turns rank into a min-cut (\thmref{Theorem}{II.10}).
\end{enumerate}
The rest is real algebraic geometry, invariant theory, Lie generation and gradient-flow ODEs, and we say so where it happens. True but idle constructions are omitted: coends for the rank filtration, a saturation monad, sheaf language for locality, the Grothendieck construction $\int\Meth\to\mathbf{PNet}$, and coKleisli 2-cells for mixtures.

\subsubsection{How the results were checked}

Each of the \nPropositions{} numbered results was checked by adversarial model referees (Appendix~\ref{app:making-of}) against its proof, against the literature, and against the methods as published and as coded, including the source of HF PEFT at commit 85be3f3 of 30 September 2026. Of these, 14 stood as first written and 27 were repaired. A second round, with one mathematical and one faithfulness referee for each result, re-examined 26 of the repaired results and narrowed many of them. Its objections mostly concerned a side clause: a missing hypothesis, a remark that went past its proof, or a description of a method that matched only one implementation. For example, HF PEFT's LoftQ initialiser turned out to ignore the LoRA scale $\alpha/r$ (\thmref{Proposition}{V.2}). Only the narrowed forms are stated in this paper. Some results are new, such as the HRA apex; many are known and re-derived with credit, as the contributions above record.

\subsubsection{How to read this paper}

\paragraph{Plan}
Exposé~\ref{sec:slice} sets up the pointed slice and its invariants. Exposé~\ref{sec:image} studies the image germ: initialisations and their strata (\ref{sec:strata}), group-type methods (\ref{sec:erlangen}), and cut and volume (\ref{sec:cut}). Exposé~\ref{sec:fibres} treats fibres, metrics and training dynamics, Exposé~\ref{sec:lens} the backpropagation lens, Exposé~\ref{sec:base} base change, and Exposé~\ref{sec:merge} extensions and mergeability. Exposé~\ref{sec:examples} works through fourteen methods, Exposé~\ref{sec:atlas} gives the classification and Exposé~\ref{sec:rosetta} the Rosetta stone. Exposé~\ref{sec:predictions} lists falsifiable predictions, Exposé~\ref{sec:analogy} separates analogy from theorem and collects open problems, Exposé~\ref{sec:related} discusses related work and Exposé~\ref{sec:conclusion} concludes.

\paragraph{Labels}
Table~\ref{tab:labels} lists the labels. Every numbered item keeps the number it has on the companion site. Definitions, remarks and results share one counter within each exposé, so Definition I.2 is followed by Remark I.3 and Observation I.4.

\begin{table}[t]
\centering\small
\def\thesisbar#1{\textcolor{#1}{\rule[-0.35ex]{1.1pt}{2.2ex}}\enspace}
\caption{How to read the labels. The coloured rule is the one that marks the environment in the text.}
\label{tab:labels}
\begin{tabularx}{\linewidth}{>{\raggedright\arraybackslash}p{0.29\linewidth}X}
\toprule
\headrow \textbf{Label} & \textbf{What it means here}\\
\midrule
\thesisbar{tide}Theorem, Proposition, Lemma, Corollary & Proved in this paper, with the complete proof in Appendix~\ref{app:proofs}; the statement is set in italic. A heading that says \emph{known} and names authors marks a result due to them, which we re-derive in our language, adding only its reading for fine-tuning.\\
\thesisbar{tide!70}Observation & True and formally easy; kept because it organises.\\
\thesisbar{ochre}Definition & A definition.\\
\thesisbar{rulegray}Remark, Example & Commentary and instances, in upright type.\\
\thesisbar{wongpurple}Conjecture, [Conj] & A conjecture.\\
\thesisbar{wongvermillion!80}Prediction, [Pred] & A falsifiable prediction, stated with its optimizer scope.\\
\tikz[baseline=0pt]\draw[inkgray,line width=0.9pt,dash pattern=on 2.5pt off 2pt](0,-0.35ex)--(0,1.85ex);\enspace Analogy & Organising language with no theorem behind it. \emph{Noether}, \emph{Erlangen}, \emph{gauge} and \emph{moduli} are meant this way beyond what is proved (Exposé~\ref{sec:analogy}).\\
\textcolor{tide}{\sffamily\scriptsize\bfseries\textls[120]{WHAT THIS EXPLAINS}} & What a result says about methods as published and as implemented.\\
{[Num]} & Checked numerically in float64 (autograd Jacobian ranks, exact optimizer runs) by the ancillary script described in Appendix~\ref{app:repro}. The companion site's figures recompute their numbers in the browser.\\
{[Sketch]} & Only a proof sketch is given, in Appendix~\ref{app:proofs}; the claim is not counted as proved.\\
verified, repaired & Review status of a numbered result, recorded in the ancillary file \texttt{propositions.json}. Repaired results were corrected or narrowed, and only that form is stated.\\
\bottomrule
\end{tabularx}
\end{table}

\paragraph{Conventions}
These carry real weight in the arguments. Table~\ref{tab:notation} collects the notation.
\begin{itemize}
\item \emph{Linear box.} A linear box computes $y=Wx$ with $W\in\R^{m\times n}$, $m=d_{\mathrm{out}}$ and $n=d_{\mathrm{in}}$. Row $i$ of $W$ is output neuron $i$.
\item \emph{Two Gram matrices, two sides.} $K_{\mathrm{out}}(W)=WW\T$ is the neuron Gram matrix and $K_{\mathrm{in}}(W)=W\T W$ the input Gram matrix. A right multiplication $W\mapsto WR$ with $R\in\Ort(n)$ (the \emph{input side}) fixes $K_{\mathrm{out}}$. A left multiplication $W\mapsto PW$ fixes $K_{\mathrm{in}}$. Papers that store $W\T$ swap the words \emph{input} and \emph{output}; we translate every method into the convention above.
\item \emph{Ambient category.} Maps and dynamics live in $\Diff$. Images are taken in $\Set$. Every image we meet is semialgebraic, or definable in $\R_{\exp}$ when softmax is involved, so dimensions, smooth points and tangent cones are well defined.
\end{itemize}

\begin{table}[tbp]
\centering\small
\caption{Notation.}
\label{tab:notation}
\begin{tabularx}{\linewidth}{>{\raggedright\arraybackslash}p{0.19\linewidth}X}
\toprule
\headrow \textbf{Symbol} & \textbf{Meaning}\\
\midrule
$W$ & A linear box $y=Wx$, $W\in\R^{m\times n}$, with $m=d_{\mathrm{out}}$ and $n=d_{\mathrm{in}}$. Row $i$ is output neuron $i$.\\
$\Theta$ & Weight space, a real vector space of adapted and frozen boxes.\\
$\theta_0$, $W_0$ & The pretrained point; $W_0$ for one box.\\
$\rho$, $(Q,q_0)$ & The method, a pointed smooth map from trainable parameters $Q$ started at $q_0$, with $\rho(q_0)=\theta_0$.\\
$\Meth(\Theta,\theta_0)$ & The pointed slice $\Diff_*/(\Theta,\theta_0)$, whose objects are methods $M=(Q,q_0,\rho)$. A morphism $h:M\to N$ is a pointed smooth map with $\rho_N\circ h=\rho_M$, read ``$N$ simulates $M$''.\\
$\lvert M\rvert$ & The budget, $\dim Q$.\\
$d(M)$ & The expressive dimension $\dim\Im M=\max_q\rk d\rho_q$. The gap $\lvert M\rvert-d(M)$ is gauge waste.\\
$S_1$ & The first-order image $d\rho_{q_0}(T_{q_0}Q)$. The base point is regular if $\dim S_1=d(M)$, an apex if smaller.\\
$K$ & The induced cometric $D\rho\,\Lambda\,D\rho\T$, with $\Lambda\succ0$ the learning rates. Gradient flow is $\dot\theta=-K\nabla_\theta L$.\\
$K_{\mathrm{out}}$, $K_{\mathrm{in}}$ & The neuron Gram matrix $WW\T$ and the input Gram matrix $W\T W$.\\
$\Mr$, $\MM_r$ & Matrices of rank at most $r$; the smooth stratum of rank exactly $r$, of dimension $r(m+n-r)$. The singular locus of $\Mr$ is $\MM_{\le r-1}$.\\
$\mathrm{LoRA}_r$ & $\rho(B,A)=W_0+sBA$ with $B\in\R^{m\times r}$, $A\in\R^{r\times n}$ and $s=\alpha/r$.\\
$G$, $\GL_r$ & The weight gradient $\nabla_WL$; the LoRA gauge, $g\cdot(B,A)=(Bg^{-1},gA)$.\\
$T_m$, $T_m^+$ & The diagonal torus of invertible diagonal matrices; its identity component $\{\diag(d):d_i>0\}$.\\
$\mathrm{Diag}_m$ & The monoid of all diagonal matrices.\\
$\mathrm{Mon}_m$ & The monomial matrices, $T_m\rtimes S_m$.\\
$B_r$ & The signed permutations, $\{\pm1\}^r\rtimes S_r$, of order $2^rr!$.\\
$\CO(n)$ & The conformal orthogonal group $\R_{>0}\cdot\Ort(n)$; $\CO^+(2)\cong\mathbb C^\times$.\\
\bottomrule
\end{tabularx}
\end{table}

\paragraph{Proofs and appendices}
The main text contains no proofs. Complete proofs of all numbered results are in Appendix~\ref{app:proofs}, in the same order as the statements, one subsection per result. External theorems that the proofs use are quoted in Appendix~\ref{app:auxiliary}. Appendix~\ref{app:background} collects the background the paper assumes, on categories, monoidal categories and lenses, manifolds, germs and cones, and groups, symmetry and conservation; its entries keep the numbers A.1--D.34 they carry on the companion site, and readers new to category theory or Lie groups can start there. Appendix~\ref{app:catalogue} describes the catalogue and Appendix~\ref{app:repro} the numerical checks.

\paragraph{The companion site}
This paper is the long-form version of the interactive site at \url{https://xiaoyulics.com/peft-atlas/}. The two share every number, so ``Theorem III.10'' names the same statement here, on the site and in the ancillary file \texttt{propositions.json}. The site states each result with its proof beneath it and has 20 interactive figures that recompute their numbers in the browser, among them an orbit explorer for LoRA+, a Gram lab for orthogonal methods, a gauge lab, Noether hyperbolas and a merge game. This paper gives the complete text, collects the proofs in Appendix~\ref{app:proofs}, and replaces the interactive figures by static ones. The catalogue, the arrows and obstructions, the numbered results with their review status, and the numerical checks accompany this paper as data files. When citing a result, please give its number; for a result marked known, please also cite the authors named in its heading, to whom the result itself is due.

\section{The pointed slice}\label{sec:slice}

We model a weight-space PEFT method as a pointed map into the pretrained weights. The frozen model and full fine-tuning sit
at the two ends of an interval, and sums and products of such maps already give methods in use. This exposé sets up the
category and its first invariants. The theorems begin when we look closely at one image near one point, which is where
Exposé~\ref{sec:image} starts.

\subsubsection{Pretrained models and methods}\label{sec:slice-methods}

A pretrained network is a function with a parameter slot and a chosen value in it. Its natural home is the bicategory of
parametrised maps (\bgref{B.16}).

\begin{definition}{I.1}{Para; pretrained model}
For a cartesian category $\mathcal C$, the bicategory $\Para(\mathcal C)$ is defined as follows
\citep{fong2017backprop,cruttwell2021categorical}:
\begin{itemize}
  \item its 1-cells $X\to Y$ are pairs $(P,\,f:P\times X\to Y)$;
  \item its 2-cells $(P,f)\Rightarrow(Q,g)$ are maps $r:Q\to P$ with $g=f\circ(r\times X)$.
\end{itemize}
A \textbf{pretrained model} is a pointed 1-cell $F=(\Theta,\theta_0,f)$ of $\Para(\Diff)$. Its \textbf{realisation} is
$\Phi_f:\Theta\to C^\infty(X,Y)$, $\theta\mapsto f(\theta,-)$. Fine-tuning is a path $\theta(t)$ with $\theta(0)=\theta_0$.
\end{definition}

$\Para$ and the lens reading of backpropagation are due to \citet{fong2017backprop} and \citet{cruttwell2021categorical};
\citet{gavranovic2024position} develop a broader programme on the same foundations.

\begin{definition}{I.2}{methods and simulations}
A weight-space \textbf{PEFT method} at $\theta_0$ is a pointed smooth map $M=(Q,q_0,\rho)$, $\rho:Q\to\Theta$, with
$\rho(q_0)=\theta_0$. Usually $\dim Q\ll\dim\Theta$, and $\theta_0$ (or a splitting of it) is frozen inside $\rho$. These
form the pointed slice
\begin{equation*}
  \Meth(\Theta,\theta_0):=\Diff_*/(\Theta,\theta_0).
\end{equation*}
\begin{itemize}
  \item A \textbf{morphism} $h:M\to N$ is a pointed smooth map with $\rho_N\circ h=\rho_M$. Read it as ``$N$ simulates $M$''.
    Then $N$ reproduces every weight that $M$ reaches, together with its parameter path. A \emph{weak} morphism drops the
    condition $h(q_0)=q_0'$.
  \item The \textbf{budget} is $\lvert M\rvert=\dim Q$.
  \item $M$ is \textbf{additive} if $\rho=\theta_0+\delta$ with $\delta(q_0)=0$ and $\delta$ independent of $\theta_0$.
  \item $M$ is of \textbf{action type} if $\rho(q)=\gamma(q)\cdot\theta_0$ for a map $\gamma$ into a group acting on $\Theta$.
\end{itemize}
\end{definition}

\begin{example*}{four objects}
On one $m\times n$ box, LoRA \citep{hu2021lora} is $\rho(B,A)=W_0+sBA$ on $Q=\R^{m\times r}\times\R^{r\times n}$, pointed
at $(0,A_0)$. It is additive, with budget $r(m+n)$. OFT \citep{qiu2023controlling} with the exact Cayley map
(\bgref{D.7}) is $\rho(Q)=W_0\,C(Q)$ for skew $Q$ (block-diagonal in OFT), pointed at $Q=0$. It is of action type, with
$\gamma=C$ landing in $\SO(n)$ on the input side. The frozen model is $(\mathrm{pt},*,\theta_0)$, the inclusion of the
single point $\theta_0$, and full fine-tuning is $(\Theta,\theta_0,\id)$, the identity of $\Theta$.
\end{example*}

\begin{figure}[tbp]
\centering
{\small
\begin{tikzcd}[column sep=6.2em,row sep=0.5em]
  (\mathrm{pt},*) \arrow[r,"{*\,\mapsto\, q_0}"] \arrow[rr,bend left=24,inkgray,"{*\,\mapsto\,\theta_0}"]
  & (Q_M,q_0) \arrow[r,tide,"\rho_M","\textsf{merge}"']
  & (\Theta,\theta_0) \\
  |[inkgray]| \textsf{Frozen (initial)} & |[inkgray]| M & |[inkgray]| \textsf{Full FT (terminal)}
\end{tikzcd}}

\smallskip
{\sffamily\footnotesize\color{inkgray}(a) the interval [frozen, full FT]}

\bigskip
\begin{minipage}[b]{0.42\linewidth}\centering
{\small
\begin{tikzcd}[column sep=1.6em,row sep=3.0em]
  (Q_M,q_0) \arrow[rr,"h"] \arrow[dr,tide,"\rho_M"'] & & (Q_N,q_0') \arrow[dl,tide,"\rho_N"] \\
  & (\Theta,\theta_0) &
\end{tikzcd}}

\smallskip
{\sffamily\footnotesize\color{inkgray}(b) $N$ simulates $M$: $\rho_N\circ h=\rho_M$}
\end{minipage}\hfill
\begin{minipage}[b]{0.56\linewidth}\centering
{\small
\begin{tikzcd}[column sep=0.6em,row sep=3.0em]
  (\R^{m\times r},0) \arrow[rr,"{B\,\mapsto\,(B,A_0)}"] \arrow[dr,tide,"{W_0+sBA_0}"']
  & & (\R^{m\times r}\times\R^{r\times n},(0,A_0)) \arrow[dl,tide,"{W_0+sBA}"] \\
  & (\R^{m\times n},W_0) &
\end{tikzcd}}

\smallskip
{\sffamily\footnotesize\color{inkgray}(c) LoRA simulates LoRA-FA}
\end{minipage}
\caption{The pointed slice $\Meth(\Theta,\theta_0)$.
(a)~The interval of weight-space methods. For a method $M$, the first arrow picks the initialisation $q_0$; it lies over
$\Theta$ because $M$ starts at $\theta_0$. The second arrow is $\rho_M$ itself, the merge into the base weights, and the
composite is the inclusion of $\theta_0$. The frozen model is initial and full fine-tuning terminal
(\thmref{Observation}{I.4}). A pointing defect, $\rho_M(q_0)\neq\theta_0$, leaves the triangle commuting only up to that
defect (Exposé~\ref{sec:base}). Adapters and prefixes fall outside the interval because they extend the network
(\thmref{Definition}{VI.1}).
(b)~A simulation is a pointed smooth map $h$ that makes the triangle over $(\Theta,\theta_0)$ commute; a weak simulation
drops $h(q_0)=q_0'$.
(c)~An instance: LoRA pointed at $(0,A_0)$ simulates LoRA-FA through $h(B)=(B,A_0)$.}
\label{fig:interval}
\end{figure}
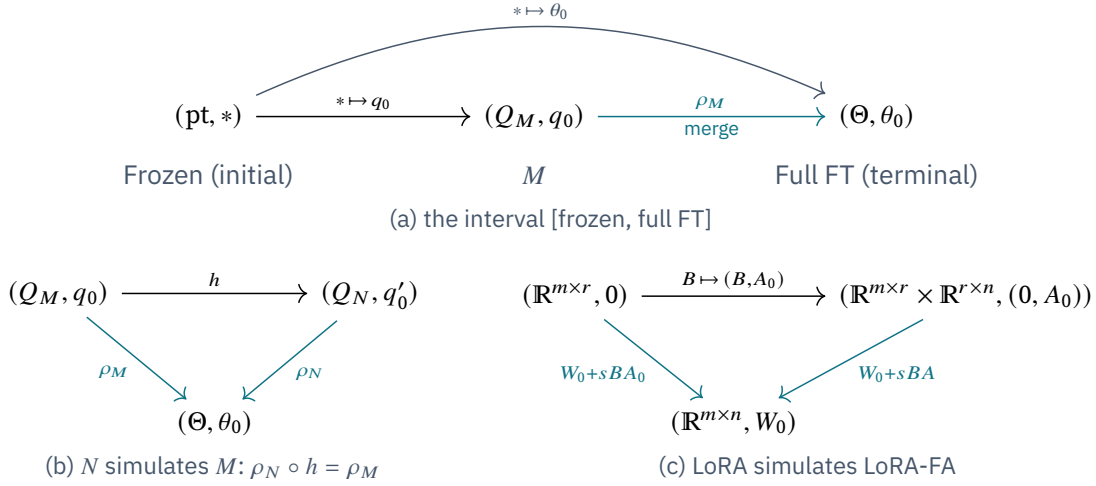

\begin{example*}{one arrow}
Take LoRA-FA \citep{zhang2023lora}, $B\mapsto W_0+sBA_0$ with $A_0$ frozen, pointed at $B=0$. The pointed map
$h(B)=(B,A_0)$ satisfies $\rho_{\rm LoRA}\circ h=\rho_{\rm FA}$, so LoRA pointed at $(0,A_0)$ simulates LoRA-FA
(Figure~\ref{fig:interval}c).
\end{example*}

Random-subspace training, $\rho(z)=\theta_0+Pz$ with a fixed random matrix $P$, is how \citet{li2018measuring} measured
intrinsic dimension and how \citet{aghajanyan2020intrinsic} fine-tuned language models in small subspaces. In the slice it
is a linear object, with $d(M)=\lvert M\rvert$ (\thmref{Definition}{I.5}) when $P$ has full column rank.

\subsubsection{What Para adds}\label{sec:slice-para}

\begin{remark}{I.3}{orientation, and what Para adds}
Each $\rho$ is a 2-cell $F\Rightarrow F^\rho:=(Q,f\circ(\rho\times X))$ \emph{out of} $F$. A simulation $h:M\to N$ is a
2-cell $F^{\rho_N}\Rightarrow F^{\rho_M}$, since $\rho_N\circ h=\rho_M$. Hence $\Meth(\Theta,\theta_0)$ is the
\textbf{opposite of the pointed coslice of reparametrisation 2-cells under $F$} (\bgref{A.7}). The slice over $F$ is a
different category, and its objects are extensions (Exposé~\ref{sec:merge}).

$\Meth$ depends only on $(\Theta,\theta_0)$ and not on $f$. So for weight-space methods, $\Para$ is bookkeeping. The
network enters in exactly three places, and we use $\Para$ only there:
\begin{itemize}
  \item extensions and mergeability (Exposé~\ref{sec:merge}; \thmref{Definition}{VI.1}, \thmref{Theorem}{VI.3});
  \item the realisation $\Phi_f$, through functional images and descent (\thmref{Observation}{II.15});
  \item the backpropagation lens (Exposé~\ref{sec:lens}; \thmref{Definition}{IV.1}, \thmref{Proposition}{IV.2}).
\end{itemize}
\end{remark}

Calling $\Para$ bookkeeping concerns this slice only. In the three places listed we use the framework of
\citet{fong2017backprop} and \citet{cruttwell2021categorical} directly, including their lens 2-functor in
Exposé~\ref{sec:lens}.

This is also why the next exposés work in weight space. Reach in $\Theta$ is a property of the image, and the cometric
$K=D\rho\,\Lambda\,D\rho\T$ that shapes training is built from $\rho$ and the optimizer's rates $\Lambda$. The network
enters training through the gradient it hands to the method, and enters expressivity through $\Phi_f$ once we ask about
functions instead of weights. So Exposés~\ref{sec:image} and~\ref{sec:fibres} work in $\Theta$ and flag the places where
$f$ matters.

\subsubsection{Frozen and full fine-tuning}\label{sec:slice-interval}

\begin{observation}{I.4}{universal properties}
In $\Meth(\Theta,\theta_0)$:

(a) the frozen model $(\mathrm{pt},*,\theta_0)$ is initial, and full fine-tuning $(\Theta,\theta_0,\id)$ is terminal. The
unique morphism $M\to{}$Full FT is $\rho_M$ itself.

(b) If the fibre product $Q_M\times_\Theta Q_N$ (\bgref{A.11}) exists in $\Diff$, it is the product $M\times N$. It exists,
for instance, when the set-level fibre product is an embedded submanifold of $Q_M\times Q_N$: a clean intersection
(\bgref{C.7}), or linear or affine $\rho$'s as in \thmref{Proposition}{I.8}(a). Transversality $\rho_M\pitchfork\rho_N$ is
never the reason when $\lvert M\rvert+\lvert N\rvert<\dim\Theta$, since at $(q_0,q_0')$ it would force
$\lvert M\rvert+\lvert N\rvert\ge\dim\Theta$. In $\Set$ the fibre product always exists and
$\Im(M\times N)=\Im M\cap\Im N$. Whenever the product exists in $\Meth$, its underlying set is this set-level fibre
product, so the image identity holds in every case. The product can fail to exist. On $\Theta=\R$ ($1\times1$ matrices),
$\mathrm{LoRA}_1\times\text{Frozen}$ would have the cross $\{BA=0\}\subset\R^2$ as underlying set. No manifold structure
on the cross has the universal property, because both axes would have to lift smoothly through the point over the origin,
which forces rank $2$ there.

(c) $\Im$ is a functor (\bgref{A.4}) to the poset (\bgref{A.16}) of subsets of $\Theta$ containing $\theta_0$. Conversely,
$\Im M\subseteq\Im N$ yields a morphism of the set-level slice $\Set_*/(\Theta,\theta_0)$, though not necessarily a smooth
one. Take $\rho_M(t)=t$ and $\rho_N(u)=u^3$ on $\R$, pointed at $0$. The images agree, yet no smooth $h:\R\to\R$ satisfies
$\rho_N\circ h=\id$, that is, $h(t)^3=t$.
\end{observation}

\noindent The proof is in Appendix~\ref{proof:I.4}. The obstruction in (b) covers every pointing of $\mathrm{LoRA}_1$,
including the standard one $(0,a_0)$ with $a_0\neq0$, whose base point is a smooth point of the cross; the obstruction sits
at the origin.

\begin{explains}
Every weight-space method pointed at $\theta_0$ lies in the interval [frozen, full FT]. Adapters, prefixes and ReFT
\citep{wu2024reft} are extensions outside $\Meth$ (Exposé~\ref{sec:merge}). ``Merge into the base weights'' is the
canonical morphism to the terminal object. In exact arithmetic on a full-precision base this is definitional, since every
object of $\Meth$ is mergeable by construction. Type-preserving merges into quantised bases are lossy
(\thmref{Proposition}{V.2}(d)), combining several adapters is a separate question (\thmref{Proposition}{V.4}), and the
remaining zero-overhead questions live on extensions (Exposé~\ref{sec:merge}). What both $M$ and $N$ can express,
$\Im M\cap\Im N$, always exists as a set. The categorical product, when it exists, is the universal method simulated by
both, and its image is exactly this set. \thmref{Proposition}{I.8}(a) computes a genuine one (LoRA-XS).
\thmref{Proposition}{I.8}(b) and \thmref{Theorem}{II.7} compute meets, that is, intersections in the poset of subsets of
$\Theta$ containing $\theta_0$; they are not products in $\Meth$.
\end{explains}

\paragraph{Defaults that do not start at the checkpoint}
Some defaults move the weights before training starts. In Hugging Face PEFT \citep{xmangrulkar2022peft} the default
FourierFT \citep{gao2024parameter} draws its spectrum from a standard normal, and the default vector bank and logits of
VB-LoRA \citep{li2024vb} are random too. LoRA with \texttt{init\_lora\_weights=False}, LoRA-XS with its small random $R$, and every method on
a quantised base also start away from $\theta_0$. Each lives over its own starting weight or carries a pointing defect,
which Exposé~\ref{sec:base} measures.

\paragraph{Lifting obstructions beyond first order}
The gap between $\Set$ and $\Diff$ in (c) is a smooth lifting obstruction. \thmref{Observation}{I.6} below detects its
first-order instances. In the cube-root example, $S_1(M)=\R\not\subseteq S_1(N)=0$. Higher-order obstructions need finer
invariants. The maps $\rho_M(t)=t^3$ and $\rho_N(u)=u^5$ have equal images and $S_1=0$ on both sides, yet the only lift,
$t^{3/5}$, is not $C^1$.

\subsubsection{Invariants of a method}\label{sec:slice-invariants}

The budget $\lvert M\rvert$ counts parameters. Two other numbers say more. One is the dimension of what a method
reaches; the other is the dimension of what it reaches to first order at the start, which is what the first gradient steps
use. For LoRA with $r=16$
on a $4096\times4096$ matrix the budget is $131{,}072$, the image has dimension $r(m+n-r)=130{,}816$, and from zero
initialisation the first steps use only $mr=65{,}536$ directions (\thmref{Theorem}{II.2}, \thmref{Proposition}{II.12}).

\begin{definition}{I.5}{invariants of an object}
For $M=(Q,q_0,\rho)$:
\begin{itemize}
  \item the image $\Im M=\rho(Q)$;
  \item the \textbf{expressive dimension} $d(M)=\dim\Im M=\max_q\rk d\rho_q$;
  \item the \textbf{gauge waste} $g(M)=\lvert M\rvert-d(M)$;
  \item the \textbf{first-order image} $S_1(M)=d\rho_{q_0}(T_{q_0}Q)$, and the \textbf{first-order deficit}, which is
    $d(M)-\dim S_1(M)$;
  \item the \textbf{gauge group} $\{\varphi\in\mathrm{Diff}(Q):\rho\circ\varphi=\rho\}$, which is unpointed, since a gauge
    transformation moves $q_0$ inside its fibre. In practice we exhibit a Lie group (\bgref{D.1}) acting on $Q$ that
    preserves $\rho$ and is locally transitive on generic fibres, and call it the \textbf{generic gauge}.
\end{itemize}
The base point $\theta_0$ is \textbf{regular} for $M$ if the deficit is $0$, and an \textbf{apex} if the deficit is
positive. We say \textbf{vertex} for the subset-level notion, the cone point of an image germ (\bgref{C.10}), and keep
\textbf{apex} for the base-point notion. The two differ in general. At a crossing of two smooth branches the image is
singular, yet the deficit can be $0$. In every case below with a positive deficit, the image germ is (diffeomorphic to) a
cone with vertex $\theta_0$; the exception is DoRA \citep{liu2024dora}, which inherits LoRA's deficit through its torus
factor. A cone germ that is a $C^1$ manifold at its vertex equals its tangent space (\bgref{C.2}) there, so it is linear.
Hence the vertex of a non-linear cone is a singular point.
\end{definition}

\begin{remark*}{the word ``gauge''}
``Gauge'' here means the symmetry group of a parametrisation, as in \thmref{Definition}{I.5}. There is no connection, and
away from the full-rank stratum the fibres jump. The word borrows, as an analogy, the physicist's picture of redundant
coordinates that leave every observation unchanged. Only the group and its action enter the theorems.
\end{remark*}

If $N$ simulates $M$, whatever $M$ can do at its first step, $N$ can do too.

\begin{observation}{I.6}{$S_1$ is functorial}
If $M\to N$ is a (pointed) morphism, then $S_1(M)\subseteq S_1(N)$. In particular, isomorphic methods have equal $S_1$ and
equal first-order deficit. Monotonicity holds only along pointed morphisms. Along a weak morphism, such as a gauge map
moving $q_0$ inside its fibre, $S_1$ can change.
\end{observation}

\noindent The proof is in Appendix~\ref{proof:I.6}.

\begin{explains}
$S_1$ is the set of weight-space directions available to the linearised (lazy, NTK) dynamics at initialisation, since the
lazy-regime kernel $J_f\,d\rho\,\Lambda\,d\rho\T J_f\T$ has range $J_f(S_1)$. That kernel also depends on $\Lambda$ and on
the network Jacobian $J_f$, not only on $S_1$. For random-$A_0$ zero-init LoRA it approximates the full fine-tuning kernel
\citep{malladi2022kernel}, so a large first-order deficit need not cost much on the data. $S_1$ is an isomorphism
invariant that separates initialisations the image cannot, namely zero-init $\mathrm T_A$ against $\mathrm T_B$, and the
choice of $\operatorname{row}A_0$ (\thmref{Theorem}{II.2}). The image already separates split initialisations, since it
determines $P$.
\end{explains}

The kernel view of language-model fine-tuning, and the closeness of LoRA's kernel to full fine-tuning's, are due to
\citet{malladi2022kernel}. \citet{hayou2024impact} compared the zero initialisations $B_0=0$ and $A_0=0$ as training
procedures; in the slice they are already different objects.

\subsubsection{Sums}\label{sec:slice-sums}

Weight space is a vector space, so running two methods side by side and adding their updates gives a method again.

\begin{observation}{I.7}{additive monoidal structure}
$\Theta$ is a vector space, so every object of $\Meth(\Theta,\theta_0)$ can be written $\rho=\theta_0+\delta$ with
$\delta:=\rho-\theta_0$ and $\delta(q_0)=0$. Put
\begin{equation*}
  M\oplus N:=\big(Q_M\times Q_N,\ (q_0,q_0'),\ \theta_0+\delta_M+\delta_N\big).
\end{equation*}
This is a symmetric monoidal structure (\bgref{B.1}) with the frozen model as unit. Its image is the Minkowski sum
$\theta_0+(\Im\delta_M+\Im\delta_N)$ (\bgref{B.14}). It differs from the categorical product, whose image, when the
product exists, is the intersection $\Im M\cap\Im N$ (\thmref{Observation}{I.4}(b)). ``Additive'' in
\thmref{Definition}{I.2} ($\delta$ independent of $\theta_0$) is a condition on \emph{families}
$\theta_0\mapsto M(\theta_0)$; it matters for base change (Exposé~\ref{sec:base}) and plays no role here.

Write $\mathrm{LoRA}_k$ for LoRA with frozen residual $\theta_0$; for split pointings the sum below has residual
$\theta_0-P-P'$. Concatenation $h(B_1,A_1,B_2,A_2)=([B_1\ B_2],[A_1;A_2])$ is an isomorphism
$\mathrm{LoRA}_r\oplus\mathrm{LoRA}_s\cong\mathrm{LoRA}_{r+s}$ of unpointed objects. It is a pointed isomorphism onto
$\mathrm{LoRA}_{r+s}$ pointed at $([B_0\ B_0'],[A_0;A_0'])$. For $\mathrm T_A$ pointings ($B_0=0$, $\rk A_0=r$), the
target pointing has $\rk[A_0;A_0']=r+s$ if and only if $\operatorname{row}A_0\cap\operatorname{row}A_0'=0$; it lies in the
stratum $\mathrm T_A$ of \thmref{Theorem}{II.2} when moreover $r+s<\min(m,n)$.

Standard pointings ($\mathrm T_A$, or symmetrically $\mathrm T_B$) exist only for $r<\min(m,n)$ (\thmref{Theorem}{II.2}),
so the monoidal question is pointwise. If $F(r)$ and $F(2r)$ are standard ($2r<\min(m,n)$), then
$F(r)\oplus F(r)\not\cong F(2r)$ as pointed objects; for $2r\ge\min(m,n)$ no standard $F(2r)$ exists. Hence no assignment
$r\mapsto\mathrm{LoRA}_r$ with $F(1)$ and $F(2)$ standard is \textbf{strong monoidal} (\bgref{B.15}) into pointed objects.
For $r\neq s$ and standard pointings, $F(r)\oplus F(s)\cong F(r+s)$ as pointed objects exactly when the row space of the
target's $A_0$ is the direct sum of the two source row spaces. The assignment is strong monoidal on unpointed objects, up to
weak morphisms, and on pointed objects with degenerate pointings, namely the origin $(B_0,A_0)=(0,0)$, where $S_1=0$, or
the stacked pointings $F(r)=F(1)^{\oplus r}$. Concatenation is strictly associative and unital, so discrete $(\N,+)$
suffices for a plain strong monoidal functor. \emph{Symmetric} strong monoidality needs the permutation groupoid, since
swapping and then concatenating permutes the blocks.

With per-adapter scales $s_r$ ($\alpha/r$ for LoRA, $\alpha/\sqrt r$ for rsLoRA) and any target scale $s_*$, the map
$h=([B_1\ B_2],[c_1A_1;c_2A_2])$ with $c_i=s_{r_i}/s_*$, where $r_1=r$ and $r_2=s$, is an isomorphism over $\Theta$. It is
a linear bijection and preserves row spaces, so the pointing criterion is unchanged.
\end{observation}

\noindent The proof is in Appendix~\ref{proof:I.7}. The negative claim rests on $\dim S_1$, which any pointed
isomorphism preserves (\thmref{Observation}{I.6}). It is $mr$ for $F(r)\oplus F(r)$ and $2mr$ for a standard $F(2r)$.
[Num] At $m=7$, $\dim S_1=28$ for a standard $\mathrm{LoRA}_4$, against $14$ for $\mathrm{LoRA}_2\oplus\mathrm{LoRA}_2$
with the same $A_0$ and $28$ with independent $A_0$'s; $F(1)\oplus F(2)$ and a standard $F(3)$ both give $21$
(Appendix~\ref{app:repro}).

\paragraph{The \texttt{cat} merge of Hugging Face PEFT}
Hugging Face PEFT's \texttt{add\_weighted\_adapter(..., combination\_type="cat")} \citep{xmangrulkar2022peft} evaluates the
rescaled concatenation $h$ at trained points. It takes $s_*=1$. It sets the merged \texttt{lora\_alpha} to the new rank
$r+s$, multiplies each $A$ block by $w_is_{r_i}$, so that $c_i=w_is_{r_i}$, and leaves the $B$ blocks alone. It is the
isomorphism $h$ when all weights $w_i=1$ and the merged adapter has \texttt{use\_rslora=False}. In PEFT releases up to
v0.21.1 the merged adapter inherits \texttt{use\_rslora} from the first source. For rsLoRA sources
\citep{kalajdzievski2023rank} its scale is then $(r+s)/\sqrt{r+s}=\sqrt{r+s}$, and the merged update is $\sqrt{r+s}$ times
too large. PR \#3449 on the main branch fixes this by setting \texttt{use\_rslora=False} on the merged adapter.

\begin{explains}
RoSA \citep{nikdan2024rosa} (LoRA $\oplus$ a fixed-mask sparse method) and GraLoRA \citep{jung2025gralora} ($\oplus$ of
$k^2$ block-embedded LoRAs) are $\oplus$-words, and their expressivity is a Minkowski sum. ReLoRA \citep{lialin2023relora}
is $\mathrm{LoRA}^{\oplus K}$ unrolled in time, apart from its full-rank warm start. A sum of $k$ rank-$r$ LoRAs, each with
residual $\theta_0$, is isomorphic in $\Meth$ to one LoRA of rank $kr$, pointed at the concatenated initialisation. Under
$\alpha/r$ scaling the isomorphism rescales the $A$ blocks, so training at equal $\alpha$ and learning rate follows a
different trajectory. The \texttt{cat} merge is the isomorphism $h$ evaluated at trained points; the merged adapter, if
unfrozen, is $\mathrm{LoRA}_{r+s}$.
\end{explains}

\paragraph{What is and is not a sum}
LoRA-soup CAT \citep{prabhakar2024lora} freezes the trained modules and learns layer-wise weights, so it is a $\oplus$ of
$k$ frozen line methods $\alpha_i\mapsto\alpha_iB_iA_i$. Its per-layer image is a span of dimension at most $k$
(generically exactly $k$), much smaller than the image of $\mathrm{LoRA}_{kr}$. UniPELT \citep{mao2021unipelt} and MAM
\citep{he2021unified} are \textbf{not} $\oplus$-words in $\Meth(\Theta,\theta_0)$. Their adapter and prefix parts live on
extensions (Exposé~\ref{sec:merge}), and UniPELT's gates depend on the input. The obstruction in
\thmref{Observation}{I.7} is the reuse of one pointing. With a fixed pointing, stacking the same $A_0$ twice gives a
rank-deficient frame, which $S_1$ distinguishes from a standard $\mathrm{LoRA}_{2r}$.

\subsubsection{Products}\label{sec:slice-products}

Freeze $A$ at $A_0$ and train $B$, or freeze $B$ at $B_0$ and train $A$. The product of these two one-sided methods is
LoRA-XS. As a pullback square over $\Theta$ it reads
\begin{equation*}
\begin{tikzcd}[column sep=6.5em,row sep=2.8em]
  \R^{r\times r} \arrow[r,"{R\,\mapsto\, B_0R}"] \arrow[d,"{R\,\mapsto\, RA_0}"'] \arrow[dr,phantom,"\lrcorner",pos=0.09]
  & Q_{\rm FA}=\R^{m\times r} \arrow[d,tide,"{B\,\mapsto\, W_0+BA_0}"] \\
  Q_{\rm FB}=\R^{r\times n} \arrow[r,tide,"{A\,\mapsto\, W_0+B_0A}"] & \Theta=\R^{m\times n}
\end{tikzcd}
\end{equation*}
and both composites send $R$ to $W_0+B_0RA_0$.

\begin{proposition}{I.8}{a product that produces a named method; an image-level meet}
(a) Let $A_0\in\R^{r\times n}$ have full row rank and $B_0\in\R^{m\times r}$ full column rank. Define the one-sided methods
\begin{equation*}
  \mathrm{FA}(A_0):B\mapsto W_0+BA_0,\qquad \mathrm{FB}(B_0):A\mapsto W_0+B_0A,
\end{equation*}
both pointed at $0$. Then
\begin{equation*}
  \mathrm{FA}(A_0)\times\mathrm{FB}(B_0)\;\cong\;\big(\R^{r\times r},\ 0,\ R\mapsto W_0+B_0RA_0\big).
\end{equation*}
The product supplies this sandwich for \emph{any} frames. Assume $\rk W_0\ge r$ ($\sigma_r>0$) and take the SVD frames
$B_0=U_r$, $A_0=\Sigma_rV_r\T$. The result is \textbf{LoRA-XS pointed at $R=0$}.

(b) Let $W_0$ have full row rank (so $m\le n$), and let $\mathrm{LoRA}_r$ carry its standard pointing ($B_0=0$, image
$W_0+\Mr$). The image-level meet of $\mathrm{LoRA}_r$ with the output-scaling method $\ell\mapsto\diag(\ell)W_0$ is
$\{\diag(\ell)W_0:\#\{i:\ell_i\neq1\}\le r\}$, that is, ``rescale at most $r$ output neurons''.

\upshape The LoRA-XS paper \citep{baazy2024lora} writes $xW+xARB$ in row-vector convention with $A=U_r\Sigma_r$ and
$B=V_r\T$. Transposed, the update is $B_0R\T A_0$, so the isomorphism is $R\mapsto\alpha R\T$. LoRA-XS initialises
$R\sim\mathcal N(0,\sigma^2)$ with $\sigma=10^{-5}$, a negligible pointing defect. The output-scaling method in (b) is the
key/value part of (IA)$^3$ \citep{liu2022few}. On the FFN, (IA)$^3$ scales the \emph{input} of $W_{\rm down}$,
$W\mapsto W\diag(\ell)$; the mirror statement then needs full column rank, or holds generically for $r<d$, where $d$ is the
number of rows of $W_{\rm down}$.
\end{proposition}

\noindent The proof is in Appendix~\ref{proof:I.8}. In (a) the fibre product is the $r^2$-dimensional linear subspace
$\{(B_0R,RA_0)\}$, and if $\rk W_0<r$ the frame $\Sigma_rV_r\T$ is rank-deficient and the LoRA-XS instance breaks.
[Num] The fibre product of (a) has dimension $4=r^2$ at $(m,n,r)=(6,8,2)$ (Appendix~\ref{app:repro}).

\begin{explains}
LoRA-XS is the largest method that factors through \emph{both} frozen frames. In the frames of an initialisation it is the
overlap that a split initialisation double-counts (\thmref{Proposition}{II.3}); it is LoRA-XS proper when those frames span
the top-$r$ singular subspaces of $W_0$, as at a PiSSA initialisation \citep{meng2024pissa}. The meet in (b) is only
set-level. For $n\ge2$ the product does not exist in $\Diff$, because its fibre product is singular at the origin
$(b,a,\ell)=(0,0,\mathbf 1)$, by the rank argument of \thmref{Observation}{I.4}(b) (Appendix~\ref{proof:I.8}). Once a support is fixed, the meet is
just (IA)$^3$ masked to $r$ neurons; compare sparse activation scaling \citep{xstoehr2024activation}.
\end{explains}

\paragraph{The shape of the meet in (b)}
The meet is the union of the $\binom mr$ coordinate $r$-planes ``rescale at most $r$ output neurons''. For $r=1$ the fibre
product is the union of the linear spaces $\{b=0,\ \ell=\mathbf 1\}$ and $\{a=0,\ \ell=\mathbf 1\}$ with $m$
two-dimensional sheets $\{b=ce_i,\ a=\lambda w_i,\ \ell_i=1+sc\lambda\}$, where $w_i$ is row $i$ of $W_0$. Under the
standard pointing ($b=0$, $a_0$ generic) the base point is a smooth point of $\{b=0,\ \ell=\mathbf 1\}$, so the singularity
sits away from it, at the origin.

\subsubsection{The dictionary so far}\label{sec:slice-dictionary}

Table~\ref{tab:slice-dictionary} collects the translations made in this exposé. The full dictionary is the Rosetta stone
of Exposé~\ref{sec:rosetta}.

\begin{table}[!htb]
\centering
\caption{The dictionary for Exposé~\ref{sec:slice}.}
\label{tab:slice-dictionary}
\small\renewcommand{\arraystretch}{1.15}
\begin{tabularx}{\linewidth}{>{\raggedright\arraybackslash}p{0.27\linewidth}>{\raggedright\arraybackslash}X>{\raggedright\arraybackslash}p{0.2\linewidth}}
\toprule
\headrow \textbf{In PEFT} & \textbf{In the pointed slice} & \textbf{Statement} \\
\midrule
pretrained checkpoint & base point $\theta_0$ of a pointed 1-cell $(\Theta,\theta_0,f)$ & \thmref{Definition}{I.1} \\
a weight-space PEFT method & an object $\rho:(Q,q_0)\to(\Theta,\theta_0)$ of $\Meth(\Theta,\theta_0)$ & \thmref{Definition}{I.2} \\
``$N$ can do whatever $M$ can'' & a morphism $h:M\to N$, $\rho_N\circ h=\rho_M$; at the level of sets, image inclusion
  & \thmref{Definition}{I.2}, \thmref{Observation}{I.4}(c) \\
frozen model, full fine-tuning & initial and terminal objects & \thmref{Observation}{I.4}(a) \\
merging into full-precision weights & the canonical morphism $\rho_M$ to the terminal object & \thmref{Observation}{I.4}(a) \\
effective number of parameters & expressive dimension $d(M)$; the rest is gauge waste & \thmref{Definition}{I.5} \\
what the first steps can use & first-order image $S_1$, a functorial invariant & \thmref{Observation}{I.6} \\
RoSA, GraLoRA, ReLoRA stages & $\oplus$-words, with a Minkowski sum as image & \thmref{Observation}{I.7} \\
HF \texttt{cat} merge & the concatenation isomorphism evaluated at trained points & \thmref{Observation}{I.7} \\
LoRA-XS & the product $\mathrm{FA}(A_0)\times\mathrm{FB}(B_0)$ & \thmref{Proposition}{I.8} \\
adapters, prefixes, ReFT & extensions of the network, outside $\Meth$ & \thmref{Definition}{VI.1} \\
\bottomrule
\end{tabularx}
\end{table}

\section{The image germ: expressivity}\label{sec:image}

Expressivity is read off the germ of a method's image at the pretrained point. An initialisation decides whether that
point sits at the apex of a cone, on a smooth sheet of one, or just off the image.

Exposé~\ref{sec:slice} turned a PEFT method into a pointed map $\rho:(Q,q_0)\to(\Theta,\theta_0)$. Its image
$\Im M=\rho(Q)$ is everything it can reach. Fine-tuning travels only a short way from the pretrained weights, so what
matters is the image near $\theta_0$, its \emph{germ} at $\theta_0$ (\bgref{C.8}).

The germ takes few shapes (Figure~\ref{fig:teaser}). It is a linear space when $\rho$ is affine, as for BitFit
\citep{benzaken2021bitfit}, LoRA-FA \citep{zhang2023lora}, LoRA-XS \citep{baazy2024lora}, FourierFT
\citep{gao2024parameter} and (IA)$^3$ \citep{liu2022few}. It is a cone (\bgref{C.10}) when $\rho$ is multihomogeneous
and non-linear and starts where the update vanishes, as LoRA's $BA$ \citep{hu2021lora} does at zero initialisation. It
is a piece of a group orbit for exactly orthogonal methods such as exact-Cayley OFT \citep{qiu2023controlling}. DoRA
\citep{liu2024dora} adds an output gain to LoRA's cone (\thmref{Proposition}{II.8}), and HRA \citep{yuan2024bridging}
meets that cone with a rotation orbit (\thmref{Theorem}{II.7}), each under its stated hypotheses. Adapters and prefixes
change the network itself, and their germs live among functions (Exposé~\ref{sec:merge}).

Four invariants of the germ carry most of the story, and isomorphic methods share them (\thmref{Observation}{I.6}).
\begin{itemize}
  \item The \textbf{first-order image} $S_1(M)=d\rho_{q_0}(T_{q_0}Q)$, the directions a first step can take.
  \item The \textbf{type of the base point}: \emph{regular} when the first-order deficit $d(M)-\dim S_1(M)$ vanishes,
    an \emph{apex} when it is positive (\thmref{Definition}{I.5}).
  \item The \textbf{maximum rank} of an update, bounded by a minimum cut (\thmref{Theorem}{II.10}).
  \item The \textbf{expressive dimension} $d(M)=\dim\Im M$, which falls short of the parameter count $|M|$ by the
    gauge waste $g(M)=|M|-d(M)$.
\end{itemize}

Subsection~\ref{sec:strata} asks where a method starts, Subsection~\ref{sec:erlangen} what a group-type method can never
change (\thmref{Theorem}{II.5}, \thmref{Corollary}{II.6}), and Subsection~\ref{sec:cut} how rank differs from dimension
(\thmref{Proposition}{II.12}). Expressivity here means reach, and finding a reachable point is left to
Exposé~\ref{sec:fibres}.

Which functions a LoRA-adapted network can represent is studied by \citet{zeng2023expressive}. The intrinsic dimension
of \citet{li2018measuring} and \citet{aghajanyan2020intrinsic} depends on the data, while $d(M)$ is a property of the
method alone.

\subsection{Initialisations: pointings and strata}\label{sec:strata}

\begin{figure}[tbp]
  \centering
  \includegraphics[width=\linewidth]{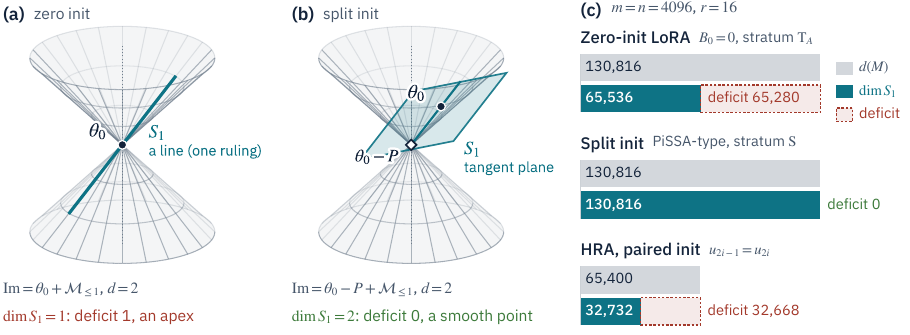}
  \caption{\textbf{The apex and the smooth point.} Panels (a) and (b) draw the symmetric slice $S=\left(\begin{smallmatrix}x+y&z\\ z&x-y\end{smallmatrix}\right)$ of $2\times2$ matrices, in which rank $\le1$ is the grey double cone $x^2=y^2+z^2$, and the model $\rho(c,\phi)=W_{\rm res}+c\,v_\phi v_\phi\T$, $v_\phi=(\cos\phi,\sin\phi)$, which plays LoRA ($c$ for $B$, $v_\phi$ for $A$); its image is the whole cone, so $d=2$. (a)~Zero initialisation, $c_0=0$ and $W_{\rm res}=\theta_0$: the base point $\theta_0$ is the vertex of the cone, and the first-order image $S_1$ (teal) is a single line, one ruling, so the deficit is $1$ and $\theta_0$ is an apex (\thmref{Definition}{I.5}, \thmref{Theorem}{II.2}(c)). (b)~A split (PiSSA-type) initialisation, $c_0\ne0$ and $W_{\rm res}=\theta_0-P$ with $P=c_0v_{\phi_0}v_{\phi_0}\T$: the vertex moves to $\theta_0-P$ (open diamond), $\theta_0$ lies on a smooth sheet of the cone, and $S_1$ is the tangent plane there, so the deficit is $0$. Both values of $\dim S_1$ are Jacobian ranks of the model. (c)~For $m=n=4096$ and $r=16$, each pair of bars compares the expressive dimension $d(M)$ (grey) with $\dim S_1$ (teal); the dashed red box is the gap between them, the first-order deficit. Zero-init LoRA, stratum $\mathrm T_A$, has $\dim S_1=mr=65{,}536$ against $d=r(m+n-r)=130{,}816$; a split initialisation has $\dim S_1=d$ (\thmref{Theorem}{II.2}(b)); HRA's paired initialisation $u_{2i-1}=u_{2i}$, $k=r/2$, has $\dim S_1=kn-\tfrac{k(k+1)}2=32{,}732$ against $d=rn-\tfrac{r(r+1)}2=65{,}400$, so it too starts at an apex (\thmref{Theorem}{II.7}(b)). The script checks each closed form against a Jacobian rank at $(m,n,r)=(11,9,4)$.}
  \label{fig:apex}
\end{figure}

Every LoRA recipe decides where training starts. LoRA \citep{hu2021lora} sets $B_0=0$ and draws $A_0$ at random. PiSSA
\citep{meng2024pissa} takes both factors from the top of the SVD of $W_0$ and subtracts their product from the frozen
weight, LoRA-GA \citep{wang2024lorab} does the same with the first gradient, and CorDA \citep{yang2024corda}, OLoRA
\citep{buyukakyuz2024olora}, MiLoRA \citep{wang2024milora} and EVA \citep{paischer2024parameter} choose otherwise. Each
recipe is a \emph{pointing}, a parameter $q_0$ with the frozen residual that makes $\rho(q_0)=\theta_0$, and two
pointings of one formula can be different objects of $\Meth(\Theta,\theta_0)$. Figure~\ref{fig:apex} draws the typical
cases.

The answers occupy the rest of this subsection. For $r<\min(m,n)$, full-rank LoRA initialisations come in exactly three
kinds up to reparametrisation, classified as sets by $\operatorname{row}A_0$ (when $B_0=0$), by $\operatorname{col}B_0$
(when $A_0=0$) or by the subtracted product $B_0A_0$ (PiSSA-type splits), as \thmref{Theorem}{II.2} shows. Zero-factor
starts sit at the apex of a cone with $mr$ or $nr$ first-order directions, and split starts at a smooth point with all
$r(m+n-r)$. For a $4096\times4096$ matrix at rank $16$ that is $130{,}816$ against $65{,}536$
(\thmref{Proposition}{II.3}). Neither kind reaches everything the other reaches. No exact start of LoRA$_r$ reaches an
update of rank above $2r$ (\thmref{Theorem}{II.1}), and a factor behind a closed gate, such as $A$ when $B_0=0$, does
not move the weights at first order (\thmref{Lemma}{II.4}).

\subsubsection{Translates of one shape}\label{sec:strata-translates}

An additive method adds updates from a fixed shape $C$, for LoRA the cone $\Mr$ of matrices of rank at most $r$
(\bgref{C.12}). Pointing it at $q$ slides the shape so that $\delta(q)$ lands on $\theta_0$, and reachable updates
become differences of two points of $C$ (\bgref{B.14}).

\begin{theorem}{II.1}{pointings of an additive method}
Let $M=(Q,q_0,\theta_0+\delta)$ be additive with shape $C=\delta(Q)$. Write $-v+C$ for a translate, and
$C-C:=\{c-c':c,c'\in C\}$ for the Minkowski difference.

(a) For each $q\in Q$, the re-pointed method $M_q:=(Q,q,\theta_0-\delta(q)+\delta(\cdot))$, with frozen residual
$\theta_0-\delta(q)$, is pointed at $\theta_0$, and $\Im M_q=\theta_0+(-\delta(q)+C)$.

(b) $\bigcup_q\Im M_q=\theta_0+(C-C)$.

(c) For LoRA$_r$ on a single linear box with $s\ne0$, $C-C=\MM_{\le\min(2r,m,n)}$; for several boxes, $C-C$ is the
product of the per-box varieties $\MM_{\le\min(2r,m_i,n_i)}$. Hence \textbf{no exact pointing of LoRA$_r$ (frozen
residual $\theta_0-\delta(q)$), whether zero-init or split (PiSSA, OLoRA, CorDA, LoRA-GA), makes an update
$\theta-\theta_0$ of rank $>2r$ reachable in any adapted matrix.} More robustly, from \emph{any} starting point $q_0$ of
LoRA$_r$, pointed or not, $\rho(q)-\rho(q_0)=\delta(q)-\delta(q_0)$ has rank $\le\min(2r,m,n)$. Quantised-residual
schemes (QLoRA, LoftQ, QPiSSA) are pointed only up to a defect that depends on the order of operations ($e$ for QLoRA;
\thmref{Proposition}{V.2}(b) for the others). Measured from $\theta_0$ rather than from their starting weight, their
updates contain this defect, whose rank is generically at least $\min(m,n)-r$. The bound $2r$ is attained by split
pointings when $2r\le\min(m,n)$.
\end{theorem}

The theorem is elementary, and its rank-$2r$ consequence is known in practice. \citet{meng2024pissa} convert a trained
PiSSA adapter into a rank-$2r$ LoRA through
\begin{equation}\label{eq:pissa-to-lora}
  BA-B_0A_0=\begin{bmatrix}B & B_0\end{bmatrix}\begin{bmatrix}A\\ -A_0\end{bmatrix},
\end{equation}
and HF PEFT ships the conversion. The theorem adds that the conversion loses nothing and that rank $2r$ is needed in
general.

\begin{explains}
An initialisation of an additive method chooses which \emph{translate} of the shape passes through $\theta_0$.
Splitting initialisations (PiSSA-type) buy ``spectral replacement'' updates $Y-P$, which can have rank $2r$. They pay by
losing generic updates of every rank $\ge1$ (\thmref{Theorem}{II.2}(d)). Moving a weight matrix by more than rank $2r$
needs rebasing, as in ReLoRA \citep{lialin2023relora} (\thmref{Theorem}{V.3}), or another shape.
\end{explains}

\subsubsection{Three strata}\label{sec:strata-three}

Isomorphisms of methods are pointed diffeomorphisms over $\Theta$ (\thmref{Definition}{I.2}), such as the gauge
$(B,A)\mapsto(Bg^{-1},gA)$ with $g\in\GL_r$ (\bgref{D.21}). They preserve the image and $S_1$
(\thmref{Observation}{I.6}) but may change training (Exposé~\ref{sec:fibres}). Up to isomorphism, a full-rank LoRA
initialisation is determined by one subspace or one matrix.

\begin{theorem}{II.2}{three strata of LoRA initialisations}
Let $\delta(B,A)=BA$ with $1\le r<\min(m,n)$. Consider pointed LoRA objects in three distinguished strata of full-rank
pointings (pointings whose nonzero factors have full rank). Only $\mathrm S$ is open and dense in the space of
pointings; $\mathrm T_A$ and $\mathrm T_B$ are nowhere dense.
\begin{itemize}
  \item $(\mathrm T_A)$: $B_0=0$ and $\rk A_0=r$;
  \item $(\mathrm T_B)$: $A_0=0$ and $\rk B_0=r$;
  \item $(\mathrm S)$: $\rk B_0A_0=r$, with frozen residual $\theta_0-B_0A_0$.
\end{itemize}

(a) \textbf{Classification.} Within a stratum, $M_{q_0}\cong M_{q_0'}$ if and only if, respectively,
$\operatorname{row}A_0=\operatorname{row}A_0'$; $\operatorname{col}B_0=\operatorname{col}B_0'$; or $B_0A_0=B_0'A_0'$.
Objects in different strata are never isomorphic. As a \emph{set}, the isomorphism classes of the three strata
therefore form
\begin{equation}\label{eq:strata-moduli}
  \Gr(r,\R^n)\ \sqcup\ \Gr(r,\R^m)\ \sqcup\ \MM_r(m\times n).
\end{equation}

(b) \textbf{First order.} $\dim S_1$ is $mr$, $nr$ and $r(m+n-r)$ respectively, while $d=r(m+n-r)$ in every case. The
deficits are $r(n-r)$, $r(m-r)$ and $0$.

(c) \textbf{Geometry.} In $\mathrm T_A$ and $\mathrm T_B$, $\Im=\theta_0+\Mr$, whose vertex $\theta_0$ is an
\textbf{apex}. In $\mathrm S$, $\Im=(\theta_0-P)+\Mr$ with $P=B_0A_0$, and $\theta_0$ is a \textbf{smooth point}.

(d) \textbf{Incomparability.} The images of $\mathrm S$-type and $\mathrm T$-type objects are incomparable.

(e) \textbf{Scope.} The three strata do not exhaust the pointings of LoRA$_r$. Outside them lie: $B_0=0$ with
$\rk A_0<r$; $A_0=0$ with $\rk B_0<r$; $B_0=A_0=0$; $0<\rk B_0A_0<r$; and pointings with $B_0,A_0$ both nonzero and
$B_0A_0=0$. HF PEFT ships one of the last kind: \texttt{init\_lora\_\allowbreak weights=\allowbreak"orthogonal"} (which requires $r$ even)
lies in the stratum $(\operatorname{rk}B_0,\operatorname{rk}A_0,\operatorname{rk}B_0A_0)=(r/2,r/2,0)$, an apex with
$\dim S_1=\tfrac r2(m+n-\tfrac r2)$. It is isomorphic to no object of the three strata. Its image $\theta_0+\Mr$ rules
out $\mathrm S$, and its $S_1$ determines $(\operatorname{col}B_0,\operatorname{row}A_0)$, of dimensions $(r/2,r/2)$,
against $(0,r)$ in $\mathrm T_A$ and $(r,0)$ in $\mathrm T_B$. The dimension of $S_1$ alone does not separate it, since
at $(m,n,r)=(4,5,2)$ it equals $mr$. This placement holds in exact arithmetic; HF casts the two factors to the base
dtype separately, and in bf16 the result is a generic pointing with a tiny defect. For every pointing, the image
determines $P=B_0A_0$ (its vertex is $\theta_0-P$), and $S_1=\{XA_0+B_0Y\}$ determines
$(\operatorname{row}A_0,\operatorname{col}B_0)$ whenever these are proper subspaces. The classification is set-level:
in the space of pointings $\mathrm T_A$ lies in the closure of $\mathrm S$, since $(\varepsilon B',A_0)\to(0,A_0)$, so
the quotient is not a topological disjoint union.
\end{theorem}

Numerically, $\dim S_1=(14,10,20)$ for $(\mathrm T_A,\mathrm T_B,\mathrm S)$ at $(m,n,r)=(7,5,2)$, and $\dim S_1=11$
for HF's \texttt{orthogonal} init (Appendix~\ref{app:repro}). Here $\Gr$ is the Grassmannian (\bgref{C.15}), $\MM_r$ the
smooth stratum of $\Mr$ (\bgref{C.12}), and smooth and singular points are those of \bgref{C.9}.

\begin{remark*}{a moduli space of initialisations, as an analogy}
Calling the three spaces in \eqref{eq:strata-moduli} a moduli space (\bgref{C.20}) borrows a word. The bijection is of
sets only. $\mathrm T_A$ lies in the closure of $\mathrm S$, since $(\varepsilon B',A_0)\to(0,A_0)$, so the quotient is
not a topological disjoint union, and no universal family is built.
\end{remark*}

Most LoRA initialisation schemes in the literature are points of these three strata (Table~\ref{tab:strata}).
LoRA-FA and MiCA freeze one factor and are sub-objects (\bgref{A.15}), and the last three rows of the table lie outside
the strata.

\begin{table}[!tbp]
  \centering
  \footnotesize
  \caption{LoRA initialisation schemes as points of the strata of \thmref{Theorem}{II.2}, with the sub-objects and
    the exceptions. Conventions: $y=Wx$, $W\in\R^{m\times n}$; HF is Hugging Face PEFT.}
  \label{tab:strata}
  \renewcommand{\arraystretch}{1.15}
  \begin{tabularx}{\linewidth}{@{}>{\raggedright\arraybackslash}p{3.5cm}>{\raggedright\arraybackslash}p{2.35cm}>{\raggedright\arraybackslash}X@{}}
    \toprule
    \headrow \textbf{scheme} & \textbf{stratum} & \textbf{point} \\
    \midrule
    vanilla LoRA \citep{hu2021lora} (\texttt{Linear}, \texttt{Conv}) & $\mathrm T_A$ &
      a random point of $\Gr(r,n)$: Haar (\bgref{D.14}) under Gaussian $A_0$, only approximately Haar under
      Kaiming-uniform \\
    vanilla LoRA on \texttt{nn.Embedding} (loralib and HF default) & $\mathrm T_B$ &
      $A_0=0$, $B_0\sim\mathcal N(0,1)$ in the convention $y=Wx$ with $m$ the embedding dimension; deficit $r(m-r)$ \\
    EVA \citep{paischer2024parameter} & $\mathrm T_A$ &
      the top-$r$ principal subspace of the input activations, at its per-layer rank \\
    LoRA-FA \citep{zhang2023lora} & sub-object of $\mathrm T_A$ &
      the same point with $A$ frozen (an arrow $\mathrm{FA}(A_0)\to\mathrm{LoRA}_{(0,A_0)}$) \\
    ``Init[B]'' \citep{hayou2024impact} & $\mathrm T_B$ & a random point of $\Gr(r,m)$ \\
    MiCA \citep{rudiger2026mica} & sub-object of $\mathrm T_B$ &
      $B_0$ the minor left-singular frame of $W_0$, frozen: $\mathrm{FB}(B_0)\to\mathrm{LoRA}_{(B_0,0)}$, a linear
      image at a regular point (HF also allows $r=\min(m,n)$, outside the theorem) \\
    PiSSA \citep{meng2024pissa} / MiLoRA \citep{wang2024milora} & $\mathrm S$ &
      $P$ the top / bottom rank-$r$ SVD part of $W_0$ (HF absorbs the scale: $P=sB_0A_0$) \\
    OLoRA \citep{buyukakyuz2024olora} & $\mathrm S$ &
      $P=sQ_rR_r$ in HF, from a truncated QR factorisation of $W_0$ \\
    CorDA \citep{yang2024corda} & $\mathrm S$ &
      $P$ from the SVD of $W_0C$, where $C$ is the activation covariance \\
    LoRA-GA \citep{wang2024lorab} & $\mathrm S$ & $P$ from the SVD of the first gradient \\
    Astra \citep{liu2026astra} & $\mathrm S$ &
      $P=VV\T W_0$ (times $s$ in HF and the official code, which do not absorb the scale), with $V$ eigenvectors of the
      output activations (the centred covariance's tail in the paper, the uncentred second moment's top $r$ in HF and
      the official code), when $\rk V\T W_0=r$ \\
    HF \texttt{orthogonal} & outside: $(r/2,r/2,0)$ &
      an apex with $\dim S_1=\tfrac r2(m+n-\tfrac r2)$, for even $r$ in exact arithmetic \\
    QLoRA \citep{dettmers2023qlora}, LoftQ \citep{li2023loftq} & defect, not a point &
      zero-init (QLoRA) or approximately split (LoftQ) over a quantised base (\thmref{Proposition}{V.2}) \\
    HF \texttt{init\_lora\_\allowbreak weights=\allowbreak False} & unpointed & random $A$ and $B$, with no subtraction \\
    \bottomrule
  \end{tabularx}
\end{table}

\begin{explains}
Zero-init LoRA starts at a \emph{singular point}. At first order the $A$-direction is dead, so LoRA and LoRA-FA (which
freezes $A$ and trains $B$; \citet{zhang2023lora} use AdamW), started from the same $A_0$, have the same $K_{q_0}$ under
plain gradient steps and the same first-order trajectory under Adam (\thmref{Lemma}{II.4}). Splitting initialisations
start at a smooth point with full first-order reach. We conjecture that this is one mechanism behind the faster early
progress reported for PiSSA and LoRA-GA. LoRA-GA attributes its speed-up to gradient alignment, and MiLoRA also starts
at a smooth point, so the smooth start alone does not account for the reported gains. The strata classify
\emph{expressivity together with $S_1$}. What an optimizer additionally sees is the subject of
\thmref{Corollary}{III.13}.
\end{explains}

\citet{hayou2024impact} compare the two zero-factor starts, ``Init[A]'' and ``Init[B]'', through their training
dynamics; here $S_1$ already tells them apart. The facts about $\Mr$ used in the proof are classical (\bgref{C.12}).

\begin{remark*}{the first-order deficit and the data}
$S_1$ counts directions in weight space. The lazy-regime kernel $J_f\,d\rho\,\Lambda\,d\rho\T J_f\T$ also involves the
network Jacobian $J_f$ and the optimizer's metric $\Lambda$, and for random-$A_0$ zero-init LoRA it approximates the
full fine-tuning kernel \citep{malladi2022kernel}. A large first-order deficit need not cost much on the data.
\end{remark*}

\subsubsection{Counting first-order directions}\label{sec:strata-count}

A first step of LoRA from $(B,A)$ adds $s(\dot BA+B\dot A)$, a matrix with rows in $\operatorname{row}A$ plus one with
columns in $\operatorname{col}B$. Counting both families and their overlap gives $\dim S_1$.

\begin{proposition}{II.3}{first-order rank formula: smoothness, alignment and scale}
For $\kappa(B,A)=W_{\rm base}+sBA$ with $s\ne0$,
\begin{equation}\label{eq:rank-formula}
  \rk d\kappa_{(B,A)}=m\operatorname{rk}A+n\operatorname{rk}B-\operatorname{rk}A\operatorname{rk}B\qquad\text{for all }(B,A),
\end{equation}
and $\{XA\}\cap\{BY\}=\{BRA:R\in\R^{r\times r}\}$. For full-rank frames this overlap is the tangent space of the
product $\mathrm{FA}(A)\times\mathrm{FB}(B)$ of \thmref{Proposition}{I.8}(a), re-based at $\kappa(B,A)$, whose image is
the meet of the two one-sided images: a generalised LoRA-XS. It is LoRA-XS proper when $(B,A)$ span the top-$r$
singular subspaces of $W_0$ ($\sigma_r>0$), as at a PiSSA init; bottom frames (MiLoRA) give a generalised LoRA-XS.

Under gradient flow $\dot B=-\eta_B\nabla_BL$, $\dot A=-\eta_A\nabla_AL$, with $G=\nabla_WL$,
\begin{equation}\label{eq:gf-rate}
  \tfrac{d}{dt}L=-s^2\big(\eta_B\|GA\T\|_F^2+\eta_A\|B\T G\|_F^2\big)
\end{equation}
at every time, in particular at $(B_0,A_0)$. To first order in the step sizes the same holds for full-batch gradient
descent, and in expectation for unbiased minibatch SGD. Under Adam ($\varepsilon\to0$, with bias correction) the first
step changes the loss by
\begin{equation}\label{eq:adam-rate}
  -\lvert s\rvert\big(\eta_B\|GA_0\T\|_1+\eta_A\|B_0\T G\|_1\big)+O(\eta^2),
\end{equation}
with entrywise $\ell_1$ norms. This first-order term is linear in $\lvert s\rvert$, against quadratic under gradient
flow.
\end{proposition}

Numerically, the rank formula \eqref{eq:rank-formula} holds in $400$ random cases, including rank-deficient factors,
and the rate \eqref{eq:gf-rate} matches finite differences (Appendix~\ref{app:repro}).

\begin{example*}{one attention projection of a 7B model}
At $4096\times4096$ with $r=16$, zero-factor initialisations (LoRA, QLoRA, EVA, LoRA-FA) have $mr=65{,}536$ first-order
directions, and split ones have $r(m+n-r)=130{,}816$, the dimension of LoRA's image.
\end{example*}

\begin{explains}
Zero-factor initialisations (LoRA, QLoRA, EVA, LoRA-FA) have $mr$ first-order directions; split initialisations have
$r(m+n-r)$. ``Better initialisation'' in the literature mixes three distinct mechanisms, and the proposition separates
them.
\end{explains}

\paragraph{Smoothness} PiSSA and LoRA-GA start off the apex, so their first step sees all $r(m+n-r)$ directions.

\paragraph{Alignment} Under gradient flow, among $A_0$ with $A_0A_0\T=cI$, the term $\eta_B\|GA_0\T\|_F^2$ of the
initial rate \eqref{eq:gf-rate}, and with it the whole rate at a zero-$B$ start, is maximised by the top-$r$ right
singular space of $G$. LoRA-GA takes exactly this for $A_0$, but takes $B_0$ from the left singular vectors
$r+1,\dots,2r$ (its Algorithm~1; HF's default \texttt{direction="ArB2r"}). $G$ then has no component in the overlap
$\{B_0RA_0\}$, so the first gradient step approximates the rank-$2r$ truncation of $G$. LoRA-GA optimises gradient
approximation, and gives up part of the $B$-term of the rate to do so. EVA uses the top principal subspace of the input
activations. Its Theorem~3.2 assumes the inputs independent of the upstream errors and bounds the expected per-sample
squared gradient norm with the uncentred $\mathbb E[xx\T]$. That quantity is not the rate $\|\mathbb E[G]A_0\T\|^2$, and
HF's EVA uses mean-centred PCA, so EVA's frame is a data-only proxy. Alignment is a gradient-flow and SGD notion. Under
Adam the first-step rate $\|GA_0\T\|_1$ of \eqref{eq:adam-rate} depends on the basis of $\operatorname{row}A_0$, not
only on the subspace.

\paragraph{Scale} $\|GA_0\T\|_F^2$ grows with $\|A_0\|^2$, and initial scale trades off against learning rate
(\thmref{Theorem}{III.10}). PiSSA's $S_r^{1/2}$-weighted factors change it. LoRA-GA's paper scales both factors by
$d_{\rm out}^{1/4}/\sqrt\gamma$; HF's default \texttt{scale="stable"} scales only $B$ (which also carries
$S_{r+1:2r}$), so there $\|GA_0\T\|_F^2$ is unchanged. HF's Kaiming default has $A_0A_0\T\approx I/3$, against
$A_0A_0\T=I$ for EVA with \texttt{whiten=False}. Comparing rates across initialisations needs $A_0A_0\T$ normalised.

\medskip
LoRA-GA uses all three mechanisms, with alignment on the $A$ side only. The gradient-aligned frame is due to
\citet{wang2024lorab}, the activation frame to \citet{paischer2024parameter}, and the spectral split to
\citet{meng2024pissa}; \thmref{Proposition}{II.3} only sorts their effects into three terms.

\subsubsection{The zero gate}\label{sec:strata-gate}

At a zero-$B$ start, $BA$ is linear in $B$ and $B_0=0$, so moving $A$ alone changes nothing. The same holds whenever a
trainable gate starts closed. Below, $K=D\rho\,\Lambda\,D\rho\T$ is the cometric on weights induced by an optimizer's
metric $\Lambda$ (\bgref{D.23}, Exposé~\ref{sec:fibres}).

\begin{lemma}{II.4}{zero gate}
Let $Q=U\times V$ with $U$ a vector space, let $\rho=\theta_0+\beta(u,v)$ with $\beta$ smooth, and suppose
$\beta(u_0,v)=0$ for all $v$; for example $\beta(\cdot,v)$ linear and $u_0=0$. Then $D_v\rho_{q_0}=0$. Hence
$\nabla_v(L\circ\rho)(q_0)=0$, $S_1=D_u\beta(u_0,v_0)(U)$ ($=\beta(U,v_0)$ in the linear case), and the cometric at
$q_0$ is $K_{q_0}=D_u\beta\,\Lambda_{uu}\,D_u\beta\T$, where $\Lambda_{uu}$ is the $uu$-block of $\Lambda$; no
block-diagonality is needed, since the $v$-columns of $D\rho$ vanish. The weight trajectory $\theta(t)$ agrees with that
of the method with $v$ frozen at $v_0$ to first order in $t$ (gradient flow), or in the step size (one or more Adam
steps). The parameter $v$ itself may still move: under terms not of the form $L\circ\rho$ (regularisers, decoupled
weight decay), and under Adam, whose step-2 move of $v$ is $O(\eta)$ although its effect on $\theta$ is $O(\eta^2)$.

\emph{Function-level variant.} For an extension with realised function $F(u,v,x)$, if $F(u_0,v,x)=F_0(x)$ for all $v$
and $x$, then $D_vF=0$ at $q_0$, with the same conclusions at the level of functions. No linearity is needed.
\end{lemma}

Numerically, under Adam with $\eta\in\{10^{-2},10^{-3},10^{-4}\}$, the step-2 move of $A$ in zero-init LoRA scales as
$\eta$, while the deviation of $\Delta W$ from the frozen-$A$ method scales as $\eta^2$ (Appendix~\ref{app:repro}).

Instances, assuming no weight decay on $v$:
\begin{itemize}
  \item LoRA \citep{hu2021lora}, with $u=B$.
  \item VeRA \citep{kopiczko2023vera}, with $u=b$: $\theta$ first moves only along $b$.
  \item AdaLoRA \citep{zhang2023adalora}, whose update is $P\diag(\lambda)Q$, with $u=\lambda$, the diagonal of
    singular values, started at $0$. $\theta$ initially moves only along the $r$ rank-one directions $p_iq_i\T$, so $\dim S_1=r$ for
    generic $P,Q$. $P$ and $Q$ do move at step one, through AdaLoRA's orthogonality penalty $R(P,Q)$ (HF default
    \texttt{orth\_reg\_weight=0.5}), which has no first-order effect on $\theta$.
  \item LoHa \citep{hyeonwoo2021fedpara} with a zero factor (HF zero-initialises \texttt{hada\_w2\_b}): only that factor
    moves.
  \item By the function-level variant, the gate of LLaMA-Adapter \citep{zhang2023llama}, which is the only parameter
    that moves at first order, and bottleneck adapters whose up-projection starts at zero, where only that factor
    moves (\citet{houlsby2019parameter} start near that point).
\end{itemize}
HRA's paired initialisation has a different form and shows the same first-order collapse
(\thmref{Theorem}{II.7}(b)).

\citet{zhu2024asymmetry} find over whole runs that a random, untrained $A$ does nearly as well as a tuned one, and that
training $B$ alone gives better generalisation bounds. \thmref{Lemma}{II.4} is only the first-order statement.

\subsubsection{Off the point: quantised bases}\label{sec:strata-defect}

QLoRA \citep{dettmers2023qlora} quantises the frozen weight, so it is LoRA pointed at $\kappa\theta_0$, with
\emph{pointing defect} $e=\kappa\theta_0-\theta_0$ (\thmref{Definition}{V.1}, \thmref{Proposition}{V.2}(a)). As $e$ is
generically of full rank, $\theta_0$ lies outside $\Im\mathrm{LoRA}_r(\kappa\theta_0)$ unless $\rk e\le r$, which is
why \thmref{Theorem}{II.1} speaks of \emph{exact} pointings. When the initialisation also splits, the order of
operations decides the defect (\thmref{Proposition}{V.2}(b)). In Table~\ref{tab:defects}, $W$ is the full-precision
weight, $e=\kappa W-W$ and $\delta_0=\delta(q_0)$.

\begin{table}[!tbp]
  \centering
  \footnotesize
  \caption{Pointing defects of quantised-base LoRA pipelines (\thmref{Proposition}{V.2}(b)).}
  \label{tab:defects}
  \renewcommand{\arraystretch}{1.15}
  \begin{tabularx}{\linewidth}{@{}>{\raggedright\arraybackslash}p{3.3cm}>{\raggedright\arraybackslash}X>{\raggedright\arraybackslash}p{4.3cm}@{}}
    \toprule
    \headrow \textbf{pipeline} & \textbf{examples} & \textbf{defect $\rho(q_0)-W$} \\
    \midrule
    quantise, then zero-init & QLoRA \citep{dettmers2023qlora} & $e$ \\
    split, then quantise the residual &
      QPiSSA \citep{meng2024pissa}; HF's recommended PiSSA/CorDA workflow; LoftQ's later quantisation steps &
      $\kappa(W_{\rm res})-W_{\rm res}$ \\
    quantise, then fit the error &
      LoftQ \citep{li2023loftq} with one iteration (HF default \texttt{loftq\_iter=1}); HF
      \texttt{replace\_lora\_\allowbreak weights\_loftq} &
      $e-\mathrm{SVD}_r(e)$, the part of $e$ left after its best rank-$r$ approximation (unit LoRA scale) \\
    quantise, split, re-quantise & HF \texttt{olora\_init} on bitsandbytes weights &
      $e+\kappa(r')-r'$, with $r'=\kappa W-\delta_0$ \\
    \bottomrule
  \end{tabularx}
\end{table}

\begin{remark*}{LoftQ's scale in HF PEFT}
HF PEFT's LoftQ initialiser does not divide by the LoRA scale $s$ ($\alpha/r$, or $\alpha/\sqrt r$ under rsLoRA), so
for $s\ne1$ the deployed defect is $e-s\,\mathrm{SVD}_r(e)$. At $s=2$ its Frobenius norm equals $\|e\|$, and the fit
buys nothing over QLoRA in that norm.
\end{remark*}

Quantised bases for LoRA come from \citet{dettmers2023qlora}, fitting the adapter to the quantisation error from
\citet{li2023loftq}, and the quantised PiSSA variant from \citet{meng2024pissa}.

The strata of \thmref{Theorem}{II.2} classify what a start can express. The dynamics see a finer quotient. For a
zero-$B$ start, SGD sees $A_0\T A_0$ and Adam sees $A_0$ up to signed permutations of its rows
(\thmref{Corollary}{III.13}). A dynamical classification of initialisations, including the degenerate pointings, is
open (Exposé~\ref{sec:analogy}).

\subsection{Erlangen: invariants of group-type methods}\label{sec:erlangen}

\begin{flushright}
\small\itshape A geometry is what a group leaves unchanged. Before asking how a method learns,\\
ask what it cannot help keeping.\\[2pt]
{\upshape\footnotesize after Klein}
\end{flushright}

A method that moves the pretrained weight by a group element can never change that group's invariants. Orbits also
place HRA as a meet, DoRA as a saturation and the scaling methods on a ladder.

\paragraph{In brief}
\begin{itemize}
  \item A method whose weights are always $W_0R$, with $R$ exactly orthogonal, can never change $WW\T$, so neuron norms,
    angles and singular values survive any number of merges in exact arithmetic. HF PEFT's default OFT, BOFT's scale and
    GOFT's default scaler break this.
  \item For injective $W_0$ and even $r\le n-2$, HRA reaches exactly the rank-$r$ LoRA updates of the form $W_0R$,
    $R\in\SO(n)$. Its HF PEFT default initialisation is an apex.
  \item DoRA reaches exactly the weights of LoRA plus a free gain per output neuron. From the same start, any gap between
    the two comes from training.
  \item (IA)$^3$, SSF's scales and DoRA's magnitude lie in the image of rank-one HiRA, $W_0\odot(J+BA)$, which for $W_0$
    without zeros is LoRA seen through $X\mapsto W_0\odot X$.
\end{itemize}

Klein's Erlangen programme of 1872 \citep{xklein1872vergleichende} (\bgref{D.17}) defines a geometry by its group of
motions and studies what the motions leave unchanged. OFT, GOFT, HRA and ETHER multiply the pretrained matrix by an
orthogonal matrix, (IA)$^3$ and SSF's scale multiply it by a diagonal one, and DoRA's magnitude does the same to a LoRA
weight. These are methods of \emph{action type} (\thmref{Definition}{I.2}), $\rho(q)=\gamma(q)\cdot W_0$ with $\gamma$
landing in a group. Whatever that group cannot change, the method cannot change, in training or after any number of
merges.

Klein's programme is the model. \citet{bronstein2021geometric} made it the organising principle of geometric deep
learning, for architectures. Here it is applied to the fine-tuning map.

\begin{analogy*}{what ``Erlangen'' means here}
For a group-type method the Erlangen reading is a theorem, since its image lies in one orbit. LoRA's cone $W_0+\Mr$ has a
singular vertex at $W_0$ (\thmref{Theorem}{II.2}(c)), so it is no orbit; for cones the word names the shape of the image
and its covariance group (\thmref{Definition}{II.14}).
\end{analogy*}

One convention matters. A linear box computes $y=Wx$ with $W\in\R^{m\times n}$, and row $i$ of $W$ is output neuron $i$.
The two Gram matrices, over neurons and over inputs, are
\begin{equation*}
  K_{\mathrm{out}}(W)=WW\T,\qquad K_{\mathrm{in}}(W)=W\T W.
\end{equation*}
A right multiplication $W\mapsto WR$ acts on the input side and fixes $K_{\mathrm{out}}$; a left multiplication
$W\mapsto PW$ acts on the output side and fixes $K_{\mathrm{in}}$. Papers that store $W\T$ swap these words, and we
translate every method into this convention.

\subsubsection{Complete invariants}\label{sec:erlangen-invariants}

A complete invariant of a group action (\bgref{D.15}) agrees on two matrices exactly when one can be carried to the
other. For the five actions in PEFT it is classical. We write $T_m$ for the invertible diagonal matrices,
$T_m^+=\{\diag(d):d_i>0\}$ for its identity component (\bgref{D.4}), and $\odot$ for the entrywise product.

\begin{theorem}{II.5}{complete invariants of the basic actions; classical}
Let $W,W'\in\R^{m\times n}$.
\begin{enumerate}[label=(\arabic*)]
  \item \emph{Right orthogonal (input side).} $W'=WR$ for some $R\in\Ort(n)$ iff $WW\T=W'W'^{\top}$. For $\SO(n)$, when
    $\rk W=n$, also require that the then unique $R$ have $\det R=1$.
  \item \emph{Left orthogonal.} $W'=PW$ for some $P\in\Ort(m)$ iff $W\T W=W'^{\top}W'$.
  \item \emph{Two-sided.} $W'=PWR$ for some $P\in\Ort(m)$ and $R\in\Ort(n)$ iff $W$ and $W'$ have the same singular
    values, counted with multiplicity.
  \item \emph{Left torus.} If $W$ has no zero row, $W'=DW$ for some $D\in T_m^+$ iff $w_i'\in\R_{>0}\,w_i$ for every $i$.
    For the full torus $T_m$, whose signs are unconstrained, the complete invariant is the set of zero rows together with
    the row \emph{lines}.
  \item \emph{Full Hadamard torus.} $W'=E\odot W$ for some $E\in(\R^\times)^{m\times n}$ iff $W$ and $W'$ have the same
    zero pattern.
\end{enumerate}
\end{theorem}

Item (1) is the orbit-separation consequence of the first fundamental theorem of invariant theory for $\Ort(n)$
(\bgref{D.16}); for a compact group, invariants separate orbits. Equivalently, it is Witt-type extension of isometries.
The two-sided diagonal torus $\{\diag(a)\,X\,\diag(b)\}\cong(T_m\times T_n)/\R^\times$ that appears as LoHa's gauge
(\thmref{Proposition}{II.12}) is a different group from the full Hadamard torus in (5).

\paragraph{Caveat for methods}
No PEFT method realises (4) or (5) as a group action on all of its parameters. (IA)$^3$ entries and DoRA magnitudes are
unconstrained reals initialised at $1$ or at the row norms, so the relevant group in (4) is $T_m$ or its closure.
HiRA's factor $J+BA$ can vanish entrywise. These methods can create new zeros, so for them only a one-way statement
transfers: zeros of $W_0$ remain zero. Table~\ref{tab:erlangen-invariants} lists the five actions with the methods that
live in each.

\begin{table}[t]
\centering\small
\caption{The five basic actions, their complete invariants (\thmref{Theorem}{II.5}) and the methods whose images lie in
one orbit or orbit closure.}
\label{tab:erlangen-invariants}
\begin{tabularx}{\linewidth}{@{}l l >{\raggedright\arraybackslash}p{3.3cm} >{\raggedright\arraybackslash}X@{}}
\toprule
\headrow \textbf{action} & \textbf{group} & \textbf{complete invariant} & \textbf{methods that live here}\\
\midrule
$W\mapsto WR$, input side & $\Ort(n)$ & $K_{\mathrm{out}}=WW\T$ & exact-Cayley OFT and block OFT, HRA, and GOFT and ETHER
  as published\\
$W\mapsto PW$, output side & $\Ort(m)$ & $K_{\mathrm{in}}=W\T W$ & LyCORIS diag-OFT; HF PEFT OFT up to version 0.13\\
$W\mapsto PWR$ & $\Ort(m)\times\Ort(n)$ & singular values with multiplicity & any composite of the two rows above\\
$W\mapsto DW$ & $T_m^+$, $T_m$ & row rays; zero rows and row lines & (IA)$^3$ on $k,v$, SSF's scale, DoRA's magnitude
  (closure only)\\
$W\mapsto E\odot W$ & $(\R^\times)^{m\times n}$ & zero pattern & HiRA (one way only: $Z(W_0)\subseteq Z(W)$)\\
\bottomrule
\end{tabularx}
\end{table}

\begin{explains}
Moving a method from the input side to the output side trades $K_{\mathrm{out}}$ for $K_{\mathrm{in}}$. Letting a scale
change sign trades rays for lines, and letting it reach zero leaves only the zeros of $W_0$.
\end{explains}

The interactive companion has a \emph{Gram lab} for these invariants. It fixes an invertible $4\times4$ matrix $W_0$ with
one zero entry and distinct singular values. The reader picks one of ten methods and drags, types or randomises its
parameters. Each lamp is a float64 test, to $10^{-10}$, that an invariant of $W=\rho(q)$ (the neuron Gram matrix, the
cosines, the input Gram matrix, the singular values, the row lines, the zeros of $W_0$) still equals its value at $W_0$,
and a further lamp tests $\rk\Delta W\le2$. A lamp guaranteed by a theorem is marked as such, and a red lamp names the
hypothesis that was broken. Heatmaps, neuron arrows and strips show $K_{\mathrm{out}}$, the cosines, $K_{\mathrm{in}}$,
$W$, $\sigma(W)$, the rank of $\Delta W$ and $\dim S_1$ against $d$. OFT's Cayley--Neumann switch (HF's default), BOFT's
$s_1\mapsto-s_1$, (IA)$^3$'s $\ell_2\mapsto0$ and DoRA's $B\mapsto0$ each break or restore a hypothesis.

\subsubsection{What exactly orthogonal methods can never do}\label{sec:erlangen-orthogonal}

If every weight a method reaches is $W_0R$ with $R$ orthogonal, \thmref{Theorem}{II.5}(1) and (3) freeze the neuron Gram
matrix and the singular values, and merging cannot help, since products of orthogonal matrices are orthogonal. The work
lies in saying which implementations are exactly orthogonal, and on which side.

\begin{corollary}{II.6}{what exactly orthogonal methods can never do}
Let a method have image in $W_0\cdot\Ort(n)$ (input side), and work in exact arithmetic. This covers:
\begin{itemize}
  \item OFT and block OFT with an exactly orthogonal parametrisation: the original paper's exact Cayley map, and HF PEFT
    from 0.16 with \texttt{use\_cayley\_\allowbreak neumann=False} on Linear and Conv targets;
  \item GOFT with strict Givens rotations and no scaler, as in the paper (\texttt{no\_scaling=True} in the official
    peft-givens code);
  \item HRA, and ETHER as defined in its paper (not ETHER+);
  \item any number of merge-and-restart cycles of these.
\end{itemize}
Then:
\begin{itemize}
  \item it preserves $K_{\mathrm{out}}$, hence every neuron norm and the angle between any two nonzero neurons, and so the
    hyperspherical energy, which depends only on these angles;
  \item it preserves the whole singular spectrum of $W_0$;
  \item block-diagonal OFT with input blocks $J$, with the same partition in every cycle, preserves each partial Gram
    matrix $W_{:,J}W_{:,J}^{\top}$.
\end{itemize}
\end{corollary}

Several methods that look orthogonal fail the hypothesis. Each item below is a separate computation, and
Table~\ref{tab:erlangen-orthogonal} summarises them.

\begin{remark*}{scope of Corollary II.6; these statements do not follow from its hypothesis}
\begin{itemize}
  \item \emph{BOFT} as implemented in HF PEFT, $W=\diag(s)W_0R\T$ with $R$ exactly orthogonal, has
    $K_{\mathrm{out}}(W)=\diag(s)K_{\mathrm{out}}(W_0)\diag(s)$. So $\cos'_{ij}=\operatorname{sign}(s_is_j)\cos_{ij}$
    whenever $s_is_j\neq0$. It preserves $\lvert\cos\rvert$ when every $s_i$ is nonzero, and preserves $\cos_{ij}$
    whenever $s_is_j>0$. Hence it preserves every cosine when all $s_i$ are nonzero and share a sign (for instance
    $s\in T_m^+$, the constraint in the re-scaled OFT of \citet{qiu2023controlling}, Sec.~3.5); when no cosine of $W_0$
    is zero, that condition is also necessary. HF's \texttt{boft\_s} is an unconstrained real initialised at $1$. At the
    default \texttt{boft\_n\_\allowbreak butterfly\_factor=1} the butterfly permutation is the identity, BOFT is block OFT times an
    output scale, and each input block satisfies
    $W_{:,J}W_{:,J}^{\top}=\diag(s)(W_0)_{:,J}(W_0)_{:,J}^{\top}\diag(s)$. HF PEFT OFT 0.14 to 0.15 (exact Cayley, input
    side) carried a trainable output scale \texttt{oft\_s} and has the same form.
  \item \emph{GOFT's official code} by default (\texttt{no\_scaling=False}) adds a trainable input-side diagonal $g$,
    initialised at ones. Then $W=W_0\diag(g)G$ and $K_{\mathrm{out}}(W)=W_0\diag(g)^2W_0\T$, whatever the Givens part $G$
    is. This is the paper's GOFT* ablation, and it is not covered.
  \item \emph{Output-side implementations} (LyCORIS diag-OFT at its defaults; HF PEFT OFT up to 0.13, which merged to
    $R\T W_0$; HF OFT on embedding layers; ETHER with \texttt{flip\_side=True}) have image in $\Ort(m)\cdot W_0$. They
    preserve $K_{\mathrm{in}}$, the partial input Grams $W_{I,:}^{\top}W_{I,:}$ and the spectrum, and they do not
    preserve $K_{\mathrm{out}}$. ETHER's official code with \texttt{ether\_nb>1} applies a different reflection to each
    output-row group, which keeps neuron norms and within-group angles but neither the cross-group cosines nor the
    spectrum.
  \item \emph{HF PEFT's default OFT} since 0.18 (\texttt{use\_cayley\_\allowbreak neumann=True}, 5 terms) builds
    $R=I+2Q+2Q^2+2Q^3+Q^4=C(Q)(I-Q^4)$, where $C(Q)=(I+Q)(I-Q)^{-1}$ is the exact Cayley map (HF's exact branch
    evaluated at $-Q$). Hence $R\T R=(I-Q^4)^2\ne I$, and the singular values of $R$ are $\lvert1-\theta_k^4\rvert$,
    where $\pm i\theta_k$ are the eigenvalues of $Q$: a systematic contraction for small $Q$, and $R$ is singular when
    some $\theta_k=1$. Already the adapted weight has $K_{\mathrm{out}}=W_0(I-Q^4)^2W_0\T$, a drift of order
    $\lVert Q\rVert^4$, and merge cycles compound it. For $\lVert Q\rVert_{\mathrm{op}}\le2^{1/4}$ every factor is a
    contraction, so the drift is a monotone shrinkage of neuron norms and singular values (\thmref{Theorem}{V.3},
    corollary 4). In 0.16 and 0.17 the last coefficient was $2$, so $R\T R=I+4Q^6+4Q^8$, still not orthogonal, with
    drift of order $\lVert Q\rVert^6$. (HF's docstring says the default is \texttt{False}; the dataclass default is
    \texttt{True}.)
  \item \emph{Quasi-orthogonal variants} preserve neither $K_{\mathrm{out}}$ nor its normalised form exactly: qGOFT,
    whose $2\times2$ blocks are arbitrary matrices with only a soft orthogonality penalty; ETHER+, which is two-sided and
    whose factor $I-uu\T+vv\T$ is singular when $u\perp v$; and the default HF OFT above. Once $n>2$, a single
    non-orthogonal $2\times2$ block, conformal or not, generically changes some cosine.
  \item \emph{Finite precision.} The invariance is exact only in exact arithmetic; low-precision merges add a rounding
    drift that grows like a random walk over merge cycles.
\end{itemize}
\end{remark*}

\paragraph{Numerical check}
With $s=(-0.5,1,\dots,1)$, BOFT changes a cosine by $1.5$ while $\lvert\cos\rvert$ is fixed to $4\times10^{-16}$, and an
all-negative $s$ fixes every cosine to $10^{-15}$. GOFT with a scaler $g=1\pm0.05$ moves cosines and singular values. A
conformal block $\sqrt2\,\mathrm{Rot}(0.3)$ on two of $8$ input coordinates changes some cosine, by $0.17$. HF's default
OFT satisfies $R\T R=(I-Q^4)^2$ to $5\times10^{-16}$, with $\lVert R\T R-I\rVert=1.5\times10^{-2}$ already for block
size $8$ and skew entries of size about $0.07$, and it is singular at $\theta=1$ (Appendix~\ref{app:repro}).

\begin{table}[t]
\centering\footnotesize
\caption{Implementations that look orthogonal, read against \thmref{Corollary}{II.6}. Each row is a separate computation
(see the remarks after the corollary).}
\label{tab:erlangen-orthogonal}
\begin{tabularx}{\linewidth}{@{}>{\raggedright\arraybackslash}p{2.9cm} >{\raggedright\arraybackslash}p{3.4cm}
  >{\raggedright\arraybackslash}p{1.7cm} >{\raggedright\arraybackslash}X@{}}
\toprule
\headrow \textbf{implementation} & \textbf{map} & \textbf{exactly orthogonal?} & \textbf{what it keeps}\\
\midrule
BOFT in HF PEFT; HF PEFT OFT 0.14--0.15 & $\diag(s)\,W_0R\T$, $R$ exactly orthogonal, $s$ free, initialised at $1$ & only $R$
  & $K_{\mathrm{out}}\mapsto\diag(s)K_{\mathrm{out}}\diag(s)$, and block Grams likewise. $\lvert\cos_{ij}\rvert$
  survives when $s_is_j\ne0$, $\cos_{ij}$ when $s_is_j>0$. At HF's default single butterfly factor, BOFT is block OFT
  times $s$.\\
GOFT, official code default & $W_0\,\diag(g)\,G$, with a trainable input scale $g$ & no & Nothing on this list, since
  $K_{\mathrm{out}}=W_0\diag(g)^2W_0\T$.\\
output-side OFT (LyCORIS diag-OFT at its defaults; HF PEFT OFT up to 0.13) & image in $\Ort(m)\cdot W_0$ & yes, on the
  other side & $K_{\mathrm{in}}$, the partial input Grams and the spectrum. Not $K_{\mathrm{out}}$.\\
HF PEFT default OFT since 0.18 (\texttt{use\_cayley\_\allowbreak neumann=True}) & $R=I+2Q+2Q^2+2Q^3+Q^4=C(Q)(I-Q^4)$, with $C$ an
  exact Cayley map & no, $R\T R=(I-Q^4)^2$ & Nothing exactly. The adapted weight already has
  $K_{\mathrm{out}}=W_0(I-Q^4)^2W_0\T$, and for small $Q$ each merge shrinks $K_{\mathrm{out}}$ and the spectrum
  further. (In 0.16--0.17, $R\T R=I+4Q^6+4Q^8$.)\\
qGOFT & free $2\times2$ blocks with a soft orthogonality penalty & no & Neither $K_{\mathrm{out}}$ nor the cosines. Once
  $n>2$, one non-orthogonal block, conformal or not, generically moves a cosine.\\
ETHER+ & two-sided, $I-uu\T+vv\T$, singular when $u\perp v$ & no & Neither $K_{\mathrm{out}}$ nor the cosines.\\
\bottomrule
\end{tabularx}
\end{table}

\paragraph{Credit}
\citet{qiu2023controlling} introduced orthogonal fine-tuning, motivated by preserving hyperspherical energy
\citep{liu2018learning}; their re-scaled OFT (Sec.~3.5) adds the positive per-neuron scale that HF BOFT carries. BOFT is
due to \citet{liu2023parameter}, building on butterfly factorisations \citep{dao2019learning}; GOFT and qGOFT are due to
\citet{ma2024parameter}, with an official peft-givens implementation; ETHER and ETHER+ to \citet{bini2024ether}; HRA to
\citet{yuan2024bridging}. LyCORIS is described by \citet{yeh2023navigating}. The Cayley--Neumann default, BOFT's scale
and the output-side merge are read from the Hugging Face PEFT source \citep{xmangrulkar2022peft} (the OFT configuration
and the BOFT layer).

\begin{explains}
``Orthogonal fine-tuning preserves knowledge'' corresponds to preserving a complete invariant: in exact arithmetic an
exactly orthogonal method can \emph{never} change spectra or neuron geometry, through any number of merges
(\thmref{Theorem}{V.3}, corollary 7), up to rounding. So a variant that must change them adds a non-orthogonal factor
(the per-neuron scale of Qiu et al.'s re-scaled OFT, as in HF BOFT and HF OFT 0.14 to 0.15, or GOFT's default input
scaler). HF's default OFT is only close to orthogonal; with block size $8$ and fresh skew generators with entries near $0.05$ in
every cycle, its merges shrink $K_{\mathrm{out}}$ by about $0.2\%$, $1.3\%$, $6\%$ and $22\%$ after $1$, $10$, $50$ and
$200$ cycles (relative Frobenius change, averaged over draws). The statement
concerns reach. It does not show that preserving the invariant causes knowledge retention, and the control study of
ETHER \citep{bini2024ether} found a non-orthogonal multiplicative variant comparably stable. Earlier claims that BOFT or
qGOFT preserve the full Gram matrix are corrected here.
\end{explains}

\subsubsection{HRA is a meet of two images}\label{sec:erlangen-hra}

A Householder reflection $H_u=I-2uu\T/\lVert u\rVert^2$ is orthogonal and differs from $I$ by rank one. So
$W_0H_{u_1}\cdots H_{u_r}$ lies in the orthogonal orbit of $W_0$ and in LoRA's cone $W_0+\Mr$, and for injective $W_0$
and even $r$ the Cartan--Dieudonn\'e theorem (\bgref{D.11}) says that HRA fills the cone's intersection with the
rotation orbit. HF PEFT repeats each reflection at initialisation, so to first order each pair can only turn its own
vector against the rest of the space, and HRA starts at an apex, as zero-init LoRA does (\thmref{Theorem}{II.2}).

\begin{theorem}{II.7}{HRA is the meet of LoRA with the rotation orbit, and it starts at an apex}
Let $W_0\in\R^{m\times n}$ be injective ($\rk W_0=n$), and let $r$ be even with $2\le r\le n-2$. Define
$\mathrm{HRA}_r:(u_1,\dots,u_r)\mapsto W_0H_{u_1}\cdots H_{u_r}$ with $u_i\ne0$, where
$H_u=I-2uu\T/\lVert u\rVert^2$. Write $\mathrm{OFT}_{\SO}:=W_0\cdot\SO(n)$ for the rotation orbit. (The $\Ort(n)$-orbit
would add the odd component: $W_0H_{e_1}$ lies in $W_0+\Mr$ but not in $\Im\mathrm{HRA}_r$.)
\begin{enumerate}[label=(\alph*)]
  \item \textbf{Meet.} $\Im\mathrm{HRA}_r=(W_0+\Mr)\cap W_0\cdot\SO(n)$. Further, $d(\mathrm{HRA}_r)=rn-\tfrac{r(r+1)}2$,
    so the generic fibre has dimension $\tfrac{r(r+1)}2=rn-d$. This gauge is non-linear. The evident continuous global
    symmetries are the $r$ scalings $u_i\mapsto c_iu_i$; the braid group also acts globally by Hurwitz moves
    $(u_i,u_{i+1})\mapsto(H_{u_i}u_{i+1},u_i)$, since $H_aH_b=H_{H_ab}H_a$.
  \item \textbf{First-order collapse.} At the paired initialisation $u_{2i-1}=u_{2i}=w_i$ ($i\le k=r/2$, with the $w_i$
    independent and $U=\spanop\{w_i\}$),
    \begin{equation*}
      S_1=W_0\cdot\{\Omega\in\so(n):P_{U^\perp}\Omega P_{U^\perp}=0\},\qquad\dim S_1=kn-\tfrac{k(k+1)}2<d(\mathrm{HRA}_r).
    \end{equation*}
  \item \textbf{Vertex.} The germ of $\Im\mathrm{HRA}_r$ at $W_0$ is diffeomorphic to the germ at $0$ of
    $\Sigma_r=\{\Omega\in\so(n):\rk\Omega\le r\}$, the skew-determinantal cone (\bgref{C.13}), which is not linear. So
    $W_0$ is a singular point (the vertex) of the image. Together with (b), the paired initialisation is an \textbf{apex}.
  \item \textbf{Scope.}
    \begin{itemize}
      \item \emph{Not a product.} (a) is a meet in the poset of subsets of $\Theta$ containing $W_0$
        (\thmref{Observation}{I.4}(c)). HRA is not the
        categorical product of $\mathrm{LoRA}_r$ and $\mathrm{OFT}_{\SO}$: the set-level fibre product
        $\{(B,A,R):W_0+BA=W_0R\}$ has local dimension $d+r^2=rn+\tfrac{r(r-1)}2\ne rn$ over a generic point of the
        meet. (It is not pure-dimensional: over $W_0$ alone it contains $\{(B,0,I)\}$, of dimension $mr$.)
      \item \emph{Exactly orthogonal OFT is smaller.} The Cayley map never produces a rotation with eigenvalue $-1$.
        For exact full-block Cayley OFT (the original paper; HF with \texttt{use\_cayley\_\allowbreak neumann=False}),
        $(W_0+\Mr)\cap\Im\mathrm{OFT}$ is dense in $\Im\mathrm{HRA}_r$ but not equal to it: $W_0H_{e_1}H_{e_2}$, with
        $H_{e_1}H_{e_2}=\diag(-1,-1,1,\dots,1)$, is missing. For exact block Cayley OFT with $b<n$ the intersection is a
        proper subset of $\Im\mathrm{HRA}_r$.
      \item \emph{HF's default OFT is incomparable.} HF PEFT's default OFT is not orthogonal (\thmref{Corollary}{II.6};
        \texttt{use\_cayley\_\allowbreak neumann=True}). Since $R\T R=(I-Q^4)^2=I$ forces $Q=0$ near $Q=0$, its image meets
        $\Im\mathrm{HRA}_r$ near $W_0$ only at $W_0$, while its intersection with $W_0+\Mr$ contains non-orthogonal
        points such as $W_0(I+2Q+2Q^2+2Q^3+Q^4)$ with $\rk Q=2$.
      \item \emph{Implementation.} The theorem concerns HRA with $\lambda<\infty$ (HF \texttt{apply\_GS=False}). The
        strictly orthogonal Gram--Schmidt variant has the smaller image $W_0(I-2P_U)$, of dimension $r(n-r)$. The
        paired initialisation is HF's default (\texttt{repeat\_interleave}) and is not stated in the HRA paper. HF HRA
        acts on \texttt{in\_features}, so injectivity requires \texttt{out\_features}${}\ge{}$\texttt{in\_features}
        (and full column rank); the theorem does not cover \texttt{down\_proj} or GQA key/value projections. The
        Gram--Schmidt variant never contains $W_0$, since $I-2P_U\ne I$, and at HF's default paired initialisation it
        runs Gram--Schmidt on duplicated columns, which is numerically degenerate.
    \end{itemize}
\end{enumerate}
\end{theorem}

\paragraph{Numerical check}
At $n=9$, the paired-initialisation Jacobian has rank $8$ against generic rank $15$ for $r=2$, and $15$ against $26$
for $r=4$ (Appendix~\ref{app:repro}).

\paragraph{Credit}
The sharp Cartan--Dieudonn\'e theorem, that a rotation moving a $k$-dimensional space is a product of exactly $k$
reflections, is due to \citet{xscherk1950decomposition}; see \citet{xartin1957geometric}. Householder parametrisations
in deep learning go back to \citet{mhammedi2016efficient}, and Lie-algebra charts such as the Cayley map to
\citet{lezcanocasado2019cheap}. \citet{yuan2024bridging} proposed HRA as a bridge between low-rank and orthogonal
adaptation, which the theorem makes precise. The paired initialisation is read from the HF PEFT HRA layer
\citep{xmangrulkar2022peft}.

\begin{explains}
\begin{itemize}[leftmargin=1.2em]
  \item HRA's ``bridge between low-rank and orthogonal adaptation'' is, at the level of images, the meet of LoRA with the
    rotation orbit $W_0\cdot\SO(n)$. It is a meet of images and no product of methods. It strictly contains
    LoRA${}\cap{}$OFT for exactly orthogonal (Cayley) OFT, as in the original paper or HF with
    \texttt{use\_cayley\_\allowbreak neumann=False}; for HF's default Cayley--Neumann OFT the two are incomparable. Compare
    \thmref{Proposition}{I.8}(a), where LoRA-XS \citep{baazy2024lora} is a genuine categorical product
    $\mathrm{FA}\times\mathrm{FB}$.
  \item HF recommends an even $r$ because its paired initialisation needs pairs, and for odd $r$ it falls back to a
    random Kaiming initialisation. The determinant identity $\det R=(-1)^r$ explains why no odd-$r$ HRA can start at
    $W_0$, so that fallback is a pointing defect.
  \item New here (earlier analyses labelled this base point smooth): \textbf{HRA's default initialisation is an apex}. At
    first order, each pair can only rotate its own vector against the rest of the space. This is the multiplicative twin
    of LoRA's dead $A$-direction (\thmref{Lemma}{II.4}), so HRA belongs in the apex column beside LoRA. Whether the
    collapse slows early training is open (\S\ref{sec:analogy}).
  \item Rebasing commutes with the meet, so after $K$ cycles re-HRA reaches $(W_0+\mathcal M_{\le Kr})\cap W_0\cdot\SO(n)$
    (\thmref{Theorem}{V.3}, corollary 5).
\end{itemize}
\end{explains}

In the Gram lab, choosing HRA and resetting to $q_0$ gives $\dim S_1=3$ against $d=5$, the values of (a) and (b) at
$n=4$, $r=2$. Moving $u_2$ keeps the Gram, spectrum and rank lamps lit together, and the preset $u=(e_1,e_2)$ lands on
$W_0H_{e_1}H_{e_2}$, which exact Cayley OFT never reaches.

\subsubsection{LOFT fixes one leaf of the meet}\label{sec:erlangen-loft}

The proof of \thmref{Theorem}{II.7} (\S\ref{proof:II.7}) already shows how the meet is built. It is the union of
the orbits $W_0\cdot\SO(V)$ over the $r$-dimensional input subspaces $V$, one subgroup orbit for each point of the
Grassmannian $\Gr(r,n)$ (\bgref{C.15}). Call each orbit a \emph{leaf}. Leaves overlap along lower-rank strata, so the
leaves cover the meet without partitioning it; a generic point lies on exactly one leaf, fixed by its moved space
$\operatorname{im}(R-I)$. HRA's reflection vectors move the leaf as it trains. LOFT \citep{zhao2026loft} freezes the
leaf and makes the choice of leaf the design variable, which the paper calls the adaptation support.

\begin{remark*}{2026: LoCO}
One component of LoCO \citep{nguyen2026loco} is the Cayley rotation of a low-rank skew generator $UV\T-VU\T$, with
$U,V\in\R^{n\times r}$. For injective $W_0$ and $2r\le n-2$, its image is $\Im\mathrm{HRA}_{2r}$ minus the points $W_0R$
at which $R$ has eigenvalue $-1$, so like HRA it moves its leaf while it trains. Writing the generator as $ZJZ\T$ with
$Z=[U\ V]$ shows that its generic fibre is an orbit of the linear group $\mathrm{Sp}(2r,\R)$, acting by $Z\mapsto ZM$.
The non-linear gauge of \thmref{Theorem}{II.7}(a) is therefore a property of the reflection parametrisation; LoCO's
component reaches an open dense part of the same set with the linear gauge $\mathrm{Sp}(2r,\R)$. A sum of several LoCO components is in general only approximately
orthogonal, and then leaves the meet.
\end{remark*}

\begin{example*}{LOFT: a leaf, and how to choose it}
Let $P_r\in\R^{r\times n}$ have orthonormal rows spanning $V$, and let $\Ort(V)=\{I+P_r\T(T-I)P_r:T\in\Ort(r)\}$ be the
orthogonal maps that fix $V^\perp$ pointwise. LOFT trains $\rho(E)=W_0\big(I+P_r\T(Q(E)-I)P_r\big)$ with $E\in\so(r)$,
$E_0=0$, so $\lvert M\rvert=r(r-1)/2$. We read it with an exactly orthogonal $Q$ (the exponential, or the Cayley map
$C$); the released code defaults to a five-term Cayley--Neumann series with $R\T R=(I-X^6)^2$, which is only
approximately orthogonal, so like HF's default OFT (\thmref{Corollary}{II.6}) that version leaves its leaf.
\begin{enumerate}[label=(\alph*)]
  \item \emph{One leaf.} If $W_0P_r\T$ has rank $r$, then
    $\big(W_0+W_0P_r\T\,\R^{r\times r}P_r\big)\cap W_0\cdot\Ort(n)=W_0\cdot\Ort(V)$. For the principal support
    $P_r=V_r\T$ the first set is the image of LoRA-XS (this needs $\sigma_r(W_0)>0$), so the leaf is
    LoRA-XS${}\cap W_0\cdot\Ort(n)$. Principal-support LOFT reaches an open dense part of its component $W_0\cdot\SO(V)$;
    for injective $W_0$ and a Cayley $Q$, its image is exactly LoRA-XS${}\cap{}$the image of exact full-block Cayley
    OFT.
  \item \emph{Leaves over the Grassmannian.} If $W_0$ is injective,
    $(W_0+\Mr)\cap W_0\cdot\SO(n)=\bigcup_{V\in\Gr(r,n)}W_0\cdot\SO(V)$, which for even $r$ is $\Im\mathrm{HRA}_r$
    (\thmref{Theorem}{II.7}(a)). LOFT's image is the whole leaf $W_0\cdot\SO(V)$ with the exponential, and an open dense
    part of it with a Cayley map, which never reaches eigenvalue $-1$.
  \item \emph{A regular base point.} If $W_0P_r\T$ has rank $r$, then $S_1=W_0P_r\T\so(r)P_r$ has dimension
    $r(r-1)/2=d(M)$: LOFT starts at a smooth point of its leaf and has no gauge, while HRA's paired initialisation is an
    apex (\thmref{Theorem}{II.7}(b)). For $r\ge4$ there is no smooth pointed arrow LOFT${}\to{}$HRA at that
    initialisation, although the images are nested (obstruction test O3 of \S\ref{sec:atlas}).
  \item \emph{Choosing the leaf.} If $Q'(0)=\id$ (the exponential, or $E\mapsto C(E/2)$), the gradient pulled back to $E$
    at $\theta_0$ is $P_r\skewop(W_0\T G)P_r\T$ with $G=\nabla L(W_0)$; the Cayley map $C$ itself doubles it, which does
    not change the best support. Its squared norm is at most $2\sum_{k\le r/2}\mu_k^2$, where $\pm i\mu_k$ are the
    eigenvalues of $\skewop(W_0\T G)$, with equality when $V$ spans the invariant subspace of the top $\lfloor r/2\rfloor$
    pairs.
\end{enumerate}
\end{example*}

Appendix~\ref{proof:loft} proves (a)--(d).

\paragraph{Numerical check}
For $n=8$, $m=10$ and $r=2,4,6$, the leaf identity, the union, the arrows into exact Cayley OFT and LoRA-XS, and
$\dim S_1=r(r-1)/2$ all hold to about $10^{-15}$ (Appendix~\ref{app:repro}).

\paragraph{Credit}
LOFT, the SkewGrad and GradSVD supports, part (d) and its bound (their Proposition~2), the reading of each support as
selecting a Lie subgroup (their Remark~1) and the suggestion of dynamic supports (their Remark~3) are from
\citet{zhao2026loft}. Their Table~1 writes full and block OFT, GOFT, BOFT, HRA and PSOFT as choices of support, width,
depth and transform in a multi-factor version of the same primitive. Parts (a)--(c) place the single-factor method inside
\thmref{Theorem}{II.7}. PSOFT is due to \citet{wu2025efficient}; LoRA-XS to \citet{baazy2024lora}.

\begin{explains}
Orthogonal methods differ in which leaves they use and how they combine them. Block OFT, GOFT and BOFT multiply rotations
of fixed coordinate subspaces; HRA moves its leaf while it trains; PSOFT takes the principal leaf and adds two tori; LOFT
takes one leaf, chosen before training from $W_0$ or from the gradient at $\theta_0$. By part (d), SkewGrad is the
orthogonal counterpart of the alignment mechanism of \thmref{Proposition}{II.3}, and GradSVD transplants the
gradient-based frames of LoRA-GA \citep{wang2024lorab}. With the leaf frozen, merge-and-restart gains nothing. With
finitely many supports revisited across restarts, the reach is the orbit of the connected group whose Lie algebra
(\bgref{D.8}) the $\so(V_k)$ generate (\thmref{Theorem}{V.3}(b)); numerically, generic supports with $r\ge2$ generate all
of $\SO(n)$ once $\lceil n/r\rceil$ of them are used. The principal support does not move when it is re-selected after
a merge, and data-chosen supports have to be checked case by case.
\end{explains}

\subsubsection{DoRA is a torus saturation of LoRA}\label{sec:erlangen-dora}

DoRA normalises each row of LoRA's matrix $V=W_0+BA$ and multiplies row $i$ by a learned magnitude $\mu_i$. That is one
row scaling, by $\mu_i/\lVert v_i\rVert$, so every DoRA weight is a diagonal matrix times a LoRA weight, and conversely
up to a repair for vanishing rows. We write $\mathrm{Diag}_m$ for the monoid of diagonal matrices (\bgref{D.2}),
$\mathrm{Mon}_m=T_m\rtimes S_m$ for the monomial matrices (\bgref{D.3}), and $N$ for row normalisation.

\begin{proposition}{II.8}{DoRA is LoRA followed by an output scale, at the level of images}
Assume $1\le r\le\min(m,n)$ and $s\ne0$. Let $N$ normalise rows; DoRA's magnitude $\mu\in\R^m$ is per output neuron, as
in HF PEFT and the official code (the paper's notation says ``column-wise'', in the transposed convention). Images refer
to the merged (eval-mode) weight; HF's train-mode forward with \texttt{lora\_dropout}${}>0$ is not of the form $Wx$. Let
$W_0$ have no zero row, let $\mathcal U$ be the open set of matrices with no zero row, and let $\mathrm{Diag}_m$ be the
monoid of all diagonal matrices. Then
\begin{equation*}
  \Im\mathrm{DoRA}_r=\{\diag(\mu)\,N(V):V\in(W_0+\Mr)\cap\mathcal U,\ \mu\in\R^m\}=\mathrm{Diag}_m\cdot(W_0+\Mr),
\end{equation*}
which is exactly the image of ``$\mathrm{LoRA}_r$ followed by a trainable output scale $\ell\in\R^m$''. With $\mu$
restricted to $(\R^\times)^m$, or equivalently after intersecting with $\mathcal U$:
$\Im\mathrm{DoRA}_r\cap\mathcal U=T_m\cdot(W_0+\Mr)\cap\mathcal U$. Moreover:
\begin{itemize}
  \item $d(\mathrm{DoRA}_r)\le\min\big(mn,\ r(m+n-r)+m\big)$. At a parameter point with $\mu\in(\R^\times)^m$ and
    $X=sBA\in\MM_r$, with $U=\operatorname{col}X$ and $R=\operatorname{row}X$, the Jacobian rank is
    $r(m+n-r)+m-\dim\ker\big(\delta\mapsto P_{U^\perp}\diag(\delta)W_0P_{R^\perp}\big)$; where some $\mu_i=0$ it is
    smaller when $r\ge2$, and for $r=1$ when some row $w_{0,i}$ with $\mu_i=0$ lies outside $R$. For generic $W_0$, $d=\min\big(mn,\ r(m+n-r)+m\big)$; in particular $d=r(m+n-r)+m$ iff
    $(m-r)(n-r)\ge m$, which holds for all practical shapes. For special $W_0$ it is smaller: for $W_0=ac\T$ with all
    $a_i\ne0$ and $r<n$, $d=r(m+n-r)+m-r$, strictly below the generic value iff $r<m$ and $r\le n-2$.
  \item The image covariance group (\thmref{Definition}{II.14}) contains $\mathrm{Mon}_m\times\GL_n$.
  \item At $B=0$, in HF's magnitude coordinates, $\Delta W=\big(\diag(\mu_i/\lVert w_{0,i}\rVert)-I\big)W_0$, which can
    have rank up to $\rk W_0$, so DoRA escapes LoRA's rank bound.
\end{itemize}
\end{proposition}

\paragraph{Numerical check}
The Jacobian rank is $9=mn$ at $(m,n,r)=(3,3,2)$, where $r(m+n-r)+m$ gives $11$; it is $36$ at $(8,8,2)$, matching, and
$34$ there when one $\mu_i=0$; it is $24$ for a rank-one $W_0$ at $(6,6,2)$ (Appendix~\ref{app:repro}).

\paragraph{Credit}
DoRA is due to \citet{liu2024dora}. Its magnitude--direction split follows weight normalisation
\citep{salimans2016weight}. From a different direction, \citet{zhang2025primacy} argue that the magnitude of the update
is the main driver of LoRA's performance.

\begin{explains}
At the level of images, DoRA's ``magnitude--direction decoupling'' is LoRA plus $m$ free output gains. What DoRA changes
beyond that lives in its training dynamics. Under gradient flow or SGD there are two different geometries. The exact
gradient of the normalised parametrisation projects the direction gradient row by row, which is a change of metric. The
detached backward of the DoRA paper (Sec.~4.3) and of HF PEFT (\texttt{weight\_norm.detach()}) removes that projection;
the resulting update is a non-gradient pseudo-gradient that no metric describes, and it need not be a descent direction.
Both the paper and HF train with AdamW, under which neither variant follows a fixed metric, and the coordinates matter
in their own right: $\mu_i$ starts at $\lVert w_{0,i}\rVert$ and $\ell_i$ at $1$, so an equal step on $\mu_i$ and on a
gain $\ell_i$ changes the row gain by different amounts.
\end{explains}

\paragraph{Experiment D}
With one optimizer (AdamW, same learning rate, weight decay and data order) and the same $W_0$, compare four arms:
(i) LoRA + gain $\ell$ with $\ell_0=\mathbf 1$; (ii) LoRA + gain in DoRA's coordinates,
$W=\diag(\mu_i/\lVert w_{0,i}\rVert)(W_0+sBA)$ with $\mu_0=(\lVert w_{0,i}\rVert)_i$, which isolates the
optimizer-coordinate effect without normalisation; (iii) DoRA with the exact backward; (iv) DoRA with HF's detached
backward. Since the images are equal, any gap between arms comes from the parametrisation, the backward rule or the
optimizer coordinates, and the four arms separate these. The DoRA paper's own ablation reports (iii) and (iv) within
$0.2$ points. \thmref{Prediction}{X.2} uses the equal images to set up this controlled comparison.

The covariance group $\mathrm{Mon}_m\times\GL_n$ decides where \thmref{Observation}{II.15} applies. One orthogonal
rotation of the residual stream of a gain-fused pre-norm model, as in SliceGPT \citep{xashkboos2024slicegpt}, acts on the
input side of $q$, $k$, $v$, gate and up, where DoRA is $\GL_n$-covariant, so DoRA on those boxes reaches the same
functions after the rotation as before; on $o$ and down the sufficient condition fails. Training outcomes are a separate
question.

\subsubsection{The torus ladder}\label{sec:erlangen-ladder}

Write a weight as $W_0\odot(J+Z)$, with $J$ the all-ones matrix. Output scalings use $Z=(\ell-\mathbf 1)\mathbf 1\T$, of
rank one; two-sided scalings use $ab\T-J$, of rank at most two; HiRA allows any $Z=BA$ of rank at most $r$. These
choices form a ladder of images, and no rung can fill a zero of $W_0$.

\begin{proposition}{II.9}{the torus ladder}
\begin{enumerate}[label=(\alph*)]
  \item \textbf{HiRA.} $\rho(B,A)=W_0\odot(J+BA)$, where $J$ is the all-ones matrix; HF PEFT has $B_0=0$ and no scale
    factor on Linear and Conv layers (on embeddings it uses $A_0=0$). For every $W_0$,
    $\Im\mathrm{HiRA}_r=W_0+D_{W_0}(\Mr)$ with $D_{W_0}:X\mapsto W_0\odot X$. If $W_0$ has no zero entries,
    $D_{W_0}\in\GL(\Theta)$, so inside the slice at $\theta_0$ HiRA is LoRA transported by a diagonal automorphism: it has
    the same $d=r(m+n-r)$ (for $r\le\min(m,n)$) and the same gauge. In all cases $\rk\Delta W\le r\cdot\rk W_0$. The
    transport depends on $\theta_0$ (\thmref{Proposition}{II.13}).
  \item \textbf{Scalings.} Output-side scalings, $\diag(\ell)W_0=W_0\odot(J+(\ell-\mathbf 1)\mathbf 1\T)$, and
    input-side scalings, $W_0\diag(\ell)=W_0\odot(J+\mathbf 1(\ell-\mathbf 1)\T)$, have images in $\Im\mathrm{HiRA}_1$.
    Output-side examples: (IA)$^3$ on keys and values, the post-linear scale of SSF, DoRA's magnitude. Input-side
    examples: (IA)$^3$'s feed-forward vector, which scales the input of $W_{\mathrm{down}}$, and a LayerNorm gain
    $\gamma$ folded into the next linear map. The fold is exact when the norm's output is consumed only by linear maps:
    the input norms of pre-norm blocks, and a final norm before an untied head (in post-norm BERT, the last block's norm
    feeds only the head). It does not apply to Gemma-style post-sublayer norms, which HF's default LN-tuning targets for
    Gemma~2 and~3, to QK-norms, or to post-norm residual paths. Where it applies, $\gamma$ is tied across all consumers of
    the norm (Q, K and V, or gate and up) rather than being a free point per box, and with tied embeddings a fold into
    the head breaks the tie. LayerNorm biases and SSF's shift fold into the next bias, not into $W$. Two-sided scalings
    $\diag(a)W_0\diag(b)=W_0\odot(ab\T)$ lie in $\Im\mathrm{HiRA}_2$. As a morphism, $h(\ell)=(\ell-\mathbf 1,\mathbf
    1\T)$ lands in $\mathrm{HiRA}_1$ pointed at $(0,\mathbf 1\T)$, which is not HiRA's random-$A$ default pointing.
  \item \textbf{Orbit dimension.} The two-sided torus orbit $\{\diag(a)W_0\diag(b)\}$ has dimension $m+n-c(W_0)$, where
    $c(W_0)$ is the number of connected components of the bipartite support graph of $W_0$, an isolated vertex counting
    as its own component.
  \item \textbf{RoAd.} RoAd$_1$ (the paper's main variant, HF \texttt{road\_1}) is $W=RW_0$ with $R$ block-diagonal in
    scaled rotations $\alpha_i\mathrm{Rot}(\theta_i)$ of output pairs. It is the left action of
    $\CO^+(2)^{m/2}\cong(\mathbb C^\times)^{m/2}$: it is \textbf{(IA)$^{\mathbf 3}$ over $\mathbb C$}. It multiplies the
    input Gram matrix of each output pair by $\alpha_i^2$. Since $\alpha_i$ may be $0$, its image is the orbit closure,
    a real-linear space of dimension exactly $m$ whenever no output-pair block of $W_0$ is zero. RoAd$_2$ and RoAd$_4$
    (HF \texttt{road\_2}, \texttt{road\_4}) let each block range over all of $M_2(\R)$. They realise the block-diagonal
    action of the monoid $M_2(\R)^{m/2}$ (its group part $\GL_2^{m/2}$ is non-abelian), with linear image
    $\{\mathrm{blockdiag}(R_i)W_0\}$ of dimension $2m$ when every pair block has rank $2$. For them the Gram statement
    and ``over $\mathbb C$'' fail, and RoAd$_4$ carries a positive-dimensional gauge (two dimensions per row). HF pairs
    output $j$ with $j+g/2$ inside each group of size $g$ (default $64$) rather than adjacent outputs as in the paper,
    and also rotates the bias; the statements above do not depend on the pairing.
  \item \textbf{Zeros and covariance.} Every $W$ reachable by HiRA, (IA)$^3$ or SSF vanishes wherever $W_0$ does,
    $Z(W_0)\subseteq Z(W)$, and this persists under merging. New zeros can appear ($\ell_i=0$, or $(J+BA)_{ij}=0$), so
    the zero pattern itself is not invariant. The image covariance group of output-scaled boxes ((IA)$^3$ on $k,v$;
    post-linear SSF) is $\mathrm{Mon}_m\times\GL_n$, and that of input-scaled boxes ((IA)$^3$ on $W_{\mathrm{down}}$) is
    $\GL_m\times\mathrm{Mon}_n$. HiRA's is exactly $\mathrm{Mon}_m\times\mathrm{Mon}_n$ ($=\GL_1\times\GL_1$ when
    $m=n=1$).
\end{enumerate}
\end{proposition}

\paragraph{Numerical check}
The RoAd$_1$ orbit has dimension $8$ and the RoAd$_2$ image dimension $16$ at $m=8$ (Appendix~\ref{app:repro}).

\paragraph{Credit}
(IA)$^3$ is due to \citet{liu2022few}, SSF to \citet{lian2022scaling}, HiRA to \citet{huang2025hira}, RoAd to
\citet{liao20243}, and LayerNorm tuning to \citet{zhao2023tuning}. Feature-wise scaling as a building block goes back at
least to FiLM \citep{perez2017film}.

\begin{explains}
Each scaling method here is a rung $W_0\odot(J+Z)$: (IA)$^3$, SSF's scales and the gain part of LN-tuning on norms that
feed only linear maps lie in $\mathrm{HiRA}_1$, on one side or the other, and HiRA is LoRA in the Hadamard frame of
$W_0$. DoRA's image is the $\mathrm{Diag}_m$-orbit of LoRA's image (image level only; \thmref{Proposition}{II.8}). DoRA
and RoAd can fill zeros of $W_0$, so they become rungs only when $W_0$ is zero-free. Then every image germ can be written
as $W_0\odot(J+Z)$, so the ladder by itself constrains nothing there and serves as bookkeeping. RoAd$_1$ is complex
(IA)$^3$ and RoAd$_{2,4}$ are block-$M_2(\R)$ (IA)$^3$; for generic $W_0$ neither lies on a low-rank rung. HiRA's ``high
rank'' is a property of a frame: its transport $D_{W_0}$ depends on $\theta_0$ and does not preserve rank. Two re-based
cycles reach $W_0\odot(J+Z_1)\odot(J+Z_2)$, whose relative update $Z_1+Z_2+Z_1\odot Z_2$ has rank up to
$\min(2r+r^2,m,n)$, against $2r$ for LoRA (\thmref{Proposition}{II.13}).
\end{explains}

With HF defaults, (IA)$^3$ scales the output side of the attention projections it targets ($k$ and $v$, or $q$ and $v$)
and the input side of down, so the residual rotation above, acting on the other side, leaves its reachable functions
unchanged. HiRA's covariance group stops at $\mathrm{Mon}_m\times\mathrm{Mon}_n$ and fails the sufficient condition of
\thmref{Observation}{II.15}. That alone does not prove that its reachable functions change; Conjecture~A (stated after
\thmref{Observation}{II.15}) says they do, and \thmref{Prediction}{X.4} is the test.

Corollary~II.6, Theorem~II.7 and Propositions~II.8 and~II.9 were repaired after two rounds of adversarial review; the
referee record in \S\ref{sec:analogy} lists what changed.

The groups of this exposé see orbits. LoRA's cone is governed instead by the rank of the update, read off a cut in a
tensor network, while the size of an image is a dimension, which behaves like a volume. Rank and volume can move in
opposite directions, the subject of \S\ref{sec:cut}.

\subsection{Cut and volume}\label{sec:cut}

\begin{flushright}
\small\itshape Rank lives on cuts and dimension lives in volumes.\\
Count the wrong one and you will compare the wrong things.\\[2pt]
{\upshape\footnotesize a working maxim}
\end{flushright}

At a fixed budget, the narrowest cut of a method's diagram bounds the rank it can reach. The dimension of what it can
reach is a separate count, and the two can move in opposite directions.

\paragraph{In brief}
\begin{itemize}
  \item A method's update rank is bounded by the narrowest cut of its tensor-network diagram. The dimension $d(M)$ of
    what it can reach is its parameter count minus the directions that change nothing.
  \item The two can move in opposite directions. At the budget of LoRA$_8$ on a $4096\times4096$ layer, LoHa$_{4,4}$
    reaches rank $16$ and at least $8{,}159$ fewer dimensions; GraLoRA with $k=2$ reaches rank $16$ at LoRA's
    dimension, with every block capped at rank $4$.
  \item Kronecker sums are LoRA across another cut of the same four-leg tensor, and HiRA is LoRA after
    $X\mapsto W_0\odot X$ (for $W_0$ without zero entries). Such changes of frame keep dimension and move rank.
  \item LoRA respects every change of a layer's coordinates and so reaches the same set of functions after a
    function-preserving rotation of the network; PiSSA, MiLoRA and CorDA splits respect only rotations and uniform
    scalings.
\end{itemize}

PEFT papers usually report a method's expressiveness as a rank and its cost as a parameter count. A third number, the
dimension of the set of weights the method can reach, usually goes unreported.

\begin{figure}[tbp]
  \centering
  \includegraphics[width=\linewidth]{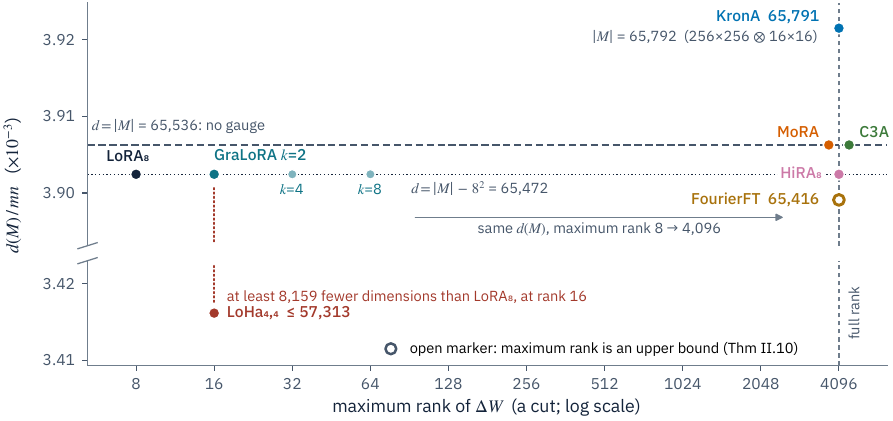}
  \caption{\textbf{Rank is not volume} (\thmref{Theorem}{II.10}, \thmref{Proposition}{II.12}). One $4096\times4096$ layer at the budget of LoRA$_8$ \citep{hu2021lora}, $\lvert M\rvert=8(m+n)=65{,}536$. Each method is placed by the largest rank of $\Delta W$ it can reach (horizontal, logarithmic scale), which is bounded by the narrowest cut of its diagram (\thmref{Theorem}{II.10}), and by its expressive dimension as a fraction of the layer, $d(M)/(mn)$ (vertical, in units of $10^{-3}$, with the axis broken between $3.41$ and $3.90$), from the gauge census of \thmref{Proposition}{II.12}. The dashed line is $d=\lvert M\rvert$, a method without gauge, and the dotted line is $d=\lvert M\rvert-8^2=65{,}472$, the dimension of LoRA$_8$ with its $\GL_8$ gauge. Along the dotted line the maximum rank runs from $8$ for LoRA$_8$, through GraLoRA \citep{jung2025gralora} with $k=2$, $4$ and $8$ blocks (the last two light), to full rank for HiRA$_8$ \citep{huang2025hira}. MoRA \citep{jiang2024mora} and C3A \citep{chen2024parameter} have no gauge and reach full rank; their points coincide and are drawn on either side of the full-rank line. FourierFT \citep{gao2024parameter} loses one dimension for each sampled conjugate pair, and its maximum rank is known only through a bound (open marker). KronA \citep{edalati2022krona} is shown at the factor shapes closest to the budget. LoHa$_{4,4}$ \citep{hyeonwoo2021fedpara}, below the break, sits at the proved upper bound of its dimension: it reaches rank $16$ with at least $8{,}159$ fewer dimensions than LoRA$_8$ (\thmref{Proposition}{II.12}(c)). The parameters behind each point are in \S\ref{app:repro-figures}.}
  \label{fig:rankvol}
\end{figure}

Draw a method as a network of boxes and wires. The rank of its update is limited by the narrowest place where the
network can be cut in two, so rank is a \emph{cut} invariant. The dimension counts the independent directions in which
the method can move the weights; we call it a \emph{volume} invariant, loosely, since no measure is computed. In
Figure~\ref{fig:rankvol}, several methods raise one while lowering the other.

\subsubsection{Rank is a cut}\label{sec:cut-rank}

\begin{figure}[tbp]
\centering
\begin{tikzpicture}[
  font=\small,
  box/.style={draw=ink, rounded corners=1.5pt, minimum width=8.5mm, minimum height=6.5mm, inner sep=2pt, line width=0.6pt},
  frozen/.style={box, fill=inkgray!14},
  train/.style={box, draw=tide, fill=tide!16, line width=0.8pt},
  wire/.style={draw=ink, line width=0.7pt},
  lab/.style={font=\footnotesize, text=inkgray},
  cutline/.style={draw=seal, line width=1pt, dash pattern=on 3pt off 2pt},
  altcut/.style={draw=inkgray!70, line width=0.6pt, densely dotted}
]
  \node[draw=inkgray, dashed, rounded corners=2pt, minimum width=13mm, minimum height=9mm] (hole) at (2.1,0) {};
  \node[lab] at (hole) {hole};
  \draw[wire] (0,0) -- (hole.west);
  \draw[wire] (hole.east) -- (4.2,0);
  \node[lab, anchor=south] at (0.55,0.08) {$V=\R^n$};
  \node[lab, anchor=south] at (3.65,0.08) {$U=\R^m$};
  \node[lab, anchor=north] at (2.1,-1.05) {(a) $W_0$ cut out of a linear layer};
  \begin{scope}[xshift=5.2cm]
    \node[frozen] (W0) at (0.9,0) {$W_0$};
    \draw[wire] (0,0) -- node[above,lab]{$n$} (W0.west);
    \draw[wire] (W0.east) -- node[above,lab]{$m$} (1.8,0);
    \node at (2.2,0) {$+$};
    \node at (2.6,0) {$s$};
    \node[train] (A) at (4.1,0) {$A$};
    \node[train] (B) at (6.4,0) {$B$};
    \draw[wire] (2.85,0) -- (A.west);
    \draw[wire] (A.east) -- (B.west);
    \draw[wire] (B.east) -- (7.95,0);
    \node[lab, anchor=south] at (3.05,0.08) {$n$};
    \node[lab, anchor=south] at (4.82,0.08) {$r$};
    \node[lab, anchor=south] at (7.1,0.08) {$m$};
    \draw[altcut] (3.4,0.5) -- (3.4,-0.5) node[below, lab]{$n$};
    \draw[altcut] (7.45,0.5) -- (7.45,-0.5) node[below, lab]{$m$};
    \draw[cutline] (5.25,0.62) -- (5.25,-0.62);
    \node[font=\footnotesize, text=seal, anchor=south] at (5.25,0.62) {cut of weight $r$};
    \node[lab, anchor=north] at (4.0,-1.05) {(b) the LoRA plug $W_0+sBA$};
  \end{scope}
\end{tikzpicture}
\caption{\textbf{LoRA as a plug.} (a) Cutting $W_0$ out of a linear layer $y=W_0x$ leaves a hole $V\to U$, with
  $V=\R^n$ and $U=\R^m$. (b) LoRA fills the hole with the frozen box $W_0$ (grey) plus the trainable chain $sBA$ (teal),
  a formal sum, over which rank is subadditive. The dashed cut puts $A$ on the input side and $B$ on the output side and
  crosses one wire of dimension $r$, so $\rk(sBA)\le r$ (\thmref{Theorem}{II.10}). The dotted cuts put both boxes on one
  side and cross an open leg, with weights $n$ and $m$; the minimum over cuts is $\min(n,r,m)$.}
\label{fig:loraplug}
\end{figure}
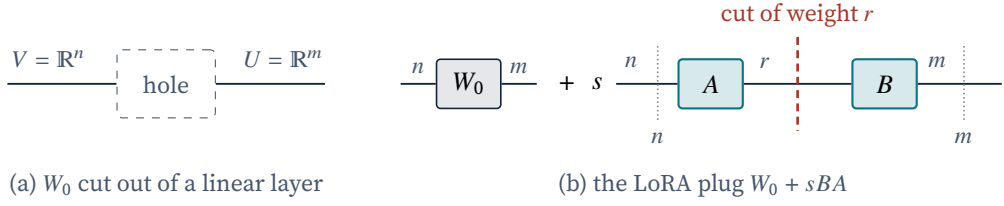

In LoRA every signal crosses the inner wire of dimension $r$ between $A$ and $B$, so $\Delta W=BA$ has rank at most $r$
(Figure~\ref{fig:loraplug}). The argument works for any method drawn as a tensor network (\bgref{B.7}), a string diagram
(\bgref{B.3}) in the symmetric monoidal category $(\FinVect,\otimes)$. Split the boxes into an input side and an output
side, and the network factors through the wires that cross.

\begin{theorem}{II.10}{cut-rank bound; standard}
Let $\Gamma$ be a tensor network, evaluated in the symmetric monoidal category $(\FinVect,\otimes)$, with input legs $V$
and output legs $U$. Every wire and open leg is assumed incident to a vertex: insert an identity box on any bare strand,
or equivalently count each bare input--output wire as crossing every cut. A crossing hyperedge (copy spider) contributes
its dimension once. Partition the vertices into $\mathcal I\sqcup\mathcal O$. The weight $w$ of the cut is the product
of the dimensions of the internal wires between the two sides and of the open legs attached on the wrong side (input legs
at $\mathcal O$, output legs at $\mathcal I$). Then
\begin{equation*}
  \rk\llbracket\Gamma\rrbracket\le\min_{\mathrm{cuts}}w,
\end{equation*}
and rank is subadditive over formal sums. The bound refers to a chosen presentation of the network.
\end{theorem}

\paragraph{Numerical check}
No violation in $283$ random networks (Appendix~\ref{app:repro}).

\paragraph{Credit}
String diagrams go back to \citet{xjoyal1991geometry}, and the copy spider (\bgref{B.6}) to the Frobenius algebra of a
basis \citep{xcoecke2013new}. Cut bounds on rank are standard tensor-network reasoning (\bgref{B.8};
\citealp{kolda2009tensor,cichocki2016low}). The PEFT reading is ours.

Without the convention on bare strands, $I_d\otimes BA$ would violate the bound. The bound also depends on the drawing. A
naive expansion of HRA's product of reflections has $2^r-1$ terms, while the telescoping presentation
\begin{equation*}
  R-I=\sum_{k=1}^{r}(H_k-I)\,H_{k+1}\cdots H_r
\end{equation*}
writes $R-I$ as $r$ terms of rank at most one and gives the sharp bound $r$.

\paragraph{Consequences}
For one box, \thmref{Theorem}{II.10} gives the bounds in Table~\ref{tab:cutbounds}. Relative to $W_0$, splitting
initialisations give $\Delta W=BA-B_0A_0$, of rank at most $2r$ (\thmref{Theorem}{II.1}). The Givens bound uses the same
telescoping presentation as HRA, each gate contributing a term of rank at most two, and HiRA's bound factors $W_0$
through a bond of dimension $\rk W_0$.

\begin{table}[t]
  \centering
  \footnotesize
  \caption{Cut bounds of \thmref{Theorem}{II.10} on $\rk\Delta W$ for one $m\times n$ box, with the presentation each
    bound uses.}
  \label{tab:cutbounds}
  \renewcommand{\arraystretch}{1.12}
  \begin{tabularx}{\linewidth}{@{}>{\raggedright\arraybackslash}p{4.6cm}>{\raggedright\arraybackslash}X>{\raggedright\arraybackslash}p{2.6cm}@{}}
    \toprule
    \headrow \textbf{method} & \textbf{presentation used} & \textbf{bound on $\rk\Delta W$}\\
    \midrule
    LoRA family, $BA$ \citep{hu2021lora} & one bond of dimension $r$ & $r$\\
    split initialisations (PiSSA type), relative to $W_0$ & $\Delta W=BA-B_0A_0$, two bonds & $2r$\\
    HRA \citep{yuan2024bridging} & telescoping sum $R-I=\sum_k(H_k-I)H_{k+1}\cdots H_r$ & $r$\\
    Givens circuit with $g$ gates \citep{ma2024parameter} & the same telescoping sum & $2g$\\
    ReLoRA after $K$ phases \citep{lialin2023relora} & formal sum of $K$ LoRAs & $Kr$\\
    GraLoRA \citep{jung2025gralora} & sum of $k^2$ block LoRAs & $kr$\\
    LoHa \citep{hyeonwoo2021fedpara} & Hadamard product through copy spiders & $r_1r_2$\\
    HiRA \citep{huang2025hira} & $W_0$ factored through a bond of dimension $\rk W_0$ & $r\,\rk W_0$\\
    LoKr \citep{yeh2023navigating} & $C\otimes BA$, inner rank $r'$ & $\rk C\cdot r'$\\
    FourierFT with $n_c$ coefficients \citep{gao2024parameter} & sum of real cosine patterns, each of rank $\le2$ &
      $\min(2n_c,m,n)$\\
    MoRA (reshape) \citep{jiang2024mora}, C3A \citep{chen2024parameter}, MPO adapters \citep{xliu2021enabling} & no
      narrow cut & up to full rank\\
    \bottomrule
  \end{tabularx}
\end{table}

\begin{explains}
A large min-cut is necessary for high rank and does not guarantee it. The ``high-rank PEFT'' literature consists of
networks whose min-cut avoids an $r$-dimensional bottleneck. Whether the bound is attained is a separate question, and
attaining it can cost volume.
\end{explains}

\subsubsection{Kronecker adapters cut the other way}\label{sec:cut-kronecker}

If $m=m_1m_2$ and $n=n_1n_2$, an $m\times n$ matrix is a tensor with output legs $U_1,U_2$ and input legs $V_1,V_2$
(\bgref{B.10}). LoRA's rank is the rank across $\{U_1U_2\}\mid\{V_1V_2\}$; a sum of $k$ Kronecker products has rank at
most $k$ across $\{U_1V_1\}\mid\{U_2V_2\}$. The two cuts disagree sharply, since $I\otimes I$ is one Kronecker term of
full rank.

\begin{theorem}{II.11}{Kronecker incomparability; the linear algebra is standard, the PEFT reading is new}
Write $\R^m=\R^{m_1}\otimes\R^{m_2}$ and $\R^n=\R^{n_1}\otimes\R^{n_2}$, and view $\Delta\in\R^{m\times n}$ as a 4-leg
tensor. Let $\operatorname{rk}$ be the rank across the cut $\{U_1U_2\}\mid\{V_1V_2\}$ and $\operatorname{rk}_\otimes$
the rank across $\{U_1V_1\}\mid\{U_2V_2\}$. Put
\begin{equation*}
  \kappa_1:=\min(m_1,m_2)\cdot\min(n_1,n_2),\qquad\nu:=\min(m_1,n_1)\cdot\min(m_2,n_2).
\end{equation*}

(a) $\operatorname{rk}_\otimes\Delta\le k$ iff $\Delta=\sum_{i\le k}C_i\otimes D_i$. Every $\Delta$ has
$\operatorname{rk}_\otimes\Delta\le\min(m_1n_1,m_2n_2)$.

(b) $\operatorname{rk}(C\otimes D)=\operatorname{rk}C\cdot\operatorname{rk}D$, while
$\operatorname{rk}_\otimes(C\otimes D)=1$ for $C,D\ne0$. So $\nu$ is the largest rank of a single Kronecker product.

(c) If $u\in\R^m$ and $v\in\R^n$ have Schmidt ranks $\rho_u,\rho_v$ (\bgref{B.12}), then
$\operatorname{rk}_\otimes(uv\T)=\rho_u\rho_v$. A generic rank-one update therefore has Kronecker rank $\kappa_1$. For
aligned splits, $(m_1-m_2)(n_1-n_2)\ge0$, which include equal square splits and every split HF PEFT's factorisation
produces (it puts the smaller factor first on both sides), $\kappa_1=\min(m_1n_1,m_2n_2)$ is the maximal Kronecker rank.

(d) \emph{Incomparability.} For $1\le r<\nu$ and $1\le k<\kappa_1$, the sets $\Mr$ and
$\mathcal K_{\le k}:=\{\Delta:\operatorname{rk}_\otimes\Delta\le k\}$ of unconstrained $k$-term Kronecker sums are
incomparable, for every split. A generic rank-one update lies in the first and not the second, and $C\otimes D$ with
full-rank factors, of Kronecker rank $1$ and rank $\nu$, lies in the second and not the first. For equal square splits
($m=n=N$, all factors $s$), $\kappa_1=\nu=s^2=N$, and $I_N$ is such a product.

(e) \emph{Constrained Kronecker methods.} Let $\Im M\subseteq W_0+\mathcal K_{\le k}$. Then
$\Im\mathrm{LoRA}_r\not\subseteq\Im M$ whenever $k<\kappa_1$, for every pointing of $\mathrm{LoRA}_r$. Give
$\mathrm{LoRA}_r$ its zero-init pointing, with image $W_0+\Mr$. Then $\Im M\not\subseteq\Im\mathrm{LoRA}_r$ iff some
$\theta\in\Im M$ has $\rk(\theta-W_0)>r$. For LoKr, $\Delta W=C\otimes BA$ with $C\in\R^{m_1\times n_1}$,
$B\in\R^{m_2\times r'}$ and $A\in\R^{r'\times n_2}$, put $r'_{\mathrm{eff}}=\min(r',m_2,n_2)$. The maximum rank is
$\min(m_1,n_1)\cdot r'_{\mathrm{eff}}$, attained. So $\Im\mathrm{LoKr}_{r'}\subseteq\Im\mathrm{LoRA}_r$ iff
$r\ge\min(m_1,n_1)\,r'_{\mathrm{eff}}$, and otherwise the two images are incomparable, provided $\kappa_1>1$. In HF PEFT
the second LoKr factor is low-rank only when its rank parameter satisfies $r'<\max(m_2,n_2)/2$; otherwise it is full,
and LoKr is unconstrained KronA, with maximum rank $\nu=\min(m_1,n_1)\min(m_2,n_2)$. On a $768$-wide layer
($24\otimes32$) the maximum rank therefore jumps from $360$ at $r'=15$ to $768$ at $r'=16$. With
\texttt{decompose\_both}, $C$ is also capped at rank $r'$.
\end{theorem}

\paragraph{Numerical check}
Schmidt ranks $(1,3)\to3$, $(2,3)\to6$, $(3,3)\to9$, each with ordinary rank $1$. For the $2\times4$ split of $\R^8$,
$\operatorname{rk}_\otimes(uv\T)=4$, the maximal Kronecker rank there. LoKr with $s=4$, $r'=1$ never exceeds rank $4$,
and at $(m_1,m_2,n_1,n_2)=(2,8,8,2)$ with inner dimension $4$, LoKr never exceeds rank
$4=\min(m_1,n_1)\,r'_{\mathrm{eff}}$ (Appendix~\ref{app:repro}).

\paragraph{Credit}
The rearrangement $\mathcal R(C\otimes D)=\vect C\,\vect D\T$ (\bgref{B.11}) and the nearest-Kronecker-product problem
are due to Van Loan and Pitsianis~\citeyearpar{loan1993approximation}. Kronecker-sum layers entered deep learning with
PHM layers \citep{zhang2021beyond} and Compacter \citep{mahabadi2021compacter}; KronA \citep{edalati2022krona} and LoKr
\citep{yeh2023navigating} brought them to weight updates.

\begin{explains}
Unconstrained Kronecker sums (KronA with full factors, sums of Kronecker terms) are LoRA across a different cut of the
same 4-leg tensor, so calling them higher-rank LoRAs mixes two cuts. With aligned splits, a generic rank-one update needs
the maximal number of Kronecker terms, and with fewer than $\kappa_1$ terms a Kronecker sum misses it. These methods also
rest on a reshape that the network does not have. Constrained Kronecker methods behave differently. LoKr with inner rank
$r'$ lies inside a LoRA image only when the LoRA rank is at least $\min(m_1,n_1)\,r'$, about $\sqrt N\,r'$: $192$ on
$768$-wide and $512$ on $4096$-wide layers at HF's default $r'=8$. At practical ranks, and at equal parameter budget,
LoKr and $\mathrm{LoRA}_r$ are incomparable (e).
\end{explains}

Compacter's PHM weights are Kronecker-structured weights inside a nonlinear bottleneck adapter and do not update $W_0$, and
MPO adapters \citep{xliu2021enabling} are a multi-core reparametrisation that is lossless at full bond dimension. The image
comparison does not transfer to either without more work.

\paragraph{Prediction}
On targets $W_0+uv\T$ with generic $u,v$, and a loss whose only minimiser is the target (for instance regression on
full-rank inputs), LoRA$_1$ reaches zero loss, while no Kronecker sum with fewer than $\kappa_1$ terms does. HF PEFT's
balanced factorisation is non-square for many widths ($768\to24\times32$ with $\kappa_1=576$, the maximal Kronecker rank;
$2048\to32\times64$; $5120\to64\times80$) but square for $1024$ and $4096$ ($32\times32$, $64\times64$), where
$\kappa_1=N$. At an equalised budget neither family should dominate the other across tasks.

\begin{figure}[tbp]
  \centering
  \includegraphics[width=\linewidth]{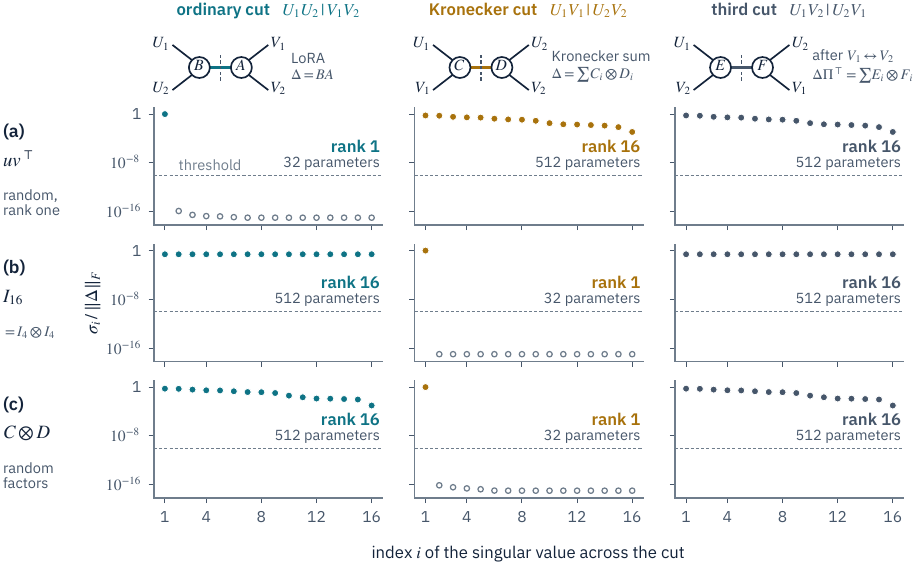}
  \caption{\textbf{Cut the tensor} (\thmref{Theorem}{II.11}). A $16\times16$ matrix $\Delta$ is read as a four-leg tensor with output legs $U_1,U_2$ and input legs $V_1,V_2$, through $\R^{16}=\R^4\otimes\R^4$ on both sides. The legs can be split into two pairs in three ways (columns): the ordinary cut $\{U_1U_2\}\mid\{V_1V_2\}$, across which LoRA factors as $\Delta=BA$; the Kronecker cut $\{U_1V_1\}\mid\{U_2V_2\}$, across which a sum of Kronecker products $\sum_iC_i\otimes D_i$ factors after the rearrangement of Van Loan and Pitsianis~\citeyearpar{loan1993approximation}; and the third cut $\{U_1V_2\}\mid\{U_2V_1\}$, the Kronecker cut with the two input legs swapped. The glyph above each column draws that factorisation as two boxes joined by a bond across the cut. Each panel shows the operator-Schmidt spectrum across one cut, the singular values $\sigma_i$ of the corresponding $16\times16$ reshape of $\Delta$ divided by $\lVert\Delta\rVert_F$, on a logarithmic scale, for (a)~a random rank-one update $uv\T$, (b)~the identity $I_{16}=I_4\otimes I_4$, and (c)~$C\otimes D$ with random $4\times4$ factors. Filled markers lie above the rank threshold (dashed line), and open markers are numerically zero. The ranks are $(1,16,16)$ for $uv\T$ and $(16,1,16)$ for $I_{16}$ and for $C\otimes D$. So a generic rank-one update has the maximal Kronecker rank $\kappa_1=16$, and a Kronecker product of full-rank factors has the maximal rank $\nu=16$, which makes zero-init LoRA$_r$ and $k$-term Kronecker sums incomparable for $r,k<16$ (\thmref{Theorem}{II.11}(b)--(d)). LoRA$_1$ fits $uv\T$ with $32$ parameters where a Kronecker sum needs $16$ terms and $512$ parameters, and for $I_{16}$ and $C\otimes D$ the two costs trade places.}
  \label{fig:cut}
\end{figure}

Figure~\ref{fig:cut} makes the three cuts of a $16\times16$ matrix concrete. On the $4\otimes4$ split, a generic $uv\T$
has rank $1$ and Kronecker rank $16$. LoRA$_1$ fits it with $32$ parameters, and a Kronecker sum needs $16$ terms and
$512$ parameters. For $I_{16}=I_4\otimes I_4$ the costs trade places.

\subsubsection{Dimension is a volume}\label{sec:cut-volume}

Moving $q\in Q$ along a fibre of $\rho$ changes nothing. For LoRA, $(Bg^{-1},gA)$ gives the same weight as $(B,A)$ for
every $g\in\GL_r$, so $r^2$ of the $r(m+n)$ parameters are spent on a symmetry and the image has dimension $r(m+n-r)$.
This is the expressive dimension $d(M)$ of \thmref{Definition}{I.5}, and $\lvert M\rvert-d(M)$ is the gauge waste.
``Gauge'' means the symmetry group of a parametrisation (\bgref{D.21}). There is no connection, and away from the
full-rank stratum the fibres jump in dimension.

\paragraph{Credit}
The quotient $(\R^{m\times r}_*\times\R^{r\times n}_*)/\GL_r$ is the classical geometry of fixed-rank matrices
\citep{mishra2012fixed}. \citet{putterman2024learning} build learners on LoRA weights that respect this $\GL_r$ symmetry.

The census below covers LoRA \citep{hu2021lora}, GraLoRA \citep{jung2025gralora}, LoHa \citep{hyeonwoo2021fedpara}, HRA
\citep{yuan2024bridging}, DoRA \citep{liu2024dora}, VeRA \citep{kopiczko2023vera}, KronA \citep{edalati2022krona},
tensor-train and Tucker adapters (\bgref{B.9}), and the linear methods LoRA-FA \citep{zhang2023lora}, LoRA-XS
\citep{baazy2024lora}, MiSS \citep{kang2024miss}, FourierFT \citep{gao2024parameter}, C3A \citep{chen2024parameter} and
masks.

\begin{proposition}{II.12}{gauge census: rank is a cut, dimension is a volume}
\begin{enumerate}[label=(\alph*)]
  \item $d(M)\le\lvert M\rvert-\dim(\text{generic gauge orbit})$, with equality when the gauge acts locally
    transitively on generic fibres.
  \item The values in Table~\ref{tab:census} hold for one $m\times n$ box, with $r\le\min(m,n)$ throughout except for
    VeRA (whose published RoBERTa-base setting has $r=1024>768$). In the LoHa row,
    \begin{equation*}
      F=r_1(m+n-r_1)+r_2(m+n-r_2)-(m+n-1)
    \end{equation*}
    is a proved upper bound for $d$. The gauge column lists the exhibited Lie group. It acts locally transitively on
    generic fibres away from the $mn$ cap, but not where the cap binds; for LoHa at $(m,n,r_1,r_2)=(4,4,2,2)$ the orbit
    has dimension $15$ and the fibre $16$. BitFit does not touch the $m\times n$ box; it is linear in the biases. Hybrid
    GraLoRA (\texttt{hybrid\_r}${}>0$) is not covered by the table. FourierFT is injective only when its sampled
    frequencies contain no conjugate pair $(u,v),(-u,-v)$; otherwise $d=\lvert M\rvert-\#\{\text{pairs}\}$. At LLM sizes
    this is close to $\lvert M\rvert$, but at $768\times768$ with $1000$ frequencies a pair occurs in about half of the
    draws (HF's default seed $777$ gives exactly two pairs there).
  \item \emph{Equal-budget trades.} Assume $2r\le\min(m,n)$, $r^2\le\min(m,n)$ and $2r^2<m+n-1$. LoHa$_{r,r}$ and
    LoRA$_{2r}$ both cost $2r(m+n)$, but LoHa reaches at least $(m+n-1)-2r^2$ \emph{fewer} dimensions (exactly that
    many when $F$ is attained [Num]). For $r\ge3$ it gains rank up to $r^2>2r$ in exchange. For $r\le2$, $r^2\le2r$, so
    $\Im\mathrm{LoHa}_{r,r}\subsetneq\Im\mathrm{LoRA}_{2r}$ and LoHa is strictly dominated. When $2r^2\ge m+n-1$ (still with $2r\le\min(m,n)$) the
    bound no longer separates them, and wherever $F$ is attained LoHa$_{r,r}$ matches or exceeds LoRA$_{2r}$ in
    dimension as well as in rank, for instance at $(m,n,r)=(9,9,3)$, at $r=64$ on $4096\times4096$, and at
    LyCORIS-style $r=32$ on $320$- to $640$-wide layers. GraLoRA$_k$ costs the same as LoRA$_r$, reaches the same
    number of dimensions, and has maximum rank $\min(kr,m,n)$, but caps every block at rank $r/k$. A generic rank-$r$
    update has rank $r$ on every block, so for $k\ge2$ and $kr\le\min(m,n)$ the images of LoRA$_r$ and GraLoRA$_k$ are
    incomparable.
\end{enumerate}
\end{proposition}

\begin{table}[t]
  \centering
  \footnotesize
  \caption{The gauge census of \thmref{Proposition}{II.12}(b) for one $m\times n$ box: budget, exhibited Lie gauge with
    the dimension of its generic orbit, expressive dimension and maximum rank. [Num] marks values checked as float64
    Jacobian ranks (Appendix~\ref{app:repro}); $F$ is defined in \thmref{Proposition}{II.12}(b).}
  \label{tab:census}
  \renewcommand{\arraystretch}{1.15}
  \begin{tabularx}{\linewidth}{@{}>{\raggedright\arraybackslash}p{3.0cm}>{\raggedright\arraybackslash}p{1.6cm}>{\raggedright\arraybackslash}X>{\raggedright\arraybackslash}p{3.2cm}>{\raggedright\arraybackslash}p{2.1cm}@{}}
    \toprule
    \headrow \textbf{method} & \textbf{budget $\lvert M\rvert$} & \textbf{exhibited Lie gauge (orbit dim)} &
      \textbf{$d(M)$} & \textbf{max rank}\\
    \midrule
    LoRA$_r$ & $r(m+n)$ & $\GL_r$ ($r^2$) & $r(m+n-r)$ & $r$\\
    GraLoRA ($k\times k$ blocks, rank $r/k$ each; HF default \texttt{hybrid\_r=0}) & $r(m+n)$ & $\prod\GL_{r/k}$
      ($r^2$) & $r(m+n-r)$ & $\min(kr,m,n)$\\
    LoHa$_{r_1,r_2}$ & $(r_1+r_2)(m+n)$ & $\GL_{r_1}\times\GL_{r_2}\times{}$diagonal torus ($r_1^2+r_2^2+m+n-1$) &
      $\min(mn,F)$ [Num]; $F$ is a proved upper bound & $\min(r_1r_2,m,n)$\\
    HRA$_r$ ($\rk W_0=n$, \texttt{apply\_GS=False}; even $r$ is needed only for HF's identity init) & $rn$ & non-linear;
      fibre dimension $r(r+1)/2$; exhibited Lie gauge $(\R^\times)^r$ & $rn-r(r+1)/2$ (for $r\in\{n-1,n\}$ this is
      $\dim\SO(n)$ [Num]) & $r$\\
    DoRA$_r$ & $r(m+n)+m$ & $\GL_r$ for generic $W_0$ with $(m-r)(n-r)\ge m$; larger otherwise (dimension $6>r^2$ at
      $(m,n,r)=(3,3,2)$, and $r^2+r$ for rank-one $W_0$) & $\min\bigl(mn,\,r(m+n-r)+m\bigr)$ for generic $W_0$
      (\thmref{Proposition}{II.8}) & up to full\\
    VeRA & $m+r$ & $\R^\times$ & $m+r-1$ [Num] & $\min(r,m,n)$\\
    KronA $A\otimes B$ & $m_1n_1+m_2n_2$ & $\R^\times$ & $m_1n_1+m_2n_2-1$ & $\min(m_1,n_1)\cdot{}$\newline$\min(m_2,n_2)$\\
    TT / Tucker (admissible, minimal ranks) & $\sum$ cores & $\prod_e\GL_{r_e}$ & $\lvert M\rvert-\sum_er_e^2$ & up to
      full\\
    linear methods (LoRA-FA, LoRA-XS, MiSS in every HF mode, FourierFT, C3A, masks) & $\lvert M\rvert$ & translations
      by the kernel & rank of the linear map; $=\lvert M\rvert$ when injective & varies\\
    \bottomrule
  \end{tabularx}
\end{table}

\paragraph{Numerical check}
LoHa dimensions $38$, $59$, $93$, $101$ at $(m,n,r)=(8,7,2)$, $(12,10,2)$, $(12,10,3)$, $(20,16,2)$, and $16=mn$ at
$(4,4,2)$, where the formula gives $17$ and $d(\mathrm{LoRA}_4)=16$ too. LoHa$_{3,3}$ against LoRA$_6$: $73$ against
$72$ at $(9,9,3)$. GraLoRA dimensions $48$, $64$, $108$ at $(m,n,r,k)=(8,8,4,2)$, $(12,8,4,2)$, $(12,12,6,3)$; GraLoRA
generic rank $8=kr$ at $(16,16,4,2)$ and $12=\min(m,n)$ at $(12,12,6,3)$ (Appendix~\ref{app:repro}).

\paragraph{Credit}
LoHa is the low-rank Hadamard product of FedPara \citep{hyeonwoo2021fedpara}; GraLoRA is due to \citet{jung2025gralora}.
The Tucker gauge and dimension are due to \citet{xkoch2010dynamical} and \citet{xuschmajew2013geometry}, and the TT case
to \citet{xholtz2012manifolds}, building on \citet{oseledets2011tensor}. The intrinsic-dimension experiments of
\citet{li2018measuring} and \citet{aghajanyan2020intrinsic} measure how many random directions a task needs; $d(M)$
counts the directions a method's image has, a complementary notion.

LoHa$_{r,r}$ maps into LoRA$_{r^2}$ through the face-splitting product (\bgref{B.13}), one of the arrow rules of
Exposé~\ref{sec:atlas}; for $r\le2$ this places LoHa$_{r,r}$ inside LoRA$_{2r}$.

\begin{explains}
Rank is a \emph{cut} invariant and dimension a \emph{volume} invariant, and the literature conflates them. When $r\ge3$
and $2r^2$ is small next to $m+n$, as on LLM projections at the usual ranks, LoHa reaches high rank on a thin, curved
subvariety, because its Hadamard product spends a whole torus on gauge. When $2r^2\ge m+n-1$, as in LyCORIS's usual
Stable Diffusion settings, it matches or exceeds LoRA$_{2r}$ in dimension too. GraLoRA keeps LoRA's dimension and
multiplies the maximum rank by $k$, but caps every block at rank $r/k$. It moves the cut, and the block caps pay for the
higher maximum rank.
\end{explains}

\subsubsection{Transport of structure}\label{sec:cut-transport}

An invertible linear change of frame fixing $\theta_0$ carries dimension, gauge, first-order image and degree along,
since they are defined from $\rho$ and its differential (\bgref{A.17}). It carries rank along only when it respects rows
and columns, which the Kronecker rearrangement, Fourier and wavelet frames and the Hadamard rescaling by $W_0$ do not.

\begin{proposition}{II.13}{transport of structure}
For $T\in\GL(\Theta)$ put $\tau_T(\theta)=\theta_0+T(\theta-\theta_0)$ and $T_*M:=(Q,q_0,\tau_T\circ\rho)$.
\begin{itemize}
  \item $T_*$ is a strict automorphism of $\Meth(\Theta,\theta_0)$, with inverse $(T^{-1})_*$. It maps $S_1$ to $T\,S_1$
    and preserves $d$, $g$, the gauge group, the first-order deficit, polynomial degree, the regular or apex type and
    the germ type of the image. It does not preserve rank, and, for the family extension below, it does not preserve the
    covariance group.
  \item $K^{T_*M}=T\,K^M\,T\T$ for every $T$. For any update rule that queries the loss only through
    $q\mapsto L(\rho(q))$, with its own state on $Q$, $T_*M$ trained on $L$ and $M$ trained on $L\circ\tau_T$ have
    identical $q$-trajectories, so their $\theta$-trajectories are conjugate by $\tau_T$. Rules that read
    $\nabla_\theta L$ and solve a least-squares problem on $\Theta$ (LoRA-Pro), and $\theta$-level weight decay or
    clipping, are covered only for orthogonal $T$. Orthogonality of $T$ is exactly the condition for $\tau_T$ to be a
    Frobenius isometry; for rules of the first kind it is needed only to keep the Frobenius spectrum of $K$, or an
    isometry-invariant loss class such as $\lVert\theta-\theta^\star\rVert^2$, unchanged.
  \item Coordinatewise optimizers (SGD, Adam) are also equivariant under coordinate permutations of $Q$. The Kronecker
    identification below needs this, because it composes $\mathcal R^{-1}_*$ with the vec reshape of $Q$.
  \item \emph{Families and rebasing.} For $T$ independent of the base point, define the family extension
    $(T_*M)(\theta):=(Q,q_0,\theta+T(\rho_\theta-\theta))$. For an additive family with fixed shape $C$
    (\thmref{Definition}{I.2}), $T_*M$ is additive with shape $TC$, so transport commutes with rebasing:
    $R_K(T_*M)=\tau_T(R_K(M))$ (\thmref{Theorem}{V.3}(a)). For action-type families this can fail even for $T=2I$,
    because the transported family acts by $I+T(g-I)$. For OFT on $\Theta=\R^{1\times2}$, two cycles of
    $\theta\mapsto\theta(2R-I)$ with $R=\operatorname{rot}(\pi)$ reach $9\theta_0$, outside the transported circle.
    Transports that depend on $\theta_0$ may fail (HiRA below) or may not (LoRA transported by the SVD frame of
    $\theta_0$, since $\Mr$ is $\Ort(m)\times\Ort(n)$-invariant). The covariance group (\thmref{Definition}{II.14}) is a
    property of a family and is compared through this extension.
\end{itemize}
Instances.
\begin{itemize}
  \item \emph{Kronecker family.} An isomorphism of slices over $\Theta'=\R^{m_1n_1\times m_2n_2}$ and over $\Theta$:
    $\sum_{i\le k}C_i\otimes D_i=\mathcal R^{-1}_*\mathrm{LoRA}_k$, with $Q$ reshaped by vec, dynamics included for
    coordinatewise optimizers. KronA with unconstrained factors is $\mathcal R^{-1}_*\mathrm{LoRA}_1$. LoKr is
    $\mathcal R^{-1}_*$ of $\mathrm{LoRA}_1$ with the second factor constrained to $\vect\MM_{\le r'}$ when
    $r'<\max(m_2,n_2)/2$ (HF's switch); otherwise HF's second factor is full and LoKr is KronA. LoKr's
    \texttt{decompose\_both} option also constrains the first factor.
  \item \emph{Linear methods as coordinate subspaces in transported frames.}
  \begin{itemize}
    \item FourierFT: sparse in the real cosine half of the real Fourier frame. With scaling $c$ (default $150$), HF's
      update $\Delta W=c\operatorname{Re}(\operatorname{ifft2}(S))$ is the transport of a mask, with $g=0$, only when the sampled index set has no conjugate pair $(u,v),(-u,-v)$; otherwise $g$ is the
      number of such pairs. Its image lies in the point-symmetric subspace $\Delta W_{j,l}=\Delta W_{-j,-l}$, and the
      frame is orthogonal only up to scale (self-conjugate frequencies have twice the squared norm).
    \item WaveFT: sparse in a wavelet frame, when HF's pad--IDWT--crop map is invertible (plausible for Haar on even
      shapes; unverified for other wavelet families).
    \item C3A: block-circulant. Over $\R$ the frame is given by the powers $P^k$ of the cyclic shift (Gram matrix
      $bI$). Over $\mathbb C$, $\operatorname{circ}(w)=F^{-1}\diag(Fw)F$ for first column $w$, and the real form has
      $2\times2$ rotation--scaling blocks. HF adds a factor $1/\texttt{in\_features}$.
    \item LoRA-XS, SVFT, TinyLoRA: masks or random subspaces in the SVD frame of $W_0$, a $\theta_0$-dependent
      transport.
    \item SHiRA, FISH, Trainable Tokens: masks in the standard frame.
    \item MiSS/Bone in its default (balance) mode: $\Delta W=D(\mathbf 1_g\T\otimes I_r)$, which is LoRA-FA with a
      frozen \emph{summing} projection. Its BAT mode, $\Delta W=\operatorname{blocks}(W_0)\,M+M$ blockwise, is linear in
      the trained block $M$ but depends on $\theta_0$, so it is not additive.
    \item These frame readings of FourierFT and C3A are for \texttt{init\_weights=True}. HF's defaults start off
      $\theta_0$: FourierFT draws its spectrum from $\mathcal N(0,1)$ with scaling $150$ (at $768^2$ with $1000$
      frequencies, $\lVert\Delta W_0\rVert_F\approx4.3$, about $14\%$ of a std-$0.04$ $W_0$), and C3A is
      Xavier-uniform. These are objects with a pointing defect.
  \end{itemize}
  \item \emph{HiRA} $=(D_{W_0})_*\mathrm{LoRA}$ inside the slice at $\theta_0$, when $W_0$ has no zero entries, with
    $D_{W_0}:X\mapsto W_0\odot X$. Because $D_{W_0}$ depends on $\theta_0$, re-based HiRA is not transported ReLoRA. Two
    cycles reach $W_0\odot(J+Z_1)\odot(J+Z_2)$, whose relative update $Z_1+Z_2+Z_1\odot Z_2$ has rank up to
    $\min(2r+r^2,m,n)$, attained, where transported ReLoRA stays at rank $\le2r$.
\end{itemize}
\end{proposition}

\paragraph{Numerical check}
$\lVert K^{T_*M}-TK^MT\T\rVert=2\times10^{-14}$ for a Gaussian $T$. Adam on $T_*M$ with $L$ and on $M$ with $L\circ\tau_T$
agree to $8\times10^{-11}$. Two HiRA cycles give relative-update rank $3$, $8$, $15$ at $r=1,2,3$ ($m=n=30$)
(Appendix~\ref{app:repro}).

\paragraph{Credit}
The transported methods are FourierFT \citep{gao2024parameter}, WaveFT \citep{bilican2025exploring}, C3A
\citep{chen2024parameter}, LoRA-XS \citep{baazy2024lora}, SVFT \citep{lingam2024svft}, TinyLoRA
\citep{morris2026learning}, SHiRA \citep{bhardwaj2024sparse}, FISH \citep{sung2021training}, Trainable Tokens
\citep{team2025trainable}, MiSS \citep{kang2024miss} and HiRA \citep{huang2025hira}; LoRA-Pro is due to
\citet{wang2024lorac}. \citet{si2024see} organise PEFT around the SVD frame of $W_0$, a related unification.

\begin{explains}
``High rank'' claims (MoRA, HiRA, LoKr, C3A, SHiRA, WaveFT) are not $\GL(\Theta)$-invariant, although rank is
$\GL_m\times\GL_n$-invariant. The transports in play ($\mathcal R$, the real Fourier (cosine) frame, $D_{W_0}$) do not
preserve the network function, and \thmref{Theorem}{II.11} shows that KronA and $\mathrm{LoRA}_r$ have incomparable
images when $r<\min(m_1,n_1)\min(m_2,n_2)$ and $\kappa_1>1$. Transport preserves every isomorphism invariant of the
method ($d$, $g$, the gauge group, $\dim S_1$, the deficit, the regular or apex type, the germ type and the degree) and
conjugates its dynamics. Relative to the network's $\GL_m\times\GL_n$ structure it changes, in particular, rank,
covariance, $S_1$ as a subspace, and the image itself. In Figure~\ref{fig:rankvol} HiRA therefore sits at LoRA's height
and at full rank (\thmref{Proposition}{II.9}(a)).
\end{explains}

\paragraph{Prediction}
At matched learning-rate sweeps, the performance gap between a transported method and its source (Kronecker sums against
LoRA; Fourier or wavelet methods against coordinate masks) is explained by how well each image aligns with the task's
update, measured for instance by projecting a full fine-tuning $\Delta W$ onto each image, more than by optimizer effects.

\subsubsection{The plug, and a reporting rule}\label{sec:cut-plug}

Cut $W_0$ out of a linear layer and a hole $V\to U$ remains. A weight-space method fills it with a small network of
frozen and trainable boxes, a \emph{plug} (Figure~\ref{fig:loraplug}). Every invariant above is read off the plug: the
budget (\thmref{Definition}{I.2}), the min-cut, the Jacobian rank (\thmref{Definition}{I.5}), the gauges on its wires,
and the merge type, set by whether an activation sits on the path (\thmref{Theorem}{VI.3}). The plug is a weight-space
cousin of the unified view of \citet{he2021unified}, which puts adapters, prefix tuning and LoRA in one design space.

Table~\ref{tab:onebudget} spends one budget three ways on one $4096\times4096$ box (\thmref{Proposition}{II.12}). The
methods differ by at least $8{,}063$ dimensions and by a factor of four in rank. The LoHa row uses the upper bound $F$,
whose equality is checked numerically at small sizes and open in general.

\begin{table}[t]
  \centering
  \footnotesize
  \caption{One $4096\times4096$ box and one budget, $\lvert M\rvert=131{,}072$, spent three ways
    (\thmref{Proposition}{II.12}).}
  \label{tab:onebudget}
  \renewcommand{\arraystretch}{1.12}
  \begin{tabularx}{\linewidth}{@{}lrrrX@{}}
    \toprule
    \headrow \textbf{method} & \textbf{$d(M)$} & \textbf{fibre $\lvert M\rvert-d(M)$} & \textbf{max rank} &
      \textbf{block structure}\\
    \midrule
    LoRA$_{16}$ & $130{,}816$ & $256$ & $16$ & none\\
    GraLoRA, $k=2$ & $130{,}816$ & $256$ & $32$ & each of $4$ blocks has rank $\le8$\\
    LoHa$_{8,8}$ & $\le122{,}753$ & $\ge8{,}319$ & $64$ & none\\
    \bottomrule
  \end{tabularx}
\end{table}

\paragraph{Prediction}
Benchmarks should report $d(M)$ beside $\lvert M\rvert$, with the maximum rank and the cut it refers to. At budgets of a
few thousand parameters, rankings at equal \emph{effective} dimension can change.

A small case shows every readout at once. On an $8\times8$ box, LoHa$_{2,2}$ has $64$ parameters, rank at most $4$,
Jacobian rank $41$ and a gauge orbit of dimension $23=4+4+15$ that fills the fibre. LoRA$_4$ costs the same and reaches
$7$ more dimensions, with LoHa's image strictly inside it (\thmref{Proposition}{II.12}(c), since $2r^2=8<15=m+n-1$). The
plug workbench of the companion site recomputes these invariants in float64 for fifteen presets and for plugs the reader
edits: the budget, the degree, the min-cut bound by brute force over vertex bipartitions beside the rank attained, $d(M)$
as a Jacobian rank, the gauge orbit with each generator checked to lie in $\ker d\rho$, and the merge verdict.

\subsubsection{Covariance and descent}\label{sec:cut-covariance}

A method is usually one recipe applied to every pretrained weight. Rename a layer's input basis by $h$ and its output
basis by $g$, so that $W_0$ becomes $gW_0h^{-1}$; the method is covariant under $(g,h)$ if its image moves the same way.

\begin{definition}{II.14}{covariance group}
For a method defined for every $W_0$ (a natural family), the \textbf{covariance group}
$\mathcal G_M\subseteq\GL_m\times\GL_n$ is the largest subgroup with $\Im M(gW_0h^{-1})=g\,\Im M(W_0)\,h^{-1}$. When
the method depends on random frames or on data, the identity is required in distribution, with the data transformed.
$\mathcal G_M$ is the \emph{geometry the method lives in}, in Klein's sense.
\end{definition}

Klein defined a geometry by its group of motions and studied what the group leaves unchanged
\citep{xklein1872vergleichende} (\bgref{D.17}). For group-type methods this reading is a theorem (\S\ref{sec:erlangen}).
For cones such as LoRA's image it is a naming device, and the content is the shape of the image and its covariance group.

LoRA's image $W_0+\Mr$ is covariant under all of $\GL_m\times\GL_n$, since rank ignores bases, while a coordinate mask
has a much smaller group. This matters because networks have symmetries of their own, such as permutations of hidden
neurons, rotations of the residual stream and the query--key gauge of attention, which change the weights and keep the
function.

\begin{observation}{II.15}{descent through function-preserving reparametrisations}
Let $M=\prod_bM_b$ be a box-wise method: each $M_b$ is a member of a natural family with covariance group
$\mathcal G_{M_b}$, and no trainable parameters or random frames are shared across boxes. Let $G_{\mathrm{arch}}$ act on
$\Theta=\prod_b\Theta_b\times\Theta_{\mathrm{frozen}}$, preserving this split and acting on each adapted box by
$W_b\mapsto g_bW_bh_b^{-1}$ with $(g_b,h_b)\in\mathcal G_{M_b}$. Assume $f(g\theta,-)=f(\theta,-)$ on a
$G_{\mathrm{arch}}$-stable set containing $\Im M(\theta_0)$. Then $\Im M(g\theta_0)=g\cdot\Im M(\theta_0)$, and the set
of \emph{functions} reachable by $M$ from $g\theta_0$ equals the set reachable from $\theta_0$. For methods with random
frames, the equality holds in distribution, provided the frames are drawn independently per box. Boxes that share a
frame need a pathwise condition: a law-preserving map $T_{(g,h)}$ on frame space with
$\Im M_b(g_bWh_b^{-1};TF)=g_b\,\Im M_b(W;F)\,h_b^{-1}$ for all $W$ and $F$, with the same $T$ on every box that shares
$F$ (marginal covariance in law, as in \thmref{Definition}{II.14}, is not enough). For data-dependent methods, require
covariance pathwise in the data, and that the box-$b$ data at $g\theta_0$ is the \thmref{Definition}{II.14} transform
under $(g_b,h_b)$ of the data at $\theta_0$. This holds for intertwiner gauges such as hidden-unit permutations,
query--key and value--output head gauges and residual rotations. The statement concerns reachable sets only; training
trajectories and outcomes can still differ, for instance under Adam or non-isotropic initialisations.

Examples of $G_{\mathrm{arch}}$:
\begin{itemize}
  \item hidden-neuron permutations;
  \item the query--key gauge $(W_Q,W_K)\mapsto(gW_Q,g^{-\top}W_K)$ per head. With no positional encoding or an absolute
    one, $g\in\GL_{d_k}$. Under RoPE, $g$ must commute with every RoPE rotation, which leaves the per-frequency $2\times2$
    rotation--scalings $(\mathbb C^\times)^{d_k/2}$, shared across a GQA group. Under partial RoPE on $r$ of the $d_k$
    dimensions the group is $(\mathbb C^\times)^{r/2}\times\GL_{d_k-r}$. Under QK-norm, $g$ must in addition be
    orthogonal when the norm's $\varepsilon>0$ (conformal, $g\in\CO(d_k)$, when $\varepsilon=0$) and satisfy
    $g\T\Gamma_qR_t\Gamma_kg=\Gamma_qR_t\Gamma_k$ for every RoPE rotation $R_t$, where $\Gamma_q,\Gamma_k$ are the gains.
    OLMo~2 normalises $q$ and $k$ across all heads jointly;
  \item a single global orthogonal rotation $Q$ of the residual stream of a pre-norm transformer whose norm gains can all
    be fused into the linear layers they feed, after LayerNorm is converted to RMSNorm (SliceGPT's computational
    invariance; QuaRot's offline rotation). SliceGPT's per-block rotations $Q_\ell$ need extra residual matrices and
    change the architecture. Sandwich and post-sublayer norms (Gemma~2 and~3, OLMo~2) are excluded, because their gains
    cannot be fused and $\diag(\gamma)$ does not commute with $Q$. The comparison is between the gain-fused model and the
    fused-and-rotated model: gain fusion is a diagonal input rescaling outside $G_{\mathrm{arch}}$ that already changes
    PiSSA's split and block OFT's image (\thmref{Proposition}{II.16}), and converting LayerNorm to RMSNorm changes LoRA's
    reachable set on the residual writers.
\end{itemize}
\end{observation}

\paragraph{Numerical check}
A tied rank-one update shared by two boxes violates the conclusion (relative distance up to $0.30$), while independent
per-box LoRA satisfies it. A Haar frame shared by two boxes, each covariant in law, gives a joint statistic of $0.44$ from
$\theta_0$ against $0.81$ from $g\theta_0$. Under RoPE a generic $g\in\GL_8$ changes attention scores by up to $313$, a
RoPE-commuting $g$ by $5\times10^{-14}$ (Appendix~\ref{app:repro}).

\paragraph{Credit}
RoPE is due to \citet{xsu2024roformer}. Residual-stream rotations are the device of SliceGPT \citep{xashkboos2024slicegpt}
and of QuaRot \citep{xashkboos2024quarot}. The architectures with sandwich or post-sublayer norms are Gemma~2
\citep{xgemma2024gemma2}, Gemma~3 \citep{xgemma2025gemma3} and OLMo~2 \citep{xolmo2024furious}.

\begin{explains}
Descent is decided by whether each induced pair $(g_b,h_b)$ lies in $\mathcal G_{M_b}$. For one-sided methods
((IA)$^3$, DoRA, OFT, BOFT) this is a question of side. For coordinate-structured methods (LoHa with $r\ge2$, HiRA, masks,
FourierFT, Kronecker methods) it is a question of group. They descend through hidden-unit monomial gauges, since
$P(X\odot Y)Q=(PXQ)\odot(PYQ)$ for permutation matrices $P,Q$ and a diagonal scaling can be absorbed into one factor, and
they do not descend through rotations. A global residual rotation acts on the input side of $q$, $k$, $v$, gate and up,
and on the output side of $o$ and down.
\end{explains}

Table~\ref{tab:descent} sorts common methods by this criterion. Failing a sufficient condition does not prove that the
reachable function sets differ.

\begin{table}[t]
  \centering
  \footnotesize
  \caption{Methods against the sufficient condition of \thmref{Observation}{II.15}. For the residual rotation the
    comparison is between the gain-fused model and the fused-and-rotated model.}
  \label{tab:descent}
  \renewcommand{\arraystretch}{1.15}
  \begin{tabularx}{\linewidth}{@{}>{\raggedright\arraybackslash}p{2.6cm}>{\raggedright\arraybackslash}X>{\raggedright\arraybackslash}X@{}}
    \toprule
    \headrow \textbf{gauge} & \textbf{satisfy the hypothesis (reachable functions unchanged)} & \textbf{fail the
      sufficient condition}\\
    \midrule
    single global residual rotation (SliceGPT style) & LoRA, full FT, BitFit, PiSSA; (IA)$^3$ with HF defaults, which
      scales the output side of $k$ (or $q$, for Qwen, Gemma and Phi) and $v$ and the input side of down, while the
      rotation acts on the other side, where (IA)$^3$ is $\GL$-covariant; DoRA on $q$, $k$, $v$, gate and up; OFT on $o$
      and down (no scale); LoHa$_1$, whose image is $W_0+\MM_{\le1}$ & block OFT and BOFT on residual-input projections
      (the block normaliser contains no Hadamard or PCA rotation); BOFT on $o$ and down (its output scale
      \texttt{boft\_s}); DoRA on $o$ and down; LoHa with $r\ge2$; HiRA; coordinate masks; FourierFT\\
    query--key and value--output head gauges & LoRA & (IA)$^3$ on $q$ or $k$ and on $v$; OFT on $o$\\
    \bottomrule
  \end{tabularx}
\end{table}

\begin{conjecture}{A}{reachable sets under a residual rotation}
Against a gain-fused but unrotated control, fine-tuning the fused-and-rotated, function-identical model measurably
changes the reachable function sets, and the results, of block OFT/BOFT on residual-input projections, LoHa with
$r\ge2$, HiRA and coordinate masks, because LLM residual streams have a privileged basis.
\end{conjecture}

QuaRot's per-head Hadamard on $W_v$ and $W_{\mathrm{out}}$ is a value--output gauge in $G_{\mathrm{arch}}$. Its online
Hadamards (across heads, before down\_proj, on keys) preserve the realised function but insert new operations, so they
change the architecture map and lie outside $G_{\mathrm{arch}}$. The optimizer adds a further layer. Adam is only
$B_r$-equivariant (\thmref{Theorem}{III.12}) and HF's Kaiming-uniform $A$ is not rotation-invariant in law, so LoRA
\emph{trained with Adam} need not descend even though its image does.

\subsubsection{Covariant data-aware splits}\label{sec:cut-splits}

SVD-based initialisations raise the same question. PiSSA, MiLoRA and CorDA split $W_0$ into a trainable part and a
frozen residual by a truncated SVD of $W_0$ or, for CorDA, of $W_0$ weighted by the input second moment. Truncated SVD
commutes only with rotations and uniform scalings, so these splits depend on the input coordinates, and whitening
removes that dependence.

\begin{proposition}{II.16}{covariant data-aware splits}
Let $C=\mathbb E[xx\T]\succ0$ be the input second moment, so that under $x\mapsto hx$ we have $W\mapsto Wh^{-1}$ and
$C\mapsto hCh\T$. Let $LL\T=C$ and $p:=\min(m,n)$. Write $[M]_r$ for the rank-$r$ truncated SVD,
$\langle M\rangle_r:=M-[M]_{p-r}$ for the bottom $r$ singular components, and $\CO(n)=\R_{>0}\cdot\Ort(n)$. Assume
$1\le r<p$ and the spectral gap each split needs: $\sigma_r>\sigma_{r+1}$ of $W_0L$ in (a); in (b),
$\sigma_r>\sigma_{r+1}$ of $W_0C$ or $W_0$ for top-$r$ splits, and $\sigma_{p-r}>\sigma_{p-r+1}$ for bottom-$r$ splits.
Covariance groups are taken over the generic set where these gaps hold.

(a) The whitened split
\begin{equation*}
  X^*=\arg\min_{\rk X\le r}\lVert(W_0-X)L\rVert_F=[W_0L]_rL^{-1}=\Pi_rW_0,
\end{equation*}
where $\Pi_r$ is the top-$r$ eigenprojector of the output second moment $W_0CW_0\T$, is $\GL_n$-covariant on the input
side and $\CO(m)$-covariant on the output side; its joint group is $\CO(m)\times\GL_n$. This is the classical
reduced-rank-regression (SVD-LLM) solution.

(b) The formula-level CorDA splits, IPM ($X=[W_0C]_rC^{-1}$, the minimiser of $\lVert(W_0-X)C\rVert_F$) and KPM (the
bottom-$r$ components of $W_0C$ mapped back by $C^{-1}$, $X=\langle W_0C\rangle_rC^{-1}$), with or without CorDA's
trace-relative damping $C+\kappa(\tr C/n)I$, and the PiSSA and MiLoRA splits ($[W_0]_r$ and the bottom-$r$ part
$\langle W_0\rangle_r$), have input-side covariance group exactly $\CO(n)$; jointly, $\CO(m)\times\CO(n)$.

(c) \emph{Scope.} An absolute damping $C+\lambda I$ reduces the input-side group of (a) and of (b) to $\Ort(n)$. The
official and HF CorDA code divide each calibration sample by the absolute value of its largest entry before
accumulating $C$, which keeps only positive scalings and coordinate permutations, so (b) describes the CorDA formulas and
does not cover shipped CorDA under rotations.
\end{proposition}

\paragraph{Numerical check}
Under $h=3Q$ with $Q$ orthogonal, CorDA (IPM and bottom-$r$ KPM), PiSSA and bottom-$r$ MiLoRA move covariantly to
$4\times10^{-15}$. Under a generic $h$ only the whitened split does ($2\times10^{-15}$, against $0.63$ and $1.1$). At
$r=n$ PiSSA is $\GL_n$-covariant. An absolute damping $0.1$ breaks the whitened split under a generic $h$ ($0.14$), and
under $h=3Q$ (about $2\times10^{-2}$), while trace-relative damping keeps it ($3\times10^{-15}$)
(Appendix~\ref{app:repro}).

\paragraph{Credit}
The splits are CorDA \citep{yang2024corda}, PiSSA \citep{meng2024pissa} and MiLoRA \citep{wang2024milora}. The whitened
truncation is the one used by SVD-LLM \citep{xwang2025svdllm} for compression.

\begin{explains}
A design rule from the Erlangen programme. A sufficient condition for a data-aware Type~II split to be independent of how
a layer's input coordinates are chosen is that it be a left multiple of $W_0$ by a matrix-valued function of the output
second moment, $X=\Phi(W_0CW_0\T)\,W_0$. The whitened (SVD-LLM-style) split and the eigenprojector split of Astra
\citep{liu2026astra} are instances; Astra's paper form, a tail eigenprojector of the output covariance, has the same form
with $C$ replaced by the centred input covariance, which transforms in the same way. Depending on the data only
through $W_0CW_0\T$ is not enough, since PiSSA satisfies that vacuously and is not $\GL_n$-covariant. Non-conformal
reparametrisations, such as the diagonal rescalings from norm-gain folding or SmoothQuant \citep{xxiao2023smoothquant},
change the CorDA and PiSSA splits.
\end{explains}

Three caveats apply. An absolute damping $C+\lambda I$ reduces the covariance to $\Ort(n)$, while CorDA's trace-relative
damping keeps $\CO(n)$. The covariance is one-sided. No rank-$r$ split that depends only on $(W_0,C)$ can be
$\GL_m\times\GL_n$-covariant, since the stabiliser of a normal form contains an orthogonal group that fixes only
full-rank or zero splits, and whitening with the output second moment itself is degenerate (every singular value of
$(W_0CW_0\T)^{-1/2}W_0L$ is $1$). Descent through hidden-layer $\GL$ gauges would need an output-side statistic
independent of $(W_0,C)$, such as a backpropagated-gradient second moment. Training dynamics (SGD, Adam) remain
non-covariant.

EVA \citep{paischer2024parameter} is not a split. It is Type~I ($B_0=0$, $W_0$ not split) and points
$\operatorname{row}A_0$ at the top-$r$ centred principal subspace of the inputs. At a fixed rank its image $W_0+\Mr$ is
$\GL_m\times\GL_n$-covariant under \thmref{Definition}{II.14}, and only its pointing $\operatorname{row}A_0$ is
$\CO(n)$-covariant. With HF's default \texttt{rho=2.0} the per-layer rank is redistributed by explained-variance ratios,
which are only $\CO(n)$-invariant, so the image becomes $\GL_m\times\CO(n)$-covariant.

\begin{conjecture}{C}{whitened splits}
Whitened CorDA or PiSSA is more robust to function-preserving input reparametrisations, and no worse otherwise.
\end{conjecture}

Rank is a cut, dimension a volume and covariance a group. All three can be computed before training. The next exposé
turns from what a method can reach to how it moves there.

\section{Fibres and metrics: dynamics}\label{sec:fibres}

Expressivity lives on the image. Training lives on the fibres, through a metric. The image of a method fixes what it can reach, while its fibres and the optimizer's metric fix how it gets there. So two objects of $\Meth$ can be isomorphic and still train differently, and methods that look different can train identically.

Training happens upstairs, in the parameter space $Q$, where each fibre of $\rho$ collects the parameters that give one weight. The optimizer's metric on $Q$, pushed down to weight space, is what the model feels. Every statement below names its optimizer.

The exposé has four movements. Training with rates $\Lambda$ moves the weights by $\dot\theta=-K\nabla_\theta L$, with the cometric $K=D\rho\,\Lambda\,D\rho^\top$, which depends on where the parameters sit in their fibre (\thmref{Definition}{III.1} to \thmref{Proposition}{III.4}). Gradient flow conserves LoRA's charge $\Phi=B^\top B/\eta_B-AA^\top/\eta_A$, which sorts initialisations (\thmref{Theorem}{III.5} to \thmref{Proposition}{III.8}); when the charge is isotropic, one scale $\sigma^*$ separates a LoRA-FA-like regime from balanced dynamics (\thmref{Theorem}{III.9}). A 2-torus of rescalings acts on LoRA's knobs $(\alpha,\eta_A,\eta_B,\sigma_A)$ without changing training. Under Adam with $\varepsilon=0$, no decay or clipping and one shared schedule, LoRA+ with ratio $\lambda$ is plain LoRA with $\alpha$ multiplied by $\lambda$ (\thmref{Theorem}{III.10}, \thmref{Corollary}{III.11}, after \citealp{schulman2025lora}). Finally, SGD respects the subgroup $\Ort(r)$ of the gauge and Adam only the signed permutations. So from a zero-$B$ start the image depends only on $\operatorname{row}A_0$, the SGD trajectory on $A_0^\top A_0$, and the Adam trajectory on $A_0$ up to signed row permutations (\thmref{Theorem}{III.12}, \thmref{Corollary}{III.13}).

\subsubsection{One experiment first}\label{sec:fibres-hook}

Figure~\ref{fig:orbit} shows the cleanest consequence. Train with Adam at $\varepsilon=0$ from $B_0=0$, with no weight decay or clipping, one schedule for both parameter groups, and the same data order and dropout masks. Then LoRA+ with ratio $\lambda$ ($\eta_B=\lambda\eta_A$) and plain LoRA with $\alpha$ multiplied by $\lambda$, at the same $\eta_A$ and $A_0$, produce the same weights at every step (\thmref{Theorem}{III.10}, \thmref{Corollary}{III.11}). This LoRA/Adam statement is due to \citet{schulman2025lora}, for the LoRA+ of \citet{hayou2024lora}; \thmref{Theorem}{III.10} says what is added here.

\subsubsection{The metric a method carries}\label{sec:fibres-cometric}

A learning rate is a metric in disguise. Gradient descent with rates $\Lambda$ moves $q$ by $-\Lambda\nabla_q$, and pushed down by $D\rho$ this is the weight gradient multiplied by a positive semidefinite operator.

\begin{definition}{III.1}{trained method; cometric; three equivalences}
A \emph{trained method} is a method together with optimizer data: a learning-rate operator $\Lambda\succ0$ (per-block rates), an initial scale, and an update rule. Gradient flow $\dot q=-\Lambda\nabla_q(L\circ\rho)$ moves the weights by
\[\dot\theta=-K_q\nabla_\theta L,\qquad K_q=D\rho_q\,\Lambda\,D\rho_q^\top.\]
We call $K_q$ the \textbf{induced cometric}. For a backward method with compression $c_t$ and anchor $a_t$ (\thmref{Definition}{IV.1}), $K_t=a_tc_t$.

Two methods are
\begin{itemize}
\item \emph{expressively equivalent} if their images agree;
\item \emph{first-order equivalent} if their $K_{q_0}$ agree;
\item \emph{dynamically equivalent} if they produce the same $\theta$-trajectory for every loss.
\end{itemize}
\end{definition}

Cometrics compose the way derivatives do.

\begin{observation}{III.2}{chain rule for cometrics}
Let $\sigma:Q'\to Q$ and $\rho:Q\to X$ be smooth, with $X=\Theta$ or a function or output space. Train $Q'$ by gradient flow with a constant rate operator $\Lambda'\succ0$, and put $K^\sigma_{q'}:=D\sigma_{q'}\Lambda'D\sigma_{q'}^\top$. Then
\[K^{\rho\sigma}_{q'}=D\rho_{\sigma(q')}\,K^\sigma_{q'}\,D\rho_{\sigma(q')}^\top,\]
and no metric on $Q$ is used. $K$ is a degenerate cometric (positive semidefinite, of rank $\le\dim Q'$); when $\sigma$ is not injective, $K^\sigma$ is a field along $\sigma$ and fails to be a cometric on $Q$.

For LoRA \citep{hu2021lora}, $\rho(B,A)=W_0+sBA$ with $s=\alpha/r$ ($\alpha/\sqrt r$ for rsLoRA, \citealp{kalajdzievski2023rank}), block rates $\Lambda=\diag(\eta_BI,\eta_AI)$ and no LoRA dropout, under gradient flow (equivalently, to first order in the step size for GD),
\[K_{(B,A)}(G)=s^2\big(\eta_B\,GA^\top A+\eta_A\,BB^\top G\big).\]
Under Adam there is no state-independent linear cometric; \thmref{Theorem}{III.10} and \thmref{Theorem}{III.12} treat Adam.
\end{observation}

\noindent\labelsf{Num}\enspace Finite-difference Jacobians agree with the formula to $6\times10^{-9}$ (Appendix~\ref{app:repro}).

\begin{explains}
For LoRA, the term $\eta_B\,GA^\top A$ lets the update reach all outputs while filtering its inputs through $\operatorname{row}A$, and $\eta_A\,BB^\top G$ does the reverse. At $B_0=0$ only the first term survives, so the first-step cometric $s^2\eta_B\,GA_0^\top A_0$ is fixed by $A_0$ (\thmref{Corollary}{III.13}).

\smallskip
Take $X$ to be the prefix space (Exposé~\ref{sec:merge}). HF \texttt{PrefixTuning} with \texttt{prefix\_projection=True} (the MLP $\sigma$ of \citealp{li2021prefix}) and with \texttt{prefix\_projection=False} (the P-tuning v2 setting of \citealp{liu2021p}, ignoring its classification head) have the same prefix image whenever \texttt{encoder\_hidden\_size}${}\ge{}$\texttt{num\_virtual\_tokens}${}-1$. Under gradient flow their cometrics on prefix space are $D\sigma\Lambda'D\sigma^\top$ and $\Lambda$, so the reparametrisation replaces the flat metric on prefix space by a point-dependent one. Prefix methods are extensions, so the target here is the prefix (or function) space rather than $\Theta$. This alone does not account for the instability that Li and Liang observed under AdamW.
\end{explains}

Backpropagation as a lens, the source of this chain rule, is due to \citet{fong2017backprop} and \citet{cruttwell2021categorical}.

\subsubsection{When a change of chart does not matter}\label{sec:fibres-naturality}

A gauge transformation changes the parameters and keeps the weights. It leaves training unchanged exactly when the cometrics agree, which for LoRA's gauge $\GL_r$ singles out $\Ort(r)$.

\begin{theorem}{III.3}{gradient descent is natural exactly where cometrics agree; for LoRA's linear gauge, exactly along $\Ort(r)$}
(a) Let $\varphi:Q\to Q'$ be a diffeomorphism with $\rho'\circ\varphi=\rho$, and let $\Lambda,\Lambda'$ be the rate operators on $Q,Q'$. Under gradient flow, the two methods induce the same $\theta$-velocity at $q$ and at $\varphi(q)$ for every loss iff $K^\rho_q=K^{\rho'}_{\varphi(q)}$, that is,
\[D\rho'_{\varphi(q)}\big(D\varphi_q\Lambda D\varphi_q^\top-\Lambda'\big)D\rho'^\top_{\varphi(q)}=0.\]
This holds whenever $D\varphi\Lambda D\varphi^\top=\Lambda'$, which is an isometry when $\Lambda=\Lambda'=I$. That condition is necessary when $D\rho'_{\varphi(q)}$ is injective, and fails to be necessary in general. Every linear gauge of a linear parametrisation is natural; for instance, for $\rho(x,y)=\theta_0+x+y$ the shear $(x,y)\mapsto(2x,y-x)$ preserves $\rho$ and $K$ without being orthogonal. So is every pointwise-orthogonal non-linear LoRA gauge $\varphi(q)=\varphi_{g(q)}(q)$ with $g(q)\in\Ort(r)$.

(b) For LoRA's linear gauge $\varphi_g(B,A)=(Bg^{-1},gA)$ with block-scalar rates, at points where $A$ has full row rank ($B$ arbitrary, including $B=0$) and $r<n$, the criterion holds iff $g\in\Ort(r)$. The same holds symmetrically where $B$ has full column rank and $r<m$.
\end{theorem}

\noindent\labelsf{Num}\enspace At a random point a non-orthogonal $g$ gives a cometric mismatch of about $18$, an orthogonal one about $10^{-15}$ (Appendix~\ref{app:repro}).

\begin{explains}
No learner with a fixed Euclidean metric respects all invertible reparametrisations. For LoRA's linear gauge, plain gradient descent defines a family of training dynamics indexed by the cosets $\Ort(r)\backslash\GL_r$, which $g\mapsto g^\top g$ identifies with $\Sym^+_r$. Under gradient descent, LoRA+'s rate ratio $\lambda$ corresponds to the scalar gauge $g=\lambda^{1/4}I$ together with an overall change of rate. rsLoRA rescales $s$, equivalently the learning rates and the initialisation; this is a move on the hyperparameter torus and involves no choice of gauge (\thmref{Theorem}{III.10}, \thmref{Corollary}{III.11}). Riemannian preconditioning and LoRA-Pro remove the dependence on the representative (\thmref{Proposition}{III.4}). Under AdamW, as used in practice, the relevant group is the group $B_r$ of signed permutations (\thmref{Theorem}{III.12}).
\end{explains}

The $\GL_r$ symmetry of $W=BA$ and the quotient geometry of fixed-rank matrices are classical \citep{mishra2012fixed}. \citet{putterman2024learning} build learners on LoRA weights that respect it.

Some published optimizers use instead a cometric that depends only on the weight, at least in an idealised form.

\begin{proposition}{III.4}{cometrics that descend to the rank-$r$ manifold; known in part}
Write $U=\operatorname{col}B$ and $V=\operatorname{row}A$. Statements (a), (b) and the first sentence of (d) are on the full-rank locus; (c) needs only $\rk A_0=r$, with $B$ arbitrary, including $B_0=0$. All statements concern gradient flow, or (S)GD to first order, on the stated gradients, undamped and without Adam. Cometrics include the learning rate.

(a) Scaled GD, $\delta B=-\eta\nabla_B(AA^\top)^{-1}$ and $\delta A=-\eta(B^\top B)^{-1}\nabla_A$, is $\GL_r$-equivariant with first-order cometric $K(G)=\eta s^2(P_UG+GP_V)$.

(b) LoRA-Pro's equivalent gradient $\tilde g=P_UG+(I-P_U)GP_V$ is the Frobenius-orthogonal projection $\Pi_{T_{W-W_0}\MM_r}(G)$. The flow of its SGD variant (LoRA-Pro Alg.~1) is the Riemannian gradient flow (\bgref{D.24}) on $W_0+\MM_r$.

(c) LoRA-FA's corrected gradient $s^{-2}\nabla_B(A_0A_0^\top)^{-1}$, fed to SGD with $A_0$ frozen, gives $K(G)=\eta\,GP_V$ with $V=\operatorname{row}A_0$, which depends on $A_0$ only through $V\in\Gr(r,n)$. Its original form, plain SGD on $B$, has $K(G)=s^2\eta_BGA_0^\top A_0$.

(d) The cometrics of (a) and (b) are functions of $W$ on the full-rank locus and define vector fields tangent to $W_0+\MM_r$. The field of (c) is $G\mapsto\eta GP_V$ with $V=\operatorname{row}A_0$. Its flow is the Frobenius gradient flow on the slice $W_0+\{XA_0\}=W_0+\R^{m\times n}P_V$, of dimension $mr$, which contains $W_0$ and lies in $W_0+\Mr$. The rank-$r$ part of the slice, $N_V=\{\operatorname{row}(W-W_0)=V\}$, is open and dense in it and is a submanifold of $W_0+\MM_r$. On $N_V$ the field is tangent to $\MM_r$ and equals $\eta GP_{\operatorname{row}(W-W_0)}$, a function of $W$. So (c) descends too, as a degenerate cometric of rank $mr$ (less than $r(m+n-r)$ when $r<n$) whose flow keeps $\operatorname{row}(W-W_0)=\operatorname{row}A_0$ fixed. The cometric of plain GD depends on the representative; it is invariant only under $\Ort(r)$.
\end{proposition}

The algorithms are known. Scaled GD is due to \citet{xtong2021accelerating} and \citet{zhang2024riemannian}, and LoRA-Pro to \citet{wang2024lorac}. LoRA-FA in its frozen-$A$ form is \citet{zhang2023lora}; the corrected gradient of (c) comes from the 2026 revisions (v2 and v3) of that arXiv record and from the LoRA-FA optimizer added to HF PEFT \citep{xmangrulkar2022peft} in 2025. The $\GL_r$-invariance of the scaled-GD metric and its descent to the quotient $\MM_r$ are classical \citep{mishra2012fixed,xmishra2016riemannian}.

\noindent\labelsf{Num}\enspace Gauge defects are about $10^{-15}$ for (a), against $52$ for plain GD. LoRA-FA with AdamW from $A_0$ and from $gA_0$ differs by $O(1)$ after a few steps, against $10^{-15}$ with SGD (Appendix~\ref{app:repro}).

\paragraph{Practice} Each published method departs from these idealisations, and the departures change the symmetry (\thmref{Theorem}{III.12}).
\begin{itemize}
\item The standard start $B_0=0$ lies off the full-rank locus, where (a) and (b) are undefined without damping. \citet{zhang2024riemannian} use $(B^\top B+\delta I)^{-1}$ with an unspecified $\delta>0$, which is only $\Ort(r)$-equivariant. Their LLM experiments report both damped scaled GD and scaled AdamW, in which the preconditioner is applied before the Adam moments ($B_r$). HF's \texttt{create\_riemannian\_optimizer} damps as well.
\item LoRA-Pro's experiments use its AdamW variant (Alg.~2, with $m\times n$ moments), $\dot W=-\Pi_T(\mathrm{Adam}(\Pi_TG))$. It descends to $W_0+\MM_r$ without being a gradient flow, and its discrete update with the Sylvester-optimal free matrix (\bgref{D.33}) is $\GL_r$-invariant only to first order. The released code also damps both inverses ($\operatorname{pinv}(\cdot+10^{-8}I)$, so $\Ort(r)$), replaces the step at $B_0=0$ by a LoRA-FA-type step, and runs with HF's default global-norm clipping.
\item LoRA-FA as published (v3, Alg.~1) and HF's \texttt{create\_lorafa\_optimizer} apply AdamW after the correction (HF also damps with $10^{-8}I$). The trajectory is then invariant only under signed row permutations of $A_0$ (\thmref{Theorem}{III.12}(b)), so it depends on the basis and scale of $A_0$ as well as on $\operatorname{row}A_0$.
\end{itemize}

\begin{remark*}{2026: Riemannion}
Riemannion \citep{bogachev2025lora} defines its optimizer on $\MM_r$ directly. Its variable is the point $X\in\MM_r$, stored as thin factors, and apart from weight decay each step uses only $X$, the projected gradient and the momentum: the tangent projection of \thmref{Proposition}{III.4}(b), momentum carried along by the same projection, a Muon step \citep{xjordan2024muon} that sets the nonzero singular values of the momentum to $1$ and projects again, and a truncated SVD as retraction. No Gram matrix is inverted, so nothing needs damping, and with weight decay off, the $W$-trajectory, momentum included, does not depend on how $X$ is factored. The published Algorithm~4 writes its weight decay through the orthonormal factors, as a term $\gamma A_LB_R^\top$, so this holds for $\gamma=0$. Its non-linearity acts on a weight-space matrix, while Adam's acts coordinatewise on the factors, which is what cuts the gauge down to $B_r$ (\thmref{Theorem}{III.12}).
\end{remark*}

\begin{explains}
Under (S)GD each method is summarised by its induced weight-space cometric (Table~\ref{tab:cometrics}). LoRA, scaled GD and LoRA-Pro share the image $W_0+\Mr$ and differ only in cometric. LoRA-FA is the sub-object $W_0+\{XA_0\}$ (\thmref{Corollary}{III.13}). GaLore (with a gradient-SVD or a random projector) and MeZO are full-parameter methods, listed for comparison. Flora proper compresses only its momentum, takes full-space Adafactor steps, and has no cometric of this form (Exposé~\ref{sec:lens}).
\end{explains}

\begin{table}[t]
\centering\small\renewcommand{\arraystretch}{1.12}
\caption{Induced weight-space cometrics under gradient flow or SGD (\thmref{Observation}{III.2}, \thmref{Proposition}{III.4}). The last three rows are full-parameter methods, listed for comparison; $P$ and $Q$ are GaLore's left and right projectors, $A_t$ a random projector, $z$ the random direction of MeZO.}\label{tab:cometrics}
\begin{tabularx}{\linewidth}{@{}>{\raggedright\arraybackslash}p{0.47\linewidth}>{\raggedright\arraybackslash}X@{}}
\toprule\headrow
\textbf{Method (gradient flow or SGD)} & \textbf{Cometric $K(G)$}\\
\midrule
LoRA \citep{hu2021lora} & $s^2(\eta_BGA^\top A+\eta_ABB^\top G)$\\
LoRA-FA, original / corrected \citep{zhang2023lora} & $s^2\eta_BGA_0^\top A_0$ / $\eta GP_V$\\
scaled GD, undamped Riemannian preconditioning \citep{xtong2021accelerating,zhang2024riemannian} & $\eta s^2(P_UG+GP_V)$\\
LoRA-Pro, SGD variant \citep{wang2024lorac} & $\eta\,\Pi_{T\MM_r}G$\\
GaLore, one period, scale $\alpha$ \citep{zhao2024galore}; \thmref{Proposition}{IV.2} & $\eta\alpha PP^\top G$, or $\eta\alpha\,GQQ^\top$ when $m>n$ (default $\alpha=0.25$)\\
GaLore with a random projector $A_t$, one period (the compressed SGD of \citealp{hao2024flora}) & $\eta GA_t^\top A_t$\\
MeZO \citep{malladi2023fine} & $\eta zz^\top G$, equal to $\eta G$ in expectation\\
\bottomrule
\end{tabularx}
\end{table}

``Transformation invariance'' in the sense of Definition~1 of LoRA-RITE \citep{yen2024lora} means that the $W$-update is invariant under the gauge ($W$-level descent), which is weaker than parameter-level $\GL_r$-equivariance. LoRA-RITE's algorithm satisfies the stronger property on the full-rank locus (\thmref{Theorem}{III.12}).

\subsubsection{Isometries and charges}\label{sec:fibres-charges}

Split LoRA's gauge in two. With per-block scalar rates, the antisymmetric generators rotate the rank channels and preserve the rate metric, so the optimizer cannot feel them. Along the symmetric ones, gradient flow conserves a quadratic charge (\bgref{D.26}). For LoRA the charges form one matrix $\Phi$, whose vanishing is the balancedness of deep linear networks (\bgref{D.28}).

\begin{theorem}{III.5}{Noether charges and isometries for cotangent signals; known, \citet{zhao2022symmetries}}
Let a Lie group act linearly on $Q=\R^N$ with $\rho\circ k=\rho$, let $\Lambda\succ0$ be constant, and let $\gamma(t)\in T^*_{\rho(q(t))}\Theta$ be any cotangent signal, gradient or not. Consider the continuous-time updates
\[\dot q=-\Lambda\,D\rho_q^\top\gamma(t).\]

(i) For every generator $\xi$ with $\Lambda^{-1}\xi$ symmetric, the charge $J_\xi(q)=\tfrac12\langle q,\Lambda^{-1}\xi q\rangle$ is conserved.

(ii) For every generator $\xi$ with $\Lambda^{-1}\xi$ antisymmetric, $k=e^{\tau\xi}$ preserves $\langle\cdot,\Lambda^{-1}\cdot\rangle$, so $K_{kq}=K_q$ along the orbit; for the same cotangent $\gamma$, the $\theta$-velocities at $q$ and at $kq$ agree. Suppose moreover that $\gamma$ depends on $q$ only through $(t,\rho(q))$, as $\gamma=\nabla_\theta L$ or an open-loop signal does, and that solutions are unique ($\gamma$ measurable in $t$ and locally Lipschitz in $\theta$, $\rho$ of class $C^{1,1}_{\mathrm{loc}}$). Then the $\theta$-trajectories from $q_0$ and from $kq_0$ coincide. If $\gamma$ uses Euclidean norms on $Q$, as global-norm clipping does, the trajectory statement holds when, in addition, $k$ is a Euclidean isometry, and can fail otherwise. So (i) and the velocity statement of (ii) hold for arbitrary $\gamma$, while the trajectory statement needs a gradient-like signal.

(iii) When $\mathfrak g$ is closed under $\xi\mapsto\Lambda\xi^\top\Lambda^{-1}$, every generator splits into a part of type (i) and a part of type (ii), a Cartan decomposition $\mathfrak g=\mathfrak k\oplus\mathfrak p$ (\bgref{D.27}). This holds for LoRA's $\gl_r$ with per-block scalar rates. Without closure, there are generators of neither type. With per-rank-channel rates $\Lambda(B,A)=(BD_B,D_AA)$, where $D_B=\diag(\eta_{B,j})$ and $D_A=\diag(\eta_{A,j})$, the generator $\xi_X$ is of type (i) iff $XD_B^{-1}$ and $D_A^{-1}X$ are both symmetric, and of type (ii) iff both are antisymmetric. An off-diagonal pair $(i,j)$ contributes one charge and one isometry iff $\eta_{B,i}\eta_{A,i}=\eta_{B,j}\eta_{A,j}$, and $\mathfrak g$ is closed iff all the products $\eta_{B,j}\eta_{A,j}$ are equal. In that case one gets a $D$-conjugate of $\so_r\oplus\Sym_r$ whose isometries are not all Euclidean unless the rates are
block-scalar, so the clipping caveat of (ii) applies. For pairwise distinct products, only the diagonal $X$ give charges and no nonzero $X$ gives an isometry.
\end{theorem}

Part (i) is \citet{zhao2022symmetries}, Prop.~5.1, including the table of symmetric and antisymmetric generators and symmetric learning rates. Laws that hold for every loss and dataset are characterised by \citet{marcotte2023abide}, and the principle goes back to \citet{du2018algorithmic}, \citet{arora2018optimization} and \citet{kunin2020neural}. New here are only the remark that $\gamma$ need not be a gradient, which is formally immediate (along the realised trajectory, $\gamma(t)$ is the gradient of the time-dependent linear loss $\langle\gamma(t),\theta\rangle$), and the PEFT instances below.

\noindent\labelsf{Num}\enspace With a non-gradient, $q$-dependent $\gamma$ on LoRA, $\Phi$ drifts by $1.6\times10^{-12}$. With Euclidean clipping and a non-Euclidean isometry $k$, the $\theta$-trajectories from $q_0$ and $kq_0$ differ by $0.06$. With per-channel rates $\eta_B=(1,1,1)$ and $\eta_A=(1,1,2)$ at $r=3$, closure fails, but $X=E_{12}+E_{21}$ still gives a conserved charge (Appendix~\ref{app:repro}).

\paragraph{Instances} All instances assume gradient flow, or SGD without momentum, with per-block rates. AdamW and heavy-ball momentum fall outside the theorem, since their buffers store $D\rho$ at past points.
\begin{itemize}
\item \textbf{LoRA.} The generators $\xi_X(B,A)=(BX,-XA)$ with $\Lambda=\diag(\eta_B,\eta_A)$ satisfy: $\Lambda^{-1}\xi_X$ is symmetric iff $X=X^\top$. So the $r(r+1)/2$ charges are the entries of
\[\Phi=\frac{B^\top B}{\eta_B}-\frac{AA^\top}{\eta_A},\]
which is ``balancedness'', the $\mathfrak p=\Sym_r$ half of $\gl_r=\so_r\oplus\Sym_r$. The $\so_r$ generators act by Euclidean isometries, so (ii) holds for LoRA even under global-norm clipping.
\item \textbf{DyLoRA} \citep{valipour2022dylora}. The gauge reduces to the torus, so only $\diag\Phi$ is conserved.
\item \textbf{VeRA} \citep{kopiczko2023vera}. The gauge is $(b,d)\mapsto(cb,d/c)$, with charge $\|b\|^2/\eta_b-\|d\|^2/\eta_d$. NOLA \citep{koohpayegani2023nola} is analogous.
\item \textbf{AdaLoRA} \citep{zhang2023adalora} ($P\Lambda Q$), without its orthogonality regulariser and between rank-budget masking events. Each torus edge gives a charge, $\|p_k\|^2-\lambda_k^2$ and $\|q_k\|^2-\lambda_k^2$, at equal rates. The regularised objective (HF default \texttt{orth\_reg\_weight=0.5}) breaks these charges.
\item \textbf{HRA} \citep{yuan2024bridging}. The rescaling $u_i\mapsto cu_i$ gives a conserved $\|u_i\|^2$.
\item \textbf{ReLU bottleneck adapters} \citep{houlsby2019parameter}. Positive per-unit rescalings give one charge per hidden unit.
\item \textbf{TT adapters} (for instance \citealp{yang2024loretta,anjum2024tensor}). One conserved matrix per bond.
\item \textbf{DoRA's detached-norm backward \citep{liu2024dora} and global-norm gradient clipping.} Both still have the form $\Lambda D\rho_{\mathrm{LoRA}}^\top\gamma$, with $\gamma$ the detached or clipped signal, so $\Phi$ is still conserved.
\end{itemize}

Under discrete GD with rates $\eta_B,\eta_A$, one step changes $\Phi$ by exactly
\[\eta_B\,AG^\top GA^\top-\eta_A\,B^\top GG^\top B,\]
so the drift of $\eta\Phi$, or the drift relative to $\Phi$, is $O(\eta^2)$ per step. Adam is not of the form $\Lambda D\rho^\top\gamma$, so the drift of $\Phi$ under Adam measures how far Adam departs from the gauge geometry; this is a prediction (Exposé~\ref{sec:predictions}).

\begin{explains}
Conservation needs only the shape $\dot q=-\Lambda D\rho^\top\gamma$, so it holds for any loss and any data. With $\eta$ the common scale of the rates, a GD step changes $\eta\Phi$ by $O(\eta^2)$ and an Adam step by $O(\eta)$, as panel (a) of Figure~\ref{fig:noether} shows.
\end{explains}

\begingroup\renewcommand{\label}[1]{}%
\begin{analogy*}{Noether; gauge}
\emph{Noether} names the passage from a symmetry to a conserved quantity, here proved for gradient flow without a Lagrangian. \emph{Gauge} means the symmetry group of a parametrisation; no connection is involved.
\end{analogy*}
\endgroup

\begin{figure}[tbp]
\centering
\includegraphics[width=\linewidth]{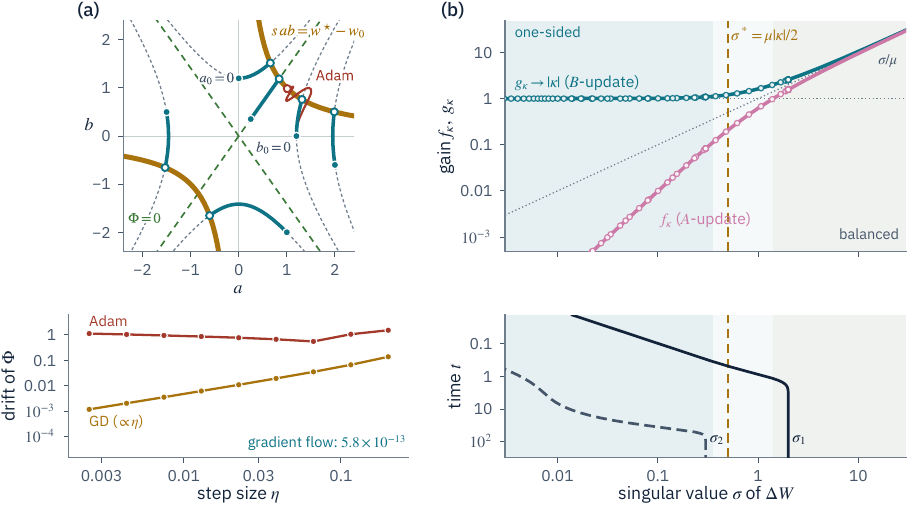}
\caption{\textbf{Noether hyperbolas and the crossover} (\thmref{Theorem}{III.5}, \thmref{Corollary}{III.6}, \thmref{Theorem}{III.9}). (a)~Scalar LoRA, $w=w_0+s\,ba$ with $s=1$, loss $\tfrac12(w-w^\star)^2$, $w^\star-w_0=1$ and rates $\eta_A=1$, $\eta_B=2$. Top: six gradient-flow trajectories (teal) run from the filled dots along their own charge hyperbolas $b^2/\eta_B-a^2/\eta_A=\Phi_0$ (dashed grey; the slice $\Phi=0$ dashed green) to the fibre $s\,ab=w^\star-w_0$ (ochre). From the zero-$B$ start $(1.2,0)$, Adam (red) leaves its hyperbola and lands at another point of the fibre. Bottom: the drift $\max_{t\le10}\lvert\Phi(t)-\Phi_0\rvert$ from the same start against the step size $\eta$ (learning rates $\eta\eta_A$, $\eta\eta_B$). A GD step changes the charge at order $\eta^2$, so over a fixed time the GD drift is proportional to $\eta$ and vanishes in the flow limit. An Adam step changes the charge at first order in $\eta$, and its drift stays of order one at every step size: Adam lies outside \thmref{Theorem}{III.5} (\thmref{Corollary}{III.6}). (b)~Matrix LoRA with $m=n=6$, $r=2$, loss $\tfrac12\lVert W-W^\star\rVert_F^2$, $B_0=0$ and $A_0A_0^\top=I$ exactly, so that $\Phi=-\kappa I$ with $\kappa=1$ and the crossover lies at $\sigma^*=\mu\lvert\kappa\rvert/2=0.5$ (dashed ochre). Top: the closed-form gains $g_\kappa(\sigma^2/\mu^2)$ (teal, the $B$-update) and $f_\kappa(\sigma^2/\mu^2)$ (purple, the $A$-update) of \thmref{Theorem}{III.9}, with the plateau $\lvert\kappa\rvert$ and the balanced asymptote $\sigma/\mu$ dotted. The hollow circles, read off the factors along the integrated flow for both singular directions, lie on the curves. Shading marks the one-sided regime, the crossover, and the balanced regime (grey). Bottom: the singular values $\sigma_1(t)$ and $\sigma_2(t)$ of $\Delta W$, with time running downward on a log scale. $\sigma_1$ crosses $\sigma^*$ and settles in the balanced regime; $\sigma_2$ stays below $\sigma^*$, where the $A$-update supplies only $7\%$ of its growth rate, as in LoRA-FA. Integrators and tolerances are in \S\ref{app:repro-figures}.}\label{fig:noether}
\end{figure}

A conserved charge is fixed at time zero, so it labels initialisations. Weight decay moves it on purpose, and its two conventions move it toward different targets.

\begin{corollary}{III.6}{charge at initialisation; weight decay}
(a) $\Phi_0=B_0^\top B_0/\eta_B-A_0A_0^\top/\eta_A$ is the value of the conserved charge, fixed by the initialisation and the rates. Since $\Ort(r)$ acts on $\Phi$ by conjugation and preserves $\theta$-trajectories (\thmref{Theorem}{III.5}(ii)), only the spectrum of $\Phi_0$ labels a $\theta$-trajectory.

\smallskip
{\upshape\small\renewcommand{\arraystretch}{1.12}\noindent
\begin{tabularx}{\linewidth}{@{}>{\raggedright\arraybackslash}p{0.5\linewidth}>{\raggedright\arraybackslash}X@{}}
\toprule\headrow
\textbf{Initialisation} & \textbf{Charge $\Phi_0$}\\
\midrule
HF default LoRA on Linear and Conv layers ($B_0=0$; Kaiming-uniform $A_0$ with entries $U(-n^{-1/2},n^{-1/2})$, $n$ the fan-in, which is \texttt{in\_channels} times the kernel size for Conv) & $-A_0A_0^\top/\eta_A$, with $\mathbb{E}[A_0A_0^\top]=\tfrac13I_r$ and fluctuations $O(n^{-1/2})$ on and off the diagonal\\
HF default LoRA on embedding layers ($A_0=0$, normal $B_0$) & $+B_0^\top B_0/\eta_B$ (Init[B]-type)\\
HF \texttt{mica} \citep{rudiger2026mica} ($A_0=0$, frozen minor-singular $B_0$) & $+B_0^\top B_0/\eta_B$ (Init[B]-type, with $B$ frozen)\\
EVA \citep{paischer2024parameter} (orthonormal rows; default \texttt{whiten=False}) & $-I/\eta_A$ exactly\\
PiSSA in HF PEFT \citep{meng2024pissa}, and MiLoRA \citep{wang2024milora} if implemented like \texttt{pissa\_init} with the bottom-$r$ singular values (MiLoRA is not in HF PEFT); $B_0^\top B_0=A_0A_0^\top=S_r/s$, with $S_r$ the top-$r$, resp.\ bottom-$r$, singular values and $s$ the LoRA scale & $(S_r/s)(1/\eta_B-1/\eta_A)$, zero \textbf{iff} $\eta_A=\eta_B$ (when $S_r\ne0$)\\
OLoRA \citep{buyukakyuz2024olora} ($B_0=Q_r$, $A_0=R_r$) & $I/\eta_B-R_rR_r^\top/\eta_A$\\
``Init[B]'' \citep{hayou2024impact} ($A_0=0$) & $+B_0^\top B_0/\eta_B$\\
\bottomrule
\end{tabularx}}
\smallskip

(b) Let $G$ be any cotangent signal, with the scale $s$ absorbed into it. Under gradient flow, if decay acts per unit time and identically on both factors,
\[\dot B=-\eta_BGA^\top-\lambda B,\qquad\dot A=-\eta_AB^\top G-\lambda A,\]
then $\dot\Phi=-2\lambda\Phi$. With learning-rate-coupled decay (PyTorch SGD \texttt{weight\_decay}; the decay term of PyTorch AdamW), $\dot B=-\eta_B(GA^\top+\lambda B)$ and likewise for $A$, so that $\dot\Phi=-2\lambda(B^\top B-AA^\top)$. Then $\Phi(t)=e^{-2\eta\lambda t}\Phi(0)$ holds when $\eta_A=\eta_B=\eta$. For $\eta_A\ne\eta_B$ it fails in general, since $-2\lambda(B^\top B-AA^\top)$ is not a function of $\Phi$. With an actual Adam(W) gradient term, $\Phi$ drifts even without decay, at first order in the step size; Adam lies outside \thmref{Theorem}{III.5}. So neither law describes AdamW training.
\end{corollary}

\noindent\labelsf{Num}\enspace Testing $\Phi(t)=e^{-2\eta_A\lambda t}\Phi(0)$ at $\eta_B/\eta_A=4$ with coupled decay gives a relative error of $10.7$; in the valid cases the error is that of the integrator, between $10^{-8}$ and $2\times10^{-4}$. One Adam sign step moves $\Phi$ by an amount proportional to the learning rate, while an SGD step moves $\eta\Phi$ at second order (Appendix~\ref{app:repro}).

\begin{explains}
\begin{itemize}
\item Zero-product initialisations keep their charge under gradient flow, and that charge is maximally unbalanced (\thmref{Proposition}{III.8}). The \emph{relative} imbalance $\|\Phi\|/(\|B\|^2/\eta_B+\|A\|^2/\eta_A)$ decays as the adapter grows (\thmref{Theorem}{III.9}).
\item HF PiSSA always has $\mu_0=B_0^\top B_0-A_0A_0^\top=0$ (unweighted-balanced), and $\Phi_0=0$ only at equal rates. So it starts on the slice that coupled decay targets, and off the per-time-decay slice when $\eta_A\ne\eta_B$.
\item Both decay conventions gauge-fix at equilibrium, because each is a $\Lambda$-gradient flow of $L$ plus a fibre-wise Kempf--Ness potential (\bgref{D.30}). Per-time decay is the flow of $L+\tfrac\lambda2(\|B\|^2/\eta_B+\|A\|^2/\eta_A)$; it drives $\Phi\to0$ exponentially, onto the rate-weighted slice $B^\top B/\eta_B=AA^\top/\eta_A$. Coupled decay is the flow of $L+\tfrac\lambda2(\|B\|^2+\|A\|^2)$; its equilibria lie on the unweighted balanced slice $\mu=B^\top B-AA^\top=0$ of \thmref{Proposition}{III.7}(a), with no closed-form law for $\Phi$ unless $\eta_A=\eta_B$. The two slices coincide iff $\eta_A=\eta_B$. In particular, coupled decay also gauge-fixes at LoRA+ rates, onto $\mu=0$.
\end{itemize}
\end{explains}

\noindent\labelsf{Num}\enspace At $\eta_B/\eta_A=4$, coupled decay converges to $\mu=0$ to within $10^{-12}$ while $\Phi$ stays of order $1$; per-time decay converges to $\Phi=0$ while $\mu$ stays of order $1$ (Appendix~\ref{app:repro}).

The names Init[A] and Init[B], and the finding that these zero-product starts train differently, are due to \citet{hayou2024impact}. Exponential decay of such charges under weight decay is the rescale-symmetry law of \citet{kunin2020neural}.

\subsubsection{The balanced slice}\label{sec:fibres-balance}

In Kempf--Ness theory, the least-norm vectors on a closed orbit of a reductive group form one orbit of its compact part, and they are where the moment map vanishes (\bgref{D.29}, \bgref{D.30}). For LoRA the moment map is $\mu=B^\top B-AA^\top$, and the least-norm representatives are the balanced ones.

\begin{proposition}{III.7}{Kempf--Ness slice; Loewner floor; known, \citet{xkempf1979length,xsrebro2004maximum}}
(a) On a full-rank $\GL_r$-orbit, the critical points of $\|B\|^2+\|A\|^2$ are exactly the balanced representatives, those with $\mu=B^\top B-AA^\top=0$. They form one $\Ort(r)$-orbit, they are the minima, and the minimum value is $2\|BA\|_*$.

(b) Assume gradient flow with block-scalar rates $\eta_A,\eta_B$, no weight decay, a loss that depends on $(B,A)$ only through $BA$, and any cotangent signal of the form $\Lambda D\rho^\top\gamma$ (dropout, minibatches and clipping included). From $B_0=0$,
\[B^\top B=\tfrac{\eta_B}{\eta_A}\big(AA^\top-A_0A_0^\top\big)\succeq0,\]
so $AA^\top\succeq A_0A_0^\top$ for all time; the $A$-factor never shrinks below its initialisation in the Loewner order. $AA^\top(t)$ need not be monotone in $t$; only this floor is guaranteed. Discrete steps break the bound by $O(\eta^2)$ per step. Under Adam it is not guaranteed and fails empirically. With per-time decay $\lambda$, only $AA^\top\succeq e^{-2\lambda t}A_0A_0^\top$ holds.
\end{proposition}

Part (a) is the theorem of \citet{xkempf1979length}, in the real form of \citet{xrichardson1990minimum}; the nuclear-norm identity is due to \citet{xsrebro2004maximum}. BaLoRA \citep{castin2026balanced} turns the slice into an algorithm, moving after every step to a product-preserving representative with $B^\top B=AA^\top$ diagonal.

\noindent\labelsf{Num}\enspace On $5$ random orbits at $(m,n,r)=(7,5,3)$ the minimum equals $2\|BA\|_*$ to within $10^{-14}$ (Appendix~\ref{app:repro}).

\begin{explains}
Balancedness is a precise notion: it is the Kempf--Ness slice of the gauge action. An explicit penalty $\lambda(\|B\|^2+\|A\|^2)$ is equivalent at its minimisers to a $(2\lambda/s)$-nuclear-norm penalty on $\Delta W=sBA$ restricted to rank $\le r$. SGD-coupled decay with coefficient $\lambda$ is the flow of $L+\tfrac\lambda2(\|B\|^2+\|A\|^2)$ and matches a $(\lambda/s)$-nuclear-norm penalty (\citealp{xsrebro2004maximum}, whose argument also covers rank-deficient factorisations). In both cases the stationary points lie on $\mu=0$. With per-time decay at unequal rates the relevant slice is $\Phi=0$ (\thmref{Corollary}{III.6}). Decoupled AdamW decay is not covered.
\end{explains}

The default start is as unbalanced as possible, and no start meets all three natural wishes.

\begin{proposition}{III.8}{the balance--escape trilemma}
No LoRA initialisation is simultaneously
\begin{enumerate}[label=(\alph*)]
\item identity-at-init without a residual split ($B_0A_0=0$),
\item balanced ($B_0^\top B_0=cA_0A_0^\top$ for some $c>0$: unweighted, $c=1$, or rate-weighted, $\Phi_0=0$ with $c=\eta_B/\eta_A$), and
\item non-stationary, meaning that $D\rho$ at $(B_0,A_0)$ is not the zero map; equivalently, $\dot W(0)\ne0$ for some parameter velocity, under any first-order optimizer.
\end{enumerate}
Moreover, if $B_0A_0=0$ then $\langle B_0^\top B_0,A_0A_0^\top\rangle_F=0$, so
\[\|B_0^\top B_0-cA_0A_0^\top\|_F^2=\|B_0^\top B_0\|_F^2+c^2\|A_0A_0^\top\|_F^2,\]
and zero-product initialisations are maximally unbalanced.
\end{proposition}

\begin{explains}
The field splits into two families of initialisation.
\begin{itemize}
\item \textbf{Type I} keeps $\rho$ and picks $q_0$ in the apex fibre $\{BA=0\}$: LoRA, EVA, Init[B]. By the identity above these are maximally unbalanced.
\item \textbf{Type II} changes $\rho$ by an exact split $\theta_0=\theta_{\mathrm{res}}+\delta(q_0)$: PiSSA, MiLoRA, OLoRA, LoRA-GA \citep{wang2024lorab}, CorDA \citep{yang2024corda}. LoftQ \citep{li2023loftq} and QPiSSA \citep{meng2024pissa} are approximate splits over a quantised residual, so they are not identity-at-init (\thmref{Proposition}{V.2}).
\end{itemize}
The proposition forces every balanced, non-stationary, identity-at-init LoRA initialisation to be Type II. The converse fails: OLoRA ($B_0^\top B_0=I$, $A_0A_0^\top=R_rR_r^\top$) and CorDA are Type II and generally unbalanced. A frozen residual is the price of any nonzero $B_0A_0$, and so in particular of balance, as in PiSSA and MiLoRA at equal rates. By \thmref{Theorem}{II.2}, Type II objects are different objects, incomparable with LoRA.
\end{explains}


\subsubsection{Two regimes and a crossover}\label{sec:fibres-crossover}

When the charge is a multiple of the identity, the LoRA flow has a closed form. Below a scale $\sigma^*$, one factor
carries the motion, as in LoRA-FA \citep{zhang2023lora}; above it, the factors share the work, as in balanced deep
linear networks.

\begin{theorem}{III.9}{closed-form isotropic-charge LoRA dynamics; known, \citet{xtarmoun2021understanding} and \citet{xmin2021explicit}}
Let $\rho(B,A)=W_{\mathrm{res}}+sBA$ be trained by gradient flow with rates $\eta_A,\eta_B$, and suppose the charge is
isotropic, $\Phi=-\kappa I_r$. Put $\mu=s\sqrt{\eta_A\eta_B}$ (a rate, unrelated to the moment map $\mu$ of
\thmref{Proposition}{III.7}), and on the locus where $Z=W-W_{\mathrm{res}}$ has rank $r$ write its compact SVD
$Z=U\Sigma V^\top$. Then
\[\dot W=-\mu^2\big[G\,V g_\kappa(\Sigma^2/\mu^2)V^\top+U f_\kappa(\Sigma^2/\mu^2)U^\top G\big],\]
where
\[f_\kappa(x)=\sqrt{x+\kappa^2/4}-\kappa/2,\qquad g_\kappa=f_\kappa+\kappa .\]
The identity holds pointwise on $\{\Phi=-\kappa I,\ \rk Z=r\}$; gradient flow is used only to keep $\Phi$ fixed. The
two terms under the root are equal at the \emph{crossover scale}
\[\sigma^*=\mu\lvert\kappa\rvert/2.\]
\begin{itemize}
\item For $\sigma\ll\sigma^*$ with $\kappa>0$, $f\to0$ and $g\to\kappa$, so $\dot W\approx-\mu^2\kappa\,GP_V$. These
  are LoRA-FA dynamics. The contribution of $A$ to $\dot W$ is $O(\sigma^2/(\mu^2\kappa))$, although $A$ itself still
  drifts slowly. For $\kappa<0$ the roles swap, $\dot W\approx-\mu^2\lvert\kappa\rvert P_UG$, and only $A$ effectively
  moves (Init[B]).
\item For $\sigma\gg\sigma^*$, $f\approx g\approx\sigma/\mu$, so
  \[\dot W\approx-\mu\big[G(Z^\top Z)^{1/2}+(ZZ^\top)^{1/2}G\big].\]
  These are the balanced dynamics of Arora, Cohen and Hazan.
\item \emph{Off the rank-$r$ locus.} Write $\dot W=-\mu^2(GS+TG)$ with $S=A^\top A/\eta_A$ and $T=BB^\top/\eta_B$, so
  that on the rank-$r$ locus $S=Vg_\kappa(\Sigma^2/\mu^2)V^\top$ and $T=Uf_\kappa(\Sigma^2/\mu^2)U^\top$. For
  $\kappa\ge0$, $T=f_\kappa(ZZ^\top/\mu^2)$ still holds as a full matrix function, but $S$ gains the term
  $\kappa(P_{\operatorname{row}A}-P_V)$; symmetrically, for $\kappa<0$, $T$ gains
  $\lvert\kappa\rvert(P_{\operatorname{col}B}-P_U)$. At the default-LoRA start $Z=0$ ($B_0=0$) this leaves
  $S=\kappa P_{\operatorname{row}A_0}$ and $T=0$, so $\dot W=-\mu^2\kappa\,GP_{\operatorname{row}A_0}$, and the
  $\sigma\ll\sigma^*$ regime extends continuously to $t=0$.
\end{itemize}
\end{theorem}

The closed form is due to \citet{xtarmoun2021understanding}, who call this charge ``$\lambda$-balanced'', and to
\citet{xmin2021explicit}; see also \citet{xdomine2025lazy} and \citet{xkunin2024get}. The balanced regime is that of
\citet{arora2018optimization}. The PEFT reading is ours. Panel (b) of Figure~\ref{fig:noether} shows the closed-form
singular-value dynamics and the crossover scale $\sigma^*$.

\noindent\labelsf{Num}\enspace The formula matches direct integration to $10^{-12}$, and the pointwise identity holds to
$10^{-15}$ for $\kappa\in\{-2,-0.3,0,0.3,2,10\}$ and all shape cases (Appendix~\ref{app:repro}).

\begin{explains}
Call this the PEFT reading; it is scoped to gradient flow and SGD.
\begin{itemize}
\item \emph{Default LoRA.} With $B_0=0$ and $A_0A_0^\top=cI$, $\kappa=c/\eta_A$ and
  $\sigma^*=\tfrac{sc}{2}\sqrt{\eta_B/\eta_A}$; HF's default gives $c\approx1/3$.
\item \emph{LoRA+} ($\lambda=\eta_B/\eta_A$) multiplies $\sigma^*$ by $\sqrt\lambda$ at fixed $\eta_A$, and its
  balanced-regime rate is the geometric mean $s\sqrt{\eta_A\eta_B}$. LoRA-FA is the endpoint $\lambda\to\infty$ of the
  LoRA+ family.
\item \emph{PiSSA at equal rates} has $\kappa=0$ and is exactly balanced forever under gradient flow; SGD drifts by
  $O(\eta^2)$ per step, and Adam does not preserve it.
\item \emph{Exactness.} The theorem is exact for EVA- and LoRA-GA-type orthonormal frames. For Kaiming initialisations
  it holds only approximately. The eigenvalues of $A_0A_0^\top$ lie near $c(1\pm\sqrt{r/n})^2$
  (\citealp{xmarchenko1967distribution}; \auxref{B.35}), so the crossover is a band of relative width $O(\sqrt{r/n})$ rather than a single
  scale. At $r=256$, $n=4096$, the eigenvalues of $3A_0A_0^\top$ span roughly $0.57$ to $1.57$.
\item \emph{Random $A$.} The one-sided regime fits the finding of \citet{zhu2024asymmetry} that a random, untrained $A$
  does nearly as well as a tuned one; the theorem says that this regime lasts while $\Delta W$ stays well below
  $\sigma^*$.
\end{itemize}
\end{explains}

\noindent\labelsf{Prediction}\enspace Under SGD, singular values of $\Delta W$ switch from one-sided to two-sided growth as they cross
$\sigma^*$ (\thmref{Prediction}{X.7}).

\subsubsection{The hyperparameter torus}\label{sec:fibres-torus}

Rescale block $i$ by $c_i$ and divide $s$ by the matching power of the $c_i$. The method is the same object of $\Meth$,
and training is the same if every step rescales alike. Adam ignores the size of a block's gradient, so its rate must
scale like $c_i$. An SGD step scales with the gradient, which shrinks by $c_i$, so its rate must scale like $c_i^2$.

\begin{theorem}{III.10}{the hyperparameter torus; the LoRA/Adam case is known, \citet{schulman2025lora}}
Let $\rho(q)=\theta_b+s\,\delta(q_1,\dots,q_k)$ with $\delta$ multihomogeneous of multidegree $(d_1,\dots,d_k)$. Train
block $i$ from $q_i(0)=\sigma_iu_i$ with Adam \citep{xkingma2015adam} ($\beta_1\in[0,1)$, $\beta_2\in(0,1)$, bias
correction allowed) at learning rate $\eta_i(t)$, on the loss $L\circ\rho$ with no $q$-dependent regulariser. Assume no
gradient clipping (or per-block clipping with thresholds rescaled as below), common random numbers (the same data order
and dropout masks in both runs), and exact arithmetic (the result is bit-exact in floating point when every $c_i$ is a
power of $2$). Then the trajectory $t\mapsto\rho(q(t))$ is invariant under the torus action of $c\in\R_{>0}^k$:
\[\big(s,\ \eta_i(t),\ \sigma_i,\ \varepsilon_i,\ \lambda_i,\ C_i\big)\mapsto\Big(s\prod_ic_i^{-d_i},\ c_i\eta_i(t),\ c_i\sigma_i,\ \varepsilon_i/c_i,\ \lambda_i/c_i,\ C_i/c_i\Big).\]
Here $\varepsilon_i$ is a per-block Adam $\varepsilon$ (with the convention $0/0=0$ at $\varepsilon=0$), $\lambda_i$ a
per-block decoupled (AdamW) weight decay, and $C_i$ a per-block clipping threshold. Under SGD, with or without
heavy-ball or Nesterov momentum, the same holds with $\eta_i(t)\mapsto c_i^2\eta_i(t)$ and coupled $L^2$ decay
$\lambda_i\mapsto\lambda_i/c_i^2$. In particular, any learning-rate schedule shared by all blocks, warm-up included, is
preserved.
\end{theorem}

The LoRA/Adam case, with the $\varepsilon>0$ extension and the two effective hyperparameters, is due to
\citet{schulman2025lora} (\emph{LoRA Without Regret}). New here are arbitrary multihomogeneous $\delta$ and the SGD and
heavy-ball $c^2$ law.

\noindent\labelsf{Num}\enspace Trajectory differences are exactly $0.0$ for power-of-two $c$ and about $10^{-15}$ otherwise,
including a warm-up schedule and rescaled $\varepsilon$ and decay. With $\varepsilon$ or decay held fixed, or with
global-norm clipping, they are $10^{-3}$ to $O(1)$ (Appendix~\ref{app:repro}).

\begin{explains}
Only the torus orbit of the hyperparameters reaches the weights. LoRA is the case $k=2$, $d=(1,1)$; $\delta$ may be any
multilinear update, such as that of VeRA \citep{kopiczko2023vera} in its two scaling vectors or that of LoHa
\citep{hyeonwoo2021fedpara} in its four factors.
\end{explains}

\begin{corollary}{III.11}{LoRA+ is $\alpha$ in disguise; rsLoRA is LoRA; (a) and the LoRA+ reading are known, \citet{schulman2025lora}}
For linear LoRA modules ($k=2$, $d=(1,1)$, $\sigma_B=0$), under the hypotheses of \thmref{Theorem}{III.10}:

(a) \emph{Effective hyperparameters.} Under Adam the torus invariants of $(s,\eta_A,\eta_B,\sigma_A)$ are
$(s\eta_A\eta_B,\ \sigma_A/\eta_A)$; the quantities $\alpha\cdot\mathrm{init}_A\cdot\mathrm{LR}_B$ and
$\mathrm{init}_A/\mathrm{LR}_A$ of Schulman et al.\ are equivalent generators. Under SGD the invariants are
$(s^2\eta_A\eta_B,\ \sigma_A^2/\eta_A)$.

(b) \emph{LoRA+.} Under Adam with $\varepsilon=0$, no weight decay, no gradient clipping and a learning-rate schedule
shared by both groups, LoRA+ with ratio $\lambda$ at scale $s$ produces \emph{exactly} the trajectory of plain LoRA at
scale $\lambda s$ (same $\eta_A$, same $A_0$). For $\varepsilon>0$ the equivalence needs a per-block
$\varepsilon_B'=\lambda\varepsilon$, and with AdamW a per-group decay $\lambda\cdot\mathrm{wd}$ on $B$. Under SGD it
holds at scale $\sqrt\lambda\,s$, or equivalently with plain LoRA at the single rate $\sqrt{\eta_A\eta_B}$ and
$A_0\mapsto\lambda^{1/4}A_0$.

(c) \emph{rsLoRA.} $\rho_{\mathrm{rs}}(B,A)=\rho_{\mathrm{LoRA}}(\sqrt r\,B,A)$ is an isomorphism of objects. Up to the
hyperparameter torus, rsLoRA can be represented as LoRA at the same $\alpha$ with $\eta_B$ multiplied by $\sqrt r$ under
Adam (by $r$ under SGD), or equally as LoRA with $\alpha$ multiplied by $\sqrt r$ at the same rates. The reading ``same
object, different metric'' is a choice of representative within the orbit. Part (c) concerns a fixed rank; it does not
address rsLoRA's rank-stability prescription, how $\alpha$ should scale with $r$.
\end{corollary}

LoRA+ is due to \citet{hayou2024lora}, rsLoRA to \citet{kalajdzievski2023rank}, and (a) with the LoRA+ reading to
\citet{schulman2025lora}. On the empirical side, \citet{lee2026learning} find that nine LoRA variants reach similar peak
performance once the learning rate is tuned.

\noindent\labelsf{Num}\enspace The maximum trajectory difference is exactly $0.0$ at $\varepsilon=0$ for power-of-two
$\lambda$ and about $10^{-14}$ otherwise, against $0.93$ for the unrescaled control. It is $5\times10^{-9}$ at a shared
$\varepsilon=10^{-8}$, and $0.6$ with global-norm clipping at $0.5$ (Appendix~\ref{app:repro}).

\begin{explains}
LoRA's four knobs are coordinates on a 4-dimensional space acted on by a 2-torus, and only the orbit matters. LoRA+,
rsLoRA and $\alpha$-tuning therefore collapse into one family. The slogan depends on the chart. Holding $\eta_A$ fixed,
LoRA+ is ``$\alpha$ multiplied by $\lambda$''; holding $\eta_B$ fixed, it is plain LoRA with $\eta_A$ divided by
$\lambda$, a slower $A$-timescale. In particular, Adam does not make the LoRA+ ratio non-absorbable.
\end{explains}

\noindent\labelsf{Prediction}\enspace Any gain of LoRA+ over LoRA with $\alpha$ re-tuned by the factor $\lambda$ must come from the
ingredients that break the extended torus: global-norm clipping (on by default in the HF Trainer,
\texttt{max\_\allowbreak grad\_\allowbreak norm=1.0}), $\varepsilon$ or weight decay held equal across parameter groups (as in HF PEFT's LoRA+
optimizer), and learning-rate schedules that differ between groups. Shared schedules and warm-up preserve the torus.
HF's LoRA+ optimizer treats embedding adapters and 1-D parameters separately, so the statement covers linear modules
only (\thmref{Prediction}{X.1}).

\begin{figure}[tbp]
\centering
\includegraphics[width=\linewidth]{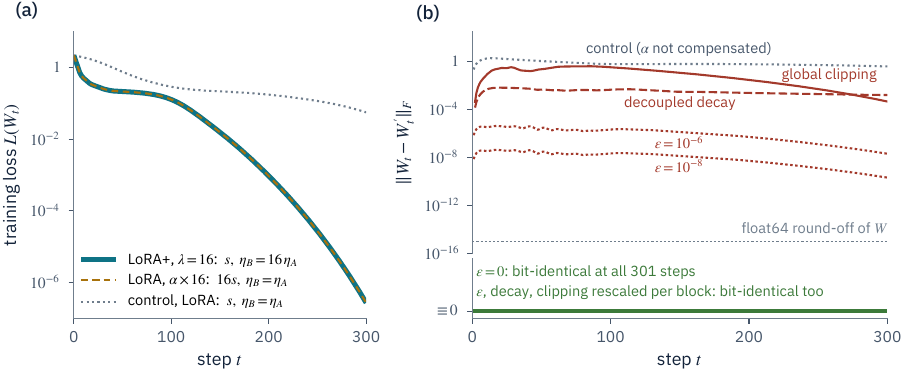}
\caption{\textbf{LoRA+ is $\alpha$ in disguise} (\thmref{Theorem}{III.10}, \thmref{Corollary}{III.11}). A seeded toy LoRA regression with $B_0=0$ is trained by Adam in float64, once as LoRA+ with ratio $\lambda=16$ (scale $s$, $\eta_B=16\,\eta_A$) and once as plain LoRA with $\alpha$ multiplied by $16$ (scale $16s$, $\eta_B=\eta_A$), at the same $\eta_A$ and $A_0$. (a)~At $\varepsilon=0$, without weight decay or clipping, the two loss curves coincide; the control, plain LoRA at the uncompensated scale $s$, ends far above them. (b)~The weight gap $\lVert W_t-W'_t\rVert_F$ between the two runs at every step, on a log scale above an exact-zero floor. At $\varepsilon=0$ the runs are bit-identical at every step: the torus element relating them, $c=(1,1/16)$, is a power of two and rescales floating-point numbers exactly. Each red curve switches on one ingredient with a single value shared by both parameter groups: Adam's $\varepsilon$, decoupled weight decay, or global-norm clipping. Each breaks the hyperparameter torus; the control is dotted. Rescaling the same ingredients per block, as \thmref{Theorem}{III.10} prescribes, restores the exact zero (green floor). The values used are in \S\ref{app:repro-figures}.}
\label{fig:orbit}
\end{figure}

\subsubsection{What the optimizer can see}\label{sec:fibres-hierarchy}

An optimizer is equivariant under a gauge element $g$ if starting from $(Bg^{-1},gA)$ gives the transported trajectory.
The Euclidean metric of SGD respects rotations of the rank channels. Adam, acting coordinatewise, respects only the
signed permutations.

\begin{remark*}{2026: privacy noise}
Privacy noise fits the same pattern. DP-SGD \citep{xabadi2016deep} on the factors adds isotropic Gaussian noise $\xi$ to
the clipped factor gradients, and $\xi g^{-1}$ has the law of $\xi$ only for orthogonal $g$, so it respects $\Ort(r)$
and no larger subgroup. PRISM \citep{wang2026prism} clips and noises the projection of each gradient onto the tangent
space of $\MM_r$ at $BA$, which depends only on $\operatorname{col}B$ and $\operatorname{row}A$. For full-rank factors
the privatised gradient then has the same law for every factorisation of $BA$. Its Algorithm 1 then applies damped
$r\times r$ preconditioners (its Eqs.~24--26), which the paper also calls gauge invariant. They respect only $\Ort(r)$.
Under $(B,A)\mapsto(cB,A/c)$ the preconditioned factor steps do not change, so $\delta B\,A$ scales by $1/c$ and
$B\,\delta A$ by $c$. The released code returns to balanced factors, $B^\top B=AA^\top$, after every step, where the
remaining gauge is $\Ort(r)$.
\end{remark*}

\begin{theorem}{III.12}{the optimizer gauge hierarchy}
On LoRA's full-rank locus, with hyperparameters shared by all rank channels, the largest subgroup of the gauge $\GL_r$
under which an update rule is equivariant, at the level of parameters, is as follows.

(a) SGD and GD with any block rates, with or without heavy-ball momentum: $\Ort(r)$.

(b) Adam and AdamW at fixed learning rates ($\beta_1,\beta_2\in[0,1)$, any $\varepsilon\ge0$, any weight decay): the
signed permutations $B_r$, of order $2^rr!$. A positive diagonal $g=\diag(d)$ is a symmetry only \emph{jointly} with
per-rank-channel hyperparameters, and only without global-norm clipping. At $\varepsilon=0$ and without weight decay it
suffices to rescale the rates, $\eta_{B,j}\mapsto\eta_{B,j}/d_j$ and $\eta_{A,j}\mapsto d_j\eta_{A,j}$. With
$\varepsilon>0$ one must also give each channel its own $\varepsilon_{B,j}=d_j\varepsilon$ and
$\varepsilon_{A,j}=\varepsilon/d_j$. With decoupled decay multiplied by the learning rate (PyTorch AdamW) one must also
set $\lambda_{B,j}=d_j\lambda$ and $\lambda_{A,j}=\lambda/d_j$; under the convention of \citet{xloshchilov2019decoupled},
where decay is scaled only by the schedule multiplier, $\lambda$ needs no change.

(c) Undamped scaled GD ($\delta=0$; the Riemannian preconditioner of Tong, Ma and Chi and of Zhang and Pilanci, with or
without heavy-ball momentum): all of $\GL_r$. The $W$-trajectory from gauge-related initial states (with zero or
correspondingly transformed momentum) is independent of the representative; a step may drop rank, so the iteration maps
into $W_0+\Mr$.
\end{theorem}

Scaled GD is due to \citet{xtong2021accelerating} and \citet{zhang2024riemannian}. LoRA-RITE \citep{yen2024lora} makes
gauge invariance a design requirement for LoRA optimizers and reports that it pays.

\paragraph{Published variants} Damped scaled GD ($\delta>0$) is only $\Ort(r)$-equivariant. Wrapped in AdamW, as in
Zhang and Pilanci's scaled AdamW (their practical variant) and in HF's \texttt{create\_\allowbreak riemannian\_\allowbreak optimizer}, it is
$B_r$-equivariant by (b). The first-order $W$-level dynamics of LoRA-Pro \citep{wang2024lorac} descend
(\thmref{Proposition}{III.4}(b)), and so do those of its variant with free matrix $X=0$; its published parameter update
with the Sylvester-optimal $X$ is only $\Ort(r)$-equivariant, and its discrete $W$-step depends on the representative at
$O(\eta^2)$. LoRA-RITE states only $W$-level invariance (its Definition 1). Its update (unmagnified gradients,
transported matrix moments, then multiplication by $(R_B^\top)^+$) is $\GL_r$-equivariant at the parameter level on the
full-rank locus: under a gauge change its internal quantities change only by orthogonal matrices, and the final
multiplication supplies the gauge element (Appendix~\ref{proof:III.12}, which follows its Algorithm~1 line by line; the
resulting step is Eq.~15 of \citet{yen2024lora}). LoRA-RITE is therefore an Adam-type optimizer in class (c).

\noindent\labelsf{Num}\enspace For Adam, the defects are $10^{-15}$ for signed permutations, $11.3$ for a positive
diagonal, $6.6$ for a rotation, and $5\times10^{-16}$ for the diagonal with jointly rescaled per-channel rates at
$\varepsilon=0$. With $\varepsilon=10^{-8}$ left unrescaled the joint torus fails by about $2\%$. LoRA-RITE (its
Algorithm 1) from $(B,A)$ and from $(Bg^{-1},gA)$ with a generic $g$ agrees to $3\times10^{-14}$ over 30 steps
(referee replication). LoRA-Pro with the Sylvester $X$ has a parameter-level defect of $1.2$ under a non-orthogonal $g$,
against $2\times10^{-15}$ with $X=0$ or with an orthogonal $g$ (Appendix~\ref{app:repro}).

\begin{explains}
There is a hierarchy $B_r\subset\Ort(r)\subset\GL_r$ of optimizer geometries, strict for $r\ge2$. $\GL_r$-equivariant
rules, started from gauge-related states, define an optimizer on the \emph{model} rather than on a chart. More generally
any rule that maps gauge orbits to gauge orbits does (for instance a step followed by a canonical refactorisation of
$W$), while $B_r$- and $\Ort(r)$-equivariant rules in general do not. Plain Adam, the default, sees a finite shadow of
the gauge. LoRA+ and rsLoRA act on the hyperparameter torus (\thmref{Theorem}{III.10}) and leave the gauge of a fixed
method alone; balanced or orthonormal initialisations choose a point in the fibre. The positive torus is a symmetry of
the \emph{hyperparameters}; Adam at fixed rates does not have it.
\end{explains}

\begin{figure}[tbp]
\centering
\includegraphics[width=\linewidth]{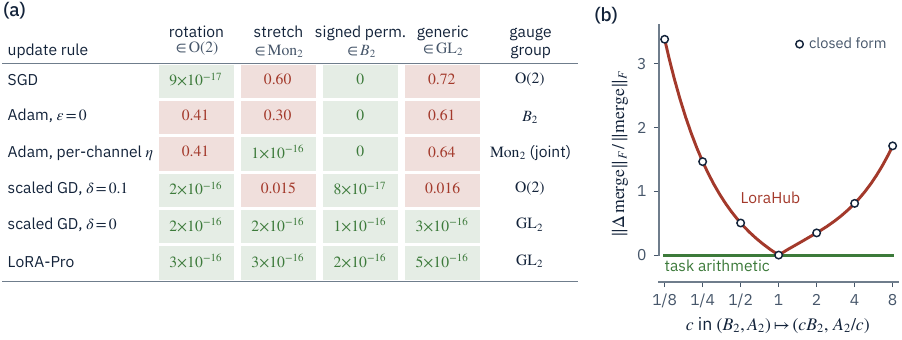}
\caption{\textbf{Update rules that see the gauge} (\thmref{Theorem}{III.12}, \thmref{Proposition}{III.4}, \thmref{Proposition}{V.4}). (a)~At a fixed seeded LoRA state $(B,A)$ with loss gradient $G$, each cell is the relative gauge defect $\lVert\delta W(g)-\delta W(I)\rVert_F/\lVert\delta W(I)\rVert_F$ of one update rule: the change in its first-order weight update $\delta W=\delta B\,A+B\,\delta A$ when the factors are replaced by the gauge-equivalent pair $(Bg^{-1},gA)$. The columns are a rotation, a stretch, a signed permutation and a generic $g\in\GL_2$; under each name is the smallest of the groups $B_2=\Ort(2)\cap\mathrm{Mon}_2$, $\Ort(2)$, $\mathrm{Mon}_2$ (monomial matrices) and $\GL_2$ that contains it. Green cells vanish up to float64 round-off, and they occur exactly where $g$ lies in the rule's gauge group (right column): $\Ort(2)$ for SGD and for damped scaled GD, $B_2$ for Adam, $\mathrm{Mon}_2$ for Adam whose per-rank-channel rates are rescaled jointly with a stretch, and all of $\GL_2$ for undamped scaled GD and for LoRA-Pro with the Sylvester-optimal free matrix, whose weight updates depend only on $W$. Red cells are genuine defects. (b)~Three LoRA modules $(B_i,A_i)$ with weights $w_i$ are merged after module~2 alone is re-gauged to $(cB_2,A_2/c)$. Task arithmetic \citep{ilharco2022editing}, $\sum_iw_iB_iA_i$, does not move. The LoraHub merge \citep{huang2023lorahub}, $(\sum_iw_iB_i)(\sum_jw_jA_j)$, does, and the open circles, the closed form of \thmref{Proposition}{V.4}, match the recomputed change. The settings are in \S\ref{app:repro-figures}.}
\label{fig:gauge}
\end{figure}

\subsubsection{What a zero-\texorpdfstring{$B$}{B} start chooses}\label{sec:fibres-zero-b}

Default LoRA starts at $B_0=0$ with a random full-rank $A_0$. The image remembers only $\operatorname{row}A_0$
(\thmref{Theorem}{II.2}). Training remembers what the optimizer's gauge group cannot move.

\begin{corollary}{III.13}{what a zero-$B$ initialisation really chooses}
At $q_0=(0,A_0)$ with $\rk A_0=r$, with hyperparameters shared by the rank channels and the same data order and dropout
masks, two choices of $A_0$ give the same $\theta$-trajectory whenever the optimizer's gauge group relates them. The
resulting invariant of $A_0$ is:
\begin{itemize}
\item $\operatorname{row}A_0\in\Gr(r,n)$ for an update rule that is $\GL_r$-equivariant at every point of its
  trajectory, including $B=0$ and rank-deficient $B$, run without global-norm clipping. An example is
  $\delta B=-\eta\nabla_B(AA^\top)^{-1}$, $\delta A=-\eta(AA^\top)\nabla_A$, which needs only $\rk A=r$;
\item the PSD matrix $A_0^\top A_0$ for SGD, which is exactly the initial preconditioner in
  \[\dot W(0)=-s^2\eta_BGA_0^\top A_0;\]
\item $A_0$ up to signed permutation of its rows for Adam.
\end{itemize}
The undamped scaled-GD optimizers of \thmref{Theorem}{III.12}(c) are not defined at $B=0$, since they need
$(B^\top B)^{-1}$. LoRA-RITE uses pseudo-inverses and so is defined there, but its $\GL_r$-equivariance is established
only on the full-rank locus. Their damped forms are only $\Ort(r)$-equivariant, so in practice their invariant is a
$\delta$-dependent function of $A_0^\top A_0$, and their Adam-wrapped forms (Riemannian-preconditioned AdamW, HF's
LoRA-FA optimizer) have the invariant $A_0$ modulo $B_r$. Conversely, different invariants give different dynamics, for
SGD and for Adam at $\varepsilon=0$.

Random-$A$ LoRA and EVA are different points of the $B=0$ stratum $\{(0,A):\rk A=r\}$ of the apex fibre
$\rho^{-1}(\theta_0)=\{(B,A):BA=0\}$; Init[B] lies in another stratum of the same fibre. LoRA-FA is a different
\emph{object}, the sub-object $\mathrm{FA}(A_0)\hookrightarrow\mathrm{LoRA}_{(0,A_0)}$ (\thmref{Proposition}{I.8}). More
precisely, EVA is a point of the apex fibre of $\mathrm{LoRA}_{r_l}$ at its per-layer redistributed rank and adjusted
$\alpha$, with orthonormal rows only when \texttt{whiten=False}.
\end{corollary}

Appendix~\ref{proof:III.13} proves the converse (different invariants give different dynamics) by comparing first
steps. \citet{hayou2024impact} showed that the two zero-product starts, random $A$ with
$B=0$ and random $B$ with $A=0$, train differently; here they lie in different strata of one apex fibre, with charges of
opposite sign. EVA \citep{paischer2024parameter} and random-$A$ LoRA share a stratum.

\noindent\labelsf{Num}\enspace SGD from $A_0$ and from $QA_0$ (with $Q$ orthogonal) agree to $10^{-15}$, and from $A_0$
and $DA_0$ (with $D$ diagonal) differ by $0.81$. Adam from $A_0$ and from $PA_0$ (with $P$ a signed
permutation) agree to $10^{-16}$, and from $A_0$ and $QA_0$ differ by $0.98$. The example rule from $A_0$ and from
$gA_0$ agrees to $3\times10^{-15}$. LoRA-RITE's Algorithm 1, run as written with its pseudo-inverses from $B_0=0$, gives
the same weights from $A_0$ and from $gA_0$ for a generic $g$ to $10^{-13}$, and weights that differ by $O(1)$ for a
different row space (Appendix~\ref{app:repro}).

\begin{explains}
The strata of \thmref{Theorem}{II.2} are the expressive quotient. The dynamics see a finer quotient, and how fine depends
on the optimizer (Table~\ref{tab:zero-b}). Undamped scaled GD is undefined at $B=0$, so scaled-GD optimizers enter in
damped ($\Ort(r)$) or AdamW-wrapped ($B_r$) form.
\end{explains}

\begin{table}[t]
\centering
\caption{What an update rule reads off $A_0$ from a zero-$B$ start (\thmref{Corollary}{III.13}).}
\label{tab:zero-b}
\small
\begin{tabularx}{\linewidth}{>{\raggedright\arraybackslash}X l >{\raggedright\arraybackslash}p{4.6cm}}
\toprule
\headrow \textbf{Update rule} & \textbf{Equivariance group} & \textbf{What it reads off $A_0$ at $B_0=0$}\\
\midrule
$\GL_r$-equivariant along the whole trajectory (the example rule; LoRA-RITE's Algorithm 1, checked numerically) & $\GL_r$ & $\operatorname{row}A_0\in\Gr(r,n)$\\
SGD, GD & $\Ort(r)$ & $A_0^\top A_0$\\
damped scaled GD & $\Ort(r)$ & $A_0^\top A_0$ (through a $\delta$-dependent function)\\
Adam, AdamW; scaled AdamW (Zhang and Pilanci; HF's Riemannian optimizer with AdamW); HF's LoRA-FA optimizer & $B_r$ & $A_0$ up to signed permutation of its rows\\
\bottomrule
\end{tabularx}
\end{table}

Two problems stay open. One is the apex fibre modulo $B_r$ for Adam, stratified by $\Phi_0$, degenerate pointings
included. The other is an Adam analogue of \thmref{Theorem}{III.9}, a conserved or monotone quantity for sign-like
updates.

Exposé~\ref{sec:lens} reads the same cometric from the backward side of the lens, where methods that compress gradients
have cometrics of their own.

\section{The lens: tangent and cotangent methods}\label{sec:lens}

Backpropagation pairs every map with its reverse derivative, so a method can shape what the model expresses on the
forward side or compress the gradient on the backward side. For linear methods and optimizers that read only their
gradients (no weight decay, no clipping of the full gradient), the two sides give the same weights, and GaLore becomes
one-sided LoRA with a moving frame and carried optimizer state.

\subsubsection{Two sides of one port}\label{sec:lens-hook}

A layer does two jobs in training. Forward, it turns parameters and inputs into outputs. Backward, it turns a cotangent
at the output into cotangents at the input and at the parameters. Backpropagation keeps the pair together, and a
category theorist calls it a \emph{lens} (\bgref{B.17}).

Every method in Exposés~\ref{sec:slice} to~\ref{sec:fibres} changes the forward job, replacing the parameter port
$\Theta$ by a smaller space $Q$ and a map $\rho:Q\to\Theta$. A second family edits only the backward job. GaLore
\citep{zhao2024galore} compresses the weight gradient $G$ to $P^\top G$, with $P$ from an SVD of $G$, runs Adam on that
small matrix and maps the result back with $P$. Every weight trains; only the optimizer state is small. LISA
\citep{pan2024lisa} and MeZO \citep{malladi2023fine} follow the same pattern.

When $\rho$ is linear the two families meet. The pullback of the forward method is then the compression of a backward
method, and an optimizer that reads only its gradients cannot tell the two apart. The equivalence between GaLore and
one-sided LoRA is due to \citet{xtorrobahennigen2025duality}, who prove a duality between linear gradient
transformations and adapters (their Thm~1 and Cor.~3, with weight decay in their App.~B). The frozen-$A$ case under
SGD, with random projections, appears earlier in \citet{hao2024flora}. Our additions are the lens bookkeeping, the
explicit hypotheses on the optimizer, and the table that sorts the cotangent family
(Table~\ref{tab:lens-schedule}).

\subsubsection{Backpropagation as a lens}\label{sec:lens-functor}

In the language of \thmref{Definition}{I.1}, a layer is a parametrised map $f:P\times X\to Y$. Its reverse derivative
$\mathsf R[f](x,y)=J_f(x)^\top y$ takes an output cotangent to an input cotangent, parameters included, and
backpropagation is the 2-functor
\begin{equation}\label{eq:lens-functor}
  \Para(\mathsf R):\Para(\mathcal C)\longrightarrow\Para(\Lens(\mathcal C)),\qquad f\longmapsto(f,\mathsf R[f]).
\end{equation}
The parametrised-map construction and the reading of backpropagation as a functor into lenses are due to
\citet{fong2017backprop} and \citet{cruttwell2021categorical} (\bgref{B.16}, \bgref{B.17}).

A reparametrisation $\rho$ is a 2-cell, so the functor turns it into a lens $(\rho,\mathsf R[\rho])$, and a
reparametrisation therefore acts on both sides of the parameter port. Its tangent side is the image, which is what the
method can \emph{express} (Exposé~\ref{sec:image}). Its cotangent side $\mathsf R[\rho](q,\gamma)=D\rho_q^\top\gamma$
pulls the weight gradient back to $Q$ and, with the rates $\Lambda$, induces the cometric $K=D\rho\,\Lambda\,D\rho^\top$
(\thmref{Definition}{III.1}, \bgref{D.23}), which is how the method \emph{moves}. A lens need not come from a map,
though, and a backward half written down directly gives the second family. The two kinds of lens on the parameter port
are
\[
\begin{tikzcd}[column sep=large,row sep=normal]
  Q \arrow[r,"\rho"] & \Theta \\
  T^*_qQ & T^*_{\rho(q)}\Theta \arrow[l,"D\rho_q^{\top}"']
\end{tikzcd}
\qquad\qquad
\begin{tikzcd}[column sep=large,row sep=normal]
  & \Theta \arrow[r,"\id_\Theta"] & \Theta \\
  T_\theta\Theta & S \arrow[l,"a_t"'] & T^*_\theta\Theta \arrow[l,"c_t"']
\end{tikzcd}
\]
on the left a forward method, whose backward half is forced by its forward half, and on the right a backward method,
whose forward half is the identity and whose backward half is chosen freely.

\begin{definition}{IV.1}{forward and backward methods}
\begin{itemize}
  \item A \textbf{forward} method acts on the parameter port by a lens in the image of $\mathsf R$, namely
  $(\rho,\mathsf R[\rho])$.
  \item A \textbf{backward}, or cotangent-side, method acts by a lens whose forward part is $\id_\Theta$ and whose
  backward part is a \emph{compression} $c_t:T^*_\theta\Theta\to S$ into a state space $S$, paired with an
  \emph{anchor} $a_t:S\to T_\theta\Theta$. The update is
  \begin{equation}\label{eq:lens-backward-update}
    \theta\leftarrow\theta-\eta\,a_t\big(\mathrm{opt}(c_t(\nabla L))\big).
  \end{equation}
\end{itemize}
Table~\ref{tab:lens-backward} lists examples.
\end{definition}

\begin{table}[t]
\centering\footnotesize
\caption{Backward methods as compression--anchor pairs (\thmref{Definition}{IV.1}). The last row is the only reading
of Flora as published in the form of \thmref{Definition}{IV.1}, the trivial lens with all compression inside
$\mathrm{opt}$.}\label{tab:lens-backward}
\begin{tabularx}{\linewidth}{@{}>{\raggedright\arraybackslash}p{0.27\linewidth}>{\raggedright\arraybackslash}X>{\raggedright\arraybackslash}p{0.24\linewidth}@{}}
\toprule
\headrow \textbf{Method} & \textbf{Compression $c_t$} & \textbf{Anchor $a_t$} \\
\midrule
GaLore \citep{zhao2024galore} & $P_t^\top$, with $P_t$ from a gradient SVD & $P_t$ \\
GaLore with a random projector \citep{xtorrobahennigen2025duality,hao2024flora} & as GaLore, with $P_t$ random & $P_t$ \\
LISA \citep{pan2024lisa} & layer-block coordinate projection & the same projection \\
MeZO \citep{malladi2023fine} & $g\mapsto\langle z_t,g\rangle$, estimated by finite differences without a backward pass &
$\lambda\mapsto\lambda z_t$ \\
\midrule
Flora as published \citep{hao2024flora} & identity: only the momentum or accumulation buffer is compressed, by a random
$A_t$, inside $\mathrm{opt}$ & identity (full-space Adafactor step) \\
\bottomrule
\end{tabularx}
\end{table}

A backward method has its own cometric $K_t=a_tc_t$ (\thmref{Definition}{III.1}), for instance $K(G)=PP^\top G$ for
GaLore under SGD. The update must lie in the image of the anchor. Read with their projections as anchors, Flora as
published \citep{hao2024flora}, APOLLO \citep{zhu2024apollo} and Fira \citep{chen2024fira} make updates that leave
$\operatorname{im}a$, so they are not backward methods in the sense of \thmref{Definition}{IV.1} beyond the trivial
reading with compression and anchor the identity, which the last row of Table~\ref{tab:lens-backward} records for
Flora. Flora compresses its momentum or gradient accumulator with a
random projection and hands the decompressed tensor to Adafactor \citep{shazeer2018adafactor}, which acts in the full
space, so its update need not lie in the image of the projection. On resampling it transports its momentum to the new
frame. It fits \thmref{Definition}{IV.1} with a nontrivial anchor only in an idealised form with an SGD or heavy-ball
base.

\subsubsection{The degree-one collapse}\label{sec:lens-collapse}

If $\rho(C)=W_k+\mathcal LC$, training $C$ and running the backward method with $c=\mathcal L^*$, $a=\mathcal L$ feed
the optimizer the same gradients $\mathcal L^*G$ and so make the same weight updates. A curved $\rho$, such as LoRA's
$BA$, separates them after one step, and so does an optimizer that reads the parameter tensor, which is $C$ in one
procedure and $W$ in the other.

\begin{proposition}{IV.2}{the degree-one collapse; known, \citet{xtorrobahennigen2025duality} for GaLore (their Thm~1
and Cor.~3, App.~B for weight decay) and \citet{hao2024flora} for random projections}
Let $\rho_k(C)=W_k+\mathcal LC$ with $\mathcal L$ linear and $C_0=0$. Let $\mathrm{opt}$ be a first-order optimizer,
with no coupled or decoupled weight decay, whose state and output depend only on the sequence of gradients it is fed,
and which does not read the parameter tensor: SGD, heavy-ball or Nesterov momentum, Adam or AMSGrad, or Adafactor with
\texttt{scale\_\allowbreak parameter=False} (fixed or relative step, with its factorisation applied to the compressed gradient in
the same tensor shape as $C$). Run both procedures below with the same implementation of $\mathrm{opt}$, on the same
minibatches, and from the same optimizer state, fresh or carried over. Then they produce the same $W$-trajectory over a
period:
\begin{itemize}
  \item training $C$ with $\mathrm{opt}$ and merging at the end of the period;
  \item the backward method with $c=\mathcal L^*$ and $a=\mathcal L$, started at $W_k$.
\end{itemize}
AdamW with $\lambda>0$, LAMB and LARS trust ratios, parameter-scaled Adafactor (\texttt{scale\_\allowbreak parameter=True}, the
default in HF and in \texttt{galore\_torch}'s \texttt{GaLoreAdafactor}) and global-norm clipping of the full gradient
(such as HF Trainer's default \texttt{max\_grad\_norm=1.0}, whenever it triggers) all break the identity. Injectivity
of $\mathcal L$ is not needed.

In particular, on each matrix that GaLore projects (its other parameters train in full in both procedures), GaLore with
a projector $P_k$ fixed on each period equals one-sided LoRA $W_k+P_kC$ (frozen $B=P_k$, or frozen $A$ when projecting
on the right), merged and re-based every period, with $C$'s optimizer state (moments and step counter) carried over
unchanged across re-basings, with no reset. $P_k$ must be exactly GaLore's matrix, signs and column order included; its
span alone does not suffice, because the state is reused without transport. ReLoRA-style resets give a different
trajectory. The identity is exact with weight decay $0$, no full-gradient clipping, and an optimizer from the admissible
class; \texttt{galore\_torch}'s \texttt{GaLoreAdafactor} defaults to \texttt{scale\_\allowbreak parameter=True} and is excluded.
\texttt{GaLoreAdamW} is the transformers-style AdamW, which adds $\varepsilon$ to $\sqrt v$ before the bias
correction, so the LoRA side must use the same implementation. GaLore's scale $\alpha$ multiplies the optimizer output,
so it acts as a learning-rate factor; a LoRA-style scale inside $\rho$ gives a different trajectory under Adam with
$\varepsilon>0$.
\end{proposition}

\begin{remark*}{scale and implementation}
Moving GaLore's $\alpha$ inside $\rho$ as a LoRA scale multiplies the compressed gradient by $\alpha$. Under Adam this
gives GaLore's trajectory only at $\varepsilon=0$, or with the LoRA side's $\varepsilon$ multiplied by $\alpha$ as well.
This is the exchange of scale for learning rate and $\varepsilon$ on the hyperparameter torus of
\thmref{Theorem}{III.10}. Both sides must also run the same implementation: \texttt{galore\_torch}'s AdamW adds
$\varepsilon$ to $\sqrt v$ before the bias correction, as in transformers, and swapping in \texttt{torch.optim.Adam} on
the LoRA side moves the trajectory measurably (see the check below).
\end{remark*}

\noindent\labelsf{Num}\enspace In a toy check under Adam the two procedures agree to $7\times10^{-16}$ over one
period, and to $5\times10^{-15}$ over four periods with the state carried on both sides. Against the official
\texttt{galore\_torch} optimizers (least squares, $W\in\R^{6\times10}$, $r=2$, $T=7$, four periods, total movement
$0.046$), the two procedures agree to $8\times10^{-17}$ with the state carried over. Resetting the LoRA side's state
separates them by $0.029$, weight decay $0.1$ by $0.020$, HF-default Adafactor by $0.030$ (against $6\times10^{-17}$
with \texttt{scale\_\allowbreak parameter=False}), and flipping the sign of one column of $P_k$, with the state carried, by $0.12$.
Using \texttt{torch.optim.Adam} on the LoRA side moves the trajectory by about $4\times10^{-5}$, and placing $\alpha$
inside $\rho$ by $0.019$ at $\varepsilon=10^{-6}$, with exact agreement at $\varepsilon=0$. The companion site reruns a
check of this kind live (Appendix~\ref{app:repro}).

\begin{explains}
Beyond one-sided LoRA, GaLore adds two ingredients: a frame read off the gradient's SVD, and the carry-over of optimizer
state across re-basings. The carry-over involves no change of basis. Moments written in the coordinates of $P_{k-1}$
are read as coordinates for $P_k$. Projection-aware state updates, such as LDAdam \citep{xrobert2025ldadam}, and Flora
transport the state instead. The equivalence classifies the cotangent methods whose optimizer satisfies its hypotheses.
Table~\ref{tab:lens-schedule} sorts them by where the projector comes from and how often it changes; its LoRA and
ReLoRA row is a contrast row outside the equivalence.
\end{explains}

\begin{table}[t]
\centering\footnotesize
\caption{The cotangent family by projector (rows) and schedule (columns). \thmref{Proposition}{IV.2} covers the first
three rows under its optimizer hypotheses, except EVA and LoRA-GA, which initialise two-sided LoRA and sit in the frozen
column only for their first step. The last row is a contrast row outside the equivalence.}\label{tab:lens-schedule}
\begin{tabularx}{\linewidth}{@{}>{\raggedright\arraybackslash}p{0.19\linewidth}>{\raggedright\arraybackslash}X>{\raggedright\arraybackslash}X>{\raggedright\arraybackslash}X@{}}
\toprule
\headrow \textbf{Projector / schedule} & \textbf{Frozen} & \textbf{Re-based periodically} & \textbf{Resampled every step} \\
\midrule
random & LoRA-FA \citep{zhang2023lora}, with $K\propto GA_0^\top A_0$ under SGD$^{a}$ & GaLore with a random projector
\citep{xtorrobahennigen2025duality,hao2024flora}; idealised Flora$^{b}$ & MeZO \citep{malladi2023fine} (rank $1$,
global), up to its finite-difference estimator \\
\addlinespace
gradient or data SVD & EVA \citep{paischer2024parameter}, LoRA-GA \citep{wang2024lorab} (first step only) & GaLore
\citep{zhao2024galore} & GaLore with $T=1$ \\
\addlinespace
coordinate blocks & BitFit \citep{benzaken2021bitfit}, FISH \citep{sung2021training}, LN-tuning
\citep{zhao2023tuning} & LISA \citep{pan2024lisa} & block coordinate descent \\
\addlinespace
two-sided, state-dependent (contrast) & LoRA \citep{hu2021lora} & ReLoRA \citep{lialin2023relora} & none in use \\
\bottomrule
\end{tabularx}
\par\smallskip
\begin{minipage}{\linewidth}\scriptsize
$^{a}$\,For the original method under SGD. HF PEFT's LoRA-FA optimizer preconditions by $(A_0A_0^\top+\delta I)^{-1}$,
which in the undamped limit gives the row-space projector $GP_V$ of \thmref{Proposition}{III.4}(c); its Adam layer
makes the update nonlinear in $G$.\quad
$^{b}$\,Flora's Alg.~2 with an SGD or heavy-ball base, with the momentum transported at resampling instead of carried
over; Flora as published, with an Adafactor base, is outside \thmref{Definition}{IV.1} and outside
\thmref{Proposition}{IV.2}.
\end{minipage}
\end{table}

\thmref{Proposition}{IV.2} covers the first three rows of Table~\ref{tab:lens-schedule} under its optimizer hypotheses,
except EVA and LoRA-GA, which initialise two-sided LoRA and sit in the frozen column only for their first step; it
covers MeZO up to the finite-difference error of its estimator. LoRA's map $(B,A)\mapsto W_0+sBA$ has degree
two, so its cometric moves with the state (\thmref{Observation}{III.2}) and the collapse does not apply. The other
frozen entries gain nothing from merge-and-restart (\thmref{Theorem}{V.3}, corollary~2); their re-based neighbours move
the frame and can reach all of weight space. The LoRA-FA cometric holds for the original method of
\citet{zhang2023lora} under SGD; the LoRA-FA optimizer of HF PEFT \citep{xmangrulkar2022peft} gives the row-space projector of
\thmref{Proposition}{III.4}(c), as noted under the table.

Flora proper is not in the table. Its Adafactor step acts on the full gradient, with parameter scaling by default, so
its steps are not confined to $\operatorname{im}P_t$ and \thmref{Proposition}{IV.2} does not apply; like LDAdam, it
transports its optimizer state. Only the idealised form with an SGD or heavy-ball base sits in the random, re-based
cell, and there its momentum is transported rather than carried.

\subsubsection{When a backward method is a forward one}\label{sec:lens-integrability}

A backward method chooses at each weight a space of directions $D_\theta=\operatorname{im}a(\theta)$. It reads as a
forward method only if these directions fit together into $k$-dimensional surfaces, the images of immersive
parametrisations (\bgref{C.4}). Frobenius' theorem tests this (\bgref{D.25}, \auxref{B.31}). The bracket $[X,Y]$, the gap by which a
small loop along $X$, $Y$, $-X$, $-Y$ fails to close, must stay in $D$. Discrete steps and the optimizer's metric ask
for more.

\begin{proposition}{IV.3}{integrability; after Frobenius' theorem}
Let $D$ be a smooth distribution of constant rank $k$ on an open set $U\subseteq\Theta$, for instance the anchor
distribution $D_\theta=\operatorname{im}a(\theta)$ of a time-independent backward method. Then $D$ is involutive on
$U$ if and only if through every $\theta\in U$ passes an injective immersion $\rho_\theta:(\R^k,0)\to(U,\theta)$ with
$\operatorname{im}D\rho_\theta=D$ along its image. A single such $\rho$ through $\theta_0$ gives involutivity only
along $\operatorname{im}\rho$. For example, $D=\spanop\{\partial_x,\ \partial_y+xz\,\partial_z\}$ on $\R^3$ has
$[\partial_x,\partial_y+xz\,\partial_z]=z\,\partial_z\notin D$ off $z=0$, yet the plane $z=0$ is an integral manifold.

The dynamic content needs separate hypotheses.
\begin{itemize}
  \item In continuous time, if $D$ is involutive, a backward flow with $\dot\theta\in D_\theta$ stays on the leaf
  through $\theta_0$.
  \item In the discrete update \eqref{eq:lens-backward-update} of \thmref{Definition}{IV.1}, the iterates stay on a
  leaf for all step sizes and all optimizer outputs $s$ (all losses) if and only if that leaf contains
  $(\theta+D_\theta)\cap U$ at every iterate, that is, if and only if the leaves are affine; an anchor constant over
  the period gives this (\thmref{Proposition}{IV.2}).
  \item Under gradient flow, dynamic equivalence with a forward lens $(\rho,\mathsf R[\rho])$ forces matching frames,
  $a_tc_t=D\rho\,\Lambda\,D\rho^\top$ along the trajectory. Hence $K=ac$ is symmetric positive semidefinite and, for
  injective $a$, $c=Ma^\top$ with $M$ symmetric positive semidefinite on $S$, possibly depending on $\theta$. If in
  addition $\rho$ is an immersion with $\dim Q=\rk D$, then $\rho$ is a local isometry from $(Q,\Lambda^{-1})$ onto the
  leaf with metric $(aMa^\top)^{-1}|_D$, so that metric must be flat. Forward methods with gauge fibres escape this
  requirement: $\rho(B)=BB^\top$ on symmetric matrices induces the curved Bures--Wasserstein metric, and balanced
  full-rank LoRA also induces a curved leaf metric. For discrete steps, exact agreement for every step size further
  forces $\rho$ to be affine along each horizontal line $q+t\,\Lambda D\rho_q^\top g$.
\end{itemize}

\emph{Instances.} Every projector-type backward method of Table~\ref{tab:lens-backward} (GaLore with any period $T$
and any projector, its random-projector variant, LISA, MeZO) has an anchor that does not depend on $\theta$ within its
period. Its distribution is constant, hence trivially involutive with affine leaves (vacuously at $T=1$, where a period
is a single step). Under the optimizer hypotheses of \thmref{Proposition}{IV.2} (no weight decay of either kind, no
full-gradient clipping; GaLore and MeZO need weight decay $0$, and LISA's AdamW with positive decay violates them),
these methods are rebasing schemes of linear forward methods with period $T$, including $T=1$, whose reach
\thmref{Theorem}{V.3} describes. Each MeZO step is a step of the linear method $\theta_t+\lambda z_t$, up to its SPSA
estimator, and GaLore with $T=1$ is one-sided LoRA re-based every step. What separates these methods from LoRA-type
methods is the resampling schedule together with the carry-over of optimizer state; integrability plays no part in it.
Flora as published (Adafactor base, with a full-rank current-gradient term in the code), APOLLO (a
low-rank-estimated scaling applied to the full gradient) and Fira (a scaled full-rank residual) are not of the form of
\thmref{Definition}{IV.1} with their projections as anchors, because their updates leave $\operatorname{im}a$.

Non-involutivity proper needs a $\theta$-dependent anchor. Full-batch GaLore with $T=1$, read as the distribution
$D_W=\operatorname{Hom}(\R^n,U_r(\nabla L(W)))$ (or $\operatorname{Hom}(V_r(\nabla L(W)),\R^m)$ when $m\ge n$, as
GaLore projects) wherever a singular-value gap exists, is involutive if and only if $U_r(\nabla L(W))$ (respectively
$V_r(\nabla L(W))$) is invariant along $D$. That fails generically for every $r\ge1$ once both matrix dimensions are at
least $2$. For every $r$ with $1\le r<\min(m,n)$ it fails for $L=\tfrac12\lVert W\rVert_F^2$, and numerically it fails
for least squares and cross-entropy with several samples. It holds, for instance, for single-sample least squares, and
automatically only when the distribution is a line field ($nr=1$). Minibatch GaLore with $T=1$ and MeZO violate
time-independence, because their anchors depend on step-wise randomness.
\end{proposition}

\noindent\labelsf{Num}\enspace For full-batch GaLore with $T=1$, the relative normal part of the bracket is $0.08$ to
$0.94$ at $r=1$ and $(m,n)\in\{(4,3),(5,4)\}$, for least-squares, tanh-regression and cross-entropy losses with several
samples (an independent replication), and $0.21$ to $0.95$ at $(m,n,r)=(4,3,2)$. It vanishes only for $n=1$ or for
single-sample least squares. At $W=\diag(2,1)$ with $L=\tfrac12\lVert W\rVert_F^2$ and the frame $X_i=u(W)e_i^\top$,
the bracket is $[X_1,X_2]=-\tfrac13e_2e_1^\top\notin D$, and the general diagonal formula behind it was checked by
central differences for several $(m,n,r)$. The cometric of $B\mapsto BB^\top$ takes the same value at $B$ and at $BR$,
$R\in\Ort(2)$, to $10^{-12}$. The companion site recomputes the bracket live at $W\in\R^{3\times4}$
(Appendix~\ref{app:repro}).

\begin{explains}
The lens 2-functor \eqref{eq:lens-functor} separates tangent-side from cotangent-side methods. Frobenius integrability
is the test for when a $\theta$-dependent anchor foliates weight space by images of forward methods. For the PEFT zoo it
does no discriminating work. Every cotangent method whose update lies in $\operatorname{im}a$, whose optimizer does not
read the parameter, and which has no decoupled decay on $W$ and no clipping of the full gradient (GaLore with any
projector and MeZO at weight decay $0$, LISA without decay) is, period by period, a forward linear method on a moving
base (Exposé~\ref{sec:base}). At $T=1$ this reading is automatic.
\end{explains}

\begin{analogy*}{the lens duality}
``Tangent side'' and ``cotangent side'' are exact words here. A forward method acts on tangent vectors through
$D\rho$, and a backward method acts on cotangent vectors through $c$. Calling GaLore and one-sided LoRA \emph{dual} is
exact only at degree one, and only for optimizers that do not read the parameter tensor, with no weight decay. Beyond
that range the word organises the zoo and proves nothing.
\end{analogy*}

\subsubsection{What the backward pass must remember}\label{sec:lens-cone}

Fine-tuning memory consists of weights, gradients and optimizer state, and the activations stored for the backward
pass. Shrinking $Q$ shrinks the gradients and the optimizer state. As for the activations, a frozen linear box's reverse
derivative $W^\top y$ ignores its input, and cotangents are needed only downstream of something trainable, so only a
cone through the network costs activation memory.

\begin{observation}{IV.4}{backward reach and the backprop cone; an upper bound}
Assume reverse mode in which each box keeps its own residual, with no checkpointing across boxes, and take boxes at the
granularity where adapters are separate boxes (for LoRA-FA, $A_0$ and $B$ separately). A \emph{VJP residual} is an
input, an output, a sufficient statistic such as a ReLU mask, or a recipe to recompute one of these inside the box.
Then it suffices to keep a VJP residual for
\begin{itemize}
  \item each trainable box whose parameter-VJP depends on its input: linear and multiplicative boxes such as $W$, $A$,
  $B$ and (IA)$^3$ scales; additive boxes such as biases and prompts need nothing;
  \item each \emph{nonlinear} box, bilinear boxes such as the attention products included, on a directed path from a
  trainable box to the loss.
\end{itemize}
Deterministic frozen linear boxes, and all boxes off this cone, need nothing; stochastic linear boxes on the cone
(dropout) need their mask or random-number state. Alternatively, the $k$ coordinates of $P^\top\nabla L$ can be
computed exactly by $k$ forward-mode JVPs, with no reverse pass and $O(1)$-layer activation memory, each JVP costing a
small constant multiple (about $1.5$ to $3$) of a forward pass. This always uses less activation memory than
un-checkpointed reverse mode, but it is time-competitive with one VJP (about $3$ forward passes) only for $k$ of about
$1$ or $2$. It also needs $P$ before the passes, so it cannot produce GaLore's SVD refresh, where $P$ depends on
$\nabla L$, more cheaply than a backward pass ($\nabla L$ itself would take $\lvert\theta\rvert$ JVPs).
\end{observation}

\begin{explains}
The bound reads off the activation memory of the zoo.
\begin{itemize}
  \item \textbf{LST} \citep{sung2022lst}. The side network reads the backbone through trainable down-projection
  ladders, so the backbone's backward pass never runs.
  \item \textbf{LoRA-FA} \citep{zhang2023lora}. With \texttt{lora\_dropout=0} (the HF default) it stores $A_0x$ ($r$
  numbers per token) instead of $x$ ($n$ numbers); with dropout on, the $n$-element mask is stored too. The saving is
  realised only where no neighbouring nonlinear box already keeps $x$.
  \item \textbf{Top-layer adaptation} truncates the cone, under the no-checkpointing assumption. With HF's reentrant
  gradient checkpointing, PEFT calls \texttt{enable\_input\_require\_grads()}, which makes the embedding outputs require
  gradients, so cotangents run down to the embeddings and top-layer adaptation no longer truncates the cone.
  \item \textbf{LISA} \citep{pan2024lisa} trains the embedding and the LM head at every step, so its cone is the whole
  network and it gets no truncation. On the cone, the unsampled layers' frozen linear boxes drop their input residuals
  (about $20\%$ of a block's activations at LLaMA proportions without FlashAttention), and relative to LoRA it also
  drops the adapter residuals, in line with the LISA paper's report of less activation memory than LoRA. Most of its
  saving against full fine-tuning is in gradients and optimizer state. A variant that froze the embedding would
  truncate the cone at its lowest sampled layer.
  \item \textbf{Prompt tuning} \citep{lester2021power} has a $Q$ as small as that of VeRA \citep{kopiczko2023vera},
  FourierFT \citep{gao2024parameter} or (IA)$^3$ \citep{liu2022few}, and the maximal cone, the whole network, which it
  shares with every method that trains something in the first layer. Its prompt tokens also enlarge every stored
  activation.
  \item \textbf{MeZO} \citep{malladi2023fine} is the case $k=1$, with the JVP estimated by a central difference and no
  reverse pass. Its estimator is unbiased for the gradient of the Gaussian-smoothed loss and, for a loss with Lipschitz
  Hessian, has bias $O(\varepsilon^2)$ against $\nabla L$; ReLU networks lack a Lipschitz Hessian. Exact unbiasedness,
  $\mathbb E[(\nabla L\cdot z)z]=\nabla L$, holds for the forward gradient computed with an exact JVP. GaLore, its
  random-projector variant, Flora and LISA all take the reverse route.
\end{itemize}
\end{explains}

Ladder side-tuning is due to \citet{sung2022lst}, prompt tuning to \citet{lester2021power}, and the zeroth-order
estimator in this setting to \citet{malladi2023fine}. That reverse derivatives of linear maps ignore their inputs is the
observation behind many activation-saving designs; the lens of \citet{cruttwell2021categorical} makes it visible box by
box.

So every cotangent method meeting the hypotheses of \thmref{Proposition}{IV.2} is, period by period, a linear forward
method on a moving base. What it reaches, and what a quantiser does to the base, are the subject of
Exposé~\ref{sec:base}.

\section{Base change}\label{sec:base}

\begin{flushright}
\small\itshape Study each object over a base, so that the base can be changed.\\
What survives the change is what the object really is.\\[2pt]
{\upshape\footnotesize after Grothendieck}
\end{flushright}

Quantising and restarting move the point a method is attached to, and combining adapters builds one update from
several. Each move comes down to one computation, of the defect a quantiser leaves, the monoid one cycle generates, or the
gauge a combination rule must ignore.

\paragraph{In brief}
\begin{itemize}
  \item QLoRA is LoRA anchored at the quantised weight $\kappa\theta_0$, so it misses $\theta_0$ by the quantisation
    error $e$, which is generically of full rank. LoftQ and QPiSSA are built to shrink this defect, by quantising and
    splitting in opposite orders. HF PEFT's LoftQ initialiser ignores the LoRA scale $s=\alpha/r$, so with its default single
    iteration it beats QLoRA in Frobenius norm only for $0<s<2$.
  \item With a fixed shape, merge-and-restart reaches exactly the monoid one cycle generates. ReLoRA reaches full rank
    after $\lceil\min(m,n)/r\rceil$ cycles, while fixed linear shapes such as BitFit, (IA)$^3$ or a fixed mask gain
    nothing.
  \item Exactly orthogonal block OFT on a fixed partition never leaves $W_0\cdot\prod\SO(b)$, and re-merged BOFT
    reaches all rotations only at full butterfly depth. HF's default Cayley--Neumann OFT is not orthogonal, and with
    small generators each merge can only shrink neuron norms and singular values.
  \item A rule combining LoRA factors is well defined only if it ignores each adapter's $\GL_r$ gauge. Task arithmetic
    on the products passes. LoraHub and HF's factor-wise \texttt{linear} merge fail through cross terms, and become
    exact only when the adapters share a factor.
\end{itemize}

\subsubsection{The point moves}\label{sec:base-hook}

Every method so far has been an arrow into one fixed point $(\Theta,\theta_0)$. In practice the point moves.
Quantisation moves it once, ReLoRA \citep{lialin2023relora} at every merge, and task arithmetic
\citep{ilharco2022editing}, LoraHub \citep{huang2023lorahub}, TIES \citep{yadav2023ties} and DARE
\citep{yu2023language} build a new update from several adapters. A PEFT method is a formula that makes sense over any
base point, so it is a family $\theta_0\mapsto M(\theta_0)$, and this exposé relates its members. Four kinds of move
occur (Table~\ref{tab:base-moves}).

\begin{table}[t]
\centering\footnotesize
\caption{Four ways the base point moves, and what each one asks us to compute.}\label{tab:base-moves}
\begin{tabularx}{\linewidth}{@{}>{\raggedright\arraybackslash}p{0.2\linewidth}>{\raggedright\arraybackslash}X>{\raggedright\arraybackslash}X>{\raggedright\arraybackslash}p{0.17\linewidth}@{}}
\toprule
\headrow \textbf{Move} & \textbf{Examples} & \textbf{What to compute} & \textbf{Where} \\
\midrule
a reparametrisation that keeps the function & permuting hidden units; SliceGPT-style rotations of the residual
stream \citep{xashkboos2024slicegpt} & nothing new; methods are carried along by composition &
\thmref{Definition}{V.1}(a), \thmref{Observation}{II.15} \\
\addlinespace
a quantiser & QLoRA, LoftQ, QPiSSA, QA-LoRA & the pointing defect, and the invariants that move with the base &
\thmref{Proposition}{V.2} \\
\addlinespace
merge and restart & ReLoRA, re-merged OFT and BOFT, re-HRA; GaLore period by period & the monoid generated by one
cycle & \thmref{Theorem}{V.3} \\
\addlinespace
combining adapters & task arithmetic, LoraHub, TIES, DARE, KnOTS & whether the rule is a function of the adapters, and
which symmetries it commutes with & \thmref{Proposition}{V.4} \\
\bottomrule
\end{tabularx}
\end{table}

\begin{definition}{V.1}{base change and rebasing}
\begin{itemize}
  \item[(a)] A \emph{function-preserving} pointed reparametrisation $\phi:(\Theta',f',\theta_0')\to(\Theta,f,\theta_0)$,
  with $f\circ(\phi\times\id_X)=f'$, induces by postcomposition
  \[\phi_!:\Meth(\Theta',\theta_0')\to\Meth(\Theta,\theta_0),\qquad(Q,q_0,\rho)\mapsto(Q,q_0,\phi\circ\rho).\]
  \item[(b)] A map $\kappa:\Theta\to\Theta$ that does \emph{not} preserve the function, such as a quantiser, moves the
  base point. Its \textbf{pointing defect} is $e=\kappa\theta_0-\theta_0$.
  \item[(c)] A \textbf{rebasing scheme} (merge-and-restart) picks $\theta_{k+1}\in\Im M(\theta_k)$. Its reachable sets
  are $R_K$, the points $\theta_K$ it can reach in $K$ cycles, and $R_\infty=\bigcup_KR_K$.
\end{itemize}
\end{definition}

Only part (a) is base change in the algebraic geometer's sense (\bgref{A.19}), and it costs nothing; whether $\phi_!$
turns LoRA into LoRA is the covariance question of \thmref{Observation}{II.15}. Parts (b) and (c) borrow the word, and
the work lies there and in the combination rules. Figure~\ref{fig:rebase} shows the restart computations of
\thmref{Theorem}{V.3}.

\begin{remark*}{what ``base change'' means here}
In algebraic geometry a family over a base $S$ pulls back along any map $S'\to S$, and the statements worth having are
the ones that survive the change. Part (a) of \thmref{Definition}{V.1} is the standard companion of pullback on slice
categories, composition with a map of bases. A quantiser changes the network's function, so there is no commuting
square to transport along, and we compute the defect it leaves instead. A rebasing scheme is an iteration, and we
compute its reachable set. ``Gauge'' keeps the sense of Exposé~\ref{sec:fibres}, the symmetry group of a
parametrisation (\bgref{D.21}); no connection is involved.
\end{remark*}

\subsubsection{Quantisation and splitting}\label{sec:base-quant}

For an additive method (\thmref{Definition}{I.2}) whose starting point does not depend on the base, such as zero-init
LoRA \citep{hu2021lora}, quantising the base only translates the image, so QLoRA \citep{dettmers2023qlora} is LoRA
with its vertex at $\kappa\theta_0$. A quantisation error is spread over all singular directions, so $e$ is generically
of full rank and the image misses $\theta_0$. Methods whose starting point is computed from the base, such as PiSSA
\citep{meng2024pissa}, CorDA \citep{yang2024corda}, OLoRA \citep{buyukakyuz2024olora}, LoftQ \citep{li2023loftq} and
DoRA \citep{liu2024dora}, meet two operations that do not commute, and the order fixes the defect (tabulated in
Table~\ref{tab:defects} of Exposé~\ref{sec:strata}).

\begin{remark*}{2026: SRR}
SRR \citep{cho2026preserve} joins the first two orders of \thmref{Proposition}{V.2}(b) by splitting the rank budget.
It puts $\operatorname{SVD}_k(W)$ into the adapter, quantises $W_{\rm res}=W-\operatorname{SVD}_k(W)$ and fits the
quantisation error with the other $r-k$ ranks. Its defect is therefore $e_k$ minus its best rank-$(r-k)$
approximation, where $e_k=\kappa(W_{\rm res})-W_{\rm res}$, and $\operatorname{SVD}_k$ and best approximations are taken
in its activation-scaled norm. With that scaling set to the identity, $k=r$ gives the split-first defect of QPiSSA and
$k=0$ that of one-step LoftQ at $s=1$. SRR picks $k$ per matrix from a closed-form proxy.
\end{remark*}

\begin{proposition}{V.2}{quantisation and splitting}
Write $W$ for the full-precision weight, $\kappa$ for the quantiser, $e=\kappa W-W$, $\delta_0=\delta(q_0)$, and $s$
for the LoRA scale $\alpha/r$ ($\alpha/\sqrt r$ under rsLoRA, \citealp{kalajdzievski2023rank}).

(a) Additive families commute with translation, $M(\theta+v)=\tau_v\circ M(\theta)$. Hence QLoRA is LoRA pointed at
$\kappa\theta_0$, with defect $e$, and it needs no new weight-space structure. Since $-e$ is generically of full rank,
$\theta_0\notin\Im\mathrm{LoRA}_r(\kappa\theta_0)$ unless $\rk e\le r$.

(b) For splitting families derived from an SVD or from data, the order of operations matters, and three orders occur
in practice.
\begin{itemize}
  \item \emph{Split, then quantise the residual} (QPiSSA; HF's recommended PiSSA and CorDA workflow; the quantisation
  half-step of LoftQ for $t\ge2$, with $W_{\rm res}=W-\delta(q_{t-1})$). The defect is $\kappa(W_{\rm res})-W_{\rm res}$.
  \item \emph{Quantise, then fit the error} (LoftQ with one iteration, HF's default \texttt{loftq\_iter=1}; HF
  \texttt{replace\_\allowbreak lora\_\allowbreak weights\_\allowbreak loftq}). HF sets the factors from
  $\operatorname{SVD}_r(W-\kappa W)$ without dividing by $s$, unlike its PiSSA and OLoRA initialisations, so
  $\delta_0=-s\operatorname{SVD}_r(e)$ and the defect is $e-s\operatorname{SVD}_r(e)$, with
  \[\lVert e-s\operatorname{SVD}_r(e)\rVert^2=\lVert e-\operatorname{SVD}_r(e)\rVert^2+(1-s)^2\lVert\operatorname{SVD}_r(e)\rVert^2.\]
  The defect is $e$ minus its best rank-$r$ approximation (of rank $\rk e-r$) iff $s=1$, which is HF's bare default
  $r=\alpha=8$. For $s=2$ its norm equals $\lVert e\rVert$, no better than QLoRA. With a callback,
  \texttt{replace\_\allowbreak lora\_\allowbreak weights\_\allowbreak loftq} can roll a module back to the zero
  initialisation, whose defect is $e$.
  \item \emph{Quantise, split, re-quantise} (HF \texttt{olora\_init} on bitsandbytes weights). The defect is
  $e+\kappa(r')-r'$ with $r'=\kappa W-\delta_0$.
\end{itemize}
Quantise-then-split has defect exactly $e$ iff the frozen part equals $\kappa W-\delta_0$, for instance $\kappa W$ kept
quantised minus a frozen copy of the rank-$r$ $\delta_0$, at a cost of about $r(m+n)$ extra numbers. Re-quantising the
residual, as HF \texttt{olora\_init} does, generically adds $\kappa(r')-r'$.

Action families whose pointing does not depend on the base (OFT at $R=I$, BOFT at $R=I$ with unit scale, HiRA at
$B=0$), evaluated at $\kappa\theta_0$, simply carry the defect $e$. Methods whose pointing is computed from the base
have an order ambiguity of their own. DoRA's magnitude initialised from $\kappa\theta_0$ gives defect $e$, while
initialised from $\theta_0$ it gives
\[\diag\big(\lVert w_{0,i}\rVert/\lVert(\kappa\theta_0)_i\rVert\big)\,\kappa\theta_0-\theta_0.\]
For action families a base change also moves the absolute invariants, along with the base point. NF4's exact zero code
enlarges the zero set that HiRA must preserve, and exactly orthogonal (exact-Cayley) OFT preserves the Gram matrix of
$\kappa\theta_0$ rather than that of $\theta_0$; HF's default Cayley--Neumann OFT preserves it only approximately.

(c) LoftQ alternates two updates on $\lVert\theta_0-Q-LR\rVert$, an exact rank-$r$ SVD step and a round-to-nearest
quantisation step with blockwise scales recomputed from its input. The second is not an exact projection, so the
objective need not decrease monotonically. With $s=1$ the output is pointed up to the attained residual. It is then an
\emph{approximate section}, a choice $(Q,q_0)$ with reduced $\lVert\rho(q_0)-\theta_0\rVert$. In general, with $Q_T$
and the low-rank fit $L_TR_T$ after $T$ iterations, HF's deployed defect is
\[(Q_T+L_TR_T-W)+(s-1)L_TR_T,\]
because HF's internal residual also omits $s$. The official LoftQ checkpoints use $r=64$ and $\alpha=16$, so $s=1/4$,
and then the deployed defect can grow with $T$ while the attained residual falls.

(d) For a generic high-precision adapter $\Delta$, fine-tune-then-quantise and quantise-then-fine-tune do not commute.
The first lands on the quantisation grid. In the second, the exactly merged weight $\kappa\theta_0+\Delta$ leaves the
quantised type. For a uniform fixed-scale INT-$k$ grid, $\kappa\theta_0+\Delta$ stays on the grid iff $\Delta$ lies on
the scale lattice (within range), so there the obstruction is the off-grid update. NF4 is non-uniform and not closed
under addition, and absmax grids whose scales are recomputed are not closed either, so for QLoRA's NF4 the grid itself is
part of the obstruction. Type-preserving merges, such as HF PEFT's dequantise--add--requantise for bitsandbytes 4- and
8-bit layers, re-quantise with recomputed block scales, so they stay in the quantised type but are lossy. Under affine
group quantisation QA-LoRA is the structured exception. Its group-pooled adapter is constant on each quantisation group
and merges into the zero-points, so the merged model stays INT-$k$.
\end{proposition}

\paragraph{Numerical check} On a $256\times256$ NF4 example whose weight has a dominant low-rank part, with $r=8$,
$s=1$ and $\lVert e\rVert=5.65$, QPiSSA's defect is $1.45$, one-step LoftQ's is $5.22$, and the LoftQ objective rises
in $27$ of $60$ half-steps. QPiSSA's large advantage there relies on the dominant low-rank part; on a Gaussian weight
all three defects are close to $\lVert e\rVert$. With $s=2$ the one-step LoftQ defect equals $\lVert e\rVert$ exactly,
and adding an NF4-representable update to an NF4 weight leaves the NF4 grid in most entries
(Appendix~\ref{app:repro}).

\paragraph{LoftQ as deployed} The official LoftQ checkpoints were made with $r=64$, $\alpha=16$ and five iterations,
so $s=1/4$. In a referee's simulation of that configuration on a Gaussian $512\times512$ NF4 weight, the attained
residual was $0.67\lVert e\rVert$ and the deployed defect $0.96\lVert e\rVert$. For $T\ge2$, HF's internal residual also
omits $s$, so the objective in \thmref{Proposition}{V.2}(c) is not the defect HF starts from. With a callback,
\texttt{replace\_\allowbreak lora\_\allowbreak weights\_\allowbreak loftq} can roll a module back to the zero
initialisation, and that module's defect is then $e$.

\paragraph{Credit} Training LoRA over a 4-bit NF4 base is due to QLoRA \citep{dettmers2023qlora}, building on the
blockwise absmax quantisation of \citet{dettmers20218}. Fitting the adapter to the quantisation error is LoftQ
\citep{li2023loftq}, and the split-first variant is QPiSSA \citep{meng2024pissa}. The data-aware and QR splits are CorDA
\citep{yang2024corda} and OLoRA \citep{buyukakyuz2024olora}. The group-pooled adapter that merges into the zero-points
is QA-LoRA \citep{xu2023qa}. HiRA is due to \citet{huang2025hira}, and OFT to \citet{qiu2023controlling}. The pipelines
are as implemented in HF PEFT \citep{xmangrulkar2022peft}.

\begin{explains}
Over a quantised base, exactly orthogonal OFT keeps the neuron geometry of $\kappa\theta_0$ (\thmref{Corollary}{II.6})
and HiRA keeps the zeros NF4 created (\thmref{Proposition}{II.9}(e)). QA-LoRA makes merging commute with quantisation
by living inside the quantiser's own degrees of freedom.
\end{explains}

\subsubsection{Restarting: what merge-and-restart can reach}\label{sec:base-rebase}

Merge the adapter, restart at the merged point, repeat. Each cycle adds one more copy of an additive shape or applies
one more group element, so the long-run reach is whatever one cycle \emph{generates}. A rank-$r$ cone generates every
matrix, and a subspace, subgroup or submonoid generates only itself. Block-orthogonal factors generate the rotations of
each connected piece of their block hypergraph, because the bracket of infinitesimal rotations in the planes $(a,c)$
and $(c,b)$ is the infinitesimal rotation in the plane $(a,b)$.

\begin{figure}[tbp]
\centering
\includegraphics[width=\linewidth]{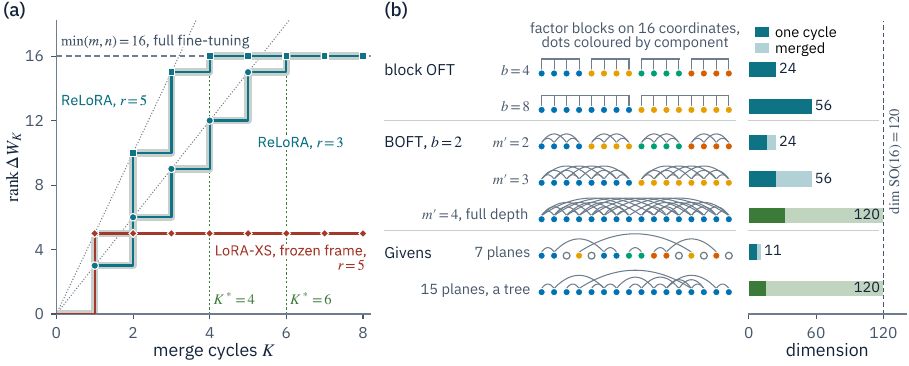}
\caption{\textbf{Rebasing reach} (\thmref{Theorem}{V.3}). (a)~The rank staircase for $m\times n=24\times16$: the rank of $\Delta W_K=\sum_{k\le K}B_kA_k$ after $K$ merge-and-restart cycles of ReLoRA, each adding a fresh random rank-$r$ product, for $r=3$ (circles) and $r=5$ (squares). The measured ranks (teal) lie on the bound $\min(Kr,m,n)$ (thick grey; dotted, the line $Kr$) at every $K$, so ReLoRA reaches full rank exactly at $K^*=\lceil16/r\rceil$ cycles, and not before (corollary~1, \thmref{Theorem}{V.3}(a)). LoRA-XS with its frames frozen at the injective base $W_0$ (red, $r=5$) stays at rank $5$: an idempotent method gains nothing from rebasing (corollary~2). (b)~Seven families of block-orthogonal factors on $n=16$ coordinates: block OFT on a fixed partition with blocks of size $b=4$ and $b=8$, BOFT with $b=2$ and $m'=2,3,4$ butterfly factors, and Givens rotations in $7$ planes and in $15$ planes forming a spanning tree. Arcs are two-element blocks and brackets larger blocks; dots are coloured by the connected component of the hypergraph the blocks generate, and hollow dots are isolated coordinates. The light bars give the dimension $\sum_C\binom{\lvert C\rvert}{2}$ of the generated Lie algebra (\thmref{Theorem}{V.3}(c)), and the dark bars the dimension reached in one cycle, against $\dim\mathrm{SO}(16)=120$ (dashed), with green bars for the families that reach it. Only a connected hypergraph reaches $120$; for BOFT that takes the full depth $m'=4$ (corollary~4). The bar values are in \S\ref{app:repro-figures}.}\label{fig:rebase}
\end{figure}

\begin{theorem}{V.3}{rebasing closure}
(a) \emph{Additive methods with a fixed shape $C\ni0$}, with frames, masks and compression operators frozen at
$\theta_0$ across cycles. Then $R_K=\theta_0+C^{+K}$, the $K$-fold Minkowski sum, and
$R_\infty=\theta_0+\operatorname{Mon}(C)$, the additive monoid generated by $C$ (\bgref{D.10}). If $C$ is closed under
real scalings, $R_\infty=\theta_0+\spanop C$.

(b) \emph{Action methods}, where one cycle applies an element of a set $S$ with $1\in S\subseteq G$; for scaling
methods, $S\subseteq\operatorname{End}(\Theta)$ is a subset containing the identity. Then $R_K=S^K\theta_0$, and
$R_\infty$ is the orbit of the monoid generated by $S$. Suppose $S$ lies in the subgroup generated by connected Lie
subgroups $H_1,\dots,H_p$ (\bgref{D.8}) and contains a neighbourhood of $1$ in each. Then the monoid generated by $S$ is
the connected Lie subgroup $H$ with $\operatorname{Lie}(H)=\operatorname{Lie}\langle\mathfrak h_1,\dots,\mathfrak
h_p\rangle$.

(c) \emph{Block-orthogonal factors.} Let the factors have coordinate blocks $I_j\subseteq\{1,\dots,n\}$ (block OFT,
BOFT factors, Givens planes). Then
\[\operatorname{Lie}\langle\so(I_j)\rangle_j=\bigoplus_C\so(C),\]
summed over the connected components $C$ of the hypergraph $(\{1,\dots,n\},\{I_j\})$.

\emph{Corollaries.}
\begin{enumerate}[label=\arabic*.,leftmargin=1.6em]
  \item \textbf{ReLoRA} \citep{lialin2023relora}. $R_K=W_0+\MM_{\le\min(Kr,m,n)}$. It reaches full fine-tuning iff
  $K\ge\lceil\min(m,n)/r\rceil$.
  \item \textbf{Idempotent methods} gain nothing from rebasing. These are the methods whose shape is exactly a
  subspace, subgroup or submonoid independent of the base, with frames, masks and compression operators frozen across
  cycles. Examples are LoRA-FA with a fixed $A$ \citep{zhang2023lora}, LoRA-XS with frozen frames
  \citep{baazy2024lora}, MoRA with fixed operators \citep{jiang2024mora}, FourierFT with fixed indices
  \citep{gao2024parameter}, WaveFT \citep{bilican2025exploring}, C3A \citep{chen2024parameter}, MiSS
  \citep{kang2024miss} (default balance mode; bat mode is base-dependent but still idempotent, since its blockwise maps
  $W\mapsto(W+I)(I+M)-I$ form a monoid), SHiRA \citep{bhardwaj2024sparse}, BitFit \citep{benzaken2021bitfit}, FISH
  with a fixed mask \citep{sung2021training}, LN-tuning \citep{zhao2023tuning}, (IA)$^3$ \citep{liu2022few}, SSF
  \citep{lian2022scaling} and RoAd \citep{liao20243}. Block OFT with a fixed partition and an exact Cayley map gains
  only the measure-zero set of block rotations with an eigenvalue $-1$, so $R_\infty=W_0\cdot\prod\SO(b)=\overline{R_1}$.
  Constrained OFT (COFT, an $\varepsilon$-ball) with an exact Cayley map gains up to $\prod\SO(b)$; HF's
  \texttt{coft=True} keeps the Cayley--Neumann map by default and so falls under corollary~4. Methods that recompute
  frames or masks from the merged weight (an SVD-recomputing LoRA-XS, ReMoRA's alternating operators, resampled
  indices) do gain.
  \item \textbf{VeRA} \citep{kopiczko2023vera}. $\mathrm{VeRA}^\infty=\text{LoRA-FA}(A)$ whenever the frozen $B$ has no
  zero entries.
  \item \textbf{OFT family.} Block OFT with an exact Cayley map is stuck in $W_0\cdot\prod\SO(b)$. HF's default
  Cayley--Neumann factor is $R=C(Q)(I-Q^4)$, with $C$ the exact Cayley map (\bgref{D.7}), so $R\T R=(I-Q^4)^2$ and $R$
  is not orthogonal. Its singular values are $\lvert1-\mu_k^4\rvert$, where $\pm i\mu_k$ are the eigenvalues of $Q$.
  In the operating regime $\lVert Q\rVert_{\rm op}\le2^{1/4}$ (every $\mu_k\le2^{1/4}$), the regime the truncated series
  is meant for, every factor is a contraction. Re-merging then decreases $K_{\rm out}$ in the Loewner order and no
  singular value grows, so the drift is monotone shrinkage, and the reachable set lies in $W_0$ times the block
  contraction semigroup. With unconstrained $Q$ the factors include singular ones ($\mu_k=1$), and re-merging
  generates $\prod\big(\GL^+(b)\cup\MM_{\le b-2}\big)$, or $\prod\big(\CO^+(2)\cup\{0\}\big)$ when $b=2$; this clause
  is established at the level of a proof sketch. Iterated BOFT (exact Cayley in HF) with $m'$ butterfly factors of
  block size $b$ reaches
  \[\diag(\R^m)\cdot W_0\cdot P\Big(\prod\SO\big(b\,2^{m'-1}\big)\Big)P\T,\]
  because the factor hypergraph has components of size $b\,2^{m'-1}$. For injective $W_0$ this is
  $\diag(\R^m)\cdot W_0\cdot\SO(n)$ iff $b\,2^{m'-1}=n$, the full-depth butterfly $m'=1+\log_2(n/b)$. HF's default \texttt{boft\_\allowbreak n\_\allowbreak butterfly\_\allowbreak factor=1} is
  block OFT times $\diag(s)$. A \emph{single} full-depth BOFT has dimension at most $m'\tfrac nb\binom b2$, which is
  below $\dim\SO(n)$ when $m'(b-1)<n-1$; its matrices are dense, while the set they form has positive codimension.
  \item \textbf{Re-HRA} (HF \texttt{apply\_GS=False}), with $W_0$ injective and $r$ even. Then
  $R_K=(W_0+\MM_{\le Kr})\cap W_0\cdot\SO(n)$, so \emph{rebasing commutes with the meet} of \thmref{Theorem}{II.7}.
  This multiplicative twin of ReLoRA reaches $W_0\cdot\SO(n)$ in at most $\lceil n/r\rceil$ cycles.
  \item \textbf{GaLore (with a gradient-SVD or a random projector), LISA and MeZO}, under the optimizer hypotheses of
  \thmref{Proposition}{IV.2}, are rebasing schemes of \emph{linear} methods whose shape moves. That is why, as a set
  over the sequences of shapes they can choose, they reach all of $\Theta$, while a fixed linear shape $L$ reaches
  only $\theta_0+L$ (corollary~2). Low rank is the per-period obstruction only for GaLore; LISA and MeZO already make
  full-rank per-matrix updates and gain coverage. Flora proper takes full-space steps with a compressed momentum, so
  it has no per-period low-rank shape.
  \item \textbf{Merging never rescues an exactly orthogonal method from its absolute invariants.} In exact arithmetic,
  every exactly orthogonal input-side rebasing preserves $K_{\rm out}$ and the spectrum, and every output-side one
  preserves $K_{\rm in}$ and the spectrum (\thmref{Corollary}{II.6}).
\end{enumerate}
\end{theorem}

\paragraph{Numerical check} For the butterfly hypergraph at $n=64$, $b=4$, the components have sizes $4,8,16,32,64$
for $m'=1,\dots,5$, so it is connected only at $m'=5$; an independent check with HF's own permutation code gives the
same sizes. The Lie algebra generated by the Cayley--Neumann factors at $b=4$ has dimension $16=\dim\gl(4)$. With skew
entries of size about $0.05$, $K_{\rm out}$ moves by $0.2\%$, $1\%$, $6\%$ and $22\%$ (relative) after $1$, $10$, $50$
and $200$ default-OFT merges, decreasing in the Loewner order at every merge. Random skew $Q$ with
$\lVert Q\rVert_{\rm op}\le2^{1/4}$ never give $\lVert R\rVert_{\rm op}>1$, and at $b=2$ the generator
$Q=\left(\begin{smallmatrix}0&-1\\1&0\end{smallmatrix}\right)$ gives $R=0$ (Appendix~\ref{app:repro}).

\paragraph{Credit} Merge-and-restart training is ReLoRA \citep{lialin2023relora}. The orthogonal methods are OFT
\citep{qiu2023controlling}, with its constrained variant COFT, BOFT \citep{liu2023parameter}, whose butterflies come from
\citet{dao2019learning}, and HRA \citep{yuan2024bridging}, whose Householder products go back to
\citet{mhammedi2016efficient}. Cayley parametrisations of orthogonal layers are studied in
\citet{lezcanocasado2019cheap}. The group theory is classical. It uses Yamabe's theorem on arcwise connected subgroups
\citep{xyamabe1950arcwise} (\bgref{D.9}) and Scherk's sharp form of the Cartan--Dieudonné theorem
\citep{xscherk1950decomposition} (\bgref{D.11}). That GaLore \citep{zhao2024galore} is one-sided LoRA re-based every
period is due to \citet{xtorrobahennigen2025duality}, and the random-projection case to Flora \citep{hao2024flora}; see
\thmref{Proposition}{IV.2}. The other moving-shape methods are LISA \citep{pan2024lisa} and MeZO
\citep{malladi2023fine}.

\begin{table}[t]
\centering\footnotesize
\caption{What merge-and-restart reaches (\thmref{Theorem}{V.3} and its corollaries).}\label{tab:base-rebasing}
\begin{tabularx}{\linewidth}{@{}>{\raggedright\arraybackslash}p{0.25\linewidth}>{\raggedright\arraybackslash}p{0.19\linewidth}>{\raggedright\arraybackslash}X>{\raggedright\arraybackslash}p{0.2\linewidth}@{}}
\toprule
\headrow \textbf{Method under rebasing} & \textbf{One cycle} & \textbf{After $K$ cycles, or $R_\infty$} &
\textbf{Gain from restarting} \\
\midrule
ReLoRA \citep{lialin2023relora} & $W_0+\Mr$ & $W_0+\MM_{\le\min(Kr,m,n)}$ & yes; full FT at
$K\ge\lceil\min(m,n)/r\rceil$ \\
\addlinespace
LoRA-FA (fixed $A$), LoRA-XS (frozen frames), BitFit, (IA)$^3$, SSF, fixed masks & a subspace or submonoid & the same
set & none (idempotent) \\
\addlinespace
VeRA \citep{kopiczko2023vera} ($B$ without zero entries) & $W_0+\Lambda_bB\Lambda_dA$ & $\text{LoRA-FA}(A)$ & yes, up to
LoRA-FA \\
\addlinespace
block OFT \citep{qiu2023controlling}, exact Cayley, fixed partition & $W_0\cdot\prod\mathrm{Cay}(b)$ &
$W_0\cdot\prod\SO(b)=\overline{R_1}$ & measure zero \\
\addlinespace
COFT, exact Cayley & an $\varepsilon$-ball in each block & up to $W_0\cdot\prod\SO(b)$ & within the blocks \\
\addlinespace
block OFT, HF Cayley--Neumann default & not orthogonal & a contraction monoid for small $Q$ & drifts; $K_{\rm out}$
shrinks \\
\addlinespace
BOFT \citep{liu2023parameter}, $m'$ butterfly factors, exact Cayley & products of butterfly factors &
$\diag\cdot W_0\cdot P\big(\prod\SO(b\,2^{m'-1})\big)P\T$ & $\SO(n)$ iff $b\,2^{m'-1}=n$ \\
\addlinespace
re-HRA \citep{yuan2024bridging} ($W_0$ injective, $r$ even) & $(W_0+\Mr)\cap W_0\cdot\SO(n)$ &
$(W_0+\MM_{\le Kr})\cap W_0\cdot\SO(n)$ & yes; $W_0\cdot\SO(n)$ within $\lceil n/r\rceil$ \\
\addlinespace
GaLore, LISA, MeZO & a linear shape that moves & all of $\Theta$, as a set & yes, because the shape moves \\
\bottomrule
\end{tabularx}
\end{table}

\begin{explains}
ReLoRA is \thmref{Observation}{I.7} unrolled in time, since a sum of $K$ rank-$r$ LoRAs is one LoRA of rank $Kr$, so it
walks the group its cone generates. A fixed additive shape gains from restarts only when it is not closed under
addition, as LoRA's cone and VeRA's shape are not. Exact-Cayley block OFT can never escape its $n/b$ partial-Gram
invariants however often it is re-merged. Butterfly permutations exist precisely to connect the block hypergraph, and
they connect it fully only at full depth.
\end{explains}

\begin{remark*}{2026: POET-X}
POET-X \citep{qiu2026poet}, a pretraining method whose $W_0$ is a random initialisation, changes the partition between
cycles, on both sides of the weight. Each round trains a block-diagonal factor conjugated by a random permutation on
each side, merges both into the weight and redraws the permutations. With exact Cayley blocks every merged weight lies
in $\SO(m)\,W_0\,\SO(n)$, so it keeps the singular values of $W_0$ (\thmref{Theorem}{II.5}), and by
\thmref{Theorem}{V.3}(b) and (c) blocks that together connect all coordinates of a side generate all of $\SO$ on that
side. Its blocks are in fact $(I+Q)(I+Q+Q^2+Q^3)$, the Cayley--Neumann factor of corollary~4, so while every
$\mu_k\le2^{1/4}$ no merge raises a singular value, and the spectrum can only drift down.
\end{remark*}

\begin{example*}{one projection of a 7B model}
Take a $4096\times4096$ projection. ReLoRA with $r=16$ reaches full fine-tuning after $256$ cycles and not before. For
injective $W_0$, re-HRA with $r=8$ reaches $W_0\cdot\SO(4096)$ within $512$ cycles. BOFT with $b=4$ needs $m'=11$
butterfly factors before re-merging reaches that orbit up to output scaling; at HF's default of one factor it stays in
$W_0\cdot\prod\SO(4)$ up to the scale. A single 11-factor BOFT has dimension at most $11\cdot1024\cdot6=67{,}584$,
against $\dim\SO(4096)=8{,}386{,}560$.
\end{example*}

\paragraph{Prediction} With an exact Cayley map and a fixed partition, re-merging adds nothing to block OFT's reach
beyond closure, so re-merged exact-Cayley block OFT and COFT on a fixed partition never leave $W_0\cdot\prod\SO(b)$.
Enlarging the reach needs a connected block hypergraph (partitions that change between cycles, or full-depth
butterflies) or a COFT ball. LoRA-type cones gain from rebasing with a fixed shape, while linear additive shapes do not
(corollary~2). These are statements about reach; whether extra reach improves a trained model is the experiment
(\thmref{Prediction}{X.6}).

\subsubsection{Combining: rules that must ignore the gauge}\label{sec:base-combine}

A LoRA module ships as factors $(B,A)$, the model sees only $BA$, and $(Bg^{-1},gA)$ gives the same product for every
$g\in\GL_r$. A combination rule should return the same model for every choice of factors, or two people with the same
adapters can merge them into different models. Task arithmetic adds the products and passes. LoraHub adds the $B$'s,
adds the $A$'s and multiplies, and fails, because its cross terms $B_iA_l$ see a rescaling $(cB,A/c)$ of one module.
The merge panel of Figure~\ref{fig:gauge} shows the LoraHub merge moving under that rescaling while task arithmetic
stays fixed.

\begin{proposition}{V.4}{combination rules: gauge and naturality}
(a) A rule $\Phi\big((B_i,A_i)_i\big)$ for combining LoRA factors descends to a function of the adapters $(B_iA_i)_i$
iff it is constant on the fibres of $(B_i,A_i)\mapsto(B_iA_i)$. On the full-rank locus this is equivalent to
$\prod_i\GL_r$-invariance. It is also equivalent for $\Phi$ continuous on the whole factor space; this clause rests on
the unique closed orbit in each fibre, which lies in the closure of every orbit of the fibre (\bgref{D.31}).
\begin{itemize}
  \item Task arithmetic $\sum_iw_is_iB_iA_i$, with $s_i$ HF's \texttt{scaling} ($\alpha_i/r_i$, or $\alpha_i/\sqrt{r_i}$
  under rsLoRA), is invariant and lands in $\mathrm{LoRA}_{\sum_ir_i}$. It is HF's \texttt{cat} with merged scale $1$
  (\thmref{Observation}{I.7}).
  \item LoraHub's factor-wise rule $\big(\sum_iw_iB_i\big)\big(\sum_jw_jA_j\big)$ is not invariant. Replacing
  $(B_2,A_2)$ by $(cB_2,A_2/c)$ changes it by
  \[\sum_{l\ne2}w_lw_2\big[(c-1)B_2A_l+(c^{-1}-1)B_lA_2\big].\]
  HF's \texttt{add\_\allowbreak weighted\_\allowbreak adapter(..., combination\_\allowbreak type="linear")} is a
  factor-wise rule of the same type. It weights $A_i$ by $\operatorname{sign}(w_i)\sqrt{\lvert w_is_i\rvert}$ and $B_i$
  by $\sqrt{\lvert w_is_i\rvert}$ and sums the factors separately, so it has cross terms. On adapters that share a
  frozen $A$ it is still not exact up to a global factor, since the relative weights become
  $\sqrt{\lvert w_is_i\rvert}$. HF's non-SVD \texttt{ties}, \texttt{dare\_*} and \texttt{magnitude\_prune} variants
  also act on factors; the \texttt{*\_svd} variants act on full updates.
  \item Poly and MHR are different. Their skill inventories are trained jointly through the product-of-sums map, which
  is invariant under the diagonal $\GL_r$, one common $g$ for every skill. Numerically, for fixed generic routing with
  fewer skills than tasks and full-rank factors, this is the whole continuous fibre. With at least as many skills as
  tasks, or once the routing is counted as a parameter, there are further symmetries (skill permutations, and
  re-mixings $\alpha\mapsto\alpha M$ with $M\mathbf 1=\mathbf 1$). Their cross terms are a design choice of a jointly
  trained parametrisation, and raise no question of well-definedness.
\end{itemize}

(b) \emph{Well-definedness versus exactness.} A joint gauge fixing makes a factor-wise rule well defined. For
rank-preserving product-of-sums rules, exactness (agreement with task arithmetic for all weights $w$) needs degree one
in $w$, which for injective factors forces a shared factor; \texttt{cat}, which concatenates and raises the rank, is
exact without one. A frozen shared factor (FFA-LoRA; LoRA-FA) gives $\big(\sum_iw_iB_i\big)A=\sum_iw_iB_iA$ exactly.
LoraHub with a shared $A$ still returns $\big(\sum_jw_j\big)\sum_iw_iB_iA$, exact only up to the global factor
$\sum_jw_j$, since its weights are unnormalised. A per-module SVD canonical form leaves a $(\mathbb Z/2)^r$ of sign
flips per module, plus rotations within repeated singular values, and these change the cross terms. What is needed is
a \emph{joint} frame that produces a shared factor, as KnOTS's joint re-factorisation of the concatenated updates
through a shared $U\Sigma$ does. Federated LoRA shows the difference. FedAvg-LoRA clients start each round from the
same broadcast $(B,A)$, so their failure,
\[\tfrac12(B_1+B_2)\cdot\tfrac12(A_1+A_2)\ne\tfrac12(B_1A_1+B_2A_2),\]
is bilinearity, and it persists in a fully fixed gauge.

(c) \emph{Naturality.} On updates with $N$ coordinates, write $T_N$, $B_N$ and $\mathrm{Mon}_N$ for the invertible
diagonal, signed permutation and monomial matrices. Task arithmetic is $\GL(\Theta)$-equivariant on updates. DARE is
coordinatewise and magnitude-free, $D(m\odot\tau)=m\odot(D\tau)$ for every invertible diagonal $D$ and the same mask
$m$. So DARE is exactly equivariant under the torus $T_N$, sign flips included, for each mask, and under permutations in
distribution; its distributional group is exactly $\mathrm{Mon}_N$. TIES with global trimming (top-$k$ magnitude, sign
election by total mass as in the TIES paper and HF's default \texttt{"total"}, disjoint mean) is equivariant under
$\R^\times\cdot B_N$, homotheties times signed permutations, generically (off magnitude ties at the threshold). With
HF's \texttt{majority\_\allowbreak sign\_\allowbreak method="frequency"}, count ties are broken to $+1$ and the group
drops to $\R_{>0}\cdot S_N$. Per-tensor trimming changes the group to $\prod_b\big(\R^\times\cdot B_{N_b}\big)$,
extended by permutations of equal-size blocks. DARE followed by full TIES (with trimming) is
$\R^\times\cdot B_N$-equivariant in distribution; HF's and MergeKit's \texttt{dare\_ties} has no magnitude trim, so
with \texttt{"total"} election it is exactly $T_N$-equivariant per mask, and its group contains $\mathrm{Mon}_N$. None
of the sparsifying rules (DARE, TIES, DARE-TIES) is $\Ort(N)$-equivariant. Applied to factors and summed across adapters
(HF's non-SVD variants), all of them are gauge-dependent, and under independent per-adapter gauges even DARE fails,
through the cross terms of (a). DARE's masking commutes with $T_r$ exactly for each mask and with $\mathrm{Mon}_r$ in
distribution, so a single masked adapter, or a factor-wise sum under one gauge shared by all adapters, is
$T_r$-invariant per mask and $\mathrm{Mon}_r$-invariant in distribution.
\end{proposition}

\paragraph{Numerical check} DARE with a random non-unit diagonal $D$ is equivariant to $2\times10^{-15}$ per mask.
TIES is homogeneous to $9\times10^{-16}$ and fails by $5.6$ under a generic positive diagonal. The general LoraHub
change formula matches with three modules to $2\times10^{-15}$. On the factors of two adapters, a diagonal gauge on one
of them moves factor-wise DARE by an amount of order one, while the same gauge shared by all adapters leaves it fixed to
$10^{-15}$. For Poly with $3$ skills, $r=2$ and $6\times6$ weights, the fibre has dimension $32$ with $2$ tasks and
$4=\dim\GL_2$ with $4$ tasks (Appendix~\ref{app:repro}).

\paragraph{HF options} With HF's \texttt{majority\_\allowbreak sign\_\allowbreak method="frequency"}, sign counts are
integers, so ties are common, and they are broken towards $+1$. Sign flips and negative homotheties therefore fail for
\texttt{ties} and \texttt{dare\_ties} alike.

\paragraph{Credit} The rules studied here are task arithmetic \citep{ilharco2022editing}, LoraHub
\citep{huang2023lorahub}, TIES \citep{yadav2023ties}, DARE \citep{yu2023language} and KnOTS \citep{stoica2024model},
whose joint SVD is the shared frame of (b); \texttt{dare\_ties} is as implemented in HF PEFT
\citep{xmangrulkar2022peft} and MergeKit \citep{xgoddard2024mergekit}. Jointly trained skill inventories are Poly
\citep{ponti2022combining} and MHR \citep{caccia2022multi}. Federated averaging of LoRA factors is used in FedIT
\citep{zhang2023building}, and the frozen shared factor that removes its bilinearity is FFA-LoRA
\citep{sun2024improving}. That functions of LoRA weights should respect their $\GL_r$ symmetry is the starting point of
the $\GL$-equivariant networks of \citet{putterman2024learning}; for full models, hidden-unit permutations play that
role in Git Re-Basin \citep{ainsworth2022git}. The unique closed orbit in each fibre is classical invariant theory.
\citet{xkempf1979length} characterise closed orbits by norm-minimising points (\bgref{D.30}), here the balanced
factorisations of \thmref{Proposition}{III.7}, and \citet{xrichardson1990minimum} treat real groups. For optimizers,
the same question leads to \citet{zhang2024riemannian} and \citet{yen2024lora}; see \thmref{Theorem}{III.12}.

\begin{explains}
Post-hoc factor merging of independently trained modules, as in LoraHub and HF's \texttt{linear}, is not a function of
the adapters. It depends on how each module happened to be factorised, which nobody controls after training. Federated
LoRA fails through bilinearity instead, and FFA-LoRA's frozen shared factor repairs it. KnOTS's joint
re-factorisation produces a shared factor. \textbf{Design rule: combine LoRAs on products, or only after a joint
re-factorisation that produces a shared factor.}
\end{explains}

\paragraph{Prediction} LoraHub behaves like update arithmetic, up to the factor $\sum_jw_j$, when its modules share
the $A$-seed and $A$ stays close to its initialisation. That regime is empirical and favoured by a large ratio
$\eta_B/\eta_A$, as in LoRA+ \citep{hayou2024lora}. LoraHub degrades when the modules come from independent seeds
(\thmref{Prediction}{X.9}).

\subsubsection{What base change leaves for the network}\label{sec:base-next}

Every move here stayed inside weight space, where merging into the base is the canonical arrow to the terminal object
of \thmref{Observation}{I.4} and always exists. Adapters, prefixes and ReFT change the network itself, and for them
merging is a lifting problem whose obstructions the activation function decides. That is Exposé~\ref{sec:merge}.

\section{Extensions and mergeability}\label{sec:merge}

\begin{flushright}\small\itshape
When a lift fails, look for the place where it fails,\\
and give the obstruction a name.\\[3pt]
{\upshape\footnotesize after Grothendieck}
\end{flushright}

Adapters, prefixes, prompts and ReFT enlarge the network before they train, so merging them back is a lifting problem.
Seven local rewrite rules give sufficient conditions for a merge and name the side condition that fails when they stop.

Merging an extension back means finding a weight of the original network that computes the trained function. The rules
slide an edit along the wires of a block until a weight absorbs it. They certify a merge when the edit acts alike at
every position, the absorbing weights are unshared, shifts have bias slots and, under RoPE, scales are constant on rotary
pairs. A stuck rewrite shows only a box-local obstruction. Three examples show what the exposé finds. The feed-forward
scale of (IA)$^3$ always folds into $W_{\mathrm{down}}$; it folds into $W_{\mathrm{up}}$ under SwiGLU for every scale
and under ReLU for nonnegative ones, and under GELU only entries $0$ and $1$ fold. A prefix is a gated parallel adapter
whose gate depends on the input, so for generic frozen weights it cannot be folded into the head's own weights. Per
hidden vector, LoReFT is LoRA on an inserted identity box, and since it acts at selected positions no merge is known.

\subsubsection{The weight that absorbs the edit}\label{sec:merge-hook}

(IA)$^3$ \citep{liu2022few} rescales the hidden units of a feed-forward block by a learned vector $\ell$, which one
would like to fold into a frozen weight after training. The weight that reads the hidden units, $W_{\mathrm{down}}$,
absorbs $\ell$ under every activation. The weight that produces them, $W_{\mathrm{up}}$, absorbs it under SwiGLU for
every $\ell$, under ReLU when no entry is negative, and under GELU only for entries $0$ or $1$.

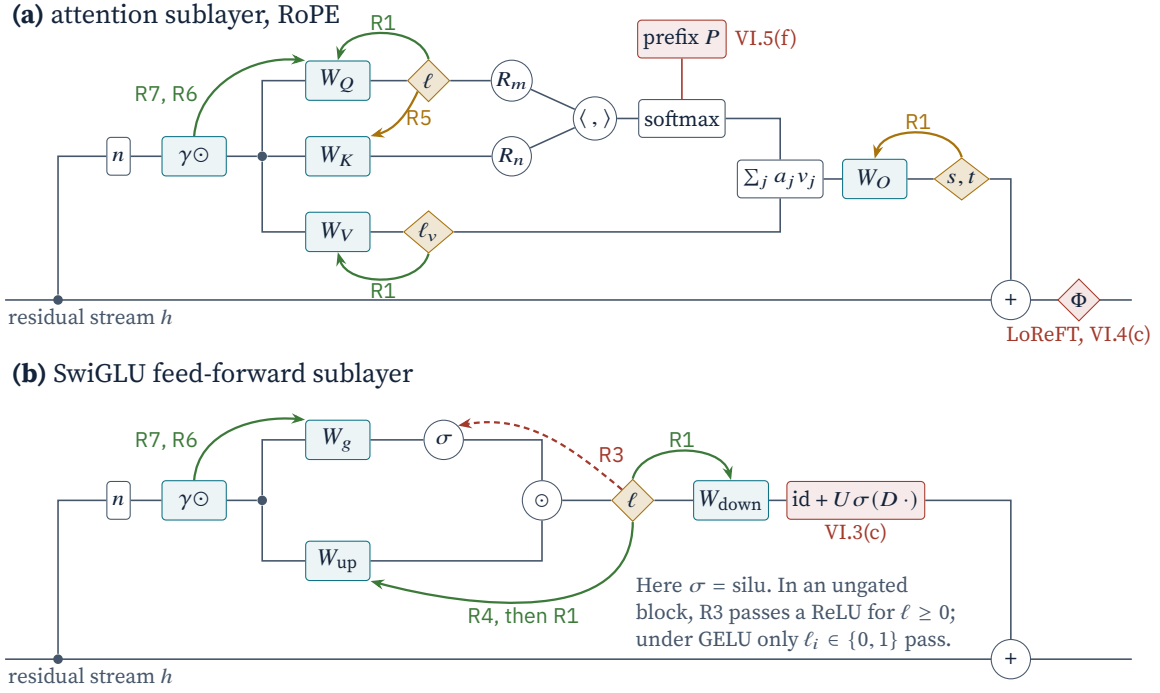
\begin{figure}[tbp]
\centering
\begin{tikzpicture}[x=1cm,y=1cm,
  wire/.style={inkgray,line width=0.6pt},
  bx/.style={draw=tide,fill=tide!8,rounded corners=1.5pt,minimum height=5.2mm,minimum width=8.5mm,inner sep=2pt,
    font=\footnotesize,text=ink},
  op/.style={draw=inkgray,fill=white,circle,inner sep=0.6pt,minimum size=5.2mm,font=\scriptsize,text=ink},
  nb/.style={draw=inkgray,fill=white,rounded corners=1.5pt,minimum height=5mm,inner sep=2pt,font=\scriptsize,text=ink},
  ed/.style={draw=ochre,fill=ochre!18,diamond,aspect=1.3,inner sep=0.5pt,minimum size=5.6mm,font=\scriptsize,text=ink},
  bad/.style={draw=seal,fill=seal!10,rounded corners=1.5pt,minimum height=5mm,inner sep=2pt,font=\scriptsize,text=ink},
  cp/.style={circle,fill=inkgray,inner sep=0pt,minimum size=3.6pt},
  ok/.style={-{Stealth[length=1.8mm]},moss,line width=0.85pt},
  cond/.style={-{Stealth[length=1.8mm]},ochre,line width=0.85pt},
  no/.style={-{Stealth[length=1.8mm]},seal,line width=0.85pt,dash pattern=on 2.2pt off 1.6pt},
  rl/.style={font=\sffamily\scriptsize,inner sep=1.2pt},
  note/.style={font=\scriptsize,text=inkgray,align=left,inner sep=1pt}]
\begin{scope}
\node[anchor=west,font=\small,text=ink] at (-0.25,3.75) {\textbf{(a)} attention sublayer, RoPE};
\draw[wire] (-0.2,0) -- (14.7,0);
\node[note,anchor=north west] at (-0.2,-0.06) {residual stream $h$};
\node[cp] (a0) at (0.5,0) {};
\node[nb] (an) at (1.3,1.9) {$n$};
\node[bx] (ag) at (2.3,1.9) {$\gamma\odot$};
\node[cp] (a1) at (3.2,1.9) {};
\node[bx] (WQ) at (4.2,2.9) {$W_Q$};
\node[bx] (WK) at (4.2,1.9) {$W_K$};
\node[bx] (WV) at (4.2,0.9) {$W_V$};
\node[ed] (lq) at (5.4,2.9) {$\ell$};
\node[ed] (lv) at (5.4,0.9) {$\ell_v$};
\node[op] (Rm) at (6.5,2.9) {$R_m$};
\node[op] (Rn) at (6.5,1.9) {$R_n$};
\node[op] (dot) at (7.6,2.4) {$\langle\,,\rangle$};
\node[nb] (sm) at (8.75,2.4) {softmax};
\node[nb] (mix) at (10.05,1.6) {$\sum_j a_jv_j$};
\node[bx] (WO) at (11.3,1.6) {$W_O$};
\node[ed] (ssf) at (12.45,1.6) {$s,t$};
\node[op] (ap) at (13.1,0) {$+$};
\node[ed,draw=seal,fill=seal!12] (reft) at (14.0,0) {$\Phi$};
\node[bad] (P) at (8.75,3.45) {prefix $P$};
\draw[wire] (a0) |- (an);
\draw[wire] (an) -- (ag);
\draw[wire] (ag) -- (a1);
\draw[wire] (a1) |- (WQ);
\draw[wire] (a1) -- (WK);
\draw[wire] (a1) |- (WV);
\draw[wire] (WQ) -- (lq);
\draw[wire] (lq) -- (Rm);
\draw[wire] (WK) -- (Rn);
\draw[wire] (WV) -- (lv);
\draw[wire] (Rm) -- (dot);
\draw[wire] (Rn) -- (dot);
\draw[wire] (dot) -- (sm);
\draw[wire] (sm.east) -| (mix.north);
\draw[wire] (lv.east) -| (mix.south);
\draw[wire] (mix) -- (WO);
\draw[wire] (WO) -- (ssf);
\draw[wire] (ssf.east) -| (ap.north);
\draw[seal,line width=0.6pt] (P) -- (sm);
\draw[ok] (lq.north) to[out=105,in=75] node[rl,above,text=moss]{R1} (WQ.north);
\draw[cond] (lq.south west) to[out=-115,in=15] node[rl,right,pos=0.45,text=ochre]{R5} (WK.north east);
\draw[ok] (lv.south) to[out=-105,in=-75] node[rl,below,text=moss]{R1} (WV.south);
\draw[ok] (ag.north) to[out=85,in=175] node[rl,above left,pos=0.2,text=moss]{R7, R6} (WQ.north west);
\draw[cond] (ssf.north) to[out=105,in=75] node[rl,above,text=ochre]{R1} (WO.north);
\node[note,anchor=west,text=seal] at (P.east) {\ VI.5(f)};
\node[note,anchor=north,text=seal] at (reft.south) {LoReFT, VI.4(c)};
\end{scope}
\begin{scope}[yshift=-4.75cm]
\node[anchor=west,font=\small,text=ink] at (-0.25,3.75) {\textbf{(b)} SwiGLU feed-forward sublayer};
\draw[wire] (-0.2,0) -- (14.7,0);
\node[note,anchor=north west] at (-0.2,-0.06) {residual stream $h$};
\node[cp] (b0) at (0.5,0) {};
\node[nb] (bn) at (1.3,2.1) {$n$};
\node[bx] (bg) at (2.3,2.1) {$\gamma\odot$};
\node[cp] (b1) at (3.2,2.1) {};
\node[bx] (Wg) at (4.2,2.9) {$W_g$};
\node[bx] (Wu) at (4.2,1.3) {$W_{\mathrm{up}}$};
\node[op] (sig) at (5.6,2.9) {$\sigma$};
\node[op] (had) at (6.9,2.1) {$\odot$};
\node[ed] (l) at (8.1,2.1) {$\ell$};
\node[bx] (Wd) at (9.4,2.1) {$W_{\mathrm{down}}$};
\node[bad] (ad) at (11.05,2.1) {$\id+U\sigma(D\,\cdot)$};
\node[op] (bp) at (13.1,0) {$+$};
\draw[wire] (b0) |- (bn);
\draw[wire] (bn) -- (bg);
\draw[wire] (bg) -- (b1);
\draw[wire] (b1) |- (Wg);
\draw[wire] (b1) |- (Wu);
\draw[wire] (Wg) -- (sig);
\draw[wire] (sig.east) -| (had.north);
\draw[wire] (Wu.east) -| (had.south);
\draw[wire] (had) -- (l);
\draw[wire] (l) -- (Wd);
\draw[wire] (Wd) -- (ad);
\draw[wire] (ad.east) -| (bp.north);
\draw[ok] (l.north) to[out=75,in=105] node[rl,above,text=moss]{R1} (Wd.north);
\draw[ok] (l.south) to[out=-90,in=-15] node[rl,below,pos=0.55,text=moss]{R4, then R1} (Wu.south east);
\draw[no] (l.north west) to[out=140,in=10] node[rl,above right,pos=0.18,text=seal]{R3} (sig.north east);
\draw[ok] (bg.north) to[out=85,in=175] node[rl,above left,pos=0.2,text=moss]{R7, R6} (Wg.north west);
\node[note,anchor=north,text=seal] at (ad.south) {VI.3(c)};
\node[note,anchor=west] at (8.1,0.6) {Here $\sigma=\mathrm{silu}$. In an ungated\\ block, R3 passes a ReLU for $\ell\ge0$;\\ under GELU only $\ell_i\in\{0,1\}$ pass.};
\end{scope}
\end{tikzpicture}
\caption{The merge rules of \thmref{Theorem}{VI.3} on a pre-LN transformer block, drawn as a string diagram with the
residual stream along the bottom: (a) the attention sublayer under RoPE, (b) a SwiGLU feed-forward sublayer. Tide boxes
are trainable boxes of the base network, $\bullet$ is a copy node, $\odot$ the product node of the gated block, and
ochre diamonds are edits inserted by a method. Moss arrows are rewrites licensed for every parameter value, ochre arrows
rewrites licensed under a side condition, and seal marks what the rules do not merge. A query scaling $\ell$ fuses into
$W_Q$ by R1; R5 moves it onto $W_K$ only if $\ell$ is constant on each rotary pair and across a grouped-query sharing
group. A value scaling $\ell_v$ fuses into $W_V$. A norm gain is copied to every reader (R7) and absorbed (R6). SSF's
scale and shift $(s,t)$ fuse into $W_O$ only if $W_O$ has a bias slot. The (IA)$^3$ scale fuses into
$W_{\mathrm{down}}$ (R1), and through the product node into $W_{\mathrm{up}}$ for every $\ell$ (R4, then R1); it does not
slide through the SiLU gate, and in an ungated block R3 carries it through a ReLU only for $\ell\ge0$. A prefix enters the
softmax with an input-dependent gate (\thmref{Theorem}{VI.5}(f)), a nonlinear adapter is a non-affine edit
(\thmref{Theorem}{VI.3}(c)), and LoReFT acts at selected positions of the residual stream
(\thmref{Proposition}{VI.4}(c)); R1--R7 merge none of the three. The companion site's merge game runs these rewrites
interactively, with a float64 check of each licensed merge on seeded weights.}\label{fig:merge}
\end{figure}

The activation belongs to the network, which a weight-space method never sees (\thmref{Remark}{I.3}). Adapters,
prefixes, prompts and ReFT add boxes the pretrained network lacked, and whether they can be removed depends on their
surroundings. Figure~\ref{fig:merge} runs the rules of this exposé on one block.

\subsubsection{Extensions, and merging as a lift}\label{sec:merge-extensions}

An adapter $y\mapsto y+U\sigma(Dy)$ \citep{houlsby2019parameter} adds weights $(U,D)$, and at $U=0$ the network
computes the old function. A prefix \citep{li2021prefix} adds key--value pairs to every softmax, and ReFT
\citep{wu2024reft} edits hidden states at chosen positions. Each time $\Theta$ grows to $\Theta\times\Theta'$ and $f$ to
a larger $\tilde f$. A prefix always takes softmax mass, and on a head with $W_oW_v\ne0$ no prefix leaves the output
unchanged, so prefixes have no neutral setting there (\thmref{Theorem}{VI.5}(f)).

\begin{definition}{VI.1}{extensions; mergeability}
\begin{itemize}
  \item An \textbf{extension} is a function-preserving pointed embedding
  $\iota:(\Theta,f,\theta_0)\to(\Theta\times\Theta',\tilde f,(\theta_0,\theta_0'))$, $\iota(\theta)=(\theta,\theta_0')$,
  with $\tilde f(\theta,\theta_0',-)=f(\theta,-)$. The point $\theta_0'$ is the \textbf{neutral point} (``identity at
  initialisation'').
  \item An \textbf{unpointed extension} has no neutral point.
  \item A method on an extension is \textbf{mergeable} if there is $\mu:Q\to\Theta$ with
  $\Phi_{\tilde f}\circ\rho=\Phi_f\circ\mu$: a lift along the realisation.
  \item It is \textbf{box-locally mergeable} if the modified box function stays in the box's function class: linear
  maps for a linear box, affine maps when the box has a bias slot. \thmref{Theorem}{VI.3} gives sufficient conditions,
  with explicit side conditions (position-uniformity, no weight sharing, bias slots), under which this yields a merge
  into the base network.
\end{itemize}
\end{definition}

A merge is an arrow $\mu$ that makes the square of realisations commute:
\[
\begin{tikzcd}[column sep=5.5em,row sep=2.6em]
  Q \arrow[r,"\mu"] \arrow[d,"\rho"'] & \Theta \arrow[d,"\Phi_f"] \\
  \Theta\times\Theta' \arrow[r,"\Phi_{\tilde f}"'] & C^\infty(X,Y)
\end{tikzcd}
\]
A weight-space method merges by $\mu=\rho$, the canonical arrow to full fine-tuning (\thmref{Observation}{I.4}(a)), so
the real questions live on extensions, in the slice of $\Para$ over the pretrained model (\thmref{Remark}{I.3}).
Box-local mergeability can be read off formulas, and the classification of Exposé~\ref{sec:atlas} records it as a merge
type:
\begin{itemize}
  \item $\mathsf{M1}$: the edit merges into its own box;
  \item $\mathsf{M1}{\rightarrow}$: the rules move it into a neighbouring box, under side conditions;
  \item $\mathsf{Mq}$: an exact merge exists but leaves the quantised type;
  \item $\mathsf{M}\infty$: no merge; the rules are blocked box-locally, or the edit lies outside them.
\end{itemize}

\begin{explains}
``Zero inference latency'' runs two questions together: whether the original architecture can compute the trained
function at all, and whether a local edit finds the weight. A zero up-projection, or the zero gate of LLaMA-Adapter,
makes training start at the pretrained function, and by the function-level form of \thmref{Lemma}{II.4} only $U$, or the
gate, moves at first.
\end{explains}

Adapters are due to \citet{houlsby2019parameter}, prefix tuning to \citet{li2021prefix}, prompt tuning to
\citet{lester2021power} and ReFT to \citet{wu2024reft}. \citet{he2021unified} first placed adapters, prefixes and LoRA in
one frame, as modifications of hidden states. The construction $\Para$ is from \citet{fong2017backprop} and
\citet{cruttwell2021categorical} (\bgref{B.16}).

\subsubsection{The activation fixes the privileged basis}\label{sec:merge-basis}

An edit moves through a network by sliding along wires, and a linear map slides through a box when it commutes with the
box. Activations are the bottleneck. Permutations commute with every elementwise $\sigma$, and for ReLU so do positive
rescalings, since $\mathrm{ReLU}(ct)=c\,\mathrm{ReLU}(t)$ for $c>0$. The lemma says how little else does.

\begin{lemma}{VI.2}{the activation fixes the privileged basis; classical, \citet{xsussmann1992uniqueness},
\citet{xchen1993geometry}, \citet{xgodfrey2022symmetries}}
Let $\sigma:\R\to\R$ be an elementwise activation that is not affine, and let $L\in\GL_d$ commute with it as maps on
$\R^d$: $L\,\sigma^{\times d}=\sigma^{\times d}L$.

(a) If $\sigma\in C^2$, then $L=PD$ is monomial.

(b) If $\sigma$ is ReLU or leaky ReLU with slope $\alpha\ne1$, then $L=PD$ with $D$ a positive diagonal.

(c) If moreover $\sigma$ is real-analytic at $0$ (\bgref{C.22}) with $\sigma(t)=\sum a_kt^k$, each diagonal entry $c$ of
$D$ satisfies $ca_k=c^ka_k$ for all $k$. So $c=1$ for GELU and SiLU, since $\sigma''(0)\neq0$, and $c=\pm1$ for tanh.
\end{lemma}

The lemma is classical. \citet{xsussmann1992uniqueness} and \citet{xchen1993geometry} found the weight-space symmetries of
tanh networks, and \citet{xgodfrey2022symmetries} computed the linear maps that commute with each common activation.
Permutation symmetry is what \citet{ainsworth2022git} use to align whole models before merging them.

So rules R2 and R3 below exhaust the invertible linear maps that slide through $\sigma$: permutations for GELU and SiLU,
and positive monomials for ReLU. In gated blocks such as SwiGLU every diagonal slides through the linear branch (rule
R4), so their hidden-unit group is the full monomial group $\mathrm{Mon}_{d_{\mathrm{ff}}}$.

\begin{explains}
Within the lemma's scope (global linear intertwiners of an ungated elementwise $\sigma$), the symmetries a method may
ignore are \emph{derived} from $\sigma$. For ungated FFNs (BERT, GPT-2, ViT), GELU and SiLU give $S_d$, tanh gives $B_d$,
and ReLU gives positive monomials. Gated FFNs (SwiGLU and GeGLU, as in LLaMA, Mistral, Qwen and Gemma) are not
elementwise, and their hidden-unit group is the full $\mathrm{Mon}_{d_{\mathrm{ff}}}$, acting through the linear branch
(R4 of \thmref{Theorem}{VI.3}). Affine, bias-absorbing symmetries can be larger still: sigmoid satisfies
$\sigma(-z)=1-\sigma(z)$. The privileged basis of LayerNorm gains comes from the diagonal gain parametrisation, and
$\sigma$ plays no part in it. A gain-free RMSNorm is $\Ort(d)$-equivariant, which SliceGPT \citep{xashkboos2024slicegpt} and QuaRot
\citep{xashkboos2024quarot} use to rotate the residual stream (\thmref{Observation}{II.15}).
\end{explains}

\noindent\labelsf{Prediction}\enspace Over random function-preserving permutations of hidden units, LoRA's outcome
variance equals its seed variance, since the pretrained order is exchangeable. FourierFT \citep{gao2024parameter}, LoKr
\citep{yeh2023navigating} and MoRA \citep{jiang2024mora} show larger variance, and so do BOFT \citep{liu2023parameter}
and OFT \citep{qiu2023controlling} on modules whose input side is the hidden layer (\texttt{down\_proj}). On
\texttt{up\_proj} and \texttt{gate\_proj} a hidden-unit permutation only reorders BOFT's output rows, so BOFT is exactly
equivariant there. \thmref{Prediction}{X.3} turns this into a permutation test.

\subsubsection{Seven local rules}\label{sec:merge-calculus}

Draw a transformer block as a string diagram (\bgref{B.3}), with a copy node where a wire feeds several boxes and a
product node in a gated block. An edit is a small box on a wire. To merge, slide it by equalities of maps until a
trainable box absorbs it; where it gets stuck names the failing side condition. A positional rotation (RoPE,
\citealp{xsu2024roformer}), a shared weight, a missing bias slot or a second reader of a wire can each block a slide, and
all four occur in production models.

\begingroup\renewcommand{\label}[1]{}%
\begin{analogy*}{what ``string diagram'' and ``spider'' mean here}
The diagrams are string diagrams of a symmetric monoidal category, read literally. Copy is the cartesian diagonal
$x\mapsto(x,x)$ (\bgref{B.4}), which commutes with every map, and the product node is $\odot$. ``Spider'' is only a name
for both; they fail the Frobenius law (\bgref{B.6}), which is never used.
\end{analogy*}
\endgroup

String diagrams for monoidal categories go back to \citet{xjoyal1991geometry}; spiders are from \citet{xcoecke2013new}.

\begin{theorem}{VI.3}{the merge calculus; known in part}
Work with activation-side diagrams in a symmetric monoidal category of continuous, piecewise-smooth maps with copy and
discard nodes. (We call copy and the Hadamard product ``spiders'' by analogy only; in this cartesian setting they do not
satisfy the Frobenius law.) A \emph{trainable box} is a node whose weight ranges over a stated class: all linear maps,
or all affine maps when the box has a bias slot. R2 and R4--R7 are identities of maps; R1 is a closure property and R3 a
characterisation.

\smallskip
{\upshape\footnotesize\renewcommand{\arraystretch}{1.15}\noindent
\begin{tabularx}{\linewidth}{@{}>{\raggedright\arraybackslash}p{0.17\linewidth}>{\raggedright\arraybackslash}p{0.33\linewidth}>{\raggedright\arraybackslash}X@{}}
\toprule\headrow
\textbf{Rule} & \textbf{Statement} & \textbf{When it holds}\\
\midrule
R1 (fusion) & composites and parallel sums of linear (affine) boxes are linear (affine) & the operators involved are
position-independent\\
R2 (permutation slide) & $P\sigma^{\times d}=\sigma^{\times d}P$ & every elementwise $\sigma$\\
R3 (positive-diagonal slide) & $\diag(\ell)\,\sigma^{\times d}=\sigma^{\times d}\diag(\ell)$ for all $\ell\ge0$ & iff
$\sigma$ is positively homogeneous (ReLU, leaky ReLU; not GELU, SiLU, tanh); such $\sigma$ are piecewise linear\\
R4 (spider slide) & $\diag(\ell)(a\odot b)=(\diag(\ell)a)\odot b$ & any $\ell$\\
R5 (cup slide) & $\langle Mq,k\rangle=\langle q,M^\top k\rangle$ & always. Under RoPE,
$\langle R_mMq,R_nk\rangle=\langle q,M^\top R_{n-m}k\rangle$ holds for every $M$, but the slide onto the key side,
$\langle R_mMq,R_nk\rangle=\langle R_mq,R_nM^\top k\rangle$ for all $m,n$, holds iff $M^\top$ commutes with every
$R_{n-m}$ that occurs; for diagonal $M=\diag(\ell)$ and distinct frequencies, iff $\ell$ is constant on each rotary pair
(in either pairing convention)\\
R6 (norm absorption) & $W(\gamma\odot n(x)+\beta)=(W\diag\gamma)\,n(x)+W\beta$ & always; the shift $W\beta$ needs a
bias slot, and an RMSNorm gain has no $\beta$\\
R7 (copy slide) & applying $D$ and then copying equals copying and then applying $D$ to every copy & always\\
\bottomrule
\end{tabularx}}
\smallskip

Consequences:

(a) \textbf{Affine edits.} Let a method insert a position-uniform affine edit $E_q$ on a wire $w$ of the base network
$f$, for every $q\in Q$. Suppose every reader of $w$ is a trainable, unshared box (no weight tying) whose class is closed
under precomposition with $E_q$; a shift needs a bias slot. Then the method merges, by $W_c\mapsto W_cE_q$ for all
readers $c$ at once. Dually, it merges into an untied trainable box that writes $w$ whenever $E_q$ composed with that box
stays in its class. Absorption into a box that the extension itself inserts does not count. Edits that violate these
hypotheses fall outside (a): on the reader side, R1--R7 do not merge position-selective edits or edits on a wire that is
also read by a skip connection, a norm or the network output. Whole-network non-mergeability is open
(Exposé~\ref{sec:analogy}). Position-selective edits (ReFT/LoReFT, and position-gated LoRA variants such as Activated
LoRA) are not box-locally mergeable. Residual-stream edits need case analysis: an edit merges into the box that produces
the wire when that box can absorb it (an untied embedding table; for a fixed-position edit at layer $0$, a learned
absolute positional table), a shift folds into a writer's bias when one exists, a diagonal edit after a post-LN folds
into that LN's affine parameters, and an orthogonal edit in an RMSNorm model merges globally, SliceGPT/QuaRot-style. No
merge is known for LoReFT's non-orthogonal low-rank edit.

(b) \textbf{Diagonal methods.} A diagonal method merges into a box reached by R1--R7 if all of the following hold: the
rules move \emph{every} copy of $\diag(\ell)$ into trainable boxes for all $\ell$ in the method's parameter domain (with
$\ell\ge0$ when R3 is used); no position-dependent operator lies on the path (for RoPE, $\ell$ is constant on each rotary
pair); the target box is unshared, or $\ell$ is constant across the sharing group; and every shift has a bias slot.
Conversely, the RoPE condition is necessary for sliding a query scaling onto $W_K$ when $W_Q$ and $W_K$ have full rank and
at least two occurring relative offsets have rotation angles that are not multiples of $\pi$. Without such genericity
hypotheses the conditions are not necessary: a dead rotary pair, or zero entries of $\ell$ under ReLU, can still merge.

(c) \textbf{Non-affine edits.} For a linear or affine box, an edit whose modified box function is non-affine in the box
input, for some parameter value, is not box-locally mergeable. For linear and affine boxes this is the definition; for
non-linear boxes such as attention, see \thmref{Theorem}{VI.5}(f).

(d) \textbf{Type relativity.} A generic high-precision update is off the quantisation grid, so exact merging of a QLoRA
or LoftQ update leaves the quantised type, a type-preserving merge is lossy, and QA-LoRA's group-constant adapters are
the exception (\thmref{Proposition}{V.2}(d)).
\end{theorem}

The rules are elementary, and the individual merges are known: (IA)$^3$ by \citet{liu2022few} (Sec.~3.3), SSF by
re-parameterisation in \citet{lian2022scaling}, and LoRA by linearity in \citet{hu2021lora}. Moving scales through ReLU,
SwiGLU's linear branch and norm gains, and the GELU obstruction, are in SmoothQuant \citep{xxiao2023smoothquant} and AWQ
\citep{xlin2024awq}. New here is only the packaging as one merge calculus for PEFT, with explicit side conditions. QLoRA
is \citet{dettmers2023qlora}, LoftQ \citet{li2023loftq}, QA-LoRA \citet{xu2023qa} and Activated LoRA
\citet{greenewald2025activated}. \thmref{Theorem}{VI.3}, \thmref{Proposition}{VI.4} and \thmref{Theorem}{VI.5} were
repaired after two rounds of adversarial review; the referee record (\S\ref{sec:analogy-referee}) lists what changed.

\noindent\labelsf{Num}\enspace Under RoPE, $\langle R_mMq,R_nk\rangle=\langle q,M^\top R_{n-m}k\rangle$ holds for a
generic $M$ to $10^{-15}$ while the slide onto the key side fails. A query scaling slid onto $W_K$ leaves a $7.6\%$
least-squares residual unless $\ell$ is pair-constant. Folding post-LN gains into $W_1$ changes the output. Legal
rewrites agree to $10^{-15}$ (Appendix~\ref{app:repro}).

Under SwiGLU, R4 moves $\ell$ onto the linear branch and R1 fuses it into $W_{\mathrm{up}}$, for $\ell$ of either sign:
\begin{equation}\label{eq:merge-swiglu}
  \diag(\ell)\big(\mathrm{silu}(W_gx)\odot W_ux\big)=\mathrm{silu}(W_gx)\odot\big(\diag(\ell)\,W_ux\big)
  =\mathrm{silu}(W_gx)\odot\big((\diag(\ell)\,W_u)\,x\big).
\end{equation}

\paragraph{Corollaries: scaling and reparametrisation methods}
All weight-space reparametrisation methods are mergeable, because their nonlinearities sit on the parameter side (Cayley
in OFT \citep{qiu2023controlling}, the norm in DoRA \citep{liu2024dora}, $u/\lVert u\rVert$ in HRA
\citep{yuan2024bridging}). In HF PEFT practice \citep{xmangrulkar2022peft}, an adapter on one side of a tied embedding
and LM head is not merged faithfully (\texttt{ensure\_weight\_tying=False} by default). The scaling methods merge as
follows.
\begin{itemize}
  \item (IA)$^3$ on keys and values is an output scaling of $W_K,W_V$ and fuses into them. On the FFN, HF applies it to
  the input of $W_{\mathrm{down}}$, and it fuses there. Both hold for every $\sigma$.
  \item Alternatively, the FFN scaling fuses into $W_{\mathrm{up}}$ under ReLU when $\ell\ge0$, a restriction of the
  domain (IA)$^3$ trains over, and \emph{under SwiGLU/GeGLU for every $\ell$}, by R4 applied to
  $\mathrm{silu}(W_gx)\odot W_ux$, as in \eqref{eq:merge-swiglu}. Under GELU it does \emph{not} fuse into
  $W_{\mathrm{up}}$, except for entries with $\ell_i\in\{0,1\}$, when the pre-activations range over $\R$ (the argument is
  in Appendix~\ref{proof:VI.3}). The $W_{\mathrm{up}}$ statement is practically idle, since $W_{\mathrm{down}}$ always
  works, and AWQ's code already migrates scales this way (into \texttt{up\_proj} under SwiGLU, into \texttt{fc1} under
  ReLU, and through an inserted scaled activation under GELU).
  \item A query scaling fuses into $W_Q$ (R1). It moves onto $W_K$ (R5) only without RoPE or with $\ell$ constant on
  rotary pairs, and only without grouped-query (GQA) sharing or with $\ell$ constant across the group.
  \item SSF fuses into the preceding linear box when that box has a bias slot; in bias-free LLaMA-type layers its shift
  needs a new bias. LN-tuning edits existing LN parameters, so it is trivially mergeable. Folding a LayerNorm's
  $\gamma,\beta$ into the following box works in pre-LN blocks, where the norm feeds only the next boxes, when that box
  has a bias slot for $W\beta$; an RMSNorm gain has no $\beta$ and folds without one (the SliceGPT/QuaRot fold). Neither
  folds into a tied embedding or LM head without breaking the tie.
\end{itemize}
Not box-locally mergeable: nonlinear Houlsby, Pfeiffer and parallel adapters; routed mixtures (MoLE, LoRAMoE, X-LoRA),
unless the gate is constant; AdapterFusion, even with a constant gate, since it combines nonlinear adapters; LST;
prefix-type methods (\thmref{Theorem}{VI.5}).

The nonlinear adapters are \citet{houlsby2019parameter}, \citet{pfeiffer2020adapterfusion} (with AdapterFusion) and the
parallel adapter of \citet{he2021unified}. The routed mixtures are MoLE \citep{wu2024mixture}, LoRAMoE
\citep{dou2023loramoe} and X-LoRA \citep{buehler2024x}. LST is \citet{sung2022lst} and LN-tuning \citet{zhao2023tuning}.
Masking (IA)$^3$ to a few neurons connects to the sparse activation scaling of \citet{xstoehr2024activation}.

\begin{explains}
Exact merging into existing weights is certified by local rewriting, a sufficient test together with box-local
obstructions, under explicit side conditions (position-uniformity, no sharing, bias slots), and the rewrite also tells
you \emph{which} weight absorbs the update. It is not a decision procedure for whole-network mergeability; a stuck rewrite
certifies only a box-local obstruction. RoPE pins a query scaling to $W_Q$ unless it is pair-constant. Bias-free layers
leave a shift no slot, and a new bias, though nearly free at inference, changes the architecture. Post-LN feeds the
normalised wire to the residual too, so R7 cannot carry its gains forward. Zero inference latency is a weaker goal (a new
bias costs almost nothing), and for quantised bases the type caveat (d) applies.
\end{explains}

\begin{remark*}{residual-stream edits that do merge}
Some edits outside (a) merge anyway, all position-uniform. A diagonal edit $h\mapsto s\odot h+t$ after a post-LN folds
into that norm's gain and shift, $\gamma'=s\odot\gamma$ and $\beta'=s\odot\beta+t$. A shift $h\mapsto h+t$ after a pre-LN
block with biases folds into its FFN output bias. An orthogonal edit $h\mapsto Qh$ after block $l$ of an untied
pre-RMSNorm model merges globally, SliceGPT-style: rotate the embedding table, fold the gains of blocks $1,\dots,l$ into
their input weights composed with $Q^\top$, compose their output weights with $Q$, and leave later blocks unchanged. No
merge is known for LoReFT's edit at interior positions.
\end{remark*}

\subsubsection{Linear insertions, and ReFT}\label{sec:merge-reft}

A linear inserted module fuses at once by R1, and the question is which weight-space method it becomes. ReFT, read one
hidden vector at a time, becomes LoRA on a box the network lacks, the identity, written in homogeneous coordinates
$(h,1)$ as $[I_d\mid0]$ so that its last column holds the bias.

\begin{proposition}{VI.4}{linear insertions, and ReFT}
(a) A linear adapter in parallel with a single linear box, $W_0x+sUDx$, \emph{is} LoRA: it has the same $\rho$
\citep[Table~1]{he2021unified}. A linear adapter parallel to a whole sublayer (He et al.'s parallel adapter around
attention or the FFN) edits the weightless skip connection rather than a base box, and is not box-locally mergeable when
the sublayer is non-affine.

(b) Let $W_0\in\R^{m\times n}$ be bias-free (as LLaMA-type $W_o$ and $W_{\mathrm{down}}$ are), and $r\ge1$. A serial
linear adapter after $W_0$, $(I+UD)W_0$ with $U\in\R^{m\times r}$ and $D\in\R^{r\times m}$ (scale absorbed), has image
\[W_0+\{Y\in\Mr:\operatorname{row}Y\subseteq\operatorname{row}W_0\}.\]
It is expressively equivalent to LoRA iff $\rk W_0=n$, and weakly (unpointed) isomorphic to LoRA iff $W_0$ is invertible.
The isomorphism $(U,D)\mapsto(U,DW_0)$ sends a base point $(U_0,D_0)$ to $(U_0,D_0W_0)$. For a fixed $\mathrm T_A$
pointing $(0,A_0)$ with $\rk A_0=r$ and $1\le r<m=n$, a pointed isomorphism exists iff $W_0$ is invertible and
$\operatorname{row}A_0=\operatorname{row}(D_0W_0)$, and then $U_0=0$ automatically. With a bias $b_0$, the serial adapter
also moves the bias by $UDb_0$, so on the affine box its image differs from weight-only LoRA's. Houlsby's near-identity
initialisation (weights from a zero-mean Gaussian with standard deviation $10^{-2}$, truncated at two standard
deviations) makes practical serial adapters slightly unpointed; the pointed statement applies to zero-up
initialisations.

(c) LoReFT \citep{wu2024reft} is $\Phi(h)=h+R^\top(Wh+b-Rh)$, with $R\in\R^{r\times d}$ having orthonormal rows, applied
at selected positions. On one hidden vector it is the extension by an \textbf{identity box} followed by a rank-$\le r$
affine edit,
\[\Phi(h)=\big(I+R^\top(W-R)\big)h+R^\top b,\]
which is LoRA$_r$ on the \emph{affine} identity box $[I_d\mid0]$ in homogeneous coordinates, with $B=R^\top$ constrained
to the Stiefel manifold and $A=[W-R\mid b]$. Its image is $(I,0)+\MM_{\le r}(d\times(d+1))$, of dimension $r(2d+1-r)$.
The bias is confined to $\operatorname{col}B$, so for $r<d$ this is not LoRA$_r{}\oplus{}$BitFit, whose dimension is larger
by $d-r$. It is simulated, through $h(R,W,b)=(R^\top,[W-R\mid b])$, by LoRA$_r$ on that affine box pointed at
$(R_0^\top,0)$, a $\mathrm T_B$ pointing with $\operatorname{col}B_0=\operatorname{row}R_0$; this homogeneous-coordinate
LoRA has a bias column in $A$, unlike HF LoRA. It is not simulated by $\mathrm T_A$-pointed LoRA$_r$, since its
$\dim S_1=r(d+1)$ exceeds $rd$. It is not isomorphic to LoRA$_r$: its parameter count is smaller by $r(r+1)/2$, and its
residual symmetry is $\Ort(r)$, acting by $(R,W,b)\mapsto(OR,OW,Ob)$. It is pointed iff $W=R$ and $b=0$. The reference
implementation (pyreft) initialises $W$ and $b$ by the default \texttt{nn.Linear} init, independently of $R$, so
practical LoReFT has a pointing defect $R^\top((W_0-R_0)h+b_0)$: at initialisation it replaces the $R$-component of $h$
by a random affine function, and the simulation above is then only a weak morphism. pyreft also applies dropout to the
whole output, $h$ included, so the training-time map is not $\Phi$.
\end{proposition}

\noindent\labelsf{Num}\enspace The Jacobian rank is $22=r(2d+1-r)$ at $d=6$, $r=2$, against $26$ for
LoRA$_r{}\oplus{}$bias. The serial adapter's image has dimension $7$ at $(m,n,r)=(5,3,1)$ and at $(4,8,1)$, against $11$
for LoRA at $(4,8,1)$ (Appendix~\ref{app:repro}).

\begin{explains}
Per hidden vector, LoReFT (and DiReFT) adds no new geometric type. Its image is that of LoRA$_r$ on an affine identity
box the architecture did not have. \citet{wu2024reft} already describe DiReFT as LoRA applied to hidden representations
at certain positions, and name position as what separates ReFT from PEFT; the new content here is the image dimension,
the bias confinement, the $\Ort(r)$ gauge and the initialisation defect. ReFT as a framework also allows nonlinear
interventions (pyreft's \texttt{act\_fn}), to which this does not apply. The main obstruction to merging is
position-selectivity: LoReFT acts only at selected prompt positions, outside \thmref{Theorem}{VI.3}(a), so no
position-independent weight can absorb it. Even applied at every position, at interior residual-stream sites the stream
is read by the skip connection and by every later block, and no linear box of the original architecture sits there (at
the embedding output an untied embedding table could absorb a position-uniform edit, \thmref{Theorem}{VI.3}(a)). By
\thmref{Definition}{VI.1} it is box-locally mergeable into its own inserted box, which does not help.
\end{explains}

LoReFT, DiReFT and the reference implementation pyreft are due to \citet{wu2024reft}. Serial adapters are due to
\citet{houlsby2019parameter}. The parallel form, and the reading of LoRA as a linear adapter parallel to one box, are in
\citet{he2021unified}. LoRA is \citet{hu2021lora}.

\subsubsection{Attention is a homomorphism followed by a perspective map}\label{sec:merge-attention}

Fix a query $q$. Attention over a context $K$ of key--value pairs computes
\begin{equation}\label{eq:merge-attn}
  \mathrm{Attn}_q(K)=\frac{\sum_{(k,v)\in K}e^{\langle q,k\rangle/\sqrt d}\,v}{\sum_{(k,v)\in K}e^{\langle q,k\rangle/\sqrt d}}.
\end{equation}
The pair (total weight, weighted sum of values) is additive in the context, and only the division $(Z,S)\mapsto S/Z$, the
perspective map of convex analysis (\bgref{D.34}), is nonlinear. With a prefix $P$ before the content $C$, the output
averages what $P$ alone and $C$ alone would give, weighted by their shares of the mass, so the prefix is a parallel
branch gated by its share.

\begin{figure}[tbp]
\centering
\includegraphics[width=\linewidth]{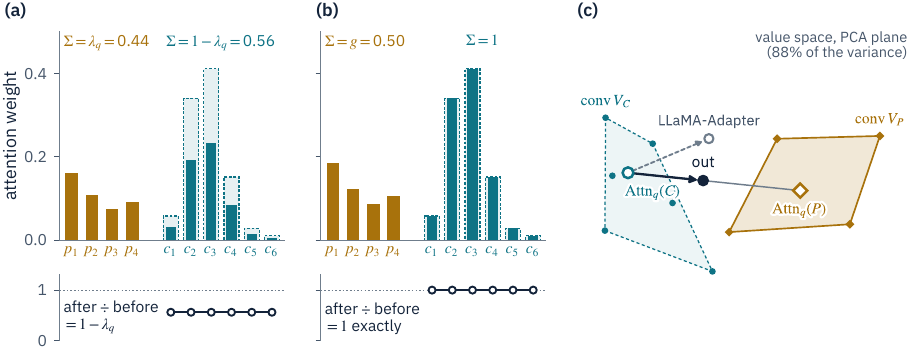}
\caption{\textbf{A prefix is a gate} (\thmref{Theorem}{VI.5}). One query $q$ attends to six content tokens $C$ and four prefix tokens $P$, all seeded synthetic data. (a)~Prefix tuning \citep{li2021prefix}, one softmax over $P\uplus C$. The prefix (ochre) takes the mass $\lambda_q=Z_q(P)/(Z_q(P)+Z_q(C))$, and each content weight (teal) is its value without the prefix (dashed outline) times $1-\lambda_q$. The strip below plots the ratio after/before for every content token; it is the same for all six, so the content weights keep their ratios and their order (part~(c), due to \citet{petrov2023prompting}). (b)~LLaMA-Adapter \citep{zhang2023llama} normalises $P$ and $C$ separately and adds the prefix branch with a gate $g$: the prefix weights sum to $g$, and the content weights equal the no-prefix ones exactly (part~(d)). (c)~The value vectors projected on their two leading principal axes, with the hulls of the content values (dashed) and of the prefix values (ochre). The prefix-tuning output (black dot) lies on the segment from $\mathrm{Attn}_q(C)$ to $\mathrm{Attn}_q(P)\in\mathrm{conv}\,V_P$, the fraction $\lambda_q$ of the way (parts~(b) and~(c); the identity~(b) is due to \citet{he2021unified}), and a larger $\lambda_q$ moves it further towards the prefix hull. The LLaMA-Adapter output $\mathrm{Attn}_q(C)+g\,\mathrm{Attn}_q(P)$ (open grey) is an additive step that leaves the segment. The rigidity holds within one layer, relative to the same layer without the prefix: prefixes at lower layers change the queries, keys and values of higher ones. The values used are in \S\ref{app:repro-figures}.}
\label{fig:prefix}
\end{figure}

\begin{theorem}{VI.5}{attention is a homomorphism followed by a perspective map; known in part}
Fix a query $q$. Multisets of key--value pairs form a commutative monoid under $\uplus$. Define
\[\Sigma_q(K)=\Big(\sum e^{\langle q,k\rangle/\sqrt d},\ \sum e^{\langle q,k\rangle/\sqrt d}\,v\Big).\]

(a) $\Sigma_q$ is a monoid homomorphism into $(\R_{\ge0}\times\R^{d_v},+)$, landing in $\{Z>0\}\cup\{(0,0)\}$. On
nonempty multisets, $\mathrm{Attn}_q=\pi\circ\Sigma_q$ with the partial map $\pi(Z,S)=S/Z$, defined for $Z>0$.

(b) \textbf{Prefix is a gated parallel adapter.} For nonempty $P$ and $C$,
\[\mathrm{Attn}_q(P\uplus C)=(1-\lambda_q)\,\mathrm{Attn}_q(C)+\lambda_q\,\mathrm{Attn}_q(P),\qquad
\lambda_q=\frac{Z_q(P)}{Z_q(P)+Z_q(C)}.\]

(c) \textbf{Per-layer pattern rigidity.} Within a single layer, for fixed $q$ and a fixed content multiset $C$: the
content attention weights are multiplied by $1-\lambda_q$, their ratios are unchanged, and the output lies on the segment
from $\mathrm{Attn}_q(C)$ to $\mathrm{Attn}_q(P)\in\operatorname{conv}V_P$, where $V_P$ is the set of prefix values. This
rigidity is relative to the same layer's computation without the prefix; it is relative to the pretrained pattern only at
the lowest prefixed layer. Lower-layer prefixes change the queries and the content keys and values of higher layers, so
attention patterns at the level of the network are \emph{not} rigid (\citealp[§6]{petrov2023prompting}; the same authors
show that prefix-tuning a pretrained transformer can be a universal approximator, \citealp{petrov2024prompting}).

(d) \textbf{LLaMA-Adapter} normalises $P$ and $C$ separately and adds a zero-initialised gate. It computes
$\mathrm{Attn}_q(C)+\tanh(g_l)\,\mathrm{Attn}_q(P)$, with one gate per head in the paper; HF PEFT uses one raw
zero-initialised scalar per layer, without $\tanh$. It uses two perspective maps, leaves the content weights exactly
unchanged, and is pointed at $g=0$. Its prompt keys and values are the frozen $W_k,W_v$ applied to one learned prompt per
adapted layer, and it acts in the top layers only.

(e) \textbf{Prompt $\to$ deep prefix in causal decoders.} Assume that the soft prompt comes first, that attention is
causal, and that content positions are offset by the prompt length $m$ in both methods (as HF PEFT does via
\texttt{cache\_position}). Then at every layer the states at prompt positions depend only on $p$, and
$h:p\mapsto(K_j(p),V_j(p))_j$ satisfies
\[\Phi\circ\rho_{\mathrm{prefix}}\circ h=\Phi\circ\rho_{\mathrm{prompt}}\]
as functions of the content tokens, read at content positions. The two methods live on different extensions, so this is
an arrow of the unpointed slice over the realisation $C^\infty(X_{\mathrm{content}},Y_{\mathrm{content}})$, not of
$\Meth$. Under RoPE without the offset, a different $h$ works: rotate each prefix key by $-m$ (positions
$-m,\dots,-1$). For additive relative biases (ALiBi, T5) the answer depends on how the implementation places the prefix
keys. The construction does not apply in bidirectional encoders of depth $\ge2$.

(f) \textbf{Obstructions.} Call a head \emph{standard} if it is vanilla softmax attention without torch-style
\texttt{add\_bias\_kv} (which would absorb a one-token prefix), \texttt{add\_zero\_attn}, learned attention sinks or
softmax-plus-one. On length-one contexts, every weight setting of a standard head is affine in $x$. Some prefix makes the
prefixed head non-affine on length-one contexts iff $W_o\ne0$ and ($W_q\ne0$ or $W_k^\top b_q\ne0$); for bias-free heads,
iff $W_o\ne0$ and $W_q\ne0$. For such frozen weights and generic prefixes the prefixed head is not affine, because the
gate $\lambda(x)$ is not constant. So prefix tuning and P-tuning v2 are not box-locally mergeable into a standard head.
When $W_oW_v\ne0$, no prefix is neutral, so prompt and prefix tuning are unpointed.
\end{theorem}

The identity is standard. Part (b) is due to \citet{he2021unified}; see also \citet{chen2022inducer}. Part (c) and the
soft-prompt-to-prefix map in (e) are due to \citet[§2.2 and §6]{petrov2023prompting}, informally already
\citet[§7.2]{li2021prefix}. LLaMA-Adapter is \citet{zhang2023llama}, P-tuning v2 is \citet{liu2021p}, and prompt tuning
is \citet{lester2021power}; on what prompts can express, see \citet{wang2023universality}. Ours are the monoid
organisation, the explicit position hypotheses and the slice-category typing of (e).

\noindent\labelsf{Num}\enspace On a 3-layer, 2-head causal decoder, prompt-to-prefix agreement is $9\times10^{-16}$ with
RoPE and the offset, against $1.3$ without it and $2\times10^{-15}$ with keys rotated by $-m$. A head with $W_o=0$ stays
affine although $W_q^\top W_k\ne0$, and one with $W_q\ne0$, $W_q^\top W_k=0$ does not (referee replication). The identities
(a)--(d) and the non-affinity in (f) are checked to rounding error (Appendix~\ref{app:repro}); Figure~\ref{fig:prefix}
draws them for one query.

\begin{explains}
At the lowest prefixed layer the content attention equals the pretrained one, rescaled by $1-\lambda_q$. At higher layers
it changes only through the modified inputs, to first order additively in each position
\citep[§6]{petrov2023prompting}, so new query--key interactions cannot be imposed directly at a layer. No network-level
impossibility follows. LLaMA-Adapter's zero-init attention is a structural change; categorically, its contribution is
\emph{pointedness}. Prompt tuning maps into P-tuning v2 in causal decoders. \citet{petrov2023prompting} and
\citet{li2021prefix} assert that the inclusion is strict. It is strict in a one-layer, one-token example, where a prompt
fixes the prefix value as a function of its key, and we do not prove the generic statement. Encoders remain open.
\end{explains}

\subsubsection{Where each method lands}\label{sec:merge-summary}

Table~\ref{tab:merge-summary} collects the verdicts of this exposé.

\begin{table}[tp]
\centering\footnotesize\renewcommand{\arraystretch}{1.15}
\caption{Extensions and their merge types. The verdicts are box-local (\thmref{Definition}{VI.1}), and the attention
results are per layer.}\label{tab:merge-summary}
\begin{tabularx}{\linewidth}{@{}>{\raggedright\arraybackslash}p{0.18\linewidth}>{\raggedright\arraybackslash}p{0.21\linewidth}>{\raggedright\arraybackslash}p{0.16\linewidth}>{\raggedright\arraybackslash}p{0.07\linewidth}>{\raggedright\arraybackslash}X@{}}
\toprule\headrow
\textbf{Method} & \textbf{What it adds} & \textbf{Neutral point} & \textbf{Merge} & \textbf{Reason}\\
\midrule
LoRA, DoRA, OFT, HRA & nothing; weight-space reparametrisations & their pointing in $\Meth$ & $\mathsf{M1}$ &
nonlinearity on the parameter side; $\mu=\rho$\\
(IA)$^3$ on $k$, $v$ & output scaling of $W_K$, $W_V$ & $\ell=\mathbf 1$ & $\mathsf{M1}$ & R1\\
(IA)$^3$ on the FFN & input scaling of $W_{\mathrm{down}}$ & $\ell=\mathbf 1$ & $\mathsf{M1}$, $\mathsf{M1}{\rightarrow}$
& R1 into $W_{\mathrm{down}}$; R4 into $W_{\mathrm{up}}$ under SwiGLU, R3 under ReLU with $\ell\ge0$\\
SSF & scale and shift after a box & $\gamma=\mathbf 1$, $\beta=0$ & $\mathsf{M1}{\rightarrow}$ & R1 into the preceding box
if it has a bias slot\\
LN-tuning & new values of existing $\gamma,\beta$ & the pretrained values & $\mathsf{M1}$ & in place; folding forward
needs pre-LN, and $\beta$ a bias slot\\
linear adapter parallel to one box & $sUDx$ beside $W_0x$ & $U=0$ & $\mathsf{M1}$ & \thmref{Proposition}{VI.4}(a): it is
LoRA\\
serial linear adapter & $(I+UD)$ after $W_0$ & $U_0D_0W_0=0$ & $\mathsf{M1}$ & \thmref{Proposition}{VI.4}(b)\\
Houlsby, Pfeiffer, nonlinear parallel adapters & $U\sigma(D\,\cdot)$ & $U=0$; Houlsby's init starts near it &
$\mathsf{M}\infty$ & non-affine edit, \thmref{Theorem}{VI.3}(c)\\
routed mixtures (MoLE, LoRAMoE, X-LoRA) & input-dependent gates over LoRA modules & varies & $\mathsf{M}\infty$ &
\thmref{Theorem}{VI.3}(c); mergeable if the gate is constant\\
AdapterFusion, LST & a gated mix of nonlinear adapters; a side network & none & $\mathsf{M}\infty$ & non-affine edit,
\thmref{Theorem}{VI.3}(c)\\
LoReFT & identity box with a rank-$\le r$ affine edit, at selected positions & $W=R$, $b=0$; pyreft's init misses it &
$\mathsf{M}\infty$ & position-selective, outside \thmref{Theorem}{VI.3}(a); no merge known\\
prefix tuning, P-tuning v2 & key--value pairs in every layer & none, generically & $\mathsf{M}\infty$ & box-locally,
\thmref{Theorem}{VI.5}(f)\\
prompt tuning & soft tokens at the input & none, generically & $\mathsf{M}\infty$ & a deep prefix in causal decoders
(\thmref{Theorem}{VI.5}(e)), so \thmref{Theorem}{VI.5}(f)\\
LLaMA-Adapter & a separately normalised, gated prompt in the top layers & $g=0$, \thmref{Theorem}{VI.5}(d) &
$\mathsf{M}\infty$ & as in \thmref{Theorem}{VI.5}(f), once the prompt has two or more tokens\\
QLoRA, LoftQ & nothing; LoRA on a quantised base & pointed off $\theta_0$, with a defect & $\mathsf{Mq}$ &
\thmref{Theorem}{VI.3}(d), \thmref{Proposition}{V.2}(d); QA-LoRA stays on the grid\\
\bottomrule
\end{tabularx}
\end{table}

The verdicts are box-local, and the attention results are per layer. Two questions remain open
(Exposé~\ref{sec:analogy}): whether prefix tuning is non-mergeable for the whole network, and whether prompt tuning is
strictly weaker than P-tuning v2 in general causal decoders, together with how the two relate in encoders.

\section{Worked examples}\label{sec:examples}

\newcommand{\exVIIlabel}[1]{{\sffamily\footnotesize\bfseries\color{inkgray}\textls[80]{\MakeUppercase{#1}}}}
\newcommand{\exVIIsrc}[2]{{\small\color{inkgray}#1\enspace\textperiodcentered\enspace{\sffamily\footnotesize #2}}}
\newcommand{\exVIIrefs}[1]{\par\smallskip\noindent{\footnotesize\color{inkgray}{\sffamily\bfseries Results.}\ #1}}
\newenvironment{exVIIlines}
  {\begin{description}[leftmargin=0pt,labelindent=0pt,itemsep=3pt,topsep=2pt,font=\exVIIlabel]}
  {\end{description}}

This exposé reads fourteen methods through the three invariants and the merge layer. For each method it says what the
method reaches, how training moves it, what a moved base point does and how it merges, and it names the numbered
result behind every claim.

A method is a pointed map $\rho:(Q,q_0)\to(\Theta,\theta_0)$. Its image germ says what it can express, its fibres
together with the optimizer's metric say how it moves, base change says what a moved $\theta_0$ does, and extensions
cover adapters and prompts. The fourteen cards below apply that toolkit; Exposé~\ref{sec:atlas} turns the answers into
coordinates and arrows.

Each card gives the object $(Q,q_0,\rho)$ for one linear layer $y=Wx$ with $W\in\R^{m\times n}$, and then reads it
through the invariants. The \exVIIlabel{Image} line is the image germ at $\theta_0$ (Exposé~\ref{sec:image}); the
\exVIIlabel{Fibre} line is the fibre, the gauge and the induced metric (Exposés~\ref{sec:fibres} and~\ref{sec:lens});
the \exVIIlabel{Base} line is base change, meaning pointings, quantised bases and merge-and-restart
(Exposé~\ref{sec:base}); and the \exVIIlabel{Merge} line is the extension and the merge type
(Exposé~\ref{sec:merge}). The prefix card adds an \exVIIlabel{Arrows} line. Lines that an invariant leaves empty or
trivial are omitted. The merge types are those of Exposé~\ref{sec:atlas}, where M1 merges into its own box, M1$\to$ into a
neighbouring box under side conditions, Mq leaves the quantised type when merged exactly, and M$\infty$ is not
box-locally mergeable. The header of each card names the source paper and the modification kind used in the
catalogue, and its last line lists the results its claims rest on. Their complete proofs are in
Appendix~\ref{app:proofs}; claims marked [Num] are numerical checks (Appendix~\ref{app:repro}).

\paragraph{Credits}
Balancedness is due to \citet{du2018algorithmic} and \citet{arora2018optimization}, conserved charges to
\citet{zhao2022symmetries}, and the balanced slice to \citet{xkempf1979length}. The crossover is the closed form of
\citet{xtarmoun2021understanding} and \citet{xmin2021explicit}. The comparison of zero-$B$ and zero-$A$ starts is due to
\citet{hayou2024impact}, and scaled GD to \citet{xtong2021accelerating} and \citet{zhang2024riemannian}. GaLore as
one-sided LoRA with carried state is due to \citet{xtorrobahennigen2025duality}. Prefixes as gated adapters are due to
\citet{he2021unified}, and the rigid content ratios and the prompt-to-prefix map to \citet{petrov2023prompting}. The
merge of (IA)$^3$ into $W_{\rm down}$ is in \citet{liu2022few}, and scale migration through SwiGLU is used by AWQ
\citep{xlin2024awq}. LoKr's transport is the rearrangement of \citet{loan1993approximation}. The $2r$ bound underlies
the PiSSA-to-LoRA conversion of \citet{meng2024pissa}.

\begin{example*}{LoRA}\phantomsection\label{ex:lora}%
\exVIIsrc{\citet{hu2021lora}}{additive}
\[
  \rho(B,A)=W_0+sBA,\qquad q_0=(0,A_0),\qquad B\in\R^{m\times r},\ A\in\R^{r\times n}.
\]
\begin{exVIIlines}
\item[Image] The cone $W_0+\Mr$, of dimension $r(m+n-r)$ for $r<\min(m,n)$. With $\rk A_0=r$ the first-order image is
  $S_1=\{XA_0\}$, of dimension $mr$, so $\theta_0$ is an apex with deficit $r(n-r)$ (Figure~\ref{fig:apex}a).
\item[Fibre] Gauge $\GL_r$, acting by $(B,A)\mapsto(Bg^{-1},gA)$. Under gradient flow with block learning rates
  $\eta_B,\eta_A$ and no dropout, the induced cometric is $K(G)=s^2(\eta_B\,GA\T A+\eta_A\,BB\T G)$. The Lie algebra splits as $\gl_r=\so_r\oplus\Sym_r$. The
  isometric half $\so_r$ leaves $K$ unchanged, and the $\Sym_r$ half carries the charge
  $\Phi=B\T B/\eta_B-AA\T/\eta_A$, conserved under gradient flow and equal to $-A_0A_0\T/\eta_A$ from the zero-$B$
  start. For isotropic $\Phi=-\kappa I$ the flow is solved in closed form
  \citep{xtarmoun2021understanding,xmin2021explicit}. With $\mu=s\sqrt{\eta_A\eta_B}$ and the crossover scale
  $\sigma^*=\mu\lvert\kappa\rvert/2$, zero-$B$ LoRA is one-sided well below $\sigma^*$, where effectively only $B$ moves
  as in LoRA-FA, and well above it follows the balanced dynamics of \citet{arora2018optimization}
  (Figure~\ref{fig:noether}). On the full-rank locus, with channel-uniform hyperparameters, SGD keeps $\Ort(r)$ of the
  gauge, Adam and AdamW keep the signed permutations $B_r$, and undamped scaled GD keeps all of $\GL_r$
  (Figure~\ref{fig:gauge}). Under Adam with $\varepsilon=0$, no weight decay, no gradient clipping and a learning-rate
  schedule shared by both blocks, LoRA+ with learning-rate ratio $\lambda$ follows exactly the trajectory of LoRA with
  $\alpha$ multiplied by $\lambda$ \citep{schulman2025lora} (Figure~\ref{fig:orbit}).
\item[Base] $K$ merge-and-restart cycles reach exactly $W_0+\MM_{\le\min(Kr,m,n)}$, so all of $\Theta$ at
  $K^*=\lceil\min(m,n)/r\rceil$. Exact full-rank pointings form three strata (zero $B$, zero $A$, and a split that
  subtracts $B_0A_0$ from the frozen weight). Their isomorphism classes are fixed by $\operatorname{row}A_0$,
  $\operatorname{col}B_0$ and $B_0A_0$ respectively, so as a set they are $\Gr(r,n)\sqcup\Gr(r,m)\sqcup\MM_r$.
\item[Merge] M1, since the trained update $sBA$ adds to $W_0$.
\end{exVIIlines}
\exVIIrefs{\thmref{Theorem}{II.2} \textperiodcentered{} \thmref{Proposition}{II.3} \textperiodcentered{}
  \thmref{Observation}{III.2} \textperiodcentered{} \thmref{Theorem}{III.5} \textperiodcentered{}
  \thmref{Corollary}{III.6} \textperiodcentered{} \thmref{Theorem}{III.9} \textperiodcentered{}
  \thmref{Corollary}{III.11} \textperiodcentered{} \thmref{Theorem}{III.12} \textperiodcentered{}
  \thmref{Theorem}{V.3}}
\end{example*}

\begin{example*}{DoRA}\phantomsection\label{ex:dora}%
\exVIIsrc{\citet{liu2024dora}}{hybrid}
\[
  \rho(\mu,B,A)=\diag(\mu)\,N(W_0+sBA),\qquad q_0=(\mu_0,0,A_0),\qquad \mu_{0,i}=\lVert w_{0,i}\rVert.
\]
Here $N$ normalises each row to unit length and $w_{0,i}$ is the $i$-th row of $W_0$.
\begin{exVIIlines}
\item[Image] For $W_0$ without zero rows and $r\le\min(m,n)$ the image is
  $\mathrm{Diag}_m\cdot(W_0+\Mr)=\{\diag(\ell)(W_0+X):\rk X\le r\}$, exactly LoRA followed by a free per-output gain.
  Its dimension is $\min(mn,\,r(m+n-r)+m)$ for generic $W_0$ [Num], and smaller for special $W_0$, for instance
  $r(m+n-r)+m-r$ when $W_0$ has rank one. Its updates can exceed rank $r$. From $B_0=0$ the base point is an apex, as
  for LoRA.
\item[Fibre] The gauge contains $\GL_r$, acting on $(B,A)$ as for LoRA. For generic $W_0$ with $(m-r)(n-r)\ge m$ the
  budget $r(m+n)+m$ exceeds the dimension by exactly $r^2$, so $\GL_r$ is the whole continuous gauge. Gradient flow
  conserves the LoRA charge $\Phi$, also through the detached norm. The two backward rules give different geometries.
  The exact gradient of the row normalisation projects the direction gradient row by row, which is a change of metric.
  The detached backward of the DoRA paper (its Sec.~4.3) and of HF PEFT (\texttt{weight\_norm.detach()}) removes that
  projection and gives a non-gradient update, which need not be a descent direction. Under AdamW, which both the paper
  and HF use, neither variant follows a fixed metric, and the magnitude coordinates matter in their own right. Since $\mu_i$
  has size about $\lVert w_{0,i}\rVert$ and an output gain $\ell_i$ has size $1$, an equal step on each changes the
  row gain by different amounts.
\item[Merge] M1. Since the images agree, any gap between DoRA and LoRA plus an output gain, started at the same
  weights, comes from training dynamics. Experiment~D (\thmref{Proposition}{II.8}, \thmref{Prediction}{X.2}) separates
  its three possible sources, the parametrisation, the backward rule and the optimizer coordinates.
\end{exVIIlines}
\exVIIrefs{\thmref{Proposition}{II.8} \textperiodcentered{} \thmref{Proposition}{II.12} \textperiodcentered{}
  \thmref{Theorem}{III.5} \textperiodcentered{} \thmref{Prediction}{X.2}}
\end{example*}

\begin{example*}{OFT\,/\,BOFT}\phantomsection\label{ex:oft}%
\exVIIsrc{\citet{qiu2023controlling}; \citet{liu2023parameter}}{multiplicative}
\[
  W=W_0R\T\ \ \text{(OFT)},\qquad W=\diag(s)\,W_0R\T\ \ \text{(BOFT)},\qquad R_0=I,\ s_0=\mathbf 1.
\]
Here $R$ is block-diagonal with Cayley blocks $C(Q)=(I+Q)(I-Q)^{-1}$, $Q$ skew-symmetric of size $b$ (OFT), or a
product of $m'$ butterfly factors (BOFT).
\begin{exVIIlines}
\item[Image] With exact Cayley maps the image is an orbit in the level set of the neuron Gram matrix
  $K_{\rm out}=WW\T$, so every neuron norm, every angle between neurons and every singular value of $W_0$ stays fixed,
  and block OFT also fixes the $n/b$ partial Grams $W_{:,J}W_{:,J}\T$. BOFT has
  $K_{\rm out}(W)=\diag(s)K_{\rm out}(W_0)\diag(s)$. It keeps $\lvert\cos\rvert$ between neurons when every
  $s_i\ne0$, and the cosines themselves when all $s_i$ share a sign; with one butterfly factor it sends each block Gram
  to its $\diag(s)$-rescaling. HF's default Cayley--Neumann OFT, $R=C(Q)(I-Q^4)$, is only approximately orthogonal,
  since $R\T R=(I-Q^4)^2$. The base point is regular. For BOFT this holds when the factors' plane sets are disjoint
  (for instance $b=2$), so that the factor Lie algebras form a direct sum.
\item[Fibre] For exact-Cayley block OFT the gauge is the stabiliser of $W_0$ in the block group, which is trivial
  when $W_0$ is injective.
\item[Base] Exact-Cayley block OFT on a fixed partition is idempotent up to closure, and stays in
  $W_0\cdot\prod\SO(b)$ however often it is merged. Iterated BOFT with $m'$ factors reaches
  $\diag\cdot W_0\cdot P\big(\prod\SO(b\,2^{m'-1})\big)P\T$ for a fixed permutation $P$, and, for injective $W_0$,
  $\diag\cdot W_0\cdot\SO(n)$ only at full butterfly depth. At $n=64$
  and $b=4$ the block hypergraph has components of sizes $4,8,16,32,64$ for $m'=1,\dots,5$, so it is connected only at
  $m'=5$ [Num] (Figure~\ref{fig:rebase}b). With small generators ($\lVert Q\rVert_{\rm op}\le2^{1/4}$) each merge of
  HF's default is a contraction, so re-merging can only shrink neuron norms and singular values.
\item[Merge] M1.
\end{exVIIlines}
\exVIIrefs{\thmref{Theorem}{II.5} \textperiodcentered{} \thmref{Corollary}{II.6} \textperiodcentered{}
  \thmref{Theorem}{V.3}}
\end{example*}

\begin{example*}{(IA)$^3$}\phantomsection\label{ex:ia3}%
\exVIIsrc{\citet{liu2022few}}{multiplicative}
\[
  W=\diag(\ell)\,W_0\ \ (W_K,\,W_V),\qquad W=W_0\,\diag(\ell)\ \ (W_{\rm down}),\qquad \ell_0=\mathbf 1.
\]
\begin{exVIIlines}
\item[Image] Linear, the closure of the orbit of $W_0$ under a torus of one-sided diagonal scalings. Zeros of $W_0$
  stay zero, and row lines (column lines on $W_{\rm down}$) stay fixed while all $\ell_i\ne0$. The image lies inside
  that of rank-1 HiRA, $\mathrm{HiRA}_1$. Its covariance group is $\mathrm{Mon}\times\GL$ on the output-scaled boxes
  and $\GL\times\mathrm{Mon}$ on $W_{\rm down}$.
\item[Base] Idempotent.
\item[Merge] M1$\to$. It fuses into $W_K$, $W_V$ and $W_{\rm down}$ for every activation. The FFN scale fuses
  alternatively into $W_{\rm up}$ under SwiGLU or GeGLU for every $\ell$, through the linear branch, a scale migration
  that AWQ already uses; under ReLU for $\ell\ge0$; and under GELU only for entries $\ell_i\in\{0,1\}$.
\end{exVIIlines}
\exVIIrefs{\thmref{Theorem}{II.5} \textperiodcentered{} \thmref{Proposition}{II.9} \textperiodcentered{}
  \thmref{Theorem}{VI.3}}
\end{example*}

\begin{example*}{VeRA}\phantomsection\label{ex:vera}%
\exVIIsrc{\citet{kopiczko2023vera}}{additive}
\[
  \rho(b,d)=W_0+\Lambda_bB\Lambda_dA,\qquad q_0=(0,d_0).
\]
Here $\Lambda_b=\diag(b)\in\R^{m\times m}$ and $\Lambda_d=\diag(d)\in\R^{r\times r}$ are trained, while
$B\in\R^{m\times r}$ and $A\in\R^{r\times n}$ are frozen, random and shared across layers.
\begin{exVIIlines}
\item[Image] A toric cone of dimension $m+r-1$ [Num], with apex at $\theta_0$. At $b_0=0$ only $b$ receives a
  gradient, so the first-order image has dimension at most $m$, below $m+r-1$ when $r\ge2$.
\item[Fibre] The gauge is $\R^\times$, with $c$ acting by $(b,d)\mapsto(cb,d/c)$, and gradient flow conserves the
  charge $\lVert b\rVert^2/\eta_b-\lVert d\rVert^2/\eta_d$.
\item[Base] If $B$ has no zero entry, rebasing reaches exactly LoRA-FA$(A)$, the method $W_0+\{XA\}$, which is the
  target of the arrow $h(b,d)=\Lambda_bB\Lambda_d$ from VeRA to LoRA-FA$(A)$.
\item[Merge] M1.
\end{exVIIlines}
\exVIIrefs{\thmref{Lemma}{II.4} \textperiodcentered{} \thmref{Proposition}{II.12} \textperiodcentered{}
  \thmref{Theorem}{III.5} \textperiodcentered{} \thmref{Theorem}{V.3}}
\end{example*}

\begin{example*}{LoHa\,/\,LoKr}\phantomsection\label{ex:loha}%
\exVIIsrc{\citet{hyeonwoo2021fedpara}; \citet{yeh2023navigating}}{additive}
\[
  W_0+(B_1A_1)\odot(B_2A_2)\ \ \text{(LoHa)},\qquad W_0+C\otimes BA\ \ \text{(LoKr)}.
\]
\begin{exVIIlines}
\item[Image] LoHa$_{r,r}$ has updates of rank at most $\min(r^2,m,n)$, by the cut-rank bound. Its dimension is at most
  $\min\big(mn,\,2r(m+n)-2r^2-(m+n-1)\big)$, with equality below $mn$ [Num]. The gauge is $\GL_r\times\GL_r$ times the
  two-sided diagonal torus $(B_1,A_1,B_2,A_2)\mapsto(D_1B_1,A_1D_2,D_1^{-1}B_2,A_2D_2^{-1})$, and their joint generic
  orbits have dimension $2r^2+m+n-1$. HF zero-initialises one factor, so the base point is an apex at which only that
  factor moves to first order. At the common budget $2r(m+n)$ of LoHa$_{r,r}$ and LoRA$_{2r}$ three regimes occur. For
  $r\le2$, $r^2\le2r$, so the image of LoHa$_{r,r}$ lies inside that of LoRA$_{2r}$, strictly when $2r^2<m+n-1$. For
  $r\ge3$ and $2r^2<m+n-1$, LoHa trades at least $(m+n-1)-2r^2$ dimensions for rank up to $r^2$. When
  $2r^2\ge m+n-1$, for instance at $r=64$ on $4096\times4096$ or at LyCORIS-style $r=32$ on $320$- to $640$-wide
  layers, it matches or exceeds LoRA$_{2r}$ in dimension as well as in rank. At $131{,}072$ parameters on a
  $4096\times4096$ matrix, LoRA$_{16}$ reaches $130{,}816$ dimensions and LoHa$_{8,8}$ at most $122{,}753$
  (Figure~\ref{fig:rankvol}).
\item[Image] A $k$-term Kronecker sum $\sum_iC_i\otimes D_i$ is a matrix of rank at most $k$ after the Van
  Loan--Pitsianis rearrangement $\mathcal R$, which sends $C\otimes D$ to $\vect(C)\vect(D)\T$. So LoKr is a
  constrained LoRA$_1$ transported by $\mathcal R^{-1}$, which is LoRA across a different cut of the same four-leg
  tensor. Write $m=m_1m_2$ and $n=n_1n_2$. For $C\in\R^{m_1\times n_1}$ and inner rank $r'\le\min(m_2,n_2)$, LoKr's
  maximum rank is $\min(m_1,n_1)\,r'$, attained. It lies inside zero-init LoRA$_r$ iff $r\ge\min(m_1,n_1)\,r'$, and
  is otherwise incomparable with it when $\min(m_1,m_2)\min(n_1,n_2)>1$ (Figure~\ref{fig:cut}). In HF PEFT the second
  factor is low-rank only when $r<\max(m_2,n_2)/2$; above that LoKr is an unconstrained Kronecker product.
\item[Merge] M1.
\end{exVIIlines}
\exVIIrefs{\thmref{Theorem}{II.10} \textperiodcentered{} \thmref{Theorem}{II.11} \textperiodcentered{}
  \thmref{Proposition}{II.12} \textperiodcentered{} \thmref{Proposition}{II.13}}
\end{example*}

\begin{example*}{FourierFT}\phantomsection\label{ex:fourierft}%
\exVIIsrc{\citet{gao2024parameter}}{additive}
\[
  W_0+\alpha\,\operatorname{Re}F^{-1}(c),\qquad c\ \text{supported on $n_f$ fixed frequencies}.
\]
Here $F$ is the two-dimensional discrete Fourier transform and $\alpha$ a fixed scaling (HF's default $150$).
\begin{exVIIlines}
\item[Image] A coordinate mask transported into the real cosine frame. Its image lies in the point-symmetric
  subspace $\Delta W_{j,l}=\Delta W_{-j,-l}$. Started at $\theta_0$ ($c_0=0$, HF's \texttt{init\_weights=True}) it is linear
  and regular. Its dimension is $n_f$ minus the number of sampled conjugate pairs $(u,v),(-u,-v)$, each of which adds
  one gauge dimension. At LLM sizes this is close to $n_f$, but at $768\times768$ with $1000$ frequencies a pair occurs
  in about half the draws, and HF's default seed $777$ gives exactly two. HF's default random spectrum
  ($\mathcal N(0,1)$) starts elsewhere, with a defect about $14\%$ the size of a $768\times768$ $W_0$ of entry standard
  deviation $0.04$ at $1000$ frequencies. Its covariance group $\Pi_{\rm cyc}$ excludes most neuron permutations, so
  permuting hidden units can change what it reaches.
\item[Base] Idempotent with fixed indices.
\item[Merge] M1.
\end{exVIIlines}
\exVIIrefs{\thmref{Proposition}{II.12} \textperiodcentered{} \thmref{Proposition}{II.13} \textperiodcentered{}
  \thmref{Observation}{II.15} \textperiodcentered{} \thmref{Theorem}{V.3}}
\end{example*}

\begin{example*}{BitFit}\phantomsection\label{ex:bitfit}%
\exVIIsrc{\citet{benzaken2021bitfit}}{selective}
\[
  b\mapsto b_0+\beta\ \ \text{on every bias},\qquad W=W_0.
\]
\begin{exVIIlines}
\item[Image] A translation of the bias block, linear and regular. It has no gauge, so its expressive dimension equals
  its parameter count.
\item[Base] Idempotent.
\item[Merge] Nothing to merge, since BitFit trains the model's own biases. Llama and Mistral have no biases, so it
  trains nothing there.
\end{exVIIlines}
\exVIIrefs{\thmref{Proposition}{II.12} \textperiodcentered{} \thmref{Theorem}{V.3}}
\end{example*}

\begin{example*}{Adapters}\phantomsection\label{ex:adapter}%
\exVIIsrc{\citet{houlsby2019parameter}}{architectural}
\[
  x\mapsto x+U\sigma(Dx),\qquad D\in\R^{r\times d},\ U\in\R^{d\times r}.
\]
\begin{exVIIlines}
\item[Image] Function level. The adapter is an extension of the network, with neutral point $U=0$ (zero-up
  initialisation). Houlsby et al.'s own near-identity initialisation, with weights drawn from a zero-mean Gaussian of
  standard deviation $10^{-2}$ truncated at two standard deviations, starts near the neutral point and leaves a small
  defect.
\item[Fibre] Under Houlsby's GELU the gauge is only the permutations of hidden units. A ReLU, as in the Pfeiffer and
  parallel adapters, adds positive per-unit rescalings, giving $S_r\ltimes\R_{>0}^r$. At $U=0$ only $U$ moves to first
  order (the zero gate, function-level variant).
\item[Merge] M$\infty$, since it is not box-locally mergeable for non-linear $\sigma$. With $\sigma=\id$, an adapter parallel to a
  single linear box is LoRA, with the same $\rho$; one parallel to a whole sublayer edits the weightless skip
  connection and stays M$\infty$ when the sublayer is non-affine. A serial linear adapter $(I+UD)W_0$ after a
  bias-free $W_0$ reaches exactly the rank-$\le r$ updates whose rows lie in $\operatorname{row}W_0$, which is LoRA's
  image iff $\rk W_0=n$.
\end{exVIIlines}
\exVIIrefs{\thmref{Lemma}{II.4} \textperiodcentered{} \thmref{Definition}{VI.1} \textperiodcentered{}
  \thmref{Lemma}{VI.2} \textperiodcentered{} \thmref{Theorem}{VI.3} \textperiodcentered{}
  \thmref{Proposition}{VI.4}}
\end{example*}

\begin{example*}{Prefix\,/\,prompt tuning}\phantomsection\label{ex:prefix}%
\exVIIsrc{\citet{li2021prefix}; \citet{lester2021power}}{augmentation}
\[
  \mathrm{Attn}_q(P\uplus C)=(1-\lambda_q)\,\mathrm{Attn}_q(C)+\lambda_q\,\mathrm{Attn}_q(P),\qquad
  \lambda_q=\frac{Z_q(P)}{Z_q(P)+Z_q(C)}.
\]
Here $P$ and $C$ are the prefix and content multisets of key--value pairs, and
$Z_q(K)=\sum_{k\in K}e^{\langle q,k\rangle/\sqrt d}$.
\begin{exVIIlines}
\item[Image] Prefix and prompt tuning are unpointed extensions on the sequence wire, in the sense of
  \thmref{Theorem}{VI.5}(f): on a head with $W_oW_v\ne0$ no prefix is neutral. A prefix acts as an input-gated
  parallel adapter on attention. Within its layer it multiplies the content attention weights by $1-\lambda_q$ and
  keeps their ratios, and the output moves on the segment towards $\mathrm{Attn}_q(P)\in\operatorname{conv}V_P$
  (Figure~\ref{fig:prefix}). This rigidity is relative to the same layer without the prefix, and relative to the
  pretrained pattern only at the lowest prefixed layer.
\item[Arrows] In causal decoders whose content positions are offset by the prompt length, as in HF PEFT, every soft
  prompt is realised exactly by the deep prefix made of its own per-layer keys and values. This is an arrow of the
  unpointed slice, from prompt tuning into deep prefix tuning (P-tuning v2). \citet{petrov2023prompting} and
  \citet{li2021prefix} assert that the inclusion is strict. It is strict in a one-layer, one-token example, where a
  prompt fixes the prefix value as a function of its key; the generic statement is not proved here, and encoders remain
  open. LLaMA-Adapter \citep{zhang2023llama} normalises the prefix separately and adds a zero-initialised gate, so it
  leaves the content attention weights unchanged and supplies the missing neutral point.
\item[Merge] M$\infty$. For a standard softmax head with $W_o\ne0$ and $W_q\ne0$, a generic prefix makes the head
  non-affine even on single-token inputs, so prefix tuning cannot be folded into the head's own weights.
\end{exVIIlines}
\exVIIrefs{\thmref{Definition}{VI.1} \textperiodcentered{} \thmref{Theorem}{VI.5}}
\end{example*}

\begin{example*}{PiSSA\,/\,LoftQ\,/\,QLoRA}\phantomsection\label{ex:pissa}%
\exVIIsrc{\citet{meng2024pissa}; \citet{li2023loftq};\newline \citet{dettmers2023qlora}}{additive}
\begin{gather*}
  \rho(B,A)=W_{\rm res}+sBA,\qquad sB_0A_0=[W_0]_r,\qquad W_{\rm res}=W_0-[W_0]_r\qquad\text{(PiSSA)},\\
  \rho(B,A)=\kappa\theta_0+sBA,\qquad B_0=0\qquad\text{(QLoRA)}.
\end{gather*}
Here $[X]_r$ is the best rank-$r$ approximation of $X$, $W_{\rm res}$ is frozen, and $\kappa$ quantises and
dequantises. \citet{meng2024pissa} take $s=1$.
\begin{exVIIlines}
\item[Image] PiSSA is a split (Type II) start. Its image is the cone with vertex $W_0-[W_0]_r$, and $\theta_0$ is a
  smooth point of it (Figure~\ref{fig:apex}b). Its image and zero-init LoRA's are incomparable. No exact pointing of
  LoRA$_r$, zero-init or split, makes an update of rank above $2r$ reachable, and split pointings attain $2r$ when
  $2r\le\min(m,n)$.
\item[Fibre] Gauge $\GL_r$. HF's PiSSA factors give $\Phi_0=(S_r/s)(1/\eta_B-1/\eta_A)$, so the start is balanced,
  $\Phi_0=0$, at equal rates. By the balance--escape trilemma a balanced, non-stationary, identity-at-init start needs
  the frozen residual.
\item[Base] QLoRA is LoRA pointed at $\kappa\theta_0$, with generically full-rank defect $e=\kappa\theta_0-\theta_0$,
  so its image misses $\theta_0$ unless $\rk e\le r$ (Figure~\ref{fig:teaser}). Splitting and error fitting change
  the defect, and the order of operations matters. QPiSSA splits, then quantises the residual, and leaves
  $\kappa(W_{\rm res})-W_{\rm res}$. One-step LoftQ quantises, then fits the error. HF's LoftQ initialisation ignores
  the LoRA scale $s=\alpha/r$, so the defect is $e-s[e]_r$. It is $e$ minus its best rank-$r$ approximation only when
  $s=1$, and for $s=2$, as with $\alpha=2r$, its norm equals $\lVert e\rVert$.
\item[Merge] PiSSA merges into its box (M1). On a quantised base the type is Mq. Exact merges of generic
  high-precision adapters leave the quantised type, and type-preserving merges are lossy; QA-LoRA's group-constant
  adapters under affine group quantisation are the exception \citep{xu2023qa}.
\end{exVIIlines}
\exVIIrefs{\thmref{Theorem}{II.1} \textperiodcentered{} \thmref{Theorem}{II.2} \textperiodcentered{}
  \thmref{Corollary}{III.6} \textperiodcentered{} \thmref{Proposition}{III.8} \textperiodcentered{}
  \thmref{Proposition}{V.2}}
\end{example*}

\begin{example*}{ReLoRA}\phantomsection\label{ex:relora}%
\exVIIsrc{\citet{lialin2023relora}}{optimizer}
\[
  \theta_{k+1}=\theta_k+sB_kA_k,\qquad \text{fresh }(B_k,A_k)\text{ at each stage}.
\]
\begin{exVIIlines}
\item[Base] Rebased LoRA. After $K$ stages it reaches exactly $W_0+\MM_{\le\min(Kr,m,n)}$, so all of $\Theta$ at
  $K^*=\lceil\min(m,n)/r\rceil$ (Figure~\ref{fig:rebase}a). Over weight space it is $\mathrm{LoRA}_r^{\oplus K}$
  unrolled in time.
\item[Fibre] Each stage is an apex with gauge $\GL_r$.
\item[Merge] Each stage merges (M1).
\end{exVIIlines}
\exVIIrefs{\thmref{Observation}{I.7} \textperiodcentered{} \thmref{Theorem}{V.3}}
\end{example*}

\begin{example*}{GaLore}\phantomsection\label{ex:galore}%
\exVIIsrc{\citet{zhao2024galore}}{optimizer}
\[
  W_{t+1}=W_t-\eta\,P_t\,\mathrm{opt}\big(P_t\T G_t\big),\qquad P_t\in\R^{m\times r}.
\]
Here $G_t$ is the weight gradient, and $P_t$ is refreshed from its SVD every $T$ steps.
\begin{exVIIlines}
\item[Fibre] A backward lens with compression $P_t\T$ and anchor $P_t$. Under an optimizer that reads only its
  gradients, such as Adam with weight decay $0$ and no full-gradient clipping, each period is exactly one-sided LoRA
  $W_k+P_kC$ with $C_0=0$, merged and re-initialised at the end of the period, with its optimizer state carried over
  rather than reset or transported. Decoupled weight decay and full-gradient clipping break this identity.
\item[Base] The anchor is constant on each period, so it is integrable trivially, including at $T=1$, and
  integrability does not separate GaLore from LoRA. What separates them is the resampling schedule together with the
  carried state. As a set over its possible sequences of projectors, the moving shape reaches all of $\Theta$.
\item[Merge] Not applicable, since GaLore writes $W$ itself.
\end{exVIIlines}
\exVIIrefs{\thmref{Definition}{IV.1} \textperiodcentered{} \thmref{Proposition}{IV.2} \textperiodcentered{}
  \thmref{Proposition}{IV.3} \textperiodcentered{} \thmref{Theorem}{V.3}}
\end{example*}

\begin{example*}{Merging}\phantomsection\label{ex:merging}%
\exVIIsrc{\citet{ilharco2022editing}; \citet{huang2023lorahub}}{composition}
\[
  \theta_0+\textstyle\sum_i\lambda_i\tau_i\ \ \text{(task arithmetic)},\qquad
  \big(\textstyle\sum_iw_iB_i\big)\big(\textstyle\sum_jw_jA_j\big)\ \ \text{(LoraHub)}.
\]
\begin{exVIIlines}
\item[Fibre] Task arithmetic is $\GL(\Theta)$-natural. TIES-Merging \citep{yadav2023ties}, with global trimming and
  sign election by total mass, is natural under homotheties times signed permutations. DARE \citep{yu2023language} is
  natural under every invertible coordinate rescaling for each mask and under permutations in law, so its
  distributional group is the monomial group. None of these sparsifying rules is natural under rotations.
\item[Fibre] Post-hoc factor-wise rules (LoraHub, and HF's \texttt{linear} merge with its non-SVD \texttt{ties} and
  \texttt{dare} variants) depend on each adapter's gauge, since replacing $(B_2,A_2)$ by $(cB_2,A_2/c)$ changes
  LoraHub's result by $\sum_{l\ne2}w_lw_2\big[(c-1)B_2A_l+(c^{-1}-1)B_lA_2\big]$ (Figure~\ref{fig:gauge}). A shared
  factor, such as a common frozen $A$, or a joint frame that produces one, such as the joint $U\Sigma$ of KnOTS
  \citep{stoica2024model}, removes the gauge dependence. With a shared $A$, LoraHub returns
  $(\sum_jw_j)\sum_iw_iB_iA$, exact up to the global factor $\sum_jw_j$.
\item[Merge] Merging an adapter into a full-precision base is the canonical arrow to the terminal object, full
  fine-tuning.
\end{exVIIlines}
\exVIIrefs{\thmref{Observation}{I.4} \textperiodcentered{} \thmref{Proposition}{V.4}}
\end{example*}

Exposé~\ref{sec:atlas} turns the answers on these cards into seven coordinates, each read off a method's formula
without training, and into the arrows between methods.

\section{The classification}\label{sec:atlas}

A method is a pointed map $\rho:(Q,q_0)\to(\Theta,\theta_0)$. Its image germ says what it can express
(Exposé~\ref{sec:image}), its fibres and the optimizer's metric how it moves (Exposé~\ref{sec:fibres}), base change what
a moved $\theta_0$ does (Exposé~\ref{sec:base}), and extensions cover adapters and prompts (Exposé~\ref{sec:merge}).
Exposé~\ref{sec:examples} applied this toolkit to fourteen methods. This exposé turns the answers into coordinates and
arrows.

Seven coordinates, each read off the formula, place all \nMethods{} catalogued methods in a periodic table
(Table~\ref{tab:periodic};
\S\ref{sec:atlas-axes} and \S\ref{sec:atlas-table}). Some combinations of coordinates are empty by theorem; for
instance, no linear method starts at an apex. A method $N$ simulates a method $M$ when a pointed smooth map $h$ gives $\rho_N\circ h=\rho_M$.
Five rules produce every arrow, and the atlas records \nArrows{} of them, each with its map $h$; three tests rule arrows
out, and PiSSA and LoRA, for instance, have no arrow either way (\S\ref{sec:atlas-arrows}). Parameters are not reach. At
$131{,}072$ parameters on a $4096\times4096$ matrix, LoRA$_{16}$ reaches $130{,}816$ directions and LoHa$_{8,8}$ at most
$122{,}753$ (\S\ref{sec:atlas-budget}).

\newsavebox{\fbAtlasPeriodicBox}%
\begin{table}[!tbp]
\centering
\caption{The periodic table of the \nCore{} core methods. Rows are kinds (axis~1 of \S\ref{sec:atlas-axes}), with
reparametrisations split by how they couple to $W_0$: additively, by a group action, by a Hadamard product or by
selecting coordinates. Columns are the shapes of the image germ (axis~2). A dot marks an empty cell. All seven
coordinates of the methods named here are in Table~\ref{tab:periodic-methods}, and those of every catalogued method in
Table~\ref{tab:catalogue}.}\label{tab:periodic}
\sbox{\fbAtlasPeriodicBox}{\begin{minipage}{1.3\linewidth}
\begingroup\scriptsize\setlength{\tabcolsep}{3pt}\renewcommand{\arraystretch}{1.15}
\begin{tabularx}{\linewidth}{@{}>{\raggedright\arraybackslash}p{1.9cm}>{\raggedright\arraybackslash}X>{\raggedright\arraybackslash}X>{\raggedright\arraybackslash}X>{\raggedright\arraybackslash}X>{\raggedright\arraybackslash}X>{\raggedright\arraybackslash}X>{\raggedright\arraybackslash}X@{}}
\toprule\headrow
\textbf{Kind} & \textbf{Lin} & \textbf{Cone} & \textbf{Cone$^{*}$} & \textbf{Orb} & \textbf{Sat} & \textbf{Meet} & \textbf{Fun}\\\midrule
R, additive & C3A, FFA-LoRA, FourierFT, FRoD, InfLoRA, LoRA-FA, LoRA-SB, LoRA-XS, MiCA, MiSS (formerly Bone), MoRA, OSF, SVFT, TinyLoRA, WaveFT & AdaLoRA, AdaMSS, BD-LoRA, DEFT, DeLoRA, DyLoRA, EVA, FacT, GraLoRA, KaSA, KronA, LoHa, LoKr, LongLoRA, LoRA, LoRA+, LoRA-Pro, LoRA-RITE, NOLA, O-LoRA, PEANuT, PVeRA, QA-LoRA, QLoRA, RandLoRA, Riemannian LoRA, RoSA, rsLoRA, Uni-LoRA, VB-LoRA, VeRA & Astra, CorDA, LoftQ, LoRA-GA, LoRETTA, MiLoRA, OLoRA, PiSSA, PRISM & \textcolor{inkgray!70}{\textperiodcentered} & \textcolor{inkgray!70}{\textperiodcentered} & \textcolor{inkgray!70}{\textperiodcentered} & \textcolor{inkgray!70}{\textperiodcentered}\\
R, action & (IA)\textsuperscript{3}, PEQA, RED, SSF & \textcolor{inkgray!70}{\textperiodcentered} & \textcolor{inkgray!70}{\textperiodcentered} & GOFT, LOFT, OFT, OFTv2, Piggyback, qGOFT, RoAd & BOFT, DoRA, PSOFT & HRA, LoCO & \textcolor{inkgray!70}{\textperiodcentered}\\
R, Hadamard & \textcolor{inkgray!70}{\textperiodcentered} & GLoRA, HiRA & \textcolor{inkgray!70}{\textperiodcentered} & \textcolor{inkgray!70}{\textperiodcentered} & \textcolor{inkgray!70}{\textperiodcentered} & \textcolor{inkgray!70}{\textperiodcentered} & \textcolor{inkgray!70}{\textperiodcentered}\\
R, selective & BEFT, BitFit, Child-Tuning, FISH Mask, Linear probing, LN Tuning, LT-SFT, SHiRA, Super-Tuning (Super / Supra), Surgical FT, Trainable Tokens & Diff Pruning & \textcolor{inkgray!70}{\textperiodcentered} & \textcolor{inkgray!70}{\textperiodcentered} & \textcolor{inkgray!70}{\textperiodcentered} & \textcolor{inkgray!70}{\textperiodcentered} & \textcolor{inkgray!70}{\textperiodcentered}\\
E (pointed ext.) & Custom Diffusion, Textual Inversion & \textcolor{inkgray!70}{\textperiodcentered} & \textcolor{inkgray!70}{\textperiodcentered} & \textcolor{inkgray!70}{\textperiodcentered} & \textcolor{inkgray!70}{\textperiodcentered} & \textcolor{inkgray!70}{\textperiodcentered} & Adapter (Houlsby), Compacter, ControlNet, LLaMA-Adapter, LoReFT, Parallel Adapter, Pfeiffer Adapter, ShadowPEFT\\
E$^{0}$ (unpointed) & \textcolor{inkgray!70}{\textperiodcentered} & \textcolor{inkgray!70}{\textperiodcentered} & \textcolor{inkgray!70}{\textperiodcentered} & \textcolor{inkgray!70}{\textperiodcentered} & \textcolor{inkgray!70}{\textperiodcentered} & \textcolor{inkgray!70}{\textperiodcentered} & AutoPrompt, Cartridges, CoOp, CPT, LST, MAM Adapter, MPT, P-Tuning, P-Tuning v2, Prefix-Tuning, Prompt Tuning, UniPELT, VPT\\
B (backward) & Flora, GaLore, LISA, LP-FT, MeZO & ReLoRA & \textcolor{inkgray!70}{\textperiodcentered} & \textcolor{inkgray!70}{\textperiodcentered} & \textcolor{inkgray!70}{\textperiodcentered} & \textcolor{inkgray!70}{\textperiodcentered} & \textcolor{inkgray!70}{\textperiodcentered}\\
P (post hoc) & DARE, Intruder-dimension reduction, KnOTS, LCM-LoRA, LoRA Soups (CAT), Task Arithmetic, TIES-Merging, WiSE-FT & LoraHub & \textcolor{inkgray!70}{\textperiodcentered} & \textcolor{inkgray!70}{\textperiodcentered} & \textcolor{inkgray!70}{\textperiodcentered} & \textcolor{inkgray!70}{\textperiodcentered} & Proxy-Tuning\\
F$_T$ (family) & \textcolor{inkgray!70}{\textperiodcentered} & HyperDreamBooth, Laplace-LoRA, MHR, Polytropon (Poly), Punica, S-LoRA, Text-to-LoRA (T2L) & \textcolor{inkgray!70}{\textperiodcentered} & \textcolor{inkgray!70}{\textperiodcentered} & \textcolor{inkgray!70}{\textperiodcentered} & \textcolor{inkgray!70}{\textperiodcentered} & AdapterFusion, aLoRA, Arrow, HydraLoRA, HyperFormer, Lily, LoRAMoE, MoLE, PHATGOOSE, X-LoRA\\
\bottomrule
\end{tabularx}\endgroup
\end{minipage}}%
\ifdim\dimexpr(\ht\fbAtlasPeriodicBox+\dp\fbAtlasPeriodicBox)*100/130\relax>0.86\textheight
  \resizebox*{!}{0.86\textheight}{\usebox{\fbAtlasPeriodicBox}}%
\else
  \resizebox{\linewidth}{!}{\usebox{\fbAtlasPeriodicBox}}%
\fi
\end{table}

\subsection{Seven axes, each decidable from the formula}\label{sec:atlas-axes}

\citet{he2021unified} gave adapters four design axes (functional form, insertion form, modified representation and
composition function), and \citet{chen2023parameter} searched a similar design space. The seven axes below extend the
idea to weight-space, backward and post-hoc methods, and each axis is tied to a result.

\makeatletter
\begin{fb@rulebox}[ochre]{1.1pt}\noindent{\normalfont\bfseries\scshape Definition}\ {\normalfont(coordinates of a
method).}\ \upshape
The \emph{coordinates} of a method $M=(Q,q_0,\rho)$ are its answers to the seven questions below, each read off $\rho$,
$q_0$ and the box that $\rho$ acts on, with no training.
\end{fb@rulebox}
\makeatother

The axes and their values are as follows. Table~\ref{tab:axes} states the question each axis asks and defines every
value, Table~\ref{tab:axes-examples} gives the coordinates of LoRA, (IA)$^3$ and prefix tuning, and the ancillary file
\texttt{catalogue.json} repeats the definitions with examples for every value.
\begin{enumerate}[label=\arabic*.]
  \item \textbf{Kind.} R (reparametrisation) $\cdot$ E (pointed extension) $\cdot$ E$^\circ$ (unpointed extension)
  $\cdot$ B (backward lens or schedule) $\cdot$ P (post-hoc operation) $\cdot$ F$_T$ (family over tasks, inputs or
  tenants).
  \item \textbf{Shape} of the image germ. Lin ($\rho$ affine) $\cdot$ Cone (multihomogeneous, non-linear) $\cdot$
  Cone$^*$ (translated cone, split pointing) $\cdot$ Orb (group orbit) $\cdot$ Sat (group $\times$ cone) $\cdot$ Meet
  (orbit $\cap$ cone) $\cdot$ Fun (function level).
  \item \textbf{Base point}, read from $S_1$. regular $\cdot$ apex (deficit $>0$) $\cdot$ neutral $\cdot$ defect
  ($\rho(q_0)\neq\theta_0$) $\cdot$ unpointed.
  \item \textbf{Gauge}, from wire insertions and spider slides, checked by Jacobian rank. none $\cdot$ scalar $\cdot$
  torus $\cdot$ $\GL$ $\cdot$ $\GL\times{}$torus $\cdot$ non-linear.
  \item \textbf{Covariance group} (the Klein geometry of the image). $\GL\times\GL$ $\cdot$ $\CO\times\CO$ $\cdot$
  $\GL\times\CO$ $\cdot$ $\mathrm{Mon}\times\GL$ $\cdot$ $\mathrm{Mon}\times\mathrm{Mon}$ $\cdot$ $\Pi_X$ (imposed
  structure) $\cdot$ frame (random or data, in law).
  \item \textbf{Merge type.} M1 (into its box) $\cdot$ M1$\to$ (into a neighbour via R1--R7, under side conditions)
  $\cdot$ Mq (an exact merge leaves the quantised type) $\cdot$ M$\infty$ (not box-locally mergeable).
  \item \textbf{Rebasing closure.} idempotent $\cdot$ $\to$span $\cdot$ $\to$group $\cdot$ schedule.
\end{enumerate}

\begin{table}[t]
\centering\footnotesize
\caption{The seven questions, answered for three methods. Every answer is read off the formula with no training; the
values of each axis are defined in Table~\ref{tab:axes}.}\label{tab:axes-examples}
\begin{tabularx}{\linewidth}{@{}>{\raggedright\arraybackslash}p{2.0cm}>{\raggedright\arraybackslash}X%
>{\raggedright\arraybackslash}p{2.9cm}>{\raggedright\arraybackslash}p{2.5cm}>{\raggedright\arraybackslash}p{2.25cm}@{}}
\toprule\headrow
\textbf{Axis} & \textbf{Question} & \textbf{LoRA} & \textbf{(IA)$^3$} & \textbf{Prefix tuning}\\\midrule
1\enspace Kind & Where does it act? & R & R & E$^\circ$\\
2\enspace Shape & What is the image germ? & Cone, $d=r(m+n-r)$ & Lin, a torus-orbit closure & Fun\\
3\enspace Base point & Is $\rho(q_0)=\theta_0$? Is $\dim S_1=d(M)$? & apex, deficit $r(n-r)$ & regular & unpointed\\
4\enspace Gauge & Which reparametrisations of $Q$ leave $\rho$ unchanged? & $\GL_r$ & none & non-linear (Li--Liang's
MLP)\\
5\enspace Covariance & For which $(g,h)$ is $\Im M(gW_0h^{-1})=g\,\Im M(W_0)\,h^{-1}$? & $\GL\times\GL$ &
$\mathrm{Mon}\times\GL$; $\GL\times\mathrm{Mon}$ on $W_{\rm down}$ & $\GL\times\GL$, function level\\
6\enspace Merge & Does the update fuse into its box, or by R1--R7 into a neighbour? & M1 & M1$\to$ & M$\infty$\\
7\enspace Rebasing & What does merge-and-restart reach? & $\to$span, all of $\Theta$ at $K=\lceil\min(m,n)/r\rceil$ &
idempotent & --\\
\bottomrule
\end{tabularx}
\end{table}

\subsubsection{Deciding each axis}\label{sec:atlas-decide}

Each axis has a procedure that uses only $\rho$, $q_0$ and the box.
\begin{itemize}
  \item \emph{Kind.} Check whether the trained model is $f(\rho(q),-)$ with the original architecture (R), whether new
  boxes are inserted and some parameter value makes them the identity (E) or none does (E$^\circ$), whether only the
  gradient path or the schedule changes (B), whether the inputs are trained checkpoints (P), and whether the adapter
  depends on a task, an input or a tenant (F$_T$). See \thmref{Definition}{I.2}, \thmref{Definition}{IV.1} and
  \thmref{Definition}{VI.1}.
  \item \emph{Shape.} Read the degree of $\rho-\theta_0$ in $q$. An affine map gives Lin and a multihomogeneous
  non-linear one gives Cone. A group acting on $\theta_0$ gives Orb, a torus acting on another method's image gives Sat
  (\thmref{Proposition}{II.8}), and an intersection gives Meet (\thmref{Theorem}{II.7}).
  \item \emph{Base point.} Evaluate $\rho(q_0)$ and compute $S_1=d\rho_{q_0}(T_{q_0}Q)$, then compare $\dim S_1$ with
  the generic rank $d(M)$ (\thmref{Definition}{I.5}). For LoRA this is \thmref{Theorem}{II.2}(b).
  \item \emph{Gauge.} Insert $gg^{-1}$ on each internal wire between trainable boxes and slide diagonals through copy
  spiders, then confirm $\lvert M\rvert-d(M)$ by a Jacobian rank at a random point
  (\thmref{Proposition}{II.12}(a)).
  \item \emph{Covariance.} Test the identity of \thmref{Definition}{II.14} on generators of $\GL_m\times\GL_n$, in
  distribution when frames are random or come from data.
  \item \emph{Merge.} Run the rewrite rules R1--R7 of \thmref{Theorem}{VI.3} with their side conditions, adding the type
  condition of \thmref{Proposition}{V.2}(d) for a quantised base. The rules are sufficient conditions. They certify M1
  and M1$\to$, and M$\infty$ records a box-local obstruction. Whether a whole network admits some other merge is left
  open.
  \item \emph{Rebasing.} Apply \thmref{Theorem}{V.3}(a) to additive shapes and \thmref{Theorem}{V.3}(b) and (c) to
  actions.
\end{itemize}

\subsubsection{Conventions}\label{sec:atlas-conventions}

\begin{itemize}
  \item \emph{Null coordinates.} ``--'' (null in the classified data) means that the axis does not apply, and no axis
  has a separate ``not applicable'' value. The merge axis keeps a legacy key \texttt{na}, used for backward lenses and
  schedules, which write the full weight directly and so have nothing to merge; it means the same as ``--''.
  \item \emph{When an axis applies.} Shape and gauge need a parameter space, so they are ``--'' for closed-form
  post-hoc operations with no tunable coefficient (Fisher merging without mixing weights, RegMean, uniform model soups),
  which are single points. Base point is ``--'' only for methods with no base point of their own, the serving families
  S-LoRA and Punica. Rebasing needs a weight-space merge, so it is ``--'' for every M$\infty$ method; function-level
  closures belong in the notes. Covariance applies to every method with a formula.
  \item \emph{Function-level covariance.} For extensions and families (shape Fun), covariance is the function-level
  analogue of \thmref{Definition}{II.14}: the inserted or generated family on each adapted wire is invariant under
  conjugation by $(g,h)$ on that wire. Routed mixtures are included, since router and experts transform together.
  \item \emph{Post-hoc operations.} The parameter space of a P operation is its tunable coefficients ($\lambda$ in task
  arithmetic, TIES, DARE, LoRA soups and CAT, LoraHub), and its covariance is the equivariance of the combination rule
  (\thmref{Proposition}{V.4}).
  \item \emph{Nearest keys.} Entries without an exact key are coded to the nearest one: $\CO\times\mathrm{Borel}\to\Pi$
  (OLoRA); $\Ort(r)\to\GL$ (LoReFT, PoLAR); stabiliser $\to$ none (OFT); $\CO(m)\times\GL(n)\to\CO\times\CO$ (Astra,
  whitened splits); $\Ort(m)\times\Ort(n)\to\CO\times\CO$ (LoRA-GA); $\R^\times B_m\times\R^\times B_n\to
  \mathrm{Mon}\times\mathrm{Mon}$ (LoRA-SB); $\R^\times B_N$ and the torus $T_N\to\mathrm{Mon}\times\mathrm{Mon}$ (TIES,
  DARE). One-sided keys ($\mathrm{Mon}\times\GL$, $\GL\times\CO$) name the group up to exchanging the two sides; the
  tables of \S\ref{sec:atlas-table} write (output side) $\times$ (input side).
  \item \emph{Instances.} Arrows and obstructions relate instances of methods: a rank, a pointing, a hyperparameter
  regime. The instance is recorded in the note that accompanies each arrow and obstruction in the ancillary files
  (\S\ref{sec:atlas-checks}).
\end{itemize}

\begin{explains}
Methods with different shapes or base-point types are never isomorphic, however alike their formulas
(\thmref{Observation}{I.4}(c), \thmref{Observation}{I.6}).
\end{explains}

\begingroup\footnotesize\setlength{\tabcolsep}{3pt}\renewcommand{\arraystretch}{1.12}
\begin{longtable}{@{}>{\raggedright\arraybackslash}p{2.75cm}>{\raggedright\arraybackslash}p{12.4cm}@{}}
\caption{The seven axes. For each axis, the question it answers and the definition of every value, with the test that
decides it where one is needed. ``--'' (a null coordinate) means that the axis does not apply.}\label{tab:axes}\\
\toprule\headrow\textbf{Value} & \textbf{Definition}\\\midrule
\endfirsthead
\toprule\headrow\textbf{Value} & \textbf{Definition}\\\midrule
\endhead
\bottomrule\endlastfoot
\multicolumn{2}{@{}p{\linewidth}@{}}{\textbf{\color{tide}1\enspace Kind} (where the method acts). \emph{Does the method
reparametrise the existing weights, extend the architecture (with or without a neutral point), act only on the backward
pass or the schedule, operate post hoc on several trained points, or form a family over an auxiliary base $T$?}}\\[2pt]
R\newline reparametrisation & A pointed smooth map $\rho:(Q,q_0)\to(\Theta,\theta_0)$ into the existing weight space, an
object of $\Meth(\Theta,\theta_0)=\Diff_*/(\Theta,\theta_0)$. Test: the trained model is $f(\rho(q),-)$ with the
original architecture $f$.\\
E\newline pointed extension & A function-preserving embedding $\iota(\theta)=(\theta,\theta_0')$ into a larger
architecture $\tilde f$ with a neutral point $\theta_0'$, so that $\tilde f(\theta,\theta_0',-)=f(\theta,-)$, followed by a
reparametrisation of $\Theta\times\Theta'$. Test: new boxes are inserted, and some parameter value makes them the
identity.\\
E$^\circ$\newline unpointed extension & An extension with no neutral point: no parameter value reproduces the frozen
function (for prompts and prefixes, \thmref{Theorem}{VI.5}(f) proves this on a head with $W_oW_v\ne0$). Test: try to solve
$\tilde f(\theta_0,p,-)=f(\theta_0,-)$ for $p$.\\
B\newline backward lens or schedule & The forward map is $\id_\Theta$ and the update is
$\theta\leftarrow\theta-\eta\,a_t\big(\mathrm{opt}(c_t(\nabla L))\big)$ with a compression $c_t$ and an anchor $a_t$; or
the method is a sequence of merge-and-restart stages (a rebasing scheme); or it is an optimizer that compresses only its
own state (Flora). Test: the method changes how gradients or optimizer state are compressed, or when the base is
re-pointed, and leaves the parametrisation alone.\\
P\newline post-hoc operation & An operation $\Theta^N\to\Theta$ (or on tuples of factors) applied to several trained
points. Test: the inputs are already-trained adapters or checkpoints.\\
F$_T$\newline family over a base $T$ & A $T$-point $a:T\to Q$ of a method, giving $\rho\circ a:T\to\Theta$, where $T$
is a set of tasks (hypernetworks), the inputs $X$ (routing), a simplex (mixing) or a set of tenants $U$ (serving). Test:
the adapter depends on an auxiliary variable.\\\midrule
\multicolumn{2}{@{}p{\linewidth}@{}}{\textbf{\color{tide}2\enspace Shape} of the image germ (expressivity). \emph{What
is the germ at $\theta_0$ of the reachable set $\Im\rho$? Read it from the formula: is $\rho$ affine, multihomogeneous,
a group action, a group acting on a cone, an intersection, or defined only at function level?}}\\[2pt]
Lin\newline linear (flat) & $\rho$ is affine in $q$, so $\Im\rho$ is an affine subspace, and $d(M)$ is the rank of the
linear map, which equals $\lvert M\rvert$ when the map is injective (FourierFT loses one dimension per sampled conjugate
frequency pair). Every point is regular, and rebasing is idempotent when frames, masks and operators are frozen across
cycles.\\
Cone\newline cone with vertex at the start & $\rho-\rho(q_0)$ is multihomogeneous (or a union of linear spaces), and the
image is a non-linear cone with vertex at the start $\rho(q_0)$, which is then a singular point (an apex). The start is
$\theta_0$ for pointed methods, and the quantised or truncated base $\theta_0+e$ for QLoRA-type defects
(\thmref{Proposition}{V.2}). Determinantal, Segre, Hadamard and tensor-network varieties are included.\\
Cone$^*$\newline translated cone (split) & $\rho=\theta_{\rm res}+\delta(q)$ with $\delta$ multihomogeneous and $q_0$
chosen so that $\delta(q_0)=\theta_0-\theta_{\rm res}$ has full rank. The image is a cone whose apex $\theta_{\rm res}$
is not $\theta_0$, and $\theta_0$ is a smooth point (Type II objects).\\
Orb\newline group orbit & $\rho(q)=\gamma(q)\cdot\theta_0$ with $\gamma(q)$ in a group (or monoid) acting linearly on
$\Theta$. Exactly orthogonal actions preserve the Gram matrix and the spectrum. Torus and Hadamard actions keep only the
zeros of $W_0$ (and its row lines while the scales are nonzero), since unconstrained scales can create new zeros.\\
Sat\newline torus saturation & A group or monoid acting on a cone or an orbit: $\Im M=T\cdot\Im N$ for a torus or
diagonal monoid $T$ and a method $N$.\\
Meet\newline meet (pullback) & The image is an intersection of a cone and an orbit (an image-level meet), or the object
is a fibre product of two methods (a categorical product, which exists only when the set-level fibre product is a
manifold).\\
Fun\newline function level & There is no weight-space image: the method lives on an extension and is compared through the
realisation $\Phi_f$.\\\midrule
\multicolumn{2}{@{}p{\linewidth}@{}}{\textbf{\color{tide}3\enspace Base point} (the position of $\theta_0$). \emph{Is
$\theta_0=\rho(q_0)$, and if so, is the first-order image $S_1=d\rho_{q_0}(T_{q_0}Q)$ as large as the expressive
dimension $d(M)$?}}\\[2pt]
regular & $\dim S_1=d(M)$, a first-order deficit of $0$: linear methods, orbit methods with an injective orbit map, and
split (Type II) initialisations.\\
apex\newline first-order deficient & $\dim S_1<d(M)$. For cone-shaped germs $\theta_0$ is the vertex. LoRA's deficit is
$r(n-r)$; HRA's paired initialisation has $\dim S_1=kn-k(k+1)/2$ against $d=rn-r(r+1)/2$, with $k=r/2$. DoRA inherits
LoRA's deficit through its torus factor.\\
neutral\newline pointed extension & An extension evaluated at its neutral point: zero up-projection, zero gate, identity
box.\\
defect\newline pointing defect & $\rho(q_0)\neq\theta_0$, by design or by base change; the defect $e=\rho(q_0)-\theta_0$
is part of the data.\\
unpointed & No parameter value reproduces the frozen function.\\\midrule
\multicolumn{2}{@{}p{\linewidth}@{}}{\textbf{\color{tide}4\enspace Gauge} (the generic fibre of $\rho$). \emph{Which
transformations of the trainable parameters leave $\rho$ unchanged, and what is the dimension $\lvert M\rvert-d(M)$ of
the generic fibre? Read it off by inserting $gg^{-1}$ on internal wires between trainable boxes and sliding diagonals
through spiders, and confirm it by the generic Jacobian rank.}}\\[2pt]
none\newline trivial & $\rho$ is injective near generic points. There are no charges, and factor-wise operations are
well defined.\\
scalar\newline one scalar slide & A one-parameter group $(b,d)\mapsto(cb,d/c)$ through frozen boxes, with one conserved
charge under gradient flow.\\
torus & A diagonal group (per rank channel or per hidden unit), whose charges are diagonal entries.\\
$\GL$\newline general linear on internal wires & $\GL_r$, or a product of general linear groups on bonds, acting by
$(Bg^{-1},gA)$. The compact half $\so(r)$ gives isometries, and the non-compact half $\Sym_r$ gives $r(r+1)/2$
charges.\\
$\GL\times{}$torus\newline with a Hadamard torus & $\GL$ on each factor plus the torus $(D_1XD_2,D_1^{-1}YD_2^{-1})$
sliding through the copy spider, of dimension $r_1^2+r_2^2+m+n-1$.\\
non-linear & The generic fibre is not an orbit of a linear group action.\\\midrule
\multicolumn{2}{@{}p{\linewidth}@{}}{\textbf{\color{tide}5\enspace Covariance group} (the Erlangen geometry of the
image). \emph{For which $(g,h)\in\GL_m\times\GL_n$ does $\Im M(gW_0h^{-1})=g\,\Im M(W_0)\,h^{-1}$ hold for every $W_0$
(in distribution if random frames or data are involved)?}}\\[2pt]
$\GL\times\GL$\newline basis-free & The image transforms with every change of basis on both sides. It satisfies the
sufficient condition of \thmref{Observation}{II.15} for every box-wise linear function-preserving gauge.\\
$\CO\times\CO$\newline spectral frame & The method uses an SVD or another orthogonally equivariant frame of $W_0$, of
gradients or of data. Truncated SVDs are also equivariant under positive scalings, so the group is
$\CO(m)\times\CO(n)=(\R_{>0}\Ort(m))\times(\R_{>0}\Ort(n))$ (\thmref{Proposition}{II.16}). Methods that use bare
singular frames without singular values (LoRA-GA) are scale-invariant rather than covariant and get only
$\Ort(m)\times\Ort(n)$; they are coded with this nearest key.\\
$\GL\times\CO(n)$\newline input-side orthogonal & A right action of $\Ort(n)$ or a subgroup of it. The covariance on the
input side is the normaliser of the structure group: $\CO(n)$ for full OFT, the block normaliser for block OFT.\\
$\mathrm{Mon}\times\GL$\newline one-sided torus & A diagonal torus acts on one side, so only monomial changes of basis
are allowed there.\\
$\mathrm{Mon}\times\mathrm{Mon}$\newline coordinate geometry & Hadamard or support structure in the standard basis on
both sides. It satisfies the sufficient condition of \thmref{Observation}{II.15} for neuron permutations but not for
rotations.\\
$\Pi_X$\newline imposed structure & The method imposes a reshape (a tensor factorisation), a block partition, a cyclic
order or a pairing that the network does not have, and so breaks even neuron permutations (the test of
\thmref{Lemma}{VI.2}).\\
frame\newline random or data frame & Covariance holds only in law, through the distribution of frozen random matrices or
of data statistics.\\\midrule
\multicolumn{2}{@{}p{\linewidth}@{}}{\textbf{\color{tide}6\enspace Merge type} (lifting along the realisation).
\emph{After training, does the modified box function stay in the box's function class, and if not, can the rewrite
rules R1--R7 of \thmref{Theorem}{VI.3} (fusion, permutation slide, positive-diagonal slide, spider slide, cup slide, norm
absorption, copy slide) move it into a neighbouring linear box?}}\\[2pt]
M1\newline fuses into its own box & All non-linearity is on the parameter side, and the merge is the canonical map $\rho$
to the terminal object.\\
M1$\to$\newline fuses into a neighbour & A diagonal method merges by R1--R7 into a neighbouring linear box under side
conditions: a position-independent path (under RoPE, scales constant on rotary pairs), an unshared target box (no tying,
and no GQA sharing unless the scale is constant across the group), and a bias slot for shifts. (IA)$^3$ always fuses into
$W_K$, $W_V$ and $W_{\rm down}$; $W_{\rm up}$ is an alternative only under SwiGLU or GeGLU (every scale) or ReLU
(nonnegative scales), and under GELU only for scales in $\{0,1\}$. The rules give sufficient conditions; they do not
decide whole-network mergeability.\\
Mq\newline leaves the quantised type & An exact merge of a generic high-precision update leaves the quantised type (NF4
and recomputed-scale absmax grids are not even closed under addition). Type-preserving merges (dequantise, add and
requantise, as HF PEFT does for bitsandbytes layers) stay on the grid but are lossy. QA-LoRA's group-constant adapters
are the exception.\\
M$\infty$\newline obstructed & The activation side is non-affine in the input for some parameter value (non-linear
adapters, softmax-gated prefixes, input-dependent routing), or the edit acts only at selected positions (ReFT,
position-gated LoRA variants). This is box-local non-mergeability under the rewrite calculus of
\thmref{Theorem}{VI.3}. Whole-network non-mergeability is open, and some residual-stream edits do merge: into an untied
embedding table, a writer's bias or a post-LN affine, or globally for orthogonal edits in RMSNorm models.\\\midrule
\multicolumn{2}{@{}p{\linewidth}@{}}{\textbf{\color{tide}7\enspace Rebasing closure} (the reach of merge-and-restart).
\emph{What does the reachable set $R_K=\theta_0+C^{+K}$ (additive) or $R_K=S^K\theta_0$ (action) converge to as the
number $K$ of merge-and-restart cycles grows?}}\\[2pt]
idempotent & The shape is exactly a subspace, subgroup or submonoid independent of the base, with frames, masks and
operators frozen across cycles, so cycles gain nothing (up to closure). Recomputing frames or masks from the merged
weight does gain.\\
$\to$span\newline converges to the span & For a cone closed under real scaling, $R_\infty=\theta_0+\spanop C$. For LoRA,
$R_K=W_0+\MM_{\le\min(Kr,m,n)}$, and full fine-tuning is reached exactly at $K=\lceil\min(m,n)/r\rceil$.\\
$\to$group\newline converges to a generated group & For action methods, $R_\infty$ is the orbit of the monoid generated
by one cycle's image. It is the connected Lie subgroup generated when that image contains a neighbourhood of $e$ in each
generating subgroup (\thmref{Theorem}{V.3}(b)). For block-orthogonal factors it is $\prod_C\SO(C)$ over the connected
components $C$ of the block hypergraph. HF's Cayley--Neumann OFT does not satisfy the hypothesis: for small generators it
generates only a contraction semigroup.\\
schedule\newline a rebasing scheme itself & The method is defined as a sequence of base changes with a moving shape, and
its reach is the sum of the shapes visited.\\
\end{longtable}\endgroup

\subsection{The periodic table}\label{sec:atlas-table}

Table~\ref{tab:periodic}, at the head of this exposé, places the \nCore{} core methods by kind and by the shape of the image germ, and splits
reparametrisations by how they touch $W_0$: additively, by a group action, by a Hadamard product or by selecting
coordinates. The split refines the additive, selective, reparametrisation-based and hybrid classes of
\citet{lialin2023scaling}; the survey of \citet{han2024parameter} uses a similar split.

Some combinations of coordinates are empty by theorem. An affine $\rho$ has the same differential everywhere, so no linear method
is an apex, and in the catalogue none is. Corollary~2 of \thmref{Theorem}{V.3} ties idempotent rebasing to shapes that
are fixed subspaces, subgroups or submonoids, and of the catalogue's $92$ idempotent methods, $84$ have linear shapes and
$7$ are orbits. The other, WASI \citep{nguyen2025efficient}, trains on the rank-$K$ variety $\MM_{\le K}$ itself, a set
that does not move with the base.

\subsubsection{The table by method}\label{sec:atlas-methods}

Table~\ref{tab:periodic-methods} gives all seven coordinates of the methods the theory works with, together with the
expressive dimension and the maximum rank. In it, $d$ is the expressive dimension, $\rho_{\max}$ the maximum rank, and $k$
(in GraLoRA) the number of blocks per side. ``$\Pi$'' means imposed structure and ``fr'' a random or data frame.
Covariance is written (output side) $\times$ (input side). F$_X$ and F$_U$ are families over inputs and over tenants,
coded F$_T$ in the catalogue, and LoRA$_{(B_0,A_0)}$ is LoRA pointed at $(B_0,A_0)$.

\begingroup\scriptsize\setlength{\tabcolsep}{1.6pt}\renewcommand{\arraystretch}{1.1}%
\newcommand*{\ptn}[1]{\textsuperscript{\sffamily\bfseries\color{tide}#1}}%
\begin{longtable}{@{}>{\raggedright\arraybackslash}p{1.7cm}>{\raggedright\arraybackslash}p{1.5cm}%
>{\raggedright\arraybackslash}p{3.2cm}>{\raggedright\arraybackslash}p{1.6cm}>{\raggedright\arraybackslash}p{1.6cm}%
>{\raggedright\arraybackslash}p{1.75cm}>{\raggedright\arraybackslash}p{1.15cm}>{\raggedright\arraybackslash}p{2.0cm}@{}}
\caption{The periodic table by method: kind, shape (with $d$ and $\rho_{\max}$), base point, gauge, covariance, merge
type and rebasing closure. The notation is explained before the table, and the marks \textdagger, $\approx$ and (fn.)
and the numbered notes after it.}\label{tab:periodic-methods}\\
\toprule\headrow\textbf{Method} & \textbf{Kind} & \textbf{Shape ($d$; $\rho_{\max}$)} & \textbf{Base pt.} &
\textbf{Gauge} & \textbf{Covariance} & \textbf{Merge} & \textbf{Rebasing}\\\midrule
\endfirsthead
\toprule\headrow\textbf{Method} & \textbf{Kind} & \textbf{Shape ($d$; $\rho_{\max}$)} & \textbf{Base pt.} &
\textbf{Gauge} & \textbf{Covariance} & \textbf{Merge} & \textbf{Rebasing}\\\midrule
\endhead
\bottomrule\endlastfoot
Frozen / Full FT & R (initial / terminal) & $\{\theta_0\}$ / $\Theta$ & -- / regular & -- & all / $\GL\times\GL$ & -- &
id\\
LoRA & R & Cone $r(m{+}n{-}r)$; $r$ & apex & $\GL_r$ & $\GL\times\GL$ & M1 & $\to\Theta$ at $K=\lceil\min(m,n)/r\rceil$\\
rsLoRA, LoRA+ & R\ptn{1} & $=$ LoRA & apex & $\GL_r$ & $\GL\times\GL$ & M1 & $\to\Theta$\\
LoRA-Pro, Riem.\ precond., LoRA-RITE & R + metric\ptn{2} & $=$ LoRA & apex & $\GL_r$\ptn{2} & $\GL\times\GL$ & M1 &
$\to\Theta$\\
DeLoRA & R & Cone $\subseteq\Mr$, $\lVert\Delta W\rVert_F\le\lambda\lVert W_0\rVert$ & apex (HF) / split (paper) &
non-linear, fibre $r^2{+}1$\ptn{3} & $\GL\times\GL$ & M1 & $\to\Theta$\\
GraLoRA & R & $\prod$ block cones, $r(m{+}n{-}r)$; $\min(kr,m,n)$ & apex & $\prod\GL_{r/k}$ & $\Pi$ (blocks) & M1 &
$\to\Theta$\\
DyLoRA & R & filtration of cones & apex & $T_r$ & $\GL\times\GL$ & M1 & $\to\Theta$\\
AdaLoRA & R & budgeted union of cones ($P\Lambda Q$) & apex ($\dim S_1=r$) & non-linear, fibre $r(r{+}1)$\ptn{4} &
$\GL\times\GL$ & M1 & $\to\Theta$\\
LoRA-FA (v1/v2) & R & Lin $\{XA_0\}$, $mr$ & regular & none & $\GL\times\mathrm{fr}$ & M1 & id\\
MiSS / Bone & R & Lin $=$ LoRA-FA$(\mathbf 1\T\otimes I_r)$ (balance mode) & regular & none & $\GL\times\Pi$ & M1 &
id\\
MiCA & R & Lin $\{U_{\min}Y\}$ & regular & none & $\CO\times\CO$ & M1 & id\\
LoRA-XS / SVFT / TinyLoRA & R & Lin in SVD frame ($r^2$ / $\lvert\Omega\rvert$ / $u$) & regular (LoRA-XS, TinyLoRA
$\approx$) & none & $\CO\times\CO$ & M1 & id (frames frozen)\\
LoRA-SB & R & Lin $BRA$ in gradient frame, $r^2$ & defect (by design) & none & $\mathrm{Mon}\times\mathrm{Mon}$\ptn{5} &
M1 & id\\
VeRA & R & toric cone, $m{+}r{-}1$; $r$ & apex & $\R^\times$ & $\mathrm{Mon}\times\mathrm{fr}$ & M1 & $\to$
LoRA-FA$(A)$\\
NOLA & R & bilinear cone, $k{+}l{-}1$ & apex & $\R^\times$ & fr & M1 & $\to\spanop\{B_iA_j\}$\\
RandLoRA & R & sum of VeRA-type cones; full rank & apex & torus & fr & M1 & $\to\Theta$\\
VB-LoRA, Uni-LoRA & R & bank / random-tied LoRA images & defect ($\approx$) & discrete & $\Pi$ / fr\ptn{6} & M1 &
$\to\Theta$\\
DoRA & R & Sat $\mathrm{Diag}_m\cdot{}$Cone, $\min(mn,\allowbreak r(m+n-r)+m)$; full & apex & $\GL_r$ &
$\mathrm{Mon}\times\GL$ & M1 & $\to\Theta$\\
PiSSA / MiLoRA & R (Type II) & translated cone & regular (smooth) & $\GL_r$ & $\CO\times\CO$ & M1 & $\to\Theta$\\
OLoRA & R (Type II) & translated cone (QR) & regular (smooth) & $\GL_r$ & $\CO\times\mathrm{Borel}$\ptn{7} & M1 &
$\to\Theta$\\
LoRA-GA / CorDA & R (Type II) & translated cone (gradient / covariance) & regular (smooth) & $\GL_r$ & $\Ort\times\Ort$
/ $\CO\times\CO$ (data)\ptn{8} & M1 & $\to\Theta$\\
Astra & R (Type II) & translated cone (output-activation second moment) & regular (smooth) & $\GL_r$ & $\CO\times\GL$
(data)\ptn{9} & M1 & $\to\Theta$\\
EVA & R (Type I) & $=$ LoRA\ptn{10} & apex & $\GL_r$ & $\GL\times\GL$\ptn{10} & M1 & $\to\Theta$\\
QLoRA & R at $\kappa\theta_0$ & cone at $\kappa\theta_0$ & defect $e$ & $\GL_r$ & Mon (grid) & Mq & $\to\Theta$\\
LoftQ & R (approx.\ Type II) & translated cone over a quantised residual & defect\ptn{11} & $\GL_r$ & Mon & Mq &
$\to\Theta$\\
LoHa & R & Hadamard variety, $\le\min(mn,\allowbreak 2r(m+n)-2r^2-(m+n-1))$; $\min(r^2,m,n)$ & apex & $\GL_r^2\times{}$torus
& $\mathrm{Mon}\times\mathrm{Mon}$ & M1 & $\to\Theta$\\
LoKr / KronA & R & $\mathcal R^{-1}$-transported cone; $\rk C\cdot r'$ & apex & $\R^\times(\times\GL_{r'})$ &
$\Pi_\otimes$ & M1 & $\to\Theta$\\
TT / Tucker (FacT, LoTR; LoRETTA) & R & meet of transported cones\ptn{12} & apex / regular\ptn{12} & $\prod\GL_{r_e}$ &
$\Pi_\otimes$ & M1 & $\to\Theta$\\
MoRA & R & Lin, $\hat r^2$; high rank & regular & none & $\Pi$ (blocks) & M1 & id (fixed operators)\\
HiRA & R & $(D_{W_0})_*$Cone; $r\rk W_0$ & apex & $\GL_r$ & $\mathrm{Mon}\times\mathrm{Mon}$ & M1 & $\to\Theta$ (not
transported ReLoRA)\\
FourierFT / WaveFT & R & Lin (sparse in the real Fourier / wavelet frame) & regular\ptn{13} & none\ptn{13} &
$\Pi_{\rm cyc}$ / $\Pi$ & M1 & id (fixed indices)\\
C3A & R & Lin (block-circulant), $mn/b$; full & regular & none & $\Pi_{\rm cyc}$ & M1 & id\\
SHiRA / FISH & R & Lin (fixed coordinate mask) & regular & none & $\mathrm{Mon}\times\mathrm{Mon}$ & M1 & id\\
Trainable Tokens & R & Lin (fixed row mask on the embedding and a tied head) & regular & none &
$\GL\times\mathrm{Mon}$\ptn{14} & M1 & id\\
Diff pruning & R & $\bigcup$ $k$-sparse subspaces & apex & none & $\mathrm{Mon}\times\mathrm{Mon}$ & M1 & $\to\Theta$\\
BitFit / LN-tuning & R & Lin (bias / LN parameters) & regular & none & $\GL\times\GL$ / Mon & M1 / M1$\to$ & id\\
OFT (block) & R & Orb of $\prod\SO(b)$ (exact Cayley); quasi-orthogonal (HF)\ptn{15} & regular & stabiliser\ptn{15} &
$\GL\times N_{\rm blk}$ & M1 & id up to closure (exact); $K_{\rm out}$ shrinks (HF)\ptn{15}\\
BOFT (HF) & R & $\diag(\R^m)\times{}$butterfly product & regular\textdagger & inter-factor & $\mathrm{Mon}\times\Pi$ & M1
& $\to\diag\cdot W_0\cdot{}$\hspace{0pt}$\prod\SO(b\,2^{m'-1})$; $\SO(n)$ only at full depth\\
GOFT (strict) / qGOFT & R & Givens product / free $2\times2$ blocks\ptn{16} & regular & none / torus\ptn{16} &
$\GL\times\Pi$ & M1 & $\to W_0\SO(n)$ / $W_0\R^{n\times n}$\ptn{16}\\
HRA & R & Meet $(W_0{+}\Mr)\cap W_0\SO(n)$ (image level), $rn-\tfrac{r(r+1)}2$ & \textbf{apex} & non-linear, fibre
$\tfrac{r(r+1)}2$ & $\GL\times\CO$ & M1 & $\to W_0\SO(n)$ in $\lceil n/r\rceil$\\
(IA)$^3$ / SSF / RED & R & Lin $=$ closure of a torus orbit & regular & none & $\mathrm{Mon}\times\GL$ /
$\GL\times\mathrm{Mon}$\ptn{17} & M1$\to$ & id\\
RoAd & R & RoAd$_1$: orbit closure of $(\mathbb C^\times)^{m/2}$, $m$; RoAd$_{2,4}$: Lin, $2m$ & regular & none
(RoAd$_4$: per block) & $\mathrm{Mon}_{\rm pair}\times\GL$ & M1 & id\\
Houlsby / Pfeiffer & E & Fun (serial, non-linear) & neutral\ptn{18} & $S_r$ / $S_r\ltimes\R_{>0}^r$\ptn{18} &
$\GL\times\GL$ (fn.) & M$\infty$ & --\\
Parallel adapter & E & Fun\ptn{19} & neutral & $S_r\ltimes\R^r_{>0}$\ptn{19} & $\GL\times\GL$ (fn.) & M$\infty$\ptn{19}
& --\\
Compacter & E & Fun with PHM weights\ptn{20} & neutral & torus\ptn{20} & $\Pi_\otimes$ & M$\infty$ & --\\
LST & E$^\circ$ & Fun (side network)\ptn{21} & unpointed\ptn{21} & $\GL$ (side network) & $\GL\times\GL$ (fn.) &
M$\infty$ & --\\
LLaMA-Adapter & E & Fun (separately normalised prefix) & neutral ($g=0$) & none & $\GL\times\GL$ (fn.) & M$\infty$ &
--\\
LoReFT & E\ptn{22} & LoRA$_r$ on $\mathrm{Aff}(I_d)$, $r(2d{+}1{-}r)$ & defect\ptn{22} & $\Ort(r)$\ptn{22} &
$\GL\times\GL$ (fn.) & M$\infty$ & --\\
Prompt tuning / P-tuning v1 & E$^\circ$ & Fun (input state)\ptn{23} & unpointed & none / non-linear\ptn{23} &
$\GL\times\GL$ (fn.) & M$\infty$ & --\\
Prefix / P-tuning v2 & E$^\circ$ & Fun (gated parallel adapter, per layer) & unpointed & non-linear / none\ptn{24} &
$\GL\times\GL$ (fn.) & M$\infty$ & --\\
MPT / CPT & E$^\circ$ & $P^*\odot uv\T$ / ball\ptn{25} & unpointed / pointed\ptn{25} & torus / none\ptn{25} &
$\GL\times\GL$ / $\CO\times\CO$ (fn.) & M$\infty$ & --\\
GaLore & B & moving Lin $\{P_tX\}$ & regular & none & $\CO\times\CO$ (gradient) & -- & $\to\Theta$\\
GaLore with a random projector & B & moving Lin $\{XA_t\}$ & regular & none & fr & -- & $\to\Theta$\\
Flora & B\ptn{26} & Lin $=\Theta$\ptn{26} & regular & none & fr (in law)\ptn{26} & -- & --\\
LISA & B & moving layer blocks & regular & none & $\GL\times\GL$\ptn{27} & -- & $\to\Theta$\\
MeZO & B & moving random line & regular & none & $\Ort(\Theta)$ in law & -- & $\to\Theta$\\
ReLoRA & B & cone per stage $\to\MM_{\le Kr}$ & apex per stage & $\GL_r$ per stage & $\GL\times\GL$ & M1 & schedule\\
Task arithmetic & P & Lin $\theta_0+\sum\lambda_i\tau_i$ & regular ($\lambda=0$) & none & $\GL(\Theta)$ & M1 & id\\
TIES / DARE & P & Lin in $\lambda$ (coordinatewise rule) & regular & none & $\R^\times\cdot B_N$ /
$\mathrm{Mon}_N$\ptn{28} & M1 & id\\
LoRA soups / CAT & P & $\oplus$ of frozen line methods\ptn{29} & regular & none & $\GL\times\GL$ & M1 & id\\
LoraHub / Poly / MHR & P / F$_T$ & factor-wise product of sums & apex & $\prod\GL_r$, broken by LoraHub\ptn{30} &
$\GL\times\GL$ / $\Pi$\ptn{30} & M1 per task & --\\
MoLE / LoRAMoE / X-LoRA & F$_X$ & input-dependent mixture & neutral / unpointed\ptn{31} & per expert / scalar /
none\ptn{31} & $\GL\times\GL$ (fn.)\ptn{31} & M$\infty$\ptn{31} & --\\
S-LoRA / Punica & F$_U$ & LoRA cone per tenant over a shared base & -- & $\prod_u\GL_{r_u}$ & $\GL\times\GL$ & M1 per
tenant\ptn{32} & --\\
Text-to-LoRA / HyperFormer & F$_T$ & a $T$-point of LoRA / of non-linear adapters & apex / defect\ptn{33} & $\GL_r$ /
$\GL$\ptn{33} & $\GL\times\GL$ & M1 / M$\infty$\ptn{33} & --\\
UniPELT / MAM & E$^\circ$\ptn{34} & Fun, not a $\oplus$-word\ptn{34} & unpointed\ptn{34} & $\GL$\ptn{34} &
$\GL\times\GL$ (fn.) & M$\infty$\ptn{34} & --\\
\end{longtable}\endgroup

\paragraph{Notes on the table}
\textdagger\ BOFT's identity is regular when the factors' plane sets are disjoint (for example $b=2$), because then the
factor Lie algebras form a direct sum. ``$\approx$'' marks implementations whose default initialisation is near, but not
exactly at, $\theta_0$: the FourierFT and C3A defaults in HF PEFT, VB-LoRA, Uni-LoRA, TinyLoRA, and LoRA-XS
($R_0\sim\mathcal N(0,10^{-10})$). They are objects of the unpointed slice with a pointing defect, small for LoRA-XS and
TinyLoRA but not for HF's FourierFT ($\mathcal N(0,1)$ spectrum with scaling $150$, about $14\%$ of a std-$0.04$ $W_0$
at $768^2$ with $1000$ frequencies), and the table classifies their exactly pointed variants. ``$\CO\times\CO$'': frames
built from truncated SVDs are equivariant under positive scalings as well as orthogonal maps, so the spectral-frame
geometry is $\CO(m)\times\CO(n)$ (\thmref{Proposition}{II.16}); bare singular frames used without their singular values
(LoRA-GA) are scale-invariant rather than covariant, which leaves $\Ort(m)\times\Ort(n)$. ``(fn.)'' marks the
function-level analogue of \thmref{Definition}{II.14}: the inserted or generated family on each adapted wire is
invariant under conjugation by $(g,h)$ on that wire. ``--'' means that the axis does not apply (null in the classified
data). Where the classified data refines an entry (Kind, Gauge, Covariance and Merge of LST, UniPELT and MAM, MPT,
Compacter, Houlsby, DeLoRA, AdaLoRA, LoRETTA, LoRA-SB, Trainable Tokens, VB-LoRA, LISA, HyperFormer, S-LoRA and Punica),
the table follows the data.

Dimensions hold under the hypotheses of the results that compute them. HRA's is for $W_0$ of full column rank and even
$r\le n-2$ (\thmref{Theorem}{II.7}); DoRA's is for generic $W_0$ with no zero rows, and it drops for special $W_0$
(\thmref{Proposition}{II.8}); GraLoRA's is for HF's default \texttt{hybrid\_r=0} (\thmref{Proposition}{II.12}); and
LoHa's entry is an upper bound, with equality checked numerically below the $mn$ cap (\thmref{Proposition}{II.12}).

\paragraph{Notes by row}
The superscript numbers in Table~\ref{tab:periodic-methods} refer to these notes.
{\small
\begin{enumerate}[label={\sffamily\bfseries\color{tide}\arabic*},leftmargin=1.6em,itemsep=1pt,topsep=3pt]
  \item rsLoRA and LoRA+ are LoRA up to the hyperparameter torus.
  \item LoRA-Pro, Riemannian preconditioning and LoRA-RITE are reparametrisations with a descending metric (their SGD
  forms). The $\GL_r$ gauge is respected only undamped; LoRA-Pro's parameter step keeps $\Ort(r)$, and LoRA-RITE keeps
  $\GL_r$ on the full-rank locus (\thmref{Theorem}{III.12}).
  \item DeLoRA: its per-term scalings form a $2r$-torus inside the fibre; for $r=1$ the fibre is that torus.
  \item AdaLoRA: the fibre consists of two torus edges and the absorption of $\Lambda$.
  \item LoRA-SB: the covariance is $\R^\times B_m\times\R^\times B_n$, because the frame is the SVD of a sign-gradient.
  \item VB-LoRA: $\Pi$, the chunk partition tied across boxes by the shared bank. Uni-LoRA: a random frame.
  \item OLoRA: the QR frame makes the input side Borel. The catalogue key is $\Pi$, since the order of the inputs
  matters.
  \item LoRA-GA: $\Ort\times\Ort$, from data; its bare singular frames are scale-invariant and depend on the sign
  convention. CorDA: $\CO\times\CO$, from data, at the level of the formula.
  \item Astra: the split is $\Phi(W_0CW_0\T)W_0$ (\thmref{Proposition}{II.16}).
  \item EVA is LoRA with $\mathrm{row}\,A_0$ spanned by principal components of the activations. Its covariance is
  $\GL\times\GL$ at a fixed rank (frame $\CO$), and $\GL\times\CO$ with HF's default \texttt{rho=2}.
  \item LoftQ: the defect is the attained residual when $s=1$; HF ignores the scale $s=\alpha/r$.
  \item TT and Tucker methods sit at the apex for zero-core initialisations. LoRETTA's compensated split initialisation
  is regular, and its shape is a meet of translated cones (Cone$^*$).
  \item FourierFT and WaveFT are regular with \texttt{init\_\allowbreak weights=True}; HF's FourierFT default is a
  defect ($\approx$). The gauge gains one dimension for each sampled conjugate pair.
  \item Trainable Tokens: $\GL$ on the embedding dimensions, Mon on the vocabulary.
  \item OFT: with the exact Cayley map the shape is the orbit of $\prod\SO(b)$ in the level set of $K_{\rm out}$, and
  rebasing is the identity up to closure. HF's default Cayley--Neumann map is only quasi-orthogonal, and its merges
  shrink $K_{\rm out}$ monotonically when $\lVert Q\rVert_{\rm op}\le2^{1/4}$. The gauge is the stabiliser of $W_0$,
  trivial for injective $W_0$; the catalogue key is ``none''.
  \item GOFT and qGOFT: qGOFT replaces each Givens rotation by an arbitrary $2\times2$ block, and its gauge is a torus of
  dimension $d-2$. GOFT's official code adds an input scaler by default (\thmref{Corollary}{II.6}), and the row is for
  strict GOFT. When the planes connect all coordinates, as in GOFT's layout, re-merging reaches $W_0\SO(n)$ for GOFT.
  For qGOFT it reaches $W_0\R^{n\times n}$, the matrices whose columns lie in $\operatorname{col}W_0$: its blocks
  include singular ones and ones of negative determinant, and on connected planes they generate the monoid of all
  $n\times n$ matrices (\thmref{Theorem}{V.3}(b)).
  \item (IA)$^3$, SSF and RED: $\mathrm{Mon}\times\GL$ for output scalings, $\GL\times\mathrm{Mon}$ for scaling
  $W_{\rm down}$.
  \item Houlsby and Pfeiffer adapters are neutral with a zero up-projection; Houlsby's own near-identity initialisation
  leaves a small defect. Houlsby's gauge is only $S_r$ (GELU in the reference code, swish in AdapterHub); Pfeiffer's is
  $S_r\ltimes\R_{>0}^r$ (ReLU).
  \item A parallel adapter with $\sigma=\id$, parallel to one linear box, is LoRA and merges as M1. The gauge
  $S_r\ltimes\R^r_{>0}$ is for ReLU, as in the code of He et al.
  \item Compacter: the PHM weights are Kronecker-structured and shared. The torus consists of per-term slides; a $\GL$
  mixing of the PHM index leaves the rank-$\rho$ class.
  \item LST: the side network is fed through trainable down-projection ladders. The up-projection has rank at most
  $d/r$, so no parameter value reproduces $f$.
  \item LoReFT is an extension at an affine identity box over positions. The pyreft initialisation leaves a defect;
  LoReFT is neutral iff $W=R$ and $b=0$. The catalogue key for its gauge is $\GL$.
  \item Prompt tuning has no gauge. P-tuning v1 adds an LSTM or MLP 2-cell, and its gauge is non-linear, through that
  encoder.
  \item Prefix tuning has a non-linear gauge through the MLP of Li and Liang; P-tuning v2 trains the prefix directly and
  has none.
  \item MPT's prompt is $P^*\odot uv\T$, and MPT is unpointed. CPT's prompt lies in a ball around the demonstrations and
  is pointed at the in-context-learning context. MPT's gauge is a torus of dimension $\ell+d$ with $P^*$ trainable
  ($\R^\times$ with $P^*$ frozen); CPT has none.
  \item Flora compresses only its momentum and takes full-space Adafactor steps (\thmref{Proposition}{IV.2}); its random
  frame is the momentum projection, in law.
  \item LISA: each stage frees whole boxes.
  \item TIES elects signs by total mass, and its covariance is $\R^\times\cdot B_N$. DARE's is $\mathrm{Mon}_N$: the
  torus exactly for each mask, the permutations in law.
  \item CAT is a $\oplus$ of frozen line methods, $\spanop\{B_iA_i\}$ per layer, inside LoRA$_{kr}$.
  \item LoraHub, which combines adapters post hoc, is gauge-dependent and breaks $\prod\GL_r$; Poly and MHR, trained
  jointly, keep a diagonal $\GL_r$. The covariance is $\GL\times\GL$ on LoraHub's products and $\Pi$ (blocks) for MHR.
  \item LoRAMoE is neutral, MoLE and X-LoRA are unpointed. The gauge is $\GL_r$ per expert for LoRAMoE, a scalar for
  MoLE (only the gate trains) and none for X-LoRA (only the head trains). Router and experts transform together.
  Merging is M1 when the gate is constant.
  \item S-LoRA and Punica compute relative to a shared base, and each tenant's fibre is a LoRA cone. Each tenant is
  left unmerged by design, to share the base batch.
  \item Text-to-LoRA is at the apex; HyperFormer's random hypernetworks leave a defect. The gauge is $\GL_r$ on the
  generated rank wire for Text-to-LoRA and $\GL$ on the task-embedding wire for HyperFormer, and factor targets are
  gauge-sensitive. Merging is M1 per task for Text-to-LoRA and M$\infty$ for HyperFormer.
  \item UniPELT's gates also make it an F$_X$. Its adapter and prefix parts sit on extensions, so it is not a
  $\oplus$-word. It is unpointed: the prefix has no neutral value, and UniPELT's gate scales the prefix keys rather
  than their softmax mass. The $\GL$ gauge comes from UniPELT's LoRA summand and MAM's prefix MLP. Merging is M$\infty$,
  also for UniPELT's input-gated LoRA part.
\end{enumerate}}

\subsubsection{Checks on the table}\label{sec:atlas-checks}

Several entries rest on numerical checks reported in Appendix~\ref{app:repro}: the GraLoRA and LoHa dimensions and the
LoHa orbit, the HRA apex (by the Jacobian rank at the paired initialisation), the RoAd orbit dimension, the VeRA rebasing
span, the DoRA dimension caps, the FourierFT conjugate pairs, the LoraHub and Poly fibres, and the Cayley--Neumann
contraction [Num].

The classified data pass the following structural checks. Every arrow and every obstruction joins two catalogued
methods and names a valid rule or test. A check at the level of method identifiers flags $26$ pairs that carry both an
arrow and an obstruction, and $8$ arrows from a method to itself. Each flagged pair concerns two instances of the
target: a different rank, pointing, hyperparameter regime or variant. For example, PiSSA$_r$ maps into LoRA$_{2r}$ at the
degenerate pointing with product $0$ (\thmref{Theorem}{II.1}), while the O3 obstruction concerns LoRA$_r$ at its apex,
whose image is incomparable with PiSSA's (\thmref{Theorem}{II.2}(d)). Each self-arrow is an inclusion between instances
of one family, such as LoRA$_r\hookrightarrow{}$LoRA$_{r+1}$, ReLoRA$^K\hookrightarrow{}$ReLoRA$^{K+1}$ or BOFT with
$m'$ butterfly factors inside BOFT with $m'+1$.

The data deviate from the conventions of \S\ref{sec:atlas-conventions} in eight recorded places, which are kept as
published in the ancillary files.
\begin{enumerate}[label=(\arabic*)]
  \item Sixteen function-level families and post-hoc compositions carry no covariance (LoRAMoE, MixLoRA, HydraLoRA,
  ICEdit, MoV, Lily, MoLE, X-LoRA, PHATGOOSE, Arrow, LoraRetriever, LoRA-Flow, SMEAR, GOAT, AdapterSoup and LoRA
  Switch). Under the conventions they carry their expert family's function-level value: $\GL\times\GL$ for dense LoRA
  experts and routers, and $\mathrm{Mon}\times\GL$ for MoV.
  \item Initial-state tuning and state-offset tuning are M$\infty$ with rebasing ``idempotent''; under the conventions
  their rebasing is ``--''.
  \item Flora is coded as a per-period linear method with rebasing ``schedule''. Flora takes full-space steps and
  compresses only its momentum (\thmref{Proposition}{IV.2}), so its per-step shape is all of $\Theta$ and its rebasing
  is ``--'', as in Table~\ref{tab:periodic-methods}.
  \item EVA is coded $\GL\times\GL$, which holds at a fixed rank; HF's default \texttt{rho=2} gives $\GL\times\CO(n)$
  (\thmref{Proposition}{II.16}).
  \item OFTv2's rebasing note says that its Cayley--Neumann factors generate $\prod\GL^+(b)$ and that $K_{\rm out}$
  drifts. \thmref{Theorem}{V.3} gives a contraction semigroup, with monotone shrinkage of $K_{\rm out}$, for small
  generators.
  \item Fisher merging's optional mixing weights would make it a coefficient family under the post-hoc convention.
  \item The entries MoV / MoLoRA and Child-Tuning (its D and F variants) each cover two variants.
  \item The ``core'' flag is not a coordinate. It marks every method named in Table~\ref{tab:periodic-methods} plus a
  curated set.
\end{enumerate}

\subsubsection{The whole catalogue}\label{sec:atlas-catalogue}

Everything above rests on a catalogue of \nMethods{} methods and \nPapers{} papers, checked against arXiv and, for
library claims, against Hugging Face PEFT \citep{xmangrulkar2022peft} as of 30 September 2026; \nHF{} of the methods
ship in that library. Each method entry gives the formula, budget, locus, kind, merge verdict, initialisation, relations,
sources, coordinates and a categorical reading, most with arrows and obstructions. Appendix~\ref{app:catalogue} lists
every method with its seven coordinates and its reference (Table~\ref{tab:catalogue}) and every arrow with its map $h$
(Table~\ref{tab:arrows}); the ancillary files \texttt{catalogue.csv}, \texttt{arrows.csv} and
\texttt{obstructions.csv} carry the full entries. Table~\ref{tab:atlas-counts} counts the methods per coordinate value.
Cones and linear shapes account for $240$ of the \nMethods{} methods, and apex and regular base points are about equally
common.

Earlier maps made this one possible: the surveys of \citet{lialin2023scaling}, \citet{xu2023parameter},
\citet{han2024parameter} and \citet{mao2024survey}, the delta-tuning study of \citet{ding2022parameter}, and the
libraries AdapterHub \citep{pfeiffer2020adapterhub}, Adapters \citep{poth2023adapters} and Hugging Face PEFT
\citep{xmangrulkar2022peft}, whose source settled many details.

\begin{table}[t]
\centering\footnotesize
\caption{The classification of all \nMethods{} catalogued methods: the number of methods with each coordinate value,
over all methods / over the \nCore{} core methods. ``--'' counts null coordinates, including the legacy merge key
\texttt{na}. Per-method values are in Table~\ref{tab:catalogue}.}\label{tab:atlas-counts}
\begin{tabularx}{\linewidth}{@{}>{\raggedright\arraybackslash}p{1.9cm}>{\raggedright\arraybackslash}X@{}}
\toprule\headrow
\textbf{Axis} & \textbf{Value: all / core}\\\midrule
Kind & R $208/85$ $\cdot$ E $33/10$ $\cdot$ E$^\circ$ $27/13$ $\cdot$ B $29/6$ $\cdot$ P $31/11$ $\cdot$ F$_T$ $54/17$\\
Shape & Lin $106/45$ $\cdot$ Cone $134/43$ $\cdot$ Cone$^*$ $20/9$ $\cdot$ Orb $13/7$ $\cdot$ Sat $5/3$ $\cdot$ Meet
$2/2$ $\cdot$ Fun $93/32$ $\cdot$ -- $9/1$\\
Base point & regular $113/56$ $\cdot$ apex $111/40$ $\cdot$ neutral $40/13$ $\cdot$ defect $59/13$ $\cdot$ unpointed
$57/18$ $\cdot$ -- $2/2$\\
Gauge & none $139/62$ $\cdot$ scalar $9/6$ $\cdot$ torus $40/9$ $\cdot$ $\GL$ $152/51$ $\cdot$ $\GL\times{}$torus $3/1$
$\cdot$ non-linear $23/10$ $\cdot$ -- $16/3$\\
Covariance & $\GL\times\GL$ $166/53$ $\cdot$ $\CO\times\CO$ $47/20$ $\cdot$ $\GL\times\CO$ $12/5$ $\cdot$
$\mathrm{Mon}\times\GL$ $21/10$ $\cdot$ $\mathrm{Mon}\times\mathrm{Mon}$ $52/18$ $\cdot$ $\Pi$ $49/19$ $\cdot$ frame
$19/10$ $\cdot$ -- $16/7$\\
Merge & M1 $242/98$ $\cdot$ M1$\to$ $11/5$ $\cdot$ Mq $11/2$ $\cdot$ M$\infty$ $97/32$ $\cdot$ -- $21/5$\\
Rebasing & idempotent $92/43$ $\cdot$ $\to$span $130/45$ $\cdot$ $\to$group $12/7$ $\cdot$ schedule $29/9$ $\cdot$ --
$119/38$\\
\bottomrule
\end{tabularx}
\end{table}

\begin{figure}[tbp]
\centering
\includegraphics[width=\linewidth]{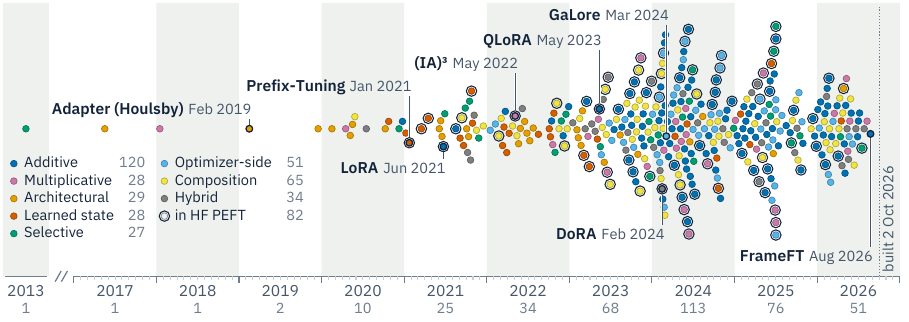}
\caption{\textbf{The catalogue in time.} Each of the \nMethods{} catalogued methods is one dot, placed at the month of its paper's first arXiv submission, read from the YYMM prefix of its identifier; within a month the dots are spread evenly in the order of their arXiv numbers, and the six entries without an identifier sit at mid-year. The dots form a centred beeswarm, coloured by modification kind (legend, with the number of entries of each kind) and ringed when the method ships in Hugging Face PEFT (\nHF{} entries). The one entry before 2017, linear probing on DeCAF features \citep{donahue2013decaf}, sits left of the axis break. The number under each year counts its entries, from $10$ in 2020 to $113$ in 2024; the busiest month is February 2024, with $20$. The 2026 count ends at the catalogue's build date, 2~October~2026 (dotted line). Labels mark the bottleneck adapter \citep{houlsby2019parameter}, prefix tuning \citep{li2021prefix}, LoRA \citep{hu2021lora}, (IA)$^3$ \citep{liu2022few}, QLoRA \citep{dettmers2023qlora}, DoRA \citep{liu2024dora}, GaLore \citep{zhao2024galore} and the newest entry, FrameFT \citep{adepu2026fine}. Additive weight edits are the largest kind ($120$ entries), followed by composition and routing ($65$) and optimizer-side methods ($51$).}\label{fig:timeline}
\end{figure}

Figure~\ref{fig:timeline} dates each method by its first arXiv submission. Landmarks come in rough order, with
bottleneck adapters in 2019 \citep{houlsby2019parameter}, prefixes \citep{li2021prefix} and LoRA's cone
\citep{hu2021lora} in 2021, the torus of (IA)$^3$ in 2022 \citep{liu2022few}, base change with QLoRA in 2023
\citep{dettmers2023qlora}, and GaLore's moving frame in 2024 \citep{zhao2024galore}, the year of $113$ of the
\nMethods{} methods.

\subsubsection{What a budget buys}\label{sec:atlas-budget}

\thmref{Proposition}{II.12} separates the budget $\lvert M\rvert$, what a method trains, from the expressive dimension
$d(M)$, the number of directions in which it can move the weights. The gap is gauge waste (\thmref{Definition}{I.5}): per
matrix, $r^2$ for LoRA, one for VeRA (numerically), zero for injective linear methods, and at least $2r^2+m+n-1$ for
LoHa$_{r,r}$ (\thmref{Proposition}{II.12}). At $131{,}072$ parameters on a $4096\times4096$ matrix, LoRA$_{16}$ reaches
$130{,}816$ directions at rank up to $16$, and LoHa$_{8,8}$ at most $122{,}753$ at rank up to $64$. Rankings at matched
budget and at matched $d(M)$ can differ (\thmref{Prediction}{X.10}). Figure~\ref{fig:rankvol} compares maximum rank and
$d(M)$ at equal budget for eight methods on such a layer, and the companion site's budget calculator computes both counts
for the adapted matrices of common base models.

\subsection{Arrow rules: when \texorpdfstring{$N$}{N} simulates \texorpdfstring{$M$}{M}}\label{sec:atlas-arrows}

``Generalises'' and ``is a special case of'' get one reading here. $N$ simulates $M$ when a pointed smooth map
$h:Q_M\to Q_N$ satisfies $\rho_N\circ h=\rho_M$ (\thmref{Definition}{I.2}), so $N$ reaches every weight that $M$
reaches. A morphism $M\to N$ is exhibited by an explicit pointed $h$. Five rules generate every arrow in the atlas, and
three tests rule arrows out (Table~\ref{tab:atlas-rules}).

\begin{table}[t]
\centering\footnotesize
\caption{The arrow rules A1--A5, transport T and the obstruction tests O1--O3. An arrow $M\to N$ is a pointed smooth
map $h:Q_M\to Q_N$ with $\rho_N\circ h=\rho_M$.}\label{tab:atlas-rules}
\begin{tabularx}{\linewidth}{@{}>{\raggedright\arraybackslash}p{1.8cm}>{\raggedright\arraybackslash}X%
>{\raggedright\arraybackslash}p{3.65cm}>{\raggedright\arraybackslash}p{4.5cm}@{}}
\toprule\headrow
\textbf{Rule} & \textbf{Statement} & \textbf{The map $h$, or the test} & \textbf{Examples}\\\midrule
\textbf{A1}\newline interval & Frozen${}\to M\to{}$Full FT for every weight-space $M$. & $h=\mathrm{const}_{q_0}$; $h=\rho_M$ &
Merging is the arrow to Full FT.\\
\textbf{A2}\newline freezing & Freezing coordinates at their base values gives a sub-object. & $h(q')=(q',q_0'')$ &
$\mathrm{FA}(A_0)\hookrightarrow{}$LoRA pointed at $(0,A_0)$; MiCA${}\hookrightarrow{}$LoRA pointed at $(B_0,0)$.\\
\textbf{A3}\newline factorisation & $\rho_M=\rho_N\circ h$ for an explicit pointed map $h$, usually polynomial. & VeRA${}\to
\mathrm{FA}(A)$: $h(b,d)=\Lambda_bB\Lambda_d$. LoHa$_{r_1,r_2}\to{}$LoRA$_{r_1r_2}$: face-splitting products.
(IA)$^3\to{}$HiRA$_1$ pointed at $(0,\mathbf 1\T)$: $h(\ell)=(\ell-\mathbf 1,\mathbf 1\T)$. & Isomorphisms:
rsLoRA${}\cong{}$LoRA by $h=(\sqrt r\,B,A)$; MiSS (balance mode) ${}\cong\mathrm{FA}(\mathbf 1\T\otimes I_r)$; a linear
adapter parallel to one layer ${}\cong{}$LoRA.\\
\textbf{A4}\newline products & A product comes with its two projections. & $R\mapsto B_0R$, $R\mapsto RA_0$ &
LoRA-XS${}\to\mathrm{FA}(A_0)$, $\mathrm{FB}(B_0)$. HRA's image lies in those of OFT$_{\SO}$ and LoRA$_r$, an
image-level inclusion only.\\
\textbf{A5}\newline sums & $M\to M\oplus N$, and $K$ stages embed in $K+1$. & $q\mapsto(q,q_0^N)$ &
LoRA$_r\oplus{}$LoRA$_s\cong{}$LoRA$_{r+s}$ (unpointed); ReLoRA$^K\to{}$ReLoRA$^{K+1}$.\\\midrule
\textbf{T}\newline transport & $T_*$ for $T\in\GL(\Theta)$ is an automorphism of the slice, giving one object in two frames. & no
arrow; record the frame $T$ & Sums of Kronecker products are $\mathcal R^{-1}_*{}$LoRA; HiRA is $(D_{W_0})_*{}$LoRA at
$\theta_0$ when $W_0$ has no zero entry.\\\midrule
\textbf{O1}\newline image & No arrow if $\Im M\not\subseteq\Im N$, as when $N$ keeps an invariant that $M$ breaks. & compare
images & LoRA${}\not\to{}$exact-Cayley OFT, which keeps $K_{\rm out}$.\\
\textbf{O2}\newline size & No arrow if $d(M)>d(N)$ or $\rho_{\max}(M)>\rho_{\max}(N)$. & compare dimension and maximum rank &
LoRA$_r\not\to{}$LoRA-FA, since $r(m+n-r)>mr$.\\
\textbf{O3}\newline first order & No arrow if $S_1(M)\not\subseteq S_1(N)$, or if $M$ and $N$ lie in different strata of
\thmref{Theorem}{II.2}. & compare $S_1$ and strata & $\mathrm T_A$-${}\not\to{}\mathrm T_B$-pointed LoRA; PiSSA and
LoRA, either way; LoRA${}\not\to{}$AdaLoRA, though on one matrix their images agree.\\
\bottomrule
\end{tabularx}
\end{table}

\subsubsection{The five rules}\label{sec:atlas-rules}

Write LoRA$_{(B_0,A_0)}$ for LoRA pointed at $(B_0,A_0)$, and let $\mathrm{FA}(A_0):B\mapsto W_0+BA_0$ and
$\mathrm{FB}(B_0):A\mapsto W_0+B_0A$ be the two one-sided LoRAs of \thmref{Proposition}{I.8}.
\begin{itemize}
  \item \textbf{A1, interval.} Frozen${}\to M\to{}$Full FT, with $h=\mathrm{const}_{q_0}$ and $h=\rho_M$ respectively.
  \item \textbf{A2, freezing gives sub-objects.} Freezing a block of coordinates $q''$ at its base value $q_0''$
  gives a monomorphism $h(q')=(q',q_0'')$. Examples: $\mathrm{FA}(A_0)\hookrightarrow{}$LoRA$_{(0,A_0)}$, $\mathrm{FB}(B_0)\hookrightarrow{}$
  LoRA, MiCA${}\hookrightarrow{}$LoRA$_{(B_0,0)}$, and DyLoRA$_k\hookrightarrow{}$DyLoRA$_{k+1}$ by zero padding.
  \item \textbf{A3, factorisation through a polynomial map.} If $\rho_M=\rho_N\circ h$ for an explicit smooth pointed $h$,
  then $M\to N$; one checks $h(q_0^M)=q_0^N$. Examples:
  \begin{itemize}
    \item VeRA${}\to\mathrm{FA}(A)$, $h=(\Lambda_bB\Lambda_d)$;
    \item NOLA${}\to{}$LoRA$_r$, $h=(\sum_i\alpha_iB_i,\sum_j\beta_jA_j)$;
    \item LoHa$_{r_1,r_2}\to{}$LoRA$_{r_1r_2}$, by the face-splitting product
    $h=\big(B_1\bullet B_2,(A_1\T\bullet A_2\T)\T\big)$ (\bgref{B.13});
    \item LoRA$_r\hookrightarrow{}$LoRA$_{r+1}$, $h=\big([B\ 0],[A;a_0]\big)$;
    \item DyLoRA$_k\hookrightarrow{}$DyLoRA$_{k+1}$;
    \item LoRA-XS${}\to{}$PiSSA, $h(R)=\big(U_r(S_r+R)S_r^{-1/2},\ S_r^{1/2}V_r\T\big)$;
    \item HRA$_r\to{}$LoRA$_r$ pointed at $h(q_0)$, via $R-I=\sum_iH_1\cdots H_{i-1}(H_i-I)$;
    \item (IA)$^3\to{}$HiRA$_1$ pointed at $(0,\mathbf 1\T)$, $h(\ell)=(\ell-\mathbf 1,\mathbf 1\T)$;
    \item isomorphisms: MiSS${}\cong\mathrm{FA}(\mathbf 1\T\otimes I_r)$; rsLoRA${}\cong{}$LoRA, $h=(\sqrt r\,B,A)$; a
    linear adapter parallel to one linear box${}\cong{}$LoRA (\thmref{Proposition}{VI.4}).
  \end{itemize}
  \item \textbf{A4, products and meets.} A product comes with two projections, and it is universal among the methods
  that both factors simulate. Examples: LoRA-XS${}\to\mathrm{FA}(A_0)$ and LoRA-XS${}\to\mathrm{FB}(B_0)$, at the level of
  objects; HRA${}\to{}$OFT$_{\SO}$ and HRA${}\to{}$LoRA$_r$, at the level of images only. An intersection of images
  gives only inclusions of images, not arrows of $\Meth$; for example $\Im{}$LoRA$_r\cap\Im{}$(IA)$^3$ exists only as a
  set, and there is no product in $\Diff$.
  \item \textbf{A5, sums.} $M\to M\oplus N$, $q\mapsto(q,q_0^N)$, for additive $N$ pointed at $\theta_0$; for example
  LoRA$_r\to{}$LoRA$_r\oplus{}$sparse (RoSA, \citealp{nikdan2024rosa}). Also
  LoRA$_r\oplus{}$LoRA$_s\cong{}$LoRA$_{r+s}$, unpointed (\thmref{Observation}{I.7}), by concatenation with the
  $A$-blocks rescaled under $\alpha/r$ scales; it is pointed exactly at the concatenated initialisation, which is a
  standard zero-$B$ pointing iff the stacked $A_0$ has full row rank and $r+s<\min(m,n)$. Rebasing stages give
  $M^{\oplus K}\to M^{\oplus(K+1)}$.
\end{itemize}

\paragraph{Transport is not an arrow}
$T_*$ for $T\in\GL(\Theta)$ is an automorphism of the slice (\thmref{Proposition}{II.13}). Methods related by
transport, such as the Kronecker family and LoRA (across $\Theta'\cong\Theta$), are ``the same object in another
frame''. Sums of $k$ Kronecker terms are $\mathcal R^{-1}_*{}$LoRA$_k$ after the Van Loan--Pitsianis rearrangement
(\thmref{Theorem}{II.11}); HiRA is $(D_{W_0})_*{}$LoRA inside the slice at $\theta_0$ when $W_0$ has no zero entries
(\thmref{Proposition}{II.9}); FourierFT is a mask in the real cosine Fourier frame, a transport only without sampled
conjugate pairs and only for \texttt{init\_weights=True}. Transports that depend on $\theta_0$ (HiRA, SVD frames) agree
with their source only at the base point. For additive families and a base-independent $T$, transport also commutes
with merge-and-restart; for action-type or $\theta_0$-dependent transports it need not.

\subsubsection{Three tests}\label{sec:atlas-tests}

There is no arrow $M\to N$ if any of the following holds.
\begin{itemize}
  \item \textbf{O1, image.} $\Im M\not\subseteq\Im N$, for instance because $N$ preserves an absolute invariant (Gram
  matrix, spectrum, row lines, zero pattern) that $M$ breaks. Examples: LoRA${}\not\to{}$OFT, LoRA${}\not\to{}$(IA)$^3$,
  DoRA${}\not\to{}$LoRA$_r$.
  \item \textbf{O2, dimension and cut-rank.} $d(M)>d(N)$, or $\rho_{\max}(M)>\rho_{\max}(N)$. Examples:
  LoHa$_{r,r}$ does not map into LoRA$_{r'}$ for $r'<r^2$, since its generic rank is $r^2$; and LoRA$_r$ does not map
  into $k$-term Kronecker sums for $k<\min(m_1,m_2)\min(n_1,n_2)$, the Kronecker rank of a generic rank-one update
  (\thmref{Theorem}{II.11}).
  \item \textbf{O3, first order and base point.} $S_1(M)\not\subseteq S_1(N)$, or $M$ and $N$ lie in different strata
  of \thmref{Theorem}{II.2} (Type I apex against Type II smooth point); the $S_1$ test applies even when the images
  agree. Examples: $\mathrm T_A$-stratum LoRA${}\not\to{}\mathrm T_B$-stratum LoRA; PiSSA${}\not\to{}$LoRA and
  LoRA${}\not\to{}$PiSSA.
\end{itemize}

\paragraph{Why the rules give arrows and the tests rule them out}
Each rule and test is an instance of a result stated earlier. In each rule $h$ is smooth and pointed, and
$\rho_N\circ h=\rho_M$ is checked by multiplying out. A1 is \thmref{Observation}{I.4}(a), since
$\rho_M\circ\mathrm{const}_{q_0}=\theta_0$ and $\id\circ\rho_M=\rho_M$. A2 holds because freezing $q''$ at $q_0''$
leaves the rest of the formula unchanged. For A3, take VeRA, where $W_0+\Lambda_bB\Lambda_dA=W_0+(\Lambda_bB\Lambda_d)A$
and $h$ sends $b_0=0$ to the base point $0$ of LoRA-FA. A4 is the universal property of the fibre product
(\thmref{Observation}{I.4}(b), \thmref{Proposition}{I.8}(a)); for a meet such as HRA only the inclusion of images holds
(\thmref{Theorem}{II.7}). A5 is the identity
\begin{equation}\label{eq:atlas-sum}
  \rho_{M\oplus N}(q,q_0^N)=\theta_0+\delta_M(q)+\delta_N(q_0^N)=\rho_M(q),
\end{equation}
from \thmref{Observation}{I.7}.

An arrow gives $\Im M=\rho_N(h(Q_M))\subseteq\Im N$, which is O1. Images are semialgebraic, or definable when softmax
enters (\bgref{C.17}), so dimension is monotone under inclusion; and every update in $\Im M-\theta_0$ also lies in
$\Im N-\theta_0$, so its rank is bounded by $\rho_{\max}(N)$. Together these give O2. The chain rule at $q_0$ gives
$S_1(M)\subseteq S_1(N)$ (\thmref{Observation}{I.6}), the first half of O3. For the strata, the images of $\mathrm T$-type
and $\mathrm S$-type objects are incomparable (\thmref{Theorem}{II.2}(d)). Inside the $\mathrm T$ strata,
$S_1(\mathrm T_A)=\{XA_0\}$ contains matrices whose column space leaves $\mathrm{col}\,B_0'$, so it does not lie in
$S_1(\mathrm T_B)=\{B_0'Y\}$, and symmetrically.

Transport $T_*$ is an automorphism of the slice with inverse $(T^{-1})_*$ (\thmref{Proposition}{II.13}). It identifies
$M$ and $T_*M$ as one object in two frames and does not by itself give an arrow between them.

\paragraph{Which published relations survive}
Some become exact. VeRA is LoRA-FA$(A)$ with a constrained factor, and rsLoRA and a linear parallel adapter are LoRA.
Others are weaker than they sound. HRA's arrow to LoRA$_r$ lands on a degenerate pointing ($B_0A_0=0$) outside the strata
of \thmref{Theorem}{II.2}, and LoRA is a special case of AdaLoRA only unpointed, since their first-order images have
dimensions $mr$ and $r$ (\thmref{Lemma}{II.4}).

\subsubsection{The recorded arrows and obstructions}\label{sec:atlas-graph}

\begin{figure}[tbp]
\centering
\includegraphics[width=\linewidth]{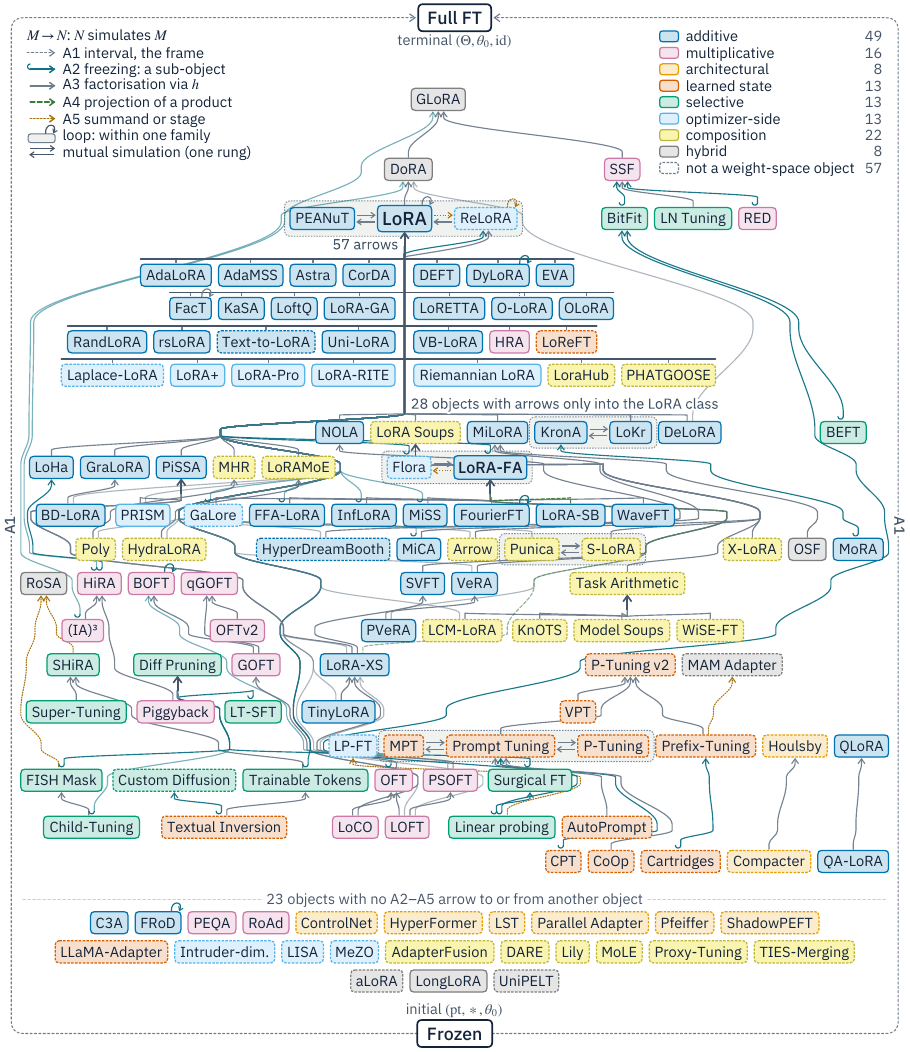}
\caption{\textbf{The category of core methods.} A layered Hasse diagram (\bgref{A.16}), from Frozen (initial, bottom) to Full FT (terminal, top); the dashed outer frame stands for the A1 arrows Frozen${}\to M\to{}$Full FT of every weight-space method $M$ (\thmref{Observation}{I.4}(a)). Chips are the \nCore{} core methods, coloured by modification kind and dashed for the $57$ that are not objects of the weight-space slice. An arrow $M\to N$ points up and means that $N$ simulates $M$ through a pointed smooth map $h$ with $\rho_N\circ h=\rho_M$ (\thmref{Definition}{I.2}). All $161$ recorded arrows between two core methods are drawn, coloured by rule (Table~\ref{tab:atlas-rules}): $29$ by freezing (A2, hooked tail), $122$ by factorisation (A3), $2$ by products (A4) and $8$ by sums or stages (A5). The $7$ loops land on another member of a family, as $\mathrm{LoRA}_r\to\mathrm{LoRA}_{r+1}$ does, and the $5$ classes of mutually simulating families share a rung in a grey box. Arrows into a method with four or more incoming arrows from outside its class end in one trunk ($57$ into LoRA), and the $28$ methods whose arrows all land in the LoRA class form the comb beneath it. The $26$ lighter arrows are composites, also reached by a longer path of arrows.}\label{fig:graph}
\end{figure}

The atlas records \nArrows{} arrows, each with its map $h$ (Table~\ref{tab:arrows} in Appendix~\ref{app:catalogue}):
$329$ by factorisation (A3), $61$ by freezing (A2), $11$ by sums (A5) and $2$ by products (A4). The interval arrows of
A1 hold for every weight-space method and are not listed. The atlas also records \nObstructions{} obstructions, each
naming the test that rules an arrow out (Table~\ref{tab:obstructions}). Of the recorded arrows, $161$ join two core
methods, and Figure~\ref{fig:graph} draws the category of core methods as a layered Hasse diagram (\bgref{A.16}).

\begin{table}[t]
\centering\footnotesize
\caption{The \nObstructions{} recorded obstructions $M\not\to N$, by the test that rules the arrow out. The full list,
with the failing invariant of each, is the ancillary file \texttt{obstructions.csv}.}\label{tab:obstructions}
\begin{tabular}{@{}lrl@{}}\toprule\headrow\textbf{Test} & \textbf{Count} & \textbf{What fails}\\\midrule
O1 & 288 & an image invariant (Gram matrix, spectrum, row lines, zero pattern) or image inclusion\\
O2 & 129 & dimension or cut-rank\\
O3 & 68 & first-order image or base-point stratum\\
\bottomrule\end{tabular}

\end{table}

The atlas is meant to grow. A new method enters by answering the seven questions and exhibiting its arrows, each with its
map $h$; a disputed comparison is settled by an explicit $h$ or by one of the three tests.

\section{Rosetta stone}\label{sec:rosetta}

\begin{flushright}
\small\itshape A half-seen analogy between two theories is the most fertile thing a mathematician can hold.\\
One day the hunch turns into certainty, and the twin theories show the source they share.\\[2pt]
{\upshape\footnotesize after Weil's Rosetta-stone letter of 1940}
\end{flushright}

This exposé is a dictionary from fine-tuning phrases to the objects of the relative theory, with the result behind each
pairing that needs one.

\paragraph{In brief}
\begin{itemize}
  \item LoRA+ is $\alpha$ in disguise. Under Adam with $\varepsilon=0$, no weight decay or clipping and one schedule,
    LoRA+ with ratio $\lambda$ trains exactly like LoRA with $\alpha$ multiplied by $\lambda$, given the same data order
    and dropout masks \citep{schulman2025lora}.
  \item Under Adam without weight decay or full-gradient clipping, GaLore is one-sided LoRA, restarted every period with
    its optimizer state kept \citep{xtorrobahennigen2025duality}. QLoRA is LoRA attached at the quantised weight, off
    $\theta_0$ by a generically full-rank error.
  \item Read from the mathematical side, some constructions are known methods. In the SVD frames of $W_0$, when
    $\rk W_0\ge r$, the product of the two one-sided LoRAs is LoRA-XS, and for injective $W_0$ and even $r\le n-2$
    LoRA's image meets the rotation orbit of $W_0$ in the image of HRA (without Gram--Schmidt, the HF default).
\end{itemize}

A fine-tuning paper says that LoRA+ gives $B$ a larger learning rate than $A$. The relative theory says that, under the
hypotheses of \thmref{Theorem}{III.10}, LoRA+ lies on the torus orbit of plain LoRA with a larger $\alpha$. Passing
between such sentences lets each side borrow the other's theorems. Table~\ref{tab:rosetta}, at the end of this exposé,
is a dictionary for the passage. It lists all 36 entries, each a PEFT phrase, its categorical or geometric counterpart,
and a gloss citing the results behind it; the hypotheses live in the cited statements. The sections below read six
entries closely, one from each of Exposés~\ref{sec:slice} to~\ref{sec:merge}, and then read the dictionary from the
right.

\subsubsection{How to read an entry}\label{sec:rosetta-read}

An entry claims that a phrase names a definite object, on which claims about the phrase can be checked. There are three
strengths (Table~\ref{tab:rosetta-strengths}).

\begin{table}[t]
\centering\small
\caption{The three strengths of a dictionary entry.}\label{tab:rosetta-strengths}
\renewcommand{\arraystretch}{1.15}
\begin{tabularx}{\linewidth}{@{}>{\raggedright\arraybackslash}p{1.7cm}>{\raggedright\arraybackslash}p{4.4cm}>{\raggedright\arraybackslash}X@{}}
  \toprule
  \headrow \textbf{Strength} & \textbf{What the entry asserts} & \textbf{Example}\\
  \midrule
  Definition & The phrase gets a precise, computable meaning. & ``Effective number of parameters'' is $d(M)$, the
    generic rank of $d\rho$ (\thmref{Definition}{I.5}).\\
  Theorem & The phrase hides a claim, proved under stated hypotheses. & ``No initialisation buys rank above $2r$''
    means that on one $m\times n$ box the exact pointings of $\mathrm{LoRA}_r$ together reach
    $\theta_0+\MM_{\le\min(2r,m,n)}$ (\thmref{Theorem}{II.1}).\\
  Analogy & A borrowed word names a pattern and adds nothing to the theorem behind it. & ``Noether charge'', ``gauge'',
    ``Erlangen'', ``moduli space''.\\
  \bottomrule
\end{tabularx}
\end{table}

\paragraph{Credit}
The language of parametrised maps and lenses on our right-hand side comes from \citet{fong2017backprop} and
\citet{cruttwell2021categorical}. Inside PEFT, \citet{he2021unified} wrote an earlier dictionary of this kind, reading
adapters, prefix tuning and LoRA as modifications of one hidden state.

The box below fixes each borrowed word's sense. Exposé~\ref{sec:analogy} says more
(Table~\ref{tab:analogy}).

\begingroup\renewcommand{\label}[1]{}%
\begin{analogy*}{the borrowed words}
\begin{itemize}
  \item \emph{Noether}: conserved charges from symmetries, proved for continuous-time flows with constant rates
    (\thmref{Theorem}{III.5}); Adam and momentum break them.
  \item \emph{Gauge}: the symmetry group of a parametrisation, with no connection.
  \item \emph{Erlangen}: a theorem for group-type methods (\thmref{Theorem}{II.5}), and a name for the shape of cones
    such as LoRA's image.
  \item \emph{Moduli space}: LoRA initialisations in three full-rank strata, classified only as a set
    (\thmref{Theorem}{II.2}).
  \item \emph{Lens duality}: exact only at degree one, for optimizers that read neither the parameters nor the full
    gradient (\thmref{Proposition}{IV.2}).
\end{itemize}
\end{analogy*}
\endgroup

Most of the right-hand column concerns images, which measure expressivity. Learnability is a separate property, and
entries with dynamical claims state their optimizer.

\subsubsection{Six entries, one from each exposé}\label{sec:rosetta-six}

\paragraph{I. ``Method $N$ can do whatever $M$ can''}

\begin{example*}{entries 3 and 4, Exposé~\ref{sec:slice}}
\labelsf{PEFT}\enspace ``method $N$ can do whatever $M$ can''; the frozen model and full fine-tuning.\par
\noindent\labelsf{Category theory\,\textperiodcentered\,geometry}\enspace a morphism $h:M\to N$ of the pointed slice, a
pointed smooth map with $\rho_N\circ h=\rho_M$; the initial and terminal objects.
\end{example*}

The weak reading asks that $N$ reach every weight $M$ reaches. The useful one asks that $N$ also follow $M$'s parameter
paths, through a map $h$ with $\rho_N\circ h=\rho_M$. Methods at $\theta_0$ with such arrows form the pointed slice
$\Meth(\Theta,\theta_0)$ (\thmref{Definition}{I.2}). The frozen model is initial, full fine-tuning is terminal, and the
arrow from $M$ to full fine-tuning is $\rho_M$ itself (\thmref{Observation}{I.4}(a)). Image inclusion gives only a map of
sets over $\Theta$, which need not be smooth (\thmref{Observation}{I.4}(c)).

\begin{explains}
Every weight-space method pointed at $\theta_0$ lies between the frozen model and full fine-tuning, and in exact
arithmetic on a full-precision base its merge is that canonical arrow. Merging questions remain for extensions such as
adapters, prefixes and ReFT (reading VI below), for quantised bases (\thmref{Proposition}{V.2}) and for combinations of
adapters (\thmref{Proposition}{V.4}).
\end{explains}

\paragraph{A first-order test, and defaults that start elsewhere}
The first-order image $S_1$ detects part of the gap between sets and smooth maps. Take $\rho_M(t)=t$ and
$\rho_N(u)=u^3$ on $\R$, both pointed at $0$. The images agree, but $S_1(M)=\R$ is not inside $S_1(N)=0$, so no smooth
simulation $M\to N$ exists (\thmref{Observation}{I.4}(c), \thmref{Observation}{I.6}). Several initialisations start
away from $\theta_0$ and carry a pointing defect (Exposé~\ref{sec:base}): HF's default FourierFT spectrum and VB-LoRA
bank, \texttt{init\_lora\_weights=False}, LoRA-XS's small random $R$, and quantised bases.

\paragraph{II. ``The effective number of parameters''}

\begin{example*}{entries 6 and 11, Exposé~\ref{sec:image}}
\labelsf{PEFT}\enspace expressivity, the effective number of parameters; redundant parameters.\par
\noindent\labelsf{Category theory\,\textperiodcentered\,geometry}\enspace the image germ of $\rho$ at $\theta_0$ and its
dimension $d(M)$; the gauge group (automorphisms of $Q$ over $\Theta$) and the generic fibre.
\end{example*}

A parameter count measures $Q$, while expressivity lives in the image. The difference is spent on the fibres of $\rho$,
along which the gauge group moves. LoRA spends $r^2$ parameters there, since $(Bg^{-1},gA)$ and $(B,A)$ give the same
weights for every $g\in\GL_r$.

With $\lvert M\rvert=\dim Q$ and $d(M)=\dim\Im M$ (\thmref{Definition}{I.5}),
\begin{equation}\label{eq:rosetta-gauge}
  d(M)\le\lvert M\rvert-\dim(\text{generic gauge orbit}),
\end{equation}
with equality when the gauge is locally transitive on generic fibres (\thmref{Proposition}{II.12}(a)). For
$\mathrm{LoRA}_r$ on an $m\times n$ box with $r\le\min(m,n)$, $d=r(m+n-r)$. Compare $\mathrm{LoHa}_{r,r}$ with
$\mathrm{LoRA}_{2r}$ at their shared budget $2r(m+n)$, with $2r\le\min(m,n)$. If $2r^2<m+n-1$, LoHa reaches at least
$(m+n-1)-2r^2$ fewer dimensions; if moreover $r^2\le\min(m,n)$, it trades them for rank up to $r^2$ when $r\ge3$, and
it lies strictly inside $\mathrm{LoRA}_{2r}$ when $r\le2$ (\thmref{Proposition}{II.12}(c)). If $2r^2\ge m+n-1$, as for
$r=64$ on $4096\times4096$ or for $r=32$ on layers 320 to 640 wide, LoHa matches or beats $\mathrm{LoRA}_{2r}$ in
rank, and in dimension by its numerically checked dimension formula. Rank, in turn, is bounded by a minimum cut of the
method's tensor network (\thmref{Theorem}{II.10}).

\begin{explains}
Report $d(M)$ beside $\lvert M\rvert$ in every comparison (\thmref{Prediction}{X.10}).
\end{explains}

\paragraph{Credit}
LoHa is the Hadamard-product parametrisation of FedPara \citep{hyeonwoo2021fedpara}. Its dimension bound is proved, and
equality is checked numerically (Exposé~\ref{sec:analogy}, Appendix~\ref{app:repro}).

\paragraph{III. ``The LoRA+ ratio, $\alpha$, the init scale, rsLoRA''}

\begin{example*}{entry 23, Exposé~\ref{sec:fibres}}
\labelsf{PEFT}\enspace the LoRA+ ratio, $\alpha$, the initial scale, rsLoRA.\par
\noindent\labelsf{Category theory\,\textperiodcentered\,geometry}\enspace coordinates on the orbit space of a
hyperparameter torus.
\end{example*}

Multiplying $B$ by $c>0$ and dividing the scale $s$ by $c$ leaves the weights unchanged, and Adam with $\varepsilon=0$
ignores a positive rescaling of a block's gradient history. Rescaling the block's learning rate too then reproduces the
run step for step, and the per-block factors form a torus acting on the hyperparameters.

\thmref{Theorem}{III.10} makes this exact for any multihomogeneous method, assuming no parameter-dependent regulariser,
no global clipping, the same data order and dropout masks, and exact arithmetic. For LoRA
(\thmref{Corollary}{III.11}(b)), under Adam with $\varepsilon=0$, no weight decay or clipping and one schedule for both
groups, LoRA+ with ratio $\lambda$ at scale $s$ produces exactly the trajectory of plain LoRA at scale $\lambda s$, with
the same $\eta_A$ and $A_0$. For $\varepsilon>0$ this needs $\varepsilon_B'=\lambda\varepsilon$, and with AdamW a decay
$\lambda\cdot\mathrm{wd}$ on $B$. Under SGD the scale is $\sqrt\lambda\,s$.

\begin{explains}
Only the torus orbit of $(s,\eta_A,\eta_B,\sigma_A)$ matters. Under Adam it is fixed by $s\eta_A\eta_B$ and
$\sigma_A/\eta_A$ (\thmref{Corollary}{III.11}(a)), so at a fixed rank LoRA+, rsLoRA and $\alpha$-tuning are one family. A
gain of LoRA+ over LoRA with $\alpha$ multiplied by $\lambda$ needs something that breaks the torus, such as global-norm
clipping (a Hugging Face Trainer default), an $\varepsilon$ or decay shared across groups, or different schedules
(\thmref{Prediction}{X.1}). Figure~\ref{fig:orbit} runs both trainings.
\end{explains}

\paragraph{Credit}
The LoRA and Adam case, with its $\varepsilon>0$ extension and the two effective hyperparameters, is due to
\citet{schulman2025lora}, for the method of \citet{hayou2024lora}. The torus for any multihomogeneous method and the
$c^2$ law for SGD are added here.

\paragraph{The same fibres under gradient flow}
Entry 22 reads the same fibres under gradient flow. With block-scalar rates the gauge yields the conserved charge
\begin{equation}\label{eq:rosetta-charge}
  \Phi=\frac{B\T B}{\eta_B}-\frac{AA\T}{\eta_A}
\end{equation}
(\thmref{Theorem}{III.5}), after \citet{du2018algorithmic}, \citet{arora2018optimization}, \citet{zhao2022symmetries}
and \citet{marcotte2023abide}. Balanced points $B\T B=AA\T$ form the Kempf--Ness slice (\thmref{Proposition}{III.7};
\citealp{xkempf1979length,xrichardson1990minimum}).

\paragraph{IV. ``GaLore versus LoRA''}

\begin{example*}{entry 25, Exposé~\ref{sec:lens}}
\labelsf{PEFT}\enspace GaLore versus LoRA.\par
\noindent\labelsf{Category theory\,\textperiodcentered\,geometry}\enspace cotangent-side and tangent-side lens
2-cells, which coincide for linear charts (the degree-one collapse).
\end{example*}

Backpropagation makes each map a lens, running forward from parameters to outputs and backward from output gradients to
parameter gradients. LoRA restricts the forward side. GaLore compresses the gradient before the optimizer and expands
its output after.

Let $\rho_k(C)=W_k+\mathcal LC$ with $\mathcal L$ linear and $C_0=0$. Let the optimizer's state and output depend only
on the gradients fed to it, with one implementation on both sides (SGD, heavy-ball or Nesterov momentum, Adam without
weight decay, Adafactor without parameter scaling). Then training $C$ and merging at the end of a period gives the same
$W$-trajectory as the backward method with compression $\mathcal L^*$ and anchor $\mathcal L$, started at $W_k$ from the
same optimizer state (\thmref{Proposition}{IV.2}). Decoupled weight decay, LAMB and LARS trust ratios, parameter-scaled
Adafactor and full-gradient clipping break the identity.

\begin{explains}
Under these hypotheses, GaLore with a projector $P_k$ fixed on each period is one-sided LoRA $W_k+P_kC$, merged and
restarted every period with $C$'s optimizer state carried over unchanged. It adds two things to one-sided LoRA, a frame
from the gradient's SVD and that carry-over, which reads moments written for $P_{k-1}$ as if written for $P_k$. LDAdam
\citep{xrobert2025ldadam} transports them instead.
\end{explains}

\paragraph{Credit}
The identity for GaLore \citep{zhao2024galore} is due to \citet{xtorrobahennigen2025duality}, and its SGD case with a
frozen random factor to \citet{hao2024flora}.

\paragraph{The rest of the cotangent family}
Under the same hypotheses the identity (\thmref{Proposition}{IV.2}, \thmref{Proposition}{IV.3}) covers LoRA-FA and
GaLore with random projectors, MeZO up to its finite-difference estimate, GaLore with gradient SVDs, and BitFit and LISA
(without weight decay) with coordinate blocks. Flora as published, APOLLO and Fira update outside their projected
subspace and are not covered.

\paragraph{V. ``QLoRA and LoftQ''}

\begin{example*}{entry 28, Exposé~\ref{sec:base}}
\labelsf{PEFT}\enspace QLoRA, LoftQ.\par
\noindent\labelsf{Category theory\,\textperiodcentered\,geometry}\enspace base change along a map that does not
preserve the function; a pointing defect; an approximate section.
\end{example*}

A quantiser $\kappa$ moves the pretrained point to $\kappa\theta_0$ and changes the function. Additive methods commute
with translation, $M(\theta+v)=\tau_v\circ M(\theta)$, so QLoRA \citep{dettmers2023qlora} is LoRA attached at
$\kappa\theta_0$, with pointing defect $e=\kappa\theta_0-\theta_0$ (\thmref{Definition}{V.1}), and $\theta_0$ lies
outside its image unless $\rk e\le r$ (\thmref{Proposition}{V.2}(a)). Splitting initialisations come in three orders
with three different defects (\thmref{Proposition}{V.2}(b); Table~\ref{tab:rosetta-orders}). Write $W$ for the
full-precision weight, $\delta_0$ for the initial adapter, $W_{\mathrm{res}}=W-\delta_0$ and $s=\alpha/r$.

\begin{table}[t]
\centering\small
\caption{Three orders of quantising and splitting, and the pointing defect $\rho(q_0)-W$ each leaves
(\thmref{Proposition}{V.2}(b)).}\label{tab:rosetta-orders}
\renewcommand{\arraystretch}{1.15}
\begin{tabularx}{\linewidth}{@{}>{\raggedright\arraybackslash}p{3.6cm}>{\raggedright\arraybackslash}p{4.4cm}>{\raggedright\arraybackslash}X@{}}
  \toprule
  \headrow \textbf{Order} & \textbf{Used by} & \textbf{Pointing defect}\\
  \midrule
  split, then quantise the residual & QPiSSA; HF's PiSSA and CorDA workflow &
    $\kappa(W_{\mathrm{res}})-W_{\mathrm{res}}$\\
  quantise, then fit the error & LoftQ with one iteration (HF default \texttt{loftq\_iter=1}) &
    $e-s\,\mathrm{SVD}_r(e)$, as HF does not divide the fit by $s$; at $s=1$, $e$ minus its best rank-$r$
    approximation\\
  quantise, split, re-quantise & HF \texttt{olora\_init} on bitsandbytes weights & $e+\kappa(r')-r'$, with
    $r'=\kappa W-\delta_0$\\
  \bottomrule
\end{tabularx}
\end{table}

\begin{explains}
Measured from its own start, QLoRA obeys the rank bound of any LoRA (\thmref{Theorem}{II.1}(c)); measured from
$\theta_0$, its update also contains the generically full-rank defect. Since $e-\mathrm{SVD}_r(e)$ is orthogonal to
$\mathrm{SVD}_r(e)$, one-step LoftQ in HF beats QLoRA in Frobenius norm only for $0<s<2$. Further LoftQ iterations need
not decrease their objective, and at $s=1$ the output is an approximate section, pointed up to its attained residual
(\thmref{Proposition}{V.2}(c)).
\end{explains}

\paragraph{Credit}
QLoRA is due to \citet{dettmers2023qlora}, LoftQ to \citet{li2023loftq}, and PiSSA and QPiSSA to
\citet{meng2024pissa}.

\paragraph{VI. ``Merging an adapter into the weights''}

\begin{example*}{entry 5, Exposé~\ref{sec:merge}}
\labelsf{PEFT}\enspace merging an adapter into the weights; zero inference latency.\par
\noindent\labelsf{Category theory\,\textperiodcentered\,geometry}\enspace for weight-space methods, the canonical arrow to
the terminal object; for extensions, a lift along the realisation $\Phi_f$, certified box by box by rewriting.
\end{example*}

Reading I settled merging for weight-space methods on a full-precision base. Adapters, prefixes and (IA)$^3$ scalings
insert new boxes and live on an extension of $f$, so merging one is a lifting problem along the realisation
$\theta\mapsto f(\theta,-)$.

A method $\rho$ on an extension $\tilde f$ is mergeable if some $\mu:Q\to\Theta$ has
$\Phi_{\tilde f}\circ\rho=\Phi_f\circ\mu$ (\thmref{Definition}{VI.1}). \thmref{Theorem}{VI.3} gives sufficient
conditions box by box, through seven rewrite rules with side conditions such as position-uniform edits, unshared
absorbing weights and bias slots for shifts. For instance, HF applies the feed-forward vector $\ell$ of (IA)$^3$
\citep{liu2022few} to the input of $W_{\mathrm{down}}$, where it always fuses. It fuses into $W_{\mathrm{up}}$ instead
under SwiGLU or GeGLU for every $\ell$, and under ReLU when $\ell\ge0$. Under GELU, an entry $\ell_i$ on a nonzero row
of $W_{\mathrm{up}}$ fuses there only if it is $0$ or $1$.

\begin{explains}
``Zero inference latency'' is certified by local rewriting, a sufficient test that also names the absorbing weight; the
$W_{\mathrm{up}}$ route is idle in practice, as $W_{\mathrm{down}}$ always works. Nonlinear adapters, input-routed
mixtures and prefixes do not merge box-locally. At its layer a prefix is a gated parallel adapter
(\thmref{Theorem}{VI.5}(b)), and it is not box-locally mergeable for generic frozen weights and prefixes
(\thmref{Theorem}{VI.5}(f)); whole-network merging of prefixes is open.
\end{explains}

\paragraph{Credit}
The (IA)$^3$ merge is noted by \citet{liu2022few}, and the gated-adapter reading of prefix tuning is due to
\citet{he2021unified}. Scale migration through SwiGLU, ReLU and norm layers, with the GELU obstruction, already appears
in SmoothQuant \citep{xxiao2023smoothquant} and AWQ \citep{xlin2024awq}. The activation symmetries behind slides R2 and
R3 (\thmref{Lemma}{VI.2}) are classical \citep{xsussmann1992uniqueness,xchen1993geometry,xgodfrey2022symmetries}.

\subsubsection{Reading from the right}\label{sec:rosetta-reverse}

Some mathematical constructions are methods already in use.
\begin{itemize}
  \item The product $\mathrm{FA}(A_0)\times\mathrm{FB}(B_0)$ of the two one-sided LoRAs is the sandwich
    $R\mapsto W_0+B_0RA_0$. In the SVD frames of $W_0$, with $\sigma_r>0$, it is LoRA-XS pointed at $R=0$
    (\thmref{Proposition}{I.8}(a); \citealp{baazy2024lora}).
  \item For injective $W_0$ and even $r$ with $2\le r\le n-2$, LoRA's image meets the rotation orbit $W_0\cdot\SO(n)$
    exactly in the image of HRA without Gram--Schmidt, the HF default (\thmref{Theorem}{II.7};
    \citealp{yuan2024bridging}). This is a meet of sets, and HRA is not the categorical product.
  \item Transport along a fixed $T\in\GL(\Theta)$ is an automorphism of the slice. It carries coordinate masks to
    FourierFT \citep{gao2024parameter} (with no conjugate frequency pair sampled), WaveFT \citep{bilican2025exploring}
    (when its padded inverse wavelet transform is invertible) and C3A \citep{chen2024parameter}, each started at zero,
    and to LoRA-XS and SVFT \citep{lingam2024svft} in the SVD frame of $W_0$, which depends on $\theta_0$. The inverse
    of the rearrangement of \citet{loan1993approximation} carries $\mathrm{LoRA}_k$ to unconstrained sums of $k$
    Kronecker products (\thmref{Proposition}{II.13}).
  \item The monoidal sum $\oplus$ adds updates, so its image is a Minkowski sum. RoSA \citep{nikdan2024rosa}, GraLoRA
    \citep{jung2025gralora} and the stages of ReLoRA \citep{lialin2023relora}, unrolled in time, are words in $\oplus$
    (\thmref{Observation}{I.7}).
\end{itemize}

\begin{explains}
Products and meets can be tried on other pairs too. Meets of images always exist as sets of weights, while products can
fail to exist (\thmref{Observation}{I.4}(b)). Which meets admit a polynomial chart, as LoRA-XS and HRA's image do, is an
open problem (\S\ref{sec:analogy-open}, problem~4).
\end{explains}

\subsubsection{The full dictionary}\label{sec:rosetta-table}

Table~\ref{tab:rosetta} gives every entry. Result numbers in the glosses point to the statements whose hypotheses the
entry inherits; entries without a number are definitions or bookkeeping.

\begingroup
\footnotesize
\renewcommand{\arraystretch}{1.18}
\setlength{\LTpre}{6pt}\setlength{\LTpost}{6pt}
\begin{longtable}{@{}>{\raggedleft\arraybackslash}p{0.45cm}>{\raggedright\arraybackslash}p{2.75cm}>{\raggedright\arraybackslash}p{4.6cm}>{\raggedright\arraybackslash}p{6.15cm}@{}}
\caption{Rosetta stone: the full dictionary from PEFT phrases to their categorical or geometric counterparts. The gloss
names the results behind each pairing; their hypotheses are part of the entry.}\label{tab:rosetta}\\
\toprule
\headrow \textbf{\#} & \textbf{PEFT} & \textbf{Categorical / geometric} & \textbf{Gloss}\\
\midrule
\endfirsthead
\multicolumn{4}{@{}l}{\footnotesize\itshape Table~\ref{tab:rosetta}, continued}\\
\toprule
\headrow \textbf{\#} & \textbf{PEFT} & \textbf{Categorical / geometric} & \textbf{Gloss}\\
\midrule
\endhead
\midrule
\multicolumn{4}{@{}r}{\footnotesize\itshape continued on the next page}\\
\endfoot
\bottomrule
\endlastfoot
1 & pretrained checkpoint & base point $\theta_0:1\to\Theta$; a pointed 1-cell $(\Theta,\theta_0,f)$ of
  $\Para(\Diff)$ & Fine-tuning moves a point; everything is measured relative to it (\thmref{Def.}{I.1}).\\
2 & PEFT method & object of the pointed slice $\Meth(\Theta,\theta_0)=\Diff_*/(\Theta,\theta_0)$, equivalently the
  opposite of the pointed coslice of reparametrisation 2-cells under $F$ & A method is a pointed map
  $\rho:(Q,q_0)\to(\Theta,\theta_0)$. The slice depends only on $(\Theta,\theta_0)$; $\Para$ is bookkeeping for
  weight-space methods (\thmref{Rem.}{I.3}).\\
3 & ``method $N$ can do whatever $M$ can'' & morphism (simulation) $h:M\to N$ with $\rho_N\circ h=\rho_M$ & Smooth
  simulations also carry parameter paths; at the level of sets they reduce to image inclusion (\thmref{Obs.}{I.4}).\\
4 & frozen model / full fine-tuning & initial / terminal object & Every weight-space method lies in the interval
  [frozen, full FT]; adapters, prefixes and ReFT are extensions, not objects of $\Meth$ (\thmref{Obs.}{I.4}).\\
5 & merging an adapter into the weights & the canonical morphism to the terminal object (it is $\rho$ itself); for
  extensions, a lift along the realisation $\Phi_f$ & For weight-space methods merging is definitional; for extensions
  zero inference overhead is a lifting problem, certified box-locally by rewriting (sufficient conditions and box-local
  obstructions) under side conditions such as position-uniformity, no weight sharing and bias slots
  (\thmref{Thm}{VI.3}).\\
6 & expressivity / effective number of parameters & image germ of $\rho$ at $\theta_0$; its dimension $d(M)$, the
  generic rank of $d\rho$ & Rank is a cut invariant; dimension is a volume invariant (\thmref{Thm}{II.10},
  \thmref{Prop.}{II.12}).\\
7 & lazy / NTK-regime capability & first-order image $S_1=d\rho_{q_0}(T_{q_0}Q)$, a functorial isomorphism invariant &
  The lazy kernel has range $J_f(S_1)$; $S_1$ separates zero-init pointings ($\mathrm T_A$ vs $\mathrm T_B$, the row
  space of $A_0$) that the image cannot, while the image already separates split inits (\thmref{Obs.}{I.6}).\\
8 & zero-init vs PiSSA-type init & apex (singular vertex) vs smooth point of the image; Type~I (new $q_0$ in the apex
  fibre) vs Type~II (new $\rho$ by splitting) & Zero-$B$ init loses $r(n-r)$ first-order directions (zero-$A$ init
  $r(m-r)$); split inits are different objects, incomparable with LoRA (\thmref{Thm}{II.2}, \thmref{Prop.}{III.8}).\\
9 & the zoo of LoRA initialisations & points of three distinguished strata of full-rank pointings, classified as a set
  by $\Gr(r,n)\sqcup\Gr(r,m)\sqcup\MM_r$ & Vanilla LoRA on linear layers, EVA, Init[B], HF's default LoRA on embedding
  layers, PiSSA, MiLoRA, OLoRA, CorDA, LoRA-GA and Astra are points; LoRA-FA and MiCA are frozen-factor sub-objects of
  the zero-$B$ and zero-$A$ strata; HF's \texttt{orthogonal} init lies outside the three strata, and LoftQ/QLoRA are
  defect pointings. The optimizer sees a finer quotient (\thmref{Thm}{II.2}, \thmref{Cor.}{III.13}).\\
10 & no init buys rank above $2r$ & union over pointings, $\bigcup_q\Im M_q=\theta_0+(C-C)=\theta_0+\MM_{\le\min(2r,m,n)}$
  & An exact initialisation picks a translate of the shape; from any starting point the update has rank at most $2r$.
  This is the known rank-$2r$ PiSSA-to-LoRA conversion (\thmref{Thm}{II.1}).\\
11 & redundant parameters & gauge group, the automorphisms of $Q$ over $\Theta$; generic fibre & LoRA wastes $r^2$
  parameters; LoHa wastes at least $r_1^2+r_2^2+m+n-1$ below the $mn$ cap (exactly that many, numerically), and more
  where the cap binds (\thmref{Prop.}{II.12}).\\
12 & high-rank PEFT (LoHa, LoKr, HiRA, MoRA, C3A) & tensor networks whose min-cut avoids the bond; transport of
  structure along $\GL(\Theta)$ & High rank is not $\GL(\Theta)$-invariant (though rank is $\GL_m\times\GL_n$-invariant)
  and need not mean large volume; a large min-cut is necessary for high rank and is not sufficient
  (\thmref{Thms}{II.10}--\hyperref[res:II.11]{II.11}, \thmref{Prop.}{II.13}).\\
13 & Kronecker adapters (unconstrained Kronecker sums, KronA, LoKr) & LoRA across a different cut of the same 4-leg
  tensor (Van Loan--Pitsianis rearrangement, a coordinate permutation) & For aligned splits (equal square splits, and
  every split HF PEFT's factorisation produces), a generic rank-one update has maximal Kronecker rank, so unconstrained
  Kronecker sums with fewer terms and zero-init $\mathrm{LoRA}_r$ below the rank of a full-rank Kronecker product are
  incomparable. LoKr with inner rank $r'\le\min(m_2,n_2)$ sits inside zero-init $\mathrm{LoRA}_r$ exactly when
  $r\ge\min(m_1,n_1)\,r'$; Compacter's PHM weights live in a nonlinear adapter (\thmref{Thm}{II.11}).\\
14 & ``new method = old method'' & transport $T_*$ along $T\in\GL(\Theta)$: an automorphism of the category, not an
  arrow inside it & FourierFT (when no conjugate frequency pair is sampled), WaveFT, C3A, LoRA-XS, SVFT and SHiRA are
  coordinate masks in transported frames, and HiRA (for $W_0$ without zero entries) is LoRA in a transported frame; for
  LoRA-XS, SVFT and HiRA the frame depends on $\theta_0$ (\thmref{Prop.}{II.13}).\\
15 & orthogonal fine-tuning preserves knowledge & the image lies in a fibre of the complete invariant $W\mapsto WW\T$
  (first fundamental theorem for $\Ort(n)$) & In exact arithmetic, exactly orthogonal input-side methods (OFT with an
  exact Cayley map, strict GOFT without its optional scaler, HRA) can never change the spectrum or neuron angles,
  however often they are merged; HF's default Cayley--Neumann OFT is only approximately orthogonal
  (\thmref{Cor.}{II.6}).\\
16 & HRA bridges LoRA and OFT & image-level meet with the rotation orbit, $(W_0+\Mr)\cap W_0\,\SO(n)$, for injective
  $W_0$ and even $r\le n-2$; not a product of methods & LoRA meets exact-Cayley OFT in a strictly smaller set, and HF's
  default Cayley--Neumann OFT, which is not orthogonal, is incomparable with HRA. At HRA's paired identity init the
  image germ is the skew-determinantal cone, so that init is an apex (\thmref{Thm}{II.7}).\\
17 & LoRA-XS & categorical product $\mathrm{FA}(A_0)\times\mathrm{FB}(B_0)$ of the two one-sided LoRAs, in the SVD
  frames of $W_0$ (with $\rk W_0\ge r$) & The overlap that a split init double-counts (\thmref{Prop.}{I.8},
  \thmref{Prop.}{II.3}).\\
18 & DoRA's magnitude/direction decoupling & diagonal saturation $\mathrm{Diag}_m\cdot(W_0+\Mr)$, the image of LoRA
  followed by a trainable output scale & Equal images, for $W_0$ without zero rows. Under gradient flow the exact
  gradient changes the metric and the detached backward is a non-gradient update; under AdamW the magnitude
  coordinates matter too (\thmref{Prop.}{II.8}).\\
19 & (IA)$^3$, SSF, LN tuning, RoAd, HiRA & the scalings and HiRA are rungs of the Hadamard-torus ladder
  $W_0\odot(J+Z)$; $\mathrm{RoAd}_1$ is the complex torus $(\mathbb C^\times)^{m/2}$, and $\mathrm{RoAd}_2$ and
  $\mathrm{RoAd}_4$ apply arbitrary $2\times2$ blocks (group part block-$\GL_2$) & On every rung zeros of $W_0$ stay
  zero. RoAd can fill them, so it is a rung only for zero-free $W_0$, and for generic $W_0$ it lies on no low-rank rung.
  Covariance $\mathrm{Mon}_m\times\GL_n$ on output-scaled boxes, $\GL_m\times\mathrm{Mon}_n$ on \texttt{down\_proj},
  exactly $\mathrm{Mon}_m\times\mathrm{Mon}_n$ for HiRA, never $\GL_m\times\GL_n$ (\thmref{Prop.}{II.9}).\\
20 & ``the geometry a method lives in'' & covariance group $\mathcal G_M$ (Klein): the largest group of pairs $(g,h)$
  with $\Im M(gW_0h^{-1})=g\,\Im M(W_0)\,h^{-1}$ & Descent depends on whether each induced pair $(g,h)$ lies in
  $\mathcal G_M$, a question of side for one-sided methods and of group for coordinate-structured ones. LoRA and
  (IA)$^3$ (HF defaults) provably keep their reachable function sets under a global residual rotation of a gain-fused
  pre-norm model; block OFT, LoHa ($r\ge2$), HiRA and masks fail the sufficient condition (\thmref{Obs.}{II.15},
  \thmref{Conj.}{A}).\\
21 & how a method trains & induced cometric $K=D\rho\,\Lambda\,D\rho\T$ (pulled-back gradient of the lens) & Two
  isomorphic objects can train differently; gradient descent is natural exactly where cometrics agree, which for LoRA's
  linear gauge means $\Ort(r)$ (\thmref{Thm}{III.3}).\\
22 & balancedness $B\T B=AA\T$ & Noether charge of the non-compact half $\Sym_r$ of $\gl_r$; the Kempf--Ness slice of
  the gauge action & Conserved under gradient flow with constant rates for any cotangent signal of the form
  $\Lambda\,D\rho\T\gamma$ (DoRA's detached norm, clipping), after \citet{zhao2022symmetries} (\thmref{Thm}{III.5},
  \thmref{Prop.}{III.7}).\\
23 & LoRA+ ratio, $\alpha$, init scale, rsLoRA & coordinates on the orbit space of a hyperparameter torus & Under Adam
  with $\varepsilon=0$, no weight decay, no clipping and a shared schedule, LoRA+ with ratio $\lambda$ is exactly LoRA
  with $\alpha$ times $\lambda$ (the LoRA case is \citealp{schulman2025lora}); at a fixed rank, rsLoRA is isomorphic to
  LoRA (\thmref{Thm}{III.10}, \thmref{Cor.}{III.11}).\\
24 & transformation-invariant optimizer (Riemannian preconditioning, ScaledGD, LoRA-Pro, LoRA-RITE) & update rule
  equivariant under the whole gauge $\GL_r$, so it descends to the quotient $\MM_r$ & On the full-rank locus, undamped
  scaled GD and LoRA-RITE are $\GL_r$-equivariant in the parameters; damped variants and LoRA-Pro's step are only
  $\Ort(r)$-equivariant, and Adam only $B_r$-equivariant (\thmref{Prop.}{III.4}, \thmref{Thm}{III.12}).\\
25 & GaLore vs LoRA & cotangent-side vs tangent-side lens 2-cells, which coincide for linear charts (the degree-one
  collapse) & Without weight decay or full-gradient clipping, GaLore is one-sided LoRA with a moving frame, with
  optimizer state carried across re-basings rather than transported (\thmref{Prop.}{IV.2}, credited to
  \citealp{xtorrobahennigen2025duality}).\\
26 & when a gradient-projection method is really a reparametrisation & Frobenius integrability of the anchor
  distribution & GaLore, LISA and MeZO keep a $\theta$-independent anchor within each period, hence affine leaves, and
  under the hypotheses of \thmref{Prop.}{IV.2} are rebasing schemes of linear forward methods; Frobenius integrability
  matters only for $\theta$-dependent anchors such as full-batch GaLore with $T=1$ (\thmref{Prop.}{IV.3}).\\
27 & activation memory & backward reach of the composite lens (the backprop cone) & An upper bound: LST avoids the
  backbone's backward pass; LoRA-FA stores $A_0x$; LISA's cone is the whole network because its embedding is always
  trained, though its frozen layers need not store their linear-input residuals; MeZO has no reverse pass
  (\thmref{Obs.}{IV.4}).\\
28 & QLoRA / LoftQ & base change along a map that does not preserve the function; pointing defect; approximate section &
  Split-then-quantise, quantise-then-fit (LoftQ's default, whose HF init ignores the LoRA scale) and
  quantise-split-requantise leave three different defects (\thmref{Prop.}{V.2}).\\
29 & ReLoRA, re-OFT, iterated BOFT/HRA & the monoid (or Lie subgroup) generated by the reachable displacements & ReLoRA
  reaches exactly $W_0+\MM_{\le\min(Kr,m,n)}$; exact-Cayley block OFT is stuck; iterated BOFT reaches the orbit
  $W_0\,\SO(n)$, up to output scaling, only at full butterfly depth; re-HRA reaches $(W_0+\MM_{\le Kr})\cap W_0\,\SO(n)$
  (\thmref{Thm}{V.3}).\\
30 & task arithmetic vs TIES/DARE & $\GL(\Theta)$-natural vs coordinatewise operations (TIES with sign election by
  total mass: homotheties times signed permutations; DARE: the monomial group, its permutations in distribution) &
  Coordinate merges depend on the basis; factor-space merges depend on the gauge (\thmref{Prop.}{V.4}).\\
31 & LoraHub / Poly factor mixing & a rule on factors that is not invariant under $\prod\GL_r$, hence not a function of
  the adapters & Post-hoc LoraHub (and HF's \texttt{linear} combination) needs a joint gauge fixing that produces a
  shared factor (KnOTS) or a shared frozen factor (FFA-LoRA); Poly/MHR are trained jointly and, for fixed generic
  routing with fewer skills than tasks, have only the diagonal $\GL_r$ symmetry, which is checked numerically
  (\thmref{Prop.}{V.4}).\\
32 & adapters (Houlsby, parallel), LLaMA-Adapter, LoReFT & pointed extensions: function-preserving embeddings with a
  neutral point & LoReFT is LoRA on an inserted affine identity box (bias confined to $\operatorname{col}B$), applied at
  selected positions; pyreft's default init is not neutral (\thmref{Prop.}{VI.4}).\\
33 & prompt / prefix tuning & unpointed extensions on the sequence wire & Prefix = gated parallel adapter (attention =
  monoid homomorphism followed by a perspective map); content-attention ratios are rigid only within the layer where the
  prefix acts; prompt tuning maps into deep prefix tuning in causal decoders with matching position offsets
  (\thmref{Thm}{VI.5}).\\
34 & privileged neuron basis & automorphism group of the elementwise activation (the $\sigma$-spider) & For ungated
  FFNs: permutations for GELU/SiLU, signed permutations for tanh, positive monomials for ReLU; gated FFNs (SwiGLU,
  GeGLU) have the full monomial group (\thmref{Lemma}{VI.2}).\\
35 & RoSA, GraLoRA, HF \texttt{cat} merges, ReLoRA stages & words in the convolution monoidal product $\oplus$ of
  additive methods (Minkowski sum of images) & $\oplus$ is not the categorical product, whose image is the
  intersection when it exists. LoRA-soup CAT is a $\oplus$ of frozen line methods (a span), and UniPELT/MAM are not
  $\oplus$-words (\thmref{Obs.}{I.7}).\\
36 & \texttt{target\_modules} / applying a method to every layer & assigning a 2-cell to each generator box and
  composing horizontally; $T$-points for routing, tasks and tenants & Non-local methods (VB-LoRA, Compacter, FacT,
  Uni-LoRA) share parameters across boxes and are not generator-wise.\\
\end{longtable}
\endgroup

\section{Falsifiable predictions}\label{sec:predictions}

\begin{flushright}
\small\itshape Say what you proved, say what you only named,\\
and say what would show you wrong.\\[2pt]
{\upshape\footnotesize a working maxim}
\end{flushright}

This exposé states eleven experiments that could show the framework wrong, each for a named optimizer.

\paragraph{In brief}
\begin{itemize}
  \item A fair LoRA+ baseline multiplies $\alpha$ by the ratio $\lambda$. Under Adam, with $\varepsilon$, decay and
    per-block clipping thresholds rescaled and one shared schedule, the two runs give identical weights (after
    \citealp{schulman2025lora}).
  \item LoRA's charge $\Phi=B\T B/\eta_B-AA\T/\eta_A$ is constant under gradient flow and, when a step size $\eta$
    multiplies fixed block rates, moves by $O(\eta^2)$ per SGD step, a cheap diagnostic of what the optimizer adds.
  \item Some claims need no training run. Generic rank-one targets separate LoRA from Kronecker adapters, exact block
    OFT re-merged on a fixed partition never leaves $W_0\cdot\prod\SO(b)$, and LoraHub's merge depends on gauge choices
    unless the modules share a factor.
\end{itemize}

\newenvironment{predmeta}{\par\small\begin{description}[font=\normalfont,leftmargin=0pt,labelindent=0pt,
  itemsep=2pt,topsep=5pt,parsep=0pt,labelsep=0.6em]}{\end{description}}
\newcommand{\predfield}[1]{\item[\labelsf{#1}]}

\subsubsection{Why the theory should risk something}\label{sec:predictions-why}

A framework that only re-describes known methods cannot be wrong, and so it earns little trust. The claims below say what
a training run will do. Each prediction carries the optimizer scope of the result it rests on; a claim about images
needs none, because no run can leave the image. Table~\ref{tab:predictions} lists them. Each keeps the number it carries on
the companion site, and the subsections group them by subject (optimizers, images, symmetry, base change and merging),
so the numbers do not run in order.

\begin{table}[t]
\centering\small
\caption{The eleven predictions, the optimizer scope each inherits, the results it rests on, and its kind.}
\label{tab:predictions}
\renewcommand{\arraystretch}{1.15}
\begin{tabularx}{\linewidth}{@{}l>{\raggedright\arraybackslash}X>{\raggedright\arraybackslash}p{3.7cm}>{\raggedright\arraybackslash}p{3.0cm}>{\raggedright\arraybackslash}p{1.9cm}@{}}
  \toprule
  \headrow \textbf{No.} & \textbf{Prediction} & \textbf{Optimizer scope} & \textbf{Rests on} & \textbf{Kind}\\
  \midrule
  \hyperref[res:X.1]{X.1} & LoRA+ against $\alpha$ re-tuned by $\lambda$ & Adam, AdamW & \thmref{Thm}{III.10},
    \thmref{Cor.}{III.11} & theorem\\
  \hyperref[res:X.2]{X.2} & DoRA against LoRA + output gain & AdamW, shared by all arms & \thmref{Prop.}{II.8},
    \thmref{Conj.}{D} & conjecture\\
  \hyperref[res:X.3]{X.3} & Hidden-unit permutations & coordinatewise (SGD, Adam) & \thmref{Lemma}{VI.2},
    \thmref{Obs.}{II.15} & theorem + data\\
  \hyperref[res:X.4]{X.4} & Residual-stream rotations & reach: any; outcomes: depend on optimizer and init &
    \thmref{Obs.}{II.15},
    \thmref{Conj.}{A} & conjecture\\
  \hyperref[res:X.5]{X.5} & Rank-one targets & any (image claim) & \thmref{Thm}{II.11} & theorem\\
  \hyperref[res:X.6]{X.6} & Re-based OFT & any (reach); exact Cayley map & \thmref{Thm}{V.3} & theorem\\
  \hyperref[res:X.7]{X.7} & The crossover scale $\sigma^*$ & gradient flow, SGD & \thmref{Thm}{III.9} & theorem + data\\
  \hyperref[res:X.8]{X.8} & Drift of the charge $\Phi$ & gradient flow, SGD, Adam & \thmref{Thm}{III.5} & theorem +
    data\\
  \hyperref[res:X.9]{X.9} & LoraHub with shared or independent seeds & post hoc; regime set by training &
    \thmref{Prop.}{V.4} & theorem + data\\
  \hyperref[res:X.10]{X.10} & Rankings at equal effective dimension & any (bookkeeping) & \thmref{Prop.}{II.12} &
    theorem + data\\
  \hyperref[res:X.11]{X.11} & Whitened data-aware splits & split: any; training is not covariant &
    \thmref{Prop.}{II.16}, \thmref{Conj.}{C} & conjecture\\
  \bottomrule
\end{tabularx}
\end{table}

Read the kind column as a warning label. \emph{Theorem} means the sharp half cannot fail inside its hypotheses.
\emph{Theorem + data} pairs a proved mechanism with an empirical claim about its size. \emph{Conjecture} means we
expect it and have not proved it. The toys check mechanisms in float64 on small matrices, and the experiments on real
models are still to be run. Each prediction below lists its scope, the results that support it, a test, the outcome
that would refute it, and the toy or figure that shows its mechanism.

\subsubsection{Predictions about optimizers}\label{sec:predictions-optimizers}

LoRA has four knobs: the scale $s=\alpha/r$, the rates $\eta_A,\eta_B$ and the initial size $\sigma_A$ of $A$. A
two-dimensional torus of rescalings moves all four without changing any weight. LoRA+ turns one knob, the ratio
$\lambda=\eta_B/\eta_A$, and under Adam that turn lands on the orbit of plain LoRA with a larger $\alpha$.

\begin{prediction}{X.1}{LoRA+ against a re-tuned $\alpha$}
Any gain of LoRA+ with ratio $\lambda$ over plain LoRA with $\alpha$ multiplied by $\lambda$ (same $\eta_A$, same $A_0$)
comes from the ingredients that break the extended torus: global-norm clipping, an $\varepsilon$ or weight decay shared
by the two parameter groups, and schedules that differ between the groups. With all of them off, or rescaled per block,
the two runs produce the same weights at every step in exact arithmetic.
\begin{predmeta}
  \predfield{Scope} Adam and AdamW ($\beta_1\in[0,1)$, $\beta_2\in(0,1)$, bias correction allowed); linear LoRA modules
    from $B_0=0$; no regulariser on the factors; common data order and dropout masks. A shared schedule, warm-up
    included, preserves the torus. HF PEFT's LoRA+ optimizer treats embeddings and 1-D parameters separately, so only
    linear modules are covered.
  \predfield{Support} \thmref{Theorem}{III.10} and \thmref{Corollary}{III.11}(b), both verified. For $\varepsilon>0$
    the LoRA run needs $\varepsilon_B=\lambda\varepsilon$ on $B$, and with AdamW a decay $\lambda\cdot\mathrm{wd}$ on
    $B$. Under SGD the matching scale is $\sqrt\lambda\,s$.
  \predfield{Test} Train LoRA+ with $\lambda=16$ against LoRA with $\alpha$ multiplied by 16, $\varepsilon$ on $A$ and
    $16\varepsilon$ on $B$, no decay and no clipping; the weights agree bit for bit, since 16 is a power of two. Then
    switch on one torus breaker at a time (shared $\varepsilon$, shared AdamW decay, \texttt{max\_grad\_norm=1.0},
    group-specific schedules) and record the gap each opens.
  \predfield{Would fail if} a gap beyond rounding appeared with every hypothesis met. A standard HF run clips by default
    and its LoRA+ optimizer shares $\varepsilon$ and decay, so there the experiment measures how much of the gain each
    breaker carries.
  \predfield{Toy} Figure~\ref{fig:orbit} runs LoRA+ with ratio $\lambda$ against LoRA with $\alpha\cdot\lambda$ under
    Adam with $\varepsilon=0$. The loss curves coincide and the weight gap is exactly zero, and the gap jumps when a
    shared $\varepsilon>0$, decoupled weight decay or global-norm clipping is switched on.
\end{predmeta}
\end{prediction}

\begin{remark*}{why the test avoids $\varepsilon=0$}
Literal $\varepsilon=0$ needs the convention $0/0=0$. At $B_0=0$ the gradient of $A$ vanishes, and a stock Adam divides
zero by zero there.
\end{remark*}

\paragraph{Credit}
The LoRA and Adam case, its extension to $\varepsilon>0$ and the two effective hyperparameters are due to
\citet{schulman2025lora}, for the method of \citet{hayou2024lora}. The torus for any multihomogeneous method, and the
$c^2$ law for SGD, are added here.

Under gradient flow LoRA conserves the charge $\Phi=B\T B/\eta_B-AA\T/\eta_A$, and when $\Phi$ is isotropic the
dynamics have a closed form. Small singular values of the update grow through $B$ alone, as in LoRA-FA; large ones grow
through both factors in balance.

\begin{prediction}{X.7}{the crossover scale}
Under SGD, the singular values of $\Delta W$ switch from one-sided to two-sided growth as they cross
$\sigma^*=\mu\lvert\kappa\rvert/2$, where $\mu=s\sqrt{\eta_A\eta_B}$ and $\Phi=-\kappa I_r$. For default LoRA with
$B_0=0$ and $A_0A_0\T=cI$, $\kappa=c/\eta_A$ and
\begin{equation}\label{eq:pred-crossover}
  \sigma^*=\frac{sc}{2}\sqrt{\frac{\eta_B}{\eta_A}}.
\end{equation}
LoRA+ with ratio $\lambda$ multiplies $\sigma^*$ by $\sqrt\lambda$.
\begin{predmeta}
  \predfield{Scope} Gradient flow exactly; SGD without momentum up to the $O(\eta^2)$ drift of the charge per step. For
    Kaiming initialisations $A_0A_0\T$ is only close to $cI$ (HF's default has $c\approx1/3$), and the crossover becomes
    a band of relative width $O(\sqrt{r/n})$. Adam is not covered.
  \predfield{Support} \thmref{Theorem}{III.9}, verified; known from the linear-network literature, with the PEFT
    reading ours.
  \predfield{Test} Train with SGD from orthonormal rows of $A_0$ scaled by $\sqrt c$, so that the charge is exactly
    isotropic. Take the SVD $\Delta W=U\Sigma V\T$ at intervals and read off $f_i=u_i\T(BB\T/\eta_B)u_i$ and
    $g_i=v_i\T(A\T A/\eta_A)v_i$. Below $\sigma^*$, $f_i\approx0$ and $g_i\approx\kappa$; above it,
    $f_i\approx g_i\approx\sigma_i/\mu$. Repeat at ratio $\lambda$.
  \predfield{Would fail if} the gains departed from the closed-form values $f_i=f_\kappa(\sigma_i^2/\mu^2)$ and
    $g_i=g_\kappa(\sigma_i^2/\mu^2)$ of \thmref{Theorem}{III.9}, with $f_\kappa(x)=\sqrt{x+\kappa^2/4}-\kappa/2$ and
    $g_\kappa=f_\kappa+\kappa$, while $\Phi$ stayed isotropic. If $\Phi$ has drifted, the test says nothing.
  \predfield{Toy} Figure~\ref{fig:noether}(b) plots the closed-form isotropic-charge singular-value dynamics and the
    crossover $\sigma^*$.
\end{predmeta}
\end{prediction}

\paragraph{Credit}
The closed form is due to \citet{xtarmoun2021understanding}, who call the charge ``$\lambda$-balanced'', and to
\citet{xmin2021explicit}; see also \citet{xdomine2025lazy} and \citet{xkunin2024get}. The balanced regime above
$\sigma^*$ is the dynamics of \citet{arora2018optimization}.

\begin{explains}
LoRA+ moves $\sigma^*$ up by $\sqrt\lambda$, so more of the update spends longer in the LoRA-FA regime. PiSSA at equal
rates has $\kappa=0$, and under gradient flow it is balanced from the first step.
\end{explains}

Part of LoRA's gauge is a symmetry of $\rho$, and gradient flow conserves its charge whatever signal drives it. A
discrete step conserves it up to second order in the step size, and nothing forces Adam's sign-like step to conserve it.

\begin{prediction}{X.8}{drift of the charge}
$\Phi$ is constant under gradient flow, moves by $O(\eta^2)$ per SGD step at fixed block rates $\eta_A,\eta_B$, and
drifts under Adam. We expect its drift under Adam to track other measures of how far Adam breaks LoRA's gauge.
\begin{predmeta}
  \predfield{Scope} Gradient flow driven by any signal $\Lambda D\rho\T\gamma$, which covers minibatches, LoRA dropout,
    global-norm clipping and DoRA's detached norm; SGD without momentum or weight decay. Momentum, Adam and AdamW fall
    outside \thmref{Theorem}{III.5}.
  \predfield{Support} \thmref{Theorem}{III.5} (repaired; known after \citealp{zhao2022symmetries}); weight decay moves
    the charge on purpose (\thmref{Corollary}{III.6}). With a step size $\eta$ multiplying the block rates, a step
    $B\leftarrow B-\eta\eta_B\nabla_B$, $A\leftarrow A-\eta\eta_A\nabla_A$ changes the charge by exactly
    \begin{equation}\label{eq:pred-drift}
      \Delta\Phi=\eta^2\big(\eta_B\,\nabla_B\T\nabla_B-\eta_A\,\nabla_A\nabla_A\T\big).
    \end{equation}
    The identity uses only that both factor gradients come from one weight-space gradient $G$, so it covers
    minibatches, LoRA dropout on the input of $A$, and global clipping when $\nabla_A,\nabla_B$ denote the gradients
    actually applied.
  \predfield{Test} Log $\Phi$ during training. Under SGD at step sizes $\eta$, $\eta/2$, $\eta/4$, the per-step change
    should match \eqref{eq:pred-drift} to rounding and shrink about fourfold each time. Under Adam, compare
    $\lVert\Phi_t-\Phi_0\rVert$ (our suggestion) with the gap between runs started at $A_0$ and $gA_0$, $g\notin B_r$,
    which by \thmref{Corollary}{III.13} can differ.
  \predfield{Would fail if} the SGD drift departed from \eqref{eq:pred-drift}, so that the update was not a plain
    gradient step, or $\Phi$-drift under Adam showed no relation to any other measure of gauge breaking.
  \predfield{Toy} Figure~\ref{fig:noether}(a): scalar LoRA rides its hyperbola $b^2/\eta_B-a^2/\eta_A=\Phi$ under
    gradient flow, and the charge drifts under discrete GD and under Adam.
\end{predmeta}
\end{prediction}

\paragraph{Credit}
The conservation law, with its symmetric and antisymmetric halves, is Prop.~5.1 of \citet{zhao2022symmetries}; laws
independent of loss and data are in \citet{marcotte2023abide}. The principle goes back to \citet{du2018algorithmic},
\citet{arora2018optimization} and \citet{kunin2020neural}.

\subsubsection{Predictions about images}\label{sec:predictions-images}

DoRA splits each output neuron into a length and a direction. At the level of images that adds exactly one free gain per
output neuron to LoRA, so anything else that separates the two happens in training, where Adam notices the coordinates a
parameter is written in.

\begin{prediction}{X.2}{DoRA against LoRA + output gain}
\phantomsection\label{res:D}{\bfseries\scshape Conjecture D}\ (DoRA against a gain in DoRA's
coordinates). Started from the same weights and trained with AdamW at the same settings, LoRA followed by a trainable
output gain matches DoRA \citep{liu2024dora} once the gain is trained in DoRA's coordinates ($\mu_i$ starting at
$\lVert w_{0,i}\rVert$ in place of $\ell_i$ starting at $1$).
\begin{predmeta}
  \predfield{Scope} AdamW, as in the DoRA paper and HF PEFT, with one learning rate, decay and data order for every arm.
    Images refer to the merged weight, with $r\le\min(m,n)$ and $W_0$ free of zero rows. Every arm trains $r(m+n)+m$
    parameters.
  \predfield{Support} \thmref{Proposition}{II.8} (repaired):
    $\Im\mathrm{DoRA}_r=\mathrm{Diag}_m\cdot(W_0+\Mr)$, the image of $\mathrm{LoRA}_r$ followed by an output gain. The
    first form of Conjecture~D, that any gap comes from the parametrisation, followed from this equality and could not
    fail; the form above can.
  \predfield{Test} Start four arms at $W_0$ with the same $A_0$, as in Experiment~D of Exposé~\ref{sec:erlangen}:
    (i) LoRA with an output gain, $\ell_0=\mathbf 1$; (ii) the gain in DoRA's coordinates,
    $W=\diag(\mu_i/\lVert w_{0,i}\rVert)(W_0+sBA)$, $\mu_{0,i}=\lVert w_{0,i}\rVert$; (iii) DoRA with the exact gradient
    of its normalisation; (iv) DoRA with the detached norm of the DoRA paper (Sec.~4.3) and HF PEFT
    (\texttt{weight\_norm.detach()}). Since the images are equal, any gap between arms comes from the parametrisation,
    the backward rule or the optimizer coordinates, and the four arms separate these. This design is ours.
  \predfield{Would fail if} arm (ii) stayed clearly behind arm (iv). Comparing (iii) with (iv) then shows how much of
    the gap the detached backward carries.
  \predfield{Toy} None for training. The companion site's Gram lab shows the image-level half: at $B=0$ DoRA acts as an
    output torus, and its update escapes rank $r$.
\end{predmeta}
\end{prediction}

A rank-one update writes one new feature into many neurons, and LoRA does this with a single rank. A Kronecker sum
measures rank across a different cut of the same four-leg tensor, where, for the splits HF PEFT uses, a generic rank-one
update is as complex as any matrix can be.

\begin{prediction}{X.5}{rank-one targets}
Split $m=m_1m_2$ and $n=n_1n_2$. On targets $W_0+uv\T$ with generic $u,v$, and a loss whose only minimiser is the
target, $\mathrm{LoRA}_1$ can reach zero loss while no sum of fewer than
\begin{equation}\label{eq:pred-kappa}
  \kappa_1=\min(m_1,m_2)\min(n_1,n_2)
\end{equation}
Kronecker products can. For HF PEFT's splits $\kappa_1$ is the largest possible Kronecker rank: $d$ for its square
splits of $d=1024$ or $4096$ (for instance the $64\times64$ split of $4096$), and $576$ for its $24\times32$ split of
$768$. For LoRA ranks below $\min(m_1,n_1)\min(m_2,n_2)$ and fewer than $\kappa_1$ terms, neither image contains the
other.
\begin{predmeta}
  \predfield{Scope} The negative half concerns images and holds for every optimizer; the positive half asks the
    optimizer to find a point that exists.
  \predfield{Support} \thmref{Theorem}{II.11} (repaired), for unconstrained Kronecker sums and for LoKr, since $C\otimes
    BA$ is one Kronecker term. With Schmidt decompositions $u=\sum_i\sigma_ix_i\otimes y_i$ and
    $v=\sum_j\tau_jz_j\otimes w_j$, the best $k$-term sum leaves the squared error $\sum\sigma_i^2\tau_j^2$ over all but
    the $k$ largest products $\sigma_i\tau_j$.
  \predfield{Test} On one linear layer with whitened inputs and more samples than parameters (or with the loss
    $\tfrac12\lVert W-W_0-uv\T\rVert_F^2$), train $\mathrm{LoRA}_1$ and sums of $k=1,\dots,\kappa_1$ unconstrained
    Kronecker products, each with its own learning-rate sweep, and plot the final loss against $k$ beside the floor.
  \predfield{Would fail if} a sum of fewer than $\kappa_1$ terms went below the floor. $\mathrm{LoRA}_1$ stalling above
    zero would be an optimization failure, against which the framework claims nothing.
  \predfield{Toy} Figure~\ref{fig:cut}: for a random rank-one update the operator-Schmidt spectrum has one nonzero value
    across the ordinary cut and $\kappa_1$ across the Kronecker cut.
\end{predmeta}
\end{prediction}

\paragraph{Credit}
The rearrangement and the nearest-Kronecker-product problem are due to \citet{loan1993approximation} (\bgref{B.11}). The
PEFT reading, that Kronecker sums are LoRA across another cut, is ours. Kronecker adapters in practice are KronA
\citep{edalati2022krona} and LoKr \citep{yeh2023navigating}.

Parameter count measures what a method pays, expressive dimension what it can reach. The difference is gauge, parameters
that move without moving the weights. It ranges from one parameter for VeRA through $r^2$ for $\mathrm{LoRA}_r$ to at
least $2r^2+m+n-1$ for $\mathrm{LoHa}_{r,r}$, which has the budget of $\mathrm{LoRA}_{2r}$.

\begin{prediction}{X.10}{effective dimension}
Benchmarks should report the expressive dimension $d(M)$ beside the budget $\lvert M\rvert$. At budgets of a few thousand
parameters, rankings at equal effective dimension can differ from rankings at equal parameter count.
\begin{predmeta}
  \predfield{Scope} Optimizer-free bookkeeping; the change of ranking is the empirical claim.
  \predfield{Support} \thmref{Proposition}{II.12} (repaired). $\mathrm{LoRA}_r$ wastes $r^2$ parameters, VeRA
    \citep{kopiczko2023vera} one (numerically, Appendix~\ref{app:repro}; for $r\le n$ and generic frozen factors also
    by \bgref{C.14}) and FourierFT one per sampled conjugate pair. At the budget of $\mathrm{LoRA}_{2r}$, with
    $2r\le\min(m,n)$, $r^2\le\min(m,n)$ and $2r^2<m+n-1$, $\mathrm{LoHa}_{r,r}$ reaches at least $(m+n-1)-2r^2$ fewer
    dimensions, and is strictly dominated for $r\le2$. When $2r^2\ge m+n-1$, as for $r=64$ on $4096\times4096$ or $r=32$
    on 320- to 640-wide layers, LoHa matches or beats $\mathrm{LoRA}_{2r}$ in dimension too, by its numerically checked
    dimension formula.
  \predfield{Test} On a small-budget benchmark, compare methods at matched $\lvert M\rvert$ and at matched $d(M)$,
    computing $d$ by a float64 Jacobian rank where no formula is proved. Report the maximum rank too, and name its cut.
  \predfield{Would fail if} the two rankings agreed at every budget tested, so that gauge is too small a share of the
    budget to matter there.
  \predfield{Toy} Figure~\ref{fig:rankvol} places LoRA, LoHa, GraLoRA, HiRA, KronA, MoRA, FourierFT and C3A at one
    budget on a $4096\times4096$ layer by maximum rank and by $d(M)/(mn)$. The companion site's budget calculator sets
    $\lvert M\rvert$ beside $d(M)$ on real base models.
\end{predmeta}
\end{prediction}

\subsubsection{Predictions about symmetry}\label{sec:predictions-symmetry}

The hidden units of a feed-forward block carry no order. Permute them with the matching rows and columns, and the network
computes the same function. A method that never looks at unit order should not notice. A method built on a frame, a
reshape or a block partition of the hidden coordinates should.

\begin{prediction}{X.3}{the permutation test}
Over random function-preserving permutations of hidden units, LoRA's outcome variance equals its seed variance.
FourierFT, LoKr and MoRA show larger variance, and so do BOFT and OFT on modules whose input side is the hidden layer
(\texttt{down\_proj}). On \texttt{up\_proj} and \texttt{gate\_proj} the permutation only reorders BOFT's output rows, so
BOFT is exactly equivariant there.
\begin{predmeta}
  \predfield{Scope} The LoRA half holds for any coordinatewise optimizer (SGD, Adam, AdamW), which commutes with
    coordinate permutations of $Q$ (the third item of \thmref{Proposition}{II.13}), and an initialisation of $A$ with i.i.d.\ entries. The
    rest is empirical.
  \predfield{Support} \thmref{Lemma}{VI.2} (classical): the linear maps that commute with GELU or SiLU are the
    permutations, and with ReLU the positive monomials. Gated SwiGLU or GeGLU blocks admit the full monomial group
    through their linear branch (rule R4 of \thmref{Theorem}{VI.3}). \thmref{Observation}{II.15} turns covariance under
    those maps into equal reachable function sets.
  \predfield{Test} Apply $K$ random permutations to the rows of \texttt{up\_proj} and \texttt{gate\_proj} and the
    columns of \texttt{down\_proj}, and confirm that outputs agree to rounding. Fine-tune each method once per permuted
    copy with a fresh seed, and $K$ times on the original with $K$ seeds; compare the two variances, method by method.
  \predfield{Would fail if} LoRA's permutation variance exceeded its seed variance, pointing to an implementation tied
    to unit order, or the listed methods showed no excess.
  \predfield{Toy} No variance test. The companion site's merge game runs the rule behind the lemma: a positive diagonal
    slides through ReLU and stops at GELU.
\end{predmeta}
\end{prediction}

\paragraph{Credit}
Lemma VI.2 is classical \citep{xsussmann1992uniqueness,xchen1993geometry,xgodfrey2022symmetries}. The methods under
test are FourierFT \citep{gao2024parameter}, LoKr \citep{yeh2023navigating}, MoRA \citep{jiang2024mora}, BOFT
\citep{liu2023parameter} and OFT \citep{qiu2023controlling}.

\begin{explains}
If the prediction holds, seed variance understates the variance of methods tied to unit order across equally good
checkpoints.
\end{explains}

SliceGPT rotates the residual stream of a model without changing what it computes. A method whose image rotates with the
weights reaches the same functions afterwards; a method built on the stream's coordinates may lose or gain reach.

\begin{prediction}{X.4}{the rotation test}
Take a pre-norm transformer with its norm gains fused into the linear maps they feed and LayerNorm converted to RMSNorm,
and rotate its residual stream by one global orthogonal matrix. This provably leaves unchanged the reachable function
sets of LoRA, of (IA)$^3$ with HF defaults, and of DoRA on the input-rotated boxes ($q$, $k$, $v$, gate, up).
\thmref{Conjecture}{A}, restated: against the gain-fused, unrotated model, the rotation measurably changes the
fine-tuning results of block OFT/BOFT on residual-input projections, LoHa ($r\ge2$), HiRA and coordinate masks, because
LLM residual streams have a privileged basis.
\begin{predmeta}
  \predfield{Scope} The proved half concerns reachable function sets, for every optimizer and for box-wise methods with
    no shared parameters or random frames. Training outcomes are exactly unchanged only for (IA)$^3$, and in law for
    LoRA and DoRA under SGD with an isotropic initialisation. HF's Kaiming-uniform $A$ is not rotation-invariant in law,
    and Adam is only $B_r$-equivariant (\thmref{Theorem}{III.12}), so both break this. Sandwich and post-norm models
    (Gemma-2/3, OLMo-2) and SliceGPT's per-block rotations are excluded.
  \predfield{Support} \thmref{Observation}{II.15} (repaired), a sufficient condition. The conjectured methods fail it,
    which by itself proves nothing.
  \predfield{Test} Keep the gain-fused model as control, rotate a copy, and confirm that outputs agree to rounding.
    Fine-tune every method on both over several seeds. LoRA's reach provably does not change, so its change in results
    measures the share of the optimizer and the initialisation.
  \predfield{Would fail if} the listed methods changed no more than LoRA does.
\end{predmeta}
\end{prediction}

\paragraph{Credit}
The rotation is that of SliceGPT \citep{xashkboos2024slicegpt}. In QuaRot \citep{xashkboos2024quarot}, the per-head
Hadamard fused offline into $W_v$ and $W_{\mathrm{out}}$ is a value--output head gauge, which
\thmref{Observation}{II.15} covers; the online Hadamards preserve the function but insert new operations, outside its
hypothesis.

PiSSA and CorDA choose their split from a truncated SVD, which follows the coordinates under rotations and uniform
scalings and changes when they are stretched unevenly. Whitening the input first removes that dependence from the split.

\begin{prediction}{X.11}{whitened splits}
\thmref{Conjecture}{C}, restated: whitened CorDA or PiSSA is more robust to function-preserving input
reparametrisations, and no worse otherwise.
\begin{predmeta}
  \predfield{Scope} The covariance of the split is optimizer-free; the robustness of trained results is the empirical
    part. Absolute damping $C+\lambda I$ cuts the covariance to $\Ort(n)$, while CorDA's trace-relative damping keeps
    $\CO(n)$. The released CorDA code divides each calibration sample by its largest entry, which breaks even rotations,
    so only the published formulas are covered.
  \predfield{Support} \thmref{Proposition}{II.16} (repaired), for $C=\mathbb E[xx\T]\succ0$, $1\le r<\min(m,n)$ and
    spectral gaps. The whitened split $\Pi_rW_0$, with $\Pi_r$ the top-$r$ eigenprojector of $W_0CW_0\T$, is
    $\GL_n$-covariant on the input side. The PiSSA and MiLoRA splits and the paper-form CorDA splits have input-side
    group exactly $\CO(n)$, so norm-gain folding or SmoothQuant rescalings change them.
  \predfield{Test} Prepare function-preserving versions of one checkpoint that differ by non-conformal input maps
    (folded RMSNorm gains, or a per-channel rescaling with its inverse folded into the producer). Initialise PiSSA,
    paper-form CorDA and the whitened split on each, fine-tune with one optimizer, and compare the spread across
    versions; compare all three on the unmodified checkpoint for the ``no worse'' half.
  \predfield{Would fail if} PiSSA and CorDA varied no more than the whitened split across versions, or the whitened
    split did worse on the unmodified checkpoint.
\end{predmeta}
\end{prediction}

\paragraph{Credit}
The splits are PiSSA \citep{meng2024pissa}, MiLoRA \citep{wang2024milora} and CorDA \citep{yang2024corda}; the whitened
form is the truncation-aware whitening of SVD-LLM \citep{xwang2025svdllm}. EVA \citep{paischer2024parameter} starts at
$B_0=0$ and splits nothing.

\subsubsection{Predictions about base change and merging}\label{sec:predictions-base}

Merging and restarting lets a method compose its own moves. For an additive cone such as ReLoRA's, composition keeps
buying rank. For block OFT with a fixed partition, a product of block rotations is again a block rotation.

\begin{prediction}{X.6}{re-based OFT}
With an exact Cayley map and a fixed partition, merge-and-restart OFT reaches nothing beyond the closure of one cycle,
$W_0\cdot\prod\SO(b)$, so fixed blocks gain nothing. Partitions whose hypergraph is connected generate $\SO(n)$, so re-based
exact OFT gains reach only if its partition changes between cycles or its butterflies reach full depth. LoRA-type cones
gain from rebasing with a fixed shape; fixed linear shapes (masks, BitFit, (IA)$^3$) gain nothing.
\begin{predmeta}
  \predfield{Scope} Reachable sets, hence optimizer-free, and silent on trained performance. HF PEFT's default
    Cayley--Neumann OFT is not orthogonal; with small generators ($\lVert Q\rVert_{\mathrm{op}}\le2^{1/4}$) each merge
    is a contraction, so repeated merging can only shrink neuron norms and singular values. Iterated exact BOFT with
    $m'$ butterfly factors reaches $W_0\cdot\SO(n)$, up to its output scale, only at full depth $b\,2^{m'-1}=n$.
  \predfield{Support} \thmref{Theorem}{V.3}(c) (repaired) and its corollaries 2 and 4.
  \predfield{Test} Run $K$ cycles of merge-and-restart OFT with \texttt{use\_cayley\_neumann=False} and a fixed
    partition, logging the partial Grams $W_{:,J}W_{:,J}\T$ and the loss. Repeat with a fresh random partition each
    cycle, and with ReLoRA at the same per-cycle budget. Only the fixed-partition run keeps its partial Grams fixed to
    rounding.
  \predfield{Would fail if} a partial Gram moved beyond rounding under the fixed partition, so that the map was not
    exactly orthogonal. Gains in loss from restarts alone are outside the claim.
  \predfield{Toy} Figure~\ref{fig:rebase}(b) shows the components of the hypergraph generated by block-OFT, butterfly
    and Givens factors on 16 coordinates and the dimension of the Lie algebra they generate against
    $\dim\so(16)=120$; a fixed block partition never connects the hypergraph.
\end{predmeta}
\end{prediction}

\paragraph{Credit}
OFT \citep{qiu2023controlling} and BOFT \citep{liu2023parameter} introduced the block and butterfly parametrisations;
ReLoRA \citep{lialin2023relora} introduced merge-and-restart for LoRA.

LoraHub mixes adapters factor by factor. A factor belongs to no adapter in particular, since $(B,A)$ and $(cB,A/c)$ give
the same update, so the mix depends on choices the adapters do not record unless the modules share a factor.

\begin{prediction}{X.9}{LoraHub seeds}
LoraHub's factor-wise merge $(\sum_iw_iB_i)(\sum_jw_jA_j)$ behaves like update arithmetic, up to the factor
$\sum_jw_j$, when its modules share the $A$-seed and $A$ stays close to its initialisation. It degrades when the modules
come from independent seeds.
\begin{predmeta}
  \predfield{Scope} The algebra is post hoc and optimizer-free. Whether $A$ stays near $A_0$ is empirical; a large
    $\eta_B/\eta_A$, as in LoRA+, favours it.
  \predfield{Support} \thmref{Proposition}{V.4} (repaired). Replacing $(B_2,A_2)$ by $(cB_2,A_2/c)$ changes the merge by
    \begin{equation}\label{eq:pred-lorahub}
      \sum_{l\ne2}w_lw_2\big[(c-1)B_2A_l+(c^{-1}-1)B_lA_2\big],
    \end{equation}
    so it is not a function of the adapters. With one shared frozen $A$ it returns $(\sum_jw_j)\sum_iw_iB_iA$ exactly.
    HF's \texttt{add\_weighted\_adapter} with \texttt{linear}, and its non-SVD \texttt{ties} and \texttt{dare\_*}
    variants, are factor-wise rules of the same kind.
  \predfield{Test} Train $N$ modules on $N$ tasks from a shared $A$-seed and from independent seeds. Merge each set by
    LoraHub's rule at fixed weights and compare with $(\sum_jw_j)\sum_iw_iB_iA_i$, as matrices and on the downstream
    metric, logging $\lVert A_i-A_0\rVert/\lVert A_0\rVert$.
  \predfield{Would fail if} independent seeds merged as well as shared ones, so that the cross terms $B_iA_j$, $i\ne j$,
    cost nothing in practice.
  \predfield{Toy} Figure~\ref{fig:gauge}, merge panel: a gauge of one module moves the LoraHub merge and leaves task
    arithmetic unchanged.
\end{predmeta}
\end{prediction}

\paragraph{Credit}
LoraHub is due to \citet{huang2023lorahub}. FFA-LoRA \citep{sun2024improving} freezes a shared factor, and KnOTS
\citep{stoica2024model} re-factorises the modules jointly through a shared $U\Sigma$. Task arithmetic
\citep{ilharco2022editing} combines updates and has no gauge to break.

\begin{explains}
A safe rule combines LoRAs on the products $B_iA_i$, or on factors only when the modules share one.
\end{explains}

\subsubsection{One more expectation, and how to read a failure}\label{sec:predictions-failure}

One expectation is not on the list. By \thmref{Proposition}{II.13}, unconstrained Kronecker sums are LoRA after a
rearrangement of entries, and FourierFT (with no sampled conjugate pair), WaveFT (when HF's pad--IDWT--crop map is
invertible) and C3A are coordinate masks in a fixed frame. An optimizer that sees the loss only through the trainable
parameters runs each exactly as its source on a transported loss. At matched learning-rate sweeps we therefore expect the
gap between a transported method and its source to follow how well each image aligns with the task's update, measured by
projecting a full fine-tuning $\Delta W$ onto each image.

A failure inside the hypotheses of \hyperref[res:X.1]{X.1}, \hyperref[res:X.5]{X.5} or \hyperref[res:X.6]{X.6}, or of a
proved half, would mean an error in a proof, and we would mark that result refuted; a failure outside them shows which
hypothesis matters. A failure of \hyperref[res:X.2]{X.2}, \hyperref[res:X.4]{X.4} or \hyperref[res:X.11]{X.11} would
refute \thmref{Conjecture}{D}, \thmref{Conjecture}{A} or \thmref{Conjecture}{C}, respectively.

\section{What is analogy, and what is open}\label{sec:analogy}

This exposé gives a plain account of which borrowed words carry theorems, what is open, whose results are re-derived
here, and what the referees changed. Referees repaired 27 of the \nPropositions{} results, and a second review narrowed
19 of those again.

\subsubsection{What the borrowed words mean}\label{sec:analogy-words}

Words borrowed from geometry and physics compress a great deal, and each also carries connotations no theorem here
supports. Table~\ref{tab:analogy} gives, for seven such words, the sense each has in this paper, the results behind it,
and what it leaves out.

\begingroup
\footnotesize
\renewcommand{\arraystretch}{1.18}
\setlength{\LTpre}{6pt}\setlength{\LTpost}{8pt}
\begin{longtable}{@{}>{\raggedright\arraybackslash}p{2.1cm}>{\raggedright\arraybackslash}p{3.75cm}>{\raggedright\arraybackslash}p{4.5cm}>{\raggedright\arraybackslash}p{3.75cm}@{}}
\caption{Analogy versus theorem: seven borrowed words, their meaning here, what is proved under each, and what is not
claimed.}\label{tab:analogy}\\
\toprule
\headrow \textbf{Word} & \textbf{Meaning here} & \textbf{Proved} & \textbf{Not claimed}\\
\midrule
\endfirsthead
\multicolumn{4}{@{}l}{\footnotesize\itshape Table~\ref{tab:analogy}, continued}\\
\toprule
\headrow \textbf{Word} & \textbf{Meaning here} & \textbf{Proved} & \textbf{Not claimed}\\
\midrule
\endhead
\midrule
\multicolumn{4}{@{}r}{\footnotesize\itshape continued on the next page}\\
\endfoot
\bottomrule
\endlastfoot
\textbf{Grothendieck} & The relative point of view: a method is a map $\rho:(Q,q_0)\to(\Theta,\theta_0)$, methods
  form the pointed slice $\Meth(\Theta,\theta_0)$, and changing $\theta_0$ is base change. & Frozen model initial and full
  fine-tuning terminal; products are fibre products in $\Diff$ and can fail to exist (\thmref{Obs.}{I.4}); a symmetric
  monoidal sum (\thmref{Obs.}{I.7}); automorphisms by transport (\thmref{Prop.}{II.13}); pushforward along
  function-preserving maps (\thmref{Def.}{V.1}(a)). & No schemes, sheaves, descent theory or stacks. Quantisation and
  rebasing borrow only the word ``base change'', and idle constructions such as $\int\Meth\to\mathbf{PNet}$ are left
  out. ``Exposé'' and the rising sea are homage.\\
\textbf{Erlangen} & After \citet{xklein1872vergleichende}, a method's geometry is its covariance group $\mathcal G_M$
  (\thmref{Def.}{II.14}, \bgref{D.17}); what a group-type method can never change is a complete invariant of the
  action. & Complete invariants of the orthogonal, torus and Hadamard actions (\thmref{Thm}{II.5}); exactly orthogonal
  input-side methods keep $K_{\mathrm{out}}$ and the spectrum through any number of merges (\thmref{Cor.}{II.6},
  \thmref{Thm}{V.3}); covariance groups of named methods (\thmref{Prop.}{II.9}(e), \thmref{Prop.}{II.16}). & For cones
  such as LoRA's image the word only names the shape and its covariance. Keeping an invariant concerns reach; that it
  causes knowledge retention is not shown.\\
\textbf{Noether} & The pattern that turns a continuous symmetry of $\rho$ into a conserved quantity. &
  \thmref{Thm}{III.5}, for gradient flow driven by any signal $\Lambda D\rho\T\gamma$, a mild extension of
  \citet{zhao2022symmetries}; for LoRA the charges are the entries of $\Phi$. & No Lagrangian. Discrete steps drift by
  $O(\eta^2)$ relative to the charge, and Adam and momentum break the law.\\
\textbf{Gauge} & The symmetry group of a parametrisation (\thmref{Def.}{I.5}, \bgref{D.21}); for LoRA, $\GL_r$ acting by
  $(Bg^{-1},gA)$. & Generic gauges and fibre dimensions (\thmref{Prop.}{II.12}); the part each optimizer respects,
  $B_r\subset\Ort(r)\subset\GL_r$ (\thmref{Thm}{III.12}); the gauge dependence of factor-wise merging
  (\thmref{Prop.}{V.4}). & No connection or curvature. Fibres are gauge orbits only on the full-rank stratum; away from it
  the fibres jump.\\
\textbf{Moduli} & A set of isomorphism classes (\bgref{C.20}). & Three strata of full-rank LoRA pointings are classified,
  as a set, by $\Gr(r,n)\sqcup\Gr(r,m)\sqcup\MM_r$ (\thmref{Thm}{II.2}). & No topology: the natural quotient topology is
  not a disjoint union ($\mathrm T_A$ lies in the closure of $\mathrm S$). No stack is built, and the strata miss some
  pointings (HF's \texttt{orthogonal} init, defect pointings).\\
\textbf{Fibre} & The preimage $\rho^{-1}(\theta)$, the parameters that give the same weights, as in the title. & Fibre
  dimensions (\thmref{Prop.}{II.12}(a)); which points of the apex fibre $\{BA=0\}$ an optimizer can tell apart
  (\thmref{Cor.}{III.13}). & $\rho$ is no fibre bundle; the fibre dimension jumps between strata.\\
\textbf{Lens} & Backpropagation as the 2-functor $\Para(\mathsf R)$, which sorts tangent-side from cotangent-side methods
  (\bgref{B.17}). & The degree-one collapse: GaLore with weight decay $0$ is one-sided LoRA with a moving frame and
  carried optimizer state (\thmref{Prop.}{IV.2}). & The duality is exact only at degree one, for optimizers that do not
  read the parameter tensor, with no weight decay; decoupled decay, default Adafactor and full-gradient clipping break
  it.\\
\end{longtable}
\endgroup

\paragraph{Credit}
$\Para$ and the lens reading of backpropagation are due to \citet{fong2017backprop} and \citet{cruttwell2021categorical}.
The ETHER control study that tempers the knowledge-retention reading of orthogonal methods is \citet{bini2024ether}.

Three scopes travel with every claim. Expressivity is a different property from learnability. Only
Exposé~\ref{sec:fibres} is dynamical, and each of its results names its optimizer. Mergeability is proved box by box
under explicit side conditions (position-uniformity, no weight sharing, bias slots); the rewrite rules give sufficient
conditions only, and whole-network statements are open.

Most of the mathematics is real algebraic geometry, invariant theory, Lie generation and gradient-flow ODEs. The category
theory does its own work in seven places: the interval [frozen, full FT] and simulation; invariants finer than the
image; a product that yields LoRA-XS, and meets that organise others; base change along maps that change the function;
the lens; naturality of gradient descent and of combination rules; and rank as a min-cut.

Some statements rest on less than a full proof. The equalities in the dimensions of LoHa (\thmref{Proposition}{II.12})
and VeRA are checked numerically, and the bounds are proved; for VeRA with $r\le n$ and generic frozen factors the
equality also follows from the toric-cone description of \bgref{C.14}, and the generic equality for DoRA
(\thmref{Proposition}{II.8}) is proved. The case of unconstrained generators in corollary 4 of \thmref{Theorem}{V.3} is
recorded as a proof sketch, which Appendix~\ref{proof:V.3} completes. The continuous case of
\thmref{Proposition}{V.4}(a) and the converse in \thmref{Corollary}{III.13} are proved in Appendix~\ref{app:proofs}, and
\thmref{Proposition}{IV.3} is argued around Frobenius' theorem (\bgref{D.25}). The ancillary script
\texttt{framework\_checks.py} re-runs the main numbers in about a minute (Appendix~\ref{app:repro}).

\subsubsection{Open problems}\label{sec:analogy-open}

Each problem grew out of a proof that stopped short, and each is stated so that a solution would be recognisable.

\begingroup\emergencystretch=2em
\begin{enumerate}[label=\arabic*.,leftmargin=1.8em,itemsep=4pt]
  \item \textbf{LoHa's dimension.} \thmref{Proposition}{II.12} proves
    \begin{equation}\label{eq:open-loha}
      d(\mathrm{LoHa}_{r_1,r_2})\le\min\big(mn,\ r_1(m+n-r_1)+r_2(m+n-r_2)-(m+n-1)\big),
    \end{equation}
    with equality checked only numerically. Prove it in general, and decide whether the Hadamard variety is closed.
  \item \textbf{Initialisations, dynamically.} \thmref{Theorem}{II.2} classifies three strata by expressivity, and
    \thmref{Corollary}{III.13} shows that Adam sees $A_0$ only up to signed row permutations. Classify the apex fibre
    modulo $B_r$ for Adam, stratified by the charge $\Phi_0$ (\thmref{Corollary}{III.6}), including the degenerate
    pointings outside the three strata of \thmref{Theorem}{II.2}.
  \item \textbf{HRA's apex in practice.} HRA's paired initialisation, HF PEFT's default, starts at an apex
    (\thmref{Theorem}{II.7}(b), for injective $W_0$ and even $r\le n-2$), with $\dim S_1=kn-\tfrac{k(k+1)}2$, $k=r/2$,
    against $d=rn-\tfrac{r(r+1)}2$. Decide whether this apex is a practical first-order bottleneck that slows early
    training, by comparison with a regular initialisation at matched budget.
  \item \textbf{Charts for meets.} LoRA-XS (\thmref{Proposition}{I.8}(a)) and HRA's image (\thmref{Theorem}{II.7}(a))
    are meets with a polynomial chart; the meet of LoRA with output scaling (\thmref{Proposition}{I.8}(b)) is a union of
    coordinate planes. Characterise the pairs of named methods whose meet is parametrisable by a polynomial chart.
  \item \textbf{The frontier.} Incomparable pairs are common: split against zero-init LoRA (\thmref{Theorem}{II.2}(d)),
    Kronecker sums or GraLoRA against LoRA (\thmref{Theorem}{II.11}, \thmref{Proposition}{II.12}(c)). Find the size-$k$
    frontier of $\Meth$, whose maximal elements are pairwise incomparable.
  \item \textbf{An Adam analogue of the crossover.} \thmref{Theorem}{III.5} and \thmref{Theorem}{III.9} hold for
    gradient flow. Find a conserved or monotone quantity for sign-like updates, whose symmetry is only $B_r$
    (\thmref{Theorem}{III.12}).
  \item \textbf{Prefixes and prompts.} For a vanilla softmax head with $W_o\ne0$ and $W_q\ne0$, a generic prefix is not
    box-locally mergeable (\thmref{Theorem}{VI.5}(f)); prove or refute its non-mergeability for the whole network. In causal decoders every soft prompt is a deep prefix
    (\thmref{Theorem}{VI.5}(e)). \citet{petrov2023prompting} and \citet{li2021prefix} assert that prompt tuning is
    strictly weaker than deep prefix tuning (P-tuning v2, \citealp{liu2021p}), and the inclusion is strict in a
    one-layer, one-token example. Prove the generic strictness, and settle encoders.
\end{enumerate}
\endgroup

Conjectures \hyperref[res:A]{A}, \hyperref[res:C]{C} and \hyperref[res:D]{D} are open as well. They appear in
Exposé~\ref{sec:predictions} as Predictions \hyperref[res:X.4]{X.4}, \hyperref[res:X.11]{X.11} and
\hyperref[res:X.2]{X.2}.

\subsubsection{Debts: known results we re-derive}\label{sec:analogy-known}

Much of what this paper proves was known in another language or setting. A re-derivation earns its place only if the new
language adds something, so Table~\ref{tab:known} states what we add next to each debt.

\begingroup
\footnotesize
\renewcommand{\arraystretch}{1.18}
\setlength{\LTpre}{6pt}\setlength{\LTpost}{8pt}
\begin{longtable}{@{}>{\raggedright\arraybackslash}p{2.35cm}>{\raggedright\arraybackslash}p{6.15cm}>{\raggedright\arraybackslash}p{5.6cm}@{}}
\caption{Known results we re-derive, whom they are due to, and what is added here.}\label{tab:known}\\
\toprule
\headrow \textbf{Here} & \textbf{Due to} & \textbf{Added here}\\
\midrule
\endfirsthead
\multicolumn{3}{@{}l}{\footnotesize\itshape Table~\ref{tab:known}, continued}\\
\toprule
\headrow \textbf{Here} & \textbf{Due to} & \textbf{Added here}\\
\midrule
\endhead
\midrule
\multicolumn{3}{@{}r}{\footnotesize\itshape continued on the next page}\\
\endfoot
\bottomrule
\endlastfoot
\thmref{Def.}{I.1}, Exposé~\ref{sec:lens} & \citet{fong2017backprop}; \citet{cruttwell2021categorical} & The pointed
  slice of methods; tangent and cotangent sides.\\
\thmref{Thm}{II.1} & The PiSSA-to-LoRA conversion of \citet{meng2024pissa}, in HF PEFT \citep{xmangrulkar2022peft} & The
  rank-$2r$ bound for every exact pointing, and from any start.\\
\thmref{Thm}{II.5} & Classical: the first fundamental theorem for $\Ort(n)$; Witt's extension theorem (\bgref{D.16}) &
  Which invariant each group-type method keeps.\\
\thmref{Thm}{II.10} & The standard cut-rank bound for tensor networks (\bgref{B.8}) & A rank census of PEFT plugs.\\
\thmref{Thm}{II.11} & \citet{loan1993approximation}; the operator-Schmidt decomposition (\bgref{B.12}) & Kronecker sums
  as LoRA across another cut; the LoKr inclusion criterion.\\
\thmref{Prop.}{II.12}, TT/Tucker row & \citet{xkoch2010dynamical}; \citet{xuschmajew2013geometry};
  \citet{xholtz2012manifolds} & The same count, at minimal ranks, beside other PEFT families.\\
\thmref{Prop.}{III.4}, \thmref{Thm}{III.12}(c) & \citet{xtong2021accelerating}; \citet{zhang2024riemannian}; LoRA-Pro
  \citep{wang2024lorac}; LoRA-FA \citep{zhang2023lora}; descent to the quotient: \citet{mishra2012fixed},
  \citet{xmishra2016riemannian} & One cometric table for (S)GD; the hierarchy $B_r\subset\Ort(r)\subset\GL_r$.\\
\thmref{Thm}{III.5} & \citet{zhao2022symmetries}; \citet{marcotte2023abide}; the principle: \citet{du2018algorithmic},
  \citet{arora2018optimization}, \citet{kunin2020neural} & The formally immediate remark that the signal need not be a
  gradient; PEFT instances (DoRA's detached norm, clipping, VeRA, AdaLoRA without its regulariser, HRA).\\
\thmref{Prop.}{III.7} & \citet{xkempf1979length}, in the real form of \citet{xrichardson1990minimum};
  \citet{xsrebro2004maximum} & Balance as the Kempf--Ness slice of LoRA's gauge; two decay conventions, two slices
  (\thmref{Cor.}{III.6}).\\
\thmref{Thm}{III.9} & \citet{xtarmoun2021understanding}; \citet{xmin2021explicit}; \citet{xdomine2025lazy};
  \citet{xkunin2024get} & The PEFT reading of $\sigma^*$, its shift under LoRA+, PiSSA's balance, the Kaiming band.\\
\thmref{Thm}{III.10}, \thmref{Cor.}{III.11} & \citet{schulman2025lora}, for LoRA+ \citep{hayou2024lora} & Any
  multihomogeneous method; the SGD $c^2$ law; per-block $\varepsilon$, decay and clipping.\\
\thmref{Prop.}{IV.2} & \citet{xtorrobahennigen2025duality}; for random projections, Flora \citep{hao2024flora} & The
  projector-by-schedule table of cotangent methods; optimizer state carried across re-basings.\\
\thmref{Prop.}{IV.3} & Frobenius' theorem (\bgref{D.25}) & When a $\theta$-dependent anchor foliates weight space by
  images of forward methods; GaLore, LISA and MeZO keep a fixed anchor per period, so the test does not separate
  them.\\
\thmref{Lemma}{VI.2} & \citet{xsussmann1992uniqueness}; \citet{xchen1993geometry}; \citet{xgodfrey2022symmetries} &
  Which symmetries a method may ignore; the monomial group of gated FFNs.\\
\thmref{Thm}{VI.3} & (IA)$^3$ merging: \citet{liu2022few}, Sec.~3.3; SSF's re-parameterisation \citep{lian2022scaling};
  LoRA's linearity \citep{hu2021lora}; scale migration through SwiGLU: AWQ \citep{xlin2024awq} & One calculus of
  sufficient conditions, with explicit side conditions.\\
\thmref{Thm}{VI.5} & The identity is standard; (b) \citet{he2021unified}; (c) and the prompt-to-prefix map (e)
  \citet{petrov2023prompting}, Sec.~2.2, who also show that prefix-tuning a pretrained transformer can be a universal
  approximator \citep{petrov2024prompting} & The homomorphism-then-perspective organisation; explicit position hypotheses
  and the slice typing of (e).\\
\end{longtable}
\endgroup

\paragraph{Credit}
We also build on work without re-deriving it. \citet{malladi2022kernel} found the lazy kernel of random-$A_0$ LoRA close
to that of full fine-tuning, which \thmref{Observation}{I.6} reads through $S_1$. The Init[B] scheme in the table of
\thmref{Theorem}{II.2} is from \citet{hayou2024impact}. With a fixed support, the meet of \thmref{Proposition}{I.8}(b)
is (IA)$^3$ masked to $r$ neurons; compare the sparse activation scaling of \citet{xstoehr2024activation}.

\subsubsection{The referee record}\label{sec:analogy-referee}

The framework was synthesised from four proposals and three referee reports. Every numbered result then went to two
independent referees, who checked it against its proof, the literature and the Hugging Face PEFT source, that is, for
its proof and for its fit to the methods as published and as coded. Of the \nPropositions{} results, 14 stood as first
written, with wording fixes, and 27 were repaired by an added hypothesis, a narrowed claim, a replaced proof or a
withdrawn reading. A second review re-examined 26 of the repaired statements in 48 reports. It raised a major objection
to 18 of them, mostly to a side clause: a missing hypothesis, a remark that went past its proof, or a description of a
method that matched only one implementation. For example, HF PEFT's LoftQ initialiser turned out to ignore the LoRA
scale $\alpha/r$ (\thmref{Proposition}{V.2}). In all it found a false or overreaching clause in 19 statements, and those
were narrowed again. Only the final statements appear in this paper; the referee notes of both rounds are in the
ancillary file \texttt{propositions.json}. Table~\ref{tab:referee} records what changed, result by result.

\begingroup
\footnotesize
\renewcommand{\arraystretch}{1.15}
\setlength{\LTpre}{6pt}\setlength{\LTpost}{8pt}
\begin{longtable}{@{}>{\raggedright\arraybackslash}p{3.45cm}>{\raggedright\arraybackslash}p{1.75cm}>{\raggedright\arraybackslash}p{8.9cm}@{}}
\caption{The referee record: the status of each numbered result after both rounds of review, and what the referees
changed.}\label{tab:referee}\\
\toprule
\headrow \textbf{Result} & \textbf{Status} & \textbf{What the referees changed}\\
\midrule
\endfirsthead
\multicolumn{3}{@{}l}{\footnotesize\itshape Table~\ref{tab:referee}, continued}\\
\toprule
\headrow \textbf{Result} & \textbf{Status} & \textbf{What the referees changed}\\
\midrule
\endhead
\midrule
\multicolumn{3}{@{}r}{\footnotesize\itshape continued on the next page}\\
\endfoot
\bottomrule
\endlastfoot
\thmref{Obs.}{I.4}, universal properties & repaired & Products need a fibre product in $\Diff$ and can fail to exist
  ($\mathrm{LoRA}_1\times\text{Frozen}$ on $1\times1$ weights).\\
\thmref{Obs.}{I.6}, $S_1$ is functorial & verified & Monotonicity stated only along pointed morphisms.\\
\thmref{Obs.}{I.7}, additive monoidal structure & repaired twice & $r\mapsto\mathrm{LoRA}_r$ is strong monoidal only
  unpointed or at degenerate pointings ($\dim S_1$ is $mr$ against $2mr$); HF's \texttt{cat} merge described exactly.\\
\thmref{Prop.}{I.8}, a product and a meet & verified & LoRA-XS needs $\rk W_0\ge r$; the set-level meet in (b) is no
  new method.\\
\thmref{Thm}{II.1}, pointings of an additive method & verified & Stated for exact pointings, with a robust form;
  credited to \citet{meng2024pissa}.\\
\thmref{Thm}{II.2}, three strata of LoRA initialisations & repaired & Full-rank pointings only, and only as a set; HF's
  \texttt{orthogonal} init and QLoRA/LoftQ defects lie outside.\\
\thmref{Prop.}{II.3}, first-order rank formula & repaired & Scale and gradient flow made explicit; Adam's first step has
  an $\ell_1$ rate; LoRA-GA's $A_0$, not its $B_0$, maximises the rate.\\
\thmref{Lemma}{II.4}, zero gate & verified & Hypotheses stated; a function-level variant for LLaMA-Adapter; AdaLoRA
  instance corrected.\\
\thmref{Thm}{II.5}, complete invariants & verified & Methods that can create zeros ((IA)$^3$, HiRA) keep only ``zeros of
  $W_0$ stay zero''.\\
\thmref{Cor.}{II.6}, what orthogonal methods never do & repaired twice & Exactly orthogonal input-side maps only; BOFT
  keeps cosines up to sign; HF's Cayley--Neumann OFT and GOFT's scaler excluded.\\
\thmref{Thm}{II.7}, HRA as a meet; its apex & repaired twice & An image-level meet with $W_0\cdot\SO(n)$, no product;
  strictly above LoRA $\cap$ OFT only for exactly orthogonal OFT.\\
\thmref{Prop.}{II.8}, DoRA's image & repaired twice & $\mathrm{Diag}_m\cdot(W_0+\Mr)$ for $W_0$ without zero rows; the
  rank formula needs nonzero magnitudes; Conjecture~D made falsifiable.\\
\thmref{Prop.}{II.9}, the torus ladder & repaired & $\mathrm{RoAd}_1$ is (IA)$^3$ over $\mathbb C$; only zeros of $W_0$
  persist; (IA)$^3$'s FFN vector acts on the input of $W_{\mathrm{down}}$.\\
\thmref{Thm}{II.10}, cut-rank bound & verified & A large min-cut is necessary for high rank and is not sufficient.\\
\thmref{Thm}{II.11}, Kronecker incomparability & repaired twice & $\kappa_1$ is the maximal Kronecker rank for aligned
  splits; LoKr with $r'\le\min(m_2,n_2)$ lies in $\mathrm{LoRA}_r$ iff $r\ge\min(m_1,n_1)\,r'$.\\
\thmref{Prop.}{II.12}, gauge census & repaired twice & GraLoRA's maximum rank is $\min(kr,m,n)$; LoHa loses dimensions
  to $\mathrm{LoRA}_{2r}$ only when $2r^2<m+n-1$.\\
\thmref{Prop.}{II.13}, transport of structure & repaired twice & Any $T\in\GL(\Theta)$; commutes with rebasing for
  additive families and fixed $T$; FourierFT needs no conjugate pair.\\
\thmref{Obs.}{II.15}, descent & repaired twice & Reachable sets only, for box-wise methods without shared frames;
  residual rotations need a gain-fused pre-norm model.\\
\thmref{Prop.}{II.16}, covariant splits & repaired twice & MiLoRA and CorDA-KPM are bottom-$r$ splits; trace-relative
  damping keeps $\CO(n)$; shipped CorDA code is not covered.\\
\thmref{Obs.}{III.2}, chain rule for cometrics & verified & Scope stated: gradient flow, no LoRA dropout, constant
  rates.\\
\thmref{Thm}{III.3}, naturality of gradient descent & verified & The title ``natural only along isometries'' was false in
  general and was replaced.\\
\thmref{Prop.}{III.4}, cometrics that descend & repaired & Undamped (S)GD only; the published methods use AdamW or
  damping.\\
\thmref{Thm}{III.5}, Noether charges & repaired twice & Isometries need the signal to factor through $(t,\rho(q))$;
  relabelled known; the per-channel-rate remark needs distinct rate products.\\
\thmref{Cor.}{III.6}, charge at initialisation; decay & repaired & The decay commentary was reversed: coupled decay tends
  to $\mu=0$, per-time decay to $\Phi=0$.\\
\thmref{Prop.}{III.7}, Kempf--Ness slice & verified & Uniqueness up to $\Ort(r)$ proved; the Loewner floor renamed.\\
\thmref{Prop.}{III.8}, balance--escape trilemma & verified & Balance generalised to $B_0\T B_0=cA_0A_0\T$.\\
\thmref{Thm}{III.9}, isotropic-charge dynamics & verified & Pointwise on the rank-$r$ locus; the Kaiming crossover is a
  band.\\
\thmref{Thm}{III.10}, the hyperparameter torus & verified & Hypotheses explicit; extended to schedules, per-block
  $\varepsilon$, decay and clipping.\\
\thmref{Cor.}{III.11}, LoRA+ and rsLoRA & verified & A shared warm-up keeps the torus; linear modules only.\\
\thmref{Thm}{III.12}, optimizer gauge hierarchy & repaired twice & Adam respects $B_r$, positive diagonals only with
  per-channel rates, $\varepsilon$ and decay; damping cuts $\GL_r$ to $\Ort(r)$.\\
\thmref{Cor.}{III.13}, what zero-$B$ chooses & repaired & The apex fibre is $\{BA=0\}$; the $\GL_r$ case needs
  equivariance along the whole trajectory.\\
\thmref{Prop.}{IV.2}, degree-one collapse & repaired & Decoupled decay, default Adafactor and full clipping break it;
  GaLore carries optimizer state.\\
\thmref{Prop.}{IV.3}, integrability & repaired twice & The ``only if'' was false; the top-$r$ gradient field is
  non-involutive for every $r\ge1$.\\
\thmref{Obs.}{IV.4}, the backprop cone & repaired twice & An upper bound; forward mode pays off only for $k=O(1)$.\\
\thmref{Prop.}{V.2}, quantisation and splitting & repaired twice & Three orders, three defects; HF's LoftQ init ignores
  the LoRA scale; QA-LoRA merges on-grid.\\
\thmref{Thm}{V.3}, rebasing closure & repaired twice & BOFT reaches $\SO(n)$ only at full depth; HF's Cayley--Neumann
  merges contract.\\
\thmref{Prop.}{V.4}, combination rules & repaired twice & The ``iff'' needs full rank or continuity; factor-wise DARE is
  gauge-dependent too.\\
\thmref{Lemma}{VI.2}, activations fix the basis & verified & Gated FFNs have the full monomial group.\\
\thmref{Thm}{VI.3}, the merge calculus & repaired twice & Sufficient conditions only; RoPE needs scales constant on
  rotary pairs.\\
\thmref{Prop.}{VI.4}, linear insertions, ReFT & repaired twice & LoReFT has dimension $r(2d+1-r)$ and a pyreft pointing
  defect; the serial-adapter criterion needs $\rk A_0=r$.\\
\thmref{Thm}{VI.5}, attention as a homomorphism & repaired twice & Rigidity is per layer; the prompt-to-prefix map
  credited to \citet{petrov2023prompting}.\\
\end{longtable}
\endgroup

A repair is a record of an error that was caught. We keep the record public so that a reader can weigh each result by
its history, and so that the next error, when someone finds it, has an obvious place to go.

\section{Related work}\label{sec:related}

This paper draws on three bodies of work: the PEFT literature, with its surveys and unified views; the theory of factorised models, their symmetries and their optimisation; and applied category theory. We take them in turn and say, for each, what we use and where our questions differ. Results that we re-derive from earlier work carry that credit in their own headings and in the contributions of Exposé~\ref{sec:thesis}; here we place the paper as a whole. The catalogue of Appendix~\ref{app:catalogue} cites all \nMethods{} methods and \nPapers{} papers it classifies, and this section names the representatives of each family.

\paragraph{Surveys, unified views and design spaces}
The closest earlier unified frame is that of \citet{he2021unified}. They read adapters \citep{houlsby2019parameter}, prefix tuning \citep{li2021prefix} and LoRA \citep{hu2021lora} as modifications of specific hidden states that differ along four axes: functional form, insertion form, modified representation and composition function. Their observation that prefix tuning is a softmax-gated parallel adapter is part (b) of \thmref{Theorem}{VI.5}, and their MAM adapter, like UniPELT \citep{mao2021unipelt}, is a hybrid in our classification. The delta-tuning study of \citet{ding2022parameter} sorts methods into addition-, specification- and reparametrisation-based families and adds optimisation and optimal-control perspectives. The guide of \citet{lialin2023scaling}, the reviews of \citet{xu2023parameter} and \citet{han2024parameter}, the LoRA survey of \citet{mao2024survey}, the vision survey of \citet{xin2024parameter}, the prompting survey of \citet{liu2021pre} and the modular view of \citet{pfeiffer2023modular} drew the family boundaries that our catalogue starts from.

Other work searches over configurations. \citet{chen2023parameter} parametrise a design space by layer grouping, parameter allocation, tunable groups and strategy assignment, and AutoPEFT \citep{zhou2023autopeft} and NOAH \citep{zhang2022neural} search it automatically. \citet{si2024see} organise PEFT as subspace tuning relative to the SVD frame of $W_0$, which is close in spirit to our transport of structure (\thmref{Proposition}{II.13}), and \citet{fu2022effectiveness} read many methods as sparse fine-tuning of an equivalent model. Recent comparisons under careful learning-rate tuning \citep{he2026unified,lee2026learning} find that vanilla LoRA matches most of its variants; \thmref{Corollary}{III.11} gives one exact mechanism behind such coincidences. Libraries made the field comparable in practice: AdapterHub \citep{pfeiffer2020adapterhub}, Adapters \citep{poth2023adapters}, LLM-Adapters \citep{hu2023llm} and Hugging Face PEFT \citep{xmangrulkar2022peft}, whose source settled many of the conventions used here.

These works classify methods by where a module is inserted and how it is composed, and compare them by benchmark. We classify by invariants of the map $\rho$ into weight space. Two methods with different code can then be shown to have the same image, as DoRA and LoRA followed by an output scale do when $W_0$ has no zero row (\thmref{Proposition}{II.8}). Two methods can be shown incomparable, as LoRA and GraLoRA are at equal budget (\thmref{Proposition}{II.12}) and zero-init LoRA and Kronecker sums with few terms are (\thmref{Theorem}{II.11}). Each of the seven coordinates of Exposé~\ref{sec:atlas} is read off a formula and tied to a result.

\paragraph{Low-rank adaptation and its variants}
LoRA \citep{hu2021lora} trains $W_0+sBA$ from $B_0=0$. Its variants change the scale (rsLoRA, \citealp{kalajdzievski2023rank}), the per-factor learning rates (LoRA+, \citealp{hayou2024lora}), which factor is trained (LoRA-FA, \citealp{zhang2023lora}; the asymmetry study of \citealp{zhu2024asymmetry}), or the split into magnitude and direction (DoRA, \citealp{liu2024dora}, after weight normalisation, \citealp{salimans2016weight}; BiDoRA, \citealp{qin2024bidora}; DeLoRA, \citealp{bini2025delora}). Others change the shape of the factors: shared or random bases (VeRA, \citealp{kopiczko2023vera}; NOLA, \citealp{koohpayegani2023nola}; VB-LoRA, \citealp{li2024vb}; Tied-LoRA, \citealp{renduchintala2023tied}; PRoLoRA, \citealp{wang2024prolora}; RandLoRA, \citealp{albert2025randlora}; Uni-LoRA, \citealp{li2025uni}), a small core between frozen or orthonormal frames (LoRA-XS, \citealp{baazy2024lora}; SVFT, \citealp{lingam2024svft}; LoRA-SB, \citealp{ponkshe2024initialization}; PoLAR, \citealp{lion2025polar}; TinyLoRA, \citealp{morris2026learning}), a single factor and its transpose (SingLoRA, \citealp{bensaid2025singlora}), or one matrix shared across input shards (MiSS, \citealp{kang2024miss}).

A further group changes the training procedure. ReLoRA \citep{lialin2023relora}, COLA \citep{xia2024chain}, PeriodicLoRA \citep{meng2024periodiclora} and ScaLoRA \citep{zhang2025scalora} merge and restart; Delta-LoRA \citep{zi2023delta} also updates $W_0$; LoRA-Pro \citep{wang2024lorac}, AltLoRA \citep{yu2025altlora} and LoFT \citep{tastan2025loft} match the full-model gradient or optimizer state; Flat-LoRA \citep{li2024flat} seeks flat minima. \citet{schulman2025lora} map a low-regret regime in which LoRA matches full fine-tuning, and prove the LoRA/Adam case of \thmref{Theorem}{III.10}. Our results place several of these variants exactly: LoRA-XS is the categorical product of the two one-sided methods obtained by freezing LoRA's frames at the SVD frames of $W_0$ (\thmref{Proposition}{I.8}), LoRA+ and rsLoRA are points of a hyperparameter torus under the stated optimizer conditions (\thmref{Corollary}{III.11}), ReLoRA reaches exactly the updates of rank at most $\min(Kr,m,n)$ after $K$ cycles (\thmref{Theorem}{V.3}), and on the full-rank locus the SGD variant of LoRA-Pro follows, under gradient flow, the Riemannian gradient flow on $W_0+\MM_r$ (\thmref{Proposition}{III.4}).

\paragraph{Multiplicative and orthogonal methods}
Orthogonal fine-tuning (OFT, \citealp{qiu2023controlling}) multiplies $W_0$ by a learned orthogonal matrix to preserve hyperspherical energy \citep{liu2018learning}, following orthogonal over-parametrised training \citep{liu2020orthogonal}. BOFT \citep{liu2023parameter} builds the rotation from butterfly factors \citep{dao2019learning}; OFTv2 \citep{qiu2025orthogonal} rotates inputs and replaces the Cayley inverse by a Neumann series; GOFT and qGOFT \citep{ma2024parameter} use Givens rotations, ETHER \citep{bini2024ether} hyperplane reflections and HRA \citep{yuan2024bridging} products of Householder reflections, a parametrisation that goes back to \citet{mhammedi2016efficient}. PSOFT \citep{wu2025efficient} confines the rotation to a principal subspace, LoCO \citep{nguyen2026loco} uses Cayley rotations of low-rank skew generators, and LOFT \citep{zhao2026loft} chooses the support of the rotation from the task. Spectral methods rotate or rescale the singular frame of $W_0$ (SVDiff, \citealp{han2023svdiff}; Spectral Adapter, \citealp{zhang2024spectral}; SVF, \citealp{sun2025transformer}), and POET-X \citep{qiu2026poet} trains weights as orthogonal equivalence transforms of a fixed matrix. Lie-algebra charts such as the exponential and Cayley maps are studied by \citet{lezcanocasado2019cheap}, and LieRA \citep{si2025generalized} lifts multiplicative updates to tensors through the exponential map.

Diagonal and affine scalings form the other half of the multiplicative family: (IA)$^3$ \citep{liu2022few}, SSF \citep{lian2022scaling}, RED \citep{wu2024advancing} and LayerNorm tuning \citep{zhao2023tuning}, preceded by FiLM \citep{perez2017film}. RoAd \citep{liao20243} applies $2\times2$ rotations, and HiRA \citep{huang2025hira} and ABBA \citep{singhal2025abba} use Hadamard products with $W_0$ or with a second learned factor. These papers motivate orthogonality and scaling by what they preserve. We make the preserved quantities exact. The complete invariants of the basic actions (\thmref{Theorem}{II.5}) say what an exactly orthogonal method can never change in exact arithmetic (\thmref{Corollary}{II.6}), also after any number of merges (\thmref{Theorem}{V.3}), and they separate exact methods from approximately orthogonal implementations such as the Cayley--Neumann default of HF PEFT. HRA's bridge between low-rank and orthogonal adaptation becomes, for $W_0$ of full column rank and even $r\le n-2$, an image-level meet with an apex at the default initialisation (\thmref{Theorem}{II.7}), and the torus ladder (\thmref{Proposition}{II.9}) places one- and two-sided rescalings inside the images of rank-one and rank-two HiRA.

\paragraph{Structured shapes}
A second line changes the shape of the update. Kronecker products enter with PHM layers \citep{zhang2021beyond} and Compacter \citep{mahabadi2021compacter}, and KronA \citep{edalati2022krona}, KAdaptation \citep{he2022parameter}, LoKr \citep{yeh2023navigating} and KRAdapter \citep{albert2025higher} apply them to weight updates. Hadamard products of low-rank factors come from FedPara \citep{hyeonwoo2021fedpara}, known as LoHa in LyCORIS. Tensor-network updates include LoRETTA \citep{yang2024loretta}, TT-LoRA \citep{anjum2024tensor}, LoTR \citep{bershatsky2024lotr}, FacT \citep{jie2022fact}, the Tucker-core FLoRA \citep{si2024maintaining}, LoRTA \citep{hounie2024lorta}, MetaTT \citep{lopezpiqueres2025metatt} and QuanTA \citep{chen2024quanta}, resting on the decompositions surveyed by \citet{kolda2009tensor}, the tensor train of \citet{oseledets2011tensor} and the tensor networks of \citet{cichocki2016low} and \citet{wang2023tensor}. Square and block shapes appear in MoRA \citep{jiang2024mora}, MELoRA \citep{ren2024melora}, GraLoRA \citep{jung2025gralora}, MoRe \citep{tan2024more}, built on Monarch matrices \citep{dao2022monarch,qiu2024compute}, and DiaBlo \citep{gurses2025diablo}. Fixed-frame coordinate methods include FourierFT \citep{gao2024parameter}, WaveFT \citep{bilican2025exploring}, LoCA \citep{du2025loca}, SSH \citep{shen2025ssh} and the circulant C3A \citep{chen2024parameter}. Sparse supports appear in SHiRA \citep{bhardwaj2024sparse}, RoSA \citep{nikdan2024rosa}, SLTrain \citep{han2024sltrain} and DSEE \citep{chen2021dsee}, and in the selective methods BitFit \citep{benzaken2021bitfit}, diff pruning \citep{guo2020parameter} and FISH \citep{sung2021training}; Sine-LoRA \citep{ji2024efficient} raises rank with an elementwise nonlinearity.

Many of these papers aim for high rank at a small budget, and we separate the two quantities in that aim. Rank is bounded by a min-cut of a tensor network (\thmref{Theorem}{II.10}), and dimension by the budget minus the generic gauge orbit (\thmref{Proposition}{II.12}), so a method can gain rank at the same dimension, as GraLoRA does, or gain rank while it loses dimension, as LoHa does in the regime $2r^2<m+n-1$. The rearrangement of \citet{loan1993approximation} makes unconstrained Kronecker sums LoRA across a different cut of the same tensor (\thmref{Theorem}{II.11}), and the gauges and dimensions of Tucker and tensor-train shapes are those of \citet{xkoch2010dynamical}, \citet{xuschmajew2013geometry} and \citet{xholtz2012manifolds}.

\paragraph{Initialisation and rank allocation}
Initialisation methods choose the frames that zero-initialised LoRA leaves free, or split $W_0$. PiSSA \citep{meng2024pissa} and MiLoRA \citep{wang2024milora} move the principal or minor singular components of $W_0$ into the adapter, OLoRA \citep{buyukakyuz2024olora} uses a QR split, and CorDA \citep{yang2024corda} and Astra \citep{liu2026astra} use activation statistics. LoRA-GA \citep{wang2024lorab}, LoRA-One \citep{zhang2025lorab} and GoRA \citep{he2025gora} align the adapter with the first gradient, EVA \citep{paischer2024parameter} with activations, KaSA \citep{wang2024kasa} and MiCA \citep{rudiger2026mica} with minor singular directions, LoRA-Null \citep{tang2025put} with a null space of activations, and KeepLoRA \citep{luo2026keeplora} with a residual subspace; \citet{zhang2025primacy} argue that the magnitude of the update matters most. \citet{hayou2024impact} showed that the two zero-product starts, Init[A] and Init[B], train differently, and \citet{li2024crucial} show that a Nyström initialisation changes the convergence rate of scaled gradient descent. Rank allocation is a separate axis, with AdaLoRA \citep{zhang2023adalora}, DyLoRA \citep{valipour2022dylora}, SoRA \citep{ding2023sparse}, IncreLoRA \citep{zhang2023increlora}, ALoRA \citep{liu2024alora}, AutoLoRA \citep{zhang2024autolora}, ElaLoRA \citep{chang2025elalora}, LoRA-drop \citep{zhou2024lora}, the dynamic-rank DoRA of \citet{mao2024dora}, FlexLoRA \citep{liu2026flexlora} and MatryoshkaLoRA \citep{modoranu2026matryoshkalora}; PLoP \citep{hayou2025plop} chooses where adapters go.

These works compare initialisations by training outcomes. We first compare them as objects. The three strata of \thmref{Theorem}{II.2} are told apart by $S_1$ before any step is taken, zero-factor starts are apexes of a cone, and no exact pointing reaches rank above $2r$ (\thmref{Theorem}{II.1}). The $2r$ bound underlies the PiSSA-to-LoRA conversion of \citet{meng2024pissa}, and split initialisations attain it when $2r\le\min(m,n)$. The optimizer then decides how much of the chosen frame training actually sees (\thmref{Corollary}{III.13}).

\paragraph{Quantised bases and function-preserving reparametrisations}
QLoRA \citep{dettmers2023qlora} trains LoRA over a 4-bit base, building on blockwise quantisation \citep{dettmers20218}. LoftQ \citep{li2023loftq} fits the adapter to the quantisation error, and LQ-LoRA \citep{guo2023lq}, IR-QLoRA \citep{qin2024accurate}, ApiQ \citep{liao2024apiq}, LoQT \citep{loeschcke2024loqt} and SRR \citep{cho2026preserve} refine how the error is split; QA-LoRA \citep{xu2023qa}, PEQA \citep{kim2023memory} and L4Q \citep{jeon2024l4q} keep the merged model quantised. Outside PEFT, SmoothQuant \citep{xxiao2023smoothquant} and AWQ \citep{xlin2024awq} migrate scales through activations, SliceGPT \citep{xashkboos2024slicegpt} and QuaRot \citep{xashkboos2024quarot} rotate the residual stream without changing the network function, and SVD-LLM \citep{xwang2025svdllm} uses the whitened truncation that reappears in \thmref{Proposition}{II.16}. Base change (Exposé~\ref{sec:base}) separates the two kinds of map. Function-preserving reparametrisations act on the slice and leave reachable function sets unchanged under the hypotheses of \thmref{Observation}{II.15}, while quantisers move the base point and leave explicit, order-dependent defects (\thmref{Proposition}{V.2}).

\paragraph{Theory of LoRA and of fine-tuning}
\emph{Intrinsic dimension.} \citet{li2018measuring} measured how many random directions a task needs, \citet{aghajanyan2020intrinsic} found that fine-tuning a language model needs few, \citet{qin2021exploring} found a shared low-dimensional subspace for soft prompts, and \citet{gurari2018gradient} observed that gradients concentrate in a small subspace early in training. These notions depend on the data, while the expressive dimension $d(M)$ is a property of the method alone.
\emph{Kernels.} \citet{malladi2022kernel} found that prompt-based fine-tuning often follows kernel dynamics, in which LoRA approximately preserves the kernel; the background is the neural tangent kernel \citep{xjacot2018neural} and lazy training \citep{xchizat2019lazy}. \citet{jang2024lora} show that LoRA in the NTK regime has no spurious local minima, and \citet{li2026learning} extend the kernel view to zeroth-order fine-tuning.
\emph{Expressivity.} \citet{zeng2023expressive} bound the rank LoRA needs to represent a smaller target network.

\emph{Dynamics.} \citet{hayou2024lora} derive LoRA+ from an infinite-width analysis in the spirit of \citet{xyang2021feature}, and \citet{chen2026learning} extend it to joint limits in width and rank. \citet{xu2025understanding} analyse gradient flow for LoRA on matrix factorisation, \citet{kang2026uniform} spectral flows of Muon type, \citet{zhang2026hidden} the role of the scale factor and \citet{mu2025convergence} convergence rates of LoRA gradient descent.
\emph{Landscapes.} \citet{kim2025lora} show that LoRA training converges either to a low-rank, small-magnitude global minimiser or to a qualitatively different solution, in the tradition of benign landscapes for factorised problems \citep{xburer2003nonlinear,xge2017no} and of implicit low-rank bias in matrix factorisation \citep{xgunasekar2017implicit,arora2019implicit}.

Empirically, LoRA and full fine-tuning differ in what they learn and forget \citep{biderman2024lora,shuttleworth2024lora}, fine-tuning has its own scaling laws \citep{zhang2024scaling}, and reinforcement-learning fine-tuning changes small and structured parts of the network \citep{mukherjee2025reinforcement,zhu2025path,shenfeld2025rl}. Our dynamical results are exact statements under stated optimizer hypotheses (Exposé~\ref{sec:fibres}). They complement asymptotic and statistical analyses, and they say nothing about generalisation.

\paragraph{Weight-space symmetry and conserved quantities}
The weight-space symmetries of tanh networks were found by \citet{xsussmann1992uniqueness} and \citet{xchen1993geometry}, and \citet{xgodfrey2022symmetries} computed the linear maps that commute with each common activation; \thmref{Lemma}{VI.2} restates these facts. Rescaling symmetries affect sharpness \citep{xdinh2017sharp} and motivate optimizers that respect them, such as Path-SGD \citep{xneyshabur2015path}. Under gradient flow, symmetries give conservation laws: balancedness of linear layers \citep{du2018algorithmic,arora2018optimization}, the Noether-type laws of \citet{kunin2020neural} and \citet{xtanaka2021noether}, the conserved quantities of \citet{zhao2022symmetries}, whose Prop.~5.1 is the core of \thmref{Theorem}{III.5}, the complete count of independent laws by \citet{marcotte2023abide}, and its extension beyond Euclidean flows, momentum included, by \citet{xmarcotte2024keep}. Symmetry teleportation \citep{xzhao2022teleportation} moves along symmetry orbits to speed up optimisation. Imbalance at initialisation governs the closed-form dynamics of \citet{xtarmoun2021understanding} and \citet{xmin2021explicit}, which \thmref{Theorem}{III.9} re-derives for LoRA, after the exact solutions of \citet{xsaxe2014exact} for deep linear networks; the lazy and rich regimes are analysed by \citet{xdomine2025lazy} and \citet{xkunin2024get}.

Kempf--Ness theory \citep{xkempf1979length,xrichardson1990minimum} characterises closed orbits by norm-minimising points, which for LoRA are the balanced factorisations (\thmref{Proposition}{III.7}), with the nuclear-norm identity of \citet{xsrebro2004maximum}; Balanced LoRA \citep{castin2026balanced} turns this slice into an algorithm. Permutation symmetry is used to align and merge whole models \citep{ainsworth2022git,xentezari2022role,xsingh2020model} and to build networks that process weights \citep{xnavon2023equivariant,xzhou2023permutation}, and \citet{putterman2024learning} build learners on LoRA weights that respect the $\GL_r$ symmetry. Our additions are the PEFT instances. Per-block rates split LoRA's gauge into isometries and conserved charges (\thmref{Theorem}{III.5}), the optimizer decides which part of the gauge descends to the weights (\thmref{Theorem}{III.12}), and combination rules must respect the gauge (\thmref{Proposition}{V.4}). FedRot-LoRA \citep{zhang2026fedrot} and FLoRG \citep{meng2026florg} meet the same gauge in federated averaging, and PRISM \citep{wang2026prism} in differentially private training \citep{xabadi2016deep}.

\paragraph{Riemannian and preconditioned optimisation on low-rank manifolds}
Optimisation over fixed-rank matrices has a mature Riemannian theory \citep{xabsil2008optimization,xvandereycken2013low}, including the quotient geometry of factorisations modulo $\GL_r$ \citep{mishra2012fixed,xmishra2016riemannian} and dynamical low-rank approximation \citep{xkoch2010dynamical}; natural gradients \citep{xamari1998natural} and Riemannian gradient flows of deep linear networks \citep{xbah2022learning} belong to the same circle of ideas. Scaled gradient descent \citep{xtong2021accelerating} preconditions each factor by the Gram matrix of the other, and \citet{zhang2024riemannian} bring it to LoRA. LoRA-RITE \citep{yen2024lora} makes invariance under reparametrisation of the factors a design requirement, LoRA-S \citep{zheng2025lora} lifts matrix-space optimizers through the quotient, Riemannion \citep{bogachev2025lora} takes a Muon step \citep{xjordan2024muon} on $\MM_r$ directly, and PoLAR \citep{lion2025polar} trains orthonormal frames by Riemannian landing; \citet{aliev2026riemannian} study the Riemannian structure of structured orthogonal classes. We sort these update rules, together with SGD, Adam \citep{xkingma2015adam} and AdamW \citep{xloshchilov2019decoupled}, by the largest subgroup of the gauge they respect (\thmref{Theorem}{III.12}). On LoRA's full-rank locus with channel-uniform hyperparameters, this is $\Ort(r)$ for SGD, the signed permutations for Adam and AdamW, and all of $\GL_r$ for undamped scaled GD and LoRA-RITE. \thmref{Proposition}{III.4} identifies which cometrics descend to the rank-$r$ manifold.

\paragraph{Gradient projection and memory}
GaLore \citep{zhao2024galore} projects gradients onto a low-rank subspace that is refreshed periodically. Its relatives include GaLore 2 \citep{su2025galore}, Q-GaLore \citep{zhang2024q}, Fira \citep{chen2024fira}, APOLLO \citep{zhu2024apollo}, Online Subspace Descent \citep{liang2024memory}, LDAdam \citep{xrobert2025ldadam} and LoRA-Pre \citep{wang2026taming}, which compresses the momentum; Flora \citep{hao2024flora} uses random projections, and \citet{he2024subspace} analyse when such methods converge. Memory is also saved by compressing optimizer state \citep{shazeer2018adafactor,dettmers20218,zhang2024adam}, by zeroth-order training (MeZO, \citealp{malladi2023fine}; see the benchmark of \citealp{zhang2024revisiting}), by fused updates (LOMO, \citealp{lv2023full}; AdaLomo, \citealp{lv2023adalomo}), by sampling layers (LISA, \citealp{pan2024lisa}; BAdam, \citealp{luo2024badam}), by side networks (LST, \citealp{sung2022lst}) and by compressing activations (VeLoRA, \citealp{miles2024velora}). \citet{xtorrobahennigen2025duality} proved the duality between linear gradient transformations and adapters that \thmref{Proposition}{IV.2} restates in lens form with explicit optimizer hypotheses. The integrability condition (\thmref{Proposition}{IV.3}) and the backward cone (\thmref{Observation}{IV.4}) sort the rest of this family. In continual learning, gradient projection memory \citep{saha2021gradient} and orthogonal gradient descent \citep{farajtabar2019orthogonal} project updates away from the subspaces that earlier tasks use, an idea that O-LoRA \citep{wang2023orthogonal} and InfLoRA \citep{liang2024inflora} carry over to adapters.

\paragraph{Prompts and prefixes, and their limits}
Soft prompts \citep{lester2021power,liu2021gpt}, deep prefixes \citep{li2021prefix,liu2021p} and their variants (\citealp{vu2021spot,asai2022attempt,wang2023multitask,shi2023dept,razdaibiedina2023residual}; in vision, \citealp{jia2022visual,zhou2021learning,khattak2022maple}) extend the network and leave its weights in place. \citet{petrov2023prompting} show that a prefix cannot change the relative attention pattern over the content and can only bias the output of an attention layer, while \citet{petrov2024prompting} show that prefix-tuning a pretrained transformer can still be a universal approximator. \citet{wang2023universality} prove universality and limitations of prompt tuning, \citet{oymak2023role} analyse prompt-attention, \citet{le2024revisiting} the statistical benefit of the prefix reparametrisation, and \citet{galim2024parameter} show that for state-space models prefix tuning is bounded by initial-state tuning. \citet{chen2022inducer} connect prefixes to adapters through kernel estimators, \citet{khashabi2021prompt} and \citet{su2021transferability} study what soft prompts encode and how they transfer, and \citet{mondal2026can} give exact resource-capped tests for when a prefix can reproduce a LoRA update at one head.

\thmref{Theorem}{VI.5} organises the attention side of this work as a monoid homomorphism followed by a perspective map. It recovers the gated-adapter reading of \citet{he2021unified} and the per-layer rigidity of \citet{petrov2023prompting}, adds that LLaMA-Adapter's \citep{zhang2023llama} separate softmax leaves the content attention weights unchanged, and shows that for a vanilla softmax head with $W_o\ne0$ and $W_q\ne0$ a generic prefix cannot be folded into the head's own weights. Representation edits (ReFT, \citealp{wu2024reft}; RED, \citealp{wu2024advancing}; LoFiT, \citealp{yin2024lofit}) and steering directions \citep{li2023inference,zou2023representation} also act on activations, and \citet{adila2026weight} relate weight updates to activation shifts directly; \thmref{Proposition}{VI.4} treats linear insertions and ReFT.

\paragraph{Model merging and composition}
Weight averaging \citep{wortsman2022model,wortsman2021robust}, Fisher-weighted merging \citep{matena2021merging}, RegMean \citep{jin2022dataless}, task arithmetic \citep{ilharco2022editing} and its tangent-space analysis \citep{ortizjimenez2023task}, TIES \citep{yadav2023ties}, DARE \citep{yu2023language}, AdaMerging \citep{yang2023adamerging} and TSV-Merge \citep{gargiulo2024task} combine full fine-tuning updates, and MergeKit \citep{xgoddard2024mergekit} packages many of these rules. Linear mode connectivity \citep{frankle2019linear} and permutation alignment \citep{ainsworth2022git,xentezari2022role} help explain when averaging works, and surveys and large-scale studies map the field \citep{yang2024model,yadav2024matters,wang2025model}. For adapters, LoraHub \citep{huang2023lorahub}, LoRA Soups \citep{prabhakar2024lora}, AdapterSoup \citep{chronopoulou2023adaptersoup}, PEM arithmetic \citep{zhang2023composing}, ZipLoRA \citep{shah2023ziplora}, Mix-of-Show \citep{gu2023mix} and orthogonal adaptation \citep{po2023orthogonal} combine modules. KnOTS \citep{stoica2024model}, Core Space \citep{panariello2025accurate} and Compress-then-Merge \citep{he2026compress} first express the modules in a shared frame, and OrthoMerge \citep{yang2026orthogonal} averages OFT rotations in the Lie algebra. Poly \citep{ponti2022combining}, MHR \citep{caccia2022multi} and routed mixtures \citep{wu2024mixture,dou2023loramoe,buehler2024x} train combinations jointly, and serving systems such as S-LoRA \citep{sheng2023s} and Punica \citep{chen2023punica} keep many adapters separate at inference. \citet{liu2026appeal} find that adaptive merging of recycled LoRAs gives limited benefit over training a new adapter.

We contribute two exact constraints to this literature. A rule for combining LoRA factors is a function of the adapters only when it is constant on factorisation fibres; task arithmetic on the products satisfies this, and post-hoc product-of-sums rules such as LoraHub's do not (\thmref{Proposition}{V.4}). Folding a trained edit into existing weights is certified by a local rewrite calculus whose side conditions depend on the activation (\thmref{Theorem}{VI.3}); it gives sufficient conditions and box-local obstructions, and SmoothQuant and AWQ already use the same scale migration.

\paragraph{Category theory and geometry for machine learning}
\citet{fong2017backprop} defined learners as the morphisms of a symmetric monoidal category, \citet{fong2019lenses} related learners to lenses, \citet{cockett2019reverse} axiomatised reverse derivative categories, and \citet{cruttwell2021categorical} gave a semantics of gradient-based learning in $\Para(\Lens(\mathcal C))$, which we use directly for the lens of Exposé~\ref{sec:lens}. Parametrised optics \citep{capucci2021foundations}, learners' languages \citep{xspivak2022learners}, reverse derivative ascent \citep{xwilson2021reverse} and the categorical account of automatic differentiation by \citet{xelliott2018simple} belong to the same programme, which \citet{shiebler2021category} survey. \citet{gavranovic2024position} propose monads valued in a 2-category of parametric maps as an algebraic theory of architectures, developed at length in \citet{gavranovic2024fundamental}. Geometric deep learning \citep{bronstein2021geometric}, after group-equivariant \citep{xcohen2016group} and gauge-equivariant \citep{xcohen2019gauge} networks, organises architectures by their symmetries in the spirit of Klein's programme \citep{xklein1872vergleichende}. The calculus of string diagrams we use was put on a rigorous footing by \citet{xjoyal1991geometry}, and the spiders we use for copying come from \citet{xcoecke2013new}.

These works give categorical semantics to architectures and learning algorithms. We study a narrower object, the fine-tuning map relative to one fixed checkpoint, as an object of the slice $\Meth(\Theta,\theta_0)$. The category theory is used in the seven places listed in Exposé~\ref{sec:thesis}, and the rest of the work is real algebraic geometry, invariant theory, Lie theory and gradient-flow analysis. Our Erlangen language applies Klein's idea to the fine-tuning map, where geometric deep learning applies it to architectures, and our gauge is the symmetry group of a parametrisation, with no connection, unlike the gauge of gauge-equivariant networks (Exposé~\ref{sec:analogy}).

\section{Conclusion}\label{sec:conclusion}

Fine-tuning moves a point. Treating a PEFT method as a pointed map $\rho:(Q,q_0)\to(\Theta,\theta_0)$ into weight space, and studying the map more than the module that implements it, turns many informal comparisons between methods into questions with exact answers. Three invariants of the map do most of the work. The image germ at $\theta_0$ governs what a method can reach, through its first-order image $S_1$, its expressive dimension $d(M)$, its cut-rank and the type of its base point. The fibres, together with the optimizer's metric, govern how it trains. Base change governs how it travels when it is initialised by splitting, quantised, restarted or merged. Extensions such as adapters and prefixes add one further layer, where the network itself enters and merging becomes a lifting problem.

Several results follow, each with the scope of its proof. Zero-initialised LoRA starts at the apex of a cone, and so does HRA at the paired initialisation of HF PEFT when $W_0$ has full column rank and $r\le n-2$ is even (\thmref{Theorem}{II.2}, \thmref{Theorem}{II.7}). No exact initialisation lets LoRA reach an update of rank above $2r$ (\thmref{Theorem}{II.1}). Exactly orthogonal methods never change the singular values of a weight in exact arithmetic, however often they are merged (\thmref{Corollary}{II.6}, \thmref{Theorem}{V.3}). Rank is a cut and dimension is a volume, so at equal budget a structured method can gain rank at the same dimension, as GraLoRA does, or trade dimension for rank, as LoHa does when $2r^2<m+n-1$ and $r\ge3$ (\thmref{Proposition}{II.12}). Under Adam with no $\varepsilon$, weight decay or clipping and one shared schedule, LoRA+ follows exactly the trajectory of LoRA with $\alpha$ multiplied by the learning-rate ratio (\thmref{Corollary}{III.11}; the LoRA/Adam case is due to \citealp{schulman2025lora}), and on LoRA's full-rank locus Adam with channel-uniform hyperparameters respects only the signed permutations inside the $\GL_r$ gauge (\thmref{Theorem}{III.12}). Whether a scaling adapter folds into the weights before it depends on the activation, and a local rewrite calculus certifies when it does (\thmref{Theorem}{VI.3}). The atlas reads the same coordinates off \nMethods{} methods from \nPapers{} papers and records \nArrows{} simulation arrows and \nObstructions{} obstructions, each backed by an explicit map or a named test.

Several of these results are short once the setting is in place, which is what the rising sea on the title page refers to (Appendix~\ref{app:making-of}). In the slice, re-pointing an additive method sweeps out the difference set $C-C$ of its update shape, and for LoRA$_r$ that set is the matrices of rank at most $2r$. The bound on every exact initialisation is then a few lines (\thmref{Theorem}{II.1}). Once LoRA's factors are seen to carry a torus of rescalings that fixes the weight trajectory, LoRA+ is a point on that torus (\thmref{Corollary}{III.11}). In the lens, GaLore without weight decay becomes one-sided LoRA on a moving base by a one-line induction (\thmref{Proposition}{IV.2}). Read off a tensor diagram, the maximum rank of a structured adapter is a min-cut (\thmref{Theorem}{II.10}). Three of these four facts were known before this paper, from the PiSSA-to-LoRA conversion \citep{meng2024pissa}, \citet{schulman2025lora} and \citet{xtorrobahennigen2025duality}. What the setting adds is a short proof, an exact scope and the same proof for every method of the same shape. Other results did not dissolve. The classification of initialisations (\thmref{Theorem}{II.2}) and rebasing closure (\thmref{Theorem}{V.3}) required real algebraic geometry and Lie theory, and their proofs say where.

For practitioners these statements become checks that can be made before training. Compare methods at equal expressive dimension and cut-rank as well as at equal budget. Ask whether an initialisation is an apex, and which part of its frame the optimizer will see (\thmref{Corollary}{III.13}). Before treating a new hyperparameter as a new method, check whether it lies on a hyperparameter torus. Before merging adapters factor by factor, check that the rule is constant on factorisation fibres (\thmref{Proposition}{V.4}). Before relying on an orthogonal method to keep a spectrum, check that its implementation is exactly orthogonal. Implementations matter throughout. Several results distinguish a method as published from its HF PEFT implementation, such as OFT's Cayley--Neumann default, HRA's paired initialisation and LoftQ's treatment of the scale $\alpha/r$ (\thmref{Proposition}{V.2}), and the catalogue records which form each statement covers.

The limits are part of the statements. Expressivity is not learnability, and only Exposé~\ref{sec:fibres} is dynamical. Its conservation laws hold under gradient flow, drift under discrete steps by an amount of second order in the step size relative to the charge, and do not cover Adam or momentum. The classification of initialisations is set-level and in exact arithmetic. Mergeability is certified box by box under explicit side conditions, and the rewrite rules give sufficient conditions and local obstructions; they are not a decision procedure for whole networks. Words such as Noether, Erlangen, gauge and moduli are analogies beyond what is proved, and Exposé~\ref{sec:analogy} says where each one stops.

The falsifiable predictions of Exposé~\ref{sec:predictions} say where the theory can be tested, each with its optimizer scope. The open problems of Exposé~\ref{sec:analogy} say where it should grow next: a dynamical classification of initialisations modulo the optimizer's shadow of the gauge, a conserved or monotone quantity for sign-like updates such as Adam's, the frontier of pairwise incomparable methods at a fixed budget, and whole-network statements about prefix tuning. We hope the frame serves as a habit of asking three questions of any new method: what its map can reach, how its fibres meet the optimizer, and what happens to it when the base point moves.


\bibliographystyle{fibresnat}
\bibliography{bib/refs}

\clearpage
\fbappendix
\section*{\LARGE Technical appendix}
\phantomsection\addcontentsline{toc}{section}{Technical appendix}
\markboth{Technical appendix}{}

The appendices fall into three groups. Each can be read on its own, and every entry, proof and table points back to the
place in the main text that uses it.

\paragraph{Preliminaries}
Appendix~\ref{app:background} collects the mathematical background the exposés assume: categories, slices and base
change; monoidal categories, string diagrams and lenses; manifolds, germs and cones; and groups, symmetry and
conservation laws. Its 93 entries keep the numbers A.1--D.34 they carry on the companion site, and each states a notion
with its hypotheses, shows it on a fine-tuning example and lists the results that rely on it. Appendix~\ref{app:auxiliary}
quotes the external theorems that the proofs use, each in the exact form in which it is applied and with its source.

\paragraph{Theory}
Appendix~\ref{app:proofs} gives complete proofs of the numbered results, in the order of the main text and with one
subsection per result. Each subsection restates its result, in the form recorded after two rounds of review, before the
proof.

\paragraph{Data and reproducibility}
Appendix~\ref{app:catalogue} describes the catalogue of \nMethods{} methods from \nPapers{} papers. It lists every method
with its seven coordinates and its reference, and every one of the \nArrows{} recorded simulation arrows with its map $h$.
Appendix~\ref{app:repro} documents the numerical checks marked [Num] in the main text and explains how to rerun them from
the data files that accompany this paper.

\paragraph{Process}
Appendix~\ref{app:making-of} records how this paper and its companion site were made: who did what, how the literature was searched and checked, how the theory was generalised and refereed, and what the work cost in agent runs and tokens.

\providecommand{\fbAusedin}[1]{\par\smallskip\noindent{\small\color{inkgray}{\sffamily\bfseries\color{tide}Used in.}\enspace #1\par}\medskip}

\section{Mathematical background}\label{app:background}

This appendix is for readers who know linear algebra, multivariable calculus, basic probability and the PEFT literature,
and who have not studied category theory, Lie groups or real algebraic geometry. It skips what ML readers already use,
such as the SVD, rank, orthogonal matrices, eigenvalues, Jacobians, gradients, Adam, LoRA and the transformer block. Each
of the 93 entries gives a short definition or statement with its hypotheses, a small example from fine-tuning, the places
in the paper that use it, and usually a textbook for further reading. Definitions carry an ochre rule and standard facts
quoted from the literature a teal one; the one remark, \bgref{A.20}, is set like a definition.

Part A covers categories, slices and base change, Part B monoidal categories, string diagrams and lenses, Part C
manifolds, cones and real algebraic geometry, and Part D groups, symmetry and conservation laws. The parts can be read in
any order, since each entry refers to the entries it depends on at their first use. The exposés cite an entry at the
first use of its term, as in \bgref{A.8}, and each entry ends with the results that rely on it.

A few symbols from the exposés recur in the examples. A method $M=(Q,q_0,\rho)$ is a smooth map $\rho:Q\to\Theta$ from
trainable parameters to weights with $\rho(q_0)=\theta_0$ (\bgref{A.8}), and $\lvert M\rvert=\dim Q$ is its parameter
budget. The expressive dimension $d(M)$, the first-order image $S_1$ and the first-order deficit are defined in
\bgref{C.3}; the base point is \emph{regular} for $M$ when the deficit is $0$ and an \emph{apex} when it is positive. LoRA
\citep{hu2021lora} is $\rho(B,A)=W_0+sBA$ with $B\in\R^{m\times r}$ and $A\in\R^{r\times n}$, and its full-rank
pointings fall into the strata $\mathrm T_A$, $\mathrm T_B$ and $\mathrm S$ of \bgref{C.19}. \emph{Block-scalar rates}
give each factor one learning rate, $\eta_B$ for $B$ and $\eta_A$ for $A$ (\bgref{D.23}).

Table~\ref{tab:bg-index} lists every entry with its area and the first places it is used.

{\footnotesize\setlength{\tabcolsep}{4pt}\renewcommand{\arraystretch}{1.08}
\begin{longtable}{@{}>{\raggedright\arraybackslash}p{0.8cm}>{\raggedright\arraybackslash}p{4.9cm}>{\raggedright\arraybackslash}p{2.35cm}>{\raggedright\arraybackslash}p{5.95cm}@{}}
\caption{Index of the background entries, with the area of each and the first places in the paper that use it.}
\label{tab:bg-index}\\
\toprule\headrow\textbf{No.} & \textbf{Entry} & \textbf{Area} & \textbf{First places used}\\\midrule
\endfirsthead
\toprule\headrow\textbf{No.} & \textbf{Entry} & \textbf{Area} & \textbf{First places used}\\\midrule
\endhead
\midrule\multicolumn{4}{r@{}}{\textit{continued on the next page}}\\
\endfoot
\bottomrule
\endlastfoot
\hyperref[bg:A.1]{A.1} & category & Categories & Table~\ref{tab:notation}, \thmref{Def.}{I.2}\\
\hyperref[bg:A.2]{A.2} & commutative diagram & Categories & Figure~\ref{fig:interval}, \thmref{Def.}{I.2}\\
\hyperref[bg:A.3]{A.3} & isomorphism & Categories & \thmref{Obs.}{I.6}, \thmref{Obs.}{I.7}\\
\hyperref[bg:A.4]{A.4} & functor; functorial invariant & Categories & \thmref{Obs.}{I.4}(c), \thmref{Obs.}{I.6}\\
\hyperref[bg:A.5]{A.5} & natural transformation; naturality & Categories & \thmref{Thm}{III.3}, \thmref{Prop.}{V.4}\\
\hyperref[bg:A.6]{A.6} & slice category & Categories & Table~\ref{tab:notation}, \thmref{Def.}{I.2}\\
\hyperref[bg:A.7]{A.7} & coslice; pointed objects; opposite category & Categories & \thmref{Def.}{I.2}, \thmref{Rem.}{I.3}\\
\hyperref[bg:A.8]{A.8} & the pointed slice $\Meth(\Theta,\theta_0)$ & Categories & Table~\ref{tab:notation}, \thmref{Def.}{I.2}\\
\hyperref[bg:A.9]{A.9} & initial and terminal objects; universal property & Limits & \thmref{Obs.}{I.4}(a), Figure~\ref{fig:interval}\\
\hyperref[bg:A.10]{A.10} & product & Limits & \thmref{Obs.}{I.4}(b), \thmref{Prop.}{I.8}\\
\hyperref[bg:A.11]{A.11} & pullback; fibre product & Limits & \thmref{Obs.}{I.4}(b), \thmref{Prop.}{I.8}\\
\hyperref[bg:A.12]{A.12} & limit; creation of limits & Limits & \thmref{Obs.}{I.4}(b)\\
\hyperref[bg:A.13]{A.13} & adjunction & Limits & \thmref{Obs.}{I.4}(b)\\
\hyperref[bg:A.14]{A.14} & two standard facts about limits & Limits & \thmref{Obs.}{I.4}(b), \S\ref{sec:slice-interval}\\
\hyperref[bg:A.15]{A.15} & monomorphism; sub-object & Limits & \thmref{Def.}{I.2}, \thmref{Cor.}{III.13}\\
\hyperref[bg:A.16]{A.16} & posets; meets; the expressivity order & Limits & \thmref{Obs.}{I.4}(c), \thmref{Prop.}{I.8}(b)\\
\hyperref[bg:A.17]{A.17} & automorphism of a category; transport of structure & Base change & \thmref{Prop.}{II.13}, Exposé~\ref{sec:cut}\\
\hyperref[bg:A.18]{A.18} & lift; lifting problem; obstruction & Base change & \thmref{Obs.}{I.4}(c), \thmref{Obs.}{I.6}\\
\hyperref[bg:A.19]{A.19} & base change; pushforward $\phi_!$; the relative point of view & Base change & Exposés~\ref{sec:thesis}--\ref{sec:slice}, \thmref{Def.}{V.1}\\
\hyperref[bg:A.20]{A.20} & constructions the paper names and leaves unused & Base change & Exposé~\ref{sec:thesis} (where the category theory earns its keep), Exposé~\ref{sec:analogy} (seven borrowed words)\\
\hyperref[bg:B.1]{B.1} & monoidal and symmetric monoidal categories & Monoidal & \thmref{Obs.}{I.7}, \thmref{Thm}{II.10}\\
\hyperref[bg:B.2]{B.2} & direct sum and tensor product in $\FinVect$ & Monoidal & \thmref{Thm}{II.10}, Exposé~\ref{sec:cut}\\
\hyperref[bg:B.3]{B.3} & string diagrams & Monoidal & Exposé~\ref{sec:cut}, \thmref{Thm}{II.10}\\
\hyperref[bg:B.4]{B.4} & cartesian categories; copy and discard & Monoidal & \thmref{Def.}{I.1}, \thmref{Thm}{VI.3}\\
\hyperref[bg:B.5]{B.5} & compact closed categories; cups, caps and wire-bending & Monoidal & \thmref{Thm}{II.11}, \thmref{Thm}{VI.3}\\
\hyperref[bg:B.6]{B.6} & Frobenius algebras and spiders & Monoidal & Exposé~\ref{sec:cut}, \thmref{Thm}{II.10}\\
\hyperref[bg:B.7]{B.7} & tensor networks, contraction, bond dimension & Tensor networks & \thmref{Thm}{II.10}, Exposé~\ref{sec:cut}\\
\hyperref[bg:B.8]{B.8} & the min-cut bound on rank & Tensor networks & \thmref{Thm}{II.10}, Exposé~\ref{sec:cut}\\
\hyperref[bg:B.9]{B.9} & tensor trains, Tucker and MPO; gauge on bonds & Tensor networks & \thmref{Prop.}{II.12}, Exposé~\ref{sec:cut}\\
\hyperref[bg:B.10]{B.10} & Kronecker product, vec, tensor product of subspaces & Tensor networks & \thmref{Thm}{II.11}, Exposé~\ref{sec:cut}\\
\hyperref[bg:B.11]{B.11} & Van Loan--Pitsianis rearrangement; Kronecker rank & Tensor networks & \thmref{Thm}{II.11}, \thmref{Prop.}{II.13}\\
\hyperref[bg:B.12]{B.12} & Schmidt and operator-Schmidt decompositions & Tensor networks & \thmref{Thm}{II.11}, Figure~\ref{fig:cut}\\
\hyperref[bg:B.13]{B.13} & rank of a Hadamard product; face-splitting product & Tensor networks & \thmref{Prop.}{II.12}, \thmref{Thm}{II.10}\\
\hyperref[bg:B.14]{B.14} & sums of methods; Minkowski sums, differences and translates & Para, lenses & \thmref{Obs.}{I.7}, \thmref{Thm}{II.1}\\
\hyperref[bg:B.15]{B.15} & strong monoidal functors; $(\N,+)$ and the permutation groupoid & Para, lenses & \thmref{Obs.}{I.7}, Exposé~\ref{sec:analogy} (the referee record)\\
\hyperref[bg:B.16]{B.16} & bicategories and the Para construction & Para, lenses & \thmref{Def.}{I.1}, \thmref{Rem.}{I.3}\\
\hyperref[bg:B.17]{B.17} & lenses, reverse derivatives and $\Para(\mathsf R)$ & Para, lenses & Exposé~\ref{sec:lens}, \thmref{Def.}{IV.1}\\
\hyperref[bg:C.1]{C.1} & smooth manifolds and smooth maps & Manifolds & \thmref{Def.}{I.2}, \thmref{Def.}{I.5}\\
\hyperref[bg:C.2]{C.2} & tangent spaces, differentials and covectors & Manifolds & Table~\ref{tab:notation}, \thmref{Def.}{I.5}\\
\hyperref[bg:C.3]{C.3} & rank, expressive dimension and first-order image & Manifolds & Table~\ref{tab:notation}, \thmref{Def.}{I.5}\\
\hyperref[bg:C.4]{C.4} & immersions, submersions and embedded submanifolds & Manifolds & \thmref{Obs.}{I.4}, \thmref{Prop.}{IV.3}\\
\hyperref[bg:C.5]{C.5} & constant-rank theorem & Manifolds & \thmref{Def.}{I.5}, \thmref{Prop.}{II.12}\\
\hyperref[bg:C.6]{C.6} & fibres and fibre dimension & Manifolds & Exposé~\ref{sec:fibres}, \thmref{Def.}{I.5}\\
\hyperref[bg:C.7]{C.7} & fibre products, transversality and clean intersection & Manifolds & \thmref{Obs.}{I.4}, \thmref{Prop.}{I.8}\\
\hyperref[bg:C.8]{C.8} & germs & Singularities & \thmref{Thm}{II.7}, Exposé~\ref{sec:image} (the image germ)\\
\hyperref[bg:C.9]{C.9} & smooth and singular points & Singularities & \thmref{Def.}{I.5}, \thmref{Thm}{II.2}\\
\hyperref[bg:C.10]{C.10} & cones and vertices & Singularities & \thmref{Def.}{I.5}, \thmref{Thm}{II.1}\\
\hyperref[bg:C.11]{C.11} & first-order image versus image & Singularities & \thmref{Def.}{I.5}, \thmref{Obs.}{I.6}\\
\hyperref[bg:C.12]{C.12} & the determinantal variety & Varieties & Table~\ref{tab:notation}, \thmref{Thm}{II.1}\\
\hyperref[bg:C.13]{C.13} & the skew-determinantal cone & Varieties & \thmref{Thm}{II.7}, \S\ref{sec:erlangen-hra}\\
\hyperref[bg:C.14]{C.14} & Segre and toric cones & Varieties & \thmref{Prop.}{II.8}, \thmref{Prop.}{II.12}\\
\hyperref[bg:C.15]{C.15} & the Grassmannian & Varieties & \thmref{Thm}{II.2}, \thmref{Thm}{II.7}\\
\hyperref[bg:C.16]{C.16} & images of smooth maps need not be manifolds & Tame geometry & \thmref{Obs.}{I.4}, \thmref{Thm}{II.2}\\
\hyperref[bg:C.17]{C.17} & semialgebraic and definable sets & Tame geometry & Exposé~\ref{sec:thesis} (where the category theory earns its keep), \S\ref{sec:atlas-rules} (arrow rules)\\
\hyperref[bg:C.18]{C.18} & Tarski--Seidenberg and definable dimension & Tame geometry & \thmref{Def.}{I.5}, \S\ref{sec:atlas-rules} (arrow rules)\\
\hyperref[bg:C.19]{C.19} & stratification and local dimension & Tame geometry & \thmref{Thm}{II.2}, \thmref{Thm}{II.7}\\
\hyperref[bg:C.20]{C.20} & moduli sets and moduli spaces & Tame geometry & \thmref{Thm}{II.2}, Exposé~\ref{sec:rosetta}\\
\hyperref[bg:C.21]{C.21} & generic properties & Tame geometry & \thmref{Thm}{II.2}, \thmref{Prop.}{II.8}\\
\hyperref[bg:C.22]{C.22} & real-analytic functions and their zero sets & Tame geometry & \thmref{Lem.}{VI.2}, \thmref{Thm}{VI.5}\\
\hyperref[bg:D.1]{D.1} & Lie groups; the matrix groups of this paper & Lie theory & Table~\ref{tab:notation}, \thmref{Def.}{I.5}\\
\hyperref[bg:D.2]{D.2} & tori, diagonal monoids and saturation & Lie theory & \thmref{Thm}{II.5}, \thmref{Prop.}{II.8}\\
\hyperref[bg:D.3]{D.3} & permutations, monomial matrices, semidirect products, normalisers & Lie theory & \thmref{Prop.}{II.8}, \thmref{Prop.}{II.9}\\
\hyperref[bg:D.4]{D.4} & connectedness and the identity component & Lie theory & \thmref{Thm}{V.3}, \S\ref{sec:erlangen-invariants}\\
\hyperref[bg:D.5]{D.5} & Lie algebras; the commutator; $\gl$, $\so$, $\Sym$, $\wedge$ & Lie theory & \thmref{Thm}{II.7}, \thmref{Thm}{III.5}\\
\hyperref[bg:D.6]{D.6} & the exponential map; one-parameter subgroups & Lie theory & \thmref{Thm}{III.5}, \thmref{Prop.}{III.7}\\
\hyperref[bg:D.7]{D.7} & the Cayley map and its Neumann truncation & Lie theory & \thmref{Cor.}{II.6}, \thmref{Thm}{II.7}\\
\hyperref[bg:D.8]{D.8} & Lie subgroups and subalgebras; the generated Lie algebra & Lie theory & \thmref{Thm}{V.3}, Figure~\ref{fig:rebase}\\
\hyperref[bg:D.9]{D.9} & Yamabe's theorem & Lie theory & \thmref{Thm}{V.3}, \S\ref{sec:base-rebase}\\
\hyperref[bg:D.10]{D.10} & monoids, units, monoid orbits, homomorphisms & Lie theory & \thmref{Thm}{V.3}, \thmref{Prop.}{II.9}\\
\hyperref[bg:D.11]{D.11} & Cartan--Dieudonné theorem; Scherk's sharp form & Lie theory & \thmref{Thm}{II.7}, \S\ref{sec:erlangen-hra}\\
\hyperref[bg:D.12]{D.12} & group actions & Invariants & \thmref{Def.}{I.2}, Table~\ref{tab:notation}\\
\hyperref[bg:D.13]{D.13} & orbits, stabilisers, free actions & Invariants & \thmref{Thm}{II.5}, \thmref{Prop.}{II.9}\\
\hyperref[bg:D.14]{D.14} & Haar measure; invariant measures on homogeneous spaces & Invariants & \thmref{Thm}{II.2}, \S\ref{sec:strata-three}\\
\hyperref[bg:D.15]{D.15} & invariants and complete invariants & Invariants & \thmref{Thm}{II.5}, \S\ref{sec:erlangen-invariants}\\
\hyperref[bg:D.16]{D.16} & first fundamental theorem for $\Ort(n)$; Witt's extension theorem & Invariants & \thmref{Thm}{II.5}, Exposé~\ref{sec:related}\\
\hyperref[bg:D.17]{D.17} & Klein's Erlangen programme & Invariants & \thmref{Def.}{II.14}, Exposé~\ref{sec:erlangen}\\
\hyperref[bg:D.18]{D.18} & equivariance and invariance of maps, update rules and merge rules & Invariants & \thmref{Def.}{II.14}, \thmref{Obs.}{II.15}\\
\hyperref[bg:D.19]{D.19} & multihomogeneous maps; multidegree & Invariants & \thmref{Thm}{III.10}, Exposé~\ref{sec:image} (the image germ)\\
\hyperref[bg:D.20]{D.20} & orbit space of a positive torus & Invariants & \thmref{Thm}{III.10}, \thmref{Cor.}{III.11}\\
\hyperref[bg:D.21]{D.21} & gauge symmetry of a parametrisation & Gauge, metrics & \thmref{Def.}{I.5}, Table~\ref{tab:notation}\\
\hyperref[bg:D.22]{D.22} & quotients and descent; the quotient manifold theorem & Gauge, metrics & \thmref{Prop.}{V.4}, \thmref{Prop.}{III.4}\\
\hyperref[bg:D.23]{D.23} & metrics, cometrics and the pulled-back cometric & Gauge, metrics & \thmref{Def.}{III.1}, \thmref{Obs.}{III.2}\\
\hyperref[bg:D.24]{D.24} & Riemannian gradient flow; tangent projection; retraction & Gauge, metrics & \thmref{Prop.}{III.4}, \thmref{Thm}{III.12}\\
\hyperref[bg:D.25]{D.25} & distributions, Lie brackets and Frobenius' theorem & Gauge, metrics & \thmref{Prop.}{IV.3}, \S\ref{sec:lens-integrability}\\
\hyperref[bg:D.26]{D.26} & Noether's theorem for gradient flows & Conservation & \thmref{Thm}{III.5}, Figure~\ref{fig:noether}\\
\hyperref[bg:D.27]{D.27} & Cartan decomposition; polar decomposition & Conservation & \thmref{Thm}{III.5}, \thmref{Prop.}{III.7}\\
\hyperref[bg:D.28]{D.28} & balancedness; the conserved $B\T B-AA\T$ & Conservation & \thmref{Prop.}{III.7}, \thmref{Prop.}{III.8}\\
\hyperref[bg:D.29]{D.29} & the moment map & Conservation & \thmref{Prop.}{III.7}, \thmref{Cor.}{III.6}\\
\hyperref[bg:D.30]{D.30} & Kempf--Ness theorem, real form & Conservation & \thmref{Prop.}{III.7}, \thmref{Cor.}{III.6}\\
\hyperref[bg:D.31]{D.31} & the closed orbit in each fibre of $(B,A)\mapsto BA$ & Conservation & \thmref{Prop.}{V.4}, \S\ref{sec:base-combine}\\
\hyperref[bg:D.32]{D.32} & Ky Fan's maximum principle & Matrix tools & \thmref{Prop.}{II.3}, \S\ref{sec:erlangen-loft}\\
\hyperref[bg:D.33]{D.33} & Sylvester equations & Matrix tools & \thmref{Thm}{III.12}, Figure~\ref{fig:gauge}\\
\hyperref[bg:D.34]{D.34} & the perspective map; convex combinations & Matrix tools & \thmref{Thm}{VI.5}, Figure~\ref{fig:prefix}\\
\end{longtable}}

\subsection*{Part A. Categories without tears}
\phantomsection\label{app:bg-a}\addcontentsline{toc}{subsection}{Part A. Categories without tears}

Part A collects the category theory the exposés use, from the definition of a category up to base change, and
illustrates every notion in one running example, the pointed slice $\Meth(\Theta,\theta_0)$ of weight-space methods.
Entries \bgref{A.1} to \bgref{A.8} set up categories, functors and slices, \bgref{A.9} to \bgref{A.16} cover universal
properties, limits and orders, and \bgref{A.17} to \bgref{A.19} cover transport, lifting and base change, with
\bgref{A.20} listing the constructions the paper names and leaves unused.

\subsubsection*{Categories and slices}

\begin{bgentry}{A.1}{category}
A \textbf{category} $\mathcal C$ has objects and, for any two objects $X,Y$, a set $\mathcal C(X,Y)$ of
\textbf{morphisms}, or arrows, $f:X\to Y$. Arrows $f:X\to Y$ and $g:Y\to Z$ have a composite $g\circ f:X\to Z$.
Composition is associative, and every object has an identity arrow $\id_X$ with $f\circ\id_X=f=\id_Y\circ f$.
\end{bgentry}

The paper uses $\Set$ (sets and functions), $\Diff$ (smooth manifolds and smooth maps), $\FinVect$ (finite-dimensional
real vector spaces and linear maps, where tensor networks are evaluated) and the pointed categories $\Set_*$ and
$\Diff_*$ of \bgref{A.7}. In $\Meth(\Theta,\theta_0)$ the objects are methods $M=(Q,q_0,\rho)$, and an arrow $h:M\to N$
is a pointed smooth map $h:Q_M\to Q_N$ with $\rho_N\circ h=\rho_M$, read ``$N$ simulates $M$''. Simulations compose,
since $\rho_P\circ k\circ h=\rho_N\circ h=\rho_M$ for $h:M\to N$ and $k:N\to P$. For instance, $h(B)=(B,A_0)$ is an
arrow from LoRA-FA$(A_0)$ \citep{zhang2023lora} to LoRA pointed at $(0,A_0)$. The simulation arrows of the atlas are
arrows of this category, each recorded with its map $h$.

\fbAusedin{Table~\ref{tab:notation}, \thmref{Definition}{I.2}, Exposés~\ref{sec:thesis}--\ref{sec:slice},
\S\ref{sec:atlas-arrows}, Figure~\ref{fig:graph}. \textsf{Reading:} \citet[Ch.~1]{xleinster2014basic}.}

\begin{bgentry}{A.2}{commutative diagram}
A \textbf{diagram} is a family of objects joined by arrows, drawn as a directed graph. It \textbf{commutes} if any two
directed paths with the same start and the same end compose to the same arrow. A triangle commutes when one side equals
the composite of the other two, and a square commutes when its two routes between opposite corners agree.
\end{bgentry}

A simulation is a commuting triangle over $\Theta$, whose two paths from $Q_M$ to $\Theta$ are
\begin{equation*}
  Q_M\xrightarrow{\;h\;}Q_N\xrightarrow{\;\rho_N\;}\Theta\qquad\text{and}\qquad Q_M\xrightarrow{\;\rho_M\;}\Theta.
\end{equation*}
For $M=\mathrm{FA}(A_0)$, $N$ LoRA pointed at $(0,A_0)$ and $h(B)=(B,A_0)$, both paths send $B$ to the weight
$W_0+BA_0$ (scale absorbed into $B$), so the triangle commutes. Figure~\ref{fig:interval}(a) draws, for a general
method, the triangle of Figure~\ref{fig:bg-interval}, which commutes only up to a pointing defect when the method starts
away from $\theta_0$ (\bgref{A.19}). Commuting squares, used for merging on extensions, return in \bgref{A.18}.

\fbAusedin{Figure~\ref{fig:interval}, \thmref{Definition}{I.2}, \thmref{Definition}{VI.1}, \thmref{Definition}{V.1},
Exposé~\ref{sec:base}. \textsf{Reading:} \citet[\S1.1]{xleinster2014basic}.}

\begin{bgentry}{A.3}{isomorphism}
An arrow $f:X\to Y$ is an \textbf{isomorphism}, written $X\cong Y$, if some $g:Y\to X$ satisfies $g\circ f=\id_X$ and
$f\circ g=\id_Y$. The inverse $g$ is then unique. In $\Diff$ the isomorphisms are the diffeomorphisms, so a smooth
bijection need not be one, as $t\mapsto t^3$ shows. In $\Meth$ an isomorphism $M\cong N$ is a pointed diffeomorphism
$h:Q_M\to Q_N$ with $\rho_N\circ h=\rho_M$. Its inverse lies over $\Theta$ automatically, since
$\rho_M\circ h^{-1}=\rho_N$.
\end{bgentry}

The LoRA gauge (\bgref{D.21}) $(B,A)\mapsto(Bg^{-1},gA)$, $g\in\GL_r$, is an isomorphism from LoRA pointed at
$(B_0,A_0)$ to LoRA pointed at $(B_0g^{-1},gA_0)$. Other examples are rsLoRA \citep{kalajdzievski2023rank} $\cong$ LoRA
by $(B,A)\mapsto(\sqrt r\,B,A)$, and $\mathrm{LoRA}_r\oplus\mathrm{LoRA}_s\cong\mathrm{LoRA}_{r+s}$ by concatenating
factors (\bgref{B.14}). Isomorphic methods have the same image, the same $S_1$ and the same first-order deficit
(\bgref{C.3}, \thmref{Observation}{I.6}). They can still train differently, because the optimizer acts on the
coordinates of $Q$. For LoRA's gauge with block-scalar rates (one rate per factor), gradient flow is unchanged exactly
when $g\in\Ort(r)$, at points where $A$ has full row rank and $r<n$ (\thmref{Theorem}{III.3}; \bgref{D.23}).

\fbAusedin{\thmref{Observation}{I.6}, \thmref{Observation}{I.7}, \S\ref{sec:strata-three}, \thmref{Theorem}{II.2},
\thmref{Proposition}{VI.4}, \S\ref{sec:atlas-arrows}.}

\begin{bgentry}{A.4}{functor; functorial invariant}
A \textbf{functor} $F:\mathcal C\to\mathcal D$ sends each object $X$ to an object $FX$ and each arrow $f:X\to Y$ to an
arrow $Ff:FX\to FY$, with $F(g\circ f)=Fg\circ Ff$ and $F(\id_X)=\id_{FX}$. A functor sends isomorphisms to
isomorphisms, so $FX$ is an isomorphism invariant of $X$. A \textbf{forgetful} functor discards structure, as
$\Diff_*\to\Diff$ discards the base point.
\end{bgentry}

The image is a functor $\Im$ from $\Meth(\Theta,\theta_0)$ to the subsets of $\Theta$ containing $\theta_0$, ordered by
inclusion (a poset, \bgref{A.16}), because an arrow $h:M\to N$ gives $\Im M=\rho_N(h(Q_M))\subseteq\Im N$. In the same
way the chain rule at $q_0$ makes $S_1$ a functor to the linear subspaces of $\Theta$, with $S_1(M)\subseteq S_1(N)$
along every pointed arrow. Read backwards, functoriality gives the atlas test O1 and the first half of O3
(\S\ref{sec:atlas-tests}). If $\Im M\not\subseteq\Im N$ or $S_1(M)\not\subseteq S_1(N)$, then $N$ does not simulate
$M$. Base change $\phi_!$ (\bgref{A.19}) and transport $T_*$ (\bgref{A.17}) are functors between slices.

\fbAusedin{\thmref{Observation}{I.4}(c), \thmref{Observation}{I.6}, \S\ref{sec:atlas-arrows},
\thmref{Proposition}{II.13}, \thmref{Definition}{V.1}, \S\ref{sec:lens-functor}. \textsf{Reading:}
\citet[\S1.2]{xleinster2014basic}.}

\begin{bgentry}{A.5}{natural transformation; naturality}
For functors $F,G:\mathcal C\to\mathcal D$, a \textbf{natural transformation} $\alpha:F\Rightarrow G$ is a family of
arrows $\alpha_X:FX\to GX$ with $Gf\circ\alpha_X=\alpha_Y\circ Ff$ for every $f:X\to Y$. This naturality square says that
applying $\alpha$ and then moving along $f$ agrees with moving first and applying $\alpha$ after. A group $K$ can be
viewed as a category with one object whose arrows are the elements of $K$. A functor from it to $\Set$ is then a set
with a $K$-action, and a natural transformation between two such functors is exactly a $K$-equivariant map.
\end{bgentry}

The copy map $\Delta_X(x)=(x,x)$ is natural, $\Delta_Y\circ f=(f\times f)\circ\Delta_X$, which is rule R7 of the merge
calculus (\bgref{B.4}). Elsewhere in the paper, natural for a group of changes of chart means equivariant in this sense
(\bgref{D.18}). Task arithmetic \citep{ilharco2022editing} is natural for all of $\GL(\Theta)$, and DARE
\citep{yu2023language} for invertible diagonal matrices at a fixed mask (\thmref{Proposition}{V.4}(c)). Gradient flow
on LoRA commutes with the gauge $g$ exactly when the induced cometrics (\bgref{D.23}) agree, which for block-scalar rates
singles out $\Ort(r)$ (\thmref{Theorem}{III.3}).

\fbAusedin{Exposé~\ref{sec:thesis} (where the category theory earns its keep), \thmref{Theorem}{III.3},
\S\ref{sec:fibres-naturality}, \thmref{Proposition}{V.4}, \S\ref{sec:merge-calculus}, \thmref{Theorem}{VI.3}.
\textsf{Reading:} \citet[\S1.3]{xleinster2014basic}.}

\begin{bgentry}{A.6}{slice category}
For an object $X$ of $\mathcal C$, the \textbf{slice} $\mathcal C/X$ has as objects the arrows $a:A\to X$, called
objects over $X$, and as morphisms $(A,a)\to(B,b)$ the arrows $h:A\to B$ of $\mathcal C$ with $b\circ h=a$. Composition
and identities are those of $\mathcal C$. The object $X$ is the \textbf{base}, and $a$ is the \textbf{structure map}.
\end{bgentry}

$\Diff/\Theta$ consists of smooth maps $\rho:Q\to\Theta$ into weight space without base points, and its morphisms are
the weak morphisms of \thmref{Definition}{I.2}, which need not send $q_0$ to $q_0'$. A slice always has a terminal
object, the identity $(X,\id_X)$, which here is full fine-tuning. Working in a slice is the relative point of view of
\bgref{A.19}, in which an object is studied together with its map to a base.

\fbAusedin{Table~\ref{tab:notation}, \thmref{Definition}{I.2}, Exposé~\ref{sec:thesis} (where the category theory earns
its keep), Rosetta entries 3 and 4 (\S\ref{sec:rosetta-six}), the seven borrowed words of Exposé~\ref{sec:analogy}.
\textsf{Reading:} \citet[\S II.6]{xmaclane1998categories}.}

\begin{bgentry}{A.7}{coslice; pointed objects; opposite category}
The \textbf{coslice} $X/\mathcal C$ has as objects the arrows $x:X\to A$ out of $X$, and as morphisms $(A,x)\to(B,x')$
the arrows $h:A\to B$ with $h\circ x=x'$. For the one-point space $X=\mathrm{pt}$, an arrow $\mathrm{pt}\to Q$ is a point
$q_0$, so $\Diff_*:=\mathrm{pt}/\Diff$ is the category of \textbf{pointed manifolds} $(Q,q_0)$ and \textbf{pointed
maps}, those with $h(q_0)=q_0'$. Likewise $\Set_*:=\mathrm{pt}/\Set$. The \textbf{opposite category}
$\mathcal C^{\mathrm{op}}$ has the same objects and every arrow reversed, and
$(\mathcal C/X)^{\mathrm{op}}\cong X/\mathcal C^{\mathrm{op}}$.
\end{bgentry}

A method sends its initialisation to the checkpoint, $\rho(q_0)=\theta_0$, so it is an arrow of $\Diff_*$, and a weak
morphism forgets the base points and lives in $\Diff$. Opposites enter in \thmref{Remark}{I.3}. A simulation
$h:M\to N$ runs from $Q_M$ to $Q_N$. The network with $Q_M$-parameters is obtained from the one with $Q_N$-parameters by
precomposing with $h$,
\begin{equation*}
  f\circ(\rho_M\times\id_X)=f\circ(\rho_N\times\id_X)\circ(h\times\id_X),
\end{equation*}
with $X$ the input space. Read as a change of parameters for networks, $h$ therefore points the other way, from the
$Q_N$-network to the $Q_M$-network, and \thmref{Remark}{I.3} records this reversal with an opposite category
(\bgref{B.16}).

\fbAusedin{\thmref{Definition}{I.2}, \thmref{Remark}{I.3}, \thmref{Observation}{I.4}, \S\ref{sec:slice-para}.}

\begin{bgentry}{A.8}{the pointed slice $\Meth(\Theta,\theta_0)$}
Combining \bgref{A.6} and \bgref{A.7}, $\Meth(\Theta,\theta_0):=\Diff_*/(\Theta,\theta_0)$. An object is a pointed smooth
map $\rho:(Q,q_0)\to(\Theta,\theta_0)$, and a morphism $h:M\to N$ is a pointed smooth map with $\rho_N\circ h=\rho_M$.
Equivalently, an object is a factorisation $\mathrm{pt}\xrightarrow{\,q_0\,}Q\xrightarrow{\,\rho\,}\Theta$ of the point
$\theta_0:\mathrm{pt}\to\Theta$, and a morphism is a map between the middle terms that is compatible with both ends.
\end{bgentry}

On one box, LoRA is $\big(\R^{m\times r}\times\R^{r\times n},\,(0,A_0),\,(B,A)\mapsto W_0+sBA\big)$, and OFT
\citep{qiu2023controlling} is $Q\mapsto W_0\,C(Q)$ on skew matrices (block-diagonal in OFT), pointed at $0$, with $C$
the Cayley map (\bgref{D.7}). Full fine-tuning is $(\Theta,\theta_0,\id)$ and the frozen model is
$(\mathrm{pt},*,\theta_0)$. The network $f$ appears nowhere in the definition, so the category depends only on
$(\Theta,\theta_0)$ (\thmref{Remark}{I.3}). The factorisation picture already shows why the frozen model and full
fine-tuning sit at the two ends (\bgref{A.9}).

\fbAusedin{Table~\ref{tab:notation}, \thmref{Definition}{I.2}, \S\ref{sec:slice-methods}, \thmref{Remark}{I.3}, Rosetta
entries 3 and 4 (\S\ref{sec:rosetta-six}).}

\subsubsection*{Universal properties}

\begin{bgentry}{A.9}{initial and terminal objects; universal property}
An object $I$ is \textbf{initial} if every object $X$ receives exactly one arrow $I\to X$, and $T$ is \textbf{terminal}
if every $X$ has exactly one arrow $X\to T$. Each is unique up to a unique isomorphism when it exists. Characterising an
object by the arrows into or out of it in this way is called a \textbf{universal property}. In $\Set$ the empty set is
initial and a one-point set is terminal, and in $\Set_*$ and $\Diff_*$ the one-point space is both.
\end{bgentry}

\begin{figure}[tbp]
\centering
\begin{tikzpicture}[>={Stealth[length=2.3mm,width=1.8mm]},
  obj/.style={align=center, inner sep=3pt, text=ink},
  arr/.style={->, draw=ink, line width=0.6pt, shorten >=1pt, shorten <=1pt},
  lab/.style={font=\small, text=ink}]
  \node[obj] (M) at (0,1.95) {method $M$\\[1pt] $(Q,q_0)$};
  \node[obj] (F) at (-3.3,0) {Frozen\\[1pt] $(\mathrm{pt},*)$};
  \node[obj] (T) at (3.3,0) {Full FT\\[1pt] $(\Theta,\theta_0)$};
  \draw[arr] (F) -- node[above left=-1pt, lab] {$q_0$} (M);
  \draw[arr] (M) -- node[above right=-1pt, lab] {$\rho$, the merge} (T);
  \draw[arr] (F) -- node[below, lab] {$\theta_0$} (T);
\end{tikzpicture}
\caption{The interval of $\Meth(\Theta,\theta_0)$. The left arrow is the only arrow out of the initial object, the right
  arrow $\rho$ is the only arrow into the terminal object, and the triangle commutes because $\rho(q_0)=\theta_0$.
  Figure~\ref{fig:interval}(a) draws the same triangle in the main text, where a pointing defect
  $\rho_M(q_0)\neq\theta_0$ leaves it commuting only up to that defect.}
\label{fig:bg-interval}
\end{figure}
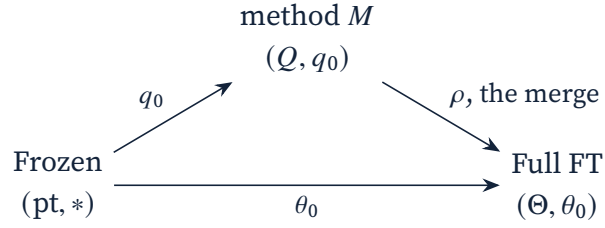

In $\Meth(\Theta,\theta_0)$ the frozen model $(\mathrm{pt},*,\theta_0)$ is initial. The only pointed map
$\mathrm{pt}\to Q$ is $*\mapsto q_0$, and it lies over $\Theta$ because $\rho(q_0)=\theta_0$. Full fine-tuning
$(\Theta,\theta_0,\id)$ is terminal, since an arrow $h$ from $M$ must satisfy $\id\circ h=\rho_M$. So the unique arrow
from $M$ to Full FT is $\rho_M$ itself, the merge, and every weight-space method that starts at $\theta_0$ sits in the
commuting triangle of Figure~\ref{fig:bg-interval}.

\fbAusedin{\thmref{Observation}{I.4}(a), \S\ref{sec:slice-interval}, Figure~\ref{fig:interval}, Rosetta entries 3 and 4
(\S\ref{sec:rosetta-six}), rule A1 in \S\ref{sec:atlas-rules}, \S\ref{sec:base-next}.}

\begin{bgentry}{A.10}{product}
A \textbf{product} of $X$ and $Y$ is an object $X\times Y$ with projections $p_X:X\times Y\to X$ and
$p_Y:X\times Y\to Y$ such that for every object $T$ and arrows $u:T\to X$, $v:T\to Y$ there is exactly one arrow
$\langle u,v\rangle:T\to X\times Y$ with $p_X\circ\langle u,v\rangle=u$ and $p_Y\circ\langle u,v\rangle=v$. It is unique
up to unique isomorphism, and it need not exist. In a slice $\mathcal C/Z$, the product of $(X,a)$ and $(Y,b)$ is the
pullback of $a$ and $b$ in $\mathcal C$ (\bgref{A.11}).
\end{bgentry}

In $\Meth$ a product $M\times N$ has arrows to $M$ and to $N$, so both factors simulate it, and every method $T$ with
arrows $u:T\to M$ and $v:T\to N$ maps to it by exactly one arrow compatible with $u$ and $v$. When it exists, its image is
$\Im M\cap\Im N$. For the one-sided LoRAs with frozen frames $B_0$ of full column rank and $A_0$ of full row rank,
\begin{equation*}
  \mathrm{FA}(A_0)\times\mathrm{FB}(B_0)\;\cong\;\big(\R^{r\times r},\,0,\,R\mapsto W_0+B_0RA_0\big),
\end{equation*}
with projections $R\mapsto B_0R$ and $R\mapsto RA_0$. In the SVD frames of $W_0$, when $\rk W_0\ge r$, this is LoRA-XS
\citep{baazy2024lora} (\thmref{Proposition}{I.8}(a)). The sum $\oplus$ of \thmref{Observation}{I.7} is a different
operation, whose image is a Minkowski sum (\bgref{B.14}).

\fbAusedin{\thmref{Observation}{I.4}(b), \thmref{Proposition}{I.8}, \S\ref{sec:slice-products},
\thmref{Proposition}{II.3}, rule A4 in \S\ref{sec:atlas-rules}, \S\ref{sec:rosetta-reverse}. \textsf{Reading:}
\citet[\S5.1]{xleinster2014basic}.}

\begin{figure}[tbp]
\centering
\begin{tikzcd}[column sep=3.2em, row sep=2.8em, arrows={line width=0.6pt}]
  T \arrow[drr, bend left=22, "v"] \arrow[ddr, bend right=22, "u"'] \arrow[dr, dashed, "\exists!\,t" description]
    & & \\
  & Q_M\times_\Theta Q_N \arrow[r, "p_N"] \arrow[d, "p_M"'] \arrow[dr, phantom, "\lrcorner", very near start]
    & Q_N \arrow[d, "\rho_N"] \\
  & Q_M \arrow[r, "\rho_M"'] & \Theta
\end{tikzcd}
\caption{A pullback square. It commutes, $\rho_M\circ p_M=\rho_N\circ p_N$, and any test object $T$ whose maps $u,v$
  agree over $\Theta$ factors through the corner by exactly one $t$. The small corner mark is the usual sign of a
  pullback.}
\label{fig:bg-pullback}
\end{figure}
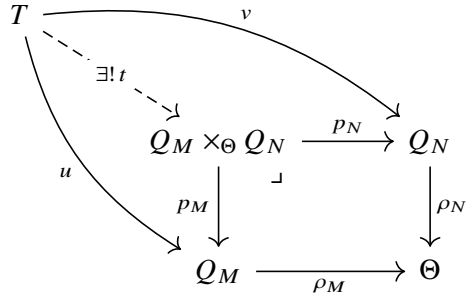

\begin{bgentry}{A.11}{pullback; fibre product}
Given $f:X\to Z$ and $g:Y\to Z$, a \textbf{pullback} is an object $P$ with arrows $p:P\to X$ and $q:P\to Y$ such that
$f\circ p=g\circ q$, and such that whenever $u:T\to X$ and $v:T\to Y$ satisfy $f\circ u=g\circ v$, exactly one
$t:T\to P$ has $p\circ t=u$ and $q\circ t=v$ (Figure~\ref{fig:bg-pullback}). In $\Set$ the pullback is the
\textbf{fibre product} $X\times_ZY=\{(x,y):f(x)=g(y)\}$. In $\Diff$, if this set is an embedded submanifold of
$X\times Y$, it is the pullback, and transversality (\bgref{C.7}) is the standard sufficient condition for that. A
pullback in $\Diff$ can fail to exist.
\end{bgentry}

Products in $\Meth$ are fibre products $Q_M\times_\Theta Q_N=\{(q,q'):\rho_M(q)=\rho_N(q')\}$. For $\mathrm{FA}(A_0)$
and $\mathrm{FB}(B_0)$, with the ranks of \bgref{A.10}, this set is the linear space $\{(B_0R,RA_0)\}$, so the product
exists. For $\mathrm{LoRA}_1$ and the frozen model on $1\times1$ weights the set is the cross $\{ba=0\}\subset\R^2$, no
manifold structure on it has the universal property, and the product does not exist (\bgref{A.13}). Pulling back along
the point $\theta_0:\mathrm{pt}\to\Theta$ gives the fibre $\rho^{-1}(\theta_0)$, the underlying set of
$M\times\text{Frozen}$ whenever that product exists.

\paragraph{HRA is a meet of images, not a product}
Take $W_0$ injective and $r$ even with $2\le r\le n-2$. The fibre product of $\mathrm{LoRA}_r$ with the rotation orbit
$W_0\cdot\SO(n)$ has local dimension $d(\mathrm{HRA}_r)+r^2=rn+r(r-1)/2$ over generic points of the meet
(\bgref{C.19}). This exceeds $rn=\lvert\mathrm{HRA}_r\rvert$ because $r\ge2$, so HRA \citep{yuan2024bridging} is not the
product of the two methods, and \thmref{Theorem}{II.7} keeps only the meet of their images (\bgref{A.16}).

\fbAusedin{\thmref{Observation}{I.4}(b), \thmref{Proposition}{I.8}, \thmref{Proposition}{II.3},
\thmref{Theorem}{II.7}(d), \thmref{Corollary}{III.13}.}

\begin{bgentry}{A.12}{limit; creation of limits}
A \textbf{cone} over a diagram $D$ in $\mathcal C$ is an object $L$ with an arrow to every object of $D$, such that each
triangle formed with an arrow of $D$ commutes. A \textbf{limit} of $D$ is a cone through which every other cone factors
by exactly one arrow. This categorical cone is unrelated to the geometric cones of \bgref{C.10}. Terminal objects,
products and pullbacks are the limits of the empty diagram, of two objects, and of two arrows with a common target. A
functor $U:\mathcal C\to\mathcal D$ \textbf{creates limits} if, for every diagram $D$ in $\mathcal C$ and every limit cone
of $U\circ D$, exactly one cone over $D$ is sent by $U$ to that limit cone, and this cone is a limit of $D$.
\end{bgentry}

The forgetful functor $\Diff_*\to\Diff$ creates limits (\bgref{A.14}). Concretely, when the underlying manifolds have a
pullback, the pullback of pointed manifolds is that pullback, pointed at the pair of base points. This is how
\thmref{Observation}{I.4}(b) computes products of methods as fibre products in $\Diff$.

\fbAusedin{\thmref{Observation}{I.4}(b). \textsf{Reading:} \citet[Ch.~V]{xmaclane1998categories};
\citet[Ch.~3]{xriehl2016category}.}

\begin{bgentry}{A.13}{adjunction}
A functor $F:\mathcal C\to\mathcal D$ is \textbf{left adjoint} to $G:\mathcal D\to\mathcal C$, and $G$ is \textbf{right
adjoint} to $F$, written $F\dashv G$, if there are bijections $\mathcal D(FX,Y)\cong\mathcal C(X,GY)$ natural in $X$ and
$Y$. Many forgetful functors are right adjoints, whose left adjoints add the forgotten structure freely.
\end{bgentry}

The forgetful functor $\Diff_*\to\Diff$ is right adjoint to $X\mapsto(X\sqcup\mathrm{pt},\mathrm{pt})$, which adds a
disjoint base point, since a pointed map out of $X\sqcup\mathrm{pt}$ is the same as an arbitrary map out of $X$ (a
manifold may have components of different dimensions). Right adjoints preserve limits (\bgref{A.14}), so a product of
methods, if it exists, is also a pullback in $\Diff$ and can be tested with unpointed maps.
\thmref{Observation}{I.4}(b) uses this to rule out $\mathrm{LoRA}_1\times\text{Frozen}$ on $1\times1$ weights. The axis
curves $t\mapsto(t,0)$ and $t\mapsto(0,t)$ in the cross would lift smoothly through the point over the origin, so the
differential of the projection from the product to the parameter plane $\R^2$ of $\mathrm{LoRA}_1$ would have rank $2$
there. That projection would then be a submersion near the point (\bgref{C.4}), and its image would contain an open set,
which the cross does not.

\fbAusedin{\thmref{Observation}{I.4}(b). \textsf{Reading:} \citet[Ch.~2 and Ch.~6]{xleinster2014basic}.}

\begin{bgfact}{A.14}{two standard facts about limits}
Let $\mathcal C$ be a category.
\begin{enumerate}[label=(\alph*)]
  \item For every object $X$ of $\mathcal C$, the forgetful functor $X/\mathcal C\to\mathcal C$ creates limits. In
    particular $\Diff_*\to\Diff$ does.
  \item A right adjoint preserves every limit that exists in its domain.
\end{enumerate}
\end{bgfact}

\paragraph{The idea behind (a) and (b)}
(a) The structure arrows $X\to D_j$ of a diagram in $X/\mathcal C$ form a cone in $\mathcal C$, which factors through
the limit by exactly one arrow, and that arrow is the structure map of the limit. (b) For $F\dashv G$,
\begin{equation*}
  \mathcal C(X,G\lim D)\cong\mathcal D(FX,\lim D)\cong\lim\mathcal D(FX,D_j)\cong\lim\mathcal C(X,GD_j),
\end{equation*}
naturally in $X$.

\medskip\noindent
\thmref{Observation}{I.4}(b) combines them. Products in $\Meth$ are pullbacks in $\Diff_*$, which by (a) exist whenever
the underlying pullback exists in $\Diff$, and by (b) and \bgref{A.13} only then. The transversality hypothesis of
\bgref{C.7} fails at the base point whenever $\lvert M\rvert+\lvert N\rvert<\dim\Theta$, with $\lvert M\rvert=\dim Q_M$
(\bgref{C.3}), so the products that exist, such as LoRA-XS, come from other structure (\bgref{A.11}).

\fbAusedin{\thmref{Observation}{I.4}(b), \S\ref{sec:slice-interval}. \textsf{Reading:}
\citet[Ch.~V]{xmaclane1998categories}; \citet[Ch.~3 and Ch.~4]{xriehl2016category}.}

\begin{bgentry}{A.15}{monomorphism; sub-object}
An arrow $m:A\to B$ is a \textbf{monomorphism} if $m\circ u=m\circ v$ implies $u=v$ for all arrows $u,v:T\to A$. A
\textbf{sub-object} of $B$ is a monomorphism into $B$, two of them being identified when they differ by an isomorphism
over $B$. Testing with maps out of one- and two-point manifolds shows that the monomorphisms of $\Diff$, and those of
$\Meth$, are exactly the arrows whose underlying map is injective. Such a map need not be an embedding, as
$t\mapsto t^3$ shows.
\end{bgentry}

Freezing coordinates at their base values gives an injective affine map, hence a sub-object, written $\hookrightarrow$ in
the atlas (rule A2). Examples are $\mathrm{FA}(A_0)\hookrightarrow\mathrm{LoRA}$ pointed at $(0,A_0)$, by
$B\mapsto(B,A_0)$, and MiCA \citep{rudiger2026mica} ${}\hookrightarrow{}$ LoRA pointed at $(B_0,0)$, by
$A\mapsto(B_0,A)$. These maps are embeddings, a stronger property than injectivity. The paper writes sub-object and never
monomorphism. In \thmref{Corollary}{III.13}, LoRA-FA is a sub-object of LoRA and so a different object, while random-$A$
LoRA and EVA \citep{paischer2024parameter} are the same map $\rho$ with two different base points in its fibre over
$\theta_0$.

\fbAusedin{\thmref{Definition}{I.2}, rule A2 in \S\ref{sec:atlas-rules}, \thmref{Corollary}{III.13}.}

\begin{bgentry}{A.16}{posets; meets; the expressivity order}
A \textbf{preorder} is a category with at most one arrow between any two objects; write $x\le y$ when there is one. It
is a \textbf{poset} if $x\le y\le x$ forces $x=y$. In a poset the product of $x$ and $y$ is their \textbf{meet}
$x\wedge y$, the greatest lower bound, and for subsets ordered by inclusion the meet is the intersection. Two elements are
\textbf{incomparable} if neither $x\le y$ nor $y\le x$, and $x$ is \textbf{maximal} if $x\le y$ implies $x=y$. Every
category has a \textbf{poset reflection}, in which $X\le Y$ when some arrow $X\to Y$ exists and $X,Y$ are identified when
arrows run both ways. A \textbf{Hasse diagram} draws a finite poset with $y$ above $x$ when $x<y$, joining them by an
edge when nothing lies strictly between.
\end{bgentry}

The poset reflection of $\Meth$, which we call the expressivity order, orders methods by simulation.
Figure~\ref{fig:graph} draws it for the core methods as a layered Hasse diagram, from the frozen model at the bottom to
full fine-tuning at the top. A meet of images always exists and need not be the image of a product. For injective $W_0$
and even $r\le n-2$, the image of HRA without Gram--Schmidt is the meet of LoRA's cone with the rotation orbit
$W_0\cdot\SO(n)$ (\bgref{D.11}), and HRA is not their product (\thmref{Theorem}{II.7}). Split and zero-init LoRA have
incomparable images (\thmref{Theorem}{II.2}(d)), so neither simulates the other. The frontier problem asks for the
maximal methods of budget at most $k$, which are automatically pairwise incomparable.

\fbAusedin{\thmref{Observation}{I.4}(c), \thmref{Proposition}{I.8}(b), \thmref{Theorem}{II.2}, \thmref{Theorem}{II.7},
\S\ref{sec:erlangen-hra}, Figure~\ref{fig:graph}, the open problems of Exposé~\ref{sec:analogy}. \textsf{Reading:}
\citet[Ch.~1 and Ch.~2]{xdavey2002introduction}.}

\subsubsection*{Base change and lifting}

\begin{bgentry}{A.17}{automorphism of a category; transport of structure}
An \textbf{automorphism} of a category $\mathcal C$ is a functor $F:\mathcal C\to\mathcal C$ with an inverse functor, so
that both composites are identity functors. It carries initial objects, products, isomorphisms and every other notion
defined from arrows alone to notions of the same kind. \textbf{Transport of structure} builds such a functor from an
invertible map of the underlying data.
\end{bgentry}

For $T\in\GL(\Theta)$ put $\tau_T(\theta)=\theta_0+T(\theta-\theta_0)$ and $T_*M:=(Q,q_0,\tau_T\circ\rho)$. An arrow
$h:M\to N$ is also an arrow $T_*M\to T_*N$, since $\tau_T\circ\rho_N\circ h=\tau_T\circ\rho_M$, so $T_*$ is a functor,
with inverse $(T^{-1})_*$. It keeps $d(M)$ and the first-order deficit (\bgref{C.3}) and the gauge group (\bgref{D.21}),
and moves $S_1$ to $TS_1$. It does not keep rank, which depends on the matrix shape of $\Theta$ that $T$ need not
respect. Sums of $k$ Kronecker products are $(\mathcal R^{-1})_*\mathrm{LoRA}_k$, LoRA on the matrix rearranged by Van
Loan and Pitsianis and carried back (\bgref{B.11}), and HiRA \citep{huang2025hira} is $(D_{W_0})_*\mathrm{LoRA}$ with
$D_{W_0}(X)=W_0\odot X$, when $W_0$ has no zero entries. The atlas records $M$ and $T_*M$ as one object in two frames,
with no arrow between them.

\fbAusedin{\thmref{Proposition}{II.13}, Exposé~\ref{sec:cut}, rule T in \S\ref{sec:atlas-rules},
\S\ref{sec:rosetta-reverse}.}

\begin{bgentry}{A.18}{lift; lifting problem; obstruction}
Given arrows $p:E\to B$ and $f:A\to B$, a \textbf{lift} of $f$ along $p$ is an arrow $g:A\to E$ with $p\circ g=f$. A
\textbf{lifting problem} asks whether a lift exists, and an \textbf{obstruction} is an invariant whose value shows that
none does. A morphism $h:M\to N$ of $\Meth$ is exactly a pointed smooth lift of $\rho_M$ along $\rho_N$.
\end{bgentry}

In $\Set$ a lift exists once $f(A)\subseteq p(E)$, by choosing preimages, so smoothness is the real constraint. For
$\rho_M(t)=t$ and $\rho_N(u)=u^3$ on $\R$ the images agree, and the only lift is the cube root, which is not
differentiable at $0$. A pointed smooth lift forces $S_1(M)\subseteq S_1(N)$ by the chain rule, so $S_1$ is a
first-order obstruction, and here $S_1(M)=\R\not\subseteq S_1(N)=0$. For $t^3$ against $u^5$ both $S_1$ vanish, yet the
only lift $t^{3/5}$ is not $C^1$, so that obstruction is invisible to $S_1$. Merging a method on an extension is again a
lifting problem, to find $\mu$ with $\Phi_f\circ\mu=\Phi_{\tilde f}\circ\rho$, a lift along the realisation $\Phi_f$
(\bgref{B.16}, \thmref{Definition}{VI.1}).

\fbAusedin{\thmref{Observation}{I.4}(c), \S\ref{sec:slice-interval}, \thmref{Observation}{I.6},
\thmref{Definition}{VI.1}, \S\ref{sec:merge-extensions}, Rosetta entry 5 (\S\ref{sec:rosetta-six}).}

\begin{bgentry}{A.19}{base change; pushforward $\phi_!$; the relative point of view}
A map of bases $\phi:S'\to S$ induces the \textbf{pushforward} $\phi_!:\mathcal C/S'\to\mathcal C/S$,
$(X,x)\mapsto(X,\phi\circ x)$, by postcomposition. When every object over $S$ has a pullback along $\phi$, it also
induces the \textbf{pullback} $\phi^*:\mathcal C/S\to\mathcal C/S'$, $(Y,y)\mapsto(Y\times_SS',\text{projection})$, and
then $\phi_!\dashv\phi^*$. In algebraic geometry, base change usually means $\phi^*$.
\end{bgentry}

\thmref{Definition}{V.1}(a) uses $\phi_!$ on pointed slices. A function-preserving pointed reparametrisation
$\phi:(\Theta',\theta_0')\to(\Theta,\theta_0)$, with $f\circ(\phi\times X)=f'$, gives
$\phi_!:\Meth(\Theta',\theta_0')\to\Meth(\Theta,\theta_0)$, $\rho\mapsto\phi\circ\rho$, for instance for a permutation
of hidden units. Whether $\phi_!$ carries LoRA to LoRA is the covariance question of \thmref{Observation}{II.15}. A
quantiser changes the function and induces no such functor, so QLoRA \citep{dettmers2023qlora} and ReLoRA
\citep{lialin2023relora} only borrow the name, and Exposé~\ref{sec:base} computes pointing defects and reachable sets for
them.

Grothendieck's relative point of view replaces a space $X$ by a morphism $X\to S$, an object over a base. Properties then
belong to the morphism, and the useful ones are those that survive pulling back along every change of base $S'\to S$. The
fibre over a point $s$ is itself the pullback along $s:\mathrm{pt}\to S$. The paper takes this stance in a modest form. A
method is a morphism over the base point $\theta_0$, and quantisation, restarts and merging are read as moves of that
base point.

\fbAusedin{Exposés~\ref{sec:thesis}--\ref{sec:slice}, Exposé~\ref{sec:thesis} (where the category theory earns its
keep), \thmref{Definition}{V.1}, Exposé~\ref{sec:base}, Rosetta entry 28 (\S\ref{sec:rosetta-six}), the seven borrowed
words of Exposé~\ref{sec:analogy}. \textsf{Reading:} \citet{xvakil2025rising}, on fibred products and base change.}

\begin{bgentry}{A.20}{constructions the paper names and leaves unused}
Coends, monads, sheaves, stacks, schemes, Grothendieck's descent theory (descent data, effective descent morphisms), the
Grothendieck construction $\int\Meth\to\mathbf{PNet}$ and coKleisli 2-cells appear in the paper only in lists of
constructions it leaves out (Exposé~\ref{sec:thesis}, where the category theory earns its keep, and
Exposé~\ref{sec:analogy}, on what the borrowed words mean). No result depends on them. The elementary descent of
\bgref{D.22}, factoring a map that is constant on orbits through the orbit space, is a different and much simpler notion,
and several results use it.
\end{bgentry}

For orientation only, the Grothendieck construction would assemble the slices of all pretrained models,
$\Meth(\Theta,\theta_0)$ for each $(\Theta,\theta_0)$, into one category of pairs (network, method), with a projection to the category $\mathbf{PNet}$ whose objects are pointed networks and whose arrows are
function-preserving reparametrisations. A monad is an endofunctor $T$ with natural transformations
$\id\Rightarrow T$ and $TT\Rightarrow T$ satisfying unit and associativity laws, and a sheaf assigns data to open sets so
that compatible local data glue to unique global data.

\fbAusedin{Exposé~\ref{sec:thesis} (where the category theory earns its keep), the seven borrowed words and what the
borrowed words mean in Exposé~\ref{sec:analogy}.}

\subsection*{Part B. Monoidal categories, string diagrams and lenses}
\phantomsection\label{app:bg-b}\addcontentsline{toc}{subsection}{Part B. Monoidal categories, string diagrams and lenses}

Part A compared methods through the arrows between them. Part B covers the algebra behind drawing a method as boxes and
wires, reading a rank bound off a cut of the drawing, and treating backpropagation as a pair of maps. Entries \bgref{B.1}
to \bgref{B.6} set up monoidal categories and their diagrams, \bgref{B.7} to \bgref{B.13} turn them into rank and
dimension counts for matrices built from small factors, and \bgref{B.14} to \bgref{B.17} cover sums of methods,
parametrised maps and lenses.

\subsubsection*{Monoidal categories}

\begin{bgentry}{B.1}{monoidal and symmetric monoidal categories}
A \textbf{monoidal category} is a category $\mathcal C$ with a functor $\otimes:\mathcal C\times\mathcal C\to\mathcal C$,
a unit object $I$, and natural isomorphisms $(A\otimes B)\otimes C\cong A\otimes(B\otimes C)$ and
$I\otimes A\cong A\cong A\otimes I$, the \textbf{coherence isomorphisms}, which satisfy the pentagon and triangle
equations. Mac Lane's coherence theorem then makes every diagram built from them commute, so brackets can be dropped. The
category is \textbf{symmetric} if it also has natural isomorphisms $\sigma_{A,B}:A\otimes B\to B\otimes A$ with
$\sigma_{B,A}\sigma_{A,B}=\id$ that are compatible with the associator (the hexagon equation). Examples are
$(\FinVect,\otimes,\R)$, $(\FinVect,\oplus,0)$ and $(\Diff,\times,\mathrm{pt})$.
\end{bgentry}

On methods, $M\oplus N$ runs the two parameter spaces side by side and adds their updates (\bgref{B.14}). This makes
$\Meth(\Theta,\theta_0)$ symmetric monoidal with the frozen model as unit. Its coherence isomorphisms are those of
$\times$ on pointed manifolds, and they lie over $\Theta$ because addition is associative and commutative.

\fbAusedin{\thmref{Observation}{I.7}, \S\ref{sec:slice-sums}, Exposé~\ref{sec:thesis} (where the category theory earns
its keep), \thmref{Theorem}{II.10}, \thmref{Theorem}{VI.3}. \textsf{Reading:} \citet[Ch.~VII and
Ch.~XI]{xmaclane1998categories}.}

\begin{bgentry}{B.2}{direct sum and tensor product in $\FinVect$}
The category $\FinVect$ has finite-dimensional real vector spaces as objects and linear maps as morphisms, and two
monoidal products.
The \textbf{direct sum} $V\oplus W$ is a \textbf{biproduct}, both the categorical product and the coproduct, with
$\dim(V\oplus W)=\dim V+\dim W$; on matrices $f\oplus g$ is block-diagonal. The \textbf{tensor product} $V\otimes W$ has
$\dim(V\otimes W)=\dim V\cdot\dim W$ and is not a categorical product; on matrices $f\otimes g$ is the Kronecker product
(\bgref{B.10}). Each bilinear map $V\times W\to U$ factors through a unique linear map $V\otimes W\to U$.
\end{bgentry}

Two boxes drawn side by side therefore mean a block-diagonal map in the activation diagrams of Exposé~\ref{sec:merge},
whose monoidal product is $\times=\oplus$, and a Kronecker product in a tensor network. Dimensions multiply along
$\otimes$, which is why the weight of a cut in \thmref{Theorem}{II.10} is a product of wire dimensions. LoRA's residual
branch $W_0+BA$ uses $\oplus$, since it sends one input to two branches and adds their outputs
(Figure~\ref{fig:bg-lora}).

\fbAusedin{\thmref{Theorem}{II.10}, Exposé~\ref{sec:cut}, \thmref{Theorem}{VI.3}, \thmref{Observation}{I.7}.}

\begin{bgentry}{B.3}{string diagrams}
A \textbf{string diagram} draws a morphism $f:A\to B$ of a monoidal category as a box with an input wire $A$ and an
output wire $B$. Composition $g\circ f$ joins boxes in series, $f\otimes g$ places them side by side, an identity is a
bare wire, the unit $I$ is drawn as no wire, and in a symmetric category $\sigma$ is a crossing. \citet{xjoyal1991geometry}
proved that the pictures are sound and complete. In a monoidal category, diagrams related by a planar deformation denote
the same morphism; in a symmetric one, diagrams with the same boxes joined in the same way do. Laws such as
$(g\otimes g')\circ(f\otimes f')=(g\circ f)\otimes(g'\circ f')$ can then be read off the picture.
\end{bgentry}

The plug of Exposé~\ref{sec:cut} (\S\ref{sec:cut-plug}) is a string diagram. LoRA is a box $A:\R^n\to\R^r$ followed by
$B:\R^r\to\R^m$, drawn beside $W_0$ (Figure~\ref{fig:bg-lora}, read left to right). The merge calculus moves an edit
along wires by equalities of maps, so a successful slide proves that the merged network computes the same function.

\fbAusedin{Exposé~\ref{sec:cut}, \thmref{Theorem}{II.10}, \S\ref{sec:cut-plug}, \S\ref{sec:merge-calculus} and the
analogy on string diagrams there, \thmref{Theorem}{VI.3}. \textsf{Reading:} \citet{xselinger2011survey}.}

\begin{bgentry}{B.4}{cartesian categories; copy and discard}
A monoidal category is \textbf{cartesian} when $\otimes$ is the categorical product $\times$ and $I$ is a terminal object
$1$. Every object then has a \textbf{copy} map $\Delta_X:X\to X\times X$, $x\mapsto(x,x)$, and a \textbf{discard} map
${!}_X:X\to1$, and copy is natural, $(f\times f)\circ\Delta_X=\Delta_Y\circ f$ for every $f:X\to Y$. Conversely, by a
theorem of \citet{xfox1976coalgebras}, a symmetric monoidal category is cartesian exactly when every object carries a
commutative comonoid, a copy and a discard map obeying coassociativity, counit and commutativity laws (\bgref{B.6}),
compatible with $\otimes$ and natural in all morphisms. $\Set$, $\Diff$ and $(\FinVect,\oplus)$ are cartesian.
$(\FinVect,\otimes)$ is not, since $x\mapsto x\otimes x$ is not linear.
\end{bgentry}

Naturality of copy is rule R7 of the merge calculus. A linear edit $E$ on a wire read by $W_Q$, $W_K$ and $W_V$ equals
$E$ applied on each of the three branches, so it is absorbed as $W_QE$, $W_KE$, $W_VE$. The paper builds $\Para(\mathcal C)$
(\bgref{B.16}) over a cartesian $\mathcal C$. The linear copy of \bgref{B.6} copies basis vectors only and is a different
map.

\fbAusedin{\thmref{Definition}{I.1}, \S\ref{sec:merge-calculus} and the analogy on string diagrams there,
\thmref{Theorem}{VI.3}.}

\begin{bgentry}{B.5}{compact closed categories; cups, caps and wire-bending}
A symmetric monoidal category is \textbf{compact closed} if every object $A$ has a dual $A^*$ with a \textbf{cup}
$\eta_A:I\to A^*\otimes A$ and a \textbf{cap} $\varepsilon_A:A\otimes A^*\to I$ satisfying the snake equations
\begin{equation*}
  (\varepsilon_A\otimes\id_A)(\id_A\otimes\eta_A)=\id_A,\qquad(\id_{A^*}\otimes\varepsilon_A)(\eta_A\otimes\id_{A^*})=\id_{A^*},
\end{equation*}
with the unitors $I\otimes A\cong A\cong A\otimes I$ of \bgref{B.1} suppressed. Drawn as bent wires, cups and caps turn
an input wire into an output wire and back. In $\FinVect$ the cap is evaluation, the cup is
$1\mapsto\sum_ie_i^*\otimes e_i$, and bending gives $\mathrm{Hom}(V,U)\cong U\otimes V^*$. With a fixed basis, a box
slid around a cup or cap comes out as its transpose.
\end{bgentry}

Bending the input wire of an $m\times n$ matrix gives its vectorisation in $\R^m\otimes\R^n$, and the Van
Loan--Pitsianis rearrangement (\bgref{B.11}) is a composite of bends and crossings. Rule R5 of the merge calculus,
$\langle Mq,k\rangle=\langle q,M\T k\rangle$, slides $M$ around the cap that computes an attention score. The cut-rank
bound uses only $\otimes$ and crossings.

\fbAusedin{\thmref{Theorem}{II.11}, \thmref{Theorem}{VI.3}, Exposé~\ref{sec:cut}. \textsf{Reading:}
\citet{xcoecke2017picturing}.}

\begin{bgentry}{B.6}{Frobenius algebras and spiders}
A \textbf{Frobenius algebra} on an object $V$ is a monoid $\mu:V\otimes V\to V$, $\eta:I\to V$ and a comonoid
$\delta:V\to V\otimes V$, $\varepsilon:V\to I$ obeying the \textbf{Frobenius law}
\begin{equation*}
  (\mu\otimes\id)(\id\otimes\delta)=\delta\mu=(\id\otimes\mu)(\delta\otimes\id).
\end{equation*}
A basis $e_1,\dots,e_n$ of $\R^n$ gives one, with \emph{copy} $\delta(e_i)=e_i\otimes e_i$, \emph{merge} (the
multiplication; unrelated to merging an adapter into the weights) $\mu(e_i\otimes e_j)=e_i$ if $i=j$ and $0$ otherwise,
$\eta=\sum_ie_i$ and $\varepsilon(e_i)=1$. It is commutative and special ($\mu\delta=\id$). For such algebras the
\textbf{spider theorem} says that every connected diagram of $\mu,\eta,\delta,\varepsilon$ with $k\ge1$ inputs and $l$
outputs equals one \textbf{spider}, which sends $e_i^{\otimes k}$ to $e_i^{\otimes l}$ and every other basis tensor to
$0$.
\end{bgentry}

Merge multiplies coordinates, $\mu(x\otimes y)=x\odot y$. So for matrices $X,Y:\R^n\to\R^m$ the Hadamard product is the
spider composite
\begin{equation*}
  X\odot Y=\mu_m\circ(X\otimes Y)\circ\delta_n.
\end{equation*}
LoHa \citep{hyeonwoo2021fedpara} is a copy spider on the input feeding $A_1$ and $A_2$, then $B_1$ and $B_2$, then a merge
spider on the output, and in \thmref{Theorem}{II.10} a spider split by the cut counts its dimension once. The copy and
$\odot$ nodes of Exposé~\ref{sec:merge} live in a cartesian category (\bgref{B.4}), where copy is $x\mapsto(x,x)$;
together they fail the Frobenius law, so there ``spider'' is only a name.

\fbAusedin{Exposé~\ref{sec:cut}, \thmref{Theorem}{II.10}, \thmref{Proposition}{II.12}, the analogy on string diagrams in
\S\ref{sec:merge-calculus}. \textsf{Reading:} \citet{xcoecke2013new}.}

\subsubsection*{Tensor networks and rank}

\begin{bgentry}{B.7}{tensor networks, contraction, bond dimension}
A \textbf{tensor network} $\Gamma$ is a string diagram in $(\FinVect,\otimes)$, possibly with spiders. Each box is a
tensor and each wire a finite-dimensional space. A wire joining two boxes means \textbf{contraction}, a sum over the index
the wire carries. The open wires (legs) are the inputs $V$ and outputs $U$, and contracting every internal wire gives the
value $\llbracket\Gamma\rrbracket:V\to U$. The dimension of an internal wire is its \textbf{bond dimension}. One value has
many presentations, and quantities read off a network, such as its cuts, depend on the presentation.
\end{bgentry}

LoRA's update is the two-box network $B\circ A$ with one bond of dimension $r$, and $\Delta W=BA$ contracts it,
$(BA)_{ij}=\sum_{k\le r}B_{ik}A_{kj}$. HiRA factors $W_0$ through a bond of dimension $\operatorname{rk}W_0$, while MoRA
\citep{jiang2024mora}, C3A \citep{chen2024parameter} and MPO adapters have no narrow cut between inputs and outputs
(Table~\ref{tab:cutbounds}, after \thmref{Theorem}{II.10}).

\fbAusedin{\thmref{Theorem}{II.10}, Exposé~\ref{sec:cut}, \S\ref{sec:cut-plug}, Exposé~\ref{sec:thesis} (where the
category theory earns its keep), the known results credited in Exposé~\ref{sec:related}. \textsf{Reading:}
\citet{xorus2014practical}.}

\begin{bgfact}{B.8}{the min-cut bound on rank}
Let $\Gamma$ be a tensor network with input legs $V$ and output legs $U$, with every wire attached to a box (put an
identity box on any bare strand). Split the boxes into an input side $\mathcal I$ and an output side $\mathcal O$. The
\textbf{weight} $w$ of the cut is the product of the dimensions of the wires joining the two sides and of the legs attached
on the wrong side (input legs at $\mathcal O$, output legs at $\mathcal I$); a spider with legs on both sides contributes
its dimension once. Then
\begin{equation*}
  \rk\llbracket\Gamma\rrbracket\le\min_{\text{cuts}}w.
\end{equation*}
For sums of networks, $\rk(X+Y)\le\rk X+\rk Y$.
\end{bgfact}

\paragraph{The idea of the proof}
Contract each side separately. The value becomes a map into the tensor product of the crossing wires and wrong-side legs
followed by a map out of it, and that space has dimension $w$.

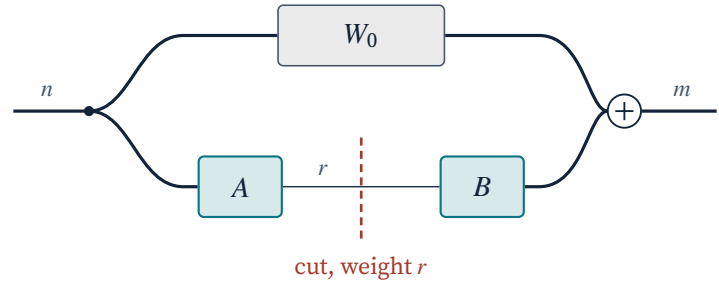
\begin{figure}[tbp]
\centering
\begin{tikzpicture}[x=1cm,y=1cm,
  box/.style={rectangle, rounded corners=2pt, minimum height=8mm, line width=0.6pt, text=ink},
  frozen/.style={box, draw=inkgray, fill=headergray, minimum width=22mm},
  train/.style={box, draw=tide, fill=tide!16, line width=0.8pt, minimum width=11mm},
  wire/.style={draw=ink, line width=1.2pt},
  thinwire/.style={draw=ink, line width=0.5pt},
  lab/.style={font=\footnotesize, text=inkgray}]
  \draw[wire] (0,0) -- (1.0,0);
  \fill[ink] (1.0,0) circle (1.9pt);
  \node[frozen] (W0) at (4.6,1.0) {$W_0$};
  \draw[wire] (1.0,0) .. controls (1.55,0) and (1.65,1.0) .. (2.25,1.0) -- (W0.west);
  \draw[wire] (W0.east) -- (6.95,1.0) .. controls (7.55,1.0) and (7.6,0) .. (7.86,0);
  \node[train] (A) at (3.0,-1.0) {$A$};
  \node[train] (B) at (6.2,-1.0) {$B$};
  \draw[wire] (1.0,0) .. controls (1.55,0) and (1.65,-1.0) .. (2.25,-1.0) -- (A.west);
  \draw[thinwire] (A.east) -- (B.west);
  \draw[wire] (B.east) -- (6.95,-1.0) .. controls (7.55,-1.0) and (7.6,0) .. (7.86,0);
  \node[circle, draw=ink, fill=white, line width=0.7pt, minimum size=4.4mm, inner sep=0pt] (plus) at (8.08,0) {};
  \draw[ink, line width=0.7pt] (7.96,0) -- (8.20,0) (8.08,-0.12) -- (8.08,0.12);
  \draw[wire] (plus.east) -- (9.3,0);
  \node[lab, anchor=south] at (0.45,0.06) {$n$};
  \node[lab, anchor=south] at (8.85,0.06) {$m$};
  \node[lab, anchor=south] at (4.1,-0.98) {$r$};
  \draw[seal, line width=1pt, dash pattern=on 3pt off 2pt] (4.6,-0.35) -- (4.6,-1.75);
  \node[font=\footnotesize, text=seal, anchor=north] at (4.6,-1.78) {cut, weight $r$};
\end{tikzpicture}
\caption{LoRA as a string diagram, read left to right, with the scale $s$ absorbed into $B$. The fork sends the input to
  both branches and the $+$ node adds their outputs. These are the copy and sum maps of the biproduct $\oplus$
  (\bgref{B.2}), so the diagram says $W=W_0+BA$. The frozen $W_0$ is grey, and the trainable $A\in\R^{r\times n}$ and
  $B\in\R^{m\times r}$ are tinted. The dashed cut separates $A$ from $B$ and crosses one wire of dimension $r$, so
  $\rk(W-W_0)=\rk BA\le r$. For $W$ itself, subadditivity over the two branches gives only $\rk W\le\rk W_0+r$.}
\label{fig:bg-lora}
\end{figure}

\medskip\noindent
For LoRA the cut between $A$ and $B$ crosses one wire, so $\rk BA\le r$ (Figure~\ref{fig:bg-lora}). For LoHa the cut
with $A_1,A_2$ on the input side crosses two bonds and gives $r_1r_2$. The bound depends on the presentation. HRA's
product of $r$ reflections expands into $2^r-1$ terms of rank at most one, while the telescoping sum
\begin{equation*}
  R-I=\sum_{k=1}^r(H_k-I)H_{k+1}\cdots H_r
\end{equation*}
gives $r$.

\fbAusedin{\thmref{Theorem}{II.10}, Exposé~\ref{sec:cut}, \S\ref{sec:cut-plug}, \thmref{Proposition}{II.12}, the known
results credited in Exposé~\ref{sec:related}. \textsf{Reading:} \citet{kolda2009tensor}.}

\begin{bgentry}{B.9}{tensor trains, Tucker and MPO; gauge on bonds}
A \textbf{tensor train} (TT) writes a tensor with legs of dimensions $n_1,\dots,n_d$ as a chain of cores
$G_k\in\R^{r_{k-1}\times n_k\times r_k}$, $r_0=r_d=1$, joined by bonds of dimensions $r_1,\dots,r_{d-1}$. An
\textbf{MPO} is the same chain with one input and one output leg on each core. \textbf{Tucker} joins a core
$C\in\R^{r_1\times\cdots\times r_d}$ to factor matrices $U_k\in\R^{n_k\times r_k}$. Inserting $g^{-1}g=\id$,
$g\in\GL_{r_e}$, on a bond $e$ and absorbing each factor into its neighbouring box leaves the value unchanged, so the
gauge (\bgref{D.21}) contains $\prod_e\GL_{r_e}$. When every $r_e$ is the rank of the matching unfolding (minimal ranks),
the tensors of these ranks form a manifold of dimension $\sum(\text{core sizes})-\sum_er_e^2$.
\end{bgentry}

LoRA is the two-core case, $B$ and $A$ joined by one bond, with gauge $(B,A)\mapsto(Bg^{-1},gA)$ and, for
$r\le\min(m,n)$, image dimension $r(m+n)-r^2$. TT and Tucker adapters have $d(M)=\lvert M\rvert-\sum_er_e^2$ at minimal
ranks (\thmref{Proposition}{II.12}), and under gradient flow each bond carries a conserved matrix
(\S\ref{sec:fibres-charges}).

\fbAusedin{\thmref{Proposition}{II.12}, Exposé~\ref{sec:cut}, \S\ref{sec:fibres-charges}. \textsf{Reading:}
\citet{oseledets2011tensor}; \citet{xholtz2012manifolds}; \citet{xkoch2010dynamical}.}

\begin{bgentry}{B.10}{Kronecker product, vec, tensor product of subspaces}
For $C\in\R^{m_1\times n_1}$ and $D\in\R^{m_2\times n_2}$, the \textbf{Kronecker product}
$C\otimes D\in\R^{m_1m_2\times n_1n_2}$ is the block matrix $[c_{ij}D]$, the matrix of $C\otimes D$ in the product basis.
It obeys the mixed-product rule $(C\otimes D)(E\otimes F)=CE\otimes DF$, and $\rk(C\otimes D)=\rk C\cdot\rk D$. The map
$u\otimes v\mapsto uv\T$ identifies $\R^m\otimes\R^n$ with $\R^{m\times n}$. For subspaces $U\subseteq\R^m$ and
$V\subseteq\R^n$, $U\otimes V$ is then the space of matrices with column space in $U$ and row space in $V$, of dimension
$\dim U\cdot\dim V$.
\end{bgentry}

LoKr's \citep{yeh2023navigating} update is $C\otimes BA$ with $C$ ranging over all $m_1\times n_1$ matrices,
$B\in\R^{m_2\times r'}$ and $A\in\R^{r'\times n_2}$. For inner rank $r'\le\min(m_2,n_2)$ its rank
$\operatorname{rk}C\cdot\operatorname{rk}(BA)$ is at most $\min(m_1,n_1)\,r'$, and this is attained. So
$\Im\mathrm{LoKr}$ lies inside the image $W_0+\Mr$ of zero-init $\mathrm{LoRA}_r$ exactly when $r\ge\min(m_1,n_1)\,r'$
(\thmref{Theorem}{II.11}(e)). At a LoRA point $(B,A)$, the directions reached by moving one factor alone overlap in
\begin{equation*}
  \{XA\}\cap\{BY\}=\operatorname{col}B\otimes\operatorname{row}A,
\end{equation*}
of dimension $\operatorname{rk}A\cdot\operatorname{rk}B$ (\thmref{Proposition}{II.3}).

\fbAusedin{\thmref{Theorem}{II.11}, Exposé~\ref{sec:cut}, \thmref{Proposition}{II.3}, \thmref{Theorem}{II.2}, the
\hyperref[ex:loha]{LoHa\,/\,LoKr example} of Exposé~\ref{sec:examples}. \textsf{Reading:}
\citet[Ch.~4]{xhorn1991topics}.}

\begin{bgfact}{B.11}{Van Loan--Pitsianis rearrangement; Kronecker rank}
Let $m=m_1m_2$ and $n=n_1n_2$, and cut $\Delta\in\R^{m\times n}$ into an $m_1\times n_1$ grid of $m_2\times n_2$ blocks
$\Delta_{ij}$. The \textbf{rearrangement} $\mathcal R(\Delta)\in\R^{m_1n_1\times m_2n_2}$ has the rows
$\vect(\Delta_{ij})\T$, ordered with $i$ running fastest. Then
\begin{equation*}
  \mathcal R(C\otimes D)=\vect(C)\,\vect(D)\T,
\end{equation*}
and $\mathcal R$ is a linear bijection that permutes entries, hence a Frobenius isometry. So the \textbf{Kronecker rank}
$\operatorname{rk}_\otimes\Delta$, the least $k$ with $\Delta=\sum_{i\le k}C_i\otimes D_i$, equals
$\rk\mathcal R(\Delta)$, and the nearest sum of $k$ Kronecker products in Frobenius norm comes from the rank-$k$
truncated SVD of $\mathcal R(\Delta)$ (Van Loan and Pitsianis~\citeyear{loan1993approximation}).
\end{bgfact}

\paragraph{The idea of the proof}
Block $(i,j)$ of $C\otimes D$ is $c_{ij}D$, so its row in $\mathcal R$ is $c_{ij}\,\vect(D)\T$. Linearity extends this
to sums, and Eckart--Young applies to $\mathcal R(\Delta)$ because $\mathcal R$ preserves the Frobenius norm.

\medskip\noindent
In string-diagram terms $\mathcal R$ bends and crosses wires, regrouping the four legs of $\Delta$ from
$U_1U_2\mid V_1V_2$ to $U_1V_1\mid U_2V_2$. A sum of $k$ Kronecker products is therefore $\mathrm{LoRA}_k$ transported by
$\mathcal R^{-1}$ (\thmref{Proposition}{II.13}), with rank taken across another cut. For $N=s^2$, $I_N=I_s\otimes I_s$
has Kronecker rank $1$ and rank $N$. Figure~\ref{fig:cut} shows the singular values across the ordinary cut, this
Kronecker cut and a third one, for a random rank-one update, the identity and a Kronecker product $C\otimes D$. In this
paper a \emph{Kronecker sum} means a sum of Kronecker products; matrix analysis uses the same name for
$C\otimes I+I\otimes D$.

\fbAusedin{\thmref{Theorem}{II.11}, \thmref{Proposition}{II.13}, Figure~\ref{fig:cut}, \thmref{Prediction}{X.5}, the
known results and key references of Exposé~\ref{sec:related}. \textsf{Reading:} \citet{xvanloan2000ubiquitous}.}

\begin{bgentry}{B.12}{Schmidt and operator-Schmidt decompositions}
Every $u\in\R^{m_1}\otimes\R^{m_2}$ has a \textbf{Schmidt decomposition} $u=\sum_{i\le\rho}\sigma_ix_i\otimes y_i$ with
$\sigma_i>0$ and orthonormal families $(x_i)$, $(y_i)$. It is the SVD of the $m_1\times m_2$ reshape of $u$
($x\otimes y\mapsto xy\T$), and the \textbf{Schmidt rank} $\rho_u$ is the rank of that reshape, at most
$\min(m_1,m_2)$. The SVD of $\mathcal R(\Delta)$ likewise gives the \textbf{operator-Schmidt decomposition}
$\Delta=\sum_{k\le K}s_kC_k\otimes D_k$, with $s_k>0$, Frobenius-orthonormal families $(C_k)$ and $(D_k)$, and
$K=\operatorname{rk}_\otimes\Delta$.
\end{bgentry}

If $u=\sum_i\sigma_ix_i\otimes y_i$ and $v=\sum_j\tau_jz_j\otimes w_j$ have Schmidt ranks $\rho_u$ and $\rho_v$, the
mixed-product rule gives
\begin{equation*}
  uv\T=\sum_{i,j}\sigma_i\tau_j\,(x_iz_j\T)\otimes(y_iw_j\T),
\end{equation*}
an operator-Schmidt decomposition with $\rho_u\rho_v$ terms, so $\operatorname{rk}_\otimes(uv\T)=\rho_u\rho_v$. A generic
rank-one update, which $\mathrm{LoRA}_1$ reaches, thus needs $\kappa_1=\min(m_1,m_2)\min(n_1,n_2)$ Kronecker terms
(\thmref{Theorem}{II.11}(c)).

\fbAusedin{\thmref{Theorem}{II.11}, Figure~\ref{fig:cut}, \thmref{Prediction}{X.5}, the known results credited in
Exposé~\ref{sec:related}. \textsf{Reading:} \citet[\S2.5]{xnielsen2000quantum}.}

\begin{bgfact}{B.13}{rank of a Hadamard product; face-splitting product}
For $X,Y\in\R^{m\times n}$, $\rk(X\odot Y)\le\rk X\cdot\rk Y$. Explicitly, if $X=B_1A_1$ and $Y=B_2A_2$ with inner
dimensions $r_1$ and $r_2$, then
\begin{equation*}
  X\odot Y=(B_1\bullet B_2)\,(A_1\T\bullet A_2\T)\T,
\end{equation*}
where the \textbf{face-splitting product} $B_1\bullet B_2\in\R^{m\times r_1r_2}$ has as row $i$ the Kronecker product of
row $i$ of $B_1$ and row $i$ of $B_2$. In particular
\begin{equation*}
  W_0\odot(BA)=\sum_{k\le r}\diag(b_k)\,W_0\,\diag(a_k),
\end{equation*}
summed over the columns $b_k$ of $B$ and the rows $a_k$ of $A$, has rank at most $r\rk W_0$
\citep[Ch.~5]{xhorn1991topics}.
\end{bgfact}

\paragraph{The idea of the proof}
$X\odot Y$ is the submatrix of $X\otimes Y$ on the rows $(i,i)$ and the columns $(j,j)$. In pictures it is the spider
composite $\mu\circ(X\otimes Y)\circ\delta$ of \bgref{B.6}, which factors through the two inner wires.

\medskip\noindent
$\mathrm{LoHa}_{r_1,r_2}$ therefore has rank at most $r_1r_2$, and $h=\big(B_1\bullet B_2,(A_1\T\bullet A_2\T)\T\big)$ is
the atlas arrow $\mathrm{LoHa}_{r_1,r_2}\to\mathrm{LoRA}_{r_1r_2}$. HiRA's update $W_0\odot BA$ has rank at most
$r\rk W_0$, a bound that allows full rank when $W_0$ has full rank (\thmref{Proposition}{II.9}(a)).

\fbAusedin{\thmref{Proposition}{II.12}, \thmref{Theorem}{II.10}, \thmref{Proposition}{II.9}, \S\ref{sec:atlas-arrows},
Rosetta entries 6 and 11 (\S\ref{sec:rosetta-six}).}

\subsubsection*{Sums, Para and lenses}

\begin{bgentry}{B.14}{sums of methods; Minkowski sums, differences and translates}
For subsets $S,T$ of a vector space, the \textbf{Minkowski sum} is $S+T=\{s+t:s\in S,\ t\in T\}$, the
\textbf{Minkowski difference} is $S-T=\{s-t:s\in S,\ t\in T\}$, a \textbf{translate} of $S$ is $v+S=\{v+s:s\in S\}$,
and $C^{+K}$ is the sum of $K$ copies of $C$. A sum is the image of $S\times T$ under addition, so for semialgebraic sets
$\dim(S+T)\le\dim S+\dim T$ (\bgref{C.18}), and likewise for differences. If $0\in C$, the additive monoid
(\bgref{D.10}) generated by $C$ is $\operatorname{Mon}(C)=\bigcup_KC^{+K}$, and if moreover $\R C=C$ it is
$\spanop C$. Writing $\rho_M=\theta_0+\delta_M$, the sum of methods
\begin{equation*}
  M\oplus N=\big(Q_M\times Q_N,\,(q_0,q_0'),\,\theta_0+\delta_M+\delta_N\big)
\end{equation*}
has image $\theta_0+(\Im\delta_M+\Im\delta_N)$, and $\oplus$ is a symmetric monoidal structure with the frozen model as
unit (\bgref{B.1}).
\end{bgentry}

Concatenation $([B_1\ B_2],[A_1;A_2])$ is an isomorphism $\mathrm{LoRA}_r\oplus\mathrm{LoRA}_s\cong\mathrm{LoRA}_{r+s}$
of unpointed objects, with the $A$-blocks rescaled to the target scale when the scales differ, as under $\alpha/r$
scaling (\thmref{Observation}{I.7}). It is also a pointed isomorphism onto $\mathrm{LoRA}_{r+s}$ pointed at the
concatenated initialisation. RoSA \citep{nikdan2024rosa} and GraLoRA \citep{jung2025gralora} are $\oplus$-words. The
categorical product of \thmref{Observation}{I.4}, when it exists, has image $\Im M\cap\Im N$ (\bgref{A.10}). Merging and
restarting $K$ times with a fixed shape $C$ reaches $\theta_0+C^{+K}$. Since
$\Mr^{+K}=\mathcal M_{\le\min(Kr,m,n)}$, ReLoRA reaches every matrix after $\lceil\min(m,n)/r\rceil$ cycles
(\thmref{Theorem}{V.3}).

Re-pointing an additive method $\rho=\theta_b+\delta$ with shape $C=\Im\delta$ at a parameter $q$, so that
$\theta_b=\theta_0-\delta(q)$, gives the translate $\theta_0+(C-\delta(q))$ as image, and all exact pointings together
reach $\theta_0+(C-C)$. For LoRA, $\Mr$ is closed under negation and $\Mr+\Mr=\mathcal M_{\le\min(2r,m,n)}$ by
splitting an SVD (\bgref{C.12}), so $C-C=\mathcal M_{\le\min(2r,m,n)}$, which is why no exact pointing reaches an update
of rank above $2r$ (\thmref{Theorem}{II.1}).

\fbAusedin{\thmref{Observation}{I.7}, \S\ref{sec:slice-sums}, \thmref{Theorem}{II.1}, \thmref{Proposition}{II.13},
\thmref{Theorem}{V.3}, \S\ref{sec:base-rebase}, \S\ref{sec:rosetta-reverse}.}

\begin{bgentry}{B.15}{strong monoidal functors; $(\N,+)$ and the permutation groupoid}
A \textbf{strong monoidal functor} $F:(\mathcal C,\otimes,I)\to(\mathcal D,\otimes,J)$ is a functor with isomorphisms
$\phi_{A,B}:FA\otimes FB\to F(A\otimes B)$, natural in $A$ and $B$, and $\phi_0:J\to FI$, compatible with associators and
unitors. It is \textbf{symmetric} if also $F(\sigma_{A,B})\circ\phi_{A,B}=\phi_{B,A}\circ\sigma_{FA,FB}$. The
\textbf{discrete monoidal category} $(\N,+)$ has the natural numbers as objects, only identity morphisms, $r\otimes s=r+s$
and identity symmetry. The \textbf{permutation groupoid} $\mathbb P$ has the same objects, the symmetric group $S_r$ as
the automorphisms of $r$ and no other morphisms, with $\otimes$ given by $+$ and block sums of permutations, and the block
transposition in $S_{r+s}$ as symmetry.
\end{bgentry}

Concatenation (\bgref{B.14}) makes $r\mapsto\mathrm{LoRA}_r$ strong monoidal from $(\N,+)$ into unpointed methods. It is
not symmetric there, because $\phi_{s,r}\circ\sigma$ lists the blocks in the other order. On $\mathbb P$ the block
transposition acts on $\mathrm{LoRA}_{r+s}$ by permuting inner channels, $(B,A)\mapsto(BP^{-1},PA)$, and the symmetry
square commutes. Pointed objects behave differently. If $F(r)$ and $F(2r)$ are standard $\mathrm T_A$ pointings
($B_0=0$ and $A_0$ of full row rank, \bgref{C.19}), which needs $2r<\min(m,n)$, then $F(r)\oplus F(r)\not\cong F(2r)$ as
pointed objects, since $\dim S_1$ is $mr$ against $2mr$ (\thmref{Observation}{I.7}).

\fbAusedin{\thmref{Observation}{I.7}, the referee record in Exposé~\ref{sec:analogy}. \textsf{Reading:}
\citet[Ch.~XI]{xmaclane1998categories}.}

\begin{bgentry}{B.16}{bicategories and the Para construction}
A \textbf{bicategory} has objects, \textbf{1-cells} between objects and \textbf{2-cells} between parallel 1-cells.
2-cells compose vertically and horizontally, and composition of 1-cells is associative and unital up to coherent
invertible 2-cells. For a symmetric monoidal category $\mathcal C$, here always cartesian, the bicategory
$\Para(\mathcal C)$ has the objects of $\mathcal C$. A 1-cell $X\to Y$ is a pair $(P,\,f:P\times X\to Y)$, and
$(P',g):Y\to Z$ composes with it to $\big(P'\times P,\ (p',p,x)\mapsto g(p',f(p,x))\big)$. A 2-cell
$(P,f)\Rightarrow(Q,g)$ is a \textbf{reparametrisation} $r:Q\to P$ with $g=f\circ(r\times\id_X)$. A \textbf{2-functor}
(pseudofunctor) between bicategories sends objects, 1-cells and 2-cells to objects, 1-cells and 2-cells, and preserves
both compositions up to coherent invertible 2-cells.
\end{bgentry}

A linear layer is the 1-cell $\big(\R^{m\times n},\,(W,x)\mapsto Wx\big):\R^n\to\R^m$, and stacking a second layer
$\R^m\to\R^k$ composes 1-cells, with parameter space $\R^{k\times m}\times\R^{m\times n}$. A pretrained model is a
pointed 1-cell $F=(\Theta,\theta_0,f)$ of $\Para(\Diff)$, and its realisation $\Phi_f:\Theta\to C^\infty(X,Y)$,
$\theta\mapsto f(\theta,-)$, records the function at each weight. LoRA is a 2-cell
$F\Rightarrow(Q,f\circ(\rho\times\id_X))$, the network with $(B,A)$ in its parameter slot, and a simulation is a 2-cell
between two such cells (\thmref{Remark}{I.3}). An adapter extension is a 2-cell into $F$, and merging it is a lift
through $\Phi_f$ (\thmref{Definition}{VI.1}). Composition is associative only up to
$(P''\times P')\times P\cong P''\times(P'\times P)$, hence a bicategory; over a strict monoidal category it is a
2-category.

\fbAusedin{\thmref{Definition}{I.1}, \S\ref{sec:slice-para}, \thmref{Remark}{I.3}, \thmref{Definition}{VI.1},
\S\ref{sec:lens-functor}, the key references of Exposé~\ref{sec:related}. \textsf{Reading:}
\citet{xleinster1998basic}.}

\begin{bgentry}{B.17}{lenses, reverse derivatives and $\Para(\mathsf R)$}
In a cartesian category, a \textbf{lens} $(A,A')\to(B,B')$ is a pair of maps, a forward part $f:A\to B$ and a backward
part $f^\sharp:A\times B'\to A'$. Lenses compose by
\begin{equation*}
  (g,g^\sharp)\circ(f,f^\sharp)=\big(g\circ f,\ (a,c')\mapsto f^\sharp(a,g^\sharp(f(a),c'))\big)
\end{equation*}
and form the category $\Lens(\mathcal C)$. For smooth maps between open subsets of Euclidean spaces, the \textbf{reverse
derivative} is
\begin{equation*}
  \mathsf R[f](x,y)=J_f(x)\T y.
\end{equation*}
The chain rule $J_{g\circ f}(x)\T=J_f(x)\T J_g(f(x))\T$ says that
$f\mapsto(f,\mathsf R[f])$ is a functor into $\Lens$. Applied to parametrised maps it gives the 2-functor
$\Para(\mathsf R):\Para(\mathcal C)\to\Para(\Lens(\mathcal C))$ of \citet{cruttwell2021categorical}, which is
backpropagation.
\end{bgentry}

LoRA's forward part is $\rho(B,A)=W_0+sBA$, and its backward part pulls the weight gradient $G$ back to
\begin{equation*}
  \mathsf R[\rho]\big((B,A),G\big)=\big(sGA\T,\,sB\T G\big).
\end{equation*}
GaLore \citep{zhao2024galore} keeps the forward part $\id_\Theta$ and uses the compression $G\mapsto P\T G$ with anchor
$P$, a lens whose backward part is not the reverse derivative of its forward part (\thmref{Definition}{IV.1}). For linear
$\rho(C)=W_k+\mathcal LC$, $\mathsf R[\rho](C,G)=\mathcal L^*G$ does not depend on $C$, which is why the two readings
agree at degree one (\thmref{Proposition}{IV.2}).

\fbAusedin{Exposé~\ref{sec:lens}, \S\ref{sec:lens-functor}, \thmref{Definition}{IV.1}, \thmref{Proposition}{IV.2},
Rosetta entry 25 (\S\ref{sec:rosetta-six}), Exposé~\ref{sec:thesis} (where the category theory earns its keep), the
seven borrowed words of Exposé~\ref{sec:analogy}. \textsf{Reading:} \citet{cruttwell2021categorical}.}

\providecommand{\fbAusedin}[1]{\par\smallskip\noindent{\small\color{inkgray}{\sffamily\bfseries\color{tide}Used in.}\enspace #1\par}\medskip}

\subsection*{Part C. Manifolds, germs and cones}
\phantomsection\label{app:bg-c}\addcontentsline{toc}{subsection}{Part C. Manifolds, germs and cones}

Parts A and B used only the arrows of a method and its wiring. Part C turns to its geometry. A PEFT method in this paper
is a smooth map $\rho:(Q,q_0)\to(\Theta,\theta_0)$, and its expressivity is read off the set $\Im M=\rho(Q)$ near
$\theta_0$. This part collects the geometry behind that reading, from manifolds, differentials and rank for the map to
germs, cones, singular points and semialgebraic dimension for the image, which is often not a manifold.

\subsubsection*{Manifolds and smooth maps}

\begin{bgentry}{C.1}{smooth manifolds and smooth maps}
A \textbf{smooth manifold} of dimension $k$ is a space covered by \textbf{charts}, bijections from open pieces of it
onto open subsets of $\R^k$, whose changes of chart are smooth. Every manifold in this paper is an open subset of a
vector space or an embedded submanifold of one (\bgref{C.4}), or is diffeomorphic to one, as quotients (\bgref{D.22})
and the Grassmannian (\bgref{C.15}) are. A \textbf{diffeomorphism} is a smooth bijection with smooth inverse, and
$\mathrm{Diff}(Q)$ is the group of diffeomorphisms of $Q$ under composition. A \textbf{local diffeomorphism} maps an
open neighbourhood of each point diffeomorphically onto an open set; by the inverse function theorem, $\rho$ does this
near $q$ whenever $d\rho_q$ (\bgref{C.2}) is invertible.
\end{bgentry}

LoRA's \citep{hu2021lora} parameter space $\R^{m\times r}\times\R^{r\times n}$ is a vector space, and HRA's
\citep{yuan2024bridging} $\{(u_1,\dots,u_r):u_i\ne0\}$ is open in $(\R^n)^r$. The rotation group $\SO(n)$ is a compact
manifold of dimension $n(n-1)/2$, and the Cayley map (\bgref{D.7}) is a diffeomorphism from $\so(n)$ onto an open dense
subset of it. A method is a pointed smooth map, and its gauge group
$\{\varphi\in\mathrm{Diff}(Q):\rho\circ\varphi=\rho\}$ (\bgref{D.21}) contains LoRA's $(B,A)\mapsto(Bg^{-1},gA)$ for
every $g\in\GL_r$.

\fbAusedin{\thmref{Definition}{I.2}, \thmref{Definition}{I.5}, \thmref{Theorem}{II.7}, \thmref{Proposition}{IV.3}.
  \textsf{Reading:} \citet[Ch.~1 and~2]{xlee2013introduction}.}

\begin{bgentry}{C.2}{tangent spaces, differentials and covectors}
For a manifold $Q\subseteq\R^N$, the \textbf{tangent space} $T_qQ$ is the set of velocities $c'(0)$ of smooth curves $c$
in $Q$ with $c(0)=q$. It is a linear space of dimension $\dim Q$, all of $\R^N$ when $Q$ is open. The
\textbf{differential} of a smooth $\rho$ at $q$ is the linear map $d\rho_q:T_qQ\to T_{\rho(q)}\Theta$,
$c'(0)\mapsto(\rho\circ c)'(0)$. In coordinates it is the Jacobian, and the chain rule reads $d(\rho\circ
h)_q=d\rho_{h(q)}\circ dh_q$. The \textbf{cotangent space} $T^*_\theta\Theta$ is the dual of $T_\theta\Theta$. The
differential $dL_\theta$ of a loss is a \textbf{covector}, and it becomes a gradient vector $\nabla L(\theta)$ only once
an inner product is chosen.
\end{bgentry}

For LoRA, $d\rho_{(B,A)}(\dot B,\dot A)=s(\dot BA+B\dot A)$. For a rotation orbit, $T_I\SO(n)=\so(n)$ and
$T_{W_0R}\big(W_0\cdot\SO(n)\big)=W_0R\,\so(n)$. \thmref{Observation}{I.6} is the chain rule at $q_0$.
\thmref{Theorem}{III.5} drives the parameters by any cotangent signal $\gamma(t)\in T^*_{\rho(q)}\Theta$, pulled back to
$Q$ as $\gamma\circ d\rho_q$, which is $D\rho\T\gamma$ in coordinates.

\fbAusedin{the notation (Table~\ref{tab:notation}), \thmref{Definition}{I.5}, \thmref{Observation}{I.6},
  \thmref{Theorem}{II.2}, \thmref{Theorem}{II.7}, \thmref{Theorem}{III.5}, \thmref{Definition}{IV.1}. \textsf{Reading:}
  \citet[Ch.~3 and~11]{xlee2013introduction}.}

\begin{bgentry}{C.3}{rank, expressive dimension and first-order image}
The \textbf{rank} of $\rho$ at $q$ is $\rk d\rho_q$. It is lower semicontinuous, since a nonzero $k\times k$ minor of
the Jacobian stays nonzero nearby, so $\{q:\rk d\rho_q\ge k\}$ is open. The \textbf{generic rank} is $\max_q\rk
d\rho_q$. If $\rho$ is real-analytic and $Q$ is connected, the generic rank is attained on an open dense set whose
complement has measure zero (\bgref{C.22}). The paper writes $\lvert M\rvert=\dim Q$ for the parameter budget, $d(M)$
for the generic rank, $S_1=d\rho_{q_0}(T_{q_0}Q)$ for the \textbf{first-order image} and $d(M)-\dim S_1$ for the
\textbf{first-order deficit}. For definable $\rho$ on an open definable $Q$, as for every method in this paper,
\bgref{C.18}(d) shows that $d(M)=\dim\Im M$, the \textbf{expressive dimension}. The deficit vanishes exactly when $q_0$
is a point of maximal rank; the base point $\theta_0$ is then called \textbf{regular} for $M$, and an \textbf{apex} when
the deficit is positive (\thmref{Definition}{I.5}).
\end{bgentry}

\thmref{Proposition}{II.3} gives LoRA's rank at every point,
$m\operatorname{rk}A+n\operatorname{rk}B-\operatorname{rk}A\operatorname{rk}B$. At a random pair it is $r(m+n-r)=d$; at
the zero-init point $(0,A_0)$ it is $mr$, a deficit of $r(n-r)$. Since the points of lower rank have measure zero, a
float64 Jacobian rank at one random point computes $d(M)$ with probability one, up to the tolerance that decides
numerical rank. The plug workbench of the companion site (\S\ref{sec:cut-plug}) and the check script
(Appendix~\ref{app:repro}) do exactly this.

\fbAusedin{the notation (Table~\ref{tab:notation}), \thmref{Definition}{I.5}, \thmref{Proposition}{II.3},
  \thmref{Proposition}{II.12}, the plug workbench (\S\ref{sec:cut-plug}), reproducing the numbers
  (Appendix~\ref{app:repro}).}

\begin{bgentry}{C.4}{immersions, submersions and embedded submanifolds}
A smooth $\rho:Q\to\Theta$ is an \textbf{immersion} if every $d\rho_q$ is injective (rank $\dim Q$), and a
\textbf{submersion} if every $d\rho_q$ is surjective (rank $\dim\Theta$). A subset $X\subseteq\R^N$ is an
\textbf{embedded submanifold} of dimension $k$ if near each of its points it is the zero set of $N-k$ smooth functions
with linearly independent gradients. A submersion is an open map, so its image contains an open set. An
\textbf{embedding}, an injective immersion that is a homeomorphism onto its image, has an embedded submanifold as image;
an injective immersion such as the figure eight need not.
\end{bgentry}

The one-sided method $\mathrm{FA}(A_0):B\mapsto W_0+BA_0$ is an injective immersion with affine image $W_0+\{BA_0\}$
when $A_0$ has full row rank. On pairs of full rank, $(B,A)\mapsto BA$ is a submersion onto $\MM_r$ (\bgref{C.12}).
\thmref{Observation}{I.4}(b) uses openness, in the argument recalled in \bgref{A.13}. \thmref{Proposition}{IV.3} asks
for injective immersions $(\R^k,0)\to(U,\theta)$ tangent to a distribution.

\fbAusedin{\thmref{Observation}{I.4}, \thmref{Proposition}{IV.3}, integrability (\S\ref{sec:lens-integrability}).
  \textsf{Reading:} \citet[Ch.~4 and~5]{xlee2013introduction}.}

\begin{bgfact}{C.5}{constant-rank theorem}
Let $\rho:Q\to\Theta$ be smooth with $\rk d\rho_q=k$ for all $q$ near $p$. Then there are charts centred at $p$ and at
$\rho(p)$ in which $\rho$ reads $(x_1,\dots,x_{\dim Q})\mapsto(x_1,\dots,x_k,0,\dots,0)$. Hence some neighbourhood $V$
of $p$ has $\rho(V)$ an embedded $k$-dimensional submanifold, and $\rho^{-1}(\rho(p))\cap V$ an embedded submanifold of
dimension $\dim Q-k$. By \bgref{C.3} the hypothesis holds wherever the rank is maximal.
\end{bgfact}

\paragraph{The idea of the proof}
Choose coordinates in which the top-left $k\times k$ block of the Jacobian is invertible. The inverse function theorem
straightens the first $k$ output coordinates. Constant rank then forces the other outputs to depend only on those $k$,
and a second change of coordinates in the target removes them.

The paper uses this theorem without naming it. Near a point of maximal rank, $\rho$ is locally a projection onto a sheet
of dimension $d(M)$, with fibres of dimension $\lvert M\rvert-d(M)$, the gauge waste. For LoRA at a full-rank pair the
sheet is a piece of $W_0+\MM_r$ and the fibre a piece of a $\GL_r$-orbit. \thmref{Proposition}{IV.3} makes the analogous
constant-rank assumption for distributions, where Frobenius' theorem (\bgref{D.25}) takes the place of this one.

\fbAusedin{\thmref{Definition}{I.5}, \thmref{Proposition}{II.12}, \bgref{C.18}. \textsf{Reading:} \citet[Ch.~4, the rank
  theorem]{xlee2013introduction}.}

\begin{bgentry}{C.6}{fibres and fibre dimension}
The \textbf{fibre} of $\rho$ over $\theta$ is $\rho^{-1}(\theta)$, the set of parameters that give the weight $\theta$.
If $\rk d\rho_q=d(M)$, \bgref{C.5} makes the fibre through $q$ a submanifold of dimension $\lvert M\rvert-d(M)$ near
$q$. For real-analytic $\rho$ on connected $Q$ this holds for all $q$ outside a closed set of measure zero, and these
are the \textbf{generic fibres}. Fibres through points of lower rank can have a different dimension. The fibres of a
fibre bundle all look alike locally, so a map whose fibre dimension jumps is no fibre bundle.
\end{bgentry}

For LoRA on the full-rank stratum, the fibre through $(B,A)$ is the orbit $\{(Bg^{-1},gA):g\in\GL_r\}$, of dimension
$r^2=r(m+n)-r(m+n-r)$ (\bgref{D.21}). For zero-init LoRA the fibre over $\theta_0$ is $\{(B,A):BA=0\}$. It contains
$\{B=0\}$, of dimension $rn>r^2$, and is itself a union of strata, such as $\{(0,A):\rk A=r\}$
(\thmref{Corollary}{III.13}).

\fbAusedin{Exposé~\ref{sec:fibres}, \thmref{Definition}{I.5}, \thmref{Proposition}{II.12}, \thmref{Corollary}{III.13},
  Rosetta entries 6 and 11 (Exposé~\ref{sec:rosetta}), the seven borrowed words (Exposé~\ref{sec:analogy}).}

\begin{bgentry}{C.7}{fibre products, transversality and clean intersection}
For $\rho_M:Q_M\to\Theta$ and $\rho_N:Q_N\to\Theta$, the set-level \textbf{fibre product} is
$P=\{(q,q'):\rho_M(q)=\rho_N(q')\}\subseteq Q_M\times Q_N$. The maps are \textbf{transverse}, $\rho_M\pitchfork\rho_N$,
if $d\rho_M(T_qQ_M)+d\rho_N(T_{q'}Q_N)=T_\theta\Theta$ at every $(q,q')\in P$; then $P$ is an embedded submanifold of
dimension $\lvert M\rvert+\lvert N\rvert-\dim\Theta$, and it is the pullback of $\rho_M$ and $\rho_N$ in $\Diff$
(\bgref{A.11}). The intersection is \textbf{clean} if $P$ is an embedded submanifold with
$T_{(q,q')}P=\{(v,w):d\rho_Mv=d\rho_Nw\}$ at every point. Transverse maps meet cleanly, and so do affine maps.
\end{bgentry}

Transversality at $(q_0,q_0')$ needs $\lvert M\rvert+\lvert N\rvert\ge\dim\Theta$, which fails for any two small
methods, so the products the paper builds come from clean cases. The affine methods $\mathrm{FA}(A_0)$ and
$\mathrm{FB}(B_0)$, with $B_0$ of full column rank and $A_0$ of full row rank, have the linear fibre product
$\{(B_0R,RA_0)\}$ of dimension $r^2$, which is LoRA-XS \citep{baazy2024lora} (\thmref{Proposition}{I.8}). On
$\Theta=\R$, $\mathrm{LoRA}_1$ and the frozen model have the cross $\{BA=0\}$ as fibre product. It is no manifold at the
origin, and no product exists (\bgref{A.13}).

\fbAusedin{\thmref{Observation}{I.4}, \thmref{Proposition}{I.8}, \thmref{Theorem}{II.7}(d). \textsf{Reading:}
  \citet[Ch.~1]{xguillemin1974differential}; \citet[Ch.~6]{xlee2013introduction}.}

\subsubsection*{Germs, cones and singular points}

\begin{bgentry}{C.8}{germs}
Two subsets $X,Y\subseteq\Theta$ have the same \textbf{germ} at $\theta_0$ if $X\cap U=Y\cap U$ for some neighbourhood
$U$ of $\theta_0$; two maps have the same germ at $q_0$ if they agree near $q_0$. The germs $(X,\theta_0)$ and $(Y,0)$
are diffeomorphic if a diffeomorphism $\varphi$ between neighbourhoods $U\ni\theta_0$ and $V\ni0$ has
$\varphi(\theta_0)=0$ and $\varphi(X\cap U)=Y\cap V$. Local properties, such as the dimension at $\theta_0$, smoothness
at $\theta_0$ and the tangent cone (\bgref{C.11}), depend only on the germ.
\end{bgentry}

The image germ of a method is the germ of $\Im M=\rho(Q)$ at $\theta_0$. It uses all of $Q$, so it can be larger than
the image of a small neighbourhood of $q_0$. For zero-init LoRA, $sBA$ with $B$ small and $A$ arbitrary is a small
update with any row space, so the germ is that of the whole cone $\theta_0+\Mr$; parameters near $(0,A_0)$ only give
updates with row space inside $\operatorname{row}A$ for some $A$ near $A_0$. For injective $W_0$ and even $r$ with $2\le
r\le n-2$, HRA's image germ at $W_0$ is the image of the germ of $\Sigma_r$ at $0$ under the embedding $\Omega\mapsto
W_0\,C(\Omega)$ of $\so(n)$ into $\R^{m\times n}$ (\bgref{C.13}, \bgref{D.7}).

\fbAusedin{the image germ (Exposé~\ref{sec:image}), the three invariants (Exposé~\ref{sec:thesis}),
  \thmref{Theorem}{II.7}, the seven axes (\S\ref{sec:atlas-axes}).}

\begin{bgentry}{C.9}{smooth and singular points}
A point $x$ of a set $X\subseteq\R^N$ is a \textbf{smooth point} if $X\cap U$ is a $C^1$ embedded submanifold for some
neighbourhood $U$ of $x$, and a \textbf{singular point} otherwise. The singular points form the \textbf{singular locus}
$\operatorname{Sing}X$. Algebraic geometry defines $\operatorname{Sing}$ through the Jacobian of defining polynomials,
and for $\Mr$ the two notions agree. Smoothness is a property of a set. The paper's regular base point and apex, defined
by a zero or positive first-order deficit (\bgref{C.3}, \thmref{Definition}{I.5}), are properties of a method at $q_0$.
\end{bgentry}

The cross $\{BA=0\}\subseteq\R^2$ is singular at the origin and smooth elsewhere. For $1\le r<\min(m,n)$,
$\operatorname{Sing}\Mr=\MM_{\le r-1}$. A split initialisation has image $(\theta_0-P)+\Mr$ with $\rk P=r$, so
$\theta_0$ is a smooth point; zero-init LoRA starts at the singular vertex. A figure-eight curve passing through its
crossing with nonzero velocity has deficit $0$ there, though its image is singular, so smooth and regular are different
notions.

\fbAusedin{\thmref{Definition}{I.5}, \thmref{Theorem}{II.2}, Figure~\ref{fig:apex}, \thmref{Theorem}{II.7}, Erlangen
  (\S\ref{sec:erlangen}). \textsf{Reading:} \citet[Lecture~14]{xharris1992algebraic}.}

\begin{bgentry}{C.10}{cones and vertices}
A set $X$ in a vector space is a \textbf{cone} with \textbf{vertex} $v$ if $v\in X$ and $v+t(x-v)\in X$ for all $x\in X$
and $t\ge0$, and \textbf{linear} if it is an affine subspace. If $\rho=\theta_0+\delta$ with
$\delta(tu,w)=t\,\delta(u,w)$ for $t\ge0$ on a block $u$ of parameters, and $Q$ is closed under $u\mapsto tu$, then $\Im
M$ is a cone with vertex $\theta_0$. A cone that is a $C^1$ embedded submanifold near its vertex $v$ equals $v+T_vX$, so
the vertex of a non-linear cone is a singular point.
\end{bgentry}

\paragraph{Why a smooth cone is linear}
Each ray from $v$ lies in $X$, so $X\subseteq v+T_vX$. Near $v$, $X$ is a submanifold of full dimension in this affine
space, so it contains a neighbourhood of $v$ there, and scaling spreads it over all of $v+T_vX$.

$\Mr$ is a closed cone with vertex $0$, and it is not linear for $1\le r<\min(m,n)$, since a rank-$r$ and a rank-one
matrix can sum to rank $r+1$. Zero-init LoRA has $\delta=sBA$, linear in $B$, so its image $W_0+\Mr$ has singular vertex
$W_0$. Its deficit $r(n-r)$ makes $\theta_0$ an apex too; the vertex is a property of the set, the apex one of the
method. Multihomogeneous updates such as LoHa's \citep{hyeonwoo2021fedpara} give cones in the same way (\bgref{D.19}).

\fbAusedin{\thmref{Definition}{I.5}, the image germ (Exposé~\ref{sec:image}), \thmref{Theorem}{II.1},
  \thmref{Theorem}{II.2}, Figure~\ref{fig:apex}, \thmref{Theorem}{II.7}.}

\begin{bgfact}{C.11}{first-order image versus image}
The \textbf{tangent cone} of a set $X$ at $x$ is the set of limits of $t_k(x_k-x)$ with $x_k\in X$, $x_k\to x$ and
$t_k>0$; for a closed cone with vertex $v$ it is $X-v$. For a method $M$: (a) $S_1$ lies in the tangent cone of $\Im M$
at $\theta_0$; (b) $\dim S_1\le d(M)$; (c) if $\Im M$ is a $C^1$ manifold of dimension $d(M)$ near $\theta_0$, then
$S_1\subseteq T_{\theta_0}\Im M$, with equality exactly when the deficit is $0$.
\end{bgfact}

For (a), $d\rho_{q_0}(\xi)$ is the velocity of the curve $\rho(q_0+t\xi)$ inside the image; for (c), $\rho$ maps a
neighbourhood of $q_0$ into the manifold. For $\mathrm T_A$-pointed LoRA ($B_0=0$, \bgref{C.19}) the tangent cone at
$\theta_0$ is $\Mr$ itself, and inside it $S_1=\{XA_0\}$, the matrices with row space in $\operatorname{row}A_0$, is a
linear space of dimension $mr<r(m+n-r)$. Figure~\ref{fig:apex}(a) draws $S_1$ as one ruling of the cone, a line through
the apex. At a split pointing case (c) applies, and $S_1=T_P\MM_r$, the tangent plane of Figure~\ref{fig:apex}(b).

\fbAusedin{\thmref{Definition}{I.5}, \thmref{Observation}{I.6}, \thmref{Theorem}{II.2}, Figure~\ref{fig:apex}.
  \textsf{Reading:} \citet[Lecture~20]{xharris1992algebraic}.}

\subsubsection*{Matrix varieties}

\begin{bgentry}{C.12}{the determinantal variety}
$\Mr=\{X\in\R^{m\times n}:\rk X\le r\}$, the \textbf{determinantal variety}, is the zero set of all $(r+1)\times(r+1)$
minors, a closed cone with vertex $0$. Let $1\le r<\min(m,n)$. The matrices of rank exactly $r$ form an open dense
subset $\MM_r$ of $\Mr$, an embedded submanifold of dimension $r(m+n-r)$, and $\operatorname{Sing}\Mr=\MM_{\le r-1}$. If
$X\in\MM_r$ has column space $U$ and row space $V$, and $X=BA$ with $B\in\R^{m\times r}$ and $A\in\R^{r\times n}$, then
\[T_X\MM_r=\{\dot BA+B\dot A\}=\{D:P_{U^\perp}DP_{V^\perp}=0\}.\]
Also $\Mr-\Mr=\MM_{\le\min(2r,m,n)}$.
\end{bgentry}

Zero-init LoRA has image $W_0+\Mr$, of dimension $130{,}816$ for a $4096\times4096$ matrix at $r=16$. On full-rank
pairs, $(B,A)\mapsto BA$ is a submersion onto $\MM_r$ whose fibres are $\GL_r$-orbits, which identifies $\MM_r$ with the
quotient $(\R^{m\times r}_*\times\R^{r\times n}_*)/\GL_r$ (\bgref{D.22}). The difference formula is why no pointing of
$\mathrm{LoRA}_r$ reaches an update of rank above $2r$ (\bgref{B.14}, \thmref{Theorem}{II.1}).

\fbAusedin{the notation (Table~\ref{tab:notation}), translates of one shape (\S\ref{sec:strata-translates}),
  \thmref{Theorem}{II.1}, \thmref{Theorem}{II.2}, \thmref{Proposition}{II.8}, Exposé~II.C (\S\ref{sec:cut}),
  \thmref{Proposition}{III.4}. \textsf{Reading:} \citet[Lecture~9]{xharris1992algebraic}.}

\begin{bgentry}{C.13}{the skew-determinantal cone}
Every skew-symmetric matrix has even rank. For even $r$ with $2\le r\le n-2$, the \textbf{skew-determinantal cone}
$\Sigma_r=\{\Omega\in\so(n):\rk\Omega\le r\}$ is a closed cone with vertex $0$ and dimension
$r(n-r)+r(r-1)/2=rn-r(r+1)/2$, counting a range in $\Gr(r,n)$ and a nondegenerate skew form on it. Its smooth points are
the matrices of rank exactly $r$, and $\operatorname{Sing}\Sigma_r=\Sigma_{r-2}$. It is not linear, since skew matrices
of rank $r$ and $2$ can sum to rank $r+2\le n$, so its vertex is singular.
\end{bgentry}

For injective $W_0$ and even $r$, HRA's image near $W_0$ consists of the $W_0R$ with $R\in\SO(n)$ near $I$ and
$\rk(R-I)\le r$. The Cayley map satisfies $\rk(C(\Omega)-I)=\rk\Omega$ (\bgref{D.7}), so the image germ at $W_0$ is the
image of the germ of $\Sigma_r$ at $0$ under the embedding $\Omega\mapsto W_0\,C(\Omega)$, $W_0$ is a singular vertex,
and $d(\mathrm{HRA}_r)=\dim\Sigma_r$. At the paired initialisation $u_{2i-1}=u_{2i}=w_i$ ($i\le k=r/2$, with the $w_i$
independent), put $U=\spanop\{w_1,\dots,w_k\}$, so $\dim U=k$. Then $S_1=W_0\cdot\mathcal S$ with
\[\mathcal S=\{\Omega\in\so(n):P_{U^\perp}\Omega P_{U^\perp}=0\}=U\wedge\R^n\]
(\bgref{D.5}), a linear space of skew matrices of rank at most $2k=r$ and of dimension $kn-k(k+1)/2<d(\mathrm{HRA}_r)$.
It lies inside $\Sigma_r$, so $S_1$ lies inside the tangent cone $W_0\cdot\Sigma_r$ (\thmref{Theorem}{II.7}(b)).

\fbAusedin{\thmref{Theorem}{II.7}, HRA is a meet (\S\ref{sec:erlangen-hra}).}

\begin{bgentry}{C.14}{Segre and toric cones}
The \textbf{Segre cone} in $\R^p\otimes\R^q\cong\R^{p\times q}$ is the set $\{a\otimes b\}=\MM_{\le1}$ of rank-one
tensors, a closed cone of dimension $p+q-1$ that spans the whole space and is non-linear when $p,q\ge2$. It is the
simplest \textbf{toric cone}, the closure of the image of a torus $(\R^\times)^k$ under a monomial map, here
$(a,b)\mapsto(a_ib_j)_{ij}$. Since rank-one matrices span $\R^{p\times q}$, $k$ generic rank-one matrices span
$\min(k,pq)$ dimensions.
\end{bgentry}

Under the Van Loan--Pitsianis rearrangement (\bgref{B.11}), KronA's \citep{edalati2022krona} update $A\otimes B$ becomes
$\vect(A)\,\vect(B)\T$, so its image is a linear copy of a Segre cone (\thmref{Proposition}{II.12}). VeRA's
\citep{kopiczko2023vera} update $\Lambda_bB\Lambda_dA$ has entries $\sum_k(b_id_k)B_{ik}A_{kj}$, a fixed linear function
of the rank-one matrix $bd\T$. This function is injective when $B$ has no zero entry and $A$ has full row rank, so the
image is a toric cone of dimension $m+r-1$ with vertex $\theta_0$, as VeRA's entry in Exposé~\ref{sec:examples} records.
\thmref{Proposition}{II.8} uses the spanning fact for DoRA \citep{liu2024dora}, with $m$ points of the Segre cone in
$\R^{m-r}\otimes\R^{n-r}$.

\fbAusedin{\thmref{Proposition}{II.8}, \thmref{Proposition}{II.12}, \hyperref[ex:vera]{VeRA}
  (Exposé~\ref{sec:examples}). \textsf{Reading:} \citet[Lecture~2]{xharris1992algebraic}; \citet[Ch.~1]{xcox2011toric}.}

\begin{bgentry}{C.15}{the Grassmannian}
The \textbf{Grassmannian} $\Gr(r,n)$ is the set of $r$-dimensional linear subspaces of $\R^n$, a compact manifold of
dimension $r(n-r)$. One model is the full-row-rank $r\times n$ matrices up to $A\mapsto gA$, $g\in\GL_r$, through
$A\mapsto\operatorname{row}A$; another is the orthogonal projectors $P=P\T=P^2$ with $\tr P=r$. Near $V$, the graphs of
linear maps $V\to V^\perp$ give a chart onto $\operatorname{Hom}(V,V^\perp)\cong\R^{r(n-r)}$. $\Gr(r,n)$ carries a
unique $\Ort(n)$-invariant probability measure, the uniform or Haar measure, and the row space of a Gaussian $A$ has
this law (\bgref{D.14}).
\end{bgentry}

$\mathrm T_A$-pointed LoRA is classified up to isomorphism by $\operatorname{row}A_0\in\Gr(r,n)$
(\thmref{Theorem}{II.2}), and its deficit $r(n-r)$ equals $\dim\Gr(r,n)$. First steps keep the row space of the update
inside $\operatorname{row}A_0$, and the missing directions are those that tilt it. For injective $W_0$ and even $r$ with
$2\le r\le n-2$, HRA's image is the union of the orbits $W_0\cdot\SO(V)$ over $V\in\Gr(r,n)$, and its dimension
$r(n-r)+r(r-1)/2$ counts a subspace and a rotation inside it. LoRA-FA \citep{zhang2023lora}, which trains $B$ with $A_0$
frozen, sees $A_0$ only through $V=\operatorname{row}A_0$ once its gradient is corrected as in
\thmref{Proposition}{III.4}(c).

\fbAusedin{\thmref{Theorem}{II.2}, the checks (Appendix~\ref{app:repro}), \thmref{Theorem}{II.7}, LOFT
  (\S\ref{sec:erlangen-loft}) and the LOFT example there, \thmref{Proposition}{III.4}, \thmref{Corollary}{III.13}.
  \textsf{Reading:} \citet[Ch.~1]{xlee2013introduction}; \citet[Lecture~6]{xharris1992algebraic}.}

\subsubsection*{Tame sets and genericity}

\begin{bgfact}{C.16}{images of smooth maps need not be manifolds}
Even a polynomial map can have an image that is no manifold. (a) \emph{Cusps.} $t\mapsto(t^2,t^3)$ is injective with
image $\{y^2=x^3\}$, which is not $C^1$ at $0$. (b) \emph{Crossings and cones.} The cross $\{BA=0\}\subseteq\R^2$ and
the cone $\Mr$ are singular at the origin. (c) \emph{Missing limits.} $(x,y)\mapsto(x,xy)$ has image
$\{u\ne0\}\cup\{(0,0)\}$ in the $(u,v)$-plane, neither open nor closed. (d) \emph{Dense images.} The line of irrational
slope is an injective immersion into the torus with dense image.
\end{bgfact}

PEFT images show (b) routinely, as in the LoRA cone and HRA's skew cone, so the paper measures them with tools that need
no manifold structure, such as germs (\bgref{C.8}), tangent cones (\bgref{C.11}) and the dimension theory of
semialgebraic and definable sets (\bgref{C.17}, \bgref{C.18}). Whether LoHa's Hadamard image is closed, the question
raised by (c), is open.

\fbAusedin{\thmref{Observation}{I.4}, \thmref{Theorem}{II.2}, \thmref{Theorem}{II.7}, the open problems
  (Exposé~\ref{sec:analogy}).}

\begin{bgentry}{C.17}{semialgebraic and definable sets}
A \textbf{semialgebraic set} is a finite union of sets $\{x\in\R^N:p(x)=0,\ q_1(x)>0,\dots,q_k(x)>0\}$ with polynomials
$p,q_i$, and a map is semialgebraic if its graph is. A set is \textbf{definable} in $\R_{\exp}$ if a first-order formula
defines it from polynomials with real coefficients, $\exp$, equality, order, Boolean operations and quantifiers over
$\R$. \citet{xwilkie1996model} proved that $\R_{\exp}$ is \textbf{o-minimal}, meaning every definable subset of $\R$ is
a finite union of points and intervals. Semialgebraic sets are the definable sets that need no $\exp$.
\end{bgentry}

LoRA, DoRA (whose row norms involve square roots), HRA and exact-Cayley OFT \citep{qiu2023controlling} are built from
polynomials, quotients and square roots, so their images are semialgebraic by \bgref{C.18}(a). A softmax brings in
$\exp$, and maps that contain one are definable in $\R_{\exp}$.

\fbAusedin{where the category theory works (Exposé~\ref{sec:thesis}), the arrow rules (\S\ref{sec:atlas-arrows}).
  \textsf{Reading:} \citet[Ch.~2]{xbochnak1998real}; \citet{xwilkie1996model}.}

\begin{bgfact}{C.18}{Tarski--Seidenberg and definable dimension}
(a) The image of a semialgebraic set under a semialgebraic map is semialgebraic (Tarski--Seidenberg); images of
definable sets under definable maps are definable. (b) Every definable set $X$ is a finite disjoint union of $C^1$
embedded submanifolds, and $\dim X$, the largest of their dimensions, does not depend on the choice. (c) $X\subseteq Y$
implies $\dim X\le\dim Y$, $\dim f(X)\le\dim X$ for definable $f$, and $\dim(\overline X\setminus X)<\dim X$ for
nonempty $X$. (d) For a definable $C^1$ map $\rho$ on an open definable $Q\subseteq\R^k$, $\dim\rho(Q)=\max_q\rk
d\rho_q$.
\end{bgfact}

\paragraph{The idea of (d)}
Near a point of maximal rank, \bgref{C.5} gives a sheet of that dimension inside $\rho(Q)$. Conversely, a definable
section of $\rho$, $C^1$ on a top-dimensional piece of $\rho(Q)$, composes with $\rho$ to the identity there, so the
rank is at least that dimension.

Part (d) is the identity $d(M)=\dim\Im M$ of \thmref{Definition}{I.5}. Part (c) is the monotonicity behind test O2
(\S\ref{sec:atlas-tests}), since an arrow $M\to N$ gives $\Im M\subseteq\Im N$ and hence $d(M)\le d(N)$; for instance
$\mathrm{LoRA}_r\not\to\text{LoRA-FA}$ because $r(m+n-r)>mr$.

\fbAusedin{\thmref{Definition}{I.5}, the arrow rules (\S\ref{sec:atlas-arrows}), where the category theory works
  (Exposé~\ref{sec:thesis}). \textsf{Reading:} \citet[Ch.~3 and~4]{xvandendries1998tame};
  \citet[Ch.~2]{xbochnak1998real}.}

\begin{bgentry}{C.19}{stratification and local dimension}
A \textbf{stratification} of a set $X$ is a partition into finitely many embedded submanifolds, the \textbf{strata},
such that the closure of each stratum is a union of strata. The \textbf{local dimension} $\dim_xX$ is the dimension of
the germ of $X$ at $x$, the largest dimension of a stratum whose closure contains $x$. $X$ is \textbf{pure-dimensional}
if $\dim_xX$ is the same at every point. Definable sets admit stratifications, so these notions make sense for every
image in this paper.
\end{bgentry}

The cross $\{ba=0\}\subseteq\R^2$ is stratified as the origin and the four open half-axes, and it is pure of dimension
$1$. Likewise $\Mr=\MM_0\sqcup\MM_1\sqcup\dots\sqcup\MM_r$ with $\overline{\MM_k}=\MM_{\le k}$, and it is pure of
dimension $r(m+n-r)$. For $1\le r<\min(m,n)$, \thmref{Theorem}{II.2} singles out three strata of full-rank LoRA
pointings, $\mathrm T_A$ ($B_0=0$, $\rk A_0=r$), $\mathrm T_B$ ($A_0=0$, $\rk B_0=r$) and the open dense $\mathrm S$
($\rk B_0A_0=r$), and the fibre $\{BA=0\}$ over $\theta_0$ is itself stratified (\thmref{Corollary}{III.13}). The HRA
fibre product of \thmref{Theorem}{II.7}(d) has local dimension $d(\mathrm{HRA}_r)+r^2=rn+r(r-1)/2$ over generic points
of the meet and contains $\{(B,0,I)\}$, of dimension $mr$, over $W_0$, so it is not pure-dimensional when
$mr>rn+r(r-1)/2$.

\fbAusedin{\thmref{Theorem}{II.2}, the three strata (\S\ref{sec:strata-three}), \thmref{Theorem}{II.7},
  \thmref{Corollary}{III.13}, the open problems (Exposé~\ref{sec:analogy}), the seven borrowed words
  (Exposé~\ref{sec:analogy}). \textsf{Reading:} \citet[Ch.~4]{xvandendries1998tame}; \citet[Ch.~9]{xbochnak1998real}.}

\begin{bgentry}{C.20}{moduli sets and moduli spaces}
Classifying objects up to isomorphism gives a set of classes. A \textbf{moduli space} also carries a topology or
geometry for which continuous or smooth families of objects correspond to continuous or smooth maps into it, ideally
with a universal family over it from which every family is pulled back. A disjoint union of spaces $X\sqcup Y$ has the
topology in which $X$ and $Y$ are both open and closed.
\end{bgentry}

\thmref{Theorem}{II.2} classifies the three strata of LoRA pointings, as a set, by $\Gr(r,n)\sqcup\Gr(r,m)\sqcup\MM_r$.
The quotient topology does not split this way, since $(\varepsilon B',A_0)\in\mathrm S$ tends to $(0,A_0)\in\mathrm T_A$
as $\varepsilon\to0$, so $\mathrm T_A$ lies in the closure of $\mathrm S$. No universal family is built, and the paper
uses ``moduli'' as an analogy.

\fbAusedin{\thmref{Theorem}{II.2}, the Rosetta stone (Exposé~\ref{sec:rosetta}), the seven borrowed words
  (Exposé~\ref{sec:analogy}). \textsf{Reading:} \citet[Lecture~4]{xharris1992algebraic}.}

\begin{bgentry}{C.21}{generic properties}
A property of points of $\R^N$ holds \textbf{generically} if the set where it fails lies in $\{f=0\}$ for some nonzero
polynomial $f$. The set $\{f=0\}$ is closed, nowhere dense and of Lebesgue measure zero, so a generic property holds
with probability one at a random point whose law has a density, such as a Gaussian or uniform initialisation. For points
of a parametrised set, such as a Segre cone, ``generic'' refers to generic parameters. To prove genericity it suffices
to write failure as the vanishing of polynomials and exhibit one point where one of them is nonzero. A weaker notion,
\textbf{almost every} point, asks only that the failure set have Lebesgue measure zero, as a countable union of proper
algebraic subsets does.
\end{bgentry}

\thmref{Theorem}{II.2}(d) uses a generic rank-one update, \thmref{Proposition}{II.8} gives DoRA's dimension
$\min\big(mn,\,r(m+n-r)+m\big)$ for generic $W_0$, and \thmref{Proposition}{V.2} notes that QLoRA's
\citep{dettmers2023qlora} defect $e=\kappa W-W$ has full rank for almost every $W$, in the weaker sense. The quantiser
$\kappa$ is defined cell by cell rather than by one polynomial, so the argument runs cell by cell, and the failure set
is a countable union of proper algebraic subsets of quantisation cells, which has measure zero. A pretrained $W_0$ is
not random, and structured weights can lie in the exceptional set. For a rank-one $W_0=ac\T$ with all $a_i\ne0$ and
$r<n$, \thmref{Proposition}{II.8} finds DoRA's dimension $r(m+n-r)+m-r$, strictly below the generic value exactly when
$r<m$ and $r\le n-2$.

\fbAusedin{\thmref{Theorem}{II.2}, \thmref{Proposition}{II.8}, \thmref{Theorem}{II.11}, \thmref{Proposition}{II.12},
  \thmref{Proposition}{V.2}.}

\begin{bgentry}{C.22}{real-analytic functions and their zero sets}
A function is \textbf{real-analytic} on an open set if near every point it equals a convergent power series.
Polynomials, $\exp$, softmax, $\tanh$, GELU and SiLU are real-analytic, and so are sums, products, compositions and
quotients with nonvanishing denominator. On a connected open set, a real-analytic function that vanishes on a nonempty
open subset vanishes identically, and the zero set of one that is not identically zero is closed, nowhere dense and of
measure zero.
\end{bgentry}

\thmref{Lemma}{VI.2} compares Taylor coefficients at $0$. A nonzero scale $c$ with $\sigma(ct)=c\,\sigma(t)$ must
satisfy $ca_k=c^ka_k$ for every coefficient $a_k$ of $\sigma$, so $c=1$ for GELU and SiLU, whose $a_2\ne0$.
\thmref{Theorem}{VI.5}(f) uses the zero-set fact. A prefixed head is analytic in the prefix, and one prefix with a
nonzero second derivative makes the prefixes with affine output a nowhere-dense set. The same fact makes the
maximal-rank set of an analytic $\rho$ open and dense (\bgref{C.3}).

\fbAusedin{\thmref{Lemma}{VI.2}, \thmref{Theorem}{VI.5}, Rosetta entry 5 (Exposé~\ref{sec:rosetta}). \textsf{Reading:}
  \citet{xkrantz2002primer}.}

\subsection*{Part D. Groups, symmetry and conservation}
\phantomsection\label{app:bg-d}\addcontentsline{toc}{subsection}{Part D. Groups, symmetry and conservation}

Part D turns to symmetry. Most PEFT parametrisations have symmetries, since different trainable parameters can give the
same weights and some transformations of weight space carry a method's image to itself. This part collects the group
theory the paper uses to describe those symmetries, the invariants they leave untouched, and the quantities they
conserve under gradient flow.

\subsubsection*{Groups and their Lie algebras}

\begin{bgentry}{D.1}{Lie groups; the matrix groups of this paper}
A \textbf{Lie group} is a group that is also a smooth manifold, with smooth multiplication and inversion. Every closed
subgroup of $\GL_n(\R)$ is a Lie group (Cartan's closed-subgroup theorem), and all groups below are of this kind. Its
dimension is its dimension as a manifold.

\smallskip
{\centering\small\renewcommand{\arraystretch}{1.12}
\begin{tabularx}{\linewidth}{@{}llX@{}}
  \toprule
  \headrow \textbf{group} & \textbf{dimension} & \textbf{elements}\\
  \midrule
  $\GL_r$ & $r^2$ & invertible $r\times r$ matrices\\
  $\Ort(n)$, $\SO(n)$ & $n(n-1)/2$ & $R\T R=I$; for $\SO(n)$ also $\det R=1$\\
  $\CO(n)$ & $n(n-1)/2+1$ & $cR$ with $c>0$, $R\in\Ort(n)$, so $\CO(n)=\R_{>0}\cdot\Ort(n)$; equivalently
    $g\T g\in\R_{>0}I$\\
  $\mathrm{Sp}(2r,\R)$ & $r(2r+1)$ & $M\T JM=J$, with $J=\begin{pmatrix}0&I_r\\-I_r&0\end{pmatrix}$\\
  $T_m$ & $m$ & invertible diagonal $m\times m$ matrices (\bgref{D.2})\\
  \bottomrule
\end{tabularx}\par}
\end{bgentry}

LoRA's gauge is $\GL_r$, acting on factors by $g\cdot(B,A)=(Bg^{-1},gA)$. Exactly orthogonal OFT keeps the weight inside
the orbit $W_0\cdot\Ort(n)$, and HRA with even $r$ keeps it inside $W_0\cdot\SO(n)$. $\CO(n)$ is exactly the input-side
covariance group (\bgref{D.17}) of the PiSSA \citep{meng2024pissa}, MiLoRA \citep{wang2024milora} and paper-form CorDA
\citep{yang2024corda} splits. LoCO's \citep{nguyen2026loco} skew generator satisfies $UV\T-VU\T=ZJZ\T$ with $Z=[U\ V]$,
so it is unchanged under $Z\mapsto ZM$ for every $M\in\mathrm{Sp}(2r,\R)$.

\fbAusedin{the notation (Table~\ref{tab:notation}), \thmref{Definition}{I.5}, \thmref{Theorem}{II.5},
  \thmref{Proposition}{II.16}, LOFT (\S\ref{sec:erlangen-loft}), \thmref{Theorem}{III.5}, \thmref{Theorem}{V.3}.
  \textsf{Reading:} \citet[Ch.~1]{xhall2015lie}.}

\begin{bgentry}{D.2}{tori, diagonal monoids and saturation}
The \textbf{diagonal torus} $T_m=\{\diag(d):d_i\ne0\}\cong(\R^\times)^m$ is a commutative Lie group with $2^m$ connected
components, and $T_m^+=\{\diag(d):d_i>0\}\cong\R_{>0}^m$ is the component of the identity. $\mathrm{Diag}_m$, the set of
all diagonal matrices including singular ones, is a monoid under multiplication (\bgref{D.10}) whose invertible elements
form $T_m$. Identifying $\R^2$ with $\mathbb C$, the scaled rotations $a\,\mathrm{Rot}(\theta)$ with $a>0$ form
$\CO^+(2)\cong\mathbb C^\times$, through $a\,\mathrm{Rot}(\theta)\mapsto ae^{i\theta}$. If a group or monoid $G$ acts on
a space containing a set $X$, the \textbf{saturation} $G\cdot X=\{g\cdot x:g\in G,\ x\in X\}$ is the smallest $G$-stable
set containing $X$.
\end{bgentry}

(IA)$^3$ \citep{liu2022few} on keys and values maps $\ell\mapsto\diag(\ell)W_0$, so its image $\mathrm{Diag}_m\cdot W_0$
is the closure of the torus orbit $T_m\cdot W_0$. DoRA's image is the saturation $\mathrm{Diag}_m\cdot(W_0+\Mr)$ of
LoRA's cone. RoAd$_1$ \citep{liao20243} is the left action of $\CO^+(2)^{m/2}\cong(\mathbb C^\times)^{m/2}$ on pairs of
output rows, which makes it (IA)$^3$ over $\mathbb C$. LoHa's gauge contains a two-sided diagonal torus.

\fbAusedin{\thmref{Theorem}{II.5}, DoRA (\S\ref{sec:erlangen-dora}), \thmref{Proposition}{II.8},
  \thmref{Proposition}{II.9}, \thmref{Proposition}{II.12}, the hyperparameter torus (\S\ref{sec:fibres-torus}).}

\begin{bgentry}{D.3}{permutations, monomial matrices, semidirect products, normalisers}
$S_d$ is the group of permutations of $d$ objects, realised by permutation matrices. A matrix is \textbf{monomial} if
each row and each column has exactly one nonzero entry. Every monomial matrix factors uniquely as $DP$ with $D\in T_m$
and $P$ a permutation matrix, and $T_m$ is normal in the group $\mathrm{Mon}_m$ of monomial matrices, so
$\mathrm{Mon}_m=T_m\rtimes S_m$. Here a \textbf{semidirect product} $N\rtimes H$ is a group with a normal subgroup $N$
and a subgroup $H$ such that $N\cap H=\{e\}$ and $NH$ is the whole group. The \textbf{signed permutations}
$B_r=\{\pm1\}^r\rtimes S_r=\mathrm{Mon}_r\cap\Ort(r)$ are the monomial matrices with entries $\pm1$, and $\lvert
B_r\rvert=2^rr!$. The \textbf{normaliser} of a subset $H\subseteq G$ is $N_G(H)=\{g\in G:gHg^{-1}=H\}$.
\end{bgentry}

A monomial $g$ satisfies $g\,\mathrm{Diag}_m\,g^{-1}=\mathrm{Diag}_m$, and in $\GL_m$ the normaliser of $T_m$ is exactly
$\mathrm{Mon}_m$. This puts $\mathrm{Mon}_m\times\GL_n$ inside the covariance group (\bgref{D.17}) of output scalings
such as (IA)$^3$ and DoRA's magnitude. Adam \citep{xkingma2015adam} acts coordinatewise, so on LoRA's full-rank locus with
channel-uniform hyperparameters (one learning rate, $\varepsilon$ and decay per factor) it is equivariant, of LoRA's
gauge, only under $B_r\subset\Ort(r)\subset\GL_r$ (\thmref{Theorem}{III.12}). By \thmref{Lemma}{VI.2}, an invertible linear map that
commutes with a non-affine $C^2$ activation is monomial.

\fbAusedin{DoRA (\S\ref{sec:erlangen-dora}), \thmref{Proposition}{II.8}, \thmref{Proposition}{II.9},
  \thmref{Theorem}{III.12}, \thmref{Corollary}{III.13}, \thmref{Proposition}{V.4}, \thmref{Lemma}{VI.2}, the privileged
  basis (\S\ref{sec:merge-basis}).}

\begin{bgentry}{D.4}{connectedness and the identity component}
The \textbf{identity component} $G^\circ$ of a Lie group $G$ is the connected component containing $e$. It is an open
normal subgroup. $\GL_n$ has two components, told apart by the sign of the determinant, and $\GL^+(n)=\{\det>0\}$ is the
identity component; likewise $\SO(n)\subset\Ort(n)$ and $T_m^+\subset T_m$. A subset $U$ is \textbf{symmetric} if
$U^{-1}=U$. Every neighbourhood of $e$ contains a symmetric one, and in a connected group a symmetric neighbourhood $U$
of $e$ generates the group as a monoid, so every element is a finite product $u_1\cdots u_k$ with $u_i\in U$.
\end{bgentry}

In merge-and-restart of a method of action type (\bgref{D.12}), each cycle applies one element of a set $S$. If $S$
contains a neighbourhood of the identity in a connected group $H$, repeated merging reaches all of $H$
(\thmref{Theorem}{V.3}(b)). HF's \citep{xmangrulkar2022peft} Cayley--Neumann OFT factors have $\det R\ge0$, so the
invertible ones lie in $\GL^+(b)$; they generate $\prod\GL^+(b)$ as a group, or $\prod\CO^+(2)$ when $b=2$.

\fbAusedin{complete invariants (\S\ref{sec:erlangen-invariants}), \thmref{Theorem}{V.3}, the corollaries of
  \thmref{Theorem}{V.3} (\S\ref{sec:base-rebase}).}

\begin{bgentry}{D.5}{Lie algebras; the commutator; $\gl$, $\so$, $\Sym$, $\wedge$}
The \textbf{Lie algebra} of a matrix group $G\subseteq\GL_n$ is $\mathfrak g=\{X:e^{tX}\in G\text{ for all }t\in\R\}$,
the tangent space of $G$ at $I$. It is a vector space closed under the \textbf{commutator} $[X,Y]=XY-YX$. The paper uses
$\gl_r$ (all $r\times r$ matrices, for $\GL_r$), $\so(n)$ (skew matrices $\Omega\T=-\Omega$, for $\Ort(n)$ and
$\SO(n)$), the diagonal matrices (for $T_m$), $\R I\oplus\so(n)$ (for $\CO(n)$) and $\mathfrak{sp}(2r)=\{X:X\T
J+JX=0\}$. For a subspace $V\subseteq\R^n$, $\so(V)$ consists of the skew maps of $V$ extended by $0$ on $V^\perp$. The
symmetric matrices $\Sym_r$ do not form a Lie algebra, since $[S,S']$ is skew. Write $v\wedge w=vw\T-wv\T\in\so(n)$,
$U\wedge V$ for the span of the $u\wedge v$ with $u\in U$, $v\in V$, and $L_{ab}=e_a\wedge e_b=E_{ab}-E_{ba}$. Then
$[L_{ac},L_{cb}]=L_{ab}$ for distinct $a,b,c$.
\end{bgentry}

For injective $W_0$, at HRA's paired initialisation $u_{2i-1}=u_{2i}=w_i$ ($i\le k=r/2$, with independent $w_i$ and
$U=\spanop\{w_i\}$ of dimension $k$), the first-order image is $S_1=W_0\cdot\{\Omega\in\so(n):P_{U^\perp}\Omega
P_{U^\perp}=0\}=W_0\cdot(U\wedge\R^n)$, a subspace of $W_0\cdot\so(n)$ of dimension $kn-k(k+1)/2$ (\bgref{C.13}). LOFT
\citep{zhao2026loft} trains a skew $E\in\so(r)$ inside a fixed $r$-dimensional input subspace. The bracket identity lets
rotations of overlapping coordinate blocks generate rotations of their union (\bgref{D.8}).

\fbAusedin{\thmref{Theorem}{II.7}, the LOFT example and LOFT fixes one leaf (\S\ref{sec:erlangen-loft}),
  \thmref{Theorem}{III.5}, \thmref{Theorem}{V.3}. \textsf{Reading:} \citet[Ch.~3]{xhall2015lie}.}

\begin{bgentry}{D.6}{the exponential map; one-parameter subgroups}
For $X\in\mathfrak g$, the curve $t\mapsto e^{tX}$ of matrix exponentials is a \textbf{one-parameter subgroup}, a smooth
homomorphism $(\R,+)\to G$, and every continuous one-parameter subgroup has this form for a unique $X$, its
\textbf{generator}. The exponential map $\mathfrak g\to G$ has derivative $\id$ at $0$, so it maps a neighbourhood of
$0$ diffeomorphically onto a neighbourhood of $I$. It is onto for $\SO(n)$ and not onto for $\GL^+(n)$, since
$\diag(-1,-2)$ is $e^X$ for no real $X$.
\end{bgentry}

The one-parameter subgroup $t\mapsto(Be^{-tX},e^{tX}A)$ of LoRA's gauge moves along a fibre without changing $BA$.
\thmref{Theorem}{III.5} builds its isometries as $k=e^{\tau\xi}$ from generators $\xi$ with $\Lambda^{-1}\xi$
antisymmetric. LOFT may use the exponential as its map $\so(r)\to\SO(r)$, and the Kempf--Ness argument (\bgref{D.30})
differentiates the norm along $e^{t\xi}$.

\fbAusedin{\thmref{Theorem}{III.5}, \thmref{Proposition}{III.7}, LOFT (\S\ref{sec:erlangen-loft}),
  \thmref{Theorem}{V.3}. \textsf{Reading:} \citet[Ch.~2]{xhall2015lie}.}

\begin{bgentry}{D.7}{the Cayley map and its Neumann truncation}
For skew $\Omega$ the eigenvalues are imaginary, so $I-\Omega$ is invertible, and the \textbf{Cayley map} is
\[C(\Omega)=(I+\Omega)(I-\Omega)^{-1}.\]
It lands in $\SO(n)$, and $C(\Omega)-I=2\Omega(I-\Omega)^{-1}$, so $\rk(C(\Omega)-I)=\rk\Omega$. Its expansion
$C(\Omega)=I+2\Omega+2\Omega^2+\cdots$ shows that its derivative at $0$ is $2\,\id$. It is a diffeomorphism from
$\so(n)$ onto the open dense set of rotations with no eigenvalue $-1$, with inverse $R\mapsto(R-I)(R+I)^{-1}$, so it
misses, for instance, $\diag(-1,-1,1,\dots,1)$.
\end{bgentry}

Exact-Cayley OFT and block OFT are orthogonal, so they preserve $K_{\mathrm{out}}=WW\T$ (\bgref{D.15}) and the spectrum.
HF PEFT's default (since version 0.18) truncates $(I-Q)^{-1}$ to $I+Q+Q^2+Q^3$, which gives
$R=C(Q)(I-Q^4)$. As $Q^4$ is symmetric and
commutes with $C(Q)$, $R\T R=(I-Q^4)^2\ne I$, and for small $Q$ each merge shrinks $K_{\mathrm{out}}$. The rank identity
is how \thmref{Theorem}{II.7}(c) identifies HRA's image germ with the cone of skew matrices of rank at most $r$.

\fbAusedin{the thesis (Exposé~\ref{sec:thesis}), \thmref{Corollary}{II.6}, \thmref{Theorem}{II.7}, the Gram lab of the
  companion site (\S\ref{sec:erlangen-invariants}), the corollaries of \thmref{Theorem}{V.3} (\S\ref{sec:base-rebase}).}

\begin{bgentry}{D.8}{Lie subgroups and subalgebras; the generated Lie algebra}
A \textbf{Lie subgroup} of $G$ is a subgroup with a Lie group structure for which the inclusion is an injective
immersion; it need not be closed. Connected Lie subgroups $H\subseteq G$ correspond one to one to Lie subalgebras
$\mathfrak h\subseteq\mathfrak g$, with $H$ the subgroup generated by $e^{\mathfrak h}$. The Lie algebra
\textbf{generated} by subalgebras $\mathfrak h_1,\dots,\mathfrak h_p$, written $\operatorname{Lie}\langle\mathfrak
h_1,\dots,\mathfrak h_p\rangle$, is the smallest subalgebra containing them, the span of their elements and all iterated
brackets of them.
\end{bgentry}

Block OFT, BOFT \citep{liu2023parameter} factors and Givens rotations \citep{ma2024parameter} use $\so(I_j)$ for
coordinate blocks $I_j\subseteq\{1,\dots,n\}$. Since $[L_{ac},L_{cb}]=L_{ab}$, two overlapping blocks generate $\so$ of
their union, and $\operatorname{Lie}\langle\so(I_j)\rangle_j=\bigoplus_C\so(C)$, summed over the connected components
$C$ of the hypergraph with vertices $\{1,\dots,n\}$ and hyperedges $I_j$. Two vertices share a component when a chain of
hyperedges, each meeting the next, joins them. Butterfly factors connect this hypergraph only at full depth.

\fbAusedin{\thmref{Theorem}{V.3}, Figure~\ref{fig:rebase}(b), the corollaries of \thmref{Theorem}{V.3}
  (\S\ref{sec:base-rebase}), LOFT fixes one leaf (\S\ref{sec:erlangen-loft}). \textsf{Reading:}
  \citet[Ch.~19]{xlee2013introduction}; \citet[Ch.~5]{xhall2015lie}.}

\begin{bgfact}{D.9}{Yamabe's theorem}
Every path-connected subgroup of a Lie group is a Lie subgroup. Consequently, if $H_1,\dots,H_p$ are connected Lie
subgroups of $G$, the subgroup they generate is the connected Lie subgroup $H$ with
$\operatorname{Lie}(H)=\operatorname{Lie}\langle\mathfrak h_1,\dots,\mathfrak h_p\rangle$ \citep{xyamabe1950arcwise}; a
short proof is in \citet{xgoto1969arcwise}.
\end{bgfact}

The subgroup generated by the $H_i$ is path-connected, because products of paths are paths, so the theorem makes it a
Lie subgroup and the correspondence of \bgref{D.8} names it. With \bgref{D.4}, this is why a re-merged action method
whose cycles contain a neighbourhood of the identity in each $H_i$ reaches exactly the connected subgroup with Lie
algebra $\operatorname{Lie}\langle\mathfrak h_1,\dots,\mathfrak h_p\rangle$, and why exact-Cayley block-orthogonal
factors reach $\prod_C\SO(C)$.

\fbAusedin{\thmref{Theorem}{V.3}, the corollaries of \thmref{Theorem}{V.3} (\S\ref{sec:base-rebase}).}

\begin{bgentry}{D.10}{monoids, units, monoid orbits, homomorphisms}
A \textbf{monoid} is a set with an associative product and an identity. Its invertible elements form its \textbf{group
of units}. The submonoid \textbf{generated} by a subset $S$ consists of all finite products $s_1\cdots s_k$ of elements
of $S$, the empty product being the identity. A monoid acts on a set when $1\cdot x=x$ and $s\cdot(t\cdot x)=(st)\cdot
x$, and the \textbf{monoid orbit} of $x$ is $\{s\cdot x\}$. A \textbf{homomorphism} of monoids preserves products and
the identity. Examples are $\mathrm{Diag}_m$, $M_2(\R)$ and $\operatorname{End}(\Theta)$ under multiplication, with
units $T_m$, $\GL_2(\R)$ and $\GL(\Theta)$, and finite multisets under disjoint union $\uplus$. Additive monoids
$\mathrm{Mon}(C)$ are in \bgref{B.14}.
\end{bgentry}

Merge-and-restart reaches the orbit of the monoid that one cycle generates, which is $\theta_0+\mathrm{Mon}(C)$ for an
additive method $\rho=\theta_0+\delta$ with shape $C=\Im\delta$ (\bgref{B.14}) and $\langle S\rangle\cdot\theta_0$ for a
method of action type (\bgref{D.12}) whose cycle applies elements of $S$. RoAd$_2$'s blocks range over $M_2(\R)$. For a
fixed query, attention's pair (total weight, weighted sum of values) is a homomorphism from multisets of key--value
pairs to $(\R_{\ge0}\times\R^{d_v},+)$ (\bgref{D.34}).

\fbAusedin{\thmref{Theorem}{V.3}, restarting (\S\ref{sec:base-rebase}), DoRA (\S\ref{sec:erlangen-dora}),
  \thmref{Proposition}{II.9}, \thmref{Theorem}{VI.5}, attention (\S\ref{sec:merge-attention}).}

\begin{bgfact}{D.11}{Cartan--Dieudonné theorem; Scherk's sharp form}
For nonzero $u\in\R^n$, the \textbf{reflection} $H_u=I-2uu\T/\|u\|^2$ fixes the hyperplane $u^\perp$ and has $\det
H_u=-1$. Every $R\in\Ort(n)$ is a product of at most $n$ reflections (Cartan--Dieudonné). In Scherk's sharp form, if
$k=\rk(R-I)$, then $R$ is a product of exactly $k$ reflections and of no fewer. Hence $\det R=(-1)^k$, and every
$R\in\SO(n)$ is a product of $\rk(R-I)$ reflections, an even number. See \citet{xscherk1950decomposition} and
\citet[Ch.~III]{xartin1957geometric}.
\end{bgfact}

\paragraph{The idea of the proof}
A product of $j$ reflections fixes an intersection of $j$ hyperplanes, so $\rk(R-I)\le j$. Conversely, pick $x$ with
$Rx\ne x$ and put $u=Rx-x$. Then $H_u$ swaps $Rx$ and $x$ and fixes every fixed vector of $R$, so $H_uR$ fixes one more
dimension, and induction on $k$ finishes.

HRA multiplies $r$ reflections, so $\rk(R-I)\le r$, and $\det R=1$ for even $r$. Scherk's form gives the converse. For
injective $W_0$ and even $r$, HRA reaches all of $(W_0+\Mr)\cap W_0\cdot\SO(n)$, padding with pairs $H_uH_u=I$, and
re-merged HRA reaches $W_0\cdot\SO(n)$ within $\lceil n/r\rceil$ cycles.

\fbAusedin{\thmref{Theorem}{II.7}, HRA is a meet (\S\ref{sec:erlangen-hra}), the corollaries of \thmref{Theorem}{V.3}
  (\S\ref{sec:base-rebase}).}

\subsubsection*{Actions, orbits and invariants}

\begin{bgentry}{D.12}{group actions}
A (left) \textbf{action} of a group $G$ on a set $X$ is a map $(g,x)\mapsto g\cdot x$ with $e\cdot x=x$ and
$g\cdot(h\cdot x)=(gh)\cdot x$. A \textbf{right} action $x\cdot g$ satisfies $x\cdot(gh)=(x\cdot g)\cdot h$, and $g\cdot
x:=x\cdot g^{-1}$ turns it into a left one. The action is \textbf{smooth} if $G\times X\to X$ is smooth, and
\textbf{linear} if $X$ is a vector space and each map $x\mapsto g\cdot x$ is linear.
\end{bgentry}

On weights, $W\mapsto WR$ is a right action of $\Ort(n)$ on the input side and $W\mapsto PW$ a left action of $\Ort(m)$
on the output side, and $(g,h)\cdot W=gWh^{-1}$ is a left action of $\GL_m\times\GL_n$. On LoRA's factors,
$g\cdot(B,A)=(Bg^{-1},gA)$ is a linear left action of $\GL_r$. A method is of \textbf{action type} when
$\rho(q)=\gamma(q)\cdot\theta_0$ for a map $\gamma$ into a group acting on $\Theta$, as for OFT and HRA. (This $\gamma$
follows the exposés and is unrelated to the cotangent signal $\gamma(t)$ of \bgref{C.2} and \bgref{D.26}.) (IA)$^3$ has
the same form with the monoid $\mathrm{Diag}_m$ in place of a group, and DoRA's magnitude applies that monoid to LoRA's
weight.

\fbAusedin{\thmref{Definition}{I.2}, the notation (Table~\ref{tab:notation}), Erlangen (\S\ref{sec:erlangen}),
  \thmref{Theorem}{II.5}, \thmref{Theorem}{III.5}.}

\begin{bgentry}{D.13}{orbits, stabilisers, free actions}
The \textbf{orbit} of $x$ is $G\cdot x=\{g\cdot x:g\in G\}$, its closure is the \textbf{orbit closure}, and its
\textbf{stabiliser} is the closed subgroup $G_x=\{g:g\cdot x=x\}$. For a smooth action of a Lie group, $g\mapsto g\cdot
x$ induces an injective immersion $G/G_x\to X$ with image $G\cdot x$, so the orbit is an immersed submanifold of
dimension $\dim G-\dim G_x$. The \textbf{infinitesimal stabiliser} $\{\xi\in\mathfrak g:\xi\cdot x=0\}$ is the Lie
algebra of $G_x$. The action is \textbf{free} if every stabiliser is trivial, and the \textbf{generic stabiliser} is the
stabiliser at a generic point, one in a suitable dense open set. Each map $x\mapsto g\cdot x$ carries an orbit onto
itself, so all its points look alike; orbits are \textbf{homogeneous}.
\end{bgentry}

$\GL_r$ acts freely on pairs with $\rk A=r$, since $gA=A$ forces $g=I$, so LoRA's generic fibre, a single orbit, has
dimension $r^2$. LoHa's joint gauge $\GL_{r_1}\times\GL_{r_2}\times T_m\times T_n$ has a one-dimensional generic
stabiliser, so its orbits have dimension $r_1^2+r_2^2+m+n-1$. By homogeneity, the two-sided torus orbit
$\{\diag(a)W_0\diag(b)\}$ has dimension $m+n-c(W_0)$ at each of its points, with $c(W_0)$ the number of connected
components of the support graph, the bipartite graph on the $m$ rows and $n$ columns with an edge $\{i,j\}$ where
$(W_0)_{ij}\ne0$. The images of (IA)$^3$ and RoAd$_1$ are orbit closures.

\fbAusedin{Erlangen (\S\ref{sec:erlangen}), \thmref{Theorem}{II.5}, \thmref{Proposition}{II.9}, the image germ
  (Exposé~\ref{sec:image}), \thmref{Proposition}{II.12}, Rosetta entries 6 and 11 (Exposé~\ref{sec:rosetta}).
  \textsf{Reading:} \citet[Ch.~7 and~21]{xlee2013introduction}.}

\begin{bgentry}{D.14}{Haar measure; invariant measures on homogeneous spaces}
A compact Lie group $K$ carries a unique invariant probability measure, its \textbf{Haar measure}. If $K$ acts
transitively on $X$, so that $X\cong K/K_x$ is a \textbf{homogeneous space}, then $X$ carries a unique $K$-invariant
probability measure, the image of Haar measure under $k\mapsto k\cdot x$. The Grassmannian $\Gr(r,n)$ of $r$-dimensional
subspaces of $\R^n$ (\bgref{C.15}) is a homogeneous space of $\Ort(n)$.
\end{bgentry}

If $A_0\in\R^{r\times n}$ has i.i.d.\ Gaussian entries, its law is unchanged by $A_0\mapsto A_0R$ for $R\in\Ort(n)$, and
$A_0$ has rank $r$ almost surely. So $\operatorname{row}A_0$ is an $\Ort(n)$-invariant random point of $\Gr(r,n)$, and
by uniqueness it is Haar-distributed. Vanilla LoRA with a Gaussian $A_0$ therefore picks a Haar-random point of
$\Gr(r,n)$, the space that classifies its stratum $\mathrm T_A$. Kaiming-uniform entries are not rotation invariant, so
under HF's default the point is not exactly Haar-distributed.

\fbAusedin{the three strata (\S\ref{sec:strata-three}), \thmref{Theorem}{II.2}. \textsf{Reading:}
  \citet[Ch.~3]{xmattila1995geometry}.}

\begin{bgentry}{D.15}{invariants and complete invariants}
A map $f:X\to Y$ is \textbf{invariant} under an action if $f(g\cdot x)=f(x)$ for all $g$ and $x$, so it is constant on
orbits. It is a \textbf{complete invariant} if moreover $f(x)=f(x')$ implies $x'\in G\cdot x$; it then identifies the
orbit space $X/G$ with the image of $f$.

\smallskip
{\centering\small\renewcommand{\arraystretch}{1.12}
\begin{tabularx}{\linewidth}{@{}XX@{}}
  \toprule
  \headrow \textbf{action on $W\in\R^{m\times n}$} & \textbf{complete invariant}\\
  \midrule
  $W\mapsto WR$, $R\in\Ort(n)$ (input side) & $K_{\mathrm{out}}=WW\T$\\
  $W\mapsto PW$, $P\in\Ort(m)$ (output side) & $K_{\mathrm{in}}=W\T W$\\
  $W\mapsto PWR$, $P\in\Ort(m)$, $R\in\Ort(n)$ & singular values with multiplicity\\
  $W\mapsto DW$, $D\in T_m^+$, $W$ without zero rows & the rays $\R_{>0}w_i$ of the rows\\
  $W\mapsto E\odot W$, $E\in(\R^\times)^{m\times n}$ & the zero pattern\\
  \bottomrule
\end{tabularx}\par}
\end{bgentry}

If a method's image lies in one orbit, every invariant of that action stays fixed through training and merging. This is
why exactly orthogonal methods never change $K_{\mathrm{out}}$ or the spectrum (\thmref{Corollary}{II.6}). Isomorphism
invariants of methods, such as $S_1$, $d(M)$ and the first-order deficit, play the same role one level up, since methods
with different values cannot be isomorphic.

\fbAusedin{\thmref{Theorem}{II.5}, complete invariants and the Gram lab (\S\ref{sec:erlangen-invariants}),
  \thmref{Corollary}{II.6}, \thmref{Observation}{I.6}, the seven borrowed words (Exposé~\ref{sec:analogy}).}

\begin{bgfact}{D.16}{first fundamental theorem for $\Ort(n)$; Witt's extension theorem}
(First fundamental theorem.) Every polynomial in the rows $w_1,\dots,w_m\in\R^n$ of $W$ that is invariant under
$W\mapsto WR$, $R\in\Ort(n)$, is a polynomial in the inner products $\langle w_i,w_j\rangle$, the entries of $WW\T$. For
a compact group acting linearly, invariant polynomials separate orbits, so $WW\T$ is a complete invariant. (Witt's
extension theorem, Euclidean case.) A linear isometry between two subspaces of $\R^n$ extends to an element of
$\Ort(n)$. See \citet[Ch.~II]{xweyl1939classical} and \citet[Ch.~5]{xgoodman2009symmetry}.
\end{bgfact}

\paragraph{Witt gives the orbit separation directly}
If $WW\T=W'{W'}\T$, the map $\sum_ic_iw_i\mapsto\sum_ic_iw_i'$ is well defined and isometric on the row spaces, because
$\|\sum_ic_iw_i\|^2=c\T WW\T c$. Extend it by an isometry between the orthogonal complements to $R\T\in\Ort(n)$; then
$W'=WR$.

This is item (1) of \thmref{Theorem}{II.5}, the reason exactly orthogonal input-side methods are pinned to
$K_{\mathrm{out}}(W_0)$. Applied to $W\T$, it gives $K_{\mathrm{in}}$ for the output side.

\fbAusedin{\thmref{Theorem}{II.5}, the known results (Exposé~\ref{sec:analogy}).}

\begin{bgentry}{D.17}{Klein's Erlangen programme}
\citet{xklein1872vergleichende} proposed to define a geometry by a space together with a group acting on it, and to
study the properties the group preserves. Euclidean geometry keeps lengths and angles, the invariants of rigid motions.
Affine geometry keeps parallelism and ratios along parallel lines but not lengths or angles, because its group is
larger. A larger group leaves fewer invariants.
\end{bgentry}

For a group-type method the reading is a theorem, because the image lies in one orbit and so the method can change
nothing the group preserves. For a cone such as LoRA's image the word only names its \textbf{covariance group}, the
largest $\mathcal G_M\subseteq\GL_m\times\GL_n$ with $\Im M(gW_0h^{-1})=g\,\Im M(W_0)\,h^{-1}$ (\bgref{D.18}). For LoRA
this is all of $\GL_m\times\GL_n$, for (IA)$^3$ on keys and values it is $\mathrm{Mon}_m\times\GL_n$, and for the PiSSA
split it is $\CO(m)\times\CO(n)$.

\fbAusedin{Erlangen and what ``Erlangen'' means here (\S\ref{sec:erlangen}), \thmref{Definition}{II.14}, the borrowed
  words (Exposé~\ref{sec:rosetta}), the seven borrowed words (Exposé~\ref{sec:analogy}). \textsf{Reading:}
  \citet[Ch.~4]{xsharpe1997differential}.}

\begin{bgentry}{D.18}{equivariance and invariance of maps, update rules and merge rules}
A map $F:X\to Y$ between spaces with $G$-actions is \textbf{equivariant} if $F(g\cdot x)=g\cdot F(x)$, and
\textbf{invariant} if $F(g\cdot x)=F(x)$. An update rule on parameters is equivariant under a subgroup $H$ of the gauge
if, started from $h\cdot q$ with correspondingly transformed optimizer state, it produces $h$ applied to the iterates
started from $q$. A rule that uses randomness is \textbf{equivariant in distribution} if $F(g\cdot x)$ and $g\cdot F(x)$
have the same law for every $x$.
\end{bgentry}

On LoRA's full-rank locus, with channel-uniform hyperparameters, the largest such subgroup of $\GL_r$ is $\Ort(r)$ for
SGD, $B_r$ for Adam and all of $\GL_r$ for undamped scaled GD (\bgref{D.23}, \thmref{Theorem}{III.12}). An equivariant
rule started from gauge-related points traces the same weights, since $\rho(h\cdot q)=\rho(q)$. Merge rules should be
invariant under each adapter's gauge. Task arithmetic \citep{ilharco2022editing} $\sum_iw_is_iB_iA_i$ is, while
LoraHub's \citep{huang2023lorahub} $(\sum_iw_iB_i)(\sum_jw_jA_j)$ is not. On a weight space $\R^N$, DARE's
\citep{yu2023language} random drop-and-rescale is equivariant under the invertible diagonal matrices $T_N$ for each
fixed mask, and under the monomial matrices $\mathrm{Mon}_N$ (\bgref{D.3}) in distribution.

\fbAusedin{\thmref{Definition}{II.14}, \thmref{Observation}{II.15}, \thmref{Theorem}{III.12},
  \thmref{Corollary}{III.13}, Figure~\ref{fig:gauge}, \thmref{Proposition}{V.4}.}

\begin{bgentry}{D.19}{multihomogeneous maps; multidegree}
Split the parameters into blocks $q=(q_1,\dots,q_k)$. A map $\delta$ is \textbf{multihomogeneous} of
\textbf{multidegree} $(d_1,\dots,d_k)$ if $\delta(c_1q_1,\dots,c_kq_k)=c_1^{d_1}\cdots c_k^{d_k}\,\delta(q)$ for all
real $c_i$. LoRA's $BA$ has multidegree $(1,1)$, and LoHa's $(B_1A_1)\odot(B_2A_2)$ has multidegree $(1,1,1,1)$.
\end{bgentry}

If some $d_i\ge1$, the image of $\delta$ contains $0$ and is closed under multiplication by every nonnegative scalar, so
it is a cone with vertex $0$ (\bgref{C.10}). For a method $\rho=\theta_b+s\,\delta$ started where $\delta(q_0)=0$, so
that $\theta_b=\theta_0$, the image is the cone $\theta_0+s\,\Im\delta$ with vertex $\theta_0$. For $c\in\R^k_{>0}$, the
change $q_i\mapsto c_iq_i$ together with $s\mapsto s\prod_ic_i^{-d_i}$ leaves $\rho$ unchanged. That rescaling is an
isomorphism of methods and the source of the hyperparameter torus (\bgref{D.20}).

\fbAusedin{the image germ (Exposé~\ref{sec:image}), \thmref{Theorem}{III.10}, the seven axes (\S\ref{sec:atlas-axes}).}

\begin{bgentry}{D.20}{orbit space of a positive torus}
Let $\R^k_{>0}$ act on positive variables $x\in\R^N_{>0}$ by $x_j\mapsto x_j\prod_ic_i^{a_{ji}}$, with a real weight
matrix $a=(a_{ji})$. In logarithms this is translation by the column space of $a$, so the orbits are parallel affine
subspaces and the orbit space is a vector space of dimension $N-\rk a$. The monomials $\prod_jx_j^{b_j}$ with $b\T a=0$
are invariant, and a basis of such vectors $b$ gives a complete set of invariants.
\end{bgentry}

Under the hypotheses of \thmref{Theorem}{III.10}, taken with Adam at $\varepsilon=0$, no weight decay or clipping and a
shared schedule (otherwise the per-block $\varepsilon$, decay and clipping thresholds rescale as well), the torus acts
on LoRA's hyperparameters by $s\mapsto s/(c_Ac_B)$, $\eta_A\mapsto c_A\eta_A$, $\eta_B\mapsto c_B\eta_B$,
$\sigma_A\mapsto c_A\sigma_A$, with $\sigma_A$ the scale of $A$'s initialisation, without changing the weight
trajectory. The invariants are $s\eta_A\eta_B$ and $\sigma_A/\eta_A$, so only two of the four knobs matter. LoRA+
\citep{hayou2024lora} with ratio $\lambda$ replaces $\eta_B$ by $\lambda\eta_B$, which changes the invariants exactly as
replacing $\alpha$ by $\lambda\alpha$ does. Under SGD the rates scale by $c_i^2$, and the invariants become
$s^2\eta_A\eta_B$ and $\sigma_A^2/\eta_A$.

\fbAusedin{\thmref{Theorem}{III.10}, \thmref{Corollary}{III.11}, Figure~\ref{fig:orbit}, Rosetta entry 23
  (Exposé~\ref{sec:rosetta}), \thmref{Prediction}{X.1}.}

\subsubsection*{Gauge, quotients and metrics}

\begin{bgentry}{D.21}{gauge symmetry of a parametrisation}
The \textbf{gauge group} of a method $\rho:Q\to\Theta$ is the group of diffeomorphisms $\varphi$ of $Q$ with
$\rho\circ\varphi=\rho$. If a Lie group $G$ acts on $Q$ with $\rho(g\cdot q)=\rho(q)$, each orbit lies inside a fibre
$\rho^{-1}(\theta)$ (\bgref{C.6}), so $d(M)\le\lvert M\rvert-\dim(\text{generic orbit})$. The action is \textbf{locally
transitive} on a fibre when its orbits are open in the fibre, that is, when orbits and fibre have the same dimension
there; on generic fibres this gives $d(M)=\lvert M\rvert-\dim(\text{generic orbit})$, and we then call $G$ the generic
gauge. Here the word means only this symmetry group; no connection is involved.
\end{bgentry}

LoRA's $\GL_r$ acts freely and locally transitively on full-rank fibres, so $d=r(m+n)-r^2=r(m+n-r)$, and $r^2$
parameters are gauge waste. HRA's generic fibres have dimension $r(r+1)/2$, while its $r$ scalings $u_i\mapsto c_iu_i$
sweep out only $r$ of those dimensions, so the scalings do not fill the fibres; \thmref{Theorem}{II.7}(a) describes the
rest of the gauge as non-linear. For $r=2$ the fibre through $(u_1,u_2)$ has dimension $3$, made up of the two scalings
and of rotating $u_1$ and $u_2$ by a common angle inside their plane, which keeps the angle between them and hence
$H_{u_1}H_{u_2}$, the rotation of that plane by twice the angle. The plane depends on the point, so this move is not a
linear action on parameters. The \textbf{Hurwitz moves} $(u_i,u_{i+1})\mapsto(H_{u_i}u_{i+1},u_i)$ preserve the product
of reflections, since $H_aH_b=H_{H_ab}H_a$, and they satisfy the braid relations, so the braid group on $r$ strands,
unrelated to the signed permutations $B_r$, acts on HRA's parameters. These moves are discrete and add no dimension to
the fibre.

\fbAusedin{\thmref{Definition}{I.5}, invariants of a method (\S\ref{sec:slice-invariants}), the notation
  (Table~\ref{tab:notation}), rank and dimension (\S\ref{sec:cut}), \thmref{Proposition}{II.12}, \thmref{Theorem}{II.7},
  the analogy ``Noether; gauge'' (\S\ref{sec:fibres-charges}), the seven borrowed words (Exposé~\ref{sec:analogy}).}

\begin{bgfact}{D.22}{quotients and descent; the quotient manifold theorem}
For an action of $G$ on $X$, the \textbf{orbit space} $X/G$ is the set of orbits, with the map $\pi:x\mapsto G\cdot x$.
A map $f:X\to Y$ that is constant on orbits factors uniquely as $f=\bar f\circ\pi$; this is \textbf{descent}. (Quotient
manifold theorem.) If a Lie group acts smoothly, freely and properly on a manifold $X$, then $X/G$ is a manifold of
dimension $\dim X-\dim G$, $\pi$ is a submersion, and $\bar f$ is smooth whenever $f$ is. Properness means that
$(g,x)\mapsto(g\cdot x,x)$ is a proper map, one under which preimages of compact sets are compact
\citep[Thm~21.10]{xlee2013introduction}.
\end{bgfact}

$\GL_r$ acts freely and properly on pairs $(B,A)$ of full rank, and $(B,A)\mapsto BA$ identifies the quotient with
$\MM_r$, of dimension $r(m+n)-r^2$. A rule for combining LoRA factors is a function of the adapters exactly when it is
constant on the fibres of $(B_i,A_i)\mapsto(B_iA_i)$, and a cometric on factors that depends only on $W$ descends to
$\MM_r$.

\fbAusedin{\thmref{Proposition}{V.4}, \thmref{Proposition}{III.4}, \thmref{Observation}{II.15}, when a change of chart
  does not matter (\S\ref{sec:fibres-naturality}).}

\begin{bgentry}{D.23}{metrics, cometrics and the pulled-back cometric}
A \textbf{metric} on a vector space $V$ is a positive-definite inner product, equivalently a symmetric positive-definite
map $g:V\to V^*$, that is, $\langle gv,v\rangle>0$ for $v\ne0$. A \textbf{cometric} is a symmetric positive semidefinite
map $K:V^*\to V$. It turns a covector, such as a gradient $\nabla_\theta L$, into a velocity $-K\nabla_\theta L$. A
metric gives the cometric $K=g^{-1}$, while a cometric may be degenerate. Metrics pull back along immersions, and
cometrics push forward along any smooth map. For instance, gradient flow $\dot q=-\Lambda\nabla_q(L\circ\rho)$ moves the
weights by $\dot\theta=-K_q\nabla_\theta L$, with
\[K_q=D\rho_q\,\Lambda\,D\rho_q\T.\]
\end{bgentry}

For LoRA with \textbf{block-scalar rates}, one rate per factor, $\Lambda=\diag(\eta_BI,\eta_AI)$ on $(B,A)$ and
$K(G)=s^2(\eta_BGA\T A+\eta_ABB\T G)$. Two points of one fibre give the same weight velocity for every loss exactly when
their cometrics agree. At points where $A$ has full row rank and $r<n$, this holds for LoRA's linear gauge and plain
gradient descent exactly along $\Ort(r)$. A rule descends to a cometric on weights when it gives the same weight
velocity at all points of each fibre; with a point-dependent preconditioner in place of $\Lambda$, undamped scaled GD
\citep{xtong2021accelerating,zhang2024riemannian} and the SGD variant of LoRA-Pro \citep{wang2024lorac} do, on the
full-rank locus. Scaled GD multiplies the $B$-gradient by $(AA\T)^{-1}$ and the $A$-gradient by $(B\T B)^{-1}$, and
undamped means that no ridge term $\delta I$ is added before inverting (\thmref{Proposition}{III.4}).

\fbAusedin{\thmref{Definition}{III.1}, \thmref{Observation}{III.2}, \thmref{Theorem}{III.3},
  \thmref{Proposition}{III.4}, the metric a method carries (\S\ref{sec:fibres-cometric}), the three invariants
  (Exposé~\ref{sec:thesis}).}

\begin{bgentry}{D.24}{Riemannian gradient flow; tangent projection; retraction}
On a submanifold $N\subseteq\Theta$ with the metric induced by the Frobenius inner product, the Riemannian gradient of
$L|_N$ at $\theta$ is the orthogonal projection $\Pi_{T_\theta N}\nabla L(\theta)$ of the ambient gradient onto the
tangent space, and the \textbf{Riemannian gradient flow} is $\dot\theta=-\Pi_{T_\theta N}\nabla L(\theta)$. A discrete
step along the tangent space leaves $N$, so it is followed by a \textbf{retraction}, a smooth map $R_\theta$ from a
neighbourhood of $0$ in $T_\theta N$ to $N$ with $R_\theta(0)=\theta$ and $DR_\theta(0)=\id$.
\end{bgentry}

At $W=W_0+X$ with $X\in\MM_r$, $U=\operatorname{col}X$ and $V=\operatorname{row}X$, the tangent space is
$\{D:P_{U^\perp}DP_{V^\perp}=0\}$ (\bgref{C.12}) and the projection is $P_UG+GP_V-P_UGP_V$. This is LoRA-Pro's
equivalent gradient, so the SGD variant of LoRA-Pro follows the Riemannian gradient flow on $W_0+\MM_r$. Truncating the
SVD of $X+\xi$ to rank $r$ is a retraction onto $\MM_r$, the one Riemannion \citep{bogachev2025lora} uses.

\fbAusedin{\thmref{Proposition}{III.4}, when a change of chart does not matter (\S\ref{sec:fibres-naturality}),
  \thmref{Theorem}{III.12}. \textsf{Reading:} \citet[Ch.~3 and~4]{xabsil2008optimization}.}

\begin{bgfact}{D.25}{distributions, Lie brackets and Frobenius' theorem}
A \textbf{distribution} $D$ of rank $k$ on an open set $U$ assigns to each point $\theta$ a $k$-dimensional subspace
$D_\theta$ of directions, smoothly in $\theta$; the word is unrelated to probability distributions. The \textbf{Lie
bracket} of vector fields is $[X,Y]=DY\cdot X-DX\cdot Y$; flowing along $X$, $Y$, $-X$, $-Y$ for time $t$ misses the
starting point by $t^2[X,Y]+O(t^3)$. $D$ is \textbf{involutive} if $[X,Y]$ lies in $D$ whenever $X$ and $Y$ do.
(Frobenius.) $D$ is involutive iff every point lies on a $k$-dimensional \textbf{integral manifold}, a submanifold whose
tangent space at each of its points $\theta$ is $D_\theta$. A maximal connected integral manifold is a \textbf{leaf}
\citep[Ch.~19]{xlee2013introduction}. The Frobenius algebras of \bgref{B.6} and the Frobenius norm share only the name.
\end{bgfact}

On $\R^3$, $D=\spanop\{\partial_x,\ \partial_y+x\,\partial_z\}$ is not involutive, since
$[\partial_x,\partial_y+x\,\partial_z]=\partial_z\notin D$, so no surface is tangent to $D$. A backward method with
anchor $a(\theta)$ chooses $D_\theta=\operatorname{im}a(\theta)$, and it reads as a forward method only along integral
manifolds. GaLore \citep{zhao2024galore} with a projector fixed on each period, LISA \citep{pan2024lisa} and MeZO
\citep{malladi2023fine} have constant distributions, which are involutive with affine leaves. Full-batch GaLore with
$T=1$, read as $D_W=\operatorname{Hom}(\R^n,U_r(\nabla L(W)))$ wherever a singular-value gap exists, already fails to be
involutive for $L=\tfrac12\|W\|_F^2$.

\fbAusedin{when a backward method is a forward one (\S\ref{sec:lens-integrability}), \thmref{Proposition}{IV.3}.}

\subsubsection*{Conservation and balance}

\begin{bgfact}{D.26}{Noether's theorem for gradient flows}
Let a Lie group act linearly on $Q=\R^N$ with $\rho(k\cdot q)=\rho(q)$, let $\Lambda\succ0$ be a constant rate matrix,
and let $\gamma(t)$ be any time-dependent covector on $\Theta$, a gradient or not. Along every solution of $\dot
q=-\Lambda\,D\rho_q\T\gamma(t)$, and for every generator $\xi$ of the action with $\Lambda^{-1}\xi$ symmetric, the
\textbf{charge}
\[J_\xi(q)=\tfrac12\langle q,\Lambda^{-1}\xi\,q\rangle\]
is constant \citep{kunin2020neural,zhao2022symmetries,marcotte2023abide}.
\end{bgfact}

\paragraph{The one-line proof}
Invariance of $\rho$ gives $D\rho_q(\xi q)=0$. Hence $\tfrac{d}{dt}J_\xi=\langle\Lambda^{-1}\xi q,\dot
q\rangle=-\langle\xi q,D\rho_q\T\gamma\rangle=-\langle D\rho_q\,\xi q,\gamma\rangle=0$.

For scalar LoRA $w=w_0+ba$ with rates $\eta_B,\eta_A$, the charge $b^2/\eta_B-a^2/\eta_A$ is conserved, so every
trajectory runs along a hyperbola, the Noether hyperbolas of Figure~\ref{fig:noether}(a). \citet{xnoether1918invariante}
derived conservation laws from symmetries of a Lagrangian; this version needs no Lagrangian and holds for any loss and
data. For LoRA with rates $\eta_B,\eta_A$, the generators $\xi_X(B,A)=(BX,-XA)$ pass the test exactly for symmetric $X$,
and the $r(r+1)/2$ charges are the entries of $\Phi=B\T B/\eta_B-AA\T/\eta_A$. Minibatches, LoRA dropout, global-norm
clipping and DoRA's detached norm keep the required form. Momentum and Adam do not, and under discrete SGD the charge
drifts, by an amount relative to the charge of order $\eta^2$ per step.

\fbAusedin{\thmref{Theorem}{III.5}, isometries and charges and the analogy ``Noether; gauge''
  (\S\ref{sec:fibres-charges}), Figure~\ref{fig:noether}, the things one can check (Appendix~\ref{app:repro}),
  \thmref{Prediction}{X.8}, the seven borrowed words (Exposé~\ref{sec:analogy}).}

\begin{bgentry}{D.27}{Cartan decomposition; polar decomposition}
The map $\vartheta(\xi)=-\xi\T$ is an automorphism of $\gl_r$ of order two, a \textbf{Cartan involution}. Its $+1$ and
$-1$ eigenspaces give the \textbf{Cartan decomposition} $\gl_r=\so(r)\oplus\Sym_r$,
$\xi=\tfrac12(\xi-\xi\T)+\tfrac12(\xi+\xi\T)$, with $[\so,\so]\subseteq\so$, $[\so,\Sym]\subseteq\Sym$ and
$[\Sym,\Sym]\subseteq\so$. Its group version is the \textbf{polar decomposition}, which says that $(k,S)\mapsto
k\,e^{S}$ is a diffeomorphism $\Ort(r)\times\Sym_r\to\GL_r$. Hence $g\mapsto g\T g$ identifies the cosets $\Ort(r)g$
with the positive-definite matrices, $\Ort(r)\backslash\GL_r\cong\Sym^+_r$.
\end{bgentry}

With block-scalar rates (\bgref{D.23}), the antisymmetric generators $X\in\so(r)$ of LoRA's gauge act by rotations
$(BO\T,OA)$ that preserve the rate metric, so the optimizer cannot feel them; they are \textbf{isometries}. The
symmetric generators $X\in\Sym_r$ give the conserved charges of \bgref{D.26}. For general rates the split needs
$\mathfrak g$ to be closed under $\xi\mapsto\Lambda\xi\T\Lambda^{-1}$. Gradient descent on LoRA is thus a family of
dynamics indexed by $\Ort(r)\backslash\GL_r$, and the Kempf--Ness argument needs only curves $e^{t\xi}$ with $\xi$
symmetric.

\fbAusedin{\thmref{Theorem}{III.5}, isometries and charges (\S\ref{sec:fibres-charges}), the three invariants and the
  thesis (Exposé~\ref{sec:thesis}), \thmref{Proposition}{III.7}, when a change of chart does not matter
  (\S\ref{sec:fibres-naturality}). \textsf{Reading:} \citet[\S VI.2]{xknapp2002lie}; \citet[\S 7.3]{xhorn2013matrix}.}

\begin{bgentry}{D.28}{balancedness; the conserved $B\T B-AA\T$}
A factorisation $W=BA$ is \textbf{balanced} if $B\T B=AA\T$, and rate-weighted balanced if $\Phi=B\T
B/\eta_B-AA\T/\eta_A=0$, that is, $B\T B=cAA\T$ with $c=\eta_B/\eta_A$. For a deep linear network $W=W_L\cdots W_1$ the
same word means $W_{j+1}\T W_{j+1}=W_jW_j\T$ for every $j$.
\end{bgentry}

Under gradient flow with block-scalar rates, $\Phi$ is conserved (\bgref{D.26}), so rate-weighted balancedness, once it
holds, holds forever; at equal rates the two notions agree. Zero-product starts $B_0A_0=0$ have $\langle B_0\T
B_0,A_0A_0\T\rangle_F=0$, which makes them maximally unbalanced for given norms of the two Gram matrices, since the
inner product of two positive semidefinite matrices is never negative. From $B_0=0$, conservation gives $B\T
B=\tfrac{\eta_B}{\eta_A}(AA\T-A_0A_0\T)$, so $AA\T$ never drops below $A_0A_0\T$ in the Loewner order. HF's PiSSA starts
with $B_0\T B_0=A_0A_0\T$. When $\Phi$ is isotropic, singular values of $\Delta W$ well above the crossover scale
$\sigma^*$ of \thmref{Theorem}{III.9}, set by the charge, the scale and the rates, follow the balanced dynamics of
\citet{arora2018optimization}.

\fbAusedin{isometries and charges (\S\ref{sec:fibres-charges}), \thmref{Proposition}{III.7},
  \thmref{Proposition}{III.8}, \thmref{Theorem}{III.9}, the bibliography. \textsf{Reading:}
  \citet{arora2018optimization}; \citet{du2018algorithmic}.}

\begin{bgentry}{D.29}{the moment map}
Let a group $G\subseteq\GL(V)$ that is closed under transpose act on a real inner-product space $V$, and split
$\mathfrak g=\mathfrak k\oplus\mathfrak p$ into its antisymmetric and symmetric elements. The (real) \textbf{moment map}
$\mu:V\to\mathfrak p$ is defined by
\[\langle\mu(v),\xi\rangle=\tfrac12\tfrac{d}{dt}\Big|_{t=0}\|e^{t\xi}v\|^2=\langle\xi v,v\rangle\qquad(\xi\in\mathfrak p).\]
It records how the norm changes along the symmetric directions of the orbit; the antisymmetric directions preserve the
norm.
\end{bgentry}

For LoRA's gauge, $\tfrac{d}{dt}\big|_{0}\big(\|Be^{-tX}\|^2+\|e^{tX}A\|^2\big)=-2\tr(X\mu)$ for symmetric $X$, with
$\mu=B\T B-AA\T$, so this $\mu$ is the moment map up to sign. Its zero set is the balanced slice, and at equal rates
$\eta_A=\eta_B=\eta$ it equals $\eta\Phi$, a multiple of the Noether charge.

\fbAusedin{the balanced slice (\S\ref{sec:fibres-balance}), \thmref{Proposition}{III.7}, \thmref{Corollary}{III.6},
  Figure~\ref{fig:noether}.}

\begin{bgfact}{D.30}{Kempf--Ness theorem, real form}
Let $G\subseteq\GL(V)$ be a real algebraic group, cut out by polynomial equations, that is closed under transpose (such
a group is \textbf{reductive}), for instance $\GL_r$ acting on LoRA factors, and let $K=G\cap\Ort(V)$. For $v\in V$, (i)
$g=e$ is a critical point of $g\mapsto\|g\cdot v\|^2$ iff $\mu(v)=0$, and then $v$ has least norm on $G\cdot v$; (ii)
$G\cdot v$ is closed iff it contains a point with $\mu=0$; (iii) on a closed orbit, the points with $\mu=0$ form a
single $K$-orbit; (iv) the closure of every orbit contains exactly one closed orbit
\citep{xkempf1979length,xbirkes1971orbits,xrichardson1990minimum}.
\end{bgfact}

\paragraph{The idea of the proof}
For symmetric $\xi$, $\|e^{t\xi}v\|^2=\sum_ie^{2t\lambda_i}\|v_i\|^2$ in an eigenbasis of $\xi$, which is convex in $t$.
By the polar decomposition every orbit point is $k\,e^{\xi}v$ with $k\in K$ norm-preserving, so critical points are
minima and two minima differ by an element of $K$.

For $m=n=r=1$, $t\in\R^\times$ acts by $t\cdot(b,a)=(b/t,ta)$ and $\mu=b^2-a^2$. The orbits are the hyperbolas
$ba=c\ne0$, which are closed and meet $\mu=0$ in the two points $\pm(b,a)$ with $\lvert b\rvert=\lvert a\rvert$, one
$\Ort(1)$-orbit; the two punctured axes, which are not closed and avoid $\mu=0$; and the origin. For LoRA in general,
$K=\Ort(r)$, and a full-rank orbit is closed, being a whole fibre of $(B,A)\mapsto BA$. The least-norm factorisations of
a fixed $BA$ are the balanced ones, they form one $\Ort(r)$-orbit, and there $\|B\|_F^2+\|A\|_F^2=2\|BA\|_*$. Weight
decay adds a norm to the loss (\thmref{Corollary}{III.6}). With learning-rate-coupled decay, which multiplies the decay
of each factor by its rate, $\dot\Phi=-2\lambda\mu$, so the equilibria lie on $\mu=0$, and $\mu$ decays exponentially at
equal rates. With per-time decay, the same $\lambda$ for both factors, $\dot\Phi=-2\lambda\Phi$, so $\Phi\to0$.

\fbAusedin{the balanced slice (\S\ref{sec:fibres-balance}), \thmref{Proposition}{III.7}, \thmref{Corollary}{III.6},
  combining (\S\ref{sec:base-combine}), the known results (Exposé~\ref{sec:analogy}), the classification
  (Exposé~\ref{sec:atlas}).}

\begin{bgfact}{D.31}{the closed orbit in each fibre of $(B,A)\mapsto BA$}
For $\GL_r$ acting on factor pairs, fix $W$ of rank $k\le r$. Among the pairs with $BA=W$, those with $\rk B=\rk A=k$
form a single orbit. It is closed, and it lies in the closure of every orbit in the fibre $\{BA=W\}$. Consequently a
continuous $\GL_r$-invariant function is constant on each fibre of $(B,A)\mapsto BA$. For a complex reductive group
acting on a vector space, every fibre of the quotient map likewise contains exactly one closed orbit
\citep[Ch.~1]{xmumford1994geometric}.
\end{bgfact}

For $m=n=r=1$ (\bgref{D.30}), the fibre $\{ba=0\}$ consists of the origin and the two punctured axes, and the origin,
its only closed orbit, lies in the closure of both axes. This is the step behind \thmref{Proposition}{V.4}(a). A
continuous rule for combining LoRA factors that is invariant under each adapter's $\GL_r$ is automatically a function of
the adapters $B_iA_i$, also off the full-rank locus. Continuity matters. The projector onto $\operatorname{row}A$,
defined where $\rk A=r$, is invariant, yet it differs at $(0,A)$ and $(0,A')$, which have the same product $0$.

\fbAusedin{\thmref{Proposition}{V.4}, combining (\S\ref{sec:base-combine}).}

\subsubsection*{Matrix and convexity tools}

\begin{bgfact}{D.32}{Ky Fan's maximum principle}
Let $S\in\R^{n\times n}$ be symmetric with eigenvalues $\lambda_1\ge\dots\ge\lambda_n$, and let $1\le k\le n$. Then
\[\max\{\tr(PSP\T):P\in\R^{k\times n},\ PP\T=I_k\}=\lambda_1+\dots+\lambda_k,\]
attained when the rows of $P$ span a top-$k$ eigenspace. For any matrix $X$ with singular values
$\sigma_1\ge\sigma_2\ge\cdots$, applying this to $S=X\T X$ gives $\max_P\|XP\T\|_F^2=\sigma_1^2+\dots+\sigma_k^2$, and
for square $X$ also $\|PXP\T\|_F^2\le\|XP\T\|_F^2\le\sigma_1^2+\dots+\sigma_k^2$. See \citet{xfan1949theorem} and
\citet[Ch.~III]{xbhatia1997matrix}.
\end{bgfact}

Among $A_0$ with $A_0A_0\T=cI$, LoRA's initial rate $\|GA_0\T\|_F^2$ under gradient flow is largest when
$\operatorname{row}A_0$ is the top-$r$ right singular space of $G$, which is LoRA-GA's \citep{wang2024lorab} choice for
$A_0$. For LOFT, the squared norm of the pulled-back gradient $P_r\skewop(W_0\T G)P_r\T$ is at most $2\sum_{k\le\lfloor
r/2\rfloor}\mu_k^2$, where $\pm i\mu_k$ are the eigenvalues of the skew matrix $\skewop(W_0\T G)$, whose singular values
are $\mu_1,\mu_1,\mu_2,\mu_2,\dots$. The floor appears because a skew matrix compressed to an odd-dimensional subspace
has a zero eigenvalue, and Cauchy interlacing bounds the rest.

\fbAusedin{LOFT (\S\ref{sec:erlangen-loft}), \thmref{Proposition}{II.3}.}

\begin{bgfact}{D.33}{Sylvester equations}
For $S\in\R^{p\times p}$, $T\in\R^{q\times q}$ and $C\in\R^{p\times q}$, the \textbf{Sylvester equation} $SX+XT=C$ has a
unique solution $X$ for every $C$ iff $S$ and $-T$ have no eigenvalue in common, since the eigenvalues of $X\mapsto
SX+XT$ are the sums $\lambda_i(S)+\lambda_j(T)$. If $S$ and $T$ are symmetric positive definite, this holds and
$X=\int_0^\infty e^{-tS}Ce^{-tT}\,dt$. The Bartels--Stewart algorithm \citep{xbartels1972algorithm} solves the equation
through Schur forms of $S$ and $T$ \citep[Ch.~4]{xhorn1991topics}.
\end{bgfact}

LoRA-Pro's equivalent gradient leaves a free $r\times r$ matrix $X$. The published method picks it by solving $B\T
B\,X+X\,AA\T=C$, with $C$ built from the gradient. On the full-rank locus both Gram matrices are positive definite, so
$X$ exists and is unique. The first-order weight velocity does not depend on $X$, while the resulting parameter update
is only $\Ort(r)$-equivariant.

\fbAusedin{\thmref{Theorem}{III.12}, when a change of chart does not matter (\S\ref{sec:fibres-naturality}),
  Figure~\ref{fig:gauge}.}

\begin{bgentry}{D.34}{the perspective map; convex combinations}
The \textbf{perspective map} is $\pi(Z,S)=S/Z$ on $\{(Z,S)\in\R\times\R^d:Z>0\}$. It is constant on rays,
$\pi(tZ,tS)=\pi(Z,S)$ for $t>0$, and it turns sums into convex combinations,
\[\pi(Z_1+Z_2,S_1+S_2)=\lambda\,\pi(Z_1,S_1)+(1-\lambda)\,\pi(Z_2,S_2),\qquad\lambda=\frac{Z_1}{Z_1+Z_2}.\]
\end{bgentry}

For a fixed query $q$, attention is $\mathrm{Attn}_q(K)=\pi(\Sigma_q(K))$, where
\[\Sigma_q(K)=\Big(\sum e^{\langle q,k\rangle/\sqrt d},\ \sum e^{\langle q,k\rangle/\sqrt d}\,v\Big)\]
is a monoid homomorphism on multisets of key--value pairs (\bgref{D.10}). With a prefix $P$ before the content $C$, the
identity above gives
\[\mathrm{Attn}_q(P\uplus C)=(1-\lambda_q)\,\mathrm{Attn}_q(C)+\lambda_q\,\mathrm{Attn}_q(P),\]
with $\mathrm{Attn}_q(P)$ in the convex hull $\operatorname{conv}V_P$ of the prefix values, so a prefix acts as a
parallel branch gated by its share of the mass.

\fbAusedin{\thmref{Theorem}{VI.5}, attention (\S\ref{sec:merge-attention}), Figure~\ref{fig:prefix}. \textsf{Reading:}
  \citet[\S 2.3.3 and \S 3.2.6]{xboyd2004convex}.}

\providecommand{\fbBusedin}[1]{\par\smallskip\noindent{\small\color{inkgray}{\sffamily\bfseries\color{tide}Used in.}\enspace #1\par}\medskip}

\section{Auxiliary results}\label{app:auxiliary}

This appendix quotes the results from outside the paper that the proofs in Appendix~\ref{app:proofs} use. Each is
stated in the form in which it is applied, with a precise source, and is followed by the places that use it. Several are
textbook facts that a specialist applies without comment. They are listed so that every step of Appendix~\ref{app:proofs}
can be traced to a statement with hypotheses. For a few elementary facts we give the one-line reason instead of a
reference. Background definitions (manifolds, germs, group actions, lenses) are in Appendix~\ref{app:background}, and
where an entry has a companion there we name it.

Three results of the main text are external results restated in the paper's conventions and proved there:
\thmref{Theorem}{II.10} (the min-cut bound, here \auxref{B.14}), \thmref{Theorem}{III.5} (Noether charges,
\auxref{B.29}) and \thmref{Proposition}{III.7}(a) (the Kempf--Ness slice, \auxref{B.26}). The entries below give their
source forms.

The entries are grouped by subject: smooth maps and limits (B.1--B.4), real algebraic and tame geometry (B.5--B.10),
matrix analysis (B.11--B.21), Lie groups and invariant theory (B.22--B.27), dynamics, optimisation and differentiation
(B.28--B.35), three elementary facts (B.36), and spectral projectors, Riemannian submanifolds and floating-point
scaling (B.37--B.39). Notation follows the exposés: $\Mr$ is the set of $m\times n$ matrices of
rank at most $r$, $\MM_r$ those of rank exactly $r$, $P_U$ the orthogonal projector onto a subspace $U$, and
$\lVert\cdot\rVert$ the Frobenius norm unless marked otherwise. Entries of Appendix~\ref{app:background} are cited as
``Background A.1'' to ``Background D.34''; the numbers of its Part~B are unrelated to the numbers B.1--B.39 used here.

\subsubsection{Smooth maps, fibre products and limits}

\begin{auxresult}{B.1}{rank, the rank theorem and submersions}{\citep[Prop.~4.1, Thm~4.12, Prop.~4.28,
Thm~5.12]{xlee2013introduction}}
Let $F:Q\to\Theta$ be a smooth map between manifolds of dimensions $N$ and $D$.
\begin{enumerate}[label=(\alph*)]
  \item For every $k$, the set $\{q:\rk dF_q\ge k\}$ is open. In particular the set where $\rk dF_q$ takes its maximal
    value is open and nonempty, and if $\rk dF_p=D$ then $F$ is a submersion on a neighbourhood of $p$.
  \item (Rank theorem.) If $\rk dF_q=k$ for all $q$ in a neighbourhood of $p$, there are charts centred at $p$ and at
    $F(p)$ in which $F$ reads $(x_1,\dots,x_N)\mapsto(x_1,\dots,x_k,0,\dots,0)$. Hence $p$ has a neighbourhood $V$ such
    that $F(V)$ is an embedded submanifold of dimension $k$ and $F^{-1}(F(p))\cap V$ is an embedded submanifold of
    dimension $N-k$.
  \item A submersion is an open map. In particular, if $\rk dF=D$ on a neighbourhood $U$ of $p$, then $F(U)$ contains
    an open neighbourhood of $F(p)$.
\end{enumerate}
\end{auxresult}
\noindent\emph{Reason for (a).} In coordinates, $\rk dF_q\ge k$ says that some $k\times k$ minor of the Jacobian is
nonzero at $q$, an open condition; Lee's Prop.~4.1 is the case $k=D$. Companion: \bgref{C.5}.
\fbBusedin{\thmref{Observation}{I.4}(b), where the lifts of both axes force $\rk d\iota_{p^*}=2$, the rank stays $2$
near $p^*$, and the image of a neighbourhood of $p^*$ contains an open subset of $\R^2$; \thmref{Definition}{I.5}
(with \auxref{B.9}); \thmref{Proposition}{II.12}(a) (fibre dimension $\lvert M\rvert-d(M)$ near points of maximal
rank); \thmref{Proposition}{III.4}(d) (rank $r$ is an open condition on the slice). The proofs of \thmref{Proposition}{I.8}, \thmref{Theorem}{II.7} and \thmref{Proposition}{IV.3} in Appendix~\ref{app:proofs} also cite it.}

\begin{auxresult}{B.2}{transversality and fibre products}{\citep[Cor.~5.30 and Thm~6.30]{xlee2013introduction};
\citep[\S1.5]{xguillemin1974differential}}
Let $\rho_M:Q_M\to\Theta$ and $\rho_N:Q_N\to\Theta$ be smooth, with $\dim Q_M=\lvert M\rvert$ and
$\dim Q_N=\lvert N\rvert$, and let $P=\{(q,q'):\rho_M(q)=\rho_N(q')\}\subseteq Q_M\times Q_N$.
\begin{enumerate}[label=(\alph*)]
  \item If $d\rho_M(T_qQ_M)+d\rho_N(T_{q'}Q_N)=T_\theta\Theta$ at every $(q,q')\in P$, then $P$ is an embedded
    submanifold of $Q_M\times Q_N$ of dimension $\lvert M\rvert+\lvert N\rvert-\dim\Theta$.
  \item Transversality at a single point of $P$ forces $\lvert M\rvert+\lvert N\rvert\ge\dim\Theta$.
  \item If $S\subseteq X$ is an embedded submanifold, every smooth map $T\to X$ with image in $S$ is smooth as a map
    into $S$. Consequently, whenever $P$ is an embedded submanifold of $Q_M\times Q_N$ (under (a), or when $\rho_M$ and
    $\rho_N$ are affine, so that $P$ is an affine subspace), $P$ with its two projections is the pullback of $\rho_M$ and
    $\rho_N$ in $\Diff$.
\end{enumerate}
\end{auxresult}
\noindent\emph{Reason.} $P=(\rho_M\times\rho_N)^{-1}(\Delta_\Theta)$, and $\rho_M\times\rho_N$ is transverse to the
diagonal $\Delta_\Theta$, of codimension $\dim\Theta$, exactly when the condition in (a) holds; Lee's Thm~6.30(a) gives
(a). Part (b) is a dimension count, since the left side of the condition has dimension at most
$\lvert M\rvert+\lvert N\rvert$. Companion: \bgref{C.7}.
\fbBusedin{\thmref{Observation}{I.4}(b), \thmref{Proposition}{I.8}(a), \thmref{Theorem}{II.7}(d). The proof of \thmref{Proposition}{III.4} in Appendix~\ref{app:proofs} also cites it.}

\begin{auxresult}{B.3}{limits in slices and coslices; right adjoints}{\citep[\S III.4, \S V.1 Exercise~1, \S V.5
Thm~1]{xmaclane1998categories}}
Let $\mathcal C$ be a category.
\begin{enumerate}[label=(\alph*)]
  \item In the slice $\mathcal C/X$, a product of $a:A\to X$ and $b:B\to X$ is a pullback $A\times_XB$ of $a$ and $b$
    in $\mathcal C$, with its map to $X$; one exists exactly when the other does.
  \item For an object $X$, the projection from the coslice $X/\mathcal C$ (the comma category $(X\downarrow\mathcal C)$)
    to $\mathcal C$ creates limits: every limit cone in $\mathcal C$ of the underlying diagram lifts to exactly one cone
    in $X/\mathcal C$ over the given diagram, and that cone is a limit.
  \item A functor that has a left adjoint preserves all limits that exist in its domain. The forgetful functors
    $\Set_*\to\Set$ and $\Diff_*\to\Diff$ have the left adjoint $X\mapsto X\sqcup\mathrm{pt}$.
\end{enumerate}
\end{auxresult}
\noindent\emph{Form used.} A product in the pointed slice $\Meth(\Theta,\theta_0)=\Diff_*/(\Theta,\theta_0)$ is a
pullback over $(\Theta,\theta_0)$ in $\Diff_*$ by (a). By (b) a pullback in $\Diff$ of the underlying maps lifts to
$\Diff_*$ with base point $(q_0,q_0')$, and by (c) any pullback in $\Diff_*$ is a pullback in $\Diff$, so it may be
tested with unpointed curves. Companions: \bgref{A.11}, \bgref{A.12}, \bgref{A.13}.
\fbBusedin{\thmref{Observation}{I.4}(b) (products in $\Meth$ as pullbacks, tested with unpointed curves). The proof of \thmref{Proposition}{I.8} in Appendix~\ref{app:proofs} also cites it.}

\begin{auxresult}{B.4}{smooth descent along submersions; quotient manifold theorem}{\citep[Thms~4.29, 4.30,
21.10]{xlee2013introduction}}
\begin{enumerate}[label=(\alph*)]
  \item Let $\pi:X\to Y$ be a surjective smooth submersion. A map $\bar f:Y\to Z$ is smooth iff $\bar f\circ\pi$ is
    smooth, and every smooth $f:X\to Z$ that is constant on the fibres of $\pi$ factors uniquely as $f=\bar f\circ\pi$
    with $\bar f$ smooth.
  \item If a Lie group $G$ acts smoothly, freely and properly on a manifold $X$, then $X/G$ has a unique smooth structure
    for which $\pi:X\to X/G$ is a submersion, and $\dim X/G=\dim X-\dim G$.
\end{enumerate}
\end{auxresult}
\noindent\emph{Instance.} On pairs $(B,A)$ of full rank, $(B,A)\mapsto BA$ is a surjective submersion onto $\MM_r$ whose
fibres are the $\GL_r$-orbits $\{(Bg^{-1},gA)\}$ (\auxref{B.5}(b)), so a function of the factors that is constant on
these orbits is a smooth function of $W$ on the full-rank locus. Companion: \bgref{D.22}.
\fbBusedin{\thmref{Proposition}{III.4}(d) (cometrics that are functions of $W$), \thmref{Proposition}{V.4}(a)
(descent of combination rules on the full-rank locus).}

\subsubsection{Real algebraic and tame geometry}

\begin{auxresult}{B.5}{determinantal varieties}{\citep[Lectures~9, 12 and 14]{xharris1992algebraic};
\citep[Thm~3.2]{xschneider2015convergence}}
Let $1\le r<\min(m,n)$.
\begin{enumerate}[label=(\alph*)]
  \item $\Mr\subseteq\R^{m\times n}$ is the common zero set of the $(r+1)\times(r+1)$ minors. It is closed and closed
    under all real scalings, so it is a closed cone with vertex $0$, of dimension $r(m+n-r)$.
  \item $\MM_r$ is open and dense in $\Mr$ and is an embedded submanifold of dimension $r(m+n-r)$. If $X\in\MM_r$ has
    column space $U$ and row space $V$, and $X=BA$ with $B\in\R^{m\times r}$, $A\in\R^{r\times n}$, then
    \begin{equation*}
      T_X\MM_r=\{\dot BA+B\dot A\}=\{D:P_{U^\perp}DP_{V^\perp}=0\},
    \end{equation*}
    and $(B,A)\mapsto BA$ is a submersion from the pairs of full rank onto $\MM_r$ whose fibres are the orbits
    $\{(Bg^{-1},gA):g\in\GL_r\}$.
  \item At a point $X$ of rank $s<r$, with column space $U$ and row space $V$, the tangent cone of $\Mr$ is
    $\{D:\rk(P_{U^\perp}DP_{V^\perp})\le r-s\}$, which is not a linear space. Hence $\Mr$ is not a $C^1$ submanifold near
    $X$, and $\mathrm{Sing}\,\Mr=\MM_{\le r-1}$, algebraically and in the $C^1$ sense.
\end{enumerate}
\end{auxresult}
\noindent\emph{Reason for the last claim of (c).} $P_{U^\perp}DP_{V^\perp}$ ranges over all $(m-s)\times(n-s)$
matrices, and since $r-s<\min(m-s,n-s)$ two of rank $r-s$ can sum to a matrix of larger rank. The tangent cone of a
$C^1$ submanifold is its tangent space, a linear space. Companion: \bgref{C.12}.
\fbBusedin{\thmref{Theorem}{II.1}(c), \thmref{Theorem}{II.2}(a), (b), (c), (e) (the image determines its vertex),
\thmref{Proposition}{II.3}, \thmref{Proposition}{II.8} (tangent space at $(I,X)$), \thmref{Proposition}{II.12},
\thmref{Proposition}{II.16} (the translation stabiliser of $\Mr$ is $\{0\}$), \thmref{Proposition}{III.4}(b), (d),
\thmref{Proposition}{V.4}(a). The proofs of \thmref{Theorem}{II.7} and \thmref{Proposition}{II.9} in Appendix~\ref{app:proofs} also cite it.}

\begin{auxresult}{B.6}{cones that are $C^1$ at their vertex}{elementary}
Let $C\subseteq\R^N$ be a closed set with $0\in C$ and $tC\subseteq C$ for every $t>0$. If $C$ is a $C^1$ embedded
submanifold near $0$, then $C$ equals the linear subspace $T_0C$. Consequently, the vertex of a closed cone that is not a
linear subspace is a singular point.
\end{auxresult}
\noindent\emph{Reason.} The tangent cone of $C$ at $0$, the set of limits of $t_kx_k$ with $x_k\in C$, $x_k\to0$ and
$t_k>0$, equals $C$ because $C$ is a closed cone, and the tangent cone of a $C^1$ submanifold at a point is its tangent
space. Companions: \bgref{C.9}, \bgref{C.10}.
\fbBusedin{\thmref{Definition}{I.5}, \thmref{Theorem}{II.2}(a), (c), \thmref{Theorem}{II.7}(c),
\thmref{Theorem}{II.11}(e) (the closed cone $\mathcal K_{\le k}$).}

\begin{auxresult}{B.7}{real normal forms; the skew-determinantal cone}{\citep[Ch.~2]{xhorn2013matrix}}
\begin{enumerate}[label=(\alph*)]
  \item Every $\Omega\in\so(n)$ is orthogonally similar to a block-diagonal matrix with $2\times2$ blocks
    $\mu_k\left(\begin{smallmatrix}0&-1\\1&0\end{smallmatrix}\right)$, $\mu_1\ge\mu_2\ge\dots>0$, and zeros. Hence
    $\rk\Omega$ is even, the nonzero eigenvalues are $\pm i\mu_k$, the singular values are $\mu_1,\mu_1,\mu_2,\mu_2,\dots$,
    and $\Omega^4$ is symmetric positive semidefinite with eigenvalues $\mu_k^4$. A skew matrix of odd size has $0$ as an
    eigenvalue.
  \item Every $R\in\SO(n)$ is orthogonally similar to a block-diagonal matrix with $2\times2$ rotation blocks of angles
    $\varphi_k\in(0,\pi]$ and ones. Then $\rk(R-I)$ is twice the number of blocks, $\mathrm{im}(R-I)$ is the sum of the
    rotation planes, and $R=e^X$ for a skew $X$ whose eigenvalues are $\pm i\varphi_k$ and $0$.
  \item For even $r$ with $2\le r\le n-2$, the set $\Sigma_r=\{\Omega\in\so(n):\rk\Omega\le r\}$ is a closed cone of
    dimension $r(n-r)+r(r-1)/2=rn-r(r+1)/2$. Its points of rank exactly $r$ form an open dense submanifold of that
    dimension, and $\Sigma_r$ is not a linear space.
\end{enumerate}
\end{auxresult}
\noindent\emph{Reason for (c).} A skew matrix of rank $r$ is a range $V\in\Gr(r,n)$ together with a nondegenerate skew
form on $V$, an open subset of $\so(r)$; and skew matrices of ranks $r$ and $2$ can sum to rank $r+2\le n$. Companion:
\bgref{C.13}.
\fbBusedin{\thmref{Corollary}{II.6} (the Cayley--Neumann factor), \thmref{Theorem}{II.7}(a), (c), (d) (the dimension count
and the density argument), \thmref{Theorem}{V.3}, corollary~4, the LOFT example of Exposé~II.B.}

\begin{auxresult}{B.8}{Segre cones}{\citep[Lecture~2]{xharris1992algebraic}}
The Segre cone $\MM_{\le1}(p\times q)=\{ab\T:a\in\R^p,b\in\R^q\}$ is a closed cone of dimension $p+q-1$, it is not a
linear space when $p,q\ge2$, and it spans $\R^{p\times q}$. Hence $k$ generic rank-one matrices span a space of
dimension $\min(k,pq)$: the parameters $(a_1,b_1,\dots,a_k,b_k)$ for which $a_1b_1\T,\dots,a_kb_k\T$ span less form a
proper algebraic subset of $(\R^p\times\R^q)^k$.
\end{auxresult}
\noindent\emph{Reason for the last claim.} Spanning dimension at least $d$ is the nonvanishing of some $d\times d$
minor of the coefficient matrix, a polynomial in the parameters, and distinct matrix units $E_{ij}$ show that it is not
identically zero for $d=\min(k,pq)$. Companion: \bgref{C.14}.
\fbBusedin{\thmref{Proposition}{II.8} (the $m$ points $(E\T e_i)\otimes(F\T w_{0,i})$ give the generic value of
$d(\mathrm{DoRA}_r)$), \thmref{Proposition}{II.12} (the KronA image), the VeRA card of Exposé~VII.}

\begin{auxresult}{B.9}{semialgebraic and definable sets; dimension}{\citep[Ch.~2 and 9]{xbochnak1998real};
\citep[Ch.~4, 6 and 7]{xvandendries1998tame}; \citep{xwilkie1996model}}
\begin{enumerate}[label=(\alph*)]
  \item (Tarski--Seidenberg.) The image of a semialgebraic set under a semialgebraic map is semialgebraic.
    (Wilkie.) The expansion $\R_{\exp}$ of the real field by the exponential function is o-minimal, so sets and maps
    defined by first-order formulas from polynomials and $\exp$ (softmax included) are definable, and images of
    definable sets under definable maps are definable.
  \item Every definable set $X$ is a finite disjoint union of $C^1$ embedded submanifolds (cells), and $\dim X$, the
    largest of their dimensions, does not depend on the decomposition. If $X\subseteq Y$ then $\dim X\le\dim Y$, and
    $\dim f(X)\le\dim X$ for every definable map $f$.
  \item (Definable choice and generic smoothness.) A definable surjection has a definable section, and a definable map
    is $C^1$ outside a definable subset of its domain of smaller dimension.
  \item Consequently, for a definable $C^1$ map $\rho$ on an open definable set $Q\subseteq\R^k$,
    \begin{equation*}
      \dim\rho(Q)=\max_{q\in Q}\rk d\rho_q .
    \end{equation*}
\end{enumerate}
\end{auxresult}
\noindent\emph{Reason for (d).} At a point of maximal rank, the rank theorem (\auxref{B.1}(b)) puts a sheet of that
dimension inside $\rho(Q)$. Conversely, by (c) a definable section $s$ of $\rho$ is $C^1$ on an open subset $S$ of a
top-dimensional cell of $\rho(Q)$, and $\rho\circ s=\id_S$ gives $\rk d\rho_{s(y)}\ge\dim S=\dim\rho(Q)$. Companions:
\bgref{C.17}, \bgref{C.18}.
\fbBusedin{\thmref{Definition}{I.5} ($d(M)=\dim\Im M=\max_q\rk d\rho_q$), the obstruction $d(M)>d(N)$ of
Exposé~VIII.C, \thmref{Proposition}{II.12}(a). The proofs of \thmref{Theorem}{II.7}, \thmref{Proposition}{II.8}, \thmref{Proposition}{II.9}, \thmref{Proposition}{II.13} and \thmref{Theorem}{V.3} in Appendix~\ref{app:proofs} also cite it.}

\begin{auxresult}{B.10}{genericity; zero sets; the identity theorem}{\citep{xmityagin2020zero};
\citep[Ch.~1 and 2]{xkrantz2002primer}}
\begin{enumerate}[label=(\alph*)]
  \item Let $U\subseteq\R^N$ be open and connected and $f:U\to\R$ real-analytic, for instance a polynomial. If $f$
    vanishes on a nonempty open subset of $U$, then $f\equiv0$. If $f\not\equiv0$, its zero set is closed, nowhere dense
    and of Lebesgue measure zero.
  \item Hence a property that fails only on the common zero set of finitely many real-analytic functions on a connected
    domain, not all identically zero, holds on an open dense set of full measure, and with probability one under every
    law with a density. To prove such genericity it suffices to write failure as the vanishing of these functions and to
    exhibit one point where one of them does not vanish. A countable union of such exceptional sets still has measure
    zero.
  \item A function equal to a convergent power series $\sum_ka_kt^k$ near $0$ determines its coefficients. In
    particular, if $\sigma(ct)=c\,\sigma(t)$ near $0$ for some $c\ne0$, then $ca_k=c^ka_k$ for every $k$.
\end{enumerate}
\end{auxresult}
\noindent Companions: \bgref{C.21}, \bgref{C.22}.
\fbBusedin{\thmref{Theorem}{II.2}(d) (a generic rank-one update), \thmref{Proposition}{II.8}, \thmref{Theorem}{II.11}(c)
(generic Schmidt ranks), \thmref{Proposition}{II.12} (the generic rank of GraLoRA), \thmref{Proposition}{V.2}(a) ($e$
has full rank for almost every $W$, cell by cell of the quantiser), \thmref{Lemma}{VI.2}(c), \thmref{Theorem}{VI.3}
(the GELU obstruction), \thmref{Theorem}{VI.5}(f) (prefixes with affine output form a nowhere-dense analytic subset). The proofs of \thmref{Proposition}{I.8}, \thmref{Theorem}{II.1}, \thmref{Corollary}{II.6}, \thmref{Theorem}{II.7}, \thmref{Proposition}{II.13}, \thmref{Proposition}{IV.3} and \thmref{Theorem}{V.3} in Appendix~\ref{app:proofs} also cite it.}

\subsubsection{Matrix analysis}

\begin{auxresult}{B.11}{singular value decomposition; Eckart--Young--Mirsky}{\citep{xeckart1936approximation,
xmirsky1960symmetric}; \citep[Ch.~2 and 7]{xhorn2013matrix}}
\begin{enumerate}[label=(\alph*)]
  \item Every $X\in\R^{m\times n}$ of rank $k$ has a singular value decomposition $X=\sum_{i\le k}\sigma_iu_iv_i\T$ with
    $\sigma_1\ge\dots\ge\sigma_k>0$ and orthonormal families $(u_i)$, $(v_i)$; two matrices lie in one orbit of
    $(P,R)\mapsto PXR$, $P\in\Ort(m)$, $R\in\Ort(n)$, iff they have the same singular values. Applied to the
    $m_1\times m_2$ reshape of $u\in\R^{m_1}\otimes\R^{m_2}$ ($x\otimes y\mapsto xy\T$), the decomposition is the Schmidt
    decomposition $u=\sum_i\sigma_ix_i\otimes y_i$, whose number of terms, the Schmidt rank, is the rank of the reshape.
    Every matrix of rank at most $r$ factors as $QV$ with $Q\in\R^{m\times r}$ having orthonormal columns.
  \item (Eckart--Young--Mirsky.) For $0\le s<k$ the truncation $[X]_s=\sum_{i\le s}\sigma_iu_iv_i\T$ minimises
    $\lVert X-Y\rVert$ over $\rk Y\le s$, with minimum $\big(\sum_{i>s}\sigma_i^2\big)^{1/2}$, and the minimiser is unique
    iff $\sigma_s>\sigma_{s+1}$. The matrices $[X]_s$ and $X-[X]_s$ are Frobenius-orthogonal, so for every real $c$
    \begin{equation*}
      \lVert X-c[X]_s\rVert^2=\lVert X-[X]_s\rVert^2+(1-c)^2\lVert[X]_s\rVert^2 .
    \end{equation*}
  \item (Reduced-rank regression.) For $L\in\GL_n$, $\arg\min_{\rk Y\le s}\lVert(X-Y)L\rVert=[XL]_sL^{-1}$, unique iff
    $\sigma_s(XL)>\sigma_{s+1}(XL)$; and $[XL]_s=\Pi_sXL$, where $\Pi_s$ is the orthogonal projector onto the top-$s$
    eigenspace of $XLL\T X\T$.
\end{enumerate}
\end{auxresult}
\noindent\emph{Reason for (c).} Substitute $Z=YL$, which has the same rank as $Y$, and apply (b) to $XL$; the left
singular vectors of $XL$ are the eigenvectors of $XLL\T X\T$.
\fbBusedin{\thmref{Proposition}{I.8}(a) (the SVD frames of LoRA-XS), \thmref{Theorem}{II.1}(c) (splitting an SVD),
\thmref{Theorem}{II.5}(3), \thmref{Theorem}{II.11}(c) (Schmidt decompositions), \thmref{Proposition}{II.16}(a) and the
lemma in (b), \thmref{Proposition}{III.7}(a) (the balanced representative), \thmref{Theorem}{III.9} (compact SVD),
\thmref{Proposition}{V.2}(b), (c) (LoftQ's rank-$r$ step and its defect), \thmref{Proposition}{VI.4}(c) (the factorisation
$R\T V$), Prediction~X.5 (the distance to $k$-term Kronecker sums). The proofs of \thmref{Proposition}{II.8} and \thmref{Theorem}{V.3} in Appendix~\ref{app:proofs} also cite it.}

\begin{auxresult}{B.12}{rank inequalities; the Hadamard rank bound}{\citep[Ch.~4 and 5]{xhorn1991topics}}
For matrices of compatible sizes:
\begin{enumerate}[label=(\alph*)]
  \item $\rk(X+Y)\le\rk X+\rk Y$ and $\rk(XY)\le\min(\rk X,\rk Y)$;
  \item $\rk(C\otimes D)=\rk C\cdot\rk D$;
  \item (Hadamard rank bound) $\rk(X\odot Y)\le\rk X\cdot\rk Y$. Explicitly, if $X=B_1A_1$ and $Y=B_2A_2$ with inner
    dimensions $r_1$ and $r_2$, then
    \begin{equation*}
      X\odot Y=(B_1\bullet B_2)\,(A_1\T\bullet A_2\T)\T ,
    \end{equation*}
    where row $i$ of the face-splitting product $B_1\bullet B_2\in\R^{m\times r_1r_2}$ is the Kronecker product of the
    rows $i$ of $B_1$ and $B_2$. In particular $W_0\odot(BA)=\sum_{k\le r}\diag(b_k)\,W_0\,\diag(a_k)$, summed over the
    columns $b_k$ of $B$ and the rows $a_k$ of $A$, has rank at most $r\rk W_0$.
\end{enumerate}
\end{auxresult}
\noindent\emph{Reason for (c).} $X\odot Y$ is the submatrix of $X\otimes Y$ on the rows $(i,i)$ and the columns
$(j,j)$, so (b) applies. Companion: \bgref{B.13}.
\fbBusedin{\thmref{Theorem}{II.1}(c), \thmref{Proposition}{II.9}(a) (HiRA), \thmref{Theorem}{II.10} (subadditivity),
\thmref{Theorem}{II.11}(b), \thmref{Proposition}{II.12} (LoHa, GraLoRA), \thmref{Proposition}{II.13} (two HiRA cycles),
\thmref{Theorem}{V.3}, corollaries~1 and~5, the face-splitting arrow of rule A3 (Exposé~VIII.C).}

\begin{auxresult}{B.13}{the Van Loan--Pitsianis rearrangement; operator-Schmidt decomposition}{\citep{loan1993approximation};
\citep{xvanloan2000ubiquitous}}
Let $m=m_1m_2$ and $n=n_1n_2$, cut $\Delta\in\R^{m\times n}$ into an $m_1\times n_1$ grid of $m_2\times n_2$ blocks
$\Delta_{ij}$, and let $\mathcal R(\Delta)\in\R^{m_1n_1\times m_2n_2}$ have the rows $\vect(\Delta_{ij})\T$, ordered with
$i$ running fastest.
\begin{enumerate}[label=(\alph*)]
  \item $\mathcal R$ is a linear bijection that permutes entries, hence a Frobenius isometry, and
    $\mathcal R(C\otimes D)=\vect C\,\vect D\T$ for $C\in\R^{m_1\times n_1}$ and $D\in\R^{m_2\times n_2}$.
  \item Hence $\Delta=\sum_{i\le k}C_i\otimes D_i$ for some $C_i,D_i$ iff $\rk\mathcal R(\Delta)\le k$. The Kronecker
    rank $\rk_\otimes\Delta$ equals $\rk\mathcal R(\Delta)\le\min(m_1n_1,m_2n_2)$, and the nearest sum of $k$ Kronecker
    products in Frobenius norm is $\mathcal R^{-1}\big([\mathcal R(\Delta)]_k\big)$ (\auxref{B.11}(b)).
  \item The SVD of $\mathcal R(\Delta)$ gives the operator-Schmidt decomposition $\Delta=\sum_{k\le K}s_kC_k\otimes D_k$
    with $s_k>0$, Frobenius-orthonormal families $(C_k)$, $(D_k)$ and $K=\rk_\otimes\Delta$. Conversely, any expansion of
    this form with positive coefficients and Frobenius-orthonormal families is one, so its number of terms is
    $\rk_\otimes\Delta$.
\end{enumerate}
\end{auxresult}
\noindent Companions: \bgref{B.11}, \bgref{B.12}.
\fbBusedin{\thmref{Theorem}{II.11}(a)--(e), \thmref{Proposition}{II.13} (the Kronecker family as
$\mathcal R^{-1}_*\mathrm{LoRA}_k$), Prediction~X.5. The proof of \thmref{Proposition}{II.12} in Appendix~\ref{app:proofs} also cites it.}

\begin{auxresult}{B.14}{the min-cut bound for tensor networks}{\citep{xcui2016quantum}}
Let $\Gamma$ be a tensor network evaluated in $(\FinVect,\otimes)$, every wire attached to a box, and let
$\mathcal I\sqcup\mathcal O$ be a partition of its boxes. Contracting each side separately factors the value
$\llbracket\Gamma\rrbracket$ through the tensor product of the wires that cross the partition and of the open legs
attached on the wrong side, so $\rk\llbracket\Gamma\rrbracket$ is at most the product $w$ of their dimensions, and at most $\min w$ over all
partitions. For fixed bond dimensions the largest rank over all choices of tensors can be strictly smaller than the
minimal cut weight; equality holds, for instance, when every bond dimension is a power of one integer.
\end{auxresult}
\noindent\emph{Form used.} \thmref{Theorem}{II.10} states the bound with the paper's conventions (bare strands, copy
spiders counted once, formal sums) and proves it; only the upper bound is used. Companion: \bgref{B.8}.
\fbBusedin{\thmref{Theorem}{II.10} and its consequences, \thmref{Proposition}{II.12} (maximum ranks), the
``high-rank PEFT'' remark of Exposé~II.C.}

\begin{auxresult}{B.15}{manifolds of fixed Tucker and tensor-train rank}{\citep{xkoch2010dynamical}
(Tucker); \citep{xholtz2012manifolds} (TT); \citep{xuschmajew2013geometry} (hierarchical Tucker)}
\begin{enumerate}[label=(\alph*)]
  \item (Tucker.) For an admissible multilinear rank $(r_1,\dots,r_d)$ of tensors in $\R^{n_1\times\dots\times n_d}$,
    the tensors of exactly that rank form an embedded submanifold of dimension $\prod_kr_k+\sum_kr_k(n_k-r_k)$. In the
    parametrisation by a core in $\R^{r_1\times\dots\times r_d}$ and factor matrices of full column rank, with
    $\prod_kr_k+\sum_kn_kr_k$ parameters, the fibres are the orbits of $\prod_k\GL_{r_k}$, so the dimension is the
    parameter count minus $\sum_kr_k^2$.
  \item (Tensor train.) For an admissible TT rank $(r_1,\dots,r_{d-1})$, with $r_0=r_d=1$, the tensors of exactly that
    rank form an embedded submanifold of dimension $\sum_{k=1}^dr_{k-1}n_kr_k-\sum_{k=1}^{d-1}r_k^2$, the number of core
    entries minus the dimension of the bond gauge $\prod_{k=1}^{d-1}\GL_{r_k}$.
\end{enumerate}
\end{auxresult}
\noindent Here admissible means realised by some tensor; at minimal ranks the bond gauge acts freely. Companion:
\bgref{B.9}.
\fbBusedin{\thmref{Proposition}{II.12}(b), row ``TT / Tucker''.}

\begin{auxresult}{B.16}{circulant matrices and the discrete Fourier transform}{\citep[Thm~3.1]{xgray2006toeplitz}}
\begin{enumerate}[label=(\alph*)]
  \item A circulant matrix with first column $w\in\R^b$ satisfies $\mathrm{circ}(w)=F^{-1}\diag(Fw)\,F$, with $F$ the
    discrete Fourier matrix. The powers $P^0,\dots,P^{b-1}$ of the cyclic shift are a basis of the circulants, with Gram
    matrix $bI$ in the Frobenius inner product.
  \item Over $\R$, since the eigenvalues of a real circulant at conjugate frequencies are conjugate, $\mathrm{circ}(w)$ is
    orthogonally similar to a block-diagonal matrix of $2\times2$ rotation--scaling blocks and $1\times1$ blocks at the
    self-conjugate frequencies.
  \item For a real coefficient $c$ and a frequency $(u,v)$, the real part of the inverse two-dimensional transform of
    $c\,e_{(u,v)}$ is the cosine pattern
    \begin{equation*}
      \mathrm{Re}\,\mathrm{ifft2}(c\,e_{(u,v)})_{j,l}=\tfrac{c}{mn}\cos\big(2\pi(uj/m+vl/n)\big).
    \end{equation*}
    It is the same for $(u,v)$ and $(-u,-v)$, and it is even under $(j,l)\mapsto(-j,-l)$, indices taken modulo $m$
    and $n$.
\end{enumerate}
\end{auxresult}
\fbBusedin{\thmref{Proposition}{II.12} (the FourierFT gauge from conjugate pairs), \thmref{Proposition}{II.13} (FourierFT
and C3A as masks in transported frames). The proof of \thmref{Theorem}{II.10} in Appendix~\ref{app:proofs} also cites it.}

\begin{auxresult}{B.17}{Ky Fan's maximum principle; interlacing}{\citep{xfan1949theorem}; \citep[Ch.~III]{xbhatia1997matrix};
\citep[Ch.~3]{xhorn1991topics}}
\begin{enumerate}[label=(\alph*)]
  \item (Ky Fan.) Let $S\in\R^{n\times n}$ be symmetric with eigenvalues $\lambda_1\ge\dots\ge\lambda_n$ and let
    $1\le k\le n$. Then
    \begin{equation*}
      \max\{\tr(PSP\T):P\in\R^{k\times n},\ PP\T=I_k\}=\lambda_1+\dots+\lambda_k ,
    \end{equation*}
    attained when the rows of $P$ span a top-$k$ eigenspace. For $S=X\T X$ this reads
    \begin{equation*}
      \max_P\lVert XP\T\rVert^2=\sigma_1(X)^2+\dots+\sigma_k(X)^2 .
    \end{equation*}
  \item (Interlacing for compressions.) If $P\in\R^{k\times n}$ and $Q\in\R^{l\times m}$ have orthonormal rows, then
    $\sigma_j(QXP\T)\le\sigma_j(X)$ for every $j$ and every $X\in\R^{m\times n}$; for symmetric $S$,
    $\lambda_j(PSP\T)\le\lambda_j(S)$.
  \item Consequently, if $S\in\so(n)$ has eigenvalues $\pm i\mu_1,\pm i\mu_2,\dots$ with $\mu_1\ge\mu_2\ge\dots$, then
    for every $P\in\R^{k\times n}$ with orthonormal rows
    \begin{equation*}
      \lVert PSP\T\rVert^2\le2\sum_{j\le\lfloor k/2\rfloor}\mu_j^2 ,
    \end{equation*}
    with equality when the row space of $P$ contains the invariant subspace of the top $\lfloor k/2\rfloor$ rotation
    pairs of $S$.
\end{enumerate}
\end{auxresult}
\noindent\emph{Reason for (c).} $PSP\T$ is skew of size $k$, so by \auxref{B.7}(a) its singular values are
$\nu_1,\nu_1,\nu_2,\nu_2,\dots$, with an extra $0$ when $k$ is odd. By (b), $\nu_j=\sigma_{2j-1}(PSP\T)\le
\sigma_{2j-1}(S)=\mu_j$, and $\lVert PSP\T\rVert^2=2\sum_{j\le\lfloor k/2\rfloor}\nu_j^2$. If the row space contains an
invariant subspace $W$ of the top pairs, then $W^\perp$ is invariant too, and the compression is $S|_W$ plus zeros.
Companion: \bgref{D.32}.
\fbBusedin{\thmref{Proposition}{II.3} (alignment: among $A_0$ with $A_0A_0\T=cI$ the rate $\lVert GA_0\T\rVert^2$ is
largest on the top-$r$ right singular space of $G$), the LOFT example of Exposé~II.B, part (d).}

\begin{auxresult}{B.18}{singular values of products; the Loewner order}{\citep[Ch.~3]{xhorn1991topics};
\citep[Ch.~III]{xbhatia1997matrix}; \citep[Ch.~7]{xhorn2013matrix}}
\begin{enumerate}[label=(\alph*)]
  \item (Horn.) $\sigma_{i+j-1}(XY)\le\sigma_i(X)\,\sigma_j(Y)$. In particular
    \begin{equation*}
      \sigma_i(XY)\le\sigma_i(X)\,\lVert Y\rVert_{\rm op},\qquad\sigma_i(XY)\le\lVert X\rVert_{\rm op}\,\sigma_i(Y),
    \end{equation*}
    and a product of contractions (matrices with $\lVert\cdot\rVert_{\rm op}\le1$) is a contraction.
  \item If $\lVert Y\rVert_{\rm op}\le1$, then $Y\T Y\preceq I$, and $WY\T YW\T\preceq WW\T$ for every $W$.
  \item (Weyl monotonicity.) If $S\preceq T$ are symmetric, then $\lambda_i(S)\le\lambda_i(T)$ for every $i$. With (b),
    $\sigma_i(WY\T)\le\sigma_i(W)$ whenever $\lVert Y\rVert_{\rm op}\le1$.
\end{enumerate}
\end{auxresult}
\fbBusedin{\thmref{Corollary}{II.6} (the Cayley--Neumann factor contracts for small $Q$), \thmref{Theorem}{V.3},
corollary~4 (each default-OFT merge decreases $K_{\rm out}$ in the Loewner order and lowers every singular value).}

\begin{auxresult}{B.19}{polar, Cartan and $KAK$ decompositions}{\citep[Ch.~2 and 7]{xhorn2013matrix}}
\begin{enumerate}[label=(\alph*)]
  \item (Polar.) Every $g\in\GL_r$ is uniquely $g=k\,e^S$ with $k\in\Ort(r)$ and $S\in\Sym_r$, and
    $(k,S)\mapsto k\,e^S$ is a diffeomorphism $\Ort(r)\times\Sym_r\to\GL_r$. Hence $g\mapsto g\T g=e^{2S}$ identifies the
    cosets $\Ort(r)g$ with the positive definite matrices, $\Ort(r)\backslash\GL_r\cong\Sym^+_r$.
  \item (Cartan.) $\gl_r=\so(r)\oplus\Sym_r$ through $\xi=\tfrac12(\xi-\xi\T)+\tfrac12(\xi+\xi\T)$, with
    $[\so,\so]\subseteq\so$, $[\so,\Sym]\subseteq\Sym$ and $[\Sym,\Sym]\subseteq\so$.
  \item ($KAK$.) Every $g\in\GL^+(b)$ is $k_1\diag(d)\,k_2$ with $k_1,k_2\in\SO(b)$ and $d\in\R^b_{>0}$.
\end{enumerate}
\end{auxresult}
\noindent\emph{Reason.} (a) combines the polar form $g=kP$, $P$ positive definite and unique, with the fact that
$\exp:\Sym_r\to\Sym_r^+$ is a diffeomorphism (spectral theorem). (c) is the SVD with signs absorbed: if both singular
frames have determinant $-1$, flip the first column of each. Companion: \bgref{D.27}.
\fbBusedin{\thmref{Theorem}{III.3} (gradient descent on LoRA as a family indexed by $\Ort(r)\backslash\GL_r$),
\thmref{Theorem}{III.5}(iii), \thmref{Proposition}{III.7}(a) (reduction to $h(t)=f(e^{t\xi}\cdot x)$ with $\xi$ symmetric),
\thmref{Theorem}{V.3}, corollary~4 (the monoid generated by Cayley--Neumann factors).}

\begin{auxresult}{B.20}{Sylvester equations}{\citep{xbhatia1997how}; \citep[Ch.~4]{xhorn1991topics}}
For $S\in\R^{p\times p}$ and $T\in\R^{q\times q}$, the map $X\mapsto SX+XT$ on $\R^{p\times q}$ has the eigenvalues
$\lambda_i(S)+\lambda_j(T)$. So $SX+XT=C$ has a unique solution for every $C$ iff $S$ and $-T$ have no common
eigenvalue. If $S$ and $T$ are symmetric positive definite, the solution is $X=\int_0^\infty e^{-tS}Ce^{-tT}\,dt$.
\end{auxresult}
\noindent Companion: \bgref{D.33}.
\fbBusedin{\thmref{Proposition}{III.4} (practice: LoRA-Pro's Sylvester-optimal free matrix exists and is unique on the
full-rank locus), \thmref{Theorem}{III.12} (published variants). The proof of \thmref{Proposition}{IV.3} in Appendix~\ref{app:proofs} also cites it.}

\begin{auxresult}{B.21}{the nuclear norm as a factorisation norm}{\citep[Lemma~1]{xsrebro2004maximum};
see also \citet{xrecht2010guaranteed}}
For $X\in\R^{m\times n}$ and every $r\ge\rk X$,
\begin{equation*}
  \lVert X\rVert_*=\min\Big\{\tfrac12\big(\lVert B\rVert^2+\lVert A\rVert^2\big):B\in\R^{m\times r},\ A\in\R^{r\times n},
  \ BA=X\Big\}=\min_{BA=X}\lVert B\rVert\,\lVert A\rVert ,
\end{equation*}
attained at $B=U\Sigma^{1/2}O$, $A=O\T\Sigma^{1/2}V\T$ for an SVD $X=U\Sigma V\T$ padded to inner size $r$ and any
$O\in\Ort(r)$.
\end{auxresult}
\fbBusedin{\thmref{Proposition}{III.7}(a) (the minimum value $2\lVert BA\rVert_*$ on a full-rank orbit) and its commentary
(weight decay on the factors as a nuclear-norm penalty).}

\subsubsection{Lie groups and invariant theory}

\begin{auxresult}{B.22}{the Cartan--Dieudonné theorem in Scherk's sharp form}{\citep{xscherk1950decomposition};
\citep[Ch.~III]{xartin1957geometric}}
For $u\ne0$ let $H_u=I-2uu\T/\lVert u\rVert^2$, the reflection that fixes $u^\perp$; $\det H_u=-1$. A product of $j$
reflections fixes an intersection of $j$ hyperplanes, so it satisfies $\rk(R-I)\le j$. Every $R\in\Ort(n)$ is a product
of at most $n$ reflections (Cartan--Dieudonné). For the Euclidean form the bound is sharp (Scherk): if
$k=\rk(R-I)$, then $R$ is a product of exactly $k$ reflections and of no fewer. Hence $\det R=(-1)^{\rk(R-I)}$, and every
$R\in\SO(n)$ is a product of $\rk(R-I)$ reflections, an even number.
\end{auxresult}
\noindent\emph{Form used.} Scherk's theorem for a nondegenerate quadratic space adds two reflections when the moved space
$\mathrm{im}(R-I)$ is totally isotropic; a definite form has no nonzero isotropic vectors, so this case does not occur.
Companion: \bgref{D.11}.
\fbBusedin{\thmref{Theorem}{II.7}(a) ($\supseteq$, and the parity of $r$), \thmref{Theorem}{V.3}, corollary~5 (re-HRA).}

\begin{auxresult}{B.23}{the Cayley map}{\citep{xcayley1846quelques}; \citep{lezcanocasado2019cheap}}
Let $\Omega\in\so(n)$.
\begin{enumerate}[label=(\alph*)]
  \item $I-\Omega$ is invertible, since the eigenvalues of $\Omega$ are imaginary, and $C(\Omega)=(I+\Omega)(I-\Omega)^{-1}$
    lies in $\SO(n)$ and commutes with every polynomial in $\Omega$.
  \item $C(\Omega)-I=2\Omega(I-\Omega)^{-1}$, so $\rk(C(\Omega)-I)=\rk\Omega$; and $C(\Omega)=I+2\Omega+2\Omega^2+\dots$, so
    $dC_0=2\,\id$.
  \item $C$ is a diffeomorphism from $\so(n)$ onto the open dense subset $\{R\in\SO(n):-1\notin\mathrm{spec}\,R\}$, with
    inverse $R\mapsto(R-I)(R+I)^{-1}$. In particular it misses $\diag(-1,-1,1,\dots,1)$.
  \item Every element of $\SO(n)$ is a product of two elements of the image of $C$.
  \item (Neumann truncation.) $(I-Q)(I+Q+Q^2+Q^3)=I-Q^4$, so for skew $Q$
    \begin{equation*}
      R:=I+2Q+2Q^2+2Q^3+Q^4=C(Q)\,(I-Q^4),\qquad R\T R=(I-Q^4)^2 ,
    \end{equation*}
    because $Q^4$ is symmetric and commutes with the orthogonal matrix $C(Q)$; the singular values of $R$ are
    $\lvert1-\mu_k^4\rvert$, where $\pm i\mu_k$ are the eigenvalues of $Q$.
\end{enumerate}
\end{auxresult}
\noindent\emph{Reason for (d).} By \auxref{B.7}(b), $R=e^X$ with $X$ skew and eigenvalues $\pm i\varphi$,
$\varphi\in[0,\pi]$. Then $e^{X/2}$ has eigenvalues $e^{\pm i\varphi/2}\ne-1$, and $R=(e^{X/2})^2$. Companion:
\bgref{D.7}.
\fbBusedin{\thmref{Corollary}{II.6} (HF PEFT's default OFT), \thmref{Theorem}{II.7}(c) (the germ of the image at $W_0$)
and (d) (exact Cayley OFT never produces eigenvalue $-1$), \thmref{Theorem}{V.3}, corollaries~2 and~4, the LOFT example
of Exposé~II.B. The proof of \thmref{Proposition}{V.2} in Appendix~\ref{app:proofs} also cites it.}

\begin{auxresult}{B.24}{Lie subgroups, generation, and Yamabe's theorem}{\citep[Prop.~7.14, Thms~19.25,
19.26]{xlee2013introduction}; \citep{xyamabe1950arcwise}; \citep{xgoto1969arcwise}}
\begin{enumerate}[label=(\alph*)]
  \item In a connected Lie group every neighbourhood of the identity generates the group. A symmetric neighbourhood
    $U=U^{-1}$ therefore generates it as a monoid.
  \item For every Lie subalgebra $\mathfrak h$ of the Lie algebra of $G$ there is a unique connected Lie subgroup of $G$
    with Lie algebra $\mathfrak h$, namely the subgroup generated by $\exp\mathfrak h$. Lie subgroups are weakly
    embedded, so an inclusion $H\subseteq H'$ of Lie subgroups is smooth and gives $\mathrm{Lie}(H)\subseteq\mathrm{Lie}(H')$.
  \item (Yamabe.) Every arcwise connected subgroup of a Lie group is a Lie subgroup.
  \item Consequently, the subgroup generated by connected Lie subgroups $H_1,\dots,H_p$ of $G$ is the connected Lie
    subgroup $H$ with $\mathrm{Lie}(H)=\mathrm{Lie}\langle\mathfrak h_1,\dots,\mathfrak h_p\rangle$, the smallest Lie
    subalgebra containing every $\mathfrak h_i$.
\end{enumerate}
\end{auxresult}
\noindent\emph{Reason for (d).} The generated subgroup is arcwise connected, so it is a connected Lie subgroup $H$ by (c),
and $\mathrm{Lie}(H)\supseteq\mathfrak h_i$ for every $i$ by (b). The connected Lie subgroup $H'$ with Lie algebra
$\mathrm{Lie}\langle\mathfrak h_i\rangle$ contains every $H_i$, so $H\subseteq H'$, hence
$\mathrm{Lie}(H)=\mathrm{Lie}(H')$ and $H=H'$ by uniqueness. Companions: \bgref{D.4}, \bgref{D.8}, \bgref{D.9}.
\fbBusedin{\thmref{Theorem}{V.3}(b), (c) and corollary~4; the LOFT example of Exposé~II.B (supports revisited
across restarts).}

\begin{auxresult}{B.25}{first fundamental theorem for $\Ort(n)$; Witt's extension theorem}{\citep[Ch.~II]{xweyl1939classical};
\citep[Ch.~5]{xgoodman2009symmetry}; \citep[Ch.~III]{xartin1957geometric}}
\begin{enumerate}[label=(\alph*)]
  \item Every polynomial in the rows $w_1,\dots,w_m\in\R^n$ of $W$ that is invariant under $W\mapsto WR$, $R\in\Ort(n)$,
    is a polynomial in the inner products $\langle w_i,w_j\rangle$. Since the invariant polynomials of a compact group
    acting linearly separate its orbits, $WW\T$ is a complete invariant of the right $\Ort(n)$-action.
  \item (Witt, Euclidean case.) A linear isometry between two subspaces of $\R^n$ extends to an element of $\Ort(n)$.
\end{enumerate}
\end{auxresult}
\noindent\emph{Form used.} Part (b) gives the orbit statement of (a) directly: if $WW\T=W'W'^{\top}$, the map
$\sum_ic_iw_i\mapsto\sum_ic_iw_i'$ is a well-defined isometry of the row spaces, and its extension is $R\T$ with $W'=WR$.
Companion: \bgref{D.16}.
\fbBusedin{\thmref{Theorem}{II.5}(1), (2), \thmref{Corollary}{II.6}, \thmref{Corollary}{III.13} (left $\Ort(r)$-orbits of
full-rank $A_0$ are classified by $A_0\T A_0$).}

\begin{auxresult}{B.26}{the Kempf--Ness theorem in the real form used}{\citep{xkempf1979length};
\citep{xrichardson1990minimum}; \citep{xbirkes1971orbits}}
Let $V$ be a real inner-product space, $G\subseteq\GL(V)$ a real algebraic group that is closed under transpose,
$K=G\cap\Ort(V)$, and $\mathfrak p$ the symmetric elements of its Lie algebra. Define the moment map $\mu:V\to\mathfrak p$
by $\langle\mu(v),\xi\rangle=\tfrac12\tfrac{d}{dt}\big|_{t=0}\lVert e^{t\xi}v\rVert^2=\langle\xi v,v\rangle$. Then:
\begin{enumerate}[label=(\roman*)]
  \item $g=e$ is a critical point of $g\mapsto\lVert g\cdot v\rVert^2$ iff $\mu(v)=0$, and then $\lVert v\rVert\le
    \lVert g\cdot v\rVert$ for every $g\in G$;
  \item $G\cdot v$ is closed iff it meets $\mu^{-1}(0)$;
  \item if $G\cdot v$ is closed, $G\cdot v\cap\mu^{-1}(0)$ is a single $K$-orbit;
  \item the closure of every orbit contains exactly one closed orbit.
\end{enumerate}
\end{auxresult}
\noindent\emph{Form used.} For $\GL_r$ acting on $V=\R^{m\times r}\times\R^{r\times n}$ by $(Bg^{-1},gA)$ with the
Frobenius inner product, the image is closed under transpose (the adjoint of the action of $g$ is the action of
$g\T$), $K=\Ort(r)$, and $\tfrac12\tfrac{d}{dt}\big|_0(\lVert Be^{-t\xi}\rVert^2+\lVert e^{t\xi}A\rVert^2)=
\langle\xi,AA\T-B\T B\rangle$. So the moment map is $-\mu$ with $\mu=B\T B-AA\T$ as in \thmref{Proposition}{III.7}, its
zero set is the balanced slice, a full-rank orbit is closed (it is a whole fibre of $(B,A)\mapsto BA$), and by
\auxref{B.21} the least value of $\lVert B\rVert^2+\lVert A\rVert^2$ on it is $2\lVert BA\rVert_*$. Companions:
\bgref{D.29}, \bgref{D.30}.
\fbBusedin{\thmref{Proposition}{III.7}(a), \thmref{Corollary}{III.6} (decay as a Kempf--Ness potential),
\thmref{Proposition}{V.4}(a) (part (iv)).}

\begin{auxresult}{B.27}{closed orbits in the fibres of $(B,A)\mapsto BA$}{\citep[Ch.~5]{xgoodman2009symmetry};
\citep[Ch.~1]{xmumford1994geometric}; part (c) elementary}
\begin{enumerate}[label=(\alph*)]
  \item (First fundamental theorem for $\GL_r$.) The polynomial invariants of $\GL_r(\mathbb C)$ acting on
    $\mathbb C^{m\times r}\times\mathbb C^{r\times n}$ by $(Bg^{-1},gA)$ are generated by the entries of $BA$, the
    contractions of the rows of $B$ with the columns of $A$.
  \item (Reductive quotients.) For a complex reductive group acting on an affine variety, every fibre of the quotient
    map contains exactly one closed orbit, and it lies in the closure of every orbit in that fibre.
  \item (Real form used.) Fix $W\in\R^{m\times n}$ of rank $k\le r$. Among the real pairs with $BA=W$, those with
    $\rk B=\rk A=k$ form a single $\GL_r$-orbit $O_W$. It is closed and lies in the closure of every orbit in the fibre
    $\{BA=W\}$. Hence a continuous $\GL_r$-invariant function of $(B,A)$ is constant on each fibre of $(B,A)\mapsto BA$.
\end{enumerate}
\end{auxresult}
\noindent\emph{Reason for (c).} If $BA=W$ with $\rk B=\rk A=k$, then $\mathrm{col}\,B=\mathrm{col}\,W$ and
$\mathrm{row}\,A=\mathrm{row}\,W$; a gauge brings $B$ to $[U_0\ 0]$ for a fixed basis $U_0$ of $\mathrm{col}\,W$, which
fixes the first $k$ rows of $A$, and the stabiliser of $[U_0\ 0]$ clears the remaining rows, which lie in
$\mathrm{row}\,W$. The set $O_W=\{BA=W,\ \rk B\le k,\ \rk A\le k\}$ is closed. For a general pair in the fibre, split
$\R^r=C_1\oplus C_2\oplus C_3\oplus C_4$ with $\mathrm{im}\,A=C_1\oplus C_2$, $\ker B=C_2\oplus C_3$ and $\dim C_1=k$,
and let $g_t$ act by $1$, $t$, $1$, $t^{-1}$ on the four summands. As $t\to0$, $(Bg_t^{-1},g_tA)$ converges to a pair in
the fibre whose factors have rank at most $k$, so to a point of $O_W$. Companion: \bgref{D.31}.
\fbBusedin{\thmref{Proposition}{V.4}(a) (a continuous combination rule that is invariant under each adapter's $\GL_r$ is a
function of the adapters, also off the full-rank locus).}

\subsubsection{Dynamics, optimisation and differentiation}

\begin{auxresult}{B.28}{existence and uniqueness under Carathéodory conditions}{\citep[Ch.~I, Thms~5.1 and
5.3]{xhale1980ordinary}}
Let $D\subseteq\R^{N+1}$ be open and let $F:D\to\R^N$ be measurable in $t$ for each fixed $x$ and continuous in $x$ for
each fixed $t$, with $\lvert F(t,x)\rvert\le m_U(t)$ on each compact $U\subseteq D$ for an integrable $m_U$. Then through
every $(t_0,x_0)\in D$ there is an absolutely continuous solution of $\dot x=F(t,x)$, holding for almost every $t$. If
moreover $\lvert F(t,x)-F(t,y)\rvert\le k_U(t)\lvert x-y\rvert$ on each compact $U$ for an integrable $k_U$, the solution
through each point is unique.
\end{auxresult}
\noindent\emph{Form used.} If $\gamma$ is measurable in $t$ and locally Lipschitz in $\theta$, with locally integrable
bounds and Lipschitz constants, and $\rho$ is of class $C^{1,1}_{\rm loc}$, then $F(t,q)=-\Lambda\,D\rho_q\T\gamma(t,\rho(q))$
satisfies these conditions, so the trajectory from $kq_0$ is unique.
\fbBusedin{\thmref{Theorem}{III.5}(ii) (the trajectory statement).}

\begin{auxresult}{B.29}{Noether's theorem for gradient flows}{\citep{kunin2020neural}; \citep[Prop.~5.1]{zhao2022symmetries};
\citep{marcotte2023abide}; \citep{du2018algorithmic}}
Let a Lie group act linearly on $Q=\R^N$ with $\rho(k\cdot q)=\rho(q)$, and let $\Lambda\succ0$ be constant. Along every
solution of $\dot q=-\Lambda\,D\rho_q\T\gamma(t)$, where $\gamma(t)$ is any cotangent signal, gradient or not, and for every
generator $\xi$ of the action with $\Lambda^{-1}\xi$ symmetric, the charge
\begin{equation*}
  J_\xi(q)=\tfrac12\langle q,\Lambda^{-1}\xi q\rangle
\end{equation*}
is constant. For LoRA with block rates $\eta_B,\eta_A$ and generators $\xi_X(B,A)=(BX,-XA)$, $X$ symmetric, the charges
are the entries of $\Phi=B\T B/\eta_B-AA\T/\eta_A$.
\end{auxresult}
\noindent\emph{Form used.} The cited sources treat gradient flows, \citet{zhao2022symmetries} with $\Lambda=I$; their
proof uses only $D\rho_q(\xi q)=0$, so it covers arbitrary signals $\gamma$ and constant $\Lambda$, which is how
\thmref{Theorem}{III.5}(i) states and proves it. Companions: \bgref{D.26}, \bgref{D.28}.
\fbBusedin{\thmref{Theorem}{III.5}(i), \thmref{Corollary}{III.6}, \thmref{Proposition}{III.7}(b),
\thmref{Theorem}{III.9} (the charge stays at $-\kappa I$).}

\begin{auxresult}{B.30}{Riemannian gradients on embedded submanifolds}{\citep[\S3.6]{xabsil2008optimization}}
Let $N\subseteq\R^D$ be an embedded submanifold with the metric induced by the Euclidean (Frobenius) inner product, and
$f$ smooth near $N$. The Riemannian gradient of $f|_N$ at $x$ is $P_{T_xN}\nabla f(x)$, the orthogonal projection of the
Euclidean gradient onto the tangent space, and the Riemannian gradient flow is $\dot x=-P_{T_xN}\nabla f(x)$.
\end{auxresult}
\noindent\emph{Form used.} For $N=W_0+\MM_r$ at $W=W_0+X$, with $U=\mathrm{col}\,X$ and $V=\mathrm{row}\,X$,
\auxref{B.5}(b) gives the projection $G\mapsto P_UG+GP_V-P_UGP_V$. Companion: \bgref{D.24}.
\fbBusedin{\thmref{Proposition}{III.4}(b) (LoRA-Pro's equivalent gradient).}

\begin{auxresult}{B.31}{Frobenius' theorem}{\citep[Prop.~19.3, Thms~19.12 and 19.21]{xlee2013introduction}}
Let $D$ be a smooth distribution of constant rank $k$ on an open set $U$.
\begin{enumerate}[label=(\alph*)]
  \item If $N$ is an integral manifold of $D$ ($T_\theta N=D_\theta$ for $\theta\in N$) and $X,Y$ are sections of $D$,
    then $[X,Y]_\theta\in D_\theta$ for every $\theta\in N$. Hence an integrable distribution is involutive.
  \item (Frobenius.) Every involutive distribution is completely integrable: each point has a chart in which $D$ is
    spanned by the first $k$ coordinate fields. In such a chart a curve with $\dot\theta(t)\in D_{\theta(t)}$ keeps its last
    coordinates constant, so it stays in one integral manifold.
  \item (Global Frobenius.) The maximal connected integral manifolds of an involutive distribution form a foliation of
    $U$.
\end{enumerate}
\end{auxresult}
\noindent Companion: \bgref{D.25}.
\fbBusedin{\thmref{Proposition}{IV.3}.}

\begin{auxresult}{B.32}{flatness and the Bures--Wasserstein metric}{\citep[Ch.~7]{xlee2018introduction};
\citep{xtakatsu2011wasserstein}; \citep{xbhatia2019bures}}
\begin{enumerate}[label=(\alph*)]
  \item Riemannian curvature is preserved by local isometries, and a Riemannian manifold is flat iff it is locally
    isometric to an open subset of a Euclidean space.
  \item On the positive definite $n\times n$ matrices, the Bures--Wasserstein metric is the quotient of the Euclidean
    (Frobenius) metric under $B\mapsto BB\T$, whose fibres are the orbits $B\mapsto BR$, $R\in\Ort(n)$; up to a constant
    factor fixed by normalisation, $B\mapsto BB\T$ is a Riemannian submersion onto it. Its sectional curvature is
    nonnegative, and for $n\ge2$ it is positive on some planes, so the metric is not flat.
\end{enumerate}
\end{auxresult}
\noindent\emph{Form used.} By (a), an immersion $\rho$ with $\dim Q=\rk D$ that is a local isometry from the flat
$(Q,\Lambda^{-1})$ onto a leaf forces the leaf metric to be flat. The map $\rho(B)=BB\T$ has positive-dimensional fibres,
so it is no immersion; its cometric $D\rho\,D\rho\T(G)=2(G\theta+\theta G)$ at $\theta=BB\T$ does not depend on the point
of the fibre, and by (b) it is the cometric of the curved Bures--Wasserstein metric.
\fbBusedin{\thmref{Proposition}{IV.3} (dynamic equivalence with a forward lens forces a flat leaf metric when
$\dim Q=\rk D$).}

\begin{auxresult}{B.33}{the cost of forward and reverse automatic differentiation}{\citep[Ch.~3 and
4]{xgriewank2008evaluating}}
Let $L:\R^N\to\R$ be evaluated by a straight-line program.
\begin{enumerate}[label=(\alph*)]
  \item Forward mode evaluates $L(\theta)$ together with one directional derivative, $DL(\theta)[p]$, which equals
    $\langle\nabla L(\theta),p\rangle$, at a cost bounded by a small constant multiple of the cost of evaluating $L$. It
    carries each tangent alongside its value and stores no trace.
  \item Reverse mode evaluates $\nabla L(\theta)$ at a cost bounded by a small constant multiple of the cost of evaluating
    $L$, independent of $N$, but needs the intermediate values of the evaluation, stored or recomputed, during the
    reverse sweep.
\end{enumerate}
\end{auxresult}
\noindent\emph{Form used.} The $k$ coordinates $\langle p_i,\nabla L(\theta)\rangle$ of $P\T\nabla L$ cost $k$ forward
passes and no reverse pass; $\nabla L$ itself by forward mode would cost $N$ of them.
\fbBusedin{\thmref{Observation}{IV.4}.}

\begin{auxresult}{B.34}{central differences and Gaussian smoothing}{\citep{xnesterov2017random}; part (a) by Taylor's
theorem}
Let $L:\R^N\to\R$ be twice differentiable with an $\ell_2$-Lipschitz Hessian (constant $L_2$), $\varepsilon>0$ and
$z\in\R^N$.
\begin{enumerate}[label=(\alph*)]
  \item $\Big\lvert\dfrac{L(\theta+\varepsilon z)-L(\theta-\varepsilon z)}{2\varepsilon}-\langle\nabla L(\theta),z\rangle
    \Big\rvert\le\dfrac{L_2}{6}\,\varepsilon^2\lVert z\rVert^3$.
  \item For $z\sim\mathcal N(0,I_N)$, $\mathbb E\big[\langle\nabla L(\theta),z\rangle z\big]=\nabla L(\theta)$, and for
    Lipschitz $L$, $\mathbb E\big[\tfrac{1}{2\varepsilon}(L(\theta+\varepsilon z)-L(\theta-\varepsilon z))\,z\big]=\nabla
    L_\varepsilon(\theta)$, the gradient of the Gaussian smoothing $L_\varepsilon(\theta)=\mathbb E_uL(\theta+\varepsilon
    u)$.
  \item Hence the central two-point estimator is unbiased for $\nabla L_\varepsilon$, and by (a) its bias relative to
    $\nabla L$ is at most $\tfrac{L_2}{6}\varepsilon^2\,\mathbb E\lVert z\rVert^4=\tfrac{L_2}{6}\varepsilon^2N(N+2)$.
\end{enumerate}
\end{auxresult}
\fbBusedin{\thmref{Observation}{IV.4} (MeZO's estimator), \thmref{Proposition}{IV.3} (instances: each MeZO step is a step
of a linear method up to its estimator). The proof of \thmref{Definition}{IV.1} in Appendix~\ref{app:proofs} also cites it.}

\begin{auxresult}{B.35}{extreme eigenvalues of Wishart-type matrices}{\citep{xmarchenko1967distribution};
\citep{xyin1988limit}; \citep{xbai1993limit}}
Let $A\in\R^{r\times n}$ have independent, identically distributed entries with mean $0$, variance $c/n$ and finite
fourth moment. As $r,n\to\infty$ with $r/n\to y\in(0,1]$, the empirical eigenvalue
distribution of $AA\T$ converges to the Marčenko--Pastur law supported on $[c(1-\sqrt y)^2,c(1+\sqrt y)^2]$, and the
largest and smallest eigenvalues converge almost surely to $c(1+\sqrt y)^2$ and $c(1-\sqrt y)^2$.
\end{auxresult}
\noindent\emph{Form used.} HF PEFT's Kaiming-uniform $A_0$ has entries $U(-n^{-1/2},n^{-1/2})$, so $c=1/3$, and the
eigenvalues of $A_0A_0\T$ fill a band of relative width $O(\sqrt{r/n})$ around $c$. No proof depends on this entry.
\fbBusedin{the commentary on \thmref{Theorem}{III.9} (exactness for Kaiming initialisations), Prediction~X.7.}

\subsubsection{Elementary facts}

\begin{auxresult}{B.36}{three elementary facts}{elementary; part (c) in \citealp[Ch.~1]{xhorn2013matrix}}
\begin{enumerate}[label=(\alph*)]
  \item If $\sigma:\R\to\R$ is differentiable at $0$ and $\sigma(\ell t)=\ell\,\sigma(t)$ for all $\ell\ge0$ and
    $t\in\R$, then $\sigma$ is linear.
  \item A real vector space is not a finite union of proper subspaces. In particular, a hyperplane of $\R^d$ that is
    contained in a finite union of coordinate hyperplanes is one of them.
  \item For $X\in\R^{p\times q}$ and $Y\in\R^{q\times p}$, $XY$ and $YX$ have the same nonzero eigenvalues, with
    multiplicities.
\end{enumerate}
\end{auxresult}
\noindent\emph{Reasons.} (a) Taking $\ell=0$ gives $\sigma(0)=0$, and $\sigma(t)=t\,\sigma(1)$ for $t>0$ and
$\sigma(t)=-t\,\sigma(-1)$ for $t<0$; the one-sided derivatives at $0$ are $\sigma(1)$ and $-\sigma(-1)$, so
differentiability forces $\sigma(t)=t\,\sigma(1)$. ReLU and leaky ReLU satisfy the identity without being linear. (b)
If $V=V_1\cup\dots\cup V_k$ with $k$ minimal, pick $x\in V_1$ outside the other $V_i$ and $y\notin V_1$. No point $y+tx$
lies in $V_1$, so two of them, $t\ne t'$, lie in one $V_i$ with $i\ge2$, and then $x\in V_i$, a contradiction. The second
claim applies the first to $H=\bigcup_j(H\cap\{x_j=0\})$.
\fbBusedin{\thmref{Theorem}{VI.3} (rule R3: positively homogeneous activations), \thmref{Lemma}{VI.2}(b) (the kink
argument for ReLU), \thmref{Theorem}{III.9} (the nonzero eigenvalues of $T=\beta\beta\T$ are those of $\beta\T\beta$). The proof of \thmref{Corollary}{III.13} in Appendix~\ref{app:proofs} also cites it.}

\subsubsection{Spectral projectors, Riemannian submanifolds and floating point}

\begin{auxresult}{B.37}{smooth spectral projectors}{\citep[Ch.~I, \S5 and Ch.~II, \S1]{xkato1995perturbation};
\citep[Ch.~III]{xbhatia1997matrix}}
Let $S:V\to\Sym(m)$ be smooth on an open set $V$ of a real vector space, with eigenvalues
$\lambda_1(W)\ge\dots\ge\lambda_m(W)$, and let $\lambda_r(W_*)>\lambda_{r+1}(W_*)$ at some $W_*\in V$. Then the gap
$\lambda_r>\lambda_{r+1}$ persists on a neighbourhood $V_*$ of $W_*$, and on $V_*$ the orthogonal projector $\Pi(W)$ onto the
sum of the eigenspaces of $\lambda_1(W),\dots,\lambda_r(W)$ is the Riesz projector
\[\Pi(W)=\frac1{2\pi i}\oint_\Gamma\big(z-S(W)\big)^{-1}\,dz\]
for a fixed contour $\Gamma$ in $\mathbb C$ that encloses $\lambda_1(W_*),\dots,\lambda_r(W_*)$ and no other eigenvalue of
$S(W_*)$. In particular $\Pi$ is smooth on $V_*$.
\end{auxresult}
\noindent\emph{Reason.} By Weyl's perturbation inequality the eigenvalues are $1$-Lipschitz in the operator norm, so
near $W_*$ the contour still separates the top $r$ eigenvalues and $z-S(W)$ is invertible on $\Gamma$. Matrix inversion
is smooth, so the integral is smooth in $W$, and at each $W$ it is the orthogonal projector onto the enclosed eigenspaces.
\fbBusedin{\thmref{Proposition}{IV.3} (step 9: the GaLore field $\Pi(W)$ is smooth where a singular-value gap exists).}

\begin{auxresult}{B.38}{isometries, totally geodesic submanifolds and the Gauss equation}{\citep[Ch.~4, 5 and
8]{xlee2018introduction}; \citep{xkobayashi1958fixed}}
Let $(M,g)$ be a Riemannian manifold.
\begin{enumerate}[label=(\alph*)]
  \item A geodesic is determined by its initial point and initial velocity, and an isometry of $(M,g)$ maps geodesics to
    geodesics.
  \item Let $F\subseteq M$ be an embedded submanifold with the induced metric. If every geodesic of $M$ with initial
    velocity tangent to $F$ stays in $F$ for small times ($F$ is \emph{totally geodesic}), the second fundamental form of
    $F$ vanishes, and by the Gauss equation the sectional curvature of $F$ on every $2$-plane $\Pi\subseteq T_xF$ equals
    the sectional curvature of $M$ on $\Pi$.
\end{enumerate}
\end{auxresult}
\noindent\emph{Form used.} Let $\tau$ be an isometry and $F$ an embedded submanifold, open in its fixed-point set, with
$d\tau=\id$ on $TF$. By (a) a geodesic $\gamma$ tangent to $F$ and $\tau\circ\gamma$ have the same initial data, so
$\tau\circ\gamma=\gamma$ and $\gamma$ stays in $F$ for small times. So $F$ is totally geodesic, and (b) applies.
\fbBusedin{\thmref{Proposition}{IV.3} (step 6: balanced full-rank LoRA induces a curved leaf metric).}

\begin{auxresult}{B.39}{exact scaling by powers of two in floating point}{\citep[Ch.~2]{xhigham2002accuracy}}
In binary floating-point arithmetic with correctly rounded operations (IEEE~754, rounding to nearest), let $c=2^k$ with
$k\in\mathbb Z$. If all operands, exact results and rounded results below stay in the normal range (no overflow and no
underflow), then $\mathrm{fl}(cx)=cx$, and for floating-point numbers $x,y$,
\[\mathrm{fl}(cx\pm cy)=c\,\mathrm{fl}(x\pm y),\quad\mathrm{fl}(cx\cdot y)=c\,\mathrm{fl}(xy),\quad
\mathrm{fl}(cx/y)=c\,\mathrm{fl}(x/y),\quad\mathrm{fl}\big(\sqrt{c^2x}\big)=c\,\mathrm{fl}\big(\sqrt x\big).\]
\end{auxresult}
\noindent\emph{Reason.} In the normal range, multiplication by $2^k$ only shifts the exponent, so it maps the set of
floating-point numbers onto itself and scales all distances between them by $c$. Rounding to the nearest floating-point
number, with its tie-breaking rule, therefore commutes with it, and each identity follows from
$\mathrm{fl}(cz)=c\,\mathrm{fl}(z)$ applied to the exact result $z$.
\fbBusedin{\thmref{Theorem}{III.10} (bit-exact trajectories when every block factor is a power of two).}

\section{Deferred proofs}\label{app:proofs}

This appendix proves the numbered results of the main text in the order in which they appear there, one subsection per
result. Each subsection first restates the result in its final form, the form recorded after the two rounds of review,
and then proves it. External theorems enter through Appendix~\ref{app:auxiliary}, cited as Auxiliary results B.1--B.39,
and background notions through Appendix~\ref{app:background}. Exposé~\ref{sec:thesis} states its headline results
without numbers; each of them is proved below or in the later parts of this appendix under its number. Definitions I.1
and I.2 need no proof, since $\Para(\mathcal C)$ is a bicategory \citep{fong2017backprop,cruttwell2021categorical} and
$\Meth(\Theta,\theta_0)$ is a slice of a coslice; Remark I.3 and the two definitions that contain a claim, Definitions
I.5 and II.14, receive short proofs. Unless we say otherwise, parameter spaces are open definable subsets of a vector
space and the maps $\rho$ are semialgebraic or definable in $\R_{\exp}$ (\bgref{C.17}), dimensions are those of
\auxref{B.9}, ``generic'' has the meaning of \auxref{B.10}, and LoRA on one $m\times n$ box is
$\rho(B,A)=W_{\mathrm{res}}+sBA$ with $s\ne0$, $B\in\R^{m\times r}$ and $A\in\R^{r\times n}$. For subspaces
$V\subseteq\R^n$ and $U\subseteq\R^m$ we write $\mathcal R(V)=\{Z\in\R^{m\times n}:\operatorname{row}Z\subseteq V\}$ and
$\mathcal C(U)=\{Z:\operatorname{col}Z\subseteq U\}$, and $x\wedge y=xy\T-yx\T$. Values marked [Num] in a statement are
checked numerically and not proved; Appendix~\ref{app:repro} documents them.

\subsection{Remark I.3}\label{proof:I.3}

\begin{restate}{Remark}{I.3}{orientation, and what Para adds}
Each $\rho$ is a 2-cell $F\Rightarrow F^\rho:=(Q,f\circ(\rho\times X))$ \emph{out of} $F$. A simulation $h:M\to N$ is a
2-cell $F^{\rho_N}\Rightarrow F^{\rho_M}$, since $\rho_N\circ h=\rho_M$. Hence $\Meth(\Theta,\theta_0)$ is the opposite
of the pointed coslice of reparametrisation 2-cells under $F$. In particular $\Meth(\Theta,\theta_0)$ depends only on
$(\Theta,\theta_0)$ and not on $f$.
\end{restate}

\begin{proof}
Let $\mathcal P$ be the category whose objects are the 1-cells $X\to Y$ of $\Para(\Diff)$ and whose arrows are the
2-cells, composed vertically. By \thmref{Definition}{I.1} a 2-cell $(P,f)\Rightarrow(Q,g)$ is a smooth map $r:Q\to P$ with
$g=f\circ(r\times X)$. The vertical composite of $r:(P,f)\Rightarrow(Q,g)$ and $r':(Q,g)\Rightarrow(Q',g')$ is
$r\circ r':Q'\to P$, a 2-cell because $g'=g\circ(r'\times X)=f\circ((r\circ r')\times X)$.

An object of the coslice $F/\mathcal P$ is a 2-cell out of $F=(\Theta,f)$. A 2-cell from $F$ to $(Q,g)$ is a smooth
$\rho:Q\to\Theta$ with $g=f\circ(\rho\times X)$, so its target is forced to be $F^\rho$ and the object is determined by the
pair $(Q,\rho)$. An arrow from $\rho_M:F\Rightarrow F^{\rho_M}$ to $\rho_N:F\Rightarrow F^{\rho_N}$ is a 2-cell
$k:F^{\rho_M}\Rightarrow F^{\rho_N}$ whose vertical composite with $\rho_M$ is $\rho_N$. Concretely, it is a smooth map
$k:Q_N\to Q_M$ with
\begin{equation*}
  f\circ(\rho_N\times X)=f\circ(\rho_M\times X)\circ(k\times X)\qquad\text{and}\qquad\rho_M\circ k=\rho_N .
\end{equation*}
The second condition implies the first. So the arrows $(Q_M,\rho_M)\to(Q_N,\rho_N)$ of $F/\mathcal P$ are exactly the
smooth maps $k:Q_N\to Q_M$ over $\Theta$, which are the arrows $N\to M$ of $\Diff/\Theta$, and composition is reversed.
Restricting to pointed parameter spaces with $\rho(q_0)=\theta_0$ and to pointed $k$ gives the pointed coslice, which is
therefore $\Meth(\Theta,\theta_0)^{\mathrm{op}}$. The network $f$ entered only through the first condition, which is
implied by the second for every $f$, so $\Meth(\Theta,\theta_0)$ depends only on $(\Theta,\theta_0)$.
\end{proof}

\subsection{Observation I.4}\label{proof:I.4}

\begin{restate}{Observation}{I.4}{universal properties}
In $\Meth(\Theta,\theta_0)$:

(a) the frozen model $(\mathrm{pt},*,\theta_0)$ is initial, and full fine-tuning $(\Theta,\theta_0,\id)$ is terminal. The
unique morphism $M\to{}$Full FT is $\rho_M$ itself.

(b) If the fibre product $Q_M\times_\Theta Q_N$ exists in $\Diff$, it is the product $M\times N$. It exists, for instance,
when the set-level fibre product is an embedded submanifold of $Q_M\times Q_N$: a clean intersection, or linear or affine
$\rho$'s as in \thmref{Proposition}{I.8}(a). Transversality $\rho_M\pitchfork\rho_N$ is never the reason when
$\lvert M\rvert+\lvert N\rvert<\dim\Theta$, since at $(q_0,q_0')$ it would force $\lvert M\rvert+\lvert
N\rvert\ge\dim\Theta$. In $\Set$ the fibre product always exists and $\Im(M\times N)=\Im M\cap\Im N$. Whenever the product
exists in $\Meth$, its underlying set is this set-level fibre product, so the image identity holds in every case. The
product can fail to exist. On $\Theta=\R$ ($1\times1$ matrices), $\mathrm{LoRA}_1\times\text{Frozen}$ would have the cross
$\{BA=0\}\subset\R^2$ as underlying set. No manifold structure on the cross has the universal property, because both axes
would have to lift smoothly through the point over the origin, which forces rank $2$ there.

(c) $\Im$ is a functor to the poset of subsets of $\Theta$ containing $\theta_0$. Conversely, $\Im M\subseteq\Im N$ yields
a morphism of the set-level slice $\Set_*/(\Theta,\theta_0)$, though not necessarily a smooth one. Take $\rho_M(t)=t$ and
$\rho_N(u)=u^3$ on $\R$, pointed at $0$. The images agree, yet no smooth $h:\R\to\R$ satisfies $\rho_N\circ h=\id$, that
is, $h(t)^3=t$.
\end{restate}

\begin{proof}
Recall that $\Meth(\Theta,\theta_0)=\Diff_*/(\Theta,\theta_0)$ (\thmref{Definition}{I.2}).

\emph{(a)} Let $M=(Q,q_0,\rho)$. An arrow from the frozen model to $M$ is a pointed smooth map $h:\mathrm{pt}\to Q$ with
$\rho\circ h=\theta_0$. Pointedness forces $h(*)=q_0$, and the condition over $\Theta$ then holds because
$\rho(q_0)=\theta_0$. So there is exactly one such arrow. An arrow $h:M\to(\Theta,\theta_0,\id)$ is a pointed smooth map
$h:Q\to\Theta$ with $\id\circ h=\rho$. The only candidate is $h=\rho$, which is smooth and pointed.

\emph{(b)} By \auxref{B.3}(a), a product of $M$ and $N$ in the slice $\Diff_*/(\Theta,\theta_0)$ is the same as a pullback
of $\rho_M$ and $\rho_N$ in $\Diff_*$. Suppose the pullback $P$ of the underlying maps exists in $\Diff$, with projections
$\pi_M,\pi_N$. The constant maps $\mathrm{pt}\to Q_M$ and $\mathrm{pt}\to Q_N$ at $q_0$ and $q_0'$ agree over $\Theta$,
both composing to $\theta_0$, so they factor through a unique point $p_0\in P$. With base point $p_0$, $P$ is the pullback
in $\Diff_*$. For a pointed test object $T$ and pointed maps $h_M,h_N$ that agree over $\Theta$, the induced map $T\to P$
sends the base point to a point whose projections are $q_0$ and $q_0'$, and that point is $p_0$ by uniqueness. This is
the creation of limits by $\Diff_*\to\Diff$ (\auxref{B.3}(b)).

If the set-level fibre product $S=\{(q,q'):\rho_M(q)=\rho_N(q')\}$ is an embedded submanifold of $Q_M\times Q_N$, then $S$
with its two projections is the pullback in $\Diff$ (\auxref{B.2}(c)). Indeed, two smooth maps $T\to Q_M$ and $T\to Q_N$
that agree over $\Theta$ give a smooth map $T\to Q_M\times Q_N$ with image in $S$, which is smooth into $S$, and the
factorisation is unique because $S\to Q_M\times Q_N$ is injective. A clean intersection has, by definition, an embedded
submanifold $S$ (\bgref{C.7}), and for affine $\rho_M$ and $\rho_N$ on vector spaces $S$ is an affine subspace.

Transversality at $(q_0,q_0')$ asks $d\rho_M(T_{q_0}Q_M)+d\rho_N(T_{q_0'}Q_N)=T_{\theta_0}\Theta$. The left side has
dimension at most $\lvert M\rvert+\lvert N\rvert$, so transversality fails at the base point whenever
$\lvert M\rvert+\lvert N\rvert<\dim\Theta$ (\auxref{B.2}(b)).

In $\Set$ the fibre product is $S$, and its image in $\Theta$ is $\Im M\cap\Im N$: a pair $(q,q')\in S$ lies over
$\rho_M(q)=\rho_N(q')$, which belongs to both images, and a point $\theta$ of both images has preimages $q$ and $q'$ with
$(q,q')\in S$. Now suppose that a product $P$ exists in $\Meth$, and test it with the pointed $0$-manifold
$T=\{*,t\}$, pointed at $*$. A pointed map $T\to P$ is the choice of the image of $t$, that is, a point of $P$. A pair of
pointed maps $T\to Q_M$ and $T\to Q_N$ that agree over $\Theta$ is a point $(q,q')$ of $S$. The universal property makes
these correspond bijectively through $p\mapsto(\pi_M(p),\pi_N(p))$. So the canonical map $P\to Q_M\times Q_N$ is a bijection
onto $S$, and $\Im(M\times N)=\rho_M(\pi_M(P))=\Im M\cap\Im N$.

Take $\Theta=\R$ and $\mathrm{LoRA}_1:(B,A)\mapsto\theta_0+sBA$ on $\R^2$, with $s\ne0$ and frozen residual $\theta_0$,
so that its pointings are the $(B_0,A_0)$ with $B_0A_0=0$. Its set-level fibre product with the frozen model is the cross
$\mathcal X=\{(B,A):BA=0\}$. Suppose a product $P$ exists, and let $\iota:P\to\R^2$ be its projection to the parameters
of $\mathrm{LoRA}_1$. The forgetful functor $\Diff_*\to\Diff$ has a left adjoint, so it preserves this pullback
(\auxref{B.3}(c)), and $P$ is also the pullback in $\Diff$ of $\rho_{\mathrm{LoRA}_1}$ and the inclusion of $\theta_0$.
Hence every smooth curve $c:\R\to\R^2$ with $\rho\circ c\equiv\theta_0$ has a unique smooth lift $\tilde c:\R\to P$ with
$\iota\circ\tilde c=c$, and by the previous paragraph $\iota$ is a bijection onto $\mathcal X$. Let $p^*$ be the point of
$P$ over $(0,0)$. The axis curves $c_1(t)=(t,0)$ and $c_2(t)=(0,t)$ lie in $\mathcal X$, so they lift to smooth curves
through $p^*$, and $d\iota_{p^*}$ maps their velocities to $c_1'(0)=(1,0)$ and $c_2'(0)=(0,1)$. Hence
$\rk d\iota_{p^*}=2$. The rank cannot exceed $2$ and cannot drop near $p^*$ (\auxref{B.1}(a)), so $\iota$ is a submersion
on a neighbourhood $V$ of $p^*$, and $\iota(V)$ contains a nonempty open subset of $\R^2$ (\auxref{B.1}(c)). But
$\iota(V)\subseteq\mathcal X$, which contains no nonempty open set. So no product exists. The argument used only the
point over the origin, so it covers every pointing, including the standard one $(0,a_0)$ with $a_0\ne0$, whose base
point is a smooth point of $\mathcal X$. (A split pointing, with $B_0A_0\ne0$ and residual $\theta_0-sB_0A_0$, has a
hyperbola over $\theta_0$, an embedded submanifold, and there the product exists.)

\emph{(c)} If $h:M\to N$ is an arrow, then $\rho_M=\rho_N\circ h$ gives $\Im M=\rho_N(h(Q_M))\subseteq\Im N$, and both
images contain $\theta_0$; functoriality is automatic because the target is a poset. Conversely, let
$\Im M\subseteq\Im N$. Choose, by the axiom of choice, a preimage $s(\theta)\in\rho_N^{-1}(\theta)$ for each
$\theta\in\Im M$, with $s(\theta_0)=q_0'$, and put $h=s\circ\rho_M$. Then $\rho_N\circ h=\rho_M$ and $h(q_0)=q_0'$, so $h$
is an arrow of $\Set_*/(\Theta,\theta_0)$. For $\rho_M(t)=t$ and $\rho_N(u)=u^3$ both images are $\R$. A map with
$\rho_N\circ h=\id$ satisfies $h(t)^3=t$, so $h(t)=t^{1/3}$, the unique real cube root, whose difference quotient
$t^{-2/3}$ at $0$ is unbounded. So no $C^1$ arrow, and in particular no smooth one, exists.
\end{proof}

\subsection{Definition I.5}\label{proof:I.5}

\begin{restate}{Definition}{I.5}{invariants of an object}
For $M=(Q,q_0,\rho)$:
\begin{itemize}
  \item the image $\Im M=\rho(Q)$;
  \item the \textbf{expressive dimension} $d(M)=\dim\Im M=\max_q\rk d\rho_q$;
  \item the \textbf{gauge waste} $g(M)=\lvert M\rvert-d(M)$;
  \item the \textbf{first-order image} $S_1(M)=d\rho_{q_0}(T_{q_0}Q)$, and the \textbf{first-order deficit}
    $d(M)-\dim S_1(M)$;
  \item the \textbf{gauge group} $\{\varphi\in\mathrm{Diff}(Q):\rho\circ\varphi=\rho\}$, which is unpointed, since a
    gauge transformation moves $q_0$ inside its fibre. In practice we exhibit a Lie group acting on $Q$ that preserves
    $\rho$ and is locally transitive on generic fibres, and call it the \textbf{generic gauge}.
\end{itemize}
The base point $\theta_0$ is \textbf{regular} for $M$ if the deficit is $0$, and an \textbf{apex} if the deficit is
positive. We say \textbf{vertex} for the subset-level notion, the cone point of an image germ, and keep \textbf{apex} for
the base-point notion. The two differ in general. At a crossing of two smooth branches the image is singular, yet the
deficit can be $0$. In every case below with a positive deficit, the image germ is (diffeomorphic to) a cone with vertex
$\theta_0$; the exception is DoRA, which inherits LoRA's deficit through its torus factor. A cone germ that is a $C^1$
manifold at its vertex equals its tangent space there, so it is linear. Hence the vertex of a non-linear cone is a
singular point.
\end{restate}

The definition contains three claims: the two descriptions of $d(M)$ agree, the deficit is never negative (so that
regular and apex exhaust the cases), and a cone germ that is $C^1$ at its vertex is linear.

\begin{proof}
\emph{$d(M)$ is well defined.} Let $Q$ be an open definable subset of $\R^N$ and $\rho$ definable and smooth, and let
$k=\max_q\rk d\rho_q$. The set where the rank equals $k$ is open and nonempty, because a nonzero $k\times k$ minor of the
Jacobian stays nonzero nearby (\auxref{B.1}(a)). At a point $p$ of it the rank is constant nearby, so by the rank theorem
(\auxref{B.1}(b)) a small open ball $V\ni p$, a definable set, has image $\rho(V)$ an embedded $k$-dimensional
submanifold. It is definable, of dimension $k$, and contained in $\Im M$, so $\dim\Im M\ge k$ (\auxref{B.9}(b)).
Conversely, let $e=\dim\Im M$. By \auxref{B.9}(b), $\Im M$ contains a definable $C^1$ cell $Y$ of dimension $e$. By
definable choice $\rho$ has a definable section $\sigma:Y\to Q$, and by generic smoothness $\sigma$ is $C^1$ on a nonempty
open subset $Y'$ of $Y$ (\auxref{B.9}(c)). Differentiating $\rho\circ\sigma=\id_{Y'}$ at $y\in Y'$ gives
$d\rho_{\sigma(y)}\circ d\sigma_y=\id_{T_yY}$, so $\rk d\rho_{\sigma(y)}\ge e$, and $k\ge e$. Hence
$\dim\Im M=\max_q\rk d\rho_q$, which is \auxref{B.9}(d). The parameter spaces of the paper are open definable sets, for
instance $\{(u_1,\dots,u_r):u_i\ne0\}$ for HRA, and the maps are built from polynomials, quotients, square roots and, for
softmax, the exponential (\bgref{C.17}).

\emph{The deficit is nonnegative.} $\dim S_1(M)=\rk d\rho_{q_0}\le\max_q\rk d\rho_q=d(M)$. Likewise
$d(M)\le\dim Q=\lvert M\rvert$, so $g(M)\ge0$.

\emph{Cones.} Let $X\subseteq\R^D$ agree near $v$ with a cone $K$ with vertex $v$, meaning $v\in K$ and
$v+t(x-v)\in K$ for all $x\in K$ and $t\ge0$, and suppose $X$ is a $C^1$ embedded submanifold near $v$, with tangent space
$T$ at $v$. Translate so that $v=0$. Near $0$, $X$ is the graph of a $C^1$ map $\varphi$ from a neighbourhood of $0$ in
$T$ to $T^\perp$ with $\varphi(0)=0$ and $d\varphi_0=0$. For $x\in K$ and small $t>0$ the point $tx$ lies in $K$ near $0$,
hence in $X$, so $tP_{T^\perp}x=\varphi(tP_Tx)=o(t)$, and therefore $P_{T^\perp}x=0$. Thus $K\subseteq T$. Near $0$, $X$
is then a graph over $T$ that lies in $T$, so $\varphi=0$ and $X$ contains a neighbourhood $W$ of $0$ in $T$. For $y\in
T$ and small $\varepsilon>0$ we have $\varepsilon y\in W\subseteq K$, so $y=\varepsilon^{-1}(\varepsilon y)\in K$. Hence
$K=T$, and the germ of $X$ at $v$ is the germ of the affine space $v+T$. If $K$ is not an affine subspace, $X$ cannot be a
$C^1$ manifold at $v$, so $v$ is a singular point (compare \auxref{B.6}). Finally, a figure-eight curve through its
crossing with nonzero velocity has deficit $0$ there and a singular image, so vertex and apex are different notions.
\end{proof}

\subsection{Observation I.6}\label{proof:I.6}

\begin{restate}{Observation}{I.6}{$S_1$ is functorial}
If $M\to N$ is a (pointed) morphism, then $S_1(M)\subseteq S_1(N)$. In particular, isomorphic methods have equal $S_1$ and
equal first-order deficit. Monotonicity holds only along pointed morphisms. Along a weak morphism, such as a gauge map
moving $q_0$ inside its fibre, $S_1$ can change.
\end{restate}

\begin{proof}
Let $h:M\to N$ be a pointed arrow, so $\rho_N\circ h=\rho_M$ and $h(q_0)=q_0'$. The chain rule at $q_0$ gives
$d(\rho_N)_{q_0'}\circ dh_{q_0}=d(\rho_M)_{q_0}$, so
$S_1(M)=d(\rho_N)_{q_0'}\big(dh_{q_0}(T_{q_0}Q_M)\big)\subseteq d(\rho_N)_{q_0'}(T_{q_0'}Q_N)=S_1(N)$. If $h$ is an
isomorphism, $h^{-1}$ is a pointed arrow $N\to M$, so $S_1(N)\subseteq S_1(M)$ as well. Moreover $\rho_M=\rho_N\circ h$
with $h$ bijective gives $\Im M=\Im N$, so $d(M)=d(N)$ (equivalently, $\rk d(\rho_M)_q=\rk d(\rho_N)_{h(q)}$ since $dh_q$ is
invertible), and the deficits agree.

Without $h(q_0)=q_0'$ the chain rule compares $S_1(M)$ with the first-order image of $N$ at $h(q_0)$, which need not be
the base point of $N$. For instance, take LoRA with frozen residual $\theta_0$ and $1\le r<\min(m,n)$, and let $M$ and
$N$ be its pointings at $(0,A_0)$ and at $(B_0,0)$, with $A_0$ and $B_0$ of full rank; both points lie in the fibre over
$\theta_0$. The identity of $Q$ is a weak morphism $M\to N$, yet $S_1(M)=\{XA_0\}=\mathcal R(\operatorname{row}A_0)$ and
$S_1(N)=\{B_0Y\}=\mathcal C(\operatorname{col}B_0)$ are not nested: $\mathcal R(\operatorname{row}A_0)$ contains $e_iv\T$
for every $i$ and every $v\in\operatorname{row}A_0$, which is not in $\mathcal C(\operatorname{col}B_0)$ when
$e_i\notin\operatorname{col}B_0$, and symmetrically.
\end{proof}

\subsection{Observation I.7}\label{proof:I.7}

\begin{restate}{Observation}{I.7}{additive monoidal structure}
$\Theta$ is a vector space, so every object of $\Meth(\Theta,\theta_0)$ can be written $\rho=\theta_0+\delta$ with
$\delta:=\rho-\theta_0$ and $\delta(q_0)=0$. Put
\begin{equation*}
  M\oplus N:=\big(Q_M\times Q_N,\ (q_0,q_0'),\ \theta_0+\delta_M+\delta_N\big).
\end{equation*}
This is a symmetric monoidal structure with the frozen model as unit. Its image is the Minkowski sum
$\theta_0+(\Im\delta_M+\Im\delta_N)$. It differs from the categorical product, whose image, when the product exists, is
the intersection $\Im M\cap\Im N$. ``Additive'' in \thmref{Definition}{I.2} ($\delta$ independent of $\theta_0$) is a
condition on \emph{families} $\theta_0\mapsto M(\theta_0)$; it matters for base change and plays no role here.

Write $\mathrm{LoRA}_k$ for LoRA with frozen residual $\theta_0$; for split pointings the sum below has residual
$\theta_0-P-P'$. Concatenation $h(B_1,A_1,B_2,A_2)=([B_1\ B_2],[A_1;A_2])$ is an isomorphism
$\mathrm{LoRA}_r\oplus\mathrm{LoRA}_s\cong\mathrm{LoRA}_{r+s}$ of unpointed objects. It is a pointed isomorphism onto
$\mathrm{LoRA}_{r+s}$ pointed at $([B_0\ B_0'],[A_0;A_0'])$. For $\mathrm T_A$ pointings ($B_0=0$, $\rk A_0=r$), the
target pointing has $\rk[A_0;A_0']=r+s$ if and only if $\operatorname{row}A_0\cap\operatorname{row}A_0'=0$; it lies in the
stratum $\mathrm T_A$ of \thmref{Theorem}{II.2} when moreover $r+s<\min(m,n)$.

Standard pointings ($\mathrm T_A$, or symmetrically $\mathrm T_B$) exist only for $r<\min(m,n)$, so the monoidal question
is pointwise. If $F(r)$ and $F(2r)$ are standard ($2r<\min(m,n)$), then $F(r)\oplus F(r)\not\cong F(2r)$ as pointed
objects; for $2r\ge\min(m,n)$ no standard $F(2r)$ exists. Hence no assignment $r\mapsto\mathrm{LoRA}_r$ with $F(1)$ and
$F(2)$ standard is \textbf{strong monoidal} into pointed objects. For $r\neq s$ and standard pointings, $F(r)\oplus
F(s)\cong F(r+s)$ as pointed objects exactly when the row space of the target's $A_0$ is the direct sum of the two source
row spaces. The assignment is strong monoidal on unpointed objects, up to weak morphisms, and on pointed objects with
degenerate pointings, namely the origin $(B_0,A_0)=(0,0)$, where $S_1=0$, or the stacked pointings $F(r)=F(1)^{\oplus r}$.
Concatenation is strictly associative and unital, so discrete $(\N,+)$ suffices for a plain strong monoidal functor.
\emph{Symmetric} strong monoidality needs the permutation groupoid, since swapping and then concatenating permutes the
blocks.

With per-adapter scales $s_r$ ($\alpha/r$ for LoRA, $\alpha/\sqrt r$ for rsLoRA) and any target scale $s_*$, the map
$h=([B_1\ B_2],[c_1A_1;c_2A_2])$ with $c_i=s_{r_i}/s_*$, where $r_1=r$ and $r_2=s$, is an isomorphism over $\Theta$. It is
a linear bijection and preserves row spaces, so the pointing criterion is unchanged.
\end{restate}

\begin{proof}
\emph{The monoidal structure.} $M\oplus N$ is an object: $Q_M\times Q_N$ is a manifold, $\theta_0+\delta_M+\delta_N$ is
smooth and takes the value $\theta_0$ at $(q_0,q_0')$. For arrows $h:M\to M'$ and $k:N\to N'$, the pointed map $h\times k$
lies over $\Theta$ because $\delta_{M'}\circ h=\rho_{M'}\circ h-\theta_0=\delta_M$ and likewise for $k$; this makes $\oplus$
a functor. The associator $(Q_1\times Q_2)\times Q_3\cong Q_1\times(Q_2\times Q_3)$, the unitors
$\mathrm{pt}\times Q\cong Q\cong Q\times\mathrm{pt}$ and the symmetry $Q_M\times Q_N\cong Q_N\times Q_M$ are the
structure maps of the cartesian monoidal structure of $\Diff_*$. They lie over $\Theta$ because addition is associative and
commutative and the frozen model has $\delta=0$. Their coherence (pentagon, triangle, hexagon and the symmetry being an
involution) holds in $\Diff_*$, and the forgetful functor $\Meth(\Theta,\theta_0)\to\Diff_*$ is faithful, so it holds in
$\Meth(\Theta,\theta_0)$.

\emph{Image.} $\{\theta_0+\delta_M(q)+\delta_N(q')\}=\theta_0+(\Im\delta_M+\Im\delta_N)$. The categorical product has
image $\Im M\cap\Im N$ when it exists (\thmref{Observation}{I.4}(b)), and the two differ in general: for zero-init
$\mathrm{LoRA}_1$ on a $2\times2$ box, $M\oplus M$ has image $\theta_0+\MM_{\le2}$, while any product of $M$ with itself
has image $\theta_0+\MM_{\le1}$.

\emph{Concatenation.} The map $h$ only regroups coordinates, so it is a linear bijection between the parameter spaces,
and $\theta_0+s[B_1\ B_2][A_1;A_2]=\theta_0+sB_1A_1+sB_2A_2$, so it lies over $\Theta$. It is a pointed isomorphism onto
$\mathrm{LoRA}_{r+s}$ pointed at the image of $(q_0,q_0')$, the concatenated point. For split pointings with
$P=sB_0A_0$ and $P'=sB_0'A_0'$, the sum is $(\theta_0-P-P')+s[B_1\ B_2][A_1;A_2]$, which is $\mathrm{LoRA}_{r+s}$ with
residual $\theta_0-P-P'$. For $\mathrm T_A$ pointings, the concatenated pointing has $B=0$ and
$\rk[A_0;A_0']=\dim(\operatorname{row}A_0+\operatorname{row}A_0')=r+s-\dim(\operatorname{row}A_0\cap\operatorname{row}A_0')$,
which is $r+s$ exactly when the intersection is $0$. The stratum $\mathrm T_A$ of rank $r+s$ is defined for
$r+s<\min(m,n)$, which gives the last condition.

\emph{Row and column types.} We use two elementary facts about the spaces $\mathcal R(V)$ and $\mathcal C(U)$ of the
introduction, which return in \thmref{Theorem}{II.2}. First, $\mathcal R(V)=\{XA:X\in\R^{m\times k}\}$ for any $A$ of full
row rank with $\operatorname{row}A=V$, since $Z=ZA^+A$ when $\operatorname{row}Z\subseteq V$; so $\dim\mathcal
R(V)=m\dim V$, and $\mathcal R(V)$ determines $V$ as the sum of the row spaces of its elements. Likewise
$\mathcal C(U)=\{BY\}$ for $\operatorname{col}B=U$, of dimension $n\dim U$, determines $U$. Second, if $V\ne0$ and
$\mathcal R(V)=\mathcal C(U)$, then $U=\R^m$, because $\mathcal R(V)$ contains $e_iv\T$ for every $i$ and every nonzero
$v\in V$, and $e_iv\T\in\mathcal C(U)$ forces $e_i\in U$.

\emph{The negative claim.} Standard pointings are the strata $\mathrm T_A$ and $\mathrm T_B$ of \thmref{Theorem}{II.2},
which are defined only for ranks below $\min(m,n)$. Let $F(r)$ and $F(2r)$ be standard, so $2r<\min(m,n)$. If $F(r)$ is
pointed at $(0,A_0)$, then $F(r)\oplus F(r)$ is pointed at $((0,A_0),(0,A_0))$ and its first-order image is
$\{X_1A_0+X_2A_0\}=\mathcal R(\operatorname{row}A_0)$, of dimension $mr$. If $F(r)$ is in $\mathrm T_B$, it is
$\mathcal C(\operatorname{col}B_0)$, of dimension $nr$. A standard $F(2r)$ has first-order image $\mathcal R(V'')$ with
$\dim V''=2r$ or $\mathcal C(U'')$ with $\dim U''=2r$. A pointed isomorphism would make the two first-order images equal
(\thmref{Observation}{I.6}). Two row-type spaces determine their subspaces, of dimensions $r\ne2r$; two column-type spaces
likewise; and a row-type space with $V\ne0$ equals a column-type space only if the column subspace is all of $\R^m$,
impossible for $U''$ of dimension $2r<m$ (and symmetrically). So $F(r)\oplus F(r)\not\cong F(2r)$. A strong monoidal
functor $F$ would supply a pointed isomorphism $F(1)\oplus F(1)\cong F(2)$, so none exists when $F(1)$ and $F(2)$ are
standard.

\emph{The case $r\ne s$.} Let $F(r)$, $F(s)$ and $F(r+s)$ be standard of type $\mathrm T_A$, pointed at $(0,A_0)$,
$(0,A_0')$ and $(0,A_0'')$; the type $\mathrm T_B$ is the transpose. Then
$S_1(F(r)\oplus F(s))=\{XA_0+X'A_0'\}=\mathcal R(\operatorname{row}A_0+\operatorname{row}A_0')$ and
$S_1(F(r+s))=\mathcal R(\operatorname{row}A_0'')$, with $\dim\operatorname{row}A_0''=r+s$. A pointed isomorphism forces
$\operatorname{row}A_0''=\operatorname{row}A_0+\operatorname{row}A_0'$, and the dimension count makes the sum direct.
Conversely, if $\operatorname{row}A_0''=\operatorname{row}A_0\oplus\operatorname{row}A_0'$, then $[A_0;A_0']$ has full row
rank and the row space of $A_0''$, so $A_0''=g[A_0;A_0']$ for a unique $g\in\GL_{r+s}$. Concatenation followed by the gauge
$(B,A)\mapsto(Bg^{-1},gA)$, which preserves $BA$, is a pointed isomorphism $F(r)\oplus F(s)\to F(r+s)$.

\emph{Strong monoidality.} The concatenation maps $\phi_{r,s}=h$ satisfy
$\phi_{r+s,t}\circ(\phi_{r,s}\times\id)=\phi_{r,s+t}\circ(\id\times\phi_{s,t})\circ a$, both sides sending
$(B_1,A_1,B_2,A_2,B_3,A_3)$ to $([B_1\ B_2\ B_3],[A_1;A_2;A_3])$, and $\phi_{0,r}$, $\phi_{r,0}$ are the unitors, with
$\mathrm{LoRA}_0$ the frozen model. Discrete $(\N,+)$ has only identity arrows, so naturality is vacuous, and $r\mapsto
\mathrm{LoRA}_r$ is strong monoidal on unpointed objects. As maps between chosen pointings, the $\phi_{r,s}$ are weak
morphisms. They are pointed isomorphisms when every $F(r)$ is pointed at the origin, since $h$ maps origin to origin, and
when $F(r)$ is pointed at the concatenation of $r$ copies of the base point of $F(1)$, since $h$ maps concatenations to
concatenations. The symmetry of discrete $(\N,+)$ is the identity, so symmetry would require $\phi_{r,s}=\phi_{s,r}\circ
\sigma$, where $\sigma$ swaps the factors. The left side sends $(B_1,A_1,B_2,A_2)$ to $([B_1\ B_2],[A_1;A_2])$ and the
right side to $([B_2\ B_1],[A_2;A_1])$, which differ for $r,s\ge1$. On the permutation groupoid, a permutation $\pi\in S_r$
acts on $\mathrm{LoRA}_r$ by the gauge $(B,A)\mapsto(BP_\pi^{-1},P_\pi A)$, which is functorial and lies over $\Theta$, the
maps $\phi_{r,s}$ are natural for block sums of permutations, and with the block transposition $\tau_{r,s}\in S_{r+s}$ as
symmetry, $F(\tau_{r,s})\circ\phi_{r,s}=\phi_{s,r}\circ\sigma$, because permuting the channel blocks of
$([B_1\ B_2],[A_1;A_2])$ gives $([B_2\ B_1],[A_2;A_1])$. The origin and the stacked pointings are fixed by channel
permutations, so the same holds for them.

\emph{Rescaled concatenation.} $s_*[B_1\ B_2][c_1A_1;c_2A_2]=s_*c_1B_1A_1+s_*c_2B_2A_2=s_rB_1A_1+s_sB_2A_2$. The scales
are nonzero, so $h$ is a linear bijection, and $\operatorname{row}(c_iA_i)=\operatorname{row}A_i$, so the criterion for
the concatenated pointing is the same as before.
\end{proof}

\subsection{Proposition I.8}\label{proof:I.8}

\begin{restate}{Proposition}{I.8}{a product that produces a named method; an image-level meet}
(a) Let $A_0\in\R^{r\times n}$ have full row rank and $B_0\in\R^{m\times r}$ full column rank. Define the one-sided methods
\begin{equation*}
  \mathrm{FA}(A_0):B\mapsto W_0+BA_0,\qquad \mathrm{FB}(B_0):A\mapsto W_0+B_0A,
\end{equation*}
both pointed at $0$. Then
\begin{equation*}
  \mathrm{FA}(A_0)\times\mathrm{FB}(B_0)\;\cong\;\big(\R^{r\times r},\ 0,\ R\mapsto W_0+B_0RA_0\big).
\end{equation*}
The product supplies this sandwich for \emph{any} frames. Assume $\rk W_0\ge r$ ($\sigma_r>0$) and take the SVD frames
$B_0=U_r$, $A_0=\Sigma_rV_r\T$. The result is \textbf{LoRA-XS pointed at $R=0$}.

(b) Let $W_0$ have full row rank (so $m\le n$), and let $\mathrm{LoRA}_r$ carry its standard pointing ($B_0=0$, image
$W_0+\Mr$). The image-level meet of $\mathrm{LoRA}_r$ with the output-scaling method $\ell\mapsto\diag(\ell)W_0$ is
$\{\diag(\ell)W_0:\#\{i:\ell_i\neq1\}\le r\}$, that is, ``rescale at most $r$ output neurons''.

\upshape The LoRA-XS paper writes $xW+xARB$ in row-vector convention with $A=U_r\Sigma_r$ and $B=V_r\T$. Transposed, the
update is $B_0R\T A_0$, so the isomorphism is $R\mapsto\alpha R\T$. LoRA-XS initialises $R\sim\mathcal N(0,\sigma^2)$
with $\sigma=10^{-5}$, a negligible pointing defect. The output-scaling method in (b) is the key/value part of (IA)$^3$.
On the FFN, (IA)$^3$ scales the \emph{input} of $W_{\rm down}$, $W\mapsto W\diag(\ell)$; the mirror statement then needs
full column rank, or holds generically for $r<d$, where $d$ is the number of rows of $W_{\rm down}$.
\end{restate}

\begin{proof}
\emph{(a)} Write $B_0^+=(B_0\T B_0)^{-1}B_0\T$ and $A_0^+=A_0\T(A_0A_0\T)^{-1}$, so that $B_0^+B_0=I_r=A_0A_0^+$. A point of
the set-level fibre product is a pair $(B,A)$ with $BA_0=B_0A=:D$. Then $\operatorname{row}D\subseteq\operatorname{row}A_0$
and $\operatorname{col}D\subseteq\operatorname{col}B_0$, so $D=B_0B_0^+DA_0^+A_0=B_0RA_0$ with $R=B_0^+DA_0^+$, and $R$ is
unique because $B_0$ is injective and $A_0$ surjective. From $BA_0=B_0RA_0$ and $A_0A_0^+=I$ we get $B=B_0R$, and from
$B_0A=B_0RA_0$ we get $A=RA_0$. So the fibre product is the $r^2$-dimensional linear subspace $\{(B_0R,RA_0)\}$ of
$\R^{m\times r}\times\R^{r\times n}$, and $j:R\mapsto(B_0R,RA_0)$ is a linear isomorphism onto it, with inverse
$(B,A)\mapsto B_0^+B$. Both composites $R\mapsto W_0+B_0RA_0$ agree, and $j(0)=(0,0)$ is the pair of base points.

The universal property: let $T$ be a pointed manifold and $h_1:T\to\R^{m\times r}$, $h_2:T\to\R^{r\times n}$ pointed smooth
maps with $W_0+h_1(t)A_0=W_0+B_0h_2(t)$. By the previous paragraph $(h_1(t),h_2(t))=j(R(t))$ with $R(t)=B_0^+h_1(t)$, a
pointed smooth map, and it is the only map with this property because $j$ is injective. So $(\R^{r\times r},0,R\mapsto
W_0+B_0RA_0)$ with the projections $R\mapsto B_0R$ and $R\mapsto RA_0$ is the product in $\Meth(\Theta,W_0)$
(\auxref{B.2}(c), \auxref{B.3}). This holds for any frames of full rank. With $B_0=U_r$ and $A_0=\Sigma_rV_r\T$ from an SVD
of $W_0$ (\auxref{B.11}(a)), the frames have full rank exactly when $\sigma_r>0$; if $\rk W_0<r$, $\Sigma_rV_r\T$ is
rank-deficient and the hypothesis fails.

The row-vector convention: LoRA-XS acts on row vectors by $x\mapsto x(W_{\rm row}+\alpha ARB)$, where $W_{\rm row}=W_0\T$
has SVD $U'\Sigma V'^{\top}$, $A=U'_r\Sigma_r$ and $B=V_r'^{\top}$. In our convention the update is the transpose,
$\alpha V'_rR\T\Sigma_rU_r'^{\top}$. Since $W_0=V'\Sigma U'^{\top}$, the left and right singular frames of $W_0$ are
$U_r=V'_r$ and $V_r=U'_r$, so the update is $U_r(\alpha R\T)\Sigma_rV_r\T=B_0(\alpha R\T)A_0$. Hence LoRA-XS is the product
composed with the linear isomorphism $R\mapsto\alpha R\T$, which fixes $0$. A random initial $R$ makes
$\rho(R_0)-W_0=\alpha U_rR_0\T\Sigma_rV_r\T$ nonzero, of the size of $\sigma$, a pointing defect.

\emph{(b)} The image of $\mathrm{LoRA}_r$ with its standard pointing is $W_0+\Mr$ (\thmref{Theorem}{II.2}(c)), and the
output-scaling method has image $\{\diag(\ell)W_0\}$. A point $\diag(\ell)W_0$ lies in $W_0+\Mr$ iff
$\rk\big(\diag(\ell-\mathbf 1)W_0\big)\le r$. The rows of $\diag(\ell-\mathbf 1)W_0$ are $(\ell_i-1)w_i$, and the nonzero ones
are linearly independent because the rows of $W_0$ are, so the rank is $\#\{i:\ell_i\ne1\}$. The meet in the poset of
subsets is the intersection, which is the stated set.

For the input side, $W_0\diag(\ell)-W_0=W_0\diag(\ell-\mathbf 1)$ consists of the columns $(\ell_j-1)c_j$ of $W_0$, so its
rank is the rank of the columns of $W_0$ indexed by $J=\{j:\ell_j\ne1\}$. If $W_0$ has full column rank this is $\lvert
J\rvert$, and the meet is $\{W_0\diag(\ell):\lvert J\rvert\le r\}$. For $W_{\rm down}\in\R^{d\times d_{\rm ff}}$ with
$d<d_{\rm ff}$, full column rank is impossible. For generic $W_0$ every set of at most $d$ columns is independent (the
failure is the vanishing of all maximal minors of the chosen columns), so the rank is $\min(\lvert J\rvert,d)$, and when
$r<d$ the condition $\min(\lvert J\rvert,d)\le r$ is again $\lvert J\rvert\le r$.

\emph{No product in (b).} The commentary after the proposition says that, for $n\ge2$, $\mathrm{LoRA}_r$ and the
output-scaling method have no product; the argument is that of \thmref{Observation}{I.4}(b). Their set-level fibre product
is $\mathcal F=\{(B,A,\ell):W_0+sBA=\diag(\ell)W_0\}$, whatever the pointing of $\mathrm{LoRA}_r$ (with residual $W_0$).
Suppose a product $P$ exists. By the proof of \thmref{Observation}{I.4}(b) its canonical map $\iota:P\to\R^{m\times
r}\times\R^{r\times n}\times\R^m$ is a bijection onto $\mathcal F$, $P$ is also the pullback in $\Diff$, and smooth curves
in $\mathcal F$ lift. Let $p^*$ be the point over $(0,0,\mathbf 1)$. For every $B'$ and $A'$ the curves $t\mapsto(tB',0,\mathbf
1)$ and $t\mapsto(0,tA',\mathbf 1)$ lie in $\mathcal F$ and lift through $p^*$, so with $\mathrm{pr}$ the projection to the
LoRA parameters, $d(\mathrm{pr}\circ\iota)_{p^*}$ is onto $\R^{m\times r}\times\R^{r\times n}$. By \auxref{B.1}(a),(c),
$\mathrm{pr}\circ\iota$ is a submersion near $p^*$ and its image contains a nonempty open set of pairs $(B,A)$. But every pair
in $\mathrm{pr}(\mathcal F)$ has $BA=\diag(\ell-\mathbf 1)W_0/s$, so the first row of $BA$ is parallel to the first row
$w_{0,1}$ of $W_0$: the $2\times2$ minors of the $2\times n$ matrix with rows $e_1\T BA$ and $w_{0,1}\T$ vanish. For
$n\ge2$ one of these minors is a polynomial in $(B,A)$ that is not identically zero, so its zero set has empty interior
(\auxref{B.10}(a)), a contradiction. (For $m=n=1$ the fibre product is the graph of $\ell=1+sBA/W_0$, and the product
exists.)
\end{proof}

\subsection{Theorem II.1}\label{proof:II.1}

\begin{restate}{Theorem}{II.1}{pointings of an additive method}
Let $M=(Q,q_0,\theta_0+\delta)$ be additive with shape $C=\delta(Q)$. Write $-v+C$ for a translate, and
$C-C:=\{c-c':c,c'\in C\}$ for the Minkowski difference.

(a) For each $q\in Q$, the re-pointed method $M_q:=(Q,q,\theta_0-\delta(q)+\delta(\cdot))$, with frozen residual
$\theta_0-\delta(q)$, is pointed at $\theta_0$, and $\Im M_q=\theta_0+(-\delta(q)+C)$.

(b) $\bigcup_q\Im M_q=\theta_0+(C-C)$.

(c) For LoRA$_r$ on a single linear box with $s\ne0$, $C-C=\MM_{\le\min(2r,m,n)}$; for several boxes, $C-C$ is the
product of the per-box varieties $\MM_{\le\min(2r,m_i,n_i)}$. Hence \textbf{no exact pointing of LoRA$_r$ (frozen
residual $\theta_0-\delta(q)$), whether zero-init or split (PiSSA, OLoRA, CorDA, LoRA-GA), makes an update
$\theta-\theta_0$ of rank $>2r$ reachable in any adapted matrix.} More robustly, from \emph{any} starting point $q_0$ of
LoRA$_r$, pointed or not, $\rho(q)-\rho(q_0)=\delta(q)-\delta(q_0)$ has rank $\le\min(2r,m,n)$. Quantised-residual
schemes (QLoRA, LoftQ, QPiSSA) are pointed only up to a defect that depends on the order of operations ($e$ for QLoRA;
\thmref{Proposition}{V.2}(b) for the others). Measured from $\theta_0$ rather than from their starting weight, their
updates contain this defect, whose rank is generically at least $\min(m,n)-r$. The bound $2r$ is attained by split
pointings when $2r\le\min(m,n)$.
\end{restate}

\begin{proof}
\emph{(a)} $M_q$ has parameter space $Q$ and the smooth map $\rho_q=\theta_0-\delta(q)+\delta(\cdot)$, with
$\rho_q(q)=\theta_0$, so it is pointed at $\theta_0$, and $\Im M_q=\theta_0-\delta(q)+\delta(Q)=\theta_0+(-\delta(q)+C)$.

\emph{(b)} As $q$ ranges over $Q$, $\delta(q)$ ranges over $C$, so
$\bigcup_q\Im M_q=\theta_0+\bigcup_{c'\in C}(C-c')=\theta_0+(C-C)$.

\emph{(c)} On one box $C=\{sBA\}$. Since $s\ne0$ and every matrix of rank at most $\min(r,m,n)$ factors as $BA$ with inner
dimension $r$ (\auxref{B.11}(a)), $C=\MM_{\le\min(r,m,n)}$, which we write $\Mr$. For $X,Y\in C$,
$\rk(X-Y)\le\rk X+\rk Y\le2r$ and $\rk(X-Y)\le\min(m,n)$ (\auxref{B.12}(a)). Conversely, let $Z$ have rank
$k\le\min(2r,m,n)$ and SVD $Z=\sum_{i\le k}\sigma_iu_iv_i\T$ (\auxref{B.11}(a)). Put $Z_1=\sum_{i\le\min(k,r)}\sigma_iu_iv_i\T$
and $Z_2=Z-Z_1$, both of rank at most $r$. The cone $C$ is closed under negation, since $-BA=(-B)A$, so
$Z=Z_1-(-Z_2)\in C-C$. On several boxes, each with its own factors, $\delta$ is the product of the per-box maps, so $C$ is
the product of the per-box shapes and $C-C$ the product of the per-box differences.

An exact pointing of $\mathrm{LoRA}_r$ is a re-pointed method $M_q$ of (a), so its image lies in $\theta_0+(C-C)$, and in
every box the update $\theta-\theta_0$ has rank at most $\min(2r,m_b,n_b)$. For an arbitrary starting point $q_0$ and an
arbitrary residual $W_{\rm res}$, $\rho(q)-\rho(q_0)=\delta(q)-\delta(q_0)\in C-C$ with no condition at all.

For a pipeline with pointing defect $e_{\rm pt}=\rho(q_0)-\theta_0$, the update measured from $\theta_0$ is
\begin{equation*}
  \theta-\theta_0=\big(\rho(q)-\rho(q_0)\big)+e_{\rm pt},
\end{equation*}
a matrix of rank at most $2r$ plus the defect. The defects are those of \thmref{Proposition}{V.2}: $e=\kappa\theta_0-
\theta_0$ for QLoRA, $e-s\,\mathrm{SVD}_r(e)$ for one-step LoftQ, and $\kappa(W_{\rm res})-W_{\rm res}$ for QPiSSA. Each is
a quantisation error, or a quantisation error minus a matrix of rank at most $r$, so it has rank at least $\min(m,n)-r$
whenever the quantisation error involved has full rank (\auxref{B.12}(a)). Full rank is the generic case recorded in
\thmref{Proposition}{V.2}(a). For instance, a round-to-nearest quantiser with fixed scales is constant on each open cell of
constant codes, so there $e(W)=c-W$, which is rank-deficient only on a translate of $\MM_{\le\min(m,n)-1}$, a proper
algebraic subset of the cell; the cells cover all weights outside a null set (\auxref{B.10}(b)).

Attainment: let $2r\le\min(m,n)$, and take a split pointing with $P=sB_0A_0$ of rank $r$. Choose $Y=sBA$ of rank $r$ with
$\operatorname{col}Y\cap\operatorname{col}P=0$ and $\operatorname{row}Y\cap\operatorname{row}P=0$, possible since
$2r\le\min(m,n)$. Writing $Y=U_YV_Y\T$ and $P=U_PV_P\T$ with factors of full column rank $r$, we have
$Y-P=[U_Y\ U_P]\,[V_Y\ {-V_P}]\T$, and both $m\times2r$ and $n\times2r$ factors have independent columns, so
$\rk(Y-P)=2r$. The weight $\theta_0-P+Y$ lies in the image of the split pointing, with update $Y-P$ of rank $2r$.
\end{proof}

\subsection{Theorem II.2}\label{proof:II.2}

\begin{restate}{Theorem}{II.2}{three strata of LoRA initialisations}
Let $\delta(B,A)=BA$ with $1\le r<\min(m,n)$. Consider pointed LoRA objects in three distinguished strata of full-rank
pointings (pointings whose nonzero factors have full rank). Only $\mathrm S$ is open and dense in the space of pointings;
$\mathrm T_A$ and $\mathrm T_B$ are nowhere dense.
\begin{itemize}
  \item $(\mathrm T_A)$: $B_0=0$ and $\rk A_0=r$;
  \item $(\mathrm T_B)$: $A_0=0$ and $\rk B_0=r$;
  \item $(\mathrm S)$: $\rk B_0A_0=r$, with frozen residual $\theta_0-B_0A_0$.
\end{itemize}

(a) \textbf{Classification.} Within a stratum, $M_{q_0}\cong M_{q_0'}$ if and only if, respectively,
$\operatorname{row}A_0=\operatorname{row}A_0'$; $\operatorname{col}B_0=\operatorname{col}B_0'$; or $B_0A_0=B_0'A_0'$.
Objects in different strata are never isomorphic. As a \emph{set}, the isomorphism classes of the three strata therefore
form $\Gr(r,\R^n)\sqcup\Gr(r,\R^m)\sqcup\MM_r(m\times n)$.

(b) \textbf{First order.} $\dim S_1$ is $mr$, $nr$ and $r(m+n-r)$ respectively, while $d=r(m+n-r)$ in every case. The
deficits are $r(n-r)$, $r(m-r)$ and $0$.

(c) \textbf{Geometry.} In $\mathrm T_A$ and $\mathrm T_B$, $\Im=\theta_0+\Mr$, whose vertex $\theta_0$ is an
\textbf{apex}. In $\mathrm S$, $\Im=(\theta_0-P)+\Mr$ with $P=B_0A_0$, and $\theta_0$ is a \textbf{smooth point}.

(d) \textbf{Incomparability.} The images of $\mathrm S$-type and $\mathrm T$-type objects are incomparable.

(e) \textbf{Scope.} The three strata do not exhaust the pointings of LoRA$_r$. Outside them lie: $B_0=0$ with
$\rk A_0<r$; $A_0=0$ with $\rk B_0<r$; $B_0=A_0=0$; $0<\rk B_0A_0<r$; and pointings with $B_0,A_0$ both nonzero and
$B_0A_0=0$. HF PEFT ships one of the last kind: \texttt{init\_lora\_\allowbreak weights=\allowbreak"orthogonal"} (which
requires $r$ even) lies in the stratum $(\operatorname{rk}B_0,\operatorname{rk}A_0,\operatorname{rk}B_0A_0)=(r/2,r/2,0)$,
an apex with $\dim S_1=\tfrac r2(m+n-\tfrac r2)$. It is isomorphic to no object of the three strata. Its image
$\theta_0+\Mr$ rules out $\mathrm S$, and its $S_1$ determines $(\operatorname{col}B_0,\operatorname{row}A_0)$, of
dimensions $(r/2,r/2)$, against $(0,r)$ in $\mathrm T_A$ and $(r,0)$ in $\mathrm T_B$. The dimension of $S_1$ alone does
not separate it, since at $(m,n,r)=(4,5,2)$ it equals $mr$. This placement holds in exact arithmetic; HF casts the two
factors to the base dtype separately, and in bf16 the result is a generic pointing with a tiny defect. For every
pointing, the image determines $P=B_0A_0$ (its vertex is $\theta_0-P$), and $S_1=\{XA_0+B_0Y\}$ determines
$(\operatorname{row}A_0,\operatorname{col}B_0)$ whenever these are proper subspaces. The classification is set-level: in
the space of pointings $\mathrm T_A$ lies in the closure of $\mathrm S$, since $(\varepsilon B',A_0)\to(0,A_0)$, so the
quotient is not a topological disjoint union.
\end{restate}

\begin{proof}
Absorb the scale $s$ into $B$; this changes neither the strata nor the images. The space of pointings is
$\R^{m\times r}\times\R^{r\times n}$, and the residual is determined by the pointing.

\emph{Openness.} $\rk B_0A_0\le\min(\rk B_0,\rk A_0)$, and if both factors have rank $r$ then $B_0$ is injective and
$A_0$ surjective, so $\rk B_0A_0=r$. Hence $\mathrm S=\{\rk B_0=\rk A_0=r\}$ is the set where some $r\times r$ minor of
$B_0A_0$ is nonzero: it is open, and dense because its complement is a proper algebraic subset (\auxref{B.10}(a)). The
stratum $\mathrm T_A$ lies in the proper linear subspace $\{B_0=0\}$, which is closed with empty interior, so
$\mathrm T_A$ is nowhere dense, and likewise $\mathrm T_B$.

\emph{A translation argument.} We use the following fact here and in Theorem~II.11 and Proposition~II.16. Let
$1\le r<\min(m,n)$ and $D\ne0$, of rank $k$. If $k\ge r+1$ put $X=0$. Otherwise choose $X$ of rank $r+1-k\le r$ with
$\operatorname{col}X\cap\operatorname{col}D=0$ and $\operatorname{row}X\cap\operatorname{row}D=0$, possible since
$r+1\le\min(m,n)$. Factoring $D=U_DV_D\T$ and $X=U_XV_X\T$ with factors of full column rank, $D+X=[U_D\ U_X][V_D\ V_X]\T$
with both factors of full column rank, so $\rk(D+X)=r+1$. In either case $X\in\Mr$ and $D+X\notin\Mr$. Consequently,
\begin{equation}\label{eq:translation-stabiliser}
  c+\Mr=c'+\Mr\quad\Longrightarrow\quad c=c',
\end{equation}
since otherwise, with $D=c-c'$, the point $c+X=c'+(D+X)$ lies in the left side and not in the right side. The
translation stabiliser of $\Mr$ is $\{0\}$.

\emph{(a), ``if''.} In $\mathrm T_A$, let $\operatorname{row}A_0=\operatorname{row}A_0'$ with both of rank $r$. Then
$A_0'=A_0'A_0^+A_0=gA_0$ with $g=A_0'A_0^+\in\R^{r\times r}$, invertible because $\rk A_0'=r$. The map
$\varphi_g(B,A)=(Bg^{-1},gA)$ is a diffeomorphism of the parameter space with $\rho\circ\varphi_g=\rho$ (both residuals
are $\theta_0$) and $\varphi_g(0,A_0)=(0,A_0')$, a pointed isomorphism over $\Theta$. In $\mathrm T_B$,
$\operatorname{col}B_0=\operatorname{col}B_0'$ gives $B_0'=B_0k$ with $k\in\GL_r$, and $\varphi_{k^{-1}}(B_0,0)=(B_0',0)$.
In $\mathrm S$, let $B_0A_0=B_0'A_0'=P$ of rank $r$. Then
$\operatorname{col}B_0=\operatorname{col}P=\operatorname{col}B_0'$, so $B_0'=B_0k$ for a unique $k\in\GL_r$, and
$B_0kA_0'=B_0A_0$ gives $kA_0'=A_0$ because $B_0$ is injective. So $(B_0',A_0')=\varphi_{k^{-1}}(B_0,A_0)$, and the
residuals $\theta_0-P$ agree.

\emph{(a), ``only if''.} In $\mathrm T_A$, $d\rho_{(0,A_0)}(\dot B,\dot A)=\dot BA_0$, so
$S_1=\{XA_0\}=\mathcal R(\operatorname{row}A_0)$, which determines $\operatorname{row}A_0$ (proof of
\thmref{Observation}{I.7}); pointed isomorphisms preserve $S_1$ (\thmref{Observation}{I.6}). In $\mathrm T_B$,
$S_1=\mathcal C(\operatorname{col}B_0)$ determines $\operatorname{col}B_0$. In $\mathrm S$ the image is
$(\theta_0-P)+\Mr$ by (c), it is an isomorphism invariant, and by \eqref{eq:translation-stabiliser} it determines $P$.

\emph{Different strata.} $S_1(\mathrm T_A)=\mathcal R(V)$ with $\dim V=r\ge1$ and $S_1(\mathrm T_B)=\mathcal C(U)$ with
$\dim U=r<m$. They differ by the second fact in the proof of \thmref{Observation}{I.7}. (When $m\ne n$ their dimensions $mr$
and $nr$ already differ.) $\mathrm S$ differs from both because $\dim S_1(\mathrm S)=r(m+n-r)$ by (b), which exceeds $mr$
by $r(n-r)>0$ and $nr$ by $r(m-r)>0$. Isomorphism classes in $\mathrm T_A$ correspond to $\operatorname{row}A_0$, and every
$r$-dimensional subspace occurs; in $\mathrm T_B$ to $\operatorname{col}B_0$; in $\mathrm S$ to $P$, and every rank-$r$
matrix has a full-rank factorisation. This gives the stated disjoint union of sets.

\emph{(b)} At $(B_0,A_0)$, $S_1=\{XA_0+B_0Y\}$, of dimension $m\rk A_0+n\rk B_0-\rk A_0\rk B_0$ by
\thmref{Proposition}{II.3}: $mr$ in $\mathrm T_A$, $nr$ in $\mathrm T_B$ and $r(m+n-r)$ in $\mathrm S$. In every case the
image is a translate of $\Mr$ by (c), of dimension $r(m+n-r)$ (\auxref{B.5}(a)), so $d=r(m+n-r)$, and the deficits are
$r(n-r)$, $r(m-r)$ and $0$.

\emph{(c)} In $\mathrm T_A$ and $\mathrm T_B$ the residual is $\theta_0$ and the image is $\theta_0+\{BA\}=\theta_0+\Mr$.
This is a cone with vertex $\theta_0$, and it is not an affine subspace, because $\sum_{i\le r}E_{ii}$ and
$E_{r+1,r+1}$ lie in $\Mr$ while their sum has rank $r+1\le\min(m,n)$. By the cone part of \thmref{Definition}{I.5}
(\auxref{B.6}), $\theta_0$ is a singular point, and by (b) the deficit is positive, so $\theta_0$ is an apex. In
$\mathrm S$ the image is $(\theta_0-P)+\Mr$, and $\theta_0=(\theta_0-P)+P$ with $P\in\MM_r$. The stratum $\MM_r$ is open in
$\Mr$ and an embedded submanifold of dimension $r(m+n-r)$ (\auxref{B.5}(b)), so the image is a $C^1$ submanifold near
$\theta_0$, which is a smooth point; its deficit is $0$.

\emph{(d)} The $\mathrm T$-image is $\theta_0+\Mr$ and the $\mathrm S$-image $(\theta_0-P)+\Mr$ with $P\ne0$. The
translation argument with $D=P$ gives $X\in\Mr$ with $\theta_0+X$ outside the $\mathrm S$-image, and with $D=-P$ it gives
$X'\in\Mr$ with $\theta_0-P+X'$ outside the $\mathrm T$-image. (For a generic rank-one $X$, $\rk(P+X)=\rk(X-P)=r+1$
already.)

\emph{(e)} Every pointing has a rank triple $(\rk B_0,\rk A_0,\rk B_0A_0)$. The strata are $(0,r,0)$, $(r,0,0)$ and
$(r,r,r)$. If $B_0=0$ and $A_0\ne0$ is not of rank $r$, or $A_0=0$ and $B_0\ne0$ is not of rank $r$, or $B_0=A_0=0$, the
pointing is in the first three listed families. Otherwise both factors are nonzero, and either $\rk B_0A_0=r$, which
forces both ranks to be $r$ and gives $\mathrm S$, or $0<\rk B_0A_0<r$, or $B_0A_0=0$. So the list is exhaustive.

For every pointing, $S_1=\{XA_0+B_0Y\}=\{Z:Z(\operatorname{row}A_0)^\perp\subseteq\operatorname{col}B_0\}$. Indeed,
$(XA_0+B_0Y)w=B_0Yw$ for $w\in\ker A_0=(\operatorname{row}A_0)^\perp$; conversely, if
$Z\ker A_0\subseteq\operatorname{col}B_0$, then $Z=ZA_0^+A_0+B_0B_0^+Z(I-A_0^+A_0)$. Its annihilator in the Frobenius inner
product is $\{W:WA_0\T=0,\ B_0\T W=0\}$, the matrices with column space in $(\operatorname{col}B_0)^\perp$ and row space
in $(\operatorname{row}A_0)^\perp$, that is $(\operatorname{col}B_0)^\perp\otimes(\operatorname{row}A_0)^\perp$. When both
subspaces are proper, both complements are nonzero, and the column spaces of the elements of the annihilator span
$(\operatorname{col}B_0)^\perp$ while their row spaces span $(\operatorname{row}A_0)^\perp$. So $S_1$ determines
$(\operatorname{col}B_0,\operatorname{row}A_0)$ and their dimensions, and so does every pointed isomorphism class
(\thmref{Observation}{I.6}). The image is $(\theta_0-P)+\Mr$ with $P=B_0A_0$ (residual $\theta_0-P$), and it determines
$P$ by \eqref{eq:translation-stabiliser}.

HF's \texttt{orthogonal} initialisation draws one $Q\in\Ort(r)$ and sets $A_0=(G_AQ_{[0::2]})\T/10$ and
$B_0=G_BQ_{[1::2]}/10$, with $G_A\in\R^{n\times r/2}$ and $G_B\in\R^{m\times r/2}$ random and $Q_{[0::2]}$, $Q_{[1::2]}$
the even- and odd-indexed rows of $Q$. Distinct rows of $Q$ are orthogonal, so $B_0A_0=G_BQ_{[1::2]}Q_{[0::2]}\T
G_A\T/100=0$, and both factors have rank $r/2$ for generic $G_A,G_B$. The residual is $\theta_0$ and the image is
$\theta_0+\Mr$, whose vertex is singular. By \thmref{Proposition}{II.3}, $\dim S_1=m\tfrac r2+n\tfrac r2-\tfrac{r^2}4=
\tfrac r2(m+n-\tfrac r2)$, and the deficit $r(m+n-r)-\tfrac r2(m+n-\tfrac r2)=\tfrac r2(m+n-\tfrac{3r}2)$ is positive, so
the base point is an apex. Its image differs from every $\mathrm S$-image by \eqref{eq:translation-stabiliser}, and its
pair of subspace dimensions $(r/2,r/2)$, determined by $S_1$ since both subspaces are proper, differs from $(0,r)$ and
$(r,0)$; so it is isomorphic to no object of the three strata. At $(m,n,r)=(4,5,2)$, $\tfrac r2(m+n-\tfrac r2)=8=mr$.

Finally, for $B'$ of rank $r$ and $\varepsilon\ne0$, $(\varepsilon B',A_0)\in\mathrm S$, and $(\varepsilon B',A_0)\to
(0,A_0)\in\mathrm T_A$ as $\varepsilon\to0$, so $\mathrm T_A$ lies in the closure of $\mathrm S$.
\end{proof}

\subsection{Proposition II.3}\label{proof:II.3}

\begin{restate}{Proposition}{II.3}{first-order rank formula: smoothness, alignment and scale}
For $\kappa(B,A)=W_{\rm base}+sBA$ with $s\ne0$,
\begin{equation*}
  \rk d\kappa_{(B,A)}=m\operatorname{rk}A+n\operatorname{rk}B-\operatorname{rk}A\operatorname{rk}B\qquad\text{for all }(B,A),
\end{equation*}
and $\{XA\}\cap\{BY\}=\{BRA:R\in\R^{r\times r}\}$. For full-rank frames this overlap is the tangent space of the product
$\mathrm{FA}(A)\times\mathrm{FB}(B)$ of \thmref{Proposition}{I.8}(a), re-based at $\kappa(B,A)$, whose image is the meet
of the two one-sided images: a generalised LoRA-XS. It is LoRA-XS proper when $(B,A)$ span the top-$r$ singular subspaces
of $W_0$ ($\sigma_r>0$), as at a PiSSA init; bottom frames (MiLoRA) give a generalised LoRA-XS.

Under gradient flow $\dot B=-\eta_B\nabla_BL$, $\dot A=-\eta_A\nabla_AL$, with $G=\nabla_WL$,
\begin{equation*}
  \tfrac{d}{dt}L=-s^2\big(\eta_B\|GA\T\|_F^2+\eta_A\|B\T G\|_F^2\big)
\end{equation*}
at every time, in particular at $(B_0,A_0)$. To first order in the step sizes the same holds for full-batch gradient
descent, and in expectation for unbiased minibatch SGD. Under Adam ($\varepsilon\to0$, with bias correction) the first
step changes the loss by
\begin{equation*}
  -\lvert s\rvert\big(\eta_B\|GA_0\T\|_1+\eta_A\|B_0\T G\|_1\big)+O(\eta^2),
\end{equation*}
with entrywise $\ell_1$ norms. This first-order term is linear in $\lvert s\rvert$, against quadratic under gradient flow.
\end{restate}

\begin{proof}
\emph{Rank.} $d\kappa_{(B,A)}(\dot B,\dot A)=s(\dot BA+B\dot A)$, and $s\ne0$, so the image of the differential is
$\{XA\}+\{BY\}=\mathcal R(\operatorname{row}A)+\mathcal C(\operatorname{col}B)$. With $a=\rk A$ and $b=\rk B$, the two
summands have dimensions $ma$ and $nb$ (proof of \thmref{Observation}{I.7}). A matrix $D$ lies in both iff
$\operatorname{row}D\subseteq\operatorname{row}A$ and $\operatorname{col}D\subseteq\operatorname{col}B$. The space of such
$D$ is $\operatorname{col}B\otimes\operatorname{row}A$, of dimension $ab$, and every such $D$ satisfies
$D=BB^+DA^+A=B(B^+DA^+)A$, since $BB^+$ and $A^+A$ are the projectors onto $\operatorname{col}B$ and
$\operatorname{row}A$. Conversely $BRA$ lies in both summands. So $\{XA\}\cap\{BY\}=\{BRA\}$ and
$\rk d\kappa=ma+nb-ab$, for every $(B,A)$.

\emph{The overlap.} Let $B$ have full column rank and $A$ full row rank. Re-based at $W=\kappa(B,A)$, the one-sided methods
$X\mapsto W+XA$ and $Y\mapsto W+sBY$ are those of \thmref{Proposition}{I.8}(a) with frames $A$ and $sB$, so their product is
$R\mapsto W+sBRA$, a linear method whose image $W+\{BRA\}$ is its own tangent space at $W$. That image is the meet
$(W+\{XA\})\cap(W+\{BY\})$ by the identity just proved. If $\operatorname{col}B=\operatorname{col}U_r$ and
$\operatorname{row}A=\operatorname{row}V_r\T$ for the top-$r$ singular frames of $W_0$ with $\sigma_r>0$, then $B=U_rk_1$ and
$A=k_2V_r\T$ with $k_1,k_2\in\GL_r$, and $\{BRA\}=\{U_rR'\Sigma_rV_r\T\}$ with $R'=k_1Rk_2\Sigma_r^{-1}$ ranging over
$\R^{r\times r}$, which is LoRA-XS's update space (\thmref{Proposition}{I.8}(a)). For other frames, such as the bottom
singular frames of MiLoRA, the overlap is the same kind of sandwich in those frames.

\emph{Gradient flow.} With $W=\kappa(B,A)$, the chain rule gives $\nabla_B(L\circ\kappa)=sGA\T$ and
$\nabla_A(L\circ\kappa)=sB\T G$. Along the flow,
\begin{equation*}
  \tfrac{d}{dt}L=\langle G,s(\dot BA+B\dot A)\rangle=s\langle GA\T,\dot B\rangle+s\langle B\T G,\dot A\rangle
  =-s^2\big(\eta_B\|GA\T\|_F^2+\eta_A\|B\T G\|_F^2\big).
\end{equation*}
One step of full-batch gradient descent moves $W$ by $s(\Delta B\,A+B\,\Delta A)+s\,\Delta B\,\Delta A$ with
$\Delta B=-\eta_BsGA\T$ and $\Delta A=-\eta_AsB\T G$. For $L$ of class $C^2$ near $W$, the change of loss is
$\langle G,s(\Delta B\,A+B\,\Delta A)\rangle$ plus terms of second order in the step sizes, and the first term is the
expression above with $\eta$ in place of $\eta\,dt$. With an unbiased minibatch gradient $\hat G$, $\mathbb E\hat G=G$, the
first-order change is $-s^2(\eta_B\langle GA\T,\hat GA\T\rangle+\eta_A\langle B\T G,B\T\hat G\rangle)$, whose expectation is
the same expression.

\emph{Adam.} With bias correction, Adam's first moment and second moment after one step are $\hat m_1=g$ and
$\hat v_1=g\odot g$, where $g$ is the first gradient, so the first update is $-\eta\,g/(\lvert g\rvert+\varepsilon)$
entrywise, which tends to $-\eta\operatorname{sign}(g)$ as $\varepsilon\to0$ (entries with $g=0$ do not move). For $B$,
$g=sGA_0\T$, so $\delta B=-\eta_B\operatorname{sign}(s)\operatorname{sign}(GA_0\T)$, and its first-order effect on the loss is
$\langle G,s\,\delta B\,A_0\rangle=s\langle GA_0\T,\delta B\rangle=-\eta_B\lvert s\rvert\|GA_0\T\|_1$. The $A$-term is the same
with $B_0\T G$, and the cross term $s\,\delta B\,\delta A$ and the Taylor remainder are $O(\eta^2)$.

\emph{The alignment remark after the proposition.} Among $A_0$ with $A_0A_0\T=cI$, write $A_0=\sqrt c\,P$ with $P$ of
orthonormal rows. The $B$-term of the gradient-flow rate is $s^2\eta_B\|GA_0\T\|_F^2=s^2\eta_Bc\,\|GP\T\|_F^2$, and by Ky Fan's
maximum principle (\auxref{B.17}(a)) its largest value over $P$ is $s^2\eta_Bc\sum_{i\le r}\sigma_i(G)^2$, attained when
$\operatorname{row}A_0$ is a top-$r$ right singular subspace of $G$. Under Adam the first-step term
$\|GA_0\T\|_1$ is not invariant under $A_0\mapsto OA_0$ with $O\in\Ort(r)$, so it depends on the basis of
$\operatorname{row}A_0$ and not only on the subspace.
\end{proof}

\subsection{Lemma II.4}\label{proof:II.4}

\begin{restate}{Lemma}{II.4}{zero gate}
Let $Q=U\times V$ with $U$ a vector space, let $\rho=\theta_0+\beta(u,v)$ with $\beta$ smooth, and suppose
$\beta(u_0,v)=0$ for all $v$; for example $\beta(\cdot,v)$ linear and $u_0=0$. Then $D_v\rho_{q_0}=0$. Hence
$\nabla_v(L\circ\rho)(q_0)=0$, $S_1=D_u\beta(u_0,v_0)(U)$ ($=\beta(U,v_0)$ in the linear case), and the cometric at
$q_0$ is $K_{q_0}=D_u\beta\,\Lambda_{uu}\,D_u\beta\T$, where $\Lambda_{uu}$ is the $uu$-block of $\Lambda$; no
block-diagonality is needed, since the $v$-columns of $D\rho$ vanish. The weight trajectory $\theta(t)$ agrees with that
of the method with $v$ frozen at $v_0$ to first order in $t$ (gradient flow), or in the step size (one or more Adam
steps). The parameter $v$ itself may still move: under terms not of the form $L\circ\rho$ (regularisers, decoupled weight
decay), and under Adam, whose step-2 move of $v$ is $O(\eta)$ although its effect on $\theta$ is $O(\eta^2)$.

\emph{Function-level variant.} For an extension with realised function $F(u,v,x)$, if $F(u_0,v,x)=F_0(x)$ for all $v$
and $x$, then $D_vF=0$ at $q_0$, with the same conclusions at the level of functions. No linearity is needed.
\end{restate}

\begin{proof}
Since $\beta(u_0,v)=0$ for every $v$, differentiating in $v$ gives $D_v\beta(u_0,v)=0$ for every $v$, in particular at
$v_0$. So $D\rho_{q_0}=[\,D_u\beta(u_0,v_0)\ \ 0\,]$. Then $\nabla_v(L\circ\rho)(q_0)=D_v\rho_{q_0}\T\nabla L(\theta_0)=0$,
$S_1=D_u\beta(u_0,v_0)(U)$, and in the linear case $D_u\beta(0,v_0)u=\beta(u,v_0)$. For any symmetric $\Lambda$ written in
blocks,
\begin{equation*}
  K_{q_0}=\begin{bmatrix}D_u\beta&0\end{bmatrix}\begin{bmatrix}\Lambda_{uu}&\Lambda_{uv}\\ \Lambda_{vu}&\Lambda_{vv}
  \end{bmatrix}\begin{bmatrix}D_u\beta\T\\ 0\end{bmatrix}=D_u\beta\,\Lambda_{uu}\,D_u\beta\T .
\end{equation*}

\emph{Gradient flow.} For $\dot q=-\Lambda\nabla(L\circ\rho)(q)$, the weights $\theta(t)=\rho(q(t))$ satisfy
$\dot\theta(0)=-K_{q_0}\nabla L(\theta_0)$. The method with $v$ frozen, $u\mapsto\rho(u,v_0)$ with the metric
$\Lambda_{uu}$, has the same velocity at $t=0$, so the two weight trajectories agree to first order in $t$.

\emph{Adam.} Let $\eta$ be the step size and $\varepsilon>0$ fixed. At step one the $v$-gradient vanishes, so Adam's
first and second moments in $v$ vanish and $v$ does not move; the $u$-update is computed from the same gradient as for the
frozen method. So after one step the two methods have the same parameters and the same weights. At step two, the gradient
in $v$ is $D_v\beta(u_1,v_0)\T\nabla L$. Since $\beta$ is smooth and $D_v\beta(u_0,\cdot)\equiv0$, we have
$\|D_v\beta(u,v)\|=O(\|u-u_0\|)$ uniformly for $v$ near $v_0$, and $\|u_1-u_0\|=O(\eta)$, so this gradient is $O(\eta)$,
and the move of $v$ is at most of order $\eta$; it is of exact order $\eta$ when $\varepsilon$ is negligible against the
gradient, since Adam normalises it. Its effect on $\theta$ is
\begin{equation*}
  \rho(u,v)-\rho(u,v_0)=\int_0^1D_v\beta\big(u,v_0+\tau(v-v_0)\big)(v-v_0)\,d\tau=O(\|u-u_0\|\,\|v-v_0\|)=O(\eta^2).
\end{equation*}
For any fixed number $k$ of steps the same estimate propagates: the Adam update is a smooth function of the gradient
history when $\varepsilon>0$, and by induction on the step, $v_t-v_0=O(\eta)$, the $u$-iterates of the two methods
differ by $O(\eta^2)$ (their gradients differ by $O(\eta)$ and are multiplied by $\eta$), and the weights differ by
$O(\eta^2)$ by the display. For $\varepsilon\to0$ the same argument holds as long as no entry of the frozen method's
gradient vanishes in the first $k$ steps, so that the normalisation stays smooth. Terms that are not of the form
$L\circ\rho$, such as a regulariser $R(v)$ or decoupled weight decay $v\mapsto(1-\eta\lambda)v$, move $v$ already at step
one, which is why the statement concerns the weight trajectory.

\emph{Function-level variant.} If $F(u_0,v,x)=F_0(x)$ for all $v$ and $x$, then $D_vF(u_0,v_0,x)=0$ for every $x$, and the
argument above applies verbatim with the realised function in place of $\theta$.
\end{proof}

\subsection{Theorem II.5}\label{proof:II.5}

\begin{restate}{Theorem}{II.5}{complete invariants of the basic actions; classical}
Let $W,W'\in\R^{m\times n}$.
\begin{enumerate}[label=(\arabic*)]
  \item \emph{Right orthogonal (input side).} $W'=WR$ for some $R\in\Ort(n)$ iff $WW\T=W'W'^{\top}$. For $\SO(n)$, when
    $\rk W=n$, also require that the then unique $R$ have $\det R=1$.
  \item \emph{Left orthogonal.} $W'=PW$ for some $P\in\Ort(m)$ iff $W\T W=W'^{\top}W'$.
  \item \emph{Two-sided.} $W'=PWR$ for some $P\in\Ort(m)$ and $R\in\Ort(n)$ iff $W$ and $W'$ have the same singular
    values, counted with multiplicity.
  \item \emph{Left torus.} If $W$ has no zero row, $W'=DW$ for some $D\in T_m^+$ iff $w_i'\in\R_{>0}\,w_i$ for every $i$.
    For the full torus $T_m$, whose signs are unconstrained, the complete invariant is the set of zero rows together with
    the row \emph{lines}.
  \item \emph{Full Hadamard torus.} $W'=E\odot W$ for some $E\in(\R^\times)^{m\times n}$ iff $W$ and $W'$ have the same
    zero pattern.
\end{enumerate}
\end{restate}

\begin{proof}
Write $w_i,w_i'\in\R^n$ for the rows of $W$ and $W'$.

\emph{(1)} If $W'=WR$ with $R$ orthogonal, then $W'W'^{\top}=WRR\T W\T=WW\T$. Conversely, let $WW\T=W'W'^{\top}$. For
$c\in\R^m$,
\begin{equation*}
  \Big\lVert\sum_ic_iw_i\Big\rVert^2=c\T WW\T c=c\T W'W'^{\top}c=\Big\lVert\sum_ic_iw_i'\Big\rVert^2 .
\end{equation*}
Hence $\varphi:\sum_ic_iw_i\mapsto\sum_ic_iw_i'$ is well defined on $\operatorname{row}W$ (if $\sum_ic_iw_i=\sum_ic_i'w_i$, the
vector $\sum_i(c_i-c_i')w_i'$ has norm $0$), linear, isometric and onto $\operatorname{row}W'$. In particular both row spaces
have dimension $\rk WW\T$. By Witt's extension theorem (\auxref{B.25}(b)), $\varphi$ extends to some $S\in\Ort(n)$, for
instance by an isometry between the orthogonal complements, which have equal dimension. Then $Sw_i=w_i'$ for every $i$,
that is $W'=WS\T$, and $R=S\T\in\Ort(n)$.

For $\SO(n)$: if $\rk W<n$, the complement $(\operatorname{row}W)^\perp$ is nonzero, and composing the extension with a
reflection of that complement changes $\det S$ without changing $Sw_i$; so some $R\in\SO(n)$ exists as soon as some
$R\in\Ort(n)$ does. If $\rk W=n$, then $W$ has full column rank and $WR=W'$ determines $R=W^+W'$ uniquely, so
$W'\in W\cdot\SO(n)$ iff this $R$ has determinant $1$.

\emph{(2)} $W'=PW$ iff $W'^{\top}=W\T P\T$ with $P\T\in\Ort(m)$, and by (1) applied to $W\T$ and $W'^{\top}$ this happens
iff $W\T W=W'^{\top}W'$.

\emph{(3)} If $W'=PWR$, then $W'^{\top}W'=R\T(W\T W)R$ has the eigenvalues of $W\T W$, so the singular values agree with
multiplicity. Conversely, if $W=U\Sigma V\T$ and $W'=U'\Sigma V'^{\top}$ are singular value decompositions with the same
$\Sigma$ (\auxref{B.11}(a)), then $W'=(U'U\T)\,W\,(VV'^{\top})$ with both factors orthogonal.

\emph{(4)} For $D=\diag(d)$ the rows of $DW$ are $d_iw_i$. If $d\in\R^m_{>0}$, then $w_i'=d_iw_i\in\R_{>0}w_i$. Conversely,
if $w_i'=d_iw_i$ with $d_i>0$ for every $i$ ($d_i$ is unique because $w_i\ne0$), then $W'=\diag(d)W$ with
$\diag(d)\in T_m^+$. For $T_m$, the rows $d_iw_i$ with $d_i\ne0$ vanish exactly where $w_i$ does and otherwise span the line
$\R w_i$. Conversely, if $W$ and $W'$ have the same zero rows and $\R w_i'=\R w_i$ for every nonzero row, then $w_i'=d_iw_i$
with $d_i\ne0$, and with $d_i=1$ on the zero rows $W'=\diag(d)W$, $\diag(d)\in T_m$.

\emph{(5)} $(E\odot W)_{ij}=E_{ij}W_{ij}$ vanishes iff $W_{ij}$ does, because $E_{ij}\ne0$. Conversely, if $W$ and $W'$
have the same zero pattern, put $E_{ij}=W'_{ij}/W_{ij}$ on the support and $E_{ij}=1$ off it; then $E$ has no zero entry
and $W'=E\odot W$.
\end{proof}

\noindent Item (1) is also the orbit-separation statement of the first fundamental theorem for $\Ort(n)$
(\auxref{B.25}(a)); the proof above uses only the extension of isometries.

\subsection{Corollary II.6}\label{proof:II.6}

\begin{restate}{Corollary}{II.6}{what exactly orthogonal methods can never do}
Let a method have image in $W_0\cdot\Ort(n)$ (input side), and work in exact arithmetic. This covers:
\begin{itemize}
  \item OFT and block OFT with an exactly orthogonal parametrisation: the original paper's exact Cayley map, and HF PEFT
    from 0.16 with \texttt{use\_cayley\_\allowbreak neumann=False} on Linear and Conv targets;
  \item GOFT with strict Givens rotations and no scaler, as in the paper (\texttt{no\_scaling=True} in the official
    peft-givens code);
  \item HRA, and ETHER as defined in its paper (not ETHER+);
  \item any number of merge-and-restart cycles of these.
\end{itemize}
Then:
\begin{itemize}
  \item it preserves $K_{\mathrm{out}}$, hence every neuron norm and the angle between any two nonzero neurons, and so the
    hyperspherical energy, which depends only on these angles;
  \item it preserves the whole singular spectrum of $W_0$;
  \item block-diagonal OFT with input blocks $J$, with the same partition in every cycle, preserves each partial Gram
    matrix $W_{:,J}W_{:,J}^{\top}$.
\end{itemize}
\end{restate}

\begin{proof}
\emph{Coverage.} Each listed parametrisation produces $W=W_0R$ with $R\in\Ort(n)$, read in the convention $y=Wx$, where an
input-side rotation acts on the columns of $W_0$. The exact Cayley map $C(\Omega)=(I+\Omega)(I-\Omega)^{-1}$ of a skew
matrix is orthogonal (\auxref{B.23}(a)); HF PEFT's input-side OFT merges to $W_0R\T$, and $R\T$ is orthogonal too. A
block-diagonal matrix with orthogonal blocks is orthogonal. A Givens rotation is orthogonal, and so is a product of them.
HRA multiplies Householder reflections $H_u=I-2uu\T/\lVert u\rVert^2$, which are orthogonal, and ETHER as defined in its
paper multiplies $W_0$ on the input side by one reflection $I-2\hat u\hat u\T$ with $\lVert\hat u\rVert=1$. For merge and
restart, if every weight reachable in cycle $k$ is $W_{k-1}R_k$ with $R_k\in\Ort(n)$, then by induction
$W_k=W_0R_1\cdots R_k$, and $R_1\cdots R_k\in\Ort(n)$.

\emph{Gram matrix and spectrum.} For $W=W_0R$ with $R\in\Ort(n)$, $K_{\mathrm{out}}(W)=W_0RR\T W_0\T=K_{\mathrm{out}}(W_0)$
(\thmref{Theorem}{II.5}(1)). The neuron norms are $\lVert w_i\rVert^2=K_{\mathrm{out}}(W)_{ii}$, and the cosine between two
nonzero neurons is $K_{\mathrm{out}}(W)_{ij}/(K_{\mathrm{out}}(W)_{ii}K_{\mathrm{out}}(W)_{jj})^{1/2}$, so both are
preserved. The hyperspherical energy of \citet{liu2018learning} is a function of the distances between the normalised
neurons, $\lVert\hat w_i-\hat w_j\rVert^2=2-2\cos_{ij}$, so it is preserved as well. The singular values of $W$ are the
square roots of the eigenvalues of $K_{\mathrm{out}}(W)=K_{\mathrm{out}}(W_0)$ (equivalently, \thmref{Theorem}{II.5}(3)
with $P=I$).

\emph{Partial Grams.} Let $R=\operatorname{blockdiag}(R_J)_J$ be block-diagonal for a partition of the input coordinates
into blocks $J$, with orthogonal blocks. The columns of $W_0R$ indexed by $J$ are $(W_0R)_{:,J}=(W_0)_{:,J}R_J$, so
$W_{:,J}W_{:,J}^{\top}=(W_0)_{:,J}R_JR_J\T(W_0)_{:,J}^{\top}=(W_0)_{:,J}(W_0)_{:,J}^{\top}$. With the same partition in
every cycle, $R_1\cdots R_k$ is again block-diagonal with orthogonal blocks, so the identity survives merging.
\end{proof}

The remarks after the corollary in \S\ref{sec:erlangen-orthogonal} make several claims that do not follow from its
hypothesis. We prove them in the same order. Throughout, $K_{\mathrm{out}}(W)=WW\T$, and for an input-side factor $G$ we have
$K_{\mathrm{out}}(W_0G)=W_0\,GG\T\,W_0\T$.

\emph{BOFT and HF OFT 0.14--0.15.} For $W=\diag(s)W_0R\T$ with $R$ orthogonal,
\begin{equation*}
  K_{\mathrm{out}}(W)=\diag(s)\,W_0R\T RW_0\T\diag(s)=\diag(s)\,K_{\mathrm{out}}(W_0)\,\diag(s),
\end{equation*}
with entries $s_is_jK_{\mathrm{out}}(W_0)_{ij}$. If $s_is_j\ne0$ and the neurons $i,j$ of $W_0$ are nonzero,
\begin{equation*}
  \cos'_{ij}=\frac{s_is_jK_{\mathrm{out}}(W_0)_{ij}}{\lvert s_i\rvert\lvert s_j\rvert\,(K_{\mathrm{out}}(W_0)_{ii}
  K_{\mathrm{out}}(W_0)_{jj})^{1/2}}=\operatorname{sign}(s_is_j)\cos_{ij} .
\end{equation*}
So $\lvert\cos_{ij}\rvert$ is preserved whenever $s_is_j\ne0$, and $\cos_{ij}$ itself whenever $s_is_j>0$. If every $s_i$ is
nonzero and all share a sign, every cosine is preserved. Conversely, if every cosine of $W_0$ is nonzero and every cosine is
preserved, then every $s_i$ is nonzero (otherwise neuron $i$ of $W$ vanishes and its cosines are undefined) and
$\operatorname{sign}(s_is_j)=1$ for all $i\ne j$, so all $s_i$ share a sign. With the butterfly permutation equal to the
identity, $R$ is block-diagonal with orthogonal blocks $R_J$ over a partition of the input coordinates, and
$W_{:,J}=\diag(s)(W_0)_{:,J}R_J\T$, so $W_{:,J}W_{:,J}^{\top}=\diag(s)(W_0)_{:,J}(W_0)_{:,J}^{\top}\diag(s)$. HF PEFT OFT
0.14--0.15 has the same form with its output scale \texttt{oft\_s} in place of $s$, which gives the same conclusions.

\emph{GOFT with its scaler.} For $W=W_0\diag(g)G$ with $G$ orthogonal, whatever $G$ is,
\begin{equation*}
  K_{\mathrm{out}}(W)=W_0\diag(g)\,GG\T\diag(g)\,W_0\T=W_0\diag(g)^2W_0\T .
\end{equation*} This differs from
$K_{\mathrm{out}}(W_0)$ in general (for $W_0=I_n$ it is $\diag(g)^2$), so the image is not contained in $W_0\cdot\Ort(n)$
and the corollary does not apply.

\emph{Output-side implementations.} If $W=PW_0$ with $P\in\Ort(m)$, then $K_{\mathrm{in}}(W)=W_0\T P\T PW_0=K_{\mathrm{in}}(W_0)$,
and the singular values are preserved by \thmref{Theorem}{II.5}(3). If $P$ is block-diagonal with orthogonal blocks $P_I$
over a partition of the output coordinates, the rows indexed by $I$ satisfy $W_{I,:}=P_I(W_0)_{I,:}$, so
$W_{I,:}^{\top}W_{I,:}=(W_0)_{I,:}^{\top}(W_0)_{I,:}$. In general $K_{\mathrm{out}}(PW_0)=PK_{\mathrm{out}}(W_0)P\T$
differs from $K_{\mathrm{out}}(W_0)$, for instance for $W_0=E_{11}$ and $P$ a rotation moving $e_1$. ETHER's code with
\texttt{ether\_nb}${}>1$ multiplies the rows of output group $g$ on the right by a reflection $H_g$, $W_{I_g,:}=
(W_0)_{I_g,:}H_g$. Within a group, $W_{I_g,:}W_{I_g,:}^{\top}=(W_0)_{I_g,:}(W_0)_{I_g,:}^{\top}$, which keeps neuron norms
and within-group angles. Across groups the inner products become $w_{0,i}\T H_gH_{g'}w_{0,j}$. For two groups of one row
each, with $w_{0,1}=w_{0,2}=e_1$ in $\R^2$, $H_1=I$ and $H_2$ the reflection that swaps $e_1$ and $e_2$, the two neurons
become $e_1$ and $e_2$: the cosine drops from $1$ to $0$, and the singular values change from $(\sqrt2,0)$ to $(1,1)$.

\emph{HF PEFT's default OFT.} Let $Q$ be skew. Then $(I-Q)(I+Q+Q^2+Q^3)=I-Q^4$ and $I-Q$ is invertible
(\auxref{B.23}(a),(e)), so
\begin{equation*}
  R:=I+2Q+2Q^2+2Q^3+Q^4=(I+Q)(I+Q+Q^2+Q^3)=(I+Q)(I-Q)^{-1}(I-Q^4)=C(Q)\,(I-Q^4),
\end{equation*}
where all factors are polynomials in $Q$ or their inverses and commute. The matrix $Q^4$ is symmetric, since
$(Q^4)\T=(-Q)^4$, and $C(Q)$ is orthogonal, so $R\T R=(I-Q^4)C(Q)\T C(Q)(I-Q^4)=(I-Q^4)^2$. If $\pm i\theta_k$ are the
eigenvalues of $Q$, then $Q^4$ has eigenvalues $\theta_k^4$ (and $0$), so the singular values of $R$ are
$\lvert1-\theta_k^4\rvert$ (and $1$). Thus $R\T R=I$ only if every $\theta_k$ is $0$ or $2^{1/4}$; in particular
$R\T R\ne I$ for every $Q\ne0$ with $\lVert Q\rVert_{\mathrm{op}}<2^{1/4}$, and $R$ is singular when some $\theta_k=1$.
The adapted weight is $W=W_0R\T$, with $K_{\mathrm{out}}(W)=W_0R\T RW_0\T=W_0(I-Q^4)^2W_0\T$, and
$(I-Q^4)^2-I=Q^8-2Q^4$ is of order $\lVert Q\rVert^4$. When $\lVert Q\rVert_{\mathrm{op}}\le2^{1/4}$, every
$\lvert1-\theta_k^4\rvert\le1$, so $R$ is a contraction, and each merge $W\mapsto WR\T$ gives
$WR\T RW\T\preceq WW\T$ and lowers every singular value (\auxref{B.18}(b),(c)); \thmref{Theorem}{V.3}, corollary 4, records
the iteration. In PEFT 0.16 and 0.17 the last coefficient was $2$, $R=p(Q)$ with $p(x)=1+2x+2x^2+2x^3+2x^4$. Then
$R\T=p(-Q)$, and writing $p=E+O$ with even part $E=1+2x^2+2x^4$ and odd part $O=2x+2x^3$,
\begin{equation*}
  p(x)p(-x)=E^2-O^2=(1+4x^2+8x^4+8x^6+4x^8)-(4x^2+8x^4+4x^6)=1+4x^6+4x^8,
\end{equation*}
so $R\T R=I+4Q^6+4Q^8$. Its eigenvalues are $1+4\theta_k^6(\theta_k^2-1)$, so it differs from $I$ unless every $\theta_k$ is $0$
or $1$, and the deviation is of order $\lVert Q\rVert^6$.

\emph{One non-orthogonal plane changes a cosine.} We prove the following, which covers qGOFT, ETHER+ and HF's default OFT.
Let $n>2$ and let $G\in\R^{n\times n}$ be an input-side factor such that $S=GG\T$ acts as the identity on the orthogonal
complement of a plane $\Pi$, maps $\Pi$ into itself, and $S|_\Pi\ne I_\Pi$ (that is, $G$ is not orthogonal on $\Pi$).
Then for $W_0$ outside a nowhere-dense null set some cosine of $W_0G$ differs from the corresponding cosine of $W_0$.
Indeed, for two rows $w_1,w_2$ of $W_0$ put
\begin{equation*}
  \phi(W_0)=(w_1\T Sw_2)^2\,\lVert w_1\rVert^2\lVert w_2\rVert^2-(w_1\T w_2)^2\,(w_1\T Sw_1)(w_2\T Sw_2),
\end{equation*}
a polynomial in $W_0$. Where both cosines are defined, $\cos'_{12}=\cos_{12}$ forces $\phi(W_0)=0$. Choose a unit vector
$c\perp\Pi$ (possible since $n>2$) and $x\in\Pi$ with $x\T Sx\ne\lVert x\rVert^2$ (possible since $S|_\Pi-I_\Pi$ is a
nonzero symmetric map), and take $w_1=c$, $w_2=c+x$. Then $w_1\T Sw_2=1$, $w_1\T Sw_1=1$, $w_2\T Sw_2=1+x\T Sx$,
$\lVert w_2\rVert^2=1+\lVert x\rVert^2$ and $w_1\T w_2=1$, so $\phi=\lVert x\rVert^2-x\T Sx\ne0$. By \auxref{B.10}(b), $\phi\ne0$
on an open dense set of full measure, which proves the claim. A conformal block $\sqrt c\,\mathrm{Rot}$ with $c\ne1$ on two
input coordinates has $S|_\Pi=cI$, so it is covered. For qGOFT, take one block non-orthogonal and the others equal to $I$. For
ETHER+, take the output factor equal to $I$ (its two vectors equal) and the input factor $I-uu\T+vv\T$ with unit $u\ne\pm v$:
it is symmetric, equal to $I$ on $\{u,v\}^\perp$, preserves $\Pi=\spanop\{u,v\}$, and on $\Pi$ has eigenvalues
$1\pm(1-(u\T v)^2)^{1/2}$, so it is not orthogonal there; for $u\perp v$ it kills $u$ and is singular. For HF's default
OFT, take $Q=\theta(e_1e_2\T-e_2e_1\T)$ with $\theta\notin\{0,2^{1/4}\}$; then $R\T R=(I-Q^4)^2$ is $(1-\theta^4)^2I$ on
$\Pi=\spanop\{e_1,e_2\}$ and $I$ elsewhere.

\emph{Finite precision.} The last remark concerns floating-point arithmetic, and Appendix~\ref{app:repro} reports the
measured drift; in exact arithmetic the corollary holds as proved.

\subsection{Theorem II.7}\label{proof:II.7}

\begin{restate}{Theorem}{II.7}{HRA is the meet of LoRA with the rotation orbit, and it starts at an apex}
Let $W_0\in\R^{m\times n}$ be injective ($\rk W_0=n$), and let $r$ be even with $2\le r\le n-2$. Define
$\mathrm{HRA}_r:(u_1,\dots,u_r)\mapsto W_0H_{u_1}\cdots H_{u_r}$ with $u_i\ne0$, where
$H_u=I-2uu\T/\lVert u\rVert^2$. Write $\mathrm{OFT}_{\SO}:=W_0\cdot\SO(n)$ for the rotation orbit. (The $\Ort(n)$-orbit
would add the odd component: $W_0H_{e_1}$ lies in $W_0+\Mr$ but not in $\Im\mathrm{HRA}_r$.)
\begin{enumerate}[label=(\alph*)]
  \item \textbf{Meet.} $\Im\mathrm{HRA}_r=(W_0+\Mr)\cap W_0\cdot\SO(n)$. Further, $d(\mathrm{HRA}_r)=rn-\tfrac{r(r+1)}2$,
    so the generic fibre has dimension $\tfrac{r(r+1)}2=rn-d$. This gauge is non-linear. The evident continuous global
    symmetries are the $r$ scalings $u_i\mapsto c_iu_i$; the braid group also acts globally by Hurwitz moves
    $(u_i,u_{i+1})\mapsto(H_{u_i}u_{i+1},u_i)$, since $H_aH_b=H_{H_ab}H_a$.
  \item \textbf{First-order collapse.} At the paired initialisation $u_{2i-1}=u_{2i}=w_i$ ($i\le k=r/2$, with the $w_i$
    independent and $U=\spanop\{w_i\}$),
    \begin{equation*}
      S_1=W_0\cdot\{\Omega\in\so(n):P_{U^\perp}\Omega P_{U^\perp}=0\},\qquad\dim S_1=kn-\tfrac{k(k+1)}2<d(\mathrm{HRA}_r).
    \end{equation*}
  \item \textbf{Vertex.} The germ of $\Im\mathrm{HRA}_r$ at $W_0$ is diffeomorphic to the germ at $0$ of the cone
    $\Sigma_r=\{\Omega\in\so(n):\rk\Omega\le r\}$ (the skew-determinantal cone), which is not linear. So $W_0$ is a
    singular point (the vertex) of the image. Together with (b), the paired initialisation is an \textbf{apex}.
  \item \textbf{Scope.}
    \begin{itemize}
      \item \emph{Not a product.} (a) is a meet in the poset of subsets of $\Theta$ containing $W_0$
        (\thmref{Observation}{I.4}(c)). HRA is not the
        categorical product of $\mathrm{LoRA}_r$ and $\mathrm{OFT}_{\SO}$: the set-level fibre product
        $\{(B,A,R):W_0+BA=W_0R\}$ has local dimension $d+r^2=rn+\tfrac{r(r-1)}2\ne rn$ over a generic point of the
        meet. (It is not pure-dimensional: over $W_0$ alone it contains $\{(B,0,I)\}$, of dimension $mr$.)
      \item \emph{Exactly orthogonal OFT is smaller.} The Cayley map never produces a rotation with eigenvalue $-1$.
        For exact full-block Cayley OFT (the original paper; HF with \texttt{use\_cayley\_\allowbreak neumann=False}),
        $(W_0+\Mr)\cap\Im\mathrm{OFT}$ is dense in $\Im\mathrm{HRA}_r$ but not equal to it: $W_0H_{e_1}H_{e_2}$, with
        $H_{e_1}H_{e_2}=\diag(-1,-1,1,\dots,1)$, is missing. For exact block Cayley OFT with $b<n$ the intersection is a
        proper subset of $\Im\mathrm{HRA}_r$.
      \item \emph{HF's default OFT is incomparable.} HF PEFT's default OFT is not orthogonal
        (\thmref{Corollary}{II.6}); its flag is \texttt{use\_cayley\_\allowbreak neumann=\allowbreak True}. Since $R\T R=(I-Q^4)^2=I$ forces $Q=0$ near $Q=0$, its image meets
        $\Im\mathrm{HRA}_r$ near $W_0$ only at $W_0$, while its intersection with $W_0+\Mr$ contains non-orthogonal
        points such as $W_0(I+2Q+2Q^2+2Q^3+Q^4)$ with $\rk Q=2$.
      \item \emph{Implementation.} The theorem concerns HRA with $\lambda<\infty$ (HF \texttt{apply\_GS=False}). The
        strictly orthogonal Gram--Schmidt variant has the smaller image $W_0(I-2P_U)$, of dimension $r(n-r)$. The paired
        initialisation is HF's default (\texttt{repeat\_interleave}) and is not stated in the HRA paper. HF HRA acts on
        \texttt{in\_features}, so injectivity requires \texttt{out\_features}${}\ge{}$\texttt{in\_features} (and full
        column rank); the theorem does not cover \texttt{down\_proj} or GQA key/value projections. The Gram--Schmidt
        variant never contains $W_0$, since $I-2P_U\ne I$, and at HF's default paired initialisation it runs
        Gram--Schmidt on duplicated columns, which is numerically degenerate.
    \end{itemize}
\end{enumerate}
\end{restate}

\begin{proof}
For $u\ne0$ write $\hat u=u/\lVert u\rVert$. The reflection $H_u=I-2\hat u\hat u\T$ is symmetric and orthogonal, has
determinant $-1$, fixes $u^\perp$ pointwise, and $H_u-I$ has rank one. For $g\in\Ort(n)$, $gH_ug^{-1}=H_{gu}$. For
$x,y\in\R^n$ put $x\wedge y=xy\T-yx\T\in\so(n)$, and for a subspace $A\subseteq\R^n$ let
$\R^n\wedge A=\spanop\{x\wedge a:x\in\R^n,\ a\in A\}$. Because $W_0$ is injective, $X\mapsto W_0X$ is an injective linear map
$\R^{n\times n}\to\R^{m\times n}$; in particular $W_0R=W_0R'$ implies $R=R'$.

\emph{Two lemmas.} (i) For every subspace $A$ of dimension $a$,
\begin{equation*}
  \R^n\wedge A=\{\Omega\in\so(n):P_{A^\perp}\Omega P_{A^\perp}=0\},\qquad\dim(\R^n\wedge A)=an-\tfrac{a(a+1)}2 .
\end{equation*}
For $x,y\in A^\perp$ and $z\in\R^n$, $a'\in A$, $y\T(z\wedge a')x=(y\T z)(a'^{\top}x)-(y\T a')(z\T x)=0$, which gives
``$\subseteq$''. Take an orthonormal basis $e_1,\dots,e_a$ of $A$ and $e_{a+1},\dots,e_n$ of $A^\perp$. The matrices
$e_p\wedge e_q$ ($p<q$) form a basis of $\so(n)$, and $P_{A^\perp}\Omega P_{A^\perp}=0$ says exactly that the coefficients
with $p,q>a$ vanish, so the right side is spanned by the $e_p\wedge e_q$ with $p\le a$, which lie in $\R^n\wedge A$. Its
dimension is $\dim\so(n)-\dim\so(n-a)=an-a(a+1)/2$.

(ii) Let $R(u)=H_{u_1}\cdots H_{u_r}$, put $g_k=H_{u_1}\cdots H_{u_{k-1}}$ (so $g_1=I$) and $a_k=g_k\hat u_k$, and let
$A(u)=\spanop\{a_1,\dots,a_r\}$. Then the right-translated differential satisfies
\begin{equation*}
  \{dR_u(\dot u)\,R(u)^{-1}:\dot u\in(\R^n)^r\}=\R^n\wedge A(u).
\end{equation*}
Indeed, varying only $u_k$ in the direction $\dot u_k$ moves $\hat u_k$ by $\delta=(I-\hat u_k\hat u_k\T)\dot u_k/\lVert
u_k\rVert$, which ranges over $\hat u_k^\perp$, and $H_{u_k}$ by $\dot H=-2(\delta\hat u_k\T+\hat u_k\delta\T)$. Since
$H_{u_k}\hat u_k=-\hat u_k$ and $H_{u_k}\delta=\delta$, we get $\dot HH_{u_k}=2(\delta\hat u_k\T-\hat u_k\delta\T)=2\,\delta\wedge
\hat u_k$. Hence
\begin{equation*}
  \dot R\,R^{-1}=g_k\dot H\,H_{u_{k+1}}\cdots H_{u_r}\,R^{-1}=g_k(\dot HH_{u_k})g_k^{-1}=2\,(g_k\delta)\wedge a_k ,
\end{equation*}
using $H_{u_{k+1}}\cdots H_{u_r}R^{-1}=(g_kH_{u_k})^{-1}$ and $g(x\wedge y)g\T=gx\wedge gy$ for orthogonal $g$. As
$\delta$ ranges over $\hat u_k^\perp$, $g_k\delta$ ranges over $a_k^\perp$, and $a_k^\perp\wedge a_k=\R^n\wedge a_k$
because $a_k\wedge a_k=0$. Summing over $k$ gives the claim.

\emph{(a), ``$\subseteq$''.} $R=H_{u_1}\cdots H_{u_r}$ is orthogonal with $\det R=(-1)^r=1$, and it fixes
$\bigcap_iu_i^\perp$, of dimension at least $n-r$, pointwise. So $\rk(R-I)\le r$ and $W_0R-W_0=W_0(R-I)\in\Mr$.

\emph{(a), ``$\supseteq$''.} Let $W_0R=W_0+X$ with $R\in\SO(n)$ and $\rk X\le r$. Since $W_0$ is injective,
$k':=\rk(R-I)=\rk X\le r$. By the Cartan--Dieudonn\'e theorem in Scherk's sharp form (\auxref{B.22}), $R$ is a product of
exactly $k'$ reflections and $\det R=(-1)^{k'}$, so $k'$ is even. Append $(r-k')/2$ pairs $H_vH_v=I$; then $R$ is a product
of exactly $r$ reflections and $W_0R\in\Im\mathrm{HRA}_r$.

\emph{Leaves.} For $V\in\Gr(r,n)$ let $\SO(V)=\{S\in\SO(n):Sx=x\text{ for }x\in V^\perp\}$. Then
\begin{equation*}
  (W_0+\Mr)\cap W_0\cdot\SO(n)=\bigcup_{V\in\Gr(r,n)}W_0\cdot\SO(V).
\end{equation*}
If $S\in\SO(V)$, then $S-I$ vanishes on $V^\perp$, so $\rk(S-I)\le r$. Conversely, if $\rk(R-I)\le r$, the fixed space
$F=\ker(R-I)$ has dimension at least $n-r$, and any $r$-dimensional $V\supseteq F^\perp$ has $V^\perp\subseteq F$, so
$R\in\SO(V)$. A point with $\rk(R-I)=r$ lies on exactly one leaf, $V=\operatorname{im}(R-I)$.

\emph{The $\Ort(n)$-orbit.} $W_0H_{e_1}-W_0=W_0(H_{e_1}-I)$ has rank one, so $W_0H_{e_1}\in W_0+\Mr$. If
$W_0H_{e_1}=W_0R$ with $R=H_{u_1}\cdots H_{u_r}$, injectivity gives $R=H_{e_1}$, whose determinant is $-1\ne(-1)^r$.

\emph{Dimension and fibre.} $\rho=\mathrm{HRA}_r$ is a rational map on the connected open set
$Q=(\R^n\setminus\{0\})^r$, and $d\rho_u(\dot u)=W_0\,dR_u(\dot u)$ has the rank of $dR_u$, which by lemma (ii) equals
$\dim(\R^n\wedge A(u))$. By lemma (i) this is $an-a(a+1)/2$ with $a=\dim A(u)\le r$, a non-decreasing function of $a$ for
$a\le n$. For $u_k=e_k$ we get $g_k=H_{e_1}\cdots H_{e_{k-1}}$ and $a_k=e_k$, so $a=r$ is attained. By \auxref{B.9}(d),
\begin{equation*}
  d(\mathrm{HRA}_r)=\max_u\rk d\rho_u=rn-\tfrac{r(r+1)}2 .
\end{equation*}
The computation uses neither the parity of $r$ nor $r\le n-2$, so the same value holds for every $1\le r\le n$; for
$r\in\{n-1,n\}$ it equals $\dim\SO(n)$. At a point of maximal rank the fibre is locally a submanifold of dimension
$rn-d=r(r+1)/2$ (\auxref{B.1}(b)), and the points of maximal rank, those with $a_1,\dots,a_r$ independent, are open and
dense in $Q$ (\auxref{B.10}). As a check, a generic point of the meet is fixed by its moved space
$\operatorname{im}(R-I)\in\Gr(r,n)$, of dimension $r(n-r)$, and by a fixed-point-free element of $\SO(r)$ on it, an open
subset of dimension $r(r-1)/2$; the sum is $d$.

\emph{Symmetries.} $H_{cu}=H_u$ for $c\ne0$, so the scalings $(\R^\times)^r$ preserve $\rho$. From $gH_bg^{-1}=H_{gb}$
with $g=H_a$ we get $H_aH_b=H_{H_ab}H_a$, so the Hurwitz move $\sigma_i:(u_i,u_{i+1})\mapsto(H_{u_i}u_{i+1},u_i)$ preserves the
product, and $H_{u_i}u_{i+1}\ne0$. It is invertible, with inverse $(u_i,u_{i+1})\mapsto(u_{i+1},H_{u_{i+1}}u_i)$, the moves
$\sigma_i,\sigma_j$ with $\lvert i-j\rvert\ge2$ commute, and $\sigma_1\sigma_2\sigma_1=\sigma_2\sigma_1\sigma_2$: on a triple
both composites give $(a,b,c)\mapsto(H_aH_bc,\,H_ab,\,a)$, using $H_{H_ab}H_ac=H_aH_bc$. So the braid group acts. The
scalings have $r$-dimensional orbits, while the generic fibre has dimension $r(r+1)/2>r$ for $r\ge2$, and the braid moves are
discrete. So the exhibited symmetries are not locally transitive on generic fibres; the remaining fibre directions come from
refactorising $R$ inside its moved space, and this is what we mean by a non-linear gauge.

\emph{(b)} At the paired initialisation, $g_{2i-1}=(H_{w_1}H_{w_1})\cdots(H_{w_{i-1}}H_{w_{i-1}})=I$, so $a_{2i-1}=\hat w_i$
and $a_{2i}=H_{w_i}\hat w_i=-\hat w_i$. Hence $A(u)=U$, $R(u)=I$, and by lemmas (ii) and (i)
\begin{equation*}
  S_1=W_0\cdot(\R^n\wedge U)=W_0\cdot\{\Omega\in\so(n):P_{U^\perp}\Omega P_{U^\perp}=0\},\qquad\dim S_1=kn-\tfrac{k(k+1)}2,
\end{equation*}
since $W_0$ is injective. Equivalently, a direct computation for one pair gives
\begin{equation*}
  \tfrac{\dd}{\dd\varepsilon}\big(H_{w+\varepsilon a}H_{w+\varepsilon b}\big)\big|_{\varepsilon=0}
  =\tfrac2{\lVert w\rVert}\big(v\hat w\T-\hat wv\T\big),\qquad v=(I-\hat w\hat w\T)(a-b),
\end{equation*}
and because every pair equals $I$ at the base point the differential of the product is the
sum of the pair differentials. Finally, with $r=2k$,
\begin{equation*}
  d-\dim S_1=2kn-k(2k+1)-kn+\tfrac{k(k+1)}2=\tfrac{k(2n-3k-1)}2 ,
\end{equation*}
which is positive because $n\ge2k+2$ gives $2n-3k-1\ge k+3$.

\emph{(c)} By \auxref{B.23}(b),(c), the Cayley map $C(\Omega)=(I+\Omega)(I-\Omega)^{-1}$ is a diffeomorphism from $\so(n)$
onto the open set $\mathcal O=\{R\in\SO(n):-1\notin\operatorname{spec}R\}$, which contains $I$, and
$\rk(C(\Omega)-I)=\rk\Omega$. By (a), $\Im\mathrm{HRA}_r\cap W_0\cdot\mathcal O=W_0\cdot C(\Sigma_r)$. The map
$\Omega\mapsto W_0C(\Omega)$ is a diffeomorphism of $\so(n)$ onto an open subset of the embedded submanifold
$W_0\cdot\SO(n)$ (a linear injection of $\SO(n)$), and it carries $\Sigma_r$ onto $\Im\mathrm{HRA}_r$ near $W_0$. So the
germs correspond. $\Sigma_r$ is a closed cone and is not linear (\auxref{B.7}(c)): with $r=2k$, the matrices
$\sum_{j\le k}e_{2j-1}\wedge e_{2j}$ and $e_{r+1}\wedge e_{r+2}$ lie in $\Sigma_r$ (here $r+2\le n$), and their sum has rank
$r+2$. By \auxref{B.6}, $\Sigma_r$ is not a $C^1$ submanifold near $0$. If $\Im\mathrm{HRA}_r$ were a $C^1$ submanifold near
$W_0$, it would be one of $W_0\cdot\SO(n)$, and its preimage $\Sigma_r$ under the diffeomorphism would be $C^1$ near $0$.
So $W_0$ is a singular point, the vertex of the germ. A singular image alone does not give a positive deficit; the apex
statement is (b).

\emph{(d), not a product.} Let $\mathcal F=\{(B,A,R)\in\R^{m\times r}\times\R^{r\times n}\times\SO(n):W_0+BA=W_0R\}$ be the
set-level fibre product. The points of the meet of the form $W_0C(\Omega)$ with $\rk\Omega=r$ form a $d$-dimensional
submanifold $\mathcal G$, the image under the diffeomorphism of (c) of the rank-$r$ part of $\Sigma_r$, which is open and
dense in $\Sigma_r$ and of dimension $d$ (\auxref{B.7}(c)). $\mathcal G$ is open in the meet, since there $\rk(R-I)\ge r$ and
$-1\notin\operatorname{spec}R$ are open conditions, and it is dense: an $R$ in the meet with eigenvalue $-1$ is a limit of
rotations with the same moved planes and the angles $\pi$ replaced by $\pi-\varepsilon$ (\auxref{B.7}(b)). Over a point
$W_0R\in\mathcal G$, the update $W_0(R-I)$ has rank exactly $r$, so near such a point $\mathcal F$ consists of the pairs
$(B,A)$ of full rank with $BA\in W_0(R-I)$ for $W_0R\in\mathcal G$, with $R$ recovered from $BA$ by injectivity. The map
$(B,A)\mapsto BA$ is a submersion from pairs of full rank onto $\MM_r$ with fibres the $\GL_r$-orbits, of dimension $r^2$
(\auxref{B.5}(b)). A submersion is transverse to every submanifold, so by \auxref{B.2}(a), applied with target $\MM_r$,
$\mathcal F$ is there a submanifold of dimension $d+r^2=rn+r(r-1)/2$. If HRA, or any method with an
$rn$-dimensional parameter space, were the product, its underlying set would be in bijection with $\mathcal F$ through its
two smooth projections (\thmref{Observation}{I.4}(b)), so $\mathcal F$ would be the image of an $rn$-manifold $P$ under
a smooth map $\iota$ into the ambient space $E=\R^{m\times r}\times\R^{r\times n}\times\R^{n\times n}$. This is
impossible, because $\mathcal F$ contains an embedded submanifold $\mathcal N$ of dimension $D=rn+r(r-1)/2>rn$ (the part
over $\mathcal G$). Fix $p\in\mathcal N$ and let $\pi:E\to T_p\mathcal N\cong\R^D$ be the orthogonal projection, which is
$1$-Lipschitz. Its restriction to $\mathcal N$ has invertible differential at $p$, so by the inverse function theorem
$\pi(N')$ contains a closed ball $\bar B$ of positive volume for a compact neighbourhood $N'$ of $p$ in $\mathcal N$. Cover
$P$ by countably many compact sets $K_j=\varphi_j([0,1]^{rn})$, images of cubes under charts, on which $\iota\circ\varphi_j$
is Lipschitz with a constant $L_j$. Cutting the cube into $k^{rn}$ subcubes of side $1/k$ covers $\pi(\iota(K_j))$ by
$k^{rn}$ sets of diameter at most $L_j\sqrt{rn}/k$, so its outer Lebesgue measure is at most
$k^{rn}\,\omega_D\,(L_j\sqrt{rn}/k)^D$, which tends to $0$ as $k\to\infty$ because $D>rn$. The countably many null sets
$\pi(\iota(K_j))$ would then cover $\pi(N')\supseteq\bar B$, which is impossible. For the parenthetical remark: over $W_0$ itself $\mathcal F$ contains $\{(B,0,I)\}$, of dimension $mr$,
while over points of $\mathcal G$ its fibre has dimension $r^2$. So the fibres of $\mathcal F$ over the meet jump, and
whenever $mr>d+r^2$ (for instance when $m\ge n+r$, as at $(m,n,r)=(12,9,4)$, where $mr=48>42$) the local dimension of
$\mathcal F$ at these points exceeds $d+r^2$ and $\mathcal F$ is not pure-dimensional.

\emph{(d), exactly orthogonal OFT.} If $C(\Omega)x=-x$, then $y=(I-\Omega)^{-1}x$ satisfies $(I+\Omega)y=-(I-\Omega)y$,
so $2y=0$ and $x=0$; this is the statement that $C$ misses eigenvalue $-1$. Exact full-block Cayley OFT has image
$W_0\cdot C(\so(n))=W_0\cdot\mathcal O$ (input-side forms $W_0C(\Omega)\T=W_0C(-\Omega)$ give the same set). Its
intersection with $W_0+\Mr$ is $W_0\cdot C(\Sigma_r)\subseteq\Im\mathrm{HRA}_r$, which contains $\mathcal G$ and is
therefore dense in $\Im\mathrm{HRA}_r$. The point $W_0H_{e_1}H_{e_2}$ lies in $\Im\mathrm{HRA}_r$ (pad with pairs), and by
injectivity it equals $W_0R$ only for $R=\diag(-1,-1,1,\dots,1)$, which has eigenvalue $-1$; so it is missing. For block
Cayley OFT with block size $b<n$ the image consists of $W_0R$ with $R$ block-diagonal for a partition with at least two
blocks. Choose coordinates $p,q$ in different blocks and $v=(e_p+e_q)/\sqrt2$. Then $R=H_{e_p}H_v\in\SO(n)$ has
$\rk(R-I)\le2\le r$ and $Re_p=H_{e_p}e_q=e_q$, so $R$ is not block-diagonal. Hence $W_0R\in\Im\mathrm{HRA}_r$ lies outside
the image of block OFT, by injectivity.

\emph{(d), HF's default OFT.} Its image consists of the $W_0R(Q)$, $R(Q)=I+2Q+2Q^2+2Q^3+Q^4$, with $Q$ skew and
block-diagonal (the input-side form uses $R(Q)\T=R(-Q)$, of the same kind). By the proof of \thmref{Corollary}{II.6},
$R(Q)\T R(Q)=(I-Q^4)^2$ has eigenvalues $(1-\theta_k^4)^2$, so $R(Q)$ is orthogonal only if every $\theta_k\in\{0,2^{1/4}\}$.
If $\lVert Q\rVert_{\mathrm{op}}<2^{1/4}$ this forces every $\theta_k=0$, hence $Q=0$ since a skew matrix is normal. As
$\Im\mathrm{HRA}_r\subseteq W_0\cdot\Ort(n)$ and $W_0$ is injective, the parameters near $Q=0$ reach $\Im\mathrm{HRA}_r$
only at $W_0$. For $Q=\theta(e_1e_2\T-e_2e_1\T)$ with $\theta\notin\{0,2^{1/4}\}$ (one plane, inside one block),
$R(Q)-I=Q(2I+2Q+2Q^2+Q^3)$ has rank at most $2\le r$ while $R(Q)$ is not orthogonal, so $W_0R(Q)\in W_0+\Mr$ lies outside
$W_0\cdot\Ort(n)\supseteq\Im\mathrm{HRA}_r$. Conversely, an orthogonal $R(Q)$ acts on each plane of $Q$ with
$\theta_k=2^{1/4}$ as $-C(2^{1/4}J)$, a rotation by $\pi+2\arctan2^{1/4}$, and as the identity on $\ker Q$; so a rotation by a
small angle $\varphi\ne0$ in one plane, which lies in $\Im\mathrm{HRA}_r$ by (a), is not of the form $R(Q)$. The two images
are therefore incomparable.

\emph{(d), implementation.} For orthonormal $\hat u_1,\dots,\hat u_r$,
$(I-2\hat u_1\hat u_1\T)(I-2\hat u_2\hat u_2\T)=I-2\hat u_1\hat u_1\T-2\hat u_2\hat u_2\T$ because $\hat u_1\T\hat u_2=0$, and
by induction $\prod_i(I-2\hat u_i\hat u_i\T)=I-2P_U$ with $U=\spanop\{\hat u_i\}$. So the Gram--Schmidt variant has image
$\{W_0(I-2P_U):U\in\Gr(r,n)\}$. The map $U\mapsto W_0(I-2P_U)$ is injective ($P_U$ determines $U$, and $W_0$ is injective) and
is an immersion of the compact manifold $\Gr(r,n)$, so the image has dimension $r(n-r)$. It never contains $W_0$, since
$I-2P_U\ne I$ for $U\ne0$. With duplicated columns, Gram--Schmidt meets a column that lies in the span of the previous
ones, whose residual vanishes, and normalising it divides by zero up to rounding. Injectivity of
$W_0\in\R^{\texttt{out}\times\texttt{in}}$ needs $\texttt{out}\ge\texttt{in}$ and full column rank, which fails for
\texttt{down\_proj} and for GQA key and value projections.
\end{proof}

\subsection{The LOFT example of Exposé II.B}\label{proof:loft}

\begin{restate}{Example}{\unskip}{LOFT: a leaf, and how to choose it; \S\ref{sec:erlangen-loft}}
Let $P_r\in\R^{r\times n}$ have orthonormal rows spanning $V$, and let $\Ort(V)=\{I+P_r\T(T-I)P_r:T\in\Ort(r)\}$ be the
orthogonal maps that fix $V^\perp$ pointwise. LOFT trains $\rho(E)=W_0\big(I+P_r\T(Q(E)-I)P_r\big)$ with $E\in\so(r)$,
$E_0=0$, so $\lvert M\rvert=r(r-1)/2$. We read it with an exactly orthogonal $Q$ (the exponential, or the Cayley map
$C$); the released code defaults to a five-term Cayley--Neumann series with $R\T R=(I-X^6)^2$, which is only
approximately orthogonal, so like HF's default OFT (\thmref{Corollary}{II.6}) that version leaves its leaf.
\begin{enumerate}[label=(\alph*)]
  \item \emph{One leaf.} If $W_0P_r\T$ has rank $r$, then
    $\big(W_0+W_0P_r\T\,\R^{r\times r}P_r\big)\cap W_0\cdot\Ort(n)=W_0\cdot\Ort(V)$. For the principal support
    $P_r=V_r\T$ the first set is the image of LoRA-XS (this needs $\sigma_r(W_0)>0$), so the leaf is
    LoRA-XS${}\cap W_0\cdot\Ort(n)$. Principal-support LOFT reaches an open dense part of its component $W_0\cdot\SO(V)$;
    for injective $W_0$ and a Cayley $Q$, its image is exactly LoRA-XS${}\cap{}$the image of exact full-block Cayley
    OFT.
  \item \emph{Leaves over the Grassmannian.} If $W_0$ is injective,
    $(W_0+\Mr)\cap W_0\cdot\SO(n)=\bigcup_{V\in\Gr(r,n)}W_0\cdot\SO(V)$, which for even $r$ is $\Im\mathrm{HRA}_r$
    (\thmref{Theorem}{II.7}(a)). LOFT's image is the whole leaf $W_0\cdot\SO(V)$ with the exponential, and an open dense
    part of it with a Cayley map, which never reaches eigenvalue $-1$.
  \item \emph{A regular base point.} If $W_0P_r\T$ has rank $r$, then $S_1=W_0P_r\T\so(r)P_r$ has dimension
    $r(r-1)/2=d(M)$: LOFT starts at a smooth point of its leaf and has no gauge, while HRA's paired initialisation is an
    apex (\thmref{Theorem}{II.7}(b)). For $r\ge4$ there is no smooth pointed arrow LOFT${}\to{}$HRA at that
    initialisation, although the images are nested (obstruction test O3 of \S\ref{sec:atlas}).
  \item \emph{Choosing the leaf.} If $Q'(0)=\id$ (the exponential, or $E\mapsto C(E/2)$), the gradient pulled back to $E$
    at $\theta_0$ is $P_r\skewop(W_0\T G)P_r\T$ with $G=\nabla L(W_0)$; the Cayley map $C$ itself doubles it, which does
    not change the best support. Its squared norm is at most $2\sum_{k\le r/2}\mu_k^2$, where $\pm i\mu_k$ are the
    eigenvalues of $\skewop(W_0\T G)$, with equality when $V$ spans the invariant subspace of the top $\lfloor r/2\rfloor$
    pairs.
\end{enumerate}
\end{restate}

\begin{proof}
(a) A rotation $T\in\Ort(r)$ gives $W_0(I+P_r\T(T-I)P_r)=W_0+W_0P_r\T(T-I)P_r$, so ``$\supseteq$'' holds. Conversely let
$W=W_0+BXP_r$ with $B=W_0P_r\T$, and suppose $W=W_0S$ with $S\in\Ort(n)$. Then $WW\T=W_0W_0\T$, and expanding with
$P_rP_r\T=I_r$ gives $B\,(X+X\T+XX\T)\,B\T=0$. Since $B$ has full column rank, $X+X\T+XX\T=0$, that is
$(I+X)(I+X)\T=I$, so $T=I+X\in\Ort(r)$ and $W=W_0(I+P_r\T(T-I)P_r)\in W_0\cdot\Ort(V)$. For $P_r=V_r\T$ we have
$B=U_r\Sigma_r$, and $\{U_r\Sigma_rXV_r\T\}=\{U_rR\,\Sigma_rV_r\T\}$ through $R=\Sigma_rX\Sigma_r^{-1}$, which is
LoRA-XS's image. With the exponential, $T$ ranges over $\SO(r)$ and LOFT reaches the whole component $W_0\cdot\SO(V)$;
with a Cayley map it reaches the $S(T)=I+P_r\T(T-I)P_r$ with $T\in\SO(r)$ and no eigenvalue $-1$, an open dense part of
it. For injective $W_0$, $W_0S$ determines $S$, so meeting with exact full-block Cayley OFT keeps exactly the $S(T)$ with
$T\in\SO(r)$ and no eigenvalue $-1$ (the eigenvalues of $S(T)$ are those of $T$ together with $1$), which is the Cayley
image.

(b) Every $S\in\SO(V)$ has $\rk(S-I)\le r$, which gives ``$\supseteq$''. Conversely, if $W_0S-W_0$ has rank at most $r$
and $W_0$ is injective, then $\rk(S-I)\le r$. The fixed space $F=\ker(S-I)$ has dimension at least $n-r$, and $F^\perp$
is invariant. Any $r$-dimensional $V\supseteq F^\perp$ works: $S$ is the identity on $V^\perp\subseteq F$ and has
determinant $1$ on $V$. The statements about LOFT's image are those of (a).

(c) The differential at $0$ is $D\rho_0(E)=W_0P_r\T Q'(0)(E)P_r$, with $Q'(0)$ invertible. If this vanishes, then $W_0P_r\T Q'(0)(E)=0$ because $P_r$
has full row rank, and $E=0$ because $W_0P_r\T$ has full column rank. So $\dim S_1=\lvert M\rvert\ge d(M)\ge\dim S_1$. A
smooth pointed arrow would give $S_1(\mathrm{LOFT})\subseteq S_1(\mathrm{HRA})$ (\thmref{Observation}{I.6}), and by
\thmref{Theorem}{II.7}(b) and injectivity this asks every $\Omega\in\so(V)$ to satisfy $P_{U^\perp}\Omega P_{U^\perp}=0$.
Since $\dim(V\cap U^\perp)\ge r-r/2\ge2$, any two orthonormal $a,b\in V\cap U^\perp$ give $a\wedge b=ab\T-ba\T$
violating it.

(d) With $Q'(0)=\id$,
\begin{equation*}
  \tfrac{\dd}{\dd t}L\big(\rho(tE)\big)\big|_{t=0}=\langle G,W_0P_r\T EP_r\rangle=\langle P_rW_0\T GP_r\T,E\rangle
  =\langle P_r\skewop(W_0\T G)P_r\T,E\rangle,
\end{equation*}
since $E$ is skew. The squared norm of this gradient is $\lVert P_r\skewop(W_0\T G)P_r\T\rVert^2$, the compression of the
skew matrix $\skewop(W_0\T G)$ to $V$, so the bound and its equality case are \auxref{B.17}(c) with $k=r$.
\end{proof}

\subsection{Proposition II.8}\label{proof:II.8}

\begin{restate}{Proposition}{II.8}{DoRA is LoRA followed by an output scale, at the level of images}
Assume $1\le r\le\min(m,n)$ and $s\ne0$. Let $N$ normalise rows; DoRA's magnitude $\mu\in\R^m$ is per output neuron, as
in HF PEFT and the official code (the paper's notation says ``column-wise'', in the transposed convention). Images refer
to the merged (eval-mode) weight; HF's train-mode forward with \texttt{lora\_dropout}${}>0$ is not of the form $Wx$. Let
$W_0$ have no zero row, let $\mathcal U$ be the open set of matrices with no zero row, and let $\mathrm{Diag}_m$ be the
monoid of all diagonal matrices. Then
\begin{equation*}
  \Im\mathrm{DoRA}_r=\{\diag(\mu)\,N(V):V\in(W_0+\Mr)\cap\mathcal U,\ \mu\in\R^m\}=\mathrm{Diag}_m\cdot(W_0+\Mr),
\end{equation*}
which is exactly the image of ``$\mathrm{LoRA}_r$ followed by a trainable output scale $\ell\in\R^m$''. With $\mu$
restricted to $(\R^\times)^m$, or equivalently after intersecting with $\mathcal U$:
$\Im\mathrm{DoRA}_r\cap\mathcal U=T_m\cdot(W_0+\Mr)\cap\mathcal U$. Moreover:
\begin{itemize}
  \item $d(\mathrm{DoRA}_r)\le\min\big(mn,\ r(m+n-r)+m\big)$. At a parameter point with $\mu\in(\R^\times)^m$ and
    $X=sBA\in\MM_r$, with $U=\operatorname{col}X$ and $R=\operatorname{row}X$, the Jacobian rank is
    $r(m+n-r)+m-\dim\ker\big(\delta\mapsto P_{U^\perp}\diag(\delta)W_0P_{R^\perp}\big)$; where some $\mu_i=0$ it is
    smaller when $r\ge2$, and for $r=1$ when some row $w_{0,i}$ with $\mu_i=0$ lies outside $R$. For generic $W_0$, $d=\min\big(mn,\ r(m+n-r)+m\big)$; in particular $d=r(m+n-r)+m$ iff
    $(m-r)(n-r)\ge m$, which holds for all practical shapes. For special $W_0$ it is smaller: for $W_0=ac\T$ with all
    $a_i\ne0$ and $r<n$, $d=r(m+n-r)+m-r$, strictly below the generic value iff $r<m$ and $r\le n-2$.
  \item The image covariance group (\thmref{Definition}{II.14}) contains $\mathrm{Mon}_m\times\GL_n$.
  \item At $B=0$, in HF's magnitude coordinates, $\Delta W=\big(\diag(\mu_i/\lVert w_{0,i}\rVert)-I\big)W_0$, which can
    have rank up to $\rk W_0$, so DoRA escapes LoRA's rank bound.
\end{itemize}
\end{restate}

\begin{proof}
DoRA is the map $\rho(\mu,B,A)=\diag(\mu)N(V)$ with $V=W_0+sBA$, defined on the open set
$\mathcal D=\{(\mu,B,A):V\in\mathcal U\}$, where $N(V)$ divides row $i$ by $\lVert v_i\rVert$. Since $s\ne0$ and every matrix
of rank at most $r$ factors with inner dimension $r$ (\auxref{B.11}(a)), $\{sBA\}=\Mr$.

\emph{The image.} $\diag(\mu)N(V)=\diag(\mu_i/\lVert v_i\rVert)V\in\mathrm{Diag}_m\cdot(W_0+\Mr)$, which gives the first
inclusion; the first equality is the definition of $\rho$. Conversely, take $\diag(\ell)V$ with $V=W_0+X$, $\rk X\le r$, and
let $Z=\{i:v_i=0\}$. For $0<\varepsilon\le1$ put
\begin{equation*}
  V'=W_0+\Big(I-\varepsilon\sum_{i\in Z}E_{ii}\Big)X .
\end{equation*}
Its update still has rank at most $r$. Its rows outside $Z$ are those of $V$, and for $i\in Z$ we have $x_i=-w_{0,i}$, so
row $i$ of $V'$ is $w_{0,i}-(1-\varepsilon)w_{0,i}=\varepsilon w_{0,i}\ne0$. Hence $V'\in\mathcal U$. With $\ell_i'=0$ on $Z$
and $\ell'_i=\ell_i$ elsewhere, $\diag(\ell')V'=\diag(\ell)V$, since both vanish on the rows in $Z$. Taking
$\mu_i=\ell_i'\lVert v_i'\rVert$ gives $\diag(\mu)N(V')=\diag(\ell)V$. The set $\mathrm{Diag}_m\cdot(W_0+\Mr)$ is by
definition the image of $(\ell,B,A)\mapsto\diag(\ell)(W_0+sBA)$.

\emph{The torus form.} If $\mu\in(\R^\times)^m$, then $\diag(\mu_i/\lVert v_i\rVert)\in T_m$ and the result has no zero
row. Conversely, if $DV\in\mathcal U$ with $D\in T_m$ and $V\in W_0+\Mr$, then $V\in\mathcal U$ and $\mu_i=D_{ii}\lVert
v_i\rVert\ne0$ realises $DV$. Finally, $\diag(\ell)V\in\mathcal U$ forces every $\ell_i\ne0$ and $V\in\mathcal U$, so
$\Im\mathrm{DoRA}_r\cap\mathcal U=T_m\cdot(W_0+\Mr)\cap\mathcal U$, and these are exactly the values at
$\mu\in(\R^\times)^m$.

\emph{A change of coordinates.} Let $\rho'(\ell,B,A)=\diag(\ell)(W_0+sBA)$ be LoRA followed by an output scale. The map
$\varphi(\mu,B,A)=(\mu_i/\lVert v_i\rVert,B,A)$ is a diffeomorphism of $\mathcal D$ onto
$\{(\ell,B,A):W_0+sBA\in\mathcal U\}$, with inverse $(\ell,B,A)\mapsto(\ell_i\lVert v_i\rVert,B,A)$, and
$\rho=\rho'\circ\varphi$. So the Jacobian rank of $\rho$ at $p$ equals that of $\rho'$ at $\varphi(p)$, and $\mu_i=0$ iff
$\ell_i=0$.

\emph{Upper bound.} $\rho'$ factors through $(\ell,B,A)\mapsto(\ell,W_0+sBA)\in\R^m\times(W_0+\Mr)$, whose image has
dimension $m+r(m+n-r)$ (\auxref{B.5}(a); for $r=\min(m,n)$, $\Mr=\R^{m\times n}$ has dimension $mn=r(m+n-r)$). A polynomial
map does not raise dimension (\auxref{B.9}(b)), and trivially $d\le mn$.

\emph{The Jacobian rank.} The differential of $\rho'$ at $(\ell,B,A)$ is
$(\delta,\dot B,\dot A)\mapsto\diag(\delta)V+\diag(\ell)\,s(\dot BA+B\dot A)$. Let $\ell$ be invertible and $X=sBA\in\MM_r$, so
that $B$ and $A$ have full rank and $\{s(\dot BA+B\dot A)\}=T_X\MM_r=\{D:P_{U^\perp}DP_{R^\perp}=0\}$, of dimension
$r(m+n-r)$ (\auxref{B.5}(b)). Substituting $\delta=\diag(\ell)\delta'$, the image of the differential is
$\diag(\ell)\big[\{\diag(\delta')V\}+T_X\MM_r\big]$, whose dimension is that of the bracket. The first summand has dimension
$m$, because $V$ has no zero row. A matrix $\diag(\delta')V$ lies in $T_X\MM_r$ iff
$P_{U^\perp}\diag(\delta')VP_{R^\perp}=0$, and $VP_{R^\perp}=W_0P_{R^\perp}$ because $XP_{R^\perp}=0$. So the intersection
has the dimension of the kernel $K$ of $\delta\mapsto P_{U^\perp}\diag(\delta)W_0P_{R^\perp}$, and the rank is
$r(m+n-r)+m-\dim K$.

\emph{Zero magnitudes.} Let $Z_0=\{i:\ell_i=0\}\ne\emptyset$, keep $X\in\MM_r$, and write $\mathcal S=\{\diag(\delta)V\}$
and $T=T_X\MM_r$. Let $\Pi_0$ keep the rows in $Z_0$ and $\Pi_1$ the others. In the image of the differential, the rows in
$Z_0$ are $\delta_iv_i$ and involve only $\delta_{Z_0}$, while the other rows do not involve $\delta_{Z_0}$; so the image has
dimension $\lvert Z_0\rvert+\dim\Pi_1(\mathcal S+T)$, after the same substitution as above on the rows outside $Z_0$. On the
other hand, the value $F=\dim(\mathcal S+T)$ of the formula satisfies $F=\dim\Pi_1(\mathcal S+T)+\dim\big((\mathcal S+T)\cap
\ker\Pi_1\big)$, and $(\mathcal S+T)\cap\ker\Pi_1$ contains every $\sum_{i\in Z_0}e_i(c_iv_i+y_i)\T$ with $y_i\in R$, since
$e_iv_i\T\in\mathcal S$ and $e_iy\T\in T$ for $y\in R$. As $v_i-w_{0,i}\in R$, the rank at the point is at most
\begin{equation*}
  F-\sum_{i\in Z_0}\big(\dim\spanop(w_{0,i},R)-1\big)\le F-\lvert Z_0\rvert(r-1),
\end{equation*}
which is smaller than $F$ whenever $r\ge2$ or some $w_{0,i}$ with $i\in Z_0$ lies outside $R$, in particular at generic
points with some $\mu_i=0$.

\emph{The generic value.} If $r=\min(m,n)$, the image is $\mathrm{Diag}_m\cdot\R^{m\times n}=\R^{m\times n}$ and $d=mn$,
which is the claimed value. Let $r<\min(m,n)$, and let $E\in\R^{m\times(m-r)}$ and $F\in\R^{n\times(n-r)}$ have orthonormal
columns spanning $U^\perp$ and $R^\perp$. Then $P_{U^\perp}DP_{R^\perp}=0$ iff $E\T DF=0$, and
\begin{equation*}
  E\T\diag(\delta)W_0F=\sum_i\delta_i\,(E\T e_i)(F\T w_{0,i})\T ,
\end{equation*}
so $m-\dim K$ is the dimension $k$ of the span of the $m$ rank-one matrices $t_i=(E\T e_i)(F\T w_{0,i})\T$ in
$\R^{(m-r)\times(n-r)}$, and the rank is $r(m+n-r)+k$ with $k\le\min\big(m,(m-r)(n-r)\big)$. Since
$r(m+n-r)+(m-r)(n-r)=mn$, it suffices to reach $k=\min(m,(m-r)(n-r))$. A witness: choose $E$ whose rows
$p_i=E\T e_i$ are in general position (every $m-r$ of them independent; orthonormalising the columns of a generic matrix keeps
this), let $U=(\operatorname{col}E)^\perp$, $R=\spanop\{e_1,\dots,e_r\}$ and $F=[e_{r+1}\cdots e_n]$. Assign to each $i$ a
column index $c(i)\in\{1,\dots,n-r\}$: if $m\le(m-r)(n-r)$, use each index at most $m-r$ times; otherwise give each index to
exactly $m-r$ of the first $(m-r)(n-r)$ values of $i$, and arbitrary indices to the rest. Put $w_{0,i}=e_{r+c(i)}$, so $F\T w_{0,i}=e_{c(i)}$ and $t_i=p_ie_{c(i)}\T$ has
$p_i$ in column $c(i)$. Matrices supported in different columns are independent, and in one column at most $m-r$ vectors in
general position are independent, while exactly $m-r$ of them form a basis; so the $t_i$ span $\min(m,(m-r)(n-r))$
dimensions. This $W_0$ has no zero row, and with
$B,A$ of full rank spanning $U$ and $R$ and $\mu_i=\lVert v_i\rVert$ the point lies in $\mathcal D$, since $v_i=w_{0,i}+x_i$
with $x_i\in R\perp w_{0,i}$. Now write $P_{U^\perp}=I-B(B\T B)^{-1}B\T$ and $P_{R^\perp}=I-A\T(AA\T)^{-1}A$. After clearing
the denominators $\det(B\T B)$ and $\det(AA\T)$, the $k\times k$ minors of the matrix of $\delta\mapsto
P_{U^\perp}\diag(\delta)W_0P_{R^\perp}$ are polynomials in $(W_0,B,A)$, and one of them is nonzero at the witness. Viewed as a
polynomial in $(B,A)$, its coefficients are polynomials in $W_0$, not all identically zero; for $W_0$ outside their common
zero set it is nonzero on an open dense set of $(B,A)$ (\auxref{B.10}), which meets the open dense set where $B,A$ have full
rank and $W_0+sBA\in\mathcal U$. At such a point with $\ell=\mathbf 1$ the rank is $r(m+n-r)+\min(m,(m-r)(n-r))$. With the
upper bound, $d=\min(mn,r(m+n-r)+m)$ for $W_0$ outside a proper algebraic subset, and $d=r(m+n-r)+m$ iff $(m-r)(n-r)\ge m$.

\emph{Rank-one $W_0$.} Let $W_0=ac\T$ with all $a_i\ne0$ and $r<n$. At a point with $\ell$ invertible and $X\in\MM_r$,
$P_{U^\perp}\diag(\delta)ac\T P_{R^\perp}=(P_{U^\perp}(\delta\odot a))(P_{R^\perp}c)\T$. If $c\notin R$, this vanishes iff
$\delta\odot a\in U$, so $K=\diag(a)^{-1}U$ has dimension $r$ and the rank is $r(m+n-r)+m-r$. If $c\in R$, then $K=\R^m$ and
the rank is $r(m+n-r)$. The points with $\ell$ invertible, $X\in\MM_r$ and $c\notin R$ form an open dense subset of the
domain (since $r<n$), and the points of maximal rank form an open nonempty subset, so the two meet and
$d=r(m+n-r)+m-r$. This is at most $mn$, since $(m-r)(n-r-1)\ge0$. It is strictly below the generic value
$r(m+n-r)+\min(m,(m-r)(n-r))$ iff $m-r<(m-r)(n-r)$, that is iff $r<m$ and $r\le n-2$.

\emph{Covariance.} Let $g=P_\pi D$ be monomial, with $P_\pi$ a permutation matrix and $D\in T_m$, and let $h\in\GL_n$.
Then $gW_0h^{-1}$ has no zero row, and $g\,\diag(x)\,g^{-1}=\diag(P_\pi x)$, so $g\,\mathrm{Diag}_m\,g^{-1}=\mathrm{Diag}_m$;
also $g\Mr h^{-1}=\Mr$. Hence
\begin{equation*}
  g\,\big(\mathrm{Diag}_m\cdot(W_0+\Mr)\big)\,h^{-1}=\mathrm{Diag}_m\cdot(gW_0h^{-1}+\Mr)=\Im\mathrm{DoRA}_r(gW_0h^{-1}).
\end{equation*}
$\mathrm{Mon}_m\times\GL_n$ is a subgroup with this property, so it lies in the covariance group (proof of
\thmref{Definition}{II.14}).

\emph{The update at $B=0$.} HF initialises $\mu_i=\lVert w_{0,i}\rVert$ and then trains $\mu$; at $B=0$ the weight is
$\diag(\mu_i/\lVert w_{0,i}\rVert)W_0$, so $\Delta W=(\diag(\mu_i/\lVert w_{0,i}\rVert)-I)W_0$. With
$\mu_i=2\lVert w_{0,i}\rVert$ for every $i$, $\Delta W=W_0$, of rank $\rk W_0$, which exceeds $2r$ whenever $\rk W_0>2r$,
beyond the bound of \thmref{Theorem}{II.1}(c).
\end{proof}

\subsection{Proposition II.9}\label{proof:II.9}

\begin{restate}{Proposition}{II.9}{the torus ladder}
\begin{enumerate}[label=(\alph*)]
  \item \textbf{HiRA.} $\rho(B,A)=W_0\odot(J+BA)$, where $J$ is the all-ones matrix; HF PEFT has $B_0=0$ and no scale
    factor on Linear and Conv layers (on embeddings it uses $A_0=0$). For every $W_0$,
    $\Im\mathrm{HiRA}_r=W_0+D_{W_0}(\Mr)$ with $D_{W_0}:X\mapsto W_0\odot X$. If $W_0$ has no zero entries,
    $D_{W_0}\in\GL(\Theta)$, so inside the slice at $\theta_0$ HiRA is LoRA transported by a diagonal automorphism: it has
    the same $d=r(m+n-r)$ (for $r\le\min(m,n)$) and the same gauge. In all cases $\rk\Delta W\le r\cdot\rk W_0$. The
    transport depends on $\theta_0$ (\thmref{Proposition}{II.13}).
  \item \textbf{Scalings.} Output-side scalings, $\diag(\ell)W_0=W_0\odot(J+(\ell-\mathbf 1)\mathbf 1\T)$, and
    input-side scalings, $W_0\diag(\ell)=W_0\odot(J+\mathbf 1(\ell-\mathbf 1)\T)$, have images in $\Im\mathrm{HiRA}_1$.
    Output-side examples: (IA)$^3$ on keys and values, the post-linear scale of SSF, DoRA's magnitude. Input-side
    examples: (IA)$^3$'s feed-forward vector, which scales the input of $W_{\mathrm{down}}$, and a LayerNorm gain
    $\gamma$ folded into the next linear map. The fold is exact when the norm's output is consumed only by linear maps:
    the input norms of pre-norm blocks, and a final norm before an untied head (in post-norm BERT, the last block's norm
    feeds only the head). It does not apply to Gemma-style post-sublayer norms, which HF's default LN-tuning targets for
    Gemma~2 and~3, to QK-norms, or to post-norm residual paths. Where it applies, $\gamma$ is tied across all consumers of
    the norm (Q, K and V, or gate and up) rather than being a free point per box, and with tied embeddings a fold into
    the head breaks the tie. LayerNorm biases and SSF's shift fold into the next bias, not into $W$. Two-sided scalings
    $\diag(a)W_0\diag(b)=W_0\odot(ab\T)$ lie in $\Im\mathrm{HiRA}_2$. As a morphism, $h(\ell)=(\ell-\mathbf 1,\mathbf
    1\T)$ lands in $\mathrm{HiRA}_1$ pointed at $(0,\mathbf 1\T)$, which is not HiRA's random-$A$ default pointing.
  \item \textbf{Orbit dimension.} The two-sided torus orbit $\{\diag(a)W_0\diag(b)\}$ has dimension $m+n-c(W_0)$, where
    $c(W_0)$ is the number of connected components of the bipartite support graph of $W_0$, an isolated vertex counting
    as its own component.
  \item \textbf{RoAd.} RoAd$_1$ (the paper's main variant, HF \texttt{road\_1}) is $W=RW_0$ with $R$ block-diagonal in
    scaled rotations $\alpha_i\mathrm{Rot}(\theta_i)$ of output pairs. It is the left action of
    $\CO^+(2)^{m/2}\cong(\mathbb C^\times)^{m/2}$: it is \textbf{(IA)$^{\mathbf 3}$ over $\mathbb C$}. It multiplies the
    input Gram matrix of each output pair by $\alpha_i^2$. Since $\alpha_i$ may be $0$, its image is the orbit closure,
    a real-linear space of dimension exactly $m$ whenever no output-pair block of $W_0$ is zero. RoAd$_2$ and RoAd$_4$
    (HF \texttt{road\_2}, \texttt{road\_4}) let each block range over all of $M_2(\R)$. They realise the block-diagonal
    action of the monoid $M_2(\R)^{m/2}$ (its group part $\GL_2^{m/2}$ is non-abelian), with linear image
    $\{\mathrm{blockdiag}(R_i)W_0\}$ of dimension $2m$ when every pair block has rank $2$. For them the Gram statement
    and ``over $\mathbb C$'' fail, and RoAd$_4$ carries a positive-dimensional gauge (two dimensions per row). HF pairs
    output $j$ with $j+g/2$ inside each group of size $g$ (default $64$) rather than adjacent outputs as in the paper,
    and also rotates the bias; the statements above do not depend on the pairing.
  \item \textbf{Zeros and covariance.} Every $W$ reachable by HiRA, (IA)$^3$ or SSF vanishes wherever $W_0$ does,
    $Z(W_0)\subseteq Z(W)$, and this persists under merging. New zeros can appear ($\ell_i=0$, or $(J+BA)_{ij}=0$), so
    the zero pattern itself is not invariant. The image covariance group of output-scaled boxes ((IA)$^3$ on $k,v$;
    post-linear SSF) is $\mathrm{Mon}_m\times\GL_n$, and that of input-scaled boxes ((IA)$^3$ on $W_{\mathrm{down}}$) is
    $\GL_m\times\mathrm{Mon}_n$. HiRA's is exactly $\mathrm{Mon}_m\times\mathrm{Mon}_n$ ($=\GL_1\times\GL_1$ when
    $m=n=1$).
\end{enumerate}
\end{restate}

\begin{proof}
\emph{(a)} $W_0\odot(J+BA)=W_0+W_0\odot(BA)$, and $\{BA\}=\Mr$, so $\Im\mathrm{HiRA}_r=W_0+D_{W_0}(\Mr)$. In the standard
basis $E_{ij}$ of $\Theta=\R^{m\times n}$, $D_{W_0}$ is diagonal with entries $(W_0)_{ij}$, so it is invertible iff no entry of
$W_0$ vanishes. Then, with $\tau(\theta)=W_0+D_{W_0}(\theta-W_0)$, HiRA is $\tau\circ\rho_{\mathrm{LoRA}}$ for LoRA with residual
$W_0$ and scale $1$. Since $\tau$ is an affine bijection, $D(\tau\circ\rho)=D_{W_0}\circ D\rho$ has the same rank as $D\rho$ at
every point, so $d$ is unchanged (\auxref{B.9}(d)), and $\tau\circ\rho\circ\varphi=\tau\circ\rho$ iff $\rho\circ\varphi=\rho$,
so the gauge group is unchanged; this is the transport $(D_{W_0})_*$ of \thmref{Proposition}{II.13}. For
$r\le\min(m,n)$, $d(\mathrm{LoRA}_r)=r(m+n-r)$ (\auxref{B.5}). For the rank, with columns $b_k$ of $B$ and rows $a_k$ of
$A$, $W_0\odot(BA)=\sum_{k\le r}\diag(b_k)\,W_0\,\diag(a_k)$ (\auxref{B.12}(c)), a sum of $r$ matrices of rank at most
$\rk W_0$. The map $D_{W_0}$ is built from $W_0=\theta_0$, which is the dependence on the base point.

\emph{(b)} Entrywise, $(\diag(\ell)W_0)_{ij}=\ell_i(W_0)_{ij}=(W_0)_{ij}\big(1+(\ell_i-1)\big)$, so
$\diag(\ell)W_0=W_0\odot(J+(\ell-\mathbf 1)\mathbf 1\T)$, and $(\ell-\mathbf 1)\mathbf 1\T=BA$ with $B=\ell-\mathbf 1$ and
$A=\mathbf 1\T$, a point of $\mathrm{HiRA}_1$. The input side is the transpose computation. For two-sided scalings,
$\diag(a)W_0\diag(b)=W_0\odot(ab\T)$ and $ab\T-J=[a,\,-\mathbf 1]\,[b,\,\mathbf 1]\T$ has rank at most $2$. The map
$h(\ell)=(\ell-\mathbf 1,\mathbf 1\T)$ is smooth, sends $\mathbf 1$ to $(0,\mathbf 1\T)$ and satisfies
$\rho_{\mathrm{HiRA}_1}\circ h(\ell)=\diag(\ell)W_0$, so it is a pointed morphism into $\mathrm{HiRA}_1$ pointed there.
For a norm whose output $\gamma\odot\hat x$ (with $\hat x$ the normalised input) feeds only linear maps $W$,
$W(\gamma\odot\hat x)=(W\diag(\gamma))\hat x$, so the gain is an input-side scaling of each consumer, the same $\gamma$ for
all of them; a bias $\beta$ gives $W\beta$, a change of the next bias. When the norm's output is added to the residual
stream (post-sublayer norms, post-norm residual paths) or is used inside attention scores (QK-norms), no linear map of the
adapted box absorbs $\gamma$. With tied embeddings the head is $E\T$, and replacing it by $E\T\diag(\gamma)$ unties it.

\emph{(c)} The orbit map $\Psi(a,b)=\diag(a)W_0\diag(b)$ of $T_m\times T_n$ has differential at $(\mathbf 1,\mathbf 1)$
\begin{equation*}
  (\alpha,\beta)\mapsto\diag(\alpha)W_0+W_0\diag(\beta)=\big((\alpha_i+\beta_j)(W_0)_{ij}\big)_{ij}.
\end{equation*}
Its kernel is cut out by $\alpha_i+\beta_j=0$ along every edge $(i,j)$ of the bipartite support graph. On a connected
component the value at one vertex determines all the others (with $\beta=-\alpha$ across each edge), and an isolated
vertex carries one free coordinate; so the kernel has dimension $c(W_0)$ and the rank is $m+n-c(W_0)$. Because
$\Psi(g g')=g\cdot\Psi(g')$ for the commutative group $T_m\times T_n$, the differential has this rank at every point of the
group, so the orbit, an immersed submanifold of dimension $\dim(T_m\times T_n)$ minus that of the stabiliser, has
dimension $m+n-c(W_0)$.

\emph{(d)} Identify an output pair, a $2\times n$ block $X$ with rows $x_1,x_2$, with the complex row $z=x_1+\mathrm ix_2\in
\mathbb C^n$. For $\mathrm{Rot}(\theta)=\left(\begin{smallmatrix}\cos\theta&-\sin\theta\\ \sin\theta&\cos\theta\end{smallmatrix}
\right)$ the rows of $\alpha\mathrm{Rot}(\theta)X$ are $\alpha(\cos\theta\,x_1-\sin\theta\,x_2)$ and
$\alpha(\sin\theta\,x_1+\cos\theta\,x_2)$, which correspond to $\alpha e^{\mathrm i\theta}z$. So RoAd$_1$ multiplies each pair
by a complex scalar, the left action of $\CO^+(2)^{m/2}\cong(\mathbb C^\times)^{m/2}$, the complex analogue of (IA)$^3$'s
real output scaling. The input Gram matrix of the pair becomes
$(\alpha\mathrm{Rot}X)\T(\alpha\mathrm{Rot}X)=\alpha^2X\T X$. As $(\alpha,\theta)$ ranges over $\R\times\R$, $\alpha e^{\mathrm i\theta}$
ranges over all of $\mathbb C$, so the image is $\prod_i\mathbb C\cdot z_i$, the closure of the orbit, a real-linear space of
dimension $2\,\#\{i:z_i\ne0\}$, equal to $m$ when no pair block vanishes. In RoAd$_2$ each row of a block has its own
polar pair, so each row of the $2\times2$ block is an arbitrary vector of $\R^2$ and the block is an arbitrary element of
$M_2(\R)$; RoAd$_4$ also reaches every element of $M_2(\R)$, with eight parameters for four entries, so its fibres have
dimension $4$ per block, two per row. The image $\{R_iX_i\}$ of one block is the set of $2\times n$ matrices whose rows lie
in $\operatorname{row}X_i$, of dimension $2\rk X_i$, so the total is $2m$ when every block has rank $2$. For $R=\diag(1,2)$
and $X=I_2$, $(RX)\T(RX)=\diag(1,4)$ is not a multiple of $X\T X$, and $R$ does not commute with $\mathrm{Rot}(\pi/2)$, so it
is not multiplication by a complex number. The pairing of outputs and the rotated bias do not enter these computations.

\emph{(e) Zeros.} $W_{ij}=(W_0)_{ij}(1+(BA)_{ij})$ for HiRA, $\ell_i(W_0)_{ij}$ or $(W_0)_{ij}\ell_j$ for (IA)$^3$, and
$\gamma_i(W_0)_{ij}$ for SSF's post-linear scale (its shift changes the bias), so $(W_0)_{ij}=0$ implies $W_{ij}=0$. After a
merge the new base $W_1$ has $Z(W_0)\subseteq Z(W_1)$, and induction gives the claim for any number of cycles. New zeros:
$\ell_i=0$ kills row $i$, and $B=-e_i$, $A=e_j\T$ give $(J+BA)_{ij}=0$.

\emph{(e) Output and input scalings.} Here $\Im M(W_0)=\mathrm{Diag}_m\cdot W_0$. For $g$ monomial and $h\in\GL_n$,
$g\,\diag(\ell)\,W_0h^{-1}=\diag(\ell')\,gW_0h^{-1}$ with $\diag(\ell')=g\diag(\ell)g^{-1}$, and $\ell'$ ranges over $\R^m$
with $\ell$, so $\mathrm{Mon}_m\times\GL_n$ is contained in the covariance group. Conversely, let $(g,h)$ be in it and take
$W_0=e_jc\T$ with $c\ne0$. Then $\Im M(e_jc\T)=\R\,e_jc\T$, and $g\,\Im M(W_0)h^{-1}=\R\,(ge_j)(h^{-\top}c)\T$ is a line,
while $\Im M(gW_0h^{-1})=\{y(h^{-\top}c)\T:\operatorname{supp}y\subseteq\operatorname{supp}(ge_j)\}$ has dimension
$\lvert\operatorname{supp}(ge_j)\rvert$. Equality for every $j$ forces every column of $g$ to have one nonzero entry, so $g$ is
monomial. The input side is the transpose.

\emph{(e) HiRA.} For permutation matrices $P,P'$ and diagonal $D,D'$ we have $P(X\odot Y)P'=(PXP')\odot(PYP')$ and
$D(X\odot Y)D'=(DXD')\odot Y$. For $g=PD$ and $h^{-1}=D'P'$ this gives
\begin{equation*}
  g\big(W_0\odot(J+Z)\big)h^{-1}=(gW_0h^{-1})\odot(J+PZP'),
\end{equation*}
and $P\Mr P'=\Mr$, so $\mathrm{Mon}_m\times\mathrm{Mon}_n$ is contained in the covariance group. Conversely, let $(g,h)$ be in
it and take $W_0=E_{ij}$, whose image $E_{ij}\odot(J+\Mr)=\R E_{ij}$ is a line. With $u=ge_i$ and $v=h^{-\top}e_j$,
covariance requires $\Im M(uv\T)=\R\,uv\T$. Suppose $uv\T$ has two nonzero entries $(k,l)\ne(k',l')$. Taking $Z=E_{kl}\in\MM_{\le1}$
gives $Y=uv\T+u_kv_lE_{kl}\in\Im M(uv\T)$. If $Y=\lambda uv\T$, the entry $(k',l')$ forces $\lambda=1$ and the entry $(k,l)$
forces $u_kv_l=0$, a contradiction. So $\lvert\operatorname{supp}u\rvert=\lvert\operatorname{supp}v\rvert=1$ for all $i,j$,
which makes $g$ and $h^{-\top}$, hence $h$, monomial. For $m=n=1$, $\mathrm{Mon}_1=\GL_1$.
\end{proof}

\subsection{Theorem II.10}\label{proof:II.10}

\begin{restate}{Theorem}{II.10}{cut-rank bound; standard}
Let $\Gamma$ be a tensor network, evaluated in the symmetric monoidal category $(\FinVect,\otimes)$, with input legs $V$
and output legs $U$. Every wire and open leg is assumed incident to a vertex: insert an identity box on any bare strand,
or equivalently count each bare input--output wire as crossing every cut. A crossing hyperedge (copy spider) contributes
its dimension once. Partition the vertices into $\mathcal I\sqcup\mathcal O$. The weight $w$ of the cut is the product
of the dimensions of the internal wires between the two sides and of the open legs attached on the wrong side (input legs
at $\mathcal O$, output legs at $\mathcal I$). Then
\begin{equation*}
  \rk\llbracket\Gamma\rrbracket\le\min_{\mathrm{cuts}}w,
\end{equation*}
and rank is subadditive over formal sums. The bound refers to a chosen presentation of the network.
\end{restate}

\begin{proof}
Fix a cut $\mathcal I\sqcup\mathcal O$. Write $V_{\mathcal I},V_{\mathcal O}$ for the tensor products of the input legs
attached to $\mathcal I$ and to $\mathcal O$, $U_{\mathcal I},U_{\mathcal O}$ likewise for the output legs, and $C$ for the
tensor product of the spaces carried by the crossing wires, each crossing hyperedge counted once. Contracting every wire
internal to $\mathcal I$ produces a tensor whose free indices are those of $V_{\mathcal I}$, $U_{\mathcal I}$ and $C$; read
it as a linear map $T_{\mathcal I}:V_{\mathcal I}\to U_{\mathcal I}\otimes C$. Likewise the $\mathcal O$ side gives
$T_{\mathcal O}:C\otimes V_{\mathcal O}\to U_{\mathcal O}$. A crossing hyperedge is a single summation index shared by
tensors on both sides, so it is summed once, over a basis of its space, when the two sides are joined. Contracting the
crossing indices joins the two halves, and up to the symmetry isomorphisms that reorder tensor factors,
\begin{equation*}
  \llbracket\Gamma\rrbracket=(\id_{U_{\mathcal I}}\otimes T_{\mathcal O})\circ(T_{\mathcal I}\otimes\id_{V_{\mathcal O}}):
  V_{\mathcal I}\otimes V_{\mathcal O}\to U_{\mathcal I}\otimes C\otimes V_{\mathcal O}\to U_{\mathcal I}\otimes U_{\mathcal O}.
\end{equation*}
A composite factors through its middle space, so its rank is at most
$\dim U_{\mathcal I}\cdot\dim C\cdot\dim V_{\mathcal O}=w$. Only the symmetric monoidal structure was used: no cups or caps.
A bare strand from an input to an output, carried by an identity box, has one of its two legs on the wrong side of every
cut, which is the convention of the statement; without it, $I_d\otimes BA$ would violate the bound. Minimising over cuts
gives the bound, and subadditivity over formal sums is $\rk(X+Y)\le\rk X+\rk Y$ (\auxref{B.12}(a)). The weight depends on
the drawing, which is why the bound refers to a presentation; compare \auxref{B.14}.
\end{proof}

\noindent\emph{The bounds of Table~\ref{tab:cutbounds}.} Each follows from a cut in the stated presentation or from
subadditivity. The LoRA family $sBA$ factors through the bond of dimension $r$. A split initialisation gives
$\Delta W=sBA-sB_0A_0$, a sum of two such terms (\thmref{Theorem}{II.1}(c)). For HRA, the telescoping identity
\begin{equation*}
  R-I=\sum_{k=1}^r\big(H_k\cdots H_r-H_{k+1}\cdots H_r\big)=\sum_{k=1}^r(H_k-I)H_{k+1}\cdots H_r
\end{equation*}
writes $R-I$ as $r$ terms of rank at most $\rk(H_k-I)=1$, so $\rk(W_0R-W_0)\le r$; a Givens rotation differs from $I$ in a
$2\times2$ block, so the same identity gives $2g$ for $g$ gates. ReLoRA after $K$ phases is a sum of $K$ rank-$r$ terms, and
GraLoRA factors through $\R^{kr}$ (proof of \thmref{Proposition}{II.12}). LoHa's Hadamard product, drawn with copy spiders on
the two legs, factors through the face-splitting bond of dimension $r_1r_2$ (\auxref{B.12}(c)). HiRA's update
$W_0\odot BA=\sum_{k\le r}\diag(b_k)W_0\diag(a_k)$ factors through $r$ copies of a bond of dimension $\rk W_0$. LoKr gives
$\rk(C\otimes BA)=\rk C\cdot\rk(BA)\le\rk C\cdot r'$ (\auxref{B.12}(b)). FourierFT's update is a sum of $n_c$ real cosine
patterns $\cos\big(2\pi(uj/m+vl/n)\big)=\operatorname{Re}(x_uy_v\T)$ with $(x_u)_j=e^{2\pi\mathrm iuj/m}$ and
$(y_v)_l=e^{2\pi\mathrm ivl/n}$ (\auxref{B.16}(c)), and $\operatorname{Re}(xy\T)=\operatorname{Re}x\operatorname{Re}y\T-
\operatorname{Im}x\operatorname{Im}y\T$ has rank at most $2$. MoRA's reshape, C3A's circulant blocks and MPO adapters have
no narrow cut in their presentation, so the theorem bounds them only by $\min(m,n)$.

\subsection{Theorem II.11}\label{proof:II.11}

\begin{restate}{Theorem}{II.11}{Kronecker incomparability; the linear algebra is standard, the PEFT reading is new}
Write $\R^m=\R^{m_1}\otimes\R^{m_2}$ and $\R^n=\R^{n_1}\otimes\R^{n_2}$, and view $\Delta\in\R^{m\times n}$ as a 4-leg
tensor. Let $\operatorname{rk}$ be the rank across the cut $\{U_1U_2\}\mid\{V_1V_2\}$ and $\operatorname{rk}_\otimes$
the rank across $\{U_1V_1\}\mid\{U_2V_2\}$. Put
\begin{equation*}
  \kappa_1:=\min(m_1,m_2)\cdot\min(n_1,n_2),\qquad\nu:=\min(m_1,n_1)\cdot\min(m_2,n_2).
\end{equation*}

(a) $\operatorname{rk}_\otimes\Delta\le k$ iff $\Delta=\sum_{i\le k}C_i\otimes D_i$. Every $\Delta$ has
$\operatorname{rk}_\otimes\Delta\le\min(m_1n_1,m_2n_2)$.

(b) $\operatorname{rk}(C\otimes D)=\operatorname{rk}C\cdot\operatorname{rk}D$, while
$\operatorname{rk}_\otimes(C\otimes D)=1$ for $C,D\ne0$. So $\nu$ is the largest rank of a single Kronecker product.

(c) If $u\in\R^m$ and $v\in\R^n$ have Schmidt ranks $\rho_u,\rho_v$, then
$\operatorname{rk}_\otimes(uv\T)=\rho_u\rho_v$. A generic rank-one update therefore has Kronecker rank $\kappa_1$. For
aligned splits, $(m_1-m_2)(n_1-n_2)\ge0$, which include equal square splits and every split HF PEFT's factorisation
produces (it puts the smaller factor first on both sides), $\kappa_1=\min(m_1n_1,m_2n_2)$ is the maximal Kronecker rank.

(d) \emph{Incomparability.} For $1\le r<\nu$ and $1\le k<\kappa_1$, the sets $\Mr$ and
$\mathcal K_{\le k}:=\{\Delta:\operatorname{rk}_\otimes\Delta\le k\}$ of unconstrained $k$-term Kronecker sums are
incomparable, for every split. A generic rank-one update lies in the first and not the second, and $C\otimes D$ with
full-rank factors, of Kronecker rank $1$ and rank $\nu$, lies in the second and not the first. For equal square splits
($m=n=N$, all factors $s$), $\kappa_1=\nu=s^2=N$, and $I_N$ is such a product.

(e) \emph{Constrained Kronecker methods.} Let $\Im M\subseteq W_0+\mathcal K_{\le k}$. Then
$\Im\mathrm{LoRA}_r\not\subseteq\Im M$ whenever $k<\kappa_1$, for every pointing of $\mathrm{LoRA}_r$. Give
$\mathrm{LoRA}_r$ its zero-init pointing, with image $W_0+\Mr$. Then $\Im M\not\subseteq\Im\mathrm{LoRA}_r$ iff some
$\theta\in\Im M$ has $\rk(\theta-W_0)>r$. For LoKr, $\Delta W=C\otimes BA$ with $C\in\R^{m_1\times n_1}$,
$B\in\R^{m_2\times r'}$ and $A\in\R^{r'\times n_2}$, put $r'_{\mathrm{eff}}=\min(r',m_2,n_2)$. The maximum rank is
$\min(m_1,n_1)\cdot r'_{\mathrm{eff}}$, attained. So $\Im\mathrm{LoKr}_{r'}\subseteq\Im\mathrm{LoRA}_r$ iff
$r\ge\min(m_1,n_1)\,r'_{\mathrm{eff}}$, and otherwise the two images are incomparable, provided $\kappa_1>1$. In HF PEFT
the second LoKr factor is low-rank only when its rank parameter satisfies $r'<\max(m_2,n_2)/2$; otherwise it is full,
and LoKr is unconstrained KronA, with maximum rank $\nu=\min(m_1,n_1)\min(m_2,n_2)$. On a $768$-wide layer
($24\otimes32$) the maximum rank therefore jumps from $360$ at $r'=15$ to $768$ at $r'=16$. With
\texttt{decompose\_both}, $C$ is also capped at rank $r'$.
\end{restate}

\begin{proof}
Index $\R^m$ by pairs $(i_1,i_2)$ and $\R^n$ by $(j_1,j_2)$. The rank across $\{U_1U_2\}\mid\{V_1V_2\}$ is the ordinary
rank of $\Delta$. The rank across $\{U_1V_1\}\mid\{U_2V_2\}$ is the rank of the rearrangement
$\mathcal R(\Delta)\in\R^{m_1n_1\times m_2n_2}$ with entries $\mathcal R(\Delta)_{(i_1j_1),(i_2j_2)}=\Delta_{(i_1i_2),(j_1j_2)}$,
a linear bijection that permutes entries, with $\mathcal R(C\otimes D)=\vect C\,\vect D\T$ (\auxref{B.13}(a)). (In this
proof and the next, $\mathcal R$ denotes this rearrangement.)

\emph{(a)} $\rk\mathcal R(\Delta)\le k$ iff $\mathcal R(\Delta)=\sum_{i\le k}x_iy_i\T$, iff
$\Delta=\sum_{i\le k}\mathcal R^{-1}(x_iy_i\T)=\sum_{i\le k}C_i\otimes D_i$ with $\vect C_i=x_i$ and $\vect D_i=y_i$. The
matrix $\mathcal R(\Delta)$ has size $m_1n_1\times m_2n_2$, which gives the second claim (\auxref{B.13}(b)).

\emph{(b)} $\rk(C\otimes D)=\rk C\cdot\rk D$ (\auxref{B.12}(b)), and $\mathcal R(C\otimes D)=\vect C\vect D\T$ has rank $1$
for $C,D\ne0$. The largest value of $\rk C\cdot\rk D$ is $\min(m_1,n_1)\min(m_2,n_2)=\nu$, attained by factors of full rank.

\emph{(c)} Take Schmidt decompositions $u=\sum_{i\le\rho_u}\sigma_ix_i\otimes y_i$ and
$v=\sum_{j\le\rho_v}\tau_jz_j\otimes w_j$ with positive coefficients and orthonormal families (\auxref{B.11}(a)). By the
mixed-product rule,
\begin{equation*}
  uv\T=\sum_{i,j}\sigma_i\tau_j\,(x_iz_j\T)\otimes(y_iw_j\T).
\end{equation*}
The two families $(x_iz_j\T)_{ij}$ and $(y_iw_j\T)_{ij}$ are Frobenius-orthonormal, because
$\langle x_iz_j\T,\,x_{i'}z_{j'}\T\rangle=(x_i\T x_{i'})\,(z_j\T z_{j'})$ and similarly for the second family, and the
$\rho_u\rho_v$ coefficients are positive. By
\auxref{B.13}(c) this is an operator-Schmidt decomposition, so $\rk_\otimes(uv\T)=\rho_u\rho_v$. The Schmidt rank of $u$ is the
rank of its $m_1\times m_2$ reshape, which equals $\min(m_1,m_2)$ outside the zero set of the maximal minors, a proper
algebraic subset (\auxref{B.10}(b)); likewise for $v$. So for generic $u,v$, $\rk_\otimes(uv\T)=\min(m_1,m_2)\min(n_1,n_2)=
\kappa_1$. If $(m_1-m_2)(n_1-n_2)\ge0$, then either $m_1\le m_2$ and $n_1\le n_2$, so $\kappa_1=m_1n_1=\min(m_1n_1,m_2n_2)$,
or $m_1\ge m_2$ and $n_1\ge n_2$, so $\kappa_1=m_2n_2=\min(m_1n_1,m_2n_2)$; in both cases $\kappa_1$ is the bound of (a).
Equal square splits are aligned, and so is any split with the smaller factor first on both sides.

\emph{(d)} A generic rank-one $uv\T$ lies in $\MM_{\le1}\subseteq\Mr$ and has $\rk_\otimes=\kappa_1>k$ by (c), so it is not in
$\mathcal K_{\le k}$. A product $C\otimes D$ with factors of full rank has $\rk_\otimes=1\le k$ and rank $\nu>r$ by (b). Neither
argument uses alignment. For equal square splits $\kappa_1=\nu=s\cdot s=N$ and $I_N=I_s\otimes I_s$.

\emph{(e)} Any pointing of $\mathrm{LoRA}_r$ has image $W_{\mathrm{res}}+\Mr$ for its frozen residual $W_{\mathrm{res}}$.
Suppose $W_{\mathrm{res}}+\Mr\subseteq W_0+\mathcal K_{\le k}$, and put $c=W_{\mathrm{res}}-W_0$, so $c+\Mr\subseteq\mathcal
K_{\le k}$. The set $\mathcal K_{\le k}$ is closed (the zero set of the $(k+1)\times(k+1)$ minors of $\mathcal R(\cdot)$) and
closed under scaling. For $X\in\Mr$ and $t>0$, $c+tX\in\mathcal K_{\le k}$, hence $c/t+X\in\mathcal K_{\le k}$, and letting
$t\to\infty$ gives $X\in\mathcal K_{\le k}$. So $\Mr\subseteq\mathcal K_{\le k}$, which contradicts (d) when $k<\kappa_1$.
The second clause restates $\Im\mathrm{LoRA}_r=W_0+\Mr$ for the zero-init pointing.

For LoKr, $\rk(C\otimes BA)=\rk C\cdot\rk(BA)\le\min(m_1,n_1)\cdot\min(r',m_2,n_2)$ (\auxref{B.12}(a),(b)), with equality
for $C$ of full rank and $BA$ of rank $r'_{\mathrm{eff}}$, which exists by \auxref{B.11}(a). If
$r\ge\min(m_1,n_1)r'_{\mathrm{eff}}$, every update has rank at most $r$ and $\Im\mathrm{LoKr}_{r'}\subseteq W_0+\Mr$.
Otherwise the maximising update has rank above $r$, so $\Im\mathrm{LoKr}_{r'}\not\subseteq\Im\mathrm{LoRA}_r$; and since
$\Im\mathrm{LoKr}_{r'}\subseteq W_0+\mathcal K_{\le1}$, the first clause with $k=1<\kappa_1$ gives
$\Im\mathrm{LoRA}_r\not\subseteq\Im\mathrm{LoKr}_{r'}$. When HF's second factor is full, LoKr is $C\otimes D$ with both
factors free, whose maximum rank is $\nu$ by (b). On a $768$-wide layer HF factors $768=24\cdot32$ on both sides, so
$m_1=n_1=24$ and $m_2=n_2=32$: for $r'=15<16$ the maximum rank is $24\cdot15=360$, and for $r'=16$ the second factor is full and
the maximum is $24\cdot32=768$. The \texttt{decompose\_both} option factors $C$ through rank $r'$, which caps $\rk C$ at $r'$.
The thresholds and factorisations are read from the HF PEFT source \citep{xmangrulkar2022peft}.
\end{proof}

\subsection{Proposition II.12}\label{proof:II.12}

\begin{restate}{Proposition}{II.12}{gauge census: rank is a cut, dimension is a volume}
\begin{enumerate}[label=(\alph*)]
  \item $d(M)\le\lvert M\rvert-\dim(\text{generic gauge orbit})$, with equality when the gauge acts locally
    transitively on generic fibres.
  \item The values in Table~\ref{tab:census} hold for one $m\times n$ box, with $r\le\min(m,n)$ throughout except for
    VeRA (whose published RoBERTa-base setting has $r=1024>768$). In the LoHa row,
    \begin{equation*}
      F=r_1(m+n-r_1)+r_2(m+n-r_2)-(m+n-1)
    \end{equation*}
    is a proved upper bound for $d$. The gauge column lists the exhibited Lie group. It acts locally transitively on
    generic fibres away from the $mn$ cap, but not where the cap binds; for LoHa at $(m,n,r_1,r_2)=(4,4,2,2)$ the orbit
    has dimension $15$ and the fibre $16$. BitFit does not touch the $m\times n$ box; it is linear in the biases. Hybrid
    GraLoRA (\texttt{hybrid\_r}${}>0$) is not covered by the table. FourierFT is injective only when its sampled
    frequencies contain no conjugate pair $(u,v),(-u,-v)$; otherwise $d=\lvert M\rvert-\#\{\text{pairs}\}$. At LLM sizes
    this is close to $\lvert M\rvert$, but at $768\times768$ with $1000$ frequencies a pair occurs in about half of the
    draws (HF's default seed $777$ gives exactly two pairs there).
  \item \emph{Equal-budget trades.} Assume $2r\le\min(m,n)$, $r^2\le\min(m,n)$ and $2r^2<m+n-1$. LoHa$_{r,r}$ and
    LoRA$_{2r}$ both cost $2r(m+n)$, but LoHa reaches at least $(m+n-1)-2r^2$ \emph{fewer} dimensions (exactly that
    many when $F$ is attained [Num]). For $r\ge3$ it gains rank up to $r^2>2r$ in exchange. For $r\le2$, $r^2\le2r$, so
    $\Im\mathrm{LoHa}_{r,r}\subsetneq\Im\mathrm{LoRA}_{2r}$ and LoHa is strictly dominated. When $2r^2\ge m+n-1$ (still with $2r\le\min(m,n)$) the
    bound no longer separates them, and wherever $F$ is attained LoHa$_{r,r}$ matches or exceeds LoRA$_{2r}$ in
    dimension as well as in rank, for instance at $(m,n,r)=(9,9,3)$, at $r=64$ on $4096\times4096$, and at
    LyCORIS-style $r=32$ on $320$- to $640$-wide layers. GraLoRA$_k$ costs the same as LoRA$_r$, reaches the same
    number of dimensions, and has maximum rank $\min(kr,m,n)$, but caps every block at rank $r/k$. A generic rank-$r$
    update has rank $r$ on every block, so for $k\ge2$ and $kr\le\min(m,n)$ the images of LoRA$_r$ and GraLoRA$_k$ are
    incomparable.
\end{enumerate}
\end{restate}

\begin{proof}
\emph{(a)} By \auxref{B.9}(d), $d(M)$ is the maximal rank of $d\rho$, attained on an open set, which is dense when $\rho$ is
real-analytic on a connected domain (\auxref{B.10}). At a point $q$ of maximal rank the fibre through $q$ is, near $q$, a
submanifold of dimension $\lvert M\rvert-d(M)$ (\auxref{B.1}(b)), and it contains the gauge orbit of $q$, an immersed
submanifold. At generic $q$ this gives $\dim(\text{orbit})\le\lvert M\rvert-d(M)$. If the gauge acts locally transitively on
generic fibres, the orbit is open in the fibre and the two dimensions agree.

\emph{(b), LoRA.} $\GL_r$ acts on pairs of full rank by $(Bg^{-1},gA)$, freely, and its orbits are the fibres of
$(B,A)\mapsto BA$ over $\MM_r$ (\auxref{B.5}(b)), so $d=r(m+n)-r^2=r(m+n-r)$. The maximum rank is $r$.

\emph{GraLoRA.} With $k\mid m,n,r$, the update is a $k\times k$ grid of blocks of size $p\times q=\tfrac mk\times\tfrac nk$, each
an independent LoRA of rank $t=r/k$, so the image is $W_0$ plus the product of $k^2$ copies of
$\MM_{\le t}(p\times q)$, of dimension $k^2\,t(p+q-t)=r(m+n-r)$ since $t\le\min(p,q)$. The gauge $\prod\GL_t$ acts freely on
generic parameters with orbits of total dimension $k^2t^2=r^2$, equal to the fibre dimension. For the rank, write block
$(i,j)$ as $U_{ij}V_{ij}\T$; then $\Delta W=\operatorname{blockdiag}(L_1,\dots,L_k)\,S$ with
$L_i=[U_{i1}\cdots U_{ik}]\in\R^{p\times r}$ and $S\in\R^{kr\times n}$ stacking the $V_{ij}\T$ placed in column block $j$, so
$\rk\Delta W\le\min(kr,m,n)$. (Placing rank-$t$ blocks only on the diagonal would give rank $r$, not $kr$.) To see that
$\min(kr,m,n)$ is attained, assume $p\le q$ (otherwise transpose) and put $a=\min(kt,p)$, so that
$\min(kr,m,n)=ka$. Write $a=k\alpha+\beta$ with $0\le\beta<k$, and let $N_{ij}=\alpha+[\,(j-i)\bmod k<\beta\,]$. Then every
row and column sum of $N$ is $a\le p\le q$, and $N_{ij}\le t$, because $\alpha\le t$ with $\alpha=t$ only when $\beta=0$. In
block row $i$ pick $N_{ij}$ distinct rows for each $j$ (in all $a\le p$ rows), in block column $j$ pick $N_{ij}$ distinct
columns for each $i$ (in all $a\le q$ columns), and put a $1$ at the matched pairs. The resulting $0/1$ matrix has at most
one $1$ in each row and column, so its rank is $\sum_{ij}N_{ij}=ka$, and its block $(i,j)$ has rank $N_{ij}\le t$, so it lies
in the image. Rank at least $ka$ is the nonvanishing of a minor, a polynomial in the block factors, so it holds for generic
parameters (\auxref{B.10}).

\emph{LoHa.} The update is $(B_1A_1)\odot(B_2A_2)$ with budget $(r_1+r_2)(m+n)$. The torus acts by
$(B_1,A_1,B_2,A_2)\mapsto(D_1B_1,A_1D_2,D_1^{-1}B_2,A_2D_2^{-1})$ and preserves the update, because
$(D_1XD_2)\odot(D_1^{-1}YD_2^{-1})=X\odot Y$ for diagonal $D_1,D_2$; together with $\GL_{r_1}\times\GL_{r_2}$ acting on the two
LoRA factors this gives an action of $G=\GL_{r_1}\times\GL_{r_2}\times T_m\times T_n$. We compute the stabiliser at generic
parameters. If $m=n=r_1$, then $r_1(m+n-r_1)=mn$ and $F\ge mn$ (since $r_2(m+n-r_2)-(m+n-1)=(r_2-1)(m+n-r_2-1)\ge0$), so the
bound $d\le\min(mn,F)$ is trivial; otherwise, by symmetry under transposition, assume $m\ge r_1+1$. Let
$(g_1,g_2,D_1,D_2)$ fix the point. From $D_1B_1g_1^{-1}=B_1$, every row of $B_1$ is a left eigenvector of $g_1$; for generic
$B_1$ any $r_1$ of its rows form a basis and the remaining rows have all coordinates nonzero in that basis, which forces a
single eigenvalue, so $g_1=cI$ and $D_1=cI$. From $g_1A_1D_2=A_1$ and nonzero columns of $A_1$, $D_2=c^{-1}I$. From
$D_1^{-1}B_2g_2^{-1}=B_2$ and $B_2$ of full column rank ($m\ge r_2$), $g_2=c^{-1}I$, and then the last equation holds. So the
stabiliser is $\{(cI,c^{-1}I,cI,c^{-1}I)\}$, of dimension $1$, and the generic orbit has dimension
$r_1^2+r_2^2+m+n-1$. By (a), $d\le(r_1+r_2)(m+n)-r_1^2-r_2^2-(m+n-1)=F$, and trivially $d\le mn$. Where $d=F$ the orbit
has the dimension of the fibre and the action is locally transitive; at $(4,4,2,2)$ the orbit has dimension $4+4+7=15$ while
the [Num] value $d=16$ gives a fibre of dimension $32-16=16$. For the rank, $\rk(X\odot Y)\le r_1r_2$ (\auxref{B.12}(c)) and
$\le\min(m,n)$. To attain $k=\min(r_1r_2,m,n)$, choose distinct pairs $(a_i,b_i)\in[r_1]\times[r_2]$ for $i\le k$, let the
first $k$ rows of $B_1$ be $e_{a_i}\T$ and the first $k$ columns of $A_1$ be $e_{a_i}$, likewise $B_2,A_2$ with the $b_i$, and
set the rest to zero. Then $(B_1A_1)_{ij}=[a_i=a_j]$ and $(B_2A_2)_{ij}=[b_i=b_j]$ for $i,j\le k$, so the Hadamard product is
$I_k$ in the top-left corner and zero elsewhere, of rank $k$; by \auxref{B.10} this rank is generic.

\emph{HRA.} The proof of \thmref{Theorem}{II.7}(a) computes $d=rn-r(r+1)/2$ and the fibre dimension $r(r+1)/2$ for every
$1\le r\le n$, with no use of parity or of $r\le n-2$; for $r\in\{n-1,n\}$ the value is $\dim\SO(n)$. The scalings
$(\R^\times)^r$ are the exhibited Lie gauge, and $W_0H_{u_1}\cdots H_{u_r}-W_0$ has rank at most $r$. Evenness of $r$ is used
only by HF's paired initialisation, which needs pairs to start at $W_0$.

\emph{DoRA.} The value of $d$ is \thmref{Proposition}{II.8}. $\GL_r$ acts on $(B,A)$ by $(Bg^{-1},gA)$, fixing $\mu$, freely on
pairs of full rank and preserving $\rho$. For generic $W_0$ with $(m-r)(n-r)\ge m$ the fibre has dimension
$r(m+n)+m-r(m+n-r)-m=r^2$, so $\GL_r$ is locally transitive on generic fibres. Otherwise $d=mn$ and the fibre has dimension
$r(m+n)+m-mn=r^2+m-(m-r)(n-r)>r^2$, which is $6$ at $(3,3,2)$; for rank-one $W_0$, $d=r(m+n-r)+m-r$ and the fibre has dimension
$r^2+r$. The update can have rank up to $\min(m,n)$ (\thmref{Proposition}{II.8}, last item).

\emph{VeRA.} The update is $\diag(b)\,B\diag(d)\,A$ with frozen random $B\in\R^{m\times r}$, $A\in\R^{r\times n}$ and trainable
$b\in\R^m$, $d\in\R^r$, so $\lvert M\rvert=m+r$, and $(b,d)\mapsto(cb,c^{-1}d)$ preserves it. The value $d(M)=m+r-1$ is
checked numerically [Num]. The rank is at most $\rk(B\diag(d)A)\le\min(r,m,n)$, and for invertible $\diag(b)$, $\diag(d)$ it
equals $\rk(BA')$ with $A'=\diag(d)A$, which is $\min(r,m,n)$ for generic frozen matrices (\auxref{B.10}). None of this uses
$r\le\min(m,n)$.

\emph{KronA.} The update $A\otimes B$ satisfies $\mathcal R(A\otimes B)=\vect A\vect B\T$, so the image is $W_0$ plus the
preimage under the linear bijection $\mathcal R$ (\auxref{B.13}(a)) of the Segre cone $\MM_{\le1}(m_1n_1\times m_2n_2)$, of
dimension $m_1n_1+m_2n_2-1$ (\auxref{B.8}). If $A\otimes B=A'\otimes B'\ne0$, then $\vect A\vect B\T=\vect A'\vect B'^{\top}$,
so $A'=cA$ and $B'=c^{-1}B$; the fibre over a nonzero update is the orbit of $\R^\times$. The maximum rank is $\nu$ by
\thmref{Theorem}{II.11}(b).

\emph{Tensor-train and Tucker adapters.} The update is a fixed reshape, a linear bijection, of a tensor in TT or Tucker
format, so $d$ is the dimension of the set of tensors of the given admissible minimal ranks, which is the parameter count
minus $\sum_er_e^2$, with the bond gauge $\prod_e\GL_{r_e}$ acting on the cores (\auxref{B.15}).

\emph{Linear methods.} For $\rho(q)=\theta_0+Lq$ with $L$ linear, $d=\rk L$, the fibres are the cosets $q+\ker L$, and
translation by $\ker L$ acts transitively on each. LoRA-FA, LoRA-XS, FourierFT, C3A and masks are of this form; so is MiSS
in each HF mode, including the BAT mode, whose update $\operatorname{blocks}(W_0)M+M$ is linear in the trained block $M$
(\thmref{Proposition}{II.13}).

\emph{FourierFT.} HF's update is $c\operatorname{Re}(\operatorname{ifft2}(S))$ with $S$ supported on a sampled set $\Omega$
of $\lvert M\rvert$ frequencies. By \auxref{B.16}(c), frequency $p=(u,v)$ contributes $s_p\kappa_p$ with
$\kappa_p(j,l)=\tfrac1{mn}\cos\big(2\pi(uj/m+vl/n)\big)$, and $\kappa_p=\kappa_{-p}$. Write $\chi_p(j,l)=e^{2\pi\mathrm
i(uj/m+vl/n)}$, so that $\langle\chi_p,\chi_{p'}\rangle=mn\,[p=p']$ and $\cos(\cdots)=\tfrac12(\chi_p+\chi_{-p})$. Hence
$\langle\kappa_p,\kappa_{p'}\rangle=0$ unless $p'=\pm p$, and every $\kappa_p\ne0$. So the patterns of distinct classes
$\{p,-p\}$ are linearly independent, and the rank of $s\mapsto\sum_ps_p\kappa_p$ is the number of classes met by $\Omega$, which
is $\lvert\Omega\rvert$ minus the number of conjugate pairs inside $\Omega$. Each pair $\{p,-p\}\subseteq\Omega$ contributes
the kernel direction $s_p=1$, $s_{-p}=-1$. The frequencies of pairs in random draws and for HF's default seed are numerical
observations.

\emph{(c)} Assume $2r\le\min(m,n)$, $r^2\le\min(m,n)$ and $2r^2<m+n-1$. Both methods cost $2r(m+n)$, and
$d(\mathrm{LoRA}_{2r})=2r(m+n-2r)$. By (b), $d(\mathrm{LoHa}_{r,r})\le F=2r(m+n-r)-(m+n-1)$, and
\begin{equation*}
  d(\mathrm{LoRA}_{2r})-F=(m+n-1)-2r^2>0 .
\end{equation*}
So LoHa reaches at least $(m+n-1)-2r^2$ fewer dimensions; since $F<d(\mathrm{LoRA}_{2r})\le mn$, the cap does not bind, and
the gap is exactly $(m+n-1)-2r^2$ when $d(\mathrm{LoHa}_{r,r})=F$. Its maximum rank is $\min(r^2,m,n)=r^2$ by (b), against
$2r$ for $\mathrm{LoRA}_{2r}$, and $r^2>2r$ iff $r\ge3$. For $r\le2$, every LoHa update has rank at most $r^2\le2r$, so
$\Im\mathrm{LoHa}_{r,r}\subseteq W_0+\MM_{\le2r}=\Im\mathrm{LoRA}_{2r}$ (both pointed at $W_0$, as HF zero-initialises one
LoHa factor), and the inclusion is strict because the dimensions differ. When $2r^2\ge m+n-1$, $F\ge d(\mathrm{LoRA}_{2r})$ and
the bound does not separate them; wherever $d(\mathrm{LoHa}_{r,r})=F$, LoHa reaches at least as many dimensions, and its
maximum rank $\min(r^2,m,n)$ is at least $\min(2r,m,n)=2r$, because $r\ge2$ here (for $r=1$, $2\ge m+n-1$ contradicts
$m,n\ge2$). The examples satisfy $2r^2\ge m+n-1$: $18\ge17$ at $(9,9,3)$, $8192\ge8191$ at $r=64$ on $4096\times4096$, and
$2048\ge m+n-1$ for $r=32$ and $m,n\le640$.

GraLoRA$_k$ has budget $k^2\cdot t(p+q)=r(m+n)$, dimension $r(m+n-r)$ and maximum rank $\min(kr,m,n)$ by (b). Let $k\ge2$ and
$kr\le\min(m,n)$. A rank-$r$ update $BA$ has blocks $B_iA_j$, with $B_i$ the rows of $B$ in block row $i$ and $A_j$ the
columns of $A$ in block column $j$; for generic $B,A$ every $B_i$ ($p\times r$, $r\le p$) and $A_j$ ($r\times q$, $r\le q$) has
rank $r$, so every block has rank $r>r/k$, and the update is not in GraLoRA's image. Conversely GraLoRA reaches rank $kr>r$,
outside $W_0+\Mr$. Both images contain $W_0$, so they are incomparable.
\end{proof}

\subsection{Proposition II.13}\label{proof:II.13}

\begin{restate}{Proposition}{II.13}{transport of structure}
For $T\in\GL(\Theta)$ put $\tau_T(\theta)=\theta_0+T(\theta-\theta_0)$ and $T_*M:=(Q,q_0,\tau_T\circ\rho)$.
\begin{itemize}
  \item $T_*$ is a strict automorphism of $\Meth(\Theta,\theta_0)$, with inverse $(T^{-1})_*$. It maps $S_1$ to $T\,S_1$
    and preserves $d$, $g$, the gauge group, the first-order deficit, polynomial degree, the regular or apex type and
    the germ type of the image. It does not preserve rank, and, for the family extension below, it does not preserve the
    covariance group.
  \item $K^{T_*M}=T\,K^M\,T\T$ for every $T$. For any update rule that queries the loss only through
    $q\mapsto L(\rho(q))$, with its own state on $Q$, $T_*M$ trained on $L$ and $M$ trained on $L\circ\tau_T$ have
    identical $q$-trajectories, so their $\theta$-trajectories are conjugate by $\tau_T$. Rules that read
    $\nabla_\theta L$ and solve a least-squares problem on $\Theta$ (LoRA-Pro), and $\theta$-level weight decay or
    clipping, are covered only for orthogonal $T$. Orthogonality of $T$ is exactly the condition for $\tau_T$ to be a
    Frobenius isometry; for rules of the first kind it is needed only to keep the Frobenius spectrum of $K$, or an
    isometry-invariant loss class such as $\lVert\theta-\theta^\star\rVert^2$, unchanged.
  \item Coordinatewise optimizers (SGD, Adam) are also equivariant under coordinate permutations of $Q$. The Kronecker
    identification below needs this, because it composes $\mathcal R^{-1}_*$ with the vec reshape of $Q$.
  \item \emph{Families and rebasing.} For $T$ independent of the base point, define the family extension
    $(T_*M)(\theta):=(Q,q_0,\theta+T(\rho_\theta-\theta))$. For an additive family with fixed shape $C$
    (\thmref{Definition}{I.2}), $T_*M$ is additive with shape $TC$, so transport commutes with rebasing:
    $R_K(T_*M)=\tau_T(R_K(M))$ (\thmref{Theorem}{V.3}(a)). For action-type families this can fail even for $T=2I$,
    because the transported family acts by $I+T(g-I)$. For OFT on $\Theta=\R^{1\times2}$, two cycles of
    $\theta\mapsto\theta(2R-I)$ with $R=\operatorname{rot}(\pi)$ reach $9\theta_0$, outside the transported circle.
    Transports that depend on $\theta_0$ may fail (HiRA below) or may not (LoRA transported by the SVD frame of
    $\theta_0$, since $\Mr$ is $\Ort(m)\times\Ort(n)$-invariant). The covariance group (\thmref{Definition}{II.14}) is a
    property of a family and is compared through this extension.
\end{itemize}
Instances.
\begin{itemize}
  \item \emph{Kronecker family.} An isomorphism of slices over $\Theta'=\R^{m_1n_1\times m_2n_2}$ and over $\Theta$:
    $\sum_{i\le k}C_i\otimes D_i=\mathcal R^{-1}_*\mathrm{LoRA}_k$, with $Q$ reshaped by vec, dynamics included for
    coordinatewise optimizers. KronA with unconstrained factors is $\mathcal R^{-1}_*\mathrm{LoRA}_1$. LoKr is
    $\mathcal R^{-1}_*$ of $\mathrm{LoRA}_1$ with the second factor constrained to $\vect\MM_{\le r'}$ when
    $r'<\max(m_2,n_2)/2$ (HF's switch); otherwise HF's second factor is full and LoKr is KronA. LoKr's
    \texttt{decompose\_both} option also constrains the first factor.
  \item \emph{Linear methods as coordinate subspaces in transported frames.} FourierFT: sparse in the real cosine half of
    the real Fourier frame. With scaling $c$ (default $150$), HF's update $\Delta W=c\operatorname{Re}(\operatorname{ifft2}(S))$
    is the transport of a mask, with $g=0$, only when the sampled index set has no conjugate pair $(u,v),(-u,-v)$;
    otherwise $g$ is the number of such pairs. Its image lies in the point-symmetric subspace
    $\Delta W_{j,l}=\Delta W_{-j,-l}$, and the frame is orthogonal only up to scale (self-conjugate frequencies have twice
    the squared norm). WaveFT: sparse in a wavelet frame, when HF's pad--IDWT--crop map is invertible (plausible for Haar
    on even shapes; unverified for other wavelet families). C3A: block-circulant; over $\R$ the frame is given by the
    powers $P^k$ of the cyclic shift (Gram matrix $bI$); over $\mathbb C$, $\operatorname{circ}(w)=F^{-1}\diag(Fw)F$ for
    first column $w$, and the real form has $2\times2$ rotation--scaling blocks; HF adds a factor
    $1/\texttt{in\_features}$. LoRA-XS, SVFT, TinyLoRA: masks or random subspaces in the SVD frame of $W_0$, a
    $\theta_0$-dependent transport. SHiRA, FISH, Trainable Tokens: masks in the standard frame. MiSS/Bone in its default
    (balance) mode: $\Delta W=D(\mathbf 1_g\T\otimes I_r)$, which is LoRA-FA with a frozen \emph{summing} projection; its
    BAT mode, $\Delta W=\operatorname{blocks}(W_0)\,M+M$ blockwise, is linear in the trained block $M$ but depends on
    $\theta_0$, so it is not additive. These frame readings of FourierFT and C3A are for \texttt{init\_weights=True}; HF's
    defaults start off $\theta_0$ (FourierFT draws its spectrum from $\mathcal N(0,1)$ with scaling $150$, so at $768^2$
    with $1000$ frequencies $\lVert\Delta W_0\rVert_F\approx4.3$, about $14\%$ of a std-$0.04$ $W_0$; C3A is
    Xavier-uniform), and these are objects with a pointing defect.
  \item \emph{HiRA} $=(D_{W_0})_*\mathrm{LoRA}$ inside the slice at $\theta_0$, when $W_0$ has no zero entries, with
    $D_{W_0}:X\mapsto W_0\odot X$. Because $D_{W_0}$ depends on $\theta_0$, re-based HiRA is not transported ReLoRA. Two
    cycles reach $W_0\odot(J+Z_1)\odot(J+Z_2)$, whose relative update $Z_1+Z_2+Z_1\odot Z_2$ has rank up to
    $\min(2r+r^2,m,n)$, attained, where transported ReLoRA stays at rank $\le2r$.
\end{itemize}
\end{restate}

\begin{proof}
\emph{The automorphism.} $\tau_T$ is an affine bijection fixing $\theta_0$ with linear part $T$, so $T_*M$ is pointed. If
$h:M\to N$ satisfies $\rho_N\circ h=\rho_M$, then $\tau_T\circ\rho_N\circ h=\tau_T\circ\rho_M$, so $h$ is also an arrow
$T_*M\to T_*N$; let $T_*$ act as the identity on arrows. This is a functor, and $(T^{-1})_*T_*=\id$ on the nose because
$\tau_{T^{-1}}\circ\tau_T=\id$. So $T_*$ is a strict automorphism.

\emph{Invariants.} $D(\tau_T\circ\rho)_q=T\,D\rho_q$ for every $q$. So $S_1(T_*M)=T\,S_1(M)$, and the rank of the differential
is unchanged at every point; hence $d$ (\auxref{B.9}(d)), $\dim S_1$, the deficit, $g=\lvert M\rvert-d$, and the regular or
apex type are unchanged. Since $\tau_T$ is injective, $\tau_T\circ\rho\circ\varphi=\tau_T\circ\rho$ iff $\rho\circ\varphi=\rho$,
so the gauge groups coincide. Composition with an affine map does not raise polynomial degree, and $\tau_{T^{-1}}$ undoes
it, so the degree is unchanged. Finally $\tau_T$ is a diffeomorphism of $\Theta$ fixing $\theta_0$ that carries $\Im M$ onto
$\Im T_*M$, so the germs at $\theta_0$ are diffeomorphic. Rank is not preserved: any $T$ with $T(E_{11})=I_n$ ($n\ge2$, $m=n$)
sends a rank-one update to a full-rank one.

\emph{Covariance is not preserved.} Take the family extension of LoRA$_1$ by $T=D_V:X\mapsto V\odot X$, with $V$ fixed,
without zero entries and of rank at least $2$; its image at $W$ is $W+V\odot\MM_{\le1}$. If $(g,h)$ were in its covariance
group for every pair, comparing the images at $gWh^{-1}$ and $W$ would give $V\odot\MM_{\le1}=g\,(V\odot\MM_{\le1})\,h^{-1}$, so
$V\odot\MM_{\le1}$ would be a union of $\GL_m\times\GL_n$-orbits, that is of rank strata $\MM_k$. It has dimension $m+n-1$,
since $D_V$ is invertible, while $\dim\MM_k=k(m+n-k)>m+n-1$ for $2\le k\le\min(m,n)$. So it would lie in $\MM_{\le1}$, and in
particular $V\odot J=V$ would have rank at most one, a contradiction. Hence the covariance group of the transported family is
a proper subgroup of $\GL_m\times\GL_n$, the covariance group of LoRA.

\emph{Cometric and dynamics.} $K^{T_*M}=D(\tau_T\rho)\Lambda D(\tau_T\rho)\T=T\,D\rho\Lambda D\rho\T\,T\T=TK^MT\T$. The method $T_*M$
trained on $L$ minimises $L\circ\tau_T\circ\rho$ on $Q$, and $M$ trained on $L\circ\tau_T$ minimises $(L\circ\tau_T)\circ\rho$,
the same function. A rule that sees the loss only through this function, from the same initial state (and, for stochastic
rules, the same random stream), therefore produces the same $q$-trajectory, and the weights are $\tau_T(\rho(q(t)))$ against
$\rho(q(t))$. A rule that reads $\nabla_\theta L$ and takes the Frobenius least-squares step on $\Theta$ moves $q$ by
$-(D\rho\T T\T TD\rho)^+D\rho\T T\T\nabla L$ under $T_*M$, and by $-(D\rho\T D\rho)^+D\rho\T T\T\nabla L$ under $M$ with loss
$L\circ\tau_T$, since $\nabla_\theta(L\circ\tau_T)=T\T\nabla L$. They agree when $T\T T=I$ and differ in general: for
$\Theta=\R^2$, $\rho(q)=qe_1$ and $T=\diag(2,1)$ the steps are $-\tfrac12\partial_1L$ and $-2\partial_1L$. Weight decay and
clipping on $\theta$ use the Frobenius norm, which $\tau_T$ preserves iff $T$ is orthogonal. For orthogonal $T$,
$TKT\T$ is similar to $K$, so the Frobenius spectrum of the cometric is kept, and $\lVert\tau_T(\theta)-\tau_T(\theta^\star)
\rVert=\lVert\theta-\theta^\star\rVert$.

\emph{Permutations of $Q$.} A coordinatewise rule updates each coordinate from that coordinate's own gradient history with
shared hyperparameters. If $q(t)$ is its trajectory for an objective $f$ from $q_0$, then $\pi q(t)$ is its trajectory for
$f\circ\pi^{-1}$ from $\pi q_0$, for any coordinate permutation $\pi$.

\emph{Families and rebasing.} For an additive family, $\rho_\theta=\theta+\delta$ with $\delta$ independent of $\theta$ and
shape $C=\delta(Q)$, the extension is $\theta+T\delta$, additive with shape $TC$. After $K$ merge-and-restart cycles an
additive family reaches $R_K=\theta_0+C+\dots+C$ ($K$ summands), and since $T$ is linear,
$\theta_0+(TC)+\dots+(TC)=\theta_0+T(C+\dots+C)=\tau_T(R_K)$. For OFT on $\R^{1\times2}$ the family is
$\{\theta R:R\in\SO(2)\}$ and its extension by $T=2I$ is $\theta\mapsto\theta+2(\theta R-\theta)=\theta(2R-I)$. With
$R=\operatorname{rot}(\pi)=-I$, two cycles give $\theta_0(-3I)(-3I)=9\theta_0$. Every rebasing of OFT stays in
$\theta_0\cdot\SO(2)$, whose transport $\{\theta_0(2S-I):S\in\SO(2)\}$ consists of vectors of norm at most $3\lVert\theta_0\rVert$,
so $9\theta_0$ lies outside it when $\theta_0\ne0$. For LoRA transported by $X\mapsto U(\theta)XV(\theta)\T$ with orthogonal
$U(\theta),V(\theta)$, the transported shape is $U\Mr V\T=\Mr$, so the transported family equals LoRA's, and both sides of
the rebasing identity are $\theta_0+\MM_{\le\min(Kr,m,n)}$ (by the splitting argument of \thmref{Theorem}{II.1}(c)).

\emph{Kronecker family.} By \auxref{B.13}(a), $\mathcal R:\Theta\to\Theta'$ is a linear bijection with
\begin{equation*}
  \mathcal R\Big(\sum_{i\le k}C_i\otimes D_i\Big)=\sum_i\vect C_i\,\vect D_i\T=BA,
\end{equation*}
where $B$ has columns $\vect C_1,\dots,\vect C_k$ and $A$ has rows $\vect D_1\T,\dots,\vect D_k\T$. So the $k$-term Kronecker method is $\mathcal R^{-1}\circ\rho_{\mathrm{LoRA}_k}$ up to the
vec relabelling of $Q$, and the argument for $T_*$ applies verbatim to an isomorphism between two weight spaces. Its
dynamics agree with LoRA$_k$'s on the pulled-back loss for rules that query the loss through $\rho$, and the vec relabelling is a
coordinate permutation, harmless for coordinatewise rules. KronA is $k=1$. For LoKr,
$\mathcal R(C\otimes BA)=\vect C\vect(BA)\T$, LoRA$_1$ with second factor in $\vect\MM_{\le r'}$, when HF factors the second
factor; otherwise it is KronA. With \texttt{decompose\_both} the first factor is factored too.

\emph{FourierFT.} By the computation in the proof of \thmref{Proposition}{II.12}, the patterns $\kappa_p$ of distinct
classes $\{p,-p\}$ are orthogonal; with the sine patterns $\sin(2\pi(uj/m+vl/n))$ of the classes with $p\ne-p$ they form an
orthogonal basis of $\R^{m\times n}$. Let $T$ send the standard basis to this basis (cosines on the positions of the
coefficients, sines elsewhere). If $\Omega$ has no conjugate pair, FourierFT is $T_*$ of the mask on $\Omega$, with $g=0$;
otherwise the two coefficients of a pair hit the same pattern and $g$ is the number of pairs. Each $\kappa_p$ is even under
$(j,l)\mapsto(-j,-l)$ (indices modulo $m,n$), so the image lies in the point-symmetric subspace. A self-conjugate
frequency has $\kappa_p=\tfrac1{mn}\chi_p$ with $\chi_p$ real, of squared norm $\tfrac1{mn}$, while for $p\ne-p$ the squared
norm is $\tfrac1{2mn}$. HF's default draws $1000$ coefficients from $\mathcal N(0,1)$ and multiplies by $150$; for generic
frequencies $\mathbb E\lVert\Delta W_0\rVert_F^2=150^2\cdot1000/(2\cdot768^2)\approx19.1$, a root-mean-square norm of
$4.37$. A single draw deviates from it by about $2\%$, since the sum of $1000$ squared standard normals has relative
standard deviation $\sqrt{2/1000}$, which gives the value $\lVert\Delta W_0\rVert_F\approx4.3$ quoted. Against
$\lVert W_0\rVert_F\approx0.04\cdot768\approx30.7$ this is about $14\%$.

\emph{Other frames.} If HF's pad--IDWT--crop map $P$ is invertible, WaveFT is $P_*$ of a mask by definition. For C3A, the
powers $P^0,\dots,P^{b-1}$ of the cyclic shift span the $b\times b$ circulants and have Gram matrix $bI$, and
$\operatorname{circ}(w)=F^{-1}\diag(Fw)F$; the real form follows from conjugate eigenvalues (\auxref{B.16}(a),(b)). So a
block-circulant update is a mask in the frame $\{E^{(\mathrm{blk})}_{ij}\otimes P^k\}$. LoRA-XS's update
$U_rR\Sigma_rV_r\T=\sum_{i,j\le r}R_{ij}\sigma_ju_iv_j\T$ is a scaled mask in the orthonormal frame $(u_iv_j\T)$ built from the SVD
of $W_0$; SVFT's update $U M V\T$ with sparse $M$ is a mask there too, and TinyLoRA's is a random subspace of it. SHiRA,
FISH and Trainable Tokens update a fixed set of entries. In MiSS's balance mode HF computes the update on an input $x$ as
$D$ times the sum of the $g$ chunks of $x$ of size $r$, that is $D(\mathbf 1_g\T\otimes I_r)x$; this is LoRA-FA with the frozen
projection $\mathbf 1_g\T\otimes I_r$. Its BAT mode is linear in $M$ with coefficients from $W_0$, so it is linear but not additive.

\emph{HiRA.} By \thmref{Proposition}{II.9}(a), HiRA is $(D_{W_0})_*\mathrm{LoRA}$ when $W_0$ has no zero entries. A merge
gives $W_1=W_0\odot(J+Z_1)$, and the next cycle $W_1\odot(J+Z_2)=W_0\odot(J+Z_1)\odot(J+Z_2)=W_0\odot(J+Z_1+Z_2+Z_1\odot Z_2)$.
With $Z_k=B_kA_k$,
\begin{equation*}
  Z_1+Z_2+Z_1\odot Z_2=\big[B_1,\ B_2,\ B_1\bullet B_2\big]\,\big[A_1;\ A_2;\ (A_1\T\bullet A_2\T)\T\big]
\end{equation*}
(\auxref{B.12}(c)), which factors through $\R^p$ with $p=2r+r^2$. Call the two factors $L\in\R^{m\times p}$ and
$L'\in\R^{p\times n}$. The rows of $L$ and the columns of $L'$ are points $(x,y,x\otimes y)$ of a variety $\mathcal V$ that
spans $\R^r\oplus\R^r\oplus\R^{r^2}$: a linear form vanishing on $\mathcal V$ vanishes on $(x,0,0)$, on $(0,y,0)$ and then on
every $(0,0,x\otimes y)$. Choosing points of $\mathcal V$ one at a time outside the span of the previous ones, $m$ rows can be
chosen to span $\min(m,p)$ dimensions, and likewise $n$ columns. If $p\le\min(m,n)$, then $L$ has full column rank and $L'$
full row rank, so $\rk LL'=p$. If $p>\min(m,n)$, say $m\le n$, choose $L$ of rank $m$, that is $L:\R^p\to\R^m$ onto; then
$L(\mathcal V)$ spans $\R^m$, so $n\ge m$ columns of $L'$ can be chosen with images spanning $\R^m$, and $\rk LL'=m$; the case
$n<m$ is the transpose. In every case rank $\min(2r+r^2,m,n)$ is attained, hence generic (\auxref{B.10}). Transported ReLoRA instead reaches $D_{W_0}(Z_1+Z_2)$, whose preimage $Z_1+Z_2$ has rank at most $2r$.
\end{proof}

\subsection{Definition II.14}\label{proof:II.14}

\begin{restate}{Definition}{II.14}{covariance group}
For a method defined for every $W_0$ (a natural family), the \textbf{covariance group}
$\mathcal G_M\subseteq\GL_m\times\GL_n$ is the largest subgroup with $\Im M(gW_0h^{-1})=g\,\Im M(W_0)\,h^{-1}$. When
the method depends on random frames or on data, the identity is required in distribution, with the data transformed.
$\mathcal G_M$ is the \emph{geometry the method lives in}, in Klein's sense.
\end{restate}

The definition contains one claim: a largest such subgroup exists.

\begin{proof}
Let $\mathcal S$ be the set of $(g,h)\in\GL_m\times\GL_n$ with $\Im M(gWh^{-1})=g\,\Im M(W)\,h^{-1}$ for every $W$ in the
domain of the family (which we assume stable under $\GL_m\times\GL_n$, or at least under the pairs considered). The pair
$(I,I)$ lies in $\mathcal S$. If $(g_1,h_1),(g_2,h_2)\in\mathcal S$, then for every $W$
\begin{equation*}
  \Im M(g_1g_2Wh_2^{-1}h_1^{-1})=g_1\,\Im M(g_2Wh_2^{-1})\,h_1^{-1}=g_1g_2\,\Im M(W)\,h_2^{-1}h_1^{-1}.
\end{equation*}
If $(g,h)\in\mathcal S$, apply its identity at $g^{-1}Wh$: $\Im M(W)=g\,\Im M(g^{-1}Wh)\,h^{-1}$, so
$\Im M(g^{-1}Wh)=g^{-1}\Im M(W)h$. So $\mathcal S$ is a subgroup, every subgroup with the property is contained in it, and
$\mathcal G_M=\mathcal S$. For random frames or data, read each identity as equality of the laws of the random sets, the
pushforwards of the frame and data laws under $F\mapsto\Im M(W;F)$, with the data transformed by the action of $(g,h)$
(for instance $x\mapsto hx$, so $C\mapsto hCh\T$). The same computation applies, with each equality an equality of laws and
with $X\mapsto gXh^{-1}$ acting on laws by pushforward; it uses only that the data transformations compose as a group
action.
\end{proof}

\subsection{Observation II.15}\label{proof:II.15}

\begin{restate}{Observation}{II.15}{descent through function-preserving reparametrisations}
Let $M=\prod_bM_b$ be a box-wise method: each $M_b$ is a member of a natural family with covariance group
$\mathcal G_{M_b}$, and no trainable parameters or random frames are shared across boxes. Let $G_{\mathrm{arch}}$ act on
$\Theta=\prod_b\Theta_b\times\Theta_{\mathrm{frozen}}$, preserving this split and acting on each adapted box by
$W_b\mapsto g_bW_bh_b^{-1}$ with $(g_b,h_b)\in\mathcal G_{M_b}$. Assume $f(g\theta,-)=f(\theta,-)$ on a
$G_{\mathrm{arch}}$-stable set containing $\Im M(\theta_0)$. Then $\Im M(g\theta_0)=g\cdot\Im M(\theta_0)$, and the set
of \emph{functions} reachable by $M$ from $g\theta_0$ equals the set reachable from $\theta_0$. For methods with random
frames, the equality holds in distribution, provided the frames are drawn independently per box. Boxes that share a
frame need a pathwise condition: a law-preserving map $T_{(g,h)}$ on frame space with
$\Im M_b(g_bWh_b^{-1};TF)=g_b\,\Im M_b(W;F)\,h_b^{-1}$ for all $W$ and $F$, with the same $T$ on every box that shares
$F$ (marginal covariance in law, as in \thmref{Definition}{II.14}, is not enough). For data-dependent methods, require
covariance pathwise in the data, and that the box-$b$ data at $g\theta_0$ is the \thmref{Definition}{II.14} transform
under $(g_b,h_b)$ of the data at $\theta_0$. This holds for intertwiner gauges such as hidden-unit permutations,
query--key and value--output head gauges and residual rotations. The statement concerns reachable sets only; training
trajectories and outcomes can still differ, for instance under Adam or non-isotropic initialisations.

Examples of $G_{\mathrm{arch}}$:
\begin{itemize}
  \item hidden-neuron permutations;
  \item the query--key gauge $(W_Q,W_K)\mapsto(gW_Q,g^{-\top}W_K)$ per head. With no positional encoding or an absolute
    one, $g\in\GL_{d_k}$. Under RoPE, $g$ must commute with every RoPE rotation, which leaves the per-frequency $2\times2$
    rotation--scalings $(\mathbb C^\times)^{d_k/2}$, shared across a GQA group. Under partial RoPE on $r$ of the $d_k$
    dimensions the group is $(\mathbb C^\times)^{r/2}\times\GL_{d_k-r}$. Under QK-norm, $g$ must in addition be
    orthogonal when the norm's $\varepsilon>0$ (conformal, $g\in\CO(d_k)$, when $\varepsilon=0$) and satisfy
    $g\T\Gamma_qR_t\Gamma_kg=\Gamma_qR_t\Gamma_k$ for every RoPE rotation $R_t$, where $\Gamma_q,\Gamma_k$ are the gains.
    OLMo~2 normalises $q$ and $k$ across all heads jointly;
  \item a single global orthogonal rotation $Q$ of the residual stream of a pre-norm transformer whose norm gains can all
    be fused into the linear layers they feed, after LayerNorm is converted to RMSNorm (SliceGPT's computational
    invariance; QuaRot's offline rotation). SliceGPT's per-block rotations $Q_\ell$ need extra residual matrices and
    change the architecture. Sandwich and post-sublayer norms (Gemma~2 and~3, OLMo~2) are excluded, because their gains
    cannot be fused and $\diag(\gamma)$ does not commute with $Q$. The comparison is between the gain-fused model and the
    fused-and-rotated model: gain fusion is a diagonal input rescaling outside $G_{\mathrm{arch}}$ that already changes
    PiSSA's split and block OFT's image (\thmref{Proposition}{II.16}), and converting LayerNorm to RMSNorm changes LoRA's
    reachable set on the residual writers.
\end{itemize}
\end{restate}

\begin{proof}
\emph{The deterministic statement.} No parameters are shared, so $\Im M(\theta)=\prod_b\Im M_b(W_b)\times\{\theta_{\mathrm{frozen}}\}$.
For $g\in G_{\mathrm{arch}}$, which preserves the split,
\begin{align*}
  \Im M(g\theta_0)&=\prod_b\Im M_b(g_bW_{0,b}h_b^{-1})\times\{g\theta_{0,\mathrm{frozen}}\}\\
  &=\prod_bg_b\,\Im M_b(W_{0,b})\,h_b^{-1}\times\{g\theta_{0,\mathrm{frozen}}\}=g\cdot\Im M(\theta_0),
\end{align*}
using $(g_b,h_b)\in\mathcal G_{M_b}$ in each factor. The reachable functions are then
\begin{equation*}
  \{\Phi_f(\theta):\theta\in\Im M(g\theta_0)\}=\{\Phi_f(g\theta):\theta\in\Im M(\theta_0)\}=\{\Phi_f(\theta):\theta\in\Im M(\theta_0)\},
\end{equation*}
since $f(g\theta,-)=f(\theta,-)$ for $\theta\in\Im M(\theta_0)$. (Stability of the set under $G_{\mathrm{arch}}$ is not needed.)

\emph{Independent frames.} With frames $F_b$ drawn independently per box, the law of the tuple of sets
$\Im M_b(g_bW_{0,b}h_b^{-1};F_b)$ is the product of the marginal laws, and by covariance in law each marginal equals the law of
$g_b\Im M_b(W_{0,b};F_b)h_b^{-1}$. So the tuple has the law of the tuple of sets $g_b\Im M_b(W_{0,b};F_b)h_b^{-1}$. The reachable function set is a fixed measurable function of the
image, and the function-level identity above holds pathwise, so the laws of the reachable function sets agree.

\emph{Shared frames.} Under the pathwise condition, $\Im M(g\theta_0;TF)=g\cdot\Im M(\theta_0;F)$ for every $F$, with the same $T$
in every box that uses $F$, and $TF$ has the law of $F$. Hence $\Im M(g\theta_0;F)$ and $g\cdot\Im M(\theta_0;F)$ have the same
law. Marginal covariance does not suffice. Take two boxes in $\R^{2\times1}$ sharing a Haar frame $F\in\Ort(2)$, with the
natural family $\Im M_b(W;F)=W+\R\,Fa(W)$, where $a(W)=(w_1^2,w_2^2)/\lVert(w_1^2,w_2^2)\rVert$ for $W\ne0$. For $g\in\Ort(2)$,
both $Fa(gW)$ and $gFa(W)$ are uniform on the circle, so each box is covariant in law under $(g,1)$. At $\theta_0=(e_1,e_2)$ the
two lines have directions $Fe_1$ and $Fe_2$, which are orthogonal, and so are those of $g\cdot\Im M(\theta_0;F)$. For $g$ the
rotation by $\pi/4$, $g\theta_0=\big((e_1+e_2)/\sqrt2,(e_2-e_1)/\sqrt2\big)$ has $a(ge_1)=a(ge_2)=(e_1+e_2)/\sqrt2$, so the two
lines of $\Im M(g\theta_0;F)$ are parallel. The joint laws differ, and the conclusion fails; the numerical check reports a
function-level instance.

\emph{Data.} Suppose each $M_b$ is covariant pathwise in its data,
\begin{equation*}
  \Im M_b\big(g_bWh_b^{-1};(g_b,h_b)\cdot D\big)=g_b\,\Im M_b(W;D)\,h_b^{-1},
\end{equation*}
and the box-$b$ data at $g\theta_0$ is $(g_b,h_b)\cdot D_b$. Then the deterministic argument applies with the data fixed. For an
intertwiner gauge the input of box $b$ in the network with weights $g\theta$ is $h_b$ times its input with weights $\theta$, for
every network input, so calibration inputs transform by $x\mapsto h_bx$ and second moments by $C\mapsto h_bCh_b\T$, as
\thmref{Definition}{II.14} requires. The examples below are of this kind: a hidden-unit permutation permutes the inputs of the
next box; the query--key and value--output gauges leave the inputs of $W_Q,W_K,W_V$ unchanged ($h=I$) and transform the input of
$W_O$; a residual rotation $Q$ rotates every normalised residual input by $Q$. Training is not covered: \thmref{Theorem}{III.12}
shows that Adam is not equivariant under general rotations of LoRA's gauge, and a Kaiming-uniform $A_0$ is not
rotation-invariant in law.

\emph{Hidden-neuron permutations.} For a permutation matrix $P$, $\sigma(Pz)=P\sigma(z)$ for an entrywise activation and
$P(u\odot v)=Pu\odot Pv$, so $(W_{\mathrm{up}},W_{\mathrm{gate}},W_{\mathrm{down}})\mapsto(PW_{\mathrm{up}},PW_{\mathrm{gate}},
W_{\mathrm{down}}P\T)$ preserves the block.

\emph{The query--key gauge.} Without positional rotation the score is $x\T W_Q\T W_Ky$, and since
$(gW_Q)\T g^{-\top}W_K=W_Q\T W_K$, every $g\in\GL_{d_k}$ works. Under RoPE the score between positions $s$ and $t$ is
$x\T W_Q\T R_{t-s}W_Ky$ with commuting rotations $R_\tau$ acting by angle $\tau\omega_j$ on the $j$-th coordinate plane. After
the gauge it is $x\T W_Q\T g\T R_{t-s}g^{-\top}W_Ky$. For $W_Q,W_K$ of full row rank, invariance for all inputs and positions
is equivalent to $g\T R_\tau g^{-\top}=R_\tau$ for every $\tau$, that is, $g\T$ commutes with every $R_\tau$. RoPE's frequencies
$\omega_j=\mathrm{base}^{-2j/d_k}$ are distinct and lie in $(0,1]\subset(0,\pi)$, so $R_1$ has the distinct eigenvalues
$e^{\pm\mathrm i\omega_j}$; a matrix commuting with $R_1$ preserves each coordinate plane and commutes with a nontrivial rotation
on it, so it is a rotation--scaling $a I_2+bJ$ there. The commutant is $(\mathbb C^\times)^{d_k/2}$ in the invertible case, and
it is closed under transposition, so the condition on $g\T$ is the same condition on $g$. In a GQA group one $W_K$ is paired with
several $W_Q$, so all of them use the same $g$. Under partial RoPE, $R_1$ is the identity on the $d_k-r$ unrotated coordinates and
has no eigenvalue $1$ on the others, so commuting matrices preserve this splitting, and the group is
$(\mathbb C^\times)^{r/2}\times\GL_{d_k-r}$.

\emph{QK-norm.} The score becomes $(\Gamma_qN(q))\T R_t(\Gamma_kN(k))$ with $q=W_Qx$, $k=W_Ky$ and
$N(v)=v\,(\lVert v\rVert^2/d_k+\varepsilon)^{-1/2}$. If $\varepsilon>0$ and $g$ is orthogonal, then $g^{-\top}=g$ and $N(gv)=gN(v)$,
so the score becomes $N(q)\T g\T\Gamma_qR_t\Gamma_kg\,N(k)$, which is invariant under the stated relation. If $\varepsilon=0$, then
$N$ removes scale, and $g=cO$ with $c>0$, $O$ orthogonal, gives $N(gq)=ON(q)$ and $N(g^{-\top}k)=ON(k)$; the relation is then
required of $O=g/c$. Conversely, let $W_Q,W_K$ have full row rank, so that $q$ and $k$ range over $\R^{d_k}$. Invariance at $t=0$
gives $\langle\Gamma_qN(gq),\Gamma_kN(g^{-\top}k)\rangle=\langle\Gamma_qN(q),\Gamma_kN(k)\rangle$ for all $q,k$. The values of $N$ span
$\R^{d_k}$ (they fill a ball when $\varepsilon>0$ and a sphere when $\varepsilon=0$), and the gains are invertible, so we may
choose $k_1,\dots,k_{d_k}$ for which the vectors $\Gamma_kN(g^{-\top}k_i)$ form a basis. Solving the resulting linear
equations gives $\Gamma_qN(gq)=A'\,\Gamma_qN(q)$ for a fixed matrix $A'$, so $N(gq)=A\,N(q)$ with $A=\Gamma_q^{-1}A'\Gamma_q$, and
$A$ is invertible because the values $N(gq)$ span $\R^{d_k}$. If $\varepsilon>0$, letting $q\to0$ (where $N(v)=v/\sqrt\varepsilon+O(\lVert v\rVert^3)$) gives $A=g$, and
then $\lVert gq\rVert=\lVert q\rVert$ for all $q$, so $g$ is orthogonal. If $\varepsilon=0$, $\lVert N(v)\rVert=\sqrt{d_k}$ for all
$v\ne0$, so $A$ is orthogonal, and $gq$ is a positive multiple of $Aq$ for every $q$, so $A^{-1}g$ has every vector as an
eigenvector and $g=cA$ with $c>0$. With $g$ orthogonal (or $O=g/c$), invariance for all $t$ and all $N(q),N(k)$ forces the
stated relation.

\emph{Residual rotations.} In a pre-norm transformer with gains fused and RMSNorm $N(h)=h\,(\lVert h\rVert^2/d+\varepsilon)^{-1/2}$,
replace the residual stream $h$ by $Qh$: embeddings $E\mapsto QE$, the writers $W_O,W_{\mathrm{down}}\mapsto QW_O,QW_{\mathrm{down}}$,
and every reader (the $q,k,v$, gate and up projections and the head) $W\mapsto WQ\T$. Since $\lVert Qh\rVert=\lVert h\rVert$,
$N(Qh)=QN(h)$, so each reader sees $WQ\T QN(h)=WN(h)$, and the function is unchanged. A gain that is not fused would sit
between $N$ and the reader as $\diag(\gamma)$, and $\diag(\gamma)Q\ne Q\diag(\gamma)$ unless $\gamma$ is constant. A
post-sublayer norm writes $h+\diag(\gamma)N(W_Oz)$, and $\diag(\gamma)N(QW_Oz)=\diag(\gamma)QN(W_Oz)\ne Q\diag(\gamma)N(W_Oz)$,
so the rotation does not preserve the function. LayerNorm subtracts the mean, the projection $I-\mathbf 1\mathbf 1\T/d$, which
commutes with $Q$ only if $Q\mathbf 1=\pm\mathbf 1$; SliceGPT therefore folds the projection into the writers and uses RMSNorm.
In the original model a LayerNorm ignores the component of the residual stream along $\mathbf 1$, so LoRA's updates of a writer
reach functions only through $(I-\mathbf 1\mathbf 1\T/d)\Delta W$; in the converted model the writer is
$(I-\mathbf 1\mathbf 1\T/d)W_O$ and an update $\Delta W$ with a component along $\mathbf 1$ is seen by every later RMSNorm, so the
reachable function sets differ. Gain fusion replaces a reader $W$ by $W\diag(\gamma)$, a non-conformal input rescaling when
$\gamma$ is not constant; by \thmref{Proposition}{II.16} it changes PiSSA's split, and it changes block OFT's image
$\{W\diag(\gamma)R\}$ against $\{WR\diag(\gamma)\}$ unless $\diag(\gamma)$ commutes with the block-orthogonal group, that is
unless $\gamma$ is constant on each block. Consecutive SliceGPT rotations $Q_{\ell-1}\ne Q_\ell$ require the matrix
$Q_{\ell-1}\T Q_\ell$ on the residual path, a new operation.
\end{proof}

\subsection{Proposition II.16}\label{proof:II.16}

\begin{restate}{Proposition}{II.16}{covariant data-aware splits}
Let $C=\mathbb E[xx\T]\succ0$ be the input second moment, so that under $x\mapsto hx$ we have $W\mapsto Wh^{-1}$ and
$C\mapsto hCh\T$. Let $LL\T=C$ and $p:=\min(m,n)$. Write $[M]_r$ for the rank-$r$ truncated SVD,
$\langle M\rangle_r:=M-[M]_{p-r}$ for the bottom $r$ singular components, and $\CO(n)=\R_{>0}\cdot\Ort(n)$. Assume
$1\le r<p$ and the spectral gap each split needs: $\sigma_r>\sigma_{r+1}$ of $W_0L$ in (a); in (b),
$\sigma_r>\sigma_{r+1}$ of $W_0C$ or $W_0$ for top-$r$ splits, and $\sigma_{p-r}>\sigma_{p-r+1}$ for bottom-$r$ splits.
Covariance groups are taken over the generic set where these gaps hold.

(a) The whitened split
\begin{equation*}
  X^*=\arg\min_{\rk X\le r}\lVert(W_0-X)L\rVert_F=[W_0L]_rL^{-1}=\Pi_rW_0,
\end{equation*}
where $\Pi_r$ is the top-$r$ eigenprojector of the output second moment $W_0CW_0\T$, is $\GL_n$-covariant on the input
side and $\CO(m)$-covariant on the output side; its joint group is $\CO(m)\times\GL_n$. This is the classical
reduced-rank-regression (SVD-LLM) solution.

(b) The formula-level CorDA splits, IPM ($X=[W_0C]_rC^{-1}$, the minimiser of $\lVert(W_0-X)C\rVert_F$) and KPM (the
bottom-$r$ components of $W_0C$ mapped back by $C^{-1}$, $X=\langle W_0C\rangle_rC^{-1}$), with or without CorDA's
trace-relative damping $C+\kappa(\tr C/n)I$, and the PiSSA and MiLoRA splits ($[W_0]_r$ and the bottom-$r$ part
$\langle W_0\rangle_r$), have input-side covariance group exactly $\CO(n)$; jointly, $\CO(m)\times\CO(n)$.

(c) \emph{Scope.} An absolute damping $C+\lambda I$ reduces the input-side group of (a) and of (b) to $\Ort(n)$. The
official and HF CorDA code divide each calibration sample by the absolute value of its largest entry before
accumulating $C$, which keeps only positive scalings and coordinate permutations, so (b) describes the CorDA formulas and
does not cover shipped CorDA under rotations.
\end{restate}

\begin{proof}
A split $X=X(W_0,C)$ of rank at most $r$ yields the method with frozen residual $W_0-X$ and image $(W_0-X)+\Mr$. Under $(g,h)$,
with $C'=hCh\T$, covariance of images means $(gW_0h^{-1}-X(gW_0h^{-1},C'))+\Mr=g(W_0-X(W_0,C))h^{-1}+\Mr$, using $g\Mr h^{-1}=\Mr$.
Since $r<p$, the translation stabiliser of $\Mr$ is $\{0\}$ (equation~\eqref{eq:translation-stabiliser}), so this is equivalent to
\begin{equation*}
  X(gW_0h^{-1},hCh\T)=g\,X(W_0,C)\,h^{-1}.
\end{equation*}

\emph{Lemma.} Let $1\le s<p$, $g\in\GL_m$ and $k\in\GL_n$. Then $[gMk]_s=g[M]_sk$ for every $M\in\R^{m\times n}$ for which both
truncations are unique iff $g\in\CO(m)$ and $k\in\CO(n)$. ``If'': for $g=aP$ and $k=bQ$ with $a,b>0$ and $P,Q$ orthogonal, an
SVD $M=\sum_i\sigma_iu_iv_i\T$ gives the SVD $gMk=\sum_i(ab\sigma_i)(Pu_i)(Q\T v_i)\T$ with the same ordering, so
$[gMk]_s=g[M]_sk$. ``Only if'': let $U_s\subseteq\R^m$ be any $s$-dimensional subspace and $y\perp U_s$ a unit vector, and
likewise $V_s\subseteq\R^n$ and $x\perp V_s$ (possible because $s<p$). Let $M_s$ have rank $s$ with column space $U_s$ and row
space $V_s$, and for small $t>0$ put $M=M_s+t\,yx\T$. Then $[M]_s=M_s$, and $gMk=gM_sk+t\,(gy)(k\T x)\T$, whose truncation is
unique for small $t$ because $gM_sk$ has rank $s$. The identity gives $[gMk]_s=gM_sk$, so the residual $t(gy)(k\T x)\T$ of the
truncation must have its column space orthogonal to that of $[gMk]_s$ and its row space orthogonal to that of $[gMk]_s$
(\auxref{B.11}(a): $[X]_s$ and $X-[X]_s$ are built from disjoint sets of singular vectors). So $gy\perp gU_s$ and
$k\T x\perp k\T V_s$. Since every pair of orthogonal unit vectors $u,y$ occurs with $u\in U_s$ and $y\perp U_s$, $g$ maps
orthogonal vectors to orthogonal vectors. Then for orthonormal eigenvectors $e,f$ of $g\T g$ with eigenvalues $\alpha,\beta$,
$0=\langle g(e+f),g(e-f)\rangle=\alpha-\beta$, so $g\T g$ is scalar and $g\in\CO(m)$; likewise $k\T\in\CO(n)$.

\emph{(a)} By reduced-rank regression (\auxref{B.11}(c)), $X^*=[W_0L]_rL^{-1}$, unique under the gap, and $[W_0L]_r=\Pi_rW_0L$
with $\Pi_r$ the top-$r$ eigenprojector of $W_0LL\T W_0\T=W_0CW_0\T$; so $X^*=\Pi_rW_0$, which does not depend on the choice of
$L$. Under $(g,h)$ the output second moment becomes $gW_0h^{-1}\,hCh\T\,h^{-\top}W_0\T g\T=gW_0CW_0\T g\T$, and with the square
root $hL$ of $hCh\T$,
\begin{equation*}
  X^*(gW_0h^{-1},hCh\T)=[gW_0L]_r\,L^{-1}h^{-1}.
\end{equation*}
This equals $gX^*h^{-1}=g[W_0L]_rL^{-1}h^{-1}$ for all $W_0$ (with gaps) iff $[gM]_r=g[M]_r$ for all $M=W_0L$, which range
over all matrices with a gap. By the lemma with $k=I$ this holds iff $g\in\CO(m)$, with no condition on $h$. So the joint group
is $\CO(m)\times\GL_n$, and its input and output parts are $\GL_n$ and $\CO(m)$.

\emph{(b), undamped.} For IPM, $X(W,C)=[WC]_rC^{-1}$ minimises $\lVert(W-X)C\rVert_F$ (\auxref{B.11}(c) with $L=C$). Under
$(g,h)$, with $M=W_0C$,
\begin{equation*}
  X(gW_0h^{-1},hCh\T)=[gW_0Ch\T]_rh^{-\top}C^{-1}h^{-1}=[gMh\T]_r\,h^{-\top}C^{-1}h^{-1},
\end{equation*}
and covariance asks this to equal $g[M]_rC^{-1}h^{-1}$, that is $[gMh\T]_r=g[M]_rh\T$. As $W_0$ varies, $M=W_0C$ ranges over all
matrices, so by the lemma covariance holds iff $g\in\CO(m)$ and $h\T\in\CO(n)$, that is $h\in\CO(n)$. KPM:
$X=\langle W_0C\rangle_rC^{-1}=W_0-[W_0C]_{p-r}C^{-1}$, and the same computation with $s=p-r\in[1,p-1]$ gives the condition
$[gMh\T]_{p-r}=g[M]_{p-r}h\T$, with the gap $\sigma_{p-r}>\sigma_{p-r+1}$. PiSSA: $[gW_0h^{-1}]_r=g[W_0]_rh^{-1}$ for all $W_0$ iff
$g\in\CO(m)$ and $h^{-1}\in\CO(n)$; MiLoRA: $\langle W\rangle_r=W-[W]_{p-r}$ reduces to the lemma at $s=p-r$. In each case the
input-side group (with $g=I$) is $\CO(n)$ and the joint group is $\CO(m)\times\CO(n)$.

\emph{(b), trace-relative damping.} Write $C_\kappa=C+\kappa(\tr C/n)I$ with $\kappa>0$ and consider $X_\kappa(W,C)=X(W,C_\kappa)$
for IPM (KPM is identical with $s=p-r$). For $h=cQ\in\CO(n)$,
$hC_\kappa h\T=c^2QCQ\T+\kappa(\tr C/n)c^2I=(hCh\T)_\kappa$, since $\tr(c^2QCQ\T)=c^2\tr C$; so damping commutes with $\CO(n)$ and
the damped splits inherit the covariance of the undamped ones. Conversely, let $(g,h)$ be in the joint group, put
$S_1=(hCh\T)_\kappa$, $S_2=C_\kappa$ and $N=W_0h^{-1}$. Covariance reads $[gNS_1]_rS_1^{-1}=g[NhS_2]_rS_2^{-1}h^{-1}$. With
$M=NS_1$, which ranges over all matrices, and $G=S_1^{-1}hS_2$, this becomes $[gM]_rG=g[MG]_r$, or $[gM'G^{-1}]_r=g[M']_rG^{-1}$
with $M'=MG$. By the lemma, $g\in\CO(m)$ and $G\in\CO(n)$ for every $C\succ0$, that is,
\begin{equation*}
  h\,S_2^2\,h\T=c(C)^2\,S_1^2\qquad\text{for every }C\succ0 .
\end{equation*}
Both sides are continuous in $C$, and $c(C)^2=\tr(hS_2^2h\T)/\tr(S_1^2)$ is too, so the identity extends to $C=vv\T$ with $v\ne0$.
There, with $w=hv$, $a=\kappa\lVert v\rVert^2/n>0$ and $b=\kappa\lVert w\rVert^2/n$,
$S_2^2=(\lVert v\rVert^2+2a)vv\T+a^2I$ and $S_1^2=(\lVert w\rVert^2+2b)ww\T+b^2I$, so
\begin{equation*}
  hh\T=\alpha_wI+\beta_w\,ww\T\qquad(\alpha_w=c^2b^2/a^2),
\end{equation*}
for every $w\in\R^n$, since $w=hv$ ranges over all vectors. So $hh\T-\alpha_wI$ has rank at most one for every $w$. If $hh\T$
were not scalar, it would have an eigenvalue $\alpha$ of multiplicity $n-1$ with $hh\T=\alpha I+\gamma ee\T$, $\gamma\ne0$; taking
$w$ neither parallel nor orthogonal to $e$, the matrix $(\alpha-\alpha_w)I+\gamma ee\T$ would have rank at most one with range
along $w$, which is impossible (for $n\ge3$ it has rank at least $n-1\ge2$ unless $\alpha_w=\alpha$, and then its range is
$\R e\ne\R w$; for $n=2$ its range is $\R e$ or $e^\perp$). So $hh\T$ is scalar and $h\in\CO(n)$. The input-side group of the damped
splits is therefore exactly $\CO(n)$, and the joint group $\CO(m)\times\CO(n)$.

\emph{(c), absolute damping.} $Q(C+\lambda I)Q\T=QCQ\T+\lambda I$ for $Q\in\Ort(n)$, so the damped splits of (a) and (b) are
$\Ort(n)$-covariant. We show that no other $h$ is in their input-side groups.

For (b): the undamped splits are invariant under $C\mapsto tC$, so the damped split at $(W,tC)$ equals the undamped split at
$(W,C+(\lambda/t)I)$, which tends to the undamped split at $(W,C)$ as $t\to\infty$ wherever the gap holds, because a truncation is
continuous where it is unique (it is the unique minimiser of a continuous function over the closed set $\MM_{\le s}$). The same
holds at the transformed data $(Wh^{-1},thCh\T)$. So every $h$ in the damped group lies in the undamped group $\CO(n)$.

For (a): with $S=(h\T h)^{-1}$, the transformed damped moment is $W_0h^{-1}(hCh\T+\lambda I)h^{-\top}W_0\T=W_0(C+\lambda S)W_0\T$, so
covariance reads $\Pi_r\big(W_0(C+\lambda S)W_0\T\big)W_0=\Pi_r\big(W_0(C+\lambda I)W_0\T\big)W_0$. Both projectors have range in
$\operatorname{col}W_0$, onto which $W_0$ maps, so this is equality of the projectors. Replacing $C$ by $tC$ and letting $t\to0$
gives $\Pi_r(W_0SW_0\T)=\Pi_r(W_0W_0\T)$ for every $W_0$ for which both have a gap at $r$. Suppose $S$ is not scalar, with
orthonormal eigenvectors $e,f$ for eigenvalues $\alpha>\beta$, and complete them to an orthonormal eigenbasis $e,f,g_3,\dots,g_n$ of
$S$. Choose orthonormal $z_1,\dots,z_p$ with $z_r=\cos\phi\,e+\sin\phi\,f$, $z_{r+1}=-\sin\phi\,e+\cos\phi\,f$ for some
$\phi\notin\tfrac\pi2\mathbb Z$, and the other $z_i$ among the $g_j$; put $W_0=\sum_{i\le p}d_ie_iz_i\T$ with $d_1>\dots>d_p>0$.
Then $W_0W_0\T=\diag(d_1^2,\dots,d_p^2)$ (padded), with top-$r$ eigenspace $\spanop\{e_1,\dots,e_r\}$, while $W_0SW_0\T$ is
diagonal except for the entry $d_rd_{r+1}(\beta-\alpha)\sin\phi\cos\phi\ne0$ in positions $(r,r+1)$ and $(r+1,r)$. Taking $d_i$ for
$i<r$ much larger and $d_i$ for $i>r+1$ much smaller than $d_r,d_{r+1}$, the top-$r$ eigenspace of $W_0SW_0\T$ is
$\spanop\{e_1,\dots,e_{r-1}\}$ plus the top eigenvector of the $2\times2$ block, which is not $e_r$; both matrices have a gap at
$r$, and the projectors differ. Hence $S$ is scalar and $h\in\CO(n)$.

It remains to exclude $h=cQ$ with $c\ne1$; since $Q$ is in the group, it suffices to exclude $h=cI$. For (a), $h=cI$ turns the
condition into $\Pi_r(W_0(C+\lambda c^{-2}I)W_0\T)=\Pi_r(W_0(C+\lambda I)W_0\T)$. Replacing $C$ by $tC$, the function
$P(\mu)=\Pi_r(W_0CW_0\T+\mu W_0W_0\T)$ would satisfy $P(\mu)=P(c^{-2}\mu)$ for all $\mu>0$, hence $P(\mu)=P(c^{-2j}\mu)$ for all
$j\in\mathbb Z$, and letting $j\to\pm\infty$ gives $P(0)=P(\infty)$, the top-$r$ eigenprojectors of $W_0CW_0\T$ and of $W_0W_0\T$
(both limits exist under gaps). Take $W_0$ rectangular-diagonal with entries $d_1>\dots>d_p>0$ and $C=\diag(c_i)$ with
$c_i=i/d_i^2$ for $i\le p$ and $c_i=1$ beyond. Then $W_0W_0\T$ has top-$r$ eigenspace $\spanop\{e_1,\dots,e_r\}$ and
$W_0CW_0\T=\diag(1,2,\dots,p)$ (padded) has top-$r$ eigenspace $\spanop\{e_{p-r+1},\dots,e_p\}$, which differ since $r<p$. For (b),
$h=cI$ gives, by scale invariance and linearity in $W$, $X(W_0,C+\lambda c^{-2}I)=X(W_0,C+\lambda I)$, and the same iteration
applied to $G(\mu)=X(W_0,C+\mu I)$ gives $G(0)=G(\infty)$ with $G(0)=X(W_0,C)$ and $G(\infty)=X(W_0,I)$. With the same $W_0$ and
$C$, $W_0C$ is rectangular-diagonal with the increasing entries $i/d_i$, so for IPM $G(0)=\sum_{i>p-r}d_iE_{ii}$ while
$G(\infty)=[W_0]_r=\sum_{i\le r}d_iE_{ii}$; for KPM the roles of top and bottom are exchanged, with the same conclusion. So
$cI$ is not in the group, and the damped input-side groups are exactly $\Ort(n)$.

\emph{(c), per-sample normalisation.} The shipped code divides a sample $x$ by $\nu(x)=\lvert\max_jx_j\rvert$. If $h=cP$ with $c>0$
and $P$ a permutation matrix, $\nu(hx)=c\,\nu(x)$, so the accumulated $C$ transforms as $hCh\T$ up to the factor $c^{-2}$, to which
the splits are insensitive, and $h\in\CO(n)$. Conversely, suppose $\nu(hx)=c\,\nu(x)$ for all $x$ and some $c>0$ ($n\ge2$). For
$x=-e_j$, $\nu(x)=0$, so $\max_i(-h_{ij})=0$ and $h\ge0$ entrywise; for $x=e_j$, $\max_ih_{ij}=c$; for $x=\mathbf 1$, every row sum of
$h$ is at most $c$. So a row $i_j$ with $h_{i_jj}=c$ has no other nonzero entry, the rows $i_j$ are distinct, and $h=cP$. Under a
rotation the normalisation does not commute with $h$, and the shipped split is not covariant. For instance, with $n=m=2$, $r=1$,
samples $x_1=e_1$ and $x_2=e_1+e_2$ (so $\nu=1$ for both and $C=x_1x_1\T+x_2x_2\T=\left(\begin{smallmatrix}2&1\\1&1\end{smallmatrix}
\right)$, up to a factor), and $h=Q$ the rotation by $\pi/4$: the rotated samples are $(1,1)/\sqrt2$ and $(0,\sqrt2)$, normalised to
$(1,1)$ and $(0,1)$, so the shipped moment is $C'=\left(\begin{smallmatrix}1&1\\1&2\end{smallmatrix}\right)$, while
$QCQ\T=\tfrac12\left(\begin{smallmatrix}1&1\\1&5\end{smallmatrix}\right)$. With $W_0=Q$, the shipped IPM split at the rotated data
is $X(I,C')=[C']_1C'^{-1}$, the top eigenprojector of $C'$, while covariance would require $X(Q,C)Q\T=Q[C]_1C^{-1}Q\T$, the top
eigenprojector of $QCQ\T$. The two symmetric matrices do not commute, so their top eigenvectors differ, and covariance fails.
\end{proof}


\subsection{Definition III.1}\label{proof:III.1}

\begin{restate}{Definition}{III.1}{trained method; cometric; three equivalences}
A \emph{trained method} is a method together with optimizer data: a learning-rate operator $\Lambda\succ0$ (per-block
rates), an initial scale, and an update rule. Gradient flow $\dot q=-\Lambda\nabla_q(L\circ\rho)$ moves the weights by
\[\dot\theta=-K_q\nabla_\theta L,\qquad K_q=D\rho_q\,\Lambda\,D\rho_q^\top.\]
We call $K_q$ the \textbf{induced cometric}. For a backward method with compression $c_t$ and anchor $a_t$
(\thmref{Definition}{IV.1}), $K_t=a_tc_t$. Two methods are \emph{expressively equivalent} if their images agree,
\emph{first-order equivalent} if their $K_{q_0}$ agree, and \emph{dynamically equivalent} if they produce the same
$\theta$-trajectory for every loss.
\end{restate}

A definition needs no proof. This subsection fixes the conventions used in all proofs of Exposé~III and checks the two
identities that the definition records.

\paragraph{Conventions}
Weight space $\Theta=\R^{m\times n}$ carries the Frobenius inner product $\langle X,Y\rangle=\tr(X^\top Y)$, and every
parameter space carries its Euclidean (Frobenius) inner product. Gradients are identified with covectors through these
inner products, and the transpose of a linear map is its adjoint. A covector on $\Theta$ is written as a matrix, usually
$G$. When $G=\nabla_\theta L$ is a weight gradient we say so; otherwise $G$ is an arbitrary cotangent signal.

LoRA is $\rho(B,A)=W_0+sBA$ with $B\in\R^{m\times r}$, $A\in\R^{r\times n}$ and $s>0$. Its differential and the
adjoint of the differential are
\begin{equation}\label{eq:C2-lora-diff}
D\rho_{(B,A)}(\dot B,\dot A)=s(\dot BA+B\dot A),\qquad D\rho_{(B,A)}^\top(G)=(sGA^\top,\ sB^\top G),
\end{equation}
since $\langle G,s(\dot BA+B\dot A)\rangle=\langle sGA^\top,\dot B\rangle+\langle sB^\top G,\dot A\rangle$. With
$G=\nabla_\theta L(\rho(B,A))$ this gives $\nabla_B(L\circ\rho)=sGA^\top$ and $\nabla_A(L\circ\rho)=sB^\top G$, written
$\nabla_B$ and $\nabla_A$. \emph{Block-scalar rates} are $\Lambda(B,A)=(\eta_BB,\eta_AA)$ with $\eta_A,\eta_B>0$.

The gauge group $\GL_r$ acts by $g\cdot(B,A)=(Bg^{-1},gA)$, and $\rho(g\cdot q)=\rho(q)$ (\bgref{D.21}). At
$q'=g\cdot q$ the weight is unchanged, hence so is the weight gradient $G$, and \eqref{eq:C2-lora-diff} gives
\begin{equation}\label{eq:C2-gauge-grad}
\nabla_{B'}=sGA^\top g^\top=\nabla_Bg^\top,\qquad \nabla_{A'}=sg^{-\top}B^\top G=g^{-\top}\nabla_A .
\end{equation}
The \emph{full-rank locus} is $\{\rk B=\rk A=r\}$. There $BA$ has rank $r$, $U:=\operatorname{col}B=\operatorname{col}(BA)$
and $V:=\operatorname{row}A=\operatorname{row}(BA)$, and the orthogonal projectors onto them are
$P_U=B(B^\top B)^{-1}B^\top$ and $P_V=A^\top(AA^\top)^{-1}A$.

An update rule is \emph{equivariant} under $g$ (\bgref{D.18}) if, for every sequence of losses (the same data order and
dropout masks in both runs), the run started from $g\cdot q_0$ with correspondingly transformed optimizer state produces
$g\cdot q_t$ at every step $t$. Equivariance gives identical weight trajectories, since $\rho(g\cdot q_t)=\rho(q_t)$. It
is checked one step at a time. If $q_t'=g\cdot q_t$, the two runs see the same weight and the same loss, hence the same
$G$, and their gradients are related by \eqref{eq:C2-gauge-grad}. It then suffices that the steps are transported,
\begin{equation}\label{eq:C2-equivariant-step}
\delta B'=\delta B\,g^{-1},\qquad\delta A'=g\,\delta A,
\end{equation}
together with the optimizer state, and induction over $t$ concludes.

\begin{proof}[Verification of the identities in Definition III.1]
By the chain rule $\nabla_q(L\circ\rho)=D\rho_q^\top\nabla_\theta L$, so gradient flow moves the weights by
$\dot\theta=D\rho_q\dot q=-D\rho_q\Lambda D\rho_q^\top\nabla_\theta L=-K_q\nabla_\theta L$. The operator $K_q$ is
symmetric, and $\langle\gamma,K_q\gamma\rangle=\langle D\rho_q^\top\gamma,\Lambda D\rho_q^\top\gamma\rangle\ge0$. Since
$\Lambda\succ0$, $\ker K_q=\ker D\rho_q^\top$, so $\rk K_q=\rk D\rho_q\le\dim Q$. Thus $K_q$ is a cometric in the sense
of \bgref{D.23}, possibly degenerate.

A backward method updates $\theta\leftarrow\theta-\eta\,a_t(\mathrm{opt}(c_t(\nabla L)))$ (\thmref{Definition}{IV.1}).
When $\mathrm{opt}$ is the identity, that is, plain (S)GD on the compressed gradient, and $\eta$ is absorbed into $a_t$,
the continuous-time update is $\dot\theta=-a_tc_t\nabla_\theta L$, which is the statement $K_t=a_tc_t$. This operator
is symmetric positive semidefinite when $a_t=c_t^\top$, as for GaLore ($P_tP_t^\top$), LISA (a coordinate projection)
and MeZO ($z_tz_t^\top$).

Finally, for two methods trained by gradient flow, dynamical equivalence implies first-order equivalence. For a linear
loss $L(\theta)=\langle\gamma,\theta\rangle$ the initial weight velocities are $-K_{q_0}\gamma$ and $-K'_{q_0'}\gamma$,
and $\gamma$ is arbitrary.
\end{proof}

\subsection{Observation III.2}\label{proof:III.2}

\begin{restate}{Observation}{III.2}{chain rule for cometrics}
Let $\sigma:Q'\to Q$ and $\rho:Q\to X$ be smooth, with $X=\Theta$ or a function or output space. Train $Q'$ by gradient
flow with a constant rate operator $\Lambda'\succ0$, and put $K^\sigma_{q'}:=D\sigma_{q'}\Lambda'D\sigma_{q'}^\top$. Then
\[K^{\rho\sigma}_{q'}=D\rho_{\sigma(q')}\,K^\sigma_{q'}\,D\rho_{\sigma(q')}^\top,\]
and no metric on $Q$ is used. $K$ is a degenerate cometric (positive semidefinite, of rank $\le\dim Q'$); when $\sigma$
is not injective, $K^\sigma$ is a field along $\sigma$ and fails to be a cometric on $Q$.

For LoRA, $\rho(B,A)=W_0+sBA$ with $s=\alpha/r$ (or $\alpha/\sqrt r$ for rsLoRA), with block rates
$\Lambda=\diag(\eta_BI,\eta_AI)$ and no LoRA dropout, under gradient flow (equivalently, to first order in the step
size for GD),
\[K_{(B,A)}(G)=s^2\big(\eta_B\,GA^\top A+\eta_A\,BB^\top G\big).\]
Under Adam there is no state-independent linear cometric; \thmref{Theorem}{III.10} and \thmref{Theorem}{III.12} treat
Adam.
\end{restate}

\begin{proof}
\emph{Chain rule.} By definition $K^{\rho\sigma}_{q'}=D(\rho\sigma)_{q'}\Lambda'D(\rho\sigma)_{q'}^\top$, and
$D(\rho\sigma)_{q'}=D\rho_{\sigma(q')}D\sigma_{q'}$. Hence
\[K^{\rho\sigma}_{q'}=D\rho_{\sigma(q')}\big(D\sigma_{q'}\Lambda'D\sigma_{q'}^\top\big)D\rho_{\sigma(q')}^\top
=D\rho_{\sigma(q')}K^\sigma_{q'}D\rho_{\sigma(q')}^\top.\]
Equivalently, the flow $\dot q'=-\Lambda'\nabla_{q'}(L\circ\rho\circ\sigma)=-\Lambda'D\sigma^\top D\rho^\top dL$ moves
$x=\rho(\sigma(q'))$ by $\dot x=-D\rho\,K^\sigma D\rho^\top dL$. In this formula
$D\rho_{\sigma(q')}^\top:T^*_xX\to T^*_{\sigma(q')}Q$ and $D\sigma_{q'}^\top:T^*_{\sigma(q')}Q\to T^*_{q'}Q'$ are dual
maps, and $K^\sigma_{q'}:T^*_{\sigma(q')}Q\to T_{\sigma(q')}Q$ is built from $\Lambda'$ alone. No inner product on $Q$
enters.

\emph{Degeneracy.} As in \S\ref{proof:III.1}, $K^{\rho\sigma}_{q'}$ is symmetric and positive semidefinite, and its
rank is at most $\rk D\sigma_{q'}\le\dim Q'$. The operator $K^\sigma_{q'}$ depends on $q'$ and not only on
$\sigma(q')$. For example, with $\sigma(a,b)=ab$ from $\R^2$ to $\R$ and $\Lambda'=I$, $K^\sigma_{(a,b)}=a^2+b^2$ takes
the value $2$ at $(1,1)$ and $17/4$ at $(2,1/2)$, two points over the same value $\sigma=1$. So for non-injective
$\sigma$, $K^\sigma$ is a field along $\sigma$ (a section of $\sigma^*(TQ\otimes TQ)$) and need not descend to $Q$.

\emph{LoRA.} By \eqref{eq:C2-lora-diff}, $\Lambda D\rho^\top(G)=(\eta_BsGA^\top,\eta_AsB^\top G)$, and applying $D\rho$
gives
\[K_{(B,A)}(G)=s\big((\eta_BsGA^\top)A+B(\eta_AsB^\top G)\big)=s^2\big(\eta_BGA^\top A+\eta_ABB^\top G\big).\]
A discrete GD step $\delta B=-\eta_BsGA^\top$, $\delta A=-\eta_AsB^\top G$ changes the weight by
\[\rho(B+\delta B,A+\delta A)-\rho(B,A)=s(\delta BA+B\delta A)+s\,\delta B\,\delta A
=-K_{(B,A)}(G)+s^3\eta_A\eta_B\,GA^\top B^\top G,\]
and the last term is of order $\eta^2$. With LoRA dropout the factors are trained on masked inputs. The same computation
then gives $K_{(B,A)}$ applied to the masked signal $\tilde G$ (for one example with output gradient $\delta$, input $x$
and mask $\mathsf m$, $\tilde G=\delta(\mathsf m\odot x)^\top$), which differs from $\nabla_\theta L$. This is why the
formula with $G=\nabla_\theta L$ assumes no dropout.

\emph{Adam.} Adam's step depends on its moment buffers, so it is not a function of $q$ and $G$ alone, and even the first
step from zero state has no linear cometric. Take $B=0$ and $A$ of full row rank. Then $\nabla_A=0$, so $A$ does not
move, and the bias-corrected first step on $B$ is $\delta B=-\eta_B\phi(sGA^\top)$, with $\phi(x)=x/(\lvert x\rvert+\varepsilon)$
entrywise and $0/0=0$. (Without bias correction the step is a positive multiple of the same expression with a rescaled
$\varepsilon$.) The weight moves by $\delta W=s\,\delta B\,A=-s\eta_B\phi(sGA^\top)A$, which is nonzero whenever
$GA^\top\ne0$, because $A$ has full row rank. Suppose $\delta W=-K(G)$ for a linear map $K$. At $\varepsilon=0$,
$\phi(cx)=\phi(x)$ for $c>0$ gives $K(G)=K(cG)=cK(G)$, so $K=0$. For $\varepsilon>0$, $\lvert\phi\rvert<1$ makes
$G\mapsto\delta W$ bounded, and a bounded linear map is $0$. Both conclusions contradict $\delta W\ne0$.
\end{proof}

\subsection{Theorem III.3}\label{proof:III.3}

\begin{restate}{Theorem}{III.3}{gradient descent is natural exactly where cometrics agree; for LoRA's linear gauge, exactly along $\Ort(r)$}
(a) Let $\varphi:Q\to Q'$ be a diffeomorphism with $\rho'\circ\varphi=\rho$, and let $\Lambda,\Lambda'$ be the rate
operators on $Q,Q'$. Under gradient flow, the two methods induce the same $\theta$-velocity at $q$ and at $\varphi(q)$
for every loss iff $K^\rho_q=K^{\rho'}_{\varphi(q)}$, that is,
\[D\rho'_{\varphi(q)}\big(D\varphi_q\Lambda D\varphi_q^\top-\Lambda'\big)D\rho'^\top_{\varphi(q)}=0.\]
This holds whenever $D\varphi\Lambda D\varphi^\top=\Lambda'$, which is an isometry when $\Lambda=\Lambda'=I$. That
condition is necessary when $D\rho'_{\varphi(q)}$ is injective, and fails to be necessary in general. Every linear gauge
of a linear parametrisation is natural; for instance, for $\rho(x,y)=\theta_0+x+y$ the shear $(x,y)\mapsto(2x,y-x)$
preserves $\rho$ and $K$ without being orthogonal. So is every pointwise-orthogonal non-linear LoRA gauge
$\varphi(q)=\varphi_{g(q)}(q)$ with $g(q)\in\Ort(r)$.

(b) For LoRA's linear gauge $\varphi_g(B,A)=(Bg^{-1},gA)$ with block-scalar rates, at points where $A$ has full row rank
($B$ arbitrary, including $B=0$) and $r<n$, the criterion holds iff $g\in\Ort(r)$. The same holds symmetrically where $B$
has full column rank and $r<m$.
\end{restate}

\begin{proof}
(a) Fix $q\in Q$ and put $\theta=\rho(q)=\rho'(\varphi(q))$. Gradient flow with rates $\Lambda$ on $Q$ moves the weight
at $q$ by $-K^\rho_q\nabla_\theta L$ (\S\ref{proof:III.1}); gradient flow with rates $\Lambda'$ on $Q'$ moves it at
$\varphi(q)$ by $-K^{\rho'}_{\varphi(q)}\nabla_\theta L$. As $L$ ranges over the linear losses $\langle\gamma,\cdot\rangle$,
the covector $\nabla_\theta L(\theta)=\gamma$ ranges over all of $T^*_\theta\Theta$, so the two velocities agree for every
loss iff the linear maps $K^\rho_q$ and $K^{\rho'}_{\varphi(q)}$ agree. Differentiating $\rho=\rho'\circ\varphi$ gives
$D\rho_q=D\rho'_{\varphi(q)}D\varphi_q$, so
\[K^\rho_q-K^{\rho'}_{\varphi(q)}=D\rho'_{\varphi(q)}\big(D\varphi_q\Lambda D\varphi_q^\top-\Lambda'\big)D\rho'^\top_{\varphi(q)},\]
which is the displayed criterion.

If $D\varphi_q\Lambda D\varphi_q^\top=\Lambda'$, the bracket vanishes. For $\Lambda=\Lambda'=I$ the condition says that
$D\varphi_q$ is orthogonal; in general it says that $\varphi$ is an isometry from $(Q,\langle\cdot,\Lambda^{-1}\cdot\rangle)$
to $(Q',\langle\cdot,\Lambda'^{-1}\cdot\rangle)$. If $D\rho'_{\varphi(q)}$ is injective, it has a left inverse $E$, and
multiplying $D\rho'MD\rho'^\top=0$ by $E$ on the left and by $E^\top$ on the right gives $M=0$; so the condition is then
necessary.

It is not necessary in general. Let $\rho(x)=\theta_0+\mathcal Lx$ be affine with linear part $\mathcal L$, and let $P$
be a linear gauge, $\mathcal LP=\mathcal L$, with $\Lambda'=\Lambda$. Then $D\rho'=\mathcal L$ and $D\varphi=P$, and
$\mathcal LP\Lambda P^\top\mathcal L^\top=\mathcal L\Lambda\mathcal L^\top$, so the criterion holds for every
$\Lambda$. For $\rho(x,y)=\theta_0+x+y$ and $\Lambda=I$, the shear $P(x,y)=(2x,y-x)$ satisfies $\mathcal LP=\mathcal L$,
while in block form $PP^\top=\begin{psmallmatrix}4&-2\\-2&2\end{psmallmatrix}\ne I$.

For a pointwise gauge $\varphi(q)=\varphi_{g(q)}(q)$ of LoRA, with $g:Q\to\GL_r$ smooth and $\varphi$ a diffeomorphism,
the product rule gives $D\varphi_q=\varphi_{g(q)}+V_q$. Here $\varphi_{g(q)}$ is the linear map
$(\dot B,\dot A)\mapsto(\dot Bg(q)^{-1},g(q)\dot A)$, and $V_q(v)=\frac{d}{d\epsilon}\big|_{\epsilon=0}\varphi_{g(q+\epsilon v)}(q)$
is tangent to the $\GL_r$-orbit of $\varphi(q)$. That orbit lies in a fibre of $\rho$, so $D\rho_{\varphi(q)}V_q=0$ and
$D\rho_{\varphi(q)}D\varphi_q=D\rho_{\varphi(q)}\varphi_{g(q)}$. The criterion depends on $D\varphi_q$ only through
$D\rho_{\varphi(q)}D\varphi_q$, so it reduces to the criterion for the linear gauge $\varphi_{g(q)}$ at $q$, which holds
for $g(q)\in\Ort(r)$ by (b).

(b) Here $Q'=Q$, $\rho'=\rho$, $\Lambda'=\Lambda$ is block-scalar and $\varphi_g(q)=g\cdot q$. By
\thmref{Observation}{III.2},
\[K_{g\cdot q}(G)=s^2\big(\eta_BGA^\top g^\top gA+\eta_ABg^{-1}g^{-\top}B^\top G\big),\]
so $K_q=K_{g\cdot q}$ iff, for every $G\in\R^{m\times n}$,
\[GX=YG,\qquad X=\eta_BA^\top(g^\top g-I)A\in\R^{n\times n},\quad Y=\eta_AB(I-g^{-1}g^{-\top})B^\top\in\R^{m\times m}.\]

\emph{Matrix-unit lemma: if $GX=YG$ for every $G\in\R^{m\times n}$, then $X=cI_n$ and $Y=cI_m$ for some $c\in\R$.} Take
$G=E_{ij}=e_ie_j^\top$. Comparing the $(k,l)$ entries of $E_{ij}X$ and $YE_{ij}$ gives $\delta_{ik}X_{jl}=Y_{ki}\delta_{jl}$.
With $k=i$ and $l\ne j$ this gives $X_{jl}=0$; with $l=j$ and $k\ne i$ it gives $Y_{ki}=0$; with $k=i$ and $l=j$ it gives
$X_{jj}=Y_{ii}$ for all $i,j$.

Now $\rk X\le r<n$, so $X=cI_n$ forces $c=0$, that is, $A^\top(g^\top g-I)A=0$. Since $A$ has full row rank, $AA^\top$ is
invertible, and multiplying by $(AA^\top)^{-1}A$ on the left and by $A^\top(AA^\top)^{-1}$ on the right gives
$g^\top g=I$. Conversely, $g\in\Ort(r)$ makes $X$ and $Y$ vanish. Only the full row rank of $A$ and $r<n$ were used, and
$B$ is arbitrary, including $B=0$. If instead $B$ has full column rank and $r<m$, then $\rk Y\le r<m$ forces $c=0$, so
$B(I-g^{-1}g^{-\top})B^\top=0$; multiplying by $(B^\top B)^{-1}B^\top$ on the left and by $B(B^\top B)^{-1}$ on the right
gives $g^{-1}g^{-\top}=I$, and again $g\in\Ort(r)$.
\end{proof}

\begin{remark*}{the family of plain-GD dynamics; LoRA+ under GD}
The formula for $K_{g\cdot q}$ depends on $g$ only through $g^\top g$, since $g^{-1}g^{-\top}=(g^\top g)^{-1}$, and
$g^\top g=h^\top h$ iff $hg^{-1}\in\Ort(r)$. So along a gauge orbit the cometrics of plain gradient descent are indexed by
$\Ort(r)\backslash\GL_r$, which $g\mapsto g^\top g$ identifies with $\Sym_r^+$ (\bgref{D.27}, \auxref{B.19}(a)). For $g=cI$ with $c>0$,
gradient flow with a single rate $\eta$ started from $g\cdot q$, pulled back by $g^{-1}$, satisfies $\dot B=c\dot B'=-\eta c^2\nabla_B$
and $\dot A=c^{-1}\dot A'=-\eta c^{-2}\nabla_A$ by \eqref{eq:C2-gauge-grad}. This is LoRA with rates
$(\eta_A,\eta_B)=(\eta c^{-2},\eta c^2)$, and the same holds step by step for GD, whose steps are linear in the gradients.
Choosing $c=\lambda^{1/4}$ and $\eta=\sqrt\lambda\,\eta_A$ gives LoRA+ with ratio $\lambda=\eta_B/\eta_A$. Under gradient
descent, LoRA+ is therefore the scalar gauge $\lambda^{1/4}I$ together with an overall change of rate.
\end{remark*}

\subsection{Proposition III.4}\label{proof:III.4}

\begin{restate}{Proposition}{III.4}{cometrics that descend to the rank-$r$ manifold; known in part}
Write $U=\operatorname{col}B$ and $V=\operatorname{row}A$. Statements (a), (b) and the first sentence of (d) are on the
full-rank locus; (c) needs only $\rk A_0=r$, with $B$ arbitrary, including $B_0=0$. All statements concern gradient
flow, or (S)GD to first order, on the stated gradients, undamped and without Adam. Cometrics include the learning rate.

(a) Scaled GD, $\delta B=-\eta\nabla_B(AA^\top)^{-1}$ and $\delta A=-\eta(B^\top B)^{-1}\nabla_A$, is $\GL_r$-equivariant
with first-order cometric $K(G)=\eta s^2(P_UG+GP_V)$.

(b) LoRA-Pro's equivalent gradient $\tilde g=P_UG+(I-P_U)GP_V$ is the Frobenius-orthogonal projection
$\Pi_{T_{W-W_0}\MM_r}(G)$. The flow of its SGD variant (LoRA-Pro Alg.~1) is the Riemannian gradient flow on
$W_0+\MM_r$.

(c) LoRA-FA's corrected gradient $s^{-2}\nabla_B(A_0A_0^\top)^{-1}$, fed to SGD with $A_0$ frozen, gives
$K(G)=\eta\,GP_V$ with $V=\operatorname{row}A_0$, which depends on $A_0$ only through $V\in\Gr(r,n)$. Its original form,
plain SGD on $B$, has $K(G)=s^2\eta_BGA_0^\top A_0$.

(d) The cometrics of (a) and (b) are functions of $W$ on the full-rank locus and define vector fields tangent to
$W_0+\MM_r$. The field of (c) is $G\mapsto\eta GP_V$ with $V=\operatorname{row}A_0$. Its flow is the Frobenius gradient
flow on the slice $W_0+\{XA_0\}=W_0+\R^{m\times n}P_V$, of dimension $mr$, which contains $W_0$ and lies in $W_0+\Mr$.
The rank-$r$ part of the slice, $N_V=\{\operatorname{row}(W-W_0)=V\}$, is open and dense in it and is a submanifold of
$W_0+\MM_r$. On $N_V$ the field is tangent to $\MM_r$ and equals $\eta GP_{\operatorname{row}(W-W_0)}$, a function of
$W$. So (c) descends too, as a degenerate cometric of rank $mr$ (less than $r(m+n-r)$ when $r<n$) whose flow keeps
$\operatorname{row}(W-W_0)=\operatorname{row}A_0$ fixed. The cometric of plain GD depends on the representative; it is
invariant only under $\Ort(r)$.
\end{restate}

\begin{proof}
In (a), (b) and the first sentence of (d), $(B,A)$ lies on the full-rank locus and the conventions of
\S\ref{proof:III.1} apply. Write $\mathcal T:=T_{W-W_0}\MM_r$, where $W-W_0=sBA$ has rank $r$, column space $U$ and row
space $V$.

(a) The steps are $\delta B=-\eta\nabla_B(AA^\top)^{-1}=-\eta s\,GA^\top(AA^\top)^{-1}$ and
$\delta A=-\eta(B^\top B)^{-1}\nabla_A=-\eta s\,(B^\top B)^{-1}B^\top G$. Hence
\begin{align*}
\delta W&=s(\delta BA+B\delta A)+s\,\delta B\,\delta A=-\eta s^2\big(GA^\top(AA^\top)^{-1}A+B(B^\top B)^{-1}B^\top G\big)+O(\eta^2)\\
&=-\eta s^2(GP_V+P_UG)+O(\eta^2),
\end{align*}
and the flow version gives $\dot W=-\eta s^2(P_UG+GP_V)$ exactly. Left multiplication by $P_U$ and right
multiplication by $P_V$ are orthogonal projections of $(\R^{m\times n},\langle\cdot,\cdot\rangle)$, so
$K(G)=\eta s^2(P_UG+GP_V)$ is symmetric positive semidefinite. For equivariance, let $q'=g\cdot q$. By
\eqref{eq:C2-gauge-grad} and $(A'A'^\top)^{-1}=(gAA^\top g^\top)^{-1}=g^{-\top}(AA^\top)^{-1}g^{-1}$,
\[\delta B'=-\eta\nabla_Bg^\top g^{-\top}(AA^\top)^{-1}g^{-1}=\delta B\,g^{-1},\]
and $(B'^\top B')^{-1}=(g^{-\top}B^\top Bg^{-1})^{-1}=g(B^\top B)^{-1}g^\top$ gives
$\delta A'=-\eta g(B^\top B)^{-1}g^\top g^{-\top}\nabla_A=g\,\delta A$. So the discrete step satisfies
\eqref{eq:C2-equivariant-step} for every $g\in\GL_r$.

(b) \emph{The tangent space.} We show that
\[\{XA+BY:\ X\in\R^{m\times r},\ Y\in\R^{r\times n}\}=\{D:\ P_{U^\perp}DP_{V^\perp}=0\}=\mathcal T.\]
The first inclusion $\subseteq$ holds because $AP_{V^\perp}=0$ and $P_{U^\perp}B=0$. Conversely, if
$P_{U^\perp}DP_{V^\perp}=0$, then
$D=P_UD+P_{U^\perp}DP_V=B\big[(B^\top B)^{-1}B^\top D\big]+\big[P_{U^\perp}DA^\top(AA^\top)^{-1}\big]A$. The common space
has dimension $mn-(m-r)(n-r)=r(m+n-r)=\dim\MM_r$ (\bgref{C.12}). The curves $t\mapsto s(B+t\dot B)(A+t\dot A)$ lie in
$\MM_r$ for small $t$, so $\{\dot BA+B\dot A\}\subseteq\mathcal T$, and equality of dimensions gives equality.

\emph{The projection.} Let $\mathcal N=\{P_{U^\perp}EP_{V^\perp}:E\in\R^{m\times n}\}$. For $D\in\mathcal T$ we
have $\langle D,P_{U^\perp}EP_{V^\perp}\rangle=\langle P_{U^\perp}DP_{V^\perp},E\rangle=0$, so
$\mathcal N\perp\mathcal T$. Moreover $G=(G-P_{U^\perp}GP_{V^\perp})+P_{U^\perp}GP_{V^\perp}$, with the first term in
$\mathcal T$ and the second in $\mathcal N$. Hence
\[\Pi_{\mathcal T}(G)=G-(I-P_U)G(I-P_V)=P_UG+GP_V-P_UGP_V=P_UG+(I-P_U)GP_V=\tilde g.\]

\emph{LoRA-Pro's SGD variant.} Algorithm~1 of \citet{wang2024lorac} takes the steps $\delta B=-\eta g^B$ and
$\delta A=-\eta g^A$ along a pair that minimises $\lVert s(g^BA+Bg^A)-G\rVert_F$ (their Theorem~2.1). As $(g^B,g^A)$
ranges over all pairs, $s(g^BA+Bg^A)$ ranges over $\mathcal T$, and the closest point of $\mathcal T$ to $G$ is
$\Pi_{\mathcal T}G$. So every minimiser satisfies $s(g^BA+Bg^A)=\Pi_{\mathcal T}G$; explicitly, the minimisers are
\[g^A=s^{-1}(B^\top B)^{-1}B^\top G+XA,\qquad g^B=s^{-1}(I-P_U)GA^\top(AA^\top)^{-1}-BX,\qquad X\in\R^{r\times r},\]
for which $sBg^A=P_UG+sBXA$ and $sg^BA=(I-P_U)GP_V-sBXA$. For every choice of the free matrix $X$,
\[\delta W=s(\delta BA+B\delta A)+s\,\delta B\,\delta A=-\eta\,\Pi_{\mathcal T}G+O(\eta^2).\]
The flow version is $\dot W=-\eta\,\Pi_{T_{W-W_0}\MM_r}\nabla_\theta L(W)$. Since $\Pi_{\mathcal T}\nabla_\theta L$ is the
Riemannian gradient of $L$ restricted to $W_0+\MM_r$ for the metric induced by the Frobenius inner product
(\bgref{D.24}, \auxref{B.30}; \citealp[Ch.~3]{xabsil2008optimization}), this is the Riemannian gradient flow on $W_0+\MM_r$, run at
rate $\eta$.

(c) Here $A=A_0$ is frozen with $\rk A_0=r$, and $B$ is arbitrary. The corrected step is
$\delta B=-\eta s^{-2}\nabla_B(A_0A_0^\top)^{-1}=-\eta s^{-1}GA_0^\top(A_0A_0^\top)^{-1}$, so
\[\Delta W=s\,\delta B\,A_0=-\eta\,GA_0^\top(A_0A_0^\top)^{-1}A_0=-\eta\,GP_V,\]
exactly, since $A_0$ does not move. $P_V$ depends only on $V=\operatorname{row}A_0$. The original LoRA-FA step
$\delta B=-\eta_B\nabla_B=-\eta_BsGA_0^\top$ gives $\Delta W=s\,\delta B\,A_0=-s^2\eta_BGA_0^\top A_0$.

(d) On the full-rank locus $P_U$ and $P_V$ are the projectors onto $\operatorname{col}(W-W_0)$ and
$\operatorname{row}(W-W_0)$, so the cometrics of (a) and (b) are functions of $W$. Their values lie in $\mathcal T$:
this is clear for $\Pi_{\mathcal T}$, and $P_{U^\perp}(P_UG+GP_V)P_{V^\perp}=0$.

For (c), the trajectory satisfies $W-W_0=sBA_0\in\{XA_0:X\in\R^{m\times r}\}$, and $\{XA_0\}=\{Z:\operatorname{row}Z\subseteq V\}=\R^{m\times n}P_V$,
a linear space of dimension $mr$. Right multiplication by $P_V$ is the Frobenius-orthogonal projection onto it, so
$\dot W=-\eta\nabla_\theta L(W)P_V$ is the gradient flow of $L$ restricted to the affine slice $W_0+\R^{m\times n}P_V$
with its flat Frobenius metric. The slice contains $W_0$ ($X=0$) and lies in $W_0+\Mr$, since $\rk XA_0\le r$. Because
$X\mapsto XA_0$ is injective, $\rk XA_0=\rk X$, and $\operatorname{row}(XA_0)=V$ iff $\rk X=r$. So
$N_V=W_0+\{XA_0:\rk X=r\}$, which is open in the slice because rank $r$ is an open condition (\auxref{B.1}(a)), and
dense because every $X\in\R^{m\times r}$ ($m\ge r$) is a limit of rank-$r$ matrices $X+tE$ as $t\to0$: if $E$ has
$I_r$ in its first $r$ rows and zeros elsewhere, the top $r\times r$ minor of $X+tE$ is a monic polynomial of degree $r$
in $t$, so it vanishes for only finitely many $t$. $N_V$ is an open
subset of an affine subspace, hence an embedded submanifold of $\R^{m\times n}$, and it lies in the embedded
submanifold $W_0+\MM_r$. The inclusion $N_V\hookrightarrow W_0+\MM_r$ is smooth by \auxref{B.2}(c), an immersion
because its composite with the inclusion $W_0+\MM_r\hookrightarrow\R^{m\times n}$ is one, and a homeomorphism onto its
image because both sets carry the subspace topology of $\R^{m\times n}$. So it is an embedding, and $N_V$ is an embedded
submanifold of $W_0+\MM_r$ (\bgref{C.4}). At $W\in N_V$, $\operatorname{row}(W-W_0)=V$, so $GP_V=GP_{\operatorname{row}(W-W_0)}$ is a
function of $W$, and, with $U_W=\operatorname{col}(W-W_0)$, $P_{U_W^\perp}(GP_V)P_{V^\perp}=0$ shows that it lies in $T_{W-W_0}\MM_r$ (it lies in
$\R^{m\times n}P_V=T N_V$ as well). The operator $G\mapsto\eta GP_V$ has rank $\dim\R^{m\times n}P_V=mr$, and
$mr<r(m+n-r)$ when $r<n$. Along the flow $W(t)$ stays in the slice, so $\operatorname{row}(W(t)-W_0)\subseteq V$ for
all $t$, with equality whenever $W(t)\in N_V$.

The last sentence of (d) is \thmref{Theorem}{III.3}(b).
\end{proof}

\begin{remark*}{the departures listed under Practice}
The symmetry statements in the Practice paragraph after \thmref{Proposition}{III.4} are proved with
\thmref{Theorem}{III.12}: damping $(B^\top B+\delta I)^{-1}$, $(AA^\top+\delta I)^{-1}$ with $\delta>0$ leaves exactly
$\Ort(r)$; an AdamW step applied after the preconditioner, as in scaled AdamW and in HF's LoRA-FA optimizer, leaves
exactly the signed permutations $B_r$; and LoRA-Pro's Sylvester-optimal free matrix leaves exactly $\Ort(r)$ at the
parameter level. For LoRA-FA with the undamped correction and AdamW at $\varepsilon=0$, the first step from $B_0=0$ is
$\Delta W=-s\eta_B\operatorname{sign}\big(GA_0^\top(A_0A_0^\top)^{-1}\big)A_0$. Replacing $A_0$ by $2A_0$, which has the
same row space, doubles this step, so the trajectory depends on $A_0$ beyond its row space (see also
\S\ref{proof:III.13}).
\end{remark*}

\subsection{Theorem III.5}\label{proof:III.5}

\begin{restate}{Theorem}{III.5}{Noether charges and isometries for cotangent signals; known, \citet{zhao2022symmetries}}
Let a Lie group act linearly on $Q=\R^N$ with $\rho\circ k=\rho$, let $\Lambda\succ0$ be constant, and let
$\gamma(t)\in T^*_{\rho(q(t))}\Theta$ be any cotangent signal, gradient or not. Consider the continuous-time updates
\[\dot q=-\Lambda\,D\rho_q^\top\gamma(t).\]

(i) For every generator $\xi$ with $\Lambda^{-1}\xi$ symmetric, the charge $J_\xi(q)=\tfrac12\langle q,\Lambda^{-1}\xi q\rangle$
is conserved.

(ii) For every generator $\xi$ with $\Lambda^{-1}\xi$ antisymmetric, $k=e^{\tau\xi}$ preserves
$\langle\cdot,\Lambda^{-1}\cdot\rangle$, so $K_{kq}=K_q$ along the orbit; for the same cotangent $\gamma$, the
$\theta$-velocities at $q$ and at $kq$ agree. Suppose moreover that $\gamma$ depends on $q$ only through $(t,\rho(q))$, as
$\gamma=\nabla_\theta L$ or an open-loop signal does, and that solutions are unique ($\gamma$ measurable in $t$ and
locally Lipschitz in $\theta$, $\rho$ of class $C^{1,1}_{\mathrm{loc}}$). Then the $\theta$-trajectories from $q_0$ and
from $kq_0$ coincide. If $\gamma$ uses Euclidean norms on $Q$, as global-norm clipping does, the trajectory statement
holds when, in addition, $k$ is a Euclidean isometry, and can fail otherwise. So (i) and the velocity statement of (ii)
hold for arbitrary $\gamma$, while the trajectory statement needs a gradient-like signal.

(iii) When $\mathfrak g$ is closed under $\xi\mapsto\Lambda\xi^\top\Lambda^{-1}$, every generator splits into a part of
type (i) and a part of type (ii), a Cartan decomposition $\mathfrak g=\mathfrak k\oplus\mathfrak p$. This holds for
LoRA's $\gl_r$ with per-block scalar rates. Without closure, there are generators of neither type. With per-rank-channel
rates $\Lambda(B,A)=(BD_B,D_AA)$, where $D_B=\diag(\eta_{B,j})$ and $D_A=\diag(\eta_{A,j})$, the generator $\xi_X$ is of
type (i) iff $XD_B^{-1}$ and $D_A^{-1}X$ are both symmetric, and of type (ii) iff both are antisymmetric. An off-diagonal
pair $(i,j)$ contributes one charge and one isometry iff $\eta_{B,i}\eta_{A,i}=\eta_{B,j}\eta_{A,j}$, and $\mathfrak g$
is closed iff all the products $\eta_{B,j}\eta_{A,j}$ are equal. In that case one gets a $D$-conjugate of
$\so_r\oplus\Sym_r$ whose isometries are not all Euclidean unless the rates are
block-scalar, so the clipping caveat of (ii) applies. For pairwise distinct
products, only the diagonal $X$ give charges and no nonzero $X$ gives an isometry.
\end{restate}

\begin{proof}
Part (i) is Noether's theorem in the form of \auxref{B.29}, which allows an arbitrary cotangent signal; we give the
short argument in full. Solutions are absolutely continuous and the identities below hold for almost
every $t$. For $\xi$ in the Lie algebra
$\mathfrak g\subseteq\gl(N)$ of the group, $\rho(e^{\tau\xi}q)=\rho(q)$ for all $\tau$; differentiating at $\tau=0$
gives the infinitesimal invariance
\begin{equation}\label{eq:C2-infinitesimal}
D\rho_q(\xi q)=0\qquad(q\in Q,\ \xi\in\mathfrak g).
\end{equation}

(i) If $\Lambda^{-1}\xi$ is symmetric, the gradient of $J_\xi$ is $\Lambda^{-1}\xi q$. Along a solution, using that
$\Lambda$ is symmetric and \eqref{eq:C2-infinitesimal},
\[\tfrac{d}{dt}J_\xi(q)=\langle\Lambda^{-1}\xi q,\dot q\rangle=-\langle\Lambda^{-1}\xi q,\Lambda D\rho_q^\top\gamma\rangle
=-\langle\xi q,D\rho_q^\top\gamma\rangle=-\langle D\rho_q\xi q,\gamma\rangle=0.\]
Nothing about $\gamma$ was used, so it may depend on $q$ in any way.

(ii) \emph{Isometry and cometric.} If $\Lambda^{-1}\xi$ is antisymmetric, then $\xi^\top\Lambda^{-1}=-\Lambda^{-1}\xi$, so
\[\tfrac{d}{d\tau}\big(e^{\tau\xi^\top}\Lambda^{-1}e^{\tau\xi}\big)=e^{\tau\xi^\top}(\xi^\top\Lambda^{-1}+\Lambda^{-1}\xi)e^{\tau\xi}=0,\]
and $k=e^{\tau\xi}$ satisfies $k^\top\Lambda^{-1}k=\Lambda^{-1}$, equivalently $k\Lambda k^\top=\Lambda$. From
$\rho\circ k=\rho$ we get $D\rho_q=D\rho_{kq}k$, so
\[K_q=D\rho_{kq}\,k\Lambda k^\top D\rho_{kq}^\top=D\rho_{kq}\Lambda D\rho_{kq}^\top=K_{kq}.\]
Since $\rho(kq)=\rho(q)$, the same covector $\gamma$ acts at both points, and the velocities $-K_q\gamma$ and
$-K_{kq}\gamma$ agree. The rest of the argument uses only the two properties $k\Lambda k^\top=\Lambda$ and $\rho\circ k=\rho$,
so it applies to every such $k$, not only to those of the form $e^{\tau\xi}$.

\emph{Trajectories.} Let $\gamma=\gamma(t,\rho(q))$ and let $q(t)$ solve $\dot q=F(t,q):=-\Lambda D\rho_q^\top\gamma(t,\rho(q))$
from $q_0$. Put $p(t)=kq(t)$. Then, using $D\rho_q^\top=k^\top D\rho_{kq}^\top$ and $\rho(p)=\rho(q)$,
\[\dot p=k\dot q=-k\Lambda k^\top D\rho_p^\top\gamma(t,\rho(q))=-\Lambda D\rho_p^\top\gamma(t,\rho(p))=F(t,p),\]
so $p$ solves the same equation from $kq_0$. Under the stated regularity, read locally uniformly in $t$, $F$ is measurable in
$t$ and locally bounded and locally Lipschitz in $p$, because $D\rho$ and $\rho$ are locally Lipschitz and $\gamma$ is
locally Lipschitz in $\theta$. By Gronwall's inequality two solutions with the same initial value coincide (\auxref{B.28}), so $p$ is the solution from
$kq_0$, and $\rho(p(t))=\rho(q(t))$ for all $t$.

\emph{Clipping.} Suppose $\gamma$ depends on $q$ through $(t,\rho(q))$ and through Euclidean norms
$\lVert D\rho_q^\top v\rVert$ of pulled-back covectors $v$ that are themselves functions of $(t,\rho(q))$. Global-norm
clipping is of this kind:
\[\gamma=\min\big(1,C/\lVert D\rho_q^\top\nabla_\theta L\rVert\big)\,\nabla_\theta L .\]
If $k$ is also a Euclidean isometry, then
$\lVert D\rho_{kq}^\top v\rVert=\lVert k^{-\top}D\rho_q^\top v\rVert=\lVert D\rho_q^\top v\rVert$, so $\gamma$ takes the
same value at $q$ and at $kq$, and the argument above goes through. Otherwise it can fail. Take $Q=\R^2$,
$\Theta=\R$, $\Lambda=\diag(1,4)$, $\rho(q)=\tfrac12q^\top\Lambda^{-1}q$ and $L(\theta)=\theta^2/2$. The group
$\{k:k^\top\Lambda^{-1}k=\Lambda^{-1}\}$ acts linearly and preserves $\rho$; its generator $\xi=\Lambda^{1/2}J\Lambda^{-1/2}$,
with $J=\begin{psmallmatrix}0&-1\\1&0\end{psmallmatrix}$, has $\Lambda^{-1}\xi=\Lambda^{-1/2}J\Lambda^{-1/2}$
antisymmetric, and $k=e^{(\pi/2)\xi}=\Lambda^{1/2}e^{(\pi/2)J}\Lambda^{-1/2}$ maps $q_0=(1,0)$ to $kq_0=(0,2)$, both with
$\theta=1/2$. Here $D\rho_q^\top\gamma=\gamma\Lambda^{-1}q$ and $\dot\theta=-\gamma\,q^\top\Lambda^{-1}q=-2\theta\gamma$.
The parameter gradients $L'(\theta)\Lambda^{-1}q$ have norms $1/2$ at $q_0$ and $1/4$ at $kq_0$, so with $C=1/10$ the
clipping factors are $1/5$ and $2/5$, and the initial velocities are $\dot\theta=-1/10$ and $\dot\theta=-1/5$. The
$\theta$-trajectories differ, while without clipping they coincide by the argument above.

(iii) \emph{Splitting.} Put $\vartheta(\xi)=-\Lambda\xi^\top\Lambda^{-1}$. It is an involution of $\gl(N)$
($\vartheta^2(\xi)=\Lambda\Lambda^{-1}\xi\Lambda\Lambda^{-1}=\xi$) and a Lie algebra automorphism:
$\vartheta([\xi,\zeta])=-\Lambda(\zeta^\top\xi^\top-\xi^\top\zeta^\top)\Lambda^{-1}=[\vartheta\xi,\vartheta\zeta]$. A
generator is of type (ii) iff $\Lambda^{-1}\xi$ is antisymmetric, iff $\Lambda\xi^\top\Lambda^{-1}=-\xi$, iff
$\vartheta\xi=\xi$; it is of type (i) iff $\vartheta\xi=-\xi$. If $\mathfrak g$ is closed under
$\xi\mapsto\Lambda\xi^\top\Lambda^{-1}$, it is $\vartheta$-stable, and
\[\xi=\tfrac12(\xi+\Lambda\xi^\top\Lambda^{-1})+\tfrac12(\xi-\Lambda\xi^\top\Lambda^{-1})\]
writes every $\xi\in\mathfrak g$ as a type (i) part (in the $-1$ eigenspace $\mathfrak p$ of $\vartheta$) plus a type (ii)
part (in the $+1$ eigenspace $\mathfrak k$). Since $\vartheta$ is an automorphism, $[\mathfrak k,\mathfrak k]\subseteq\mathfrak k$,
$[\mathfrak k,\mathfrak p]\subseteq\mathfrak p$ and $[\mathfrak p,\mathfrak p]\subseteq\mathfrak k$, which is the Cartan
decomposition (\bgref{D.27}, \auxref{B.19}(b)). If some $\xi\in\mathfrak g$ has $\Lambda\xi^\top\Lambda^{-1}\notin\mathfrak g$, then
$\vartheta\xi\ne\pm\xi$, so $\xi$ is of neither type.

\emph{LoRA.} The generators of the gauge are $\xi_X(B,A)=(BX,-XA)$ for $X\in\gl_r$ (the derivative of $\tau\mapsto e^{-\tau X}\cdot(B,A)$),
and $X\mapsto\xi_X$ is injective. For per-rank-channel rates $\Lambda(B,A)=(BD_B,D_AA)$,
\[\Lambda^{-1}\xi_X(B,A)=(BXD_B^{-1},\ -D_A^{-1}XA),\qquad (\Lambda^{-1}\xi_X)^\top(B,A)=\big(B(XD_B^{-1})^\top,\ -(D_A^{-1}X)^\top A\big),\]
because $\langle B',BM\rangle=\langle B'M^\top,B\rangle$ and $\langle A',NA\rangle=\langle N^\top A',A\rangle$. Hence
$\Lambda^{-1}\xi_X$ is symmetric iff $XD_B^{-1}$ and $D_A^{-1}X$ are symmetric, and antisymmetric iff both are
antisymmetric. Entrywise, $(XD_B^{-1})_{ij}=X_{ij}/\eta_{B,j}$ and $(D_A^{-1}X)_{ij}=X_{ij}/\eta_{A,i}$. For $i\ne j$, the
two symmetry conditions read $X_{ji}=X_{ij}\eta_{B,i}/\eta_{B,j}$ and $X_{ji}=X_{ij}\eta_{A,j}/\eta_{A,i}$; the
antisymmetry conditions are the same with a minus sign. Both admit $X_{ij}\ne0$ iff
$\eta_{B,i}/\eta_{B,j}=\eta_{A,j}/\eta_{A,i}$, that is, iff $\eta_{B,i}\eta_{A,i}=\eta_{B,j}\eta_{A,j}$, and then the pair
$(i,j)$ carries a one-dimensional space of each type. On the diagonal the symmetry conditions are empty and the
antisymmetry conditions force $X_{ii}=0$. So for pairwise distinct products the type (i) generators are exactly the
diagonal $X$, and the only type (ii) generator is $X=0$.

For closure, $\xi_X^\top(B,A)=(BX^\top,-X^\top A)$, so $\Lambda\xi_X^\top\Lambda^{-1}(B,A)=(BD_B^{-1}X^\top D_B,\ -D_AX^\top D_A^{-1}A)$.
An operator $(B,A)\mapsto(BM,-NA)$ lies in $\mathfrak g=\{\xi_Y\}$ iff $M=N$, so closure holds iff
$D_B^{-1}X^\top D_B=D_AX^\top D_A^{-1}$ for all $X$. The $(i,j)$ entries are $X_{ji}\eta_{B,j}/\eta_{B,i}$ and
$X_{ji}\eta_{A,i}/\eta_{A,j}$, which agree for all $X$ iff all products $\eta_{B,j}\eta_{A,j}$ are equal. With block-scalar
rates ($D_B=\eta_BI$, $D_A=\eta_AI$) this holds, the type (i) generators are the symmetric $X$ and the type (ii) ones the
antisymmetric $X$, and $\gl_r=\so_r\oplus\Sym_r$.

When all products equal $p$, $D_B=pD_A^{-1}$, and the type (i) generators are $X=SD_B$ with $S$ symmetric (then
$D_A^{-1}X=pD_A^{-1}SD_A^{-1}$ is symmetric too); the type (ii) generators are $X=KD_B$ with $K$ antisymmetric. With
$D=D_B^{1/2}$, conjugation $X\mapsto DXD^{-1}$ maps $SD_B$ to $DSD$ and $KD_B$ to $DKD$, so
$\mathfrak k\oplus\mathfrak p=D^{-1}(\so_r\oplus\Sym_r)D$. The isometries are the group elements $D^{-1}OD$ with
$O\in\SO(r)$, and $D^{-1}OD$ is orthogonal iff $O$ commutes with $D^2=D_B$. Unless $D_B$ is a multiple of the identity
(the block-scalar case), some of these isometries are not orthogonal, so they act on $Q$ by non-Euclidean maps and the
clipping caveat of (ii) applies.
\end{proof}

\begin{remark*}{instances and the discrete step}
Each instance listed after \thmref{Theorem}{III.5} applies (i) to the generator named there; we record the checks. All
assume gradient flow, or SGD without momentum, with per-block rates.
\begin{itemize}
\item \emph{LoRA.} For symmetric $X$, $J_{\xi_X}(B,A)=\tfrac12\big(\langle B,BX\rangle/\eta_B-\langle A,XA\rangle/\eta_A\big)=\tfrac12\tr(X\Phi)$
with $\Phi=B^\top B/\eta_B-AA^\top/\eta_A$, so the $r(r+1)/2$ charges are the entries of $\Phi$. The $\so_r$ generators
act by $(BO^\top,OA)$ with $O\in\SO(r)$, which are Euclidean isometries commuting with $\Lambda$, so (ii) holds under
global-norm clipping.
\item \emph{DyLoRA.} At each step the loss uses $\rho_b(B,A)=W_0+s_bBE_bA$ for a sampled $b$, with
$E_b=\diag(1,\dots,1,0,\dots,0)$ of rank $b$. The update is $-\Lambda D\rho_{b(t)}^\top\gamma$, and the proof of (i) needs
only $D\rho_{b(t)}(\xi q)=0$. Now $\rho_b(g\cdot q)=\rho_b(q)$ for all $(B,A)$ iff $g^{-1}E_bg=E_b$, that is, iff
$g$ preserves $\spanop(e_1,\dots,e_b)$ and $\spanop(e_{b+1},\dots,e_r)$. When every $b\in\{1,\dots,r\}$ can be sampled,
$g$ then preserves each line $\spanop(e_b)=\spanop(e_1,\dots,e_b)\cap\spanop(e_b,\dots,e_r)$, so it is diagonal. Diagonal $X$ are symmetric, which gives the charges $\diag\Phi$.
\item \emph{VeRA.} $\rho(b,d)=W_0+\diag(b)B_{\mathrm f}\diag(d)A_{\mathrm f}$ with frozen $B_{\mathrm f},A_{\mathrm f}$ is
invariant under $(b,d)\mapsto(cb,d/c)$, with generator $(b,-d)$ and charge $\tfrac12(\lVert b\rVert^2/\eta_b-\lVert d\rVert^2/\eta_d)$.
NOLA's two coefficient vectors behave in the same way.
\item \emph{AdaLoRA.} $\rho(P,\lambda,Q)=W_0+P\diag(\lambda)Q$ is invariant under $(p_k,\lambda_k)\mapsto(cp_k,\lambda_k/c)$ and
under $(\lambda_k,q_k)\mapsto(\lambda_k/c,cq_k)$; at equal rates the charges are proportional to
$\lVert p_k\rVert^2-\lambda_k^2$ and $\lVert q_k\rVert^2-\lambda_k^2$. The orthogonality regulariser
$\lVert P^\top P-I\rVert^2+\lVert QQ^\top-I\rVert^2$ is not invariant under these scalings, so the regularised loss does not
factor through $\rho$, and rank-budget masking changes $\rho$.
\item \emph{HRA.} $\rho(u_1,\dots,u_r)=W_0\prod_i(I-2u_iu_i^\top/\lVert u_i\rVert^2)$ is invariant under $u_i\mapsto cu_i$,
with charge $\tfrac12\lVert u_i\rVert^2/\eta$.
\item \emph{ReLU bottleneck adapters} $h\mapsto h+U\operatorname{relu}(Dh+\beta)$. Multiplying row $j$ of $D$ and $\beta_j$ by
$c>0$ and column $j$ of $U$ by $1/c$ preserves the function, since $\operatorname{relu}(cz)=c\operatorname{relu}(z)$. The
target is a function space, which \thmref{Observation}{III.2} allows, and the derivative of $\rho$ along the scaling
direction vanishes by homogeneity even at the kinks. Each hidden unit gives the charge
$\tfrac12\big((\lVert D_{j,:}\rVert^2+\beta_j^2)/\eta_D-\lVert U_{:,j}\rVert^2/\eta_U\big)$, with a common rate for $D$
and $\beta$.
\item \emph{TT adapters.} Neighbouring cores share a bond of dimension $r_k$, on which $\GL(r_k)$ acts by $h^{-1}$ on one
side and $h$ on the other. The symmetric generators give one conserved symmetric $r_k\times r_k$ matrix per bond, the
Gram matrix of the left unfolding of one core divided by its rate minus that of the right unfolding of the next core
divided by its rate.
\item \emph{DoRA's detached-norm backward, global-norm clipping, LoRA dropout.} DoRA writes $W=V\diag(\mathsf g/\lVert V\rVert_c)$
with $V=W_0+sBA$, column norms $\lVert V\rVert_c$ and magnitudes $\mathsf g$. Treating $\lVert V\rVert_c$ as a constant in
the backward pass gives $\nabla_B=s\gamma A^\top$ and $\nabla_A=sB^\top\gamma$ with
$\gamma=\nabla_WL\,\diag(\mathsf g/\lVert V\rVert_c)$. Clipping multiplies the factor gradients by one scalar, and dropout
replaces $G$ by the masked signal of \S\ref{proof:III.2}. In each case the update of $(B,A)$ is
$-\Lambda D\rho_{\mathrm{LoRA}}^\top\gamma$ for some $\gamma$, so $\Phi$ is conserved.
\item \emph{Discrete GD.} With $s$ absorbed into $G$, one step $B^+=B-\eta_BGA^\top$, $A^+=A-\eta_AB^\top G$ gives
\begin{gather*}
\tfrac{1}{\eta_B}B^{+\top}B^+=\tfrac{1}{\eta_B}B^\top B-(AG^\top B+B^\top GA^\top)+\eta_BAG^\top GA^\top,\\
\tfrac{1}{\eta_A}A^+A^{+\top}=\tfrac{1}{\eta_A}AA^\top-(B^\top GA^\top+AG^\top B)+\eta_AB^\top GG^\top B,
\end{gather*}
so $\Phi^+-\Phi=\eta_BAG^\top GA^\top-\eta_AB^\top GG^\top B$ exactly. This is of order $\eta$ while $\Phi$ is of order
$1/\eta$, so $\eta\Phi$, or $\Phi$ relative to its size, drifts by $O(\eta^2)$ per step.
\item \emph{Momentum and Adam.} A heavy-ball step adds a buffer of gradients taken at past points, $D\rho_{q_s}^\top G_s$
with $s<t$, and Adam rescales gradients coordinatewise. Neither step has the form $-\Lambda D\rho_q^\top\gamma$ at the
current $q$ with a constant $\Lambda$, so (i) does not apply.
\end{itemize}
\end{remark*}

\subsection{Corollary III.6}\label{proof:III.6}

\begin{restate}{Corollary}{III.6}{charge at initialisation; weight decay}
(a) $\Phi_0=B_0^\top B_0/\eta_B-A_0A_0^\top/\eta_A$ is the value of the conserved charge, fixed by the initialisation and
the rates. Since $\Ort(r)$ acts on $\Phi$ by conjugation and preserves $\theta$-trajectories (\thmref{Theorem}{III.5}(ii)),
only the spectrum of $\Phi_0$ labels a $\theta$-trajectory.

\smallskip
{\upshape\small\renewcommand{\arraystretch}{1.12}\noindent
\begin{tabularx}{\linewidth}{@{}>{\raggedright\arraybackslash}p{0.5\linewidth}>{\raggedright\arraybackslash}X@{}}
\toprule\headrow
\textbf{Initialisation} & \textbf{Charge $\Phi_0$}\\
\midrule
HF default LoRA on Linear and Conv layers ($B_0=0$; Kaiming-uniform $A_0$ with entries $U(-n^{-1/2},n^{-1/2})$, $n$ the fan-in, which is \texttt{in\_channels} times the kernel size for Conv) & $-A_0A_0^\top/\eta_A$, with $\mathbb{E}[A_0A_0^\top]=\tfrac13I_r$ and fluctuations $O(n^{-1/2})$ on and off the diagonal\\
HF default LoRA on embedding layers ($A_0=0$, normal $B_0$) & $+B_0^\top B_0/\eta_B$ (Init[B]-type)\\
HF \texttt{mica} \citep{rudiger2026mica} ($A_0=0$, frozen minor-singular $B_0$) & $+B_0^\top B_0/\eta_B$ (Init[B]-type, with $B$ frozen)\\
EVA \citep{paischer2024parameter} (orthonormal rows; default \texttt{whiten=False}) & $-I/\eta_A$ exactly\\
PiSSA in HF PEFT \citep{meng2024pissa}, and MiLoRA \citep{wang2024milora} if implemented like \texttt{pissa\_init} with the bottom-$r$ singular values (MiLoRA is not in HF PEFT); $B_0^\top B_0=A_0A_0^\top=S_r/s$, with $S_r$ the top-$r$, resp.\ bottom-$r$, singular values and $s$ the LoRA scale & $(S_r/s)(1/\eta_B-1/\eta_A)$, zero \textbf{iff} $\eta_A=\eta_B$ (when $S_r\ne0$)\\
OLoRA \citep{buyukakyuz2024olora} ($B_0=Q_r$, $A_0=R_r$) & $I/\eta_B-R_rR_r^\top/\eta_A$\\
``Init[B]'' \citep{hayou2024impact} ($A_0=0$) & $+B_0^\top B_0/\eta_B$\\
\bottomrule
\end{tabularx}}
\smallskip

(b) Let $G$ be any cotangent signal, with the scale $s$ absorbed into it. Under gradient flow, if decay acts per unit time
and identically on both factors,
\[\dot B=-\eta_BGA^\top-\lambda B,\qquad\dot A=-\eta_AB^\top G-\lambda A,\]
then $\dot\Phi=-2\lambda\Phi$. With learning-rate-coupled decay (PyTorch SGD \texttt{weight\_decay}; the decay term of
PyTorch AdamW), $\dot B=-\eta_B(GA^\top+\lambda B)$ and likewise for $A$, so that $\dot\Phi=-2\lambda(B^\top B-AA^\top)$.
Then $\Phi(t)=e^{-2\eta\lambda t}\Phi(0)$ holds when $\eta_A=\eta_B=\eta$. For $\eta_A\ne\eta_B$ it fails in general,
since $-2\lambda(B^\top B-AA^\top)$ is not a function of $\Phi$. With an actual Adam(W) gradient term, $\Phi$ drifts even
without decay, at first order in the step size; Adam lies outside \thmref{Theorem}{III.5}. So neither law describes AdamW
training.
\end{restate}

\begin{proof}
(a) \emph{Conservation and conjugation.} With block-scalar rates the entries of $\Phi$ are the charges of
\thmref{Theorem}{III.5}(i) (\S\ref{proof:III.5}, remark on instances), so under gradient flow $\Phi(t)=\Phi_0$, the value
fixed by $(B_0,A_0)$ and $(\eta_A,\eta_B)$. For $O\in\Ort(r)$, the map $k:(B,A)\mapsto(BO,O^\top A)$ is the gauge
element $g=O^\top$, and
\[\Phi(BO,O^\top A)=O^\top\tfrac{B^\top B}{\eta_B}O-O^\top\tfrac{AA^\top}{\eta_A}O=O^\top\Phi(B,A)\,O.\]
The linear map $k$ is orthogonal on $Q$ and commutes with block-scalar $\Lambda$, so $k\Lambda k^\top=\Lambda$, and
$\rho\circ k=\rho$. The trajectory argument of \thmref{Theorem}{III.5}(ii) uses only these two properties
(\S\ref{proof:III.5}), so the $\theta$-trajectories from $q_0$ and from $kq_0$ coincide for every signal that depends on
$q$ only through $(t,\rho(q))$, and also under global-norm clipping, since $k$ is a Euclidean isometry. This covers all of
$\Ort(r)$, reflections included. So initialisations whose charges are conjugate under $\Ort(r)$ produce the same
$\theta$-trajectory, and a function of $\Phi_0$ that labels $\theta$-trajectories must be invariant under
$\Phi_0\mapsto O^\top\Phi_0O$. By the spectral theorem the $\Ort(r)$-conjugacy class of the symmetric matrix $\Phi_0$ is
determined by its eigenvalues, so such a function is a function of the spectrum of $\Phi_0$.

\emph{The table.} Each entry follows by substituting the initialisation into $\Phi_0$.
\begin{itemize}
\item \emph{HF default, Linear and Conv.} $B_0=0$ gives $\Phi_0=-A_0A_0^\top/\eta_A$. HF initialises $A_0$ with
\texttt{kaiming\_uniform\_} at $a=\sqrt5$, whose bound is $\sqrt{2/(1+a^2)}\sqrt{3/n}=n^{-1/2}$, with $n$ the fan-in
(\texttt{in\_channels} times the product of the kernel sizes for Conv). An entry $U(-n^{-1/2},n^{-1/2})$ has variance
$1/(3n)$ and fourth moment $1/(5n^2)$. A diagonal entry of $A_0A_0^\top$ is a sum of $n$ independent squares, with mean
$1/3$ and variance $n\big(1/(5n^2)-1/(9n^2)\big)=4/(45n)$; an off-diagonal entry is a sum of $n$ independent products,
with mean $0$ and variance $n\cdot1/(9n^2)=1/(9n)$. So $\mathbb E[A_0A_0^\top]=\tfrac13I_r$, with fluctuations of standard
deviation $2/(3\sqrt{5n})$ on the diagonal and $1/(3\sqrt n)$ off it.
\item \emph{Embeddings, Init[B] and \texttt{mica}.} $A_0=0$ gives $\Phi_0=B_0^\top B_0/\eta_B$. For \texttt{mica}, $B$ is
frozen, so its rate is $0$ and \thmref{Theorem}{III.5} does not apply to it; the entry records the formula and its sign
type, and the training dynamics are those of the one-sided method $A\mapsto W_0+sB_0A$, along which $B_0^\top B_0$ is
constant and $AA^\top$ moves.
\item \emph{EVA.} $B_0=0$ and orthonormal rows, $A_0A_0^\top=I$, give $\Phi_0=-I/\eta_A$.
\item \emph{PiSSA in HF PEFT.} \texttt{pissa\_init} takes the top-$r$ singular triples $(U_r,S_r,V_r)$ of $W_0$, divides
$S_r$ by the scale $s$, and sets $A_0=(S_r/s)^{1/2}V_r^\top$ and $B_0=U_r(S_r/s)^{1/2}$. With $U_r^\top U_r=V_r^\top V_r=I$,
$B_0^\top B_0=A_0A_0^\top=S_r/s$, so $\Phi_0=(S_r/s)(1/\eta_B-1/\eta_A)$, which vanishes iff $\eta_A=\eta_B$ when
$S_r\ne0$. The same computation with the bottom-$r$ triples gives the MiLoRA entry.
\item \emph{OLoRA.} $B_0=Q_r$ has orthonormal columns, so $\Phi_0=I/\eta_B-R_rR_r^\top/\eta_A$.
\end{itemize}

(b) Write both decay conventions as $\dot B=-\eta_BGA^\top-c_BB$ and $\dot A=-\eta_AB^\top G-c_AA$, with $c_B=c_A=\lambda$
per unit time and $c_B=\eta_B\lambda$, $c_A=\eta_A\lambda$ when coupled. Then
\[\tfrac{d}{dt}\tfrac{B^\top B}{\eta_B}=-(AG^\top B+B^\top GA^\top)-2c_B\tfrac{B^\top B}{\eta_B},\qquad
\tfrac{d}{dt}\tfrac{AA^\top}{\eta_A}=-(B^\top GA^\top+AG^\top B)-2c_A\tfrac{AA^\top}{\eta_A},\]
and subtracting, the $G$-terms cancel:
\[\dot\Phi=-2c_B\tfrac{B^\top B}{\eta_B}+2c_A\tfrac{AA^\top}{\eta_A}.\]
Per unit time this is $\dot\Phi=-2\lambda\Phi$, so $\Phi(t)=e^{-2\lambda t}\Phi(0)$. Coupled, it is
$\dot\Phi=-2\lambda(B^\top B-AA^\top)$; when $\eta_A=\eta_B=\eta$ the right-hand side is $-2\lambda\eta\Phi$, so
$\Phi(t)=e^{-2\eta\lambda t}\Phi(0)$. For $\eta_A\ne\eta_B$ there are states with the same $\Phi$ and different
$B^\top B-AA^\top$: replacing $(B^\top B,AA^\top)$ by $(B^\top B+\eta_BC,AA^\top+\eta_AC)$ with $C\succeq0$ (realised by
suitable factors, since $m,n\ge r$) leaves $\Phi$ unchanged and changes $B^\top B-AA^\top$ by $(\eta_B-\eta_A)C$. Since
$\dot\Phi$ does not involve $G$, two such initial states have equal $\Phi(0)$ and different $\dot\Phi(0)$, so no law
$\dot\Phi=F(\Phi)$, and in particular no exponential law, holds for all trajectories. Some trajectories do satisfy one:
with $B=0$ and $G=0$, $\Phi(t)=e^{-2\eta_A\lambda t}\Phi(0)$; so the law fails in general and not always.

\emph{Adam.} One Adam step can move $\eta\Phi$ at first order. Take $m=n=r=1$, $s$ absorbed into $G$, rates
$\eta_B=\eta_A=\eta$, $(b,a)=(2,1)$, $G=1$ and zero optimizer state. The bias-corrected first step at $\varepsilon=0$ is
$\delta b=-\eta\operatorname{sign}(Ga)=-\eta$ and $\delta a=-\eta\operatorname{sign}(bG)=-\eta$, so
\[\eta\Phi^+-\eta\Phi=(2-\eta)^2-(1-\eta)^2-(4-1)=-2\eta,\]
of first order in $\eta$; the GD step of the previous remark moves $\eta\Phi$ by $O(\eta^2)$. Adam's step is not of the
form $-\Lambda D\rho^\top\gamma$, so it lies outside \thmref{Theorem}{III.5}, and neither decay law describes AdamW
training.
\end{proof}

\begin{remark*}{the two decay conventions as gauge fixings}
Per-time decay is the $\Lambda$-gradient flow of $L+\tfrac\lambda2\big(\lVert B\rVert^2/\eta_B+\lVert A\rVert^2/\eta_A\big)$,
since $-\eta_B\nabla_B\big(\tfrac\lambda2\lVert B\rVert^2/\eta_B\big)=-\lambda B$. Coupled decay is the $\Lambda$-gradient
flow of $L+\tfrac\lambda2(\lVert B\rVert^2+\lVert A\rVert^2)$. At an equilibrium of the coupled flow,
$\nabla_B=-\lambda B$ and $\nabla_A=-\lambda A$; since $B^\top\nabla_B=sB^\top GA^\top=\nabla_AA^\top$, this gives
$\lambda(B^\top B-AA^\top)=0$, so equilibria lie on the unweighted balanced slice $\mu=B^\top B-AA^\top=0$. At an
equilibrium of the per-time flow, $\eta_B\nabla_B=-\lambda B$ and $\eta_A\nabla_A=-\lambda A$, and the same identity gives
$\Phi=0$; per-time decay also drives $\Phi\to0$ exponentially by (b). The two slices
$\{B^\top B=AA^\top\}$ and $\{B^\top B=(\eta_B/\eta_A)AA^\top\}$ coincide iff $\eta_A=\eta_B$. HF's PiSSA has
$B_0^\top B_0=A_0A_0^\top$ at every rate, so it starts on the coupled-decay slice $\mu=0$, and off the per-time slice
$\Phi=0$ when $\eta_A\ne\eta_B$.
\end{remark*}

\subsection{Proposition III.7}\label{proof:III.7}

\begin{restate}{Proposition}{III.7}{Kempf--Ness slice; Loewner floor; known, \citet{xkempf1979length,xsrebro2004maximum}}
(a) On a full-rank $\GL_r$-orbit, the critical points of $\lVert B\rVert^2+\lVert A\rVert^2$ are exactly the balanced
representatives, those with $\mu=B^\top B-AA^\top=0$. They form one $\Ort(r)$-orbit, they are the minima, and the minimum
value is $2\lVert BA\rVert_*$.

(b) Assume gradient flow with block-scalar rates $\eta_A,\eta_B$, no weight decay, a loss that depends on $(B,A)$ only
through $BA$, and any cotangent signal of the form $\Lambda D\rho^\top\gamma$ (dropout, minibatches and clipping
included). From $B_0=0$,
\[B^\top B=\tfrac{\eta_B}{\eta_A}\big(AA^\top-A_0A_0^\top\big)\succeq0,\]
so $AA^\top\succeq A_0A_0^\top$ for all time; the $A$-factor never shrinks below its initialisation in the Loewner order.
$AA^\top(t)$ need not be monotone in $t$; only this floor is guaranteed. Discrete steps break the bound by $O(\eta^2)$ per
step. Under Adam it is not guaranteed and fails empirically. With per-time decay $\lambda$, only
$AA^\top\succeq e^{-2\lambda t}A_0A_0^\top$ holds.
\end{restate}

\begin{proof}
(a) Let $x=(B,A)$ have $\rk B=\rk A=r$, put $f(B,A)=\lVert B\rVert^2+\lVert A\rVert^2$, and let
$\mathcal O=\GL_r\cdot x$. Every factorisation $B'A'$ of $BA$ with inner dimension $r$ lies in $\mathcal O$: from
$\operatorname{col}B'=\operatorname{col}(BA)=\operatorname{col}B$ we get $B'=Bg^{-1}$ for a unique $g\in\GL_r$, and then
$Bg^{-1}A'=BA$ gives $A'=gA$ because $B$ is injective. As in Kempf--Ness theory (\bgref{D.30}, \auxref{B.26}), $x$ is a critical point
of $f$ on $\mathcal O$ when $e\in\GL_r$ is a critical point of $g\mapsto f(g\cdot x)$. With $e^{t\xi}\cdot x=(Be^{-t\xi},e^{t\xi}A)$,
\[\tfrac{d}{dt}\Big|_{t=0}f(e^{t\xi}\cdot x)=-2\langle B,B\xi\rangle+2\langle A,\xi A\rangle=-2\tr\big(\xi(B^\top B-AA^\top)\big)=-2\tr(\xi_s\mu),\]
where $\xi_s$ is the symmetric part of $\xi$ and the antisymmetric part drops out because $\mu$ is symmetric. This vanishes
for all $\xi$ iff $\mu=0$ (take $\xi=\mu$). So the critical points are exactly the balanced points of $\mathcal O$.

\emph{Convexity along symmetric directions.} The function $f$ is $\Ort(r)$-invariant, $f(BO^\top,OA)=f(B,A)$, and $\mu$
is $\Ort(r)$-equivariant, $\mu(O\cdot x)=O\mu(x)O^\top$. By the polar decomposition every $g\in\GL_r$ is $g=Oe^\xi$ with
$O\in\Ort(r)$ and $\xi$ symmetric (\bgref{D.27}, \auxref{B.19}(a); \citealp[\S7.3]{xhorn2013matrix}), so every point of $\mathcal O$ is
$O\cdot(e^\xi\cdot x)$ and $f(g\cdot x)=f(e^\xi\cdot x)$. For symmetric $\xi=R\diag(d)R^\top$ with $R$ orthogonal, put
$h(t)=f(e^{t\xi}\cdot x)$. Orthogonal invariance of the Frobenius norm gives
\[h(t)=\lVert BRe^{-t\diag(d)}\rVert^2+\lVert e^{t\diag(d)}R^\top A\rVert^2=\sum_i\Big(\lVert(BR)_{:,i}\rVert^2e^{-2td_i}+\lVert(R^\top A)_{i,:}\rVert^2e^{2td_i}\Big),\]
\[h''(t)=4\sum_id_i^2\Big(\lVert(BR)_{:,i}\rVert^2e^{-2td_i}+\lVert(R^\top A)_{i,:}\rVert^2e^{2td_i}\Big)\ge0 .\]
So $h$ is convex.

\emph{Balanced points are the minima.} Let $x$ be balanced and $y=O\cdot(e^\xi\cdot x)\in\mathcal O$. Then
$h'(0)=-2\tr(\xi\mu(x))=0$, and convexity gives $f(y)=h(1)\ge h(0)=f(x)$. So every balanced point is a global minimum on
$\mathcal O$; conversely every minimum is a critical point, hence balanced.

\emph{One $\Ort(r)$-orbit.} Let $x$ and $y=O\cdot(e^\xi\cdot x)$ both be balanced. Then $e^\xi\cdot x=O^{-1}\cdot y$ is
balanced by equivariance of $\mu$, so $h'(0)=0$ and $h'(1)=-2\tr\big(\xi\mu(e^\xi\cdot x)\big)=0$. Since $h'$ is
non-decreasing, $h'\equiv0$ on $[0,1]$, so $h''(0)=0$. Every column $(BR)_{:,i}$ is nonzero because $BR$ has full column
rank, so all $d_i=0$, $\xi=0$ and $y=O\cdot x$. Conversely the $\Ort(r)$-orbit of a balanced point consists of balanced
points.

\emph{Existence and value.} Let $BA=U\Sigma V^\top$ be a compact SVD (\auxref{B.11}(a)) with $\Sigma\in\R^{r\times r}$ positive diagonal. The
factorisation $(U\Sigma^{1/2},\Sigma^{1/2}V^\top)$ lies in $\mathcal O$ by the first paragraph, is balanced since both
Gram matrices equal $\Sigma$, and has $f=2\tr\Sigma=2\lVert BA\rVert_*$. This is the minimum value.

(b) Under these hypotheses the update has the form of \thmref{Theorem}{III.5}, so $\Phi$ is conserved. With $B_0=0$,
$\Phi_0=-A_0A_0^\top/\eta_A$, and $B^\top B/\eta_B-AA^\top/\eta_A=-A_0A_0^\top/\eta_A$ rearranges to
$B^\top B=\tfrac{\eta_B}{\eta_A}(AA^\top-A_0A_0^\top)$. Since $B^\top B\succeq0$, $AA^\top\succeq A_0A_0^\top$.

\emph{No monotonicity.} Take $m=n=r=1$, $a(0)=a_0>0$, $b(0)=0$ and the signal $\gamma=-1$ on $[0,1)$ and $\gamma=+1$ on
$[1,2]$, the gradients of the minibatch losses $-w$ and $w$. The flow is $\dot b=-\eta_Bs\gamma a$, $\dot a=-\eta_As\gamma b$.
On $[0,1]$, $a(t)=a_0\cosh(s\sqrt{\eta_A\eta_B}\,t)$ increases. On $[1,2]$ the system is the time reversal of the first one,
so it retraces its path and returns to $(a_0,0)$ at $t=2$, and $a$ decreases. The floor $a^2\ge a_0^2$ holds throughout,
but $a^2$ is not monotone.

\emph{Discrete steps.} By the discrete formula of \S\ref{proof:III.5}, after $k$ GD steps (with $s$ absorbed into $G_j$)
\begin{align*}
A_kA_k^\top&=\eta_A\Big(\tfrac{1}{\eta_B}B_k^\top B_k-\Phi_k\Big)\succeq-\eta_A\Phi_k
=A_0A_0^\top-\sum_{j<k}\big(\eta_A\eta_BA_jG_j^\top G_jA_j^\top-\eta_A^2B_j^\top G_jG_j^\top B_j\big)\\
&\succeq A_0A_0^\top-\eta_A\eta_B\sum_{j<k}A_jG_j^\top G_jA_j^\top,
\end{align*}
so the floor holds up to an error of order $\eta^2$ per step. The identity itself breaks at that order already at the first
step, where $B_1=-\eta_BG_0A_0^\top$ and $A_1=A_0$ give $B_1^\top B_1=\eta_B^2A_0G_0^\top G_0A_0^\top$ against
$\tfrac{\eta_B}{\eta_A}(A_1A_1^\top-A_0A_0^\top)=0$.

\emph{Adam.} Take $m=n=r=1$, $s=1$, $(b_0,a_0)=(0,1)$, minibatch gradients $G_1=1$ and $G_2=-1$, any $\varepsilon\ge0$
and bias correction. At step 1, $\nabla_a=b_0G_1=0$, so $a_1=1$, and $b_1=-\eta_B\phi(G_1a_0)$ has sign $-1$. At step 2,
$\nabla_a=b_1G_2>0$ is the first nonzero gradient of $a$, so Adam's step on $a$ is
$-\eta_A\hat m/(\sqrt{\hat v}+\varepsilon)$ with $\hat m=\nabla_a/(1+\beta_1)>0$; at $\varepsilon=0$ it equals
$-\eta_A\sqrt{1+\beta_2}/(1+\beta_1)$. So $a_2<1=a_0$ for every $\varepsilon\ge0$, and at $\varepsilon=0$ the violation is of first order in $\eta_A$.

\emph{Per-time decay.} By \thmref{Corollary}{III.6}(b), $\Phi(t)=e^{-2\lambda t}\Phi(0)=-e^{-2\lambda t}A_0A_0^\top/\eta_A$,
so $AA^\top=e^{-2\lambda t}A_0A_0^\top+\tfrac{\eta_A}{\eta_B}B^\top B\succeq e^{-2\lambda t}A_0A_0^\top$.
\end{proof}

\begin{remark*}{nuclear-norm penalties}
By (a), $\min\{\lambda(\lVert B\rVert^2+\lVert A\rVert^2):sBA=\Delta W\}=2\lambda\lVert\Delta W\rVert_*/s$ for $\Delta W$ of
rank $r$; \citet{xsrebro2004maximum} show (\auxref{B.21}) that $\min\{\tfrac12(\lVert B\rVert^2+\lVert A\rVert^2):BA=Z\}=\lVert Z\rVert_*$
for every inner dimension at least $\rk Z$, which covers rank-deficient $\Delta W$. So the explicit penalty
$\lambda(\lVert B\rVert^2+\lVert A\rVert^2)$ matches a $(2\lambda/s)$-nuclear-norm penalty at its minimisers, and coupled
decay with coefficient $\lambda$, the flow of $L+\tfrac\lambda2(\lVert B\rVert^2+\lVert A\rVert^2)$, matches a
$(\lambda/s)$-nuclear-norm penalty.
\end{remark*}

\subsection{Proposition III.8}\label{proof:III.8}

\begin{restate}{Proposition}{III.8}{the balance--escape trilemma}
No LoRA initialisation is simultaneously
\begin{enumerate}[label=(\alph*)]
\item identity-at-init without a residual split ($B_0A_0=0$),
\item balanced ($B_0^\top B_0=cA_0A_0^\top$ for some $c>0$: unweighted, $c=1$, or rate-weighted, $\Phi_0=0$ with
$c=\eta_B/\eta_A$), and
\item non-stationary, meaning that $D\rho$ at $(B_0,A_0)$ is not the zero map; equivalently, $\dot W(0)\ne0$ for some
parameter velocity, under any first-order optimizer.
\end{enumerate}
Moreover, if $B_0A_0=0$ then $\langle B_0^\top B_0,A_0A_0^\top\rangle_F=0$, so
\[\lVert B_0^\top B_0-cA_0A_0^\top\rVert_F^2=\lVert B_0^\top B_0\rVert_F^2+c^2\lVert A_0A_0^\top\rVert_F^2,\]
and zero-product initialisations are maximally unbalanced.
\end{restate}

\begin{proof}
Assume (a) and (b). Then $0=B_0^\top(B_0A_0)=(B_0^\top B_0)A_0=cA_0A_0^\top A_0$, and multiplying by $A_0^\top$ on the left
gives $c(A_0^\top A_0)^2=0$. The matrix $A_0^\top A_0$ is symmetric positive semidefinite with vanishing square, so its
eigenvalues vanish and $A_0^\top A_0=0$, that is, $A_0=0$. Then $B_0^\top B_0=cA_0A_0^\top=0$, so $B_0=0$. At $(0,0)$,
$D\rho(\dot B,\dot A)=s(\dot B\cdot0+0\cdot\dot A)=0$, which contradicts (c). The equivalence in (c) is the definition of
the zero map: some parameter velocity has $\dot W=D\rho(\dot B,\dot A)\ne0$ iff $D\rho\ne0$. At $(0,0)$ also
$\nabla_B=sGA^\top=0$ and $\nabla_A=sB^\top G=0$, so every update rule built from these gradients and the current
parameters, including Adam, momentum and weight decay started from zero state, stays at $(0,0)$.

For the identity, $\langle B_0^\top B_0,A_0A_0^\top\rangle_F=\tr(B_0^\top B_0A_0A_0^\top)=\tr\big(B_0^\top(B_0A_0)A_0^\top\big)=0$,
and expanding the square gives the displayed formula. For positive semidefinite $P$ and $N$,
$\langle P,N\rangle_F=\tr(P^{1/2}NP^{1/2})\ge0$, so $\lVert P-cN\rVert_F^2\le\lVert P\rVert_F^2+c^2\lVert N\rVert_F^2$, with
equality iff $\langle P,N\rangle_F=0$. Zero-product initialisations attain this maximum for the given norms of the two
Gram matrices.
\end{proof}

\subsection{Theorem III.9}\label{proof:III.9}

\begin{restate}{Theorem}{III.9}{closed-form isotropic-charge LoRA dynamics; known, \citet{xtarmoun2021understanding} and \citet{xmin2021explicit}}
Let $\rho(B,A)=W_{\mathrm{res}}+sBA$ be trained by gradient flow with rates $\eta_A,\eta_B$, and suppose the charge is
isotropic, $\Phi=-\kappa I_r$. Put $\mu=s\sqrt{\eta_A\eta_B}$ (a rate, unrelated to the moment map $\mu$ of
\thmref{Proposition}{III.7}), and on the locus where $Z=W-W_{\mathrm{res}}$ has rank $r$ write its compact SVD
$Z=U\Sigma V^\top$. Then
\[\dot W=-\mu^2\big[G\,V g_\kappa(\Sigma^2/\mu^2)V^\top+U f_\kappa(\Sigma^2/\mu^2)U^\top G\big],\qquad
f_\kappa(x)=\sqrt{x+\kappa^2/4}-\kappa/2,\quad g_\kappa=f_\kappa+\kappa .\]
The identity holds pointwise on $\{\Phi=-\kappa I,\ \rk Z=r\}$; gradient flow is used only to keep $\Phi$ fixed. The two
terms under the root are equal at the \emph{crossover scale} $\sigma^*=\mu\lvert\kappa\rvert/2$.
\begin{itemize}
\item For $\sigma\ll\sigma^*$ with $\kappa>0$, $f\to0$ and $g\to\kappa$, so $\dot W\approx-\mu^2\kappa\,GP_V$. These are
LoRA-FA dynamics. The contribution of $A$ to $\dot W$ is $O(\sigma^2/(\mu^2\kappa))$, although $A$ itself still drifts
slowly. For $\kappa<0$ the roles swap, $\dot W\approx-\mu^2\lvert\kappa\rvert P_UG$, and only $A$ effectively moves
(Init[B]).
\item For $\sigma\gg\sigma^*$, $f\approx g\approx\sigma/\mu$, so $\dot W\approx-\mu\big[G(Z^\top Z)^{1/2}+(ZZ^\top)^{1/2}G\big]$.
These are the balanced dynamics of Arora, Cohen and Hazan.
\item \emph{Off the rank-$r$ locus.} Write $\dot W=-\mu^2(GS+TG)$ with $S=A^\top A/\eta_A$ and $T=BB^\top/\eta_B$, so that
on the rank-$r$ locus $S=Vg_\kappa(\Sigma^2/\mu^2)V^\top$ and $T=Uf_\kappa(\Sigma^2/\mu^2)U^\top$. For $\kappa\ge0$,
$T=f_\kappa(ZZ^\top/\mu^2)$ still holds as a full matrix function, but $S$ gains the term
$\kappa(P_{\operatorname{row}A}-P_V)$; symmetrically, for $\kappa<0$, $T$ gains $\lvert\kappa\rvert(P_{\operatorname{col}B}-P_U)$.
At the default-LoRA start $Z=0$ ($B_0=0$) this leaves $S=\kappa P_{\operatorname{row}A_0}$ and $T=0$, so
$\dot W=-\mu^2\kappa\,GP_{\operatorname{row}A_0}$, and the $\sigma\ll\sigma^*$ regime extends continuously to $t=0$.
\end{itemize}
\end{restate}

\begin{proof}
Let $G$ be the signal driving the flow, $\dot B=-\eta_BsGA^\top$ and $\dot A=-\eta_AsB^\top G$; under gradient flow
$G=\nabla_\theta L$, and \thmref{Theorem}{III.5}(i) keeps $\Phi=-\kappa I$ for all time if it holds at $t=0$. Everything
below uses only the value of $\Phi$ at the point considered.

\emph{Normalisation.} Put $\beta=B/\sqrt{\eta_B}$ and $\alpha=A/\sqrt{\eta_A}$. Then
$\beta^\top\beta-\alpha\alpha^\top=\Phi=-\kappa I$ and $\beta\alpha=BA/\sqrt{\eta_A\eta_B}=Z/\mu$. Let
$T=\beta\beta^\top=BB^\top/\eta_B$ and $S=\alpha^\top\alpha=A^\top A/\eta_A$. Then
\[\dot W=s(\dot BA+B\dot A)=-s^2\big(\eta_BGA^\top A+\eta_ABB^\top G\big)=-s^2\eta_A\eta_B(GS+TG)=-\mu^2(GS+TG).\]

\emph{Two quadratic identities.} Using $\alpha\alpha^\top=\beta^\top\beta+\kappa I$ and $\beta^\top\beta=\alpha\alpha^\top-\kappa I$,
\[\tfrac{ZZ^\top}{\mu^2}=\beta\alpha\alpha^\top\beta^\top=\beta(\beta^\top\beta+\kappa I)\beta^\top=T^2+\kappa T,\qquad
\tfrac{Z^\top Z}{\mu^2}=\alpha^\top\beta^\top\beta\alpha=\alpha^\top(\alpha\alpha^\top-\kappa I)\alpha=S^2-\kappa S .\]
These hold at every point with $\Phi=-\kappa I$.

\emph{Spectra on the rank-$r$ locus.} If $\rk Z=r$, then $\beta$ and $\alpha^\top$ have full column rank, so
$\beta^\top\beta\succ0$ and $\alpha\alpha^\top\succ0$. The nonzero eigenvalues of $T=\beta\beta^\top$ are the eigenvalues of
$\beta^\top\beta=\alpha\alpha^\top-\kappa I$ (\auxref{B.36}(c)), which are positive and exceed $-\kappa$; so they exceed $\max(0,-\kappa)$.
Likewise the nonzero eigenvalues of $S$ are those of $\alpha\alpha^\top=\beta^\top\beta+\kappa I$, which exceed $\max(0,\kappa)$.
The range of $T$ is $\operatorname{col}\beta=\operatorname{col}Z=U$, and that of $S$ is $\operatorname{row}\alpha=\operatorname{row}Z=V$.

\emph{Inversion.} The polynomial $p(t)=t^2+\kappa t$ is strictly increasing on $[-\kappa/2,\infty)$, which contains the
nonzero eigenvalues of $T$ because $\max(0,-\kappa)\ge-\kappa/2$, and $p(t)>0$ there while $p(0)=0$. So $p$ is injective
on the spectrum of $T$, the eigenspaces of $T$ are those of $p(T)=ZZ^\top/\mu^2=U(\Sigma^2/\mu^2)U^\top$ (by the compact SVD of $Z$,
\auxref{B.11}(a)), and on the
eigenvector $u_i$ with $ZZ^\top u_i=\sigma_i^2u_i$ the eigenvalue of $T$ is the unique root $t>\max(0,-\kappa)$ of
$t^2+\kappa t=\sigma_i^2/\mu^2$, namely $f_\kappa(\sigma_i^2/\mu^2)$. On $U^\perp$ both $T$ and $ZZ^\top$ vanish. Hence
$T=Uf_\kappa(\Sigma^2/\mu^2)U^\top$. In the same way $q(s)=s^2-\kappa s$ is strictly increasing on $[\kappa/2,\infty)$,
the root above $\max(0,\kappa)$ of $s^2-\kappa s=x$ is $\kappa/2+\sqrt{x+\kappa^2/4}=g_\kappa(x)$, and
$S=Vg_\kappa(\Sigma^2/\mu^2)V^\top$. The compact SVD matters: $f_\kappa(0)=\lvert\kappa\rvert$ for $\kappa<0$ and
$g_\kappa(0)=\kappa$ for $\kappa>0$, so the full matrix functions $f_\kappa(ZZ^\top/\mu^2)$ and
$g_\kappa(Z^\top Z/\mu^2)$ would put nonzero values on $\ker Z^\top$ and $\ker Z$, where $T$ and $S$ vanish. Substituting
into $\dot W=-\mu^2(GS+TG)$ gives the formula. The crossover: $x=\sigma^2/\mu^2$ equals $\kappa^2/4$ iff
$\sigma=\mu\lvert\kappa\rvert/2$.

\emph{The regimes.} For $\kappa>0$, $f_\kappa(x)=x/\big(\sqrt{x+\kappa^2/4}+\kappa/2\big)\in[0,x/\kappa]$ and
$g_\kappa(x)-\kappa=f_\kappa(x)$. So $\dot W=-\mu^2\kappa\,GP_V+E$ with $E=-\mu^2\big(GVf_\kappa(\Sigma^2/\mu^2)V^\top+TG\big)$
and $\lVert E\rVert\le2(\sigma_1^2/\kappa)\lVert G\rVert$, where $\sigma_1$ is the largest singular value of $Z$. The
contribution $-\mu^2TG=s\,B\dot A$ of $A$ has gain at most $f_\kappa(\sigma_1^2/\mu^2)\le\sigma_1^2/(\mu^2\kappa)$, against
the gain $g_\kappa\ge\kappa$ of the $B$-term. The factor $A$ still moves, at speed
$\lVert\dot A\rVert\le\eta_As\lVert B\rVert_2\lVert G\rVert$ with $\lVert B\rVert_2^2=\eta_B\lVert T\rVert_2\le\eta_B\sigma_1^2/(\mu^2\kappa)$.
For $\kappa<0$ the same bounds hold with the roles exchanged: $g_\kappa(x)\in[0,x/\lvert\kappa\rvert]$ and
$f_\kappa=g_\kappa+\lvert\kappa\rvert$, so $\dot W=-\mu^2\lvert\kappa\rvert P_UG+O(\sigma_1^2/\lvert\kappa\rvert)\lVert G\rVert$.
For large $x$, $0\le\sqrt{x+\kappa^2/4}-\sqrt x\le\lvert\kappa\rvert/2$ gives $\lvert f_\kappa(x)-\sqrt x\rvert\le\lvert\kappa\rvert$
and $\lvert g_\kappa(x)-\sqrt x\rvert\le\lvert\kappa\rvert$. At $x=\sigma_i^2/\mu^2$ the relative error is at most
$\lvert\kappa\rvert\mu/\sigma_i=2\sigma^*/\sigma_i$, so for $\sigma_i\gg\sigma^*$, $f\approx g\approx\sigma_i/\mu$ and
\[\dot W\approx-\mu^2\big[GV(\Sigma/\mu)V^\top+U(\Sigma/\mu)U^\top G\big]=-\mu\big[G(Z^\top Z)^{1/2}+(ZZ^\top)^{1/2}G\big].\]
This is the end-to-end dynamics of a balanced depth-2 linear network with rate $\mu$ \citep[Thm.~1 with
$N=2$]{arora2018optimization}; at $\kappa=0$ it is exact, since $f_0(x)=g_0(x)=\sqrt x$.

\emph{Off the rank-$r$ locus.} Let $\kappa>0$ and $\rk Z<r$. The nonzero eigenvalues of $T$ are those of
$\beta^\top\beta\succeq0$, and on them $p(t)=t^2+\kappa t>0$; if $ZZ^\top v=0$ then $T(T+\kappa I)v=0$ with $T+\kappa I\succ0$,
so $Tv=0$. Hence $\ker T=\ker ZZ^\top$ and the inversion above gives $T=f_\kappa(ZZ^\top/\mu^2)$ as a full matrix function,
using $f_\kappa(0)=0$ for $\kappa\ge0$. For $S$, decompose $\operatorname{row}A=\operatorname{row}Z\oplus(\operatorname{row}A\cap\ker Z)$
orthogonally, which is possible since $\operatorname{row}Z\subseteq\operatorname{row}A$ and $(\operatorname{row}Z)^\perp=\ker Z$.
On $(\operatorname{row}A)^\perp=\ker\alpha$, $S=0$. If $v=\alpha^\top w\in\operatorname{row}A$ has $Zv=0$, then
$0=\beta\alpha\alpha^\top w=\beta(\beta^\top\beta+\kappa I)w$, so $\beta^\top\beta(\beta^\top\beta+\kappa I)w=0$; as
$\beta^\top\beta+\kappa I\succ0$ commutes with $\beta^\top\beta\succeq0$, this forces $\beta^\top\beta w=0$, hence $\beta w=0$,
and $Sv=\alpha^\top(\alpha\alpha^\top)w=\alpha^\top(\beta^\top\beta+\kappa I)w=\kappa v$. On $\operatorname{row}Z$ the
eigenvalues of $S$ exceed $\kappa$ and the inversion gives $Vg_\kappa(\Sigma^2/\mu^2)V^\top$. Together,
$S=Vg_\kappa(\Sigma^2/\mu^2)V^\top+\kappa(P_{\operatorname{row}A}-P_V)$. The case $\kappa<0$ is symmetric under
$(B,A,Z,\eta_B,\eta_A,\kappa)\mapsto(A^\top,B^\top,Z^\top,\eta_A,\eta_B,-\kappa)$, which exchanges $S$ and $T$. At $\kappa=0$ the correction terms vanish. At $Z=0$ with $B_0=0$
and $A_0A_0^\top=cI$, $\kappa=c/\eta_A>0$, $T=0$ and $S=\kappa P_{\operatorname{row}A_0}$, so
$\dot W=-\mu^2\kappa\,GP_{\operatorname{row}A_0}=-s^2\eta_BcGP_{\operatorname{row}A_0}$, which agrees with the direct value
$-s^2\eta_BGA_0^\top A_0$ because $A_0^\top A_0=cP_{\operatorname{row}A_0}$. As $Z\to0$ along the flow with $V\to\operatorname{row}A_0$,
the rank-$r$ formula tends to the same value, since $g_\kappa(0)=\kappa$ and $f_\kappa(0)=0$.
\end{proof}

\subsection{Theorem III.10}\label{proof:III.10}

\begin{restate}{Theorem}{III.10}{the hyperparameter torus; the LoRA/Adam case is known, \citet{schulman2025lora}}
Let $\rho(q)=\theta_b+s\,\delta(q_1,\dots,q_k)$ with $\delta$ multihomogeneous of multidegree $(d_1,\dots,d_k)$. Train block
$i$ from $q_i(0)=\sigma_iu_i$ with Adam ($\beta_1\in[0,1)$, $\beta_2\in(0,1)$, bias correction allowed) at learning rate
$\eta_i(t)$, on the loss $L\circ\rho$ with no $q$-dependent regulariser. Assume no gradient clipping (or per-block
clipping with thresholds rescaled as below), common random numbers (the same data order and dropout masks in both runs),
and exact arithmetic (the result is bit-exact in floating point when every $c_i$ is a power of $2$). Then the trajectory
$t\mapsto\rho(q(t))$ is invariant under the torus action of $c\in\R_{>0}^k$:
\[\big(s,\ \eta_i(t),\ \sigma_i,\ \varepsilon_i,\ \lambda_i,\ C_i\big)\mapsto\Big(s\prod_ic_i^{-d_i},\ c_i\eta_i(t),\ c_i\sigma_i,\ \varepsilon_i/c_i,\ \lambda_i/c_i,\ C_i/c_i\Big).\]
Here $\varepsilon_i$ is a per-block Adam $\varepsilon$ (with the convention $0/0=0$ at $\varepsilon=0$), $\lambda_i$ a
per-block decoupled (AdamW) weight decay, and $C_i$ a per-block clipping threshold. Under SGD, with or without heavy-ball
or Nesterov momentum, the same holds with $\eta_i(t)\mapsto c_i^2\eta_i(t)$ and coupled $L^2$ decay
$\lambda_i\mapsto\lambda_i/c_i^2$. In particular, any learning-rate schedule shared by all blocks, warm-up included, is
preserved.
\end{restate}

\begin{proof}
\emph{The rescaling is an isomorphism.} Write $c\odot q=(c_1q_1,\dots,c_kq_k)$ and $\rho'=\theta_b+s'\delta$ with
$s'=s\prod_ic_i^{-d_i}$. Multihomogeneity (\bgref{D.19}) gives
$\rho'(c\odot q)=\theta_b+s\prod_ic_i^{-d_i}\prod_ic_i^{d_i}\delta(q)=\rho(q)$ for all $q$, so $q\mapsto c\odot q$ is a
diffeomorphism over $\Theta$ that maps the base point $q(0)$ to $q'(0)=c\odot q(0)$; it is an isomorphism in $\Meth$. Let
$\ell_t=L_t\circ\rho$ and $\ell'_t=L_t\circ\rho'$ be the step-$t$ losses, with the same minibatch and dropout masks in
both runs and no $q$-dependent regulariser. Differentiating $\ell'_t(c\odot q)=\ell_t(q)$ gives
\begin{equation}\label{eq:C2-torus-grad}
\nabla_{q_i}\ell'_t(c\odot q)=c_i^{-1}\nabla_{q_i}\ell_t(q).
\end{equation}

\emph{The optimizer.} For block $i$ at step $t\ge1$, Adam(W) computes the gradient $g_t$ of $\ell_t$ at $q_{t-1}$, clips it
per block, $g_t\leftarrow g_t\min(1,C_i/\lVert g_t\rVert)$ (with factor $1$ when $g_t=0$, and no clipping step when there
is no clipping), and sets
\begin{gather*}
m_t=\beta_1m_{t-1}+(1-\beta_1)g_t,\qquad v_t=\beta_2v_{t-1}+(1-\beta_2)\,g_t\odot g_t,\\
q_t=q_{t-1}-\eta_i(t)\Big(\frac{m_t/b_{1,t}}{\sqrt{v_t/b_{2,t}}+\varepsilon_i}+\lambda_iq_{t-1}\Big),
\end{gather*}
with $m_0=v_0=0$, entrywise square root and division, $b_{j,t}=1-\beta_j^t$ with bias correction and $b_{j,t}=1$ without,
and $0/0=0$. PyTorch's AdamW first multiplies $q_{t-1}$ by $1-\eta_i(t)\lambda_i$ and then takes the Adam step, which is the
same update.

\emph{Induction.} Run the original hyperparameters from $q(0)$ and the transformed ones from $q'(0)=c\odot q(0)$, which is
what $\sigma_i\mapsto c_i\sigma_i$ means. Suppose that after step $t-1$, in every block,
\[q'_{t-1}=c_iq_{t-1},\qquad m'_{t-1}=c_i^{-1}m_{t-1},\qquad v'_{t-1}=c_i^{-2}v_{t-1},\]
which holds at $t=1$. By \eqref{eq:C2-torus-grad} the raw gradients satisfy $g'_t=c_i^{-1}g_t$. With $C'_i=C_i/c_i$ the
clipping factors agree, $\min(1,C'_i/\lVert g'_t\rVert)=\min(1,C_i/\lVert g_t\rVert)$, so the clipped gradients also satisfy
$g'_t=c_i^{-1}g_t$. The moment recursions are linear in $g_t$ and $g_t\odot g_t$, so $m'_t=c_i^{-1}m_t$ and
$v'_t=c_i^{-2}v_t$. With $\varepsilon'_i=\varepsilon_i/c_i$ the normalised step is unchanged,
\[\frac{m'_t/b_{1,t}}{\sqrt{v'_t/b_{2,t}}+\varepsilon'_i}=\frac{c_i^{-1}m_t/b_{1,t}}{c_i^{-1}\big(\sqrt{v_t/b_{2,t}}+\varepsilon_i\big)}=\frac{m_t/b_{1,t}}{\sqrt{v_t/b_{2,t}}+\varepsilon_i},\]
including the entries where $v_t=0$, at which $m_t=0$ as well and both sides are $0$. The decay term transforms as
$\lambda'_iq'_{t-1}=(\lambda_i/c_i)c_iq_{t-1}=\lambda_iq_{t-1}$. With $\eta'_i(t)=c_i\eta_i(t)$ the whole step is multiplied
by $c_i$, so $q'_t=c_iq_t$, which closes the induction. Then $\rho'(q'_t)=\rho(q_t)$ for all $t$. The value of $\eta_i(t)$
at each step was arbitrary, so any schedule works; a schedule shared by all blocks, $\eta_i(t)=\eta_i\varphi(t)$, becomes
$c_i\eta_i\varphi(t)$, the same schedule.

\emph{SGD.} With coupled $L^2$ decay the step-$t$ direction is $g_t+\lambda_iq_{t-1}$, the heavy-ball buffer is
$b_t=\mu_mb_{t-1}+(1-\tau)(g_t+\lambda_iq_{t-1})$ (with momentum $\mu_m$ and dampening $\tau$), the direction is $b_t$, or
$g_t+\lambda_iq_{t-1}+\mu_mb_t$ for Nesterov, and $q_t=q_{t-1}-\eta_i(t)\cdot(\text{direction})$. Suppose $q'_{t-1}=c_iq_{t-1}$
and $b'_{t-1}=c_i^{-1}b_{t-1}$. Then $g'_t=c_i^{-1}g_t$ and, with $\lambda'_i=\lambda_i/c_i^2$,
$\lambda'_iq'_{t-1}=c_i^{-1}\lambda_iq_{t-1}$; every quantity entering the direction is multiplied by $c_i^{-1}$, so
$b'_t=c_i^{-1}b_t$ and the direction is multiplied by $c_i^{-1}$. With $\eta'_i(t)=c_i^2\eta_i(t)$ the step is multiplied by
$c_i$, and $q'_t=c_iq_t$. Per-block clipping with $C_i/c_i$ is handled as for Adam.

\emph{Floating point.} If every $c_i$ is a power of $2$, multiplication by $c_i^{\pm1}$ and by $\prod_ic_i^{\mp d_i}$ is
exact, and it commutes with every correctly rounded addition, multiplication, division and square root, barring overflow
and underflow (\auxref{B.39}). Then the forward pass computes bit-identical weights and outputs in both
runs, the backward pass computes gradients that are exactly $c_i^{-1}$ times the original ones, and every quantity in the
induction above holds exactly in floating point, so the two trajectories agree bit for bit.
\end{proof}

\subsection{Corollary III.11}\label{proof:III.11}

\begin{restate}{Corollary}{III.11}{LoRA+ is $\alpha$ in disguise; rsLoRA is LoRA; (a) and the LoRA+ reading are known, \citet{schulman2025lora}}
For linear LoRA modules ($k=2$, $d=(1,1)$, $\sigma_B=0$), under the hypotheses of \thmref{Theorem}{III.10}:

(a) \emph{Effective hyperparameters.} Under Adam the torus invariants of $(s,\eta_A,\eta_B,\sigma_A)$ are
$(s\eta_A\eta_B,\ \sigma_A/\eta_A)$; the quantities $\alpha\cdot\mathrm{init}_A\cdot\mathrm{LR}_B$ and
$\mathrm{init}_A/\mathrm{LR}_A$ of Schulman et al.\ are equivalent generators. Under SGD the invariants are
$(s^2\eta_A\eta_B,\ \sigma_A^2/\eta_A)$.

(b) \emph{LoRA+.} Under Adam with $\varepsilon=0$, no weight decay, no gradient clipping and a learning-rate schedule
shared by both groups, LoRA+ with ratio $\lambda$ at scale $s$ produces \emph{exactly} the trajectory of plain LoRA at
scale $\lambda s$ (same $\eta_A$, same $A_0$). For $\varepsilon>0$ the equivalence needs a per-block
$\varepsilon_B'=\lambda\varepsilon$, and with AdamW a per-group decay $\lambda\cdot\mathrm{wd}$ on $B$. Under SGD it holds at
scale $\sqrt\lambda\,s$, or equivalently with plain LoRA at the single rate $\sqrt{\eta_A\eta_B}$ and
$A_0\mapsto\lambda^{1/4}A_0$.

(c) \emph{rsLoRA.} $\rho_{\mathrm{rs}}(B,A)=\rho_{\mathrm{LoRA}}(\sqrt r\,B,A)$ is an isomorphism of objects. Up to the
hyperparameter torus, rsLoRA can be represented as LoRA at the same $\alpha$ with $\eta_B$ multiplied by $\sqrt r$ under
Adam (by $r$ under SGD), or equally as LoRA with $\alpha$ multiplied by $\sqrt r$ at the same rates. The reading ``same
object, different metric'' is a choice of representative within the orbit. Part (c) concerns a fixed rank; it does not
address rsLoRA's rank-stability prescription, how $\alpha$ should scale with $r$.
\end{restate}

\begin{proof}
Here $\delta(B,A)=BA$ has multidegree $(1,1)$, and $\sigma_B=0$ means $B_0=0$, which every $c_B$ leaves unchanged.

(a) Under Adam, $(c_A,c_B)$ acts by $s\mapsto s/(c_Ac_B)$, $\eta_A\mapsto c_A\eta_A$, $\eta_B\mapsto c_B\eta_B$ and
$\sigma_A\mapsto c_A\sigma_A$. In logarithms this is translation by the column space of
\[a=\begin{pmatrix}-1&-1\\1&0\\0&1\\1&0\end{pmatrix}\qquad(\text{rows } \log s,\log\eta_A,\log\eta_B,\log\sigma_A),\]
which has rank $2$, so the orbit space has dimension $2$ and a basis of the vectors $b$ with $b^\top a=0$ gives a complete
set of invariant monomials (\bgref{D.20}). The vectors $(1,1,1,0)$ and $(0,-1,0,1)$ satisfy $b^\top a=0$ and are independent;
they give $s\eta_A\eta_B$ and $\sigma_A/\eta_A$. Since $\alpha\propto s$ at fixed $r$, Schulman et al.'s quantities are
$s\sigma_A\eta_B=(s\eta_A\eta_B)(\sigma_A/\eta_A)$ and $\sigma_A/\eta_A$, with exponent vectors $(1,0,1,1)$ and $(0,-1,0,1)$,
another basis of the same lattice of invariant monomials. Under SGD the rates scale by $c_A^2$ and $c_B^2$; the weight
matrix has rows $(-1,-1)$, $(2,0)$, $(0,2)$, $(1,0)$, again of rank $2$, and $(2,1,1,0)$ and $(0,-1,0,2)$ give the
invariants $s^2\eta_A\eta_B$ and $\sigma_A^2/\eta_A$.

(b) LoRA+ has $\eta_B=\lambda\eta_A$. Under Adam apply the torus element $c_A=1$, $c_B=1/\lambda$: it maps
$(s,\eta_A,\lambda\eta_A,\sigma_A)$ to $(\lambda s,\eta_A,\eta_A,\sigma_A)$, plain LoRA at scale $\lambda s$ with the same
$\eta_A$ and the same $A_0$; a shared schedule $\varphi(t)$ stays shared. \thmref{Theorem}{III.10} gives
$\varepsilon_B'=\varepsilon/c_B=\lambda\varepsilon$ and $\mathrm{wd}_B'=\mathrm{wd}/c_B=\lambda\cdot\mathrm{wd}$, which are
the stated per-block values, and both vanish when $\varepsilon=\mathrm{wd}=0$. Under SGD, $c_A=1$ and $c_B=\lambda^{-1/2}$
give $\eta_B'=c_B^2\lambda\eta_A=\eta_A$ and $s'=\sqrt\lambda\,s$. The single-rate form uses $c_A=\lambda^{1/4}$ and
$c_B=\lambda^{-1/4}$: then $\eta_A'=\sqrt\lambda\,\eta_A$, $\eta_B'=\lambda^{-1/2}\lambda\eta_A=\sqrt\lambda\,\eta_A$, both
equal to $\sqrt{\eta_A\eta_B}$, $s'=s$, and $\sigma_A'=\lambda^{1/4}\sigma_A$, that is, $A_0\mapsto\lambda^{1/4}A_0$.

(c) $\rho_{\mathrm{rs}}(B,A)=W_0+(\alpha/\sqrt r)BA=W_0+(\alpha/r)(\sqrt rB)A=\rho_{\mathrm{LoRA}}(\sqrt rB,A)$, and
$(B,A)\mapsto(\sqrt rB,A)$ is a pointed diffeomorphism ($0\mapsto0$ in the $B$-factor), hence an isomorphism of objects.
The rsLoRA scale $s_{\mathrm{rs}}=\alpha/\sqrt r$ is the LoRA scale for $\alpha'=\sqrt r\,\alpha$, which is the second
representation. Applying the torus element $c_A=1$, $c_B=\sqrt r$ to rsLoRA gives $s'=s_{\mathrm{rs}}/\sqrt r=\alpha/r$,
the LoRA scale at the same $\alpha$, with $\eta_B'=\sqrt r\,\eta_B$ under Adam and $\eta_B'=r\eta_B$ under SGD, and with
$\varepsilon_B'=\varepsilon/\sqrt r$ and $\mathrm{wd}_B'=\mathrm{wd}/\sqrt r$ when these are nonzero. Both representations
lie on the torus orbit of rsLoRA's hyperparameters, which is what ``up to the hyperparameter torus'' means. The rank $r$
is fixed throughout.
\end{proof}

\subsection{Theorem III.12}\label{proof:III.12}

\begin{restate}{Theorem}{III.12}{the optimizer gauge hierarchy}
On LoRA's full-rank locus, with hyperparameters shared by all rank channels, the largest subgroup of the gauge $\GL_r$
under which an update rule is equivariant, at the level of parameters, is as follows.

(a) SGD and GD with any block rates, with or without heavy-ball momentum: $\Ort(r)$.

(b) Adam and AdamW at fixed learning rates ($\beta_1,\beta_2\in[0,1)$, any $\varepsilon\ge0$, any weight decay): the
signed permutations $B_r$, of order $2^rr!$. A positive diagonal $g=\diag(d)$ is a symmetry only \emph{jointly} with
per-rank-channel hyperparameters, and only without global-norm clipping. At $\varepsilon=0$ and without weight decay it
suffices to rescale the rates, $\eta_{B,j}\mapsto\eta_{B,j}/d_j$ and $\eta_{A,j}\mapsto d_j\eta_{A,j}$. With
$\varepsilon>0$ one must also give each channel its own $\varepsilon_{B,j}=d_j\varepsilon$ and
$\varepsilon_{A,j}=\varepsilon/d_j$. With decoupled decay multiplied by the learning rate (PyTorch AdamW) one must also
set $\lambda_{B,j}=d_j\lambda$ and $\lambda_{A,j}=\lambda/d_j$; under the convention of \citet{xloshchilov2019decoupled},
where decay is scaled only by the schedule multiplier, $\lambda$ needs no change.

(c) Undamped scaled GD ($\delta=0$; the Riemannian preconditioner of Tong, Ma and Chi and of Zhang and Pilanci, with or
without heavy-ball momentum): all of $\GL_r$. The $W$-trajectory from gauge-related initial states (with zero or
correspondingly transformed momentum) is independent of the representative; a step may drop rank, so the iteration maps
into $W_0+\Mr$.
\end{restate}

\begin{proof}
We use the conventions of \S\ref{proof:III.1}: at $g\cdot q$ the gradients are \eqref{eq:C2-gauge-grad}, and equivariance
under $g$ means \eqref{eq:C2-equivariant-step} for all gradients, together with transported optimizer state. The set of
such $g$ is a subgroup. On the full-rank locus $\nabla_B=sGA^\top$ ranges over all of $\R^{m\times r}$ as $G$ ranges over
$\R^{m\times n}$, because $A$ has full row rank; this is the only use of the locus in the inclusions $\subseteq$ below.

(a) An SGD step is $\delta B=-\eta_B\nabla_B$, $\delta A=-\eta_A\nabla_A$. At $g\cdot q$,
$\delta B'=-\eta_B\nabla_Bg^\top$, and \eqref{eq:C2-equivariant-step} requires $\nabla_B(g^\top-g^{-1})=0$ for all
$\nabla_B\in\R^{m\times r}$, that is, $g^\top=g^{-1}$. Conversely, for $g\in\Ort(r)$, $\delta B'=\delta Bg^\top=\delta Bg^{-1}$
and $\delta A'=-\eta_Ag^{-\top}\nabla_A=g\,\delta A$. A heavy-ball or Nesterov buffer is a linear combination of past
gradients, all taken at points related by the same $g$, so for orthogonal $g$ it transforms like the gradients, $b'_B=b_Bg^{-1}$
and $b'_A=gb_A$, and the step is transported. The first step from a zero buffer is a positive multiple of an SGD step (the multiple is $1$ for heavy ball
and $1+\mu_m$ for Nesterov with momentum coefficient $\mu_m$), so the inclusion $\subseteq\Ort(r)$ also holds with momentum.

(b) \emph{The inclusion $\supseteq$.} Let $g$ be a signed permutation, so $g^{-1}=g^\top$ and $\nabla_{B'}=\nabla_Bg^{-1}$
permutes the columns of $\nabla_B$ and flips some signs; likewise $\nabla_{A'}=g\nabla_A$ permutes and flips rows. Adam is
coordinatewise: the first moment $m$ is permuted with the same sign flips, the second moment $v$ is permuted without sign
changes, so the normalised step is transported, and so is a decoupled or coupled decay term, which is linear in the
parameters. Hence the step satisfies \eqref{eq:C2-equivariant-step}. Since $g$ is orthogonal it preserves
$\lVert\nabla_B\rVert$ and $\lVert\nabla_A\rVert$, so global-norm clipping does not affect this inclusion.

\emph{The inclusion $\subseteq$.} Consider the first step from zero state at a full-rank point. With bias correction it is
$\delta B=-\eta\big(\phi(\nabla_B)+\lambda B\big)$ with $\phi(x)=x/(\lvert x\rvert+\varepsilon)$ entrywise and $0/0=0$;
without bias correction $\phi$ is replaced by a positive multiple of $x/(\lvert x\rvert+\varepsilon')$ with a rescaled
$\varepsilon'$, which changes nothing below. At $g\cdot q$ the decay term $\lambda Bg^{-1}$ is already transported, so
equivariance requires $\phi(\nabla_Bg^\top)=\phi(\nabla_B)g^{-1}$, row by row:
\begin{equation}\label{eq:C2-adam-row}
\phi(xg^\top)=\phi(x)h\qquad\text{for every row vector }x\in\R^{1\times r},\qquad h:=g^{-1}.
\end{equation}
(With coupled decay the argument of $\phi$ is $\nabla_B+\lambda B$; letting $\nabla_B$ grow along a fixed direction leads
to the limit \eqref{eq:C2-adam-sign} below in the same way.)

\emph{Case $\varepsilon=0$.} Then $\phi=\operatorname{sign}$ and \eqref{eq:C2-adam-row} reads
\begin{equation}\label{eq:C2-adam-sign}
\operatorname{sign}(xg^\top)=\operatorname{sign}(x)h\qquad\text{for every }x.
\end{equation}
Fix $\mathsf s\in\{\pm1\}^r$. The open orthant $\{x:\operatorname{sign}(x)=\mathsf s\}$ is not contained in the finite union
of the hyperplanes $\{x:(xg^\top)_j=0\}$, which are proper because $g$ is invertible (a finite union of hyperplanes has
empty interior), so it contains an $x$ with $xg^\top$
free of zeros. For that $x$, \eqref{eq:C2-adam-sign} gives $\mathsf sh\in\{\pm1\}^r$. Thus $\mathsf sh\in\{\pm1\}^r$ for
every $\mathsf s$. Flipping $\mathsf s_i$ changes $(\mathsf sh)_j$ by $2\mathsf s_ih_{ij}$, a difference of two elements of
$\{\pm1\}$, so $h_{ij}\in\{0,\pm1\}$. If column $j$ of $h$ had no nonzero entry, $(\mathsf sh)_j=0$; if it had $t\ge2$
nonzero entries, the choice $\mathsf s_i=h_{ij}$ on its support would give $(\mathsf sh)_j=t\ge2$. So every column of $h$
has exactly one entry, equal to $\pm1$, and these entries lie in distinct rows because $h$ is invertible. Hence $h$, and
with it $g=h^{-1}$, is a signed permutation matrix.

\emph{Case $\varepsilon>0$.} For a fixed $x$ and $\tau\to\infty$, $\phi(\tau y)\to\operatorname{sign}(y)$ entrywise for every
$y$, so applying \eqref{eq:C2-adam-row} to $\tau x$ and letting $\tau\to\infty$ gives \eqref{eq:C2-adam-sign}, and the case
$\varepsilon=0$ shows $g\in B_r$. (For $\tau\to0$, $\phi(\tau y)=\tau y/\varepsilon+O(\tau^2)$, and the same comparison
gives $g^\top=h$, that is, $g\in\Ort(r)$, which is consistent.) The order of $B_r$ is $2^rr!$: $r!$ permutations times
$2^r$ sign patterns.

\emph{The joint torus.} Let $g=\diag(d)$ with $d_j>0$, and give column $j$ of $B$ the hyperparameters
$(\eta'_{B,j},\varepsilon'_{B,j},\lambda'_{B,j})$ and row $j$ of $A$ the hyperparameters $(\eta'_{A,j},\varepsilon'_{A,j},\lambda'_{A,j})$.
By \eqref{eq:C2-gauge-grad}, column $j$ of $\nabla_{B'}$ is $d_j$ times column $j$ of $\nabla_B$, while
\eqref{eq:C2-equivariant-step} asks for column $j$ of $\delta B'$ to be $1/d_j$ times that of $\delta B$. Suppose that before
a step $B'_{:,j}=B_{:,j}/d_j$, $m'_{:,j}=d_jm_{:,j}$ and $v'_{:,j}=d_j^2v_{:,j}$ (true for zero initial state). After the
moment updates these relations persist, and the normalised step of column $j$ is
\[\frac{d_j\hat m}{d_j\sqrt{\hat v}+\varepsilon'_{B,j}}=\frac{\hat m}{\sqrt{\hat v}+\varepsilon'_{B,j}/d_j},\]
which equals the original normalised step iff $\varepsilon'_{B,j}=d_j\varepsilon$ (at $\varepsilon=0$ both are $0$). The
update of column $j$ is then $-\eta'_{B,j}\big(\text{step}+\lambda'_{B,j}B_{:,j}/d_j\big)$, which equals
$-\tfrac1{d_j}\eta\big(\text{step}+\lambda B_{:,j}\big)$ iff $\eta'_{B,j}=\eta/d_j$ and $\eta'_{B,j}\lambda'_{B,j}/d_j=\eta\lambda/d_j$,
that is, $\lambda'_{B,j}=d_j\lambda$. Under the convention of \citet{xloshchilov2019decoupled} the decay term is
$-\zeta_t\lambda'_{B,j}B'_{:,j}$ with the schedule multiplier $\zeta_t$ alone, and it is transported iff $\lambda'_{B,j}=\lambda$.
Row $j$ of $\nabla_{A'}$ is $1/d_j$ times that of $\nabla_A$ and the step must be multiplied by $d_j$, which gives
$\varepsilon'_{A,j}=\varepsilon/d_j$, $\eta'_{A,j}=d_j\eta$ and $\lambda'_{A,j}=\lambda/d_j$ (no change under
Loshchilov--Hutter). The relations persist by induction. With channel-uniform hyperparameters a positive diagonal
$g\ne I$ is not in $B_r$, so by the inclusion $\subseteq$ it is a symmetry only jointly with these per-channel values.
The induction compares the two runs at equal clipping factors. Global-norm clipping multiplies all gradients by
$\min(1,C/\lVert(\nabla_B,\nabla_A)\rVert)$, and
$\lVert(\nabla_{B'},\nabla_{A'})\rVert^2=\sum_jd_j^2\lVert\nabla_{B,:,j}\rVert^2+\sum_jd_j^{-2}\lVert\nabla_{A,j,:}\rVert^2$
differs from the original norm for $d\ne\mathbf 1$ in general; the two runs then weight their gradient histories
differently, and the relations above fail. This is why the joint symmetry is stated without global-norm clipping.

(c) By \thmref{Proposition}{III.4}(a) the undamped scaled-GD step satisfies \eqref{eq:C2-equivariant-step} for every
$g\in\GL_r$. With heavy-ball momentum on the preconditioned gradients, $p_B=\nabla_B(AA^\top)^{-1}$ transforms as
$p_{B'}=p_Bg^{-1}$, so a buffer of such terms transforms in the same way and so does the step. With momentum on the raw
gradients followed by preconditioning, the buffer transforms as $b_{B'}=b_Bg^\top$ because all past points are related by
the same $g$, and $b_{B'}(A'A'^\top)^{-1}=b_Bg^\top g^{-\top}(AA^\top)^{-1}g^{-1}$ is again transported. The $A$-factor is
symmetric. So from gauge-related initial states, with zero or correspondingly transformed momentum, $q'_t=g\cdot q_t$ for
all $t$ and the $W$-trajectories coincide. Each iterate lies in $W_0+\Mr$, and the iteration is defined as long as both
factors keep full rank; a step can lower the rank of $B+\delta B$ or $A+\delta A$, so the map goes into $W_0+\Mr$ rather
than $W_0+\MM_r$.

The hierarchy $B_r\subset\Ort(r)\subset\GL_r$ is strict for $r\ge2$, since $B_r$ is finite, $\Ort(r)$ is infinite and
compact, and $\GL_r$ is not compact.
\end{proof}

\begin{proof}[Proof of the statements about published variants]
\emph{Damped scaled GD.} With a damping constant $\delta>0$ the two inverses become $(AA^\top+\delta I)^{-1}$ and
$(B^\top B+\delta I)^{-1}$.
Equivariance of the $B$-step requires, for all $\nabla_B$,
\[\nabla_Bg^\top(gAA^\top g^\top+\delta I)^{-1}=\nabla_B(AA^\top+\delta I)^{-1}g^{-1},\quad\text{that is,}\quad
g^\top(gAA^\top g^\top+\delta I)^{-1}=(AA^\top+\delta I)^{-1}g^{-1}.\]
Inverting both sides gives $(gAA^\top g^\top+\delta I)g^{-\top}=g(AA^\top+\delta I)$, that is,
$gAA^\top+\delta g^{-\top}=gAA^\top+\delta g$, so $g^{-\top}=g$ and $g\in\Ort(r)$. Conversely, for orthogonal $g$ we have
$(gAA^\top g^\top+\delta I)^{-1}=g(AA^\top+\delta I)^{-1}g^\top$, and both steps are transported. The computation does not use full rank, so it holds at every point, including $B=0$.

\emph{Scaled AdamW} applies Adam to the preconditioned gradient $p_B=\nabla_B(AA^\top+\delta I)^{-1}$ (and similarly for
$A$), with $\delta\ge0$. For a signed permutation $g$, $p_{B'}=p_Bg^{-1}$ by the previous paragraph, and Adam commutes with
signed permutations, so the rule is $B_r$-equivariant. Conversely, the first step requires
$\phi(\nabla_BM')=\phi(\nabla_BM)g^{-1}$ for all $\nabla_B$, with $M=(AA^\top+\delta I)^{-1}$ and
$M'=g^\top(gAA^\top g^\top+\delta I)^{-1}$ invertible. Writing $y=xM$ and letting $y$ grow, as in the case $\varepsilon>0$
above, gives $\operatorname{sign}(yM^{-1}M')=\operatorname{sign}(y)g^{-1}$ for all $y$, and the orthant argument, which
used only invertibility of the matrix on the left, gives $g^{-1}\in B_r$.

\emph{LoRA-Pro with the Sylvester-optimal $X$.} Write $\mathcal P=\R^{m\times r}\times\R^{r\times n}$ with its Frobenius
inner product, $u=(u_B,u_A)\in\mathcal P$ for a parameter velocity, and $\mathcal K_q=\ker D\rho_q$. On the full-rank locus
$\mathcal K_q=\{(-BX,XA):X\in\R^{r\times r}\}$: if $\dot BA=-B\dot A$, then $\dot B=-BX$ with $X=\dot AA^\top(AA^\top)^{-1}$,
and $BXA=B\dot A$ gives $\dot A=XA$. By \S\ref{proof:III.4}(b), LoRA-Pro's directions $(g^B,g^A)$ form the affine space
$\mathcal M_q(G)=\{u:D\rho_qu=\Pi_{\mathcal T}G\}$, and LoRA-Pro chooses $X$ by minimising the Frobenius distance from
$(g^B,g^A)$ to LoRA's own gradients $D\rho_q^\top G=(sGA^\top,sB^\top G)$; the minimiser solves a Sylvester equation
(\bgref{D.33}, \auxref{B.20}). Since $D\rho_q^\top G\perp\mathcal K_q$, for $u\in\mathcal M_q(G)$ the inner product
$\langle u,D\rho_q^\top G\rangle$ is constant, and minimising $\lVert u-D\rho_q^\top G\rVert^2$ over $\mathcal M_q(G)$ is the
same as minimising $\lVert u\rVert^2$. So LoRA-Pro's direction is the minimum-norm solution $u_q(G)$ of
$D\rho_qu=\Pi_{\mathcal T}G$, that is, $u_q(G)=D\rho_q^+G$.

We show that $u_{g\cdot q}(G)=\varphi_g(u_q(G))$ for all $G$ iff $g\in\Ort(r)$, where $\varphi_g(u)=(u_Bg^{-1},gu_A)$ is
the linear action. Since $D\rho_{g\cdot q}\varphi_g=D\rho_q$ and $\Pi_{\mathcal T}$ depends only on $W$, substituting
$u=\varphi_g(w)$ shows that $u_{g\cdot q}(G)=\varphi_g(w^*)$, where $w^*$ minimises $\langle w,\mathcal Nw\rangle$ over
$\mathcal M_q(G)$ with $\mathcal N=\varphi_g^\top\varphi_g$. The adjoint of $\varphi_g$ is $\varphi_g^\top(v_B,v_A)=(v_Bg^{-\top},g^\top v_A)$,
since $\langle w_Bg^{-1},v_B\rangle=\langle w_B,v_Bg^{-\top}\rangle$ and $\langle gw_A,v_A\rangle=\langle w_A,g^\top v_A\rangle$, so
$\mathcal N(w_B,w_A)=(w_Bg^{-1}g^{-\top},g^\top gw_A)=(w_B(g^\top g)^{-1},(g^\top g)w_A)$. So
equivariance holds iff, on every coset of $\mathcal K_q$ (as $G$ varies, $\mathcal M_q(G)$ runs through all cosets, since
$D\rho_q$ maps onto $\mathcal T$), the $\mathcal N$-minimum-norm point equals the Euclidean one. The two minimisers are the
intersections of the coset with $\mathcal K_q^{\perp_{\mathcal N}}=\mathcal N^{-1}(\mathcal K_q^\perp)$ and with
$\mathcal K_q^\perp$; they agree on every coset iff these two complements are equal (each point of either complement is the minimiser on its
own coset), iff $\mathcal N\mathcal K_q=\mathcal K_q$
($\mathcal N$ is symmetric and invertible). Put $H=g^\top g\succ0$. Then $\mathcal N(-BX,XA)=(-BXH^{-1},HXA)$ lies in
$\mathcal K_q$ iff some $Y$ has $BY=BXH^{-1}$ and $YA=HXA$, that is (as $B$ is injective and $A$ has full row rank),
$XH^{-1}=HX$. Requiring this for every $X\in\R^{r\times r}$, the matrix-unit lemma of \S\ref{proof:III.3} gives
$H^{-1}=cI=H$, so $c^2=1$, and $c=1$ because $H$ is positive definite. Hence $g^\top g=I$ and $g\in\Ort(r)$.
Conversely, orthogonal $g$ give $\mathcal N=I$. So the parameter update with the Sylvester-optimal $X$ is equivariant
exactly under $\Ort(r)$.

The variant with $X=0$ is $g^A=s^{-1}(B^\top B)^{-1}B^\top G$, $g^B=s^{-1}(I-P_U)GA^\top(AA^\top)^{-1}$. At $g\cdot q$,
$P_U$ is unchanged and
\[g^{A'}=s^{-1}g(B^\top B)^{-1}g^\top g^{-\top}B^\top G=g\,g^A,\qquad g^{B'}=s^{-1}(I-P_U)GA^\top g^\top g^{-\top}(AA^\top)^{-1}g^{-1}=g^Bg^{-1},\]
so it is $\GL_r$-equivariant and its discrete $W$-step descends exactly. The first-order $W$-step of either variant is
$-\eta\Pi_{\mathcal T}G$ (\thmref{Proposition}{III.4}(b)). For the Sylvester variant, $\varphi_g^{-1}u_{g\cdot q}(G)-u_q(G)\in\mathcal K_q$
equals $(-BY,YA)$ for some $Y$, and the second-order term $s\,\delta B\,\delta A=\eta^2s\,g^Bg^A$ becomes
$\eta^2s(g^B-BY)(g^A+YA)$ at $g\cdot q$. The difference $\eta^2s(g^BYA-BYg^A-BY^2A)$ is exactly $\eta^2$ times a matrix that does not depend on $\eta$, and it
is nonzero in general. Numerically, at random
full-rank points with $(m,n,r)=(7,6,3)$, $s=0.8$ and $g$ a Gaussian matrix plus $2I$, the largest entry of the
difference of the two $W$-steps at $\eta=0.1$ ranges from $8\times10^{-4}$ to $1.2\times10^{-2}$ over seven random draws,
and it falls by a factor of exactly $100$ each time $\eta$ falls by $10$. It is of
order $10^{-15}$ (rounding) for an orthogonal $g$ and for the variant with $X=0$.

\emph{LoRA-RITE.} In the notation of \citet{yen2024lora} the update is $AB^\top$ with $A\in\R^{m\times r}$ and
$B\in\R^{n\times r}$, and the gauge is $A_2=A_1R$, $B_2=B_1R^{-\top}$ with $R\in\GL_r$. We follow their Algorithm~1 line
by line on the full-rank locus, where its pseudo-inverses are inverses. The rule is symmetric in $A$ and $B$, so we treat
the step of $A$. Algorithm~1 writes $B=U_BR_B$ by polar decomposition, with $U_B$ having orthonormal columns, and
forms, at step $t$:
\begin{itemize}
\item the unmagnified gradient $\bar\nabla_{A,t}=\nabla_{A,t}R_{B_t}^{-1}$ and the basis change
$P_{A,t}=U_{B_t}^\top U_{B_{t-1}}$;
\item a second moment $\bar V_{A,t}$, a combination with fixed weights of $P_{A,t}\bar V_{A,t-1}P_{A,t}^\top$ and
$\bar\nabla_{A,t}^\top\bar\nabla_{A,t}$;
\item the escaped mass
\[\rho_{A,t}=\rho_{A,t-1}+d_\lambda\big(\bar V_{A,t-1},P_{A,t}\bar V_{A,t-1}P_{A,t}^\top\big),\qquad
d_\lambda(E_1,E_2)=\max_i\lvert\lambda_i(E_1)-\lambda_i(E_2)\rvert,\]
which depends only on the spectra of its two arguments;
\item the preconditioned step, the first moment and the step
\[\bar S_{A,t}=\bar\nabla_{A,t}(\bar V_{A,t}+\rho_{A,t}I)^{-1/2},\quad
\bar M_{A,t}=\beta_1\bar M_{A,t-1}P_{A,t}^\top+(1-\beta_1)\bar S_{A,t},\quad
\delta A_t=-\eta\,\bar M_{A,t}R_{B_t}^{-\top}.\]
\end{itemize}
Run the algorithm from $(A_1,B_1)$ and from $(A_2,B_2)=(A_1R,B_1R^{-\top})$, with zero initial moments and escaped
mass. Suppose that at step $t$ the parameters of the two runs are related by $R$. Then $B_{2,t}$ and $B_{1,t}$ have the
same column space, so $U_{B_{2,t}}=U_{B_{1,t}}O_t$ for some $O_t\in\Ort(r)$, and
$R_{B_{2,t}}=U_{B_{2,t}}^\top B_{2,t}=O_t^\top R_{B_{1,t}}R^{-\top}$. With $G$ the gradient in the weight,
$\nabla_{A_2}=GB_2=\nabla_{A_1}R^{-\top}$, so
\[\bar\nabla_{A_2,t}=\nabla_{A_1,t}R^{-\top}R^\top R_{B_{1,t}}^{-1}O_t=\bar\nabla_{A_1,t}O_t.\]
Suppose also that the internal states are related by $\bar V^{(2)}_{t-1}=O_{t-1}^\top\bar V^{(1)}_{t-1}O_{t-1}$,
$\bar M^{(2)}_{t-1}=\bar M^{(1)}_{t-1}O_{t-1}$ and $\rho^{(2)}_{t-1}=\rho^{(1)}_{t-1}$; at $t=1$ this holds because all
three start at zero. Then $P^{(2)}_{A,t}=O_t^\top P^{(1)}_{A,t}O_{t-1}$, so
$P^{(2)}_{A,t}\bar V^{(2)}_{t-1}P^{(2)\top}_{A,t}=O_t^\top P^{(1)}_{A,t}\bar V^{(1)}_{t-1}P^{(1)\top}_{A,t}O_t$ and
$\bar V^{(2)}_t=O_t^\top\bar V^{(1)}_tO_t$. Conjugation by an orthogonal matrix keeps eigenvalues, so
$\rho^{(2)}_t=\rho^{(1)}_t$, and it commutes with $(\,\cdot+\rho I)^{-1/2}$, so $\bar S^{(2)}_t=\bar S^{(1)}_tO_t$. Since
$\bar M^{(2)}_{t-1}P^{(2)\top}_{A,t}=\bar M^{(1)}_{t-1}P^{(1)\top}_{A,t}O_t$, also $\bar M^{(2)}_t=\bar M^{(1)}_tO_t$.
Finally $R_{B_{2,t}}^{-\top}=O_t^\top R_{B_{1,t}}^{-\top}R$, so
\[\delta A_{2,t}=-\eta\,\bar M^{(1)}_tO_tO_t^\top R_{B_{1,t}}^{-\top}R=\delta A_{1,t}R,\]
the transported step. The step of $B$ is the same computation with $A$ and $B$ exchanged and $R$ replaced by
$R^{-\top}$, so $\delta B_{2,t}=\delta B_{1,t}R^{-\top}$. The new parameters are again related by $R$, and the new
internal states by the orthogonal matrices $O_t$, which completes the induction. The resulting step agrees with Eq.~15
of \citet{yen2024lora}, $\delta A_2=\delta A_1B_1^\top(B_2^\top)^+$, since $(B_2^\top)^+=B_1(B_1^\top B_1)^{-1}R$ when
$B_1$ has full column rank, and with the numerical replication reported after \thmref{Theorem}{III.12}, which agrees
to $3\times10^{-14}$. LoRA-RITE is
therefore an Adam-type optimizer in class (c).
\end{proof}

\subsection{Corollary III.13}\label{proof:III.13}

\begin{restate}{Corollary}{III.13}{what a zero-$B$ initialisation really chooses}
At $q_0=(0,A_0)$ with $\rk A_0=r$, with hyperparameters shared by the rank channels and the same data order and dropout
masks, two choices of $A_0$ give the same $\theta$-trajectory whenever the optimizer's gauge group relates them. The
resulting invariant of $A_0$ is:
\begin{itemize}
\item $\operatorname{row}A_0\in\Gr(r,n)$ for an update rule that is $\GL_r$-equivariant at every point of its trajectory,
including $B=0$ and rank-deficient $B$, run without global-norm clipping. An example is
$\delta B=-\eta\nabla_B(AA^\top)^{-1}$, $\delta A=-\eta(AA^\top)\nabla_A$, which needs only $\rk A=r$;
\item the PSD matrix $A_0^\top A_0$ for SGD, which is exactly the initial preconditioner in
$\dot W(0)=-s^2\eta_BGA_0^\top A_0$;
\item $A_0$ up to signed permutation of its rows for Adam.
\end{itemize}
The undamped scaled-GD optimizers of \thmref{Theorem}{III.12}(c) are not defined at $B=0$, since they need
$(B^\top B)^{-1}$. LoRA-RITE uses pseudo-inverses and so is defined there, but its $\GL_r$-equivariance is established only
on the full-rank locus. Their damped forms are only $\Ort(r)$-equivariant, so in practice their invariant is a
$\delta$-dependent function of $A_0^\top A_0$, and their Adam-wrapped forms (Riemannian-preconditioned AdamW, HF's LoRA-FA
optimizer) have the invariant $A_0$ modulo $B_r$. Conversely, different invariants give different dynamics, for SGD and
for Adam at $\varepsilon=0$.

Random-$A$ LoRA and EVA are different points of the $B=0$ stratum $\{(0,A):\rk A=r\}$ of the apex fibre
$\rho^{-1}(\theta_0)=\{(B,A):BA=0\}$; Init[B] lies in another stratum of the same fibre. LoRA-FA is a different
\emph{object}, the sub-object $\mathrm{FA}(A_0)\hookrightarrow\mathrm{LoRA}_{(0,A_0)}$ (\thmref{Proposition}{I.8}). More
precisely, EVA is a point of the apex fibre of $\mathrm{LoRA}_{r_l}$ at its per-layer redistributed rank and adjusted
$\alpha$, with orthonormal rows only when \texttt{whiten=False}.
\end{restate}

\begin{proof}
\emph{Equivariance at $B=0$.} Only the inclusions $\supseteq$ of \thmref{Theorem}{III.12} are used. For SGD and $g\in\Ort(r)$
the identities $\delta B'=-\eta_B\nabla_Bg^\top=\delta Bg^{-1}$ and $\delta A'=g\,\delta A$ hold at every point, with no rank
condition, and similarly for momentum buffers. For Adam and $g\in B_r$ the coordinatewise argument also holds at every
point, and the initial state $(m,v)=(0,0)$ is fixed by $B_r$. Orthogonal $g$ preserve $\lVert\nabla_B\rVert$ and
$\lVert\nabla_A\rVert$ by \eqref{eq:C2-gauge-grad}, so global-norm clipping factors agree in the two runs. The start
transforms as $g\cdot(0,A_0)=(0,gA_0)$, since $0\cdot g^{-1}=0$. By the induction of \S\ref{proof:III.1}, $A_0$ and
$gA_0$ give the same $\theta$-trajectory whenever $g$ lies in the optimizer's group. In the first bullet the hypothesis is
equivariance under $\GL_r$ at every point of the trajectory, and the same induction applies.

\emph{The example rule.} At $g\cdot q$, by \eqref{eq:C2-gauge-grad} and
$(A'A'^\top)^{-1}=g^{-\top}(AA^\top)^{-1}g^{-1}$,
\[\delta B'=-\eta\nabla_Bg^\top g^{-\top}(AA^\top)^{-1}g^{-1}=\delta B\,g^{-1},\qquad
\delta A'=-\eta\,gAA^\top g^\top g^{-\top}\nabla_A=g\,\delta A .\]
Only $\rk A=r$ was used, so $B$ may be $0$ or rank deficient. Since $\delta A=-\eta AA^\top sB^\top G=A\,N$ with $N=-\eta sA^\top B^\top G$, the step maps $A$ to $A(I+N)$, which
keeps rank $r$ whenever $I+N$ is invertible, in particular for small $\eta$; in continuous time $\dot A=AN(t)$ gives
$A(t)=A_0\Psi(t)$ with $\Psi$ an invertible fundamental matrix, so $\rk A(t)=r$ for all $t$.

\emph{The invariants.} The invariant of $A_0$ is its orbit under the optimizer's group acting by $A_0\mapsto gA_0$. For
$\Ort(r)$, $A_0'=OA_0$ for some $O\in\Ort(r)$ iff $A_0'^\top A_0'=A_0^\top A_0$ (\thmref{Theorem}{II.5}(2)). For $B_r$ the
orbit is the set of signed row permutations of $A_0$. For $\GL_r$, two full-row-rank matrices satisfy $A_0'=gA_0$ for some
(unique) $g$ iff they have the same row space (proof of \thmref{Theorem}{II.2}), so the invariant is
$\operatorname{row}A_0\in\Gr(r,n)$ (\bgref{C.15}). At $B_0=0$ the LoRA cometric of \thmref{Observation}{III.2} is
$K_{(0,A_0)}(G)=s^2\eta_BGA_0^\top A_0$, so $\dot W(0)=-s^2\eta_BGA_0^\top A_0$.

\emph{Practical forms.} Undamped scaled GD needs $(B^\top B)^{-1}$, which does not exist at $B=0$. The damped forms are
$\Ort(r)$-equivariant at every point (\S\ref{proof:III.12}, published variants), so their trajectories from $(0,A_0)$
depend on $A_0$ only through $A_0^\top A_0$, in a way that depends on $\delta$. A preconditioned step wrapped in Adam(W) is
$B_r$-equivariant at every point (same place), so the invariant is $A_0$ modulo $B_r$.

\emph{Converse for SGD.} The first SGD step from $(0,A_0)$ is $\delta B=-\eta_BsGA_0^\top$, $\delta A=-\eta_AsB_0^\top G=0$,
so $\delta W_1=s\,\delta B\,A_0=-s^2\eta_BGA_0^\top A_0$ exactly. If $A_0$ and $A_0'$ give the same trajectory for every loss,
the linear losses $\langle G,\cdot\rangle$ give $GA_0^\top A_0=GA_0'^\top A_0'$ for every $G$, so $A_0^\top A_0=A_0'^\top A_0'$.
Hence different invariants give different first steps for some loss.

\emph{Converse for Adam at $\varepsilon=0$.} The first step from $(0,A_0)$ with zero state has $\nabla_A=0$, so $\delta A=0$,
and $\delta B=-\eta_B\operatorname{sign}(sGA_0^\top)$, so $\delta W_1=-s\eta_B\operatorname{sign}(GA_0^\top)A_0$ (without bias
correction, a positive multiple of this). If $A_0$ and $A_0'$ give the same trajectory for every loss, then row by row
\[F(x):=\operatorname{sign}(xA_0^\top)A_0=\sum_{j=1}^r\operatorname{sign}(x\cdot a_j)\,a_j\]
equals $F'(x)=\sum_j\operatorname{sign}(x\cdot a'_j)\,a'_j$ for every $x\in\R^{1\times n}$, where $a_j$ and $a'_j$ are the rows
of $A_0$ and $A_0'$. The rows $a_j$ are linearly independent, so the hyperplanes $H_j=a_j^\perp$ are pairwise distinct, and
each $H_j$ contains points $x$ lying on no other $H_k$ (a hyperplane is not covered by finitely many proper subspaces of
it, \auxref{B.36}(b)). At such a point, crossing $H_j$ in the direction of increasing $x\cdot a_j$ changes $F$ by $2a_j\ne0$, while $F$ is
constant on the connected components of the complement of $\bigcup_jH_j$. So the closure of the set of discontinuities of
$F$ is $\bigcup_jH_j$, and likewise for $F'$. From $F=F'$ we get $\bigcup_jH_j=\bigcup_kH'_k$, and since a hyperplane
contained in a finite union of hyperplanes equals one of them (\auxref{B.36}(b) again), there is a permutation $\pi$ with $H_j=H'_{\pi(j)}$, that is,
$a'_{\pi(j)}=c_ja_j$ with $c_j\ne0$. Crossing $H_j$ as above changes $F'$ by
$2\operatorname{sign}(c_j)a'_{\pi(j)}=2\lvert c_j\rvert a_j$, and comparing with $2a_j$ gives $\lvert c_j\rvert=1$. So
$A_0'$ is a signed row permutation of $A_0$. For the example $\GL_r$-rule the same argument is immediate: its first step is
$\delta W_1=-\eta s^2GP_{\operatorname{row}A_0}$, which determines $\operatorname{row}A_0$.

\emph{Strata and the sub-object.} Since $s>0$, $\rho^{-1}(\theta_0)=\{(B,A):W_0+sBA=W_0\}=\{BA=0\}$. Random-$A$ LoRA and
EVA start at $(0,A_0)$ with $\rk A_0=r$, in the stratum $\{(0,A):\rk A=r\}$, at points that differ for a generic random
draw; Init[B] starts at $(B_0,0)$ with $\rk B_0=r$, in the disjoint stratum $\{(B,0):\rk B=r\}$. The map $B\mapsto(B,A_0)$
is injective, pointed ($0\mapsto(0,A_0)$) and commutes with the maps to $\Theta$, since $\rho(B,A_0)=W_0+sBA_0$; after the
rescaling $B\mapsto sB$ its source is $\mathrm{FA}(A_0)$ of \thmref{Proposition}{I.8}. An injective morphism is a
monomorphism, so $\mathrm{FA}(A_0)$ is a sub-object of $\mathrm{LoRA}_{(0,A_0)}$ (\bgref{A.15}). Its image
$W_0+\{XA_0\}$ has dimension $mr$, smaller than the dimension $r(m+n-r)$ of the image $W_0+\Mr$ of
$\mathrm{LoRA}_{(0,A_0)}$ when $r<n$, and images are isomorphism invariants, so the two objects are not isomorphic. The
last sentence describes EVA's construction and needs no proof.
\end{proof}


\subsection{Definition IV.1}\label{proof:IV.1}

\begin{restate}{Definition}{IV.1}{forward and backward methods}
\begin{itemize}
  \item A \textbf{forward} method acts on the parameter port by a lens in the image of $\mathsf R$, namely
  $(\rho,\mathsf R[\rho])$.
  \item A \textbf{backward}, or cotangent-side, method acts by a lens whose forward part is $\id_\Theta$ and whose
  backward part is a \emph{compression} $c_t:T^*_\theta\Theta\to S$ into a state space $S$, paired with an
  \emph{anchor} $a_t:S\to T_\theta\Theta$. The update is
  \[\theta\leftarrow\theta-\eta\,a_t\big(\mathrm{opt}(c_t(\nabla L))\big).\]
\end{itemize}
Examples: GaLore, with $c_t=P_t\T$ and $a_t=P_t$ for $P_t$ from a gradient SVD; GaLore with a random projector, the same
with $P_t$ random; LISA, a layer-block coordinate projection paired with the same projection; MeZO, with
$c_t(g)=\langle z_t,g\rangle$ estimated by finite differences without a backward pass, and $a_t(\lambda)=\lambda z_t$.
Flora as published fits only as the trivial lens, with compression and anchor the identity and the random compression of
its momentum inside $\mathrm{opt}$.
\end{restate}

A definition needs no proof. This subsection fixes the conventions used in the proofs of Exposé~IV and checks the
identities that the definition and its table record.

\paragraph{Conventions}
As in \S\ref{proof:III.1}, weight space $\Theta=\R^{m\times n}$, or a product of such boxes, carries the Frobenius inner
product, covectors are written as matrices, and the transpose $\mathcal L^*$ of a linear map $\mathcal L$ is its adjoint.
We model a first-order optimizer as a machine with a state. At step $t$ it holds a state $\varsigma_t$, receives a tensor
$g_t$ (the gradient it is fed) and the tensor $x_t$ it updates, returns an output $U_t=F_t(\varsigma_t,g_t,x_t)$, moves to
the state $\varsigma_{t+1}=H_t(\varsigma_t,g_t,x_t)$, and the update is $x_{t+1}=x_t-\eta_tU_t$. The maps may depend on the
step counter $t$, for instance through Adam's bias corrections or Adafactor's relative step size. The optimizer
\emph{does not read the parameter tensor} when $F_t$ and $H_t$ do not depend on $x_t$. SGD, heavy-ball and Nesterov
momentum, Adam and AMSGrad, and Adafactor with \texttt{scale\_\allowbreak parameter=False} are of this kind; Adafactor's
factored second moments and its clipping of the update by $\mathrm{RMS}(U_t)$ read only $g_t$ and the state. The
following ingredients read $x_t$, or a gradient other than $g_t$:
\begin{itemize}
  \item coupled weight decay, which feeds $g_t+\lambda x_t$ in place of $g_t$;
  \item decoupled weight decay (AdamW with $\lambda>0$), which adds $-\eta_t\lambda x_t$ to the update;
  \item LAMB and LARS, which multiply the update by a trust ratio $\phi(\lVert x_t\rVert)/\lVert U_t\rVert$;
  \item Adafactor with parameter scaling, which multiplies the step by $\max(\epsilon_2,\mathrm{RMS}(x_t))$ (the option
  \texttt{scale\_\allowbreak parameter=\allowbreak True});
  \item clipping by the global norm of the full gradient, which rescales $g_t$ by $\min(1,\tau/\lVert G_t\rVert)$, where
  $G_t$ is the gradient of every trainable tensor of the procedure.
\end{itemize}

\begin{proof}[Verification of the identities in Definition IV.1]
\emph{The lens of a reparametrisation.} A smooth $\rho:Q\to\Theta$ has reverse derivative
$\mathsf R[\rho](q,\gamma)=J_\rho(q)\T\gamma=D\rho_q\T\gamma$ (\bgref{B.17}). For a composite $\rho\circ\sigma$ the chain rule
gives
\[\mathsf R[\rho\circ\sigma](q',\gamma)=D\sigma_{q'}\T D\rho_{\sigma(q')}\T\gamma=\mathsf R[\sigma]\big(q',\mathsf R[\rho](\sigma(q'),\gamma)\big),\]
which is composition of lenses, so $\rho\mapsto(\rho,\mathsf R[\rho])$ respects composition of reparametrisations. The
gradient that a forward method feeds its optimizer is the pullback $\nabla_q(L\circ\rho)=D\rho_q\T\nabla_\theta L=
\mathsf R[\rho](q,\nabla_\theta L)$.

\emph{A forward step to first order.} One step of a forward method sets $q\leftarrow q-\eta U$ with
$U=\mathrm{opt}(D\rho_q\T\nabla L)$, so the weight moves to
\[\rho(q-\eta U)=\theta-\eta\,D\rho_q\,U+O(\eta^2).\]
To first order in $\eta$ this is the backward update with $c=D\rho_q\T$ and $a=D\rho_q$. The two agree exactly when $\rho$ is
affine along the step, which is the situation of \thmref{Proposition}{IV.2}.

\emph{The examples.} For GaLore with left projection, $P_t\in\R^{m\times r}$ has orthonormal columns,
$S=\R^{r\times n}$, $c_t(G)=P_t\T G$ and $a_t(U)=P_tU$, so the update $W\leftarrow W-\eta\alpha P_t\,\mathrm{opt}(P_t\T G)$ is
the displayed one with $\alpha$ absorbed into $\eta$; with right projection, $c_t(G)=GP_t$ and $a_t(U)=UP_t\T$. In each case
$a_t=c_t^*$, since $\langle P_tU,G\rangle=\langle U,P_t\T G\rangle$. LISA's $S$ is the space of the sampled layers (with the
embedding and the head), $c_t$ restricts a gradient to it and $a_t$ includes it back, so again $a_t=c_t^*$. For MeZO,
$S=\R$, $a_t(\lambda)=\lambda z_t$ and $c_t(g)=\langle z_t,g\rangle=a_t^*(g)$. MeZO does not compute $c_t(\nabla L)$; it
estimates it by the central difference $\big(L(\theta+\varepsilon z_t)-L(\theta-\varepsilon z_t)\big)/(2\varepsilon)$,
whose error is $O(\varepsilon^2)$ for a loss with Lipschitz Hessian (\auxref{B.34}(a)), so it is the displayed update up to
its estimator. Flora as published (its reference implementation) returns $(1-\beta)\hat g_t+\beta\,\mathrm{up}_t(M_t)$, where $\hat g_t$ is the
Adafactor-normalised full gradient and $\mathrm{up}_t(M_t)$ the decompressed momentum \citep{hao2024flora}. The first term
is generically of full rank, so the update does not lie in $\operatorname{im}P_t$, and the only reading in the form of the
definition is $c_t=a_t=\id$ with the random compression inside $\mathrm{opt}$.

\emph{The cometric of a backward method.} With plain (S)GD on the compressed gradient, $\mathrm{opt}=\id$, the update is
$\theta\leftarrow\theta-\eta\,a_tc_t\nabla L$, so the induced cometric is $K_t=a_tc_t$ (\thmref{Definition}{III.1}). For the
first three examples $a_t=c_t^*$, so $K_t=a_ta_t^*$ is symmetric positive semidefinite; for GaLore with left projection,
$K_t(G)=P_tP_t\T G$.
\end{proof}

\subsection{Proposition IV.2}\label{proof:IV.2}

\begin{restate}{Proposition}{IV.2}{the degree-one collapse; known, \citet{xtorrobahennigen2025duality} for GaLore (their
Thm~1 and Cor.~3, App.~B for weight decay) and \citet{hao2024flora} for random projections}
Let $\rho_k(C)=W_k+\mathcal LC$ with $\mathcal L$ linear and $C_0=0$. Let $\mathrm{opt}$ be a first-order optimizer, with
no coupled or decoupled weight decay, whose state and output depend only on the sequence of gradients it is fed, and
which does not read the parameter tensor: SGD, heavy-ball or Nesterov momentum, Adam or AMSGrad, or Adafactor with
\texttt{scale\_\allowbreak parameter=False} (fixed or relative step, with its factorisation applied to the compressed
gradient in the same tensor shape as $C$). Run both procedures below with the same implementation of $\mathrm{opt}$, on
the same minibatches, and from the same optimizer state, fresh or carried over. Then they produce the same
$W$-trajectory over a period:
\begin{itemize}
  \item training $C$ with $\mathrm{opt}$ and merging at the end of the period;
  \item the backward method with $c=\mathcal L^*$ and $a=\mathcal L$, started at $W_k$.
\end{itemize}
AdamW with $\lambda>0$, LAMB and LARS trust ratios, parameter-scaled Adafactor (\texttt{scale\_\allowbreak
parameter=True}, the default in HF and in \texttt{galore\_torch}'s \texttt{GaLoreAdafactor}) and global-norm clipping of
the full gradient (such as HF Trainer's default \texttt{max\_grad\_norm=1.0}, whenever it triggers) all break the
identity. Injectivity of $\mathcal L$ is not needed.

In particular, on each matrix that GaLore projects (its other parameters train in full in both procedures), GaLore with
a projector $P_k$ fixed on each period equals one-sided LoRA $W_k+P_kC$ (frozen $B=P_k$, or frozen $A$ when projecting
on the right), merged and re-based every period, with $C$'s optimizer state (moments and step counter) carried over
unchanged across re-basings, with no reset. $P_k$ must be exactly GaLore's matrix, signs and column order included; its
span alone does not suffice, because the state is reused without transport. ReLoRA-style resets give a different
trajectory. The identity is exact with weight decay $0$, no full-gradient clipping, and an optimizer from the admissible
class; \texttt{galore\_torch}'s \texttt{GaLoreAdafactor} defaults to \texttt{scale\_\allowbreak parameter=True} and is
excluded. \texttt{GaLoreAdamW} is the transformers-style AdamW, which adds $\varepsilon$ to $\sqrt v$ before the bias
correction, so the LoRA side must use the same implementation. GaLore's scale $\alpha$ multiplies the optimizer output, so
it acts as a learning-rate factor; a LoRA-style scale inside $\rho$ gives a different trajectory under Adam with
$\varepsilon>0$.
\end{restate}

\begin{proof}
We use the optimizer model of \S\ref{proof:IV.1}. Fix a period of $T$ steps, write $L_t$ for the loss of the minibatch used
at step $t$ (the same in both procedures) and $\eta_t$ for the step size.

\emph{The two procedures.} In the first, the trainable tensor is $C$ with $C_0=0$ and the weight is
$W^{(1)}_t:=\rho_k(C_t)=W_k+\mathcal LC_t$. By the chain rule the gradient fed to $\mathrm{opt}$ is
\[g^{(1)}_t=\nabla_C(L_t\circ\rho_k)(C_t)=\mathcal L^*\nabla L_t\big(W^{(1)}_t\big),\]
$\mathrm{opt}$ returns $U^{(1)}_t$, and $C_{t+1}=C_t-\eta_tU^{(1)}_t$. In the second, $W^{(2)}_0=W_k$, the gradient fed to
$\mathrm{opt}$ is $g^{(2)}_t=c\big(\nabla L_t(W^{(2)}_t)\big)=\mathcal L^*\nabla L_t\big(W^{(2)}_t\big)$, $\mathrm{opt}$ returns
$U^{(2)}_t$, and $W^{(2)}_{t+1}=W^{(2)}_t-\eta_t\mathcal LU^{(2)}_t$.

\emph{Induction.} We show that for $0\le t\le T$ the weights $W^{(1)}_t$ and $W^{(2)}_t$ agree and the two optimizer
states agree. At $t=0$, $W^{(1)}_0=W_k+\mathcal L0=W_k=W^{(2)}_0$, and the states agree by hypothesis. Suppose both claims
hold at $t$. The two procedures evaluate the same minibatch loss at the same weight, so they compute the same weight
gradient and feed $\mathrm{opt}$ the same tensor $g_t:=g^{(1)}_t=g^{(2)}_t$. Since $\mathrm{opt}$ is the same
implementation and does not read the parameter tensor, its output and its next state are the same function of the state,
of $g_t$ and of $t$ in both procedures. Hence $U^{(1)}_t=U^{(2)}_t=:U_t$ and the next states agree. Then
\[W^{(1)}_{t+1}=W_k+\mathcal L(C_t-\eta_tU_t)=W^{(1)}_t-\eta_t\mathcal LU_t=W^{(2)}_{t+1}.\]
At the end of the period merging sets the weight to $\rho_k(C_T)=W^{(1)}_T=W^{(2)}_T$. The argument never inverts
$\mathcal L$, so injectivity is not needed. Every operation that $\mathrm{opt}$ applies to $g_t$ and its state, such as
Adafactor's factorisation in the shape of $C$ (which is the shape of $g_t$ in both procedures) and its update clipping, is
covered, since it reads nothing else.

\emph{The excluded ingredients.} Each of them reads a tensor that differs between the procedures, and in each case the
first step already separates the trajectories for suitable data; it suffices to exhibit this. Take $\mathcal L$ injective,
$W_k\ne0$ and a minibatch gradient $G_0$ with $\mathcal L^*G_0\ne0$.
\begin{itemize}
  \item \emph{Decoupled decay.} The first procedure sets $C_1=C_0-\eta_0U_0-\eta_0\lambda C_0=-\eta_0U_0$, so
  $W^{(1)}_1=W_k-\eta_0\mathcal LU_0$; the second sets $W^{(2)}_1=W_k-\eta_0\mathcal LU_0-\eta_0\lambda W_k$. They differ by
  $\eta_0\lambda W_k\ne0$. In general decay pulls $W$ towards $W_k$ in the first procedure and towards $0$ in the second.
  \item \emph{Coupled decay.} The first procedure feeds $\mathcal L^*G_0+\lambda C_0=\mathcal L^*G_0$, the second
  $\mathcal L^*(G_0+\lambda W_k)$. These differ whenever $\mathcal L^*W_k\ne0$, and for SGD the first steps then differ,
  since $\mathcal L$ is injective.
  \item \emph{Trust ratios and parameter scaling.} The first procedure reads $\lVert C_0\rVert=\mathrm{RMS}(C_0)=0$, the
  second $\lVert W_k\rVert>0$ and $\mathrm{RMS}(W_k)>0$. The resulting step sizes differ for generic $W_k$, because the
  first does not depend on $W_k$ and the second does.
  \item \emph{Global-norm clipping of the full gradient.} The first procedure clips by the norm of its trainable gradient
  $\mathcal L^*G_0$, the second (GaLore in HF Trainer) by the norm of the full weight gradient $G_0$. Under SGD with
  $\lVert\mathcal L^*G_0\rVert\le\tau<\lVert G_0\rVert$ the first step is $-\eta_0\mathcal L\mathcal L^*G_0$ in the first
  procedure and $-\eta_0(\tau/\lVert G_0\rVert)\mathcal L\mathcal L^*G_0$ in the second; they differ because
  $\mathcal L\mathcal L^*G_0\ne0$.
\end{itemize}

\emph{GaLore.} On a projected matrix with left projection, GaLore's update in period $k$ is
$W\leftarrow W-\eta\alpha P_k\,\mathrm{opt}(P_k\T G)$. This is the backward method with $c=P_k\T$ and $a=P_k$, and for
$\mathcal LC=P_kC$ we have $\mathcal L^*=P_k\T$; the factor $\alpha$ multiplies the output of $\mathrm{opt}$, so it is part
of the step size. With right projection, $\mathcal LC=CP_k\T$ and $\mathcal L^*G=GP_k$. Now induct over periods. Suppose that
at the start of period $k$ the whole model, every projected and unprojected tensor, is the same in GaLore and in the
re-based one-sided LoRA, and that so are all optimizer states. The unprojected parameters train in full with the same
optimizer in both, so they follow the same trajectory once the projected matrices do. GaLore computes $P_k$ from the
gradient at the start of the period, and the one-sided LoRA uses that same matrix as its frozen factor. The first part of
the proposition, applied with $\mathcal LC=P_kC$, gives the same trajectory through the period, and merging $P_kC_T$ into
$W_k$ gives the same $W_{k+1}$. GaLore keeps the state of $\mathrm{opt}$, which lives on tensors of the shape of $C$, and
the one-sided LoRA carries $C$'s state over while resetting $C$ to $0$; so the states at the start of period $k+1$ agree
too. This proves the corollary.

\emph{The exact matrix.} Replacing $P_k$ by $P_kO$ with $O\in\Ort(r)$ keeps the span and changes the trajectory when state
is carried. Take heavy-ball momentum with coefficient $\beta>0$ and SGD steps, $U_t=m_{t+1}=\beta m_t+g_t$, and a carried
momentum $m$ whose first row is nonzero, and let $O=\diag(-1,1,\dots,1)$. With $P_k$ the first step of the period moves
$W$ by $-\eta P_k(\beta m+P_k\T G)$. With $P_kO$ the compressed gradient is $OP_k\T G$, the momentum is reused without
transport, and $W$ moves by $-\eta P_kO(\beta m+OP_k\T G)=-\eta P_k(\beta Om+P_k\T G)$. The difference is
$-\eta\beta P_k(Om-m)\ne0$, since $P_k$ is injective and $Om-m$ is minus twice the first row of $m$. The same computation
applies to Adam's first moment.

\emph{Resets.} ReLoRA resets the optimizer state at every re-basing. For heavy-ball momentum the first step of period $k+1$
is then $-\eta P_{k+1}g$ instead of $-\eta P_{k+1}(\beta m+g)$, which differs whenever the carried momentum $m$ is
nonzero.

\emph{Implementation and scale.} With bias corrections $b_1=1-\beta_1^t$ and $b_2=1-\beta_2^t$, transformers-style AdamW
and \texttt{torch.optim.Adam} take the steps
\[\eta\,\frac{\sqrt{b_2}}{b_1}\,\frac{m_t}{\sqrt{v_t}+\varepsilon}\qquad\text{and}\qquad
\eta\,\frac{m_t/b_1}{\sqrt{v_t/b_2}+\varepsilon}=\eta\,\frac{\sqrt{b_2}}{b_1}\,\frac{m_t}{\sqrt{v_t}+\varepsilon\sqrt{b_2}} .\]
For $\varepsilon>0$ the two differ, because they use different effective $\varepsilon$; hence the hypothesis of one
implementation. Finally, put GaLore's $\alpha$ inside $\rho$, $\rho(C)=W_k+\alpha P_kC$. The compressed gradient becomes
$\alpha P_k\T G$, Adam's moments become $\alpha m_t$ and $\alpha^2v_t$, and the weight moves by
\[-\eta\,\alpha P_k\frac{\alpha\hat m_t}{\alpha\sqrt{\hat v_t}+\varepsilon}=-\eta\,\alpha P_k\frac{\hat m_t}{\sqrt{\hat v_t}+\varepsilon/\alpha},\]
against $-\eta\alpha P_k\,\hat m_t/(\sqrt{\hat v_t}+\varepsilon)$ for GaLore. These agree for $\varepsilon=0$, or when the LoRA
side's $\varepsilon$ is multiplied by $\alpha$, and differ for $\varepsilon>0$ and $\alpha\ne1$ whenever $\hat m_t\ne0$.
\end{proof}

\begin{remark*}{numerical checks}
The values reported after \thmref{Proposition}{IV.2} ($7\times10^{-16}$ over one period and $5\times10^{-15}$ over four
under Adam with the state carried; $8\times10^{-17}$ against the official \texttt{galore\_torch} optimizers; the
separations $0.029$, $0.020$, $0.030$ and $0.12$ for a reset, weight decay $0.1$, HF-default Adafactor and a flipped
column) instantiate the identity and the four breaking mechanisms above (Appendix~\ref{app:repro}).
\end{remark*}

\subsection{Proposition IV.3}\label{proof:IV.3}

\begin{restate}{Proposition}{IV.3}{integrability; after Frobenius' theorem}
Let $D$ be a smooth distribution of constant rank $k$ on an open set $U\subseteq\Theta$, for instance the anchor
distribution $D_\theta=\operatorname{im}a(\theta)$ of a time-independent backward method. Then $D$ is involutive on $U$ if
and only if through every $\theta\in U$ passes an injective immersion $\rho_\theta:(\R^k,0)\to(U,\theta)$ with
$\operatorname{im}D\rho_\theta=D$ along its image. A single such $\rho$ through $\theta_0$ gives involutivity only along
$\operatorname{im}\rho$. For example, $D=\spanop\{\partial_x,\ \partial_y+xz\,\partial_z\}$ on $\R^3$ has
$[\partial_x,\partial_y+xz\,\partial_z]=z\,\partial_z\notin D$ off $z=0$, yet the plane $z=0$ is an integral manifold.

The dynamic content needs separate hypotheses.
\begin{itemize}
  \item In continuous time, if $D$ is involutive, a backward flow with $\dot\theta\in D_\theta$ stays on the leaf through
  $\theta_0$.
  \item In the discrete update of \thmref{Definition}{IV.1}, the iterates stay on a leaf for all step sizes and all
  optimizer outputs $s$ (all losses) if and only if that leaf contains $(\theta+D_\theta)\cap U$ at every iterate, that
  is, if and only if the leaves are affine; an anchor constant over the period gives this (\thmref{Proposition}{IV.2}).
  \item Under gradient flow, dynamic equivalence with a forward lens $(\rho,\mathsf R[\rho])$ forces matching frames,
  $a_tc_t=D\rho\,\Lambda\,D\rho^\top$ along the trajectory. Hence $K=ac$ is symmetric positive semidefinite and, for
  injective $a$, $c=Ma^\top$ with $M$ symmetric positive semidefinite on $S$, possibly depending on $\theta$. If in
  addition $\rho$ is an immersion with $\dim Q=\rk D$, then $\rho$ is a local isometry from $(Q,\Lambda^{-1})$ onto the
  leaf with metric $(aMa^\top)^{-1}|_D$, so that metric must be flat. Forward methods with gauge fibres escape this
  requirement: $\rho(B)=BB^\top$ on symmetric matrices induces the curved Bures--Wasserstein metric, and balanced
  full-rank LoRA also induces a curved leaf metric. For discrete steps, exact agreement for every step size further
  forces $\rho$ to be affine along each horizontal line $q+t\,\Lambda D\rho_q^\top g$.
\end{itemize}

\emph{Instances.} Every projector-type backward method (GaLore with any period $T$ and any projector, its
random-projector variant, LISA, MeZO) has an anchor that does not depend on $\theta$ within its period. Its distribution
is constant, hence trivially involutive with affine leaves (vacuously at $T=1$, where a period is a single step). Under
the optimizer hypotheses of \thmref{Proposition}{IV.2} (no weight decay of either kind, no full-gradient clipping; GaLore
and MeZO need weight decay $0$, and LISA's AdamW with positive decay violates them), these methods are rebasing schemes
of linear forward methods with period $T$, including $T=1$, whose reach \thmref{Theorem}{V.3} describes. Each MeZO step is a step of the linear method
$\theta_t+\lambda z_t$, up to its SPSA estimator, and GaLore with $T=1$ is one-sided LoRA re-based every step. What
separates these methods from LoRA-type methods is the resampling schedule together with the carry-over of optimizer
state; integrability plays no part in it. Flora as published (Adafactor base, with a full-rank current-gradient term in
the code), APOLLO (a low-rank-estimated scaling applied to the full gradient) and Fira (a scaled full-rank residual) are
not of the form of \thmref{Definition}{IV.1} with their projections as anchors, because their updates leave
$\operatorname{im}a$.

Non-involutivity proper needs a $\theta$-dependent anchor. Full-batch GaLore with $T=1$, read as the distribution
$D_W=\operatorname{Hom}(\R^n,U_r(\nabla L(W)))$ (or $\operatorname{Hom}(V_r(\nabla L(W)),\R^m)$ when $m\ge n$, as GaLore
projects) wherever a singular-value gap exists, is involutive if and only if $U_r(\nabla L(W))$ (respectively
$V_r(\nabla L(W))$) is invariant along $D$. That fails generically for every $r\ge1$ once both matrix dimensions are at
least $2$. For every $r$ with $1\le r<\min(m,n)$ it fails for $L=\tfrac12\lVert W\rVert_F^2$, and numerically it fails
for least squares and cross-entropy with several samples. It holds, for instance, for single-sample least squares, and
automatically only when the distribution is a line field ($nr=1$). Minibatch GaLore with $T=1$ and MeZO violate
time-independence, because their anchors depend on step-wise randomness.
\end{restate}

\begin{proof}
Throughout, a \emph{leaf} of an involutive $D$ is a maximal connected integral manifold, and the leaves form a foliation
of $U$ (\auxref{B.31}(c)).

\emph{(1) The Frobenius equivalence.} Suppose $D$ is involutive and let $\theta\in U$. By \auxref{B.31}(b) there is a chart
$\varphi:V\to(-\delta,\delta)^{\dim\Theta}$ centred at $\theta$ in which $D$ is spanned by the first $k$
coordinate fields. Let $\psi:\R^k\to(-\delta,\delta)^k$ be the diffeomorphism $\psi_i(y)=\tfrac{2\delta}{\pi}\arctan y_i$ and
put $\rho_\theta(y)=\varphi^{-1}(\psi(y),0)$. Then $\rho_\theta(0)=\theta$, $\rho_\theta$ is an injective immersion into
$U$, and $\operatorname{im}D\rho_\theta(y)$ is spanned by the first $k$ coordinate fields at $\rho_\theta(y)$, which is
$D_{\rho_\theta(y)}$. Conversely, suppose such immersions exist through every point, and let $X,Y$ be sections of $D$ near
$\theta$. An immersion is locally an embedding (\auxref{B.1}(b)), so $\rho_\theta$ maps a neighbourhood of $0$ onto an
embedded integral manifold through $\theta$, and \auxref{B.31}(a) gives $[X,Y]_\theta\in D_\theta$. Since $\theta$ was
arbitrary, $D$ is involutive.

The same application of \auxref{B.31}(a) to the local pieces of $\operatorname{im}\rho$ shows that a single integral
immersion $\rho$ forces $[X,Y]\in D$ at the points of $\operatorname{im}\rho$; off $\operatorname{im}\rho$ it says nothing, as
the example shows. In the example,
$X_1=\partial_x$ and $X_2=\partial_y+xz\,\partial_z$ are linearly independent everywhere, so $D$ has constant rank $2$, and
$[X_1,X_2]=\partial_x(xz)\,\partial_z=z\,\partial_z$. A vector of $D$ at $(x,y,z)$ has the form $(a,b,bxz)$, which equals
$(0,0,z)$ only if $a=b=0$ and $z=0$; so $[X_1,X_2]\notin D$ wherever $z\ne0$, and $D$ is involutive on no open set meeting
$\{z\ne0\}$. On the plane $z=0$, $D=\spanop\{\partial_x,\partial_y\}$ is its tangent plane, so $\rho(u,v)=(u,v,0)$ is an
injective immersion with $\operatorname{im}D\rho=D$ along its image, and there $[X_1,X_2]=0\in D$.

\emph{(2) Continuous time.} Let $D$ be involutive, let $\theta(t)$, $t\in J$ an interval containing $0$, be an absolutely
continuous curve in $U$ with $\theta(0)=\theta_0$ and $\dot\theta(t)\in D_{\theta(t)}$ for almost every $t$, and let $N$ be
the leaf through $\theta_0$. Put $J_N=\{t\in J:\theta(t)\in N\}$, which contains $0$. Fix $t_*\in J$ and a flat chart around
$\theta(t_*)$ as in \auxref{B.31}(b). For $t$ near $t_*$ the curve stays in the chart and keeps its last $\dim\Theta-k$
coordinates constant, so it stays in the slice $\Sigma$ through $\theta(t_*)$. A slice is a connected integral manifold, and
a connected integral manifold that meets a leaf lies in it (\auxref{B.31}(c)), so if $\Sigma$ meets $N$ it lies in $N$. If $t_*\in J_N$, the slice $\Sigma$ meets $N$ at $\theta(t_*)$, so a neighbourhood of
$t_*$ lies in $J_N$: $J_N$ is open. If $t_n\to t_*$ with $t_n\in J_N$, then for large $n$ the point $\theta(t_n)$ lies in
$\Sigma\cap N$, so $\Sigma\subseteq N$ and $t_*\in J_N$: $J_N$ is closed. Hence $J_N=J$.

\emph{(3) Discrete steps.} The update is $\theta'=\theta-\eta\,a(\theta)s$ with $\eta>0$ and $s\in S$ arbitrary. As $\eta$
and $s$ range, $-\eta\,a(\theta)s$ ranges over the linear space $\operatorname{im}a(\theta)=D_\theta$, so the possible next
iterates in $U$ form $(\theta+D_\theta)\cap U$. They all lie on the leaf $N$ through $\theta$ iff $(\theta+D_\theta)\cap
U\subseteq N$; this is the first equivalence. Call the leaves \emph{affine} when every leaf $N$ equals $(\theta+D_\theta)\cap
U$ for each $\theta\in N$. If the containment holds at every point of every leaf, then for each $\theta\in N$ a small
$k$-disc of the plane $\theta+D_\theta$ around $\theta$ lies in $N$. Take a flat chart around $\theta$ (\auxref{B.31}(b)) and
shrink the disc into it. The leaves form a foliation (\auxref{B.31}(c)), so $N$ meets the chart in countably many slices, on
each of which the last $\dim\Theta-k$ coordinates are constant. On the connected disc these coordinates are continuous and
take countably many values, so they are constant, and the disc lies in the slice through $\theta$, a $k$-dimensional integral
manifold, in which it is open by invariance of domain. Hence its tangent planes, all equal to $D_\theta$, are the planes
$D_x$ at its points, so $D$ is
locally constant along $N$, so constant on the connected $N$; every path in $N$ then has velocity in $D_\theta$, so
$N\subseteq\theta+D_\theta$, and with the containment $N=(\theta+D_\theta)\cap U$. The converse is clear. For an anchor that is constant over the period, $D_\theta=V$ is defined
on $U=\Theta$, sections of $D$ are $V$-valued functions whose derivatives and brackets are again $V$-valued, and the leaves
are the planes $\theta+V$; the condition holds, as \thmref{Proposition}{IV.2} uses.

\emph{(4) Frames.} A backward method under gradient flow ($\mathrm{opt}=\id$) moves the weight by
$\dot\theta=-a_tc_t\nabla L$; the forward method moves it by $\dot\theta=-D\rho_q\Lambda D\rho_q\T\nabla L$
(\thmref{Definition}{III.1}). We read ``for every loss'' as including time-dependent losses, as minibatch training produces.
Let both methods follow the same trajectory up to time $t_1$ for a loss $L$, and switch at $t_1$ to $L+\langle\gamma,\cdot\rangle$.
The trajectories still agree up to $t_1$, and their right derivatives at $t_1$ agree, so
$a_{t_1}c_{t_1}(\nabla L+\gamma)=K_{q(t_1)}(\nabla L+\gamma)$ for every covector $\gamma$, that is,
$a_{t_1}c_{t_1}=D\rho\,\Lambda\,D\rho\T$ at $q(t_1)$. This operator is symmetric positive semidefinite, with range in
$\operatorname{im}a=D$. If $a$ is injective, let $a^+=(a\T a)^{-1}a\T$, so $P=aa^+$ is the orthogonal projector onto
$\operatorname{im}a$, and put $M=a^+(ac)(a^+)\T$, which is symmetric positive semidefinite. Since $ac$ is symmetric with
range in $\operatorname{im}a$, $ac=P(ac)=(ac)P$, so $ac=P(ac)P=aMa\T$, and $a(c-Ma\T)=0$ gives $c=Ma\T$.

Now let $\rho$ be an immersion with $\dim Q=\rk D=k$. Then $D\rho_q\Lambda D\rho_q\T$ has rank $k$ and range in $D$, so
$\operatorname{im}D\rho_q=D_{\rho(q)}$, and $\rho$ maps small open sets of $Q$ diffeomorphically onto open pieces of leaves.
On $D$ the cometric $K=aMa\T$ is invertible as a map from $D^*$ to $D$, and its inverse is a metric $g_D$ on $D$:
$g_D(v,v)=\langle\xi,v\rangle$ whenever $K\xi=v$. For $u\in T_qQ$ put $v=D\rho_qu$. Since $D\rho_q$ is injective,
$D\rho_q\T$ is onto, so there is $\xi$ with $D\rho_q\T\xi=\Lambda^{-1}u$, and then $K\xi=D\rho_q\Lambda\Lambda^{-1}u=v$. Hence
\[g_D(D\rho_qu,D\rho_qu)=\langle\xi,D\rho_qu\rangle=\langle D\rho_q\T\xi,u\rangle=\langle\Lambda^{-1}u,u\rangle .\]
So $\rho$ is a local isometry from $Q$ with the constant metric $\Lambda^{-1}$, which is flat, onto the leaf with metric
$g_D=(aMa\T)^{-1}|_D$, and by \auxref{B.32}(a) the leaf metric is flat.

\emph{(5) Gauge fibres: $BB\T$.} Let $\rho(B)=BB\T$ on $Q=\R^{n\times n}$ with $\Lambda=I$, landing in $\Sym(n)$, and pair
covectors with $\Sym(n)$ by the trace. Then $D\rho_B(\dot B)=\dot BB\T+B\dot B\T$, and for symmetric $G$,
$\langle G,\dot BB\T+B\dot B\T\rangle=\langle2GB,\dot B\rangle$, so $D\rho_B\T G=2GB$ and
\[K(G)=D\rho_B(2GB)=2(G\theta+\theta G),\qquad\theta=BB\T .\]
This depends only on $\theta$, not on the point of the fibre $\{BR:R\in\Ort(n)\}$. On the positive definite matrices it is
invertible (\auxref{B.20}), so $D=\Sym(n)$ there. Put $a=K^{1/2}$ and $c=a\T$. The backward flow
$\dot\theta=-K(\theta)\nabla L$ and the forward flow pushed to $\theta$ solve the same ordinary differential equation with
smooth right-hand side, so they agree for every loss. The leaf metric $K^{-1}$ is a constant multiple of the
Bures--Wasserstein metric (\auxref{B.32}(b)), which is not flat for $n\ge2$. Here $\dim Q=n^2>\dim\Sym(n)$, so $\rho$ is no
immersion, in line with (4).

\emph{(6) Gauge fibres: balanced full-rank LoRA.} Take $m=n=r\ge2$, $s=1$ and equal block rates $\eta_A=\eta_B=1$, and
restrict to balanced factors, $B\T B=AA\T$, with $W=BA$ invertible; gradient flow preserves balancedness by the Noether
charge of \thmref{Theorem}{III.5}. From $B\T B=AA\T$, $W\T W=A\T(B\T B)A=(A\T A)^2$ and $WW\T=B(AA\T)B\T=(BB\T)^2$, so
$A\T A=\lvert W\rvert:=(W\T W)^{1/2}$ and $BB\T=\lvert W\T\rvert:=(WW\T)^{1/2}$, the unique positive definite square roots.
By \thmref{Observation}{III.2},
\[K_W(G)=G\,\lvert W\rvert+\lvert W\T\rvert\,G ,\]
a function of $W$ alone, invertible by \auxref{B.20}; as in (5) a backward method with $a=K^{1/2}$, $c=a\T$ reproduces the
flow, and the leaf is an open subset of $\GL_n$ with metric $g_W(X,X)=\langle K_W^{-1}X,X\rangle$. The transpose $\tau(W)=W\T$
is an isometry of $g$: $K_{W\T}(G\T)=(K_WG)\T$, so $K_{W\T}^{-1}(X\T)=(K_W^{-1}X)\T$ and
$g_{W\T}(X\T,X\T)=g_W(X,X)$. Its fixed points in $\GL_n$ are the invertible symmetric matrices, and the positive definite ones
form an open subset $F$ of them, an embedded submanifold on which $d\tau=\id$. A $g$-geodesic $\gamma$ with $\gamma(0)\in F$ and $\dot\gamma(0)$ tangent to $F$
is carried by $\tau$ to a geodesic with the same initial data (\auxref{B.38}(a)), so $\tau\circ\gamma=\gamma$ and
$\gamma$ stays in $F$ for small times. Every geodesic tangent to $F$ stays in $F$, so $F$ is totally geodesic, and by the
Gauss equation its intrinsic sectional curvatures equal those of $g$ on its tangent planes (\auxref{B.38}(b)). On $F$,
$\lvert W\rvert=\lvert W\T\rvert=W$, and the induced metric is $\langle L_W^{-1}X,X\rangle$ with $L_W(G)=GW+WG$ on $\Sym(n)$,
half the cometric of (5); this is a constant multiple of the Bures--Wasserstein metric, which has a plane of positive
sectional curvature for $n\ge2$ (\auxref{B.32}(b)). So $g$ has a plane of nonzero sectional curvature and is not flat.

\emph{(7) Discrete steps and horizontal lines.} A plain GD step of the forward method from $q$, with $g=\nabla L(\rho(q))$,
lands at $\rho(q-\eta v)$ with $v=\Lambda D\rho_q\T g$; the backward step lands at $\rho(q)-\eta\,a c\,g$. Agreement for small
$\eta$ forces $ac=D\rho_q\Lambda D\rho_q\T$ as in (4), so the backward step is $\rho(q)-\eta\,D\rho_qv$. Agreement for every
$\eta>0$, and for $-g$ as well as $g$ (another loss), says $\rho(q+tv)=\rho(q)+t\,D\rho_qv$ for all real $t$ in the domain:
$\rho$ is affine along the line $q+t\Lambda D\rho_q\T g$.

\emph{(8) Instances.} Within a period, GaLore's anchor is a fixed matrix $P_k$, LISA's a fixed coordinate inclusion, and
MeZO's the fixed vector $z_t$ for its single step; none depends on $\theta$, so (3) applies on $U=\Theta$ (vacuously when a
period is one step). Each pair has $c=a^*$ (\S\ref{proof:IV.1}), so with $\mathcal L=a$ it is the backward method of
\thmref{Proposition}{IV.2}, and under that proposition's optimizer hypotheses each period is a period of the linear forward
method $\theta_k+\mathcal LC$, merged at its end. For MeZO, $\mathcal L(\lambda)=\lambda z_t$, and the identification holds up
to the error of the central difference (\auxref{B.34}). Integrability holds for all of them, so it cannot tell them apart,
and by \thmref{Proposition}{IV.2} what distinguishes them from LoRA is how the projector changes and which state is
carried. Flora's update contains the full-rank term $(1-\beta)\hat g_t$, APOLLO's is $G\,\diag(s)$ with $G$ the full
gradient \citep{zhu2024apollo}, and Fira's adds a scaled multiple of $(I-P_tP_t\T)G$ \citep{chen2024fira}; for gradients with
$(I-P_tP_t\T)G\ne0$ each update leaves $\operatorname{im}P_t$, so none is of the displayed form with $a=P_t$.

\emph{(9) The GaLore field.} Let $L$ be smooth. On the open set where $\sigma_r(\nabla L(W))>\sigma_{r+1}(\nabla L(W))$,
let $\Pi(W)$ be the orthogonal projector onto $U_r(\nabla L(W))$, the top-$r$ eigenspace of $S(W)=\nabla L(W)\nabla L(W)\T$. It
is a smooth function of $W$ there, being a Riesz projector of $S(W)$ whose spectral gap persists (\auxref{B.37}). The fields
$X_E(W)=\Pi(W)E$, $E\in\R^{m\times n}$, span $D_W=\{Z:\operatorname{col}Z\subseteq U_r\}$. Write $\Pi'(Z)=D\Pi(W)[Z]$.
Differentiating $\Pi^2=\Pi$ gives $\Pi'\Pi+\Pi\Pi'=\Pi'$, hence $\Pi\Pi'\Pi=0$ and $(I-\Pi)\Pi'(I-\Pi)=0$. With
$N(Z):=(I-\Pi)\Pi'(Z)\Pi$ and the symmetry of $\Pi'(Z)$,
\begin{equation}\label{eq:C3-proj-deriv}
\Pi'(Z)=N(Z)+N(Z)\T ,
\end{equation}
so $\Pi'(Z)=0$ iff $N(Z)=0$. The bracket is $[X_E,X_F]=\Pi'(\Pi E)F-\Pi'(\Pi F)E$, and its component normal to $D$ is
\[(I-\Pi)[X_E,X_F]=N(\Pi E)\,\Pi F-N(\Pi F)\,\Pi E .\]
\emph{Criterion.} If $U_r$ is invariant along $D$, that is $\Pi'(Z)=0$ for $Z\in D$, the normal component vanishes and $D$ is
involutive. Conversely, let $D$ be involutive and $n\ge2$. For $u,u'\in U_r$ and $x,y\in\R^n$ linearly independent, take
$\Pi E=ux\T$ and $\Pi F=u'y\T$: involutivity gives $N(ux\T)u'\,y\T=N(u'y\T)u\,x\T$. The left side has its rows in
$\spanop\{y\}$ and the right side in $\spanop\{x\}$, so both vanish, and $N(ux\T)u'=0$. As $u'$ ranges over $U_r$ and
$N(Z)$ vanishes on $U_r^\perp$, $N(ux\T)=0$ for all $u\in U_r$ and $x\in\R^n$ (given $x$, pick $y$ independent of it), so
$N=0$ on $D$ by linearity, and $U_r$ is invariant along $D$. For right projection apply this to the transpose. The failure proved next covers every $1\le r<\min(m,n)$
with $m,n\ge2$; the remaining case with a gap is $n=r=1$, where $D_W=U_1$ is a line field, and a line field is always
involutive, since $[fX,gX]=(f\,Xg-g\,Xf)X$.

\emph{Failure for $L=\tfrac12\lVert W\rVert_F^2$.} Here $\nabla L(W)=W$. Let $p=\min(m,n)\ge2$, $1\le r<p$, and
$W=\sum_{i\le p}\sigma_ie_ie_i\T$ with $\sigma_1>\dots>\sigma_p>0$, so $U_r=\spanop\{e_1,\dots,e_r\}$ and $S=WW\T$ is
diagonal with entries $\lambda_i=\sigma_i^2$ ($i\le p$) and $0$. For a smooth symmetric family $S(t)$ with $S(0)=S$,
differentiating $[S(t),\Pi(t)]=0$ gives $[\dot S,\Pi]+[S,\dot\Pi]=0$; the $(i,j)$ entry with $i\le r<j$ reads
$-\dot S_{ij}+(\lambda_i-\lambda_j)\dot\Pi_{ij}=0$, and $\dot\Pi$ vanishes on the diagonal blocks by the identities above.
So $\dot\Pi_{ij}=\dot S_{ij}/(\lambda_i-\lambda_j)$ for $i\le r<j$. Take $Z=e_1x\T\in D$, so $\dot S=ZW\T+WZ\T$. Then
$\dot S_{1j}=x\T W\T e_j=\sigma_jx_j$ for $r<j\le p$, $\dot S_{1j}=0$ for $j>p$, and $\dot S_{ij}=0$ for $2\le i\le r<j$. Hence
\[N(e_1x\T)=\sum_{j=r+1}^{p}\frac{\sigma_jx_j}{\sigma_1^2-\sigma_j^2}\,e_je_1\T .\]
With $E=e_1e_{r+1}\T$ and $F=e_1e_1\T$ we get $N(\Pi F)=0$ and
\[(I-\Pi)[X_E,X_F]=N(e_1e_{r+1}\T)\,e_1e_1\T=\frac{\sigma_{r+1}}{\sigma_1^2-\sigma_{r+1}^2}\,e_{r+1}e_1\T\ne0 ,\]
so $D$ is not involutive at $W$. At $m=n=2$, $r=1$ and $W=\diag(2,1)$, the frame $X_i=X_{e_1e_i\T}$ gives the full bracket
$[X_1,X_2]=\Pi'(e_1e_1\T)e_1e_2\T-\Pi'(e_1e_2\T)e_1e_1\T=0-\tfrac13(e_1e_2\T+e_2e_1\T)e_1e_1\T=-\tfrac13e_2e_1\T$; the frame
$X_i=u(W)e_i\T$, with $u$ the top left singular vector, gives the same value, since $Du[e_1e_1\T]=0$ and
$Du[e_1e_2\T]=\tfrac13e_2$ at this $W$.

\emph{Genericity.} For least squares $L(W)=\tfrac12\lVert WX-Y\rVert_F^2$ the normal component of the bracket of two fixed
frame fields is a real-analytic function of $(W,X,Y)$ on the open set where the gap exists. It does not vanish identically
on the connected component containing $(W,I_n,0)$ with $W$ as above, where it is the value just computed, so it is nonzero
on an open dense subset of that component (\auxref{B.10}). The values for cross-entropy and for several samples are the
numerical checks reported with the proposition.

\emph{Single-sample least squares.} With one sample, $\nabla L(W)=(Wx-y)x\T$ has rank one and, where $Wx\ne y$,
$U_1=\spanop\{Wx-y\}$. Moving $W$ along $Z=uv\T$ with $u\in U_1$ changes $Wx-y$ by $u(v\T x)\in U_1$, so $U_1$ is constant
along $D$, and $D$ is involutive by the criterion.

\emph{Time-dependence.} Minibatch GaLore with $T=1$ draws a new minibatch, and MeZO a new $z_t$, at every step, so their
anchors are functions of $(t,\theta)$ and not of $\theta$ alone.
\end{proof}

\subsection{Observation IV.4}\label{proof:IV.4}

\begin{restate}{Observation}{IV.4}{backward reach and the backprop cone; an upper bound}
Assume reverse mode in which each box keeps its own residual, with no checkpointing across boxes, and take boxes at the
granularity where adapters are separate boxes (for LoRA-FA, $A_0$ and $B$ separately). A \emph{VJP residual} is an input,
an output, a sufficient statistic such as a ReLU mask, or a recipe to recompute one of these inside the box. Then it
suffices to keep a VJP residual for
\begin{itemize}
  \item each trainable box whose parameter-VJP depends on its input: linear and multiplicative boxes such as $W$, $A$, $B$
  and (IA)$^3$ scales; additive boxes such as biases and prompts need nothing;
  \item each \emph{nonlinear} box, bilinear boxes such as the attention products included, on a directed path from a
  trainable box to the loss.
\end{itemize}
Deterministic frozen linear boxes, and all boxes off this cone, need nothing; stochastic linear boxes on the cone
(dropout) need their mask or random-number state. Alternatively, the $k$ coordinates of $P^\top\nabla L$ can be computed
exactly by $k$ forward-mode JVPs, with no reverse pass and $O(1)$-layer activation memory, each JVP costing a small
constant multiple (about $1.5$ to $3$) of a forward pass. This always uses less activation memory than un-checkpointed
reverse mode, but it is time-competitive with one VJP (about $3$ forward passes) only for $k$ of about $1$ or $2$. It also
needs $P$ before the passes, so it cannot produce GaLore's SVD refresh, where $P$ depends on $\nabla L$, more cheaply than a
backward pass ($\nabla L$ itself would take $\lvert\theta\rvert$ JVPs).
\end{restate}

\begin{proof}
\emph{The computation.} The network is a directed acyclic graph of boxes. A box receives inputs $x$ (activations), may carry
parameters $p$, and produces an output $y=f(x,p)$; the loss is the output of the last box. Reverse mode propagates
cotangents backwards. For a box with output cotangent $\bar y$ it computes the input cotangents $\bar x=J_xf(x,p)\T\bar y$
and, for a trainable box, the parameter cotangent $\bar p=J_pf(x,p)\T\bar y$ (\bgref{B.17}). Call the \emph{cone} the set of
boxes on a directed path from a trainable box to the loss.

\emph{Pruning.} The parameter gradients are sums over paths from the parameters to the loss of products of Jacobians, so
only cotangents of wires on such paths contribute. A wire whose value does not depend on any trainable parameter carries a
cotangent that is never multiplied into a parameter gradient, so its cotangent need not be formed. Hence it suffices to
compute $\bar y$ for the outputs of boxes on the cone, $\bar x$ for those inputs of cone boxes that are produced on the
cone, and $\bar p$ for the trainable boxes. Boxes off the cone take no part in the backward sweep and need no residual.

\emph{What each box on the cone needs.} We list, for each kind of box, what the backward sweep reads besides $\bar y$ and
the weights, which are stored anyway.
\begin{itemize}
  \item A deterministic frozen linear or affine box, $y=Wx+b$: $\bar x=W\T\bar y$ does not involve $x$, so no residual.
  \item A dropout box, $y=\mathsf m\odot x/(1-p)$: $\bar x=\mathsf m\odot\bar y/(1-p)$, so it needs its mask, or the random
  state that regenerates it.
  \item A trainable linear box, $y=Wx$: $\bar W=\bar y\,x\T$, summed over tokens, so it needs its input $x$; $\bar x=W\T\bar
  y$ needs nothing more. A trainable multiplicative box, $y=\ell\odot x$: $\bar\ell=\sum\bar y\odot x$, so it needs $x$,
  and $\bar x=\ell\odot\bar y$.
  \item A trainable additive box, $y=x+b$, or a prompt, $y=[P;x]$: $\bar b=\sum\bar y$ and $\bar P$ is the slice of $\bar y$
  at the prompt positions, and $\bar x$ is $\bar y$ or its slice; no residual.
  \item A nonlinear box, $y=f(x)$, possibly with frozen parameters: $\bar x=J_f(x)\T\bar y$ depends on $x$ through
  $J_f(x)$, so it needs something from which $J_f(x)\T\bar y$ can be formed: the input, the output (for tanh, sigmoid or
  softmax), a mask (for ReLU), or a recipe to recompute one of them. A bilinear box is nonlinear in its joint input; for
  $y=a\odot b$, $\bar a=\bar y\odot b$ and $\bar b=\bar y\odot a$, and for the attention products $QK\T$ and $PV$ the same
  holds with matrix products.
\end{itemize}
A trainable nonlinear box lies on the cone, being on a path from itself to the loss, so it is covered by the second bullet,
and also by the first if its parameter-VJP reads its input. Processing the cone in reverse topological order, each box has
$\bar y$ available (it is a sum of input cotangents of its readers on the cone) together with its residual, so the sweep
computes every parameter gradient exactly. This proves that the residuals listed suffice. For LoRA-FA, $A_0$ is a frozen
linear box and needs nothing, while $B$ is a trainable linear box whose input is $A_0x$, so $B$ keeps $A_0x$.

\emph{The forward-mode route.} The $i$-th coordinate of $P\T\nabla L$ is $\langle p_i,\nabla L(\theta)\rangle=DL(\theta)[p_i]$,
the directional derivative along the $i$-th column of $P$. By \auxref{B.33}(a), forward mode computes it alongside one
evaluation of $L$ at a small constant multiple of that cost, carrying each tangent with its value and discarding both once
consumed. In a layered network this keeps the values and tangents of $O(1)$ layers alive at a time, whereas un-checkpointed
reverse mode keeps the residuals of every box on the cone until the backward sweep reaches it. The $k$ coordinates cost
$k$ such passes. One reverse pass costs about three forward passes, so with a per-JVP cost of $1.5$ to $3$ forward passes the
forward route is no slower only when $1.5k$ to $3k$ is at most about $3$, that is, for $k$ of about $1$ or $2$. The
directions $p_i$ must be fixed before the passes. GaLore's refresh computes $P$ from the SVD of the full gradient $\nabla L$
of each projected matrix, and $\nabla L$ by forward mode costs one pass per coordinate, $\lvert\theta\rvert$ passes, against
one reverse pass (\auxref{B.33}(b)).

\emph{MeZO.} The central difference estimates $DL(\theta)[z]$ with error $O(\varepsilon^2)$ when $L$ has a Lipschitz
Hessian, and is unbiased for the gradient of the Gaussian smoothing of $L$ (\auxref{B.34}).
\end{proof}

\subsection{Definition V.1}\label{proof:V.1}

\begin{restate}{Definition}{V.1}{base change and rebasing}
\begin{itemize}
  \item[(a)] A \emph{function-preserving} pointed reparametrisation $\phi:(\Theta',f',\theta_0')\to(\Theta,f,\theta_0)$,
  with $f\circ(\phi\times\id_X)=f'$, induces by postcomposition
  \[\phi_!:\Meth(\Theta',\theta_0')\to\Meth(\Theta,\theta_0),\qquad(Q,q_0,\rho)\mapsto(Q,q_0,\phi\circ\rho).\]
  \item[(b)] A map $\kappa:\Theta\to\Theta$ that does \emph{not} preserve the function, such as a quantiser, moves the
  base point. Its \textbf{pointing defect} is $e=\kappa\theta_0-\theta_0$.
  \item[(c)] A \textbf{rebasing scheme} (merge-and-restart) picks $\theta_{k+1}\in\Im M(\theta_k)$. Its reachable sets
  are $R_K$, the points $\theta_K$ it can reach in $K$ cycles, and $R_\infty=\bigcup_KR_K$.
\end{itemize}
\end{restate}

A definition needs no proof. We check that $\phi_!$ is a functor that preserves the realised functions, and record the
recursion for the reachable sets that the proofs of \thmref{Theorem}{V.3} use.

\begin{proof}[Verification of the claims in Definition V.1]
(a) Since $\phi(\theta_0')=\theta_0$, the map $\phi\circ\rho$ sends $q_0$ to $\theta_0$, so $\phi_!M$ is an object of
$\Meth(\Theta,\theta_0)$. A morphism $h:M\to N$ is a pointed smooth map with $\rho_N\circ h=\rho_M$
(\thmref{Definition}{I.2}); then $(\phi\circ\rho_N)\circ h=\phi\circ\rho_M$, so the same $h$ is a morphism
$\phi_!M\to\phi_!N$. Setting $\phi_!h=h$ preserves identities and composition, so $\phi_!$ is a functor, the
postcomposition functor of slice categories (\bgref{A.19}). It changes neither the parameter space nor the realised
functions: for $q\in Q$,
\[f\big(\phi(\rho(q)),x\big)=f'\big(\rho(q),x\big)\qquad\text{for all }x\in X,\]
so $\phi_!M$ computes exactly the functions $M$ computes, and $\Im\phi_!M=\phi(\Im M)$.

(b) is a definition. (c) Put $R_0=\{\theta_0\}$. Each cycle trains the method $M(\theta_k)$ over the current base and
merges, so the points reachable in $K+1$ cycles are
\[R_{K+1}=\bigcup_{\theta\in R_K}\Im M(\theta).\]
When every $M(\theta)$ is pointed at $\theta$, that is $\rho_\theta(q_0)=\theta$, we have $\theta\in\Im M(\theta)$, so
$R_K\subseteq R_{K+1}$ and $R_\infty$ is an increasing union.
\end{proof}

\subsection{Proposition V.2}\label{proof:V.2}

\begin{restate}{Proposition}{V.2}{quantisation and splitting}
Write $W$ for the full-precision weight, $\kappa$ for the quantiser, $e=\kappa W-W$, $\delta_0=\delta(q_0)$, and $s$ for the
LoRA scale $\alpha/r$ ($\alpha/\sqrt r$ under rsLoRA).

(a) Additive families commute with translation, $M(\theta+v)=\tau_v\circ M(\theta)$. Hence QLoRA is LoRA pointed at
$\kappa\theta_0$, with defect $e$, and it needs no new weight-space structure. Since $-e$ is generically of full rank,
$\theta_0\notin\Im\mathrm{LoRA}_r(\kappa\theta_0)$ unless $\rk e\le r$.

(b) For splitting families derived from an SVD or from data, the order of operations matters, and three orders occur in
practice.
\begin{itemize}
  \item \emph{Split, then quantise the residual} (QPiSSA; HF's recommended PiSSA and CorDA workflow; the quantisation
  half-step of LoftQ for $t\ge2$, with $W_{\rm res}=W-\delta(q_{t-1})$). The defect is $\kappa(W_{\rm res})-W_{\rm res}$.
  \item \emph{Quantise, then fit the error} (LoftQ with one iteration, HF's default \texttt{loftq\_iter=1}; HF
  \texttt{replace\_\allowbreak lora\_\allowbreak weights\_\allowbreak loftq}). HF sets the factors from
  $\operatorname{SVD}_r(W-\kappa W)$ without dividing by $s$, unlike its PiSSA and OLoRA initialisations, so
  $\delta_0=-s\operatorname{SVD}_r(e)$ and the defect is $e-s\operatorname{SVD}_r(e)$, with
  \[\lVert e-s\operatorname{SVD}_r(e)\rVert^2=\lVert e-\operatorname{SVD}_r(e)\rVert^2+(1-s)^2\lVert\operatorname{SVD}_r(e)\rVert^2.\]
  The defect is $e$ minus its best rank-$r$ approximation (of rank $\rk e-r$) iff $s=1$, which is HF's bare default
  $r=\alpha=8$. For $s=2$ its norm equals $\lVert e\rVert$, no better than QLoRA. With a callback,
  \texttt{replace\_\allowbreak lora\_\allowbreak weights\_\allowbreak loftq} can roll a module back to the zero
  initialisation, whose defect is $e$.
  \item \emph{Quantise, split, re-quantise} (HF \texttt{olora\_init} on bitsandbytes weights). The defect is
  $e+\kappa(r')-r'$ with $r'=\kappa W-\delta_0$.
\end{itemize}
Quantise-then-split has defect exactly $e$ iff the frozen part equals $\kappa W-\delta_0$, for instance $\kappa W$ kept
quantised minus a frozen copy of the rank-$r$ $\delta_0$, at a cost of about $r(m+n)$ extra numbers. Re-quantising the
residual, as HF \texttt{olora\_init} does, generically adds $\kappa(r')-r'$.

Action families whose pointing does not depend on the base (OFT at $R=I$, BOFT at $R=I$ with unit scale, HiRA at $B=0$),
evaluated at $\kappa\theta_0$, simply carry the defect $e$. Methods whose pointing is computed from the base have an order
ambiguity of their own. DoRA's magnitude initialised from $\kappa\theta_0$ gives defect $e$, while initialised from
$\theta_0$ it gives $\diag\big(\lVert w_{0,i}\rVert/\lVert(\kappa\theta_0)_i\rVert\big)\,\kappa\theta_0-\theta_0$. For
action families a base change also moves the absolute invariants, along with the base point. NF4's exact zero code
enlarges the zero set that HiRA must preserve, and exactly orthogonal (exact-Cayley) OFT preserves the Gram matrix of
$\kappa\theta_0$ rather than that of $\theta_0$; HF's default Cayley--Neumann OFT preserves it only approximately.

(c) LoftQ alternates two updates on $\lVert\theta_0-Q-LR\rVert$, an exact rank-$r$ SVD step and a round-to-nearest
quantisation step with blockwise scales recomputed from its input. The second is not an exact projection, so the
objective need not decrease monotonically. With $s=1$ the output is pointed up to the attained residual. It is then an
\emph{approximate section}, a choice $(Q,q_0)$ with reduced $\lVert\rho(q_0)-\theta_0\rVert$. In general, with $Q_T$ and
the low-rank fit $L_TR_T$ after $T$ iterations, HF's deployed defect is $(Q_T+L_TR_T-W)+(s-1)L_TR_T$, because HF's internal
residual also omits $s$. The official LoftQ checkpoints use $r=64$ and $\alpha=16$, so $s=1/4$, and then the deployed
defect can grow with $T$ while the attained residual falls.

(d) For a generic high-precision adapter $\Delta$, fine-tune-then-quantise and quantise-then-fine-tune do not commute. The
first lands on the quantisation grid. In the second, the exactly merged weight $\kappa\theta_0+\Delta$ leaves the
quantised type. For a uniform fixed-scale INT-$k$ grid, $\kappa\theta_0+\Delta$ stays on the grid iff $\Delta$ lies on
the scale lattice (within range), so there the obstruction is the off-grid update. NF4 is non-uniform and not closed under
addition, and absmax grids whose scales are recomputed are not closed either, so for QLoRA's NF4 the grid itself is part of
the obstruction. Type-preserving merges, such as HF PEFT's dequantise--add--requantise for bitsandbytes 4- and 8-bit
layers, re-quantise with recomputed block scales, so they stay in the quantised type but are lossy. Under affine group
quantisation QA-LoRA is the structured exception. Its group-pooled adapter is constant on each quantisation group and
merges into the zero-points, so the merged model stays INT-$k$.
\end{restate}

\begin{proof}
\emph{Quantisers.} We use two properties of the blockwise quantisers in practice (NF4 and INT-$k$, with a fixed scale or
the absmax scale of each block, \citealp{dettmers2023qlora,dettmers20218}). First, $\kappa$ rounds each entry, divided by
its block's scale, to the nearest point of a finite codebook and multiplies back, so its values, the \emph{representable}
matrices, form a Lebesgue-null set: a finite set for fixed scales, and for absmax scales a finite union of sets with one
free scale per block of $\beta\ge2$ entries. Second, $\kappa$ is idempotent, $\kappa(\kappa X)=\kappa X$, and it fixes
exactly the representable matrices.

(a) An additive family has $\rho_\theta(q)=\theta+\delta(q)$ with $\delta$ independent of the base and $\delta(q_0)=0$
(\thmref{Definition}{I.2}). Then $\rho_{\theta+v}(q)=v+\rho_\theta(q)=\tau_v(\rho_\theta(q))$, which is the first claim.
QLoRA trains zero-initialised LoRA over the frozen base $\kappa\theta_0$, so it is $\mathrm{LoRA}_r(\kappa\theta_0)$, with
$\rho(q_0)=\kappa\theta_0$ and defect $\rho(q_0)-\theta_0=e$. Since every matrix of rank at most $r$ has the form $sBA$
($s\ne0$), $\theta_0=\kappa\theta_0+sBA$ for some $(B,A)$ iff $\rk(-e)\le r$.

\emph{The word ``generically''.} For a fixed-scale quantiser, $\kappa$ is constant on each open cell of a decomposition of
$\Theta$ into countably many cells and a null set of boundaries, so on a cell $e=K-W$ with $K$ constant. The rank-deficient
matrices are the common zeros of the maximal minors, a proper algebraic subset, and $W\mapsto K-W$ is a bijection, so $e$
has full rank for almost every $W$ in each cell, hence for almost every $W$ (\auxref{B.10}). For an absmax quantiser with
blocks of $\beta\ge2$ consecutive entries of a row, take the cells on which every code and the position of every block
maximum are constant. With the scale stored exactly, the block maximum is represented exactly, so $e$ vanishes at those positions, and each other
entry is $e_{ij}=a\,c_{ij}-W_{ij}$ with $a$ the scale of its block and $c_{ij}$ its code; varying $W_{ij}$ alone moves
$e_{ij}$ alone. So on each cell $e$ fills an open subset of the space $Z$ of matrices vanishing at the block maxima, and its
rank is, for almost every $W$ in the cell, the largest rank in $Z$. A row of length $n$ has at most $\lceil n/\beta\rceil+1$
forced zeros, so $Z$ contains a matrix of rank $r+1$ whenever $r+1\le n-\lceil n/\beta\rceil-1$ and $r<m$ (place $r+1$
nonzero entries in distinct rows and columns, avoiding the forced zeros, greedily). Hence $\rk e>r$, and
$\theta_0\notin\Im\mathrm{LoRA}_r(\kappa\theta_0)$, for almost every $W$ in this range. Full rank holds for almost every
$W$ in each cell whose zero pattern admits a matrix of full rank; the other cells, in which the block maxima of many rows
fall in a few common columns, have positive but tiny measure (for instance, a column of zeros needs the maximum of its
block in every row, probability $\beta^{-m}$ per column for weights with independent continuous entries).

(b) In each pipeline the initial weight is $\rho(q_0)=F+\delta_0$, where $F$ is the frozen part and $\delta_0$ the initial
adapter, and the defect is $F+\delta_0-W$.
\begin{itemize}
  \item \emph{Split first.} $F=\kappa(W_{\rm res})$ with $W_{\rm res}=W-\delta_0$, so the defect is
  $\kappa(W_{\rm res})+\delta_0-W=\kappa(W_{\rm res})-W_{\rm res}$. In LoftQ's quantisation half-step for $t\ge2$ the same
  computation applies with $\delta_0$ replaced by the current fit $\delta(q_{t-1})=L_{t-1}R_{t-1}$.
  \item \emph{Quantise, then fit.} $F=\kappa W$ and HF sets $L R=\operatorname{SVD}_r(W-\kappa W)=-\operatorname{SVD}_r(e)$
  (the best rank-$r$ approximation is odd, $\operatorname{SVD}_r(-X)=-\operatorname{SVD}_r(X)$), and the forward pass
  multiplies by $s$, so $\delta_0=-s\operatorname{SVD}_r(e)$ and the defect is $e-s\operatorname{SVD}_r(e)$. Write
  $P=\operatorname{SVD}_r(e)$. By \auxref{B.11}(b), $P$ and $e-P$ are Frobenius-orthogonal, so
  $\lVert e-sP\rVert^2=\lVert(e-P)+(1-s)P\rVert^2=\lVert e-P\rVert^2+(1-s)^2\lVert P\rVert^2$. For $e\ne0$, $P\ne0$, so the
  defect equals $e-P$ iff $s=1$; then it has rank $\rk e-r$ when $\rk e\ge r$, being the sum of the singular terms beyond
  the $r$-th. HF's defaults $r=\alpha=8$ give $s=\alpha/r=1$. For $s=2$, $\lVert e-2P\rVert^2=\lVert e-P\rVert^2+\lVert
  P\rVert^2=\lVert e\rVert^2$, again by orthogonality; in general $\lVert e-sP\rVert<\lVert e\rVert$ iff $(1-s)^2<1$, that is
  $0<s<2$. A module rolled back to the zero initialisation has $\delta_0=0$ and defect $\kappa W-W=e$.
  \item \emph{Quantise, split, re-quantise.} The split is computed from $\kappa W$, the residual $r'=\kappa W-\delta_0$ is
  re-quantised, and $F=\kappa(r')$. The defect is $\kappa(r')+\delta_0-W=(\kappa(r')-r')+(\kappa W-W)=e+\kappa(r')-r'$.
\end{itemize}
For quantise-then-split with frozen part $F$, the defect $F+\delta_0-W$ equals $e=\kappa W-W$ iff $F=\kappa W-\delta_0$.
Storing $\kappa W$ quantised together with the frozen factors of $\delta_0=sL_0R_0$, which hold $r(m+n)$ numbers, realises
this $F$ exactly. Re-quantising instead adds $\kappa(r')-r'$, which vanishes iff $r'$ is representable, that is, iff $r'$
lies in the null set of representable matrices.

\emph{Action families.} OFT at $R=I$ gives $\rho(q_0)=\kappa\theta_0\,I=\kappa\theta_0$, BOFT at $R=I$ with unit scale gives
$\diag(\mathbf 1)\,\kappa\theta_0\,I=\kappa\theta_0$, and HiRA at $B=0$ gives $\kappa\theta_0+\kappa\theta_0\odot(0\cdot
A)=\kappa\theta_0$; each has defect $e$. DoRA at $B=0$ computes $\rho(q_0)=\diag\big(\mu_{0,i}/\lVert(\kappa\theta_0)_i\rVert\big)
\kappa\theta_0$, normalising the rows of its frozen base $\kappa\theta_0$ (assumed nonzero) and rescaling them by the
magnitude vector $\mu_0$. If $\mu_0$ holds the row norms of $\kappa\theta_0$, then $\rho(q_0)=\kappa\theta_0$ and the defect
is $e$; if it holds those of $\theta_0$, the defect is the displayed matrix.

\emph{Moved invariants.} HiRA's update $W\mapsto W\odot(\mathbf 1+BA)$ preserves every zero entry of its base
(\thmref{Proposition}{II.9}(e)). NF4 contains the code $0$, so every entry of $\theta_0$ whose ratio to its block's absmax
rounds to $0$ becomes an exact zero of $\kappa\theta_0$, while zero entries of $\theta_0$ stay zero; the zero set of
$\kappa\theta_0$ therefore contains that of $\theta_0$, and generically more. Exact-Cayley OFT has
$\rho(Q)\rho(Q)\T=\kappa\theta_0\,C(Q)C(Q)\T\kappa\theta_0\T=\kappa\theta_0\kappa\theta_0\T$ (\thmref{Corollary}{II.6}). HF's
default factor is $R=C(Q)(I-Q^4)$ (\auxref{B.23}(e)), so $\rho\rho\T=\kappa\theta_0(I-Q^4)^2\kappa\theta_0\T=
\kappa\theta_0\kappa\theta_0\T+O(\lVert Q\rVert^4)$.

(c) LoftQ starts from $L_0R_0=0$ and, for $t\ge1$, sets $Q_t=\kappa(\theta_0-L_{t-1}R_{t-1})$ and
$L_tR_t=\operatorname{SVD}_r(\theta_0-Q_t)$. By \auxref{B.11}(b) the SVD step minimises $\lVert\theta_0-Q_t-LR\rVert$ over
rank-$r$ products, so it never increases the objective. The quantisation step returns $\kappa(y)$ for $y=\theta_0-L_{t-1}
R_{t-1}$, which need not be the representable matrix nearest to $y$, and the objective can increase. An example with blocks
equal to the rows of a $2\times3$ matrix, absmax scales, the codebook $\{0,\pm\tfrac12,\pm1\}$ and $r=1$:
\[\theta_0=\begin{pmatrix}1&1&-\tfrac54\\ \tfrac54&1&-1\end{pmatrix},\quad
Q_1=\kappa(\theta_0)=\tfrac54\begin{pmatrix}1&1&-1\\1&1&-1\end{pmatrix},\quad
\theta_0-Q_1=\tfrac14\begin{pmatrix}-1&-1&0\\0&-1&1\end{pmatrix}.\]
Here $(\theta_0-Q_1)(\theta_0-Q_1)\T=\tfrac1{16}\left(\begin{smallmatrix}2&1\\1&2\end{smallmatrix}\right)$ has eigenvalues
$\tfrac3{16}$ and $\tfrac1{16}$, with top eigenvector $(1,1)/\sqrt2$, so $L_1R_1=\tfrac12\left(\begin{smallmatrix}1&1\\1&1
\end{smallmatrix}\right)(\theta_0-Q_1)=\tfrac18\left(\begin{smallmatrix}-1&-2&1\\-1&-2&1\end{smallmatrix}\right)$ and the
objective after the SVD step is $\lVert\theta_0-Q_1-L_1R_1\rVert^2=\tfrac1{16}$. The next quantisation step acts on
$y=\theta_0-L_1R_1=\tfrac18\left(\begin{smallmatrix}9&10&-11\\11&10&-9\end{smallmatrix}\right)$. Each row has absmax
$\tfrac{11}8$, and the ratios $\tfrac9{11}$ and $\tfrac{10}{11}$ round to $1$, so $Q_2=\tfrac{11}8\left(\begin{smallmatrix}1&1&-1\\1&1&-1
\end{smallmatrix}\right)$ and
\[\lVert\theta_0-Q_2-L_1R_1\rVert^2=\lVert y-Q_2\rVert^2=2\Big(\tfrac1{16}+\tfrac1{64}\Big)=\tfrac5{32}>\tfrac1{16}.\]
The objective rose, and $Q_1$, which is representable, is closer to $y$ than $\kappa(y)$. The deployed weight after $T$
iterations is $Q_T+sL_TR_T$, because HF stores the fitted factors and the forward pass multiplies by $s$; its defect is
\[Q_T+sL_TR_T-W=(Q_T+L_TR_T-W)+(s-1)L_TR_T .\]
For $s=1$ this is minus the attained residual, so the output is pointed up to that residual. For $s\ne1$ the second term
$(s-1)L_TR_T$ is proportional to the fit itself and does not shrink with the residual; in the referee simulation of the official configuration ($r=64$, $\alpha=16$, $s=1/4$,
five iterations, a Gaussian $512\times512$ NF4 weight) the attained residual falls to $0.67\lVert e\rVert$ while the deployed
defect rises from $0.91\lVert e\rVert$ to $0.96\lVert e\rVert$ (Appendix~\ref{app:repro}).

(d) Fine-tuning and then quantising gives $\kappa(\theta_0+\Delta)$, a value of $\kappa$. Exact merging after quantising gives
$\kappa\theta_0+\Delta$, which is representable iff $\Delta$ lies in the translate of the representable set by
$-\kappa\theta_0$, a null set; so for generic $\Delta$ it leaves the quantised type and the two orders differ. For a uniform
grid with a fixed scale $a$ per block, the representable matrices are $a$ times integer codes in range, so
$\kappa\theta_0+\Delta$ is representable iff $\Delta=a(c'-c_0)$ with $c'$ in range, that is, iff $\Delta$ lies on the scale
lattice. NF4 is not closed under addition. Its smallest positive code is $q_1\approx0.0796$, and $2q_1\approx0.1592$ is not a
code (the next codes are $0.0796$ and $0.1609$). The blocks $(1,q_1,0,\dots)$ and $(0,q_1,0,\dots)$ are representable,
with absmax scales $1$ and $q_1$, but their sum $(1,2q_1,0,\dots)$ has absmax $1$ and needs the code $2q_1$. Absmax grids
with uniform codes of step $h$ are not closed either: $(1,h,0,\dots)$ and $(h,0,0,\dots)$ are representable, while their sum
$(1+h,h,0,\dots)$ would need the code $h/(1+h)$, which is not a multiple of $h$ for $0<h<1$. A type-preserving merge
returns $\kappa(\kappa\theta_0+\Delta)$, which is representable and differs from $\kappa\theta_0+\Delta$ whenever the latter
is not, so it is lossy for generic $\Delta$. Finally, affine group quantisation stores each group $g$ as $a_gc_g+\beta_g\mathbf
1$, with integer codes $c_g$ and a high-precision scale $a_g$ and zero-point $\beta_g$. QA-LoRA pools its input over the
quantisation groups, so its update is constant on each group, $\Delta_g=\delta_g\mathbf 1$ \citep{xu2023qa}, and
$a_gc_g+\beta_g\mathbf 1+\delta_g\mathbf 1=a_gc_g+(\beta_g+\delta_g)\mathbf 1$ keeps the codes and moves only the zero-point.
\end{proof}

\subsection{Theorem V.3}\label{proof:V.3}

\begin{restate}{Theorem}{V.3}{rebasing closure}
(a) \emph{Additive methods with a fixed shape $C\ni0$}, with frames, masks and compression operators frozen at $\theta_0$
across cycles. Then $R_K=\theta_0+C^{+K}$, the $K$-fold Minkowski sum, and $R_\infty=\theta_0+\operatorname{Mon}(C)$, the
additive monoid generated by $C$. If $C$ is closed under real scalings, $R_\infty=\theta_0+\spanop C$.

(b) \emph{Action methods}, where one cycle applies an element of a set $S$ with $1\in S\subseteq G$; for scaling methods,
$S\subseteq\operatorname{End}(\Theta)$ is a subset containing the identity. Then $R_K=S^K\theta_0$, and $R_\infty$ is the
orbit of the monoid generated by $S$. Suppose $S$ lies in the subgroup generated by connected Lie subgroups
$H_1,\dots,H_p$ and contains a neighbourhood of $1$ in each. Then the monoid generated by $S$ is the connected Lie subgroup
$H$ with $\operatorname{Lie}(H)=\operatorname{Lie}\langle\mathfrak h_1,\dots,\mathfrak h_p\rangle$.

(c) \emph{Block-orthogonal factors.} Let the factors have coordinate blocks $I_j\subseteq\{1,\dots,n\}$ (block OFT, BOFT
factors, Givens planes). Then $\operatorname{Lie}\langle\so(I_j)\rangle_j=\bigoplus_C\so(C)$, summed over the connected
components $C$ of the hypergraph $(\{1,\dots,n\},\{I_j\})$.

\emph{Corollaries.}
\begin{enumerate}[label=\arabic*.,leftmargin=1.6em]
  \item \textbf{ReLoRA.} $R_K=W_0+\MM_{\le\min(Kr,m,n)}$. It reaches full fine-tuning iff $K\ge\lceil\min(m,n)/r\rceil$.
  \item \textbf{Idempotent methods} gain nothing from rebasing. These are the methods whose shape is exactly a subspace,
  subgroup or submonoid independent of the base, with frames, masks and compression operators frozen across cycles:
  LoRA-FA with a fixed $A$, LoRA-XS with frozen frames, MoRA with fixed operators, FourierFT with fixed indices, WaveFT,
  C3A, MiSS (default balance mode; bat mode is base-dependent but still idempotent, since its blockwise maps
  $W\mapsto(W+I)(I+M)-I$ form a monoid), SHiRA, BitFit, FISH with a fixed mask, LN-tuning, (IA)$^3$, SSF and RoAd. Block OFT
  with a fixed partition and an exact Cayley map gains only the measure-zero set of block rotations with an eigenvalue
  $-1$, so $R_\infty=W_0\cdot\prod\SO(b)=\overline{R_1}$. Constrained OFT (COFT, an $\varepsilon$-ball) with an exact Cayley
  map gains up to $\prod\SO(b)$; HF's \texttt{coft=True} keeps the Cayley--Neumann map by default and so falls under
  corollary~4. Methods that recompute frames or masks from the merged weight (an SVD-recomputing LoRA-XS, ReMoRA's
  alternating operators, resampled indices) do gain.
  \item \textbf{VeRA.} $\mathrm{VeRA}^\infty=\text{LoRA-FA}(A)$ whenever the frozen $B$ has no zero entries.
  \item \textbf{OFT family.} Block OFT with an exact Cayley map is stuck in $W_0\cdot\prod\SO(b)$. HF's default
  Cayley--Neumann factor is $R=C(Q)(I-Q^4)$, with $C$ the exact Cayley map, so $R\T R=(I-Q^4)^2$ and $R$ is not orthogonal.
  Its singular values are $\lvert1-\mu_k^4\rvert$, where $\pm i\mu_k$ are the eigenvalues of $Q$. In the operating regime
  $\lVert Q\rVert_{\rm op}\le2^{1/4}$, every factor is a contraction. Re-merging then decreases $K_{\rm out}$ in the
  Loewner order and no singular value grows, so the drift is monotone shrinkage, and the reachable set lies in $W_0$ times
  the block contraction semigroup. With unconstrained $Q$ the factors include singular ones ($\mu_k=1$), and re-merging
  generates $\prod\big(\GL^+(b)\cup\MM_{\le b-2}\big)$, or $\prod\big(\CO^+(2)\cup\{0\}\big)$ when $b=2$; this clause is
  established at the level of a proof sketch. Iterated BOFT (exact Cayley in HF) with $m'$ butterfly factors
  of block size $b$ reaches $\diag(\R^m)\cdot W_0\cdot P\big(\prod\SO(b\,2^{m'-1})\big)P\T$, because the factor hypergraph
  has components of size $b\,2^{m'-1}$. For injective $W_0$ this is $\diag(\R^m)\cdot W_0\cdot\SO(n)$ iff
  $b\,2^{m'-1}=n$, the full-depth butterfly $m'=1+\log_2(n/b)$. HF's default \texttt{boft\_\allowbreak n\_\allowbreak butterfly\_\allowbreak factor=1} is
  block OFT times $\diag(s)$. A \emph{single} full-depth BOFT has dimension at most $m'\tfrac nb\binom b2$, which is below
  $\dim\SO(n)$ when $m'(b-1)<n-1$; its matrices are dense, while the set they form has positive codimension.
  \item \textbf{Re-HRA} (HF \texttt{apply\_GS=False}), with $W_0$ injective and $r$ even. Then
  $R_K=(W_0+\MM_{\le Kr})\cap W_0\cdot\SO(n)$, so rebasing commutes with the meet of \thmref{Theorem}{II.7}. This
  multiplicative twin of ReLoRA reaches $W_0\cdot\SO(n)$ in at most $\lceil n/r\rceil$ cycles.
  \item \textbf{GaLore (with a gradient-SVD or a random projector), LISA and MeZO}, under the optimizer hypotheses of
  \thmref{Proposition}{IV.2}, are rebasing schemes of \emph{linear} methods whose shape moves. That is why, as a set over
  the sequences of shapes they can choose, they reach all of $\Theta$, while a fixed linear shape $L$ reaches only
  $\theta_0+L$ (corollary~2). Low rank is the per-period obstruction only for GaLore; LISA and MeZO already make full-rank
  per-matrix updates and gain coverage. Flora proper takes full-space steps with a compressed momentum, so it has no
  per-period low-rank shape.
  \item \textbf{Merging never rescues an exactly orthogonal method from its absolute invariants.} In exact arithmetic,
  every exactly orthogonal input-side rebasing preserves $K_{\rm out}$ and the spectrum, and every output-side one
  preserves $K_{\rm in}$ and the spectrum.
\end{enumerate}
\end{restate}

\begin{proof}
(a) With frames, masks and compression operators frozen, every cycle trains the same additive shape over the current
base: $\Im M(\theta)=\theta+C$ for every $\theta$. By the recursion of \S\ref{proof:V.1}, $R_1=\theta_0+C$ and
$R_{K+1}=\bigcup_{\theta\in R_K}(\theta+C)=R_K+C$, so $R_K=\theta_0+C^{+K}$ by induction. Since $0\in C$, the sets $C^{+K}$
increase, and their union is the additive monoid $\operatorname{Mon}(C)$ generated by $C$ (\bgref{D.10}); hence
$R_\infty=\theta_0+\operatorname{Mon}(C)$. If $\lambda C\subseteq C$ for every real $\lambda$, then $-C=C$, so
$\operatorname{Mon}(C)$ is closed under negation and is an additive subgroup, and
$\lambda(x_1+\dots+x_K)=\lambda x_1+\dots+\lambda x_K$ shows that it is closed under scaling. So it is a linear subspace
containing $C$ and contained in $\spanop C$, and equals $\spanop C$.

(b) Each cycle applies one element of $S$ to the current point, so $\Im M(\theta)=S\theta$, and the recursion gives
$R_K=S^K\theta_0$. Since $1\in S$, the sets $S^K$ increase and their union is the monoid $\operatorname{Mon}(S)$ generated
by $S$, so $R_\infty=\operatorname{Mon}(S)\,\theta_0$, the orbit of $\theta_0$ under that monoid. For the second statement,
let $U_i\subseteq S$ be a neighbourhood of $1$ in $H_i$. Then $U_i\cap U_i^{-1}$ is a symmetric neighbourhood of $1$ in the
connected Lie group $H_i$, so it generates $H_i$ as a monoid (\auxref{B.24}(a)), and $H_i\subseteq\operatorname{Mon}(S)$.
Each $H_i$ is a group, so the monoid generated by $H_1\cup\dots\cup H_p$ is closed under inverses and equals the subgroup
they generate, which is the connected Lie subgroup $H$ with $\operatorname{Lie}(H)=\operatorname{Lie}\langle\mathfrak
h_1,\dots,\mathfrak h_p\rangle$ (\auxref{B.24}(d)). So $H\subseteq\operatorname{Mon}(S)$. Conversely $S\subseteq H$ by
hypothesis, and $H$ is closed under products, so $\operatorname{Mon}(S)\subseteq H$.

(c) Write $L_{ab}=E_{ab}-E_{ba}$, so $\so(I)=\spanop\{L_{ab}:a,b\in I\}$. For distinct $a,b,c$,
\[L_{ac}L_{cb}=E_{ab},\qquad L_{cb}L_{ac}=E_{ba},\qquad\text{so}\qquad[L_{ac},L_{cb}]=L_{ab},\]
while $L_{ab}L_{cd}=L_{cd}L_{ab}=0$ when $\{a,b\}\cap\{c,d\}=\emptyset$. Let $\mathfrak g=\operatorname{Lie}\langle\so(I_j)
\rangle_j$. If $a\ne b$ lie in one component $C$, join them by a shortest chain $a=c_0,c_1,\dots,c_t=b$ of distinct
coordinates with consecutive ones in a common block. By induction on $t$, $L_{ab}\in\mathfrak g$: for $t=1$,
$L_{ab}\in\so(I_j)$ for a block containing both; for $t>1$, $L_{ac_{t-1}}\in\mathfrak g$ by induction and $L_{c_{t-1}b}\in
\mathfrak g$, and their bracket is $L_{ab}$. So $\bigoplus_C\so(C)\subseteq\mathfrak g$. Conversely, $\bigoplus_C\so(C)$ is a
Lie subalgebra, since each $\so(C)$ is one and elements of different components have disjoint supports and commute, and it
contains every $\so(I_j)$, because each block lies in one component. Hence $\mathfrak g=\bigoplus_C\so(C)$, of dimension
$\sum_C\binom{\lvert C\rvert}2$.

\emph{Corollary 1.} LoRA's shape $\{sBA\}$ is $\Mr$, which contains $0$ and is closed under real scalings, so (a) applies.
By \auxref{B.12}(a), a sum of $K$ matrices of rank at most $r$ has rank at most $\min(Kr,m,n)$. Conversely, a matrix of rank
$k\le\min(Kr,m,n)$ is a sum of $k$ rank-one terms of its SVD (\auxref{B.11}(a)), which split into $K$ groups of at most $r$
terms, each in $\Mr$. So $C^{+K}=\MM_{\le\min(Kr,m,n)}$. This is all of $\R^{m\times n}$ iff $Kr\ge\min(m,n)$, that is,
iff $K\ge\lceil\min(m,n)/r\rceil$.

\emph{Corollary 2.} If the shape is an additive submonoid $C$ independent of the base (in particular a linear subspace),
then $C^{+K}=C$ and $R_K=R_1$ for all $K$ by (a); if it is a submonoid $S$ of a group or of $\operatorname{End}(\Theta)$,
then $S^K=S$ and $R_K=R_1$ by (b). The listed methods are of this kind once their frames, masks and operators are frozen:
the updates of LoRA-FA ($\{XA\}$), LoRA-XS ($\{UMV\T\}$), MoRA (linear in its square matrix), FourierFT and WaveFT (fixed
coefficient positions), C3A (block-circulant matrices), MiSS in balance mode (linear in its shared block), SHiRA, BitFit,
FISH and LN-tuning (fixed coordinates) form linear subspaces; the scalings of (IA)$^3$, $W\mapsto\diag(\ell)W$, form a
monoid in $\operatorname{End}(\Theta)$; SSF's maps $y\mapsto s\odot y+t$ form a monoid under composition,
$s'\odot(s\odot y+t)+t'=(s's)\odot y+(s't+t')$; RoAd's $2\times2$ blocks range over the scaled rotations or over all
$2\times2$ matrices, both monoids; and MiSS's bat-mode maps satisfy $(W+I)(I+M)(I+M')-I$ with $(I+M)(I+M')=I+M''$,
$M''=M+M'+MM'$ again in the class, so they form a monoid.

For block OFT with an exact Cayley map and a fixed partition, $S=\prod\mathrm{Cay}(b)$ with
$\mathrm{Cay}(b)=C(\so(b))=\{R\in\SO(b):-1\notin\operatorname{spec}R\}$, open and dense in $\SO(b)$ (\auxref{B.23}(c)). Every
element of $\SO(b)$ is a product of two elements of $\mathrm{Cay}(b)$ (\auxref{B.23}(d)), so $S^2=\prod\SO(b)$, and
$R_K=W_0\cdot\prod\SO(b)$ for $K\ge2$. By continuity of $g\mapsto W_0g$ and compactness of $\prod\SO(b)$,
$\overline{R_1}=W_0\cdot\overline{\prod\mathrm{Cay}(b)}=W_0\cdot\prod\SO(b)$, and $R_\infty\setminus R_1$ lies in the image
of the Haar-null set $\{\prod_j\det(I+R_j)=0\}$. For COFT with an exact Cayley map, $S$ is the image under $C$ of a ball
about $0$, which contains a neighbourhood of $1$ in $\prod\SO(b)$ because $C$ is a local diffeomorphism at $0$; by (b) with
the single connected group $\prod\SO(b)$, $\operatorname{Mon}(S)=\prod\SO(b)$.

Methods whose shape changes with the base escape (a). For linear shapes $L_1\ne L_2$ used in two cycles (resampled indices,
alternating operators), $R_2\supseteq\theta_0+L_1+L_2\supsetneq\theta_0+L_1=R_1$ whenever $L_2\not\subseteq L_1$. For LoRA-XS
with frames recomputed from the merged weight, take $r=1$ and $W_0=\diag(2,1)$: $R_1=\{\diag(2+t,1)\}$; the point
$\diag(-\tfrac12,1)\in R_1$ has top singular direction $e_2$, so the second cycle adds $\{te_2e_2\T\}$ and reaches
$\diag(-\tfrac12,1+t)\notin R_1$.

\emph{Corollary 3.} VeRA's shape is $C=\{\Lambda_bB\Lambda_dA\}$ with $B$, $A$ frozen and $b\in\R^m$, $d\in\R^r$ trainable. It
contains $0$, is closed under real scalings ($b\mapsto\lambda b$), and lies in $\{XA:X\in\R^{m\times r}\}$. The entries of
$\Lambda_bB\Lambda_d$ are $b_iB_{ij}d_j$, so $b=e_i$, $d=e_j$ gives $B_{ij}E_{ij}$; if every $B_{ij}\ne0$, the span of the
matrices $\Lambda_bB\Lambda_d$ is $\R^{m\times r}$, and $\spanop C=\{XA\}$. By (a), $R_\infty=W_0+\{XA\}$, the image of
LoRA-FA with the same $A$.

\emph{Corollary 4.} \emph{Exact Cayley.} This is corollary~2.

\emph{The Cayley--Neumann factor.} By \auxref{B.23}(e), $R=C(Q)(I-Q^4)$, $R\T R=(I-Q^4)^2$, and the singular values of $R$ are
$\lvert1-\mu_k^4\rvert$, each twice, and $1$ for a zero eigenvalue of $Q$. So $R$ is orthogonal only when every $\mu_k^4\in
\{0,2\}$. If $\lVert Q\rVert_{\rm op}\le2^{1/4}$, then $0\le\mu_k^4\le2$ and $\lvert1-\mu_k^4\rvert\le1$, so $R$ is a contraction.
For an input-side merge $W\mapsto WR$ with block-diagonal $R$,
\[K_{\rm out}(WR)=WRR\T W\T=W(I-Q^4)^2W\T\preceq WW\T=K_{\rm out}(W),\]
because $C(Q)$ commutes with $Q^4$ and $(I-Q^4)^2\preceq I$ (\auxref{B.18}(b)); by \auxref{B.18}(c) no singular value of $W$
grows. Products of block-diagonal contractions are block-diagonal contractions (\auxref{B.18}(a)), so the reachable set lies
in $W_0$ times that semigroup.

\emph{Unconstrained $Q$.} Fix one block of size $b$ and let $S_b=\{R(Q):Q\in\so(b)\}$; since the blocks are independent and
$R(0)=I$, the monoid generated by the block-diagonal factors is $\prod\operatorname{Mon}(S_b)$. Write $Q=O(\bigoplus_k\mu_kJ
\oplus0)O\T$ with $O\in\Ort(b)$, $J=\left(\begin{smallmatrix}0&-1\\1&0\end{smallmatrix}\right)$ and $\mu_k\ge0$; then
$I-Q^4=O(\bigoplus_k(1-\mu_k^4)I_2\oplus I)O\T$.

\emph{$\subseteq$.} $\det R=\det C(Q)\det(I-Q^4)=\prod_k(1-\mu_k^4)^2\ge0$, and $\rk R=\rk(I-Q^4)=b-2\#\{k:\mu_k=1\}$. So an
invertible factor lies in $\GL^+(b)$ and a singular one has rank at most $b-2$. Products of elements of $\GL^+(b)$ stay in
it, and any product with a factor of rank at most $b-2$ has rank at most $b-2$. Hence $\operatorname{Mon}(S_b)\subseteq\GL^+
(b)\cup\MM_{\le b-2}$.

\emph{Positive scalings.} $C(-Q)=C(Q)^{-1}$ commutes with $Q^4$, so $R(Q)R(-Q)=(I-Q^4)^2=O(\bigoplus_k(1-\mu_k^4)^2I_2\oplus
I)O\T$. As $\mu_k$ ranges over $[0,\infty)$, $(1-\mu_k^4)^2$ takes every positive value ($\mu_k^4=1+\sqrt\lambda$ gives
$\lambda$). Taking $O$ a permutation matrix, $\operatorname{Mon}(S_b)$ contains every positive diagonal matrix that is
constant on the pairs of a matching of the coordinates and $1$ off it. For $b\ge3$ these generate all positive diagonal
matrices: their logarithms include every multiple of $e_i+e_j$, $i\ne j$, and $e_i=\tfrac12\big((e_i+e_j)+(e_i+e_k)-(e_j+
e_k)\big)$, so products of commuting such matrices realise every $\diag(e^x)$. For $b=2$ they are the positive scalars.

\emph{Rotations.} For $\lVert Q\rVert_{\rm op}<1$, $I-Q^4$ is positive definite with eigenvalues $1-\mu_k^4$ on the planes
of $Q$. Choose $Q'=O(\bigoplus_k\mu_k'J\oplus0)O\T$ with the same $O$ and $(1-\mu_k'^4)^2=(1-\mu_k^4)^{-1}$; then
$R(Q')R(-Q')=(I-Q^4)^{-1}$, and $C(Q)=R(Q)R(Q')R(-Q')\in\operatorname{Mon}(S_b)$. The set $\{C(Q):\lVert Q\rVert_{\rm op}<1\}$
is a neighbourhood of $I$ in $\SO(b)$ (\auxref{B.23}(c)), symmetric because $C(-Q)=C(Q)^{-1}$, so it generates $\SO(b)$ as a
monoid (\auxref{B.24}(a)). With the positive diagonals and the decomposition $\GL^+(b)=\SO(b)\,D^+\SO(b)$
(\auxref{B.19}(c)), $\GL^+(b)\subseteq\operatorname{Mon}(S_b)$ for $b\ge3$; for $b=2$, $\SO(2)\cdot\R_{>0}=\CO^+(2)$.

\emph{Singular blocks.} Let $V\subseteq\R^b$ have dimension $b-2$ and choose $O$ mapping the last coordinate plane onto
$V^\perp$, and $Q_V=O(0\oplus J)O\T$. Then $Q_V^4$ is the projector onto $V^\perp$, $C(Q_V)$ acts as $J$ on $V^\perp$ (since
$(I+J)(I-J)^{-1}=J$) and as the identity on $V$, and $R(Q_V)=C(Q_V)(I-Q_V^4)=P_V$, the orthogonal projector onto $V$. If
$P_U$ with $\dim U=k$, $1\le k\le b-2$, is in the monoid, choose $U'\subseteq U$ of dimension $k-1$ and $W'\subseteq U^\perp$
of dimension $b-1-k\le\dim U^\perp$, and put $V=U'\oplus W'$, of dimension $b-2$. Since $W'\perp U$,
$P_UP_V=P_U(P_{U'}+P_{W'})=P_{U'}$. By downward induction every orthogonal projector of rank at most $b-2$ lies in the
monoid. A matrix $X$ of rank $k\le b-2$ is $X=U\diag(\sigma_1,\dots,\sigma_k,1,\dots,1)\,P_k\,V\T$ by its SVD, with $P_k$
the projector onto the first $k$ coordinates. Changing the sign of the last column of $U$, or of the last row of $V\T$,
leaves this product unchanged, because $P_k$ kills the last coordinate; so both outer factors may be taken in $\GL^+(b)$,
and $X\in\operatorname{Mon}(S_b)$. For $b=2$, $R(\mu J)=(1-\mu^4)\,C(\mu J)$ is a real multiple of a rotation, so
$\operatorname{Mon}(S_2)\subseteq\CO^+(2)\cup\{0\}$, and $R(J)=0$. Altogether
$\operatorname{Mon}(S_b)=\GL^+(b)\cup\MM_{\le b-2}$ for $b\ge3$ and $\CO^+(2)\cup\{0\}$ for $b=2$, which is the clause; the
steps above complete the sketch recorded in the main text.

\emph{BOFT.} Group the coordinates into runs of $b/2$ consecutive coordinates, indexed by $p=0,1,\dots,2n/b-1$. Up to one fixed
relabelling of coordinates, the $l$-th butterfly factor of BOFT ($l=1,\dots,m'$) is block-diagonal with one orthogonal
$b\times b$ block, an exact Cayley image, on each pair of runs whose indices differ exactly in binary digit $l-1$
\citep{liu2023parameter,dao2019learning}; for $l=1$ these are $n/b$ consecutive blocks of size $b$, block OFT. Two runs lie
in one component of the factor hypergraph iff their indices agree in all digits from $m'$ on, so the components consist of
$2^{m'}$ runs, $b\,2^{m'-1}$ coordinates. HF's BOFT multiplies the output by a trainable scale, $W\mapsto\diag(s)\,W\,R$, and
left and right multiplications commute, so after $K$ cycles the weight is $\diag(s_K\cdots s_1)\,W_0\,R_1\cdots R_K$; the
products of scale vectors already fill $\R^m$ at $K=1$. The orthogonal part is generated by the set $S$ of products of the
$m'$ butterfly factors. Let $H_l$ be the group of all block rotations of level $l$, connected; $S$ lies in the group they
generate, and setting all but one factor to the identity shows that $S$ contains a neighbourhood of $1$ in each $H_l$, the
exact Cayley image being open. By (b) and (c), $\operatorname{Mon}(S)$ is the connected Lie subgroup with Lie algebra
$\bigoplus_C\so(C)$, which is the group of block rotations on the components, $P\big(\prod\SO(b\,2^{m'-1})\big)P\T$ for the
permutation $P$ that makes the components consecutive. This proves the reach. If $b\,2^{m'-1}=n$ there is one component and
the reach is $\diag(\R^m)\cdot W_0\cdot\SO(n)$. If $c:=b\,2^{m'-1}<n$, the group $G=P\big(\prod\SO(c)\big)P\T$ has dimension
$n(c-1)/2<n(n-1)/2$; and for $W_0$ of full column rank the reachable sets differ (for $W_0=0$ both are $\{0\}$). Indeed, if
$\diag(\R^m)W_0G=\diag(\R^m)W_0\SO(n)$, then each $R\in\SO(n)$ satisfies $W_0R=\diag(s)W_0g$ for some $s$ and $g\in G$, so
$M=Rg^{-1}$ satisfies $W_0M=\diag(s)W_0$: every row $w_i$ of $W_0$ is an eigenvector of $M\T$. These rows span $\R^n$, so $M\T$ is
diagonalisable with real eigenvalues; being orthogonal, $M$ is an orthogonal involution $I-2P_E$ whose $-1$-eigenspace $E$ is
spanned by the rows of $W_0$ it contains. So $M$ ranges over a finite set $F_0$, and $\SO(n)\subseteq F_0G$, a finite union
of submanifolds of lower dimension, which is impossible. With $m'=1$ the single level is block OFT, so HF's default
\texttt{boft\_n\_butterfly\_factor=1} is block OFT times $\diag(s)$.

\emph{A single BOFT.} Its parameters number $m'\frac nb\binom b2$, so its image has at most this dimension
(\auxref{B.9}(b)), and $m'\frac nb\frac{b(b-1)}2<\frac{n(n-1)}2$ iff $m'(b-1)<n-1$; the image then has positive codimension
in $\SO(n)$ and is Haar-null. At full depth, $2^{m'}=2n/b$ runs form one component, and for every pair of coordinates $(i,j)$
there is a chain $j=c_0,c_1,\dots,c_{m'}=i$ in which $c_{l-1}$ and $c_l$ lie in a common block of the $l$-th factor applied:
each factor lets the run index flip one binary digit, a different one for each factor, and lets the position inside the run
be chosen freely. So the entry $(i,j)$ of the product, a polynomial in the block entries, is nonzero at a product of signed
permutation matrices of determinant $1$ (one per block) that routes $j$ to $i$ along the chain. It is therefore not
identically zero on the connected group $\prod_lH_l$, and since the Cayley parametrisation is a diffeomorphism onto an open
subset of that group, the entry is nonzero for parameters outside a nowhere-dense analytic subset (\auxref{B.10}). All
$n^2$ entries are nonzero for generic parameters, so the matrices are dense.

\emph{Corollary 5.} With \texttt{apply\_GS=False}, one HRA cycle applies $H_{u_1}\cdots H_{u_r}$ on the input side. Let
$S_k=\{R\in\SO(n):\rk(R-I)\le k\}$ for even $k$. A product of $r$ reflections lies in $\SO(n)$ ($r$ even) and has
$\rk(R-I)\le r$ (\auxref{B.22}); conversely, if $R\in S_r$ then $j=\rk(R-I)$ is even and $R$ is a product of exactly $j$
reflections (\auxref{B.22}), which pairs $H_uH_u=I$ pad to $r$. So one cycle applies exactly the elements of $S_r$, and by
(b) $R_K=W_0\cdot S_r^K$. Next, $S_r^K=S_{Kr}$. For $\subseteq$, $R_1R_2-I=R_1(R_2-I)+(R_1-I)$ gives
$\rk(R_1R_2-I)\le\rk(R_2-I)+\rk(R_1-I)$ (\auxref{B.12}(a)). For $\supseteq$, write $R\in S_{Kr}$ as a product of
$j=\rk(R-I)\le Kr$ reflections, $j$ even, split them into $K$ consecutive groups of even size at most $r$, and pad each group
with pairs to size $r$. Finally, since $W_0$ is injective, $\rk(W_0R-W_0)=\rk(R-I)$, so
$W_0\cdot S_{Kr}=(W_0+\MM_{\le Kr})\cap W_0\cdot\SO(n)$. When $Kr\ge n$, $S_{Kr}=\SO(n)$, which gives the last sentence.

\emph{Corollary 6.} By \thmref{Proposition}{IV.2} and \S\ref{proof:IV.3}(8), each period of these methods is a period of
the linear method $\theta_k+\mathcal L_kC$ with shape $L_k=\operatorname{im}\mathcal L_k$, so $R_K=\theta_0+L_1+\dots+L_K$
for the chosen shapes. For GaLore with left projection, $L_k=\{Z:\operatorname{col}Z\subseteq\operatorname{col}P_k\}$, and
column spaces that together span $\R^m$ give all of $\R^{m\times n}$ after $\lceil m/r\rceil$ periods; for LISA, layer blocks
covering every layer do the same; for MeZO, $L_k=\R z_k$, and $z_k$ running through a basis give $\Theta$. A fixed linear
shape $L$ is its own additive monoid, so it reaches only $\theta_0+L$ (corollary~2). A GaLore period moves a projected matrix
by an update of rank at most $r$; a LISA period moves each sampled matrix by an unconstrained update, and a MeZO step by a
multiple of the Gaussian $z_k$, whose restriction to each matrix has full rank almost surely. Flora's steps contain a
full-rank term (\S\ref{proof:IV.1}).

\emph{Corollary 7.} For orthogonal $R$, $(WR)(WR)\T=WW\T$ and $(RW)\T(RW)=W\T W$; $WR$ and $RW$ have the singular values of $W$.
By induction over cycles, every input-side rebasing preserves $K_{\rm out}$ and every output-side one $K_{\rm in}$, and both
preserve the spectrum (\thmref{Corollary}{II.6}).
\end{proof}

\begin{remark*}{numerical checks}
The component sizes $4,8,16,32,64$ for $n=64$, $b=4$ and $m'=1,\dots,5$ reported after \thmref{Theorem}{V.3} are the
$b\,2^{m'-1}$ of the BOFT paragraph; the Loewner-monotone decrease of $K_{\rm out}$ over $200$ default-OFT merges, the
contraction bound for random skew $Q$ with $\lVert Q\rVert_{\rm op}\le2^{1/4}$, and $R=0$ at $b=2$, $Q=J$, instantiate the
Cayley--Neumann paragraph (Appendix~\ref{app:repro}).
\end{remark*}

\subsection{Proposition V.4}\label{proof:V.4}

\begin{restate}{Proposition}{V.4}{combination rules: gauge and naturality}
(a) A rule $\Phi\big((B_i,A_i)_i\big)$ for combining LoRA factors descends to a function of the adapters $(B_iA_i)_i$ iff it
is constant on the fibres of $(B_i,A_i)\mapsto(B_iA_i)$. On the full-rank locus this is equivalent to
$\prod_i\GL_r$-invariance. It is also equivalent for $\Phi$ continuous on the whole factor space; this clause rests on the
unique closed orbit in each fibre, which lies in the closure of every orbit of the fibre.
\begin{itemize}
  \item Task arithmetic $\sum_iw_is_iB_iA_i$, with $s_i$ HF's \texttt{scaling} ($\alpha_i/r_i$, or $\alpha_i/\sqrt{r_i}$
  under rsLoRA), is invariant and lands in $\mathrm{LoRA}_{\sum_ir_i}$. It is HF's \texttt{cat} with merged scale $1$.
  \item LoraHub's factor-wise rule $\big(\sum_iw_iB_i\big)\big(\sum_jw_jA_j\big)$ is not invariant. Replacing $(B_2,A_2)$ by
  $(cB_2,A_2/c)$ changes it by
  \[\sum_{l\ne2}w_lw_2\big[(c-1)B_2A_l+(c^{-1}-1)B_lA_2\big].\]
  HF's \texttt{add\_\allowbreak weighted\_\allowbreak adapter(..., combination\_\allowbreak type="linear")} is a factor-wise
  rule of the same type. It weights $A_i$ by $\operatorname{sign}(w_i)\sqrt{\lvert w_is_i\rvert}$ and $B_i$ by
  $\sqrt{\lvert w_is_i\rvert}$ and sums the factors separately, so it has cross terms. On adapters that share a frozen $A$
  it is still not exact up to a global factor, since the relative weights become $\sqrt{\lvert w_is_i\rvert}$. HF's non-SVD
  \texttt{ties}, \texttt{dare\_*} and \texttt{magnitude\_prune} variants also act on factors; the \texttt{*\_svd} variants
  act on full updates.
  \item Poly and MHR are different. Their skill inventories are trained jointly through the product-of-sums map, which is
  invariant under the diagonal $\GL_r$, one common $g$ for every skill. Numerically, for fixed generic routing with fewer
  skills than tasks and full-rank factors, this is the whole continuous fibre. With at least as many skills as tasks, or
  once the routing is counted as a parameter, there are further symmetries (skill permutations, and re-mixings
  $\alpha\mapsto\alpha M$ with $M\mathbf 1=\mathbf 1$). Their cross terms are a design choice of a jointly trained
  parametrisation, and raise no question of well-definedness.
\end{itemize}

(b) \emph{Well-definedness versus exactness.} A joint gauge fixing makes a factor-wise rule well defined. For
rank-preserving product-of-sums rules, exactness (agreement with task arithmetic for all weights $w$) needs degree one in
$w$, which for injective factors forces a shared factor; \texttt{cat}, which concatenates and raises the rank, is exact
without one. A frozen shared factor (FFA-LoRA; LoRA-FA) gives $\big(\sum_iw_iB_i\big)A=\sum_iw_iB_iA$ exactly. LoraHub with
a shared $A$ still returns $\big(\sum_jw_j\big)\sum_iw_iB_iA$, exact only up to the global factor $\sum_jw_j$, since its
weights are unnormalised. A per-module SVD canonical form leaves a $(\mathbb Z/2)^r$ of sign flips per module, plus
rotations within repeated singular values, and these change the cross terms. What is needed is a \emph{joint} frame that
produces a shared factor, as KnOTS's joint re-factorisation of the concatenated updates through a shared $U\Sigma$ does.
Federated LoRA shows the difference. FedAvg-LoRA clients start each round from the same broadcast $(B,A)$, so their failure,
$\tfrac12(B_1+B_2)\cdot\tfrac12(A_1+A_2)\ne\tfrac12(B_1A_1+B_2A_2)$, is bilinearity, and it persists in a fully fixed gauge.

(c) \emph{Naturality.} On updates with $N$ coordinates, write $T_N$, $B_N$ and $\mathrm{Mon}_N$ for the invertible diagonal,
signed permutation and monomial matrices. Task arithmetic is $\GL(\Theta)$-equivariant on updates. DARE is coordinatewise
and magnitude-free, $D(m\odot\tau)=m\odot(D\tau)$ for every invertible diagonal $D$ and the same mask $m$. So DARE is exactly
equivariant under the torus $T_N$, sign flips included, for each mask, and under permutations in distribution; its
distributional group is exactly $\mathrm{Mon}_N$. TIES with global trimming (top-$k$ magnitude, sign election by total mass
as in the TIES paper and HF's default \texttt{"total"}, disjoint mean) is equivariant under $\R^\times\cdot B_N$,
homotheties times signed permutations, generically (off magnitude ties at the threshold). With HF's
\texttt{majority\_\allowbreak sign\_\allowbreak method="frequency"}, count ties are broken to $+1$ and the group drops to
$\R_{>0}\cdot S_N$. Per-tensor trimming changes the group to $\prod_b\big(\R^\times\cdot B_{N_b}\big)$, extended by
permutations of equal-size blocks. DARE followed by full TIES (with trimming) is $\R^\times\cdot B_N$-equivariant in
distribution; HF's and MergeKit's \texttt{dare\_ties} has no magnitude trim, so with \texttt{"total"} election it is exactly
$T_N$-equivariant per mask, and its group contains $\mathrm{Mon}_N$. None of the sparsifying rules (DARE, TIES, DARE-TIES)
is $\Ort(N)$-equivariant. Applied to factors and summed across adapters (HF's non-SVD variants), all of them are
gauge-dependent, and under independent per-adapter gauges even DARE fails, through the cross terms of (a). DARE's masking
commutes with $T_r$ exactly for each mask and with $\mathrm{Mon}_r$ in distribution, so a single masked adapter, or a
factor-wise sum under one gauge shared by all adapters, is $T_r$-invariant per mask and $\mathrm{Mon}_r$-invariant in
distribution.
\end{restate}

\begin{proof}
Write $\pi\big((B_i,A_i)_i\big)=(B_iA_i)_i$, and let adapter $i$ have rank $r_i$ (the statement's $r$ when all ranks agree);
its gauge group $\GL_{r_i}$ acts by $(B_ig_i^{-1},g_iA_i)$, and $\pi$ is invariant under $\prod_i\GL_{r_i}$.

(a) \emph{Descent.} If $\Phi=\varphi\circ\pi$, then $\Phi$ is constant on the fibres of $\pi$. Conversely, if $\Phi$ is constant
on fibres, $\varphi(\pi(x)):=\Phi(x)$ is a well-defined function on the image of $\pi$.

\emph{Full-rank locus.} Let $\rk B_i=\rk A_i=r_i$ and $B_i'A_i'=B_iA_i$. The product has rank $r_i$, so $\rk B_i'=\rk A_i'=r_i$,
and the full-rank locus is a union of whole fibres. Moreover $\operatorname{col}B_i'=\operatorname{col}(B_iA_i)=
\operatorname{col}B_i$, so $B_i'=B_ig_i^{-1}$ for a unique $g_i\in\GL_{r_i}$; then $B_iA_i=B_ig_i^{-1}A_i'$, and injectivity of
$B_i$ gives $A_i'=g_iA_i$. So each fibre in the full-rank locus is a single $\prod_i\GL_{r_i}$-orbit (as in
\thmref{Theorem}{II.2}), and there constancy on fibres is the same as invariance.

\emph{Continuous rules.} Let $\Phi$ be continuous on the whole factor space and invariant, and let $x,x'$ lie in the fibre
over $(W_i)_i$, $\rk W_i=k_i\le r_i$. By \auxref{B.27}(c), in each factor the pairs with product $W_i$ and both factors of rank
$k_i$ form one $\GL_{r_i}$-orbit $O_{W_i}$, which lies in the closure of every orbit of the fibre. Applying this factor by
factor, there are gauges $g(t),g'(t)\in\prod_i\GL_{r_i}$ with $g(t)\cdot x\to o$ and $g'(t)\cdot x'\to o'$ as $t\to0$,
where $o,o'\in\prod_iO_{W_i}$, and $o'=h\cdot o$ for some gauge $h$. By invariance and continuity,
\[\Phi(x)=\lim_{t\to0}\Phi(g(t)\cdot x)=\Phi(o)=\Phi(h\cdot o)=\Phi(o')=\lim_{t\to0}\Phi(g'(t)\cdot x')=\Phi(x').\]
So $\Phi$ is constant on fibres; the converse holds because orbits lie in fibres. The review record graded this clause as a
sketch; with the degeneration of \auxref{B.27}(c), whose reason is given in Appendix~\ref{app:auxiliary}, the steps above
complete it. Invariance alone does not suffice off the full-rank locus: on the invariant open set $\{\rk A=r\}$ the
projector $P_{\operatorname{row}A}$ is $\GL_r$-invariant, since $\operatorname{row}(gA)=\operatorname{row}A$, yet $(0,A)$
and $(0,A')$ with $\operatorname{row}A\ne\operatorname{row}A'$ lie in the fibre over $0$ and receive different values. This
$\Phi$ has no continuous extension to the whole factor space.

\emph{Task arithmetic.} $\sum_iw_is_iB_iA_i$ depends only on the products, so it is invariant, and
\[\sum_iw_is_iB_iA_i=\big[\,w_1s_1B_1\ \big|\ \cdots\ \big|\ w_Ns_NB_N\,\big]\begin{bmatrix}A_1\\ \vdots\\ A_N\end{bmatrix}\]
is a LoRA update of rank at most $\sum_ir_i$ with scale $1$, the concatenation \texttt{cat} of \thmref{Observation}{I.7}.

\emph{LoraHub.} Expanding, $\big(\sum_iw_iB_i\big)\big(\sum_jw_jA_j\big)=\sum_{i,j}w_iw_jB_iA_j$. Under the gauge
$(B_2,A_2)\mapsto(cB_2,A_2/c)$ the terms with $i,j\ne2$ and the term $w_2^2B_2A_2$ are unchanged, each $w_2w_lB_2A_l$ ($l\ne2$) gains the factor $c$,
and each $w_lw_2B_lA_2$ gains $c^{-1}$, which gives the displayed change. For $c\ne1$, nonzero weights and factors for which
the matrices $B_2A_l$, $B_lA_2$ ($l\ne2$) are linearly independent (true for generic factors), the change is nonzero. HF's
\texttt{linear} combination is $\big(\sum_i\beta_iB_i\big)\big(\sum_j\alpha_jA_j\big)$ with $\beta_i=\sqrt{\lvert w_is_i\rvert}$
and $\alpha_j=\operatorname{sign}(w_j)\sqrt{\lvert w_js_j\rvert}$ \citep{xmangrulkar2022peft}, a product of sums with the same
cross terms and the same failure. If all adapters share $A$, it equals $\big(\sum_j\alpha_j\big)\sum_i\beta_iB_iA$, which is a
fixed multiple of $\sum_iw_is_iB_iA$ for all $B_i$ only if the $\beta_i$ are proportional to the $w_is_i$, that is, only if all
$\lvert w_is_i\rvert$ with $w_is_i\ne0$ are equal (and the signs agree); for instance $w_1s_1=1$ and $w_2s_2=4$ give relative
weights $1:2$ instead of $1:4$.

\emph{Poly and MHR.} For task $t$ with routing weights $\alpha_{tj}$ over skills $j$ the update is
$W_t=\big(\sum_j\alpha_{tj}B_j\big)\big(\sum_j\alpha_{tj}A_j\big)$. A common $g$ gives
$\big(\sum_j\alpha_{tj}B_jg^{-1}\big)\big(\sum_j\alpha_{tj}gA_j\big)=W_t$, so the diagonal $\GL_r$ preserves the map. Per-skill
gauges $(g_j)$ that preserve it for all routings in an open set are diagonal on any pair of mixed skills with complementary
column spaces: $W_t=\sum_{i,j}\alpha_{ti}\alpha_{tj}B_iA_j$ is a polynomial in $\alpha$, and comparing the coefficients of
$\alpha_{ti}\alpha_{tj}$, $i\ne j$, gives
\[B_i(g_i^{-1}g_j-I)A_j+B_j(g_j^{-1}g_i-I)A_i=0 .\]
If $\operatorname{col}B_i\cap\operatorname{col}B_j=0$, the two terms vanish separately, and for injective $B_i$ and surjective
$A_j$ this forces $g_i=g_j$. That for one fixed generic routing with fewer skills than tasks the continuous fibre is exactly
the diagonal orbit is the numerical check reported below. With at least as many skills as tasks the fibre is larger. If there
are more skills than tasks, a vector $v\ne0$ with $\sum_jv_j\alpha_{tj}=0$ for every $t$ exists, and $B_j\mapsto B_j+v_jX$
leaves every $\sum_j\alpha_{tj}B_j$ unchanged for every $X$. If there are equally many and the routing matrix is invertible,
the task-wise sums $(C_t,D_t)=\big(\sum_j\alpha_{tj}B_j,\sum_j\alpha_{tj}A_j\big)$ are free coordinates, and each task has its
own $\GL_r$ gauge. Counting the routing as a parameter, relabelling the skills together with the columns of $\alpha$ preserves
the map, and so does $(\alpha,B)\mapsto(\alpha M,M^{-1}B)$ for invertible $M$ with $M\mathbf 1=\mathbf 1$, where $B$ stacks the
skills (and likewise for $A$), since $(\alpha M)(M^{-1}B)=\alpha B$ and $M\mathbf 1=\mathbf 1$ keeps the rows of $\alpha$ summing
to $1$. Poly and MHR train skills and routing jointly through this map, so it is the method's own $\rho$; the question of
(a) concerns combining modules that were factorised independently.

(b) \emph{Gauge fixing.} If a section $\sigma$ of $\pi$ chooses the factors from the products, then $\Phi\circ\sigma$ is a
function of the products. \emph{Exactness.} Task arithmetic $T(w)=\sum_iw_iB_iA_i$ (scales absorbed) is linear in $w$.
Consider rank-preserving products of sums in which each adapter's factor carries its own weight $w_i$, in one sum or in
both. If both sums are weighted, as in LoraHub, the rule $P(w)$ is homogeneous of degree two in $w$, and $P(w)=T(w)$ for all
$w$ would give, replacing $w$ by $\lambda w$, $\lambda^2P(w)=\lambda T(w)$ for all $\lambda$, hence $T\equiv0$. So an exact
rule of this kind has degree one in $w$: one of its factors does not depend on $w$, say $\big(\sum_iw_iB_i\big)X$. Then exactness for all $w$ means $B_iX=B_iA_i$ for every $i$, and for injective $B_i$ this gives
$A_i=X$: the $A$-factor is shared. The symmetric case $X\big(\sum_jw_jA_j\big)$ with surjective $A_j$ forces a shared $B$.
\texttt{cat} realises $T(w)$ exactly, as shown in (a), with inner dimension $\sum_ir_i$. With a frozen shared $A$,
$\big(\sum_iw_iB_i\big)A=\sum_iw_iB_iA$, and LoraHub's rule gives $\big(\sum_iw_iB_i\big)\big(\sum_jw_j\big)A$.

\emph{SVD canonical forms.} If $B_iA_i=U_i\Sigma_iV_i\T$ and the canonical factors are $B_i=U_i\Sigma_i^{1/2}$,
$A_i=\Sigma_i^{1/2}V_i\T$ (or any convention built from $U_i,\Sigma_i,V_i$), then changing the signs of the $j$-th column of
$U_i$ and of $V_i$ gives another SVD and another canonical pair, multiplying column $j$ of $B_i$ and row $j$ of $A_i$ by $-1$.
The product $B_iA_i$ is unchanged, while a cross term $B_iA_l$ ($l\ne i$) changes by $-2\,b_{ij}a_{lj}\T$, where $b_{ij}$ is
column $j$ of $B_i$ and $a_{lj}\T$ row $j$ of $A_l$, which is nonzero generically. When $\sigma_j=\sigma_{j+1}$, rotating the
corresponding columns of $U_i$ and $V_i$ by the same $O\in\Ort(2)$ also preserves the product and changes the cross terms.
\emph{KnOTS.} Write the horizontal concatenation $[\Delta_1\mid\cdots\mid\Delta_N]=U\Sigma[V_1\T\mid\cdots\mid V_N\T]$ by an
SVD \citep{stoica2024model}; then $\Delta_i=(U\Sigma)V_i\T$ share the left factor $U\Sigma$, and
$(U\Sigma)\big(\sum_iw_iV_i\T\big)=\sum_iw_i\Delta_i$ exactly. \emph{FedAvg-LoRA.} Expanding,
\[\tfrac12(B_1+B_2)\cdot\tfrac12(A_1+A_2)-\tfrac12(B_1A_1+B_2A_2)=-\tfrac14(B_1-B_2)(A_1-A_2),\]
which is nonzero whenever $(B_1-B_2)(A_1-A_2)\ne0$, whatever gauge the broadcast pair is in.

(c) Write a merge rule as $\Psi(\tau_1,\dots,\tau_K)$ for task vectors $\tau_k\in\R^N$ (or the matrices with $N$ entries,
flattened). It is $g$-equivariant if $\Psi(g\tau_1,\dots,g\tau_K)=g\Psi(\tau_1,\dots,\tau_K)$, read for random rules either
for each realisation of the randomness (per mask) or as equality of laws (in distribution).

\emph{Task arithmetic} is linear, hence $\GL(\Theta)$-equivariant.

\emph{DARE.} $\mathrm{DARE}(\tau)=m\odot\tau/(1-p)$, with a mask $m$ of independent Bernoulli$(1-p)$ entries, $0<p<1$,
drawn independently for each task, followed by a weighted sum. For invertible diagonal $D$, $D(m\odot\tau)=m\odot(D\tau)$, so
$\mathrm{DARE}(D\tau)=D\,\mathrm{DARE}(\tau)$ for each mask, and the weighted sum preserves this. For a permutation matrix $P$,
$P(m\odot\tau)=(Pm)\odot(P\tau)$ and $Pm$ has the law of $m$, so $\mathrm{DARE}(P\tau)$ has the law of $P\,\mathrm{DARE}(\tau)$,
jointly over the tasks. Together, $\mathrm{Mon}_N=S_N\ltimes T_N$ acts equivariantly in distribution. Conversely, suppose
$\mathrm{DARE}(g\tau)$ and $g\,\mathrm{DARE}(\tau)$ have the same law for every $\tau$, and take $\tau=e_k$. The law of
$g\,\mathrm{DARE}(e_k)$ is supported on the two points $0$ and $ge_k/(1-p)$, while $\mathrm{DARE}(ge_k)$ takes the $2^q$
distinct values $m\odot ge_k/(1-p)$, each with positive probability, where $q$ is the number of nonzero entries of $ge_k$. So
$q=1$: every column of $g$ has a single nonzero entry, and $g\in\mathrm{Mon}_N$.

\emph{TIES with global trimming} \citep{yadav2023ties}. Each $\tau_k$ is trimmed to its $k_0$ entries of largest magnitude,
giving $\hat\tau_k$; the sign $\gamma_i=\operatorname{sign}\big(\sum_k\hat\tau_{k,i}\big)$ is elected by total mass; and the
output is $\lambda$ times the disjoint mean, $\frac1{\lvert A_i\rvert}\sum_{k\in A_i}\hat\tau_{k,i}$ over
$A_i=\{k:\operatorname{sign}\hat\tau_{k,i}=\gamma_i\}$. Let $g=cPS$ with $c\ne0$, $P$ a permutation matrix and $S=\diag(s_1,\dots,s_N)$ a
diagonal sign matrix, and let $j$ be the coordinate that $P$ moves to $i$. The magnitudes of $g\tau_k$ are $\lvert c\rvert$ times
the permuted magnitudes of $\tau_k$, so, off ties at the threshold, trimming keeps the corresponding entries and
$\widehat{g\tau_k}=g\hat\tau_k$. The elected sign at $i$ becomes $\operatorname{sign}(c)s_j\gamma_j$, the set $A_i$ for the
transformed input is $A_j$, and the mean at $i$ is $cs_j$ times the mean at $j$. So the output is $g$ times the output, and
TIES is $\R^\times\cdot B_N$-equivariant off ties. A non-scalar positive diagonal fails: with $N=2$, one task, $k_0=1$,
$\tau=(2,1)$ and $D=\diag(1,3)$, TIES returns $\lambda(2,0)$ on $\tau$ and $\lambda(0,3)\ne D\lambda(2,0)$ on $D\tau$.

\emph{Frequency election.} Here $\gamma_i=\operatorname{sign}\big(\sum_k\operatorname{sign}\hat\tau_{k,i}\big)$, with $0$
replaced by $+1$. Positive homotheties and permutations leave the counts unchanged or permute them, so the argument above
gives $\R_{>0}\cdot S_N$-equivariance. Sign changes fail on ties, which have positive probability because the counts are
integers. With two tasks and $\hat\tau_{1,i}=1$, $\hat\tau_{2,i}=-1$, the tie elects $+1$ and the output at $i$ is $\lambda$;
after flipping coordinate $i$ the entries are $-1,+1$, the tie again elects $+1$, and the output at $i$ is $\lambda$, not
$-\lambda$. The same example defeats $-I$.

\emph{Per-tensor trimming.} If coordinates are grouped into tensors $b$ of sizes $N_b$ and each tensor is trimmed to its own
top entries, an independent homothety $c_b$ and signed permutation inside each tensor preserve the trimmed sets, and
permuting tensors of equal size (and equal $k_0$) preserves them too; the election and the mean are coordinatewise, so the
argument above applies blockwise.

\emph{DARE followed by TIES.} For $g\in\R^\times\cdot B_N\subseteq\mathrm{Mon}_N$, the masked inputs
$(\mathrm{DARE}(g\tau_k))_k$ have the law of $(g\,\mathrm{DARE}(\tau_k))_k$, and TIES commutes with $g$ off magnitude ties
among nonzero entries (ties among entries zeroed by the mask do not matter, since a kept zero is zero), so the composite is
equivariant in distribution. HF's and MergeKit's \texttt{dare\_ties} \citep{xmangrulkar2022peft,xgoddard2024mergekit} omit the
trim: for invertible diagonal $D$ and fixed masks, $\mathrm{DARE}(D\tau_k)=D\,\mathrm{DARE}(\tau_k)$, the elected sign at $i$
is multiplied by $\operatorname{sign}d_i$, the sets $A_i$ are unchanged and the mean is multiplied by $d_i$; so it is exactly
$T_N$-equivariant per mask, and with permutations in distribution its group contains $\mathrm{Mon}_N$.

\emph{No rotations.} Let $R$ rotate the plane of $e_1,e_2$ by $30^\circ$, so $Re_1=(\cos30^\circ,\sin30^\circ,0,\dots)$. For DARE,
$R\,\mathrm{DARE}(e_1)$ takes the values $0$ and $Re_1/(1-p)$, each with two nonzero entries or none, while
$\mathrm{DARE}(Re_1)$ has exactly one nonzero entry with probability $2p(1-p)>0$; the laws differ. For TIES with one task and
$k_0=1$, $\mathrm{TIES}(e_1)=\lambda e_1$, so $R\,\mathrm{TIES}(e_1)$ has two nonzero entries, while $\mathrm{TIES}(Re_1)=
\lambda\cos30^\circ e_1$ has one. With one task, \texttt{dare\_ties} reduces to DARE, and DARE followed by trimmed TIES
returns, with positive probability, the trimmed vector $\lambda\cos30^\circ e_1/(1-p)$ on $Re_1$, which no value of
$R\,\mathrm{TIES}(\mathrm{DARE}(e_1))\in\{0,\lambda Re_1/(1-p)\}$ matches.

\emph{Factor-level rules.} HF's non-SVD variants apply the sparsifier to each $B_i$ and each $A_j$ with independent masks and
then form $\big(\sum_iv_i\,\mathrm{DARE}(B_i)\big)\big(\sum_jv_j\,\mathrm{DARE}(A_j)\big)$ (and the analogues for TIES). The
$B$-masks and the $A$-masks are independent and DARE is unbiased, $\mathbb E\,\mathrm{DARE}(X)=X$, so the expected output is the
LoraHub-type rule $\big(\sum_iv_iB_i\big)\big(\sum_jv_jA_j\big)$. A gauge $(cB_2,A_2/c)$ on one adapter changes this expectation
by the formula of (a), which is nonzero for generic factors, so the output's law changes: even DARE is not invariant under
independent per-adapter gauges, not even in distribution. On the other hand, let $g\in T_r$ act on all adapters at once,
$B_i\mapsto B_ig^{-1}$ and $A_j\mapsto gA_j$. Right multiplication by a diagonal matrix scales columns, which commutes with an
entrywise mask, so $M\odot(B_ig^{-1})=(M\odot B_i)g^{-1}$ and $M'\odot(gA_j)=g(M'\odot A_j)$, and for every choice of masks
\[\Big(\sum_iv_i\,M_i\odot B_ig^{-1}\Big)\Big(\sum_jv_j\,M_j'\odot gA_j\Big)=\Big(\sum_iv_i\,M_i\odot B_i\Big)g^{-1}g\Big(\sum_jv_j\,
M_j'\odot A_j\Big).\]
For a permutation $g=P$, $M\odot(B_iP\T)=\big((MP)\odot B_i\big)P\T$, and $MP$ has the law of $M$; with all masks permuted by the
same $P$ their joint law is unchanged, so the output's law is invariant. A single masked adapter is the case of one
summand. This gives $T_r$-invariance per mask and $\mathrm{Mon}_r$-invariance in distribution.
\end{proof}

\begin{remark*}{numerical checks}
The checks reported after \thmref{Proposition}{V.4}, namely DARE's per-mask equivariance under a random diagonal to
$2\times10^{-15}$, TIES's homogeneity and its failure by $5.6$ under a positive diagonal, the LoraHub formula with three
modules, the order-one change of factor-wise DARE under a one-adapter gauge against $10^{-15}$ under a shared one, and the Poly
fibre dimensions $32$ (two tasks) and $4=\dim\GL_2$ (four tasks) for three skills, $r=2$ and $6\times6$ weights, instantiate the
statements proved above; the last one is the numerical part of the Poly bullet (Appendix~\ref{app:repro}).
\end{remark*}

\subsection{Definition VI.1}\label{proof:VI.1}

\begin{restate}{Definition}{VI.1}{extensions; mergeability}
\begin{itemize}
  \item An \textbf{extension} is a function-preserving pointed embedding
  \[\iota:(\Theta,f,\theta_0)\to(\Theta\times\Theta',\tilde f,(\theta_0,\theta_0')),\qquad\iota(\theta)=(\theta,\theta_0'),\]
  with $\tilde f(\theta,\theta_0',-)=f(\theta,-)$. The point $\theta_0'$ is the \textbf{neutral point} (``identity at
  initialisation'').
  \item An \textbf{unpointed extension} has no neutral point.
  \item A method on an extension is \textbf{mergeable} if there is $\mu:Q\to\Theta$ with
  $\Phi_{\tilde f}\circ\rho=\Phi_f\circ\mu$: a lift along the realisation.
  \item It is \textbf{box-locally mergeable} if the modified box function stays in the box's function class: linear maps
  for a linear box, affine maps when the box has a bias slot. \thmref{Theorem}{VI.3} gives sufficient conditions, with
  explicit side conditions (position-uniformity, no weight sharing, bias slots), under which this yields a merge into the
  base network.
\end{itemize}
\end{restate}

A definition needs no proof. We check the facts that the definition and the square after it use. Here
$\Phi_f:\Theta\to C^\infty(X,Y)$, $\theta\mapsto f(\theta,-)$, is the realisation, and a box's function class consists of the
maps it can compute with some weight, applied identically at every position.

\begin{proof}[Verification of the claims in Definition VI.1]
\emph{Extensions.} The map $\iota(\theta)=(\theta,\theta_0')$ is smooth, injective and immersive, and a homeomorphism onto the
closed set $\Theta\times\{\theta_0'\}$, so it is an embedding. It is pointed, $\iota(\theta_0)=(\theta_0,\theta_0')$, and
function-preserving, $\Phi_{\tilde f}\circ\iota=\Phi_f$, which is the displayed condition.

\emph{Weight-space methods merge.} A method $(Q,q_0,\rho)$ on $\Theta$ itself, read on the trivial extension
($\Theta'$ a point), satisfies the merge condition with $\mu=\rho$, the arrow to full fine-tuning of
\thmref{Observation}{I.4}(a). More generally, a method on an extension whose $\rho$ factors as $\iota\circ\mu$ merges by
$\mu$.

\emph{Merges are lifts, and need not be unique.} The realisation $\Phi_f$ need not be injective. Every function-preserving
reparametrisation $\phi$ of the base gives another lift $\phi\circ\mu$; permutations of hidden units are examples
(\thmref{Lemma}{VI.2}).
Let a pointed method merge, with $\rho(q_0)$ the neutral point $(\theta_0,\theta_0')$. Then
\[\Phi_f(\mu(q_0))=\Phi_{\tilde f}(\theta_0,\theta_0')=\Phi_f(\theta_0),\]
so $\mu(q_0)$ realises the pretrained function, without having to equal $\theta_0$.

\emph{Box-local mergeability.} If the edit of a method modifies a single linear (affine) box of the base network, and for
every $q$ the modified box function is linear (affine), position-independent and computed by an unshared weight $W(q)$ of
that box, then replacing the box's weight by $W(q)$ and keeping the others gives a $\mu$ with $\Phi_f\circ\mu=\Phi_{\tilde
f}\circ\rho$. The side conditions are exactly what makes this replacement legitimate, and \thmref{Theorem}{VI.3}(a)
spells them out.
\end{proof}

\subsection{Lemma VI.2}\label{proof:VI.2}

\begin{restate}{Lemma}{VI.2}{the activation fixes the privileged basis; classical, \citet{xsussmann1992uniqueness},
\citet{xchen1993geometry}, \citet{xgodfrey2022symmetries}}
Let $\sigma:\R\to\R$ be an elementwise activation that is not affine, and let $L\in\GL_d$ commute with it as maps on $\R^d$:
$L\,\sigma^{\times d}=\sigma^{\times d}L$.

(a) If $\sigma\in C^2$, then $L=PD$ is monomial.

(b) If $\sigma$ is ReLU or leaky ReLU with slope $\alpha\ne1$, then $L=PD$ with $D$ a positive diagonal.

(c) If moreover $\sigma$ is real-analytic at $0$ with $\sigma(t)=\sum a_kt^k$, each diagonal entry $c$ of $D$ satisfies
$ca_k=c^ka_k$ for all $k$. So $c=1$ for GELU and SiLU, since $\sigma''(0)\neq0$, and $c=\pm1$ for tanh.
\end{restate}

\begin{proof}
Write $\ell_i$ for row $i$ of $L$; it is nonzero because $L$ is invertible. Row $i$ of the identity
$L\,\sigma^{\times d}(x)=\sigma^{\times d}(Lx)$ reads
\begin{equation}\label{eq:C3-commute-row}
\sum_jL_{ij}\,\sigma(x_j)=\sigma(\ell_i\cdot x)\qquad\text{for all }x\in\R^d .
\end{equation}
Once each row has a single nonzero entry, the positions of these entries are distinct, since otherwise a column of $L$
would vanish; so $L=PD$ with $P$ a permutation matrix and $D$ an invertible diagonal matrix.

(a) For $j\ne k$, the mixed partial $\partial_j\partial_k$ of the left side of \eqref{eq:C3-commute-row} vanishes, since each
term depends on one coordinate, and the mixed partial of the right side is $L_{ij}L_{ik}\,\sigma''(\ell_i\cdot x)$. Since
$\sigma$ is $C^2$ and not affine, $\sigma''(t_0)\ne0$ for some $t_0$, and since $\ell_i\ne0$ there is $x$ with $\ell_i\cdot
x=t_0$. Hence $L_{ij}L_{ik}=0$ for all $j\ne k$: row $i$ has a single nonzero entry.

(b) Let $\sigma(t)=t$ for $t\ge0$ and $\sigma(t)=\alpha t$ for $t<0$, with $\alpha\ne1$ ($\alpha=0$ is ReLU). Then $\sigma$ is
smooth off $0$, and not differentiable at $0$, where its one-sided derivatives $1$ and $\alpha$ differ. The left side of
\eqref{eq:C3-commute-row} is a sum of functions of separate coordinates, so it is differentiable at $x$ iff each term with
$L_{ij}\ne0$ is differentiable in $x_j$; its non-differentiability set is $\bigcup_{j:L_{ij}\ne0}\{x_j=0\}$. The right side is
smooth where $\ell_i\cdot x\ne0$, and at a point with $\ell_i\cdot x=0$ its one-sided directional derivatives along $\pm\ell_i$
are $\lVert\ell_i\rVert^2$ and $-\alpha\lVert\ell_i\rVert^2$, which are not negatives of each other, so it is not
differentiable there; its non-differentiability set is the hyperplane $\{\ell_i\cdot x=0\}$. The two sets are equal. By
\auxref{B.36}(b) the hyperplane is one of the coordinate hyperplanes $\{x_{j_0}=0\}$, and each $\{x_j=0\}$ with $L_{ij}\ne0$
is contained in it and so equal to it; hence $L_{ij}\ne0$ only for $j=j_0$. Then \eqref{eq:C3-commute-row} reads
$L_{ij_0}\sigma(t)=\sigma(L_{ij_0}t)$ for all $t$. At $t=1$ this is $L_{ij_0}=\sigma(L_{ij_0})$; if $L_{ij_0}<0$ the right side
is $\alpha L_{ij_0}$, which would force $\alpha=1$. So $L_{ij_0}>0$, and $D$ is positive.

(c) A permutation matrix commutes with every elementwise map (rule R2 of \thmref{Theorem}{VI.3}), so
$D\sigma^{\times d}=P^{-1}L\sigma^{\times d}=P^{-1}\sigma^{\times d}L=\sigma^{\times d}P^{-1}L=\sigma^{\times d}D$, that is,
$c\,\sigma(t)=\sigma(ct)$ for each diagonal entry $c\ne0$ and all $t$. Near $0$ both sides are power series,
$\sum_kca_kt^k=\sum_ka_kc^kt^k$, so $ca_k=c^ka_k$ for every $k$ (\auxref{B.10}(c)). For GELU, $\sigma(t)=t\,\Phi_{\mathcal
N}(t)$ with the standard normal distribution function $\Phi_{\mathcal N}(t)=\tfrac12+t/\sqrt{2\pi}+O(t^3)$, so
$a_2=1/\sqrt{2\pi}\ne0$; for SiLU, $\sigma(t)=t/(1+e^{-t})=\tfrac t2+\tfrac{t^2}4+O(t^4)$, so $a_2=\tfrac14\ne0$. In both cases
$c=c^2$, and $c\ne0$ gives $c=1$. For tanh, $a_3=-\tfrac13\ne0$, so $c=c^3$ and $c=\pm1$; both occur, tanh being odd.
\end{proof}

\subsection{Theorem VI.3}\label{proof:VI.3}

\begin{restate}{Theorem}{VI.3}{the merge calculus; known in part}
Work with activation-side diagrams in a symmetric monoidal category of continuous, piecewise-smooth maps with copy and
discard nodes. (We call copy and the Hadamard product ``spiders'' by analogy only; in this cartesian setting they do not
satisfy the Frobenius law.) A \emph{trainable box} is a node whose weight ranges over a stated class: all linear maps, or all
affine maps when the box has a bias slot. R2 and R4--R7 are identities of maps; R1 is a closure property and R3 a
characterisation.

\smallskip
{\upshape\footnotesize\renewcommand{\arraystretch}{1.15}\noindent
\begin{tabularx}{\linewidth}{@{}>{\raggedright\arraybackslash}p{0.17\linewidth}>{\raggedright\arraybackslash}p{0.33\linewidth}>{\raggedright\arraybackslash}X@{}}
\toprule\headrow
\textbf{Rule} & \textbf{Statement} & \textbf{When it holds}\\
\midrule
R1 (fusion) & composites and parallel sums of linear (affine) boxes are linear (affine) & the operators involved are
position-independent\\
R2 (permutation slide) & $P\sigma^{\times d}=\sigma^{\times d}P$ & every elementwise $\sigma$\\
R3 (positive-diagonal slide) & $\diag(\ell)\,\sigma^{\times d}=\sigma^{\times d}\diag(\ell)$ for all $\ell\ge0$ & iff
$\sigma$ is positively homogeneous (ReLU, leaky ReLU; not GELU, SiLU, tanh); such $\sigma$ are piecewise linear\\
R4 (spider slide) & $\diag(\ell)(a\odot b)=(\diag(\ell)a)\odot b$ & any $\ell$\\
R5 (cup slide) & $\langle Mq,k\rangle=\langle q,M^\top k\rangle$ & always. Under RoPE,
$\langle R_mMq,R_nk\rangle=\langle q,M^\top R_{n-m}k\rangle$ holds for every $M$, but the slide onto the key side,
$\langle R_mMq,R_nk\rangle=\langle R_mq,R_nM^\top k\rangle$ for all $m,n$, holds iff $M^\top$ commutes with every
$R_{n-m}$ that occurs; for diagonal $M=\diag(\ell)$ and distinct frequencies, iff $\ell$ is constant on each rotary pair
(in either pairing convention)\\
R6 (norm absorption) & $W(\gamma\odot n(x)+\beta)=(W\diag\gamma)\,n(x)+W\beta$ & always; the shift $W\beta$ needs a bias
slot, and an RMSNorm gain has no $\beta$\\
R7 (copy slide) & applying $D$ and then copying equals copying and then applying $D$ to every copy & always\\
\bottomrule
\end{tabularx}}
\smallskip

Consequences:

(a) \textbf{Affine edits.} Let a method insert a position-uniform affine edit $E_q$ on a wire $w$ of the base network $f$,
for every $q\in Q$. Suppose every reader of $w$ is a trainable, unshared box (no weight tying) whose class is closed under
precomposition with $E_q$; a shift needs a bias slot. Then the method merges, by $W_c\mapsto W_cE_q$ for all readers $c$ at
once. Dually, it merges into an untied trainable box that writes $w$ whenever $E_q$ composed with that box stays in its
class. Absorption into a box that the extension itself inserts does not count. Edits that violate these hypotheses fall
outside (a): on the reader side, R1--R7 do not merge position-selective edits or edits on a wire that is also read by a skip
connection, a norm or the network output. Whole-network non-mergeability is open. Position-selective edits (ReFT/LoReFT, and
position-gated LoRA variants such as Activated LoRA) are not box-locally mergeable. Residual-stream edits need case
analysis: an edit merges into the box that produces the wire when that box can absorb it (an untied embedding table; for a
fixed-position edit at layer $0$, a learned absolute positional table), a shift folds into a writer's bias when one exists,
a diagonal edit after a post-LN folds into that LN's affine parameters, and an orthogonal edit in an RMSNorm model merges
globally, SliceGPT/QuaRot-style. No merge is known for LoReFT's non-orthogonal low-rank edit.

(b) \textbf{Diagonal methods.} A diagonal method merges into a box reached by R1--R7 if all of the following hold: the rules
move \emph{every} copy of $\diag(\ell)$ into trainable boxes for all $\ell$ in the method's parameter domain (with $\ell\ge0$
when R3 is used); no position-dependent operator lies on the path (for RoPE, $\ell$ is constant on each rotary pair); the
target box is unshared, or $\ell$ is constant across the sharing group; and every shift has a bias slot. Conversely, the
RoPE condition is necessary for sliding a query scaling onto $W_K$ when $W_Q$ and $W_K$ have full rank and at least two
occurring relative offsets have rotation angles that are not multiples of $\pi$. Without such genericity hypotheses the
conditions are not necessary: a dead rotary pair, or zero entries of $\ell$ under ReLU, can still merge.

(c) \textbf{Non-affine edits.} For a linear or affine box, an edit whose modified box function is non-affine in the box
input, for some parameter value, is not box-locally mergeable. For linear and affine boxes this is the definition; for
non-linear boxes such as attention, see \thmref{Theorem}{VI.5}(f).

(d) \textbf{Type relativity.} A generic high-precision update is off the quantisation grid, so exact merging of a QLoRA or
LoftQ update leaves the quantised type, a type-preserving merge is lossy, and QA-LoRA's group-constant adapters are the
exception (\thmref{Proposition}{V.2}(d)).
\end{restate}

\begin{proof}
A diagram denotes a map on activations, and two diagrams related by an identity of maps compute the same network function;
a rewrite that ends with every edit inside trainable boxes of the base network therefore exhibits a merge $\mu$, the new
weights of those boxes. ``Position-independent'' means that the same map acts at every token position.

\emph{R1.} Composites and sums of linear (affine) maps are linear (affine). If each acts identically at every position, so
does the result, which is then computed by one weight (and bias) at all positions.

\emph{R2.} For a permutation $\pi$ with matrix $P$, $(P\sigma^{\times d}(x))_i=\sigma(x_{\pi^{-1}(i)})=\sigma((Px)_i)$.

\emph{R3.} Coordinatewise the identity says $\ell_i\sigma(t)=\sigma(\ell_it)$ for all $\ell_i\ge0$ and $t$; with $d=1$ this is
positive homogeneity of degree one, and conversely positive homogeneity gives the identity in every coordinate. A positively
homogeneous $\sigma$ satisfies $\sigma(t)=t\,\sigma(1)$ for $t\ge0$ and $\sigma(t)=\lvert t\rvert\,\sigma(-1)$ for $t<0$, so it
is linear on each half-line. ReLU and leaky ReLU are of this form. GELU, SiLU and tanh are differentiable at $0$ and not
linear, so by \auxref{B.36}(a) they are not positively homogeneous.

\emph{R4} holds entrywise, $\ell_ia_ib_i=(\ell_ia_i)b_i$.

\emph{R5.} $\langle Mq,k\rangle=q\T M\T k=\langle q,M\T k\rangle$. Under RoPE, $R_m$ is block-diagonal with the planar rotations
$\rho(m\omega_j)$ on the rotary pairs, in either pairing convention, so $R_m\T=R_{-m}$ and $R_m\T R_n=R_{n-m}$. Hence
$\langle R_mMq,R_nk\rangle=q\T M\T R_{n-m}k=\langle q,M\T R_{n-m}k\rangle$ for every $M$, while
$\langle R_mq,R_nM\T k\rangle=\langle q,R_{n-m}M\T k\rangle$. These agree for all $q,k$ iff $M\T R_{n-m}=R_{n-m}M\T$, and for all
occurring $(m,n)$ iff $M\T$ commutes with every occurring $R_\Delta$. For $M=\diag(\ell)$ the commutation holds pair by pair.
On a pair with entries $(\ell_a,\ell_b)$ and rotation angle $\varphi$,
\[\begin{pmatrix}\ell_a&0\\0&\ell_b\end{pmatrix}\begin{pmatrix}\cos\varphi&-\sin\varphi\\ \sin\varphi&\cos\varphi\end{pmatrix}
-\begin{pmatrix}\cos\varphi&-\sin\varphi\\ \sin\varphi&\cos\varphi\end{pmatrix}\begin{pmatrix}\ell_a&0\\0&\ell_b\end{pmatrix}
=(\ell_b-\ell_a)\sin\varphi\begin{pmatrix}0&1\\1&0\end{pmatrix},\]
which vanishes iff $\ell_a=\ell_b$ or $\varphi\in\pi\mathbb Z$. So the slide holds iff $\ell$ is constant on every pair whose
angle $\Delta\omega_j$ is not a multiple of $\pi$ for some occurring $\Delta$. For RoPE, $0<\omega_j\le1$, so every pair
qualifies as soon as the offset $\Delta=1$ occurs; distinct frequencies are not needed for diagonal $M$.

\emph{R6.} By linearity, $W(\gamma\odot n(x)+\beta)=W\diag(\gamma)\,n(x)+W\beta$. The first term is a linear box applied to the
normalised wire, and the constant $W\beta$ is absorbed by a bias slot, $b\mapsto b+W\beta$; an RMSNorm gain has $\beta=0$.

\emph{R7.} With $\Delta(x)=(x,x)$ the copy map, $\Delta\circ D=(D\times D)\circ\Delta$ for every map $D$: the naturality of the
cartesian diagonal (\bgref{B.4}).

(a) Let $E_q(x)=A_qx+c_q$ act at every position on $w$. By R7 applying $E_q$ before the copy node that feeds the readers
equals applying it on each reader's input. For a reader $c$ with weight $W_c$ and bias $b_c$,
$W_c(A_qx+c_q)+b_c=(W_cA_q)x+(W_cc_q+b_c)$, which lies in the class of $c$ by the closure hypothesis (for a linear class this
requires $c_q=0$, and for a shift a bias slot). The readers are unshared, so their weights can be changed independently; and
$E_q$ is the same at every position, so one weight serves all positions (R1). Setting $\mu(q)$ to the base weights with
$W_c\mapsto W_cA_q$ and $b_c\mapsto b_c+W_cc_q$ for every reader gives $\Phi_f\circ\mu=\Phi_{\tilde f}\circ\rho$. Dually, if
$w$ is the output of an untied box $p(x)=W_px+b_p$ and $E_q\circ p$ lies in its class, replace $p$ by $E_q\circ p$, that is,
$W_p\mapsto A_qW_p$ and $b_p\mapsto A_qb_p+c_q$ (R1). A merge must land in the base network $\Theta$, so absorption into a box
that only the extension contains is not a merge. On the reader side the rules have nothing to absorb an edit read by a skip
connection, a norm or the network output, since those readers carry no trainable weight, and R1 does not fuse a
position-selective edit into a weight that acts at every position. A position-selective edit modifies the box function at
some positions only; for parameter values at which it is not the identity, the modified computation is not one map applied
identically at every position, so it is not in the box's class, and the edit is not box-locally mergeable.

\emph{Residual-stream cases.} (i) If $w$ is the output of an untied embedding table $\mathcal E$, the dual case applies:
replacing each row $e(v)$ by $A_qe(v)+c_q$ absorbs a position-uniform affine edit. For an additive edit $h\mapsto h+c$ at one
fixed position $p_0$ of layer $0$ in a model with a learned absolute positional table, replacing row $p_0$ of that table by
itself plus $c$ absorbs it. (ii) A shift $h\mapsto h+t$ of the stream after a block whose last box has a bias $b$ is absorbed
by $b\mapsto b+t$, since the block adds its output to the stream. (iii) In a post-LN model the normalised wire is the stream,
$h=\gamma\odot n(x)+\beta$, and a diagonal edit gives $s\odot h+t=(s\odot\gamma)\odot n(x)+(s\odot\beta+t)$, absorbed by
$\gamma\mapsto s\odot\gamma$, $\beta\mapsto s\odot\beta+t$. (iv) In an untied pre-RMSNorm model, let block $j$ map the stream by
$h\mapsto h+F_j(h)$ with $F_j(h)=W^{(j)}_{\rm out}\,G_j\big(W^{(j)}_{\rm in}\diag(\gamma_j)\,n(h)\big)$, where $n(h)=h/\mathrm{rms}(h)$,
$G_j$ is the internal computation (attention with its positional encoding applied after the input projections, or the gated
MLP), and $W^{(j)}_{\rm in}$ collects the input projections. Let $Q$ be orthogonal and the edit be $h\mapsto Qh$ after block
$l$. Replace the embedding rows $e(v)$ by $Qe(v)$, and for $j\le l$ replace $W^{(j)}_{\rm in}$ by $W^{(j)}_{\rm in}\diag(\gamma_j)Q\T$,
$\gamma_j$ by $\mathbf 1$ and $W^{(j)}_{\rm out}$ by $QW^{(j)}_{\rm out}$ (and any bias $b$ of an output projection by $Qb$). Since
$\mathrm{rms}(Qh)=\mathrm{rms}(h)$, $n(Qh)=Qn(h)$, and by induction the new streams are $Qh_j$ for $j\le l$:
$Qh_{j-1}+QW^{(j)}_{\rm out}G_j\big(W^{(j)}_{\rm in}\diag(\gamma_j)Q\T Qn(h_{j-1})\big)=Q\big(h_{j-1}+F_j(h_{j-1})\big)$. After block $l$
the new network carries $Qh_l$, as the edited one does, and the later blocks are unchanged, so the outputs agree. This is the
construction of \citet{xashkboos2024slicegpt,xashkboos2024quarot}.

(b) The rules used along the path are identities of maps under their side conditions: R3 needs $\ell\ge0$, the slide of R5
under RoPE needs $\ell$ constant on the rotary pairs, and R6 needs a bias slot for the shift. Applying them moves every copy of
$\diag(\ell)$ next to a trainable box, and R1 fuses it; this is (a) applied at the box reached. An unshared target can absorb
its copy independently, and a target shared across a group absorbs a single matrix, which is possible for every member exactly
when $\ell$ is constant across the group. For the converse under RoPE, suppose the query scaling $D=\diag(\ell)$ merges into
some $W_K'$ with $W_Q$ unchanged. Equality of all logits for all inputs gives
$x\T W_Q\T DR_\Delta W_Kx'=x\T W_Q\T R_\Delta W_K'x'$ for every occurring offset $\Delta=n-m$, so
$W_Q\T(R_\Delta W_K'-DR_\Delta W_K)=0$. If $W_Q$ has full row rank, $W_Q\T$ is injective and $W_K'=R_{-\Delta}DR_\Delta W_K$ for
every occurring $\Delta$; if $W_K$ has full row rank too, it has a right inverse, and $R_{-\Delta}DR_\Delta$ is one matrix for
all occurring $\Delta$. On a rotary pair $j$, let $\Delta$ and $\Delta'$ be occurring offsets with $(\Delta-\Delta')\omega_j\notin\pi
\mathbb Z$; then $D$ commutes with $R_{\Delta-\Delta'}$ on that pair, and by the computation in R5, $\ell$ is constant on it. In
causal attention $\Delta'=0$ occurs (a query attends to its own position), so it suffices that one occurring offset has an angle
on pair $j$ that is not a multiple of $\pi$, which the hypothesis provides pair by pair. The genericity hypotheses are needed.
If $W_Q$ has a dead pair (both rows zero) and $\ell$ is constant on every other pair, then $W_Q\T DR_\Delta=W_Q\T R_\Delta D$,
since both sides vanish on the dead pair and $D$ commutes with $R_\Delta$ on the others, so $W_K'=DW_K$ works with $\ell$
arbitrary on the dead pair. Under ReLU, $\diag(\ell)\mathrm{ReLU}(Wx)=\mathrm{ReLU}(\diag(\ell)Wx)$ holds for every $\ell\ge0$,
zero entries included.

(c) For a linear or affine box the class is the linear or affine maps, so a non-affine modified box function is outside it,
which is the definition of failing box-local mergeability.

(d) is \thmref{Proposition}{V.2}(d).
\end{proof}

\paragraph{The corollaries after \thmref{Theorem}{VI.3}}
\emph{Weight-space methods} merge by $\mu=\rho$ (\S\ref{proof:VI.1}); their nonlinearities act on the parameters. If an
embedding and an LM head share one tensor, a merge that changes one side must change the other or break the tie, which is why
an adapter on one side is not merged faithfully when the tie is not maintained. \emph{(IA)$^3$ on keys and values} is
$\diag(\ell)W_Kx=(\diag(\ell)W_K)x$ and likewise for $W_V$ (R1). \emph{On the FFN}, applied to the input of $W_{\rm down}$, it is
$W_{\rm down}\diag(\ell)h=(W_{\rm down}\diag(\ell))h$ (R1); neither step involves $\sigma$. \emph{Into $W_{\rm up}$}: under ReLU,
R3 gives $\diag(\ell)\mathrm{ReLU}(W_{\rm up}x)=\mathrm{ReLU}(\diag(\ell)W_{\rm up}x)$ for $\ell\ge0$, while an entry $\ell_i<0$
cannot fuse, since $\ell_i\mathrm{ReLU}(w_i\cdot x)\le0$ and is negative somewhere, whereas $\mathrm{ReLU}(w_i'\cdot x)\ge0$. Under
SwiGLU or GeGLU, R4 and R1 give \eqref{eq:merge-swiglu} for every $\ell$. \emph{Under GELU}, suppose
$\ell_i\,\mathrm{GELU}(w_i\cdot x)=\mathrm{GELU}(w_i'\cdot x)$ for all $x$, with $w_i\ne0$ so that the pre-activation ranges over
$\R$. If $\ell_i=0$, $w_i'=0$ works. If $\ell_i\ne0$, the right side vanishes on the hyperplane $w_i\cdot x=0$, and
$\mathrm{GELU}(t)=t\,\Phi_{\mathcal N}(t)$ vanishes only at $t=0$, so $w_i'\cdot x=0$ there and $w_i'=cw_i$. Then
$\ell_i\,\mathrm{GELU}(t)=\mathrm{GELU}(ct)$ for all $t$, and comparing the coefficients $a_1=\tfrac12$ and $a_2=1/\sqrt{2\pi}$
(\auxref{B.10}(c)) gives $c=\ell_i$ and $\ell_i=\ell_i^2$, so $\ell_i=1$. Hence a fused $W_{\rm up}'$ exists iff every
$\ell_i\in\{0,1\}$. \emph{A query scaling} fuses into $W_Q$ by R1; onto $W_K$ it needs R5's slide, hence $\ell$ constant on the
rotary pairs under RoPE (necessary under the genericity hypotheses of (b)), and with grouped-query attention one $W_K$ serves
several query heads $h$, so the same argument with full-rank $W_Q^{(h)}$ forces $W_K'=D_hW_K$ for every $h$ in the group, hence
$D_h$ constant across the group when $W_K$ has full row rank. \emph{SSF} after a box $Wx+b$ gives
$s\odot(Wx+b)+t=(\diag(s)W)x+(s\odot b+t)$, which needs a bias slot when $t\ne0$. \emph{LN-tuning} changes existing parameters, so
$\mu$ is the new parameter vector. \emph{Folding a LayerNorm} into the following boxes in a pre-LN block, where the norm feeds only
those boxes, is R7 followed by R6, which needs a bias slot for $W\beta$; an RMSNorm gain has no $\beta$. Folding a gain into a
tied embedding or head changes the other side of the tie. \emph{Not box-locally mergeable}: for $\sigma$ non-affine and nonzero
$u$, $d$, the map $y\mapsto y+u\,\sigma(d\T y)$ is non-affine, so a serial Houlsby or Pfeiffer adapter after a box $W$ with
$d\T W\ne0$, and a parallel adapter $x\mapsto Wx+u\,\sigma(d\T x)$, modify the box function non-affinely for some parameter
value; a routed mixture $x\mapsto Wx+\sum_eg_e(x)B_eA_ex$ with a softmax gate is real-analytic in its input and parameters and
non-affine at a witness (two experts with $B_1A_1=I$, $B_2A_2=0$ and gate logits differing by $\gamma\T x$, $\gamma\ne0$, give
$Wx+g(\gamma\T x)\,x$ with $g$ the logistic function, non-affine along any line on which $\gamma\T x$ varies), hence non-affine for
generic parameters (\auxref{B.10}), and linear, so mergeable, when the gate is constant; AdapterFusion with a constant gate is a fixed combination of nonlinear adapters,
still non-affine; LST adds the output of a nonlinear side network; and prefix-type methods are treated in
\thmref{Theorem}{VI.5}(f). In each case (c) applies. For LLaMA-Adapter with a two-token prompt $p_1,p_2$ and gate $g\ne0$,
the prompt pairs are $k_i=W_kp_i+b_k$ and $v_i=W_vp_i+b_v$, and a head on a length-one context computes
$W_o(W_vx+b_v)+g\,W_o\mathrm{Attn}_{q(x)}(P)+b_o$ with
$\mathrm{Attn}_{q}(P)=v_2+\operatorname{logit}^{-1}\big(\langle q,W_k\delta\rangle/\sqrt d\big)\,W_v\delta$, $\delta=p_1-p_2$.
If $W_q\T W_k\ne0$ and $W_oW_v\ne0$, choose $u$ and $\delta$ with $\langle W_qu,W_k\delta\rangle\ne0$ and $W_oW_v\delta\ne0$
(each condition fails only on a proper subspace or a nowhere-dense set). Along $x=tu$ the logistic factor is a bounded,
non-constant real-analytic function of $t$, hence not affine, so the head is not affine. By \auxref{B.10} the same holds for
generic prompts with two or more tokens, which is the entry for LLaMA-Adapter in the summary table of
Exposé~\ref{sec:merge}; with one prompt token, $\mathrm{Attn}_q(P)$ is constant and the head stays affine.

\subsection{Proposition VI.4}\label{proof:VI.4}

\begin{restate}{Proposition}{VI.4}{linear insertions, and ReFT}
(a) A linear adapter in parallel with a single linear box, $W_0x+sUDx$, \emph{is} LoRA: it has the same $\rho$. A linear
adapter parallel to a whole sublayer (He et al.'s parallel adapter around attention or the FFN) edits the weightless skip
connection rather than a base box, and is not box-locally mergeable when the sublayer is non-affine.

(b) Let $W_0\in\R^{m\times n}$ be bias-free (as LLaMA-type $W_o$ and $W_{\mathrm{down}}$ are), and $r\ge1$. A serial linear
adapter after $W_0$, $(I+UD)W_0$ with $U\in\R^{m\times r}$ and $D\in\R^{r\times m}$ (scale absorbed), has image
$W_0+\{Y\in\Mr:\operatorname{row}Y\subseteq\operatorname{row}W_0\}$. It is expressively equivalent to LoRA iff $\rk W_0=n$,
and weakly (unpointed) isomorphic to LoRA iff $W_0$ is invertible. The isomorphism $(U,D)\mapsto(U,DW_0)$ sends a base point
$(U_0,D_0)$ to $(U_0,D_0W_0)$. For a fixed $\mathrm T_A$ pointing $(0,A_0)$ with $\rk A_0=r$ and $1\le r<m=n$, a pointed
isomorphism exists iff $W_0$ is invertible and $\operatorname{row}A_0=\operatorname{row}(D_0W_0)$, and then $U_0=0$
automatically. With a bias $b_0$, the serial adapter also moves the bias by $UDb_0$, so on the affine box its image differs
from weight-only LoRA's. Houlsby's near-identity initialisation (weights from a zero-mean Gaussian with standard
deviation $10^{-2}$, truncated at two standard deviations) makes practical serial adapters slightly unpointed; the pointed
statement applies to zero-up initialisations.

(c) LoReFT is $\Phi(h)=h+R^\top(Wh+b-Rh)$, with $R\in\R^{r\times d}$ having orthonormal rows, applied at selected positions.
On one hidden vector it is the extension by an \textbf{identity box} followed by a rank-$\le r$ affine edit,
\[\Phi(h)=\big(I+R^\top(W-R)\big)h+R^\top b,\]
which is LoRA$_r$ on the \emph{affine} identity box $[I_d\mid0]$ in homogeneous coordinates, with $B=R^\top$ constrained to the
Stiefel manifold and $A=[W-R\mid b]$. Its image is $(I,0)+\MM_{\le r}(d\times(d+1))$, of dimension $r(2d+1-r)$. The bias is
confined to $\operatorname{col}B$, so for $r<d$ this is not LoRA$_r{}\oplus{}$BitFit, whose dimension is larger by $d-r$. It is
simulated, through $h(R,W,b)=(R^\top,[W-R\mid b])$, by LoRA$_r$ on that affine box pointed at $(R_0^\top,0)$, a $\mathrm T_B$
pointing with $\operatorname{col}B_0=\operatorname{row}R_0$; this homogeneous-coordinate LoRA has a bias column in $A$, unlike
HF LoRA. It is not simulated by $\mathrm T_A$-pointed LoRA$_r$, since its $\dim S_1=r(d+1)$ exceeds $rd$. It is not isomorphic
to LoRA$_r$: its parameter count is smaller by $r(r+1)/2$, and its residual symmetry is $\Ort(r)$, acting by
$(R,W,b)\mapsto(OR,OW,Ob)$. It is pointed iff $W=R$ and $b=0$. The reference implementation (pyreft) initialises $W$ and $b$
by the default \texttt{nn.Linear} init, independently of $R$, so practical LoReFT has a pointing defect
$R^\top((W_0-R_0)h+b_0)$: at initialisation it replaces the $R$-component of $h$ by a random affine function, and the
simulation above is then only a weak morphism. pyreft also applies dropout to the whole output, $h$ included, so the
training-time map is not $\Phi$.
\end{restate}

\begin{proof}
(a) The adapted box computes $x\mapsto(W_0+sUD)x$, so $\rho(U,D)=W_0+sUD$ on $\R^{m\times r}\times\R^{r\times n}$, which is
LoRA's $\rho(B,A)=W_0+sBA$ under $(U,D)=(B,A)$, with the zero-up pointing $U=0$ \citep[Table~1]{he2021unified}. A linear adapter
around a whole sublayer $F$ gives $x\mapsto x+F(x)+sUDx$; the term $sUDx$ is added to the weightless skip connection, which no
box of the base network computes, so there is no base box whose function the edit modifies. For a non-affine sublayer such as
$F(x)=W_2\sigma(W_1x)$ with $\sigma$ non-affine, no base box both reads $x$ and writes linearly into the block's output: $W_1$'s
output passes through $\sigma$, and $W_2$ reads $\sigma(W_1x)$ rather than $x$. So no box of $F$ can absorb $sUDx$ within its
class. If $F(x)=Mx+c$ were affine, $x+(M+sUD)x+c$ would fuse into $M$.

(b) \emph{Image.} $\rho(U,D)=(I+UD)W_0=W_0+UDW_0$. Each $UDW_0$ has rank at most $r$ and its rows are combinations of the rows
of $W_0$. Conversely, if $\rk Y\le r$ and $\operatorname{row}Y\subseteq\operatorname{row}W_0$, then $Y=YW_0^+W_0$, since
$W_0^+W_0$ is the orthogonal projector onto $\operatorname{row}W_0$; $X=YW_0^+$ has rank at most $r$, so $X=UD$ with
$U\in\R^{m\times r}$, $D\in\R^{r\times m}$, and $Y=UDW_0$. \emph{Expressive equivalence.} LoRA's image is $W_0+\Mr$. If
$\operatorname{row}W_0=\R^n$ the two images agree. If not, pick $v\notin\operatorname{row}W_0$; then $e_1v\T\in\MM_{\le1}\subseteq
\Mr$ (here $r\ge1$) has a row outside $\operatorname{row}W_0$, so it lies in LoRA's image minus $W_0$ but not in the
adapter's.

\emph{Weak isomorphism.} If $W_0$ is invertible, then $m=n$ and $h(U,D)=(U,DW_0)$ is a linear bijection
$\R^{m\times r}\times\R^{r\times m}\to\R^{m\times r}\times\R^{r\times n}$ with $W_0+U(DW_0)=\rho(U,D)$ (LoRA with $s=1$, or $U$
rescaled), so it is an isomorphism over $\Theta$; it sends $(U_0,D_0)$ to $(U_0,D_0W_0)$. Conversely, a weak isomorphism is a
diffeomorphism $h$ with $\rho_{\mathrm{LoRA}}\circ h=\rho$, so the images agree, which forces $\rk W_0=n$
(\thmref{Observation}{I.4}(c)), and the dimensions agree, $2mr=r(m+n)$, which forces $m=n$; so $W_0$ is square of full rank.

\emph{Pointed isomorphism.} Let $1\le r<m=n$, $\rk A_0=r$, and let the adapter be pointed at $(U_0,D_0)$, that is,
$U_0D_0W_0=0$. ``If'': $W_0$ invertible gives $U_0D_0=0$. Since $\operatorname{row}(D_0W_0)=\operatorname{row}A_0$ has dimension
$r$, $D_0W_0$ and hence $D_0$ have rank $r$, so $D_0$ has a right inverse and $U_0=0$. Two $r\times n$ matrices of rank $r$ with
the same row space differ by an invertible left factor, so $A_0=gD_0W_0$ for some $g\in\GL_r$. The map
$(U,D)\mapsto(Ug^{-1},gDW_0)$ is a linear bijection over $\Theta$ that sends $(U_0,D_0)=(0,D_0)$ to $(0,A_0)$. ``Only if'': a
pointed isomorphism is a weak one, so $W_0$ is invertible, and it preserves $S_1$ (\thmref{Observation}{I.6}). Differentiating,
\begin{align*}
S_1(\text{adapter at }(U_0,D_0))&=\{(XD_0+U_0Y)W_0\}=\{XD_0W_0+U_0Y'\},\\
S_1(\mathrm{LoRA}\text{ at }(0,A_0))&=\{XA_0\},
\end{align*}
with $X\in\R^{m\times r}$ and $Y'=YW_0$ ranging over $\R^{r\times n}$ because $W_0$ is invertible; and $\{XA_0\}=\{Z:
\operatorname{row}Z\subseteq\operatorname{row}A_0\}$. If $(U_0)_{ij}\ne0$, choose $v\notin\operatorname{row}A_0$, possible
since $r<n$, and $Y'=e_jv\T$: row $i$ of $U_0Y'$ is $(U_0)_{ij}v\T$, outside $\operatorname{row}A_0$, a contradiction. So $U_0=0$,
and $\{XD_0W_0\}=\{XA_0\}$ gives $\operatorname{row}(D_0W_0)=\operatorname{row}A_0$.

\emph{Bias.} $(I+UD)(W_0x+b_0)=(W_0+UDW_0)x+(b_0+UDb_0)$. On the affine box the adapter's image contains maps with bias
$b_0+UDb_0\ne b_0$ whenever $b_0\ne0$ (choose $UD$ with $UDb_0\ne0$), while weight-only LoRA keeps the bias $b_0$. \emph{Houlsby
initialisation.} With both factors drawn at random, $U_0D_0W_0\ne0$ almost surely, so the initial weight misses $W_0$.

(c) \emph{Algebra.} Expanding, $\Phi(h)=h+R\T(W-R)h+R\T b=\big(I+R\T(W-R)\big)h+R\T b$. In homogeneous coordinates $(h,1)$ an
affine map $h\mapsto Mh+c$ is the matrix $[M\mid c]\in\R^{d\times(d+1)}$, the identity box is $[I_d\mid0]$, and
\[\big[\,I+R\T(W-R)\ \big|\ R\T b\,\big]=[I_d\mid0]+R\T\,[W-R\mid b]=[I_d\mid0]+BA,\qquad B=R\T,\ A=[W-R\mid b].\]
\emph{Image.} As $(W,b)$ ranges freely, $V=[W-R\mid b]$ ranges over all of $\R^{r\times(d+1)}$, so the image is
$[I_d\mid0]+\{R\T V\}$ over Stiefel $R\T$. Every matrix of rank at most $r$ factors as $R\T V$ with $R\T$ having orthonormal
columns (\auxref{B.11}(a)), so the image is $[I_d\mid0]+\MM_{\le r}(d\times(d+1))$, of dimension $r(d+(d+1)-r)=r(2d+1-r)$
(\bgref{C.12}). The bias $R\T b$ lies in $\operatorname{col}R\T=\operatorname{col}B$. LoRA$_r{}\oplus{}$BitFit has image
$\{[I_d+Y\mid c]:\rk Y\le r,\ c\in\R^d\}$, of dimension $r(2d-r)+d$, which exceeds $r(2d+1-r)$ by $d-r$; for $r<d$ the images
differ. \emph{Simulation.} The map $h(R,W,b)=(R\T,[W-R\mid b])$ is smooth, and LoRA$_r$ on the affine box evaluated at $h(R,W,b)$ is
$[I_d\mid0]+R\T[W-R\mid b]=\Phi$, so $h$ is a morphism to that LoRA$_r$. It sends the neutral point $(R_0,R_0,0)$ to $(R_0\T,0)$, a pointing in the stratum $\mathrm T_B$
($A_0=0$, $B_0$ of rank $r$) with $\operatorname{col}B_0=\operatorname{row}R_0$. \emph{$S_1$.} At the neutral point
$[W-R\mid b]=0$, so the derivative of $R\T[W-R\mid b]$ is $R_0\T[\dot W-\dot R\mid\dot b]$, and as $(\dot W,\dot b)$ range freely
$S_1=\{R_0\T V\}$, of dimension $r(d+1)$ because $R_0\T$ is injective. LoRA$_r$ pointed at $(0,A_0)$ with $\rk A_0=r$ has
$S_1=\{XA_0\}$, of dimension $dr$. A pointed morphism would give an inclusion of $S_1$'s (\thmref{Observation}{I.6}), impossible
since $r(d+1)>rd$.

\emph{Not isomorphic.} The parameter space is the Stiefel manifold of orthonormal $r$-frames in $\R^d$, of dimension
$rd-r(r+1)/2$, times $\R^{r\times d}\times\R^r$, of total dimension $2rd+r-r(r+1)/2$; LoRA$_r$ on the affine box has dimension
$r(2d+1)$. The difference $r(r+1)/2$ is positive, so no diffeomorphism, hence no isomorphism, exists. \emph{Symmetry.} For
$O\in\Ort(r)$, $OR$ has orthonormal rows and $(OR)\T(OW-OR)=R\T(W-R)$, $(OR)\T Ob=R\T b$, so $\Phi$ is unchanged. Conversely, if
$R\T V=R'^{\top}V'$ with $\rk V=r$, then the column space of the product is $\operatorname{col}R\T=\operatorname{col}R'^{\top}$, so
$R'=OR$ for some $O\in\Ort(r)$, and injectivity of $R\T$ gives $V'=OV$, that is, $W'=OW$ and $b'=Ob$: the generic fibres are the
$\Ort(r)$-orbits, of dimension $r(r-1)/2$, against $r^2$ for LoRA. \emph{Pointedness.} $\Phi=\id$ iff $R\T[W-R\mid b]=0$ iff
$W=R$ and $b=0$, because $R\T$ is injective.

\emph{The pyreft initialisation.} At $(R_0,W_0,b_0)$ with $W_0$, $b_0$ drawn independently of $R_0$, $\Phi_0(h)-h=R_0\T\big((W_0-R_0)
h+b_0\big)$, which is nonzero almost surely. Since $R_0R_0\T=I_r$, $R_0\Phi_0(h)=R_0h+(W_0h+b_0-R_0h)=W_0h+b_0$, while
$(I-R_0\T R_0)\Phi_0(h)=(I-R_0\T R_0)h$: the $R_0$-component of $h$ is replaced by the random affine function $W_0h+b_0$ and the
rest is kept. The map $h$ sends this initial point to $(R_0\T,[W_0-R_0\mid b_0])$, not to the base point $(R_0\T,0)$, so it is a
weak morphism only. The remark on dropout describes the implementation \citep{wu2024reft}.
\end{proof}

\subsection{Theorem VI.5}\label{proof:VI.5}

\begin{restate}{Theorem}{VI.5}{attention is a homomorphism followed by a perspective map; known in part}
Fix a query $q$. Multisets of key--value pairs form a commutative monoid under $\uplus$. Define
\[\Sigma_q(K)=\Big(\sum e^{\langle q,k\rangle/\sqrt d},\ \sum e^{\langle q,k\rangle/\sqrt d}\,v\Big).\]

(a) $\Sigma_q$ is a monoid homomorphism into $(\R_{\ge0}\times\R^{d_v},+)$, landing in $\{Z>0\}\cup\{(0,0)\}$. On nonempty
multisets, $\mathrm{Attn}_q=\pi\circ\Sigma_q$ with the partial map $\pi(Z,S)=S/Z$, defined for $Z>0$.

(b) \textbf{Prefix is a gated parallel adapter.} For nonempty $P$ and $C$,
\[\mathrm{Attn}_q(P\uplus C)=(1-\lambda_q)\,\mathrm{Attn}_q(C)+\lambda_q\,\mathrm{Attn}_q(P),\qquad
\lambda_q=\frac{Z_q(P)}{Z_q(P)+Z_q(C)}.\]

(c) \textbf{Per-layer pattern rigidity.} Within a single layer, for fixed $q$ and a fixed content multiset $C$: the content
attention weights are multiplied by $1-\lambda_q$, their ratios are unchanged, and the output lies on the segment from
$\mathrm{Attn}_q(C)$ to $\mathrm{Attn}_q(P)\in\operatorname{conv}V_P$, where $V_P$ is the set of prefix values. This rigidity
is relative to the same layer's computation without the prefix; it is relative to the pretrained pattern only at the lowest
prefixed layer. Lower-layer prefixes change the queries and the content keys and values of higher layers, so attention
patterns at the level of the network are \emph{not} rigid.

(d) \textbf{LLaMA-Adapter} normalises $P$ and $C$ separately and adds a zero-initialised gate. It computes
$\mathrm{Attn}_q(C)+\tanh(g_l)\,\mathrm{Attn}_q(P)$, with one gate per head in the paper; HF PEFT uses one raw
zero-initialised scalar per layer, without $\tanh$. It uses two perspective maps, leaves the content weights exactly
unchanged, and is pointed at $g=0$. Its prompt keys and values are the frozen $W_k,W_v$ applied to one learned prompt per
adapted layer, and it acts in the top layers only.

(e) \textbf{Prompt $\to$ deep prefix in causal decoders.} Assume that the soft prompt comes first, that attention is causal,
and that content positions are offset by the prompt length $m$ in both methods (as HF PEFT does via \texttt{cache\_position}).
Then at every layer the states at prompt positions depend only on $p$, and $h:p\mapsto(K_j(p),V_j(p))_j$ satisfies
\[\Phi\circ\rho_{\mathrm{prefix}}\circ h=\Phi\circ\rho_{\mathrm{prompt}}\]
as functions of the content tokens, read at content positions. The two methods live on different extensions, so this is an
arrow of the unpointed slice over the realisation $C^\infty(X_{\mathrm{content}},Y_{\mathrm{content}})$, not of $\Meth$.
Under RoPE without the offset, a different $h$ works: rotate each prefix key by $-m$ (positions $-m,\dots,-1$). For additive
relative biases (ALiBi, T5) the answer depends on how the implementation places the prefix keys. The construction does not
apply in bidirectional encoders of depth $\ge2$.

(f) \textbf{Obstructions.} Call a head \emph{standard} if it is vanilla softmax attention without torch-style
\texttt{add\_bias\_kv} (which would absorb a one-token prefix), \texttt{add\_zero\_attn}, learned attention sinks or
softmax-plus-one. On length-one contexts, every weight setting of a standard head is affine in $x$. Some prefix makes the
prefixed head non-affine on length-one contexts iff $W_o\ne0$ and ($W_q\ne0$ or $W_k^\top b_q\ne0$); for bias-free heads, iff
$W_o\ne0$ and $W_q\ne0$. For such frozen weights and generic prefixes the prefixed head is not affine, because the gate
$\lambda(x)$ is not constant. So prefix tuning and P-tuning v2 are not box-locally mergeable into a standard head. When
$W_oW_v\ne0$, no prefix is neutral, so prompt and prefix tuning are unpointed.
\end{restate}

\begin{proof}
Write $\epsilon_q(k)=e^{\langle q,k\rangle/\sqrt d}>0$, and for a multiset $K$ put $Z_q(K)=\sum_{(k,v)\in K}\epsilon_q(k)$ and
$S_q(K)=\sum_{(k,v)\in K}\epsilon_q(k)v$, so $\Sigma_q(K)=(Z_q(K),S_q(K))$.

(a) Sums over a multiset union split, $\Sigma_q(K\uplus K')=\Sigma_q(K)+\Sigma_q(K')$, and $\Sigma_q(\emptyset)=(0,0)$, so
$\Sigma_q$ is a monoid homomorphism. Each term of $Z_q$ is positive, so $Z_q(K)>0$ for nonempty $K$ and $Z_q(\emptyset)=0$. By
definition, $\mathrm{Attn}_q(K)=S_q(K)/Z_q(K)=\pi(\Sigma_q(K))$ for nonempty $K$; this $\pi$ is the perspective map of
\bgref{D.34}.

(b) With $\lambda=Z_P/(Z_P+Z_C)$ and $Z_P,Z_C>0$,
\[\pi(Z_P+Z_C,S_P+S_C)=\frac{S_P+S_C}{Z_P+Z_C}=\frac{Z_P}{Z_P+Z_C}\frac{S_P}{Z_P}+\frac{Z_C}{Z_P+Z_C}\frac{S_C}{Z_C}
=\lambda\,\mathrm{Attn}_q(P)+(1-\lambda)\,\mathrm{Attn}_q(C).\]

(c) In $P\uplus C$ a content pair $(k,v)\in C$ has weight $\epsilon_q(k)/(Z_P+Z_C)=(1-\lambda)\,\epsilon_q(k)/Z_C$, which is
$1-\lambda$ times its weight in $C$ alone; so all content weights scale by one factor and their ratios are unchanged. By (b)
the output lies on the segment from $\mathrm{Attn}_q(C)$ to $\mathrm{Attn}_q(P)$, with $\lambda\in(0,1)$, and
$\mathrm{Attn}_q(P)=\sum_{(k,v)\in P}(\epsilon_q(k)/Z_P)\,v$ is a convex combination of the prefix values. At the lowest
prefixed layer the query and the content pairs are computed from inputs that no prefix has touched, so they are those of the
pretrained model, and the content weights are the pretrained ones times $1-\lambda_q$. At a higher layer the inputs come from
lower layers whose outputs (b) has changed, so $q$ and $C$ themselves differ from the pretrained ones, and the statement
compares only with the same layer's computation on those inputs. The failure of network-level rigidity is shown in
\citet[\S6]{petrov2023prompting}, and universality of prefix-tuning in \citet{petrov2024prompting}.

(d) LLaMA-Adapter \citep{zhang2023llama} computes the two softmaxes separately, so its output is
$\pi(\Sigma_q(C))+\tanh(g_l)\,\pi(\Sigma_q(P))$, with two perspective maps. A content weight is $\epsilon_q(k)/Z_q(C)$, the same as
without the adapter. At $g=0$, $\tanh(0)=0$ (and HF's raw gate is $0$), so the output is $\mathrm{Attn}_q(C)$, the pretrained
one, and the method is pointed there. The remaining clauses describe how the method is built.

(e) Number the positions of the prompt-tuned model $0,\dots,m-1$ for the soft prompt $p=(p_0,\dots,p_{m-1})$ and $m,\dots,m+n-1$
for the content, and give the content the same positions $m,\dots,m+n-1$ in the prefix-tuned model. Let $K_j(p),V_j(p)$ be the
keys and values, after the positional encoding, that the prompt-tuned model forms at the prompt positions in layer $j$, as they
are cached. We show by induction over layers $j$ that (i) the hidden states of the prompt-tuned model at prompt positions
depend only on $p$, and (ii) the hidden states at content positions agree in the prompt-tuned model and in the prefix-tuned
model with prefix $h(p)=(K_j(p),V_j(p))_j$. At the input, (i) holds trivially and (ii) holds because both models embed the same
content tokens at the same positions. Suppose (i) and (ii) hold at the input of layer $j$. Under causal masking a prompt
position attends only to prompt positions, so its output depends only on prompt states, which gives (i) at the next layer. A
content position $t$ attends to the prompt positions and to the content positions $u\le t$. Its query, and the content keys
and values, are computed from equal content states with equal positional encodings, by (ii). Its prompt keys and values are
$K_j(p),V_j(p)$ in the prompt-tuned model by definition, and they are supplied verbatim by the prefix in the other. With
rotary encodings, a score is $\langle R_tq,R_sk_s\rangle$ in both models, the cached key $R_sk_s$ being the same object. So the
query and the multiset of key--value pairs are the same, and the attention outputs agree; the position-wise sublayers then give
(ii) at the next layer. Reading the content positions at the top gives $\Phi\circ\rho_{\rm prefix}\circ h=\Phi\circ
\rho_{\rm prompt}$. The prompt-tuned model and the prefix-tuned model are methods on different extensions of the base network
(input tokens in one, key--value slots in the other), so $h$ is a map over the common realisation as functions of the content,
an arrow of the slice over $C^\infty(X_{\rm content},Y_{\rm content})$.

Without the offset, the prefix-tuned model places the content at positions $0,\dots,n-1$. For RoPE put
$h'(p)=\big(R_{-m}K_j(p),V_j(p)\big)_j$, rotating the cached key from prompt position $s$ to $s-m$. A content query at position
$t-m$ then scores $\langle R_{t-m}q,R_{s-m}k_s\rangle=\langle q,R_{s-t}k_s\rangle=\langle R_tq,R_sk_s\rangle$, content--content
scores depend only on differences of positions and are unchanged, and values carry no rotation, so the induction goes through
with $h'$. For additive relative biases the score of a content query against a prefix key depends on the distance the
implementation assigns to that key, and the induction goes through exactly when these distances equal $t-s$. In a bidirectional
encoder of depth at least $2$, prompt positions attend to content positions, so already the second layer's prompt states depend on
the content, and step (i) fails.

(f) \emph{Affine without prefix.} On a context of one token $x$ the softmax has one term, equal to $1$, so a standard head
computes $W_o(W_vx+b_v)+b_o$, which is affine in $x$ for every weight setting. (With \texttt{add\_zero\_attn}, a sink or
softmax-plus-one there is a second term, and this fails.)

\emph{With a prefix.} Let the prefix be $P=\{(k_i,v_i)\}$ with $N_P\ge1$ fixed pairs, and write $q(x)=W_qx+b_q$,
$k(x)=W_kx+b_k$, $v(x)=W_vx+b_v$, the content score $\varsigma(x)=\langle q(x),k(x)\rangle/\sqrt d$ and
$Z_P(x)=\sum_i\epsilon_{q(x)}(k_i)$. By (b),
\begin{equation}\label{eq:C3-prefix-head}
\mathrm{head}_P(x)=W_o\big[(1-\lambda(x))\,v(x)+\lambda(x)\,\mathrm{Attn}_{q(x)}(P)\big]+b_o,\qquad
\lambda(x)=\frac{Z_P(x)}{Z_P(x)+e^{\varsigma(x)}}\in(0,1).
\end{equation}
\emph{Only if.} If $W_o=0$ the output is the constant $b_o$. If $W_q=0$ and $W_k\T b_q=0$, then $q=b_q$ is constant, every prefix
score $\langle b_q,k_i\rangle/\sqrt d$ is constant, and the content score $\langle b_q,W_kx+b_k\rangle/\sqrt d=\big((W_k\T b_q)\T
x+\langle b_q,b_k\rangle\big)/\sqrt d$ is constant; so $\lambda$ and $\mathrm{Attn}_q(P)$ are constant and the output is affine.

\emph{If.} Suppose $W_o\ne0$ and ($W_q\ne0$ or $W_k\T b_q\ne0$). Take all $N_P$ prefix pairs equal to $(k_1,v_1)$; then
$\mathrm{Attn}_{q(x)}(P)=v_1$ and $\lambda(x)=\operatorname{logit}^{-1}\big(\delta(x)\big)$ with
$\delta(x)=\langle q(x),k_1\rangle/\sqrt d+\log N_P-\varsigma(x)$. Restrict to a line $x=tu$. If $W_q\ne0$, choose $u$ with
$W_qu\ne0$; $\varsigma(tu)$ is a fixed polynomial of degree at most two in $t$, and $\langle q(tu),k_1\rangle$ has the slope
$\langle W_qu,k_1\rangle$, which $k_1$ can make any real number, so $k_1$ can be chosen with $\delta(tu)$ non-constant. If $W_q=0$ and
$W_k\T b_q\ne0$, choose $u$ with $\langle W_k\T b_q,u\rangle\ne0$; then $\varsigma(tu)$ has nonzero slope while the prefix score is
constant, so $\delta(tu)$ is non-constant for every $k_1$. In both cases $\lambda(t):=\lambda(tu)$ is real-analytic, takes values
in $(0,1)$ and is non-constant, so $\lambda''(t_0)\ne0$ for some $t_0$; otherwise $\lambda$ would be affine and bounded, hence
constant. Choose $v_*$ with $W_ov_*\ne0$ and put $v_1=Mv_*$. Since $v(tu)$ is affine in $t$,
\[\frac{d^2}{dt^2}\mathrm{head}_P(tu)=\lambda''(t)\big(MW_ov_*-W_ov(tu)\big)-2\lambda'(t)\,W_oW_vu ,\]
and at $t_0$ the term $M\lambda''(t_0)W_ov_*$ dominates for large $M$, so the second derivative is nonzero and the head is not
affine. For bias-free heads $W_k\T b_q=0$, which gives the second form of the criterion.

\emph{Generic prefixes.} The prefixed output is real-analytic in $(x,P)$ on the connected space of inputs and prefixes with $N_P$
pairs. It is affine in $x$ for a given $P$ iff all second partial derivatives in $x$ vanish at every $x$; this is the common zero
set, over $x$ and pairs of coordinates, of real-analytic functions of $P$. The witness above is a prefix at which one of these
functions is nonzero, so that function is not identically zero, and the prefixes with affine output lie in its zero set, which
is closed, nowhere dense and null (\auxref{B.10}). The same argument applied to $\lambda$ shows that the gate is non-constant
for generic prefixes. Box-locally, the class of a standard head is the set of functions it computes with some weights; on
length-one contexts these are all affine, while for generic prefixes the prefixed head is not, so prefix tuning and P-tuning v2
are not box-locally mergeable into the head.

\emph{No neutral prefix.} A neutral prefix would leave the head unchanged on every input, in particular on length-one contexts,
where by \eqref{eq:C3-prefix-head} this means $\lambda(x)\big(W_o\mathrm{Attn}_{q(x)}(P)-W_ov(x)\big)=0$, and since $\lambda(x)>0$,
$W_o\mathrm{Attn}_{q(x)}(P)=W_o(W_vx+b_v)$ for all $x$. The left side lies in the compact set $W_o\operatorname{conv}V_P$; if
$W_oW_v\ne0$, choose $x_0$ with $W_oW_vx_0\ne0$, and along $x=tx_0$ the right side is unbounded. So no prefix is neutral. For a
base network consisting of one such head, the extension by prefixes therefore has no neutral point, and by (e), at the first layer
of a causal decoder a soft prompt acts through its key--value pairs exactly as a one-layer prefix, so the same base network admits
no neutral prompt: prompt and prefix tuning are unpointed.
\end{proof}

\begin{remark*}{numerical checks}
The values reported after \thmref{Theorem}{VI.5} (prompt-to-prefix agreement $9\times10^{-16}$ with RoPE and the offset, $1.3$
without it and $2\times10^{-15}$ with keys rotated by $-m$; an affine head with $W_o=0$ although $W_q\T W_k\ne0$, and a non-affine
one with $W_q\ne0$ and $W_q\T W_k=0$) instantiate parts (e) and (f) as proved above: the criterion of (f) involves $W_o$, $W_q$ and
$W_k\T b_q$, not $W_q\T W_k$ (Appendix~\ref{app:repro}).
\end{remark*}

\section{The catalogue}\label{app:catalogue}

The atlas of Exposé~\ref{sec:atlas} rests on a catalogue of \nMethods{} methods and \nPapers{} papers. This appendix
says how the catalogue was assembled and fact-checked, how each method received its seven coordinates, its arrows and
its obstructions, and how these were checked in turn. It ends with the complete tables: every method with its
coordinates and its reference (Table~\ref{tab:catalogue}), every recorded arrow with its map $h$
(Table~\ref{tab:arrows}), and the arrows and obstructions counted by rule and by test
(Table~\ref{tab:app-rules-tests}).

\paragraph{Data files} The data accompany this paper in machine-readable form, generated from the same source as the tables
below and as the companion site.
\begin{description}[leftmargin=1.2em,itemsep=2pt]
  \item[\texttt{catalogue.json}] Every method (\nMethods{}) and every paper (\nPapers{}), each with its verified
  metadata (title, full author list, year of the first arXiv version, venue, links to the paper and the code) and its
  BibTeX record. A method entry adds the formula, the trainable-parameter count, the locus, the modification kind, the
  merge verdict with its reason, the initialisation, the relations to other methods, the sources consulted, the seven
  coordinates with a justification for each, the categorical reading, and the method's arrows and obstructions. The
  file also holds the definitions of the axis values, with the question that decides each axis, and the arrow rules.
  \item[\texttt{catalogue.csv}] One row per method, with the main fields of \texttt{catalogue.json} and one column per
  coordinate.
  \item[\texttt{arrows.csv}] Every recorded arrow $M\to N$ with its rule, its map $h$ and a note that names the instance
  of $N$ it lands in (rank, pointing or regime).
  \item[\texttt{obstructions.csv}] Every recorded obstruction $M\not\to N$ with the test that rules the arrow out and
  the invariant that fails.
  \item[\texttt{propositions.json}] Every numbered result (\nPropositions{} of them) with its statement, proof, review
  status, second-round verdict, public statement and referee notes.
  \item[\texttt{framework\_checks.py}] The numerical checks of Appendix~\ref{app:repro}.
\end{description}

\subsection{Families and sources}\label{app:catalogue-families}

Every entry was written once and then attacked once, by a separate reader looking for what the first got wrong. The
literature was split into the thirteen families of Table~\ref{tab:families}, then completed by four gap-filling
batches, which added the methods a completeness review found missing, and by two addenda on 2 October 2026: methods the authors reported missing, and a sweep of the 2026 literature run in two passes
(\S\ref{app:catalogue-2026}). The family boundaries owe much to earlier maps of the field: the unified view of
\citet{he2021unified}, the delta-tuning study of \citet{ding2022parameter}, the guide of \citet{lialin2023scaling}, the
survey of \citet{han2024parameter} and the modular view of \citet{pfeiffer2023modular}.

\begin{table}[t]
\centering\footnotesize
\caption{How the catalogue was assembled. Each family was written by one reader and attacked by another; the last column
(Corr.) counts the logged corrections. The families overlap, so the records were merged into the catalogue on a canonical
identifier.}\label{tab:families}
\begin{tabularx}{\linewidth}{@{}>{\raggedright\arraybackslash}p{4.1cm}>{\raggedright\arraybackslash}Xrrr@{}}
\toprule\headrow
\textbf{Family} & \textbf{What it covers} & \textbf{Methods} & \textbf{Papers} & \textbf{Corr.}\\\midrule
LoRA core & LoRA and its variants & 20 & 12 & 21\\
Initialisation and rank & Base-point splits; rank allocation & 25 & 6 & 46\\
Shape & Kronecker, Hadamard, tensor and sparse shapes & 29 & 9 & 49\\
Multiplicative & Group and monoid actions on $W_0$ & 32 & 9 & 54\\
Sharing & Shared or random bases & 28 & 4 & 46\\
Selective & Subsets of existing weights & 26 & 7 & 37\\
Adapters & Inserted modules and side networks & 31 & 10 & 30\\
Prompts & Soft prompts, prefixes and learned states & 27 & 13 & 39\\
Quantisation and memory & Quantised bases, gradient memory & 24 & 5 & 44\\
Composition & Merging, routing, mixtures & 42 & 8 & 61\\
HF PEFT & The library's tuners and utilities & 74 & 10 & 53\\
Frontier & 2025--2026: RL, reasoning, diffusion & 21 & 5 & 48\\
Theory & Theory and foundations & 0 & 50 & 23\\
Gap-filling (4 batches) & Methods a completeness review found missing & 36 & 18 & 62\\
Addendum (2 Oct 2026) & Methods the authors reported missing & 1 & 0 & 2\\
Addendum (2026 sweep, pass 1) & arXiv, January to September 2026; ICLR and ICML 2026 & 29 & 8 & 126\\
Addendum (2026 sweep, pass 2) & Further candidates, mostly ICLR, ICML and NeurIPS 2026 & 32 & 6 & 164\\\midrule
\textbf{All records} & Before merging & \textbf{477} & \textbf{180} & \textbf{905}\\
\textbf{Catalogue} & After merging duplicates & \textbf{\nMethods} & \textbf{\nPapers} & \\
\bottomrule
\end{tabularx}
\end{table}

\paragraph{Merging}
The families overlap, so the $477$ method records and $180$ paper records were merged into \nMethods{} methods and
\nPapers{} papers. Records were matched on a canonical identifier and a table of aliases, because matching by name would
fold LoRA+ \citep{hayou2024lora} into LoRA \citep{hu2021lora} and O-LoRA \citep{wang2023orthogonal} into OLoRA
\citep{buyukakyuz2024olora}. Of the \nMethods{} entries, $358$ are methods proper; the others are $17$ training recipes,
$4$ software systems, $2$ analyses and $1$ baseline (linear probing).
The first arXiv versions of the methods run from 2013 to 2026, and $47$ methods were first posted in 2026.

\paragraph{The bibliography}
The BibTeX record of each entry comes from the arXiv metadata of its identifier, with the full author list, for $493$
identifiers, and was written by hand for the nine entries without one. These $502$ records
form the catalogue part of this paper's bibliography, and the citation keys of Table~\ref{tab:catalogue} point into it.

\subsection{Fact-checking}\label{app:catalogue-verification}

Each field was checked against a primary source.
\begin{itemize}
  \item Titles, authors and first-version years were checked against the arXiv metadata.
  \item Venues were checked against the official records: OpenReview, the ACL Anthology, PMLR and CVF open access, with
  Crossref, Semantic Scholar and DBLP for the remaining cases.
  \item Formulas and parameter counts were checked against the papers, and against the official code where an
  initialisation was in doubt.
  \item Every claim about Hugging Face PEFT \citep{xmangrulkar2022peft} was checked against its source at commit
  \texttt{85be3f3} (\texttt{85be3f3a1a58c4d153ffc4631c57196430af87a4}) of 30 September 2026. The library's code
  settled many questions of convention, such as the side of $W$ on which a rotation acts, the default initialisation of
  each factor, and whether a scale such as $\alpha/r$ is applied. The column ``HF'' of Table~\ref{tab:catalogue} marks
  the \nHF{} methods available at that commit.
\end{itemize}
Every change was logged with its reason, $905$ in all (Table~\ref{tab:families}). The merge verdict of each method is
recorded beside its reason: $250$ methods fold back into the frozen weights always, $41$ under stated conditions and $91$
not at all. This verdict concerns the deployed method; the merge coordinate of Exposé~\ref{sec:atlas} refines it by the
rewrite rules of \thmref{Theorem}{VI.3}.

\subsection{The 2026 sweep and the authors' addendum}\label{app:catalogue-2026}

On 2 October 2026 the authors pointed out methods missing from the catalogue. LOFT \citep{zhao2026loft} was added at once as an addendum of its own,
checked against arXiv and the paper (two corrections logged), and classified like every other method.
Its coordinates are R, Orb, regular, none, $\CO\times\CO$, M1 and idempotent; it has arrows into OFT, LoRA-XS, PSOFT
and LoRA, and for $r\ge4$ its first-order image rules out a smooth pointed arrow into HRA at HF's paired
initialisation (test O3). Two
independent referees checked these coordinates, arrows and the obstruction, numerically and against the paper and the
code. \S\ref{sec:erlangen-loft} reads LOFT as a fixed leaf of HRA's meet.

The other gaps were sought by a sweep of the 2026 literature, run in two passes on the same day. A pool of $265$ candidates was
drawn from six slices:
\begin{itemize}
  \item three quarterly arXiv slices, for first versions posted from January to March, April to June and July to
  September 2026, with $60$ candidates each;
  \item the accepted papers of ICLR 2026, $5{,}468$ of them screened through their abstracts, with $48$ candidates;
  \item the accepted papers of ICML 2026, $6{,}797$ screened, with a second pass over ACL, EACL and CVPR 2026, together
  giving $31$ candidates;
  \item the 2026 papers citing anchor methods (LoRA, DoRA, PiSSA, QLoRA, OFT, VeRA, HRA, GaLore, PSOFT and LoRA-GA),
  ranked by citations, recency and acceptance, with the 2026 release notes and pull requests of Hugging Face PEFT,
  giving $6$ candidates.
\end{itemize}
Acceptance at ICLR and ICML 2026 was confirmed by an exact title match against the official lists of accepted papers
or by the authors' own arXiv comment. Candidates were matched against the catalogue by arXiv identifier and by name, so that near-homonyms stay
apart: the entropy-guided FlexLoRA of 2026 \citep{liu2026flexlora} is not the federated FlexLoRA of 2024
(arXiv:2402.11505), and the per-token MoLoRA of 2026 (arXiv:2603.15965) is not the MoLoRA of \citet{zadouri2023pushing}.
The first pass admitted $29$ methods and $8$ papers. The second re-screened the rest of the pool and admitted $32$ methods
and $6$ papers, mostly from ICLR, ICML and NeurIPS 2026. Each admitted entry went through the same fact-checking as the
families, with $126$ and $164$ logged corrections, and each of the $61$ new methods received an independent check of its
coordinates and arrows on the same day. The arXiv queries searched abstracts for LoRA, PEFT, adapter,
parameter-efficient and prompt tuning, so a 2026 method paper whose abstract uses none of these terms lies outside the
sweep's reach.

\subsection{Coordinates, arrows and obstructions}\label{app:catalogue-classification}

\paragraph{Assignment}
Each method was classified from its formula, in thirteen batches: ten of $32$ methods for the catalogue as it stood
on 1 October 2026, one for the authors' addendum, and one for each pass of the 2026 sweep ($29$ and $32$ methods). A first reader answered the
seven questions of \S\ref{sec:atlas-axes}, which fix the kind, the shape of the image germ, the base point, the gauge,
the covariance group, the merge type and the rebasing closure, following the decision procedures of
\S\ref{sec:atlas-decide} and the conventions of \S\ref{sec:atlas-conventions}. Each answer carries a short
justification, kept in \texttt{catalogue.json}. A null coordinate, printed ``--'', means that the axis does not apply,
for instance the shape of a method with no germ at $\theta_0$. Where an answer turns on the dimension of an image or a
fibre, or on the first-order image at the initialisation, it was confirmed by a float64 Jacobian rank at a random point
and at the base point (\bgref{C.3}).

Each arrow $M\to N$ comes with an explicit pointed smooth map $h$ satisfying $\rho_N\circ h=\rho_M$, obtained by one of
the rules A2--A5 of \S\ref{sec:atlas-rules}, and with the instance of $N$ it lands in: a rank, a pointing or a
hyperparameter regime. Each obstruction $M\not\to N$ names the test O1--O3 of \S\ref{sec:atlas-tests} that rules the
arrow out and the invariant that fails. The interval arrows $\text{Frozen}\to M\to\text{Full FT}$ of rule A1 hold for
every weight-space method and are not recorded. The \nCore{} core methods are those named in the periodic table
(Table~\ref{tab:periodic}) together with a curated set of others; the flag is not a coordinate.

\paragraph{Independent check}
A second reader re-derived every coordinate from the formula and re-checked every arrow, by recomputing
$\rho_N\circ h$ and the pointing $h(q_0^M)$, and every obstruction, by recomputing the failing invariant. Each method
carries a check note with the outcome. More than half of the notes record no change; the others record each change with
its reason, and $47$ record a numerical Jacobian computation (a rank, a fibre dimension or a deficiency).

A consolidation pass on 1 October 2026, over the $320$ methods classified by then, unified the data and touched $135$
methods, each change logged in its check note.
\begin{enumerate}[label=(\arabic*)]
  \item $137$ coordinates used a value ``n/a'' that the axis definitions do not contain. $136$ became null, matching
  the ``--'' of the tables, and the base point of GenKnowSub became ``unpointed'', as for the other parameterless
  post-hoc operations. The merge axis keeps its legacy key \texttt{na}, read as null.
  \item The row labels of the periodic table were unified across batches.
  \item The covariance of LISA and BAdam changed from $\Pi$ to $\GL\times\GL$, matching HiFT, surgical fine-tuning and
  BitFit: each stage frees whole boxes, and the layer partition breaks no neuron permutation.
  \item Four obstruction reasons now say which instance of the target they concern (DSEE, GaLore, Spectral Adapter$^A$
  and MoV, each against LoRA).
\end{enumerate}
Where a formula disagreed with an earlier draft of the periodic table, the value the formula supports was kept and the
table of Exposé~\ref{sec:atlas} was revised to agree, in thirteen places. For example, LST, UniPELT
and MAM are unpointed extensions, MPT and Compacter have torus gauges, the Houlsby adapter's GELU or swish leaves only the
permutations of hidden units, and the gauges of DeLoRA and AdaLoRA are non-linear. The $62$ later entries, LOFT and the
$61$ methods of the 2026 sweep, were classified and checked in the same way.

\paragraph{Structural checks on the final data}
For this appendix the structural checks were re-run on the ancillary files. Every arrow and every obstruction joins two
catalogued methods and names a valid rule or test; every arrow has a map $h$ and every obstruction a reason; every
coordinate is a value defined for its axis; and no arrow or obstruction is recorded twice. A check at the level of method
identifiers flags four kinds of pattern, and each was resolved by the instance it concerns.
\begin{itemize}
  \item \emph{Pairs with both an arrow and an obstruction} ($26$). In every case the two concern different instances of
  the target: a different rank, pointing, regime or variant (\S\ref{sec:atlas-checks}). For example, ABBA is isomorphic
  to LoHa pointed at the factors of $[W_0]_{r_1}$, while the O3 obstruction concerns LoHa at HF's random default
  pointing.
  \item \emph{Arrows from a method to itself} ($8$). Each is an inclusion between instances of one family, such as
  LoRA$_r\hookrightarrow{}$LoRA$_{r+1}$ or ReLoRA$^K\hookrightarrow{}$ReLoRA$^{K+1}$.
  \item \emph{Cycles} ($13$ groups of mutually reachable methods). The cycles in these groups are of four kinds.
  Eight are isomorphisms: DoRA and BiDoRA, ReLoRA and COLA, AsyLoRA and LoRA-FA, AdaptFormer and the parallel adapter,
  S-LoRA and Punica, EFT and proxy tuning, LoRAMoE and MoLoRA, and AdaMerging and LoRA soups (CAT), which differ only in
  pointing. Three join methods with equal images that are not isomorphic: prompt tuning with MPT and with P-tuning, and
  QLoRA with IR-QLoRA. Five go up in rank or in the number of stages and are not isomorphisms: LoRA with ReLoRA, LoRA
  with PEANuT, LoRA-FA with Flora, surgical fine-tuning with HiFT, and KronA with LoKr. Initial-state tuning and prefix
  tuning simulate each other on S4 state-space layers only.
  \item \emph{Obstructions $M\not\to N$ with no direct arrow but a path of arrows from $M$ to $N$} ($20$). In each,
  the obstruction concerns a fixed rank or
  pointing of the target and the path ends in another instance. FourierFT with $n_c$ frequencies, for example, maps
  into LoRA-FA$_{2n_c}$ and so into LoRA$_{2n_c}$, while the O2 obstruction concerns LoRA$_r$ at a matched budget.
\end{itemize}
No arrow contradicts an obstruction once the stated rank or pointing is taken into account. The eight places where the
data depart from the conventions of \S\ref{sec:atlas-conventions} are listed in \S\ref{sec:atlas-checks}, and
Table~\ref{tab:atlas-counts} counts the methods per coordinate value.

\begin{table}[t]
\centering\footnotesize
\caption{The \nArrows{} recorded arrows by rule (top) and the \nObstructions{} recorded obstructions by test (bottom).
The rules and tests are those of \S\ref{sec:atlas-arrows}; $161$ of the arrows join two core methods. Every arrow is
listed in Table~\ref{tab:arrows}, and every obstruction, with its failing invariant, in the ancillary file
\texttt{obstructions.csv}.}\label{tab:app-rules-tests}
\begin{tabularx}{\linewidth}{@{}lrX@{}}\toprule\headrow
\textbf{Rule} & \textbf{Count} & \textbf{How $h$ arises}\\\midrule
A2 & 61 & freezing a block of coordinates at its base value, $h(q')=(q',q_0'')$\\
A3 & 329 & an explicit smooth pointed factorisation $\rho_M=\rho_N\circ h$\\
A4 & 2 & a projection out of a fibre product\\
A5 & 11 & a sum or an added stage, $h(q)=(q,q_0^N)$\\
\bottomrule\end{tabularx}

\medskip
\begin{tabular}{@{}lrl@{}}\toprule\headrow\textbf{Test} & \textbf{Count} & \textbf{What fails}\\\midrule
O1 & 288 & an image invariant (Gram matrix, spectrum, row lines, zero pattern) or image inclusion\\
O2 & 129 & dimension or cut-rank\\
O3 & 68 & first-order image or base-point stratum\\
\bottomrule\end{tabular}

\end{table}

\subsection{The complete tables}\label{app:catalogue-tables}

Table~\ref{tab:catalogue} lists every method with its reference, the year of its first arXiv version and its seven
coordinates. A method without an arXiv version carries the year it first became public. The year in a citation is that
of the version cited, so the two can differ: LoRA-S, for instance, first appeared in 2025 and is cited as
\citet{zheng2025lora}, its ICLR 2026 version. The rows are grouped by kind, with the reparametrisations split as in the periodic table (additive, action,
Hadamard and selective), and sorted by name within each group. The abbreviations are those of
Table~\ref{tab:atlas-counts}, with ``unpt.''\ for unpointed, ``nonlin.''\ for a non-linear gauge, GL$\times$T for
$\GL$ times a Hadamard torus, O$\times$O for the spectral-frame value $\CO\times\CO$, ``idemp.''\ for idempotent and
``sched.''\ for a scheme that is itself a rebasing schedule. The merge value \texttt{na} is the legacy key for
backward methods, read as ``--''.

Table~\ref{tab:arrows} lists every recorded arrow, sorted by source, with its rule and its map $h$. Each $h$ is stated
in the notation of the target's formula in \texttt{catalogue.json}, and the instance of the target (its rank, its
pointing or its regime) is in the note column of \texttt{arrows.csv}.

\begingroup\scriptsize\setlength{\tabcolsep}{2.1pt}\renewcommand{\arraystretch}{1.08}
\begin{longtable}{@{}>{\raggedright\arraybackslash}p{2.53cm}>{\raggedright\arraybackslash}p{3.2cm}r l l l l l l l l@{}}
\caption{The complete catalogue: every method with its seven coordinates (\S\ref{sec:atlas}). Core methods are marked $\bullet$; HF marks methods available in Hugging Face PEFT at the pinned commit. ``--'' means the axis does not apply.}\label{tab:catalogue}\\
\toprule\headrow\textbf{Method} & \textbf{Reference} & \textbf{Year} & \textbf{Kind} & \textbf{Shape} & \textbf{Base pt.} & \textbf{Gauge} & \textbf{Covar.} & \textbf{Merge} & \textbf{Rebase} & \textbf{HF}\\\midrule
\endfirsthead
\multicolumn{11}{@{}l}{\footnotesize\itshape Table~\ref{tab:catalogue}, continued}\\
\toprule\headrow\textbf{Method} & \textbf{Reference} & \textbf{Year} & \textbf{Kind} & \textbf{Shape} & \textbf{Base pt.} & \textbf{Gauge} & \textbf{Covar.} & \textbf{Merge} & \textbf{Rebase} & \textbf{HF}\\\midrule
\endhead
\midrule
\multicolumn{11}{@{}r}{\footnotesize\itshape continued on the next page}\\
\endfoot
\bottomrule\endlastfoot
\multicolumn{11}{@{}l}{\sffamily\bfseries\color{tide}R, additive}\\[1pt]
ABBA & \citealp{singhal2025abba} & 2025 & R & Cone & apex & GL$\times$T & Mon$\times$Mon & M1 & span & \\
$\bullet$\,AdaLoRA & \citealp{zhang2023adalora} & 2023 & R & Cone & apex & nonlin. & GL$\times$GL & M1 & span & \checkmark\\
$\bullet$\,AdaMSS & \citealp{zheng2025adamss} & 2025 & R & Cone & apex & GL & O$\times$O & M1 & span & \checkmark\\
AdaPaD & \citealp{su2026adapad} & 2026 & R & Cone & defect & GL & GL$\times$GL & M1 & span & \\
ALoRA & \citealp{liu2024alora} & 2024 & R & Cone & apex & GL & GL$\times$GL & M1 & span & \\
AltLoRA & \citealp{yu2025altlora} & 2025 & R & Cone & apex & GL & GL$\times$GL & M1 & span & \\
ApiQ & \citealp{liao2024apiq} & 2024 & R & Cone$^{*}$ & defect & GL & Mon$\times$Mon & Mq & span & \\
\end{longtable}\endgroup

\begingroup\scriptsize\setlength{\tabcolsep}{3pt}\renewcommand{\arraystretch}{1.12}
\begin{longtable}{@{}>{\raggedright\arraybackslash}p{2.4cm}>{\raggedright\arraybackslash}p{2.4cm}c>{\raggedright\arraybackslash}p{8.6cm}@{}}
\caption{Every recorded simulation arrow $M\to N$ (rules A2--A5 of \S\ref{sec:atlas}), with its explicit map. Notes on the instances (rank, pointing) are in the ancillary file \texttt{arrows.csv}.}\label{tab:arrows}\\
\toprule\headrow\textbf{Source $M$} & \textbf{Target $N$} & \textbf{Rule} & \textbf{Pointed map $h$ with $\rho_N\circ h=\rho_M$}\\\midrule
\endfirsthead
\multicolumn{4}{@{}l}{\footnotesize\itshape Table~\ref{tab:arrows}, continued}\\
\toprule\headrow\textbf{Source $M$} & \textbf{Target $N$} & \textbf{Rule} & \textbf{Pointed map $h$ with $\rho_N\circ h=\rho_M$}\\\midrule
\endhead
\midrule
\multicolumn{4}{@{}r}{\footnotesize\itshape continued on the next page}\\
\endfoot
\bottomrule\endlastfoot
(IA)\textsuperscript{3} & DoRA & A2 & $\ell\mapsto\big(B=0,\ A=A_0,\ \mu=\ell\odot(\|w_{0,i}\|)_i\big)$\\
(IA)\textsuperscript{3} & HiRA & A3 & $\ell\mapsto(\ell-\mathbf 1,\ \mathbf 1^\top)\ \ (k,v);\qquad \ell\mapsto(\mathbf 1,\ (\ell-\mathbf 1)^\top)\ \ (W_{\rm down})$\\
ABBA & LoHa & A3 & $h(B_1,A_1,B_2,A_2)=(B_1,A_1,cB_2,A_2),\ \ c=\frac{\alpha/\sqrt{r_1r_2}}{\gamma_{\rm LoHa}}$\\
ABBA & LoRA & A3 & $h=\big(\tfrac{\alpha}{\sqrt{r_1r_2}}\,B_1\bullet B_2,\;(A_1^\top\bullet A_2^\top)^\top\big)$\\
AdaLomo & LOMO & A3 & $h=\mathrm{id}_\Theta$\\
AdaLoRA & LoRA & A3 & $h(P,\Lambda,Q)=(P\Lambda,\ Q)$\\
AdaLoRA & PoLAR & A3 & $h(P,\Lambda,Q)=\big(P(P^\top P)^{-1/2},\ (P^\top P)^{1/2}\Lambda(QQ^\top)^{1/2},\ Q^\top(QQ^\top)^{-1/2}\big)$\\
Adam-mini & LOMO & A3 & $h=\mathrm{id}_\Theta$\\
\end{longtable}\endgroup

\section{Numerical checks and reproducibility}\label{app:repro}

Claims marked [Num] in the main text are numerical checks and are not counted as proved. Most of them are re-run by one
script, \texttt{framework\_checks.py}, one of the data files that accompany this paper. This appendix says what the script
computes and how to read its output (\S\ref{app:repro-script}), lists every check with the result it supports and the
value it prints (\S\ref{app:repro-checks}, Table~\ref{tab:repro-checks}), names the few numbers of the main text that
come from elsewhere (\S\ref{app:repro-other}), records the software (\S\ref{app:repro-software}), describes how
the companion site and the figures of this paper compute what they show (\S\ref{app:repro-site}), and gives the
settings and values behind each figure (\S\ref{app:repro-figures}).

\subsection{The script}\label{app:repro-script}

The script runs with \texttt{python3 framework\_checks.py} and needs PyTorch \citep{xpaszke2019pytorch}, NumPy
\citep{xharris2020array} and SciPy \citep{xvirtanen2020scipy}. It works in float64 throughout
(\texttt{torch.set\_default\_dtype(torch.float64)}) and on the CPU. It seeds PyTorch's generator at the start and again
before several checks, so that a run prints the same numbers every time on a given machine. On a laptop (Apple M5) it
finishes in about $50$ seconds.

It computes three kinds of quantity.
\begin{itemize}
  \item \emph{Dimensions.} The expressive dimension $d(M)$ of a method is the generic rank of its Jacobian
  (\bgref{C.3}). The script forms the Jacobian of $\rho$ by automatic differentiation
  (\texttt{torch.autograd.functional.jacobian}) at a random point, or at the base point when it measures a first-order
  image $S_1$, and counts the singular values above $10^{-9}$ times the largest. The points of lower rank have measure
  zero (\bgref{C.21}, \bgref{C.22}), so this computes the generic rank with probability one, up to the tolerance.
  \item \emph{Optimizer runs.} Adam, decoupled weight decay and gradient descent are written out step by step, with the
  bias correction and the convention $0/0=0$ at $\varepsilon=0$, so that nothing is hidden in a library default. Gradient
  flows are integrated by explicit Euler steps or by the classical fourth-order Runge--Kutta method.
  \item \emph{Identities and defects.} An identity or an equivariance is tested by computing both sides and printing the
  Frobenius norm of their difference.
\end{itemize}
Each line prints a value beside its prediction. Some lines are controls, cases in which the theory predicts that an
identity fails; they show that the test would detect a failure.

\paragraph{How sharp the rank decisions are}
The script takes $45$ Jacobian ranks. In every one of them the smallest singular value it keeps is at least
$3.9\times10^{-3}$ times the largest, and the largest singular value it discards is at most $4.3\times10^{-16}$ times the
largest, which is rounding error. The threshold $10^{-9}$ lies more than six orders of magnitude from both, so no rank in
Table~\ref{tab:repro-checks} depends on its choice. We measured these margins with a copy of the script whose rank
function also records the two ratios; the copy prints the same output.

\paragraph{What changes between machines}
The ranks and dimensions are integer decisions taken with the margins above, and the exact zero of check~3 comes from
rescaling by a power of two, which is exact in IEEE arithmetic. These outputs are expected to be the same on any machine.
Values of order $10^{-15}$ are rounding errors, and their leading digits change with the processor and the
linear-algebra library. Values of order one depend on the random draws, which follow from the seeds.

\subsection{The checks, section by section}\label{app:repro-checks}

The script has thirteen sections. Sections 1 to 11 are the numerical claims of the framework, section 12 re-runs the
main numbers of the first referee round, and section 13 those of the second. Table~\ref{tab:repro-checks} lists every
check in the script's order, with the result it supports, what it computes and what it prints in the run of
2 October 2026. Every printed value agrees with its prediction, and every control fails, by amounts between
$1.6\times10^{-2}$ and $52$. Matrix shapes are written $(m,n,r)$ for an $m\times n$ weight and rank $r$; a defect is the
Frobenius norm of the difference between the two sides of the identity tested.

\begingroup\footnotesize\setlength{\tabcolsep}{3pt}\renewcommand{\arraystretch}{1.12}
\begin{longtable}{@{}>{\raggedright\arraybackslash}p{0.9cm}>{\raggedright\arraybackslash}p{1.85cm}%
>{\raggedright\arraybackslash}p{6.2cm}>{\raggedright\arraybackslash}p{5.75cm}@{}}
\caption{The checks of \texttt{framework\_checks.py}, in the script's order, with the result each one supports, what it
computes, and what it prints. Controls are cases in which the theory predicts a failure.}\label{tab:repro-checks}\\
\toprule\headrow
\textbf{Check} & \textbf{Supports} & \textbf{What the script computes} & \textbf{What it prints}\\\midrule
\endfirsthead
\toprule\headrow
\textbf{Check} & \textbf{Supports} & \textbf{What the script computes} & \textbf{What it prints}\\\midrule
\endhead
\bottomrule\endlastfoot
\multicolumn{4}{@{}l}{\sffamily\bfseries\color{tide}Sections 1--11: the numerical claims of the framework}\\[1pt]
1 & \thmref{Thm}{II.7}(b) &
HRA$_r$ without Gram--Schmidt, $u\mapsto W_0H(u_1)\cdots H(u_r)$, with an injective $W_0\in\R^{(n+2)\times n}$. Jacobian
rank at a random point, and at HF's paired initialisation ($u_{2i-1}=u_{2i}$, so $k=r/2$ distinct vectors). &
At $(n,r)=(9,2)$, $(9,4)$, $(10,2)$, $(10,4)$, $(12,6)$: generic $15$, $26$, $17$, $30$, $51$, equal to $rn-r(r+1)/2$;
paired $8$, $15$, $9$, $17$, $30$, equal to $kn-k(k+1)/2$.\\
2 & \thmref{Thm}{III.12}(b) &
The first Adam step from zero state with $\varepsilon=0$ is sign descent. Its first-order weight update $dB\,A+B\,dA$ at
$(B,A)$ and at $(Bg,g^{-1}A)$, $(m,n,r)=(6,5,3)$. &
Defect $1.4\times10^{-15}$ for a signed permutation $g$. Controls: $11.3$ for $g=\diag(2,0.5,3)$ and $6.6$ for a random
rotation. With per-channel rates $(\eta_Bd_j,\eta_A/d_j)$ the diagonal $g=\diag(d)$ gives $5.0\times10^{-16}$.\\
3 & \thmref{Cor}{III.11}, \thmref{Thm}{III.10} &
$200$ steps of bias-corrected Adam with $\varepsilon=0$ on a least-squares toy ($W_0\in\R^{7\times6}$, $r=2$, $B_0=0$):
LoRA+ with ratio $\lambda=8$ and scale $s$, against LoRA with scale $\lambda s$ and equal rates. &
Largest weight difference over the whole trajectory $0.0$, exactly, since $\lambda$ is a power of two. Control without
the rescaling: $0.93$.\\
4 & \thmref{Prop}{II.12} &
GraLoRA with $k\times k$ blocks of rank $r/k$: Jacobian rank, and the rank of $\Delta W$ at a random point. &
At $(m,n,r,k)=(8,8,4,2)$, $(12,8,4,2)$, $(12,12,6,3)$: dimension $48$, $64$, $108$, equal to $r(m+n-r)$; rank of $\Delta
W$ $8$, $8$, $12$, against LoRA's cap $4$, $4$, $6$.\\
5 & \thmref{Prop}{II.12} &
LoHa$_{r,r}$, $(B_1A_1)\odot(B_2A_2)$: Jacobian rank. &
At $(m,n,r)=(8,7,2)$, $(12,10,2)$, $(12,10,3)$: $38$, $59$, $93$, equal to the bound $2r(m+n-r)-(m+n-1)$.\\
6 & \thmref{Cor}{III.6}(b) &
Euler integration ($h=10^{-4}$, $T=2$) of LoRA's gradient flow with decay $\lambda=0.5$ at $(m,n,r)=(5,4,2)$; the final
charge $\Phi$ against its closed form. &
Coupled decay at $\eta_B/\eta_A=4$ against $e^{-2\eta_A\lambda t}\Phi(0)$: relative error $10.7$ (control). Coupled decay
at $\eta_A=\eta_B$: $2.5\times10^{-4}$. Per-time decay at $\eta_B/\eta_A=4$ against $e^{-2\lambda t}\Phi(0)$:
$2.3\times10^{-4}$. Both are integrator errors.\\
7 & \thmref{Thm}{III.3}, \thmref{Prop}{III.4} &
At a random point, $(m,n,r)=(7,6,2)$, under a generic $g\in\GL_2$: the plain-GD cometric $GA\T A+BB\T G$ and the
undamped scaled-GD cometric; and LoRA-Pro's update $\Pi=P_UG+GP_V-P_UGP_V$ paired with a random tangent vector
$T=XA+BY$. &
Gauge defect $52.3$ for plain GD (control) and $1.1\times10^{-14}$ for scaled GD, which equals $P_UG+GP_V$ to
$8.7\times10^{-16}$. $\langle G-\Pi,T\rangle=0.0$, so $\Pi$ is the orthogonal projection onto the tangent space.\\
8 & \thmref{Thm}{V.3} &
The span of $200$ random VeRA updates $\diag(b)\,B\,\diag(d)\,A$ with frozen $B$ and $A$, $(m,n,r)=(6,5,2)$. &
Dimension $12=mr$: the span is $\{XA\}$, the reach of re-merged VeRA.\\
9 & \thmref{Prop}{II.9} &
RoAd$_1$ at $m=8$, one angle and one scale per output pair acting on $W_0$ by $2\times2$ blocks: Jacobian rank. &
$8=m$, the number of parameters: an injective map with an $m$-dimensional image.\\
10 & \thmref{Prop}{II.3} &
The rank of $(dB,dA)\mapsto dB\,A+B\,dA$ with $\rk A=r_A$ and $\rk B=r_B$, at $(m,n,r)=(7,6,3)$. &
$(r_A,r_B)=(3,0)$, $(3,3)$, $(2,1)$: $21$, $30$, $18$, equal to $mr_A+nr_B-r_Ar_B$.\\
11 & \thmref{Thm}{V.3} &
The product of two rotations, each a product of two reflections in $\R^8$, as in two cycles of HRA$_2$. &
$\rk(R_1R_2-I)=4\le2+2$: subadditivity, attained.\\[2pt]
\multicolumn{4}{@{}l}{\sffamily\bfseries\color{tide}Section 12: the first referee round}\\[1pt]
12a & \thmref{Prop}{II.8} &
DoRA as in HF PEFT, $\diag(\mu_i/\lVert V_i\rVert)\,V$ with $V=W_0+BA$: Jacobian rank. &
$9=mn$ at $(3,3,2)$, where $r(m+n-r)+m=11$; $36$ at $(8,8,2)$, matching the formula; $24$ for a rank-one $W_0$ at
$(6,6,2)$, equal to $r(m+n-r)+m-r$.\\
12b & \thmref{Prop}{II.12} &
The rank of GraLoRA's $\Delta W$ at a random point. &
$8=\min(kr,m,n)$ at $(16,16,4,2)$; $12=\min(m,n)$ at $(12,12,6,3)$.\\
12c & \thmref{Prop}{II.12} &
LoHa at $(4,4,2)$: Jacobian rank. &
$16=mn$, where the bound gives $17$.\\
12d & \thmref{Prop}{VI.4} &
LoReFT on $d=6$ with $r=2$, $(R,W,b)\mapsto\big(I+R\T(W-R),\,R\T b\big)$ with the rows of $R$ orthonormalised: Jacobian
rank. &
$22=r(2d+1-r)$, against $26$ for LoRA$_r$ with a free bias.\\
12e & \thmref{Prop}{VI.4}(b) &
The serial adapter $(I+UD)W_0$ at $(m,n,r)=(5,3,1)$: Jacobian rank. &
$7$, equal to LoRA's $r(m+n-r)$, since this $W_0$ is injective.\\
12f & \thmref{Prop}{II.12}, \thmref{Prop}{II.13} &
FourierFT atoms: the real parts of inverse two-dimensional DFTs of single frequencies on $8\times8$. &
The frequencies $(1,2)$ and $(7,6)$ give the same atom; three frequencies containing one conjugate pair span $2$
dimensions.\\
12g & \thmref{Cor}{II.6}, \thmref{Thm}{V.3} &
HF's default Cayley--Neumann factor $R=I+2Q+2Q^2+2Q^3+Q^4$ for a skew $Q$ of block size $8$ with entries of size about
$0.07$; then $200$ successive merges with entries of size about $0.05$. &
$\lVert R\T R-(I-Q^4)^2\rVert=4.7\times10^{-16}$ and $\lVert R\T R-I\rVert=1.5\times10^{-2}$. Relative drift of $K_{\rm
out}$: $0.002$, $0.010$, $0.064$, $0.224$ after $1$, $10$, $50$, $200$ merges.\\
12h & \thmref{Cor}{II.6} &
BOFT's $\diag(s)\,W_0R\T$ with an exact Cayley $R$ and $s=(-0.5,1,\dots,1)$. &
Largest change of a cosine between neurons $1.50$; of an absolute cosine $5.6\times10^{-16}$.\\
12i & \thmref{Prop}{V.4} &
DARE with a fixed mask under an invertible diagonal $D$ with signs; TIES (top $20\%$ kept, sign election by total mass)
under scalings. &
$\mathrm{DARE}(D\tau)-D\,\mathrm{DARE}(\tau)$: $1.8\times10^{-15}$. TIES: $8.9\times10^{-16}$ for the factor $3.7$,
$0.0$ for $-2$, and $5.6$ for a generic positive diagonal (control).\\
12j & \thmref{Prop}{II.16} &
The whitened split $[W_0L]_rL^{-1}$, the paper-form CorDA split $[W_0C]_rC^{-1}$ and PiSSA's $[W_0]_r$ at
$(m,n,r)=(6,5,2)$: the defect of $f(W_0h^{-1},hCh\T)$ against $f(W_0,C)h^{-1}$ for an input change $h$. &
Under $h=3O$, $O\in\Ort(n)$: $2.2\times10^{-15}$, $1.8\times10^{-15}$, $8.1\times10^{-16}$. Under a generic $h$: $1.3\times10^{-15}$,
$0.63$, $1.1$. Whitened split with damping $C+0.1I$, generic $h$: $0.14$. The whitened split equals $\Pi_rW_0$ to
$5.3\times10^{-15}$.\\
12k & \thmref{Thm}{III.12}(b) &
The first Adam step with $\varepsilon=10^{-8}$ and gradients of size $10^{-6}$, under $g=\diag(0.3,1,4)$ with
per-channel rates. &
Relative defect $1.6\times10^{-2}$ with $\varepsilon$ shared (control), $4.9\times10^{-17}$ with $\varepsilon$ rescaled
per channel.\\
12l & \thmref{Thm}{III.12} &
LoRA-Pro's parameter step at $(m,n,r)=(8,7,3)$, $s=1.5$, with the Sylvester-optimal free matrix $X$ and with $X=0$. &
Non-orthogonal $g$: $1.2$ with the Sylvester $X$, $1.9\times10^{-15}$ with $X=0$. Orthogonal $g$: $1.2\times10^{-15}$
and $1.4\times10^{-15}$.\\
12m & \thmref{Thm}{II.2}, \thmref{Prop}{II.3} &
$\dim S_1$ at $(m,n,r)=(7,5,2)$ for the three strata and for HF's \texttt{orthogonal} initialisation. &
$\mathrm T_A$: $14=mr$; $\mathrm T_B$: $10=nr$; $\mathrm S$: $20=r(m+n-r)$. Orthogonal init: $\lvert B_0A_0\rvert\le
3.5\times10^{-18}$ and $\dim S_1=11=m+n-1$, the rank formula at $\rk A_0=\rk B_0=1$.\\
12n & \thmref{Obs}{I.7}, \thmref{Prop}{I.8}(a) &
$\dim S_1$ of a standard LoRA$_4$ and of LoRA$_2\oplus{}$LoRA$_2$ with one $A_0$ used twice ($m=7$, $n=6$); the dimension
of the fibre product of $B\mapsto W_0+BA_0$ and $A\mapsto W_0+B_0A$ at $(6,8,2)$. &
$28$ against $14$; fibre product $4=r^2$.\\
12o & \thmref{Prop}{II.13} &
Two HiRA cycles at $r=1$ on a $6\times5$ weight with no zero entry: the rank of the relative update
$(W_2-W_0)\oslash W_0$. &
$3$, above $2r=2$ and equal to $\min(2r+r^2,m,n)$.\\
12p & \thmref{Thm}{II.11} &
The Kronecker rank of a generic $uv\T$ for the $2\times4$ split of $\R^8$; the largest rank of a LoKr update $C\otimes
D$ with $C\in\R^{4\times4}$ and a rank-one $D\in\R^{4\times4}$ over $200$ draws. &
$4=\min(m_1,m_2)\min(n_1,n_2)$, the largest Kronecker rank for this split; $4$.\\
12q & \thmref{Thm}{V.3}(c), cor.~4 &
Connected components of the hypergraph of BOFT blocks of size $b=4$ on $n=64$ coordinates after $m'$ butterfly
factors. &
Component sizes $4$, $8$, $16$, $32$, $64$ for $m'=1,\dots,5$: connected only at full depth.\\
12r & \thmref{Cor}{III.6} &
A long gradient flow ($T=200$) from $B_0=0$ at $\eta_B/\eta_A=4$, decay $\lambda=0.1$, $(m,n,r)=(12,10,3)$: the final
sizes of $\mu=B\T B-AA\T$ and of $\Phi$, relative to $B\T B$ and to $B\T B/\eta_B$. &
Coupled decay: $\mu$ $1.0\times10^{-12}$, $\Phi$ $3.0$. Per-time decay: $\mu$ $0.75$, $\Phi$ $1.4\times10^{-12}$.\\
12s & \thmref{Cor}{III.13} &
A $\GL_r$-equivariant example rule, $B\leftarrow B-\eta\,G_B(AA\T)^{-1}$ and $A\leftarrow A-\eta\,AA\T G_A$, for $50$
steps from $B_0=0$, started at $A_0$ and at $gA_0$. &
Difference of the final weights $1.8\times10^{-15}$.\\
12t & \thmref{Thm}{V.3}, cor.~4 &
The Lie algebra generated by the logarithms of $12$ random HF Cayley--Neumann factors with $b=4$. &
Dimension $16=\dim\gl(4)$, against $\dim\so(4)=6$. This identifies the generated group only; re-merging reaches a
monoid (check 13n).\\[2pt]
\multicolumn{4}{@{}l}{\sffamily\bfseries\color{tide}Section 13: the second referee round}\\[1pt]
13a & \thmref{Obs}{I.7} &
$\dim S_1$ at $m=7$, $n=6$ for LoRA$_2\oplus{}$LoRA$_2$ with independent $A_0$'s, for $F(1)\oplus F(2)$, and for a
standard $F(3)$. &
$28$; $21$ and $21$.\\
13b & \thmref{Thm}{II.2}(e) &
HF's \texttt{orthogonal} initialisation against a $\mathrm T_A$ point at $(m,n,r)=(4,5,2)$. &
$\dim S_1=8=mr$ for both, but the two first-order images together span $14$ dimensions, so they differ.\\
13c & \thmref{Cor}{II.6} &
BOFT with every $s_i<0$; GOFT with its scaler $g=1\pm0.05$; a conformal block $\sqrt2\,\mathrm{Rot}(0.3)$ on two of eight
inputs; the Cayley--Neumann factor at $\theta=1$ with $b=2$. &
Cosine change $4.4\times10^{-16}$. GOFT: cosine change $0.044$, relative singular-value change $0.033$, and $K_{\rm out}=
W_0\diag(g)^2W_0\T$ to $6.1\times10^{-15}$. Conformal block: cosine change $0.19$. $\lVert R\rVert=0$, so $R=0$.\\
13d & \thmref{Thm}{II.7} &
The Hurwitz move $H_aH_b=H_{H_ab}H_a$ for reflections in $\R^6$. &
Defect $4.4\times10^{-16}$.\\
13e & \thmref{Prop}{II.8} &
DoRA at $(8,8,2)$ with one magnitude $\mu_i=0$: Jacobian rank. &
$34$, against $36$ when every $\mu_i\ne0$.\\
13f & \thmref{Thm}{II.11}(e) &
The largest rank of a LoKr update $C\otimes BA$ over $50$ draws. &
$(m_1,m_2,n_1,n_2)=(2,8,8,2)$ with inner dimension $4$: $4$. $(4,2,2,8)$ with inner dimension $5$: $4$. Predicted
$\min(m_1,n_1)\min(r',m_2,n_2)=4$ and $4$.\\
13g & \thmref{Prop}{II.12} &
LoHa$_{r,r}$ against LoRA$_{2r}$; the orbit of LoHa's exhibited gauge; HRA at $r\ge n-1$. &
$(9,9,3)$: LoHa $73$, LoRA$_6$ $72$. $(4,4,2)$: $16$ and $16$. LoHa at $(4,4,2)$: orbit dimension $15$, fibre dimension
$16$. HRA at $n=5$: $10$ for $r=4$ and for $r=5$, equal to $\dim\SO(5)$.\\
13h & \thmref{Prop}{II.13} &
OFT transported by $T=2I$, through two merge-and-restart cycles of the rotation $-I$ from $\theta_0=(1,0)$; HiRA after
two cycles at $r=1,2,3$ on a $30\times30$ weight. &
$\theta_2=(9,0)$, of norm $9$, off the rotation orbit of $\theta_0$. HiRA's relative-update ranks $3$, $8$, $15$, equal
to $\min(2r+r^2,30)$.\\
13i & \thmref{Prop}{II.16} &
Bottom-$r$ splits (CorDA's knowledge-preserving form on $W_0C$; MiLoRA on $W_0$) and whitened splits with absolute
damping $0.1$ and with trace-relative damping $0.1\tr C/n$, at $(m,n,r)=(7,6,2)$. &
Under $h=3O$: $3.1\times10^{-15}$, $4.0\times10^{-15}$, $2.3\times10^{-2}$, $3.0\times10^{-15}$. Under a generic $h$:
$4.9$, $0.34$, $4.6$, $5.9$. At $r=n$, PiSSA under a generic $h$: $3.1\times10^{-14}$.\\
13j & \thmref{Thm}{III.5} &
Per-channel rates $\eta_B=(1,1,1)$ and $\eta_A=(1,1,2)$ at $(m,n,r)=(5,4,3)$ on the non-quadratic loss
$\sum_{ij}\log\cosh W_{ij}-\langle Y,W\rangle$, by Runge--Kutta with $h=10^{-3}$ for $1000$ steps; one discrete GD step. &
Drift of the charge of channels $1,2$ (equal products $\eta_{B,j}\eta_{A,j}$) $4.2\times10^{-12}$; of the form of
channels $1,3$, $3.1$. One GD step changes $\Phi$ by $\eta_BAG\T GA\T-\eta_AB\T GG\T B$ to $2.4\times10^{-15}$.\\
13k & \thmref{Cor}{III.6} &
One Adam sign step: $\lVert\Delta\Phi\rVert$ divided by the learning rate. &
$20.9$, $21.1$, $21.1$ at learning rates $10^{-2}$, $10^{-3}$, $10^{-4}$: first order in the step size.\\
13l & \thmref{Prop}{IV.3} &
GaLore's field for $T=1$ at $W=\diag(2,1)$ with $L=\lVert W\rVert^2/2$: the Lie bracket of $X_i(W)=u_1(W)e_i\T$ by
automatic differentiation; the cometric of $B\mapsto BB\T$ at $B$ and at $BR$, $R\in\Ort(2)$. &
$[X_1,X_2]=-\tfrac13e_2e_1\T$ to six digits, as predicted. Cometric difference $9.5\times10^{-16}$.\\
13m & \thmref{Prop}{V.2} &
Blockwise absmax NF4 (blocks of $64$) on $256\times256$ weights with $r=8$: the defects of QLoRA, QPiSSA and one-step
LoftQ; NF4 under addition. &
Gaussian $W$: $\lVert e\rVert=0.945$, $\lVert e-\mathrm{SVD}_8(e)\rVert=0.889$, LoftQ with $s=2$: $0.945=\lVert e\rVert$,
QPiSSA $0.887$. Low rank plus noise: $\lVert e\rVert=0.592$, QPiSSA $0.227$, LoftQ with $s=1$: $0.549$. $96.7\%$ of the
entries of $\kappa(W)+\kappa(\Delta)$ lie off the NF4 grid.\\
13n & \thmref{Thm}{V.3}, cor.~4 &
The operator norm of Cayley--Neumann factors for $2000$ random skew $Q\in\R^{4\times4}$ with $\lVert Q\rVert_{\rm op}\le
2^{1/4}$; $200$ successive merges on $W\in\R^{6\times8}$. &
Largest $\lVert R\rVert_{\rm op}$: $1$ to rounding. Largest eigenvalue of $K_{\rm out}(t+1)-K_{\rm out}(t)$:
$-5.5\times10^{-9}$, so $K_{\rm out}$ decreases in the Loewner order at every merge.\\
13o & \thmref{Prop}{V.4} &
Factor-wise DARE on two adapters under a diagonal gauge of one adapter or of both; the LoraHub change formula with three
modules; the fibre of Poly with $K=3$ skills, $r=2$ and $6\times6$ weights. &
Gauge on one adapter: $39.5$ (control); on both: $5.6\times10^{-15}$. LoraHub formula: $4.0\times10^{-15}$. Poly fibre:
$32$ with $2$ tasks, $4=\dim\GL_2$ with $4$ tasks.\\
13p & \thmref{Thm}{VI.3} (R5) &
Transposition under RoPE with $d_k=8$: $\langle R_mMq,R_nk\rangle$ against $\langle q,M\T R_{n-m}k\rangle$, and against
the slide onto the key side, $\langle R_mq,R_nM\T k\rangle$. &
$2.2\times10^{-16}$; the key-side slide differs by $2.2$ (control).\\
13q & \thmref{Prop}{VI.4}(b) &
The serial adapter at $(m,n,r)=(4,8,1)$: Jacobian rank. &
$7$, against $d(\mathrm{LoRA}_1)=11$: with $m<n$ the update rows stay in $\operatorname{row}W_0$.\\
\end{longtable}\endgroup

\subsection{Numbers from other sources}\label{app:repro-other}

A few [Num] values in the main text are not printed by the script. They come from the replications that the referees ran
in the two review rounds, recorded with the review notes of each result, and from the check scripts that accompanied
the four proposals from which the framework was synthesised. They are the rank formula of \thmref{Proposition}{II.3} in
$400$ random cases and its loss rate, the Adam steps after \thmref{Lemma}{II.4}, the LOFT identities of \S\ref{sec:erlangen-loft}, the
RoAd$_2$ dimension (\thmref{Proposition}{II.9}), the $283$ random networks of \thmref{Theorem}{II.10}, VeRA's dimension
$m+r-1$ and the LoHa dimension at $(20,16,2)$ in the census of \thmref{Proposition}{II.12}, the operator-Schmidt ranks
after \thmref{Theorem}{II.11}, the transported cometric and Adam runs of \thmref{Proposition}{II.13}, the shared-frame
and RoPE tests of \thmref{Observation}{II.15}, the finite-difference cometrics of \thmref{Observation}{III.2} and
\thmref{Theorem}{III.3}, LoRA-FA under AdamW (\thmref{Proposition}{III.4}), the non-gradient signals and the clipping of
\thmref{Theorem}{III.5}, the Kempf--Ness minimum of \thmref{Proposition}{III.7}, the closed form of
\thmref{Theorem}{III.9}, the warm-up, $\varepsilon$, decay and clipping runs of \thmref{Theorem}{III.10} and
\thmref{Corollary}{III.11}, LoRA-RITE, LoRA-Pro's second-order $W$-step defect and the zero-$B$ runs of \thmref{Theorem}{III.12} and \thmref{Corollary}{III.13},
the GaLore runs of \thmref{Proposition}{IV.2}, the general bracket formula of \thmref{Proposition}{IV.3}, the NF4
example with a dominant low-rank part in \thmref{Proposition}{V.2} (check 13m runs an analogous one at another scale),
the query-scaling and post-LN tests of \thmref{Theorem}{VI.3}, and the decoder tests of \thmref{Theorem}{VI.5}. Where a
referee's replication and the script use different random draws, the main text may quote the referee's value; the
conformal block of \thmref{Corollary}{II.6}, for instance, changes some cosine by $0.17$ in the referee's draw and by
$0.19$ in the script's (check 13c).

\subsection{Software}\label{app:repro-software}

Table~\ref{tab:repro-software} lists the software behind the run of Table~\ref{tab:repro-checks}, the figures and this
document. Only the first four rows are needed to re-run the checks.

\begin{table}[t]
\centering\footnotesize
\caption{Software and hardware for the checks, the figures and the companion site.}\label{tab:repro-software}
\begin{tabularx}{\linewidth}{@{}>{\raggedright\arraybackslash}p{3.6cm}>{\raggedright\arraybackslash}p{3.7cm}>{\raggedright\arraybackslash}X@{}}
\toprule\headrow
\textbf{Component} & \textbf{Version} & \textbf{Used for}\\\midrule
Python & 3.9.6 & the check script, the figure scripts, the generators of the tables and ancillary files\\
PyTorch \citep{xpaszke2019pytorch} & 2.8.0 & float64 automatic differentiation, linear algebra, seeded random draws\\
NumPy \citep{xharris2020array} & 2.0.2 & discrete Fourier transforms and ranks in the checks; the figure scripts\\
SciPy \citep{xvirtanen2020scipy} & 1.13.1 & Sylvester equations (LoRA-Pro, check 12l) and matrix logarithms (check 12t)\\
Matplotlib \citep{xhunter2007matplotlib} & 3.9.4 & the figures of this paper\\
Hugging Face PEFT \citep{xmangrulkar2022peft} & commit \texttt{85be3f3}, 30 September 2026 & the implementation that every library claim was checked against; not needed to run the checks\\
TeX Live & 2026 (pdf\TeX{} 1.40.29, latexmk 4.88) & this document\\
D3 \citep{xbostock2011d3} & 7.9.0 & drawing the companion site's figures\\
MathJax \citep{xcervone2012mathjax} & 3.2.2 & typesetting the companion site's mathematics\\
Hardware & Apple M5, macOS 26.7, CPU only & the run reported in Table~\ref{tab:repro-checks}\\
\bottomrule
\end{tabularx}
\end{table}

\subsection{How the companion site and the figures compute what they show}\label{app:repro-site}

The companion site is one static page. A build script assembles it from one HTML file per exposé, and a second script
compiles its data file from the verified catalogue, the arXiv metadata, the classification batches, the axis definitions,
the Rosetta stone and the numbered results. The data file is loaded as a script, so the page runs from a web host, from
disk or in a sandboxed viewer without fetching its data; it loads D3, MathJax and its fonts from public CDNs. The tables of Appendix~\ref{app:catalogue} and the ancillary
files are generated from the same data.

Each figure of the site is a small module in plain JavaScript that computes every number it shows in the reader's
browser, when it first comes into view and again at every interaction; nothing is typed in by hand. JavaScript numbers
are IEEE doubles, so the arithmetic is float64, as in the script. A shared kernel provides dense matrix products,
Kronecker and Hadamard products, a cyclic Jacobi eigensolver for symmetric matrices, singular values, numerical rank,
random orthogonal matrices and a seeded generator (mulberry32, with Box--Muller normals), so every figure opens on the
same default state. Figures that must see exact zeros, such as the operator-Schmidt spectra behind Figure~\ref{fig:cut}
and the invariant lamps of the Gram lab, use a one-sided Jacobi SVD, which resolves singular values down to about
$10^{-16}$ of the norm, where an eigensolver applied to the Gram matrix stops near $10^{-8}$.
\begin{itemize}
  \item The orbit explorer trains a seeded toy regression twice, by a hand-written bias-corrected Adam or by gradient
  descent. Its sliders move on a factor-of-two grid, so every torus element it can reach rescales doubles exactly, and at
  $\varepsilon=0$ without decay or clipping its meter $\max_t\lVert W_t-W'_t\rVert_F$ reads exactly $0$, as check~3
  does.
  \item The gauge lab computes the update of SGD, Adam, Adam with per-channel rates, damped and undamped scaled GD and
  LoRA-Pro at $(B,A)$ and at $(Bg^{-1},gA)$ and prints the relative defect, as checks 2, 7, 12k and 12l do. It recomputes
  task arithmetic and LoraHub from the factors at every gauge, as check 13o does.
  \item The Noether figure integrates scalar LoRA under gradient flow by fourth-order Runge--Kutta with $h=10^{-3}$, under
  gradient descent by Euler steps and under Adam step by step, so that the charge is computed and never assumed. Its
  matrix panel uses the implicit midpoint rule, which keeps quadratic invariants to rounding, and compares the
  closed-form velocity of \thmref{Theorem}{III.9} with the integrated one.
  \item The apex figure checks its counters by a Jacobian rank on mount, reproducing the [Num] lines of
  \thmref{Theorem}{II.2} and \thmref{Theorem}{II.7}. The rank-versus-volume figure rebuilds each parametrisation, forms
  its Jacobian by central differences, which are exact here because each map is affine in every single coordinate, and
  takes its numerical rank by Householder QR with column pivoting; for a large layer it runs on a small copy with the
  same rank parameters. The plug workbench (\S\ref{sec:cut-plug}) measures dimensions the same way.
  \item The merge game checks every rewrite of \thmref{Theorem}{VI.3} on seeded random weights of a small pre-LN
  transformer block and prints the relative output difference, about $10^{-16}$ for a legal rewrite, or the residual of
  a refused one. The prefix figure computes its attention weights from seeded synthetic tokens. The rebasing figure
  measures the rank of accumulated ReLoRA updates by Jacobi SVD against $\min(Kr,m,n)$. The lens figure runs two float64
  checks on mount, of \thmref{Proposition}{IV.2} and of the bracket of \thmref{Proposition}{IV.3}.
  \item The budget calculator evaluates parameter counts from each base model's \texttt{config.json} and reproduces on
  screen the parameter count of the official checkpoint. The catalogue, the periodic table, the timeline, the category
  graph and the citation panel compute every count from the data file at load, and the graph's layered layout is a
  deterministic Sugiyama procedure.
\end{itemize}

The static figures of this paper are drawn by Python scripts kept with the authors' sources, most of them ports of the
site's modules, with NumPy and Matplotlib. They recompute every plotted number in float64 from the site's toys and seeds.
Figure~\ref{fig:orbit}, for example, ports the orbit explorer and check~3, and Figure~\ref{fig:gauge} ports the gauge lab
and checks 2, 7, 12k, 12l and 13o.

\subsection{The numbers behind the figures}\label{app:repro-figures}

The captions of the figures describe what each panel shows. This subsection records the settings and the values
behind them, as the figure scripts print them. Every value is recomputed in float64 when a script runs.

\paragraph{Figure~\ref{fig:rankvol}} GraLoRA with $k=2$, $4$ and $8$ blocks reaches ranks $16$, $32$ and $64$ at
$d=65{,}472$. HiRA$_8$, the transport $X\mapsto W_0\odot X$ of LoRA$_8$ (\thmref{Proposition}{II.13}), reaches full rank
for generic $W_0$. MoRA in its reshape form $\Delta W=I_{16}\otimes M$ with $M\in\R^{256\times256}$, and C3A with
$256\times256$ circulant blocks, are injective, so $d=\lvert M\rvert$, and both reach full rank. FourierFT with
$n_c=65{,}536$ frequencies from HF PEFT's default draw (seed $777$) has $120$ conjugate pairs, so $d=65{,}416$, and its
maximum rank is known only through the bound $\min(2n_c,m,n)$. No KronA factor shapes have budget exactly $65{,}536$. The
closest, $256\times256\otimes16\times16$, cost $65{,}792$ and give $d=65{,}791$ and rank
$\min(m_1,n_1)\min(m_2,n_2)=4{,}096$ (\thmref{Theorem}{II.11}(b)). LoHa$_{4,4}$ is placed at the proved upper bound
$F=57{,}313$; it reaches rank $16$, as GraLoRA with $k=2$ does, with at least $(m+n-1)-2r^2=8{,}159$ fewer dimensions
than LoRA$_8$ (\thmref{Proposition}{II.12}(c)). Every dimension and maximum rank shown is confirmed by float64 Jacobian
ranks and random-point ranks on a $64\times64$ copy.

\paragraph{Figure~\ref{fig:cut}} The vectors $u,v$ and the factors $C,D$ are seeded Gaussians, and the rank threshold is
$10^{-10}$. The numerically zero values are at most $1.3\times10^{-16}$; values below $10^{-17}$, exact zeros included,
are drawn at $10^{-17}$, and every nonzero value is at least $9.5\times10^{-4}$. Each reshape permutes the entries, so
the squared normalised singular values sum to $1$ on every cut. A bond of dimension $k$ costs $k(4\cdot4+4\cdot4)=32k$
parameters across any of the three cuts.

\paragraph{Figure~\ref{fig:noether}} Panel (a) integrates gradient flow by fourth-order Runge--Kutta with step $10^{-3}$.
The charge is integrated, not imposed, and stays at $\Phi_0$ to within $6\times10^{-11}$ on every run. From the zero-$B$
start $(1.2,0)$, $\Phi_0=-1.44$, and Adam with $\eta=0.1$ ends at $\Phi=-0.53$. A GD step multiplies the charge by exactly
$1-\eta^2\eta_A\eta_B\,s^2(w_k-w^\star)^2$; the GD drift is $1.2\times10^{-3}$ at $\eta=0.0025$ and $0.14$ at $\eta=0.2$,
and $5.8\times10^{-13}$ for the flow. The Adam drift stays between $0.55$ and $1.5$. In panel (b) the seeded target update
has singular values $2.0$ and $0.3$, with $s=1$ and $\eta_A=\eta_B=1$, so $\mu=s\sqrt{\eta_A\eta_B}=1$. The flow is
integrated by the implicit midpoint rule. The hollow circles are $u_i^\top(BB^\top/\eta_B)u_i$ and
$v_i^\top(A^\top A/\eta_A)v_i$; the charge stays at $-\kappa I$ to $2\times10^{-14}$, and the closed-form $\dot W$ matches
$s(\dot BA+B\dot A)$ to a relative $1.3\times10^{-13}$. The one-sided regime is where one gain exceeds ten times the
other, $\sigma<0.70\,\sigma^*$, and the balanced regime where the two gains are within a factor two,
$\sigma>2.83\,\sigma^*$. $\sigma_1$ crosses $\sigma^*$ at $t\approx0.49$ and settles at $2.0$, where the $A$-update
supplies $38\%$ of its growth rate; $\sigma_2$ settles at $0.3$, where that share is $7\%$.

\paragraph{Figure~\ref{fig:orbit}} The toy has $W_0\in\R^{8\times6}$, a rank-two target update and $r=2$. Both runs take
300 steps of a hand-written, bias-corrected Adam in float64 and reach a loss of $2.7\times10^{-7}$; the control ends at
$5.7\times10^{-2}$. At $\varepsilon=0$ the runs are bit-identical at all 301 steps (check~3 of the script). The red curves
use Adam's $\varepsilon=10^{-8}$ or $10^{-6}$ (the gap is proportional to $\varepsilon$), PyTorch-style decoupled weight
decay $0.01$, and global-norm clipping at $1$, which clips 8 steps of the LoRA+ run and 37 of the LoRA run. Their maximal
gaps are $4\times10^{-8}$, $4\times10^{-6}$, $7\times10^{-3}$ and $0.43$, against $2.0$ for the control. The per-block
rescaling sets $\varepsilon_B=16\,\varepsilon$, decay $16\times0.01$ on $B$, and $B$'s clipping threshold multiplied by
$16$.

\paragraph{Figure~\ref{fig:gauge}} The state is $B\in\R^{5\times2}$, $A\in\R^{2\times4}$. The rotation is by $30^\circ$,
the stretch is $\diag(e^{0.5},e^{-0.4})$, and the signed permutation swaps the two rank channels and flips one. Adam is
its first step from zero state, which at $\varepsilon=0$ is sign descent; the per-channel rescaling for $g=\diag(d)$ is
$\eta_{B,j}\mapsto\eta_{B,j}/d_j$, $\eta_{A,j}\mapsto d_j\,\eta_{A,j}$, and the damped scaled GD has $\delta=0.1$. The
first-order weight update is $P_UG+GP_V$ for undamped scaled GD and $P_UG+GP_V-P_UGP_V$ for LoRA-Pro. In panel (b) the
weights are $w=(0.6,0.5,0.4)$. Task arithmetic changes by less than $10^{-16}$ relative to its norm, the LoraHub merge by
up to $3.4$ times its norm at $c=1/8$, and the closed form
$\sum_{l\ne2}w_lw_2\big[(c-1)B_2A_l+(c^{-1}-1)B_lA_2\big]$ matches the recomputed change to $10^{-15}$.

\paragraph{Figure~\ref{fig:rebase}} In panel (a), ReLoRA reaches the full rank $16=\min(m,n)$ after $6$ cycles for $r=3$
and after $4$ for $r=5$. LoRA-XS adds updates $U_rR_kV_r^\top$ inside one $r^2$-dimensional subspace. In panel (b), BOFT's
level $\ell$ pairs coordinate $c$ with $c\oplus2^\ell$. Each light bar equals both the dimension of the Lie closure of the
blocks' plane rotations under brackets and the measured dimension of the reachable set after $16$ merge cycles of
exact-Cayley factors (the rank of the right-trivialised Jacobian of the product map). Block OFT reaches $24$ and $56$ in
one cycle and never more. BOFT goes from $16$ to $24$ and from $24$ to $56$ at depths $m'=2,3$, and only the full depth
$m'=1+\log_2(16/b)=4$ connects the hypergraph: a single full-depth BOFT has dimension $32$, and merging reaches $120$ after
$4$ cycles. The $7$ Givens planes leave components of sizes $4,3,2,2$ and five isolated coordinates, so they generate only
$11$ dimensions; the $15$-plane tree reaches $120$ after $9$ cycles.

\paragraph{Figure~\ref{fig:prefix}} The query has $q\in\R^4$ and $\lVert q\rVert=2.4$; the logits are $\langle q,k\rangle/2$,
and the prefix keys are scaled by $1.2$. In panel (a), $\lambda_q=0.44$, so every content weight is multiplied by $0.56$,
and the six ratios agree to within $10^{-15}$. In panel (b), $g=0.5$ and the ratios are exactly $1$. In panel (c), the two
principal axes keep $88\%$ of the variance of the values, and the LLaMA-Adapter output lies at distance $0.65$ from the
segment in $\R^4$. The script evaluates the joint softmax directly and recovers the identity of part~(b) of
\thmref{Theorem}{VI.5} to $4\times10^{-16}$.

\section{How this paper and its website were made}\label{app:making-of}

This appendix records how the paper and its companion site were produced: who did what, how the material was found, how the results were generalised, unified and checked, and what it cost.

\subsection{In brief}

\begin{itemize}
\item 1 October 2026, 17:06 AEST, to 4 October 2026, about 01:00 AEST, when the first full version of this appendix was written: about 56 hours of wall-clock time. At least one agent was running for about 23 of those hours. The rest was a 43-minute wait for an account usage reset, two interruptions, and long pauses between requests.
\item The authors directed one orchestrating session of Claude (Anthropic's Claude Opus 5.5) running in Claude Code, in short messages and a few answers to multiple-choice questions. That session wrote and launched 20 multi-agent workflows, which did most of the research, writing and checking, and two small ones that checked and redrafted this appendix. It did the planning, integration and publishing itself.
\item Those workflows started 422 agent runs (420 distinct agent calls, two of which were started twice). 358 completed, 29 failed when the account reached a spend limit, and 35 were stopped or cut off.
\item The agents processed about 117 million fresh tokens: prompts written to the cache plus generated text, of which 8.6 million were generated. They also re-read 4.1 billion tokens from the prompt cache. The orchestrating session added about 10 million fresh tokens and 0.2 billion cache reads.
\item The outputs are a website (15 sections including a version of this appendix, 20 interactive figures plus an animated hero, and a checked catalogue of 382 methods and 138 papers) and a paper of about 400 pages with the same numbering.
\item On 5 October a later session made a narrated film of the atlas, 16 minutes 44 seconds long, and a two-minute teaser. Its five agents processed about 5.0 million fresh tokens, and the session itself about 1.4 million.
\end{itemize}

\subsection{Who did what}

The authors think that when the literature on parameter-efficient fine-tuning (PEFT) already holds several hundred methods, the field gains more from the approach Grothendieck pictured as a rising sea than from one more method. This atlas catalogues 382 of them.

In his memoir \emph{Récoltes et Semailles}, written 1983--1986, Grothendieck compared a hard problem to a nut. One can strike the shell with hammer and chisel, at point after point, until it cracks. Or one can soak the nut in a softening liquid and let time pass, until the pressure of a hand is enough. On a larger scale the second way is the rising sea. It advances so slowly and quietly that nothing seems to change, until it surrounds the problem and the problem is dissolved by a theory that reaches well beyond it~\citep{xgrothendieck2022recoltes,xmclarty2007rising}. In the authors' analogy, a 383rd method would be one more strike of the chisel, and the sea is a theory in which the existing methods appear as instances of one notion.

The authors have packaged this way of working as a Claude Code skill. It asks for the natural home of a result, for objects defined by their relationships and for the relative point of view, so that hard results dissolve. Its binding rule is that every reframe must pay, by a shorter proof, a dropped assumption or a real unity, or be rejected.

This view shaped the question the authors chose, to re-found PEFT ``as Grothendieck would'' and to make the result an interactive site. They also made the main decisions of scope and presentation. Their instructions to the session were short messages: pointing out some missing methods in the literature, asking for a sweep of 2026 work, a mathematical appendix, more readable typefaces, a LaTeX paper and this appendix, supplying the author list and the title, and asking the session to resume after interruptions.

The orchestrating session did the following:

\begin{itemize}
\item planned the work and wrote the seed sketch of the theory;
\item designed the visual identity, the data pipeline, the build and test tools and the LaTeX template;
\item launched and resumed the workflows, read their reports and decided what to run next;
\item applied some changes itself, such as the new typography rules, and rebuilt, tested and published the site, and built the paper.
\end{itemize}

The sub-agents researched the literature, proposed, judged and refereed the theory, classified the methods, wrote the sections, drew and tested the figures, and reviewed each other's work. They also made the remaining decisions, including which 2026 candidates entered the catalogue and what was cut from the site in its final review.

A workflow is a script for Claude Code's Workflow tool. It starts sub-agents in parallel or in sequence, hands each a prompt and files, and keeps a journal of their results. Every sub-agent ran the same model, \texttt{claude-opus-5-5}, whose training data ends in June 2026. All had web search and fetch and a shell on one Apple M5 laptop, with Python 3.9 (NumPy, PyTorch, SciPy, Matplotlib), Node.js, headless Chrome and TeX Live 2026. Below, an independent check means one made by a separate agent with its own context and instructions.

\subsection{Finding the material}

The literature work was split into thirteen families:

\begin{itemize}
\item the Hugging Face PEFT library;
\item LoRA and its training-dynamics variants;
\item initialisation and rank allocation;
\item sharing and extreme efficiency;
\item structured shapes;
\item multiplicative and orthogonal methods;
\item adapters;
\item prompts and states;
\item selective fine-tuning;
\item quantisation and memory;
\item composition and merging;
\item theory and foundations;
\item the 2025--2026 frontier.
\end{itemize}

One agent swept each family. It worked from the arXiv API, Semantic Scholar, OpenReview, the ACL Anthology, the official conference listings and the Hugging Face PEFT source code. Each sweeper was told to read a paper's method section before writing down its formula. As each family finished, an independent fact-checker re-verified every entry:

\begin{itemize}
\item title, first author and first-version year against the arXiv API;
\item the venue against the proceedings;
\item the formula against the method section;
\item the parameter count, re-derived independently;
\item mergeability and initialisation;
\item Hugging Face PEFT support, against the repository at commit 85be3f3 (30 September 2026).
\end{itemize}

A completeness critic then compared the whole catalogue with the library and with its own knowledge of the field. It listed 34 missing methods, and 32 of them were researched and checked in four more batches. The checkers, including those of the four gap-filling batches, made 613 corrections before the first release. Merging duplicates on a canonical identifier and an alias table gave 320 methods and 124 papers. Matching by name would have folded LoRA+ into LoRA.

After the first build, the authors pointed out methods missing from the catalogue. They were added and refereed, and a dedicated sweep of 2026 work followed, in which six scouts covered:

\begin{itemize}
\item arXiv submissions from January to September 2026, one scout per quarter, by keyword queries over titles and abstracts;
\item the accepted papers of ICLR 2026 and ICML 2026, with ACL and CVPR 2026;
\item the citation graph of ten anchor PEFT papers and the 2026 Hugging Face PEFT releases.
\end{itemize}

A triage agent then chose which candidates entered the atlas. It admitted new 2026 methods with at least one signal (a 2026 venue, inclusion in Hugging Face PEFT, citations, a widely used codebase, or a structure the seven coordinates would treat differently), at most 36 per pass. The first triage saw only the first 105 of the 265 candidates not already in the atlas, the last of them cut off mid-entry, because the orchestrator had truncated the list it passed in. A second pass triaged the remaining 161 candidates and the scouts' overflow lists. Together the passes added 61 methods and 14 theory papers. They were researched and fact-checked in the same way, giving 382 methods, 138 papers and 905 corrections in total.

The BibTeX of the 502 catalogue entries is built from arXiv metadata for the 493 that have an arXiv identifier, and the other nine are built from the catalogue's own fields. Eight of those nine were corrected by hand for the paper. The paper's 130 other references, mostly mathematics and related work, were written by the agents who cited them, who were told to check every field against the publisher, DOI or arXiv page.

\subsection{From examples to a theory}

The orchestrator wrote a seed sketch of eleven numbered points, about 1,000 words, which the architects could adopt, correct or reject. Its points were:

\begin{itemize}
\item methods as reparametrisations in Para and as objects of a pointed slice;
\item frozen and full fine-tuning as initial and terminal;
\item mergeability as a lifting problem;
\item Klein's Erlangen programme for group-type methods;
\item the $\GL_r$ gauge of LoRA and its conserved charge;
\item string diagrams for structured updates;
\item a monoidal sum;
\item base change;
\item the lens duality between forward and backward methods;
\item prompts and prefixes as learned input states;
\item naturality, so that a method defined once per layer type extends to every network.
\end{itemize}

Four architect agents each developed a complete framework from this sketch, with definitions, propositions and proofs, from a different point of view:

\begin{itemize}
\item parametric lenses;
\item Grothendieck's relative point of view;
\item the Erlangen programme with gauge symmetry;
\item string diagrams.
\end{itemize}

Three judge agents scored all four proposals, each from an assigned point of view:

\begin{itemize}
\item a category theorist's, for rigour;
\item a PEFT researcher's, for faithfulness to the methods;
\item a science communicator's, for clarity and visual potential.
\end{itemize}

All three recommended the relative point of view as the backbone. One synthesis agent then built the framework on it. As the judges advised, it took the lens and cometric from the parametric-lens proposal, the covariance and optimizer gauge from the Erlangen proposal and the diagrams from the string-diagram proposal, and it cut the constructions they judged decorative. The standing rule was that a categorical construction must explain, classify or predict something about PEFT, or be cut. The result was 41 numbered results, a seven-axis classification and a dictionary between the two vocabularies.

Generalisation usually followed one pattern:

\begin{enumerate}
\item Take an observation about a single method. Zero-initialised LoRA cannot move in some directions at the first step.
\item Find the invariant it depends on. Here it is the first-order image compared with the dimension of the image germ.
\item State the result for every method with that coordinate value. Any method whose base point is the vertex of a cone has the same deficit.
\item Look for consequences the source papers do not state. Householder reflection adaptation at its default start is also an apex.
\item Test the claim numerically and send it to referees.
\end{enumerate}

The same pattern turned ``LoRA+ helps'' into an exact statement about Adam's invariance to per-block rescaling. It turned ``orthogonal methods preserve structure'' into complete invariants of each group action. It turned ``this adapter can be merged'' into a local rewriting test. The test certifies exact merging under stated side conditions, names the weight that absorbs the update, and depends on the activation, the normalisation and the bias slots.

The seven coordinates are decidable from a method's formula. That made it possible to classify the first 320 methods in ten batches of 32, and the 62 later additions in three more. In each batch a second agent re-derived every coordinate, checked every arrow's factorisation and base point by hand, and spot-checked a few gauge dimensions and first-order deficits numerically. A consolidator then reconciled the first ten batches with each other and with the theory, and the final site review made the prose agree with the data. Every arrow between methods records its explicit map h with $\rho_N\circ h=\rho_M$ (403 of them). Every recorded non-arrow names the test that rules it out (485).

\subsection{How category theory unifies the field here}

The unifying move is to study each method through its map into weight space. A weight-space method is a pointed map from its small parameter space into weight space that sends its starting point to the pretrained weights. Comparing methods then means comparing maps:

\begin{itemize}
\item a method N simulates a method M when M's map factors through N's, that is $\rho_N\circ h=\rho_M$ for a pointed map h;
\item doing nothing and full fine-tuning are the initial and terminal objects, so they bracket every weight-space method;
\item three invariants of the map carry most of the content: its image germ, its fibres with the optimizer's metric, and its behaviour under base change.
\end{itemize}

Extensions such as adapters and prefixes, which change the network itself, fit through a realisation map. Mergeability becomes the question of whether a lift exists.

Exposé XI lists what is analogy and what is theorem. The words Noether, Erlangen, gauge and Grothendieck are used in a precise, limited sense, and the paper says which.

\subsection{How the results were checked}

Verification ran in layers:

\begin{itemize}
\item \textbf{Sources.} Every catalogue entry was checked by an agent other than the one that wrote it, against arXiv metadata, the paper and the library source.
\item \textbf{Adversarial model referees, three rounds.} Each of the 41 results went to two referee agents with different lenses: mathematical correctness, and faithfulness to the papers and code. They were told to default to ``refuted'' when in doubt.
\begin{itemize}
\item Round 1 produced 82 verdicts, 45 of them major objections (41 refutations). 27 results were repaired and none was false outright.
\item Round 2 re-examined 26 of the 27 repaired results in 48 reports. Six of its 54 runs were stopped by the spend limit, so Corollary III.13 received no second-round report and four results received one; in the final site review the section editors took the missing referees' place. The reports raised 28 major objections against 18 results, and 19 results were narrowed again.
\item Each round-2 report ended with the strongest statement its referee would sign. These became one public statement per result in \texttt{propositions.json}. The writers and referees of the site and the paper were told to state nothing stronger, and checked for it.
\item A third round, four mathematical referees during the writing of the paper, made 63 further fixes.
\item Those fixes were then carried back into the site, the data and the theory files, 108 edits in all.
\end{itemize}
\item \textbf{Numerical checks.} \texttt{framework\_checks.py}, one of the paper's data files, re-runs most claims marked [Num] in about a minute. The paper's Appendix E lists the few that rest only on the referees' own replications. Each interactive figure computes its numbers in the browser. The engineers of the theory-driven figures checked their key numbers against \texttt{framework\_checks.py} or a separate NumPy computation, and in the final site review a second agent spot-checked the key numbers of every figure.
\item \textbf{Engineering checks.} Every figure was rendered headlessly in light and dark themes at desktop and phone widths, and its builder looked at the images. A whole-site test drives Chrome to check console errors, unmounted figures, overflow, MathJax errors and broken in-page links. The paper is compiled from a clean copy of its sources with pdfLaTeX alone and checked for errors and undefined references.
\item \textbf{This appendix.} Two agents checked every number and date in it against the workflow journals, the agents' transcripts and the project files, and its text was corrected where they disagreed. Its opening was then redrafted by three agents writing independently, merged by a fourth and checked by two more, one for facts and one for style.
\end{itemize}

These checks do not amount to human peer review. Every proposer, writer and referee was the same model, so their errors can be correlated, and the referees of the paper also edited the files they reviewed. A numerical check is not a proof, and every check runs on small float64 examples. No fine-tuning run on a real model was made, so the eleven predictions of Exposé X are untested. As Exposé XI records, the dimension equalities for LoHa and, in general, for VeRA are checked numerically; one clause of corollary 4 of Theorem V.3 is labelled a proof sketch, although Appendix C gives an argument; and Conjectures A, C and D are open. The catalogue is a selection and can still miss methods. Its 2026 additions came from keyword queries on arXiv, four conference lists and one citation graph, filtered by a triage agent, and the completeness critic's own knowledge of the field ends with the model's training data. In the catalogue, 51 method entries and 7 paper entries carry a medium-confidence flag. Some methods were missing until the authors noticed them. The authors are responsible for what the site and the paper say.

\subsection{Building the site and the paper}

The site is a static page. Its prose lives in fifteen HTML sections, counting a version of this appendix, and its data is generated from the verified catalogue and the theory files. Each of its 21 figure scripts was written by an engineer agent (four were finished by a second agent after the spend limit cut off the first), self-tested and later reviewed by another. The page runs from any static host, straight from disk and inside a sandboxed artifact. Its data is a script, so nothing is fetched at run time apart from D3, MathJax and the fonts, which come from public CDNs.

The typography was changed once, at the authors' request. The session set one passage of the site in four typeface schemes, and the authors chose the one it recommended, with Source Serif 4 for text, IBM Plex Sans for labels and IBM Plex Mono for numbers. Seven engineer agents then raised the text inside all 21 figure modules to at least 12 pixels where a reader must read it, and 11 pixels for tick labels. Later the authors found the display face of the title unattractive, and the title was reset in Source Serif 4 as well.

The paper uses a template designed for it: SGA-style exposés with Roman numerals, so that every result has the same number on the site and in the paper. Twenty-one writer agents converted the full 35,000-word theory into LaTeX, with complete proofs in Appendix C. Six figure engineers redrew the key figures as vector PDFs in the same typefaces. An assembler, four mathematical referees, a typesetting referee and a finalizer took it to a clean build of 393 pages, about 400 with this appendix.

\subsection{The film}

On the evening of 5 October, at the authors' request, a later session of Claude Code made a narrated film of the atlas, 16 minutes 44 seconds long, and then a two-minute teaser for X. The authors chose the narrating voice, Sulafat from Gemini 3.8 Flash TTS, and a quiet visual style from a short list of options.

The session wrote the script and checked each claim against the propositions and the paper, and a separate agent then fact-checked it adversarially. Every number on screen comes from a short computation in \texttt{video/sim/}. The opening, for example, trains LoRA+ and plain LoRA with a scaled alpha side by side, and the largest difference between their weights stays exactly zero. The session wrote a shared Manim library, a brief and one chapter, and four agents built the other seven chapters from the brief. The session then synthesised the narration beat by beat, timed each animation cue to the pauses detected in the audio, reviewed every chapter on contact sheets of its frames, and assembled the film with captions and chapter marks. The film plays on the companion site, and the title page of the paper links to it.

\subsection{Cost and interruptions}

Table~\ref{tab:making-cost} breaks the agent runs and tokens down by phase.

\begin{table}[htbp]\centering\caption{Agent runs and tokens by phase. Fresh tokens are prompts written to the cache plus generated text; re-reads from the prompt cache are not counted.}\label{tab:making-cost}\footnotesize
\begin{tabular}{@{}p{5.4cm}rrrr@{}}\toprule\headrow \textbf{Phase} & \textbf{Workflows} & \textbf{Agent runs} & \textbf{Fresh tokens} & \textbf{Generated tokens}\\\midrule
Literature catalogue (first sweep, additions, 2026 sweep) & 4 & 75 & 24.6 M & 0.43 M\\
Theory and adversarial review (rounds 1--2) & 2 & 145 & 16.3 M & 2.03 M\\
Classification, site sections and figures & 6 & 103 & 32.1 M & 2.74 M\\
Site review, appendix and typography & 4 & 49 & 22.1 M & 1.68 M\\
Paper (writing, referees, sync) & 4 & 50 & 22.0 M & 1.75 M\\
\textbf{All workflows} & \textbf{20} & \textbf{422} & \textbf{117.1 M} & \textbf{8.63 M}\\
Film (script check and chapters) & -- & 5 & 5.0 M & 0.35 M\\
\textbf{Total} & \textbf{20} & \textbf{427} & \textbf{122.1 M} & \textbf{8.98 M}\\\bottomrule\end{tabular}\end{table}

The table counts every started agent run, one transcript each: the 422 runs of the workflows and the film's five agents, a script fact-checker and four chapter builders. Cache re-reads are not counted (4.1 billion tokens for the workflows, 0.16 billion for the film's agents), and neither are the sessions that directed the work: the orchestrating session of the workflows, and the film's session, which added 1.4 million fresh tokens, 0.55 million of them generated. Also not counted are the two later workflows that checked and redrafted this appendix (nine agents), and the film's narration: about 15 minutes of speech from Gemini 3.8 Flash TTS, plus voice tests and a few retakes, which Google bills separately.

The work was interrupted three times:

\begin{itemize}
\item at 20:52 on 1 October an account spend limit made 29 agent runs fail, 12 of them mid-run and 17 within seconds of starting;
\item at about 20:40 on 2 October the host machine crashed while the paper was being written;
\item at about 23:10 on 2 October the session was cut off again, and it resumed at 02:04 the next morning.
\end{itemize}

Each time, the orchestrator inspected the workflow journals and launched fresh workflows containing only the unfinished items, telling each agent to read and complete its predecessor's partial file. The six stopped second-round referee runs were not repeated; the final site review covered them instead. A first attempt to resume in place re-ran agents that had already finished. The workflow tool replays only the longest unchanged prefix of agent calls, so that attempt was stopped within two minutes. The other process lessons were:

\begin{itemize}
\item pass large inputs to agents through files, never as truncated prompt text;
\item refereeing a repaired result again is worth it;
\item corrections found while writing the paper must flow back to the site.
\end{itemize}

\subsection{Reproducibility}

The catalogue and the checks accompany the paper as data files:

\begin{itemize}
\item \texttt{catalogue.csv} and \texttt{catalogue.json}, with \texttt{arrows.csv} and \texttt{obstructions.csv};
\item \texttt{propositions.json}, which holds each numbered result with its statement, proof, public statement and the notes of both review rounds;
\item \texttt{framework\_checks.py}, the numerical checks.
\end{itemize}

The LaTeX source carries the text and the generated tables, and the site's figure code can be read in the page. The public repository, \url{https://github.com/xiaoyulics/peft-atlas}, holds the rest of the record:

\begin{itemize}
\item the raw and verified literature files and the classification batches, in \texttt{research/};
\item the full theory text and the round-2 reports, in \texttt{theory/};
\item the build and test scripts, in \texttt{site/build}, \texttt{site/test} and \texttt{paper/tools}, and the scripts of the paper's figures, in \texttt{paper/figures/src};
\item the film's script, scenes and computations, in \texttt{video/}.
\end{itemize}

The 20 workflow scripts, which contain every agent prompt, and the agents' transcripts stay with the authors.

\end{document}